# ABSTRACT

## BLINDFOLDED SPIDER-MAN OPTIMIZATION: A SINGLE-POINT METAHEURISTIC SUITABLE FOR CONTINUOUS AND DISCRETE SPACES

In this study, we introduce a new single point metaheuristic optimization approach suitable for both continuous and discrete domains. The proposed algorithm, entitled Blindfolded Spiderman Optimization, follows a piecewise linear search trajectory where each line segment considers a move to an improved solution point. The trajectory resembles spiderman jumping from one building to the highest neighbor building in a blindfolded manner. Blindfolded Spiderman Optimization builds on top of the Buggy Pinball Optimization algorithm. Blindfolded Spiderman Optimization is tested on 16 mathematical optimization functions and one discrete problem of Unbounded Knapsack. We perform a thorough evaluation of Blindfolded Spiderman Optimization against established and state-of-the-art metaheuristic optimization methods, including Whale Optimization, Grey Wolf Optimization, Particle Swarm Optimization, Simulated Annealing, Threshold Accepting, and Buggy Pinball Optimization considering various optimization domains and dimensions. We show that Blindfolded Spiderman Optimization achieves great performance on both continuous and discrete spaces, and superior performance compared to all single-point metaheuristic approaches considered.

Satyam Mittal

December 2023

# BLINDFOLDED SPIDER-MAN OPTIMIZATION: A SINGLE-POINT METAHEURISTIC SUITABLE FOR CONTINUOUS AND DISCRETE SPACES

by

Satyam Mittal

A thesis

submitted in partial

fulfillment of the requirements for the degree of

Master of Science in Computer Science

in the College of Science and Mathematics

California State University, Fresno

December 2023

APPROVED

For the Department of Computer Science:

We, the undersigned, certify that the thesis of the following student meets the required standards of scholarship, format, and style of the university and the student's graduate degree program for the awarding of the master's degree.

Satyam Mittal
Thesis Author

Athanasios Aris Panagopoulos (Chair) | Computer Science

Shih-His Liu | Computer Science

Amith Kamath Belman | Computer Science

For the University Graduate Committee:

Dean, Division of Graduate Studies

AUTHORIZATION FOR REPRODUCTION
OF MASTER'S THESIS

______________ I grant permission for the reproduction of this thesis in part or in its entirety without further authorization from me, on the condition that the person or agency requesting reproduction absorbs the cost and provides proper acknowledgment of authorship.

____X_____ Permission to reproduce this thesis in part or in its entirety must be obtained from me.

Signature of thesis author: Satyam Mittal

ACKNOWLEDGMENTS

Firstly, I wish to extend my deepest thanks to my mentor who guided me on each step of the work, Dr. Athanasios (Thanos) Aris Panagopoulos. His guidance, support, and unparalleled expertise have been the beacon of light throughout the journey of this dissertation. His patience and mentorship have not only shaped this work but have also played a pivotal role in my personal and professional growth.

I'd also like to genuinely express my gratitude to the members of the committee who have dedicated their valuable time and provided indispensable feedback that greatly enhanced the work of this research.

A special mention to the head of the Computer Science department for fostering an environment of excellence and inquiry. Your leadership and vision have created a platform where young minds like mine can dare to dream and explore and, to Fresno State University, I am deeply thankful for providing me with the resources and the conducive environment to pursue my academic aspirations which I had in my mind when I started the journey of graduation. Being a part of this esteemed institution has been an honor and a privilege.

Lastly, I wish to acknowledge the work of Vasileios Lymperakis. His research on Buggy Pinball Optimization laid the foundation upon which this dissertation stands. His contributions to the field have been a source of inspiration. And, to Technical University of Crete which provide us the computation power to run our experiments on cloud server.

TABLE OF CONTENTS

Page

LIST OF TABLES

LIST OF FIGURES

# CHAPTER 1- INTRODUCTION

Both nature and people have consistently looked for the best answers to issues throughout history. For instance, there is a recurring theme of seeking the shortest or fastest path, from the spider's instinctive tendency of picking the tense line in its web to catch prey quickly to the smart general's creative tactic of rolling his forces down a mountain to seize a town speedily. With respect to human history, researchers were interested in finding the shortest path between two sites as early as 100 BC [1]. These tasks are known as optimization tasks where the objective is finding the optimal solution among many possible solutions. The search of optimal solutions is foundational in the world around us. Optimal solutions are not considered as random occurrences; they are the foundational concept that shapes biological behaviors and influences societal decisions. For instance, ants have evolved a complex communication mechanism based on pheromones (i.e., stigmetry) through which they are able to solve the shortest path problem among their nest and food sources [2].

Whether in science, engineering, or even philosophy, the aim is constantly for improvement—improving upon current techniques or developing new ones to improve results. Today, optimization is present in practically all areas of societal activity. Importantly, optimization techniques (such gradient descent optimization [3]) are behind the great advancements that machine learning and artificial intelligence is achieving today.

Historically, optimization tasks have been tackled through trial-and-error methods. This is supported by observations in optimization tasks performed by other members of the great ape’s family [1]. In the dawn of humans, these approaches were potentially supported by simulations. Even in the present period, physical simulations are used to visualize and solve problems, such as utilizing ropes to find the shortest path on a network or a table and ropes to find the best school placement for villages [1]. Based on such approaches humans derived with simple heuristics which are problem specific rules of thumbs that can be used to solve an optimization problem quickly and near-optimally. For instance, some heuristics include the nearest neighbor approach in finding the optimal path in a graph.

However, as the body of human knowledge advanced, the idea of optimization started to be integrated with mathematics, giving rise to more organized techniques to arrive at optimal solutions. The history of optimization in mathematics can be divided into three periods [1]: (i) initial phase, (ii) medieval phase, and (iii) the new era. This categorization is based on the fundamental techniques used to determine the maximum or minimum of a function. Mathematical optimization can be defined as:

Mathematically stating [4]: Let us consider a function f, such that

OptimizationFunc: $X \rightarrow R$ , here X represents real numbers.

Sought: Here, consider x` and x` $\in X$

OptimizationFunc (x`) >= OptimizationFunc (x) $\forall x \in X$ => this equation represents maximization

OptimizationFunc(x`) <= OptimizationFunc (x) $\forall x \in X$ => this equation represents minimization

During the initial phase, we did not have a universal approach to determine the maximum or minimum of function. Only specific methods were identified to enhance or reduce certain distinct functions, but the general approach was based on trial and error.

P. de Fermat inaugurated a new era, known as the medieval phase, in 1646 [5]. In his research work [2], a generic method has been suggested for computing the location of local minima and maxima for differentiable functions. The method says that, by setting the derivative of the differentiable function to zero one can identify the critical points (i.e., local minima and maxima, as well as the saddle points) which can be used to compute the global maxima and minima [2].

Finally, a new era of optimization was inaugurated by the discovery of linear programming. It is also worth noting that the landscape of optimization was entirely transformed through the emergence of digital computers. The ability to conduct many fast computations allowed the development of several search algorithms and optimization techniques extending beyond pure mathematical optimization towards computing science.

Notably, with the advancement of computing power, a new domain of optimization algorithms emerged which could solve problems based on searching large search spaces. These algorithms essentially combine heuristics with mathematical optimization. The new family of techniques is known as metaheuristics. While mathematical programming techniques offer assured optimality, they typically suffer in large-space problems. At the same time heuristic approaches can effectively manage large search spaces however they are problem specific. Metaheuristics can tackle large spaces and are general enough to tackle a wide range of problems, although they do generally not come with guarantees of optimality [5]. Nevertheless, they have proven to be potent in addressing extensive optimization conundrum. As said by Fred Glover [6], metaheuristics are fusion of two broader words and coined the term "metaheuristic" .

meta- μετά, Surpassing in terms of an upper-tier level.

heuristic - Greek heuriskein or ευρισχειν, and it means "to search".

Typically, metaheuristics are inspired by nature as many organisms and physical processes can solve optimization tasks in an extraordinary way. Metaheuristics can be classified into population-based and single-point methods. Population-based methods, such as Particle Swarm Optimization (PSO), Whale Optimization (WO) and Grey Wolf Optimization (GWO) have multiple agents which work toward finding the most optimum solution. These studies frequently take their cues from natural and social events, such as the hunting tactics of humpback whales, the complex management of ant colonies, or the predation tactics of grey wolves etc. Single point methods, such as Simulated Annealing (SA), Buggy Pinball Optimization (BPO or BP) and Threshold Accepting (TA) etc. have one agent which works towards finding the most optimum solution. These studies are typically inspired by physical processes such as annealing process in metallurgy, pinball game etc.

Currently, there is a strong trend in research toward population-based techniques. This inclination toward population based is due to their promising results shown in evaluation of large space continuous optimization problems. In more detail, optimization tasks can be divided into continuous and discrete (or combinatorial) optimization. The former deals with continuous functions (such as finding the minimum of a convex or non-convex function) while the latter deals with discrete problems (such as finding the shortest path between two nodes in a graph). The term non-continuous optimization is also used to highlight optimization tasks that are generally continuous but have some points of non-continuity (such as piecewise functions). Now, continuous optimization is particularly popular in the context of machine learning algorithms progress, and, hence, population-based metaheuristic is quite popular. For instance, in the research work by A. Eiben, S. Smit [7] addresses how parameter adjustment in evolutionary algorithms can be difficult and suggests methods that are like machine learning strategies for enhancing performance.

Population-based approaches are also gaining in popularity due to the enhanced computational capabilities [8] offered by hardware like TPUs and GPUs which enable the simultaneous operation of multiple agents, leading to more refined hyper-parameters and a closer reach to the global optimum. A population-based approach can be easily parallelized in this respect. There is still a lot of work being done to find the best answer in challenging continuous situations using the capabilities of machine learning and population-based algorithms.

## 1.1- Problem Statement

Given the above, whether it is worthwhile for the search community to further investigate single-point metaheuristics becomes an interesting question. Recently, a new metaheuristic approach named “Buggy Pinball Optimization” has been proposed which is a single point approach that explores the domain of single point evolution. The approach is based on a pinball game which results in a random starting point and has the capability to avoid getting stuck in local optima. The approach has shown promising results in optimization problems and is better than standard single points like SA, TA and in some cases better than PSO even in higher dimensions. However, the current research work is only applied to continuous domain optimization problems and although BPO has shown promising results in continuous domains, the performance in non-continuous domain and discrete domain remains untouched and hence paves the way for further exploration and optimization of BPO.

In this work, we built upon the prior research in BPO concerning the continuous domain, we ventured further by examining BPO's performance against established metaheuristics in continuous, non-continuous and discrete spheres. Apart from this, we propose a new single point metaheuristic approach that builds on top of BPO, by introducing memorization and enhancing the trajectory followed based on this information. This gave birth to a new approach that we name "Blindfolded Spiderman Optimization (BSO)" — an advanced iteration of the core BPO mechanism. This strategy uncovers near-ideal solutions to intricate challenges across continuous, non-continuous, and discrete domains, all while utilizing fewer trajectories compared to its predecessor, BPO. Such advancements expand the repertoire of techniques at our disposal for tackling optimization tasks. We evaluate BSO against BPO as well as against TA, PSO, WO, GWO. We show that BSO has superior performance to all single point approaches considered (including BPO) and comparable performance to the population-based approaches.

## 1.2 - Contributions

In more detail, our main contributions are listed as follows:

- Introducing a new metaheuristic approach called BSO that refines the BPO approach.
- Performing a comparative analysis of BSO with BP, SA, TA, PSO, WO, GWO for continuous and non-continuous domains considering various testbed functions and dimensions and a thorough statistical analysis of the results.
- Performing a comparative analysis of BSO with BP, SA, TA, PSO, WO, GWO for discrete domains considering the unbounded Knapsack problem In various dimensions and a thorough statistical analysis of the results.
- Providing insights and visualizations on the performance of BPO and BSO in various optimization tasks.

## 1.3 - Outline

The rest of the thesis is structured as follows: Chapter 2 provides an extensive exploration of the background and related work, dissecting several optimization algorithms, including SA, TA, PSO, WO, GWO, and BPO. In Chapter 3, we introduce BSO, a new algorithm, providing step-by-step pseudocode explanations. In Chapter 4 and 5 sections delve into experiments, case studies, and results across seventeen diverse optimization problems (continuous, non-continuous domain along with one unbounded knapsack, a discrete problem) and, showcasing BSO's trajectory visualizations. Finally, Chapter 6 concludes, encapsulating our findings and future work directions.  A supplementary material chapter (See Appendixes A, B, C, D, E, F) are also included where the results are included in detail for multiple case studies and appendixes (G, H, I, J, K, L) consist of supporting statistical significance results for results obtained in multiple case studies of optimization problems.

# CHAPTER 2 - BACKGROUND AND RELATED WORK

## 2.1- General Working of Metaheuristic

As metaheuristics are general purpose algorithms [9] which are inspired from wide variety of natural and social phenomena (e.g., observing social behavior of birds flocking, foraging behavior of ants, human process like tabu search, other unique inspirations like pinball games, natural phenomena like simulated annealing and threshold accepting etc.) and these can be applied to a large set of problems. Hence, these are optimization algorithms which do not guarantee to provide optimal solutions every time but assure us to provide good feasible optimal solutions in reasonable time.

In general, there are two categories of metaheuristic algorithms: single solution point and swarm/population based. In the former one, we have one single point, and we try to optimize that solution to get the most optimized solution whereas in the later one we have a "population" usually referred to as "agents" and all these agents work toward to get the most appropriate solution. In recent years metaheuristic has gained popularity and there many new approaches have been developed by observing the behavior of ants, bats, whales, pinball, birds, humans etc. All these search algorithms work on the same principle of exploration and exploitation [9].

Exploration [9]: It is the practice of looking for and examining potential new areas of the solution space that may hold more accurate solutions. It entails broadening the scope of the search by venturing into uncharted territory and experimenting with various approaches. Exploration strives to bypass confinement in local maxima, which represent inferior solutions found in a narrow segment of the search domain.

Exploitation [9]: On the other hand, exploitation concentrates on stepping up the finding in promising areas of the solution space. It entails using the knowledge gathered from earlier search cycles to improve and hone the solutions. Exploitation seeks to move people toward the most well-known solutions or to take advantage of areas that have so far performed well.

Metaheuristic algorithms often require repeatedly iteratively seeking and exploring the solution space using exploration technique to gradually improve the results and exploitation technique to further evaluate the best promising solution space.  Hence, by following these two strategies, metaheuristic algorithms continuously evolve and very efficiently navigate through the search space avoiding any obstacles/saddle points and reach towards the goal. They use meta heuristics for optimization even though they do not guarantee exact optimal solutions every time, because they can be widely utilized for comprehensive optimization challenges. They have the capability to bypass settling at local peaks and are adaptable to problems with multiple dimensions, hence locating the near optimal solution [9]. Additionally, they usually do not need the objective function's derivative, setting them apart from optimization methods that rely on gradients. Fig. 1, reflects a conclusive idea on global optimization and broadly categorizing metaheuristics.

These are very well-known algorithms in the field of metaheuristics and each one operates following their respective concepts. For example, SA follows the metallurgical annealing process, likewise TA is an extension of SA and follows the concept of threshold values. The roots of PSO were greatly influenced by the cooperative behaviors of social organisms, particularly the coordinated movements observed in bird flocks and fish feeding. Similarly, WO is based on humpback whales' behavior and GWO is based on social structure and hunting methods of grey wolves. Hence, we have several great algorithms based on some natural phenomena with each having their own pros and cons which we will be studying in detail in the background section later.

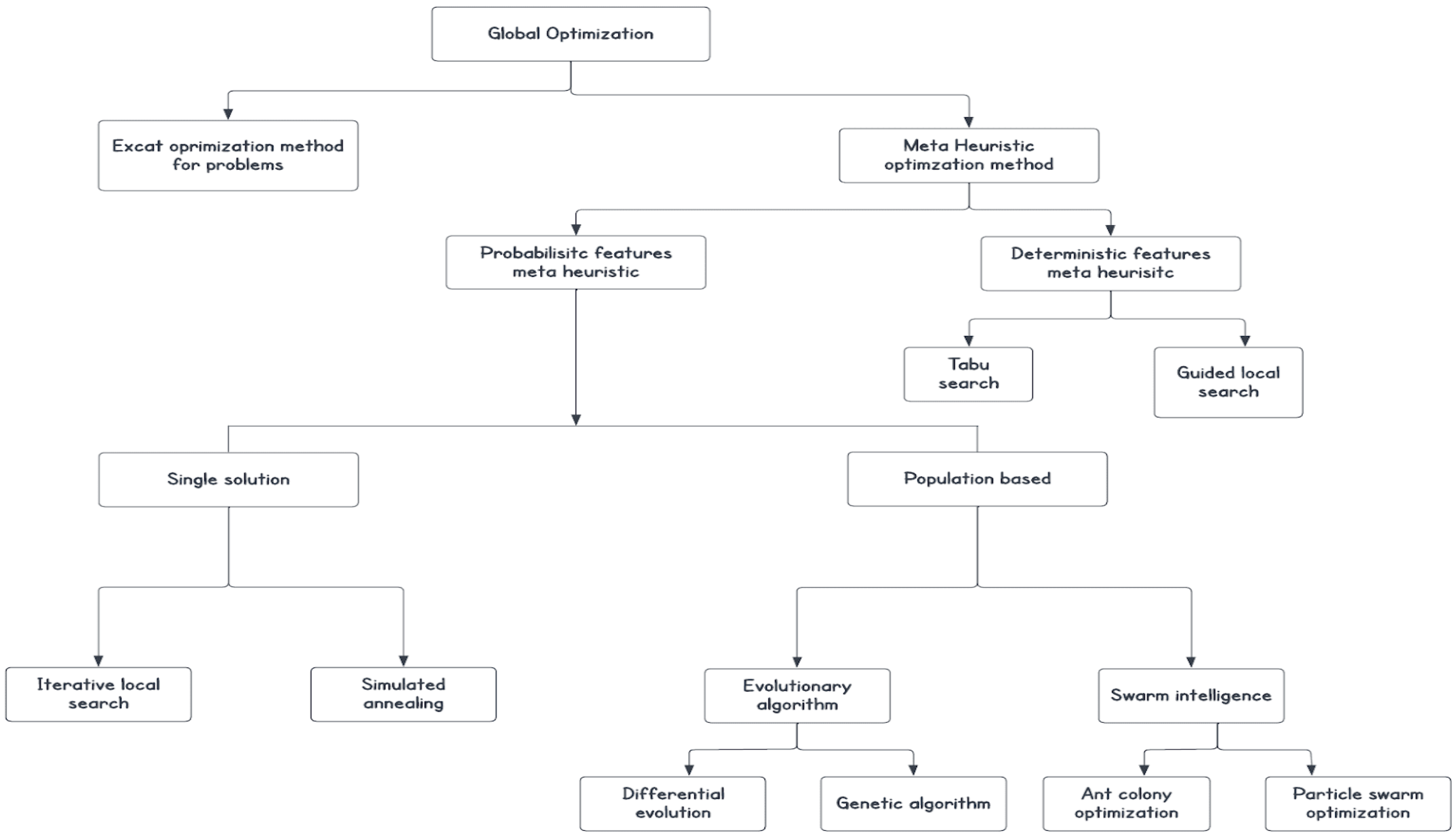

Fig. 1. Metaheuristics Categorization in terms of Global Optimization

## 2.2 - Categorization of Optimization Problems

We have been discussing optimization problems and in the former chapter we saw how optimization functions are stated mathematically. Broadly, we can categorize optimization problems as continuous, non-continuous and discrete problems.

### Continuous Optimization:

In this category, the problems are defined over a continuous domain. And within the given range, involved variables can take on an infinite number of values. The variable in the problem equation can assume real-number values within a specified range. For example, Sphere function, Minimum time taken for a person to travel from one city to another under certain conditions.

Non-Continuous Optimization:

In this category, variables specified throughout seemingly continuous domains but have discrete points or regions where they are not defined or abrupt shifts. There could be "jumps" or "breaks" in a function or issue's domain. Here, conventional continuous optimization methods might not be immediately applicable, making the problem difficult to address. For example, Piecewise-defined functions

Discrete Optimization:

In this category, optimization problems are defined over a discrete domain, which indicates that the variables only have a limited number of possible values, frequently integers. Assume values from a specific set (for example, whole numbers) and the use of calculus is not always necessary to find solutions instead, combinatorial or integer optimization methods are frequently used. For example, knapsack problem, travelling-salesman problem

Before moving ahead to further sections, let us understand the term pseudocode. A pseudocode [10] is a way of writing complex algorithm including suitable grammar and vocabulary words, so that anyone can read and understand the execution of the algorithms. We cannot directly use pseudocode as an executable program as it consists of simple and less ambiguous words [10]. Pseudocode contains steps which follows indentation like a tree structure to represents statements like if-then(conditional statements), while and for(loop statements), symbolic notation to represents mathematical equations, break stops the execution of loop statements and return statement is used to return something from a method or function. A method is a block of lines of code which are one after the another. Table 1 to Table 7 represents pseudocode for SA, TA, PSO WO, GWO, BPO and BSO metaheuristics, following the instructions mentioned in the pseudocode anyone can implement the executable algorithms using programming languages.

## 2.3 - Simulated Annealing

SA [11] is mimicked from the process used in metallurgy. In this process a material is taken and heated to a point and then allowed to cool down slowly to reach the lowest energy state. This metaheuristic is a single approach for optimizing complex problems and used to find the global min/max of complex functions/problems.

The metallurgical annealing process serves as the source of the algorithm's name and inspiration. A technique known as "annealing" causes a material to be heated and then gradually cooled down, allowing its atoms to settle into a more stable and low-energy state. The SA algorithm is an adaptation of this idea for optimization.

### Basic Concepts

SA consist of four major sub-routines i.e., Initial Solution, Schedule Temperature, Evaluation and Acceptance of Optimal Solution. The algorithm begins with some random initial solution and continues until the terminating condition is encountered.

Initial Solution: SA starts with an initial solution, which is frequently produced at random or given by a heuristic.

Schedule Temperature: The algorithm adds the idea of "temperature," which regulates the propensity to accept alternatives to the present answer that are worse. The algorithm may investigate a wider variety of possibilities because the temperature is initially set rather high.

Evaluation: By causing the existing solution to change, a new solution is created after each cycle. Swapping components, altering values, or making other changes might all be part of this disturbance.

Acceptance of Optimal Solution: The core principle of SA is that the method approves new solutions based on a combination of the current temperature and the variance in the objective function values between the present and proposed solutions. The algorithm can occasionally accept inferior solutions thanks to this likelihood, making it easier to avoid local optima.

Finally, the temperature progressively drops during the algorithm in accordance with a predetermined cooling plan. The algorithm will eventually converge to an ideal answer as a result of the decreased likelihood of accepting suboptimal options.

Advantages [12]

Because SA is so good at exploring the solution space, it works well for optimization issues involving several local optima. A large amount of exploration is possible because to the high beginning temperature, and the convergence towards the global optimum is made possible by the cooling process. In addition to this, to avoid being stuck in suboptimal portions of the search space, SA offers the capacity to escape local optima by permitting the acceptance of poorer solutions with a specific probability. And SA does not require prior knowledge of the problem's landscape.

Limitations [12]

One of the limitations is the beginning temperature, cooling schedule, and acceptance probability function can have an impact on SA performance. To get the best results, parameter adjustment is essential. Although SA can locate global optimums, the issue and parameter choices can affect how quickly it converges. The convergence of some problems could take more computing time.

Code Explanation

In SA we search the solution space randomly and use a probability function to decide which solution to accept and reject. Initially a solution is selected and modified as each iteration progresses. During every cycle of the algorithm, the proposed solution is weighed against the existing one, and using a probability metric, a choice to accept or decline is determined. Probability function is defined in terms of temperature parameters and initially temperature is set to a high value to allow more exploration in a broader way and as the temperature reduces the probability function becomes stricter and more restrictive so that algorithms start to converge either towards local min/max. SA can easily escape the local minima due the usage of probability function as it allows to select even the much worse solution than the current solution which can help in avoiding getting stuck. See Table 1 for pseudo code of SA.

Table 1. Simulated Annealing Pseudocode

```
SA():
  s = s_0        #randomly generated solution
  T = T_max
  n_d        #Initialize Neighbor Distance
  f(·)                 #objective function
  for k = 0 until terminal condition is met:
     T←cooling(T, k)    # temperature decrease rate
     solution_new←neighbour_solutions(s)
     E_s = f(s)
     E_new = f(solution_new)
     E_delta = E_new − E_s
     if E_delta < 0 or random(0,1) < exp^(−E_delta/T)
          s = solution_new
  return  s

neighbour_ solutions (solution):      # random neighbor solution
  for x_i in solution:
    neighbour_sol [i]=random(x_i-n_d, x_i+n_d)
  return neighbour_sol
```

## 2.4 - Threshold Accepting - Methodology

TA [13] is another single point meta heuristic algorithm designed to solve complex optimization problems and find global min/max of TA is like SA algorithm but instead of temperature threshold values are used to accept or reject any new solution.

### Basic Concepts

TA consist of four major sub-routines i.e., Initialization, Threshold Update, Solution Generation and Decision.

Initialization: Before executing the TA algorithm, the starting threshold value and initial solution are set. As the algorithm progresses, this threshold value becomes pivotal in gauging the probability of embracing newer solutions.

Threshold Update: TA uses a decreasing threshold value as opposed to SA, which steadily decreases the temperature parameter. This cut-off point regulates how openly the algorithm will accept novel solutions. Suboptimal solutions are ever more difficult to accept as the algorithm gets more discriminating about allowing answers as the threshold value drops over time.

Solution Generation: A new solution is created after each iteration by making certain changes to the previous one. This alteration can entail changing one or more solution components. The objective function is used to assess the new solution.

Decision: The new solution is likely to be approved if the difference is less than the threshold value.

### Advantages

TA grows more selective as the threshold value drops over time, enabling it to come out of local optima and extensively do exploration in the solution space. And TA is more flexible to various issue types and solution areas since it does not need a set cooling schedule like SA. Numerous optimization issues, including those with complicated, non-convex, and non-continuous objective functions, can benefit from the use of TA.

Limitations

In order to guarantee successful convergence, proper parameter adjustment is required. And due to its stochastic nature, TA's performance might differ across runs, much as other probabilistic algorithms.

Code Explanation

The algorithm begins with an initial temperature and a threshold value where the threshold value ins slowly and gradually decreased over time. A new solution is generated at each iteration and a certain probability determines whether to accept or reject the solution. This process is the result of difference between the values of current threshold and the new generated solution. As threshold value gradually decreases over time it allows in escaping local min. and max and allows more exploration in the search space. See Table 2 for pseudocode of TA.

Table 2. Threshold Accepting Pseudocode

```
TA():
    Initialize #rounds      # total number of rounds
    Initialize #steps       #number of steps
    Initialize τ            #threhold sequence
    f(·)                    #objective function
    s=s0
    for round =1 to #rounds do:
        for step in range(1, #steps) do:
            s_new← neighbor solution (s)
            If (f(s_new)-f(s))< τ_round
                s← s_new
      return s

neighbour_solution(s):    # randomly generated solution
  for x_i in s:
    neighbours[i] = random(xi - nd, xi + nd)
  return neighbors
```

## 2.5 - Particle Search Optimization

PSO [14], metaheuristic approach which is designed after observing social behaviour of bird flocking. In PSO, we have groups of particles that move and search for optimal points in search space. Each particle has its own best position {personal best} and as a group they have global best position. So, the movement of particles are affected and changed by their own best and global best solution found so far.

### Basic Concept

Every particle represents a location in the search domain, and the method initiates by randomly setting the positions of these particles and their speed. In each iteration, particle position is updated based on current position, velocity and global best solution till now. Each particle has a velocity which is calculated using below equation:

$$v_{i,j}(t+1) = wv_{i,j}(t) + c_1r_1\left(p_{i,j} - x_{i,j}(t)\right) + c_2r_2\left(g_j - x_{i,j}(t)\right)$$

- $v_{i,j}$: the velocity of the $i^{th}$ particle and $j^{th}$ dimension
- w: inertial weight
- $c_1, c_2$ : constant of acceleration
- $r_1, r_2$ : random number in-between [0, 1]
- $p_{i,j}$ : $i^{th}$ particle, $j^{th}$ dimension - personal best position
- $x_{i,j}(t)$ : $i^{th}$ particle, jth dimension - current position in iteration t
- $gb_j$ : group global best position in $j^{th}$ dimension.

Hence, ith particle updated position in $j^{th}$ dimension is represented as :

$$x_{i,j}(t+1) = x_{i,j}(t) + v_{i,j}(t+1)$$

This algorithm persists in its exploration and utilization until it reaches the maximum iteration count or attains a desired fitness level. We have been using PSO in multiple disciplines like machine learning, image procession, function optimization and the advantage of PSO is its simplicity of implementation and efficient way to avoid local optima. Although the convergence rate might be slower than other algorithms in the metaheuristic domain, performance depends on the number of parameters as discussed previously. See Table 3 for pseudocode of PSO.

Table 3. Particle Swarm Optimization Pseudocode

```
Initialize: #rounds, #particles, weights
 f(·)                    #objective function
    for i in range(0, #particle) :
       initialize x_i, pb, v
    if f(x_i) < f(gb) then
        update global best
    for round in range(0, #rounds):
       for p in particle:
         for d in range(1, dimension+1):
              v_d ←ωv + c_1 random(0, 1)( pb_d - x_{i,d} )
                   + c_2 random(0, 1)( gb_d - x_{i,d})
          update x_i
          if f(x_i)<f(pb) then
              update pb
                     if f(x_i )<f(gb)
                              update gb
```

Advantages [15]

The major advantage of PSO is excellence in global exploration, which enables it to find the best solutions throughout a large solution space even when there are several local optima. The efficacy of PSO's factors, like inertia weight and acceleration constants, can be adjusted for diverse challenges. Due to its adaptability, users can tailor PSO to tackle specific optimization problems.

Limitations [15]

Premature convergence, which occurs when particles converge to less-than-ideal solutions too fast and insufficiently explore the solution space, can be a problem for PSO. Even, the selection of parameters, particularly inertia weight and acceleration coefficients, has a significant impact on PSO performance. Finding the right parameter settings might be difficult.

## 2.6 - WHALE Optimization

WO [16] approach has gained prominence as a revolutionary technique to address intricate optimization challenges in the realm of metaheuristic optimization procedures. Motivated by the fascinating actions of humpback whales when hunting for nourishment, WOA is propelled by the innate wisdom of nature. This method, which was released in 2016, captures the spirit of these magnificent marine animals by channeling their intrinsic tactics into a computational framework that deals with a wide range of optimization difficulties.

The WO algorithm's fundamental function is to comprehend the complex movements of humpback whales as they travel through the vast oceans. A coordinated search of the solution space is orchestrated by the population of agents in the algorithm, much to how humpback whales coordinate their movements to capture schools of fish. This imitation of natural tactics offers a distinctive perspective through which optimization problems might be reframed and solved.

### Working of WO [16]

It starts with initializing a candidate (known as whales) population with random position. Each whale has its own position in search space and the position gets updated in each iteration by encircling prey, bubble-net attacking and searching.

Encircling prey: During this phase, every whale gravitates towards the most suitable optimized solution identified so far. A hyperparameter named 'a' regulates the magnitude of the search movement and diminishes over time to harmonize discovery and utilization. Consequently, each whale's location is recalibrated considering the top-performing whale's position and a randomly generated vector scaled by a factor named 'a'.

Bubble-net attacking: Bubble net feeding is used to create a circular region around the most optimized solution till now. This outcome is realized by adjusting the location of every whale, considering the leading whale's position in the group, combined with a random vector scaled by a factor named "A". The "A" parameter determines the radius of the circular area and diminishes progressively over time.

Searching: To explore new regions whales move randomly and this feature is achieved by updating the position of each whale based on random vector multiples by a parameter ‘C’. The hyperparameter "C" managed the step size of its random search and is kept constant throughout the algorithm.

Consequently, the crux of the bubble-net hunting technique lies in the attack. In WOA, this is termed the exploitation phase, segmented into two parts: the tightening encirclement strategy and the spiral position refresh. See Table 4 for pseudocode of WO.

Advantages

WO is excellent for issues with complicated, multi-modal, or non-convex solution spaces because of its emphasis on global exploration. Also, WO permits parameter changes, like other metaheuristics, allowing users to fine-tune the algorithm's performance for various problem domains.

Limitations

One of the limitations of WO is that WO may depend on parameter variables. The efficacy of the algorithm can be considerably impacted by selecting the right parameter settings. Moreover, WO may experience premature convergence, when the algorithm moves toward a less-than-optimized solution before thoroughly exploring the solution space, just like other global optimization methods.

Table 4. Whale Optimization Pseudocode

```
X_i (i=1,2,3,4....n)  #whale population initial position
for w in range( #whales):
   calculate value of the Fitness function
X * = best search agent chosen to start
for t in range(0,iteration_max) do:
      for i in  range(#whale)
         a, A, C, l, and p  # updated values of a, A, C, l, and p
           if1(p<0.5) then
              if2( |A|<1) then
                  # current whale agent position changes
                  D = | C.X_best (t) − X (t) |
                  X (t+1) = X_best (t) −A.D
              else if2(|A| >=1 ) then
                   D = | C. X_rand (t) − X (t) |
                   X (t+1) = X_best (t) − A.D
               end if2
           else if1(p>=0.5) then
                    # set new obtained solution value of the current whale agent
                    X (t+1) = D. e^bl. cos(2 πl)  +  X_best (t)
         t=t+1
return X*
```

## 2.7 -Grey Wolf Optimization

GWO is one of the known benchmark metaheuristics in the domain of population-based approaches proposed by Seyedali Mirjalili, Seyed Mohammad Mirjalili, and Andrew Lewis in 2014 [17]. It is modelled after the social structure and hunting methods of grey wolves in the wild. It imitates the way wolves pursue, circle, and attack their prey.

The algorithm follows the following approach where we have four different wolf kinds used in the method are “alpha, beta, delta, and omega.” The pack's leader, or alpha, oversees making decisions. Beta is the runner-up candidate and aids alpha in making decisions. While submitting to alpha and beta, delta rules over omega. The lowest rank,

Omega, must follow the others. With a population of wolves chosen at random, the algorithm begins. Each wolf stands for a potential answer to the optimization issue. By changing a group of wolves' locations iteratively, the algorithm is made to discover the best answer. Each wolf stands for a possible answer to the optimization issue.

Basic Concept

GWO consist of five major sub-routines i.e., Initialization, Pack Hierarchy, Updating Position, Fitness Evaluation, Updating Pack Position and Terminating Condition.

Initialization: Distribute a pack of grey wolves at random around the problem space. The quantity of wolves is a parameter that the user defines. And the Objective function determine each wolf's fitness (objective value) considering the provided optimization challenge. The wolves are led to better options via the fitness function.

Pack Hierarchy: Within their groups, grey wolves display a hierarchical structure. In GWO, the alpha, beta, and delta wolves stand in for this social order as shown in Fig. 2. These wolves are regarded as the pack leaders and possess the best fitness ratings out of all the wolves in the current iteration.

Updating Position: Based on their hierarchy and hunting habits, the positions of the wolves are modified after each cycle. The placements of the alpha, beta, and delta wolves, as well as a random exploration element, have an impact on this update.

Fitness Evaluation: Make sure the wolves' updated locations remain inside the specified search area. Determine if the new placements are suitable for each wolf.

Updating Pack Position and Terminating Condition: After the position updates, update the alpha, beta, and delta wolves depending on their fitness ratings. The method iterates until a predetermined stopping condition, such attaining a predetermined iterations number or arriving at a good solution, is satisfied.

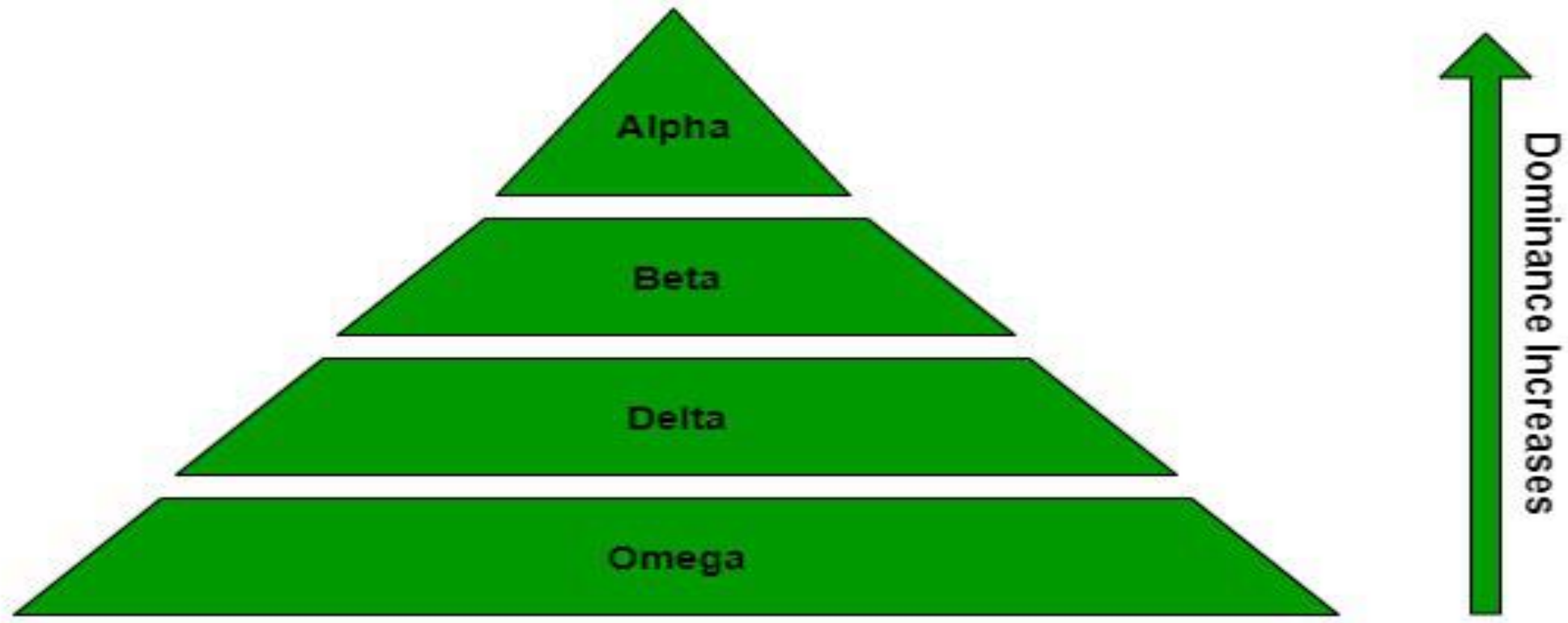

Fig. 2. Hierarchy of Wolf pack in Grey Wolf Optimization

GWO has been utilized in a variety of domains, including reactive power dispatch, machine learning, and engineering design. See Table 5 for pseudocode of GWO.

Table 5. Grey Wolf Optimization Pseudocode

```
Gwo():
        Initialization :
           a    #exploration factor
           A    #exploutation factor
           x_i, i = 1,2,3…n           # population initialization : max wolf agents
          rounds_max                  # maximum number of rounds
        calculate Fitness value for each wolf agent and sort them
        x'     # first highest fitness wolf agent
        x''    # second highest fitness wolf agent
        x'''   # third highest fitness wolf agent
        while t < #rounds_max:
            for  x_i in wolf_agents do:
                Random initialization of r1 and r2
                #Using below equation update ith search agent position
                 if(0 <= rand() <= 1) then:
                    pos_new = a (x' − x[i])
                 else:
                    if(0 <= rand() <= 0.5) then:
                       pos_new = a.(x'' − x[i])
                    else:
                       pos_new = a(x''' − x[i])
                 #set new obtained solution value for (x[i])wolf position with pos_new
             Update the  values for a , A
             Update x', x'', x'''              # update best three wolf
                     t=t+1
           return  X_α
```

## 2.8 - Buggy Pinball Optimization

The idea of BPO [18] revolves around creating virtual planes in search space and moving with progressive continuous simulation steps following the plane until a crossing between the function and the moving steps is encountered, just like a pinball game. As soon as the collision occurs a new calling of sub-routing is made. This sub-routine helps in finding the actual collision point. Once the collision point is detected, a new trajectory is initiated following this collision point and the process continues iteratively.

The imaginary plane is designed using elevation angles which start almost horizontally and become steeper over time to make convergence faster. Besides this step size reduces over time to increase the precision. Henceforth, as per the author BP can be fragmented into following sub routines:

### Working Of BPO

The core of the BPO algorithm revolves around exploring the search space through trajectory generation. The process involves taking progressive continuous evaluating steps until a hit between the ball and the function is encountered. At this point, a new sub-route is initiated to determine the collision point, from which a new trajectory begins, repeating the entire sequence. The question arises: how does BPO establish these trajectories? To address this, the author employs randomly generated directions, facilitating comprehensive exploration of the search space and preventing the BP from remaining stuck in local optima. Notably, the angle of elevation starts nearly horizontal and gradually becomes steeper over time, enhancing convergence speed. Simultaneously, the step size diminishes progressively to attain heightened precision. The overall structure of BP can be dissected into distinct segments: initialization, creation of trajectory segments, advancement through steps, iterative refinement, and incorporation of cooling schedules.

Furthermore, during the setup phase, all essential parameters are set. The ascent angle is aligned to the plane orthogonal to the y-axis, with both the round tally and step count within each round being predetermined. The design of path segments dictates the route's direction. At the start of every round's journey, the direction is chosen at random, with the ascent angle kept constant. This angle is derived by calculating the cross-product between a vector aligned with the y-axis and the path segment vector. Advancing involves identifying where the path segment vector intersects with the objective function. The process of iterative refinement is a nested procedure called during the "Advance Step" phase, aiding in pinpointing the exact cross-point of the path segment and the objective function. The temperature regulation segment defines the angle for the upcoming move and the optimal stride magnitude for the ensuing round. Presently, rudimentary linear temperature strategies are in use to determine the new ascent angle and stride dimension. Importantly, the linear temperature tactic has shown commendable results in the author's studies and insights.

The core functioning of BPO lies in five major sub-routines which as Initialization, Trajectory Creation, Progressive Stepping, Recursive Refining and Cooling Schedules.

Initialization: here, all the hyper-parameters are given values for example, elevation angles, number of rounds, number of steps for each round.

Trajectory Creation: A imaginary plane is created by choosing random points, elevation angles. Currently, the trajectory's plane is formulated by performing a cross product between a vector aligned with the y-axis and the trajectory segment vector.

$$\mathrm{y_{step}} = \sqrt{\frac{\sin^2\mathrm{a}\sum_{j=1}^{d} x_{step,j}^2}{1 - sin^2 a}}$$

Progressive Stepping: In each round, there are certain number of steps which helps in progressive stepping in the designed trajectory plane. It aids in determining the intersection of the crafted trajectory with the objective function.

Recursive Refining: Here, the author is evaluating the crossing point of designed trajectory and objective function and find the exact crossing point of both.

Cooling Schedules: in this step, new elevation angles and stepping size are determined for another round after the crossing point is identified in the recursive refining process. And as per the BPO paper, there has been promising positive results on using linear function cooling schedules.

BPO is an advancement in the domain of metaheuristic following single point approach and designed with an objective to determine globally optimized values for objective functions. BPO has been applied to continuous domain as of now and the horizon of non-continuous and discrete math's are still untouched in the BPO paper.

### Advantages

The advantages of the BPO algorithm include both its capacity to escape local optima through cooperative exploration and its propensity to explore complicated solution spaces.  The BPO technique is evidence of the inventive use of ideas from nature to resolve challenging optimization issues.

### Limitations

However, it is typically uncertain how the algorithm would behave when applied to very high dimensions like 10D, 20D etc.. The algorithm may even perform worse when applied to higher dimensions, according to preliminary data, especially when compared to population-based techniques like GWO and PSO.

BPO Pseudocode

Table 6 shows the pseudocode of BPO main function in which we have initialization of hyper parameter, initializing elevation angle, trajectory creating in while loop, and inside step while loop we have recursive refining. In Table 6.1 we have shown the pseudocode for crossing detected and Table 6.2 consist of the pseudocode for recursive refining sub-routines.

Table 6. Buggy Pinball Optimization Pseudocode

bp_main():

set setpSize = stepMax ;  a = amin ; $\#rounds_{max}$, $\#step_{max}$

$func(\cdot)$ #objective function

$\boldsymbol{x}, y$ 🡐 random initialization of starting solution

**while** r in $\#rounds_{max}$ $\boldsymbol{do}$:

$\boldsymbol{x}_{step}$= random(-1,1)

$y_{step} = \sqrt{sin^2 a \sum_{j=1}^{d} x_{step,j}^2 \Big/ 1 - sin^2 a}$

$\boldsymbol{x}_{step}$= z$\boldsymbol{x}_{step}$

$y_{step}$ = $zy_{step}$

**while** s in range($step_{max}$) **do**:

**if**(crossing_detected($\boldsymbol{x}$, $y$, $\boldsymbol{x}_{step}$, $y_{step}$, s) **then**:

$\boldsymbol{x}, y$ = recursiveRefining($\boldsymbol{x}$ $y$, $\boldsymbol{x}_{step}$, $y_{step}$, s)

break

a = update_ elevation_angle (amin, r, $\#rounds_{max}$)

stepSize = update_step_size_ (stepMax, r, $\#rounds_{max}$)

**return** $\boldsymbol{x}$,y

Table 6.1. Buggy Pinball Optimization Crossing Detection Pseudocode

```
find_the_crossing (x, y, x_step, y_step, j):
    a1 = y + (j-1) y_step – func (x + (j-1). x_step)
    b1 = y + j. y_step – func (x + j. x_step)
    return (a1 > 0 and b1 < 0) or (a1 < 0 and b1 > 0)
```

Table 6.2. Buggy Pinball Optimization Recursive Refining Pseudocode

```
recursiveRefining (x,y, x_step, y_step, j):
    x = x + x_step. j
    y = y + y_step. j
    if y – func(x) ≅ 0 then:
      return x, func (x)
    else
      x_step = x_step /2
      y_step = y_step /2
      x = x – x_step
      y = y – y_step
      if find_the_crossing (x, y, x_step, y_step, 0) then:
        return recursiveRefining (x, y , x_step, y_step, 0)
      else then:
        return recursiveRefining (x, y, x_step, y_step, 1)
```

# CHAPTER 3 – BLINDFOLDED SPIDERMAN OPTIMIZATION

The research work introduces an evolved iteration of the BPO approach and named "Blindfolded Spiderman". BSO is an extension of BPO in which, we are following the same BPO structure of designing virtual planes in search space and moving progressively with continuation steps. Unlike BPO, BSO do not have recursive refining, BSO works on the concept of memoization i.e., storing the most optimal solution achieved till now, until a better solution is encountered during the exploration of the search space. This intuition resembles the movement of spiderman.

## 3.1- Intuition Behind BSO and Pseudocode

What lies behind the choice of the moniker "Blindfolded Spiderman"? Spiderman's web-shooting endeavors are directed toward the highest building in proximity. Our new approach operates on a similar principle of trajectory formation. However, as it is impossible to know in advance which one is the highest building in proximity, in our approach spiderman is "blindfolded". As such, trajectories are created "blindly" and if they result to a higher building (i.e., a better solution than the current), they are kept. Otherwise, a new trajectory is generated. This pursuit is fueled by an amalgamation of exploration and exploitation, where the search space is thoroughly navigated and harnessed using the concept of memoization. In addition to this memoized concept, a significant departure from the original BPO algorithm is the omission of the iterative refinement process. This constitutes a notable advancement within the algorithm, streamlining its complexity and curtailing the number of required steps. Hence, heightened efficiency and reduced computational load. Our method can tackle complex continuous, non-continuous, and discrete problems.

Table 7 consist of the pseudocode for the main function of BSO where we have hyper parameter initialization, creation of trajectory plane and calling recursive refining and memoized refining sub-routines. In Table 7.1 we have the pseudocode for memoized refining, and Table 7.2 consist of the pseudocode for updating the trajectory steps until the terminating condition is fulfilled.

Table 7. Blindfolded Spiderman Pseudocode

Algorithm 1 BSO

1: **initialize:**

| | |
|---|---|
| $\Theta_{init}$ | #initial elevation angle (scalar) |
| $\Theta_{final}$ | #final elevation angle (scalar) |
| $i_{max}$ | #total number of trajectories (scalar) |
| i | #remaining number of trajectories (scalar) |
| $\lambda_{init}$ | #initial step size (scalar) |
| $\lambda_{init}$ | #final step size (scalar) |
| $s_{max}$ | #total number of steps (scalar) |
| s | #remaining number of steps (scalar) |
| $f(\cdot)$ | #objective function |
| d | #number of dimensions (scalar) |
| $\boldsymbol{domain_{min}}$ | #objective function domain lower limits (vector) |
| $\boldsymbol{domain_{max}}$ | #objective function domain upper limits (vector) |
| $\boldsymbol{x}$ | #initial points (vector) |

2: $y$ = f($\boldsymbol{x}$) #calculate y based on the initial points

3: **for** i in range(0, $i_{max}$) **do**

4: $\Theta = \Theta_{init} - \frac{i}{i_{max}}\left(\Theta_{init} - \Theta_{final}\right)$ # new elevation angle for ith trajectory

5: $\lambda = \lambda_{init} - \frac{i}{i_{max}}\left(\lambda_{init} - \lambda_{final}\right)$ # new step size for ith trajectory

6: $\boldsymbol{x}_{step}$ = random(-1,1) # random step component for each **x**

7: $y_{step} = \sqrt{\frac{sin^2\Theta \sum_{j=1}^{d} x_{step,j}^2}{1 - sin^2\Theta}}$ #Calculate y step according to $\boldsymbol{x}_{step}$ between [-1.1]

$$z = \frac{\lambda}{\sqrt{x_0^2 + x_1^2 + x_2^2 + \ldots + x_{d-1}^2 + y_{step}^2}}$$ #normalising factor as per function domain

8: $y_{step} = z y_{step}$

$y^t = y + y_{step}$ #hypothetical trajectory using normalized $y_{step}$ i.e $y^t$

9: $\boldsymbol{x}_{step} = z\boldsymbol{x}_{step}$ #normalization of $\boldsymbol{x}_{step}$ as per objection fun domain

$\boldsymbol{x}' = \boldsymbol{x} + \boldsymbol{x}_{step}$

$y' = f(\boldsymbol{x}')$ # actual $y'$ as per normalized $\boldsymbol{x}'$

10: is_underneath = False #keeps track of hypothetical $y^t$crosses the $y'$

11: if $y^t < y'$

is_underneath = True

12: **for** $s$ in range (0, $s_{max}$) **do**

#objective function crossed by trajectory setp

13: **if** is_underneath == True and $y^t >= y'$ **then**

is_underneath = False

14: **if** not is_underneath and $y' > y$ **then**

15: **break** #new solution worse than previous solution, stop the trajectory

16: $\boldsymbol{x}', y', y^t$ = update_steps($\boldsymbol{x}'$, $y'$, $\boldsymbol{x}_{step}$, $y_{step}$, $y^t$) #trajectory steps update

17: $\boldsymbol{x}$, $y$ = memoized_refining($\boldsymbol{x}'$ , $\boldsymbol{x}$, $y'$ , $y$)

18: **return** $\boldsymbol{x}$, $y$ # return the most optimized result

Table 7.1. Blindfolded Spiderman Memoized Refining

Algorithm 2 memoized_refining($\boldsymbol{x}'$ , $\boldsymbol{x}$, $y'$, $y$)

1. if $\boldsymbol{x}_i > \boldsymbol{domain}_i^{min}$ or $\boldsymbol{x}_i < \boldsymbol{domain}_i^{max}$or $y' > y$ for any i then
2. $\boldsymbol{x}' \leftarrow \boldsymbol{x}$ // return the most optimal memoized solution
3. $y' \leftarrow y$
4. return $\boldsymbol{x}'$ , $y'$

Table 7.2. Blindfolded Spiderman Update Trajectory Steps

| Algorithm 3 update_steps($\boldsymbol{x}'$`, $y'$`, $\boldsymbol{x}_{step}$, $y_{step}$, $y^t$): | | |
|---|---|---|
| 1: | $\boldsymbol{x}' = \boldsymbol{x}' + \boldsymbol{x}_{step}$ | **#next step** $\boldsymbol{x}'$ for current trajectory |
| 2: | $y^t = y^t + y_{step}$ | #stepping $y^t$ for current trajectory |
| 3: | $y'$= f($\boldsymbol{x}'$) | #$y'$ as per new $\boldsymbol{x}'$ for current trajectory |
| 4: | **return** $\boldsymbol{x}'$, $y', y^t$ | |

Pseudocode – Step by Step

BSO can be divided into five main sub routines which are Hyperparameter – Initialization, Trajectory Segment, Stepping Forward, Memorized Refining, Cooling Schedules All these subroutines have designated task to do and each one is related to one another as follows:

Hyperparameter - Initialization: One the most important sub-routine of the BSO, because we are initializing the hyper parameters used in BSO(i.e., Table 7 lines 1-2).

We have a set of hyper parameters which decides the behaviour of BSO. Now selecting values for these parameters depends on the objective function, dimensionality of the problem we will be working on and need respective tuning every time. Original step size and angle of elevation i.e., initial angle($\Theta_{init}$ ), final angle($\Theta_{final}$) are initialized. The initialization i.e., start resultant($\lambda_{init}$) , end resultant($\lambda_{final}$) of the step size should be done cautiously to avoid taking overly large steps in the search domain. With the expansion of the search domain, it's prudent to allocate larger values to the step size to ensure a harmonious balance between exploration and utilization. Via several experiments on BSO and BP author observation, for minimization - step size(λ) is given negative value, for maximization- step size is given positive value. For elevation angle(Θ), tt has been observed that if we begin with a value very close to 0, we will have a decent horizontal plane almost perpendicular to y axis and the chances of avoiding local optima increases. Hence resulting in horizontal spiderman trajectory. Unlike step size, elevation angle is kept quite low close enough to zero with increases in configuration space. As the search space increases, we reduce elevation angle to have more horizontal movement, to avoid several local optima. The $i_{max}$ dictates the number of spiderman maximum trajectories, and the number of steps ($s_{max}$) establishes the strides in each trajectory. Depending on the time limitations for the algorithm's intended runtime, the number of rounds can be maximized, ensuring optimal time utilization. Furthermore, the number of steps can be set randomly as low as possible depending on the optimization function. In lines 1-2 of Table 7, we are initializing starting points(x,y) randomly, although a suitable initial point can surely speedup the convergence rate and time, however, we can avoid local optima easily so starting from random points also works just like BP.

Trajectory Segment: Each round generates a new spiderman-trajectory where a trajectory is created and then executed from the previous optimal solution points and the continues to execute until the number of steps are completed or there is an encounter of much worse solutions and for the former and latter case, we either return the new most optimal solution point or the memoized solution from previous trajectories. In BSO, to ensure best exploration in the search space, at the beginning of each round randomization is implemented to set the trajectory direction with elevation angle being fixed. Following which, we set up a random step component for each variable which lies in between [-1,1]. For y-axis, with respect to the elevation angle, we set stepping component using following equation (Table 7 line 7):

$$y_{step} = \sqrt{\frac{sin^2\Theta \sum_{j=1}^{d} x_{step,j}^2}{1 - sin^2\Theta}}$$

The above equation is achieved as follows: As we already have details about elevation angle which is

$$sin\Theta = \frac{|\boldsymbol{n} . \boldsymbol{u}|}{|\boldsymbol{n}|\ |\boldsymbol{u}|}$$

In the above equation,"n" represents a parallel vector to y axis (0,0,0...1)and perpendicular to the fundamental plane and "u" represents the vector of trajectory segment ($x_0, x_1, x_2, x_3 \ldots . x_{d-1}, y$). As we know elevation angle and ($x_0, x_1, x_2, x_3 \ldots . x_{d-1}$) we can solve for y which in turn gives us the equation of calculating $y_{step}$. After determining the direction of trajectory, we also need to normalize the steps (line 7-9) with respect to each dimension and with respect to the previously calculated overall step size(λ). This is done by following equation (Table 7, line 7-9) where the normalization factor(z) is calculated as:

$$z = \frac{\lambda}{\sqrt{x_0^2 + x_1^2 + x_2^2 + \ldots . + x_{d-1}^2 + y_{step}^2}}$$

Once all this is done then we are ready to implement to exploit the designed trajectory space as per the pre-defined number of steps or until a worse solution is found. To explore the designed hypothetical plane, we have Stepping Forward and finally returning the most optimum solution from Memoized Refining sub-routine.

Stepping Forward: In this subroutine (Table 7, line 12-16), we began exploring the designed trajectory by applying steps value on each axis, the process of stepping continues until stopping conditions are encountered which are either the number of steps ends, or we encounter a worse solution then the memoized one. As we proceed in each step, we move downwards with respect to the y axis and y values decrease in each step. We compare the new encountered value($y'$) of y-axis with memoized one(y), and, if we encounter worse ($y'$) and the $y'$ value has crossed the hypothetical $y^t$ function value then the sub-routine returns the memoized result otherwise, these new optimized values are accepted which is closer to global optima. However, if the condition is not fulfilled then the loop will terminate when we complete the number of steps.

Memoized Refining: This is the second last subroutine of the algorithm (Table 7 line 17). This is called at the end and is utilized to return the most optimal results if we go out of bounds in the last step of stepping forward subroutine or the most recent stepping result is worse than the memoized result(i.e., calculated in last step → (s > (smax -1))). This is a normal refining to maintain the core concept of memoization in the BSO.

Conditions of refining are as follows:

- if the final resulting $\boldsymbol{x}$ values in stepping forward outside the function space (>$\boldsymbol{domain_i^{max}}$ and <$\boldsymbol{domain_i^{min}}$) than return memoized values

    or

- if the final resulting $\boldsymbol{x}_i \in [domain^{min}, domain^{max}]$but the resulting $y'$ value is worse than the memoized results (y) return y

Cooling schedules: This is another subroutine of BSO (Table 7, line 4-5). This is the point where the BSO calculates the new trajectory angle and reduced step for the new trajectory. Here, we have used linear cooling schedules like used in the BP approach. There can be various approaches for implementing cooling schedules like using logarithmic or exponentially reducing cooling schedule. The linear cooling strategy has proven effective for every optimization challenge we've examined to date. However, further investigation would be beneficial. Consequently, the concluding numbers are a fraction of 1 for step length, and from 0.1* up to 89* depending on the complexity, ranging from intricate issues to straightforward ones without local optima.

# CHAPTER 4 – EXPERIMENTS AND CASE STUDY

Each test was carried out on an identical machine to ensure unbiased comparisons. These identical machines were set up on cloud and servers were Linux containers. The configuration of the cloud server was, a processor with 40-CPU Intel(R) Xeon(R) E5-2680 v2, clocked at 2.80 GHz and the servers had 64GB RAM. To have a detailed comparative analysis standard metaheuristics were used which can be classified as Single Point and Particle Based:

Single Point:

- Simulated Annealing
- Threshold Accepting
- Buggy Pinball Optimization

Particle Based:

- Particle Swarm Optimization
- Whale Optimization
- Grey Wolf Optimization

## 4.1 - Optimization Problems

To perform the experiment, we have used standard continuous functions , non-continuous function and one discrete problem named as unbounded knapsack. And all the experiments are performed on multiple dimensions which are: 2D, 3D, 4D, 5D, 6D.

### 4.1.1 Continuous Functions [19]

The standard continuous optimization problems we have used in our work to perform comparative analysis between the algorithms are shubert, eggholder, dropwave, holdertabable, langermann, easom, ackley, rastrigin, schwefel. These are the same continuous functions used by BPO author. To ensure the scalability of the algorithms we have used these optimization problems in higher dimesions(D) which are 2D, 3D, 4D, 5D and 6D. Table 8 shows the continuous functions definitions and their corresponding 3D Plots

Table 8. Continuous Functions Definitions And 3D plots

| Name | Equation | 3D Plots |
|---|---|---|
| Dropwave | $-\frac{1+\cos(12\sqrt{x_1^2+x_2^2})}{0.5(x_1^2+x_2^2)+2}$ | 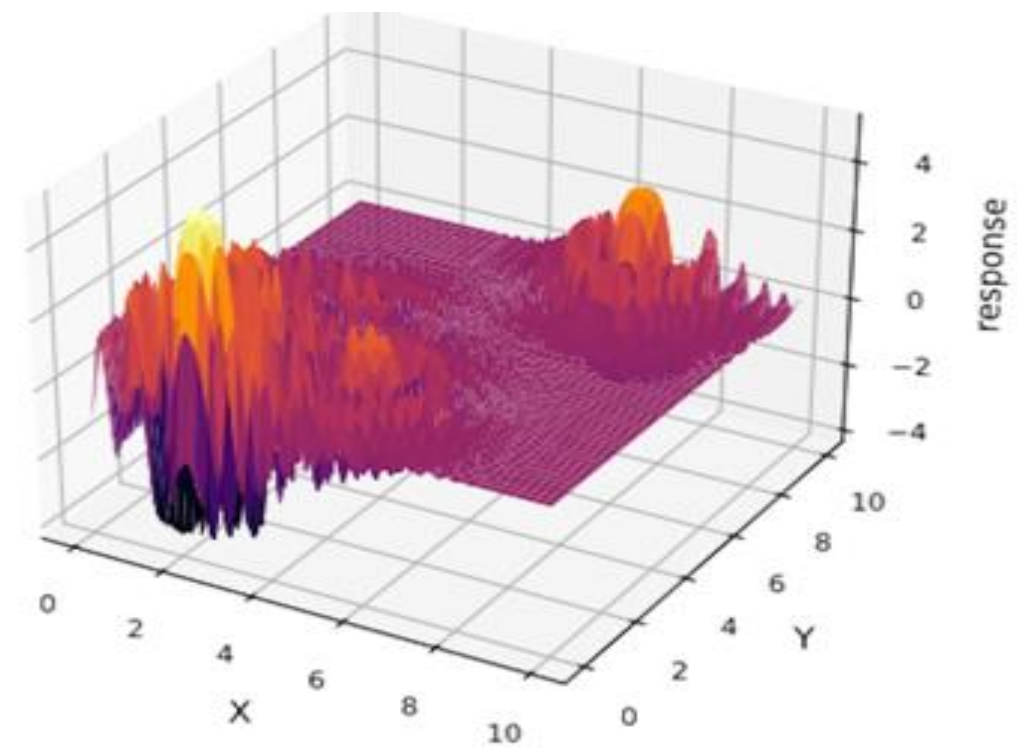 |
| Eggholder | $-x_1\ \sin\sqrt{\lvert x_1-x_2-47\rvert}$ | 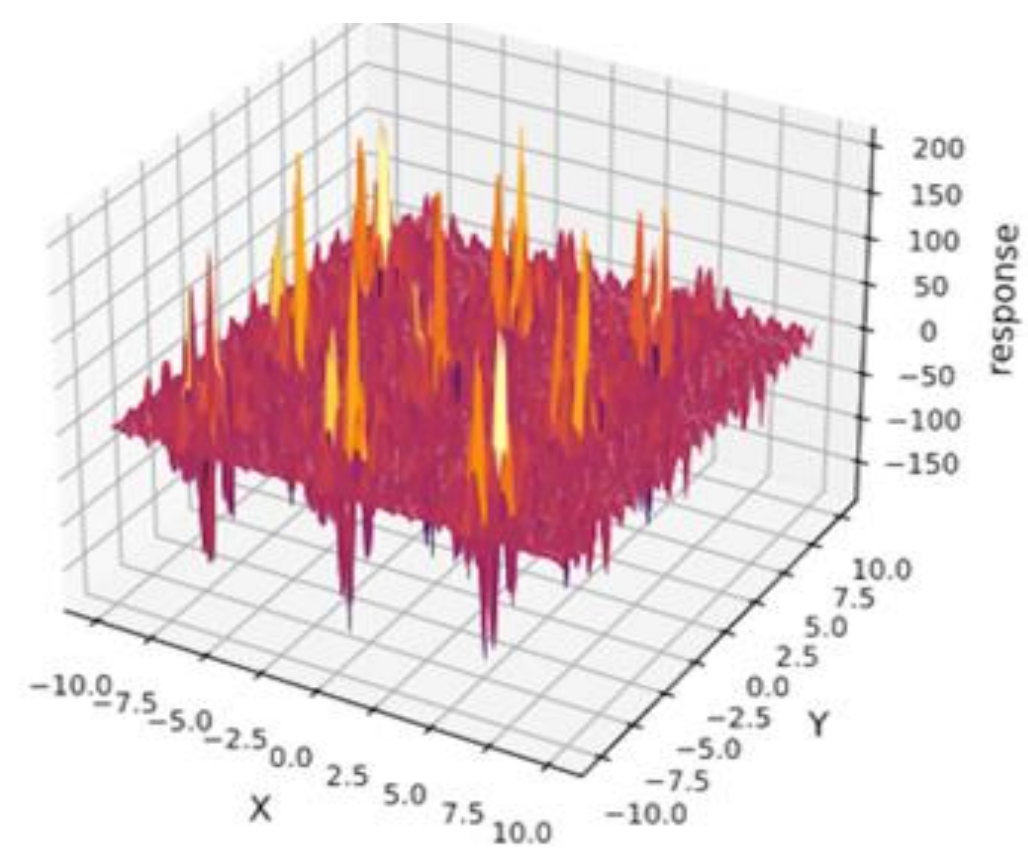 |

| Name | Equation | 3D Plots |
|---|---|---|
| Holder table | $-\left\|\boldsymbol{sin(x_1)cos(x_2)e^{\left\|1-\frac{\sqrt{x_1{}^2+x_2{}^2}}{\pi}\right\|}}\right\|$ | 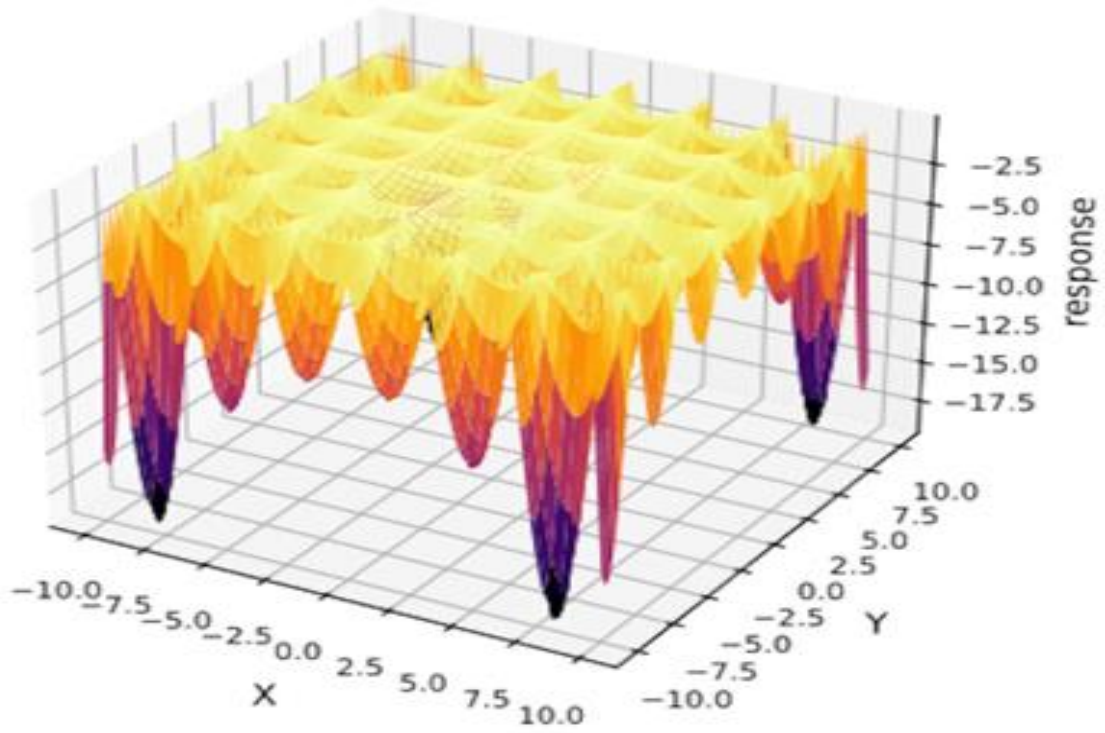  |
| Langermann | $\sum_{i=1}^{5} c1_i\, e^{-\frac{1}{\pi}\sum_{j=1}^{d}(x_j-AA_{ij})^2}\cos\left(\pi\sum_{j=1}^{d}(x_j-AA_{ij})^2\right)$ | 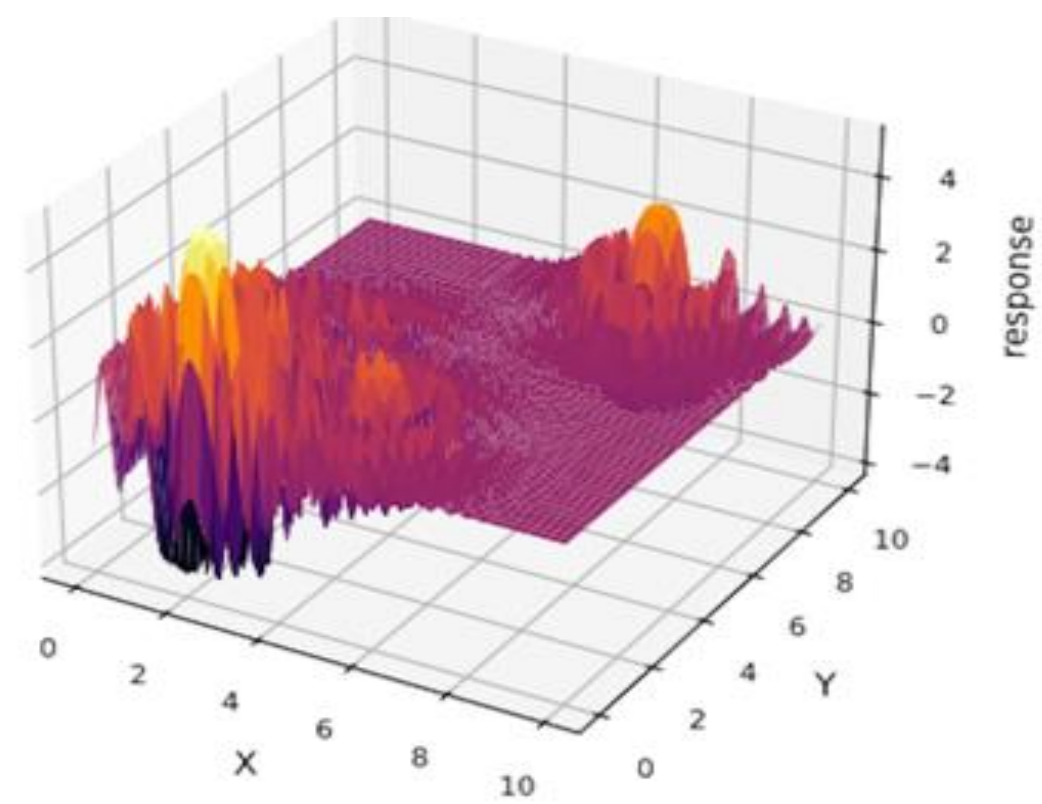  |
| Shubert | $\left(\sum_{i=1}^{5} i\cos((i+1)x_1+i)\right)\left(\boldsymbol{\sum_{i=1}^{5} vicos((i+1)x_2+i)}\right)$ | 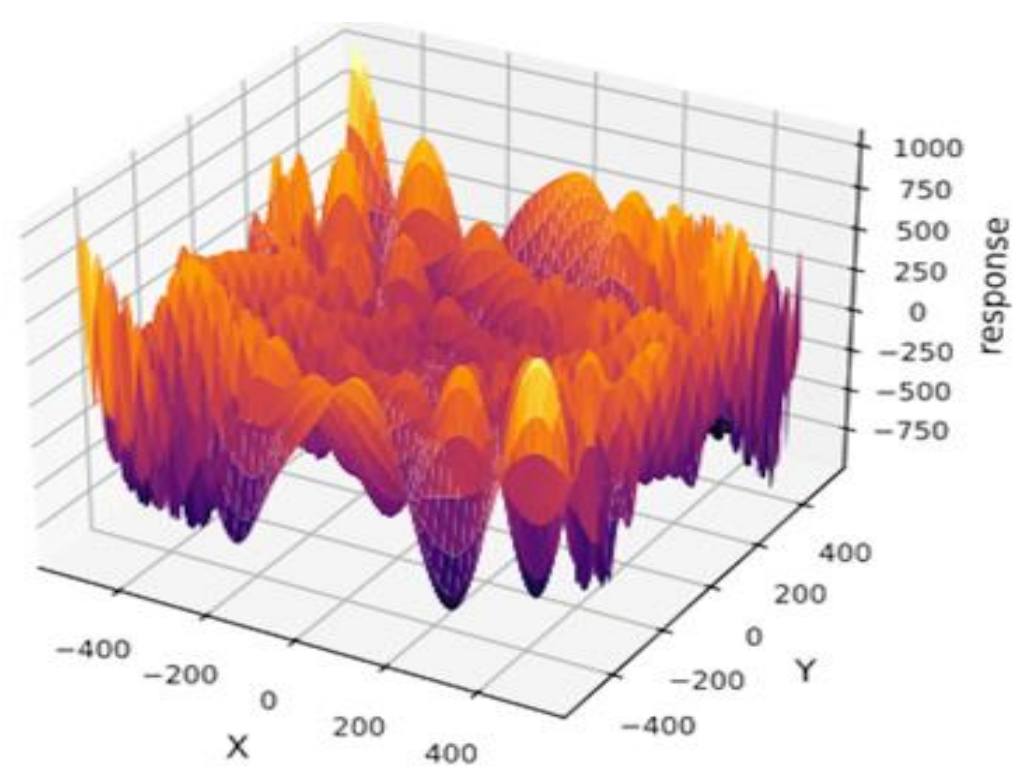  |

| Name | Equation | 3D Plots |
|---|---|---|
| Easom | $-\cos(x_1)\cos(x_2)e^{-(x_1-\pi)^2-(x_2-\pi)^2}$ | 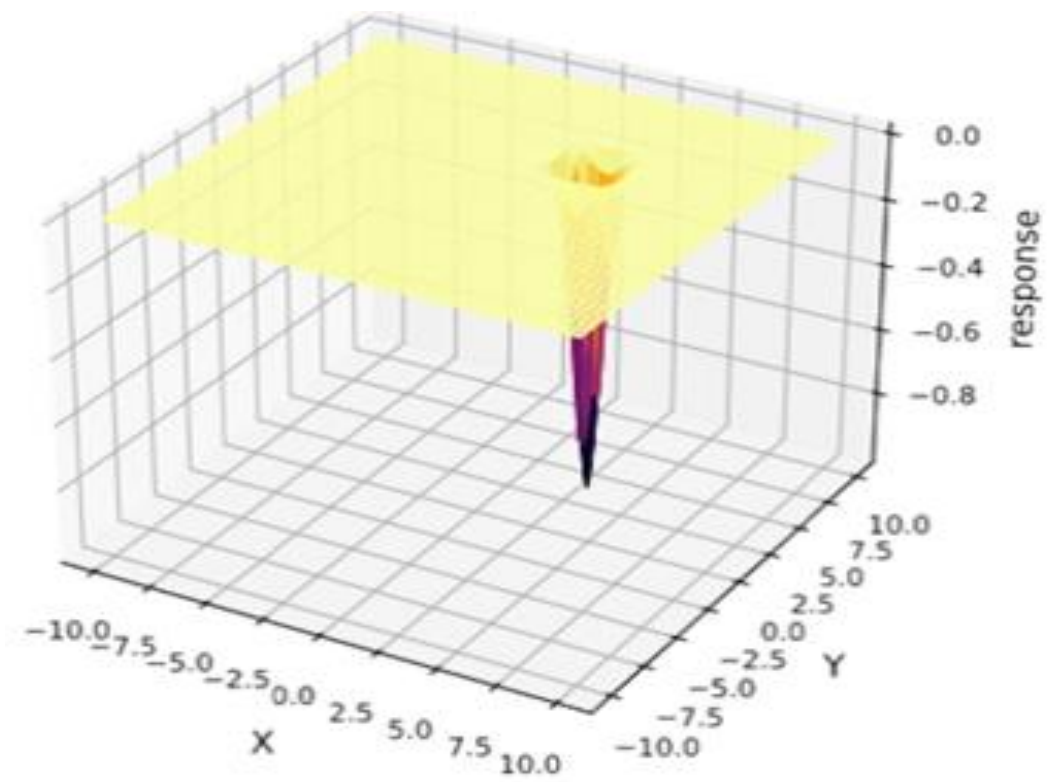 |
| Ackley | $-20e^{-0.2\sqrt{\frac{1}{d}\sum_{i=1}^{d} x_i^2}} - e^{\sqrt{\frac{1}{d}\sum_{i=1}^{d} \cos(2\pi x_i)}} + 20 + e^1$ | 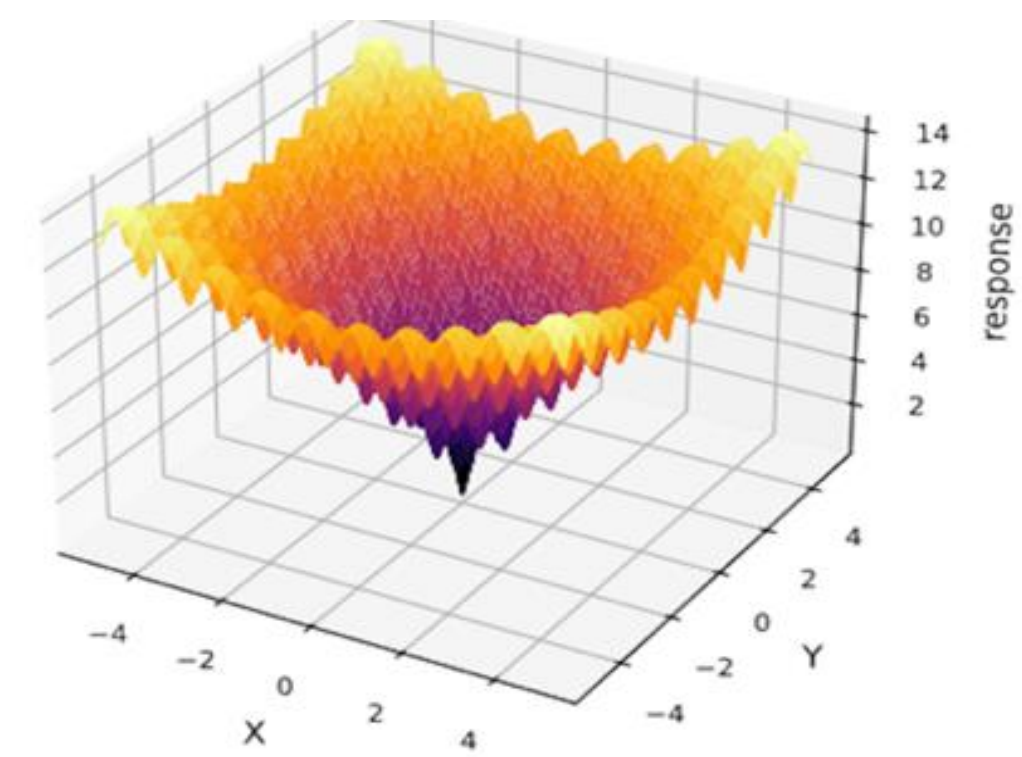 |
| Rastrigin | $10D + \sum_{i=1}^{d}\left(x_i^2 - 10\cos(2\pi x_i)\right)$ | 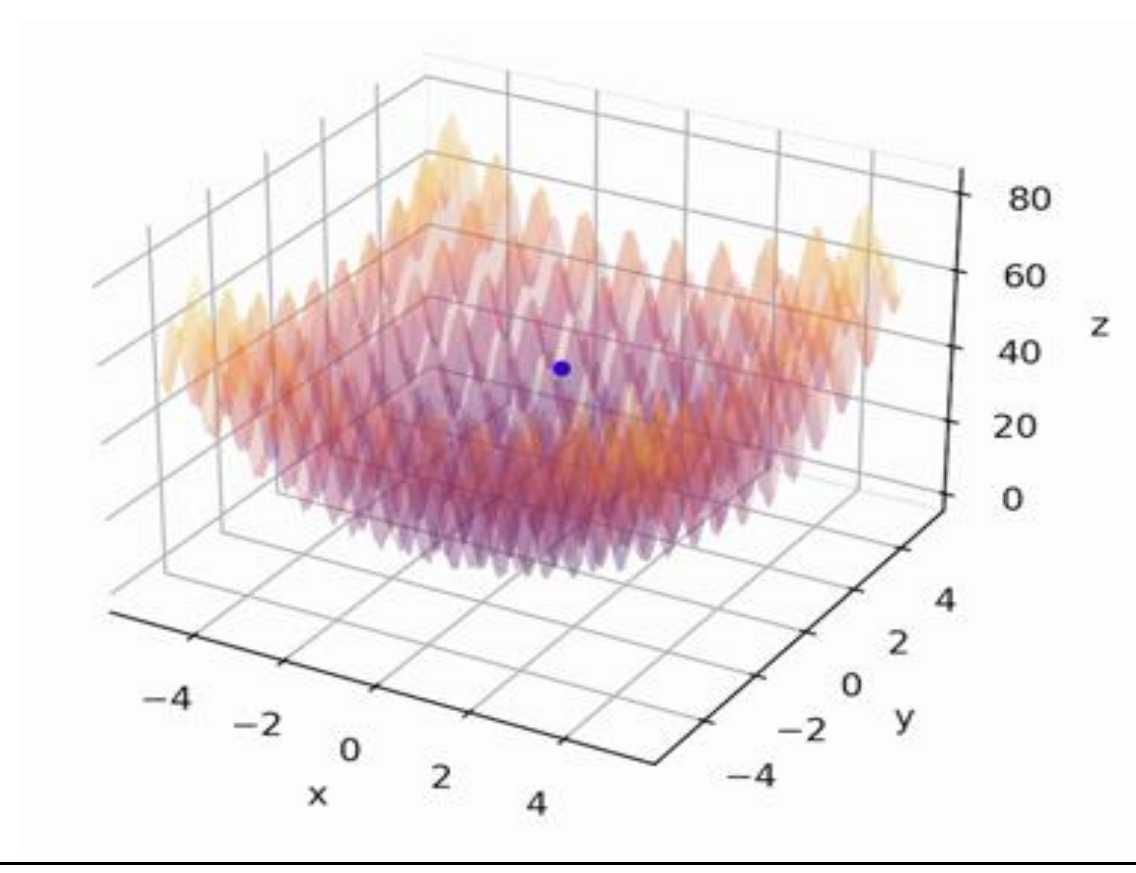 |

| Name | Equation | 3D Plots |
| --- | --- | --- |
| Schwefel | $418.9829d - \sum_{i=1}^{d} x_i \sin\left(\sqrt{\lvert x_i \rvert}\right)$ | 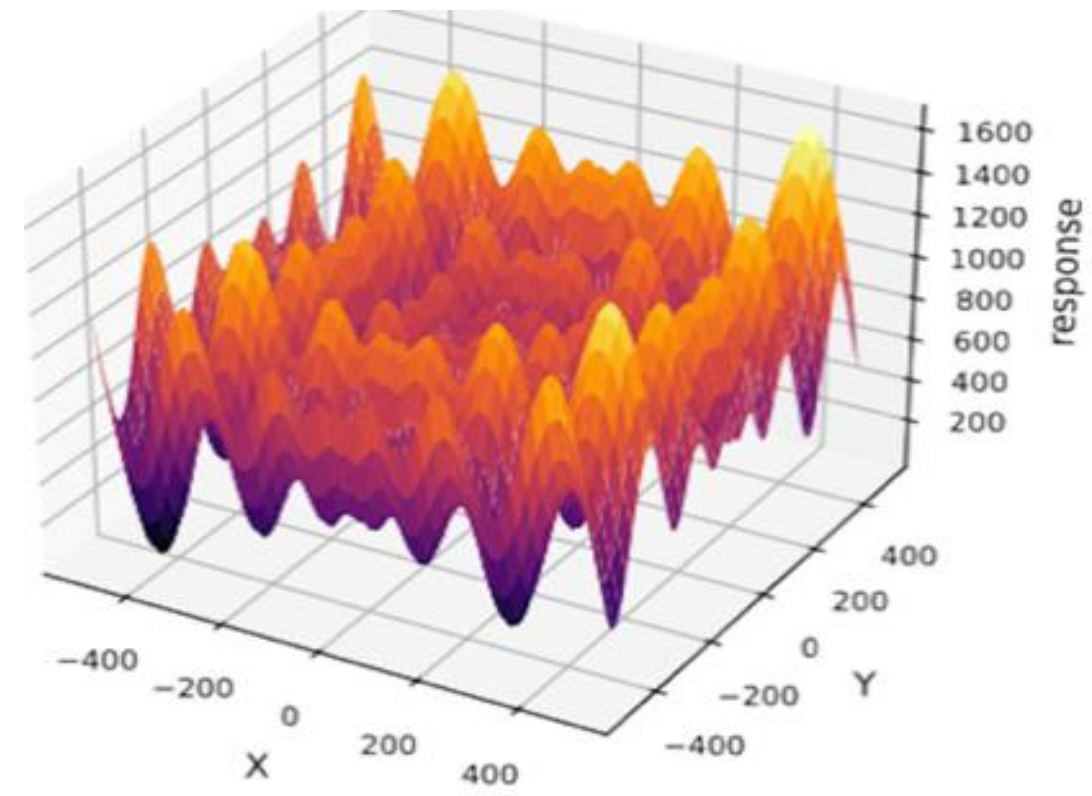 |
| Sphere | $\sum_{i=1}^{d} x_i^2$ | 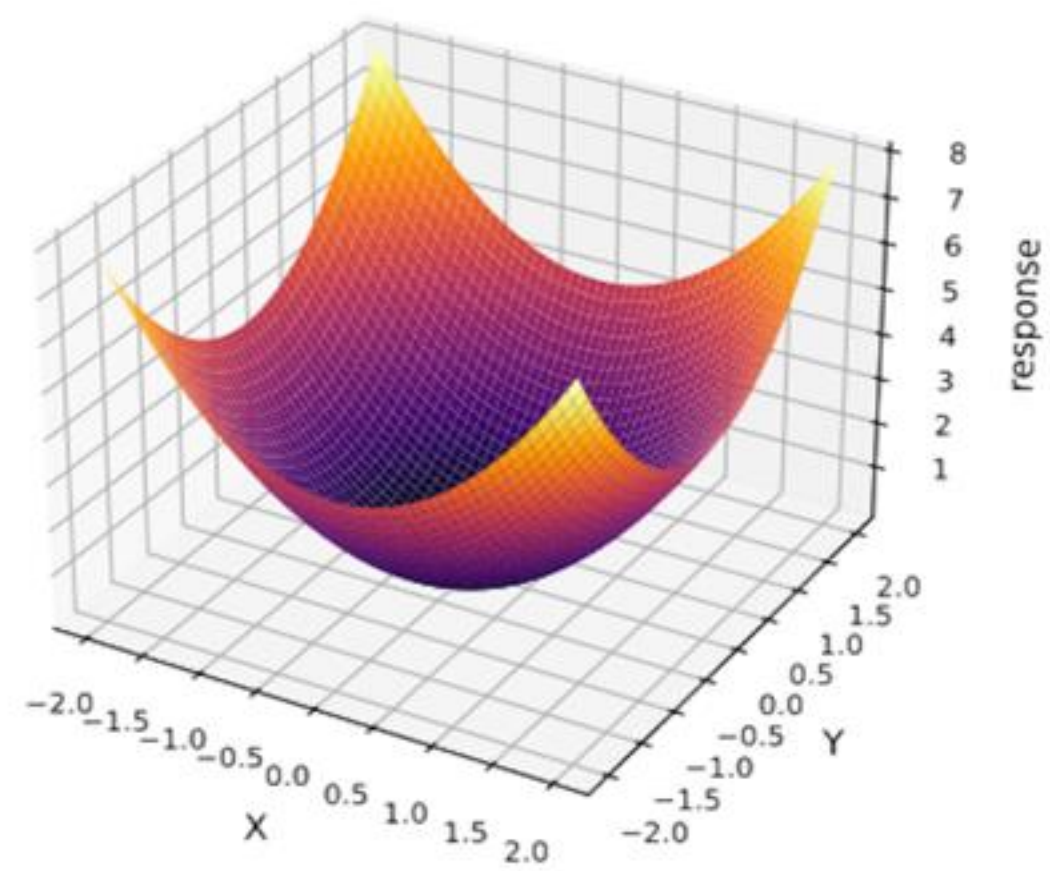 |

4.1.2 Non-Continuous Functions

The non-continuous optimization problems we have used in our work are rastrigin [20], ellipsoid [21], rosenbrock [21], quadric [21], xin-she-yang n2 [22], and step function [23]. The motive of using piecewise non-continuous problem is the fact that non-continuous optimization problems are specific case of a continuous optimization problem[24]. All the optimization problems have been considered in higher dimension which 2D, 3D, 4D, 5D, 6D. Table 8 shows the definitions of non-continuous optimization problems and their corresponding 3D plots.

Table 9. Non-Continuous Functions And 3D plots

| Name | Equations | 3D Plots |
|---|---|---|
| Ellipsoid | $\begin{array}{ll} \sum_{i=1}^{n} \frac{1}{1.1} \quad 2^{i-1}\, x^2(i) + \frac{1}{n}, & if\ sin\left(2\sum_{j=1}^{n} x(j)\right) > 0.5 \\ \sum_{i=1}^{n} 1.1 * 2^{i-1}\, x^2(i) + \frac{1}{n}, & if \sin\left(2\sum_{j=1}^{n} x(j)\right) < 0 \\ \sum_{i=1}^{n} 2^{i-1}\, x^2(i)\,, & if\ 0 \le \sin\left(2\sum_{j=1}^{n} x(j)\right) \le 0.5 \end{array}$ | 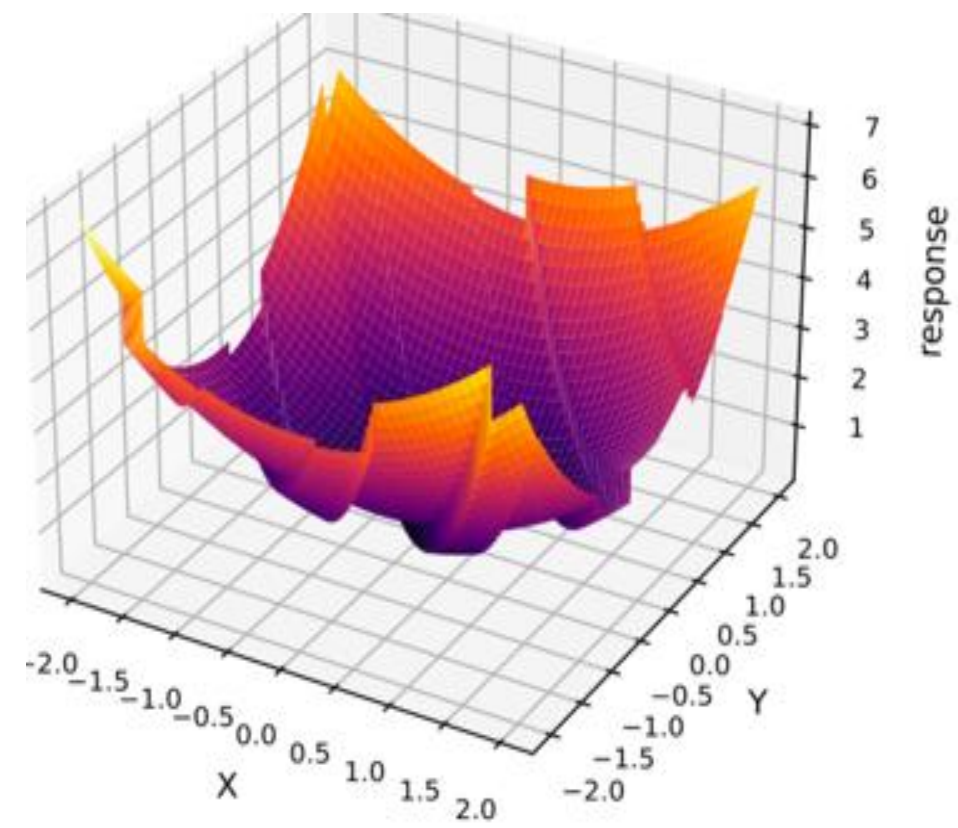 |
| Quadric | $\begin{cases} \sum_{i=1}^{n}\left(\sum_{j=1}^{i} x(j)\right)^2, & if\ sin\ (8\lVert x \rVert) > 0.5 \\ \sum_{i=1}^{n} 1.2\left(\sum_{j=1}^{i} x(j)\right)^2, & if\ sin\ (8\lVert x \rVert) < -0.5 \\ \sum_{i=1}^{n} \frac{1}{1.2}\left(\sum_{j=1}^{i} x(j)\right)^2, & if - 0.5 \le \sin(8\lVert x \rVert) \le 0.5 \end{cases}$ | 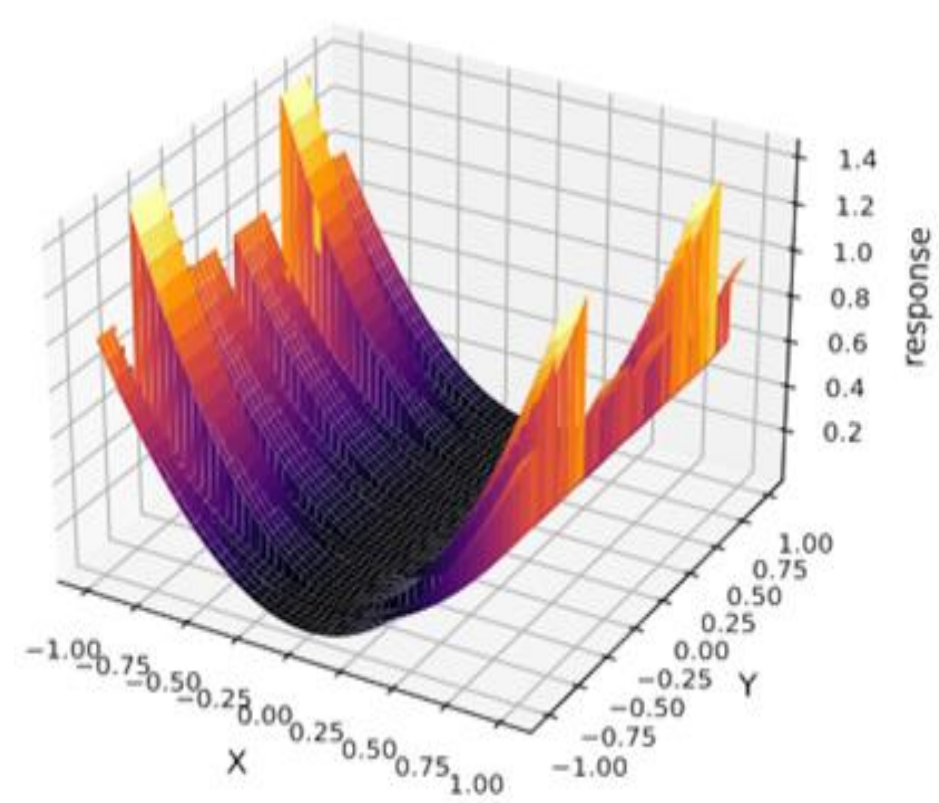 |
| Rosenbrock | $\begin{cases} \sum_{i=1}^{\frac{n}{2}} \frac{\mathbf{1}}{\mathbf{1.2}}\left(\mathbf{100}\left(x(2i) - x^2(2i-1)\right)^2 + \left(1 - x(2i-1)\right)^2\right), \\ \quad if\ 0 \le \sin\ (2\lVert x \rVert) < 2/3 \\ \sum_{i=1}^{\frac{n}{2}} \mathbf{1.2}\left(\mathbf{100}\left(x(2i) - x^2(2i-1)\right)^2 + \left(1 - x(2i-1)\right)^2\right), \\ \quad if -\frac{2}{3} \le \sin(2\lVert x \rVert) < 0 \\ \sum_{i=1}^{\frac{n}{2}} \left(\mathbf{100}\left(x(2i) - x^2(2i-1)\right)^2 + \left(1 - x(2i-1)\right)^2\right), \\ \quad if -\frac{2}{3} > \sin(2\lVert x \rVert) \ge 2/3 \end{cases}$ | 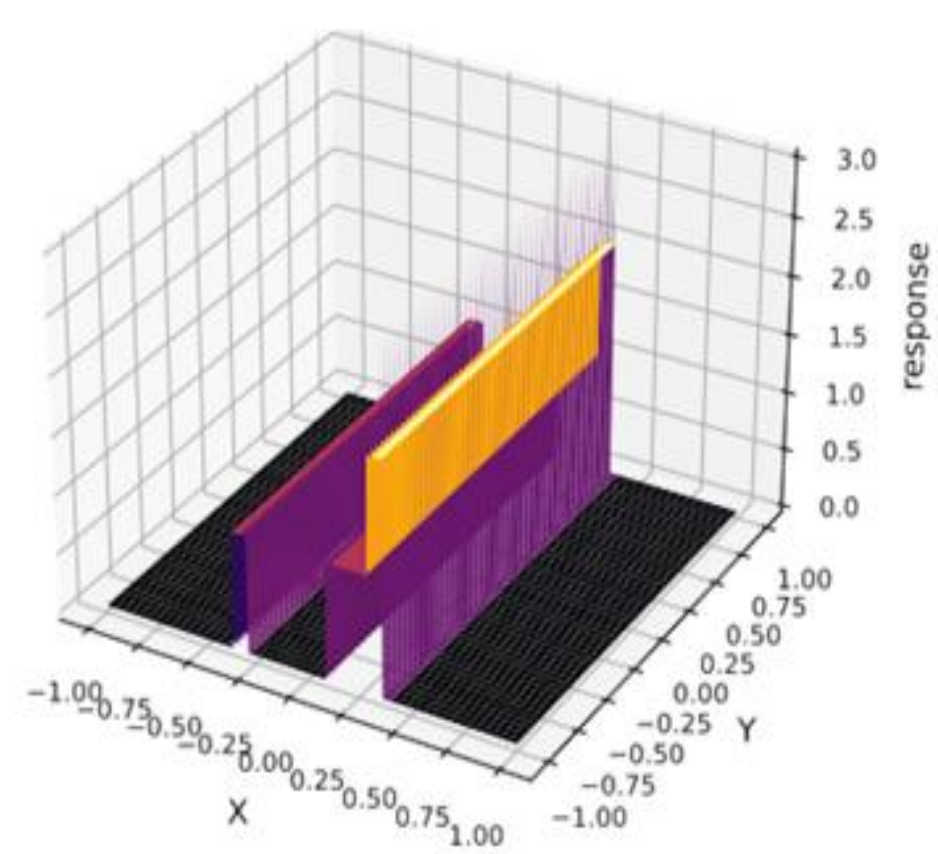 |

| Name | Equations | 3D Plots |
|---|---|---|
| Xin she yang n2 | $\sum_{i=1}^{d}\lvert x_i\rvert\, exp\left(-\sum_{i=1}^{d} sin(x_i^2)\right)$ | 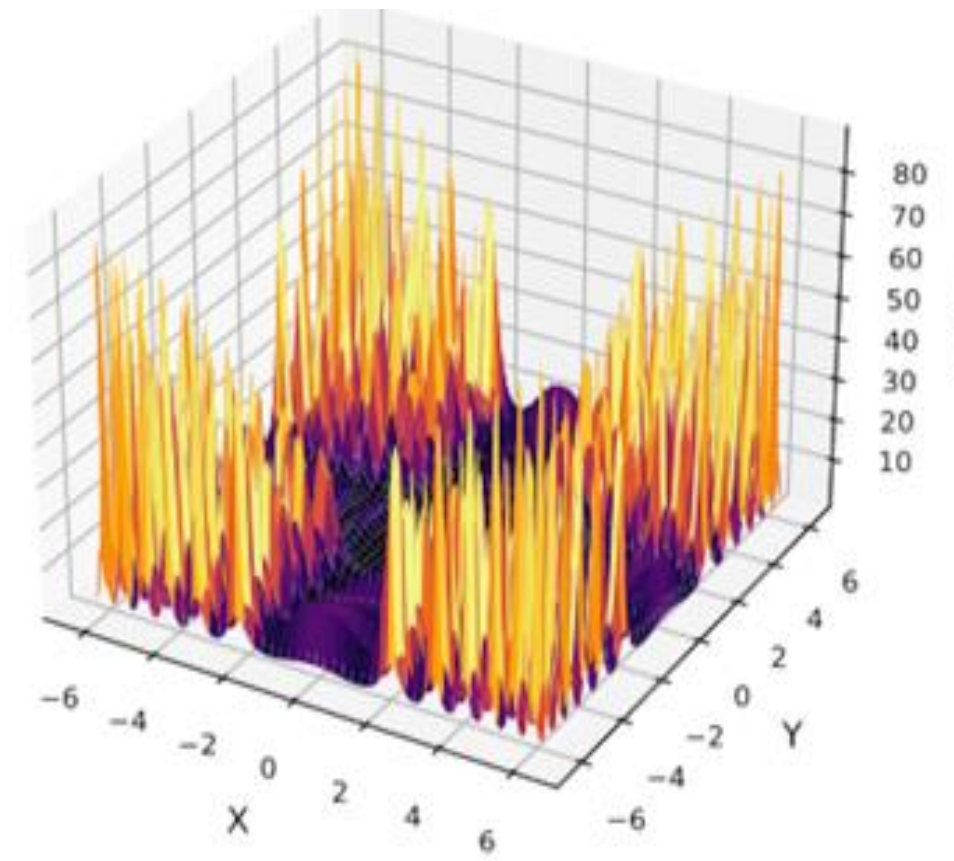 |
| Step fun | $\sum_{i=1}^{d}\{floor(\lvert x_i\rvert\})$ | 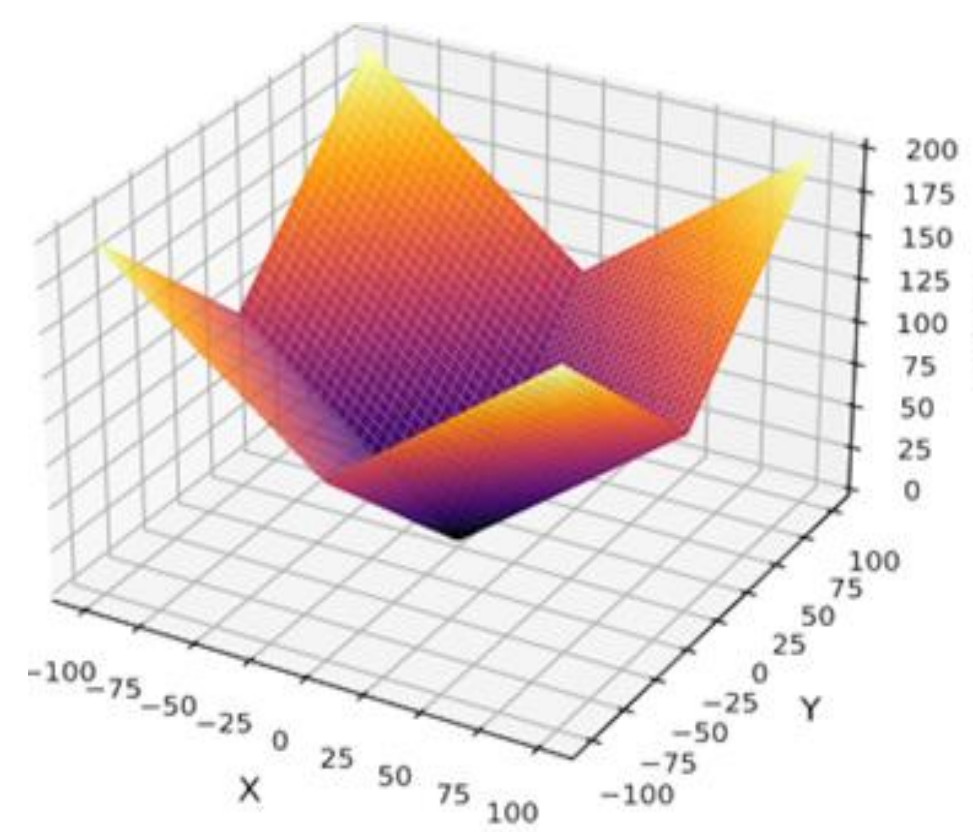 |
| Rastrigin | $\left\{\sum_{i=1}^{d}(y_i^2 - 10cos(2\pi y_i) + 10)\right.$ <br> $y_j = \begin{cases} x_j, & \lvert x_j\rvert < \frac{1}{2} \\ \frac{round(2x_j)}{2}, & \lvert x\rvert \geq \frac{1}{2} \end{cases}$ for j=1,2,3…d | 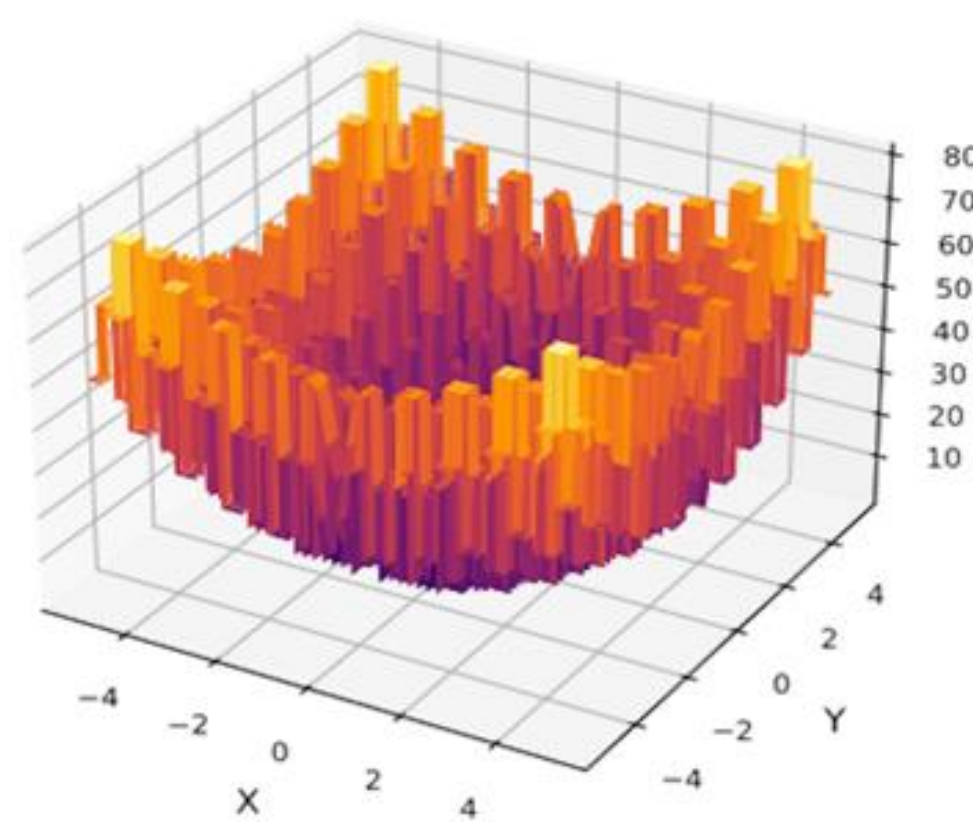 |

## 4.1.3. Discrete Problem: Unbounded Knapsack

A combinatorial problem is one that entails selecting an ideal or workable item from a limited number of discrete options while taking into consideration certain limitations or requirements. Counting, enumeration, and optimization techniques are frequently needed to tackle combinatorial issues.

Knapsack [25] is a classic example of combinatorial optimization, where we choose a subset of items with highest profit and minimal weight, while keeping in mind a capacity restriction. Finding the ideal solution to the knapsack issue may take a very long time for big cases since it is an NP-hard problem.

Knapsack problems Implementation: Knapsack may be used in real-world applications to solve many different challenges. The knapsack problem or its derivatives can be employed to strategize for diverse practical challenges, including inventory minimization, project choice dilemmas, allocation issues, and financial budgeting concerns [2].

Goal: Find the most valuable selection of items, given number of items where each individual item have values and weights, and a maximum weight constraint.

Implementation: So, in our work we have used the concept of unbounded knapsack. As the name suggests, an unbounded knapsack item can occur any number of times until the weight is less than or equal to knapsack bag weight.

Assumption:

- n : total number of items
- W : total weight of knapsack
- $w_i$ : weight of each item
- $x_i$ : represents the weight of each item , subject to $x_i$ >=0 and $x_i$ are integers
- $v_i$ belongs to $\{v_1, v_2, v_3 \ldots v_n\}$

- n: total # of items
- $v_i$ : each item value i.e., ith item
- d: number of dimensions

To implement multidimensional knapsack, we have assumed that each item represents one dimension and as we increase the number of items the corresponding number of dimensions will be increased (Fig. 3).

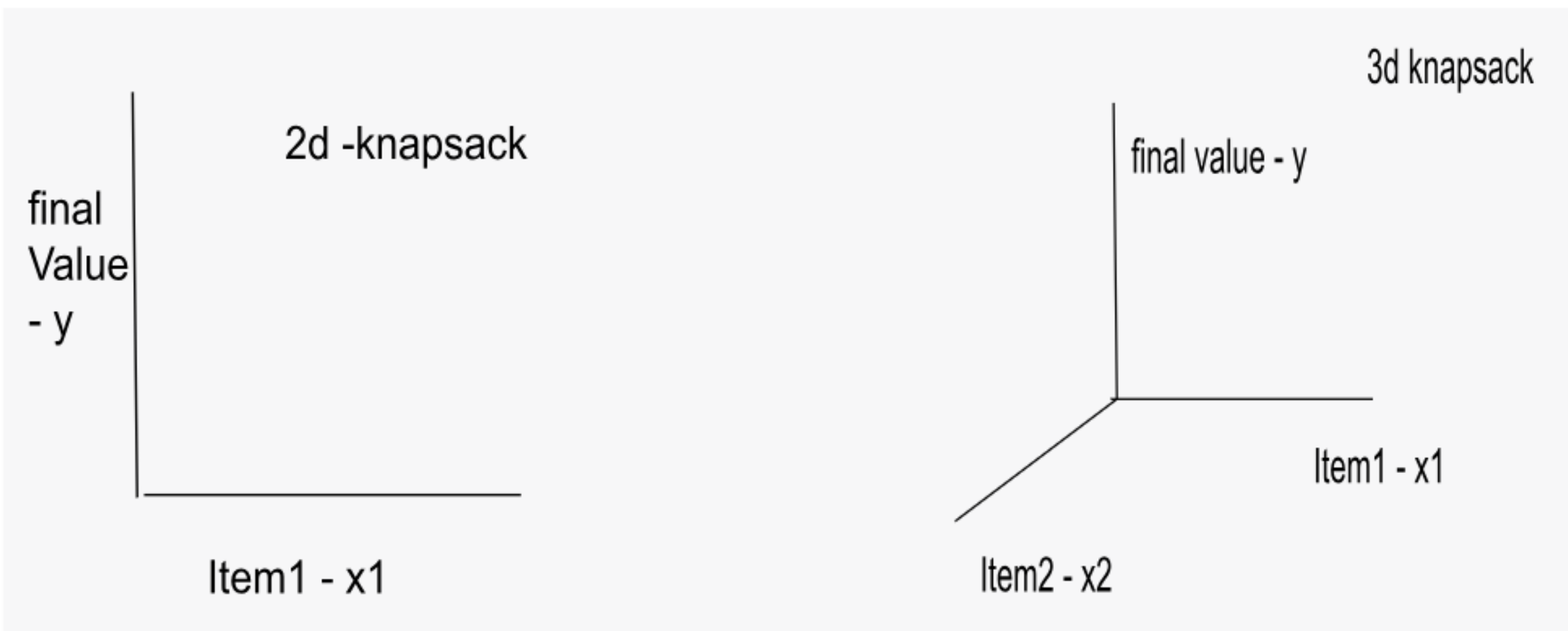

Fig. 3. Knapsack representation for items and final cost in multi-dimensions

Following the above assumption the formula to the unbounded knapsack has been used:

Mathematical formula for unbounded knapsack:

- optimal_value[$w'$] = max($\sum_{1}^{n} v_i * x_i$) subject to $\sum_{i=1}^{n} w_i * x_i < w'$ and $x_i$ > 0
    - Note the characteristics of m[W] as listed below [22]:
- optimal_value[0] is 0 (representing the total of no items, essentially, the addition of an empty set).

- optimal_value[W] equates to the maximum of ($v_i$ + optimal_value [W-$w_1$], $v_2$ + optimal_value [W-$w_2$], ..., $v_n$ + optimal_value [W-$w_n$] , given $w_i$ < W),
  - with $v_i$ representing the value of the item of type i.

In our implementation $x_i$ represents the current most optimal solution

Use of "rounding off " Operator in Metaheuristics Algorithms for Unbounded Knapsack

The rounding off operator transforms $R_n$ feasible solution into $Z_n$ feasible solution [26]. Due to its simplicity and less processing cost, this method is one of the most used methods for handling discrete variables. Its foundation is the practice of rounding to the next whole number. So, in each iteration, the possible answer is subjected to a rounding off near integer operation. As a discretization technique, it was initially used [27] to optimize reactive power and voltage management.

Likewise, all metaheuristics are usually designed for global continuous optimization problems, where the test function is a real vector and real-valued function. The knapsack challenge is a distinct optimization dilemma, wherein the target function is quantified as an integer, stemming from a binary vector.

So, following the “rounding off” strategy, we created variants of metaheuristics to run experiments on knapsack and explore the boundaries of the discrete problems. However, there are certain disadvantages to this strategy, and the chance that the answer is in an infeasible region is one of this method's drawbacks. Additionally, the fitness value in the rounded point may change significantly from the value in the original point.

## 4.2 - Case Studies

There were 100 trials for each experiment and the predetermined time was selected before the experiment began to avoid any bias and have a fair comparison with all the approaches. We ran the experiments on pre-allocated time. Table 10, Table 11, Table 12, Table 13 shows four different case studies for performing comparative analysis between the meta heuristics, in detail:

Table 10 shows Case Study 1, time allocation per trial for continuous and non-continuous functions with respect to each dimension and meta heuristics. This pre-allocated time is less than pre-allocated time by BPO author. This has been done to validate the fast convergence of BSO. Table 11 shows the Case Study 2, time allocation for non-continuous functions to perform comparative analysis between all the meta heuristics. Whereas Table 11.1 shows the Case Study 2, time allocation for continuous optimization problems for BPO and BSO. Time selection of Case Study 2 is as per BPO author. Table 12 shows Case Study 3, time allocation for discrete optimization problem and Table 13 shows the time allocation(as per BPO author) in Case Study 4 for 3D comparison of both continuous and non-continuous functions. In Case Study 4, six different times slots were chosen, and all the optimization problems mentioned in Table 8 and Table 9 were tested with 100 trials for each algorithm per-time slot.

Table 10. Case Study 1 Time Allocation for Continuous and Non-Continuous functions

| Dimension | Time Per Trial | Metaheuristic Algorithms |
|---|---|---|
| 2D | 0.005 second | SA, TA, PSO, WO, GWO , BPO, BSO |
| 3D | 1 second | SA, TA, PSO, WO, GWO , BPO, BSO |
| 4D | 15 second | SA, TA, PSO, WO, GWO , BPO, BSO |
| 5D | 2 minutes | SA, TA, PSO, WO, GWO , BPO, BSO |
| 6D | 10 minutes | SA, TA, PSO, WO, GWO , BPO, BSO |

Table 11. Case Study 2 Time Allocation for Non-Continuous Functions

| Dimension | Time Per Trial | Metaheuristic Algorithms |
|---|---|---|
| 2D | 1 second | SA, TA, PSO, WO, GWO, BPO, BSO |
| 3D | 5 second | SA, TA, PSO, WO, GWO, BPO, BSO |
| 4D | 1 minutes | SA, TA, PSO, WO, GWO, BPO, BSO |
| 5D | 5 minutes | SA, TA, PSO, WO, GWO, BPO, BSO |
| 6D | 20 minutes | SA, TA, PSO, WO, GWO, BPO, BSO |

Table 11.1. Case Study 2 Time Allocation for Continuous Functions

| Dimension | Time per Trial | Metaheuristic Algorithms |
|---|---|---|
| 2D | 1 second | BPO, BSO |
| 3D | 5 second | BPO, BSO |
| 4D | 1 minutes | BPO, BSO |
| 5D | 5 minutes | BPO, BSO |
| 6D | 20 minutes | BPO, BSO |

Table 12. Case Study 3 Time Allocation for Discrete Unbounded Knapsack Optimization

| Dimension | Time Per Trial | Metaheuristic Algorithms |
|---|---|---|
| 2D | 0.005 second | SA, TA, PSO, WO, GWO, BPO, BSO |
| 3D | 1 second | SA, TA, PSO, WO, GWO, BPO, BSO |
| 4D | 15 second | SA, TA, PSO, WO, GWO, BPO, BSO |
| 5D | 2 minutes | SA, TA, PSO, WO, GWO, BPO, BSO |
| 6D | 10 minutes | SA, TA, PSO, WO, GWO, BPO, BSO |

Table 13. Case Study 4 Time Allocation for Non-Continuous Functions

| Dimension | Time Per Trial | Metaheuristic Algorithms |
|---|---|---|
| 3D | 0.1second | SA, TA, PSO, WO, GWO , BPO, BSO |
| | 0.2sec | SA, TA, PSO, WO, GWO , BPO, BSO |
| | 0.5 sec | SA, TA, PSO, WO, GWO , BPO, BSO |
| | 1 sec | SA, TA, PSO, WO, GWO , BPO, BSO |
| | 2 sec | SA, TA, PSO, WO, GWO , BPO, BSO |
| | 5 sec | SA, TA, PSO, WO, GWO , BPO, BSO |

## 4.3 - Evaluation Metrics

Among all the metaheuristics compared in our experiments SA, TA, BP, BSO are single point whereas WO, PSO, GWO are particle based. We are trying to compare our approach with both single point and particle-based approaches and all of them work differently and have different set of hyperparameters [28] so ensuring fairness becomes top-most priority and a big challenge while selecting appropriate time.

### 4.3.1- Fair Comparison

A fair comparison suggest that we should not have biasness in our experiments and all algorithms should be given fair and equal chance to perform. So, with respect to the respective hyperparameters for each meat-heuristic approach we followed the same level of optimization/calibration. Moreover, for every-benchmark function in each dimension we performed an exhaustive search to find the best hyper-parameters in a predetermined time frame. The selected dimension range mentioned in case studies in section 4.2 allows us to evaluate the scalability of our approach in comparison to other metaheuristics and each dimension search was done in per-determined time with optimal hyper-parameters. And the same fairness of comparison is done for analysis of the knapsack implementation. After the completion of experiments the obtained results are compared on appropriate comparative metrics which are Accuracy and Precision.

**Accuracy** is determined by the proportion of instances where the corresponding metaheuristic approaches an estimate of the global optimal minimum relative to the count of all the trials.

To select an approximation of global minimum, we selected the results which are better than local minima considered as second-best for the respective continuous and non-continuous optimization functions.

$$\text{accuracy} = \frac{\text{\#of trials where algorithm converges to approximate global optima}}{\text{total \# of trials}}$$

**Precision** is calculated as the difference between the solution which converged to the approximation of the global minimum and the actual global minimum itself.

Hence, mean absolute error is calculated as

Precision (MAE) is calculated:

$$\text{MAE} = \frac{\sum_{i=1}^{n} |y - y_i|}{n}$$

- n : total # of trials
- y : global optima obtained from experiment
- $y_i$: actual global optima for that dimension

4.3.2 Statistical Significance Tests

Moreover, we have proven the results and our hypothesis statistically by using statistical significance tests. A Shapiro-Wilk Test, using a 0.05 significance threshold, suggested that considering a normal distribution (details in the appendix) is not appropriate. For evaluating the results' statistical importance, we applied the Kruskal-Wallis H test, followed by Conover's tests (details in the appendix). The step-down approach was utilized, accompanied by a Bonferroni modification for adjusting the p-value. A 0.05 threshold was selected to ascertain statistical relevance. Comprehensive statistical results for Case Study 1 have been provided in Appendix G and for Case Study 2 has been provided in Appendix H.

Corresponding Quantile-Quantile(Q-Q) plots to validate that the result data sets are not normally distributed, and the plots has been provided for Case Study 1 in Appendix I, for Case Study 2 in Appendix J, and for Case Study 4 in Appendix K. We have shown the plots for both continuous and non-continuous optimization problems for each case study.

# CHAPTER 5 -RESULTS

## 5.1- Evaluation Summary

BSO algorithms is designed to locate the global optimum in complex optimization problems. The results show that BSO has constantly achieved accuracy levels that are either to or on par with all the considered benchmark metaheuristics across various scenarios. A notable feature of BSO is that, irrespective of its randomized starting point in each trial, the BSO consistently hones on the global optima. And BSO stands out as one of the single-point algorithms that most frequently reaches absolute 100% accuracy rate when compared to the benchmarked single-point meta heuristics and even gave competitive accuracy results with particle-based metaheuristics.

Discussing precision, it is important to note that we measure it exclusively for instances where the algorithm moves toward a near estimate of the global minimum. In essence, high precision underscores the algorithm's effectiveness in closely mirroring the global optimum, given it has not been trapped by a local optimum. However, a scenario where an algorithm boasts good precision, but less high accuracy is suboptimal for global optimization tasks. This is because, despite being precise, the algorithm may not consistently identify the global optima. Nevertheless, ensuring a commendable level of precision remains crucial for global optimization methods.

In our studies, BSO's precision is either to or on par to single-point algorithms, BPO, SA and TA, in most of the scenarios. Further statistical analyses validate BSO's almost superior precision over SA and TA in every tested case and has shown on-par precision when tested against BPO. However, when comparing BSO to the WO, GWO – another algorithm tailored for continuous optimization tasks – BSO's precision appears to be marginally lesser. Whereas, when compared with PSO, BSO have shown more precision in many of the test scenarios. Subsequent tests have corroborated that this difference in precision between BSO, PSO, WO, GWO are indeed statistically significant.

Appendix A, Table A1 shows the accuracy results of Case Study 1 continuous optimization problems, and correspondingly Table A2 shows precision results. From the achieved results it can be inferred that BSO has shown an exceptionally 100% accuracy results in all the continuous optimization test cases except schwefel 6D. However even for Schwefel-6Dimension BSO performs well when compared to other single point metaheuristics like SA and TA.

Appendix B, Table B1 shows the accuracy results of Case Study 1 non-continuous optimization problems, and correspondingly Table B2 shows precision results. From the achieved results it can be inferred that BSO has shown 100% accuracy in all optimization problems except the non-continuous rastrigin functions. But, comparatively even in non-continuous rastrigin BSO has given a tough competition to all other metaheuristics and performed well in comparison to Single Point. Although practical based approaches have an upper hand, and GWO, WO has achieved 100% accuracy even for non-continuous rastrigin where all other single point metaheuristics were struggling.

Appendix C, Table C1 shows the accuracy results of Case Study 2 continuous optimization problems, and correspondingly Table C2 shows precision results. In this case study we are doing comparison for the allotted time between BP and BSO for continuous functions only because, author of BP has already published a detailed comparative study of BP for continuous optimization problem mentioned used in this research work. From the results achieved it is evident that BSO have shown 100% accuracy in all the tested continuous optimization problems.

Appendix D, Table D1 shows the accuracy results of Case Study 2 non-continuous optimization problems, and correspondingly Table D2 shows precision results. It is evident from the results, BSO has performed exceptionally well in comparison to single point and particle-based metaheuristics by achieving 100% accuracy in all problems except non-continuous rastrigin. Even though, BSO has not achieved 100% accuracy in non-continuous rastrigin like WO, GWO, but BSO has better accuracy than PSO (which is a particle-based approach) and even single point metaheuristics.

Appendix E, Tables E1 to E5, are showing the Case Study 3 unbounded-knapsack statistical significance results with non-parametric kruskall-wallis test followed by convors test and the used p value is 0.05. Here, we can see that the obtained p-values from convors test are less than 0.05 and hence we can reject the null-hypothesis. This suggests that there may be data supporting the idea that the groups under study differ from one another and we were able to obtain the optimal cost in all the dimensions.

For unbounded knapsack, we could not perform the accuracy and precision metric because of the randomness in generating Item {weights and value} for each trial for each dimension. Due to 100 trials for each dimension and a total of six dimensions we were having 600 randomly generated items with value and weights for seven metaheuristics in total, and the random selection of Items also helps in avoiding any kind of biasness (which may occur due to selection of an Item which become favorable to any benchmark metaheuristics ). Hence, due to this large dataset, it became difficult to figure out the optimal solution for each trial and hence we stick to statistical significance test instead of accuracy and precision. Moreover, all the metaheuristics ran for same-time-per-trial.

Appendix F shows the result for Case Study 4. There are four tables which are labeled as Tables F1, F2, F3 and F4. Each table shows result for accuracy and precision as follows:

- Table F1, shows the continuous optimization functions(Table 8) comparative results for accuracy

- Table F2, shows the continuous optimization functions(Table 8) comparative results for precision results.
- Table F3 , shows the non-continuous optimization functions(Table 9) comparative results for accuracy
- Table F4 shows the non-continuous optimization functions(Table 9) comparative results for precision results.

For Table F1 and table F3, we can see that BSO has shown almost 100% accuracy in all the tested scenarios except some complex optimization problem like langermann, eggholder, schwefel. But the accuracy is still better than BPO for langermann, 1-2% less than BPO in eggholder and almost equal accuracy for schwefel optimization problem. Still BSO performs better than other single point SA and TA and almost equal to swarm optimization (WO, GWO).

In Appendix L, we have shown final optimal value achieved by metaheuristics in unbounded knapsack problems for 100 trials as mentioned in Case Study 3. The details for Tables L1, L2, L3, L4, L5, L6 and L7 are as follows:

- Table L1 shows the optimal value achieved by BSO in 2D, 3D, 4D, 5D, 6D.
- Table L2 shows the optimal value achieved by BPO in 2D, 3D, 4D, 5D, 6D.
- Table L3 shows the optimal value achieved by SA in 2D, 3D, 4D, 5D, 6D.
- Table L4 shows the optimal value achieved by TA in 2D, 3D, 4D, 5D, 6D.
- Table L5 shows the optimal value achieved by PSO in 2D, 3D, 4D, 5D, 6D.
- Table L6 shows the optimal value achieved by WO in 2D, 3D, 4D, 5D, 6D.
- Table L7 shows the optimal value achieved by GWO in 2D, 3D, 4D, 5D,6D.

5.5.1- Data Set calculation

In Case Study 1and 2, we have 16 optimization problems as mentioned in Table 8 and Table 9. For each optimization problem we perform 100 trials, and we have 7 metaheuristics. So, the total number of data points are generated as follows:

6(total number of dimensions) . 100(trials) . 16(continuous and non-continuous optimization problems) . 7(metaheuristic) = 67,200 data points

67,200 . 2(number of case cases studies) = 134,400 total data point

In Case Study 3, we have 6 dimensions, 100 trials in each dimension and 7 metaheuristics, So, the total number of data points generated are:

6(total number of dimensions) . 100(trials) . 1(optimization problems) . 7(metaheuristic) = 4,200 data points

In Case Study 4, we have 16 optimization problems as mentioned in Table 8 and Table 9, for each optimization problem we perform 100 trials, and we have 7 metaheuristics. So, the total number of data points are generated as follows:

6(total number of different timeslots) . 100(trials) . 16(optimization problems) . 7(metaheuristic) = 67,200 data points.

Here, data points represent number of solutions generated for each case study to perform comparative analysis between metaheuristics. All the tests between BSO, BPO, SA, TA, PSO, WO, GWO are verified statistical test mention in section 4.3.2 and are indeed statistically significance.

## 5.2 - BSO Trajectory Visualization

In this context, we present step-by-step trajectory plots corresponding to the BSO algorithm applied to continuous Rastrigin, Ackley, non-continuous Rastrigin, and step-function optimization problems. These plots illustrate the progression of optimization through various stages of the BSO process. It is apparent that the algorithm navigates from one highly optimal point to an even better solution, eventually culminating in achieving the desired optimization goal. You will see a blue line inside the plots that blue line represents the trajectory path followed by the BSO and each dot on blue line represents the most optimal solution for that trajectory. In subsections 5.2.1, 5.2.2, 5.2.3, 5.2.4, mentioned are intermediatory plots capturing the trajectories from starting point to almost near to optimal solution.

### 5.2.1- Continuous Rastrigin Trajectory Visualization - BSO

The following step by step sequence of figures shows the trajectory created by BSO while optimizing the continuous rastrigin optimization problem in 3D. If we have "n" total number of trajectories, then:

- trajectory 1: starting point(Fig. 4)
- trajectory 2 to trajectory n-4 to trajectory n-1 - intermediate trajectories (Fig. 5 to Fig. 10)
- trajectory n - final solution(Fig. 11)

Trajectory begun from step1 and then start optimizing by locating the next optimal solution and then begun another trajectory from the most recent optimal solution.

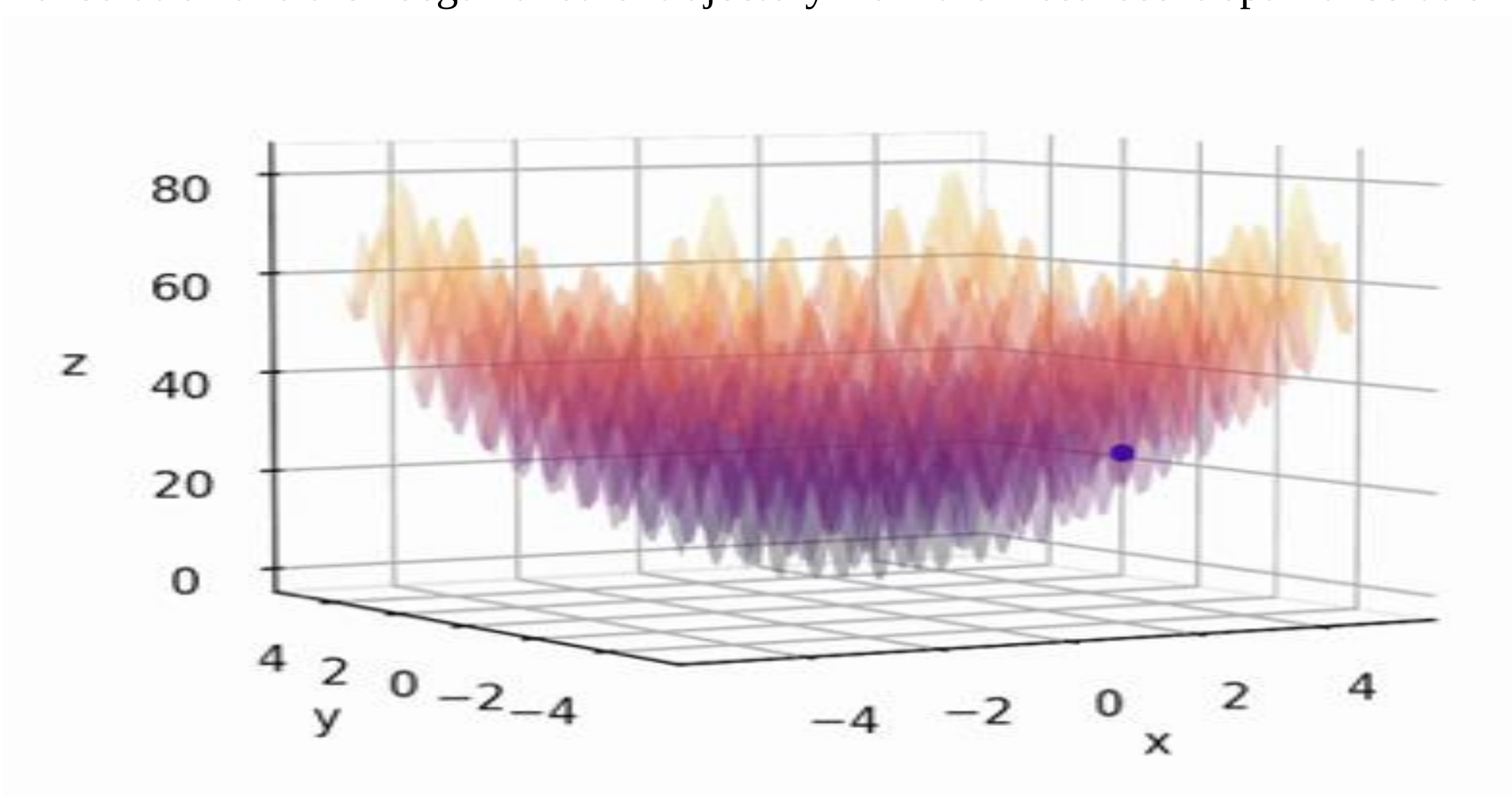

Fig. 4. BSO Starting Point(Trajectory1) For Continuous Rastrigin Functions

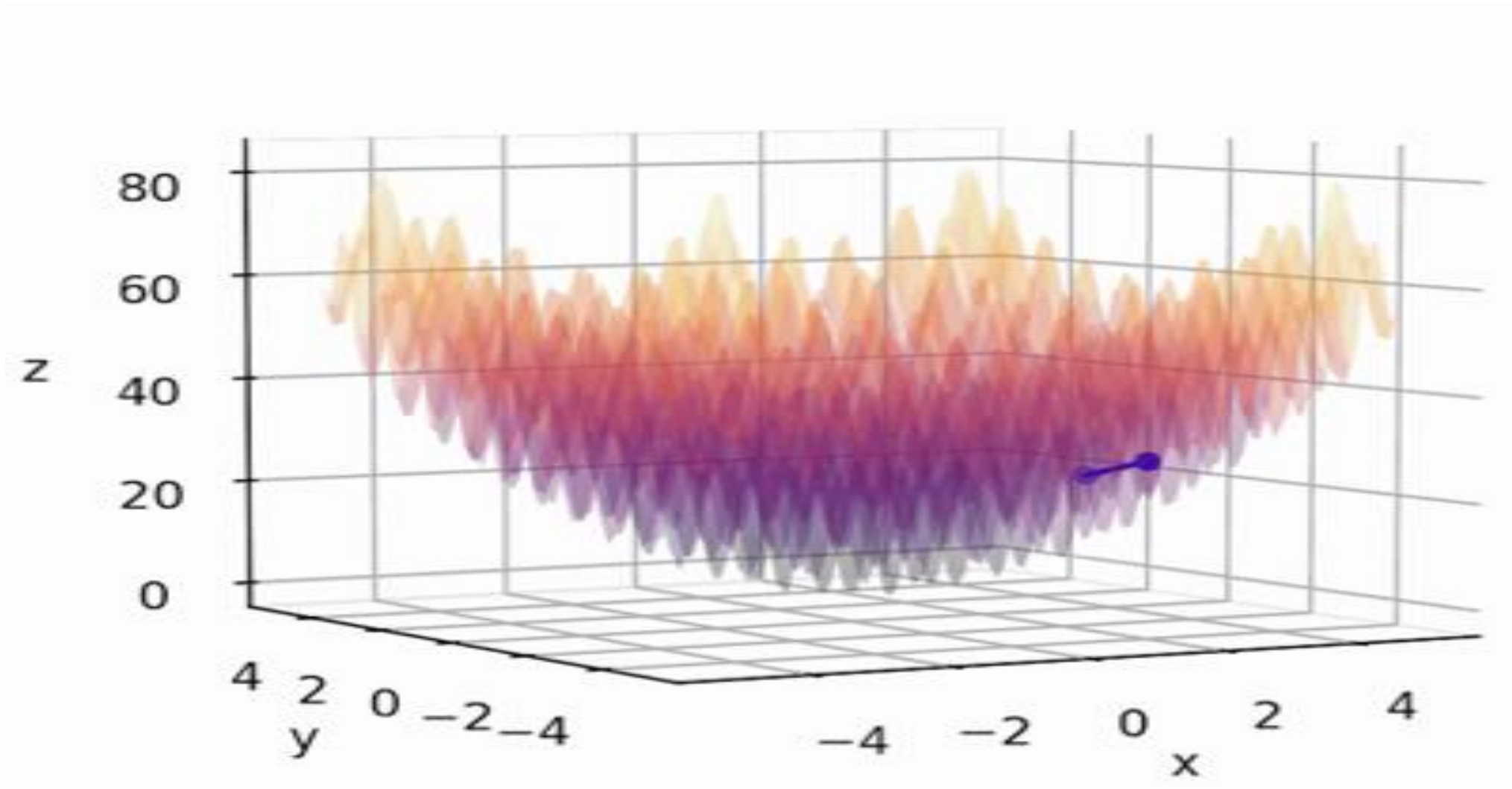

Fig. 5. BSO Intermediatory Point(Trajectory2) For Continuous Rastrigin Functions

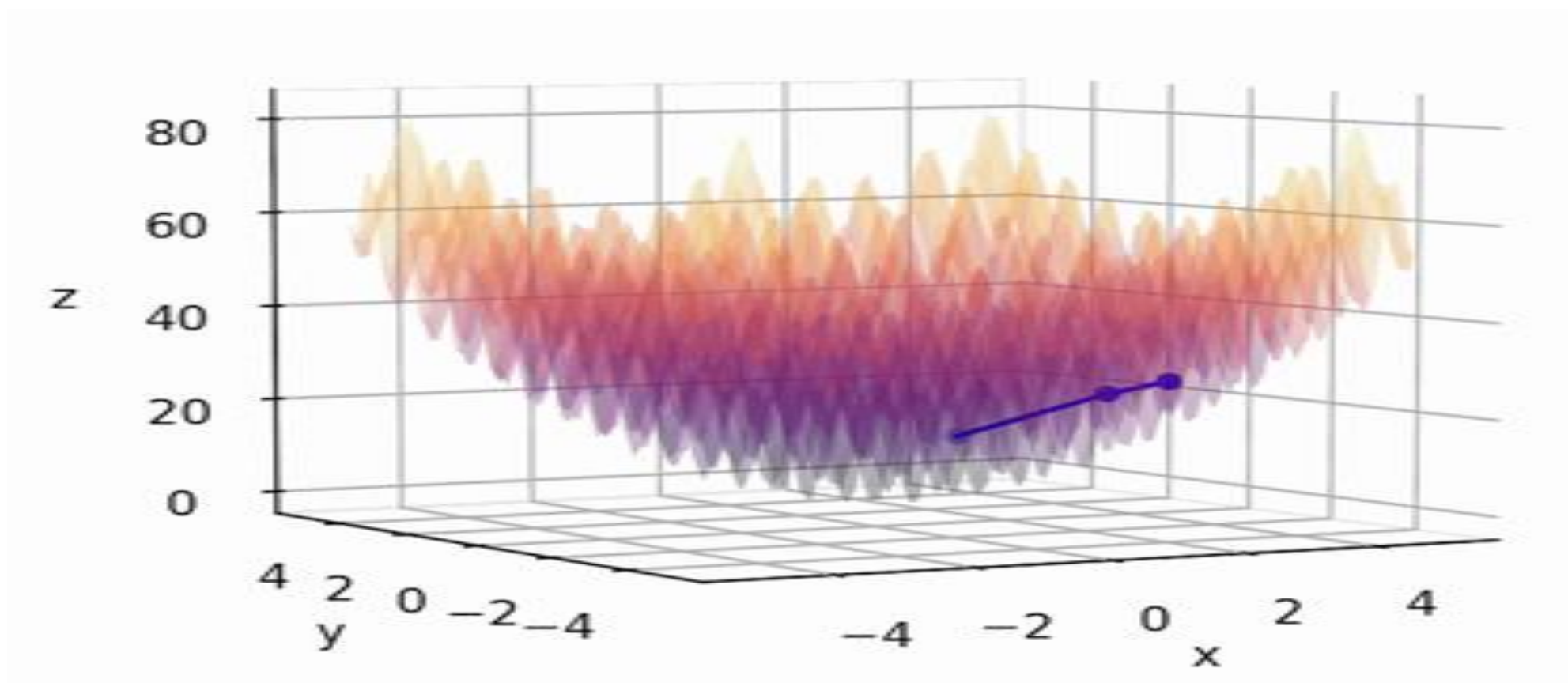

Fig. 6. BSO Intermediatory Point(Trajectory3) For Continuous Rastrigin Functions

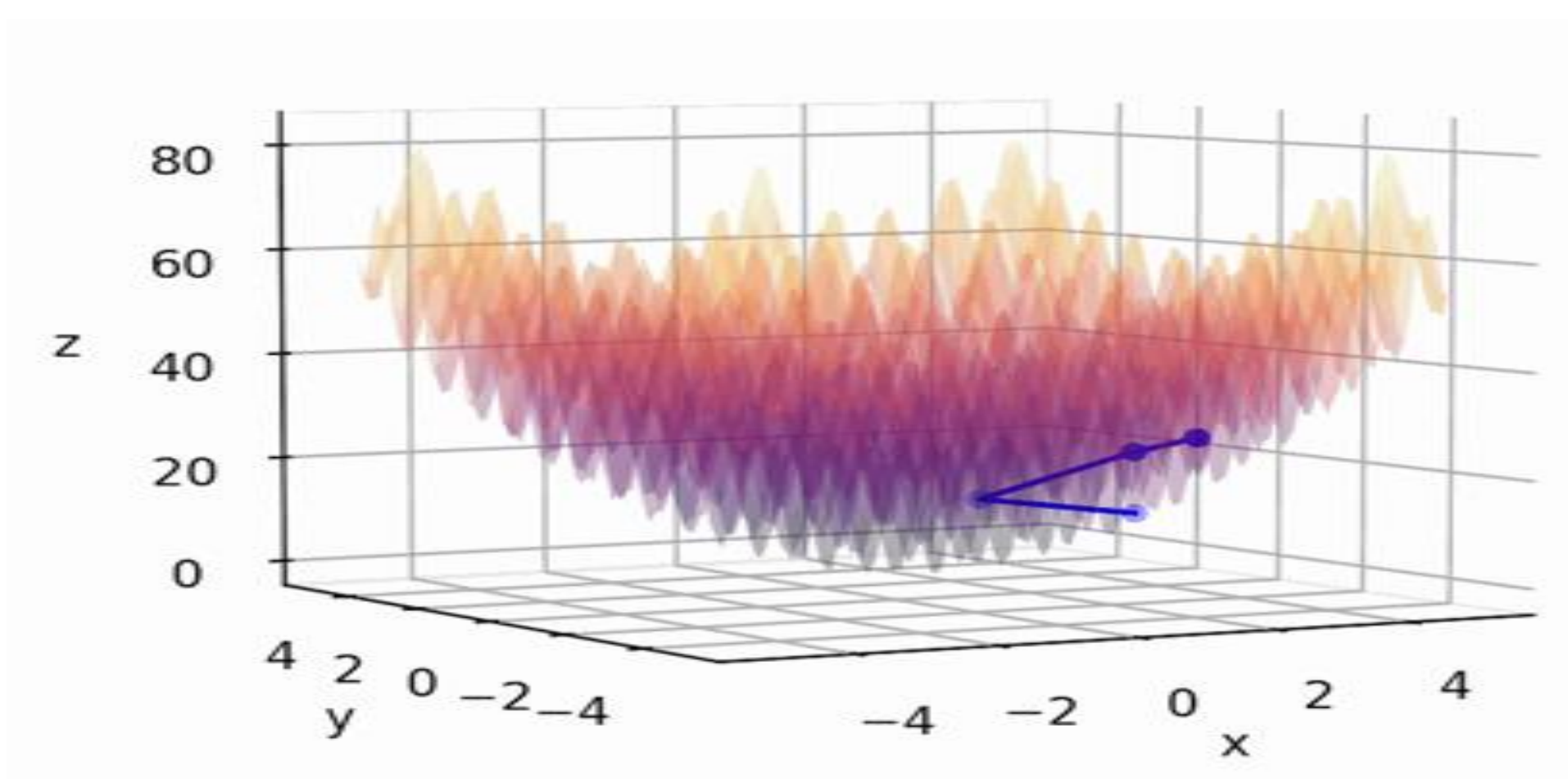

Fig. 7. BSO Intermediatory Point(Trajectory4) For Continuous Rastrigin Functions

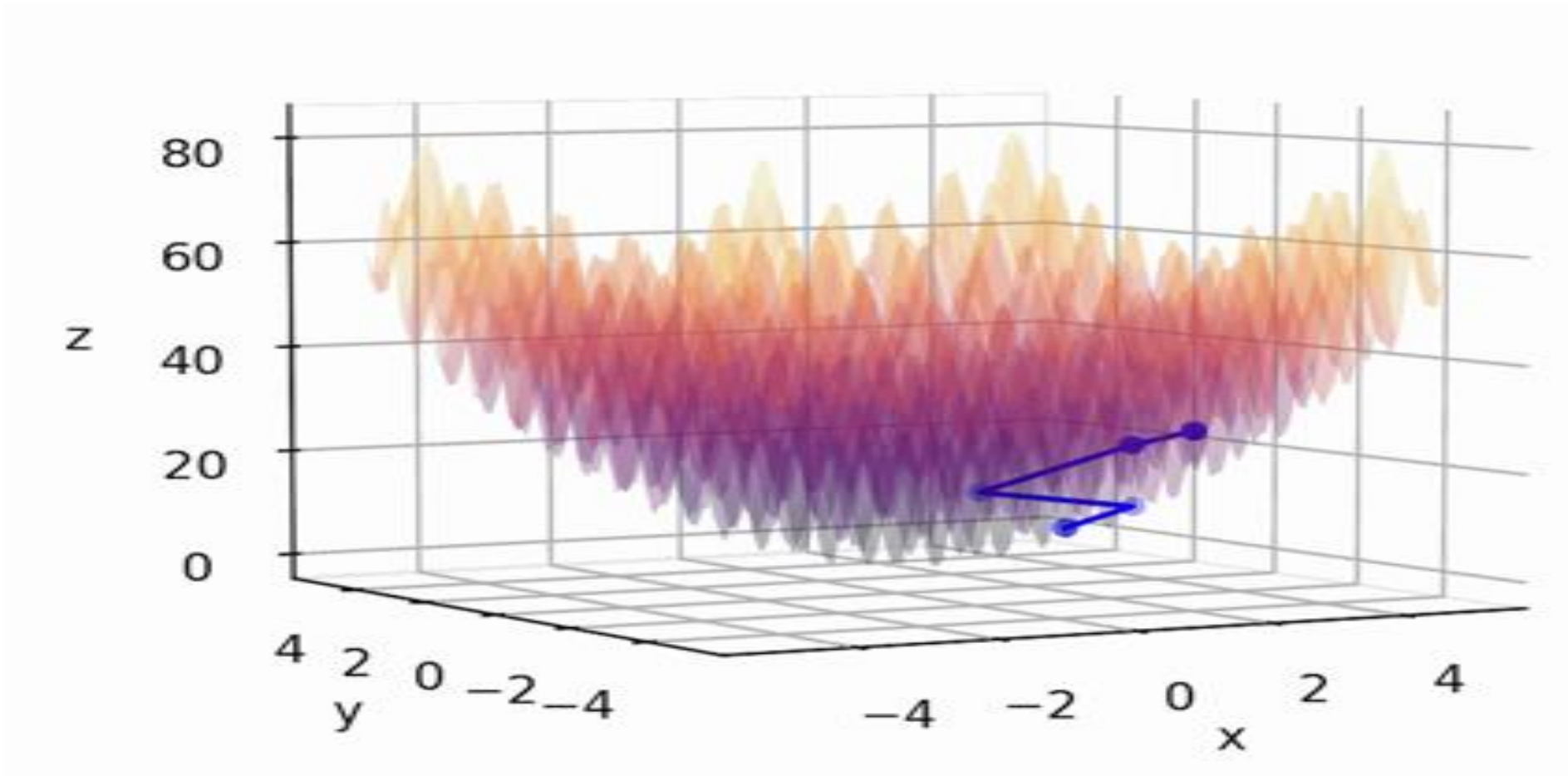

Fig. 8. BSO Intermediatory Point(Trajectoryn-3) For Continuous Rastrigin Functions

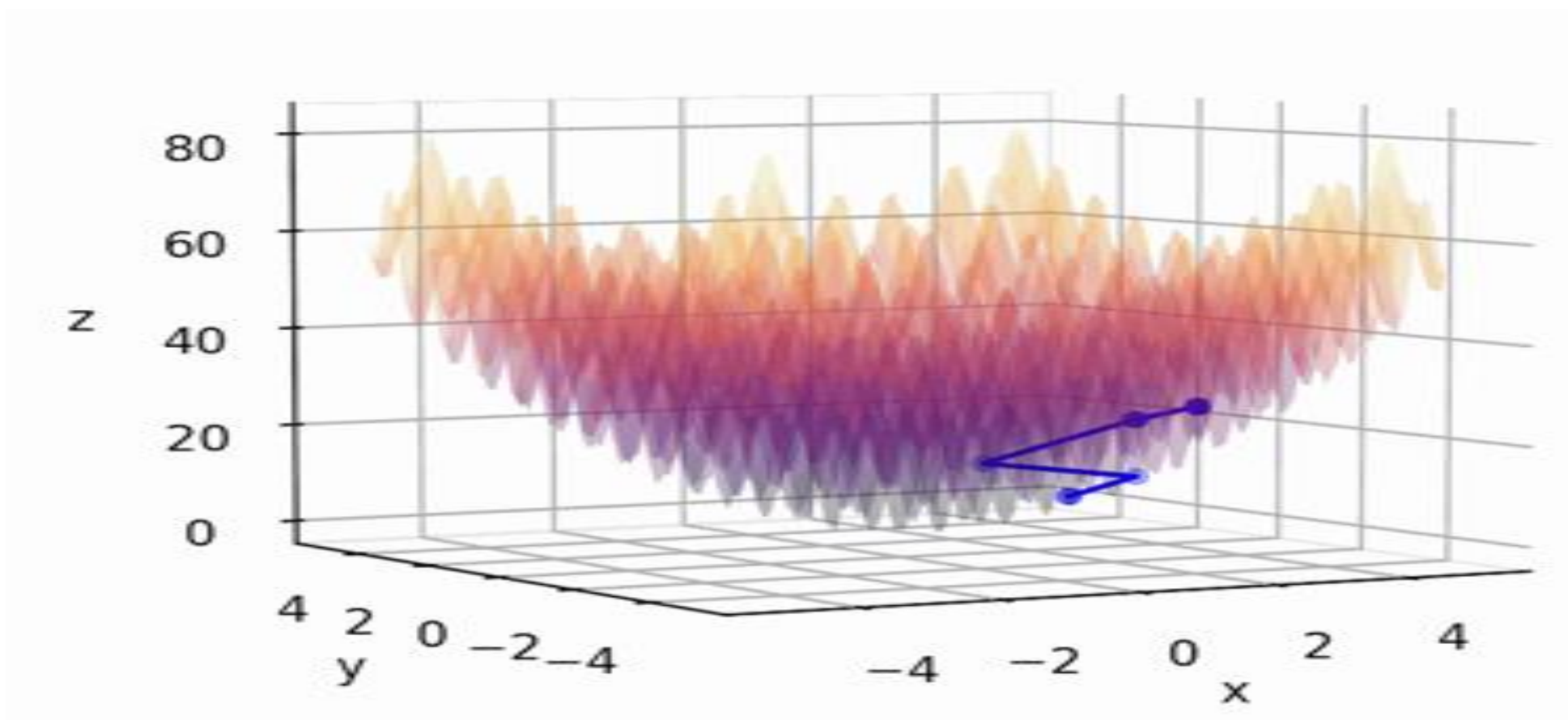

Fig. 9. BSO Intermediatory Point(Trajectory n-2) For Continuous Rastrigin Functions

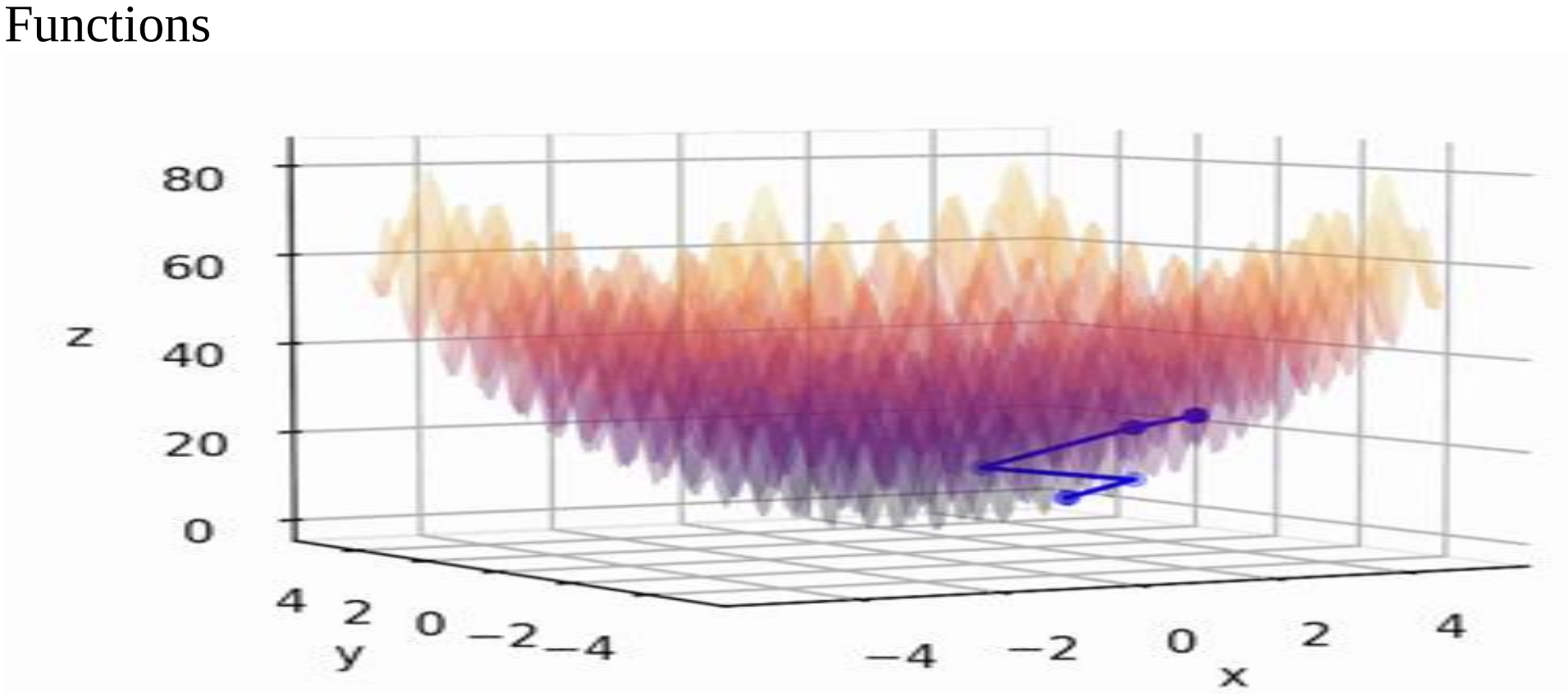

Fig. 10. BSO Intermediatory Point(Trajectory n-1) For Continuous Rastrigin Functions

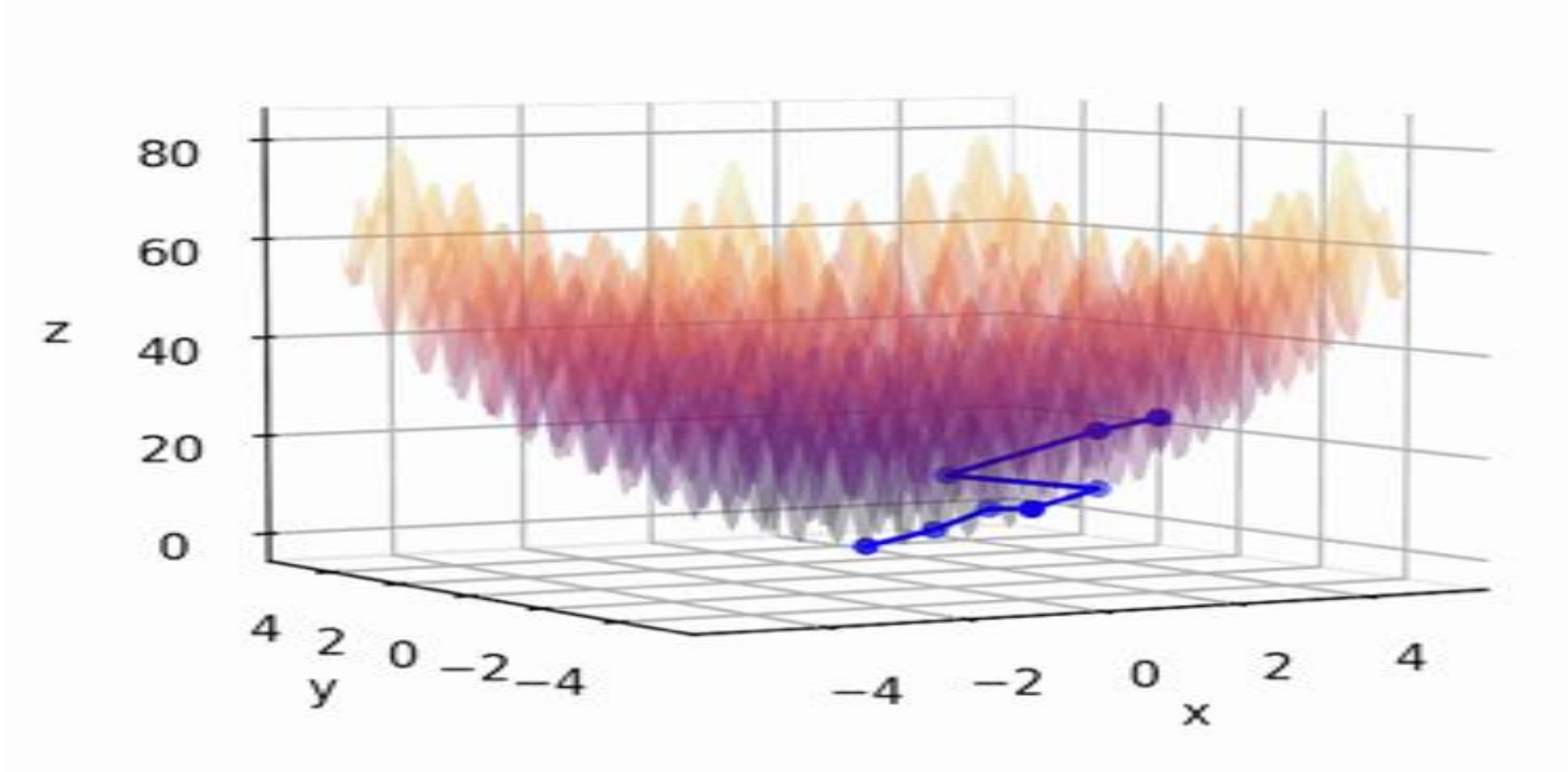

Fig. 11. BSO Final Point(Trajectory n) For Continuous Rastrigin Functions

5.2.2 - Continuous Ackley Trajectory Visualization- BSO

The following step by step sequence of figures shows the trajectory created by BSO while optimizing the continuous ackley optimization problem in 3D. If we have “n” total number of trajectories, then:

- trajectory 1: starting point(Fig. 12)
- trajectory 2 to trajectory n-3 to trajectory n-1 - intermediate trajectories(Fig. 12 to Fig. 17)
- trajectory n - final solution(Fig. 18).

Trajectory begun from step1 and then start optimizing by locating the next optimal solution and then begun another trajectory from the most recent optimal solution

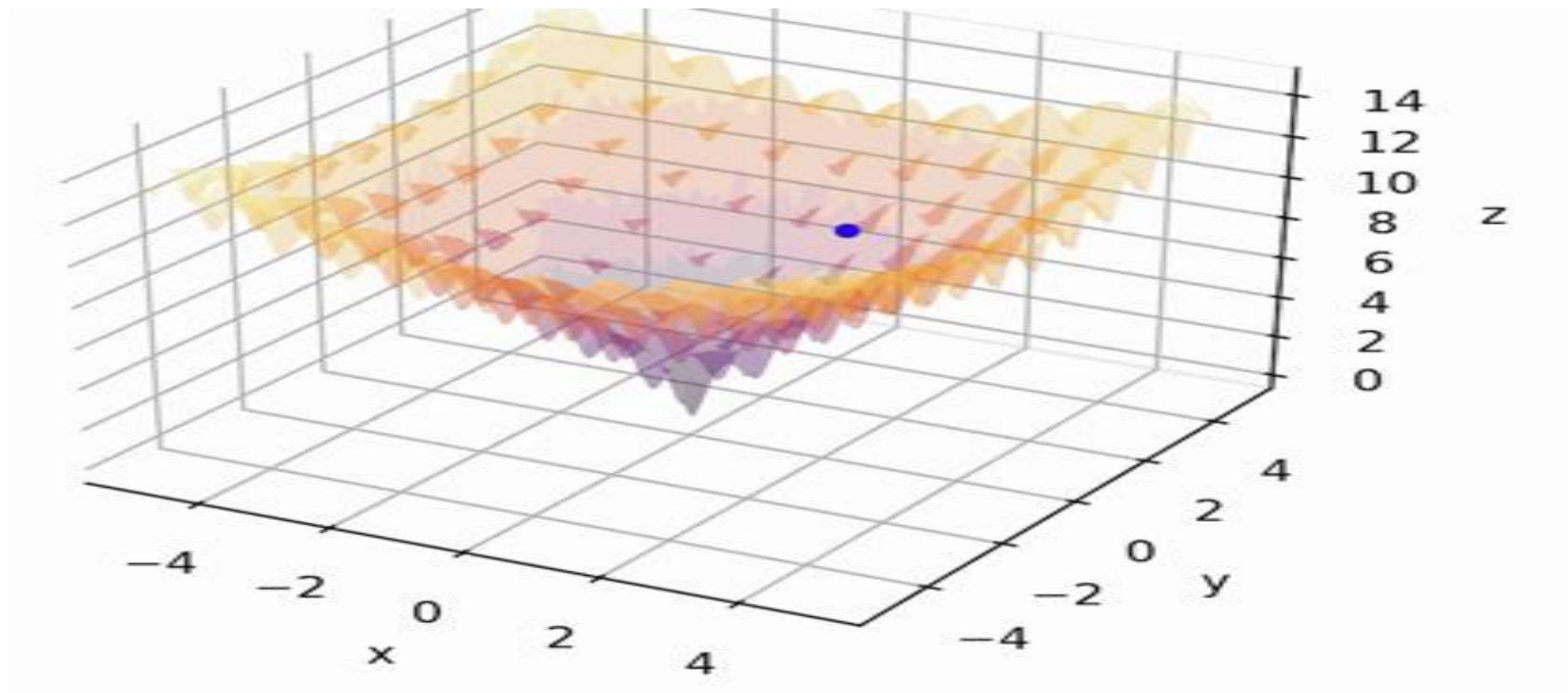

Fig. 12. BSO Starting Point(Trajectory 1) For Continuous Ackley Functions

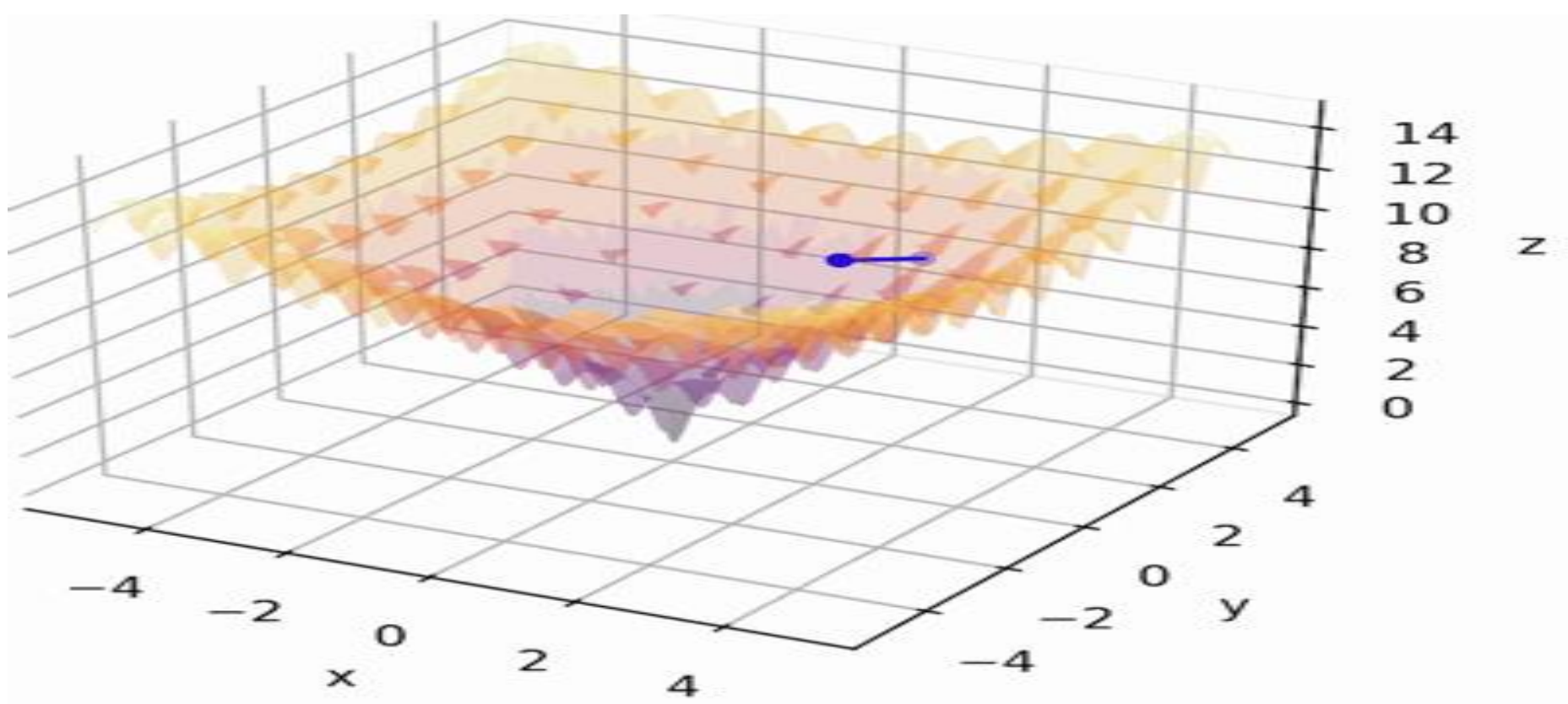

Fig. 13. BSO Intermediatory Point(Trajectory 2) For Continuous Ackley Functions

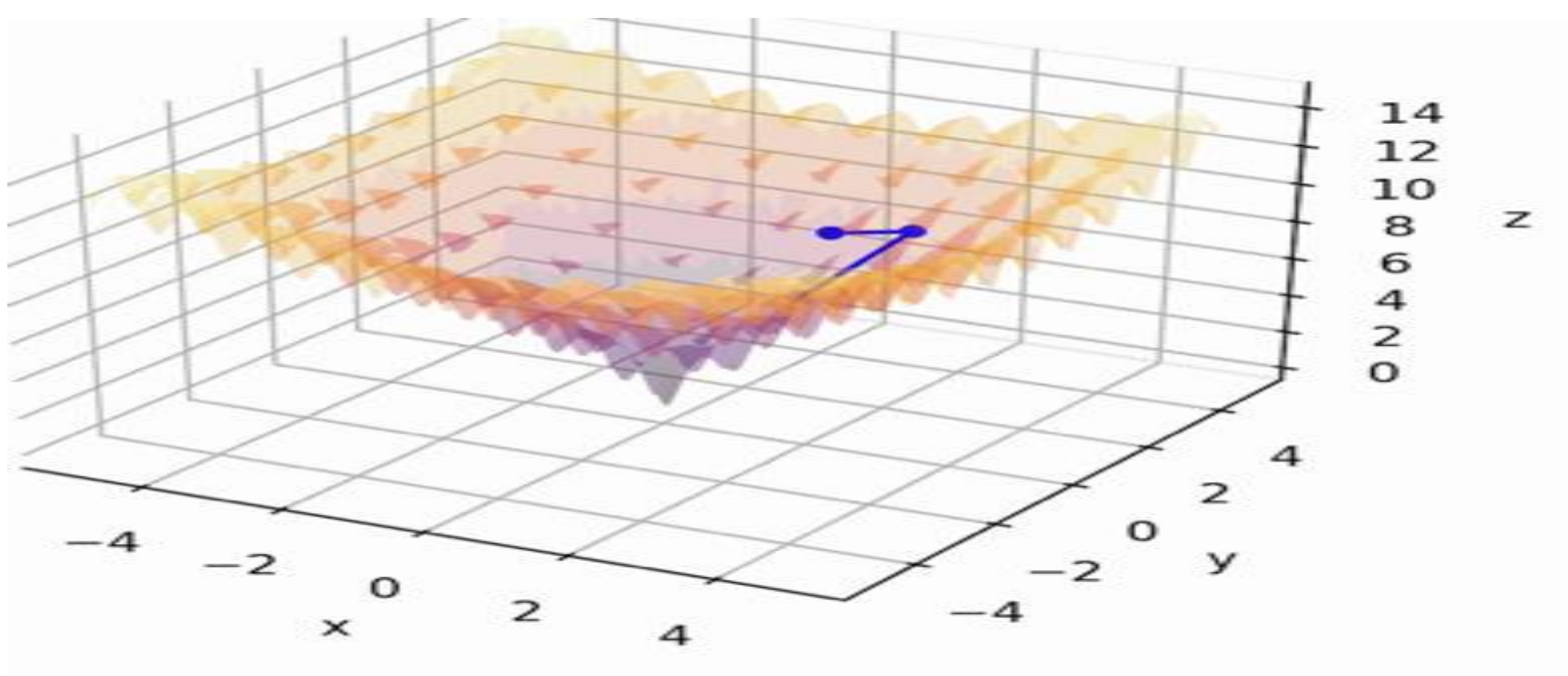

Fig. 14. BSO Intermediatory Point(Trajectory 3) For Continuous Ackley Functions

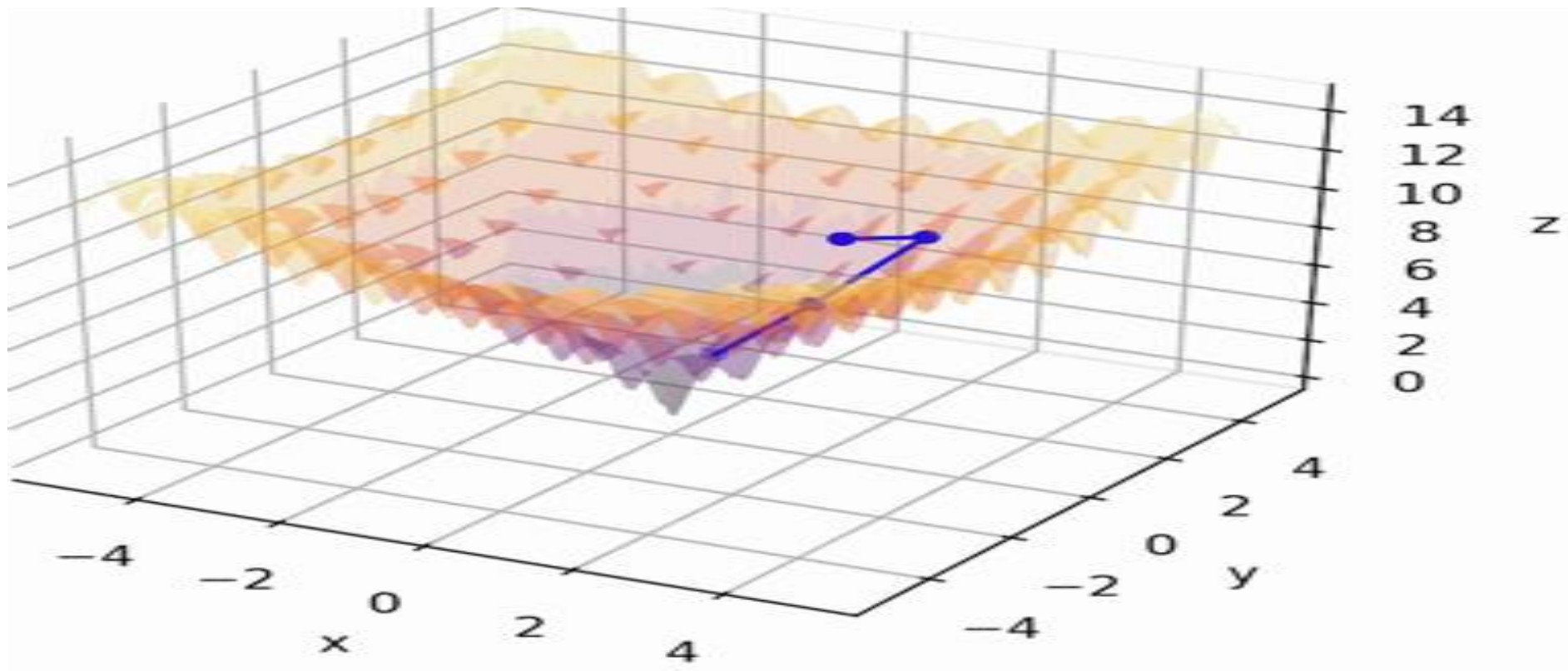

Fig. 15. BSO Intermediatory Point(Trajectory 4) For Continuous Ackley Functions

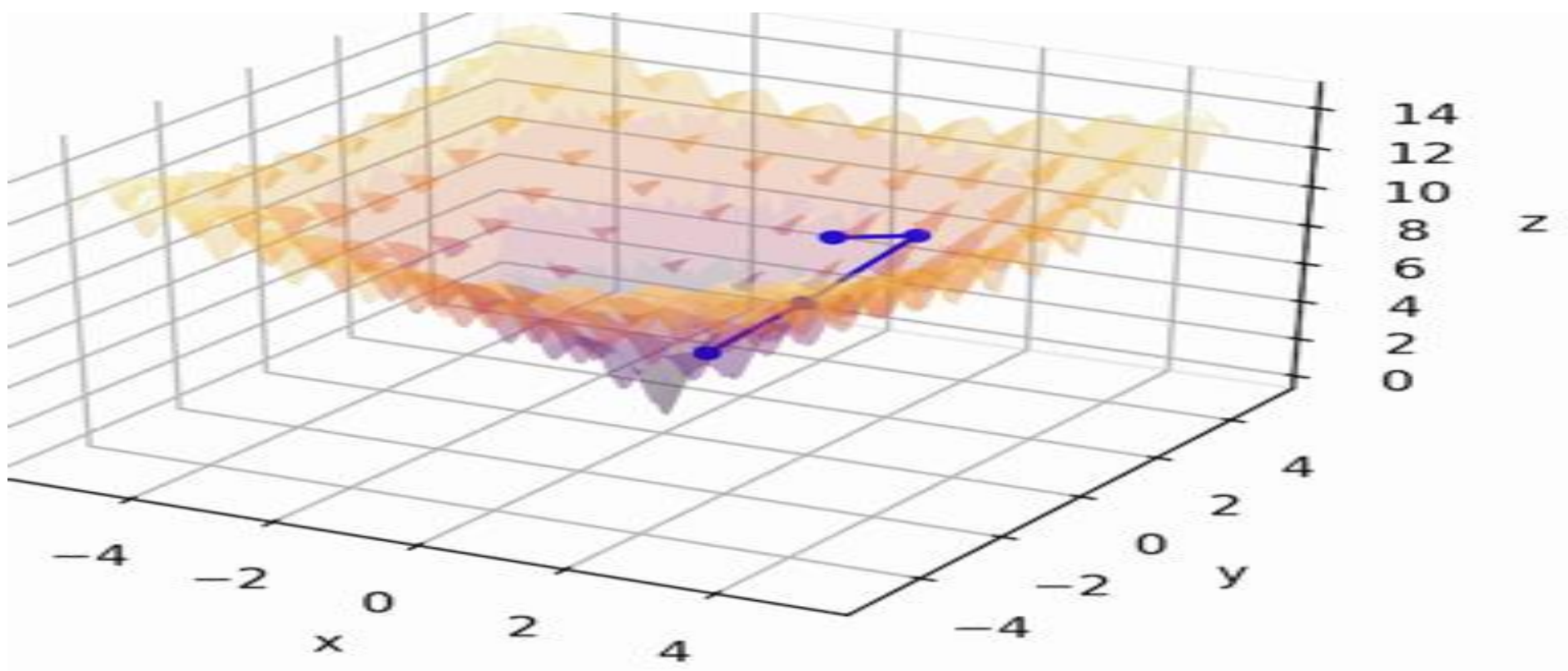

Fig. 16. BSO Intermediatory Point(Trajectory n-2) For Continuous Ackley Functions

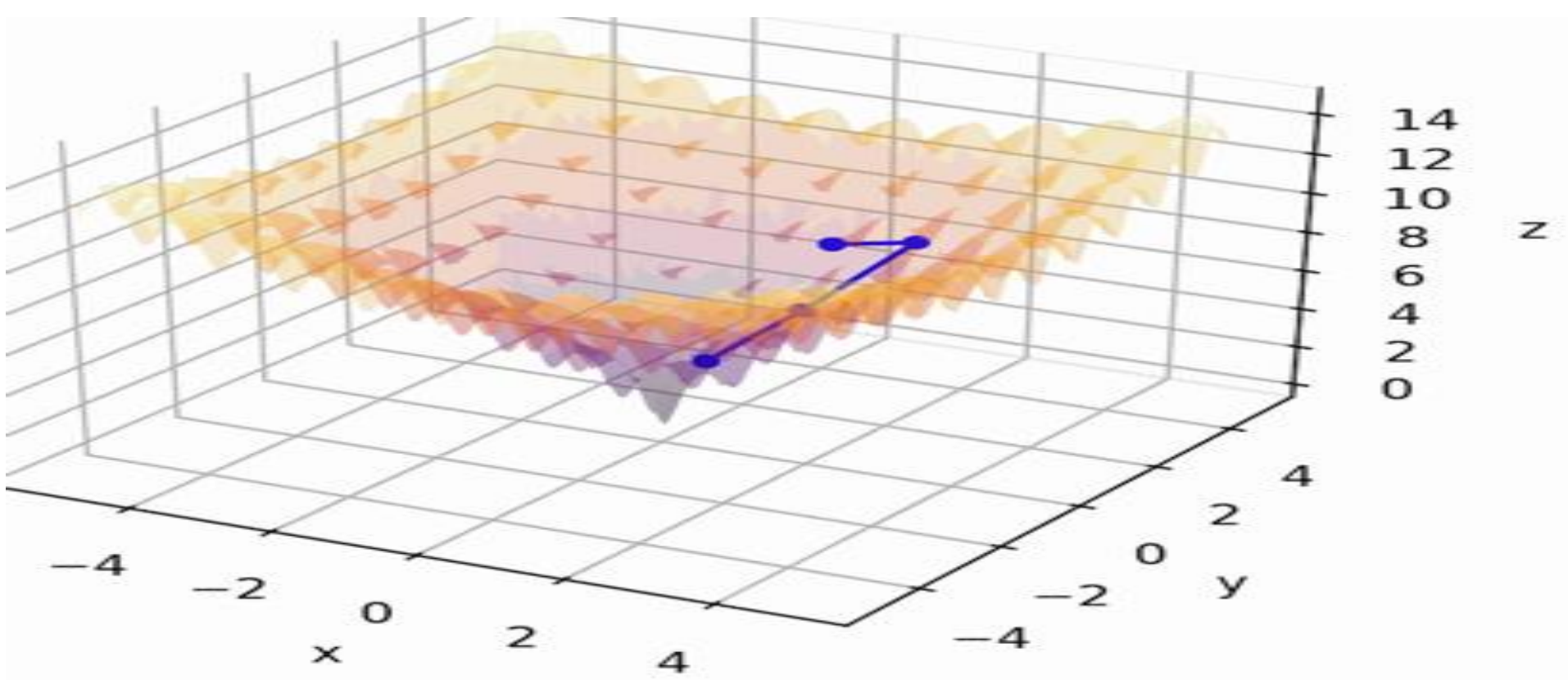

Fig. 17. BSO Intermediatory Point(Trajectory n-1) For Continuous Ackley Functions

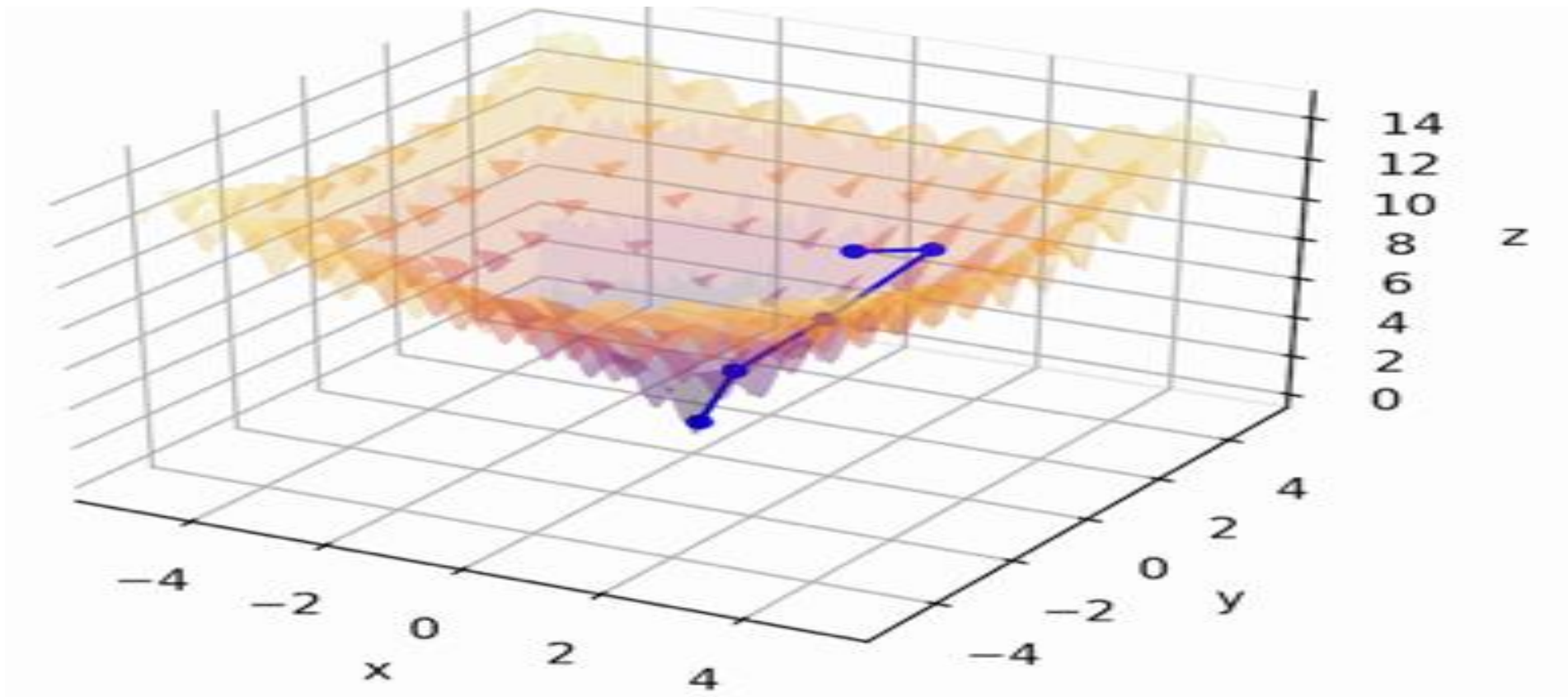

Fig. 18. BSO Final Point(Trajectory n) For Continuous Ackley Functions

5.2.3 - Non- Continuous Rastrigin
Trajectory Visualization - BSO

The following step by step sequence of figures shows the trajectory created by BSO while optimizing the non-continuous rastrigin optimization problem in 3D. If we have “n” total number of trajectories, then:

- trajectory 1: starting point(Fig. 19)
- trajectory 2 to trajectory n-2 to trajectory n-1 - intermediate trajectories(Fig. 20 to Fig. 23)
- trajectory n - final solution(Fig. 24).

Trajectory begun from step1 and then start optimizing by locating the next optimal solution and then begun another trajectory from the most recent optimal solution.

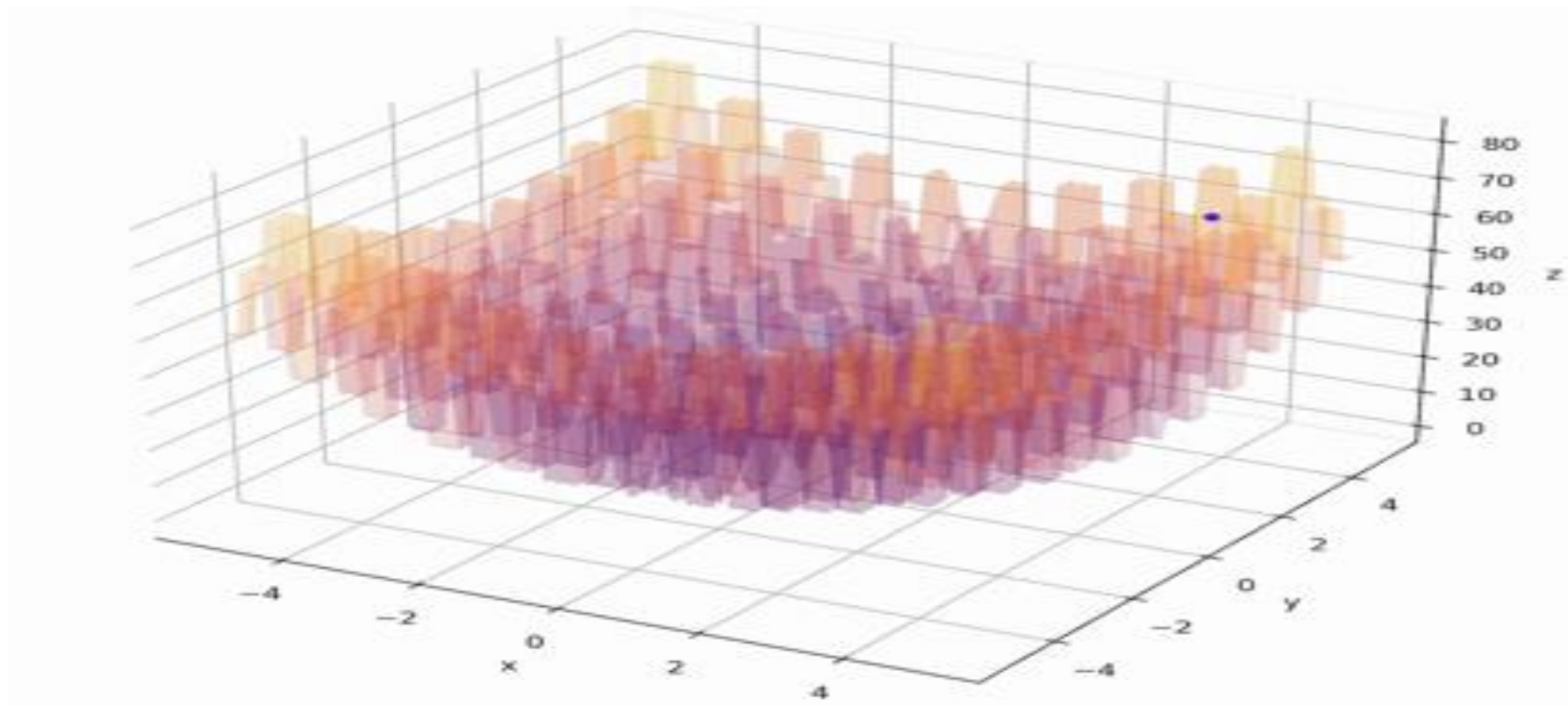

Fig. 19. BSO Starting Point(Trajectory 1) For Non-Continuous Rastrigin Functions

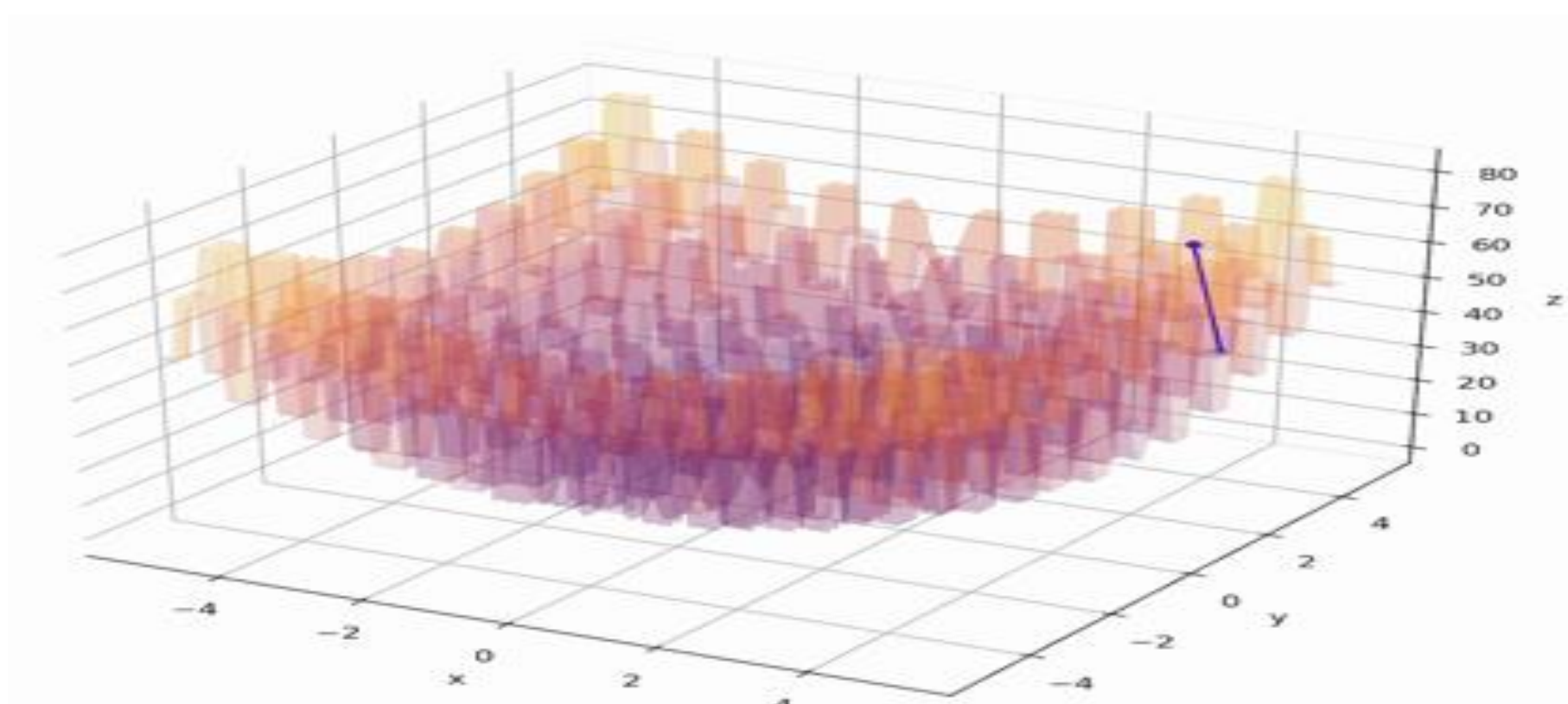

Fig. 20. BSO Intermediatory Point(Trajectory 2) For Non-Continuous Rastrigin Functions

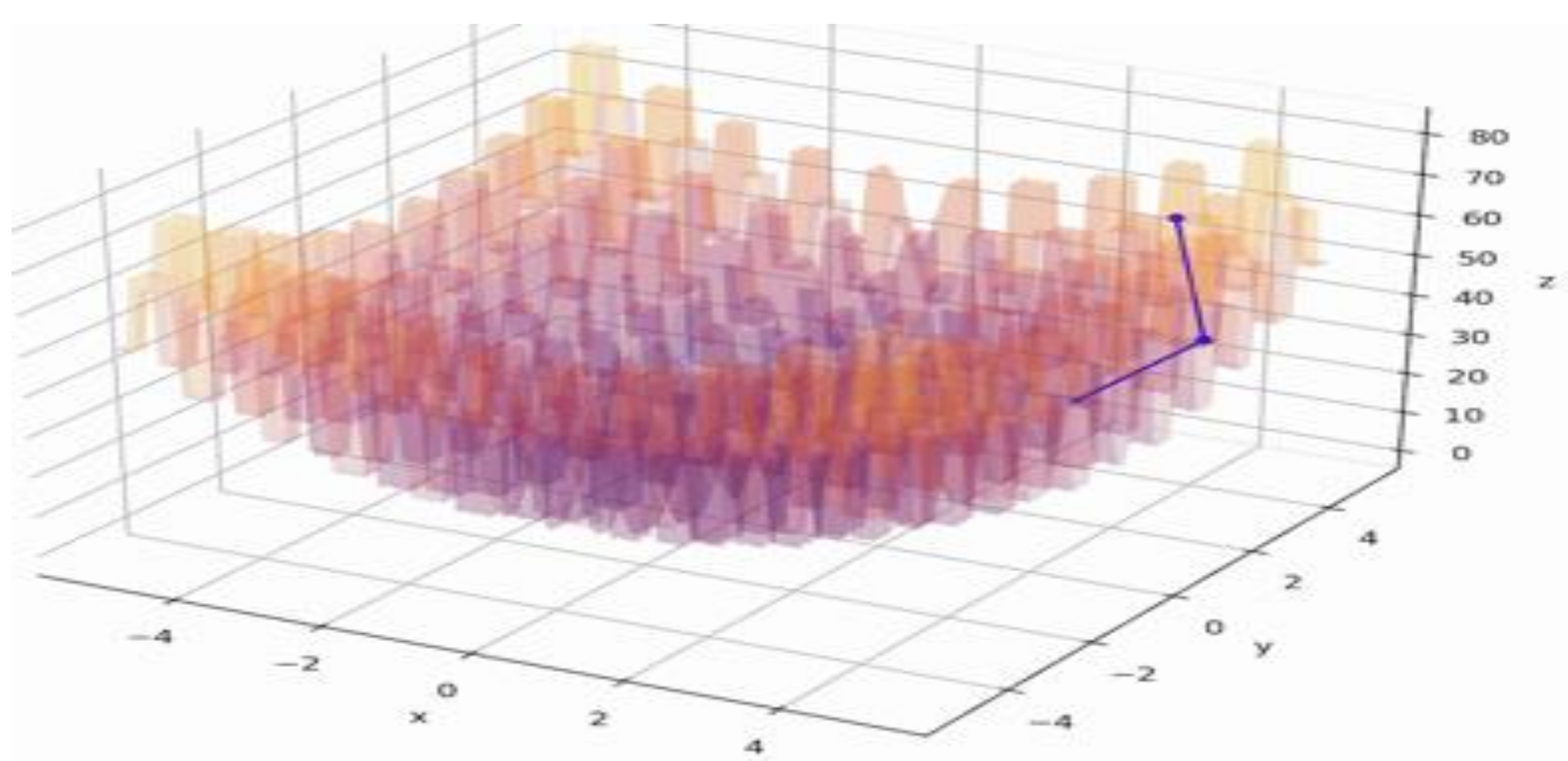

Fig. 21. BSO Intermediatory Point(Trajectory 3) For Non-Continuous Rastrigin Functions

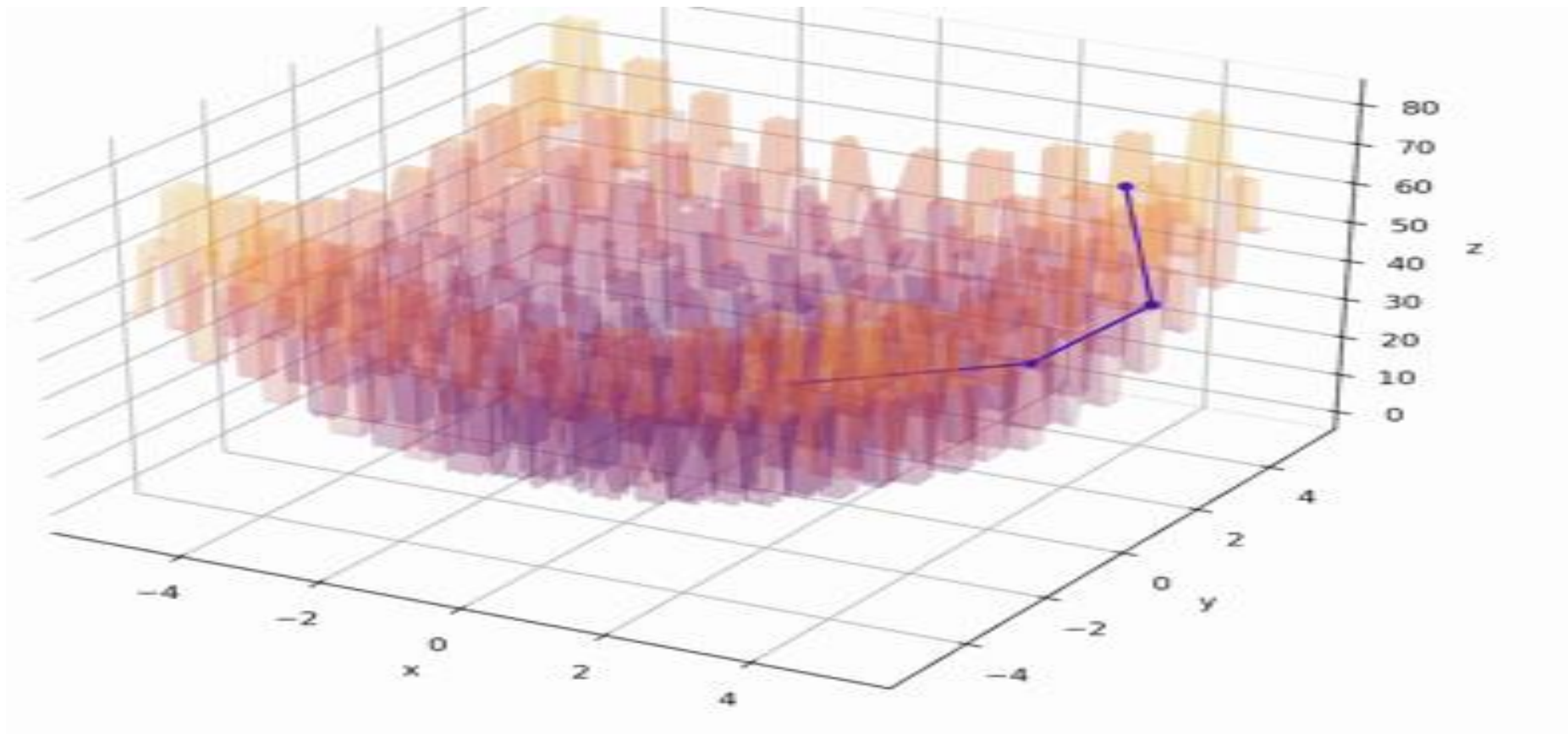

Fig. 22. BSO Starting Point(Trajectory n-2) For Non-Continuous Rastrigin Functions

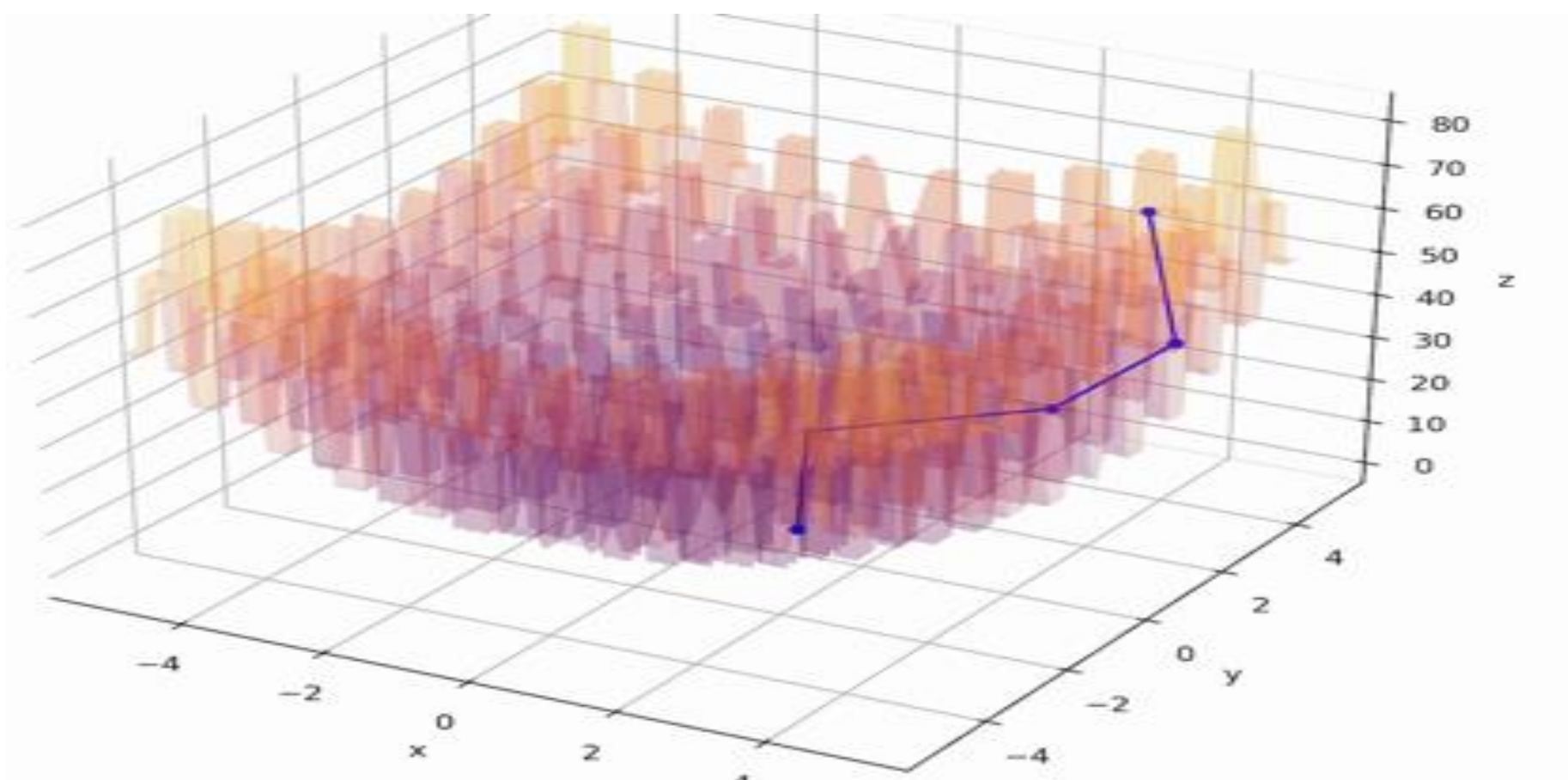

Fig. 23. BSO Starting Point(Trajectory n-1) For Non-Continuous Rastrigin Functions

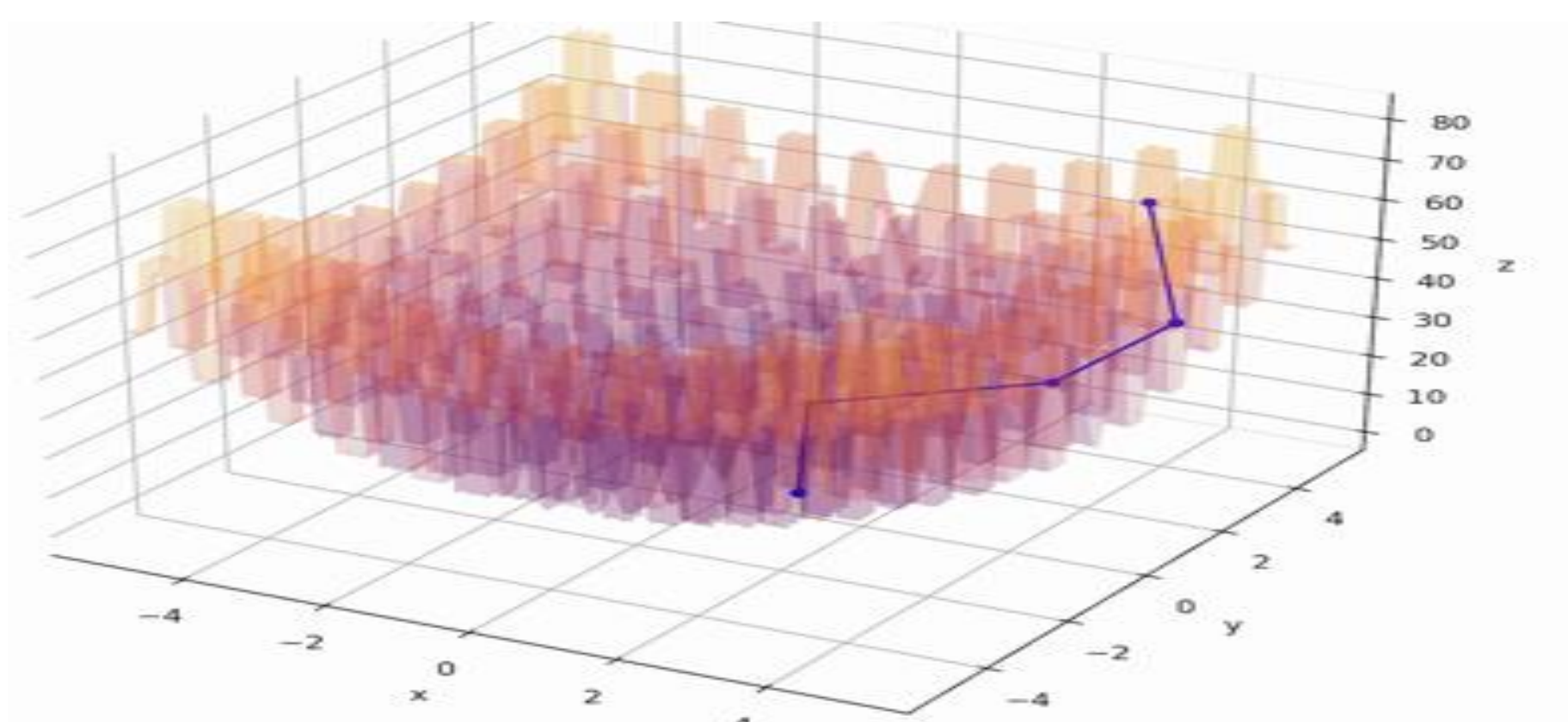

Fig. 24. BSO Final Point(Trajectory n) For Non-Continuous Rastrigin Functions

5.2.4 - Non-Continuous Step Function
Trajectory Visualization - BSO

The following step by step sequence of figures shows the trajectory created by BSO while optimizing the non-continuous step function optimization problem in 3D. If we have "n" total number of trajectories, then:

- trajectory 1: starting point(Fig. 25)
- trajectory 2 to trajectory n-4 to trajectory n-1 - intermediate trajectories(Fig. 26 to Fig. 32)
- trajectory n - final solution(Fig. 33).

Trajectory begun from step1 and then start optimizing by locating the next optimal solution and then begun another trajectory from the most recent optimal solution

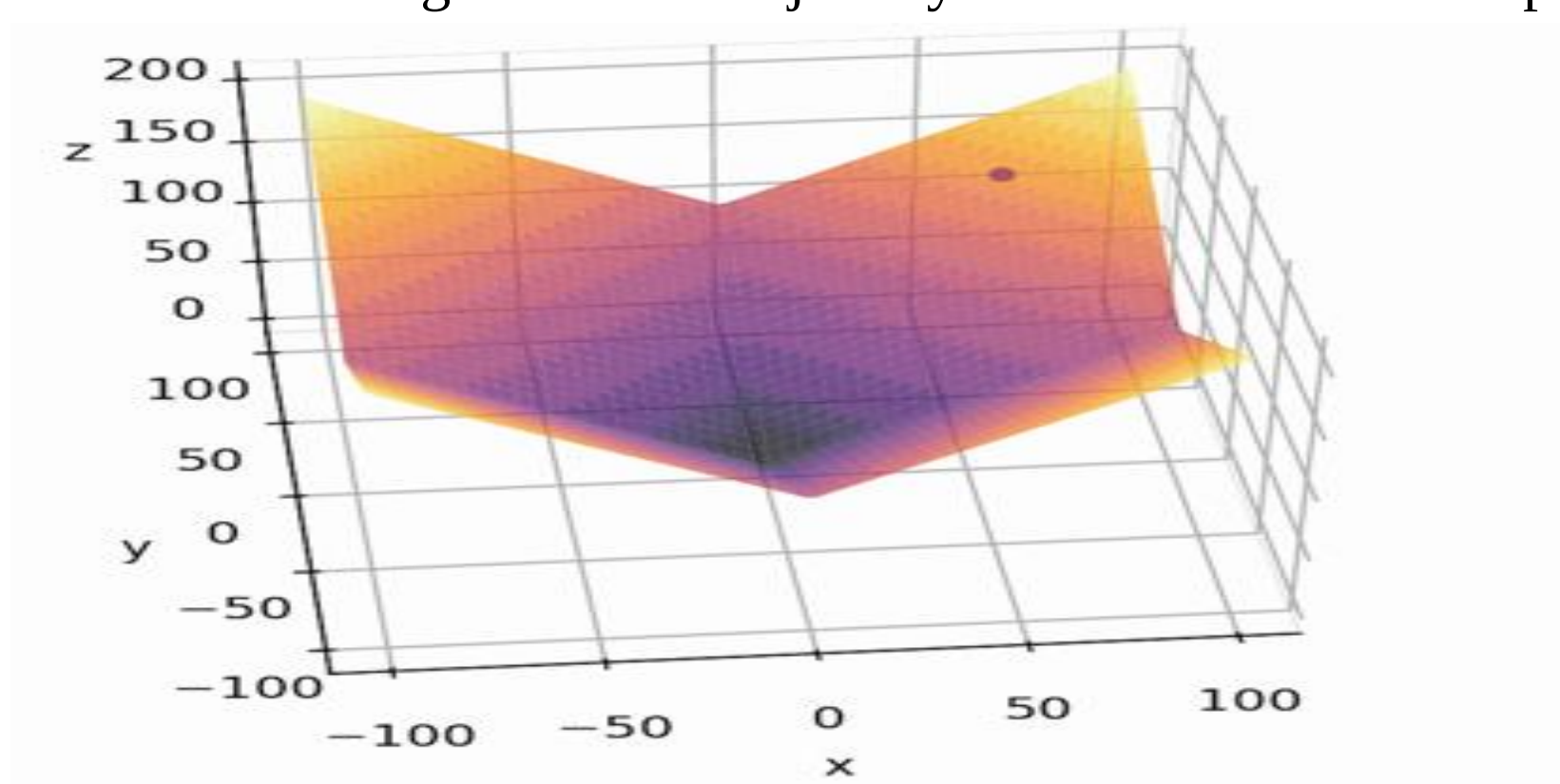

Fig. 25. BSO Starting Point (Trajectory 1) For Non-Continuous Step Function

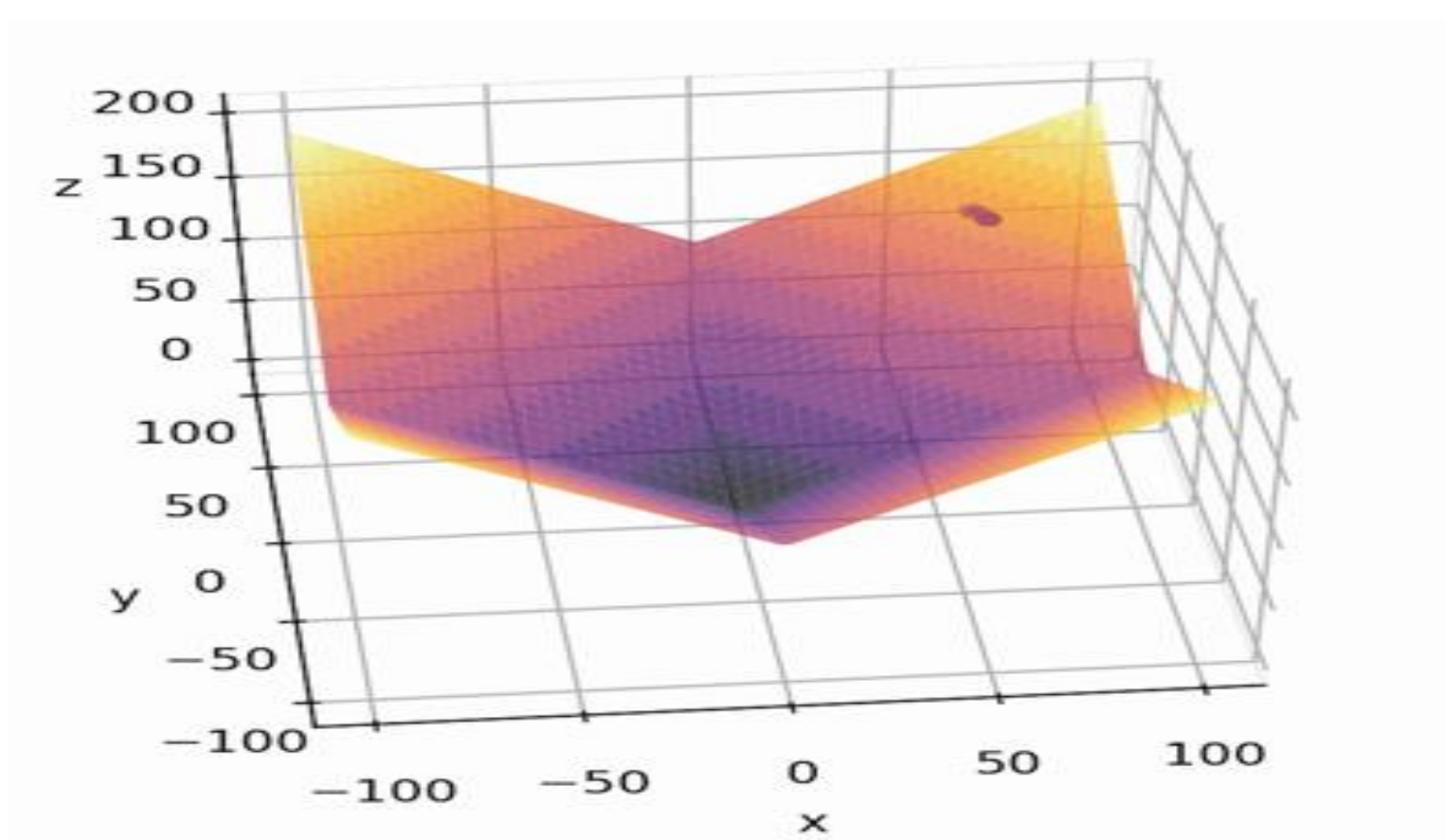

Fig. 26. BSO Intermediatory Point (Trajectory 2) For Non-Continuous Step Function

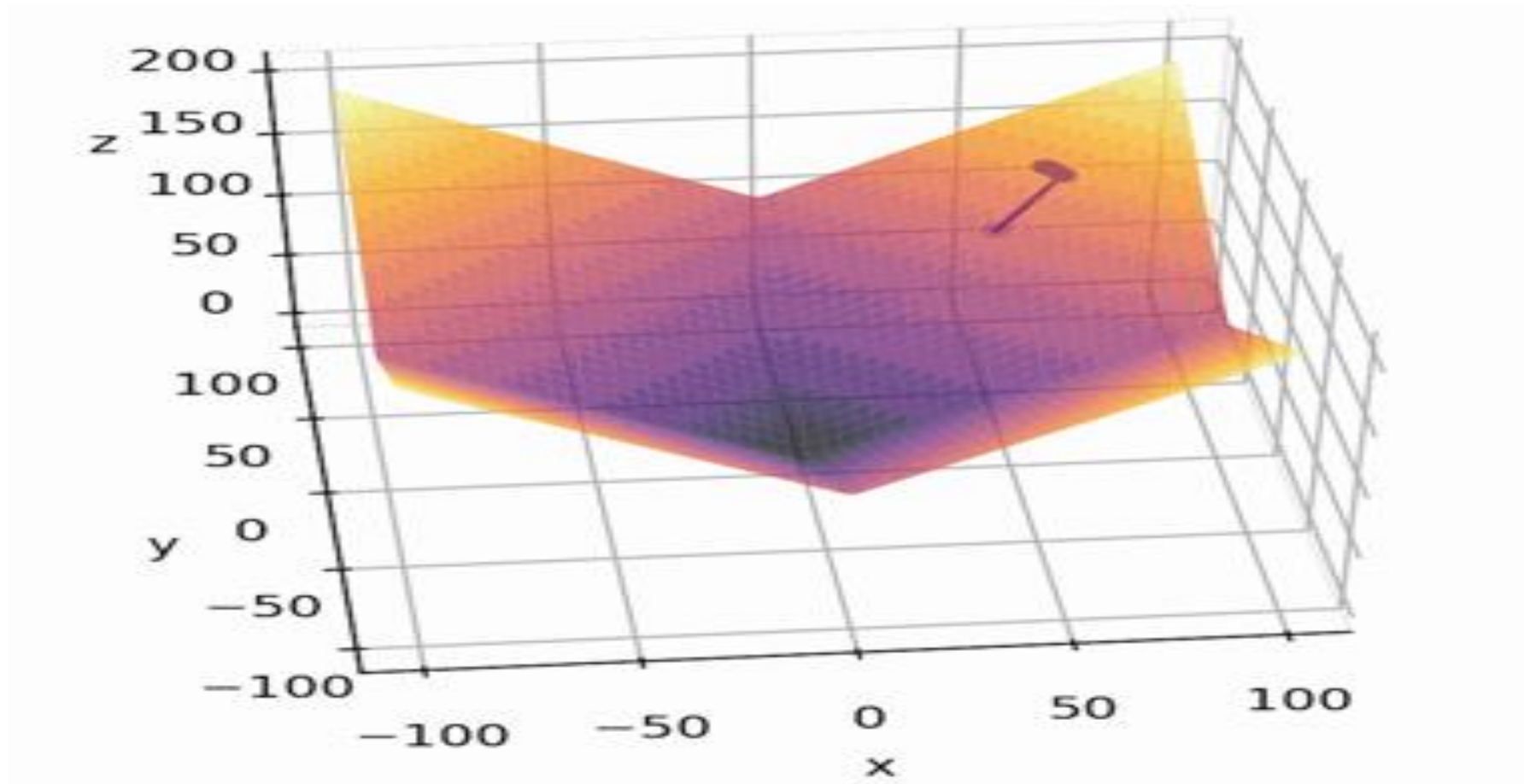

Fig. 27. BSO Intermediatory Point (Trajectory 3) For Non-Continuous Step Function

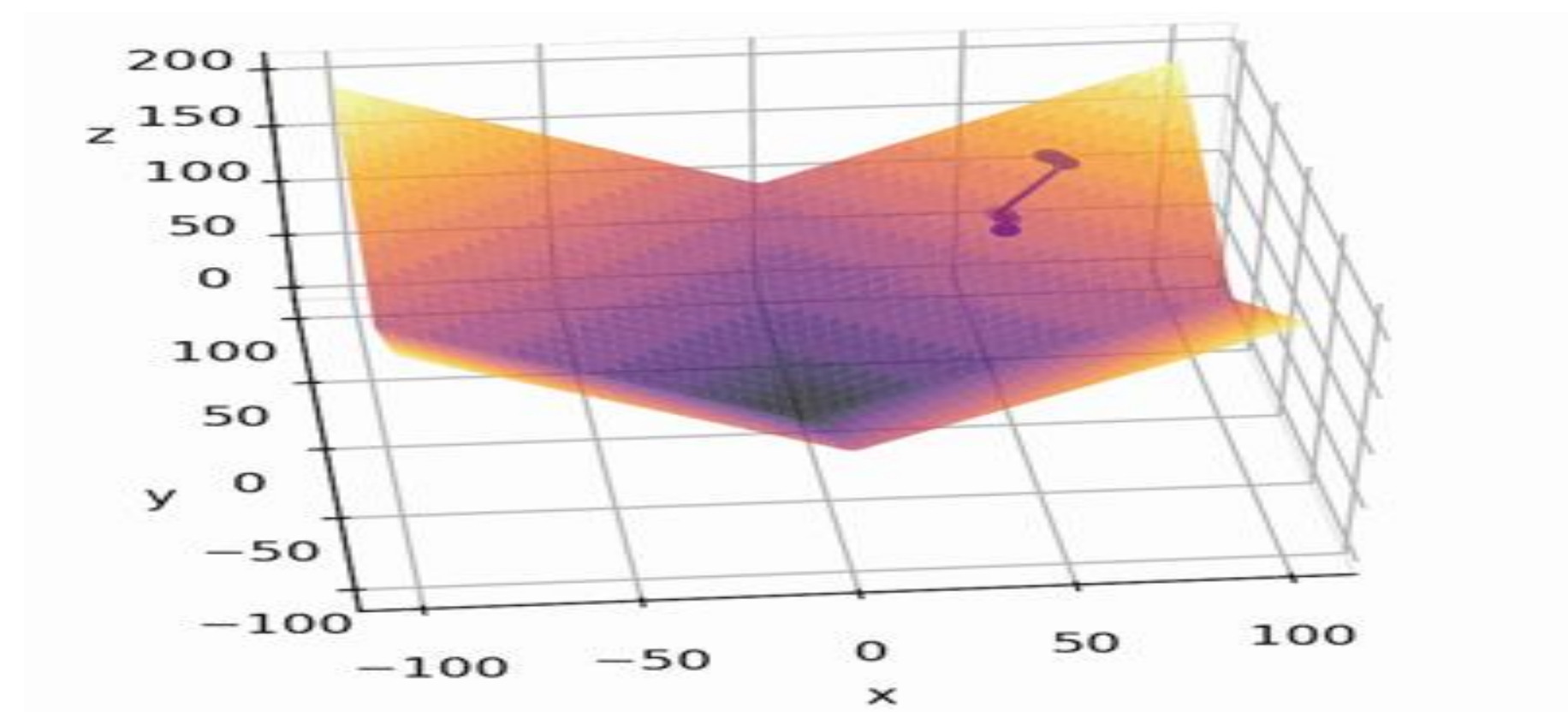

Fig. 28. BSO Intermediatory Point(Trajectory 4) For Non-Continuous Step Function

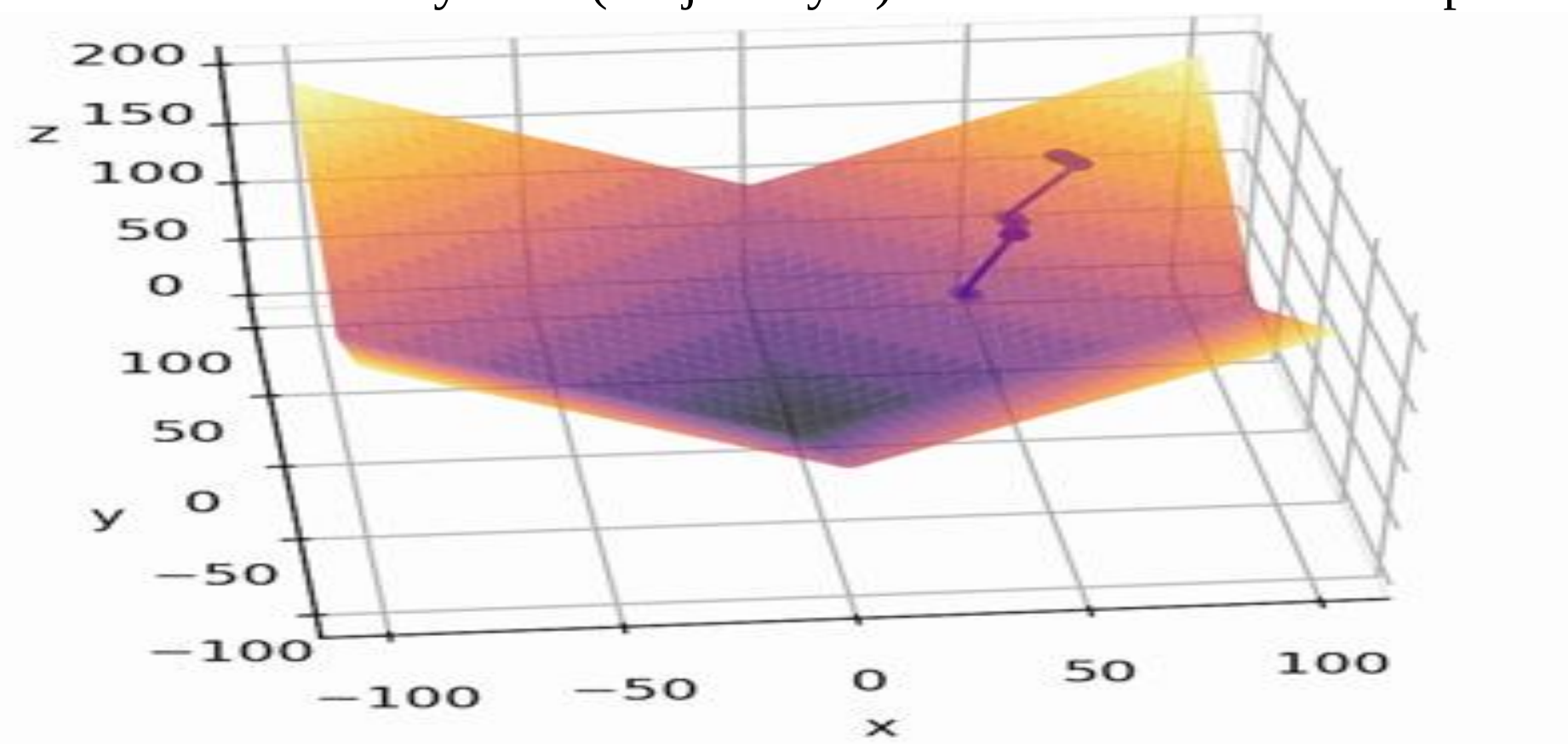

Fig. 29. BSO Intermediatory Point(Trajectory n-4) For Non-Continuous Step Function

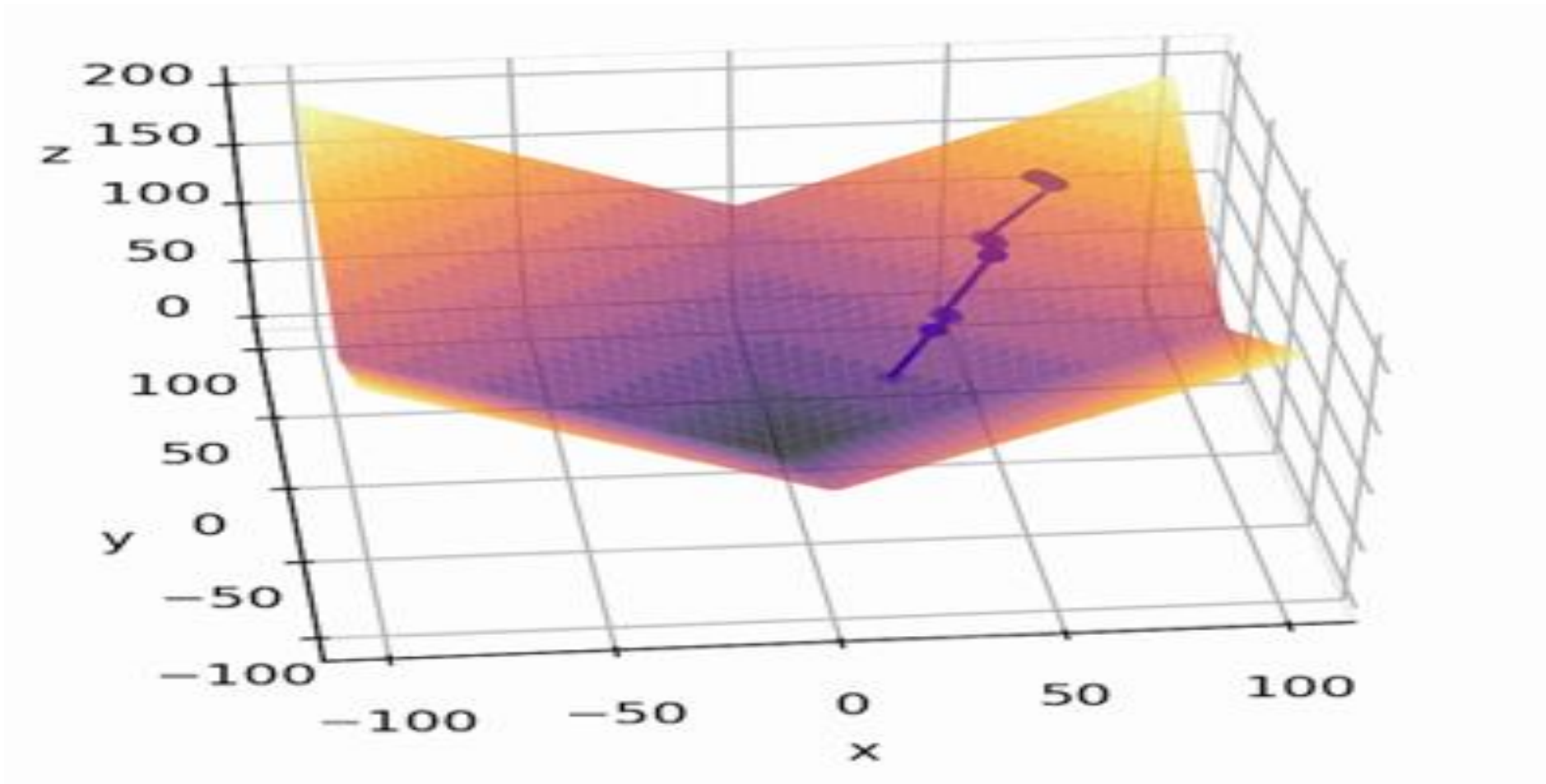

Fig. 30. BSO Intermediatory Point(Trajectory n-3) For Non-Continuous Step Function

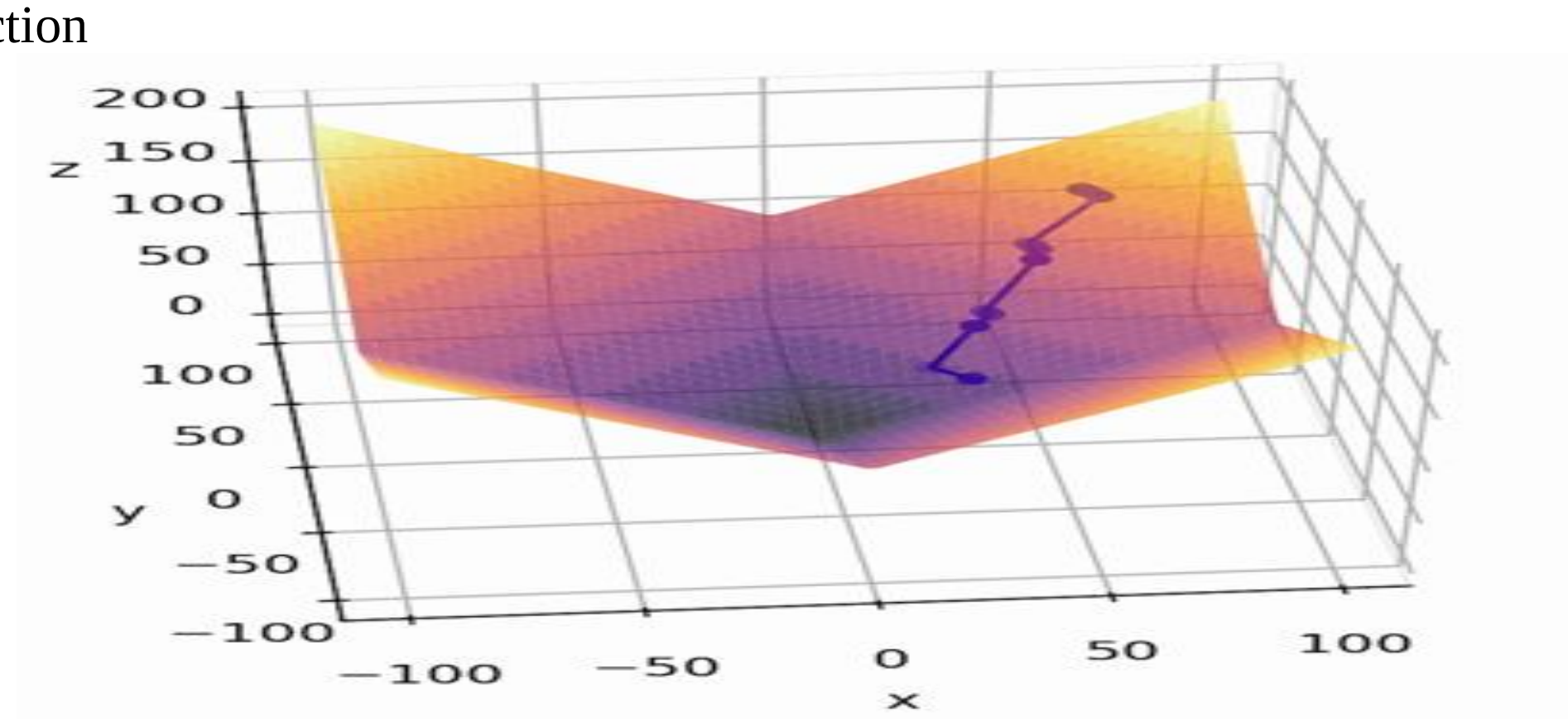

Fig. 31. BSO Intermediatory Point(Trajectory n-2) For Non-Continuous Step Function

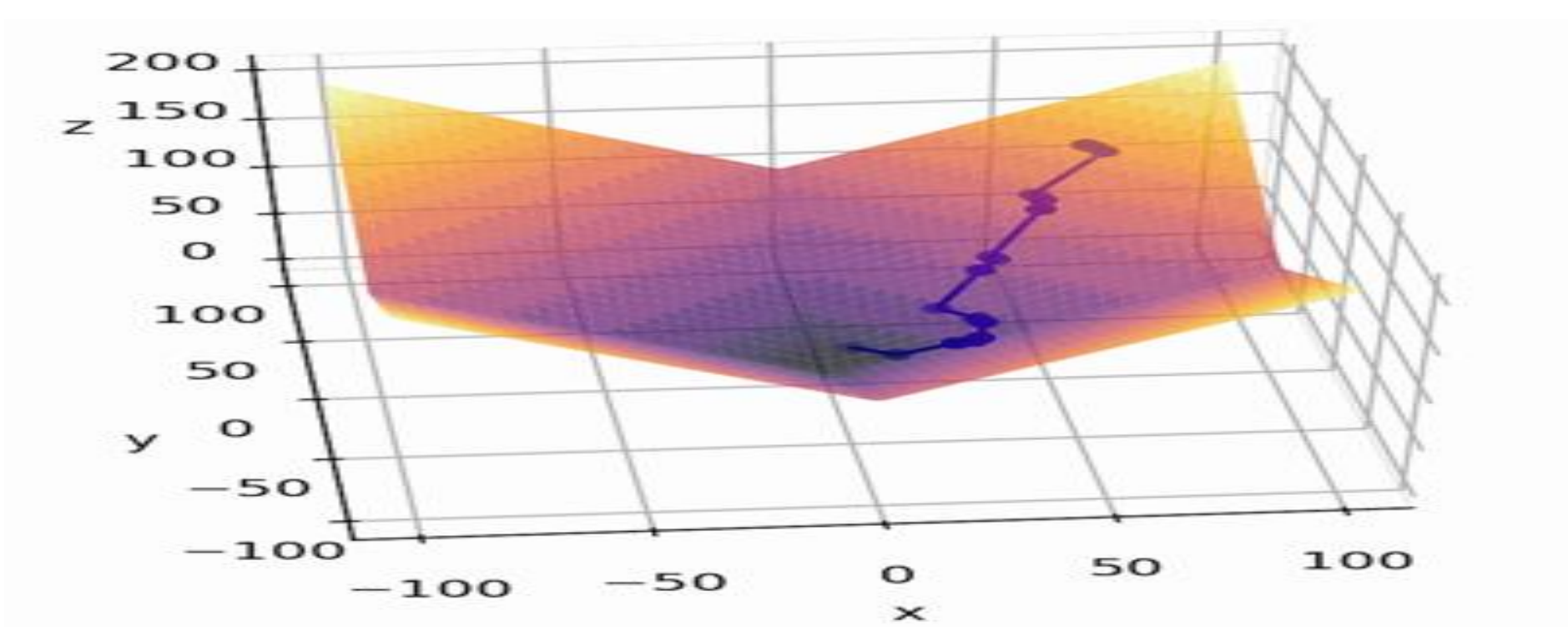

Fig. 32. BSO Intermediatory Point(Trajectory n-1) For Non-Continuous Step Function

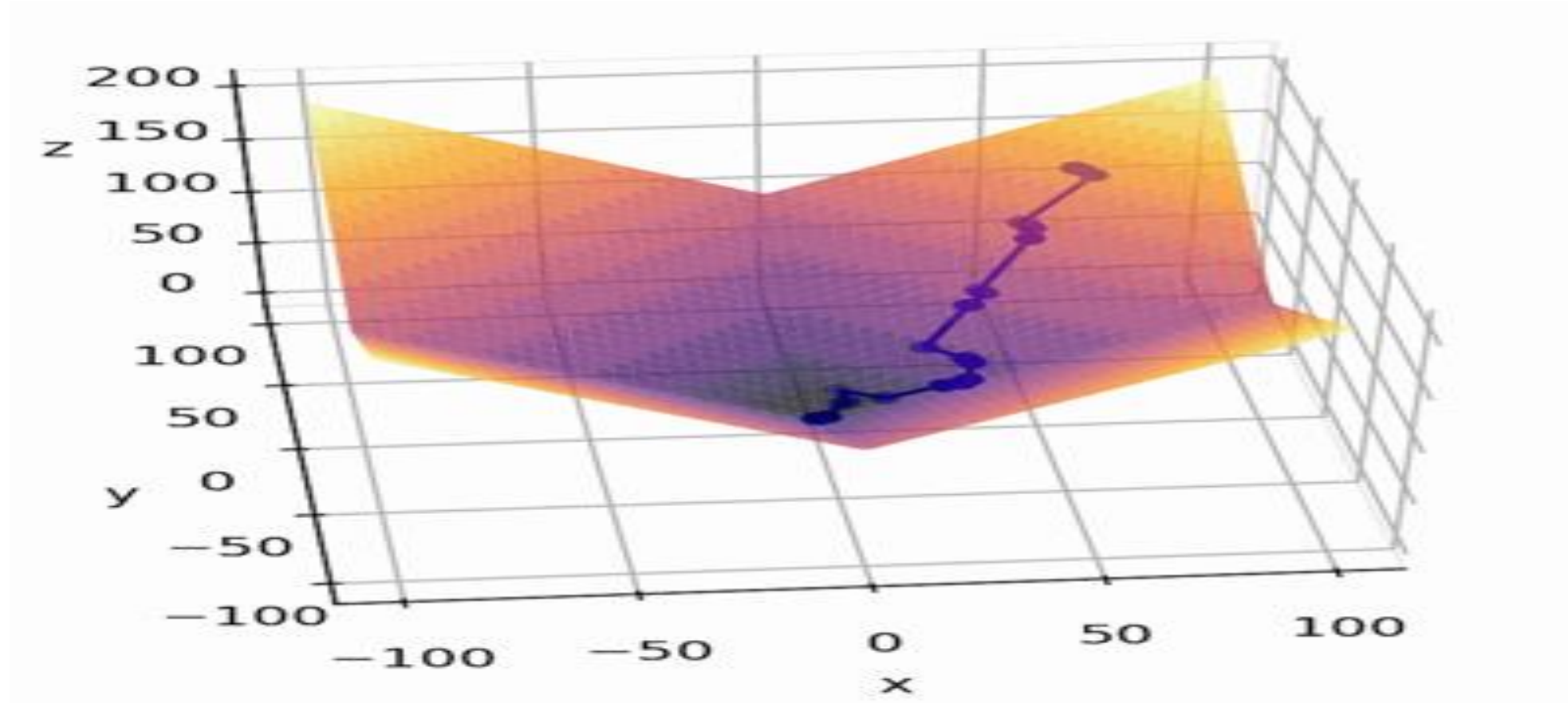

Fig. 33. BSO Final Point(Trajectory n) For Non-Continuous Step Functions

## 5.3 - Comparative Plots Between BSO And BPO Trajectories

In this context, we are showcasing the trajectory plots of BSO and BPO for continuous rastrigin and non-continuous rastrigin functions. We perform this comparison as BSO is an extension of BPO and hence we can conclude that the number of trajectories in BSO are quite less in comparison to BSO. From the table we can infer about the number of trajectories between two algorithms, the movement of trajectories during the search of near optimal solution.

We have chosen these two functions randomly but as we have seen in result Chapter 5, even though BSO has shown less than 100% accuracy in non-continuous rastrigin function but still BSO is taking a smaller number of trajectories to reach near optimal solution and less time to converge. Table 14 shows the comparison between BSO and BPO for continuous rastrigin function and Table 15 shows comparison between BSO and BPO for non-continuous rastrigin function.

Table 14. Comparison for Continuous Rastrigin Between BSO and BPO

| BSO | BPO |
| --- | --- |
| 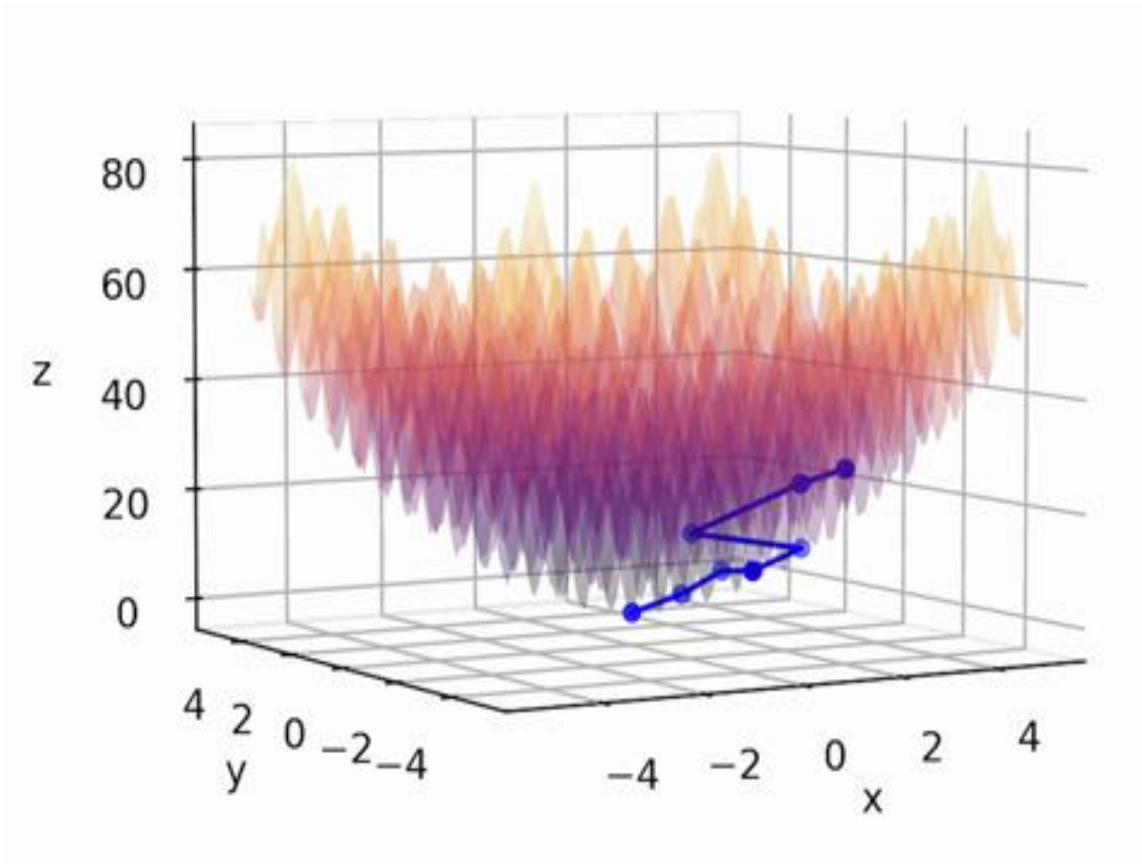 | 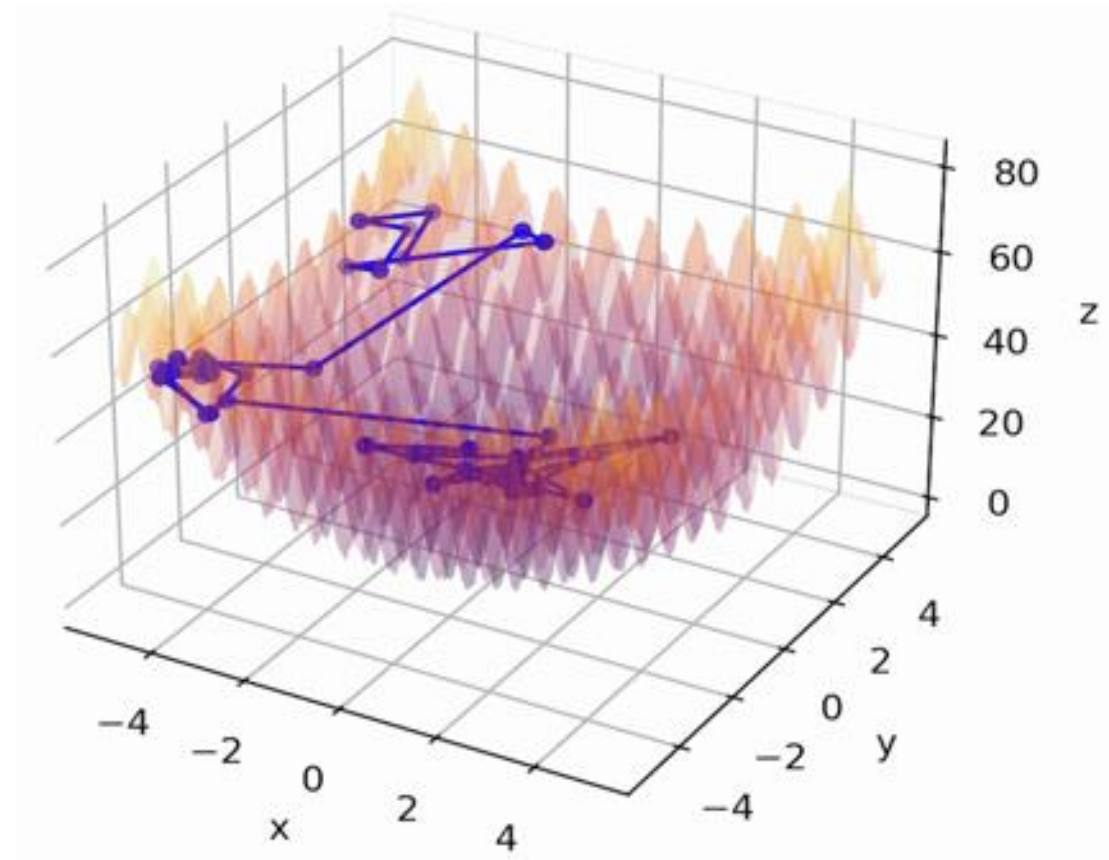 |
| Less #of trajectories | More #of trajectories |
| Exploring only the next optimal using memoized results. | Exploring all the small points until a crossing detected. |
| Less complex in tracing the trajectories. | More complex trajectory structure, even exploring what is explored. |
| Always trying to find the next most optimal solution. | Always trying to find the next most optimal solution. |
| Rejecting the results larger than intermediate most optimal result and hence reducing running complexity | Exploring all the solutions even larger than the previous one. |
| | More running time as shown by trajectory plots. |

Table 15. Comparison for Non-Continuous Rastrigin Between BSO and BPO

| BSO | BPO |
|---|---|
| 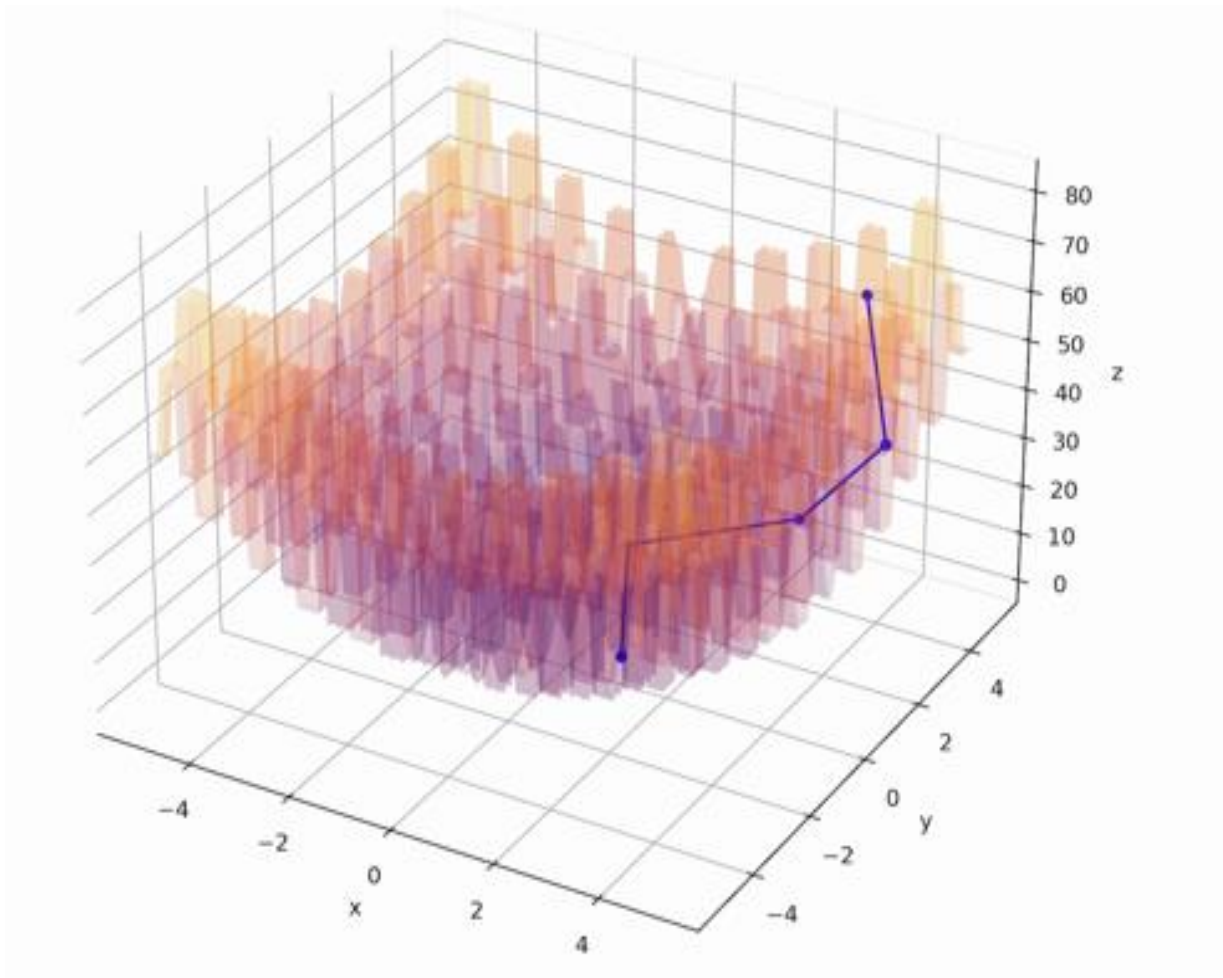 | 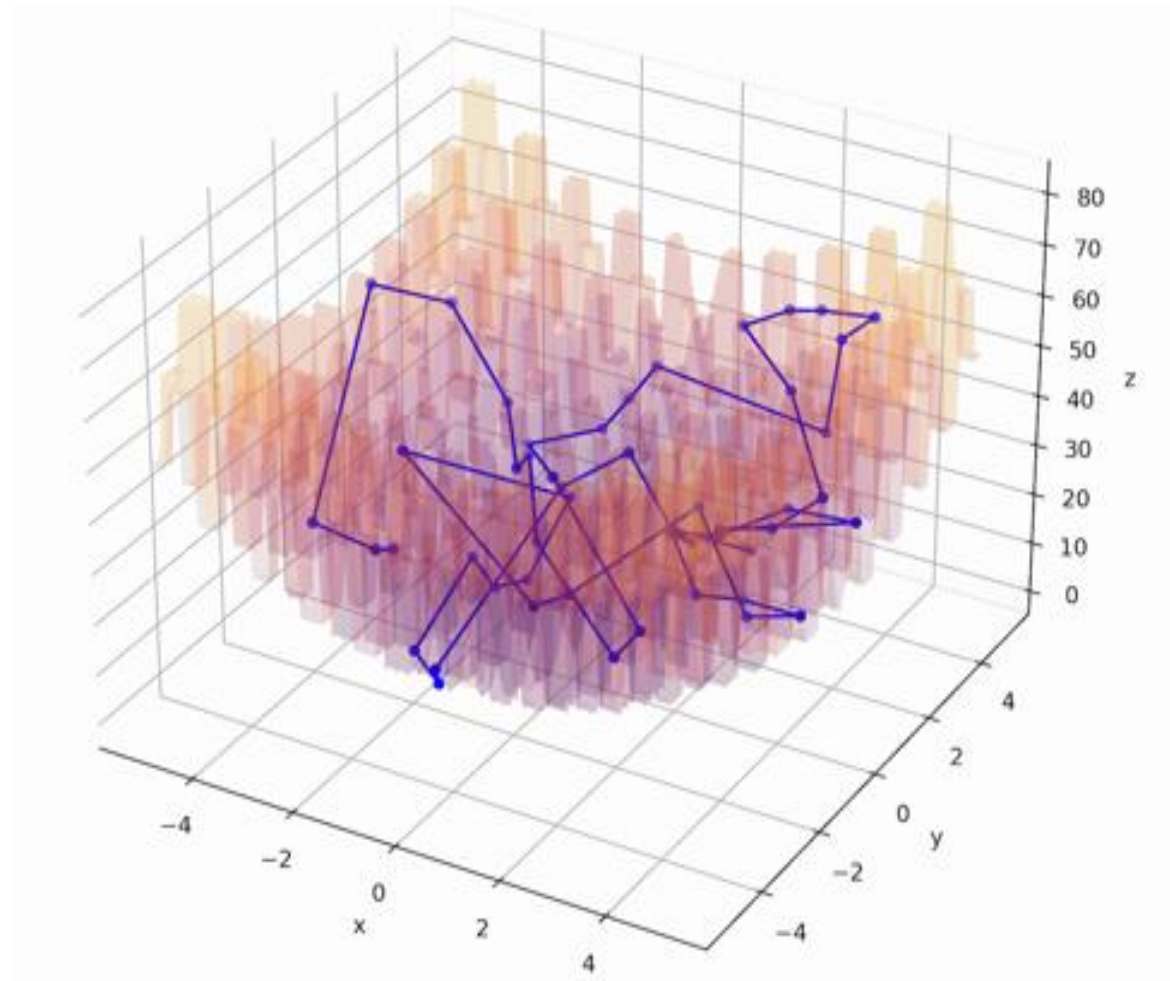 |
| Less #of trajectories | More #of trajectories |
| Exploring only the next optimal using memoized results | Exploring all the small points until a crossing detected |
| Less complex in tracing the trajectories | More complex trajectory structure, even exploring what is explored |
| Always trying to find the next most optimal solution | Exploring all the solutions even larger than the previous one |
| Rejecting the results larger than intermediate most optimal result and hence reducing running complexity | More running time as shown by trajectory plots |

# CHAPTER 6 - CONCLUSION

In this research work, we presented a new ground-breaking single point metaheuristics approach which is designed for global continuous optimization problems but have been tested on non-continuous and discrete optimization problems in multiple higher dimensions. This new algorithm which we have named as "Blindfolded Spiderman Optimization" is an extension of Buggy Pinball Optimization and mimics the behaviour of spiderman swinging from one building to another by locating the most optimal building to have a good long swing. We rigorously tested our approach against several well-established metaheuristics by fixing the time for each trial and used two different time-per-trial values respectively. By using standard evaluation metric of accuracy and precision and validating the results with parameters and non-parametric statistical significance test we concluded that the newly designed algorithm outperforms the mentioned single-point benchmarks techniques and did good when tested along particle-based approach (WO, GWO) and have shown better results than PSO in several optimization problems(continuous and non-continuous). Hence, it demonstrated strengths in handling intricate optimization functions spanning multiple dimensions.

As we look ahead, our research is more focused on further refining and testing several cooling schedules (such as exponential, logarithmic) for parameters such as step length, elevation angles to evaluate the effect on efficiency of the algorithms. Besides this, we look ahead to having a swarm of spiderman in the search space multiverse and each spiderman  tries to locate a new optimum thereby locating the best global optima and hence broadening the adaptation and applicability of the BSO. For discrete optimization (unbounded knapsack) in future work, we can BSO and benchmark metaheuristics by pre-defined Item's weight and value for each trial to have a detailed comparison on the basis of average, standard deviation and accuracy of the studied metaheuristics.

APPENDICES

# APPENDIX A: CASE STUDY-1 CONTINUOUS OPTIMIZATION PROBLEMS RESULTS

In this appendix, we are showing results obtained for continuous optimization problems of Case Study 1. Here, Table A1 shows accuracy comparison and Table A2 shows precession results comparison

Table A1. Case Study 1 Continuous Functions Results for Accuracy Comparison

| Functions | | Accuracy(%) | | | | | | |
|---|---|---|---|---|---|---|---|---|
| | D | BSO | BP | SA | TA | PSO | GWO | WO |
| Dropwave | 3D | 100 | 100 | 82 | 100 | 98 | 100 | 100 |
| Eggholder | 3D | 100 | 97 | 64 | 100 | 95 | 99 | 99 |
| Holdertable | 3D | 100 | 100 | 100 | 49 | 54 | 100 | 100 |
| Langermann | 3D | 100 | 100 | 47 | 100 | 42 | 100 | 100 |
| Shubert | 3d | 100 | 100 | 100 | 100 | 100 | 100 | 100 |
| Easom | 3d | 100 | 100 | 100 | 100 | 100 | 100 | 100 |
| Ackley | 2d | 100 | 100 | 100 | 100 | 100 | 100 | 100 |
| | 3d | 100 | 100 | 100 | 100 | 100 | 100 | 100 |
| | 4d | 100 | 100 | 100 | 100 | 100 | 100 | 100 |
| | 5d | 100 | 100 | 100 | 100 | 100 | 100 | 100 |
| | 6d | 100 | 100 | 100 | 100 | 100 | 100 | 100 |
| Rastrigin | 2d | 100 | 87 | 100 | 100 | 100 | 100 | 100 |
| | 3d | 100 | 100 | 100 | 100 | 100 | 100 | 100 |
| | 4d | 100 | 100 | 100 | 100 | 100 | 100 | 100 |
| | 5d | 100 | 100 | 100 | 100 | 95 | 100 | 100 |
| | 6d | 100 | 100 | 91 | 100 | 95 | 100 | 100 |
| Schwefel | 2d | 100 | 100 | 100 | 100 | 77 | 99 | 99 |
| | 3d | 100 | 100 | 100 | 100 | 64 | 100 | 100 |
| | 4d | 100 | 98 | 100 | 100 | 70 | 100 | 100 |
| | 5d | 100 | 100 | 100 | 99 | 75 | 100 | 100 |
| | 6d | 97 | 97 | 70 | 69 | 54 | 96 | 100 |
| Sphere | 2d | 100 | 100 | 100 | 100 | 100 | 100 | 100 |
| | 3d | 100 | 100 | 100 | 100 | 100 | 100 | 100 |
| | 4d | 100 | 100 | 100 | 100 | 100 | 100 | 100 |
| | 5d | 100 | 100 | 100 | 100 | 100 | 100 | 100 |
| | 6d | 100 | 100 | 100 | 100 | 100 | 100 | 100 |

Table A2. Case Study 1 Continuous Functions Results for Precision Comparison

| Functions | Precision(MAE) | | | | | | | |
|---|---|---|---|---|---|---|---|---|
| | D | BSO | BPO | SA | TA | PSO | GWO | WO |
| Dropwave | 3D | 1.21 E-06 | 8.97 E-05 | 7.82 E-03 | 1.94 E-03 | 8.88 E-18 | 0.00 E+00 | 0.00 E+00 |
| Eggholder | 3D | 2.88 E+00 | 1.98 E+00 | 1.45 E+00 | 3.27 E-02 | 3.54 E-05 | 2.64 E-02 | 2.73 E-02 |
| Holdertable | 3D | 2.48 E-06 | 2.63 E-06 | 9.59 E-03 | 5.05 E-03 | 1.39 E-06 | 1.33 E-04 | 2.57 E-06 |
| Langermann | 3D | 1.53 E-03 | 5.74 E-04 | 6.05 E-03 | 2.20 E-01 | 6.49 E-03 | 8.77 E-07 | 5.94 E-03 |
| Shubert | 3d | 7.79 E-03 | 5.00 E-01 | 3.46 E+00 | 1.82 E-03 | 3.22 E+00 | 5.22 E-02 | 8.75 E-06 |
| Easom | 3d | 1.91 E-07 | 1.94 E-05 | 1.06 E-02 | 3.71 E+00 | 3.22 E-17 | 4.63 E-08 | 1.12 E-06 |
| Ackley | 2d | 1.74 E-03 | 2.94 E-02 | 4.20 E-02 | 1.06 E-01 | 3.65 E-02 | 1.10 E-14 | 3.75 E-04 |
| | 3d | 4.84 E-05 | 3.42 E-04 | 6.56 E-02 | 1.02 E-01 | 4.44 E-16 | 4.44 E-16 | 5.86 E-16 |
| | 4d | 1.46 E-05 | 8.29 E-05 | 1.10 E-01 | 2.01 E-01 | 6.57 E-16 | 4.44 E-16 | 4.44 E-16 |
| | 5d | 2.52 E-06 | 1.53 E-04 | 3.29 E-01 | 5.05 E-01 | 1.15 E-15 | 4.44 E-16 | 4.44 E-16 |
| | 6d | 8.24 E-07 | 4.50 E-05 | 3.94 E-01 | 5.35 E-01 | 2.86 E-15 | 4.44 E-16 | 4.44 E-16 |
| Rastrigin | 2d | 1.18E-05 | 4.72 E-01 | 1.06 E-01 | 1.53E -02 | 2.64 E-04 | 1.64 E-04 | 1.30 E-07 |
| | 3d | 1.03E-07 | 4.54 E-01 | 5.82 E-02 | 2.46 E-02 | 1.99 E-02 | 0.00 E+00 | 0.00 E+00 |
| | 4d | 9.85E-09 | 6.60 E-04 | 2.16 E-01 | 8.33 E-02 | 2.79 E-01 | 0.00 E+00 | 0.00 E+00 |
| | 5d | 2.98E-02 | 3.74 E-07 | 1.60 E-01 | 1.89 E-01 | 3.28 E-01 | 0.00 E+00 | 0.00 E+00 |
| | 6d | 2.49E-01 | 4.90 E-07 | 4.75 E-01 | 4.37 E-01 | 2.98 E-02 | 0.00 E+00 | 0.00 E+00 |
| Schwefel | 2d | 1.80 E+00 | 1.22 E-01 | 4.20 E-01 | 1.06 E+01 | 1.90 E+01 | 1.22 E-03 | 2.77 E-05 |
| | 3d | 4.43 E-03 | 2.16 E-02 | 4.20 E-01 | 8.50 E+00 | 5.51 E-03 | 6.58 E-05 | 7.70 E-05 |
| | 4d | 2.51 E-04 | 2.66 E-02 | 3.71 E+00 | 2.88 E+01 | 1.50 E-05 | 4.85 E-05 | 3.87 E-05 |
| | 5d | 9.18 E-05 | 3.85 E-04 | 1.64 E+01 | 4.86 E+01 | 4.64 E-04 | 4.25 E-05 | 5.09 E-05 |
| | 6d | 3.25 E+00 | 1.10 E+00 | 1.18 E+01 | 5.56 E+01 | 3.49 E-04 | 1.40 E+01 | 6.36 E-05 |
| Sphere | 2d | 1.97 E-10 | 1.38 E-01 | 3.29 E-07 | 2.17 E-03 | 3.33 E-07 | 9.63 E-32 | 1.46 E-09 |
| | 3d | 1.44E-15 | 2.88 E-05 | 3.29 E-07 | 1.59 E-03 | 1.40 E-152 | 0.00 E+00 | 5.40 E-219 |
| | 4d | 3.33E-17 | 2.71 E-05 | 7.41 E-06 | 2.83 E-03 | 0.00 E+00 | 0.00 E+00 | 0.00 E+00 |
| | 5d | 4.50E-16 | 2.53 E-05 | 9.16 E-05 | 5.13 E-03 | 0.00 E+00 | 0.00 E+00 | 0.00 E+00 |
| | 6d | 9.13E-14 | 2.47 E-05 | 3.50 E-04 | 8.66 E-03 | 0.00 E+00 | 0.00 E+00 | 0.00 E+00 |

# APPENDIX B: CASE STUDY-1 NON-CONTINUOUS OPTIMIZATION PROBLEMS RESULTS

In this appendix, we are showing results obtained for non-continuous optimization problems of Case Study 1. Here, Table B1 shows accuracy comparison and Table B2 shows precession results comparison

Table B1. Case Study1 Non-Continuous Functions Results for Accuracy Comparison

| Functions | | Accuracy(%) | | | | | | |
|---|---|---|---|---|---|---|---|---|
| | D | BSO | BPO | SA | TA | PSO | GWO | WO |
| Rastrigin | 2D | 100 | 100 | 100 | 100 | 100 | 100 | 100 |
| | 3D | 100 | 100 | 100 | 100 | 100 | 100 | 100 |
| | 4D | 100 | 100 | 100 | 100 | 81 | 100 | 100 |
| | 5D | 92 | 93 | 94 | 93 | 73 | 100 | 100 |
| | 6D | 83 | 75 | 54 | 80 | 76 | 100 | 100 |
| Step Fun | 2D | 100 | 100 | 100 | 100 | 100 | 100 | 100 |
| | 3D | 100 | 100 | 100 | 100 | 100 | 100 | 100 |
| | 4D | 100 | 100 | 100 | 100 | 100 | 100 | 100 |
| | 5D | 100 | 100 | 100 | 100 | 100 | 100 | 100 |
| | 6D | 100 | 100 | 100 | 100 | 100 | 100 | 100 |
| Xin She Yang N2 | 2D | 100 | 100 | 67 | 100 | 100 | 100 | 100 |
| | 3D | 100 | 100 | 92 | 100 | 100 | 100 | 100 |
| | 4D | 100 | 100 | 100 | 100 | 94 | 100 | 100 |
| | 5D | 100 | 100 | 100 | 100 | 100 | 100 | 100 |
| | 6D | 100 | 100 | 100 | 100 | 100 | 100 | 100 |
| Rosenbrock | 2D | 100 | 100 | 100 | 100 | 100 | 100 | 100 |
| | 3D | 100 | 100 | 100 | 100 | 100 | 100 | 100 |
| | 4D | 100 | 100 | 100 | 100 | 100 | 100 | 100 |
| | 5D | 100 | 100 | 100 | 100 | 100 | 100 | 100 |
| | 6D | 100 | 100 | 100 | 100 | 100 | 100 | 100 |
| Quadric | 2D | 100 | 100 | 100 | 100 | 100 | 100 | 100 |
| | 3D | 100 | 100 | 100 | 100 | 100 | 100 | 100 |
| | 4D | 100 | 100 | 100 | 100 | 100 | 100 | 100 |
| | 5D | 100 | 100 | 100 | 100 | 100 | 100 | 100 |
| | 6D | 100 | 100 | 100 | 100 | 100 | 100 | 100 |
| | 2D | 100 | 100 | 100 | 100 | 100 | 100 | 100 |
| | 3D | 100 | 100 | 100 | 100 | 100 | 100 | 100 |
| Ellipsoid | 4D | 100 | 100 | 100 | 100 | 100 | 100 | 100 |
| | 5D | 100 | 100 | 100 | 100 | 95 | 100 | 100 |
| | 6D | 100 | 100 | 100 | 100 | 97 | 100 | 100 |

Table B2 Case Study1 Non-Continuous Functions Results for Precision Comparison

| Functions | | Precision(MAE) | | | | | | |
|---|---|---|---|---|---|---|---|---|
| | D | BSO | BPO | SA | TA | PSO | GWO | WO |
| Rastrigin | 2D | 8.11 E-05 | 4.10 E-01 | 1.65 E-02 | 4.35 E-02 | 9.23 E-02 | 6.57 E-03 | 2.04 E-07 |
| | 3D | 1.30 E-06 | 3.64 E-02 | 2.89 E-02 | 6.09 E-02 | 8.43 E-03 | 0.00 E+00 | 0.00 E+00 |
| | 4D | 6.03 E-07 | 2.41 E-06 | 7.85 E-02 | 1.15 E-01 | 0.00 E+00 | 0.00 E+00 | 0.00 E+00 |
| | 5D | 1.65 E-08 | 5.00 E-03 | 3.52 E-01 | 3.10 E-01 | 0.00 E+00 | 0.00 E+00 | 0.00 E+00 |
| | 6D | 3.25 E-08 | 5.96 E-06 | 3.17 E-01 | 3.41 E-01 | 0.00 E+00 | 0.00 E+00 | 0.00 E+00 |
| Step Fun | 2D | 1.72 E-09 | 0.00 E+00 | 0.00 E+00 | 0.00 E+00 | 0.00 E+00 | 0.00 E+00 | 0.00 E+00 |
| | 3D | 0.00 E+00 | 0.00 E+00 | 0.00 E+00 | 0.00 E+00 | 0.00 E+00 | 0.00 E+00 | 0.00 E+00 |
| | 4D | 1.56 E-09 | 0.00 E+00 | 0.00 E+00 | 0.00 E+00 | 0.00 E+00 | 0.00 E+00 | 0.00 E+00 |
| | 5D | 1.38 E-09 | 0.00 E+00 | 0.00 E+00 | 0.00 E+00 | 0.00 E+00 | 0.00 E+00 | 0.00 E+00 |
| | 6D | 3.60 E-09 | 0.00 E+00 | 0.00 E+00 | 0.00 E+00 | 0.00 E+00 | 0.00 E+00 | 0.00 E+00 |
| Xin She Yang N2 | 2D | 1.32 E-03 | 1.28 E-01 | 5.96 E-03 | 1.45 E-02 | 5.80 E-03 | 6.74 E-09 | 1.24 E-04 |
| | 3D | 4.32 E-05 | 1.28 E-01 | 9.34 E-03 | 2.35 E-02 | 1.46 E-21 | 1.59 E-220 | 7.28 E-45 |
| | 4D | 6.62 E-06 | 7.41 E-02 | 1.90 E-01 | 5.26 E-02 | 1.45 E-01 | 1.71 E-01 | 1.65 E-249 |
| | 5D | 4.35 E-02 | 1.84 E-01 | 1.07 E-01 | 9.16 E-02 | 1.25 E-01 | 9.06 E-02 | 0.00 E+00 |
| | 6D | 4.18 E-02 | 1.07 E-01 | 5.51 E-02 | 6.01 E-02 | 9.49 E-02 | 4.18 E-02 | 0.00 E+00 |
| Rosenbrock | 2D | 3.70 E-06 | 1.98 E-08 | 0.00 E+00 | 0.00 E+00 | 0.00 E+00 | 0.00 E+00 | 0.00 E+00 |
| | 3D | 8.03 E-09 | 2.42 E-04 | 0.00 E+00 | 0.00 E+00 | 0.00 E+00 | 0.00 E+00 | 0.00 E+00 |
| | 4D | 0.00 E+00 | 2.76 E-05 | 0.00 E+00 | 0.00 E+00 | 0.00 E+00 | 0.00 E+00 | 0.00 E+00 |
| | 5D | 5.76 E-09 | 0.00 E+00 | 0.00 E+00 | 0.00 E+00 | 4.69 E-05 | 0.00 E+00 | 0.00 E+00 |
| | 6D | 6.60 | 0.00 | 0.00 | 0.00 | 0.00 | 0.00 | 0.00 |
| | | E-09 | E+00 | E+00 | E+00 | E+00 | E+00 | E+00 |
| Quadric | 2D | 5.73 E-06 | 0.00 E+00 | 0.00 E+00 | 0.00 E+00 | 0.00 E+00 | 0.00 E+00 | 0.00 E+00 |
| | 3D | 2.57 E-10 | 2.04 E-05 | 4.88 E-04 | 3.55 E-04 | 4.17 E-79 | 0.00 E+00 | 0.00 E+00 |
| | 4D | 1.18 E-12 | 1.55 E-05 | 4.86 E-04 | 1.16 E-04 | 7.40 E-34 | 3.75 E-13 | 8.29 E-21 |
| | 5D | 5.72 E-14 | 6.61 E-06 | 6.26 E-04 | 5.23 E-05 | 0.00 E+00 | 2.16 E-14 | 1.17 E-25 |
| | 6D | 6.32 E-15 | 2.30 E-06 | 5.47 E-04 | 3.81 E-05 | 0.00 E+00 | 1.85 E-15 | 0.00 E+00 |
| Ellipsoid | 2D | 1.6 E-05 | 3.21E-02 | 7.8E-04 | 5.5E-03 | 1.2E-06 | 9.1E-29 | 1.9E-10 |
| | 3D | 9.1E-09 | 4.4E-05 | 1.2E-03 | 2.7E-03 | 2.1E-74 | 0.00E+00 | 0.00E+00 |
| | 4D | 2.7E-10 | 4.9E-05 | 1.6E-03 | 4.0E-03 | 0.00E+00 | 0.00E+00 | 0.00E+00 |
| | 5D | 1.3E-11 | 3.6E-05 | 3.2E-03 | 8.7E-03 | 0.00E+00 | 0.00E+00 | 0.00E+00 |
| | 6D | 4.7E-12 | 2.7E-05 | 7.0E-03 | 2.0E-02 | 0.00E+00 | 0.00E+00 | 0.00E+00 |

# APPENDIX C: CASE STUDY-2 CONTINUOUS OPTIMIZATION PROBLEMS RESULTS

In this appendix, we are showing results obtained for continuous optimization problems of Case Study 2. Here, Table C1 shows accuracy comparison and Table C2 shows precession results comparison

Table C1. Case Study 2  Continuous Functions Results for Accuracy Comparison

| Functions | | Accuracy(%) | |
|---|---|---|---|
| | D | BSO | BPO |
| Dropwave | 3D | 100 | 100 |
| Eggholder | 3D | 100 | 100 |
| Holdertable | 3D | 100 | 100 |
| Langermann | 3D | 100 | 100 |
| Shubert | 3d | 100 | 100 |
| Easom | 3d | 100 | 100 |
| Ackley | 2d | 100 | 100 |
| | 3d | 100 | 100 |
| | 4d | 100 | 100 |
| | 5d | 100 | 100 |
| | 6d | 100 | 100 |
| Rastrigin | 2d | 100 | 100 |
| | 3d | 100 | 100 |
| | 4d | 100 | 100 |
| | 5d | 100 | 100 |
| | 6d | 100 | 100 |
| Schwefel | 2d | 100 | 100 |
| | 3d | 100 | 100 |
| | 4d | 100 | 100 |
| | 5d | 100 | 100 |
| | 6d | 100 | 94 |
| Sphere | 2d | 100 | 100 |
| | 3d | 100 | 100 |
| | 4d | 100 | 100 |
| | 5d | 100 | 100 |
| | 6d | 100 | 100 |

Table C2. Case Study 2 Continuous Functions Results for Precision Comparison

| Functions | | Precision(MAE) | |
|---|---|---|---|
| | D | BSO | BPO |
| Dropwave | 3D | 7.30E-11 | 1E-4 |
| Eggholder | 3D | 1.63E+00 | 0.872 |
| Holdertable | 3D | 1.05E-05 | 1.02E-04 |
| Langermann | 3D | 1.80E-06 | 2E-6 |
| Shubert | 3d | 8.65E-06 | 0.032 |
| Easom | 3d | 1.81E-08 | 0.002 |
| Ackley | 2d | 3.73E-05 | 2E-4 |
| | 3d | 3.73E-05 | 7E-5 |
| | 4d | 3.73E-05 | 8E-5 |
| | 5d | 3.73E-05 | 9E-5 |
| | 6d | 3.73E-05 | 3E-4 |
| Rastrigin | 2d | 2.90E-07 | 7E-7 |
| | 3d | 2.90E-07 | 8E-7 |
| | 4d | 2.90E-07 | 6E-7 |
| | 5d | 2.90E-07 | 7E-7 |
| | 6d | 2.90E-07 | 7E-7 |
| Schwefel | 2d | 8.98E-05 | 2E-4 |
| | 3d | 8.98E-05 | 3E-4 |
| | 4d | 8.98E-05 | 3E-4 |
| | 5d | 8.98E-05 | 4E-4 |
| | 6d | 8.98E-05 | 3E-4 |
| Sphere | 2d | 1.80E-14 | 0.036 |
| | 3d | 1.80E-14 | 0.003 |
| | 4d | 1.80E-14 | 8E-4 |
| | 5d | 1.80E-14 | 7E-4 |
| | 6d | 1.80E-14 | 7E-4 |

# APPENDIX D: CASE STUDY-2 NON-CONTINUOUS OPTIMIZATION PROBLEMS RESULT

In this appendix, we are showing results obtained for non-continuous optimization problems of Case Study 2. Here, Table D1 shows accuracy comparison and Table D2 shows precession results comparison

Table D1. Case Study 2 Non-Continuous Functions Results for Precision Comparison

| Functions | | Accuracy(%) | | | | | | |
|---|---|---|---|---|---|---|---|---|
| | D | BSO | BPO | SA | TA | PSO | GWO | WO |
| Rastrigin | 2D | 100 | 100 | 100 | 100 | 100 | 100 | 100 |
| | 3D | 100 | 100 | 100 | 100 | 97 | 100 | 100 |
| | 4D | 100 | 100 | 100 | 100 | 89 | 100 | 100 |
| | 5D | 100 | 100 | 92 | 98 | 95 | 100 | 100 |
| | 6D | 81 | 80 | 66 | 73 | 80 | 100 | 100 |
| Step Fun | 2D | 100 | 100 | 100 | 100 | 100 | 100 | 100 |
| | 3D | 100 | 100 | 100 | 100 | 100 | 100 | 100 |
| | 4D | 100 | 100 | 100 | 100 | 100 | 100 | 100 |
| | 5D | 100 | 100 | 100 | 100 | 100 | 100 | 100 |
| | 6D | 100 | 100 | 100 | 100 | 100 | 100 | 100 |
| Xin She Yang N2 | 2D | 100 | 100 | 100 | 100 | 100 | 100 | 100 |
| | 3D | 100 | 100 | 100 | 100 | 100 | 100 | 100 |
| | 4D | 100 | 100 | 100 | 100 | 99 | 100 | 100 |
| | 5D | 100 | 100 | 100 | 100 | 100 | 100 | 100 |
| | 6D | 100 | 100 | 100 | 100 | 100 | 100 | 100 |
| Rosenbrock | 2D | 100 | 100 | 100 | 100 | 100 | 100 | 100 |
| | 3D | 100 | 100 | 100 | 100 | 100 | 100 | 100 |
| | 4D | 100 | 100 | 100 | 100 | 100 | 100 | 100 |
| | 5dD | 100 | 100 | 100 | 100 | 100 | 100 | 100 |
| | 6D | 100 | 100 | 100 | 100 | 100 | 100 | 100 |
| Quadric | 2D | 100 | 100 | 100 | 100 | 100 | 100 | 100 |
| | 3D | 100 | 100 | 100 | 100 | 100 | 100 | 100 |
| | 4D | 100 | 100 | 100 | 100 | 100 | 100 | 100 |
| | 5D | 99 | 100 | 100 | 100 | 100 | 100 | 100 |
| | 6D | 100 | 100 | 100 | 100 | 99 | 100 | 100 |

Table D2. Case Study 2 Non-Continuous Functions Results for Precision Comparison

| Functions | | Precision(MAE) | | | | | | |
|---|---|---|---|---|---|---|---|---|
| | D | BSO | BPO | SA | TA | PSO | GWO | WO |
| Rastrigin | 2d | 3.55 E-07 | 9.81 E-07 | 5.11 E-03 | 1.40 E-03 | 0.00 E+00 | 0.00 E+00 | 0.00 E+00 |
| | 3d | 5.62 | 9.80 | 1.36 | 2.24 | 0.00 | 0.00 | 0.00 |
| | | E-07 | E-07 | E-02 | E-02 | E+00 | E+00 | E+00 |
| | 4d | 2.62 E-08 | 8.36 E-07 | 6.07 E-02 | 7.24 E-02 | 0.00 E+00 | 0.00 E+00 | 0.00 E+00 |
| | 5d | 2.83 E-09 | 3.46 E-07 | 3.99 E-01 | 3.32 E-01 | 0.00 E+00 | 0.00 E+00 | 0.00 E+00 |
| | 6d | 3.51 E-10 | 5.05 E-02 | 3.87 E-01 | 3.77 E-01 | 0.00 E+00 | 0.00 E+00 | 0.00 E+00 |
| Step Fun | 2d | 1.28 E-08 | 2.03 E-09 | 2.11 E-07 | 8.24 E-09 | 9.01 E-08 | 1.11 E-06 | 1.00 E-06 |
| | 3d | 1.89 E-09 | 5.47 E-11 | 1.04 E-07 | 0.00 E+00 | 3.26 E-07 | 1.05 E-06 | 8.83 E-09 |
| | 4d | 2.42 E-09 | 1.09 E-09 | 0.00 E+00 | 0.00 E+00 | 9.13 E-08 | 8.48 E-07 | 1.32 E-10 |
| | 5d | 0.00 E+00 | 2.66 E-09 | 2.82 E-07 | 8.88 E-08 | 2.47 E-07 | 2.08 E-06 | 3.33 E-09 |
| | 6d | 2.51 E-09 | 3.22 E-09 | 3.83 E-07 | 1.66 E-08 | 2.64 E-07 | 2.24 E-08 | 3.02 E-08 |
| Xin She Yang N2 | 2d | 9.12 E-05 | 3.13 E-03 | 6.59 E-05 | 1.09 E-03 | 1.27 E-22 | 1.68 E-235 | 4.55 E-21 |
| | 3d | 8.36 E-05 | 1.83 E-03 | 6.51 E-03 | 1.24 E-02 | 3.51 E-56 | 0.00 E+00 | 1.62 E-63 |
| | 4d | 1.87 E-05 | 4.40 E-04 | 2.82 E-02 | 4.25 E-02 | 1.34 E-01 | 1.19 E-01 | 0.00 E+00 |
| | 5d | 2.63 E-02 | 5.81 E-02 | 8.80 E-02 | 9.04 E-02 | 1.34 E-01 | 9.06 E-02 | 0.00 E+00 |
| | 6d | 4.18 E-02 | 4.20 E-02 | 4.34 E-02 | 4.60 E-02 | 9.20 E-02 | 4.18 E-02 | 7.78 E-02 |
| Rosenbrock | 2d | 3.70 E-06 | 0.00 E+00 | 1.14 E+16 | 0.00 E+00 | 0.00 E+00 | 0.00 E+00 | 0.00 E+00 |
| | 3d | 0.00 E+00 | 9.21 E+10 | 0.00 E+00 | 0.00 E+00 | 0.00 E+00 | 0.00 E+00 | 0.00 E+00 |
| | 4d | 0.00 E+00 | 0.00 E+00 | 0.00 E+00 | 6.44 E-11 | 8.77 E-06 | 0.00 E+00 | 1.24 E-10 |
| | 5d | 4.45 E-06 | 0.00 E+00 | 3.05 E+15 | 2.78 E-06 | 1.34 E-05 | 3.15 E-17 | 1.99 E-10 |
| | 6d | 2.80 E-06 | 0.00 E+00 | 3.33 E+15 | 8.49 E-11 | 5.14 E-06 | 2.52 E-06 | 1.29 E-09 |
| Quadric | 2d | 0.00 E+00 | 8.69 E-11 | 1.33 E-04 | 0.00 E+00 | 1.32 E-10 | 6.11 E-11 | 0.00 E+00 |
| | 3d | 3.08 E-12 | 1.12 E-05 | 4.59 E-04 | 1.91 E-04 | 1.51 E-173 | 0.00 E+00 | 0.00 E+00 |
| | 4d | 3.50 | 1.36 | 4.55 | 5.23 | 9.86 | 1.25 | 1.87 |
| | | E-14 | E-05 | E-04 | E-05 | E-34 | E-14 | E-25 |
| | 5d | 5.86<br>E-15 | 1.86<br>E-06 | 4.71<br>E-04 | 3.51<br>E-05 | 0.00<br>E+00 | 2.07<br>E-15 | 5.61<br>E-27 |
| | 6d | 1.09<br>E-15 | 2.22<br>E-06 | 4.48<br>E-04 | 2.15<br>E-05 | 0.00<br>E+00 | 3.34<br>E-15 | 0.00<br>E+00 |

# APPENDIX E: CASE STUDY-3 DISCRETE OPTIMIZATION PROBLEMS - UNBOUNDED KNAPSACK RESULTS

In this appendix, we are showing results obtained for discrete optimization problem-Unbounded Knapsack of Case Study 3. Here, Table E1, E2, E3, E4, E5 shows statistical significance results based on precision(optimal value obtained) for dimensions 2,3,4,5,6.

Table E1. Two Dimension With P-value 2.9571175258070256e-24 < 0.05

| | BSO | BPO | SA | TA | PSO | WO | GWO |
|---|---|---|---|---|---|---|---|
| BSO | 1.000000e+00 | 2.429853e-14. | 2.197290e-15 | 2.295437e-11 | 1.112986e-09 | 1.330203e-13 | 1.000000e+00 |
| BPO | 2.429853e-14 | 1.000000e+00 | 1.000000e+00 | 1.000000e+00 | 1.000000e+00 | 1.000000e+00 | 4.212369e-13 |
| SA | 2.197290e-15 | 1.000000e+00 | 1.000000e+00 | 1.000000e+00 | 1.000000e+00 | 1.000000e+00 | 4.160598e-14 |
| TA | 2.295437e-11 | 1.000000e+00 | 1.000000e+00 | 1.000000e+00 | 1.000000e+00 | 1.000000e+00 | 3.011385e-10 |
| PSO | 1.112986e-09 | 1.000000e+00 | 1.000000e+00 | 1.000000e+00 | 1.000000e+00 | 1.000000e+00 | 1.220023e-08 |
| WO | 1.330203e-13 | 1.000000e+00 | 1.000000e+00 | 1.000000e+00 | 1.000000e+00 | 1.000000e+00 | 2.160470e-12 |
| GWO | 1.000000e+00 | 4.212369e-13 | 4.160598e-14 | 3.011385e-10 | 1.220023e-08 | 2.160470e-12 | 1.000000e+00 |

Table E2. Three Dimension With P-value 7.0305693255017766e-15 < 0.05

| | BSO | BPO | SA | TA | PSO | WO | GWO |
|---|---|---|---|---|---|---|---|
| BSO | 1.000000e+00. | 6.008048e-16 | 1.217027e-10 | 2.659593e-11 | 2.426320e-09 | 1.257057e-12 | 1.000000e+00 |
| BPO | 6.008048e-16. | 1.000000e+00 | 1.000000e+00 | 1.000000e+00 | 6.993989e-01 | 1.000000e+00 | 5.585481e-14 |
| SA | 1.217027e-10. | 1.000000e+00 | 1.000000e+00 | 1.000000e+00 | 1.000000e+00 | 1.000000e+00 | 5.334405e-09 |
| TA | 2.659593e-11 | 1.000000e+00 | 1.000000e+00 | 1.000000e+00 | 1.000000e+00 | 1.000000e+00 | 1.295478e-09 |
| PSO | 2.426320e-09 | 6.993989e-01 | 1.000000e+00 | 1.000000e+00 | 1.000000e+00 | 1.000000e+00 | 8.532992e-08 |
| WO | 1.257057e-12 | 1.000000e+00 | 1.000000e+00 | 1.000000e+00 | 1.000000e+00 | 1.000000e+00 | 7.481436e-11 |
| GWO | 1.000000e+00 | 5.585481e-14 | 5.334405e-09 | 1.295478e-09 | 8.532992e-08 | 7.481436e-11 | 1.000000e+00 |

Table E3. Four Dimension With P-value 7.1396328550292165e-12 < 0.05

| | BSO | BPO | SA | TA | PSO | WO | GWO |
|---|---|---|---|---|---|---|---|
| BSO | 1.000000 | 1.000000 | 1.000000 | 1.000000 | 5.592669e-02 | 1.000000 | 1.673616e-01 |
| BPO | 1.000000 | 1.000000 | 1.000000 | 1.000000 | 1.153599e-01 | 1.000000 | 8.322111e-02 |
| SA | 1.000000 | 1.000000 | 1.000000 | 1.000000 | 1.142531e-04 | 1.000000 | 1.000000e+00 |
| TA | 1.000000 | 1.000000 | 1.000000 | 1.000000 | 1.594106e-03 | 1.000000 | 1.000000e+00 |
| PSO | 0.055927 | 0.115360 | 0.000114 | 0.001594 | 1.000000e+00 | 0.000228 | 4.254943e-07 |
| WO | 1.000000 | 1.000000 | 1.000000 | 1.000000 | 2.281578e-04 | 1.000000 | 1.000000e+00 |
| GWO | 0.167362 | 0.083221 | 1.000000 | 1.000000 | 4.254943e-07 | 1.000000 | 1.000000e+00 |

Table E4. Five Dimension With P-value 2.2440660674257614e-13 < 0.05

| | BSO | BPO | SA | TA | PSO | WO | GWO |
|---|---|---|---|---|---|---|---|
| BSO | 1.000000 | 1.000000 | 0.582394 | 1.00000 | 0.573005 | 0.781292 | 1.000000 |
| BPO | 1.000000 | 1.000000 | 1.000000 | 1.00000 | 0.019286 | 1.000000 | 1.000000 |
| SA | 0.582394 | 1.000000 | 1.000000 | 1.00000 | 0.000243 | 1.000000 | 1.000000 |
| TA | 1.000000 | 1.000000 | 1.000000 | 1.00000 | 0.004680 | 1.000000 | 1.000000 |
| PSO | 0.573005 | 0.019286 | 0.000243 | 0.00468 | 1.000000 | 0.000411 | 0.004102 |
| WO | 0.781292 | 1.000000 | 1.000000 | 1.00000 | 0.000411 | 1.000000 | 1.000000 |
| GWO | 1.000000 | 1.000000 | 1.000000 | 1.00000 | 0.004102 | 1.000000 | 1.000000 |

Table E5. Six Dimension With P-value 2.9560489050074697e-14 < 0 .05

| | BSO | BPO | SA | TA | PSO | WO | GWO |
|---|---|---|---|---|---|---|---|
| BSO | 1.000000 | 3.464605e-02 | 1.000000 | 1.000000 | 2.938537e-05 | 1.000000 | 1.000000 |
| BPO | 0.034646 | 1.000000e+00 | 0.000972 | 0.000147 | 8.982398e-14 | 0.000021 | 0.002472 |
| SA | 1.000000 | 9.715433e-04 | 1.000000 | 1.000000 | 1.984248e-03 | 1.000000 | 1.000000 |
| TA | 1.000000 | 1.470872e-04 | 1.000000 | 1.000000 | 1.043724e-02 | 1.000000 | 1.000000 |
| PSO | 0.000029 | 8.982398e-14 | 0.001984 | 0.010437 | 1.000000e+00 | 0.044241 | 0.000771 |
| WO | 1.000000 | 2.051672e-05 | 1.000000 | 1.000000 | 4.424135e-02 | 1.000000 | 1.000000 |
| GWO | 1.000000 | 2.471855e-03 | 1.000000 | 1.000000 | 7.714280e-04 | 1.000000 | 1.000000 |

# APPENDIX F: CASE STUDY 4 CONTINUOUS AND NON-CONTINUOUS OPTIMIZATION PROBLEMS 3D COMPARISON FOR MULTIPLE TIMES

In this appendix, we are showing results obtained for continuous and non-continuous optimization problems of Case Study 4. Here, we have four tables labeled as Table F1, Table F2, Table F3, Table F4.

Table F1, shows comparative results of accuracy for continuous optimization functions. Table F2, shows comparative results of precision for continuous optimization functions. Table F3, shows comparative results of accuracy for non-continuous optimization functions and, Table F4, shows comparative results of precision for non-continuous optimization functions

Table F1. Case Study 4 Accuracy Comparative Results for Continuous Functions

| Functions | | Accuracy(%) | | | | | | |
|---|---|---|---|---|---|---|---|---|
| | Time (sec) | BSO | BPO | SA | TA | PSO | GWO | WO |
| Dropwave | 0.1 | 100 | 100 | 91 | 13 | 100 | 100 | 95 |
| | 0.2 | 100 | 100 | 100 | 13 | 100 | 100 | 98 |
| | 0.5 | 100 | 100 | 100 | 25 | 100 | 100 | 100 |
| | 1 | 100 | 100 | 100 | 41 | 100 | 100 | 100 |
| | 2 | 100 | 100 | 100 | 61 | 100 | 100 | 100 |
| | 5 | 100 | 100 | 100 | 86 | 100 | 100 | 100 |
| Eggholder | 0.1 | 85 | 99 | 48 | 89 | 52 | 90 | 93 |
| | 0.2 | 91 | 100 | 55 | 91 | 57 | 97 | 96 |
| | 0.5 | 100 | 97 | 63 | 100 | 58 | 100 | 94 |
| | 1 | 100 | 100 | 59 | 45 | 60 | 99 | 99 |
| | 2 | 100 | 100 | 49 | 59 | 49 | 97 | 98 |
| | 5 | 100 | 100 | 59 | 77 | 59 | 99 | 98 |
| Holdertable | 0.1 | 100 | 100 | 100 | 100 | 47 | 100 | 100 |
| | 0.2 | 100 | 100 | 100 | 100 | 49 | 100 | 100 |
| | 0.5 | 100 | 100 | 100 | 100 | 100 | 100 | 100 |
| | 1 | 100 | 100 | 100 | 100 | 66 | 99 | 100 |
| | 2 | 100 | 100 | 100 | 100 | 71 | 100 | 100 |
| | 5 | 100 | 100 | 100 | 100 | 100 | 100 | 100 |
| Langermann | 0.1 | 93 | 42 | 33 | 52 | 33 | 95 | 97 |
| | 0.2 | 98 | 59 | 30 | 7 | 61 | 97 | 81 |
| | 0.5 | 100 | 94 | 60 | 10 | 61 | 100 | 92 |
| | 1 | 100 | 99 | 20 | 95 | 72 | 100 | 100 |
| | 2 | 100 | 100 | 97 | 76 | 53 | 100 | 98 |
| | 5 | 100 | 100 | 100 | 81 | 61 | 100 | 100 |
| Shubert | 0.1 | 100 | 100 | 100 | 100 | 100 | 100 | 100 |
| | 0.2 | 100 | 99 | 100 | 100 | 93 | 100 | 100 |
| | 0.5 | 100 | 100 | 100 | 100 | 100 | 100 | 100 |
| | 1 | 100 | 100 | 100 | 100 | 100 | 100 | 100 |
| | 2 | 100 | 100 | 100 | 100 | 99 | 100 | 100 |
| | 5 | 100 | 100 | 100 | 100 | 100 | 100 | 100 |
| Easom | 0.1 | 100 | 100 | 100 | 100 | 100 | 100 | 100 |
| | 0.2 | 100 | 100 | 100 | 100 | 100 | 100 | 100 |
| | 0.5 | 100 | 100 | 100 | 100 | 100 | 100 | 100 |
| | 1 | 100 | 100 | 100 | 100 | 100 | 100 | 100 |
| | 2 | 100 | 100 | 100 | 100 | 100 | 100 | 100 |
| | 5 | 100 | 100 | 100 | 100 | 100 | 100 | 100 |
| Ackley | 0.1 | 100 | 100 | 100 | 100 | 100 | 100 | 100 |
| | 0.2 | 100 | 100 | 100 | 100 | 100 | 100 | 100 |
| | 0.5 | 100 | 100 | 100 | 100 | 100 | 100 | 100 |
| | 1 | 100 | 100 | 100 | 100 | 100 | 100 | 100 |
| | 2 | 100 | 100 | 100 | 100 | 100 | 100 | 100 |
| | 5 | 100 | 100 | 100 | 100 | 100 | 100 | 100 |
| Rastrigin | 0.1 | 100 | 100 | 98 | 99 | 100 | 100 | 100 |
| | 0.2 | 100 | 100 | 100 | 100 | 100 | 100 | 100 |
| | 0.5 | 100 | 100 | 100 | 100 | 100 | 100 | 100 |
| | 1 | 100 | 100 | 100 | 100 | 100 | 100 | 100 |
| | 2 | 100 | 100 | 100 | 100 | 100 | 100 | 100 |
| | 5 | 100 | 100 | 100 | 100 | 100 | 100 | 100 |
| Schwefel | 0.1 | 99 | 99 | 60 | 100 | 58 | 97 | 100 |
| | 0.2 | 99 | 100 | 100 | 100 | 100 | 100 | 100 |
| | 0.5 | 100 | 100 | 84 | 100 | 86 | 100 | 100 |
| | 1 | 100 | 100 | 80 | 100 | 92 | 100 | 100 |
| | 2 | 100 | 100 | 88 | 100 | 98 | 100 | 100 |
| | 5 | 100 | 100 | 100 | 100 | 100 | 100 | 100 |
| Sphere | 0.1 | 100 | 100 | 100 | 100 | 100 | 100 | 100 |
| | 0.2 | 100 | 100 | 100 | 100 | 100 | 100 | 100 |
| | 0.5 | 100 | 100 | 100 | 100 | 100 | 100 | 100 |
| | 1 | 100 | 100 | 100 | 100 | 100 | 100 | 100 |
| | 2 | 100 | 100 | 100 | 100 | 100 | 100 | 100 |
| | 5 | 100 | 100 | 100 | 100 | 100 | 100 | 100 |

Table F2. Case Study 4 Precision Comparative Results for Continuous Functions

| Functions | | Precision (MAE) | | | | | | |
|---|---|---|---|---|---|---|---|---|
| | Time | BSO | BPO | SA | TA | PSO | GWO | WO |
| | (sec) | | | | | | | |
| Dropwave | 0.1 | 4.11 E-06 | 1.02 E-03 | 1.43 E-02 | 3.09 E-03 | 2.48 E-04 | 1.05 E-08 | 4.46 E-11 |
| | 0.2 | 2.39 E-06 | 6.36 E-06 | 1.12 E-02 | 3.09 E-03 | 1.32 E--05 | 1.15 E-12 | 3.54 E-13 |
| | 0.5 | 2.16 E-10 | 2.69 E-06 | 5.34 E-03 | 5.83 E-03 | 1.27 E-06 | 0.00 E+00 | 1.48 E-16 |
| | 1 | 2.42 E-09 | 1.43 E-06 | 2.01 E-03 | 9.77 E-03 | 7.97 E-08 | 0.00 E+00 | 0.00 E+00 |
| | 2 | 6.99 E-11 | 1.01 E-06 | 1.62 E-03 | 1.02 E-02 | 5.73 E-09 | 0.00 E+00 | 0.00 E+00 |
| | 5 | 3.10 E-12 | 1.03 E-06 | 7.15 E-04 | 1.50 E-02 | 2.42 E-10 | 0.00 E+00 | 0.00 E+00 |
| Eggholder | 0.1 | 6.53 E-01 | 9.12 E+00 | 2.38 E+00 | 9.48 E+00 | 2.19 E-05 | 2.63 E+00 | 1.91 E-01 |
| | 0.2 | 3.28 E-01 | 4.19 E+00 | 2.21 E+00 | 9.29 E+00 | 2.29 E-05 | 1.68 E+00 | 5.45 E-02 |
| | 0.5 | 1.14 E-01 | 2.86 E+00 | 1.71 E+00 | 6.78 E+00 | 2.18 E-05 | 6.37 E-01 | 2.73 E-02 |
| | 1 | 2.54 E-06 | 2.14 E+00 | 1.01 E+00 | 5.56 E+00 | 2.24 E-05 | 1.13 E-01 | 3.62 E-05 |
| | 2 | 4.79 E-03 | 1.46 E+00 | 6.51 E-01 | 7.11 E+00 | 1.82 E-06 | 3.18 E-02 | 3.65 E-05 |
| | 5 | 2.75 E-03 | 4.75 E-01 | 4.42 E-01 | 8.16 E+00 | 2.20 E-05 | 8.63 E-03 | 3.66 E-05 |
| Holdertable | 0.1 | 2.86 E-06 | 1.61 E-04 | 2.61 E-03 | 4.66 E-02 | 8.63 E-06 | 2.51 E-03 | 2.55 E-06 |
| | 0.2 | 2.03 E-06 | 9.36 E-06 | 1.98 E-03 | 3.54 E-02 | 1.26 E-06 | 8.20 E-04 | 2.54 E-06 |
| | 0.5 | 2.32 E-06 | 5.03 E-06 | 1.17 E-03 | 2.49 E-02 | 1.24 E-06 | 1.76 E-04 | 2.54 E-06 |
| | 1 | 8.47 E-04 | 3.10 E-06 | 7.14 E-04 | 1.51 E-02 | 1.69 E-06 | 9.88 E-05 | 2.54 E-06 |
| | 2 | 2.57 E-06 | 2.70 E-06 | 3.87 E-04 | 9.98 E-03 | 1.82 E-06 | 2.54 E-06 | 4.64 E-05 |
| | 5 | 2.57 E-03 | 2.06 E-06 | 3 E-04 | 7.58 E-03 | 1.93 E-06 | 4.52 E-05 | 2.57 E-06 |
| Langermann | 0.1 | 5.93 E-03 | 1.66 E-03 | 4.64 E-03 | 4.97 E-03 | 3.67 E-03 | 4.64 E-03 | 1.15 E-02 |
| | 0.2 | 4.23 E-03 | 2.11 E-03 | 4.12 E-03 | 1.21 E-03 | 5.93 E-03 | 1.60 E-03 | 1.23 E-02 |
| | 0.5 | 2.82 | 6.49 | 7.56 | 1.69 | 6.49 | 1.11 | 1.72 |
| | | E-03 | E-03 | E-03 | E-03 | E-03 | E-05 | E-02 |
| | 1 | 4.10<br>E-05 | 1.05<br>E-03 | 7.94<br>E-03 | 1.65<br>E-03 | 4.52<br>E-03 | 1.71<br>E-06 | 9.60<br>E-03 |
| | 2 | 3.03<br>E-10 | 2.86<br>E-04 | 5.48<br>E-03 | 8.49<br>E-03 | 5.36<br>E-03 | 6.99<br>E-07 | 1.61<br>E-02 |
| | 5 | 6.36<br>E-11 | 1.41<br>E-06 | 2.83<br>E-03 | 7.92<br>E-03 | 3.39<br>E-03 | 2.75<br>E-07 | 1.04<br>E-02 |
| Shubert | 0.1 | 4.66<br>E-03 | 5<br>E-01 | 5.43<br>E-01 | 9.47<br>E-01 | 5.97<br>E-02 | 4.52<br>E-01 | 1.05<br>E-05 |
| | 0.2 | 1.06<br>E-03 | 1.19<br>E+02 | 3.31<br>E-01 | 6.24<br>E-01 | 1.01<br>E-02 | 1.71<br>E-01 | 8.36<br>E-06 |
| | 0.5 | 1.65<br>E-04 | 8.99<br>E+01 | 1.57<br>E-01 | 4.31<br>E-01 | 3.44<br>E-03 | 4.82<br>E-02 | 8.78<br>E-06 |
| | 1 | 4.20<br>E-02 | 5.33<br>E+01 | 5.89<br>E-02 | 2.76<br>E-01 | 1.02<br>E-04 | 9.88<br>E-03 | 8.82<br>E-06 |
| | 2 | 8.39<br>E-06 | 2.04<br>E+01 | 3.83<br>E-02 | 1.98<br>E-01 | 8.90<br>E-06 | 3.63<br>E-03 | 8.83<br>E-06 |
| | 5 | 7.23<br>E-06 | 4.74<br>E+00 | 1.97<br>E-02 | 1.34<br>E-01 | 8.78<br>E-06 | 3.27<br>E-03 | 8.83<br>E-06 |
| Easom | 0.1 | 1.31<br>E-06 | 3.96<br>E-05 | 9.87<br>E-04 | 6.84<br>E-03 | 3.77<br>E-04 | 7.32<br>E-07 | 6.34<br>E-04 |
| | 0.2 | 3.59 E-07 | 2.89 E-05 | 7.36 E-04 | 4.15 E-03 | 9.28 E-05 | 1.51 E-07 | 2.28 E-04 |
| | 0.5 | 1.77 E-08 | 2.36 E-05 | 4.36 E-04 | 3.59 E-03 | 7.15 E-06 | 2.71 E-08 | 4.01 E-05 |
| | 1 | 1.13 E-02 | 2.21 E-05 | 2.88 E-04 | 2.48 E-03 | 2.63 E-07 | 6.35 E-09 | 1.40 E-05 |
| | 2 | 8.25 E-05 | 1.86 E-05 | 1.82 E-04 | 1.42 E-03 | 8.56 E-09 | 1.77 E-09 | 2.32 E-06 |
| | 0.5 | 1.07 E-10 | 1.86 E-05 | 1.36 E-04 | 9.66 E-04 | 3.80 E-11 | 4.79 E-10 | 7.73 E-07 |
| Ackley | 0.1 | 1.38 E-03 | 1.04 E+00 | 6.76 E-02 | 2.36 E-01 | 5.23 E-03 | 4.44 E-16 | 1.62 E-15 |
| | 0.2 | 6.95 E-04 | 2.28 E-01 | 4.88 E-02 | 1.91 E-01 | 1.07 E-03 | 4.44 E-16 | 4.44 E-16 |
| | 0.5 | 3.46 E-04 | 2.16 E-02 | 3.44 E-02 | 1.27 E-01 | 6.26 E-04 | 4.44 E-16 | 4.44 E-16 |
| | 1 | 5.02 E-10 | 8.81 E-03 | 2.09 E-02 | 9.75 E-02 | 1.01 E-04 | 4.44 E-16 | 4.44 E-16 |
| | 2 | 5.71 E-05 | 4.17 E-03 | 1.39 E-02 | 7.96 E-02 | 2.36 E-05 | 4.44 E-16 | 4.44 E-16 |
| | 5 | 2.63 | 1.95 | 1.06 | 5.75 | 4.42 | 4.44 | 4.44 |
| | | E-05 | E-03 | E-02 | E-02 | E-06 | E-16 | E-16 |
| Rastrigin | 0.1 | 3.34 E-04 | 1.95 E-03 | 2.42 E-01 | 1.67 E-01 | 3.85 E-02 | 0.00 E+00 | 0.00 E+00 |
| | 0.2 | 8.90 E-05 | 1.71 E-03 | 1.74 E-01 | 1.81 E-01 | 4.17 E-04 | 0.00 E+00 | 0.00 E+00 |
| | 0.5 | 1.57 E-05 | 1.65 E-03 | 5.92 E-02 | 1.10 E-01 | 7.81 E-05 | 0.00 E+00 | 0.00 E+00 |
| | 1 | 4.85 E-11 | 5.02 E-05 | 3.22 E-02 | 9.30 E-02 | 1.55 E-06 | 0.00 E+00 | 0.00 E+00 |
| | 2 | 9.77 E-07 | 1.17 E-01 | 1.63 E-02 | 5.70 E-02 | 6.83 E-08 | 0.00 E+00 | 0.00 E+00 |
| | 5 | 8.52 E-09 | 1.95 E-06 | 8.49 E-03 | 3.72 E-02 | 5.04 E-09 | 0.00 E+00 | 0.00 E+00 |
| Schwefel | 0.1 | 3.16 E-05 | 5.61 E-04 | 2.41 E-01 | 2.43 E+01 | 6.6 E-03 | 4.82 E-03 | 2.46 E-03 |
| | 0.2 | 9.74 E-05 | 5.86 E-04 | 1.32 E-01 | 1.62 E+01 | 1.79 E-03 | 1.34 E-03 | 5.76 E-04 |
| | 0.5 | 3.72 E-05 | 3.60 E-04 | 8.03 E-02 | 1.19 E+01 | 5.66 E-04 | 3.24 E-04 | 1.46 E-04 |
| | 1 | 5.00 E-06 | 3.11 E-04 | 3.66 E-02 | 7.71 E+00 | 5.56 E-04 | 9.87 E-05 | 6.63 E-05 |
| | 2 | 2.59 E-05 | 2.83 E-04 | 2 E-02 | 5.63 E+00 | 2.68 E-04 | 4.25 E-05 | 3.33 E-05 |
| | 5 | 2.56 E-05 | 2.65 E-04 | 1.63 E-02 | 3.52 E+00 | 1.08 E-04 | 2.95 E-05 | 2.66 E-05 |
| Sphere | 0.1 | 7.73 E-08 | 3.13 E-05 | 7.40 E-04 | 1.38 E-02 | 2.36 E-09 | 3.57 E-155 | 7.31 E-64 |
| | 0.2 | 2.16 E-08 | 3.38 E-05 | 4.71 E-04 | 8.88 E-03 | 3.78 E-07 | 0.00 E+00 | 4.27 E-134 |
| | 0.5 | 3.08 E-09 | 2.84 E-05 | 3.25 E-04 | 6.14 E-03 | 5.15 E-09 | 0.00 E+00 | 1.07 E-267 |
| | 1 | 2.72 E-05 | 2.72 E-05 | 1.79 E-04 | 4.82 E-03 | 1.18 E-09 | 0.00 E+00 | 0.00 E+00 |
| | 2 | 1.23 E-10 | 2.77 E-05 | 1.58 E-04 | 3.25 E-03 | 1.18 E-09 | 0.00 E+00 | 0.00 E+00 |
| | 5 | 1.98 E-11 | 2.30 E-05 | 1.35 E-04 | 2.43 E-03 | 6.45 E-13 | 0.00 E+00 | 0.00 E+00 |

Table F3. Case Study 4 Accuracy Comparative Results for Non-Continuous Functions

| Functions | | Accuracy(%) | | | | | | |
|---|---|---|---|---|---|---|---|---|
| | D | BSO | BPO | SA | TA | PSO | GWO | WO |
| Ellipsoid | 0.1 | 100 | 97 | 100 | 100 | 100 | 100 | 100 |
| | 0.2 | 100 | 95 | 100 | 100 | 100 | 100 | 100 |
| | 0.5 | 100 | 98 | 100 | 100 | 100 | 100 | 100 |
| | 1 | 100 | 99 | 100 | 100 | 100 | 100 | 100 |
| | 2 | 100 | 100 | 100 | 100 | 100 | 100 | 100 |
| | 5 | 100 | 100 | 100 | 100 | 100 | 100 | 100 |
| Rastrigin | 0.1 | 99 | 98 | 100 | 100 | 100 | 100 | 100 |
| | 0.2 | 100 | 100 | 100 | 100 | 100 | 100 | 100 |
| | 0.5 | 100 | 99 | 100 | 100 | 100 | 100 | 100 |
| | 1 | 100 | 100 | 100 | 100 | 100 | 100 | 100 |
| | 2 | 100 | 100 | 100 | 100 | 100 | 100 | 100 |
| | 5 | 100 | 100 | 100 | 100 | 100 | 100 | 100 |
| Step Fun | 0.1 | 100 | 45* | 100 | 100 | 100 | 100 | 100 |
| | 0.2 | 100 | 80 | 100 | 100 | 100 | 100 | 100 |
| | 0.5 | 100 | 100 | 100 | 100 | 100 | 100 | 100 |
| | 1 | 100 | 100 | 100 | 100 | 100 | 100 | 100 |
| | 2 | 100 | 100 | 100 | 100 | 100 | 100 | 100 |
| | 5 | 100 | 100 | 100 | 100 | 100 | 100 | 100 |
| | 0.1 | 100 | 100 | 100 | 18 | 100 | 100 | 100 |
| Xin She Yang N2 | 0.2 | 100 | 100 | 100 | 12 | 100 | 100 | 100 |
| | 0.5 | 100 | 100 | 100 | 21 | 100 | 100 | 100 |
| | 1 | 100 | 100 | 100 | 12 | 100 | 100 | 100 |
| | 2 | 100 | 100 | 100 | 14 | 100 | 100 | 100 |
| | 5 | 100 | 100 | 100 | 25 | 100 | 100 | 100 |
| Rosenbrock | 0.1 | 100 | 81 | 100 | 100 | 100 | 100 | 100 |
| | 0.2 | 100 | 83 | 100 | 100 | 100 | 100 | 100 |
| | 0.5 | 100 | 81 | 100 | 100 | 100 | 100 | 100 |
| | 1 | 100 | 85 | 100 | 100 | 100 | 100 | 100 |
| | 2 | 100 | 83 | 100 | 100 | 100 | 100 | 100 |
| | 5 | 100 | 100 | 100 | 100 | 100 | 100 | 100 |
| Quadric | 0.1 | 100 | 97 | 100 | 100 | 100 | 100 | 100 |
| | 0.2 | 100 | 98 | 100 | 100 | 100 | 100 | 100 |
| | 0.5 | 100 | 100 | 100 | 100 | 100 | 100 | 100 |
| | 1 | 100 | 100 | 100 | 100 | 100 | 100 | 100 |
| | 2 | 100 | 100 | 100 | 100 | 100 | 100 | 100 |
| | 5 | 100 | 100 | 100 | 100 | 100 | 100 | 100 |

Table F4. Case Study 4 Precision Comparative Results for Non-Continuous Functions

| Functions | | Precision (MAE) | | | | | | |
|---|---|---|---|---|---|---|---|---|
| | D | BSO | BPO | SA | TA | PSO | GWO | WO |
| Rastrigin | 0.1 | 3.84 E-04 | 2.69 E-01 | 6.76 E-02 | 4.82 E-02 | 6.45 E-02 | 0.0 E+00 | 5.33 E-17 |
| | 0.2 | 2.23 E-07 | 1.07 E-01 | 2.86 E-02 | 2.69 E-02 | 1.14 E-02 | 0.0 E+00 | 0.0 E+00 |
| | 0.5 | 3.79 E-08 | 7.61 E-01 | 1.78 E-02 | 1.67 E-02 | 1.87 E-03 | 0.0 E+00 | 0.0 E+00 |
| | 1 | 5.82 E-09 | 5.74 E-01 | 1.01 E-02 | 9.10 E-03 | 5.17 E-05 | 0.0 E+00 | 0.0 E+00 |
| | 2 | 2.85 E-10 | 2.82 E-01 | 4.86 E-03 | 8.73 E-03 | 5.17 E-05 | 0.0 E+00 | 0.0 E+00 |
| | 5 | 2.19 E-10 | 2.47 E-04 | 3.03 E-03 | 6.30 E-03 | 3.22 E-08 | 0.0 E+00 | 0.0 E+00 |
| Step Fun | 0.1 | 0.00 E+00 | 0.00 E+00 | 0.00 E+00 | 0.00 E+00 | 0.00 E+00 | 0.00 E+00 | 0.00 E+00 |
| | 0.2 | 0.00 E+00 | 0.00 E+00 | 0.00 E+00 | 0.00 E+00 | 0.00 E+00 | 0.00 E+00 | 0.00 E+00 |
| | 0.5 | 0.00 E+00 | 0.00 E+00 | 0.00 E+00 | 0.00 E+00 | 0.00 E+00 | 0.00 E+00 | 0.00 E+00 |
| | 1 | 0.00 E+00 | 0.00 E+00 | 0.00 E+00 | 0.00 E+00 | 0.00 E+00 | 0.00 E+00 | 0.00 E+00 |
| | 2 | 0.00 E+00 | 0.00 E+00 | 0.00 E+00 | 0.00 E+00 | 0.00 E+00 | 0.00 E+00 | 0.00 E+00 |
| | 5 | 0.00 E+00 | 0.00 E+00 | 0.00 E+00 | 0.00 E+00 | 0.00 E+00 | 0.00 E+00 | 0.00 E+00 |
| Xin She Yang N2 | 0.1 | 7.89 E-02 | 1 E-01 | 4.20 E-02 | 3.58 E-03 | 1.17 E-02 | 7.32 E-45 | 1.9 E-35 |
| | 0.2 | 3.49 E-05 | 5.61 E-02 | 3.61 E-02 | 2.46 E-03 | 1.05 E-02 | 5.92 E-95 | 1.98 E-02 |
| | 0.5 | 1.48 E-05 | 4.05 E-02 | 2.74 E-02 | 3.11 E-03 | 2.45 E-03 | 0.00 E+00 | 2.76 E-110 |
| | 1 | 6.26 E-06 | 1.83 E-02 | 1.71 E-02 | 1.64 E-03 | 3.86 E-04 | 0.00 E+00 | 2.76 E-110 |
| | 2 | 2.1 E-03 | 1.05 E-02 | 1.1 E-03 | 1.05 E-03 | 9.49 E-05 | 2.10 E-239 | 0.00 E+00 |
| | 5 | 1.39 E-06 | 4.24 E-03 | 8.92 E-03 | 1.09 E-03 | 1.39 E-05 | 0.00 E+00 | 2.76 E-110 |
| Rosenbrock | 0.1 | 0.00 E+00 | 0.00 E+00 | 0.00 E+00 | 0.00 E+00 | 0.00 E+00 | 0.00 E+00 | 0.00 E+00 |
| | 0.2 | 0.00 | 0.00 | 0.00 | 0.00 | 0.00 | 0.00 | 0.00 |
| | | E+00 | E+00 | E+00 | E+00 | E+00 | E+00 | E+00 |
| | 0.5 | 0.00 E+00 | 0.00 E+00 | 0.00 E+00 | 0.00 E+00 | 0.00 E+00 | 0.00 E+00 | 0.00 E+00 |
| | 1 | 0.00 E+00 | 0.00 E+00 | 0.00 E+00 | 0.00 E+00 | 0.00 E+00 | 0.00 E+00 | 0.00 E+00 |
| | 2 | 0.00 E+00 | 0.00 E+00 | 0.00 E+00 | 0.00 E+00 | 0.00 E+00 | 0.00 E+00 | 0.00 E+00 |
| | 5 | 0.00 E+00 | 0.00 E+00 | 0.00 E+00 | 0.00 E+00 | 0.00 E+00 | 0.00 E+00 | 0.00 E+00 |
| Quadric | 0.1 | 8.95 E-05 | 2.77 E-05 | 5.40 E-05 | 2.61 E-04 | 4.28 E-09 | 1.38 E-121 | 1.22 E-105 |
| | 0.2 | 1.36 E-11 | 2.52 E-05 | 6.17 E-05 | 1.81 E-04 | 1.88 E-09 | 3.48 E-239 | 5.02 E-68 |
| | 0.5 | 6.27 E-13 | 2 E-05 | 6.84 E-05 | 9.82 E-05 | 7.52 E-11 | 0.00 E+00 | 4.52 E-182 |
| | 1 | 1.41 E-13 | 1.56 E-05 | 5.80 E-05 | 5.62 E-05 | 3.43 E-13 | 0.00 E+00 | 0.00 E+00 |
| | 2 | 2.83 E-06 | 1.19 E-05 | 1.36 E-02 | 3.72 E-05 | 3.42 E-15 | 0.00 E+00 | 0.00 E+00 |
| | 5 | 4.88 E-15 | 1.07 E-05 | 5.33 E-05 | 2.41 E-05 | 5.63 E-18 | 0.00 E+00 | 0.00 E+00 |
| Ellipsoid | 0.1 | 1.62 E-09 | 7.76 E-03 | 2.53 E-03 | 6.13 E-03 | 2.47 E-05 | 4.41 E-159 | 1.93 E-35 |
| | 0.2 | 2.18 E-10 | 5.32 E-05 | 2.16 E-03 | 4.78 E-03 | 2.48 E-06 | 0.00 E+00 | 8.96 E-74 |
| | 0.5 | 2.83 E-10 | 5.03 E-05 | 1.39 E-03 | 3.57 E-03 | 5.81 E-08 | 0.00 E+00 | 3.42E-184 |
| | 1 | 4.53 E-12 | 5.76 E-05 | 1.06 E-03 | 2.24 E-03 | 1.66 E-09 | 0.00 E+00 | 0.00 E+00 |
| | 2 | 0.00 E+00 | 4.90 E-05 | 1.32 E-03 | 1.33 E-03 | 1.02 E-03 | 0.00 E+00 | 0.00 E+00 |
| | 5 | 1.82 E-13 | 4.26 E-05 | 1.07 E-03 | 1.05 E-03 | 2.72 E-12 | 0.00 E+00 | 0.00 E+00 |

# APPENDIX G: STATISTICAL SIGNIFICANCE TEST RESULTS FOR CASE STUDY 1

In this appendix, we have statistical significance results for Case Study 1. All of the following statistical result table represents Convor's results for BSO, BPO, SA, TA, PSO, WO, GWO Our test criteria include, the Kruskal-Wallis H test, complemented by subsequent Conover's tests (incorporating the step-down technique with a Bonferroni correction for p-value modification), is employed. A 0.05 p-value benchmark is set for determining statistical relevance. For continuous functions the statistical test results are generated by using the result data mentioned in Table A2, and for non-continuous functions the statistical test results are generated by using the result data mentioned in Table B2

Statistical significance results for precision, continuous rastrigin function, 2 dimensions

| | BSO | BPO | SA | TA | PSO | WO | GWO |
|---|---|---|---|---|---|---|---|
| BSO | 1.000000e+00 | 4.821712e-216 | 1.210691e-144 | 1.300049e-85 | 3.617064e-15 | 2.661349e-40 | 1.318372e-95 |
| BPO | 4.821712e-216 | 1.000000e+00 | 2.594570e-36 | 6.889262e-92 | 6.991502e-173 | 3.808753e-281 | 4.150151e-321 |
| SA | 1.210691e-144 | 2.594570e-36 | 1.000000e+00 | 2.790490e-22 | 1.123175e-96 | 4.188208e-220 | 1.060203e-266 |
| TA | 1.300049e-85 | 6.889262e-92 | 2.790490e-22 | 1.000000e+00 | 2.589188e-40 | 5.886714e-167 | 3.280449e-219 |
| PSO | 3.617064e-15 | 6.991502e-173 | 1.123175e-96 | 2.589188e-40 | 1.000000e+00 | 1.342544e-85 | 1.335306e-143 |
| WO | 2.661349e-40 | 3.808753e-281 | 4.188208e-220 | 5.886714e-167 | 1.342544e-85 | 1.000000e+00 | 1.474726e-21 |
| GWO | 1.318372e-95 | 4.150151e-321 | 1.060203e-266 | 3.280449e-219 | 1.335306e-143 | 1.474726e-21 | 1.000000e+00 |

Statistical significance results for precision, continuous rastrigin function, 3 dimensions

| | BSO | BPO | SA | TA | PSO | WO | GWO |
|---|---|---|---|---|---|---|---|
| BSO | 1.000000e+00 | 5.470294e-45 | 3.474223e-160 | 6.345784e-204 | 7.749534e-15 | 2.758341e-157 | 2.758341e-157 |
| BPO | 5.470294e-45 | 1.000000e+00 | 3.146195e-73 | 9.328679e-121 | 1.912225e-90 | 1.708034e-235 | 1.708034e-235 |
| SA | 3.474223e-160 | 3.146195e-73 | 1.000000e+00 | 7.482689e-15 | 6.688325e-204 | 1.452553e-321 | 1.452553e-321 |
| TA | 6.345784e-204 | 9.328679e-121 | 7.482689e-15 | 1.000000e+00 | 6.644060e-244 | 0.000000e+00 | 0.000000e+00 |
| PSO | 7.749534e-15 | 1.912225e-90 | 6.688325e-204 | 6.644060e-244 | 1.000000e+00 | 2.395287e-110 | 2.395287e-110 |
| WO | 2.758341e-157 | 1.708034e-235 | 1.452553e-321 | 0.000000e+00 | 2.395287e-110 | 1.000000e+00 | 1.000000e+00 |
| GWO | 2.758341e-157 | 1.708034e-235 | 1.452553e-321 | 0.000000e+00 | 2.395287e-110 | 1.000000e+00 | 1.000000e+00 |

Statistical significance results for precision, continuous rastrigin function, 4 dimensions

| | BSO | BPO | SA | TA | PSO | WO | GWO |
|---|---|---|---|---|---|---|---|
| BSO | 1.000000e+00 | 2.882837e-13 | 1.382045e-65 | 2.303034e-43 | 4.017732e-01 | 9.082701e-34 | 9.082701e-34 |
| BPO | 2.882837e-13 | 1.000000e+00 | 1.218500e-26 | 2.912655e-11 | 1.328893e-21 | 1.553084e-74 | 1.553084e-74 |
| SA | 1.382045e-65 | 1.218500e-26 | 1.000000e+00 | 5.301299e-04 | 1.264124e-78. | 1.254802e-139 | 1.254802e-139 |
| TA | 2.303034e-43 | 2.912655e-11 | 5.301299e-04 | 1.000000e+00 | 2.105715e-55 | 1.308995e-115 | 1.308995e-115 |
| PSO | 4.017732e-01 | 1.328893e-21 | 1.264124e-78 | 2.105715e-55 | 1.000000e+00 | 1.245930e-23 | 1.245930e-23 |
| WO | 9.082701e-34 | 1.553084e-74 | 1.254802e-139 | 1.308995e-115 | 1.245930e-23 | 1.000000e+00 | 1.000000e+00 |
| GWO | 9.082701e-34 | 1.553084e-74 | 1.254802e-139 | 1.308995e-115 | 1.245930e-23 | 1.000000e+00 | 1.000000e+00 |

Statistical significance results for precision, continuous rastrigin function, 5 dimensions

| | BSO | BPO | SA | TA | PSO | WO | GWO |
|---|---|---|---|---|---|---|---|
| BSO | 1.000000e+00 | 7.664999e-10 | 2.699032e-33 | 5.463862e-38 | 1.000000e+00 | 1.689864e-29 | 1.689864e-29 |
| BPO | 7.664999e-10 | 1.000000e+00 | 1.518127e-08 | 2.258362e-11 | 1.724105e-08 | 5.551255e-63 | 5.551255e-63 |
| SA | 2.699032e-33 | 1.518127e-08 | 1.000000e+00 | 1.000000e+00 | 4.760436e-31 | 4.554989e-98 | 4.554989e-98 |
| TA | 5.463862e-38 | 2.258362e-11 | 1.000000e+00 | 1.000000e+00 | 1.220232e-35 | 7.890047e-104 | 7.890047e-104 |
| PSO | 1.000000e+00 | 1.724105e-08 | 4.760436e-31 | 1.220232e-35 | 1.000000e+00 | 1.046898e-31 | 1.046898e-31 |
| WO | 1.689864e-29 | 5.551255e-63 | 4.554989e-98 | 7.890047e-104 | 1.046898e-31 | 1.000000e+00 | 1.000000e+00 |
| GWO | 1.689864e-29 | 5.551255e-63 | 4.554989e-98 | 7.890047e-104 | 1.046898e-31 | 1.000000e+00 | 1.000000e+00 |

Statistical significance results for precision, continuous rastrigin function, 6 dimensions

| | BSO | BPO | SA | TA | PSO | WO | GWO |
|---|---|---|---|---|---|---|---|
| BSO | 1.000000e+00 | 6.081535e-01 | 3.052523e-33 | 5.240849e-23 | 6.172278e-87 | 1.885108e-97 | 1.885108e-97 |
| BPO | 6.081535e-01 | 1.000000e+00 | 1.183059e-43 | 2.472361e-32 | 1.340461e-74 | 5.556951e-85 | 5.556951e-85 |
| SA | 3.052523e-33 | 1.183059e-43 | 1.000000e+00 | 3.354443e-01 | 8.874723e-160 | 1.010689e-169 | 1.010689e-169 |
| TA | 5.240849e-23 | 2.472361e-32 | 3.354443e-01 | 1.000000e+00 | 5.869774e-147 | 4.020462e-157 | 4.020462e-157 |
| PSO | 6.172278e-87 | 1.340461e-74 | 8.874723e-160 | 5.869774e-147 | 1.000000e+00 | 1.000000e+00 | 1.000000e+00 |
| WO | 1.885108e-97 | 5.556951e-85 | 1.010689e-169 | 4.020462e-157 | 1.000000e+00 | 1.000000e+00 | 1.000000e+00 |
| GWO | 1.885108e-97 | 5.556951e-85 | 1.010689e-169 | 4.020462e-157 | 1.000000e+00 | 1.000000e+00 | 1.000000e+00 |

Statistical significance results for precision, continuous ackley function, 2 dimensions

| | BSO | BPO | SA | TA | PSO | WO | GWO |
|---|---|---|---|---|---|---|---|
| BSO | 1.000000e+00 | 7.405308e-56 | 3.145639e-60 | 1.887053e-100 | 2.045691e-51 | 2.789691e-09 | 1.560581e-45 |
| BPO | 7.405308e-56 | 1.000000e+00 | 1.000000e+00 | 1.147734e-13 | 1.000000e+00 | 4.804170e-92 | 3.438692e-143 |
| SA | 3.145639e-60 | 1.000000e+00 | 1.000000e+00 | 3.920738e-11 | 1.000000e+00 | 1.006290e-96 | 9.544298e-148 |
| TA | 1.887053e-100 | 1.147734e-13 | 3.920738e-11 | 1.000000e+00 | 1.623633e-16 | 1.449144e-137 | 2.439552e-186 |
| PSO | 2.045691e-51 | 1.000000e+00 | 1.000000e+00 | 1.623633e-16 | 1.000000e+00 | 3.213821e-87 | 1.891654e-138 |
| WO | 2.789691e-09 | 4.804170e-92 | 1.006290e-96 | 1.449144e-137 | 3.213821e-87 | 1.000000e+00 | 4.529828e-17 |
| GWO | 1.560581e-45 | 3.438692e-143 | 9.544298e-148 | 2.439552e-186 | 1.891654e-138 | 4.529828e-17 | 1.000000e+00 |

Statistical significance results for precision, continuous ackley function, 3 dimensions

| | BSO | BPO | SA | TA | PSO | WO | GWO |
|---|---|---|---|---|---|---|---|
| BSO | 1.000000e+00 | 9.000347e-123 | 1.145382e-192 | 2.194869e-267 | 7.936897e-06 | 4.581432e-126 | 4.581432e-126 |
| BPO | 9.000347e-123 | 1.000000e+00 | 1.092689e-32 | 1.015647e-118 | 1.845953e-93 | 6.802450e-273 | 6.802450e-273 |
| SA | 1.145382e-192 | 1.092689e-32 | 1.000000e+00 | 2.752352e-47 | 1.201654e-165 | 0.000000e+00 | 0.000000e+00 |
| TA | 2.194869e-267 | 1.015647e-118 | 2.752352e-47 | 1.000000e+00 | 1.099008e-244 | 0.000000e+00 | 0.000000e+00 |
| PSO | 7.936897e-06 | 1.845953e-93 | 1.201654e-165 | 1.099008e-244 | 1.000000e+00 | 9.669904e-155 | 9.669904e-155 |
| WO | 4.581432e-126 | 6.802450e-273 | 0.000000e+00 | 0.000000e+00 | 9.669904e-155 | 1.000000e+00 | 1.000000e+00 |
| GWO | 4.581432e-126 | 6.802450e-273 | 0.000000e+00 | 0.000000e+00 | 9.669904e-155 | 1.000000e+00 | 1.000000e+00 |

Statistical significance results for precision, continuous ackley function, 4 dimensions

| | BSO | BPO | SA | TA | PSO | WO | GWO |
|---|---|---|---|---|---|---|---|
| BSO | 1.000000e+00 | 1.491931e-39 | 1.243025e-175 | 3.814879e-231 | 9.502940e-173 | 1.198448e-181 | 1.198448e-181 |
| BPO | 1.491931e-39 | 1.000000e+00 | 9.729372e-96 | 2.943510e-158 | 1.342563e-243 | 3.272514e-251 | 3.272514e-251 |
| SA | 1.243025e-175 | 9.729372e-96 | 1.000000e+00 | 4.591159e-25 | 0.000000e+00 | 0.000000e+00 | 0.000000e+00 |
| TA | 3.814879e-231 | 2.943510e-158 | 4.591159e-25 | 1.000000e+00 | 0.000000e+00 | 0.000000e+00 | 0.000000e+00 |
| PSO | 9.502940e-173 | 1.342563e-243 | 0.000000e+00 | 0.000000e+00 | 1.000000e+00 | 1.000000e+00 | 1.000000e+00 |
| WO | 1.198448e-181 | 3.272514e-251 | 0.000000e+00 | 0.000000e+00 | 1.000000e+00 | 1.000000e+00 | 1.000000e+00 |
| GWO | 1.198448e-181 | 3.272514e-251 | 0.000000e+00 | 0.000000e+00 | 1.000000e+00 | 1.000000e+00 | 1.000000e+00 |

Statistical significance results for precision, continuous ackley function, 5 dimensions

| | BSO | BPO | SA | TA | PSO | WO | GWO |
|---|---|---|---|---|---|---|---|
| BSO | 1.000000e+00 | 8.679055e-40 | 1.801514e-138 | 1.026519e-178 | 9.828040e-119 | 4.751198e-147 | 5.171968e-90 |
| BPO | 8.679055e-40 | 1.000000e+00 | 1.447160e-57 | 8.606622e-99 | 8.632792e-197 | 1.108844e-221 | 1.523321e-170 |
| SA | 1.801514e-138 | 1.447160e-57 | 1.000000e+00 | 9.397110e-12 | 3.783271e-279 | 1.695783e-299 | 1.287766e-257 |
| TA | 1.026519e-178 | 8.606622e-99 | 9.397110e-12 | 1.000000e+00 | 1.162789e-308 | 0.000000e+00 | 7.755021e-289 |
| PSO | 9.828040e-119 | 8.632792e-197 | 3.783271e-279 | 1.162789e-308 | 1.000000e+00 | 1.347952e-05 | 1.347952e-05 |
| WO | 4.751198e-147 | 1.108844e-221 | 1.695783e-299 | 0.000000e+00 | 1.347952e-05 | 1.000000e+00 | 5.944911e-21 |
| GWO | 5.171968e-90 | 1.523321e-170 | 1.287766e-257 | 7.755021e-289 | 1.347952e-05 | 5.944911e-21 | 1.000000e+00 |

Statistical significance results for precision, continuous ackley function, 6 dimensions

| | BSO | BPO | SA | TA | PSO | WO | GWO |
|---|---|---|---|---|---|---|---|
| BSO | 1.000000e+00 | 2.141258e-28 | 1.324516e-48 | 1.955974e-88 | 3.955187e-40 | 1.497530e-102 | 5.921110e-93 |
| BPO | 2.141258e-28 | 1.000000e+00 | 4.941659e-04 | 5.557105e-27 | 6.285983e-105 | 5.532669e-169 | 6.627471e-160 |
| SA | 1.324516e-48 | 4.941659e-04 | 1.000000e+00 | 1.870231e-11 | 3.339964e-129 | 2.259779e-191 | 1.171987e-182 |
| TA | 1.955974e-88 | 5.557105e-27 | 1.870231e-11 | 1.000000e+00 | 1.949691e-169 | 2.019808e-227 | 2.285578e-219 |
| PSO | 3.955187e-40 | 6.285983e-105 | 3.339964e-129 | 1.949691e-169 | 1.000000e+00 | 1.263296e-26 | 6.482362e-20 |
| WO | 1.497530e-102 | 5.532669e-169 | 2.259779e-191 | 2.019808e-227 | 1.263296e-26 | 1.000000e+00 | 1.000000e+00 |
| GWO | 5.921110e-93 | 6.627471e-160 | 1.171987e-182 | 2.285578e-219 | 6.482362e-20 | 1.000000e+00 | 1.000000e+00 |

Statistical significance results for precision, continuous sphere function, 2 dimensions

| | BSO | BPO | SA | TA | PSO | WO | GWO |
|---|---|---|---|---|---|---|---|
| BSO | 1.000000e+00 | 6.816790e-253 | 1.972846e-83 | 1.729281e-190 | 2.151007e-71 | 1.690767e-08 | 1.451572e-8 |
| BPO | 6.816790e-253 | 1.000000e+00 | 5.345587e-139 | 2.443014e-33 | 6.112815e-151 | 9.797092e-280 | 0.000000e+00 |
| SA | 1.972846e-83 | 5.345587e-139 | 1.000000e+00 | 1.741441e-65 | 6.760225e-01 | 5.310235e-119 | 3.709336e-212 |
| TA | 1.729281e-190 | 2.443014e-33 | 1.741441e-65 | 1.000000e+00 | 2.187707e-77 | 1.448024e-221 | 1.445689e-297 |
| PSO | 2.151007e-71 | 6.112815e-151 | 6.760225e-01 | 2.187707e-77 | 1.000000e+00 | 9.814894e-107 | 1.827348e-201 |
| WO | 1.690767e-08 | 9.797092e-280 | 5.310235e-119 | 1.448024e-221 | 9.814894e-107 | 1.000000e+00 | 4.215642e-55 |
| GWO | 1.451572e-89 | 0.000000e+00 | 3.709336e-212 | 1.445689e-297 | 1.827348e-201 | 4.215642e-55 | 1.000000e+00 |

Statistical significance results for precision, continuous sphere function, 3 dimensions

| | BSO | BPO | SA | TA | PSO | WO | GWO |
|---|---|---|---|---|---|---|---|
| BSO | 1.000000e+00 | 3.986894e-122 | 2.124757e-196 | 8.997347e-283 | 2.844524e-01 | 6.847002e-146 | 6.847002e-146 |
| BPO | 3.986894e-122 | 1.000000e+00 | 1.390489e-36 | 4.925749e-140 | 5.440418e-108 | 4.464201e-287 | 4.464201e-287 |
| SA | 2.124757e-196 | 1.390489e-36 | 1.000000e+00 | 1.114317e-62 | 1.264525e-183 | 0.000000e+00 | 0.000000e+00 |
| TA | 8.997347e-283 | 4.925749e-140 | 1.114317e-62 | 1.000000e+00 | 2.020741e-272 | 0.000000e+00 | 0.000000e+00 |
| PSO | 2.844524e-01 | 5.440418e-108 | 1.264525e-183 | 2.020741e-272 | 1.000000e+00 | 1.606036e-159 | 1.606036e-159 |
| WO | 6.847002e-146 | 4.464201e-287 | 0.000000e+00 | 0.000000e+00 | 1.606036e-159 | 1.000000e+00 | 1.000000e+00 |
| GWO | 6.847002e-146 | 4.464201e-287 | 0.000000e+00 | 0.000000e+00 | 1.606036e-159 | 1.000000e+00 | 1.000000e+00 |

Statistical significance results for precision, continuous sphere function, 4 dimensions

| | BSO | BPO | SA | TA | PSO | WO | GWO |
|---|---|---|---|---|---|---|---|
| BSO | 1.000000e+00 | 4.039657e-180 | 1.912248e-112 | 1.919551e-305 | 2.195316e-206 | 2.195316e-206 | 2.195316e-206 |
| BPO | 4.039657e-180 | 1.000000e+00 | 5.420922e-30 | 1.912248e-112 | 0.000000e+00 | 0.000000e+00 | 0.000000e+00 |
| SA | 1.912248e-112 | 5.420922e-30 | 1.000000e+00 | 4.039657e-180 | 0.000000e+00 | 0.000000e+00 | 0.000000e+00 |
| TA | 1.919551e-305 | 1.912248e-112 | 4.039657e-180 | 1.000000e+00 | 0.000000e+00 | 0.000000e+00 | 0.000000e+00 |
| PSO | 2.195316e-206 | 0.000000e+00 | 0.000000e+00 | 0.000000e+00 | 1.000000e+00 | 1.000000e+00 | 1.000000e+00 |
| WO | 2.195316e-206 | 0.000000e+00 | 0.000000e+00 | 0.000000e+00 | 1.000000e+00 | 1.000000e+00 | 1.000000e+00 |
| GWO | 2.195316e-206 | 0.000000e+00 | 0.000000e+00 | 0.000000e+00 | 1.000000e+00 | 1.000000e+00 | 1.000000e+00 |

Statistical significance results for precision, continuous sphere function, 5 dimensions

| | BSO | BPO | SA | TA | PSO | WO | GWO |
|---|---|---|---|---|---|---|---|
| BSO | 1.000000e+00 | 1.530637e-119 | 2.079458e-235 | 0.000000e+00 | 1.162378e-245 | 1.162378e-245 | 1.162378e-245 |
| BPO | 1.530637e-119 | 1.000000e+00 | 3.290505e-81 | 3.102136e-235 | 0.000000e+00 | 0.000000e+00 | 0.000000e+00 |
| SA | 2.079458e-235 | 3.290505e-81 | 1.000000e+00 | 2.495726e-119 | 0.000000e+00 | 0.000000e+00 | 0.000000e+00 |
| TA | 0.000000e+00 | 3.102136e-235 | 2.495726e-119 | 1.000000e+00 | 0.000000e+00 | 0.000000e+00 | 0.000000e+00 |
| PSO | 1.162378e-245 | 0.000000e+00 | 0.000000e+00 | 0.000000e+00 | 1.000000e+00 | 1.000000e+00 | 1.000000e+00 |
| WO | 1.162378e-245 | 0.000000e+00 | 0.000000e+00 | 0.000000e+00 | 1.000000e+00 | 1.000000e+00 | 1.000000e+00 |
| GWO | 1.162378e-245 | 0.000000e+00 | 0.000000e+00 | 0.000000e+00 | 1.000000e+00 | 1.000000e+00 | 1.000000e+00 |

Statistical significance results for precision, continuous sphere function, 6 dimensions

| | BSO | BPO | SA | TA | PSO | WO | GWO |
|---|---|---|---|---|---|---|---|
| BSO | 1.000000e+00 | 1.229972e-138 | 1.501043e-293 | 0.000000e+00 | 1.414467e-293 | 1.414467e-293 | 1.414467e-293 |
| BPO | 1.229972e-138 | 1.000000e+00 | 1.578709e-138 | 1.501043e-293 | 0.000000e+00 | 0.000000e+00 | 0.000000e+00 |
| SA | 1.501043e-293 | 1.578709e-138 | 1.000000e+00 | 1.229972e-138 | 0.000000e+00 | 0.000000e+00 | 0.000000e+00 |
| TA | 0.000000e+00 | 1.501043e-293 | 1.229972e-138 | 1.000000e+00 | 0.000000e+00 | 0.000000e+00 | 0.000000e+00 |
| PSO | 1.414467e-293 | 0.000000e+00 | 0.000000e+00 | 0.000000e+00 | 1.000000e+00 | 1.000000e+00 | 1.000000e+00 |
| WO | 1.414467e-293 | 0.000000e+00 | 0.000000e+00 | 0.000000e+00 | 1.000000e+00 | 1.000000e+00 | 1.000000e+00 |
| GWO | 1.414467e-293 | 0.000000e+00 | 0.000000e+00 | 0.000000e+00 | 1.000000e+00 | 1.000000e+00 | 1.000000e+00 |

Statistical significance results for precision, continuous schwefel function, 2 dimensions

| | BSO | BPO | SA | TA | PSO | WO | GWO |
|---|---|---|---|---|---|---|---|
| BSO | 1.000000e+00 | 1.382857e-01 | 6.311482e-01 | 3.233031e-12 | 1.919199e-04 | 3.557299e-46 | 1.353640e-14 |
| BPO | 1.382857e-01 | 1.000000e+00 | 2.517620e-05 | 9.142895e-22 | 1.000000e+00 | 4.557530e-33 | 8.432851e-07 |
| SA | 6.311482e-01 | 2.517620e-05 | 1.000000e+00 | 2.378681e-06 | 1.299444e-09 | 1.849433e-57 | 1.606466e-22 |
| TA | 3.233031e-12 | 9.142895e-22 | 2.378681e-06 | 1.000000e+00 | 5.687632e-29 | 3.822585e-87 | 5.409131e-47 |
| PSO | 1.919199e-04 | 1.000000e+00 | 1.299444e-09 | 5.687632e-29 | 1.000000e+00 | 1.794086e-25 | 3.204167e-03 |
| WO | 3.557299e-46 | 4.557530e-33 | 1.849433e-57 | 3.822585e-87 | 1.794086e-25 | 1.000000e+00 | 1.006951e-11 |
| GWO | 1.353640e-14 | 8.432851e-07 | 1.606466e-22 | 5.409131e-47 | 3.204167e-03 | 1.006951e-11 | 1.000000e+00 |

Statistical significance results for precision, continuous schwefel function, 3 dimensions

| | BSO | BPO | SA | TA | PSO | WO | GWO |
|---|---|---|---|---|---|---|---|
| BSO | 1.000000e+00 | 1.663424e-86 | 2.325572e-158 | 6.299934e-182 | 9.022412e-56 | 2.023093e-09 | 4.304895e-45 |
| BPO | 1.663424e-86 | 1.000000e+00 | 3.830887e-32 | 9.490880e-54 | 7.829141e-07 | 1.679262e-50 | 1.552442e-12 |
| SA | 2.325572e-158 | 3.830887e-32 | 1.000000e+00 | 2.616091e-04 | 6.205903e-60 | 1.307567e-121 | 2.977009e-71 |
| TA | 6.299934e-182 | 9.490880e-54 | 2.616091e-04 | 1.000000e+00 | 2.503660e-84 | 2.167677e-146 | 4.037509e-96 |
| PSO | 9.022412e-56 | 7.829141e-07 | 6.205903e-60 | 2.503660e-84 | 1.000000e+00 | 1.888572e-24 | 8.129948e-01 |
| WO | 2.023093e-09 | 1.679262e-50 | 1.307567e-121 | 2.167677e-146 | 1.888572e-24 | 1.000000e+00 | 1.395716e-16 |
| GWO | 4.304895e-45 | 1.552442e-12 | 2.977009e-71 | 4.037509e-96 | 8.129948e-01 | 1.395716e-16 | 1.000000e+00 |

Statistical significance results for precision, continuous schwefel function, 4 dimensions

| | BSO | BPO | SA | TA | PSO | WO | GWO |
|---|---|---|---|---|---|---|---|
| BSO | 1.000000e+00 | 3.750523e-09 | 7.078652e-29 | 4.536025e-54 | 1.446222e-18 | 4.174770e-26 | 4.886926e-01 |
| BPO | 3.750523e-09 | 1.000000e+00 | 1.084850e-06 | 1.537161e-23 | 7.153538e-02 | 2.246142e-57 | 3.390434e-16 |
| SA | 7.078652e-29 | 1.084850e-06 | 1.000000e+00 | 5.493120e-06 | 2.193750e-01 | 6.961274e-88 | 2.648669e-39 |
| TA | 4.536025e-54 | 1.537161e-23 | 5.493120e-06 | 1.000000e+00 | 6.022570e-13 | 1.328016e-117 | 2.754270e-66 |
| PSO | 1.446222e-18 | 7.153538e-02 | 2.193750e-01 | 6.022570e-13 | 1.000000e+00 | 2.048236e-73 | 1.301667e-27 |
| WO | 4.174770e-26 | 2.246142e-57 | 6.961274e-88 | 1.328016e-117 | 2.048236e-73 | 1.000000e+00 | 2.800113e-17 |
| GWO | 4.886926e-01 | 3.390434e-16 | 2.648669e-39 | 2.754270e-66 | 1.301667e-27 | 2.800113e-17 | 1.000000e+00 |

Statistical significance results for precision, continuous schwefel function, 5 dimensions

| | BSO | BPO | SA | TA | PSO | WO | GWO |
|---|---|---|---|---|---|---|---|
| BSO | 1.000000e+00 | 2.485360e-14 | 2.212786e-67 | 1.174365e-99 | 1.081345e-46 | 5.601646e-33 | 1.000000e+0 |
| BPO | 2.485360e-14 | 1.000000e+00 | 1.230705e-26 | 4.734120e-54 | 2.797455e-12 | 2.102128e-75 | 1.308686e-12 |
| SA | 2.212786e-67 | 1.230705e-26 | 1.000000e+00 | 3.311732e-07 | 2.181114e-03 | 1.714782e-140 | 1.923647e-64 |
| TA | 1.174365e-99 | 4.734120e-54 | 3.311732e-07 | 1.000000e+00 | 2.414136e-19 | 8.217549e-172 | 1.383241e-96 |
| PSO | 1.081345e-46 | 2.797455e-12 | 2.181114e-03 | 2.414136e-19 | 1.000000e+00 | 2.046571e-118 | 5.344008e-44 |
| WO | 5.601646e-33 | 2.102128e-75 | 1.714782e-140 | 8.217549e-172 | 2.046571e-118 | 1.000000e+00 | 1.894353e-35 |
| GWO | 1.000000e+00 | 1.308686e-12 | 1.923647e-64 | 1.383241e-96 | 5.344008e-44 | 1.894353e-35 | 1.000000e+00 |

Statistical significance results for precision, continuous schwefel function, 6 dimensions

| | BSO | BPO | SA | TA | PSO | WO | GWO |
|---|---|---|---|---|---|---|---|
| BSO | 1.000000e+00 | 1.658661e-35 | 1.227284e-93 | 1.710895e-108 | 4.619853e-89 | 8.904805e-02 | 1.696293e-20 |
| BPO | 1.658661e-35 | 1.000000e+00 | 3.374550e-24 | 1.452665e-35 | 5.096039e-21 | 1.323725e-49 | 9.526024e-03 |
| SA | 1.227284e-93 | 3.374550e-24 | 1.000000e+00 | 2.017071e-01 | 1.000000e+00 | 4.736053e-110 | 5.764421e-40 |
| TA | 1.710895e-108 | 1.452665e-35 | 2.017071e-01 | 1.000000e+00 | 1.497763e-02 | 7.296107e-125 | 4.625618e-53 |
| PSO | 4.619853e-89 | 5.096039e-21 | 1.000000e+00 | 1.497763e-02 | 1.000000e+00 | 1.861232e-105 | 3.828758e-36 |
| WO | 8.904805e-02 | 1.323725e-49 | 4.736053e-110 | 7.296107e-125 | 1.861232e-105 | 1.000000e+00 | 1.677258e-32 |
| GWO | 1.696293e-20 | 9.526024e-03 | 5.764421e-40 | 4.625618e-53 | 3.828758e-36 | 1.677258e-32 | 1.000000e+00 |

Statistical significance results for precision, continuous easom function, 3 dimensions

| | BSO | BPO | SA | TA | PSO | WO | GWO |
|---|---|---|---|---|---|---|---|
| BSO | 1.000000e+00 | 3.959581e-154 | 9.309709e-218 | 1.299434e-259 | 2.252239e-53 | 3.303154e-112 | 5.703100e-08 |
| BPO | 3.959581e-154 | 1.000000e+00 | 4.142826e-30 | 1.231059e-76 | 3.084958e-58 | 5.871590e-12 | 2.501526e-120 |
| SA | 9.309709e-218 | 4.142826e-30 | 1.000000e+00 | 2.780272e-17 | 2.223319e-127 | 1.074921e-67 | 1.787693e-187 |
| TA | 1.299434e-259 | 1.231059e-76 | 2.780272e-17 | 1.000000e+00 | 3.489921e-177 | 4.516494e-119 | 2.999498e-232 |
| PSO | 2.252239e-53 | 3.084958e-58 | 2.223319e-127 | 3.489921e-177 | 1.000000e+00 | 8.581343e-23 | 7.961914e-25 |
| WO | 3.303154e-112 | 5.871590e-12 | 1.074921e-67 | 4.516494e-119 | 8.581343e-23 | 1.000000e+00 | 7.133216e-78 |
| GWO | 5.703100e-08 | 2.501526e-120 | 1.787693e-187 | 2.999498e-232 | 7.961914e-25 | 7.133216e-78 | 1.000000e+00 |

Statistical significance results for precision, continuous shubert function, 3 dimensions

| | BSO | BPO | SA | TA | PSO | WO | GWO |
|---|---|---|---|---|---|---|---|
| BSO | 1.000000e+00 | 1.357421e-184 | 6.500143e-76 | 4.589519e-118 | 1.000000e+00 | 6.184872e-71 | 1.000000e+00 |
| BPO | 1.357421e-184 | 1.000000e+00 | 1.421105e-66 | 2.015756e-29 | 1.609255e-179 | 1.599626e-279 | 2.617335e-183 |
| SA | 6.500143e-76 | 1.421105e-66 | 1.000000e+00 | 8.190468e-12 | 1.593971e-70 | 1.053935e-188 | 1.543014e-74 |
| TA | 4.589519e-118 | 2.015756e-29 | 8.190468e-12 | 1.000000e+00 | 1.547707e-112 | 1.589628e-225 | 1.163323e-116 |
| PSO | 1.000000e+00 | 1.609255e-179 | 1.593971e-70 | 1.547707e-112 | 1.000000e+00 | 2.496604e-76 | 1.000000e+00 |
| WO | 6.184872e-71 | 1.599626e-279 | 1.053935e-188 | 1.589628e-225 | 2.496604e-76 | 1.000000e+00 | 2.664968e-72 |
| GWO | 1.000000e+00 | 2.617335e-183 | 1.543014e-74 | 1.163323e-116 | 1.000000e+00 | 2.664968e-72 | 1.000000e+00 |

Statistical significance results for precision, continuous langermann function, 3 dimensions

| | BSO | BPO | SA | TA | PSO | WO | GWO |
|---|---|---|---|---|---|---|---|
| BSO | 1.000000e+00 | 2.506349e-26 | 4.092122e-50 | 2.292203e-89 | 1.347398e-14 | 3.098083e-35 | 1.198423e-08 |
| BPO | 2.506349e-26 | 1.000000e+00 | 1.264122e-05 | 8.777738e-30 | 4.135453e-02 | 9.389258e-01 | 9.331796e-06 |
| SA | 4.092122e-50 | 1.264122e-05 | 1.000000e+00 | 4.566371e-11 | 3.729818e-14 | 5.386323e-02 | 2.782461e-21 |
| TA | 2.292203e-89 | 8.777738e-30 | 4.566371e-11 | 1.000000e+00 | 2.108904e-44 | 1.883383e-21 | 1.280317e-54 |
| PSO | 1.347398e-14 | 4.135453e-02 | 3.729818e-14 | 2.108904e-44 | 1.000000e+00 | 8.420968e-06 | 9.846373e-01 |
| WO | 3.098083e-35 | 9.389258e-01 | 5.386323e-02 | 1.883383e-21 | 8.420968e-06 | 1.000000e+00 | 6.139247e-11 |
| GWO | 1.198423e-08 | 9.331796e-06 | 2.782461e-21 | 1.280317e-54 | 9.846373e-01 | 6.139247e-11 | 1.000000e+00 |

Statistical significance results for precision, continuous dropwave function, 3 dimensions

| | BSO | BPO | SA | TA | PSO | WO | GWO |
|---|---|---|---|---|---|---|---|
| BSO | 1.000000e+00 | 1.047639e-170 | 7.728643e-271 | 0.000000e+00 | 1.509750e-63 | 1.187424e-120 | 1.047378e-126 |
| BPO | 1.047639e-170 | 1.000000e+00 | 4.485948e-74 | 1.228108e-178 | 1.899214e-64 | 3.861959e-304 | 2.954692e-308 |
| SA | 7.728643e-271 | 4.485948e-74 | 1.000000e+00 | 3.516504e-62 | 7.789546e-181 | 0.000000e+00 | 0.000000e+00 |
| TA | 0.000000e+00 | 1.228108e-178 | 3.516504e-62 | 1.000000e+00 | 9.179818e-270 | 0.000000e+00 | 0.000000e+00 |
| PSO | 1.509750e-63 | 1.899214e-64 | 7.789546e-181 | 9.179818e-270 | 1.000000e+00 | 3.006530e-221 | 2.356071e-226 |
| WO | 1.187424e-120 | 3.861959e-304 | 0.000000e+00 | 0.000000e+00 | 3.006530e-221 | 1.000000e+00 | 1.000000e+00 |
| GWO | 1.047378e-126 | 2.954692e-308 | 0.000000e+00 | 0.000000e+00 | 2.356071e-226 | 1.000000e+00 | 1.000000e+00 |

Statistical significance results for precision, continuous holdertable function, 3 dimensions

| | BSO | BPO | SA | TA | PSO | WO | GWO |
|---|---|---|---|---|---|---|---|
| BSO | 1.000000e+00 | 2.589389e-28 | 7.379051e-66 | 2.926734e-100 | 2.081215e-10 | 1.608023e-21 | 1.844482e-11 |
| BPO | 2.589389e-28 | 1.000000e+00 | 3.404390e-12 | 2.481276e-36 | 2.204467e-05 | 1.028619e-80 | 1.229136e-04 |
| SA | 7.379051e-66 | 3.404390e-12 | 1.000000e+00 | 3.614243e-08 | 6.040154e-31 | 1.216770e-123 | 2.309547e-29 |
| TA | 2.926734e-100 | 2.481276e-36 | 3.614243e-08 | 1.000000e+00 | 2.166186e-61 | 9.627757e-158 | 1.854310e-59 |
| PSO | 2.081215e-10 | 2.204467e-05 | 6.040154e-31 | 2.166186e-61 | 1.000000e+00 | 9.432652e-54 | 1.000000e+00 |
| WO | 1.608023e-21 | 1.028619e-80 | 1.216770e-123 | 9.627757e-158 | 9.432652e-54 | 1.000000e+00 | 1.213926e-55 |
| GWO | 1.844482e-11 | 1.229136e-04 | 2.309547e-29 | 1.854310e-59 | 1.000000e+00 | 1.213926e-55 | 1.000000e+00 |

Statistical significance results for precision, non-continuous rastrigin function, 2 dimensions

| | BSO | BPO | SA | TA | PSO | WO | GWO |
|---|---|---|---|---|---|---|---|
| BSO | 1.000000e+00 | 3.480724e-124 | 2.007292e-29 | 2.779420e-48 | 2.258320e-48 | 1.864444e-16 | 3.249734e-28 |
| BPO | 3.480724e-124 | 1.000000e+00 | 4.487464e-56 | 3.264332e-36 | 3.949806e-36 | 5.990839e-173 | 2.403129e-188 |
| SA | 2.007292e-29 | 4.487464e-56 | 1.000000e+00 | 1.753640e-03 | 1.631123e-03 | 1.886664e-74 | 2.967361e-91 |
| TA | 2.779420e-48 | 3.264332e-36 | 1.753640e-03 | 1.000000e+00 | 1.000000e+00 | 7.193901e-97 | 7.905873e-114 |
| PSO | 2.258320e-48 | 3.949806e-36 | 1.631123e-03 | 1.000000e+00 | 1.000000e+00 | 5.701410e-97 | 6.269894e-114 |
| WO | 1.864444e-16 | 5.990839e-173 | 1.886664e-74 | 7.193901e-97 | 5.701410e-97 | 1.000000e+00 | 5.384930e-02 |
| GWO | 3.249734e-28 | 2.403129e-188 | 2.967361e-91 | 7.905873e-114 | 6.269894e-114 | 5.384930e-02 | 1.000000e+00 |

Statistical significance results for precision, non-continuous rastrigin function, 3 dimensions

| | BSO | BPO | SA | TA | PSO | WO | GWO |
|---|---|---|---|---|---|---|---|
| BSO | 1.000000e+00 | 2.199978e-126 | 4.473773e-104 | 5.085072e-154 | 1.608350e-36 | 1.375309e-46 | 1.375309e-46 |
| BPO | 2.199978e-126 | 1.000000e+00 | 2.138759e-03 | 2.039244e-05 | 6.223488e-50 | 1.956577e-210 | 1.956577e-210 |
| SA | 4.473773e-104 | 2.138759e-03 | 1.000000e+00 | 1.510533e-16 | 4.650412e-31 | 1.059461e-190 | 1.059461e-190 |
| TA | 5.085072e-154 | 2.039244e-05 | 1.510533e-16 | 1.000000e+00 | 1.137863e-76 | 3.375048e-234 | 3.375048e-234 |
| PSO | 1.608350e-36 | 6.223488e-50 | 4.650412e-31 | 1.137863e-76 | 1.000000e+00 | 1.164215e-122 | 1.164215e-122 |
| WO | 1.375309e-46 | 1.956577e-210 | 1.059461e-190 | 3.375048e-234 | 1.164215e-122 | 1.000000e+00 | 1.000000e+00 |
| GWO | 1.375309e-46 | 1.956577e-210 | 1.059461e-190 | 3.375048e-234 | 1.164215e-122 | 1.000000e+00 | 1.000000e+00 |

Statistical significance results for precision, non-continuous rastrigin function, 4 dimensions

| | BSO | BPO | SA | TA | PSO | WO | GWO |
|---|---|---|---|---|---|---|---|
| BSO | 1.000000e+00 | 4.473557e-08 | 2.960181e-56 | 3.689253e-68 | 1.790606e-12 | 6.117226e-47 | 6.117226e-47 |
| BPO | 4.473557e-08 | 1.000000e+00 | 6.192423e-27 | 8.494384e-37 | 1.378155e-36 | 1.049965e-79 | 1.049965e-79 |
| SA | 2.960181e-56 | 6.192423e-27 | 1.000000e+00 | 5.897684e-01 | 1.004792e-98 | 3.664247e-145 | 3.664247e-145 |
| TA | 3.689253e-68 | 8.494384e-37 | 5.897684e-01 | 1.000000e+00 | 2.555343e-111 | 2.637523e-157 | 2.637523e-157 |
| PSO | 1.790606e-12 | 1.378155e-36 | 1.004792e-98 | 2.555343e-111 | 1.000000e+00 | 2.758506e-14 | 2.758506e-14 |
| WO | 6.117226e-47 | 1.049965e-79 | 3.664247e-145 | 2.637523e-157 | 2.758506e-14 | 1.000000e+00 | 1.000000e+00 |
| GWO | 6.117226e-47 | 1.049965e-79 | 3.664247e-145 | 2.637523e-157 | 2.758506e-14 | 1.000000e+00 | 1.000000e+00 |

Statistical significance results for precision, non-continuous rastrigin function, 5 dimensions

| | BSO | BPO | SA | TA | PSO | WO | GWO |
|---|---|---|---|---|---|---|---|
| BSO | 1.000000e+00 | 8.591806e-10 | 3.710147e-37 | 4.039922e-33 | 8.563318e-05 | 3.844674e-37 | 3.844674e-37 |
| BPO | 8.591806e-10 | 1.000000e+00 | 6.602429e-11 | 1.720948e-08 | 3.419614e-26 | 4.185244e-72 | 4.185244e-72 |
| SA | 3.710147e-37 | 6.602429e-11 | 1.000000e+00 | 1.000000e+00 | 8.116995e-61 | 1.991361e-112 | 1.991361e-112 |
| TA | 4.039922e-33 | 1.720948e-08 | 1.000000e+00 | 1.000000e+00 | 3.615960e-56 | 1.886625e-107 | 1.886625e-107 |
| PSO | 8.563318e-05 | 3.419614e-26 | 8.116995e-61 | 3.615960e-56 | 1.000000e+00 | 1.229169e-17 | 1.229169e-17 |
| WO | 3.844674e-37 | 4.185244e-72 | 1.991361e-112 | 1.886625e-107 | 1.229169e-17 | 1.000000e+00 | 1.000000e+00 |
| GWO | 3.844674e-37 | 4.185244e-72 | 1.991361e-112 | 1.886625e-107 | 1.229169e-17 | 1.000000e+00 | 1.000000e+00 |

Statistical significance results for precision, non-continuous rastrigin function, 6 dimensions

| | BSO | BPO | SA | TA | PSO | WO | GWO |
|---|---|---|---|---|---|---|---|
| BSO | 1.000000e+00 | 3.949837e-12 | 5.370943e-33 | 1.064327e-18 | 8.331554e-16 | 4.225358e-45 | 4.225358e-45 |
| BPO | 3.949837e-12 | 1.000000e+00 | 1.891699e-06 | 1.000000e+00 | 1.403798e-48 | 9.061276e-86 | 9.061276e-86 |
| SA | 5.370943e-33 | 1.891699e-06 | 1.000000e+00 | 1.211420e-02 | 1.057539e-77 | 1.237198e-116 | 1.237198e-116 |
| TA | 1.064327e-18 | 1.000000e+00 | 1.211420e-02 | 1.000000e+00 | 9.725290e-59 | 7.485082e-97 | 7.485082e-97 |
| PSO | 8.331554e-16 | 1.403798e-48 | 1.057539e-77 | 9.725290e-59 | 1.000000e+00 | 6.123728e-10 | 6.123728e-10 |
| WO | 4.225358e-45 | 9.061276e-86 | 1.237198e-116 | 7.485082e-97 | 6.123728e-10 | 1.000000e+00 | 1.000000e+00 |
| GWO | 4.225358e-45 | 9.061276e-86 | 1.237198e-116 | 7.485082e-97 | 6.123728e-10 | 1.000000e+00 | 1.000000e+00 |

Statistical significance results for precision, non-continuous ellipsoid function, 2 dimensions

| | BSO | BPO | SA | TA | PSO | WO | GWO |
|---|---|---|---|---|---|---|---|
| BSO | 1.000000e+00 | 3.088516e-158 | 8.782423e-43 | 2.360818e-83 | 6.941234e-13 | 1.995452e-86 | 1.442281e-155 |
| BPO | 3.088516e-158 | 1.000000e+00 | 1.038481e-73 | 1.409582e-34 | 4.555928e-199 | 2.342912e-270 | 4.434437e-319 |
| SA | 8.782423e-43 | 1.038481e-73 | 1.000000e+00 | 2.981259e-12 | 1.666805e-84 | 1.885103e-170 | 6.003585e-232 |
| TA | 2.360818e-83 | 1.409582e-34 | 2.981259e-12 | 1.000000e+00 | 1.489403e-127 | 2.309255e-209 | 6.118324e-266 |
| PSO | 6.941234e-13 | 4.555928e-199 | 1.666805e-84 | 1.489403e-127 | 1.000000e+00 | 1.828908e-44 | 5.401416e-112 |
| WO | 1.995452e-86 | 2.342912e-270 | 1.885103e-170 | 2.309255e-209 | 1.828908e-44 | 1.000000e+00 | 6.205399e-30 |
| GWO | 1.442281e-155 | 4.434437e-319 | 6.003585e-232 | 6.118324e-266 | 5.401416e-112 | 6.205399e-30 | 1.000000e+00 |

Statistical significance results for precision, non-continuous ellipsoid function, 3 dimensions

| | BSO | BPO | SA | TA | PSO | WO | GWO |
|---|---|---|---|---|---|---|---|
| BSO | 1.000000e+00 | 5.937167e-58 | 1.805622e-177 | 6.506650e-215 | 7.886521e-55 | 7.249479e-183 | 9.015501e-213 |
| BPO | 5.937167e-58 | 1.000000e+00 | 3.077312e-77 | 3.875831e-119 | 2.333565e-155 | 4.694379e-268 | 1.660629e-292 |
| SA | 1.805622e-177 | 3.077312e-77 | 1.000000e+00 | 1.145415e-11 | 4.605683e-261 | 0.000000e+00 | 0.000000e+00 |
| TA | 6.506650e-215 | 3.875831e-119 | 1.145415e-11 | 1.000000e+00 | 6.807771e-292 | 0.000000e+00 | 0.000000e+00 |
| PSO | 7.886521e-55 | 2.333565e-155 | 4.605683e-261 | 6.807771e-292 | 1.000000e+00 | 2.346793e-86 | 5.240362e-120 |
| WO | 7.249479e-183 | 4.694379e-268 | 0.000000e+00 | 0.000000e+00 | 2.346793e-86 | 1.000000e+00 | 1.255149e-07 |
| GWO | 9.015501e-213 | 1.660629e-292 | 0.000000e+00 | 0.000000e+00 | 5.240362e-120 | 1.255149e-07 | 1.000000e+00 |

Statistical significance results for precision, non-continuous ellipsoid function, 4 dimensions

| | BSO | BPO | SA | TA | PSO | WO | GWO |
|---|---|---|---|---|---|---|---|
| BSO | 1.000000e+00 | 4.043858e-81 | 4.061497e-220 | 3.264344e-275 | 6.703840e-198 | 6.703840e-198 | 6.703840e-198 |
| BPO | 4.043858e-81 | 1.000000e+00 | 3.211050e-101 | 1.621901e-169 | 2.103690e-297 | 2.103690e-297 | 2.103690e-297 |
| SA | 4.061497e-220 | 3.211050e-101 | 1.000000e+00 | 7.688655e-30 | 0.000000e+00 | 0.000000e+00 | 0.000000e+00 |
| TA | 3.264344e-275 | 1.621901e-169 | 7.688655e-30 | 1.000000e+00 | 0.000000e+00 | 0.000000e+00 | 0.000000e+00 |
| PSO | 6.703840e-198 | 2.103690e-297 | 0.000000e+00 | 0.000000e+00 | 1.000000e+00 | 1.000000e+00 | 1.000000e+00 |
| WO | 6.703840e-198 | 2.103690e-297 | 0.000000e+00 | 0.000000e+00 | 1.000000e+00 | 1.000000e+00 | 1.000000e+00 |
| GWO | 6.703840e-198 | 2.103690e-297 | 0.000000e+00 | 0.000000e+00 | 1.000000e+00 | 1.000000e+00 | 1.000000e+00 |

Statistical significance results for precision, non-continuous ellipsoid function, 5 dimensions

| | BSO | BPO | SA | TA | PSO | WO | GWO |
|---|---|---|---|---|---|---|---|
| BSO | 1.000000e+00 | 1.989715e-100 | 1.125611e-248 | 0.000000e+00 | 1.460443e-235 | 1.460443e-235 | 1.460443e-235 |
| BPO | 1.989715e-100 | 1.000000e+00 | 8.707681e-117 | 5.137971e-222 | 0.000000e+00 | 0.000000e+00 | 0.000000e+00 |
| SA | 1.125611e-248 | 8.707681e-117 | 1.000000e+00 | 3.963020e-68 | 0.000000e+00 | 0.000000e+00 | 0.000000e+00 |
| TA | 0.000000e+00 | 5.137971e-222 | 3.963020e-68 | 1.000000e+00 | 0.000000e+00 | 0.000000e+00 | 0.000000e+00 |
| PSO | 1.460443e-235 | 0.000000e+00 | 0.000000e+00 | 0.000000e+00 | 1.000000e+00 | 1.000000e+00 | 1.000000e+00 |
| WO | 1.460443e-235 | 0.000000e+00 | 0.000000e+00 | 0.000000e+00 | 1.000000e+00 | 1.000000e+00 | 1.000000e+00 |
| GWO | 1.460443e-235 | 0.000000e+00 | 0.000000e+00 | 0.000000e+00 | 1.000000e+00 | 1.000000e+00 | 1.000000e+00 |

Statistical significance results for precision, non-continuous ellipsoid function, 6 dimensions

| | BSO | BPO | SA | TA | PSO | WO | GWO |
|---|---|---|---|---|---|---|---|
| BSO | 1.000000e+00 | 1.201213e-57 | 1.819397e-166 | 3.973732e-240 | 3.822985e-146 | 5.593358e-157 | 5.593358e-157 |
| BPO | 1.201213e-57 | 1.000000e+00 | 4.009437e-66 | 2.476875e-149 | 1.811163e-237 | 1.764968e-246 | 1.764968e-246 |
| SA | 1.819397e-166 | 4.009437e-66 | 1.000000e+00 | 1.437343e-41 | 2.273038e-318 | 0.000000e+00 | 0.000000e+00 |
| TA | 3.973732e-240 | 2.476875e-149 | 1.437343e-41 | 1.000000e+00 | 0.000000e+00 | 0.000000e+00 | 0.000000e+00 |
| PSO | 3.822985e-146 | 1.811163e-237 | 2.273038e-318 | 0.000000e+00 | 1.000000e+00 | 1.000000e+00 | 1.000000e+00 |
| WO | 5.593358e-157 | 1.764968e-246 | 0.000000e+00 | 0.000000e+00 | 1.000000e+00 | 1.000000e+00 | 1.000000e+00 |
| GWO | 5.593358e-157 | 1.764968e-246 | 0.000000e+00 | 0.000000e+00 | 1.000000e+00 | 1.000000e+00 | 1.000000e+00 |

Statistical significance results for precision, non-continuous xin-she-yang-n2 function, 2 dimensions

| | BSO | BPO | SA | TA | PSO | WO | GWO |
|---|---|---|---|---|---|---|---|
| BSO | 1.000000e+00 | 1.871720e-115 | 1.185292e-61 | 1.888919e-50 | 1.285647e-18 | 7.605296e-19 | 1.564376e-62 |
| BPO | 1.871720e-115 | 1.000000e+00 | 5.288421e-19 | 1.883540e-27 | 3.216852e-62 | 2.600361e-168 | 6.414649e-216 |
| SA | 1.185292e-61 | 5.288421e-19 | 1.000000e+00 | 7.256204e-01 | 1.051724e-17 | 3.282772e-115 | 6.160912e-167 |
| TA | 1.888919e-50 | 1.883540e-27 | 7.256204e-01 | 1.000000e+00 | 8.548927e-11 | 4.277143e-103 | 2.007954e-155 |
| PSO | 1.285647e-18 | 3.216852e-62 | 1.051724e-17 | 8.548927e-11 | 1.000000e+00 | 2.377592e-63 | 8.753154e-116 |
| WO | 7.605296e-19 | 2.600361e-168 | 3.282772e-115 | 4.277143e-103 | 2.377592e-63 | 1.000000e+00 | 4.560670e-18 |
| GWO | 1.564376e-62 | 6.414649e-216 | 6.160912e-167 | 2.007954e-155 | 8.753154e-116 | 4.560670e-18 | 1.000000e+00 |

Statistical significance results for precision, non-continuous xin-she-yang-n2 function, 3 dimensions

| | BSO | BPO | SA | TA | PSO | WO | GWO |
|---|---|---|---|---|---|---|---|
| BSO | 1.000000e+00 | 1.304871e-204 | 4.027639e-74 | 1.788494e-119 | 1.530682e-46 | 3.169566e-134 | 3.416902e-217 |
| BPO | 1.304871e-204 | 1.000000e+00 | 2.437348e-90 | 8.213873e-47 | 9.287353e-277 | 0.000000e+00 | 0.000000e+00 |
| SA | 4.027639e-74 | 2.437348e-90 | 1.000000e+00 | 1.374723e-13 | 9.821490e-163 | 1.500183e-241 | 1.109337e-307 |
| TA | 1.788494e-119 | 8.213873e-47 | 1.374723e-13 | 1.000000e+00 | 2.416869e-204 | 1.556021e-276 | 0.000000e+00 |
| PSO | 1.530682e-46 | 9.287353e-277 | 9.821490e-163 | 2.416869e-204 | 1.000000e+00 | 1.530682e-46 | 3.169566e-134 |
| WO | 3.169566e-134 | 0.000000e+00 | 1.500183e-241 | 1.556021e-276 | 1.530682e-46 | 1.000000e+00 | 1.530682e-46 |
| GWO | 3.416902e-217 | 0.000000e+00 | 1.109337e-307 | 0.000000e+00 | 3.169566e-134 | 1.530682e-46 | 1.000000e+00 |

Statistical significance results for precision, non-continuous xin-she-yang-n2 function, 4 dimensions

| | BSO | BPO | SA | TA | PSO | WO | GWO |
|---|---|---|---|---|---|---|---|
| BSO | 1.000000e+00 | 3.408749e-37 | 1.485953e-118 | 1.694417e-25 | 1.083724e-44 | 2.161918e-12 | 1.050554e-73 |
| BPO | 3.408749e-37 | 1.000000e+00 | 2.795964e-42 | 1.845034e-01 | 1.000000e+00 | 4.297665e-77 | 1.406931e-10 |
| SA | 1.485953e-118 | 2.795964e-42 | 1.000000e+00 | 1.019552e-55 | 6.452974e-35 | 5.603098e-161 | 2.605222e-13 |
| TA | 1.694417e-25 | 1.845034e-01 | 1.019552e-55 | 1.000000e+00 | 7.259791e-04 | 1.399916e-62 | 2.682605e-19 |
| PSO | 1.083724e-44 | 1.000000e+00 | 6.452974e-35 | 7.259791e-04 | 1.000000e+00 | 8.689683e-86 | 1.517935e-06 |
| WO | 2.161918e-12 | 4.297665e-77 | 5.603098e-16 | 1.399916e-62 | 8.689683e-86 | 1.000000e+00 | 6.920541e-117 |
| GWO | 1.050554e-73 | 1.406931e-10 | 2.605222e-13 | 2.682605e-19 | 1.517935e-06 | 6.920541e-117 | 1.000000e+00 |

Statistical significance results for precision, non-continuous xin-she-yang-n2 function, 5 dimensions

| | BSO | BPO | SA | TA | PSO | WO | GWO |
|---|---|---|---|---|---|---|---|
| BSO | 1.000000e+00 | 1.922159e-130 | 4.757042e-75 | 2.030571e-71 | 3.068075e-38 | 1.304921e-19 | 3.277149e-12 |
| BPO | 1.922159e-130 | 1.000000e+00 | 9.310666e-20 | 2.881729e-22 | 9.235469e-52 | 2.212941e-183 | 1.776961e-87 |
| SA | 4.757042e-75 | 9.310666e-20 | 1.000000e+00 | 1.000000e+00 | 1.226260e-10 | 3.196662e-130 | 9.397703e-36 |
| TA | 2.030571e-71 | 2.881729e-22 | 1.000000e+00 | 1.000000e+00 | 8.002865e-09 | 1.573295e-126 | 9.409753e-33 |
| PSO | 3.068075e-38 | 9.235469e-52 | 1.226260e-10 | 8.002865e-09 | 1.000000e+00 | 2.985668e-90 | 3.187552e-09 |
| WO | 1.304921e-19 | 2.212941e-183 | 3.196662e-130 | 1.573295e-126 | 2.985668e-90 | 1.000000e+00 | 2.611398e-54 |
| GWO | 3.277149e-12 | 1.776961e-87 | 9.397703e-36 | 9.409753e-33 | 3.187552e-09 | 2.611398e-54 | 1.000000e+00 |

Statistical significance results for precision, non-continuous xin-she-yang-n2 function, 6 dimensions

| | BSO | BPO | SA | TA | PSO | WO | GWO |
|---|---|---|---|---|---|---|---|
| BSO | 1.000000e+00 | 9.911093e-174 | 4.053012e-65 | 7.176431e-84 | 3.472307e-145 | 9.233516e-45 | 1.000000e+00 |
| BPO | 9.911093e-174 | 1.000000e+00 | 5.736379e-66 | 3.968212e-48 | 4.494305e-06 | 3.029111e-249 | 1.347266e-175 |
| SA | 4.053012e-65 | 5.736379e-66 | 1.000000e+00 | 1.704489e-02 | 7.724934e-39 | 6.985806e-152 | 4.834479e-67 |
| TA | 7.176431e-84 | 3.968212e-48 | 1.704489e-02 | 1.000000e+00 | 6.741864e-24 | 3.601907e-170 | 7.328554e-86 |
| PSO | 3.472307e-145 | 4.494305e-06 | 7.724934e-39 | 6.741864e-24 | 1.000000e+00 | 6.301915e-225 | 3.881840e-147 |
| WO | 9.233516e-45 | 3.029111e-249 | 6.985806e-152 | 3.601907e-170 | 6.301915e-22 | 1.000000e+00 | 5.144303e-43 |
| GWO | 1.000000e+00 | 1.347266e-175 | 4.834479e-67 | 7.328554e-86 | 3.881840e-147 | 5.144303e-43 | 1.000000e+00 |

Statistical significance results for precision, non-continuous step-fun function, 2 dimensions

| | BSO | BPO | SA | TA | PSO | WO | GWO |
|---|---|---|---|---|---|---|---|
| BSO | 1.000000e+00 | 2.448637e-09 | 2.448637e-09 | 2.448637e-09 | 2.448637e-09 | 2.448637e-09 | 2.448637e-09 |
| BPO | 2.448637e-09 | 1.000000e+00 | 1.000000e+00 | 1.000000e+00 | 1.000000e+00 | 1.000000e+00 | 1.000000e+00 |
| SA | 2.448637e-09 | 1.000000e+00 | 1.000000e+00 | 1.000000e+00 | 1.000000e+00 | 1.000000e+00 | 1.000000e+00 |
| TA | 2.448637e-09 | 1.000000e+00 | 1.000000e+00 | 1.000000e+00 | 1.000000e+00 | 1.000000e+00 | 1.000000e+00 |
| PSO | 2.448637e-09 | 1.000000e+00 | 1.000000e+00 | 1.000000e+00 | 1.000000e+00 | 1.000000e+00 | 1.000000e+00 |
| WO | 2.448637e-09 | 1.000000e+00 | 1.000000e+00 | 1.000000e+00 | 1.000000e+00 | 1.000000e+00 | 1.000000e+00 |
| GWO | 2.448637e-09 | 1.000000e+00 | 1.000000e+00 | 1.000000e+00 | 1.000000e+00 | 1.000000e+00 | 1.000000e+00 |

Statistical significance results for precision, non-continuous step-fun function, 3 dimensions

| | BSO | BPO | SA | TA | PSO | WO | GWO |
|---|---|---|---|---|---|---|---|
| BSO | 1.000000 | 0.226964 | 0.226964 | 1.000000 | 0.226964 | 0.226964 | 0.226964 |
| BPO | 0.226964 | 1.000000 | 1.000000 | 0.034573 | 1.000000 | 1.000000 | 1.000000 |
| SA | 0.226964 | 1.000000 | 1.000000 | 0.034573 | 1.000000 | 1.000000 | 1.000000 |
| TA | 1.000000 | 0.034573 | 0.034573 | 1.000000 | 0.034573 | 0.034573 | 0.034573 |
| PSO | 0.226964 | 1.000000 | 1.000000 | 0.034573 | 1.000000 | 1.000000 | 1.000000 |
| WO | 0.226964 | 1.000000 | 1.000000 | 0.034573 | 1.000000 | 1.000000 | 1.000000 |
| GWO | 0.226964 | 1.000000 | 1.000000 | 0.034573 | 1.000000 | 1.000000 | 1.000000 |

Statistical significance results for precision, non-continuous step-fun function, 4 dimensions

| | BSO | BPO | SA | TA | PSO | WO | GWO |
|---|---|---|---|---|---|---|---|
| BSO | 1.000000e+00 | 4.372059e-14 | 4.372059e-14 | 4.372059e-14 | 4.372059e-14 | 4.372059e-14 | 4.372059e-14 |
| BPO | 4.372059e-14 | 1.000000e+00 | 1.000000e+00 | 1.000000e+00 | 1.000000e+00 | 1.000000e+00 | 1.000000e+00 |
| SA | 4.372059e-14 | 1.000000e+00 | 1.000000e+00 | 1.000000e+00 | 1.000000e+00 | 1.000000e+00 | 1.000000e+00 |
| TA | 4.372059e-14 | 1.000000e+00 | 1.000000e+00 | 1.000000e+00 | 1.000000e+00 | 1.000000e+00 | 1.000000e+00 |
| PSO | 4.372059e-14 | 1.000000e+00 | 1.000000e+00 | 1.000000e+00 | 1.000000e+00 | 1.000000e+00 | 1.000000e+00 |
| WO | 4.372059e-14 | 1.000000e+00 | 1.000000e+00 | 1.000000e+00 | 1.000000e+00 | 1.000000e+00 | 1.000000e+00 |
| GWO | 4.372059e-14 | 1.000000e+00 | 1.000000e+00 | 1.000000e+00 | 1.000000e+00 | 1.000000e+00 | 1.000000e+00 |

Statistical significance results for precision, non-continuous step-fun function, 5 dimensions

| | BSO | BPO | SA | TA | PSO | WO | GWO |
|---|---|---|---|---|---|---|---|
| BSO | 1.000000e+00 | 1.557083e-07 | 1.557083e-07 | 1.557083e-07 | 1.557083e-07 | 1.557083e-07 | 1.557083e-07 |
| BPO | 1.557083e-07 | 1.000000e+00 | 1.000000e+00 | 1.000000e+00 | 1.000000e+00 | 1.000000e+00 | 1.000000e+00 |
| SA | 1.557083e-07 | 1.000000e+00 | 1.000000e+00 | 1.000000e+00 | 1.000000e+00 | 1.000000e+00 | 1.000000e+00 |
| TA | 1.557083e-07 | 1.000000e+00 | 1.000000e+00 | 1.000000e+00 | 1.000000e+00 | 1.000000e+00 | 1.000000e+00 |
| PSO | 1.557083e-07 | 1.000000e+00 | 1.000000e+00 | 1.000000e+00 | 1.000000e+00 | 1.000000e+00 | 1.000000e+00 |
| WO | 1.557083e-07 | 1.000000e+00 | 1.000000e+00 | 1.000000e+00 | 1.000000e+00 | 1.000000e+00 | 1.000000e+00 |

| GWO | 1.557083e-07 | 1.000000e+00 | 1.000000e+00 | 1.000000e+00 | 1.000000e+00 | 1.000000e+00 | 1.000000e+00 |
|---|---|---|---|---|---|---|---|

Statistical significance results for precision, non-continuous step-fun function, 6 dimensions

| | BSO | BPO | SA | TA | PSO | WO | GWO |
|---|---|---|---|---|---|---|---|
| BSO | 1.000000e+00 | 8.309616e-04 | 8.309616e-04 | 8.309616e-04 | 8.309616e-04 | 8.309616e-04 | 2.131064e-93 |
| BPO | 8.309616e-04 | 1.000000e+00 | 1.000000e+00 | 1.000000e+00 | 1.000000e+00 | 1.000000e+00 | 4.700806e-117 |
| SA | 8.309616e-04 | 1.000000e+00 | 1.000000e+00 | 1.000000e+00 | 1.000000e+00 | 1.000000e+00 | 4.700806e-117 |
| TA | 8.309616e-04 | 1.000000e+00 | 1.000000e+00 | 1.000000e+00 | 1.000000e+00 | 1.000000e+00 | 4.700806e-117 |
| PSO | 8.309616e-04 | 1.000000e+00 | 1.000000e+00 | 1.000000e+00 | 1.000000e+00 | 1.000000e+00 | 4.700806e-117 |
| WO | 8.309616e-04 | 1.000000e+00 | 1.000000e+00 | 1.000000e+00 | 1.000000e+00 | 1.000000e+00 | 4.700806e-117 |
| GWO | 2.131064e-93 | 4.700806e-117 | 4.700806e-117 | 4.700806e-117 | 4.700806e-117 | 4.700806e-117 | 1.000000e+00 |

Statistical significance results for precision, non-continuous rosenbrock function, 2 dimensions

| | BSO | BPO | SA | TA | PSO | WO | GWO |
|---|---|---|---|---|---|---|---|
| BSO | 1.000000e+00 | 7.692483e-36 | 4.589526e-03 | 4.589526e-03 | 4.589526e-03 | 4.589526e-03 | 4.589526e-03 |
| BPO | 7.692483e-36 | 1.000000e+00 | 2.215904e-54 | 2.215904e-54 | 2.215904e-54 | 2.215904e-54 | 2.215904e-54 |
| SA | 4.589526e-03 | 2.215904e-54 | 1.000000e+00 | 1.000000e+00 | 1.000000e+00 | 1.000000e+00 | 1.000000e+00 |
| TA | 4.589526e-03 | 2.215904e-54 | 1.000000e+00 | 1.000000e+00 | 1.000000e+00 | 1.000000e+00 | 1.000000e+00 |
| PSO | 4.589526e-03 | 2.215904e-54 | 1.000000e+00 | 1.000000e+00 | 1.000000e+00 | 1.000000e+00 | 1.000000e+00 |
| WO | 4.589526e-03 | 2.215904e-54 | 1.000000e+00 | 1.000000e+00 | 1.000000e+00 | 1.000000e+00 | 1.000000e+00 |
| GWO | 4.589526e-03 | 2.215904e-54 | 1.000000e+00 | 1.000000e+00 | 1.000000e+00 | 1.000000e+00 | 1.000000e+00 |

Statistical significance results for precision, non-continuous rosenbrock function, 3 dimensions

| | BSO | BPO | SA | TA | PSO | WO | GWO |
|---|---|---|---|---|---|---|---|
| BSO | 1.000000e+00 | 4.677303e-290 | 1.157513e-07 | 1.157513e-07 | 1.157513e-07 | 1.157513e-07 | 1.157513e-07 |
| BPO | 4.677303e-290 | 1.000000e+00 | 3.626758e-313 | 3.626758e-313 | 3.626758e-313 | 3.626758e-313 | 3.626758e-313 |
| SA | 1.157513e-07 | 3.626758e-313 | 1.000000e+00 | 1.000000e+00 | 1.000000e+00 | 1.000000e+00 | 1.000000e+00 |
| TA | 1.157513e-07 | 3.626758e-313 | 1.000000e+00 | 1.000000e+00 | 1.000000e+00 | 1.000000e+00 | 1.000000e+00 |
| PSO | 1.157513e-07 | 3.626758e-313 | 1.000000e+00 | 1.000000e+00 | 1.000000e+00 | 1.000000e+00 | 1.000000e+00 |
| WO | 1.157513e-07 | 3.626758e-313 | 1.000000e+00 | 1.000000e+00 | 1.000000e+00 | 1.000000e+00 | 1.000000e+00 |
| GWO | 1.157513e-07 | 3.626758e-313 | 1.000000e+00 | 1.000000e+00 | 1.000000e+00 | 1.000000e+00 | 1.000000e+00 |

Statistical significance results for precision, non-continuous rosenbrock function, 5 dimensions

| | BSO | BPO | SA | TA | PSO | WO | GWO |
|---|---|---|---|---|---|---|---|
| BSO | 1.000000e+00 | 8.219832e-23 | 8.219832e-23 | 8.219832e-23 | 8.219832e-23 | 8.219832e-23 | 4.319904e-86 |
| BPO | 8.219832e-23 | 1.000000e+00 | 1.000000e+00 | 1.000000e+00 | 1.000000e+00 | 1.000000e+00 | 7.377478e-146 |
| SA | 8.219832e-23 | 1.000000e+00 | 1.000000e+00 | 1.000000e+00 | 1.000000e+00 | 1.000000e+00 | 7.377478e-146 |
| TA | 8.219832e-23 | 1.000000e+00 | 1.000000e+00 | 1.000000e+00 | 1.000000e+00 | 1.000000e+00 | 7.377478e-146 |
| PSO | 8.219832e-23 | 1.000000e+00 | 1.000000e+00 | 1.000000e+00 | 1.000000e+00 | 1.000000e+00 | 7.377478e-146 |
| WO | 8.219832e-23 | 1.000000e+00 | 1.000000e+00 | 1.000000e+00 | 1.000000e+00 | 1.000000e+00 | 7.377478e-146 |
| GWO | 4.319904e-86 | 7.377478e-146 | 7.377478e-146 | 7.377478e-146 | 7.377478e-146 | 7.377478e-146 | 1.000000e+00 |

Statistical significance results for precision, non-continuous rosenbrock function, 6 dimensions

| | BSO | BPO | SA | TA | PSO | WO | GWO |
|---|---|---|---|---|---|---|---|
| BSO | 1.000000e+00 | 1.816199e-165 | 2.273229e-27 | 2.273229e-27 | 2.273229e-27 | 2.273229e-27 | 2.273229e-27 |
| BPO | 1.816199e-165 | 1.000000e+00 | 9.623855e-225 | 9.623855e-225 | 9.623855e-225 | 9.623855e-225 | 9.623855e-225 |
| SA | 2.273229e-27 | 9.623855e-225 | 1.000000e+00 | 1.000000e+00 | 1.000000e+00 | 1.000000e+00 | 1.000000e+00 |
| TA | 2.273229e-27 | 9.623855e-225 | 1.000000e+00 | 1.000000e+00 | 1.000000e+00 | 1.000000e+00 | 1.000000e+00 |
| PSO | 2.273229e-27 | 9.623855e-225 | 1.000000e+00 | 1.000000e+00 | 1.000000e+00 | 1.000000e+00 | 1.000000e+00 |
| WO | 2.273229e-27 | 9.623855e-225 | 1.000000e+00 | 1.000000e+00 | 1.000000e+00 | 1.000000e+00 | 1.000000e+00 |
| GWO | 2.273229e-27 | 9.623855e-225 | 1.000000e+00 | 1.000000e+00 | 1.000000e+00 | 1.000000e+00 | 1.000000e+00 |

Statistical significance results for precision, non-continuous quadric function, 2 dimensions

| | BSO | BPO | SA | TA | PSO | WO | GWO |
|---|---|---|---|---|---|---|---|
| BSO | 1.000000e+00 | 7.070939e-177 | 7.070939e-177 | 7.070939e-177 | 7.070939e-177 | 7.070939e-177 | 7.070939e-177 |
| BPO | 7.070939e-177 | 1.000000e+00 | 1.000000e+00 | 1.000000e+00 | 1.000000e+00 | 1.000000e+00 | 1.000000e+00 |
| SA | 7.070939e-177 | 1.000000e+00 | 1.000000e+00 | 1.000000e+00 | 1.000000e+00 | 1.000000e+00 | 1.000000e+00 |
| TA | 7.070939e-177 | 1.000000e+00 | 1.000000e+00 | 1.000000e+00 | 1.000000e+00 | 1.000000e+00 | 1.000000e+00 |
| PSO | 7.070939e-177 | 1.000000e+00 | 1.000000e+00 | 1.000000e+00 | 1.000000e+00 | 1.000000e+00 | 1.000000e+00 |
| WO | 7.070939e-177 | 1.000000e+00 | 1.000000e+00 | 1.000000e+00 | 1.000000e+00 | 1.000000e+00 | 1.000000e+00 |
| GWO | 7.070939e-177 | 1.000000e+00 | 1.000000e+00 | 1.000000e+00 | 1.000000e+00 | 1.000000e+00 | 1.000000e+00 |

Statistical significance results for precision, non-continuous quadric function, 3 dimensions

| | BSO | BPO | SA | TA | PSO | WO | GWO |
|---|---|---|---|---|---|---|---|
| BSO | 1.000000e+00 | 1.075391e-59 | 6.522977e-157 | 7.341657e-145 | 2.349967e-41 | 1.026521e-161 | 2.684776e-163 |
| BPO | 1.075391e-59 | 1.000000e+00 | 1.312251e-54 | 1.949353e-43 | 1.670592e-142 | 4.076722e-252 | 2.033882e-253 |
| SA | 6.522977e-157 | 1.312251e-54 | 1.000000e+00 | 6.166304e-01 | 9.803902e-232 | 2.075076e-322 | 0.000000e+00 |
| TA | 7.341657e-145 | 1.949353e-43 | 6.166304e-01 | 1.000000e+00 | 2.442917e-221 | 3.313948e-314 | 2.632022e-315 |
| PSO | 2.349967e-41 | 1.670592e-142 | 9.803902e-232 | 2.442917e-221 | 1.000000e+00 | 6.489845e-79 | 1.478338e-80 |
| WO | 1.026521e-161 | 4.076722e-252 | 2.075076e-322 | 3.313948e-314 | 6.489845e-79 | 1.000000e+00 | 1.000000e+00 |
| GWO | 2.684776e-163 | 2.033882e-253 | 0.000000e+00 | 2.632022e-315 | 1.478338e-80 | 1.000000e+00 | 1.000000e+00 |

Statistical significance results for precision, non-continuous quadric function, 4 dimensions

| | BSO | BPO | SA | TA | PSO | WO | GWO |
|---|---|---|---|---|---|---|---|
| BSO | 1.000000e+00 | 1.906033e-70 | 7.350649e-168 | 8.143459e-130 | 1.014049e-161 | 2.339371e-87 | 1.907194e-10 |
| BPO | 1.906033e-70 | 1.000000e+00 | 7.370997e-55 | 1.601879e-22 | 7.856301e-261 | 1.224514e-198 | 8.272286e-110 |
| SA | 7.350649e-168 | 7.370997e-55 | 1.000000e+00 | 2.795391e-10 | 0.000000e+00 | 1.553212e-278 | 6.207871e-204 |
| TA | 8.143459e-130 | 1.601879e-22 | 2.795391e-10 | 1.000000e+00 | 6.162753e-304 | 1.105586e-248 | 3.585949e-168 |
| PSO | 1.014049e-161 | 7.856301e-261 | 0.000000e+00 | 6.162753e-304 | 1.000000e+00 | 2.857509e-34 | 4.813322e-123 |
| WO | 2.339371e-87 | 1.224514e-198 | 1.553212e-278 | 1.105586e-248 | 2.857509e-34 | 1.000000e+00 | 2.146090e-49 |
| GWO | 1.907194e-10 | 8.272286e-110 | 6.207871e-204 | 3.585949e-168 | 4.813322e-123 | 2.146090e-49 | 1.000000e+00 |

Statistical significance results for precision, non-continuous quadric function, 5 dimensions

| | BSO | BPO | SA | TA | PSO | WO | GWO |
|---|---|---|---|---|---|---|---|
| BSO | 1.000000e+00 | 6.568081e-85 | 4.330204e-191 | 2.612721e-139 | 7.043150e-128 | 1.038578e-77 | 1.000000e+00 |
| BPO | 6.568081e-85 | 1.000000e+00 | 1.067563e-64 | 4.209783e-19 | 3.533556e-245 | 4.919846e-203 | 4.528032e-8 |
| SA | 4.330204e-191 | 1.067563e-64 | 1.000000e+00 | 2.825438e-19 | 0.000000e+00 | 1.201090e-289 | 3.088769e-191 |
| TA | 2.612721e-139 | 4.209783e-19 | 2.825438e-19 | 1.000000e+00 | 2.128095e-286 | 1.375472e-248 | 1.809132e-139 |
| PSO | 7.043150e-128 | 3.533556e-245 | 0.000000e+00 | 2.128095e-286 | 1.000000e+00 | 2.140315e-16 | 1.021465e-127 |
| WO | 1.038578e-77 | 4.919846e-203 | 1.201090e-289. | 1.375472e-248 | 2.140315e-16 | 1.000000e+00 | 1.501015e-77 |
| GWO | 1.000000e+00 | 4.528032e-85 | 3.088769e-191 | 1.809132e-139 | 1.021465e-127 | 1.501015e-77 | 1.000000e+00 |

Statistical significance results for precision, non-continuous quadric function, 6 dimensions

| | BSO | BPO | SA | TA | PSO | WO | GWO |
|---|---|---|---|---|---|---|---|
| BSO | 1.000000e+00 | 1.812624e-87 | 2.465727e-198 | 1.766380e-148 | 2.129548e-116 | 2.129548e-116 | 1.000000e+00] |
| BPO | 1.812624e-87 | 1.000000e+00 | 5.243584e-70 | 9.865059e-24 | 5.541615e-238 | 5.541615e-238 | 8.278623e-97 |
| SA | 2.465727e-198 | 5.243584e-70 | 1.000000e+00 | 1.969040e-18 | 2.075076e-321 | 2.075076e-321 | 1.424389e-206 |
| TA | 1.766380e-148 | 9.865059e-24 | 1.969040e-18 | 1.000000e+00 | 7.088264e-285 | 7.088264e-285 | 1.700048e-157 |
| PSO | 2.129548e-116 | 5.541615e-238 | 2.075076e-321 | 7.088264e-285 | 1.000000e+00 | 1.000000e+00 | 4.950044e-107 |
| WO | 2.129548e-116 | 5.541615e-238 | 2.075076e-321 | 7.088264e-285 | 1.000000e+00 | 1.000000e+00 | 4.950044e-107 |
| GWO | 1.000000e+00 | 8.278623e-97 | 1.424389e-206 | 1.700048e-157 | 4.950044e-107 | 4.950044e-107 | 1.000000e+00 |

# APPENDIX H: STATISTICAL SIGNIFICANCE TEST RESULTS FOR CASE STUDY 2

In this appendix, we are having the statistical significance results for Case Study 2 results. All the following statistical result table represents Convor's results for BSO, BPO, SA, TA, PSO, WO, GWO. Our test criteria include the Kruskal-Wallis H test, complemented by subsequent Conover's tests (incorporating the step-down technique with a Bonferroni correction for p-value modification), is employed. A 0.05 p-value benchmark is set for determining statistical relevance. For non-continuous functions the statistical test results are generated by using the result data mentioned in Table D2

Statistical significance results for precision, non-continuous rastrigin function, 2 dimensions

| | BSO | BPO | SA | TA | PSO | WO | GWO |
|---|---|---|---|---|---|---|---|
| BSO | 1.000000e+00 | 2.687487e-15 | 1.521904e-176 | 5.988418e-146 | 6.165908e-165 | 6.165908e-165 | 6.165908e-165 |
| BPO | 2.687487e-15 | 1.000000e+00 | 6.221816e-130 | 8.446383e-98 | 5.980762e-209 | 5.980762e-209 | 5.980762e-209 |
| SA | 1.521904e-176. | 6.221816e-130 | 1.000000e+00 | 5.437971e-07 | 0.000000e+00 | 0.000000e+00 | 0.000000e+00 |
| TA | 5.988418e-146 | 8.446383e-98 | 5.437971e-07 | 1.000000e+00 | 3.530160e-317 | 3.530160e-317 | 3.530160e-317 |
| PSO | 6.165908e-165 | 5.980762e-209 | 0.000000e+00 | 3.530160e-317 | 1.000000e+00 | 1.000000e+00 | 1.000000e+00 |
| WO | 6.165908e-165 | 5.980762e-209 | 0.000000e+00 | 3.530160e-317 | 1.000000e+00 | 1.000000e+00 | 1.000000e+00 |
| GWO | 6.165908e-165 | 4.970721e-211 | 0.000000e+00 | 3.130160e-310 | 1.000000e+00 | 1.000000e+00 | 1.000000e+00 |

- Statistical significance results for precision, non-continuous rastrigin function, 3 dimensions

| | BSO | BPO | SA | TA | PSO | WO | GWO |
|---|---|---|---|---|---|---|---|
| BSO | 1.000000e+00 | 4.248464e-12 | 1.102501e-115 | 2.963347e-136 | 3.324463e-107 | 1.411063e-119 | 1.411063e-119 |
| BPO | 4.248464e-12 | 1.000000e+00 | 4.609884e-73 | 1.471818e-93 | 1.462918e-149 | 1.956167e-161 | 1.956167e-161 |
| SA | 1.102501e-115 | 4.609884e-73 | 1.000000e+00 | 6.462972e-03 | 4.046712e-253 | 1.171375e-262 | 1.171375e-262 |
| TA | 2.963347e-136 | 1.471818e-93 | 6.462972e-03 | 1.000000e+00 | 5.761739e-269 | 4.048377e-278 | 4.048377e-278 |
| PSO | 3.324463e-107 | 1.462918e-149 | 4.046712e-253 | 5.761739e-269 | 1.000000e+00 | 6.449920e-01 | 6.449920e-01 |
| WO | 1.411063e-119 | 1.956167e-161 | 1.171375e-262 | 4.048377e-278 | 6.449920e-01 | 1.000000e+00 | 1.000000e+00 |
| GWO | 1.411063e-119 | 5.956167e-131 | 2.171375e-162 | 4.148377e-378 | 6.449920e-01 | 1.000000e+00 | 1.000000e+00 |

- Statistical significance results for precision, non-continuous rastrigin function, 4 dimensions

| | BSO | BPO | SA | TA | PSO | WO | GWO |
|---|---|---|---|---|---|---|---|
| BSO | 1.000000e+00 | 2.886912e-14 | 3.248735e-70 | 1.248165e-74 | 5.018063e-14 | 5.631500e-47 | 5.631500e-47 |
| BPO | 2.886912e-14 | 1.000000e+00 | 6.409372e-29 | 1.932936e-32 | 2.138791e-49 | 1.034844e-91 | 1.034844e-91 |
| SA | 3.248735e-70 | 6.409372e-29 | 1.000000e+00 | 1.000000e+00 | 3.110279e-116 | 2.233405e-159 | 2.233405e-159 |
| TA | 1.248165e-74 | 1.932936e-32 | 1.000000e+00 | 1.000000e+00 | 9.198146e-121 | 1.078921e-163 | 1.078921e-163 |
| PSO | 5.018063e-14 | 2.138791e-49 | 3.110279e-116 | 9.198146e-121 | 1.000000e+00 | 9.627919e-13 | 9.627919e-13 |
| WO | 5.631500e-47 | 1.034844e-91 | 2.233405e-159 | 1.078921e-163 | 9.627919e-13 | 1.000000e+00 | 1.000000e+00 |
| GWO | 5.631500e-47 | 1.134844e-81 | 1.233405e-139 | 1.078921e-163 | 9.57919e-23 | 1.000000e+00 | 1.000000e+00 |

- 

- Statistical significance results for precision, non-continuous rastrigin function, 5 dimensions

| | BSO | BPO | SA | TA | PSO | WO | GWO |
|---|---|---|---|---|---|---|---|
| BSO | 1.000000e+00 | 4.144004e-31 | 2.784310e-137 | 5.157379e-129 | 1.769628e-76 | 1.349553e-95 | 1.349553e-95 |
| BPO | 4.144004e-31 | 1.000000e+00 | 1.619866e-66 | 1.423093e-58 | 2.622378e-147 | 9.063617e-166 | 9.063617e-166 |
| SA | 2.784310e-137 | 1.619866e-66 | 1.000000e+00 | 1.000000e+00 | 5.873278e-246 | 5.547142e-261 | 5.547142e-261 |
| TA | 5.157379e-129 | 1.423093e-58 | 1.000000e+00 | 1.000000e+00 | 3.051261e-239 | 1.607246e-254 | 1.607246e-254 |
| PSO | 1.769628e-76 | 2.622378e-147 | 5.873278e-246 | 3.051261e-239 | 1.000000e+00 | 1.652793e-02 | 1.652793e-02 |
| WO | 1.349553e-95 | 9.063617e-166 | 5.547142e-261 | 1.607246e-254 | 1.652793e-02 | 1.000000e+00 | 1.000000e+00 |
| GWO | 1.349553e-95 | 8.012617e-166 | 5.541242e-161 | 2.617446e-210 | 2.652393e-06 | 1.000000e+00 | 1.000000e+00 |

Statistical significance results for precision, non-continuous rastrigin function, 6 dimensions

| | BSO | BPO | SA | TA | PSO | WO | GWO |
|---|---|---|---|---|---|---|---|
| BSO | 1.000000e+00 | 1.260800e-08 | 2.450694e-33 | 1.154589e-27 | 2.426988e-27 | 3.522396e-52 | 3.522396e-52 |
| BPO | 1.260800e-08 | 1.000000e+00 | 8.797239e-10 | 1.576130e-06 | 6.906285e-58 | 1.140869e-86 | 1.140869e-86 |
| SA | 2.450694e-33 | 8.797239e-10 | 1.000000e+00 | 1.000000e+00 | 2.908429e-95 | 6.007568e-125 | 6.007568e-125 |
| TA | 1.154589e-27 | 1.576130e-06 | 1.000000e+00 | 1.000000e+00 | 8.805129e-88 | 2.114244e-117 | 2.114244e-117 |
| PSO | 2.426988e-27. | 6.906285e-58 | 2.908429e-95 | 8.805129e-88 | 1.000000e+00 | 5.618861e-06 | 5.618861e-06 |
| WO | 3.522396e-52 | 1.140869e-86 | 1.017568e-225 | 4.214244e-127 | 4.618561e-16 | 1.000000e+00 | 1.000000e+00 |
| GWO | 3.522396e-52. | 1.140869e-86 | 6.007568e-125 | 2.114244e-117 | 5.618861e-06 | 1.000000e+00 | 1.000000e+00 |

Statistical significance results for precision, non-continuous xin-she-yang-n2 function, 2 dimensions

| | BSO | BPO | SA | TA | PSO | WO | GWO |
|---|---|---|---|---|---|---|---|
| BSO | 1.000000e+00 | 2.345611e-145 | 1.000000e+00 | 2.024910e-68 | 6.499584e-121 | 1.445260e-108 | 1.923929e-224 |
| BPO | 2.345611e-145 | 1.000000e+00 | 6.592221e-146 | 3.934683e-36 | 8.162397e-286 | 7.262676e-277 | 0.000000e+00 |
| SA | 1.000000e+00 | 6.592221e-146 | 1.000000e+00 | 5.751986e-69 | 2.385403e-120 | 5.337561e-108 | 5.740823e-224 |
| TA | 2.024910e-68 | 3.934683e-36 | 5.751986e-69 | 1.000000e+00 | 9.723701e-226 | 2.689450e-215 | 1.684807e-309 |
| PSO | 6.499584e-121 | 8.162397e-286 | 2.385403e-120 | 9.723701e-226 | 1.000000e+00 | 6.503442e-01 | 6.282261e-67 |
| WO | 1.445260e-108 | 7.262676e-277 | 5.337561e-108 | 2.689450e-215 | 6.503442e-01 | 1.000000e+00 | 5.929343e-79 |
| GWO | 1.923929e-224 | 0.000000e+00 | 5.740823e-224 | 1.684807e-309 | 6.282261e-67 | 5.929343e-79 | 1.000000e+00 |

Statistical significance results for precision, non-continuous xin-she-yang-n2 function, 3 dimensions

| | BSO | BPO | SA | TA | PSO | WO | GWO |
|---|---|---|---|---|---|---|---|
| BSO | 1.000000e+00 | 3.645071e-67 | 1.111801e-182 | 4.080101e-235 | 5.607254e-61 | 1.037647e-164 | 1.226698e-255 |
| BPO | 3.645071e-67 | 1.000000e+00 | 2.872313e-73 | 1.798941e-133 | 7.998262e-171 | 1.215943e-260 | 0.000000e+00 |
| SA | 1.111801e-182 | 2.872313e-73 | 1.000000e+00 | 4.480279e-23 | 2.577872e-270 | 0.000000e+00 | 0.000000e+00 |
| TA | 4.080101e-235 | 1.798941e-133 | 4.480279e-23 | 1.000000e+00 | 6.619216e-313 | 0.000000e+00 | 0.000000e+00 |
| PSO | 5.607254e-61 | 7.998262e-171 | 2.577872e-270 | 6.619216e-313 | 1.000000e+00 | 5.607254e-61 | 1.037647e-164 |
| WO | 1.037647e-164 | 1.215943e-260 | 0.000000e+00 | 0.000000e+00 | 5.607254e-61 | 1.000000e+00 | 5.607254e-61 |
| GWO | 1.226698e-255 | 0.000000e+00 | 0.000000e+00 | 0.000000e+00 | 1.037647e-164 | 5.607254e-61. | 1.000000e+00 |

Statistical significance results for precision, non-continuous xin-she-yang-n2 function, 4 dimensions

| | BSO | BPO | SA | TA | PSO | WO | GWO |
|---|---|---|---|---|---|---|---|
| BSO | 1.000000e+00 | 5.724047e-06 | 2.245737e-27 | 1.153374e-39 | 2.456005e-38 | 6.162373e-12 | 2.611448e-32 |
| BPO | 5.724047e-06 | 1.000000e+00 | 4.686359e-09 | 1.449314e-17 | 1.384822e-16 | 9.249377e-32 | 2.604099e-12 |
| SA | 2.245737e-27 | 4.686359e-09. | 1.000000e+00 | 1.499732e-01 | 3.300413e-01 | 3.339770e-64 | 1.000000e+00 |
| TA | 1.153374e-39 | 1.449314e-17 | 1.499732e-01 | 1.000000e+00 | 1.000000e+00 | 3.631794e-79 | 1.000000e+00 |
| PSO | 2.456005e-38 | 1.384822e-16 | 3.300413e-01 | 1.000000e+00 | 1.000000e+00 | 1.304956e-77 | 1.000000e+00 |
| WO | 6.162373e-12 | 9.249377e-32 | 3.339770e-64 | 3.631794e-79 | 1.304956e-77 | 1.000000e+00 | 2.224365e-70 |
| GWO | 2.611448e-32 | 2.604099e-12 | 1.000000e+00 | 1.000000e+00 | 1.000000e+00 | 2.224365e-70 | 1.000000e+00 |

Statistical significance results for precision, non-continuous xin-she-yang-n2 function, 5 dimensions

| | BSO | BPO | SA | TA | PSO | WO | GWO |
|---|---|---|---|---|---|---|---|
| BSO | 1.000000e+00 | 5.555661e-27 | 2.711193e-70 | 9.086074e-106 | 3.024465e-69 | 2.759058e-16 | 3.360242e-18 |
| BPO | 5.555661e-27 | 1.000000e+00 | 7.847012e-16 | 1.842828e-42 | 3.464842e-15 | 5.070821e-71 | 5.528733e-01 |
| SA | 2.711193e-70 | 7.847012e-16 | 1.000000e+00 | 1.411705e-08 | 1.000000e+00 | 3.855398e-120 | 3.359450e-24 |
| TA | 9.086074e-106 | 1.842828e-42 | 1.411705e-08 | 1.000000e+00 | 4.403679e-09 | 3.127062e-155. | 8.502844e-54 |
| PSO | 3.024465e-69 | 3.464842e-15 | 1.000000e+00 | 4.403679e-09 | 1.000000e+00 | 4.635580e-119 | 1.960125e-23 |
| WO | 2.759058e-16 | 5.070821e-71 | 3.855398e-120 | 3.127062e-155 | 4.635580e-119 | 1.000000e+00 | 7.039412e-59 |
| GWO | 3.360242e-18 | 5.528733e-01 | 3.359450e-24 | 8.502844e-54 | 1.960125e-23 | 7.039412e-59 | 1.000000e+00 |

Statistical significance results for precision, non-continuous xin-she-yang-n2 function, 6 dimensions

| | BSO | BPO | SA | TA | PSO | WO | GWO |
|---|---|---|---|---|---|---|---|
| BSO | 1.000000e+00. | 4.627729e-18 | 2.026243e-36 | 2.331975e-64 | 6.807921e-76 | 2.466879e-04 | 1.000000e+00 |
| BPO | 4.627729e-18 | 1.000000e+00 | 2.958979e-04 | 4.158324e-20 | 1.388374e-28 | 3.028699e-05 | 2.138242e-14 |
| SA | 2.026243e-36 | 2.958979e-04 | 1.000000e+00 | 1.436053e-06 | 3.081761e-12 | 6.491635e-18 | 1.454522e-31 |
| TA | 2.331975e-64 | 4.158324e-20 | 1.436053e-06 | 1.000000e+00 | 7.835147e-01 | 2.090845e-41 | 1.207080e-58 |
| PSO | 6.807921e-76 | 1.388374e-28 | 3.081761e-12 | 7.835147e-01 | 1.000000e+00 | 6.057378e-52 | 5.262196e-70 |
| WO | 2.466879e-04 | 3.028699e-05 | 6.491635e-18 | 2.090845e-41 | 6.057378e-52 | 1.000000e+00 | 1.749470e-02 |
| GWO | 1.000000e+00 | 2.138242e-14 | 1.454522e-31 | 1.207080e-58 | 5.262196e-70 | 1.749470e-02 | 1.000000e+00 |

Statistical significance results for precision, non-continuous ellipsoid function, 2 dimensions

| | BSO | BPO | SA | TA | PSO | WO | GWO |
|---|---|---|---|---|---|---|---|
| BSO | 1.000000e+00 | 3.393659e-85 | 7.207226e-123 | 9.117357e-83 | 3.549168e-31 | 6.581341e-133 | 6.581341e-133 |
| BPO | 3.393659e-85 | 1.000000e+00 | 1.733185e-09 | 1.000000e+00 | 6.448960e-156 | 1.960938e-249 | 1.960938e-249 |
| SA | 7.207226e-123 | 1.733185e-09 | 1.000000e+00 | 1.058409e-10 | 5.020202e-191 | 4.728371e-278 | 4.728371e-278 |
| TA | 9.117357e-83 | 1.000000e+00 | 1.058409e-10 | 1.000000e+00 | 1.462410e-153 | 1.692888e-247 | 1.692888e-247 |
| PSO | 3.549168e-31 | 6.448960e-156 | 5.020202e-191 | 1.462410e-153 | 1.000000e+00 | 3.359524e-62 | 3.359524e-62 |
| WO | 6.581341e-133 | 1.960938e-249 | 4.728371e-278 | 1.692888e-247 | 3.359524e-62 | 1.000000e+00 | 1.000000e+00 |
| GWO | 6.581341e-133 | 1.960938e-249 | 4.728371e-278 | 1.692888e-247 | 3.359524e-62 | 1.000000e+00 | 1.000000e+00 |

Statistical significance results for precision, non-continuous ellipsoid function, 3 dimensions

| | BSO | BPO | SA | TA | PSO | WO | GWO |
|---|---|---|---|---|---|---|---|
| BSO | 1.000000e+00 | 2.283249e-63 | 3.515786e-193 | 3.713706e-206 | 8.193803e-57 | 8.675766e-203 | 8.675766e-203 |
| BPO | 2.283249e-63 | 1.000000e+00 | 1.318025e-88 | 2.943588e-103 | 6.890282e-163 | 1.856466e-288 | 1.856466e-288 |
| SA | 3.515786e-193 | 1.318025e-88 | 1.000000e+00 | 2.210189e-01 | 1.338888e-275 | 0.000000e+00 | 0.000000e+00 |
| TA | 3.713706e-206 | 2.943588e-103 | 2.210189e-01 | 1.000000e+00 | 3.135790e-286 | 0.000000e+00 | 0.000000e+00 |
| PSO | 8.193803e-57 | 6.890282e-163 | 1.338888e-275 | 3.135790e-286 | 1.000000e+00 | 2.175487e-106 | 2.175487e-106 |
| WO | 8.675766e-203 | 1.856466e-288 | 0.000000e+00 | 0.000000e+00 | 2.175487e-106 | 1.000000e+00 | 1.000000e+00 |
| GWO | 8.675766e-203 | 1.856466e-288 | 0.000000e+00 | 0.000000e+00 | 2.175487e-106 | 1.000000e+00 | 1.000000e+00 |

Statistical significance results for precision, non-continuous ellipsoid function, 4 dimensions

| | BSO | BPO | SA | TA | PSO | WO | GWO |
|---|---|---|---|---|---|---|---|
| BSO | 1.000000e+00 | 1.776648e-75 | 1.829474e-224 | 5.504386e-259 | 2.689198e-191 | 2.689198e-191 | 2.689198e-191 |
| BPO | 1.776648e-75 | 1.000000e+00 | 3.874428e-112 | 1.756447e-154 | 3.409193e-288 | 3.409193e-288 | 3.409193e-288 |
| SA | 1.829474e-224 | 3.874428e-112 | 1.000000e+00 | 3.428101e-12 | 0.000000e+00 | 0.000000e+00 | 0.000000e+00 |
| TA | 5.504386e-259 | 1.756447e-154 | 3.428101e-12 | 1.000000e+00 | 0.000000e+00 | 0.000000e+00 | 0.000000e+00 |
| PSO | 2.689198e-191 | 3.409193e-288 | 0.000000e+00 | 0.000000e+00 | 1.000000e+00 | 1.000000e+00 | 1.000000e+00 |
| WO | 2.689198e-191 | 3.409193e-288 | 0.000000e+00 | 0.000000e+00 | 1.000000e+00 | 1.000000e+00 | 1.000000e+00 |
| GWO | 2.689198e-191 | 3.409193e-288 | 0.000000e+00 | 0.000000e+00 | 1.000000e+00 | 1.000000e+00 | 1.000000e+00 |

Statistical significance results for precision, non-continuous ellipsoid function, 5 dimensions

| | BSO | BPO | SA | TA | PSO | WO | GWO |
|---|---|---|---|---|---|---|---|
| BSO | 1.000000e+00 | 5.619532e-34 | 3.437852e-115 | 9.350953e-168 | 3.429193e-82 | 9.314998e-103 | 9.314998e-103 |
| BPO | 5.619532e-34 | 1.000000e+00 | 8.914571e-43 | 4.414015e-94 | 2.125207e-156 | 7.376985e-176 | 7.376985e-176 |
| SA | 3.437852e-115 | 8.914571e-43 | 1.000000e+00 | 1.234124e-18 | 1.179803e-232 | 2.999011e-249 | 2.999011e-249 |
| TA | 9.350953e-168 | 4.414015e-94 | 1.234124e-18 | 1.000000e+00 | 1.245472e-274 | 1.537787e-289 | 1.537787e-289 |
| PSO | 3.429193e-82 | 2.125207e-156 | 1.179803e-232 | 1.245472e-274 | 1.000000e+00 | 6.919831e-03 | 6.919831e-03 |
| WO | 9.314998e-103 | 7.376985e-176 | 2.999011e-249 | 1.537787e-289 | 6.919831e-03 | 1.000000e+00 | 1.000000e+00 |
| GWO | 9.314998e-103 | 7.376985e-176 | 2.999011e-249 | 1.537787e-289 | 6.919831e-03 | 1.000000e+00 | 1.000000e+00 |

Statistical significance results for precision, non-continuous ellipsoid function, 6 dimensions

| | BSO | BPO | SA | TA | PSO | WO | GWO |
|---|---|---|---|---|---|---|---|
| BSO | 1.000000e+00 | 5.079816e-48 | 3.706754e-144 | 1.180074e-215 | 3.384105e-121 | 3.736683e-136 | 3.736683e-136 |
| BPO | 5.079816e-48 | 1.000000e+00 | 3.517689e-54 | 8.424189e-131 | 2.617498e-207 | 2.869495e-220 | 2.869495e-220 |
| SA | 3.706754e-144 | 3.517689e-54 | 1.000000e+00 | 2.388574e-36 | 3.887171e-285 | 8.346199e-296 | 8.346199e-296 |
| TA | 1.180074e-215 | 8.424189e-131 | 2.388574e-36 | 1.000000e+00 | 0.000000e+00 | 0.000000e+00 | 0.000000e+00 |
| PSO | 3.384105e-121 | 2.617498e-207 | 3.887171e-285 | 0.000000e+00 | 1.000000e+00 | 1.768127e-01 | 1.768127e-01 |
| WO | 3.736683e-136 | 2.869495e-220 | 8.346199e-296 | 0.000000e+00 | 1.768127e-01 | 1.000000e+00 | 1.000000e+00 |
| GWO | 3.736683e-136 | 2.869495e-220 | 8.346199e-296 | 0.000000e+00 | 1.768127e-01 | 1.000000e+00 | 1.000000e+00 |

Statistical significance results for precision, non-continuous step_fun function, 2 dimensions

| | BSO | BPO | SA | TA | PSO | WO | GWO |
|---|---|---|---|---|---|---|---|
| BSO | 1.000000e+00 | 1.000000e+00 | 1.000000e+00 | 1.000000e+00 | 1.000000e+00 | 1.000000e+00 | 4.360001e-36 |
| BPO | 1.000000e+00 | 1.000000e+00 | 1.000000e+00 | 1.000000e+00 | 1.000000e+00 | 1.000000e+00 | 2.138795e-41 |
| SA | 1.000000e+00 | 1.000000e+00 | 1.000000e+00 | 1.000000e+00 | 1.000000e+00 | 1.000000e+00 | 2.749514e-37 |
| TA | 1.000000e+00 | 1.000000e+00 | 1.000000e+00 | 1.000000e+00 | 1.000000e+00 | 1.000000e+00 | 2.818508e-38 |
| PSO | 1.000000e+00 | 1.000000e+00 | 1.000000e+00 | 1.000000e+00 | 1.000000e+00 | 1.000000e+00 | 1.168653e-37 |
| WO | 1.000000e+00 | 1.000000e+00 | 1.000000e+00 | 1.000000e+00 | 1.000000e+00 | 1.000000e+00 | 6.280058e-35 |
| GWO | 4.360001e-36 | 2.138795e-41 | 2.749514e-37 | 2.818508e-38 | 1.168653e-37 | 6.280058e-35 | 1.000000e+00 |

Statistical significance results for precision, non-continuous step_fun function, 4 dimensions

| | BSO | BPO | SA | TA | PSO | WO | GWO |
|---|---|---|---|---|---|---|---|
| BSO | 1.000000 | 0.907425 | 0.000422 | 0.000422 | 0.608156 | 1.000000 | 0.632664 |
| BPO | 0.907425 | 1.000000 | 0.495933 | 0.495933 | 1.000000 | 1.000000 | 1.000000 |
| SA | 0.000422 | 0.495933 | 1.000000 | 1.000000 | 0.748431 | 0.253814 | 0.720107 |
| TA | 0.000422 | 0.495933 | 1.000000 | 1.000000 | 0.748431 | 0.253814 | 0.720107 |
| PSO | 0.608156 | 1.000000 | 0.748431 | 0.748431 | 1.000000 | 1.000000 | 1.000000 |
| WO | 1.000000 | 1.000000 | 0.253814 | 0.253814 | 1.000000 | 1.000000 | 1.000000 |
| GWO | 0.632664 | 1.000000 | 0.720107 | 0.720107 | 1.000000 | 1.000000 | 1.000000 |

Statistical significance results for precision, non-continuous step_fun function, 5 dimensions

| | BSO | BPO | SA | TA | PSO | WO | GWO |
|---|---|---|---|---|---|---|---|
| BSO | 1.000000e+00 | 6.875233e-01 | 4.223214e-04 | 2.194745e-01 | 3.374803e-01 | 4.521561e-01 | 4.364831e-52 |
| BPO | 6.875233e-01 | 1.000000e+00 | 6.636303e-01 | 1.000000e+00 | 1.000000e+00 | 1.000000e+00 | 2.795677e-41 |
| SA | 4.223214e-04 | 6.636303e-01 | 1.000000e+00 | 1.000000e+00 | 1.000000e+00 | 9.876232e-01 | 3.360173e-31 |
| TA | 2.194745e-01 | 1.000000e+00 | 1.000000e+00 | 1.000000e+00 | 1.000000e+00 | 1.000000e+00 | 3.311983e-39 |
| PSO | 3.374803e-01 | 1.000000e+00 | 1.000000e+00 | 1.000000e+00 | 1.000000e+00 | 1.000000e+00 | 5.973070e-40 |
| WO | 4.521561e-01 | 1.000000e+00 | 9.876232e-01 | 1.000000e+00 | 1.000000e+00 | 1.000000e+00 | 1.761530e-40 |
| GWO | 4.364831e-52 | 2.795677e-41 | 3.360173e-31 | 3.311983e-39 | 5.973070e-40 | 1.761530e-40 | 1.000000e+00 |

Statistical significance results for precision, non-continuous step_fun function, 6 dimensions

| | BSO | BPO | SA | TA | PSO | WO | GWO |
|---|---|---|---|---|---|---|---|
| BSO | 1.000000 | 0.020562 | 1.000000 | 1.000000 | 0.165783 | 0.814733 | 0.010014 |
| BPO | 0.020562 | 1.000000 | 0.062637 | 0.166756 | 1.000000 | 1.000000 | 1.000000 |
| SA | 1.000000 | 0.062637 | 1.000000 | 1.000000 | 0.417776 | 1.000000 | 0.032351 |
| TA | 1.000000 | 0.166756 | 1.000000 | 1.000000 | 0.927927 | 1.000000 | 0.091120 |
| PSO | 0.165783 | 1.000000 | 0.417776 | 0.927927 | 1.000000 | 1.000000 | 1.000000 |
| WO | 0.814733 | 1.000000 | 1.000000 | 1.000000 | 1.000000 | 1.000000 | 1.000000 |
| GWO | 0.010014 | 1.000000 | 0.032351 | 0.091120 | 1.000000 | 1.000000 | 1.000000 |

Statistical significance results for precision, non-continuous rosenbrock function, 2 dimensions

| | BSO | BPO | SA | TA | PSO | WO | GWO |
|---|---|---|---|---|---|---|---|
| BSO | 1.000000 | 3.408026e-04 | 1.000000e+00 | 3.408026e-04 | 3.408026e-04 | 3.408026e-04 | 3.408026e-04 |
| BPO | 0.000341 | 1.000000e+00 | 4.232395e-08 | 1.000000e+00 | 1.000000e+00 | 1.000000e+00 | 1.000000e+00 |
| SA | 1.000000 | 4.232395e-08 | 1.000000e+00 | 4.232395e-08 | 4.232395e-08 | 4.232395e-08 | 4.232395e-08 |
| TA | 0.000341 | 1.000000e+00 | 4.232395e-08 | 1.000000e+00 | 1.000000e+00 | 1.000000e+00 | 1.000000e+00 |
| PSO | 0.000341 | 1.000000e+00 | 4.232395e-08 | 1.000000e+00 | 1.000000e+00 | 1.000000e+00 | 1.000000e+00 |
| WO | 0.000341 | 1.000000e+00 | 4.232395e-08 | 1.000000e+00 | 1.000000e+00 | 1.000000e+00 | 1.000000e+00 |
| GWO | 0.000341 | 1.000000e+00 | 4.232395e-08 | 1.000000e+00 | 1.000000e+00 | 1.000000e+00 | 1.000000e+00 |

Statistical significance results for precision, non-continuous rosenbrock function, 3 dimensions

| | BSO | BPO | SA | TA | PSO | WO | GWO |
|---|---|---|---|---|---|---|---|
| BSO | 1.000000e+00 | 9.222658e-07 | 1.000000e+00 | 1.000000e+00 | 1.000000e+00 | 1.000000e+00 | 1.000000e+00 |
| BPO | 9.222658e-07 | 1.000000e+00 | 9.222658e-07 | 9.222658e-07 | 9.222658e-07 | 9.222658e-07 | 9.222658e-07 |
| SA | 1.000000e+00 | 9.222658e-07 | 1.000000e+00 | 1.000000e+00 | 1.000000e+00 | 1.000000e+00 | 1.000000e+00 |
| TA | 1.000000e+00 | 9.222658e-07 | 1.000000e+00 | 1.000000e+00 | 1.000000e+00 | 1.000000e+00 | 1.000000e+00 |
| PSO | 1.000000e+00 | 9.222658e-07 | 1.000000e+00 | 1.000000e+00 | 1.000000e+00 | 1.000000e+00 | 1.000000e+00 |
| WO | 1.000000e+00 | 9.222658e-07 | 1.000000e+00 | 1.000000e+00 | 1.000000e+00 | 1.000000e+00 | 1.000000e+00 |
| GWO | 1.000000e+00 | 9.222658e-07 | 1.000000e+00 | 1.000000e+00 | 1.000000e+00 | 1.000000e+00 | 1.000000e+00 |

Statistical significance results for precision, non-continuous rosenbrock function, 4 dimensions

| | BSO | BPO | SA | TA | PSO | WO | GWO |
|---|---|---|---|---|---|---|---|
| BSO | 1.000000 | 1.000000 | 1.000000 | 0.011504 | 0.000001 | 0.759911 | 1.000000 |
| BPO | 1.000000 | 1.000000 | 1.000000 | 0.011504 | 0.000001 | 0.759911 | 1.000000 |
| SA | 1.000000 | 1.000000 | 1.000000 | 0.011504 | 0.000001 | 0.759911 | 1.000000 |
| TA | 0.011504 | 0.011504 | 0.011504 | 1.000000 | 0.993455 | 1.000000 | 0.011504 |
| PSO | 0.000001 | 0.000001 | 0.000001 | 0.993455 | 1.000000 | 0.017229 | 0.000001 |
| WO | 0.759911 | 0.759911 | 0.759911 | 1.000000 | 0.017229 | 1.000000 | 0.759911 |
| GWO | 1.000000 | 1.000000 | 1.000000 | 0.011504 | 0.000001 | 0.759911 | 1.000000 |

Statistical significance results for precision, non-continuous rosenbrock function, 5 dimensions

| | BSO | BPO | SA | TA | PSO | WO | GWO |
|---|---|---|---|---|---|---|---|
| BSO | 1.000000e+00 | 7.107706e-02 | 1.187629e-22 | 1.000000e+00 | 1.952782e-03 | 1.000000e+00 | 1.000000e+00 |
| BPO | 7.107706e-02 | 1.000000e+00 | 2.264065e-35 | 4.429379e-02 | 2.968178e-10 | 1.000000e+00 | 1.000000e+00 |
| SA | 1.187629e-22 | 2.264065e-35 | 1.000000e+00 | 4.450430e-22 | 2.303644e-09 | 1.237048e-30 | 9.963919e-27 |
| TA | 1.000000e+00 | 4.429379e-02 | 4.450430e-22 | 1.000000e+00 | 3.495703e-03 | 8.690012e-01 | 1.000000e+00 |
| PSO | 1.952782e-03 | 2.968178e-10 | 2.303644e-09 | 3.495703e-03 | 1.000000e+00 | 1.791160e-07 | 2.197914e-05 |
| WO | 1.000000e+00 | 1.000000e+00 | 1.237048e-30 | 8.690012e-01 | 1.791160e-07 | 1.000000e+00 | 1.000000e+00 |
| GWO | 1.000000e+00 | 1.000000e+00 | 9.963919e-27 | 1.000000e+00 | 2.197914e-05 | 1.000000e+00 | 1.000000e+00 |

Statistical significance results for precision, non-continuous rosenbrock function, 6 dimensions

| | BSO | BPO | SA | TA | PSO | WO | GWO |
|---|---|---|---|---|---|---|---|
| BSO | 1.000000e+00 | 8.757386e-03 | 2.740621e-17 | 1.000000e+00 | 3.130487e-14 | 1.292632e-26 | 1.000000e+00 |
| BPO | 8.757386e-03 | 1.000000e+00 | 1.238731e-31 | 1.000000e+00 | 1.014962e-27 | 6.270877e-43 | 9.845599e-06 |
| SA | 2.740621e-17 | 1.238731e-31 | 1.000000e+00 | 2.589672e-23 | 1.000000e+00 | 3.568531e-01 | 3.570324e-12 |
| TA | 1.000000e+00 | 1.000000e+00 | 2.589672e-23 | 1.000000e+00 | 7.704164e-20 | 1.311555e-33 | 3.817845e-02 |
| PSO | 3.130487e-14 | 1.014962e-27 | 1.000000e+00 | 7.704164e-20 | 1.000000e+00 | 2.241624e-02 | 1.448267e-09 |
| WO | 1.292632e-26 | 6.270877e-43 | 3.568531e-01 | 1.311555e-33 | 2.241624e-02 | 1.000000e+00 | 1.987314e-20 |
| GWO | 1.000000e+00 | 9.845599e-06 | 3.570324e-12 | 3.817845e-02 | 1.448267e-09 | 1.987314e-20 | 1.000000e+00 |

Statistical significance results for precision, non-continuous quadric function, 2 dimensions

| | BSO | BPO | SA | TA | PSO | WO | GWO |
|---|---|---|---|---|---|---|---|
| BSO | 1.000000e+00 | 1.634570e-05 | 7.592531e-08 | 1.000000e+00 | 8.977785e-25 | 1.000000e+00 | 1.164754e-02 |
| BPO | 1.634570e-05 | 1.000000e+00 | 1.000000e+00 | 1.634570e-05 | 5.791006e-08 | 1.634570e-05 | 1.000000e+00 |
| SA | 7.592531e-08 | 1.000000e+00 | 1.000000e+00 | 7.592531e-08 | 1.296201e-05 | 7.592531e-08 | 2.590561e-01 |
| TA | 1.000000e+00 | 1.634570e-05 | 7.592531e-08 | 1.000000e+00 | 8.977785e-25 | 1.000000e+00 | 1.164754e-02 |
| PSO | 8.977785e-25 | 5.791006e-08 | 1.296201e-05 | 8.977785e-25 | 1.000000e+00 | 8.977785e-25 | 3.066715e-12 |
| WO | 1.000000e+00 | 1.634570e-05 | 7.592531e-08 | 1.000000e+00 | 8.977785e-25 | 1.000000e+00 | 1.164754e-02 |
| GWO | 1.164754e-02 | 1.000000e+00 | 2.590561e-01 | 1.164754e-02 | 3.066715e-12 | 1.164754e-02 | 1.000000e+00 |

Statistical significance results for precision, non-continuous quadric function, 3 dimensions

| | BSO | BPO | SA | TA | PSO | WO | GWO |
|---|---|---|---|---|---|---|---|
| BSO | 1.000000e+00 | 4.521355e-48 | 1.485177e-135 | 5.962003e-113 | 1.983769e-28 | 1.633378e-119 | 2.222303e-93 |
| BPO | 4.521355e-48 | 1.000000e+00 | 3.267898e-46 | 1.463885e-27 | 4.164519e-114 | 6.943998e-206 | 1.680215e-182 |
| SA | 1.485177e-135 | 3.267898e-46 | 1.000000e+00 | 1.594011e-03 | 2.805365e-199 | 1.493176e-277 | 6.244330e-258 |
| TA | 5.962003e-113 | 1.463885e-27 | 1.594011e-03 | 1.000000e+00 | 1.071480e-178 | 1.601543e-260 | 5.446107e-240 |
| PSO | 1.983769e-28 | 4.164519e-114 | 2.805365e-199 | 1.071480e-178 | 1.000000e+00 | 5.818799e-53 | 1.035028e-30 |
| WO | 1.633378e-119 | 6.943998e-206 | 1.493176e-277 | 1.601543e-260 | 5.818799e-53 | 1.000000e+00 | 1.206821e-04 |
| GWO | 2.222303e-93 | 1.680215e-182 | 6.244330e-258 | 5.446107e-240 | 1.035028e-30 | 1.206821e-04 | 1.000000e+00 |

Statistical significance results for precision, non-continuous quadric function, 4 dimensions

| | BSO | BPO | SA | TA | PSO | WO | GWO |
|---|---|---|---|---|---|---|---|
| BSO | 1.000000e+00 | 1.206360e-61 | 4.605975e-134 | 1.849665e-96 | 1.758408e-41 | 2.436137e-36 | 1.000000e+00 |
| BPO | 1.206360e-61 | 1.000000e+00 | 1.474032e-32 | 1.869680e-08 | 1.151140e-144 | 1.131042e-138 | 9.537838e-61 |
| SA | 4.605975e-134 | 1.474032e-32 | 1.000000e+00 | 1.736304e-09 | 4.530984e-212 | 9.314890e-207 | 3.993310e-133 |
| TA | 1.849665e-96 | 1.869680e-08 | 1.736304e-09 | 1.000000e+00 | 2.112880e-178 | 1.036517e-172 | 1.652349e-95 |
| PSO | 1.758408e-41 | 1.151140e-144 | 4.530984e-212 | 2.112880e-178 | 1.000000e+00 | 1.000000e+00 | 2.685441e-42 |
| WO | 2.436137e-36 | 1.131042e-138 | 9.314890e-207 | 1.036517e-172 | 1.000000e+00 | 1.000000e+00 | 4.008286e-37 |
| GWO | 1.000000e+00 | 9.537838e-61 | 3.993310e-133 | 1.652349e-95 | 2.685441e-42 | 4.008286e-37 | 1.000000e+00 |

Statistical significance results for precision, non-continuous quadric function, 5 dimensions

| | BSO | BPO | SA | TA | PSO | WO | GWO |
|---|---|---|---|---|---|---|---|
| BSO | 1.000000e+00 | 1.925277e-49 | 2.096056e-133 | 1.870197e-96 | 4.688750e-42 | 2.761498e-60 | 2.284698e-04 |
| BPO | 1.925277e-49 | 1.000000e+00 | 5.724248e-43 | 1.995559e-15 | 2.330031e-132 | 4.652434e-152 | 2.762502e-73 |
| SA | 2.096056e-133 | 5.724248e-43 | 1.000000e+00 | 3.603298e-09 | 4.510940e-212 | 4.084970e-229 | 5.019709e-158 |
| TA | 1.870197e-96 | 1.995559e-15 | 3.603298e-09 | 1.000000e+00 | 5.143457e-179 | 2.698543e-197 | 8.752575e-122 |
| PSO | 4.688750e-42 | 2.330031e-132 | 4.510940e-212 | 5.143457e-179 | 1.000000e+00 | 9.317557e-03 | 2.790418e-22 |
| WO | 2.761498e-60 | 4.652434e-152 | 4.084970e-229 | 2.698543e-197 | 9.317557e-03 | 1.000000e+00 | 1.065011e-37 |
| GWO | 2.284698e-04 | 2.762502e-73 | 5.019709e-158 | 8.752575e-122 | 2.790418e-22 | 1.065011e-37 | 1.000000e+00 |

Statistical significance results for precision, non-continuous quadric function, 6 dimensions

| | BSO | BPO | SA | TA | PSO | WO | GWO |
|---|---|---|---|---|---|---|---|
| BSO | 1.000000e+00 | 1.845533e-91 | 2.170875e-219 | 1.602106e-151 | 8.044371e-104 | 1.396072e-108 | 1.000000e+00 |
| BPO | 1.845533e-91 | 1.000000e+00 | 5.950770e-90 | 3.985618e-23 | 5.511832e-231 | 6.928929e-235 | 2.563899e-83 |
| SA | 2.170875e-219 | 5.950770e-90 | 1.000000e+00 | 9.492919e-34 | 0.000000e+00 | 0.000000e+00 | 2.155312e-212 |
| TA | 1.602106e-151 | 3.985618e-23 | 9.492919e-34 | 1.000000e+00 | 4.709313e-278 | 1.642397e-281 | 1.422742e-143 |
| PSO | 8.044371e-104 | 5.511832e-231 | 0.000000e+00 | 4.709313e-278 | 1.000000e+00 | 1.000000e+00 | 5.052448e-112 |
| WO | 1.396072e-108 | 6.928929e-235 | 0.000000e+00 | 1.642397e-281 | 1.000000e+00 | 1.000000e+00 | 8.953564e-117 |
| GWO | 1.000000e+00 | 2.563899e-83 | 2.155312e-212 | 1.422742e-143 | 5.052448e-112 | 8.953564e-117 | 1.000000e+00 |

# APPENDIX I: CASE STUDY 1 QUANTILE-QUANTILE(Q-Q) PLOTS FOR RESULT DATASET

In order to determine if a dataset adheres to a specific theoretical distribution, in our case it is normal distribution, a Q-Q (Quantile-Quantile) plot is utilized, i.e. examining whether a dataset is normally distributed. These Q-Q plots are for Case Study 1. Here, we have all the Q-Q plots for almost all optimization problems{Table 8 and Table 9} in multiple dimensions(2,3,4,5,6)D. which validates that the generated result datasets are not normally distributed. In the following plots, the blue line represents the data set generated via experiments and red line denotes the benchmark normal-distribution line. Hence, a deviation in blue line from red line represents the non-normally distributed data and, the plots are labelled as follows: **metaheuristic name - optimization function name-dimension**. For continuous functions plots we have used the precision results from Table A2 and for non-continuous functions plots we have used the precision results from Table B2

BSO Q-Q plots

Continuous functions

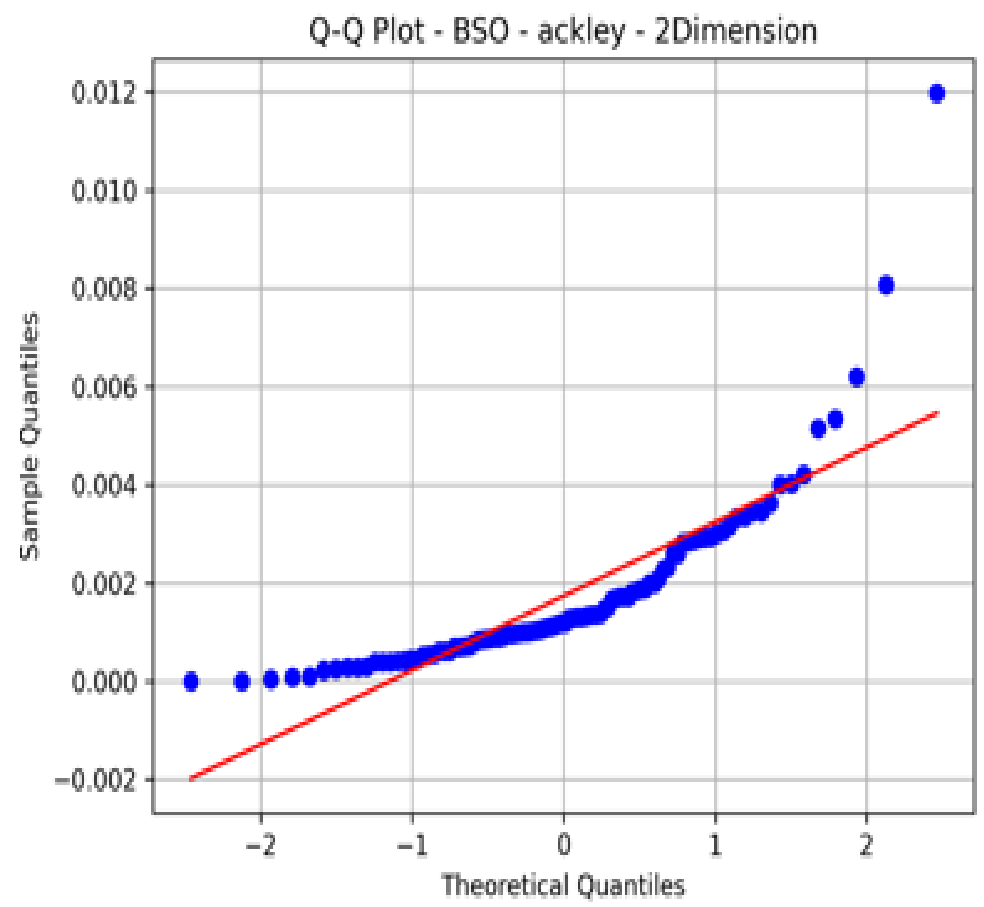
Q-Q Plot - BSO - ackley - 2Dimension
Sample Quantiles
Theoretical Quantiles

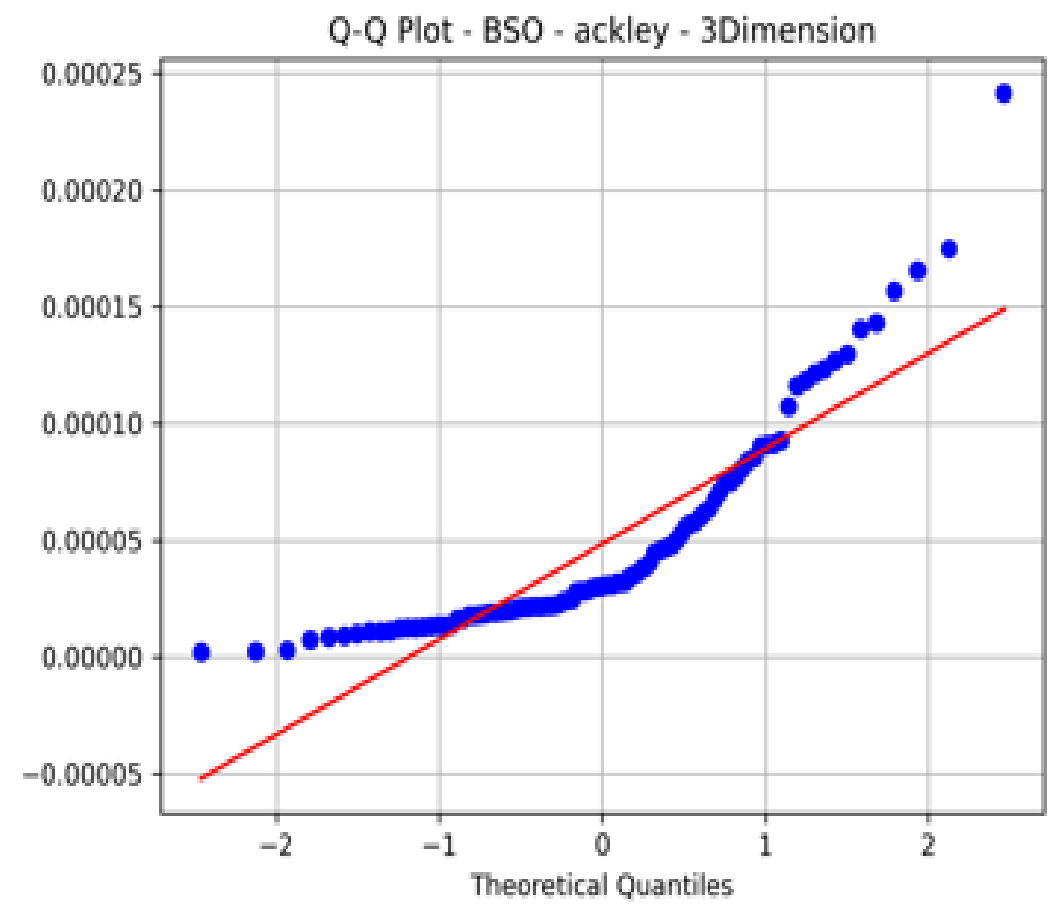
Q-Q Plot - BSO - ackley - 3Dimension
Theoretical Quantiles

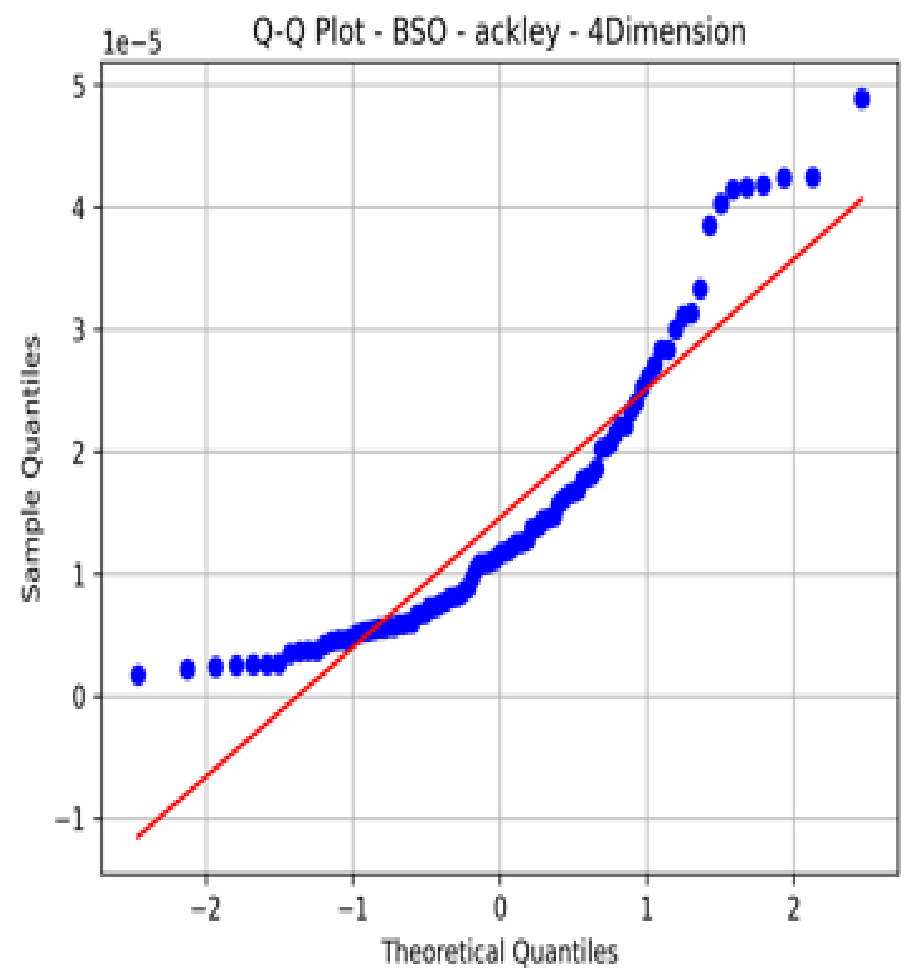
1e−5
Q-Q Plot - BSO - ackley - 4Dimension
Sample Quantiles
Theoretical Quantiles

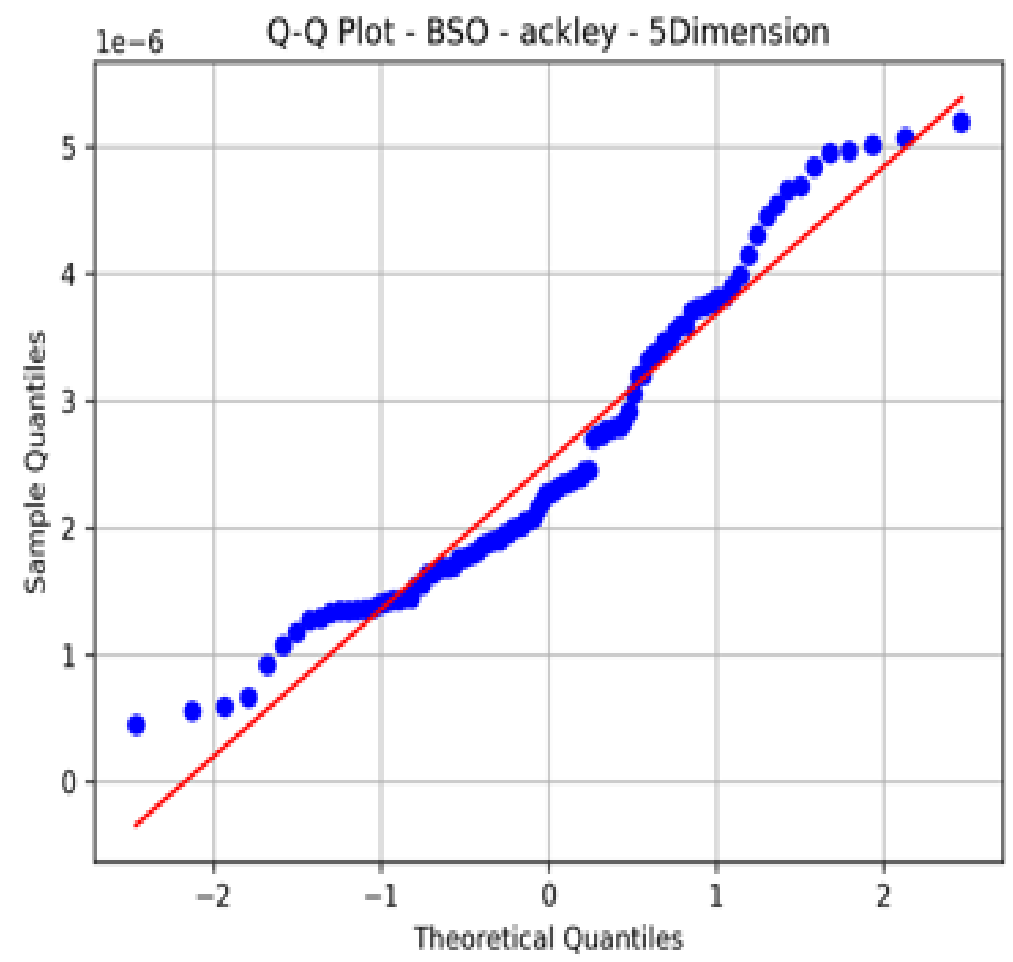
1e−6
Q-Q Plot - BSO - ackley - 5Dimension
Sample Quantiles
Theoretical Quantiles

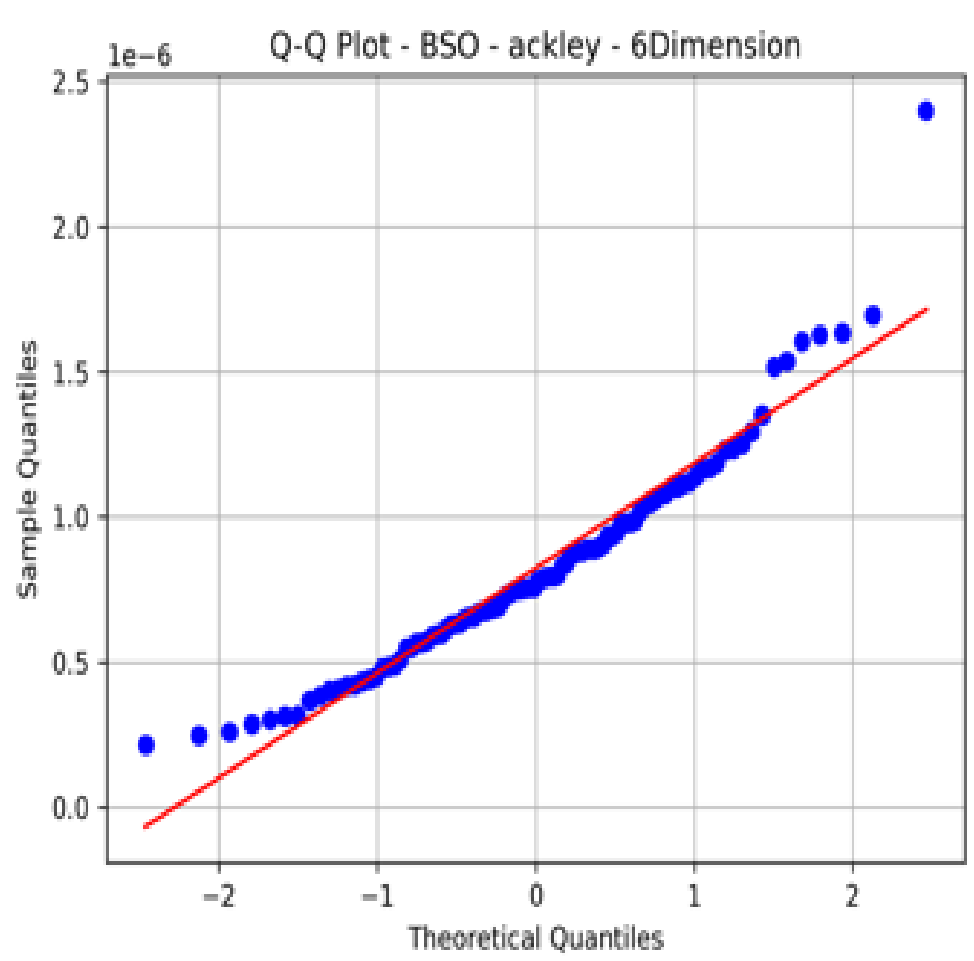
1e−6
Q-Q Plot - BSO - ackley - 6Dimension
Sample Quantiles
Theoretical Quantiles

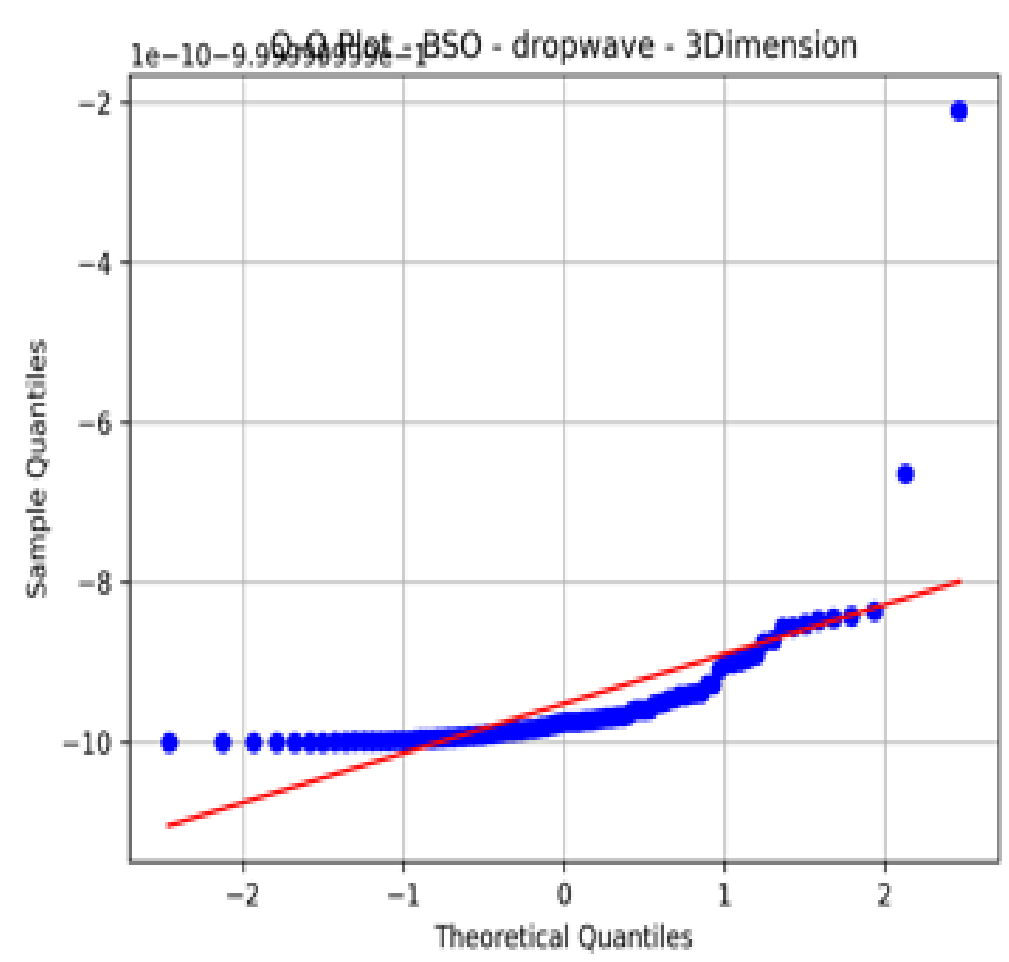
BSO - dropwave - 3Dimension
Sample Quantiles
Theoretical Quantiles

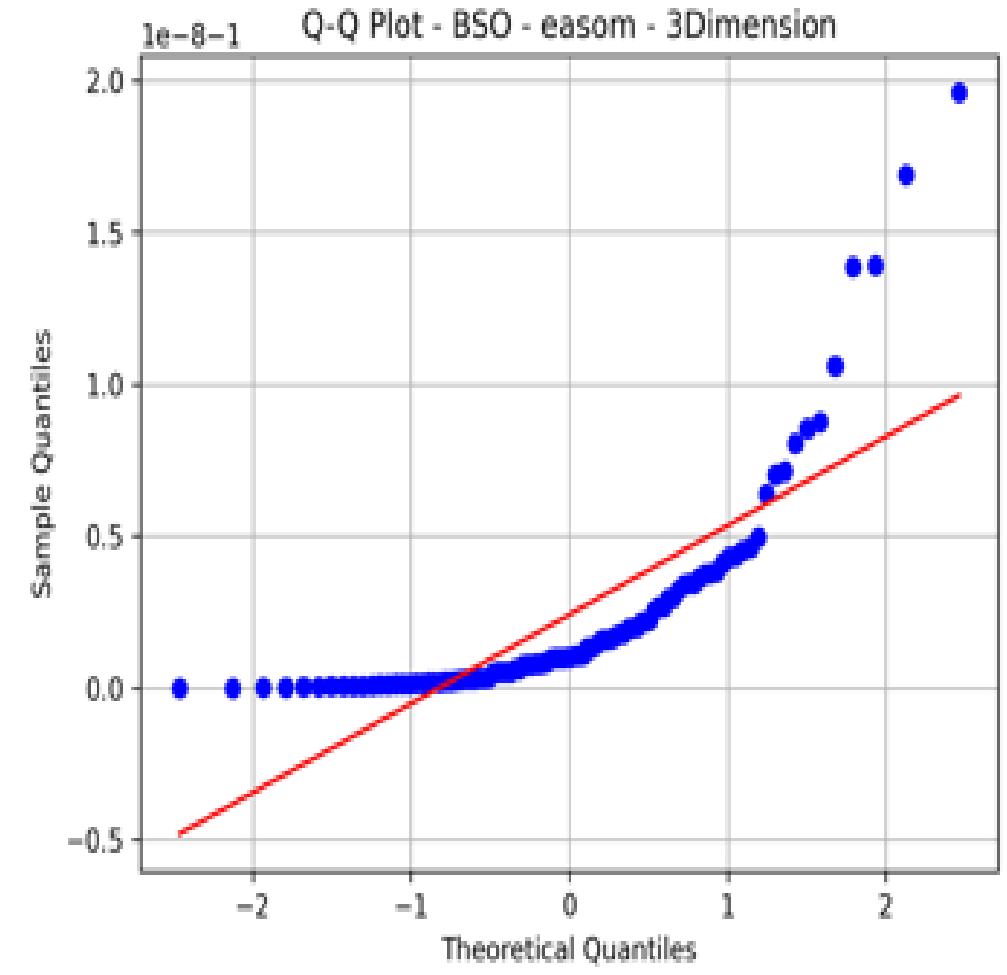

Q-Q Plot - BSO - easom - 3Dimension
1e−8−1
Sample Quantiles
Theoretical Quantiles

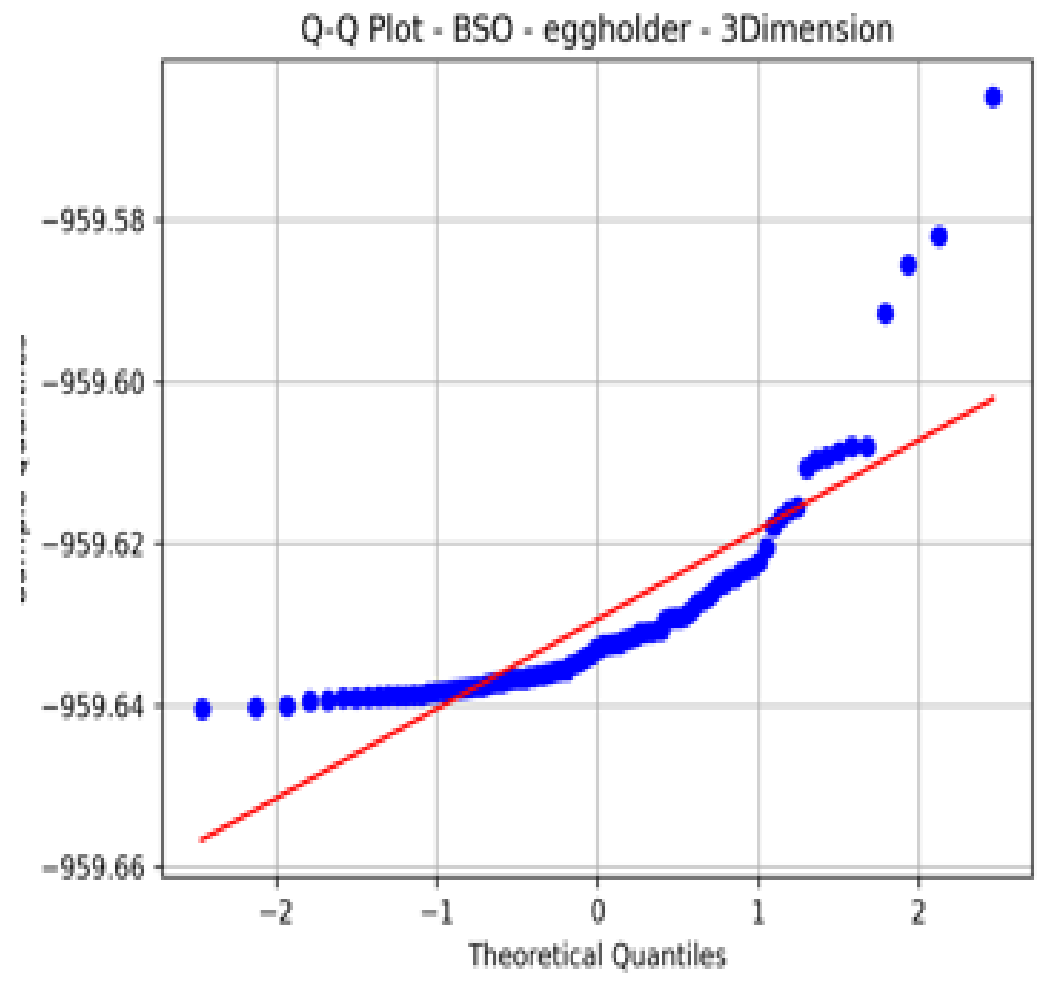

Q-Q Plot - BSO - eggholder - 3Dimension
Theoretical Quantiles

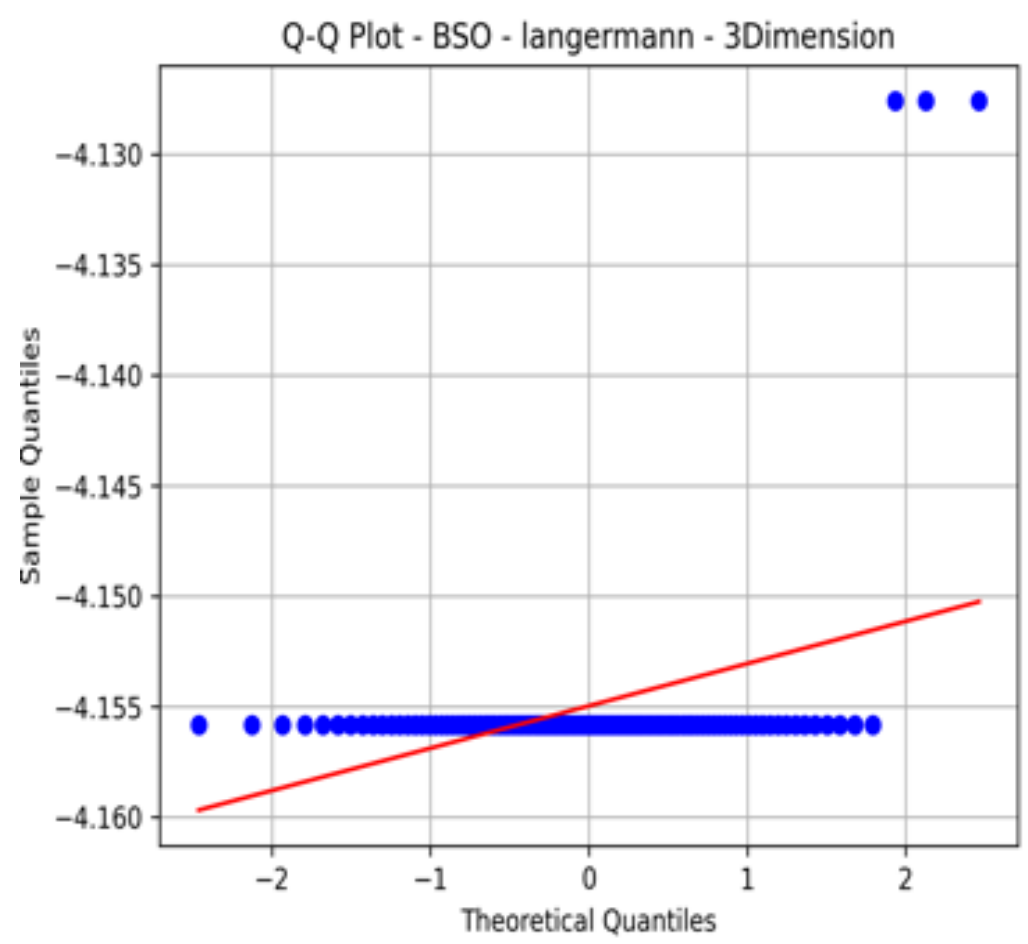

Q-Q Plot - BSO - langermann - 3Dimension
Sample Quantiles
Theoretical Quantiles

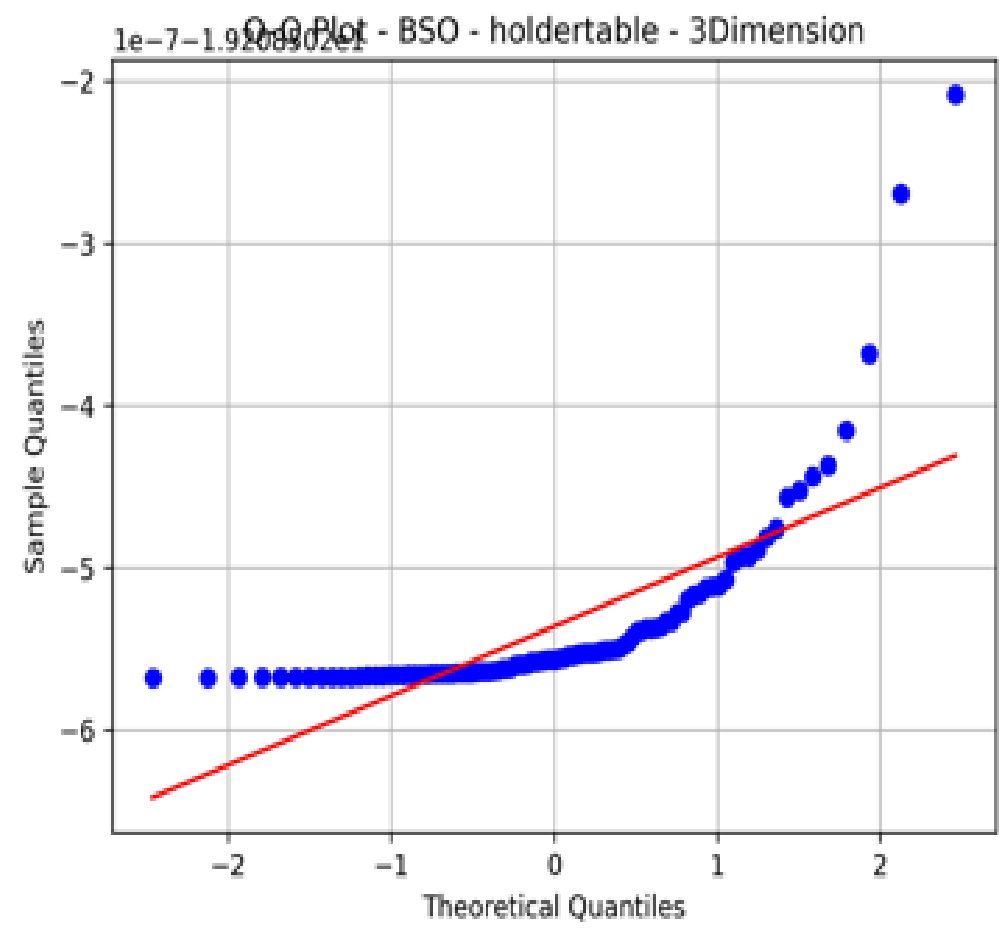

Q-Q Plot - BSO - holdertable - 3Dimension
Sample Quantiles
Theoretical Quantiles

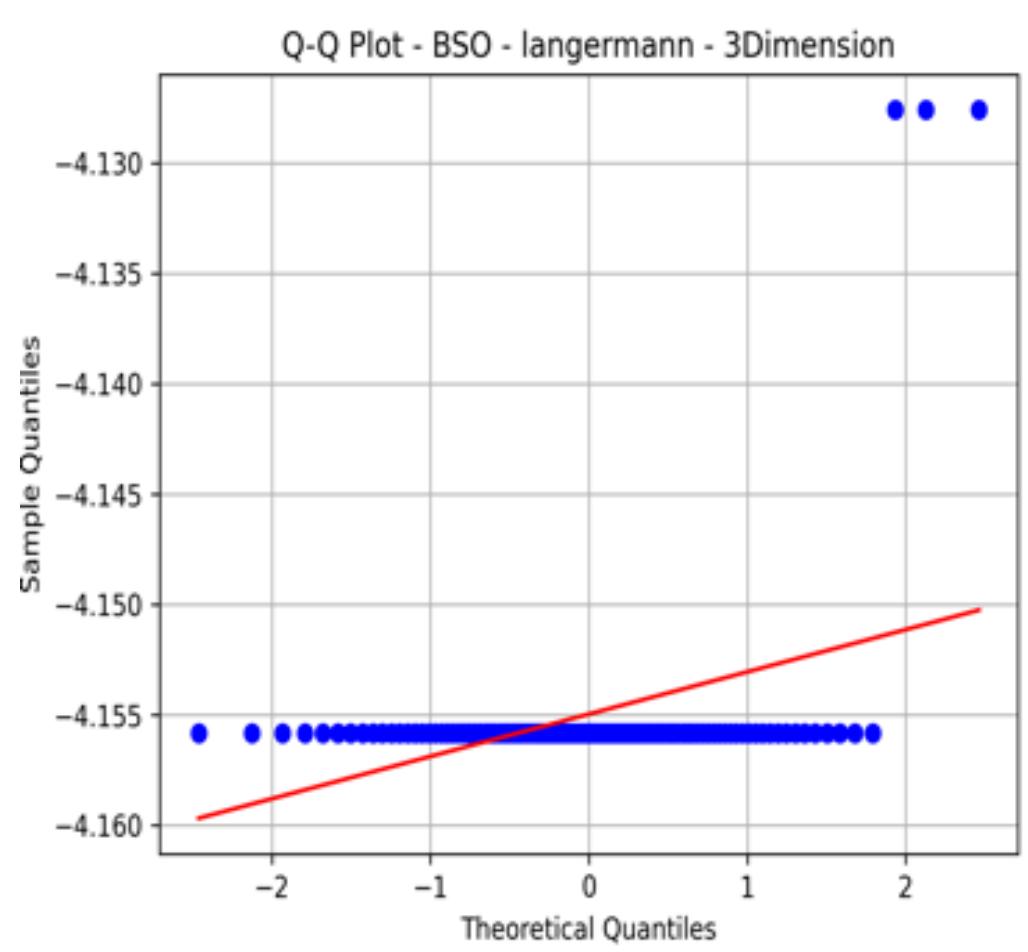

Q-Q Plot - BSO - langermann - 3Dimension
Sample Quantiles
Theoretical Quantiles

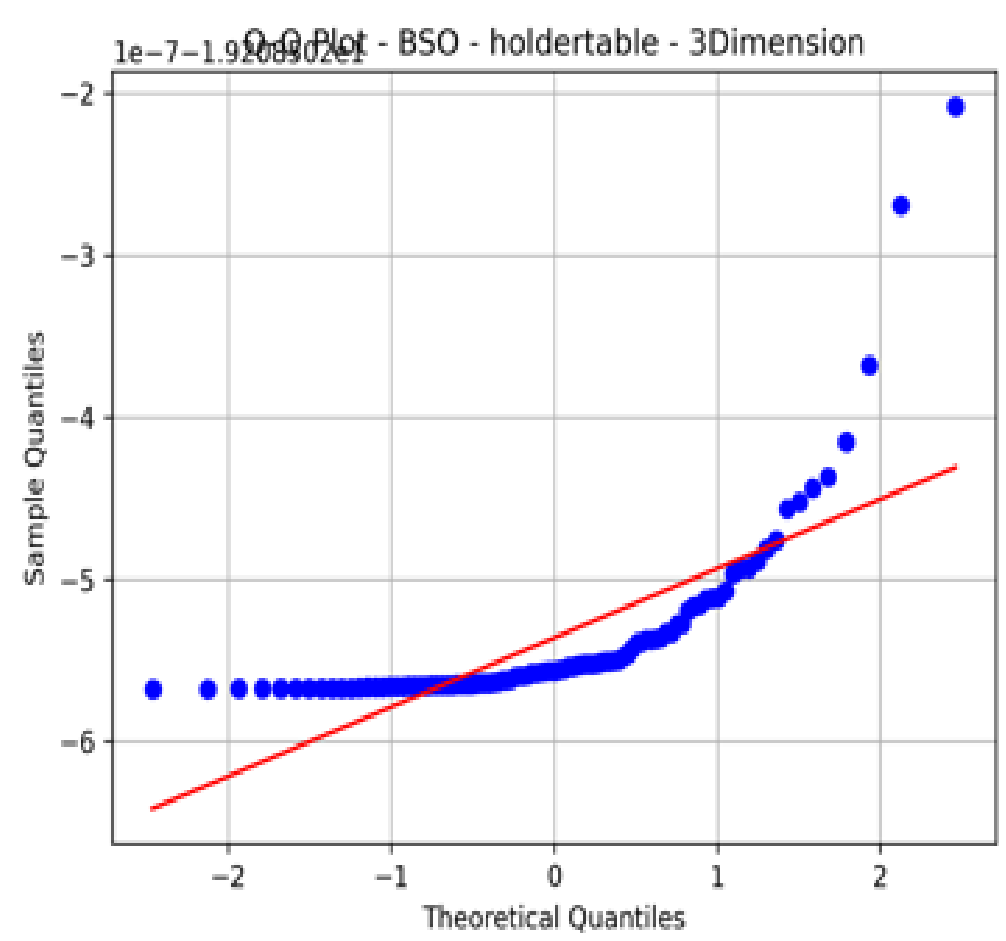

Q-Q Plot - BSO - holdertable - 3Dimension
Sample Quantiles
Theoretical Quantiles

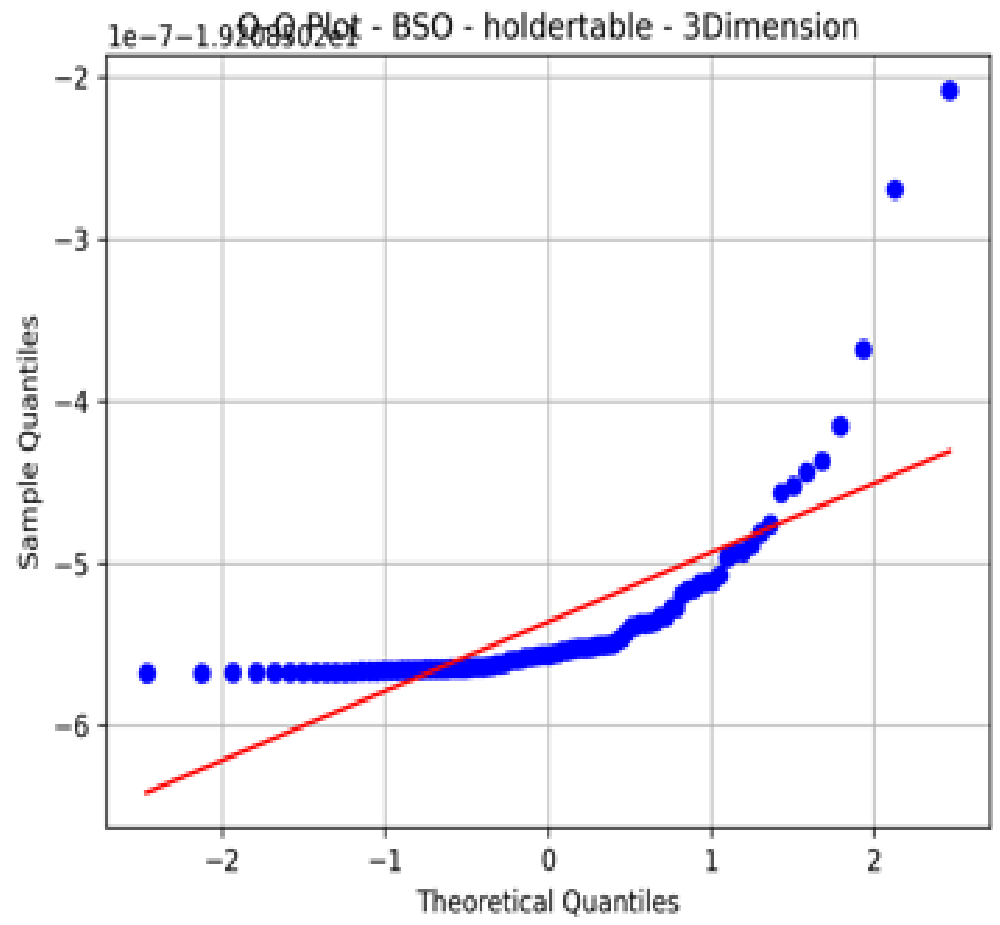
Q-Q Plot - BSO - holdertable - 3Dimension
Sample Quantiles
Theoretical Quantiles

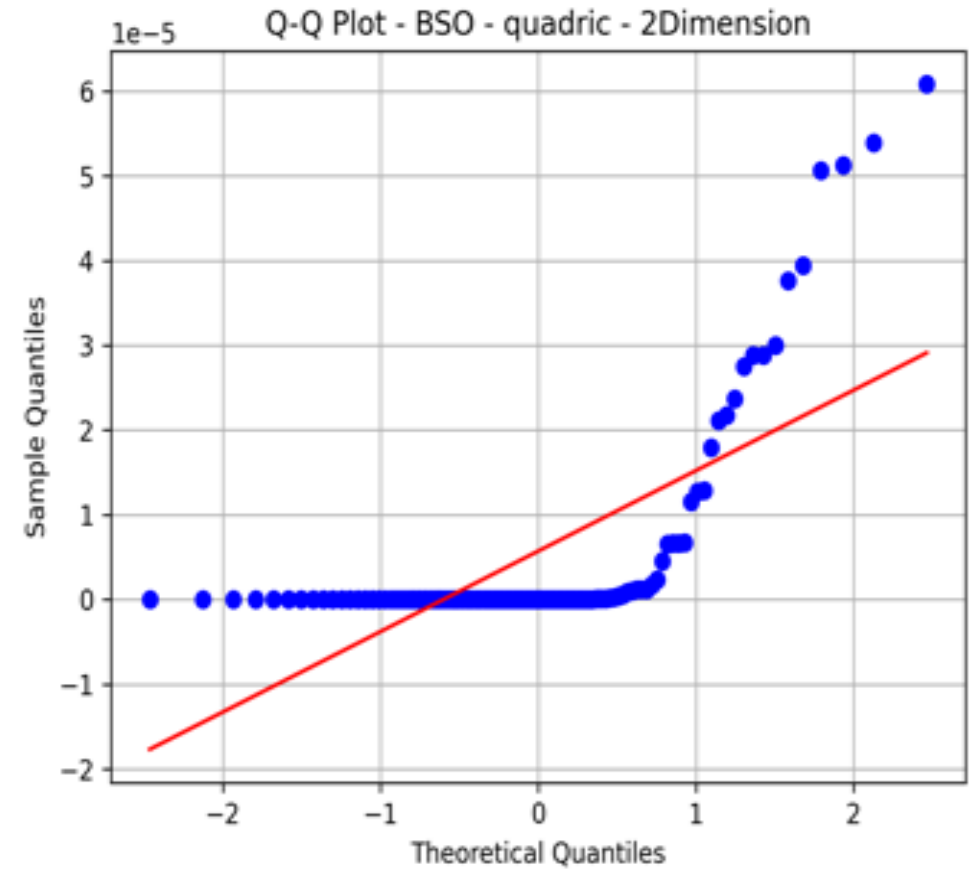
Q-Q Plot - BSO - quadric - 2Dimension
1e−5
Sample Quantiles
Theoretical Quantiles

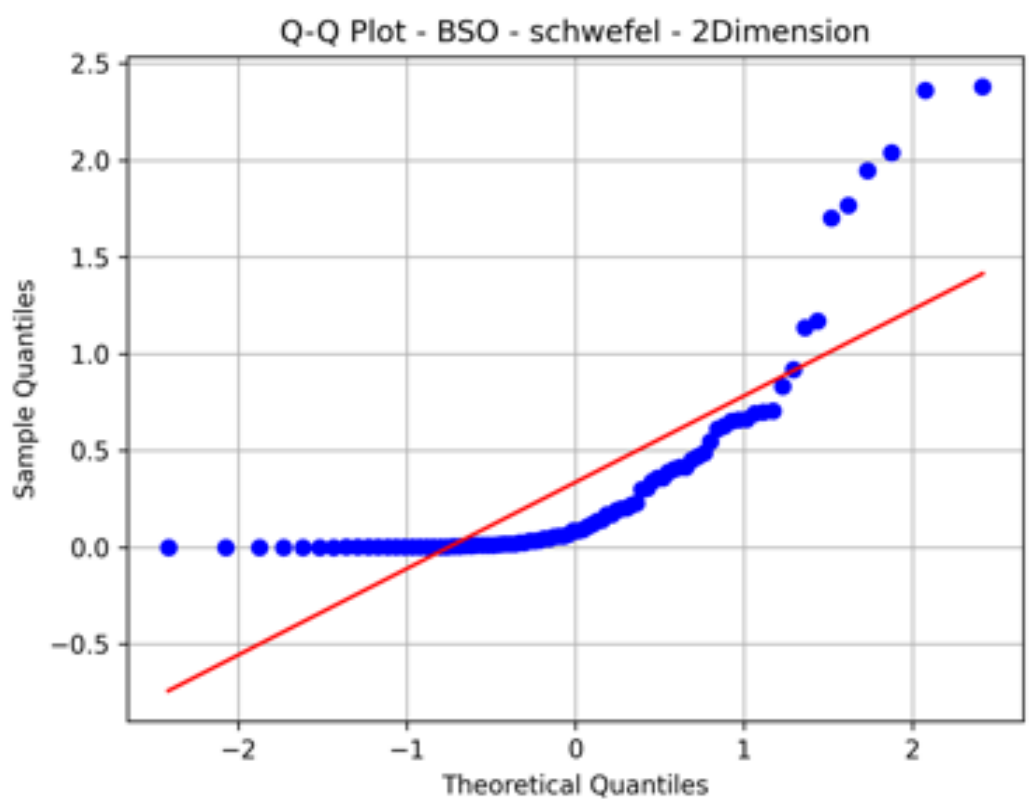
Q-Q Plot - BSO - schwefel - 2Dimension
Sample Quantiles
Theoretical Quantiles

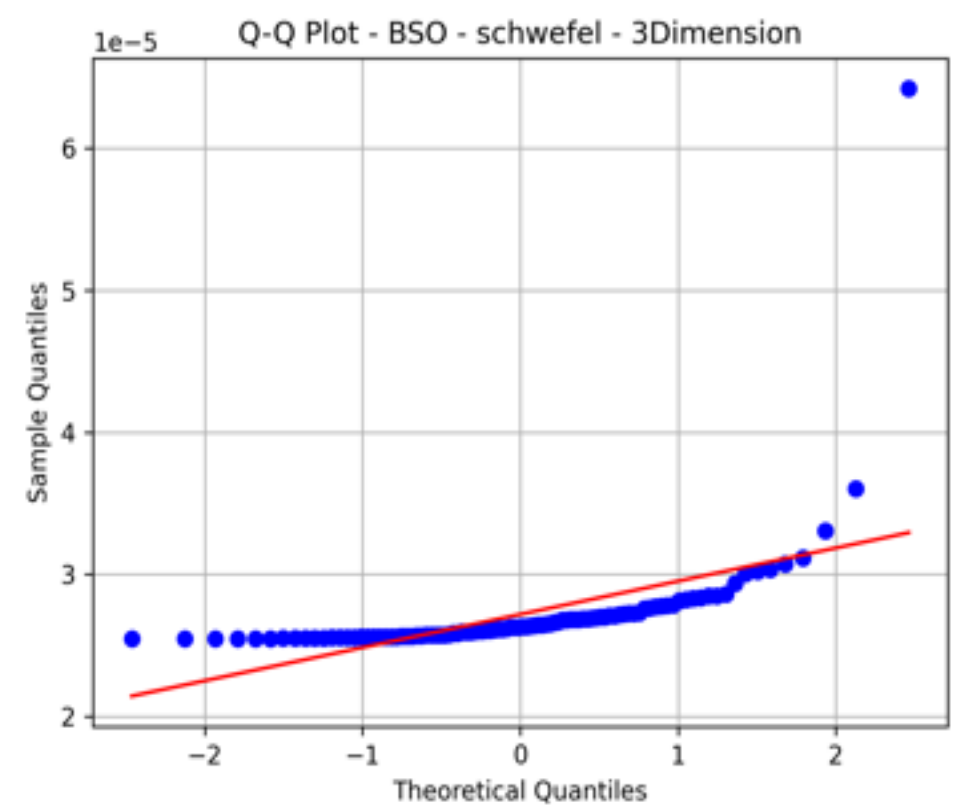
Q-Q Plot - BSO - schwefel - 3Dimension
1e−5
Sample Quantiles
Theoretical Quantiles

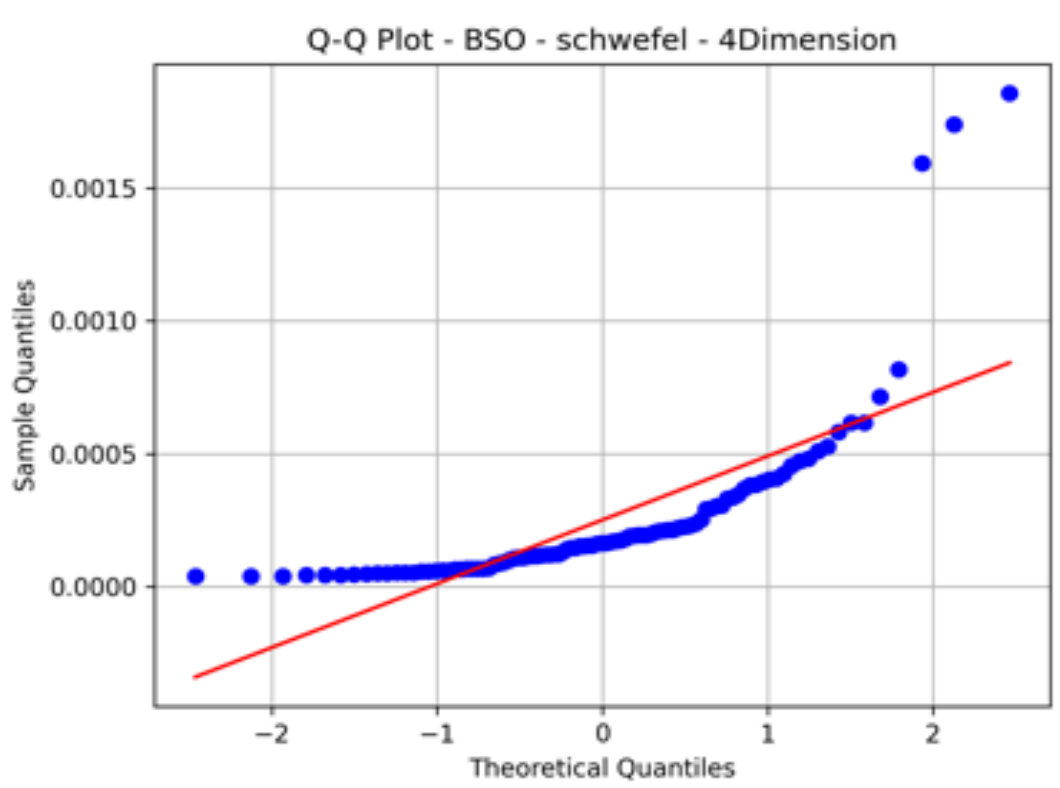
Q-Q Plot - BSO - schwefel - 4Dimension
Sample Quantiles
Theoretical Quantiles

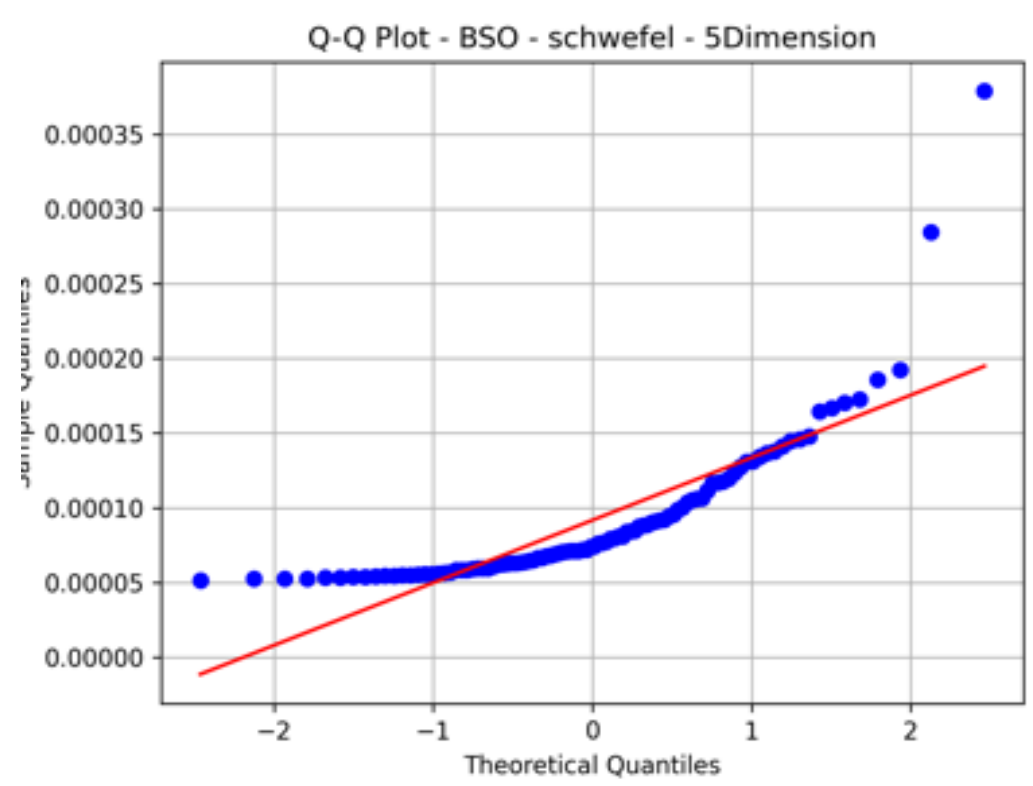
Q-Q Plot - BSO - schwefel - 5Dimension
Theoretical Quantiles

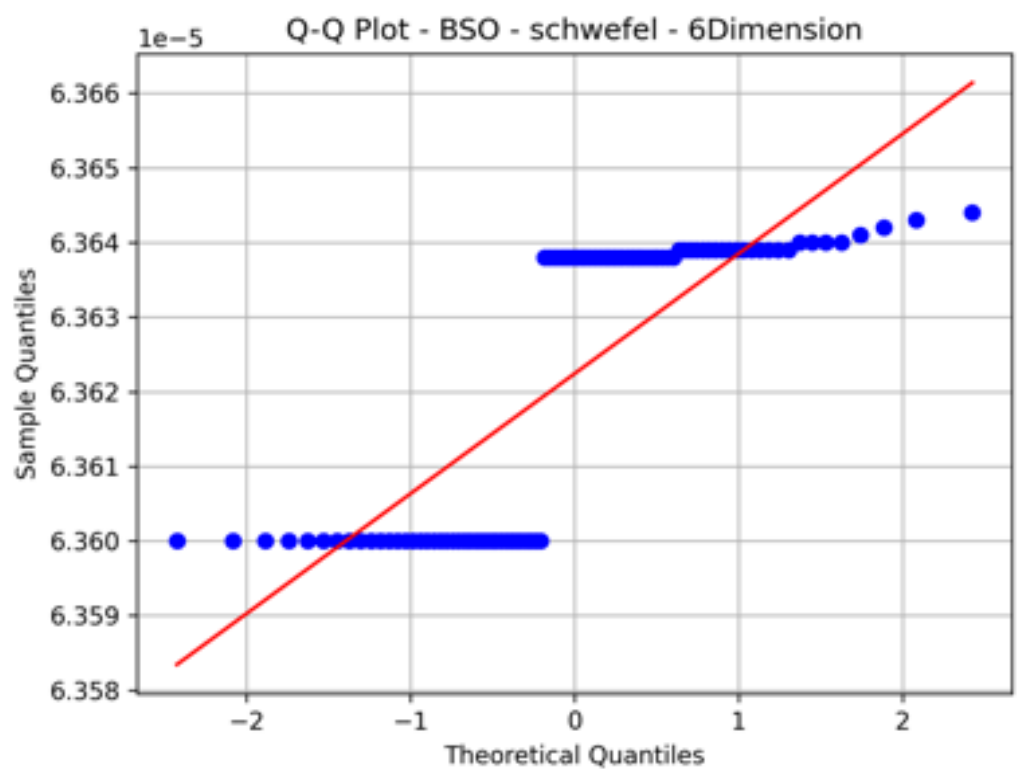
1e−5
Q-Q Plot - BSO - schwefel - 6Dimension
Sample Quantiles
Theoretical Quantiles

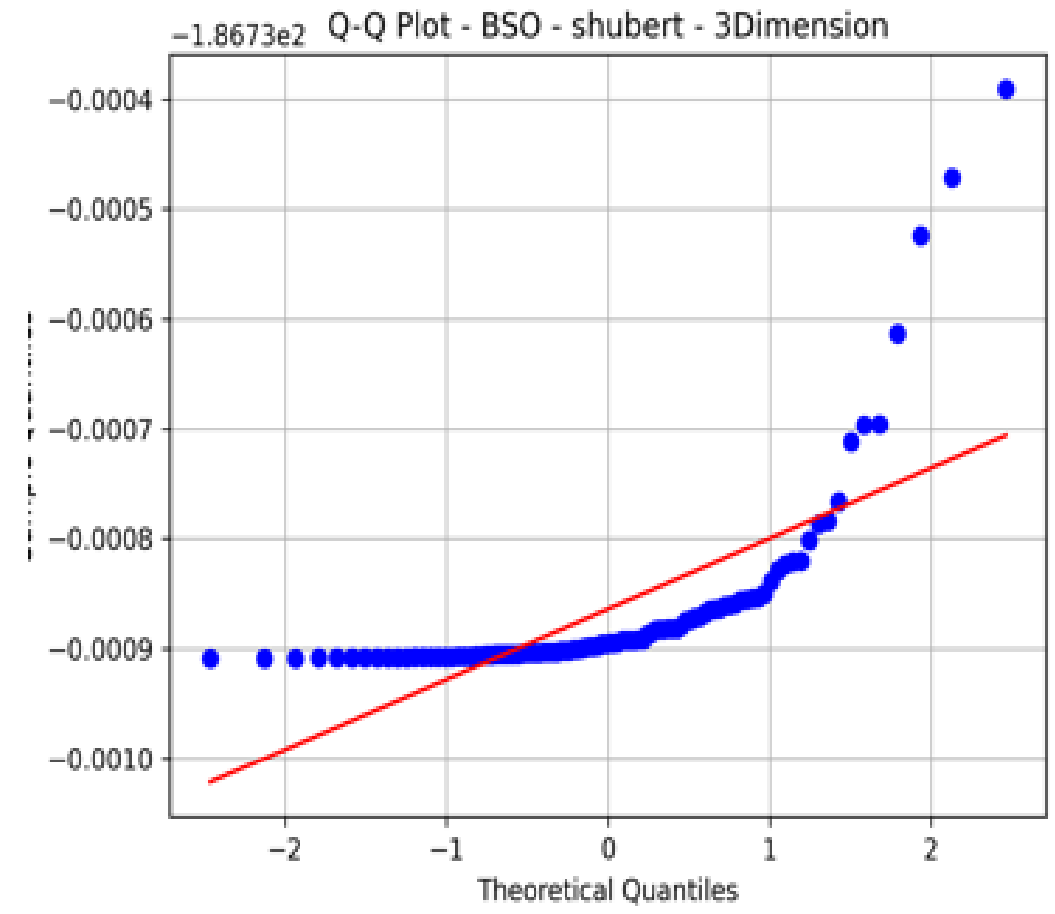
−1.8673e2
Q-Q Plot - BSO - shubert - 3Dimension
Theoretical Quantiles

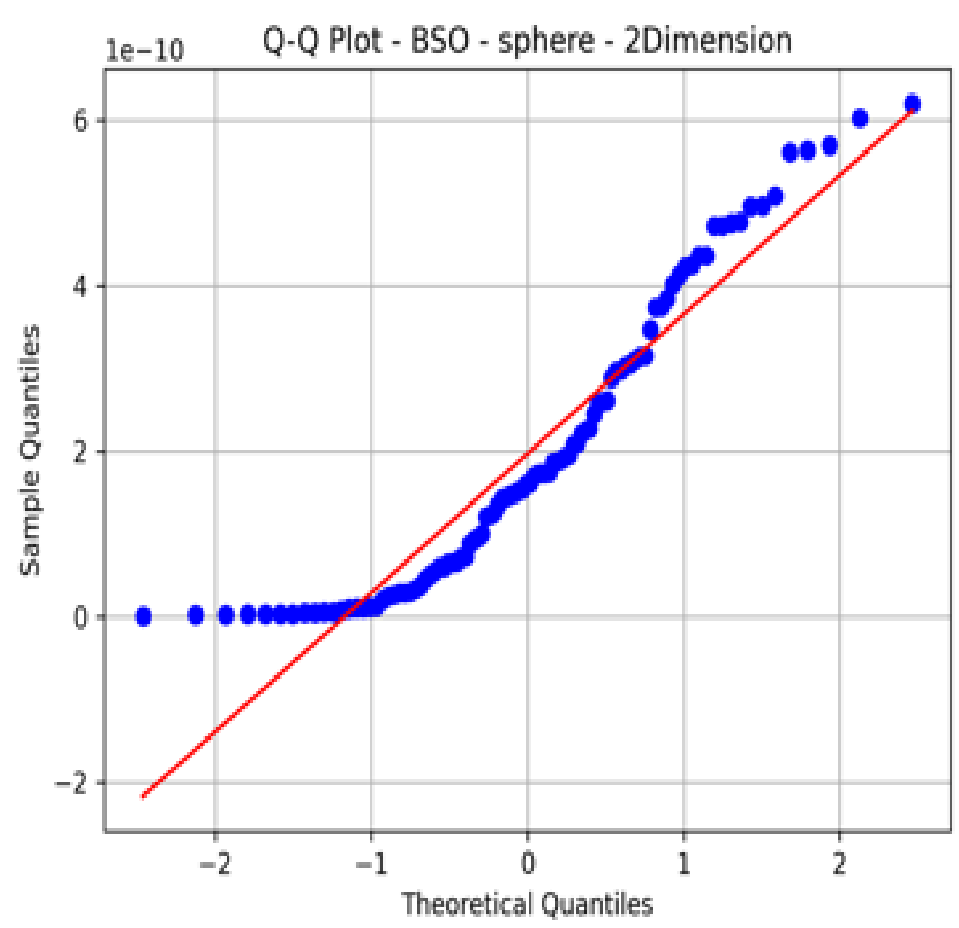
1e−10
Q-Q Plot - BSO - sphere - 2Dimension
Sample Quantiles
Theoretical Quantiles

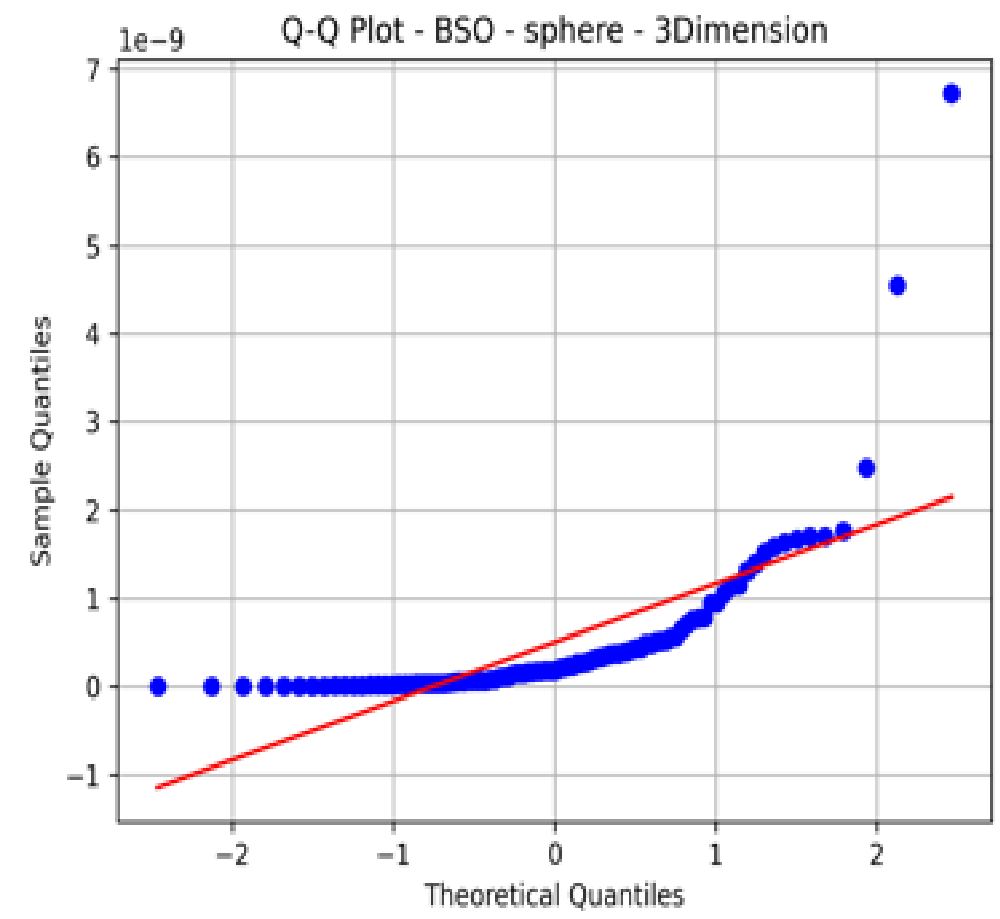
1e−9
Q-Q Plot - BSO - sphere - 3Dimension
Sample Quantiles
Theoretical Quantiles

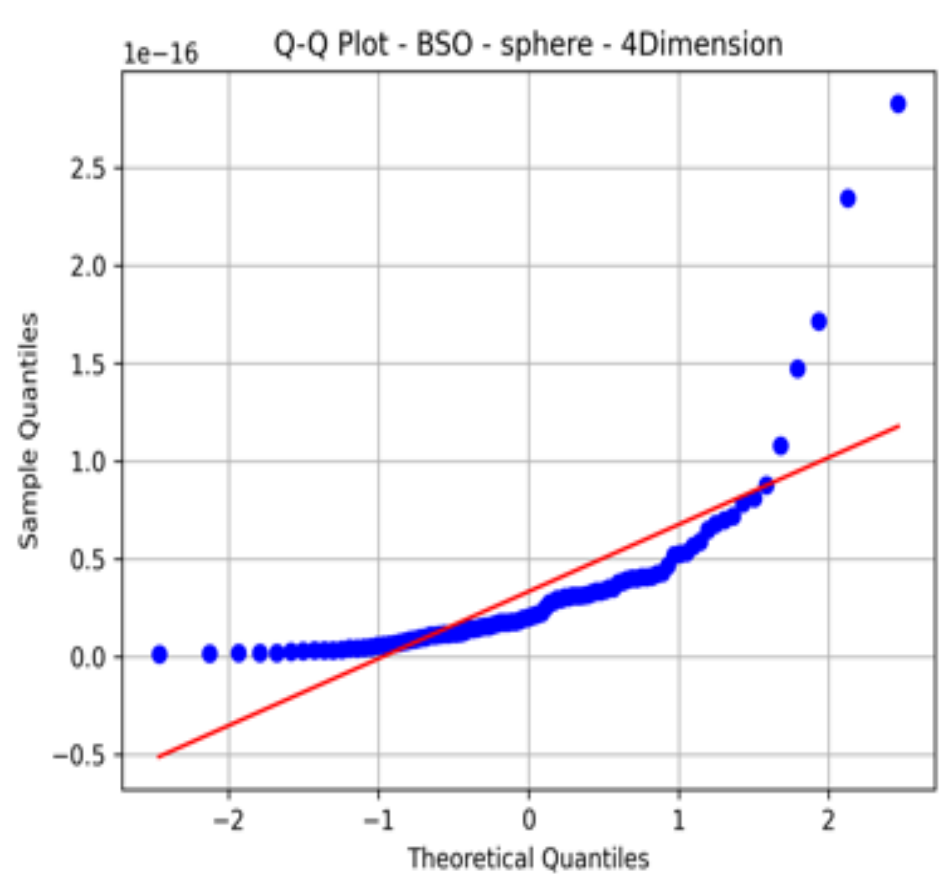
1e−16
Q-Q Plot - BSO - sphere - 4Dimension
Sample Quantiles
Theoretical Quantiles

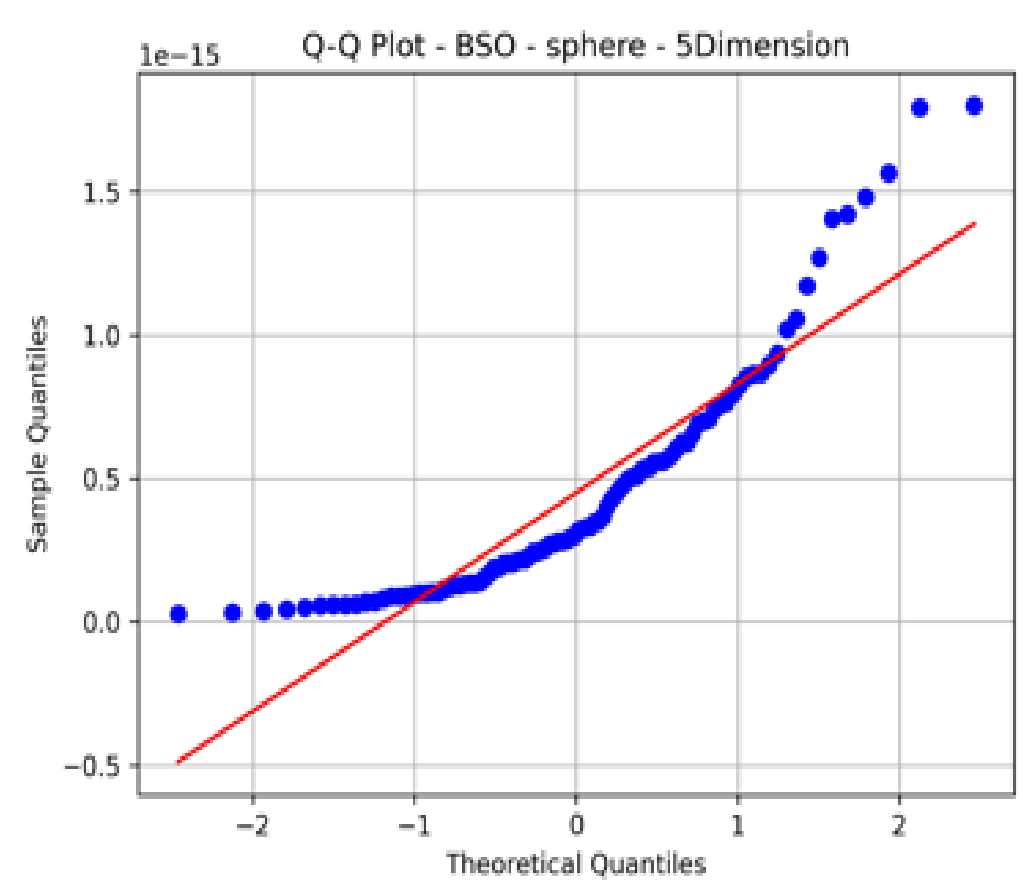
1e−15
Q-Q Plot - BSO - sphere - 5Dimension
Sample Quantiles
Theoretical Quantiles

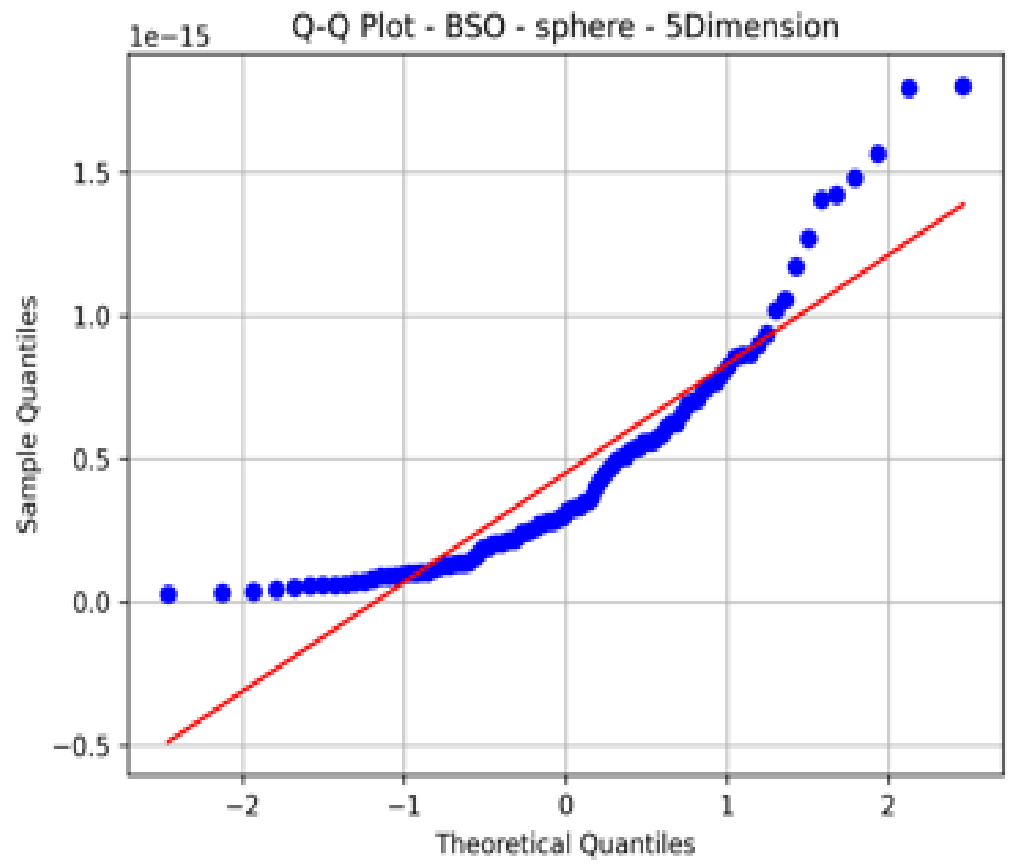
Q-Q Plot - BSO - sphere - 5Dimension
1e−15
Sample Quantiles
Theoretical Quantiles

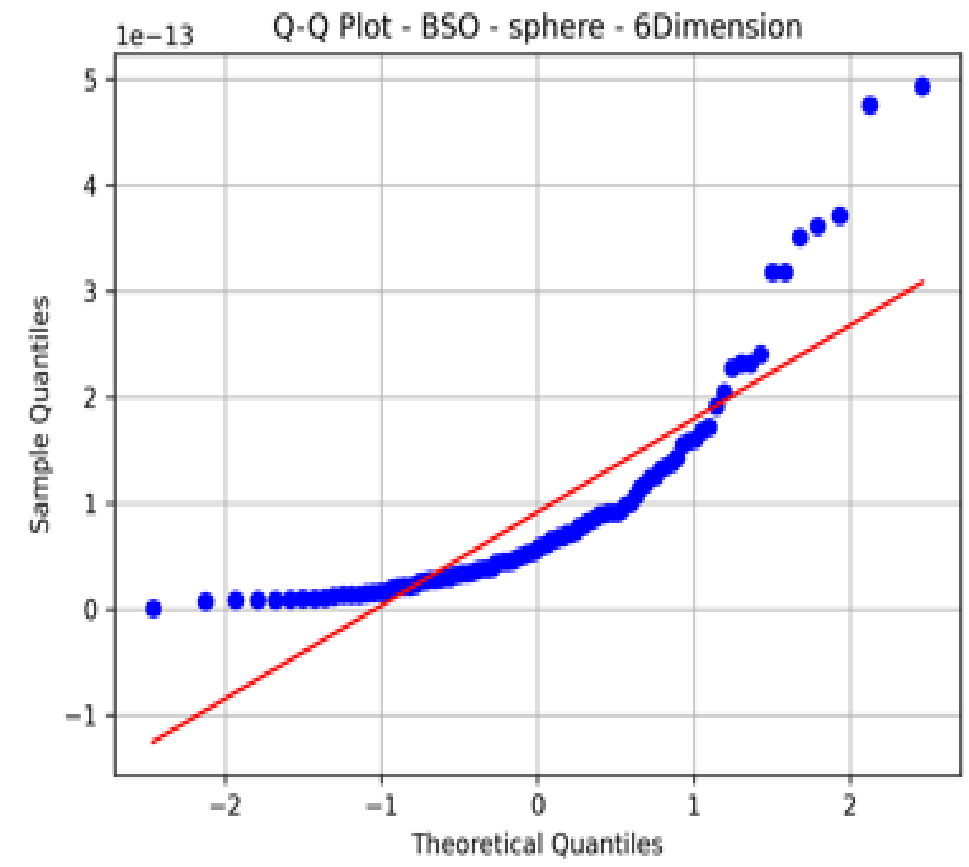
Q-Q Plot - BSO - sphere - 6Dimension
1e−13
Sample Quantiles
Theoretical Quantiles

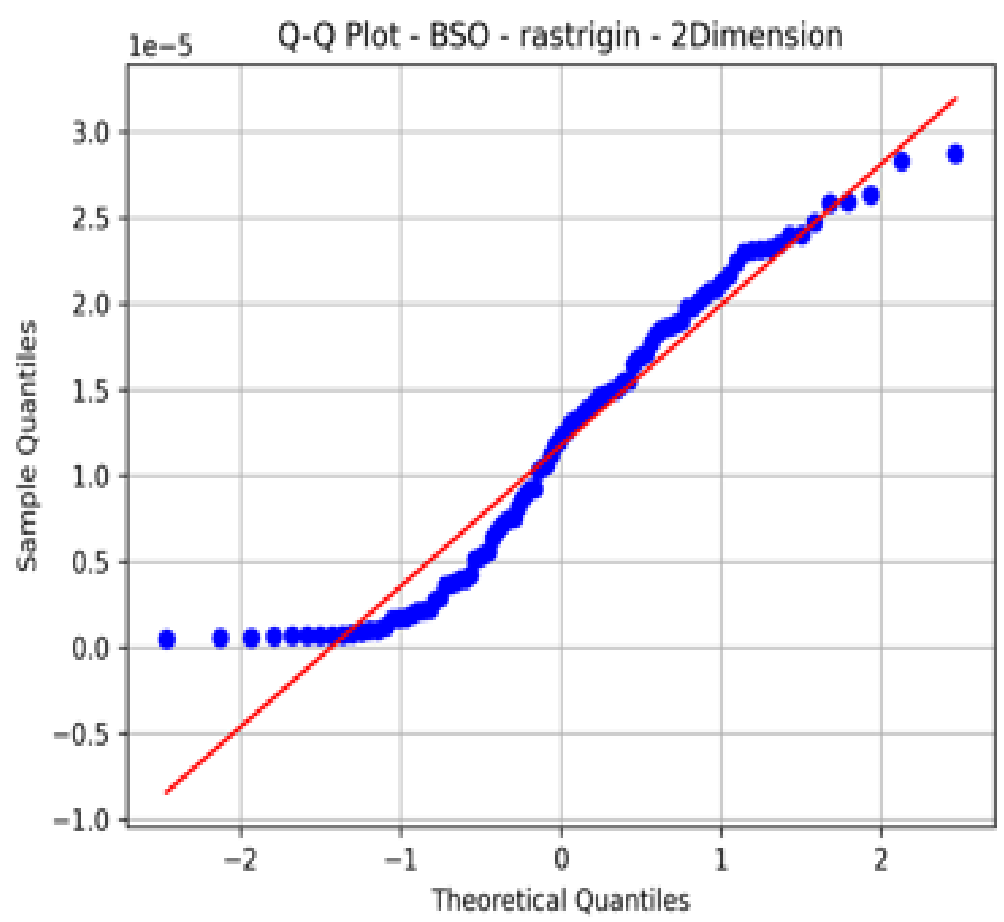
Q-Q Plot - BSO - rastrigin - 2Dimension
1e−5
Sample Quantiles
Theoretical Quantiles

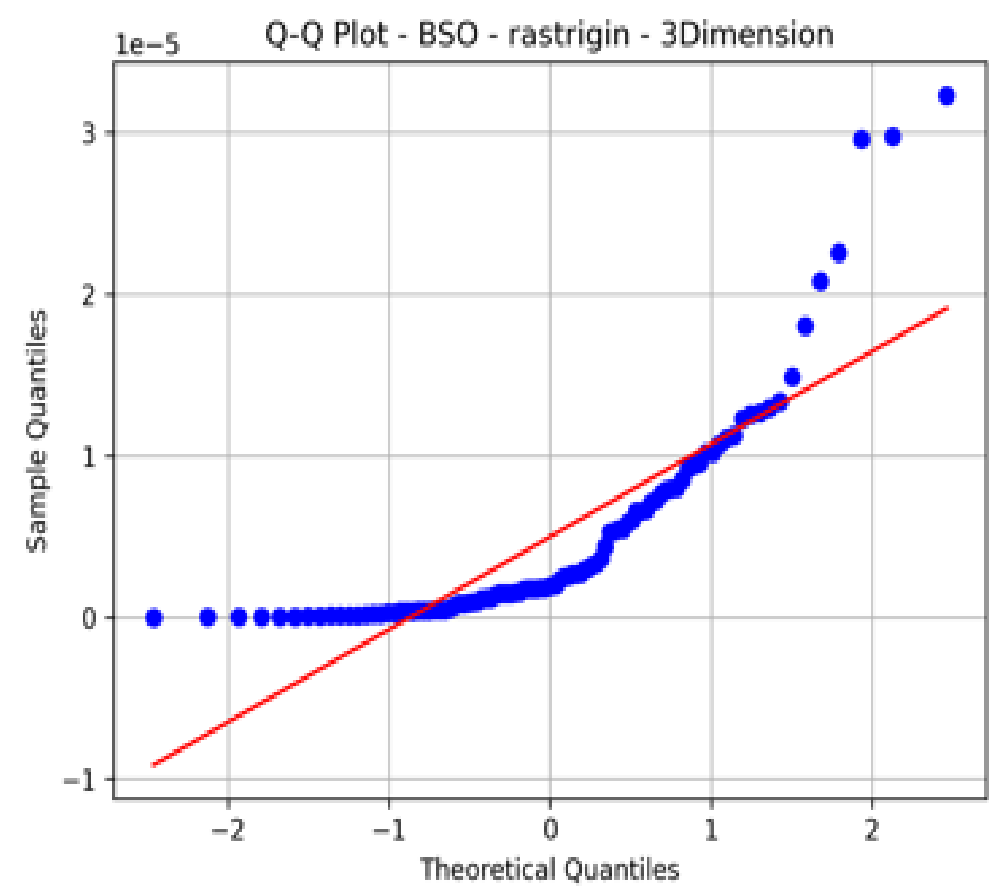
Q-Q Plot - BSO - rastrigin - 3Dimension
1e−5
Sample Quantiles
Theoretical Quantiles

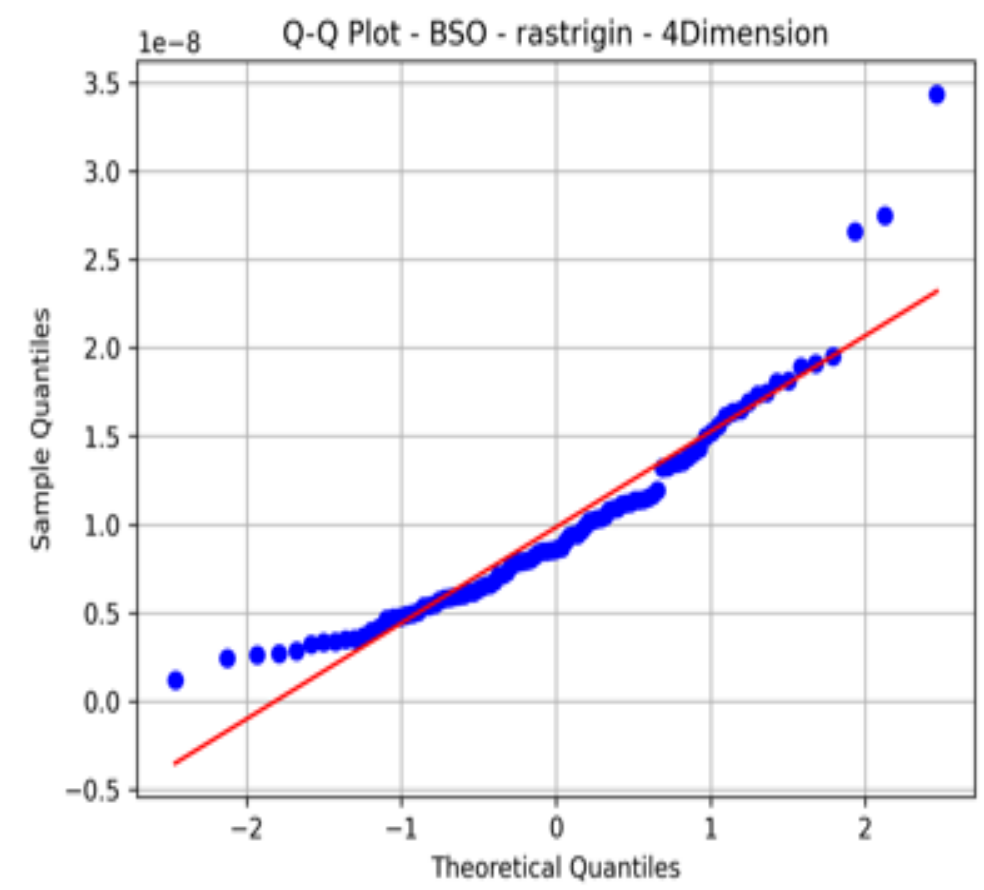
Q-Q Plot - BSO - rastrigin - 4Dimension
1e−8
Sample Quantiles
Theoretical Quantiles
3.5
3.0
2.5
2.0
1.5
1.0
0.5
0.0
−0.5
−2
−1
0
1
2

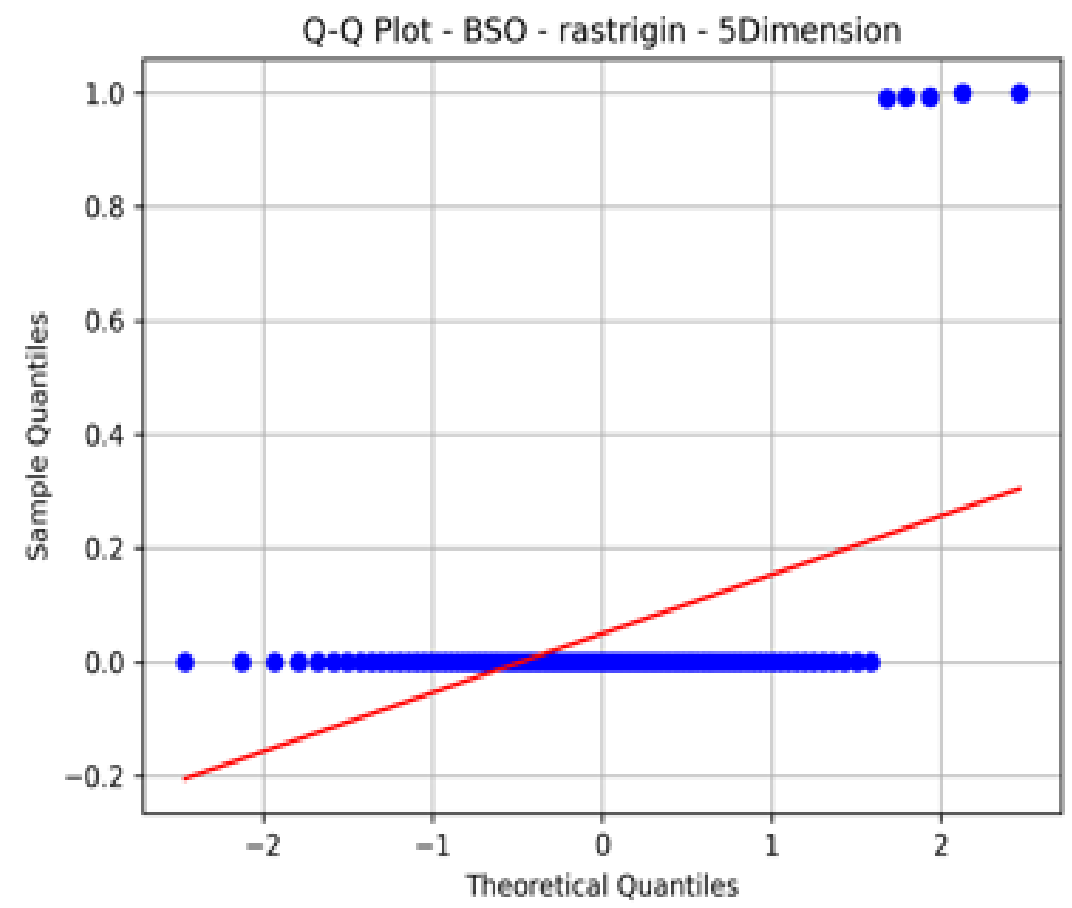
Q-Q Plot - BSO - rastrigin - 5Dimension
Sample Quantiles
Theoretical Quantiles
1.0
0.8
0.6
0.4
0.2
0.0
−0.2
−2
−1
0
1
2

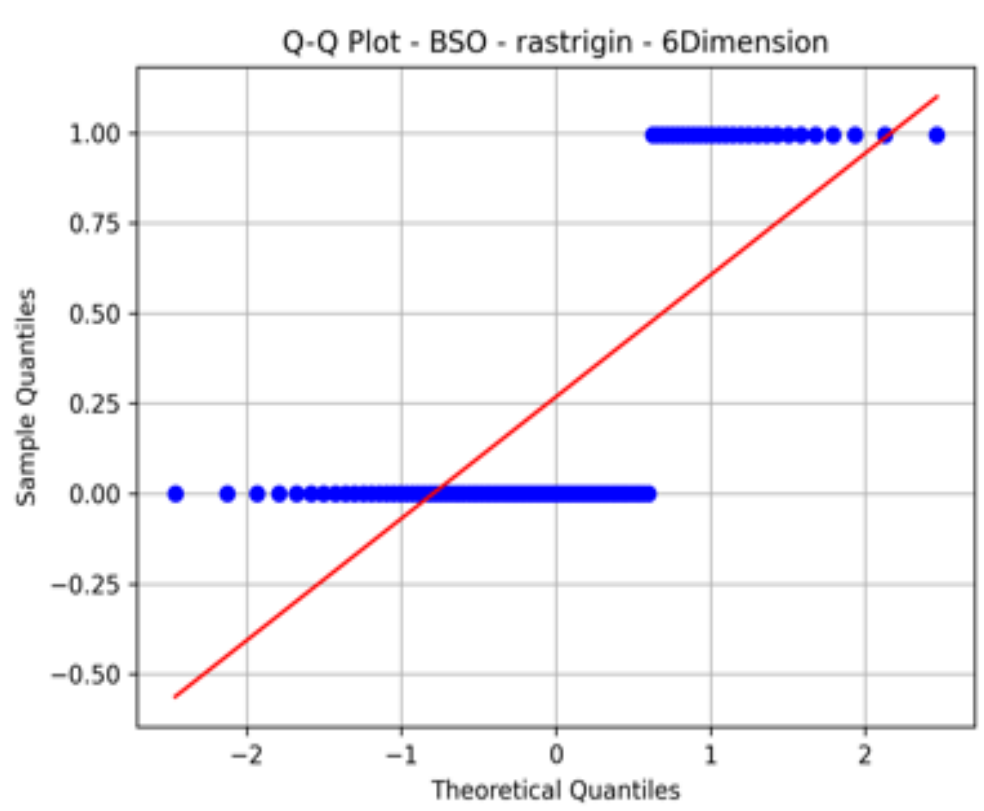
Q-Q Plot - BSO - rastrigin - 6Dimension
Sample Quantiles
Theoretical Quantiles
1.00
0.75
0.50
0.25
0.00
−0.25
−0.50
−2
−1
0
1
2

BSO - Non-Continuous Functions Q-Q Plots

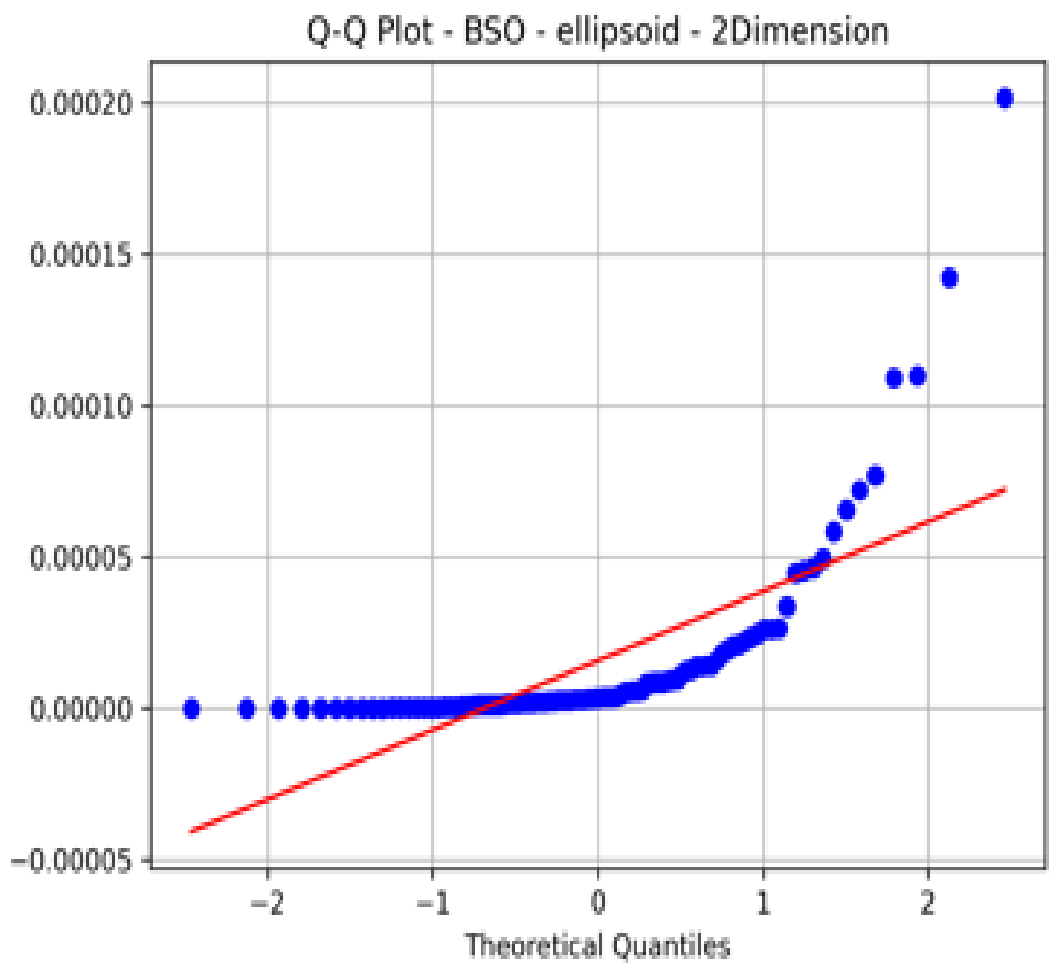

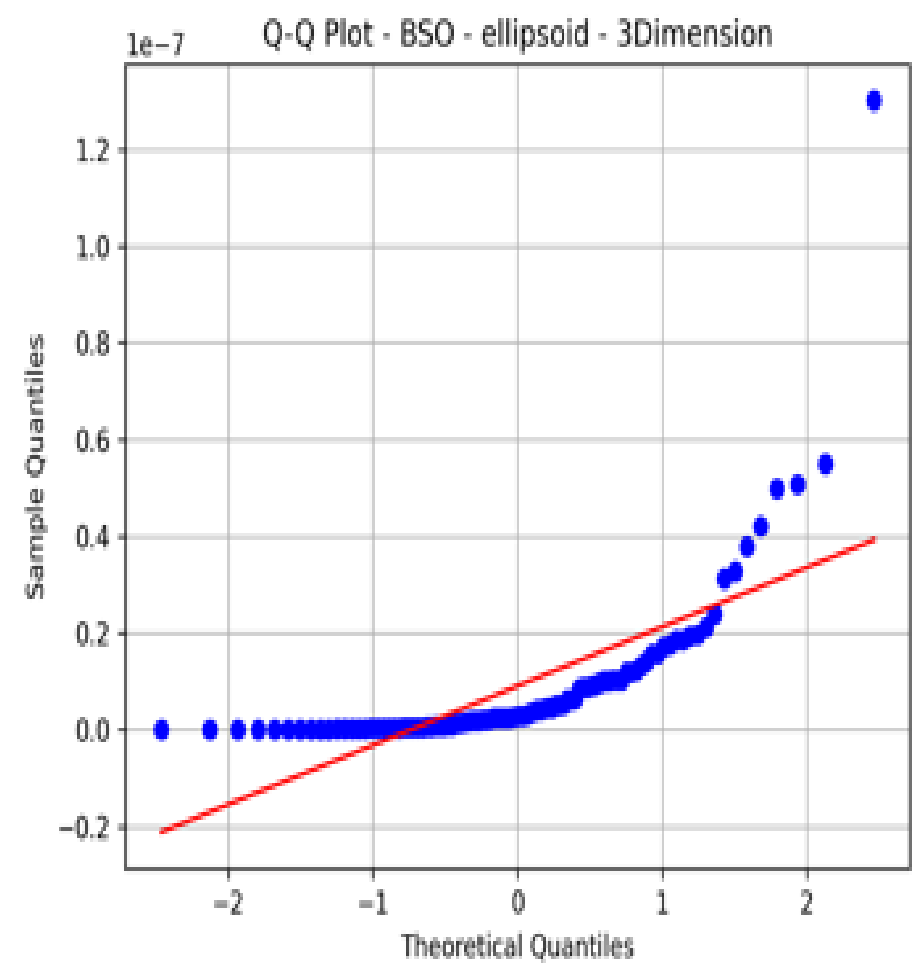

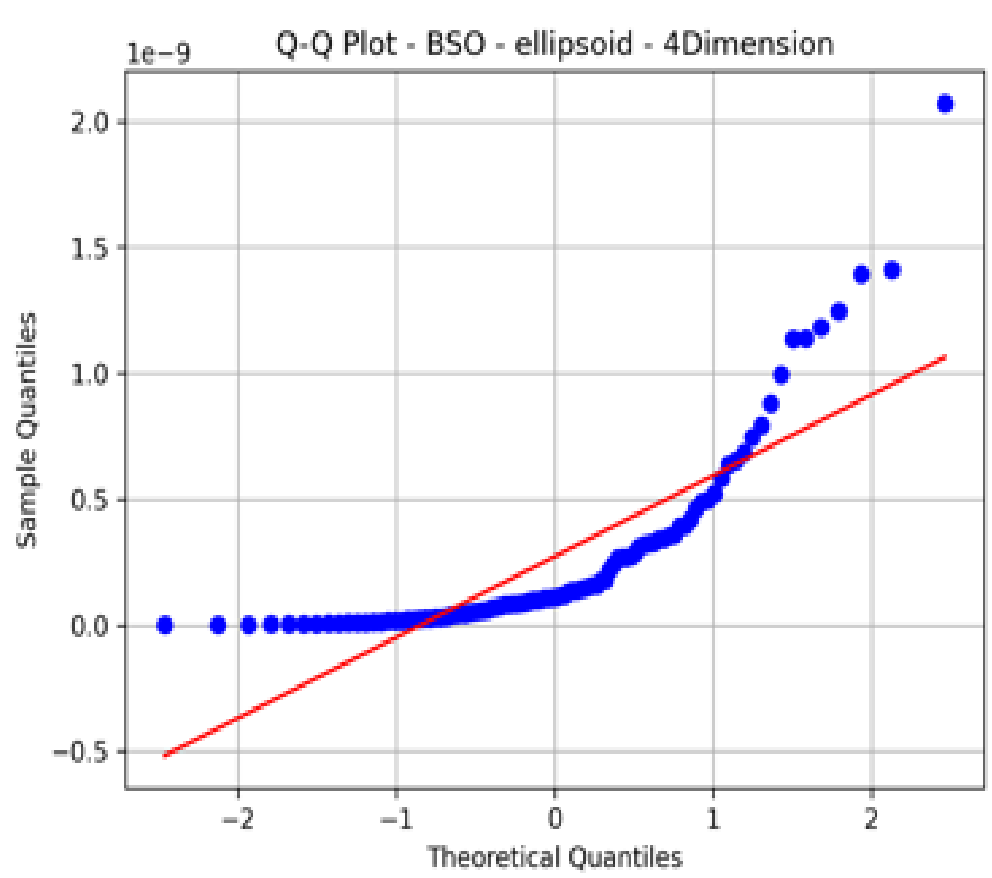

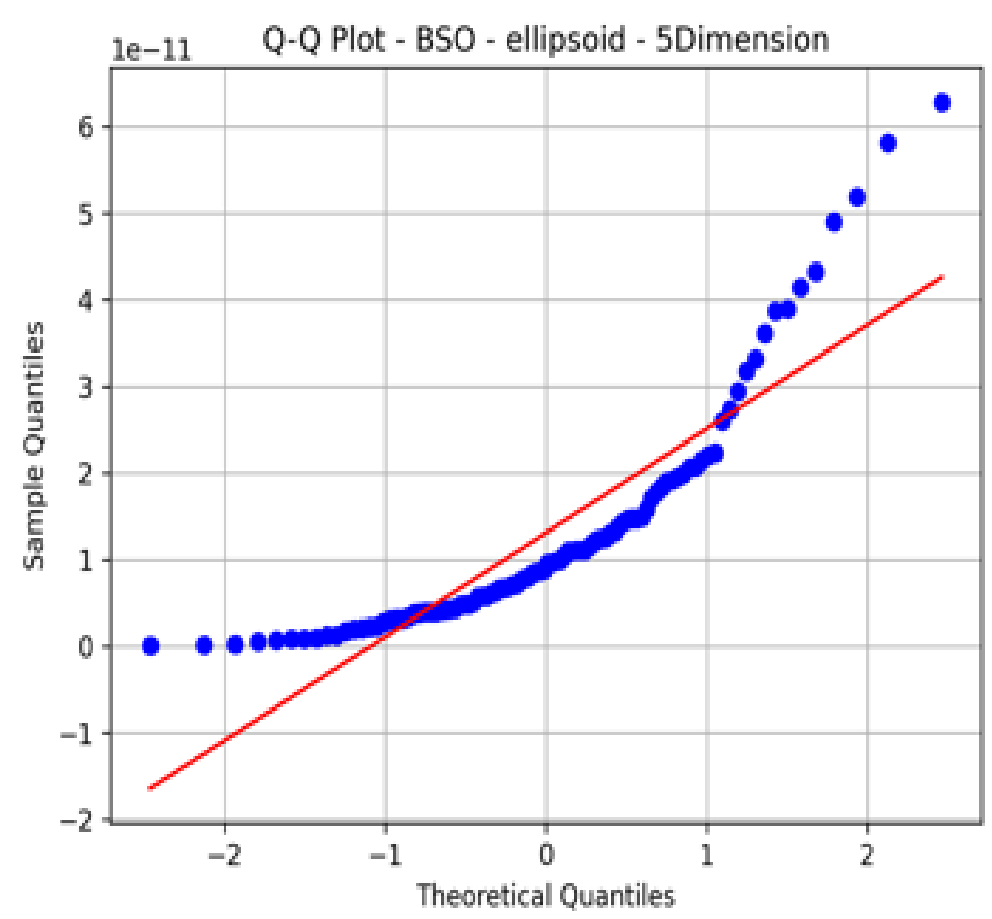

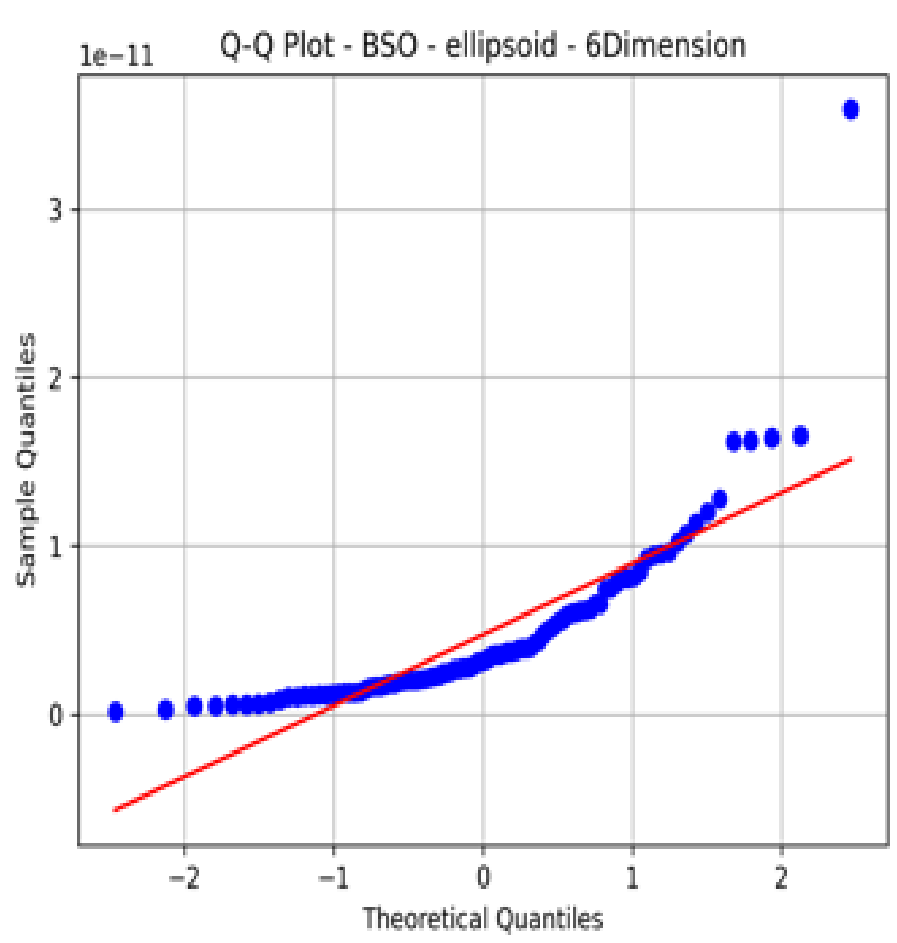

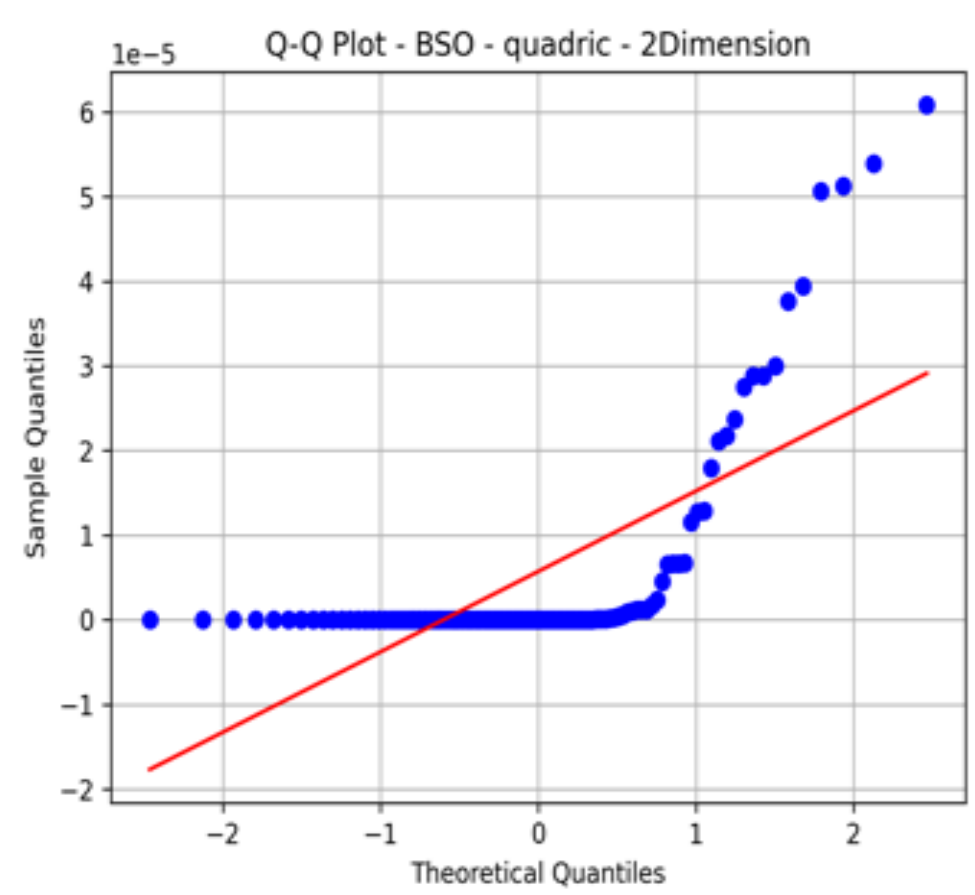

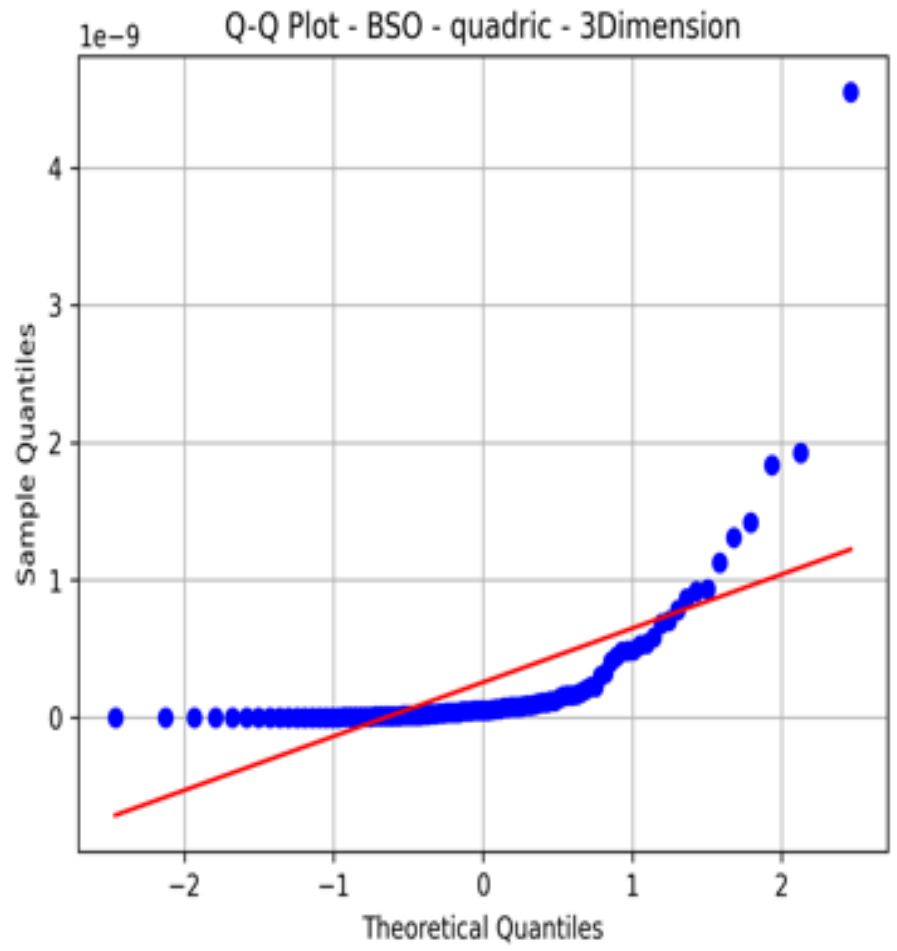
Q-Q Plot - BSO - quadric - 3Dimension
1e−9
Sample Quantiles
Theoretical Quantiles

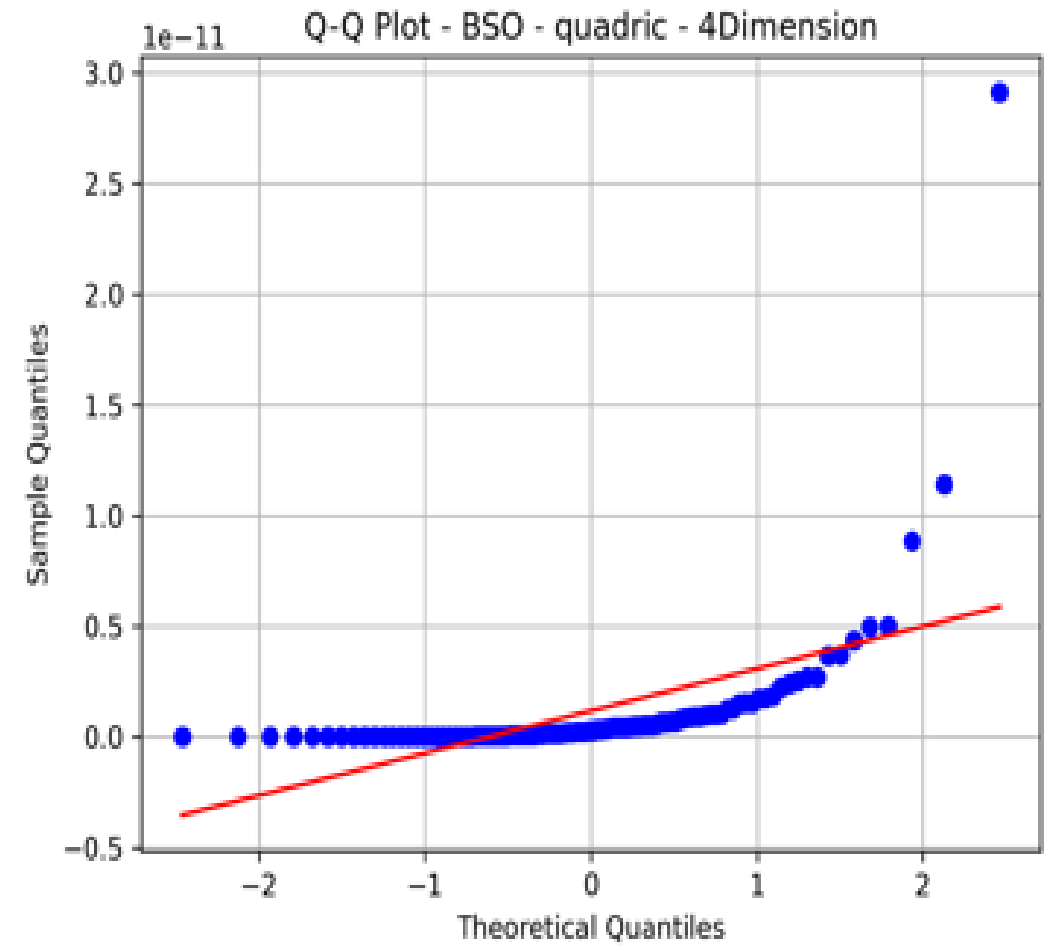
Q-Q Plot - BSO - quadric - 4Dimension
1e−11
Sample Quantiles
Theoretical Quantiles

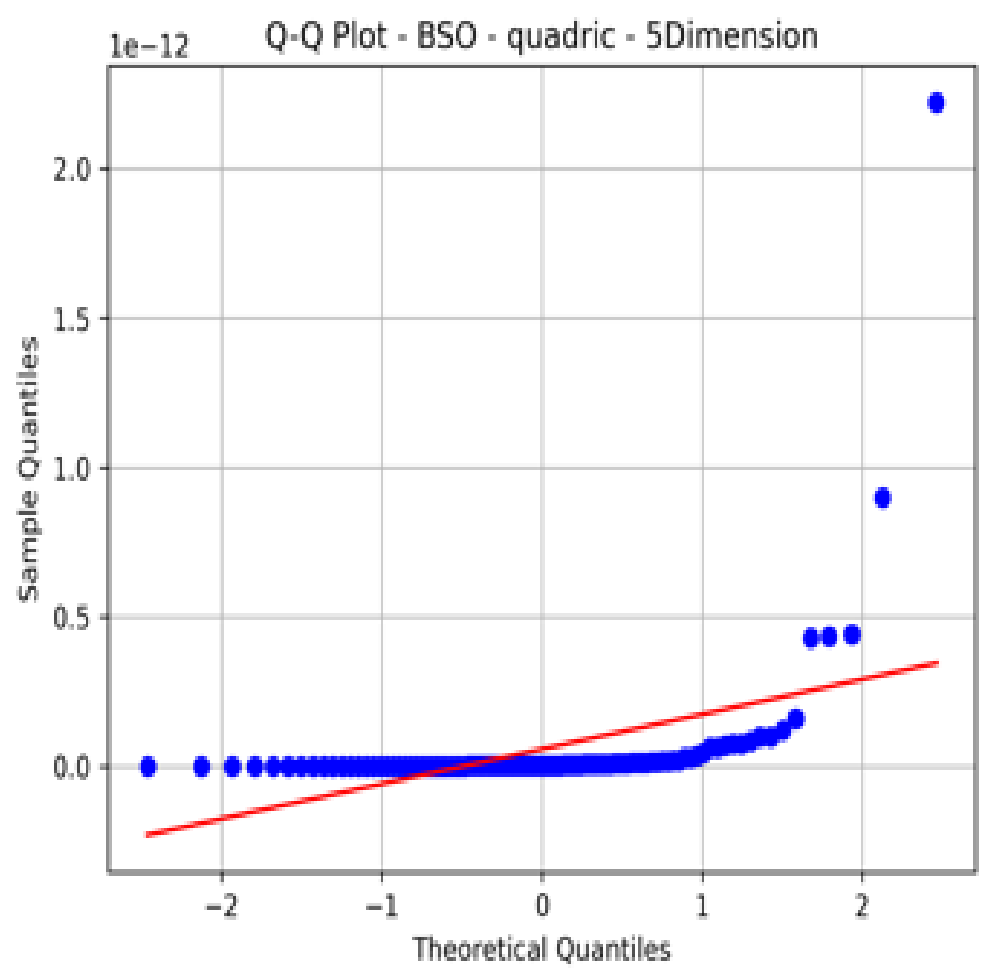
Q-Q Plot - BSO - quadric - 5Dimension
1e−12
Sample Quantiles
Theoretical Quantiles

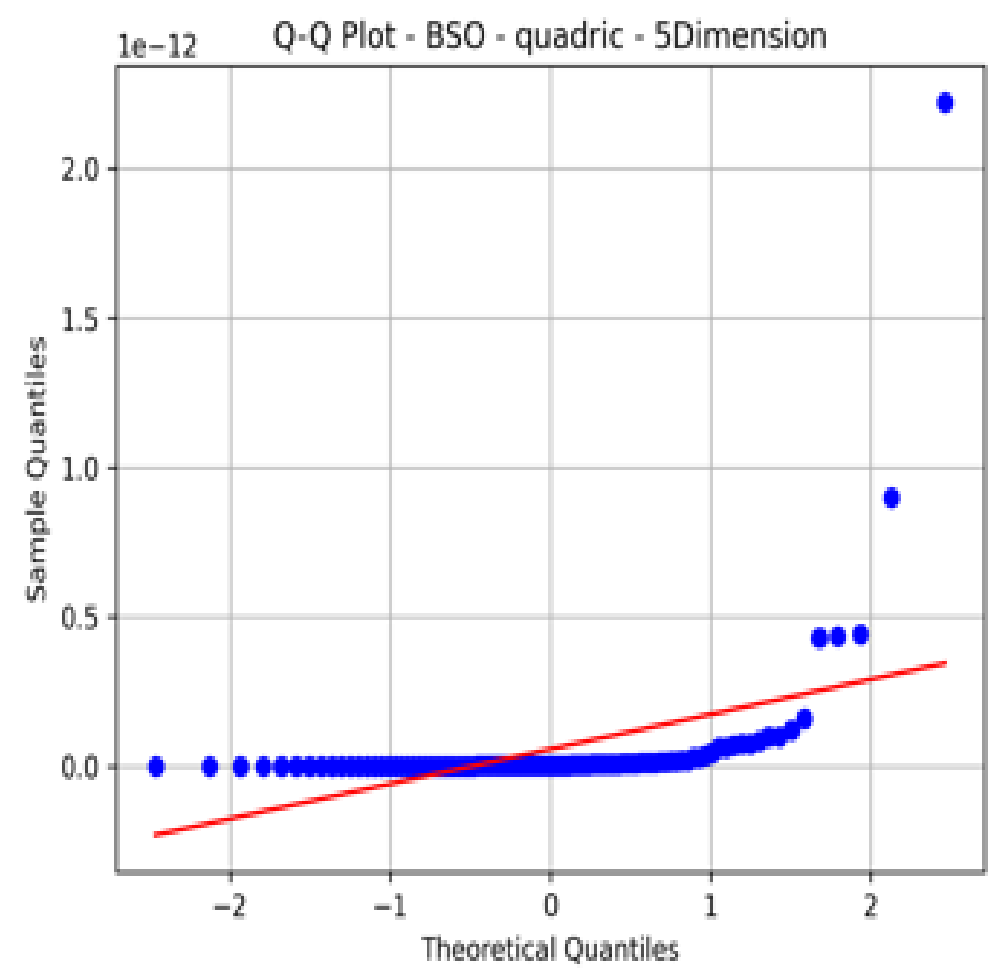
Q-Q Plot - BSO - quadric - 5Dimension
1e−12
Sample Quantiles
Theoretical Quantiles

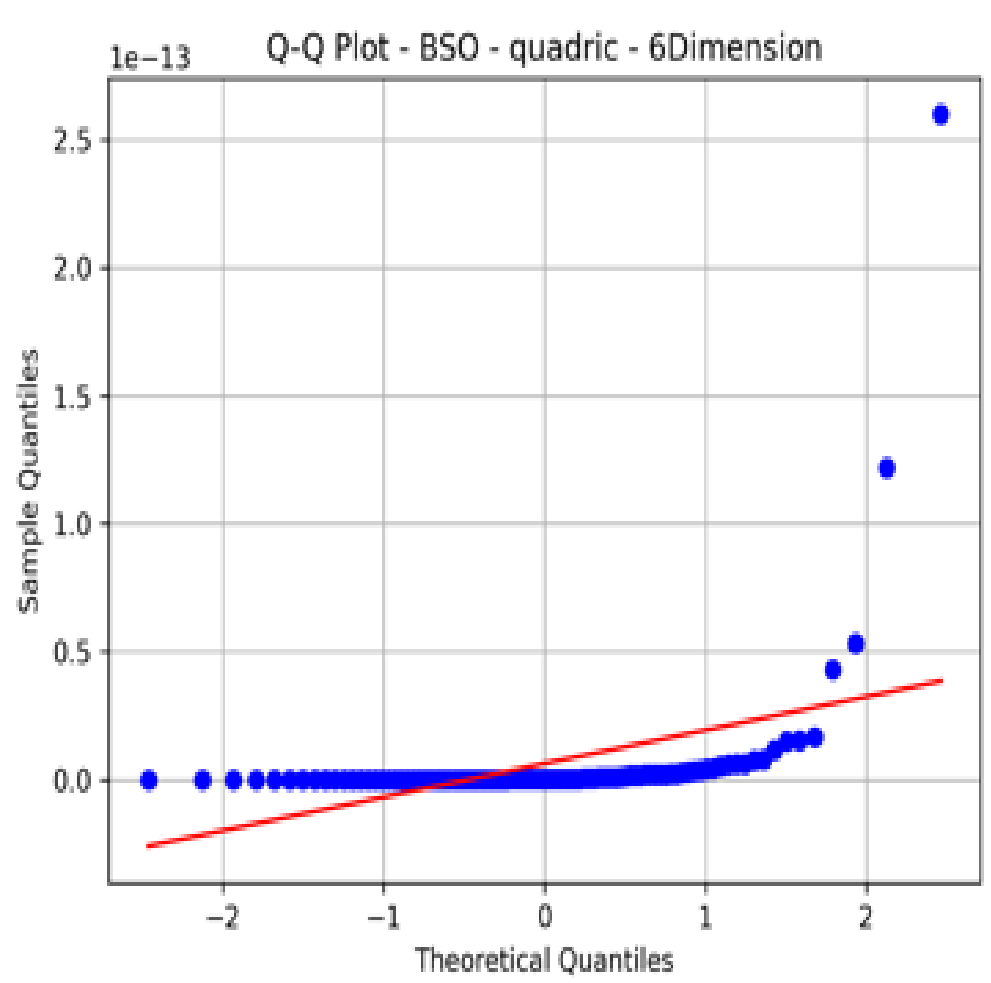
Q-Q Plot - BSO - quadric - 6Dimension
1e−13
Sample Quantiles
Theoretical Quantiles

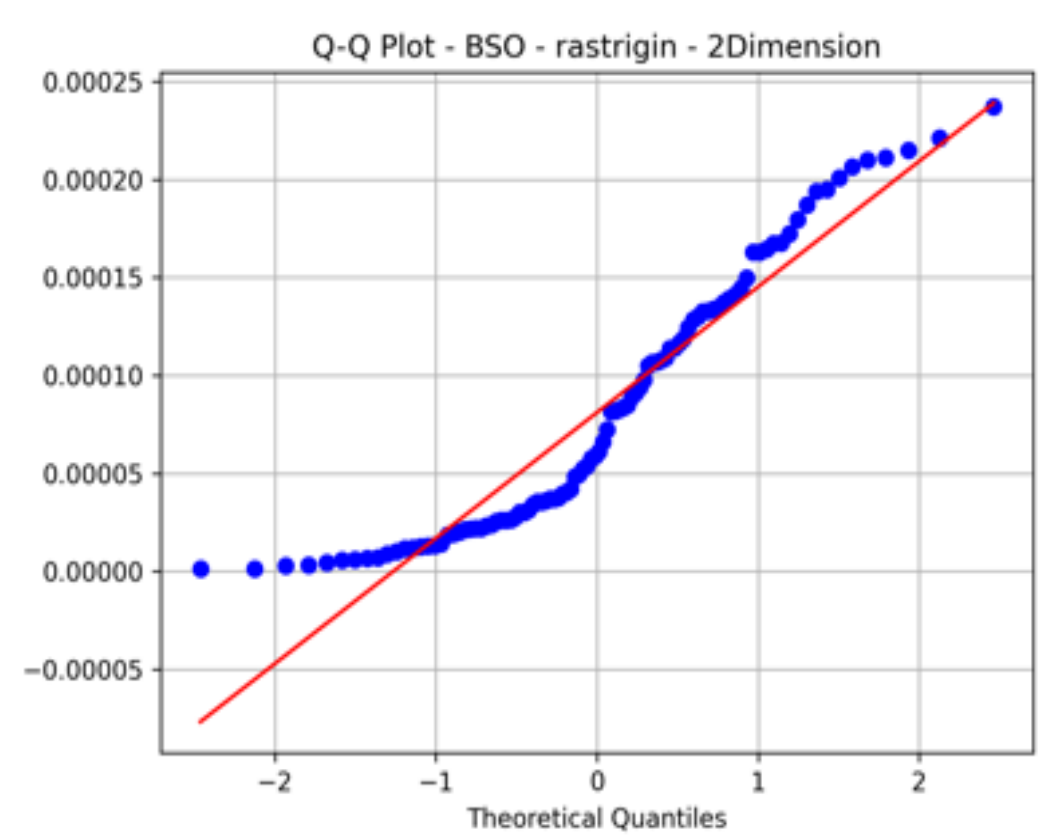
Q-Q Plot - BSO - rastrigin - 2Dimension
Theoretical Quantiles

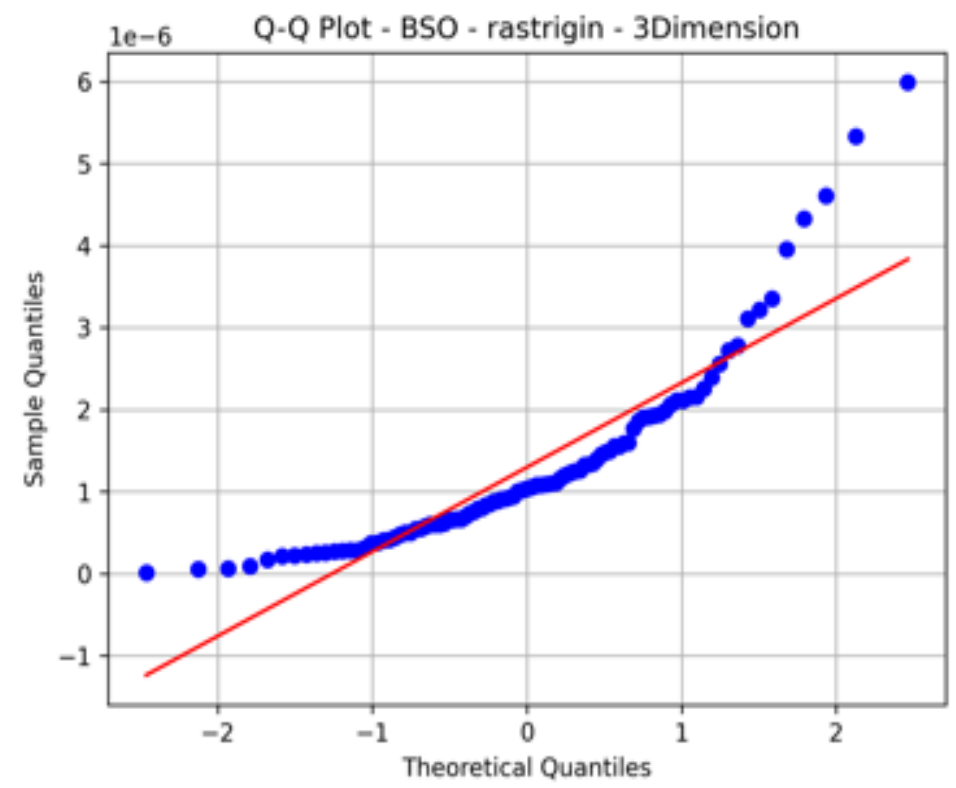
Q-Q Plot - BSO - rastrigin - 3Dimension
1e−6
Sample Quantiles
Theoretical Quantiles

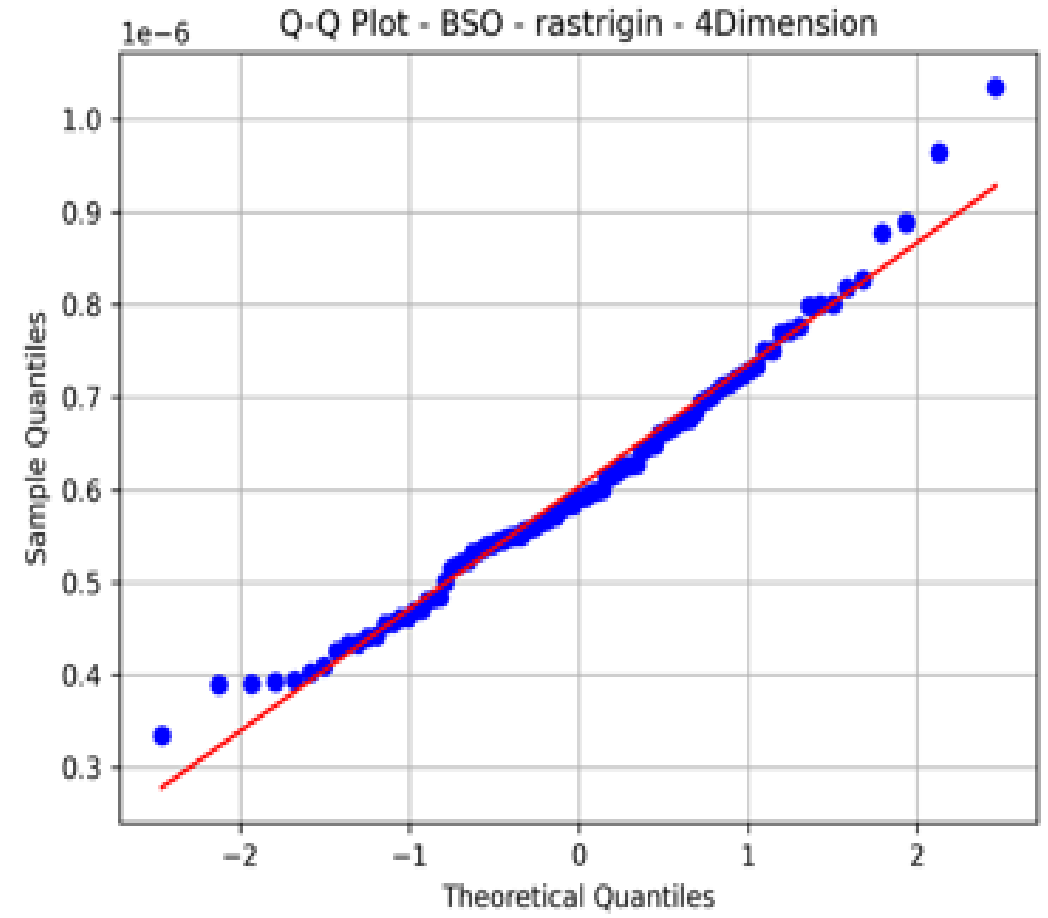
Q-Q Plot - BSO - rastrigin - 4Dimension
1e−6
Sample Quantiles
Theoretical Quantiles

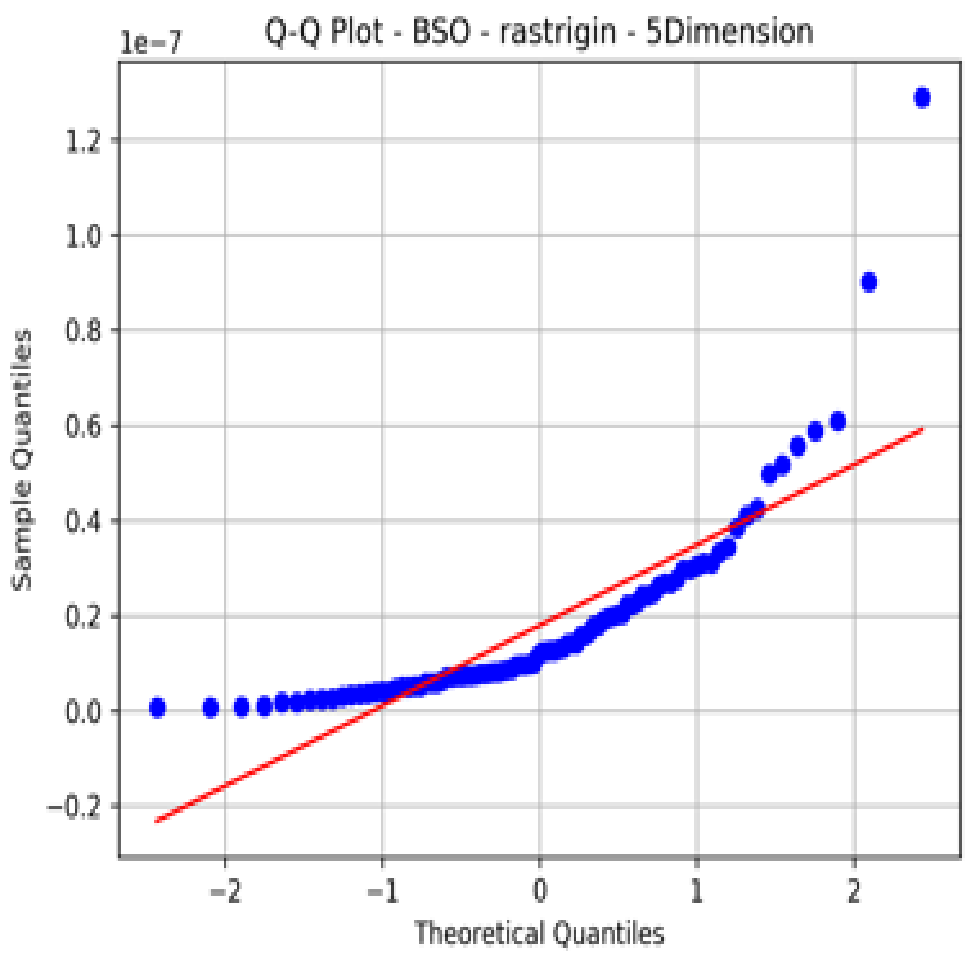
Q-Q Plot - BSO - rastrigin - 5Dimension
1e−7
Sample Quantiles
Theoretical Quantiles

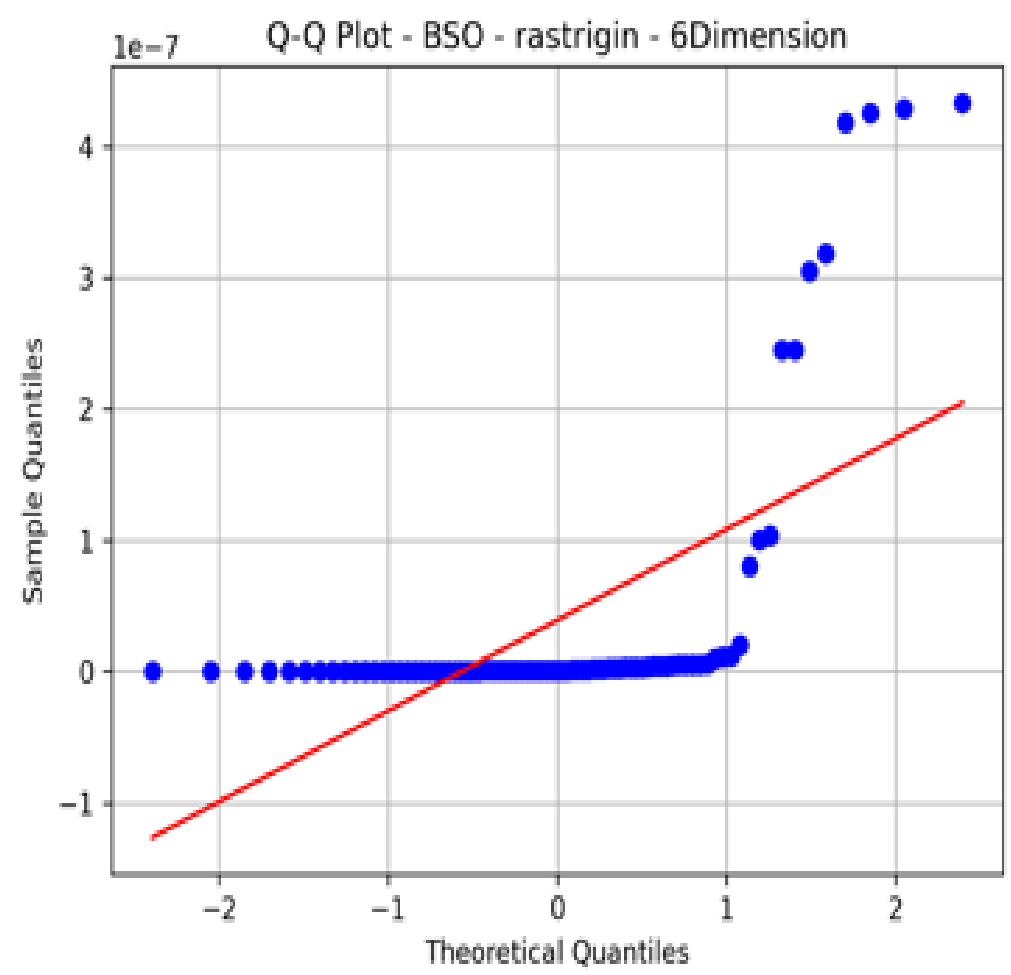
Q-Q Plot - BSO - rastrigin - 6Dimension
1e−7
Sample Quantiles
Theoretical Quantiles

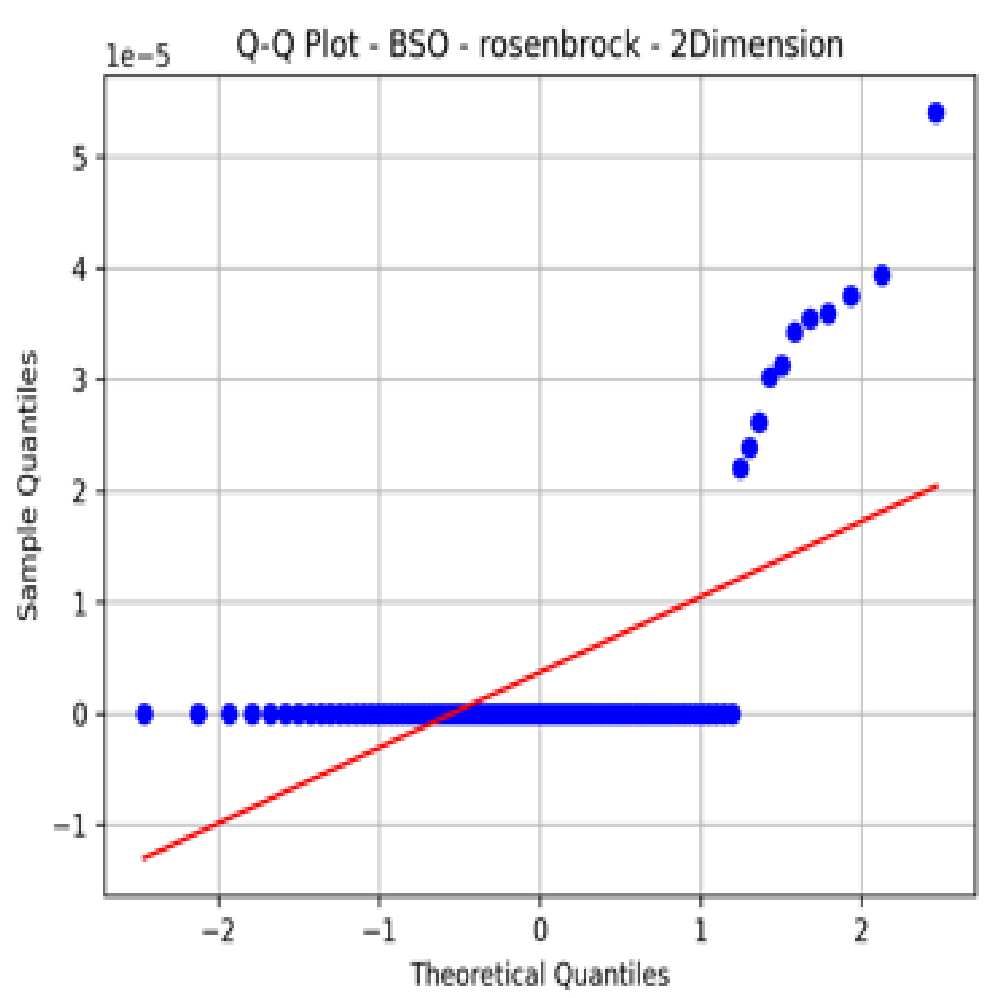
Q-Q Plot - BSO - rosenbrock - 2Dimension
1e−5
Sample Quantiles
Theoretical Quantiles

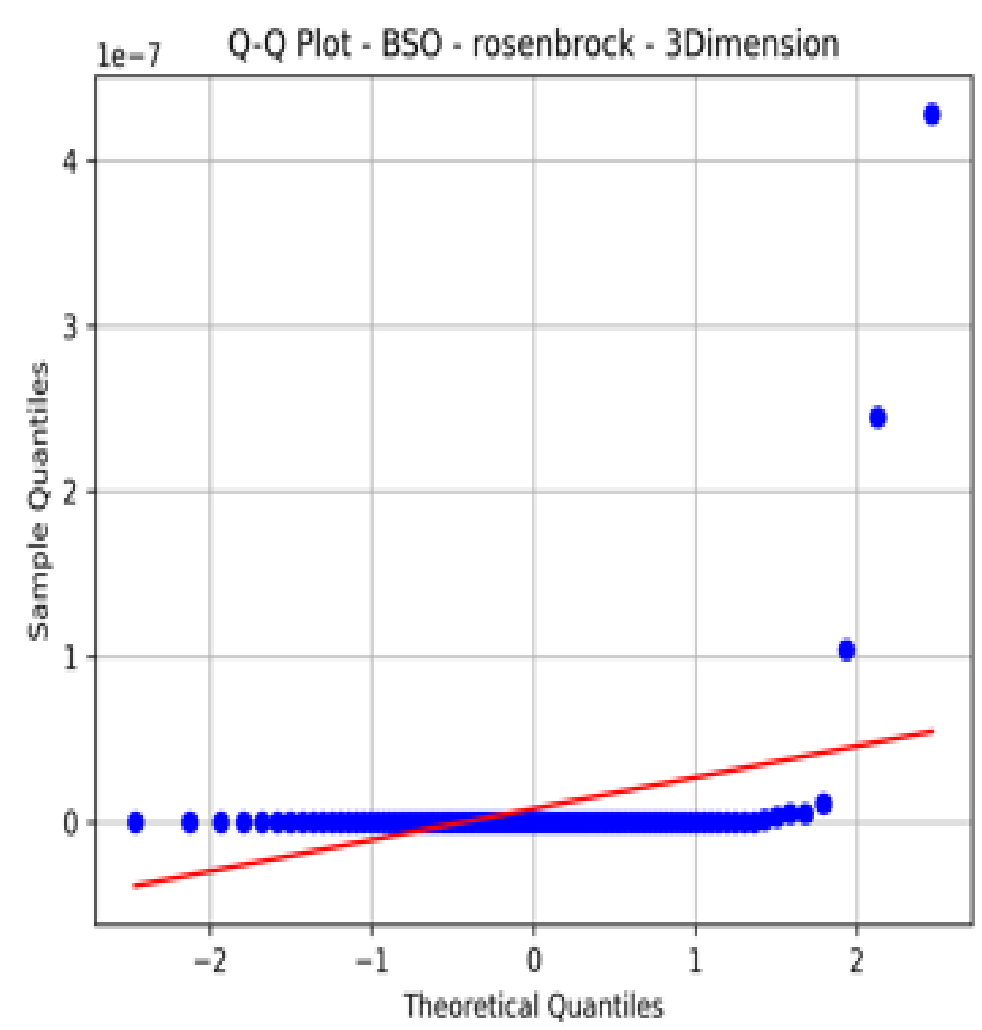
Q-Q Plot - BSO - rosenbrock - 3Dimension
1e−7
Sample Quantiles
Theoretical Quantiles

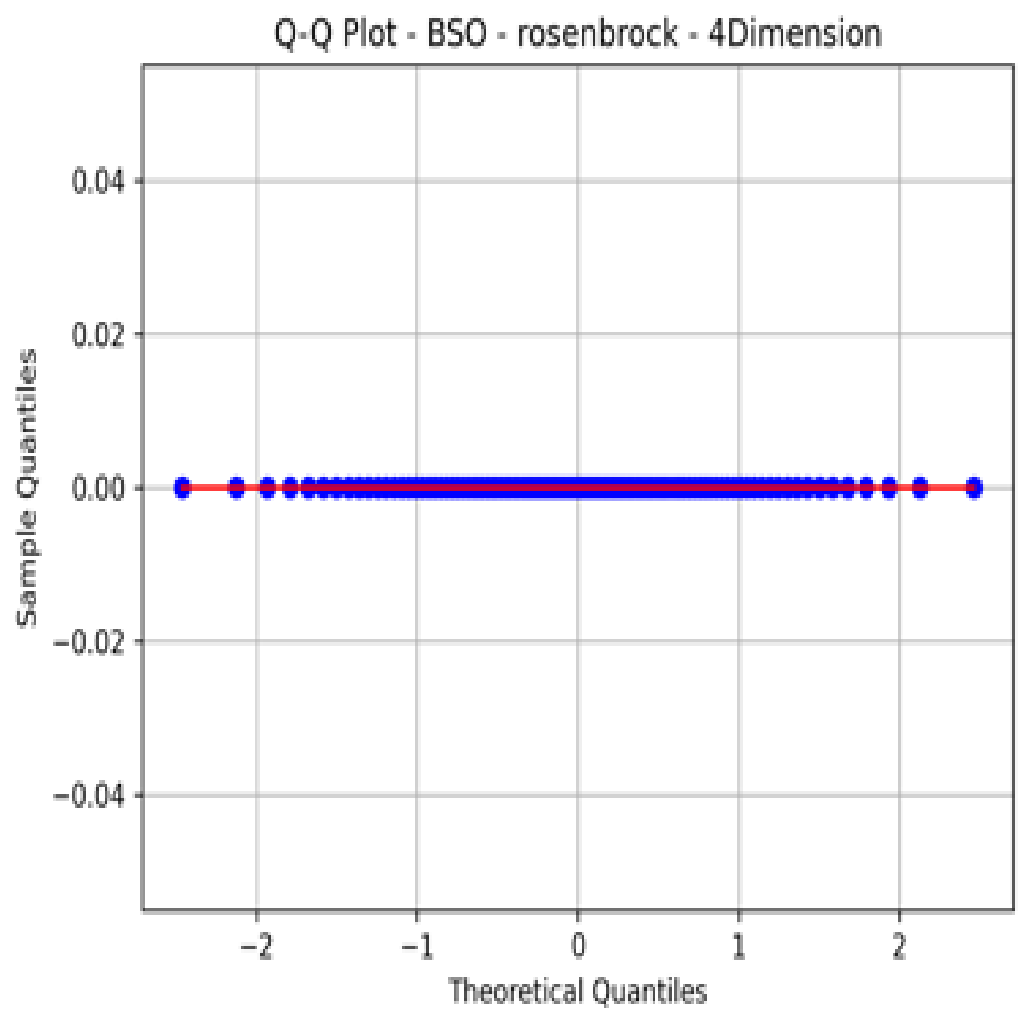
Q-Q Plot - BSO - rosenbrock - 4Dimension
Sample Quantiles
Theoretical Quantiles
0.04
0.02
0.00
−0.02
−0.04
−2
−1
0
1
2

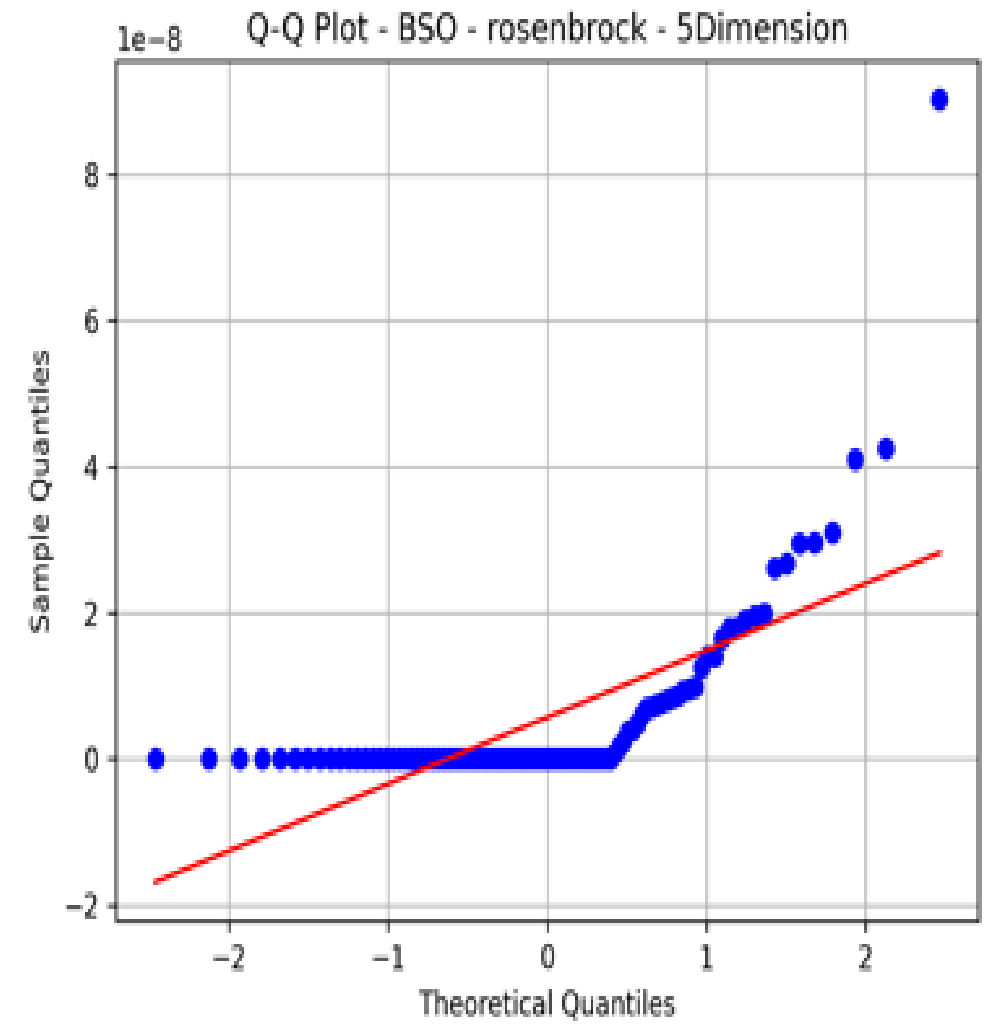
1e−8
Q-Q Plot - BSO - rosenbrock - 5Dimension
Sample Quantiles
Theoretical Quantiles
8
6
4
2
0
−2
−2
−1
0
1
2

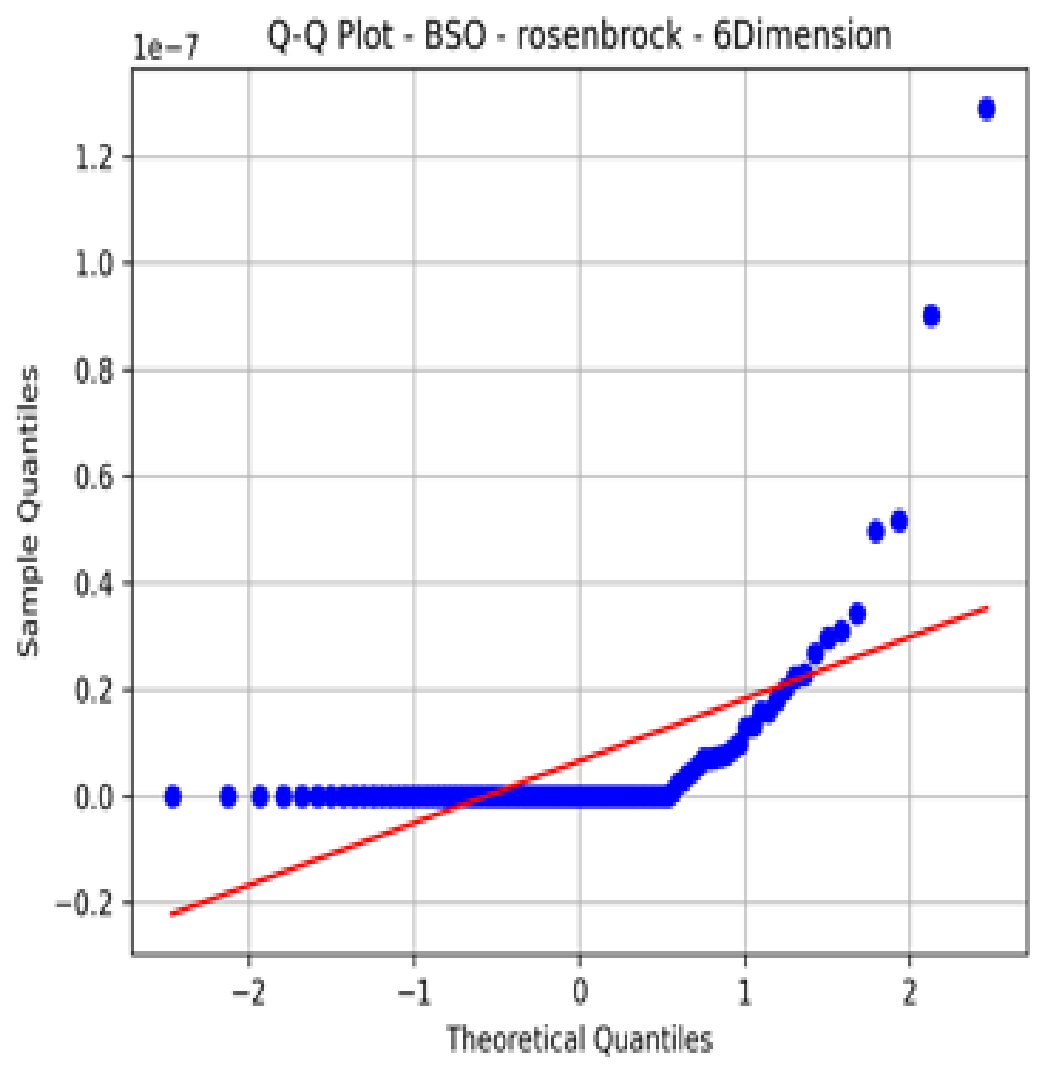
1e−7
Q-Q Plot - BSO - rosenbrock - 6Dimension
Sample Quantiles
Theoretical Quantiles
1.2
1.0
0.8
0.6
0.4
0.2
0.0
−0.2
−2
−1
0
1
2

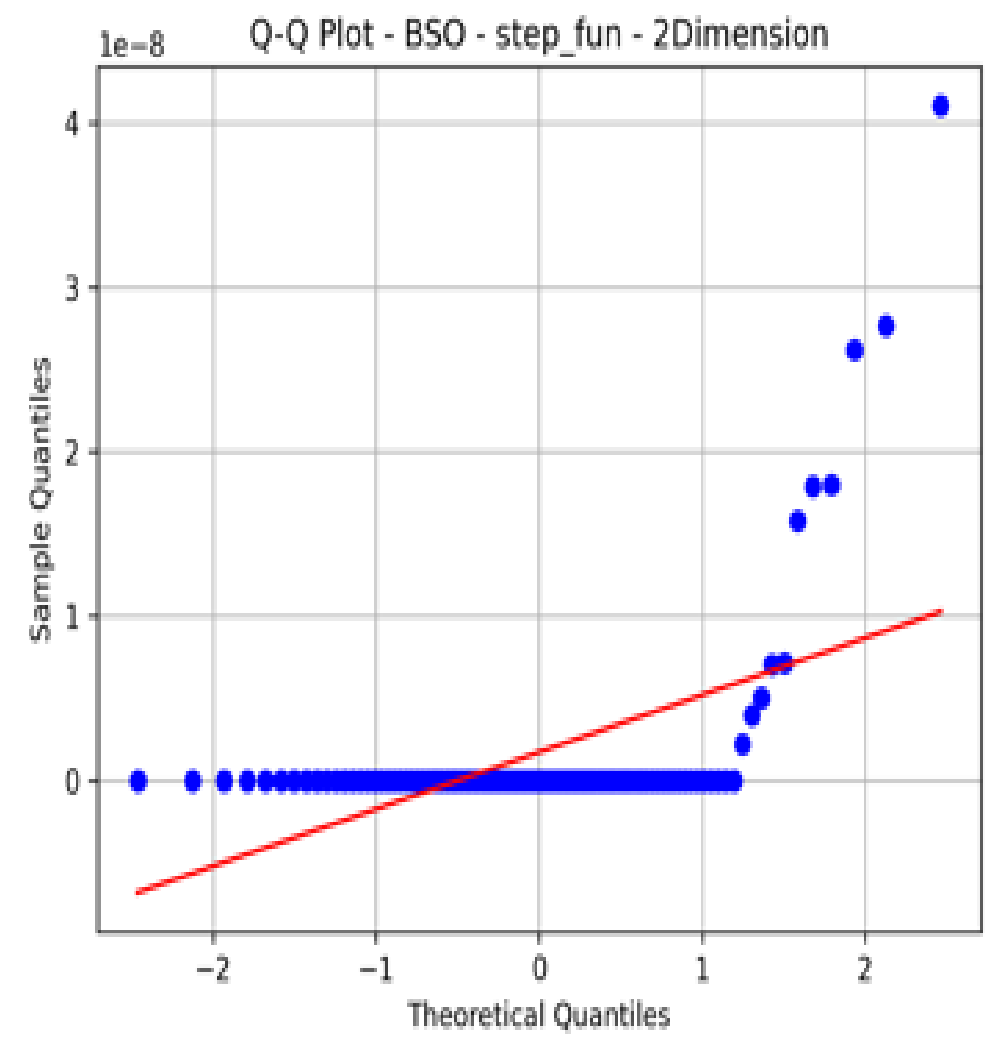
1e−8
Q-Q Plot - BSO - step_fun - 2Dimension
Sample Quantiles
Theoretical Quantiles
4
3
2
1
0
−2
−1
0
1
2

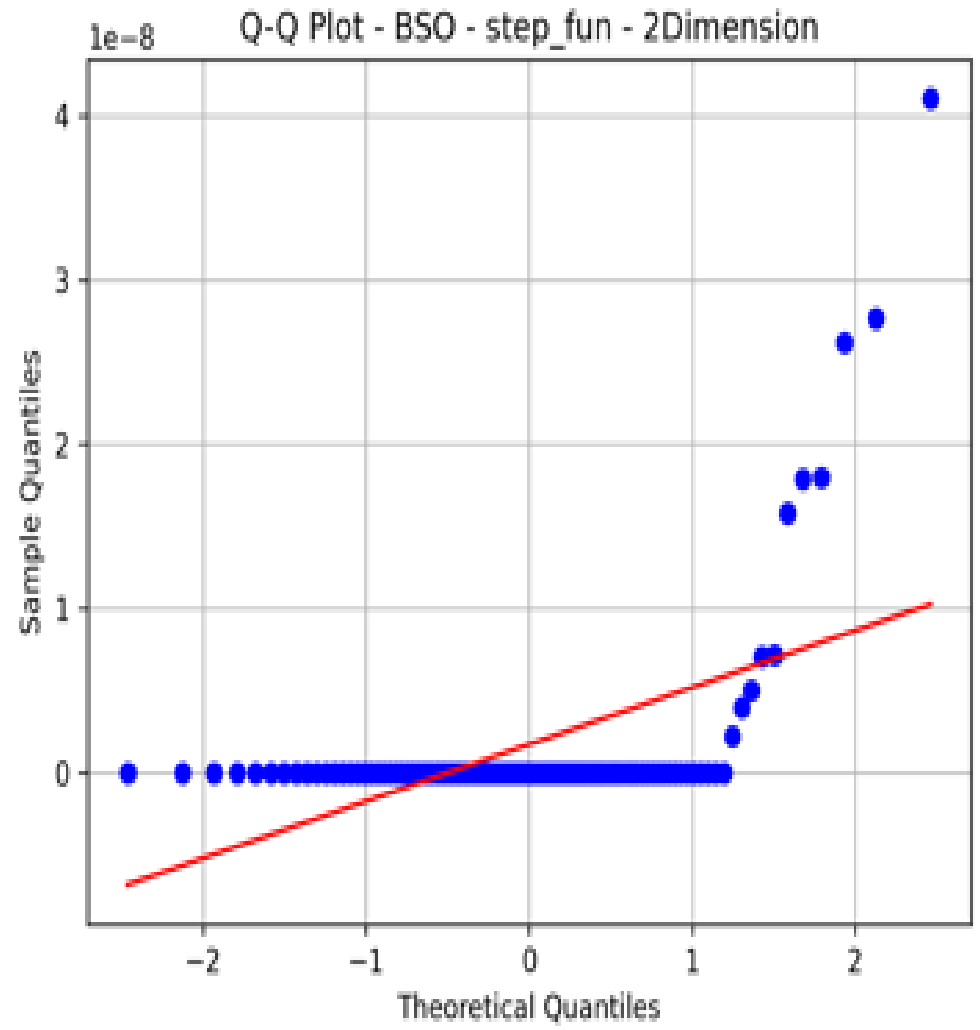
Q-Q Plot - BSO - step_fun - 2Dimension
1e−8
Sample Quantiles
Theoretical Quantiles

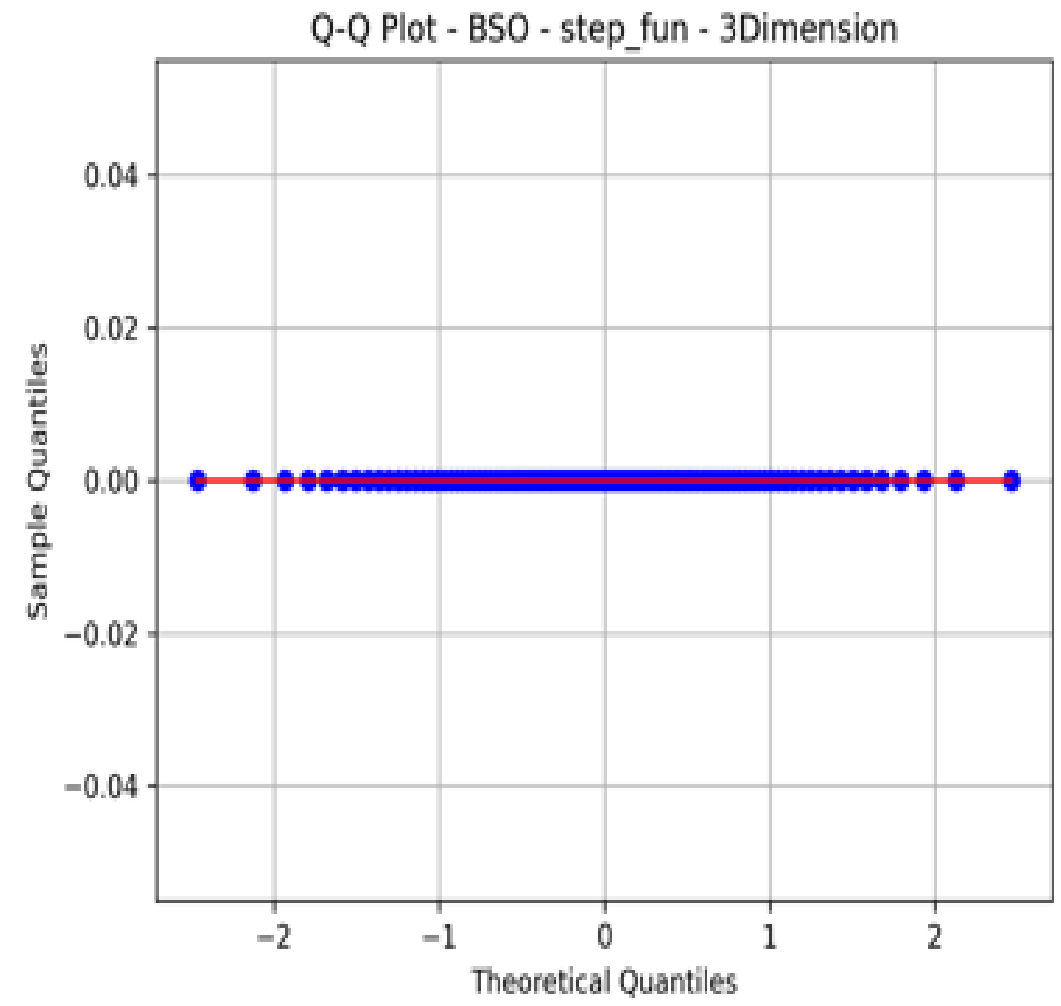
Q-Q Plot - BSO - step_fun - 3Dimension
Sample Quantiles
Theoretical Quantiles

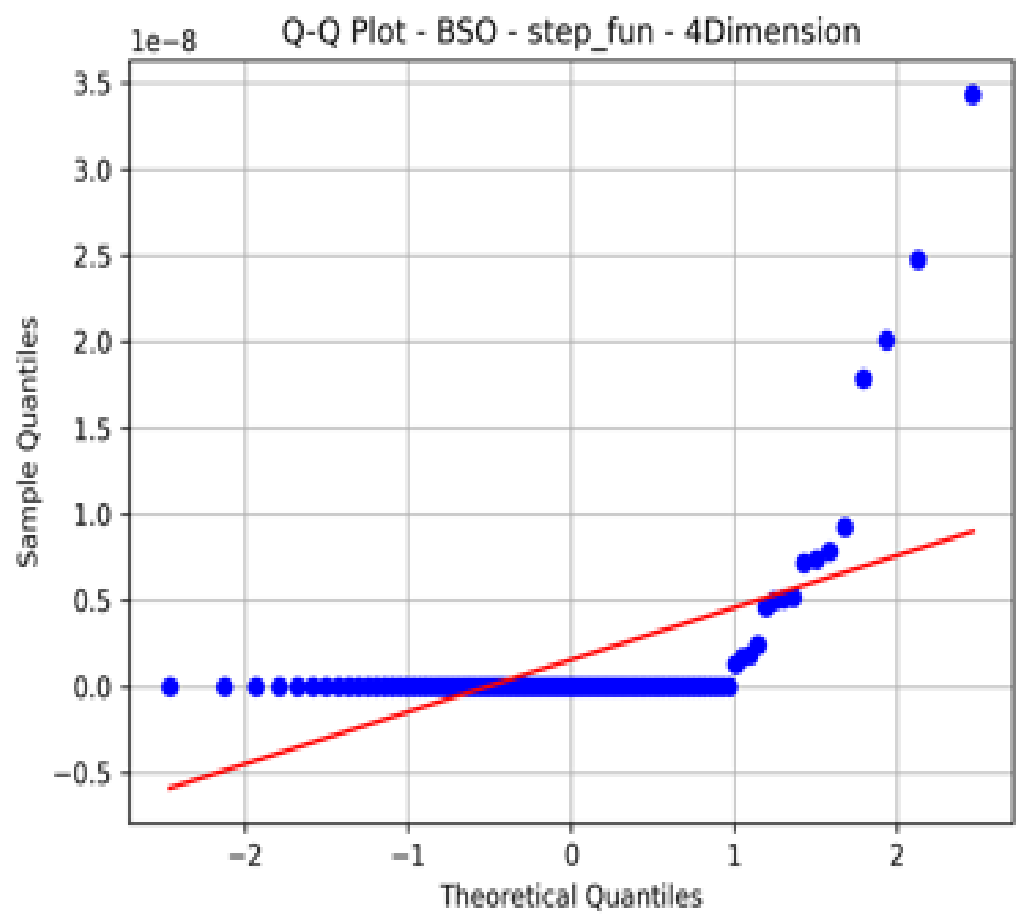
Q-Q Plot - BSO - step_fun - 4Dimension
1e−8
Sample Quantiles
Theoretical Quantiles

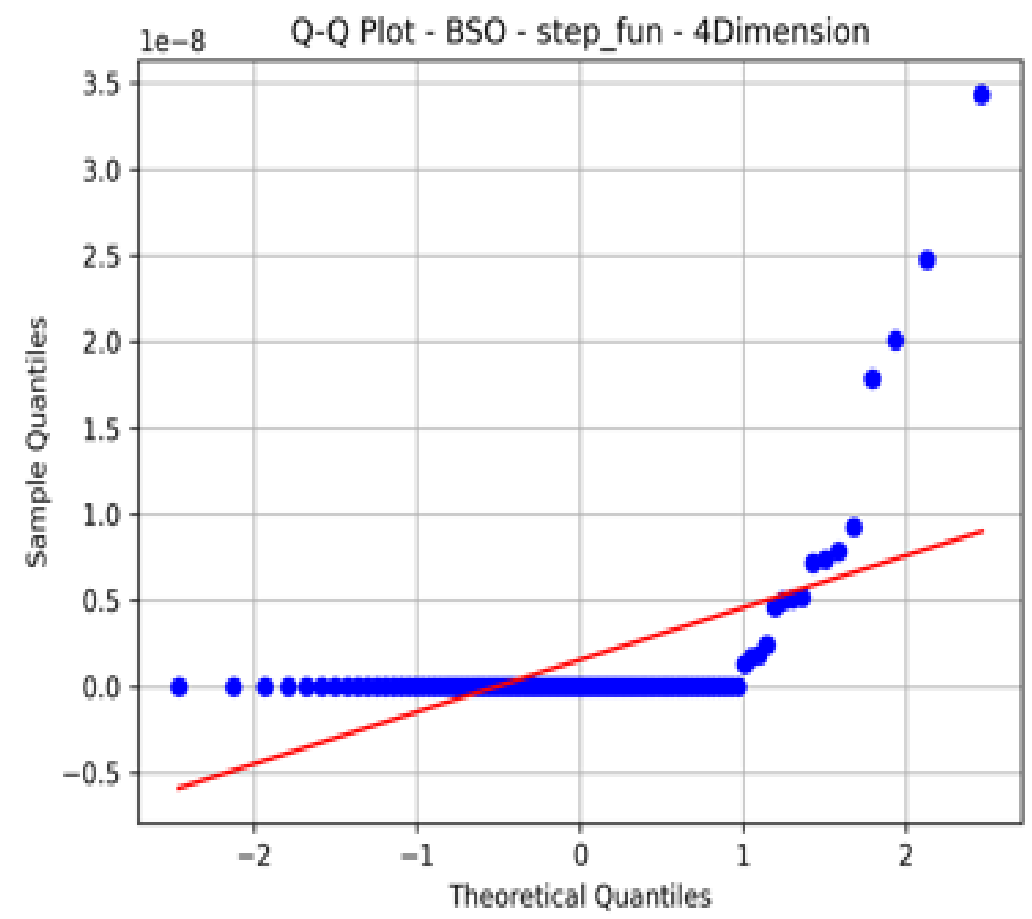
Q-Q Plot - BSO - step_fun - 4Dimension
1e−8
Sample Quantiles
Theoretical Quantiles

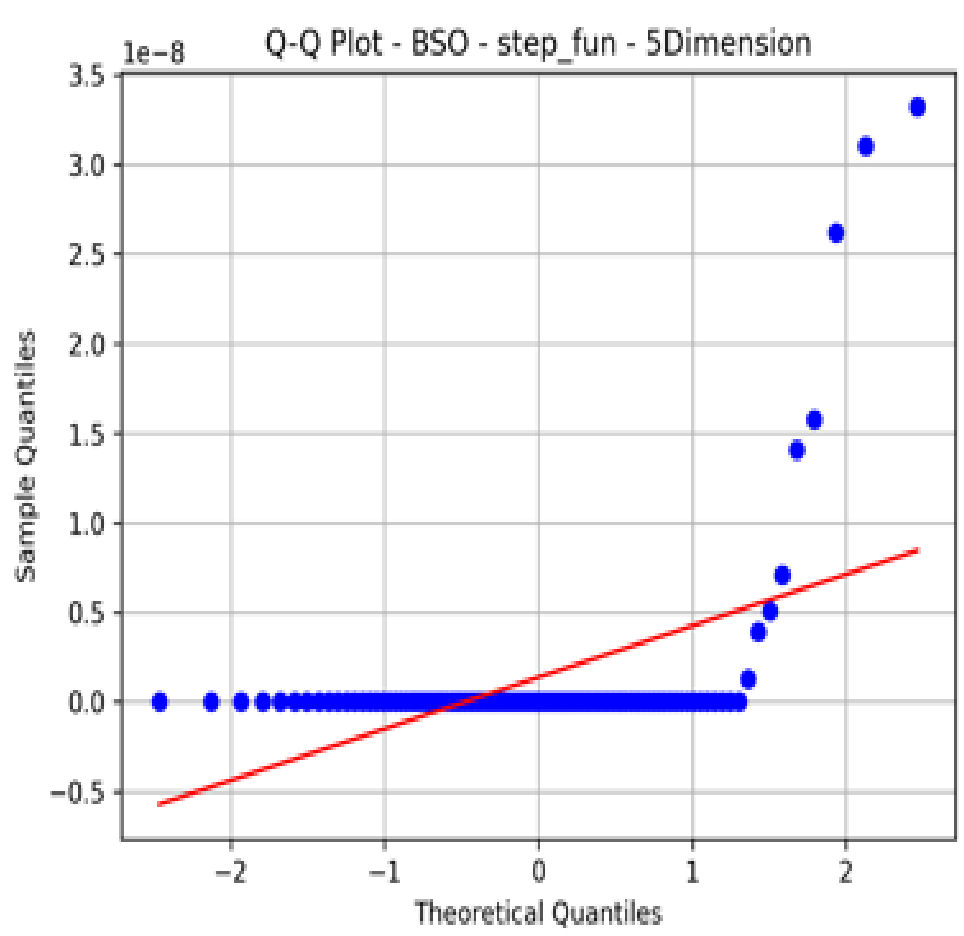
Q-Q Plot - BSO - step_fun - 5Dimension
1e−8
Sample Quantiles
Theoretical Quantiles

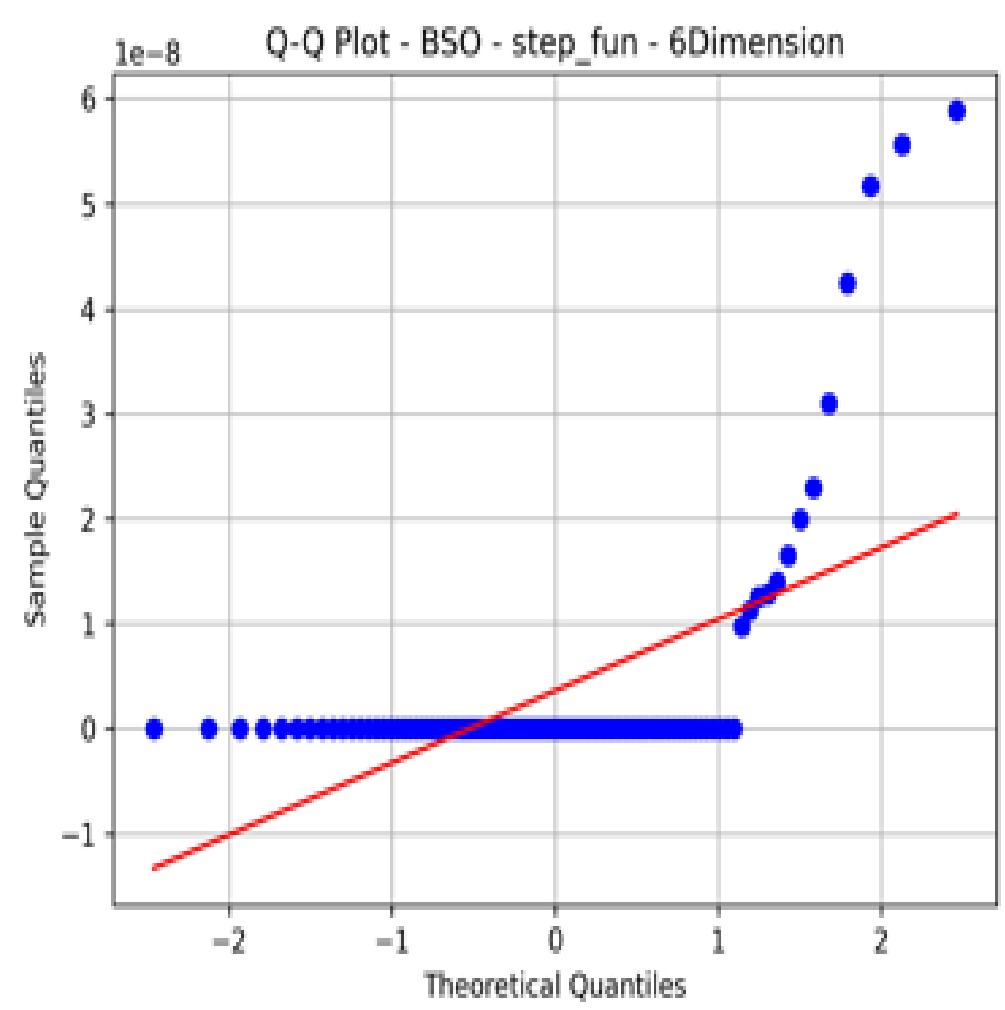
Q-Q Plot - BSO - step_fun - 6Dimension
1e−8
Sample Quantiles
Theoretical Quantiles

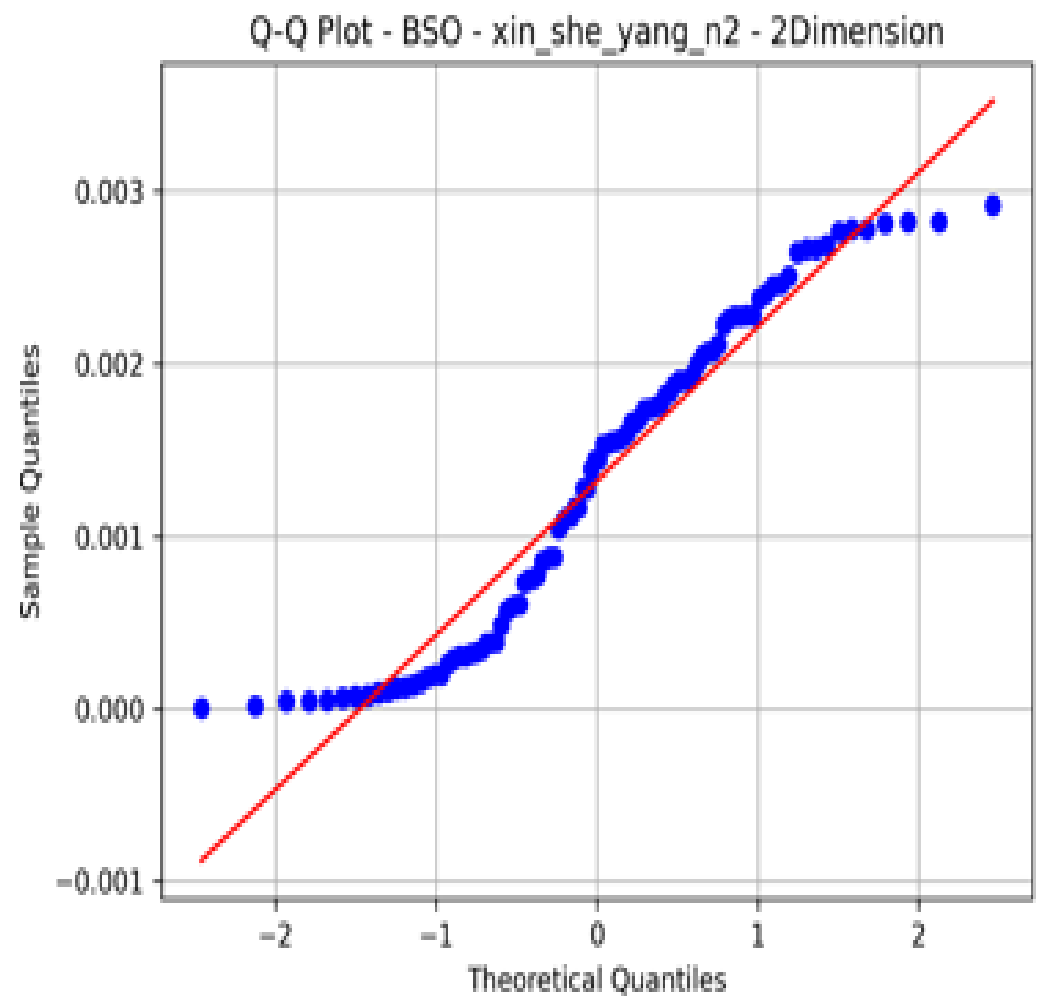
Q-Q Plot - BSO - xin_she_yang_n2 - 2Dimension
Sample Quantiles
Theoretical Quantiles
0.003
0.002
0.001
0.000
−0.001
−2
−1
0
1
2

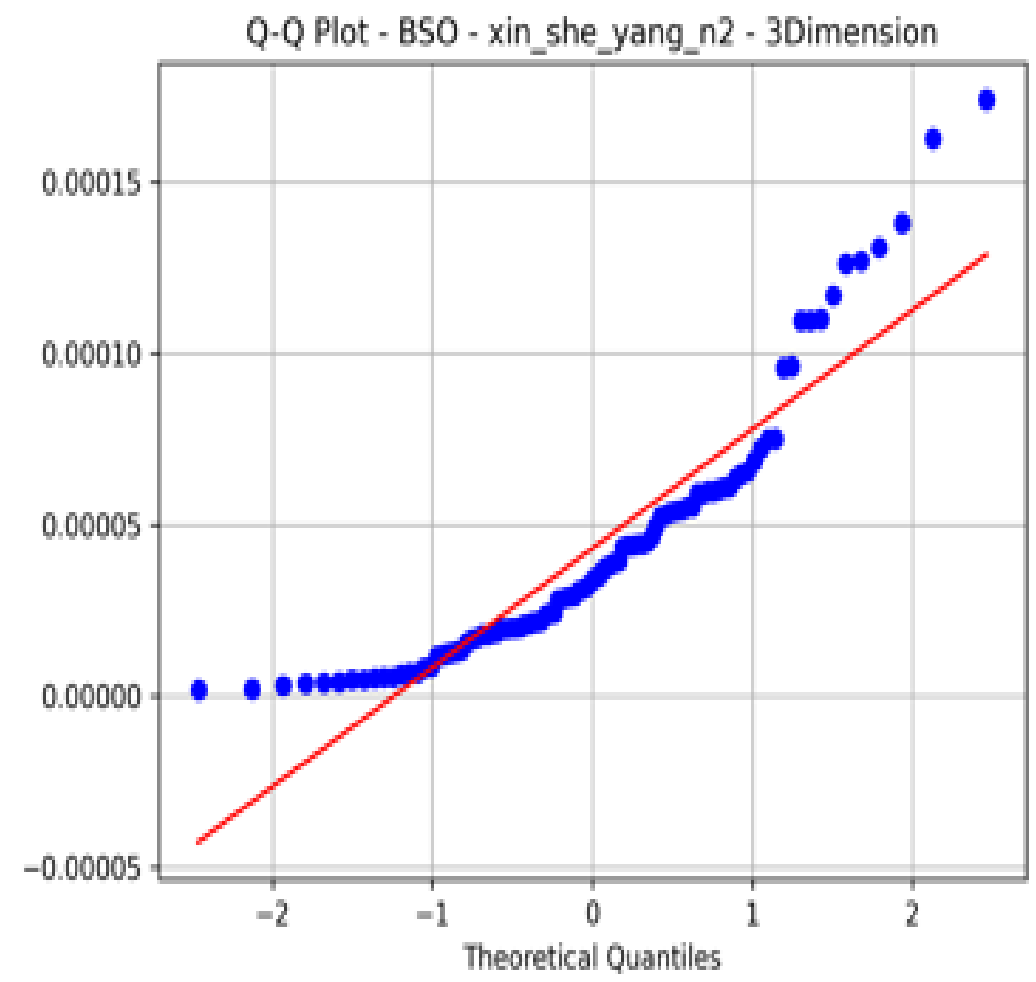
Q-Q Plot - BSO - xin_she_yang_n2 - 3Dimension
Theoretical Quantiles
0.00015
0.00010
0.00005
0.00000
−0.00005
−2
−1
0
1
2

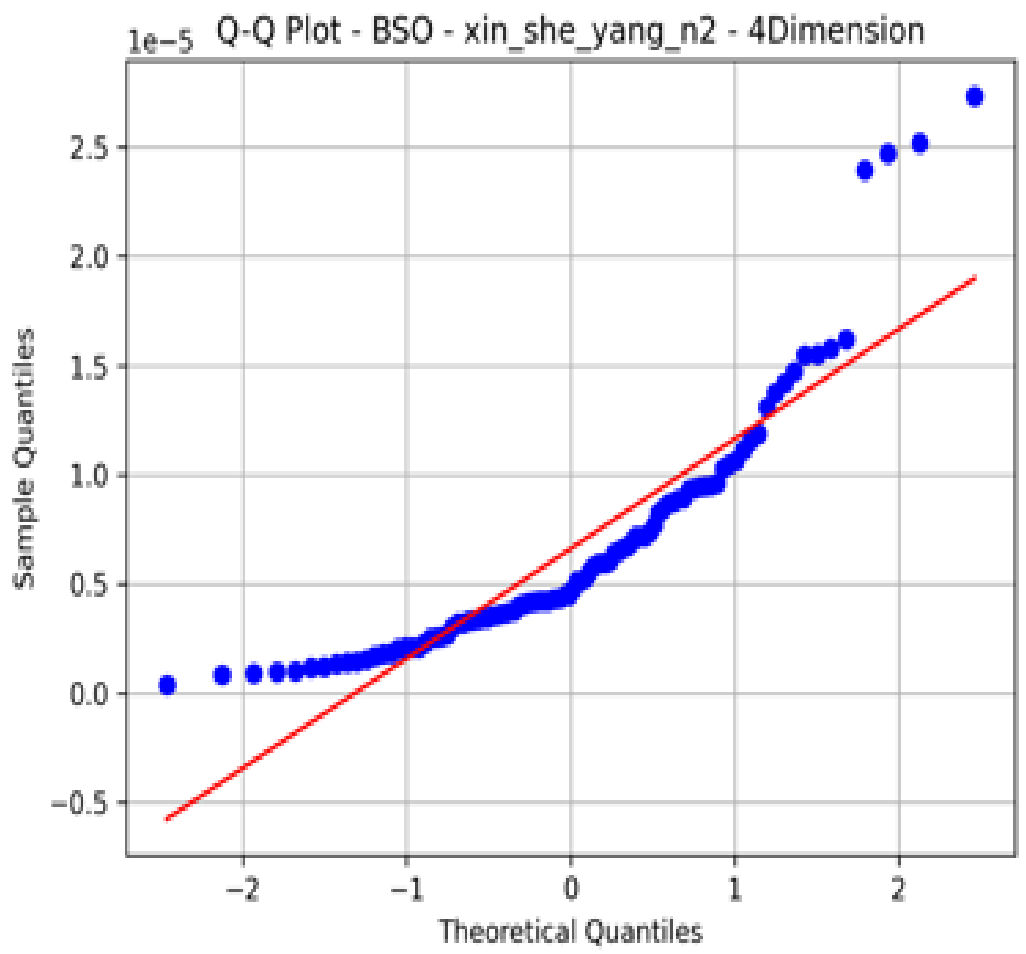
1e−5
Q-Q Plot - BSO - xin_she_yang_n2 - 4Dimension
Sample Quantiles
Theoretical Quantiles
2.5
2.0
1.5
1.0
0.5
0.0
−0.5
−2
−1
0
1
2

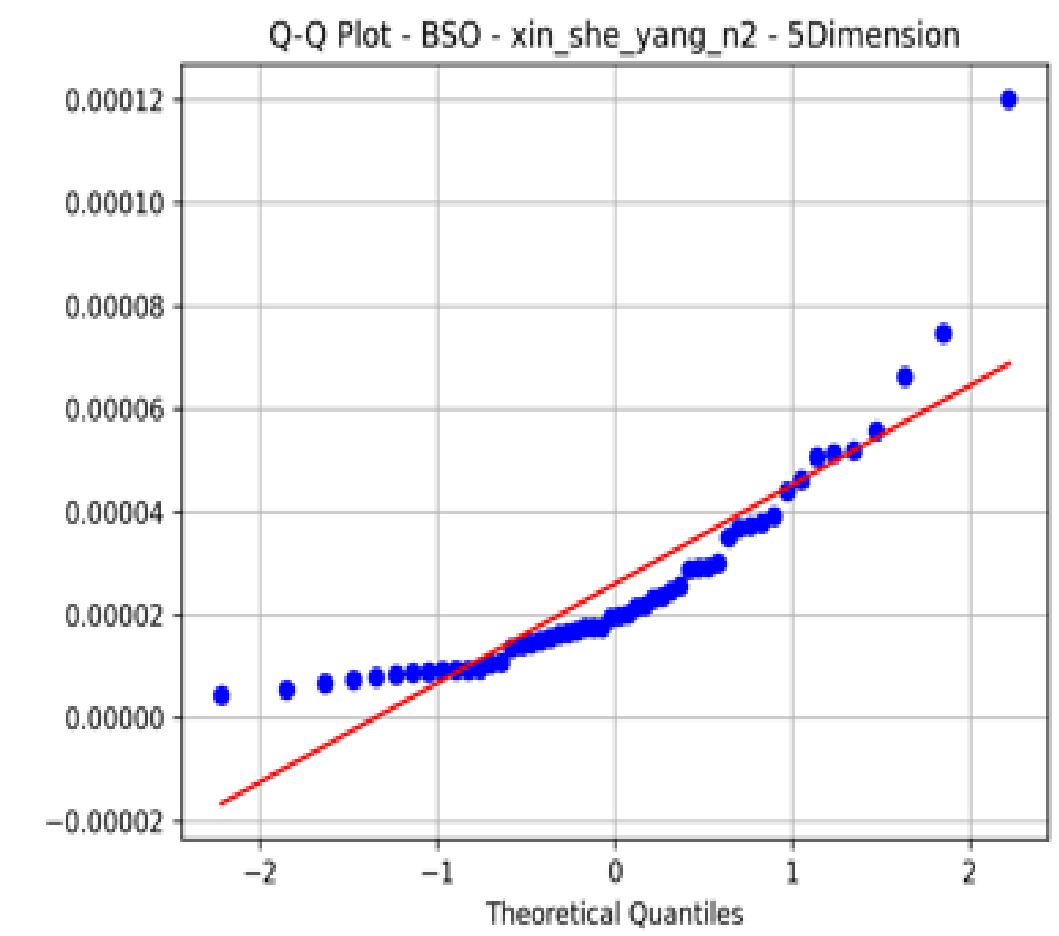
Q-Q Plot - BSO - xin_she_yang_n2 - 5Dimension
Theoretical Quantiles
0.00012
0.00010
0.00008
0.00006
0.00004
0.00002
0.00000
−0.00002
−2
−1
0
1
2

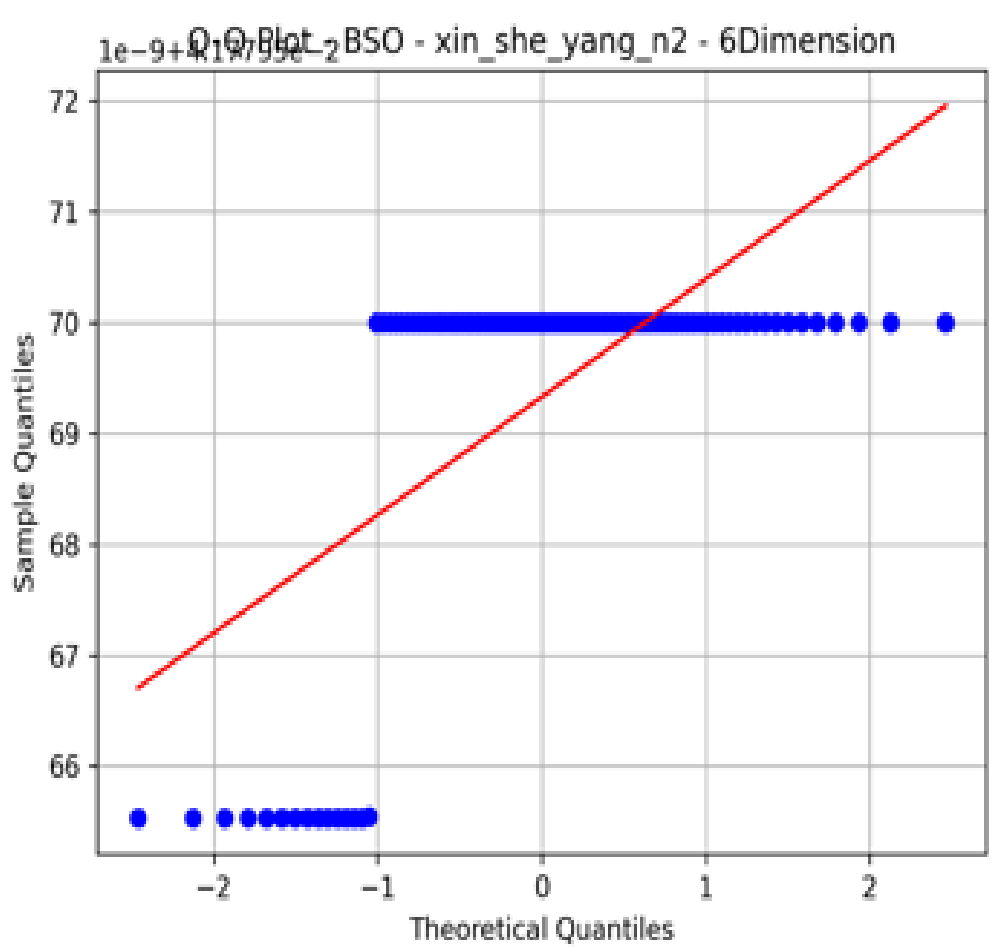
Q-Q Plot - BSO - xin_she_yang_n2 - 6Dimension
Sample Quantiles
Theoretical Quantiles
72
71
70
69
68
67
66
−2
−1
0
1
2

<u>BPO Q-Q Plots</u>

## BPO Continuous Functions Q-Q Plots

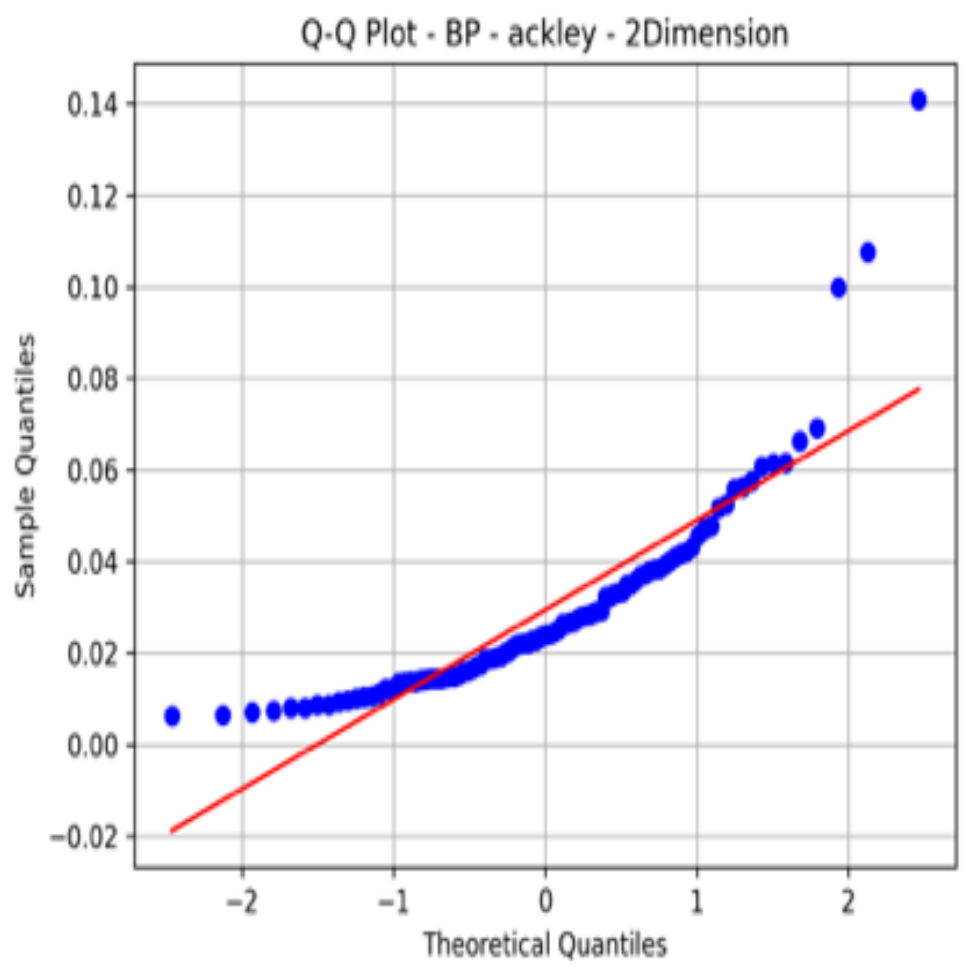

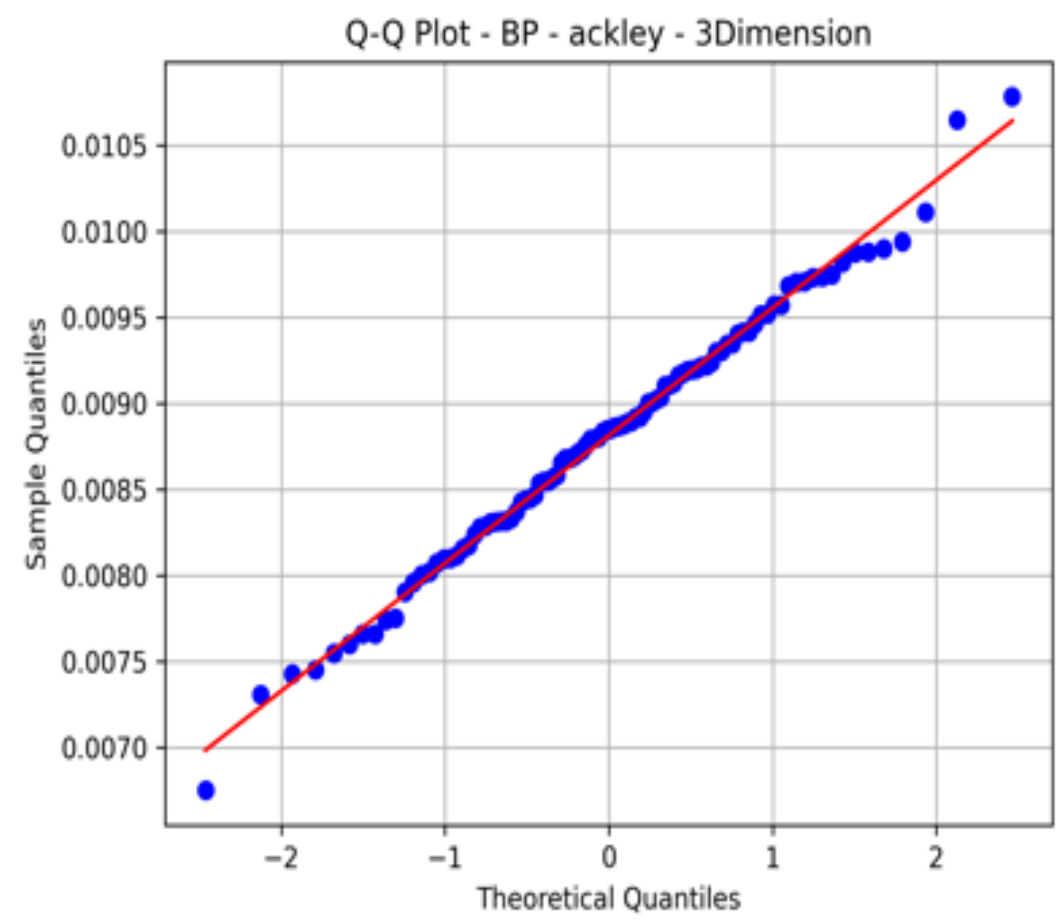

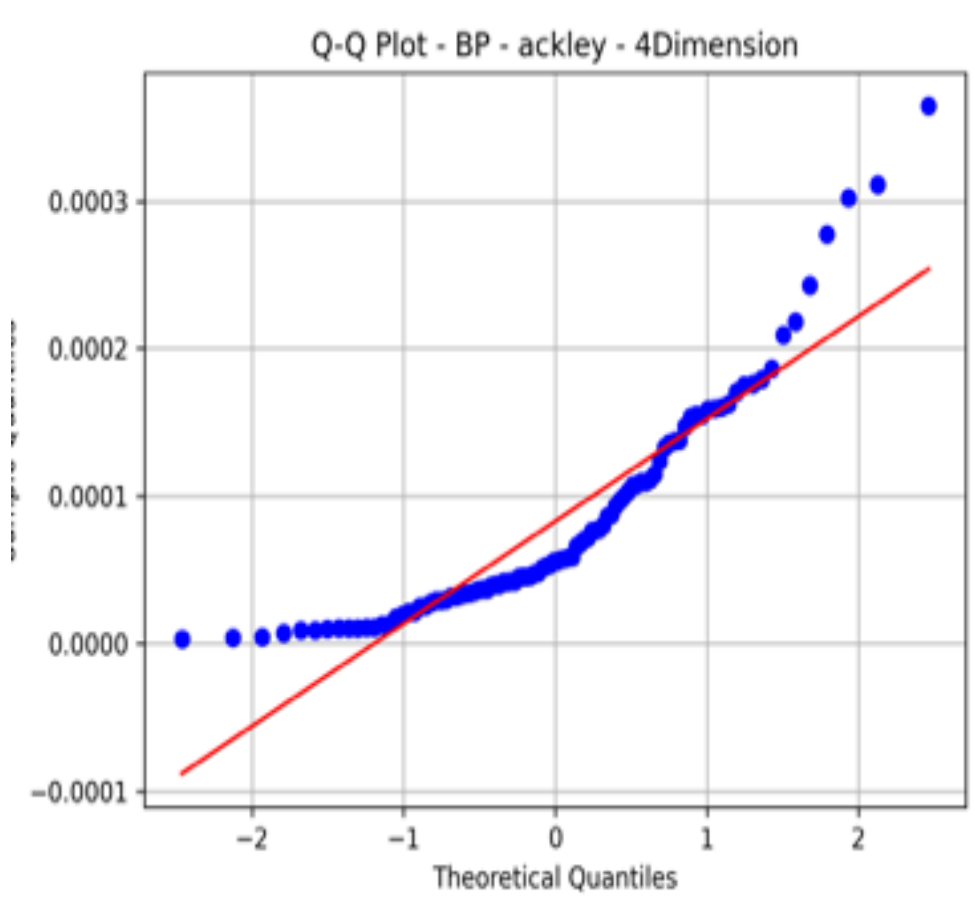

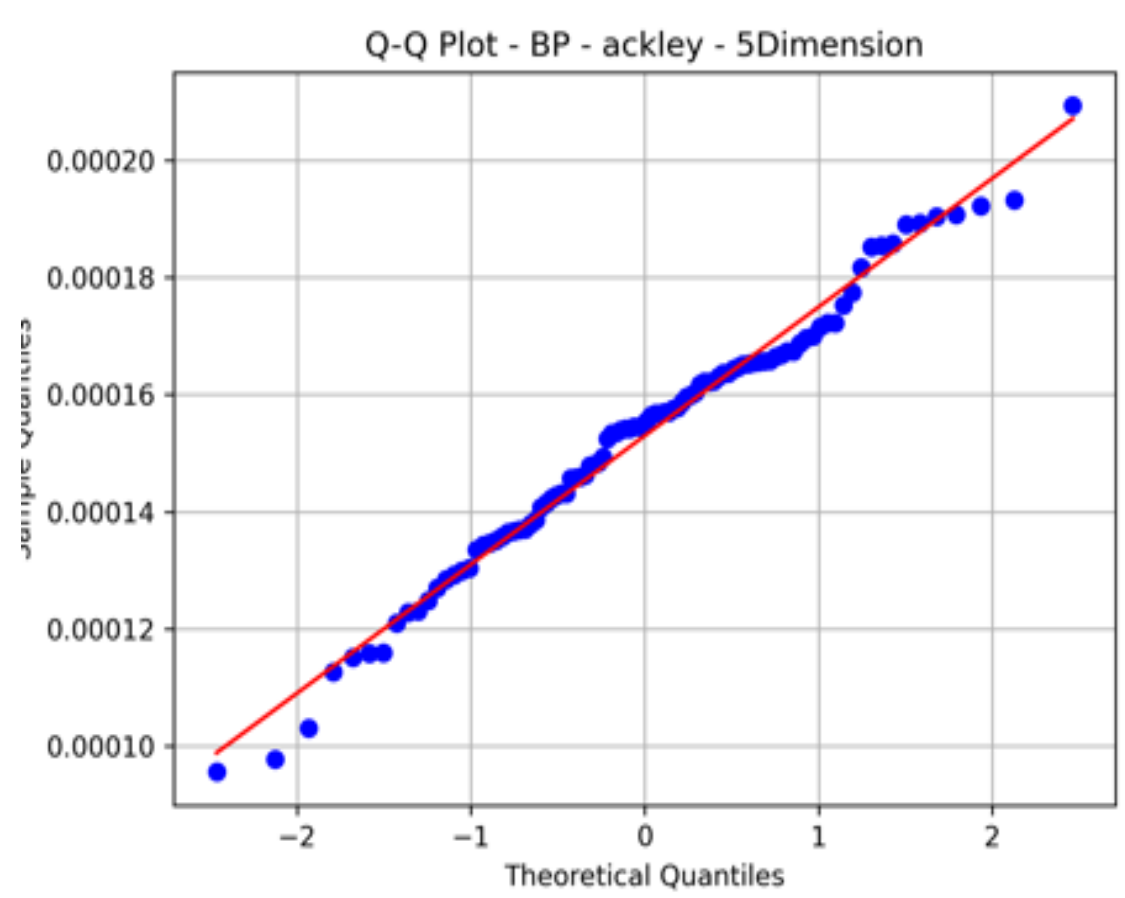

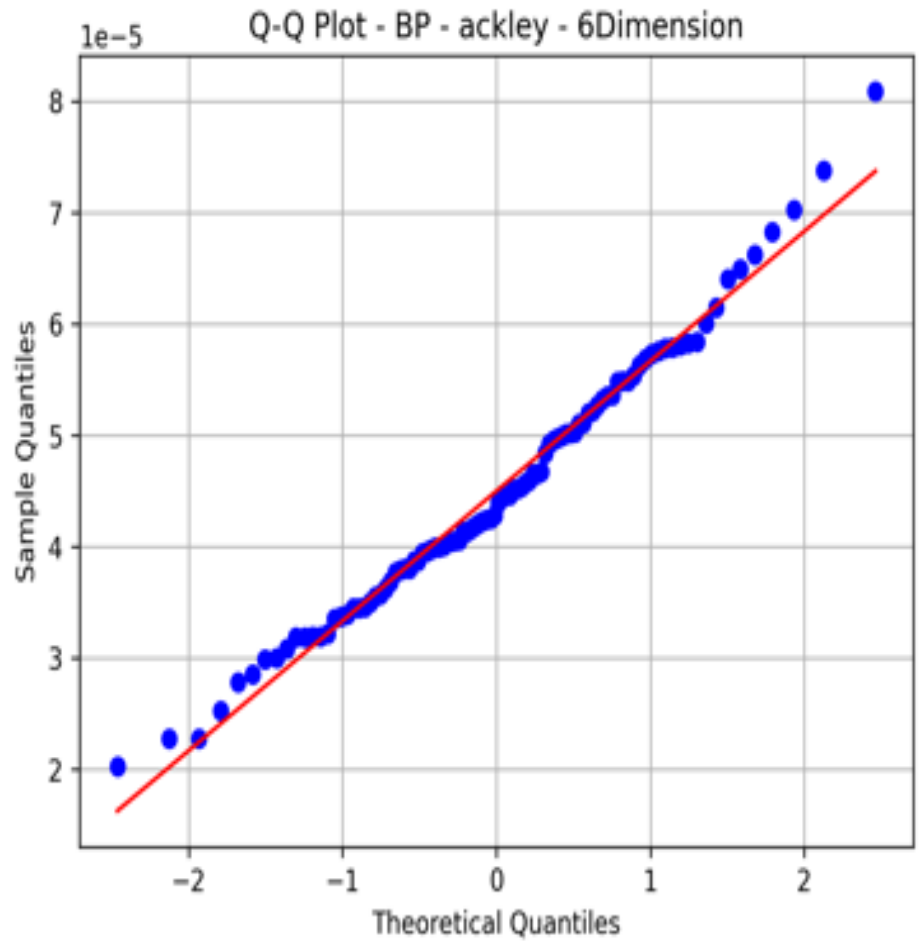
1e−5
Q-Q Plot - BP - ackley - 6Dimension
Sample Quantiles
Theoretical Quantiles

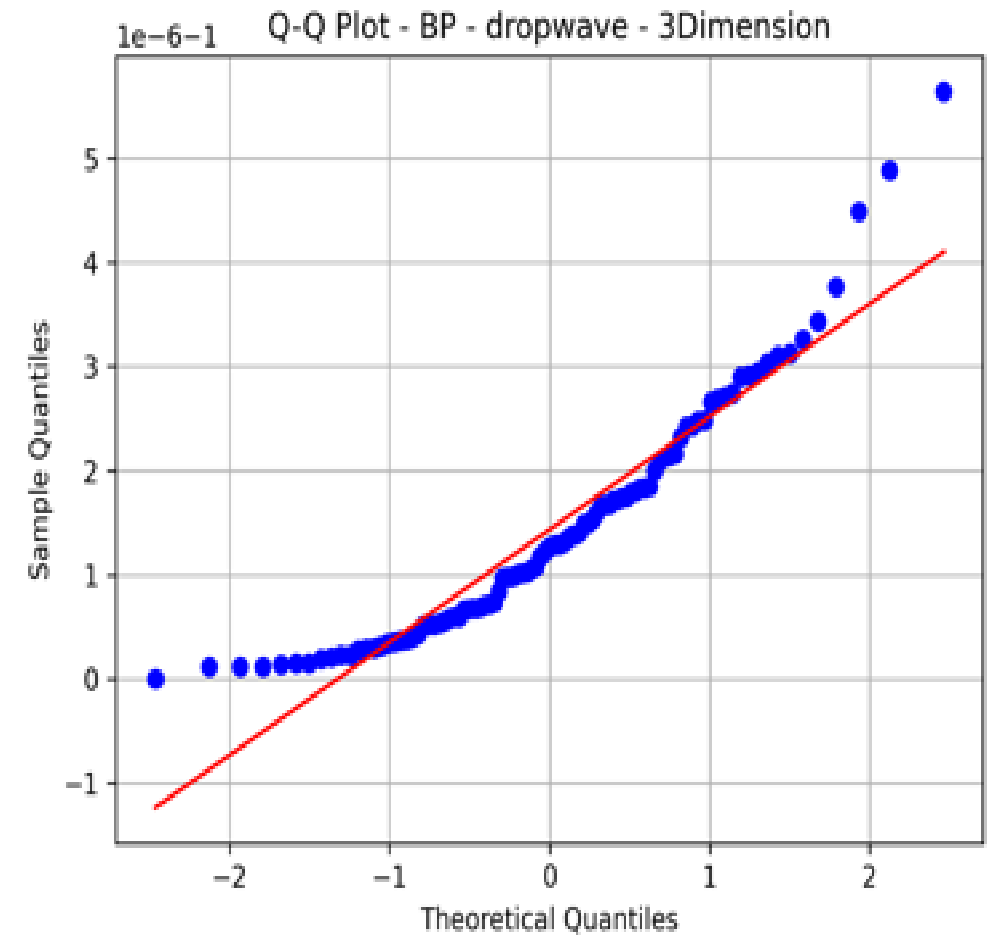
1e−6−1
Q-Q Plot - BP - dropwave - 3Dimension
Sample Quantiles
Theoretical Quantiles

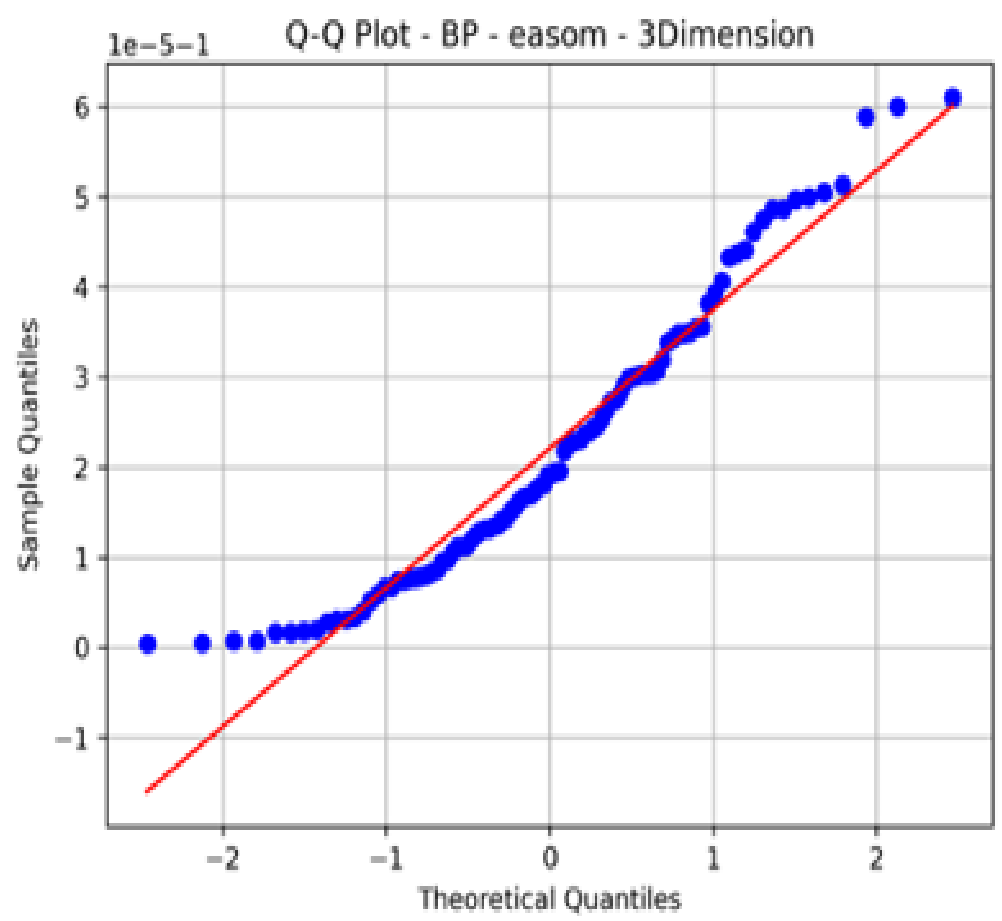
1e−5−1
Q-Q Plot - BP - easom - 3Dimension
Sample Quantiles
Theoretical Quantiles

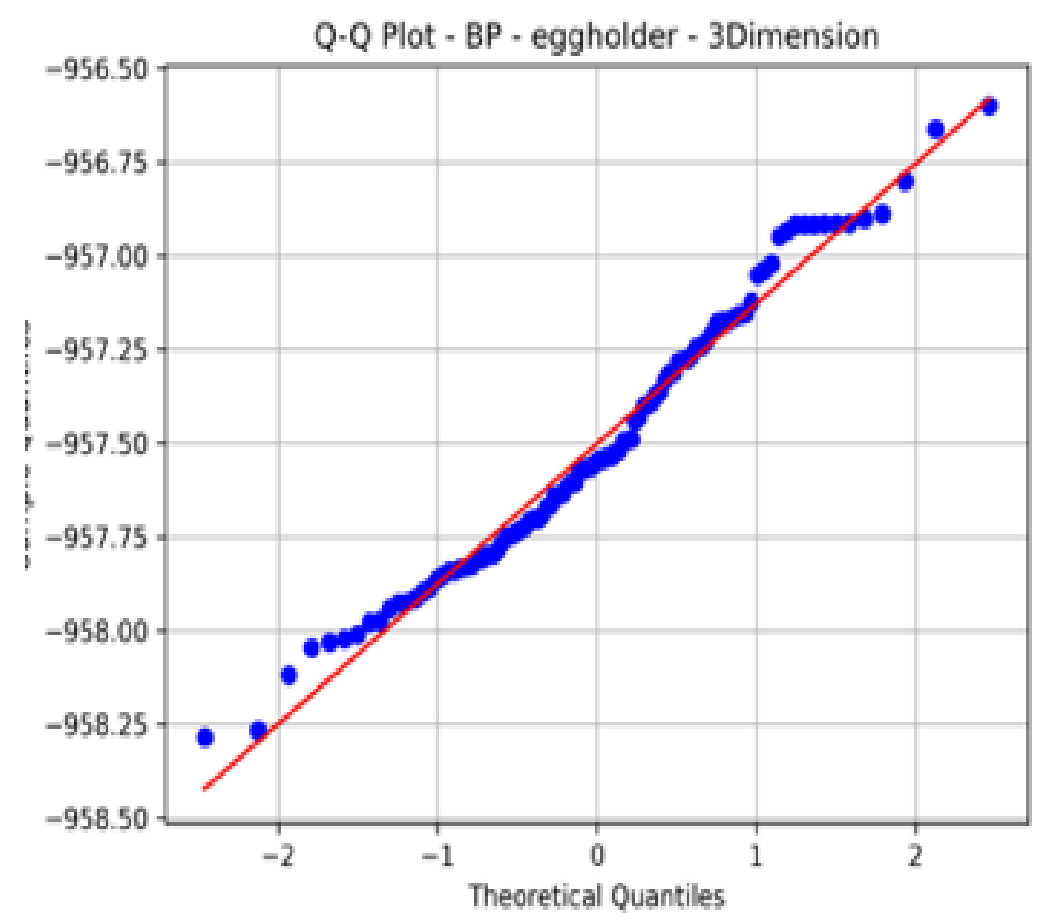
Q-Q Plot - BP - eggholder - 3Dimension
Theoretical Quantiles

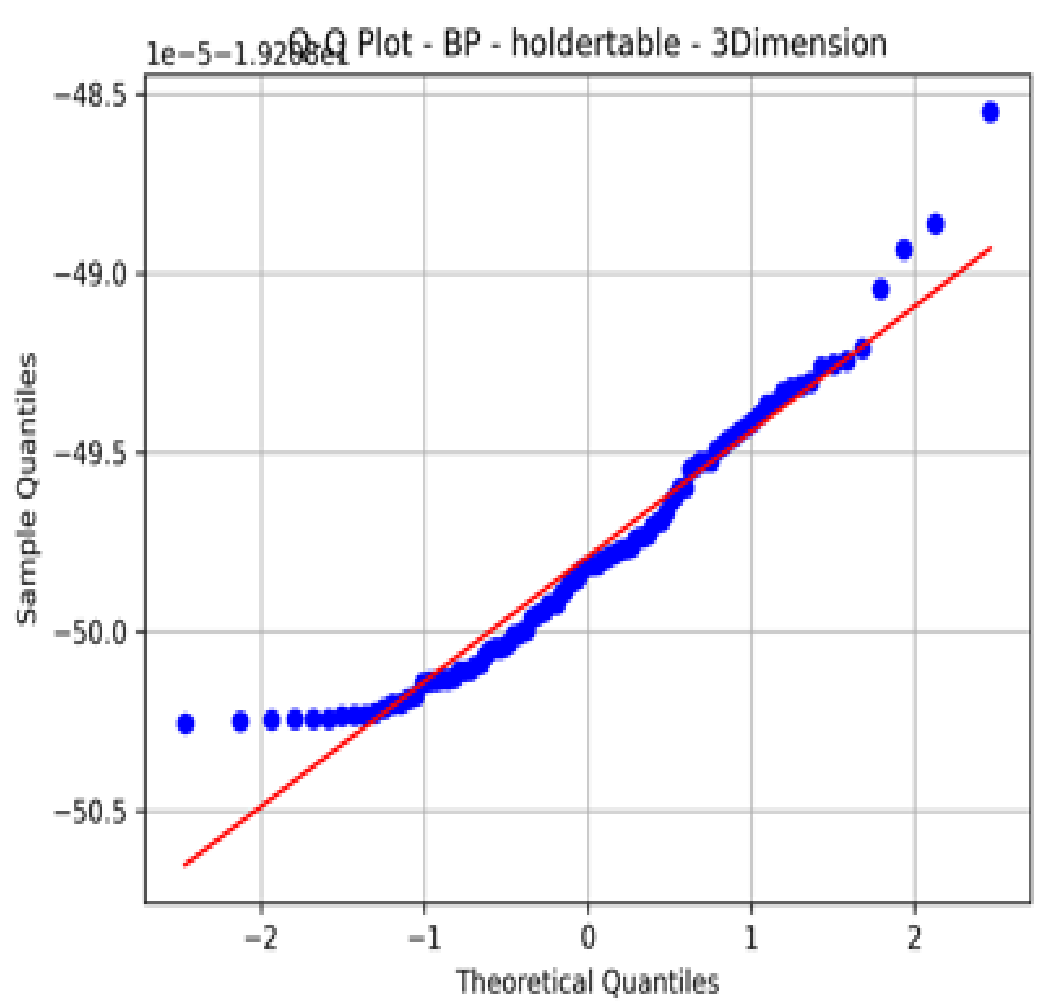
1e−5−1.9208e1
Q-Q Plot - BP - holdertable - 3Dimension
Sample Quantiles
Theoretical Quantiles

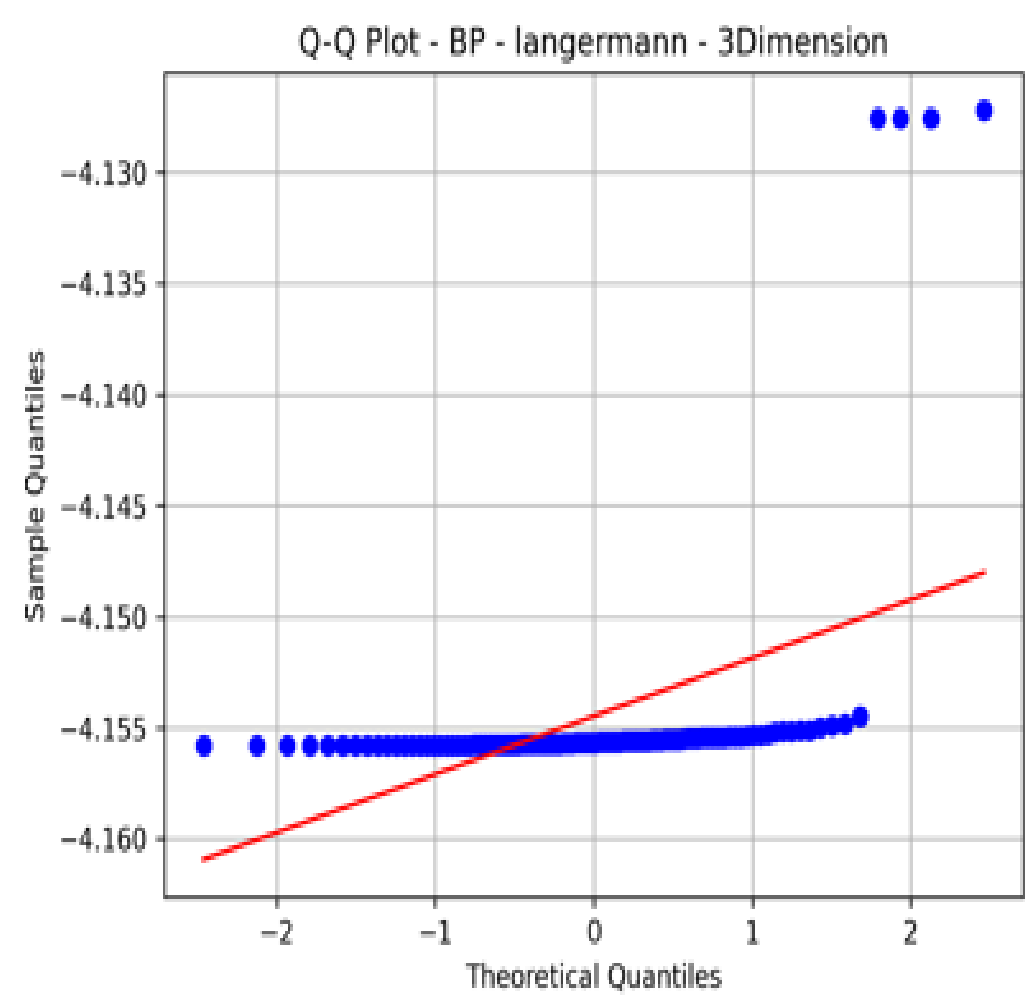
Q-Q Plot - BP - langermann - 3Dimension
Sample Quantiles
Theoretical Quantiles

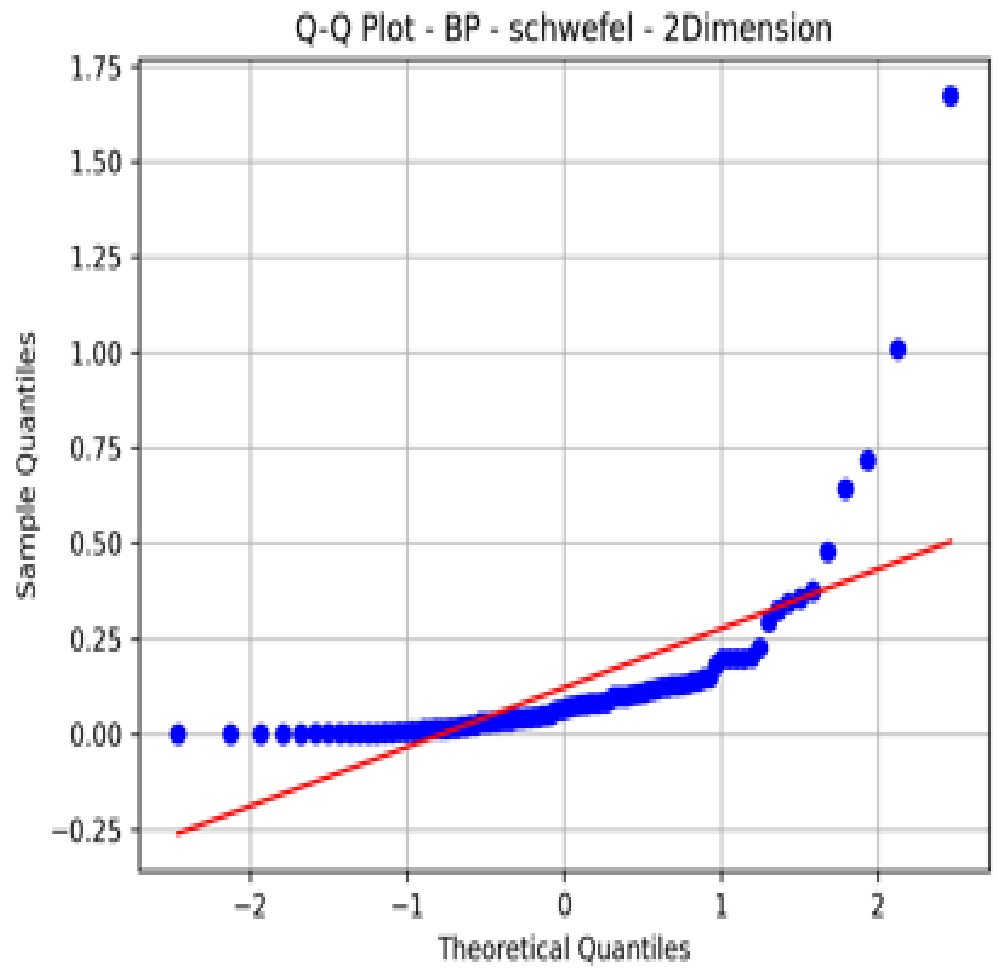
Q-Q Plot - BP - schwefel - 2Dimension
Sample Quantiles
Theoretical Quantiles

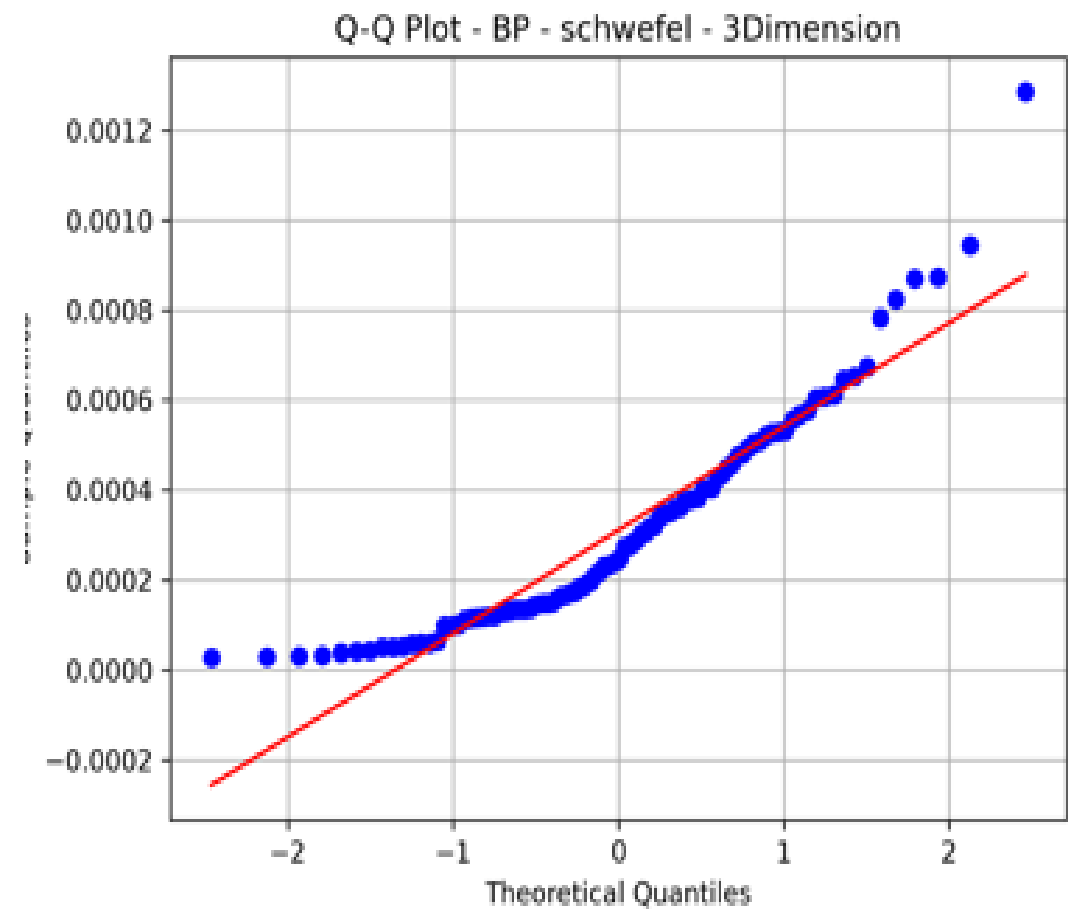
Q-Q Plot - BP - schwefel - 3Dimension
Theoretical Quantiles

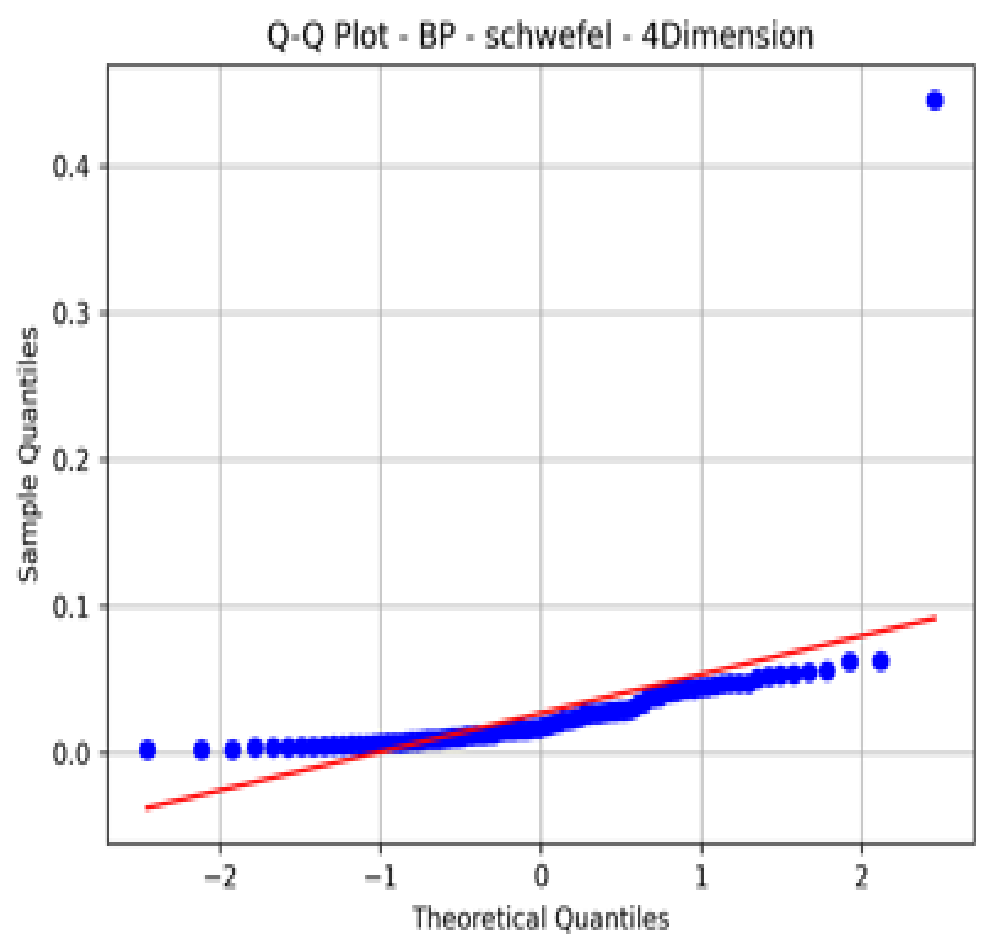
Q-Q Plot - BP - schwefel - 4Dimension
Sample Quantiles
Theoretical Quantiles

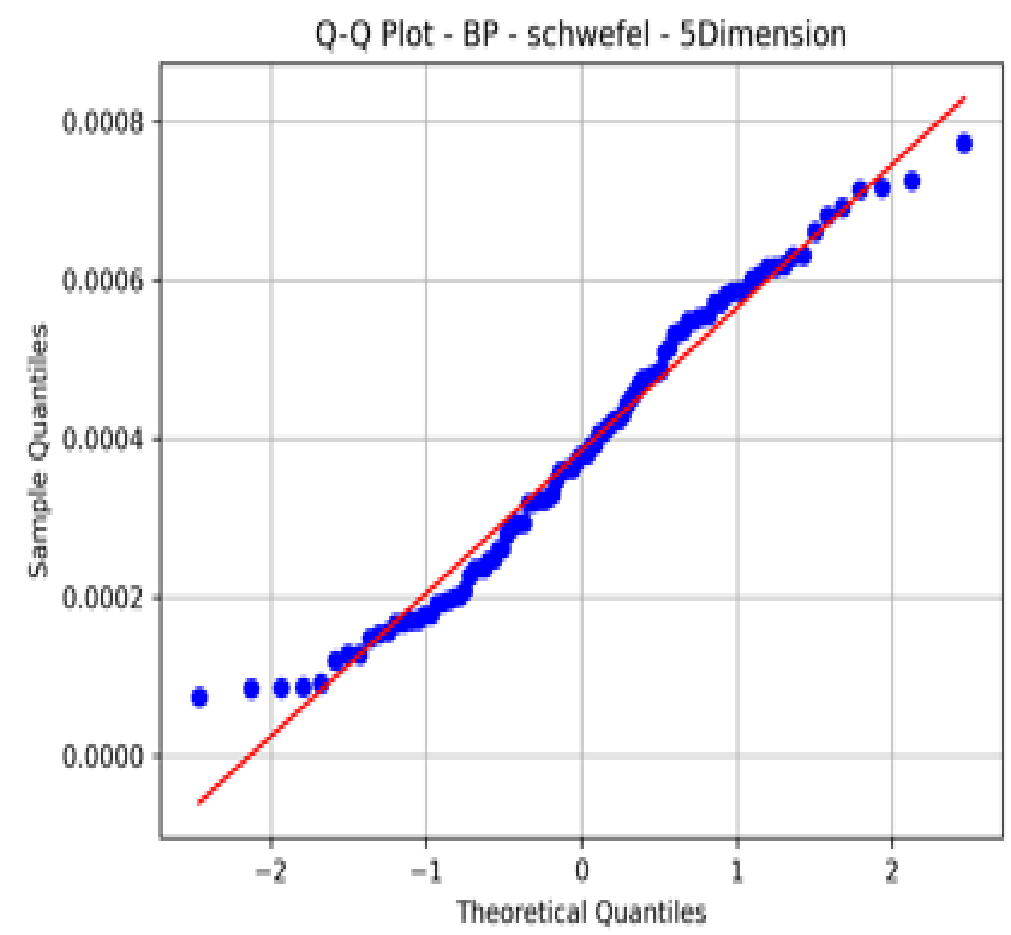
Q-Q Plot - BP - schwefel - 5Dimension
Sample Quantiles
Theoretical Quantiles

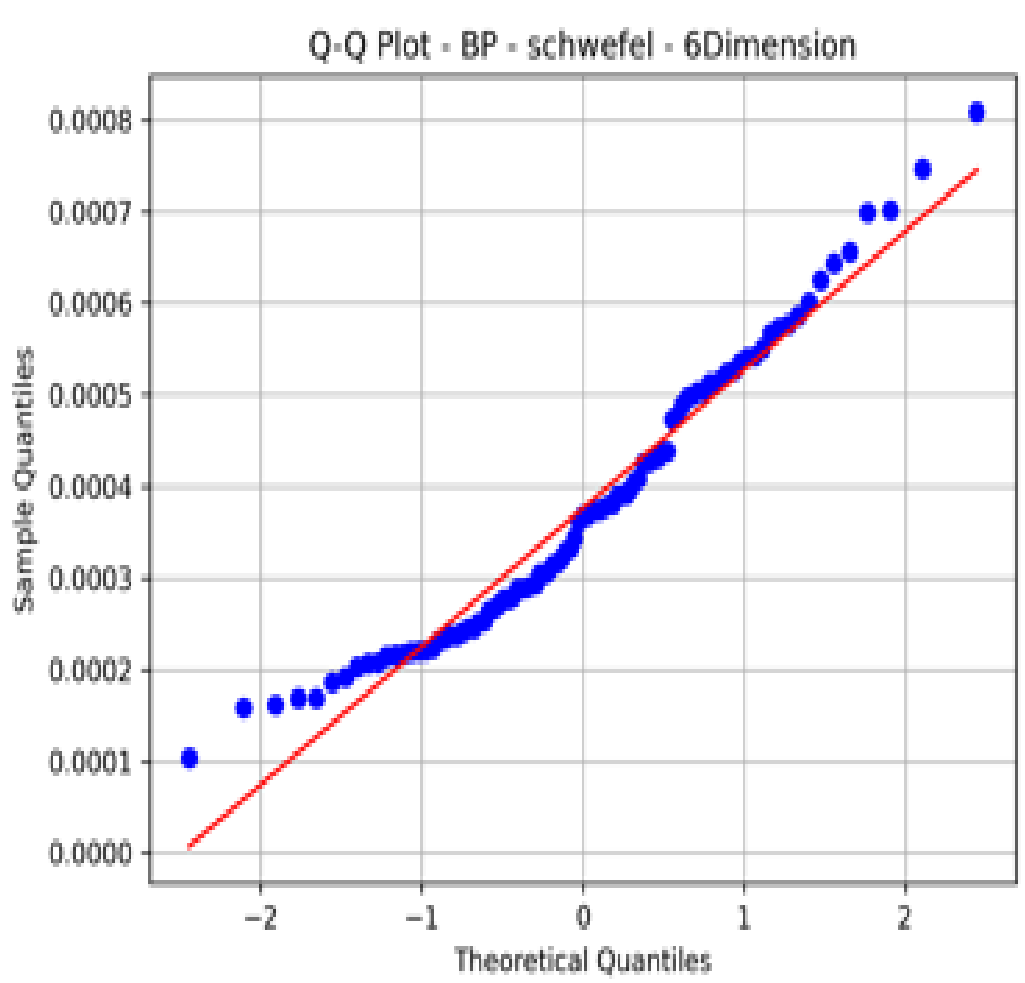
Q-Q Plot - BP - schwefel - 6Dimension
Sample Quantiles
Theoretical Quantiles

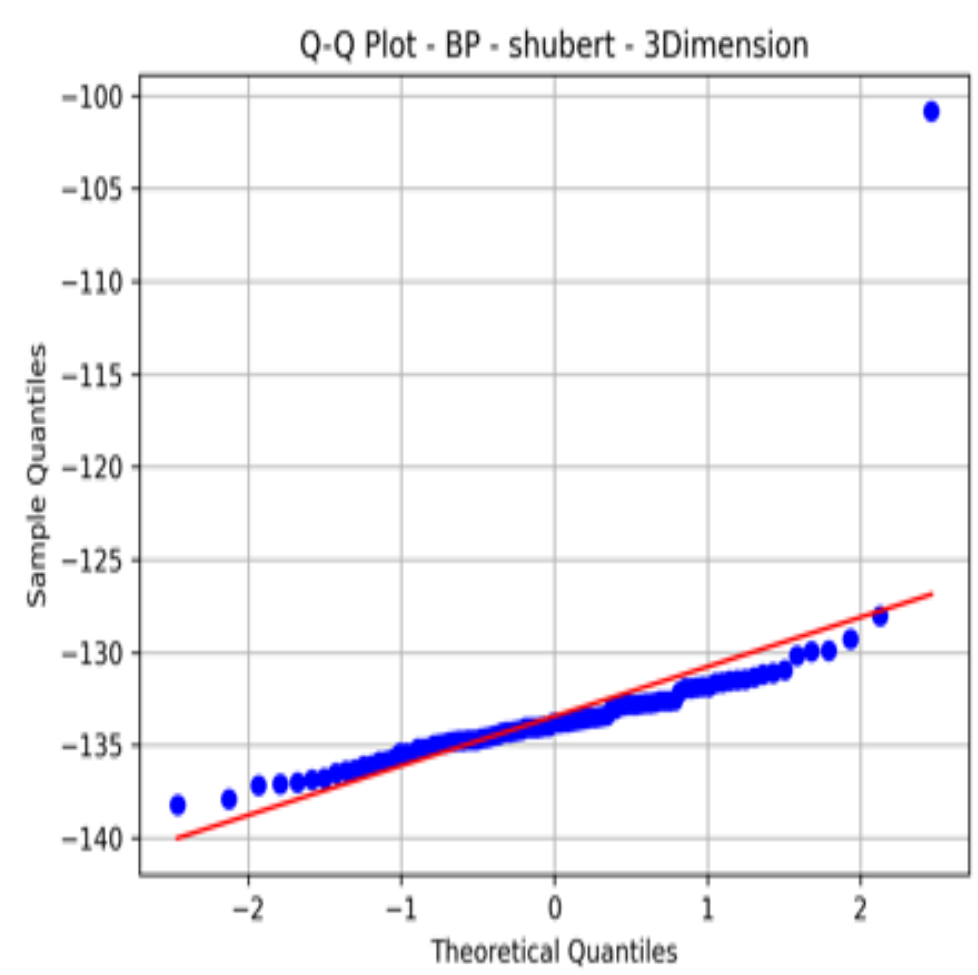
Q-Q Plot - BP - shubert - 3Dimension
Sample Quantiles
Theoretical Quantiles

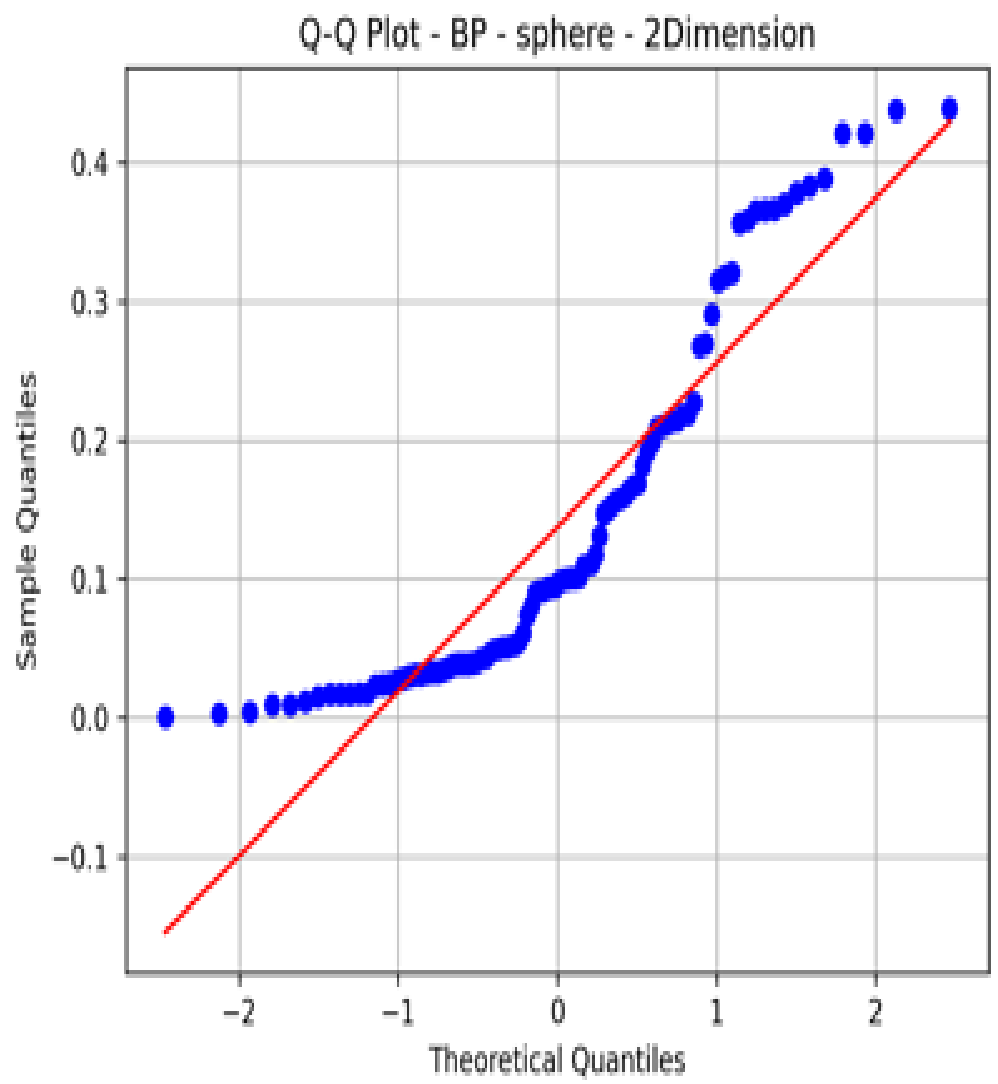
Q-Q Plot - BP - sphere - 2Dimension
Sample Quantiles
Theoretical Quantiles
0.4
0.3
0.2
0.1
0.0
−0.1
−2
−1
0
1
2

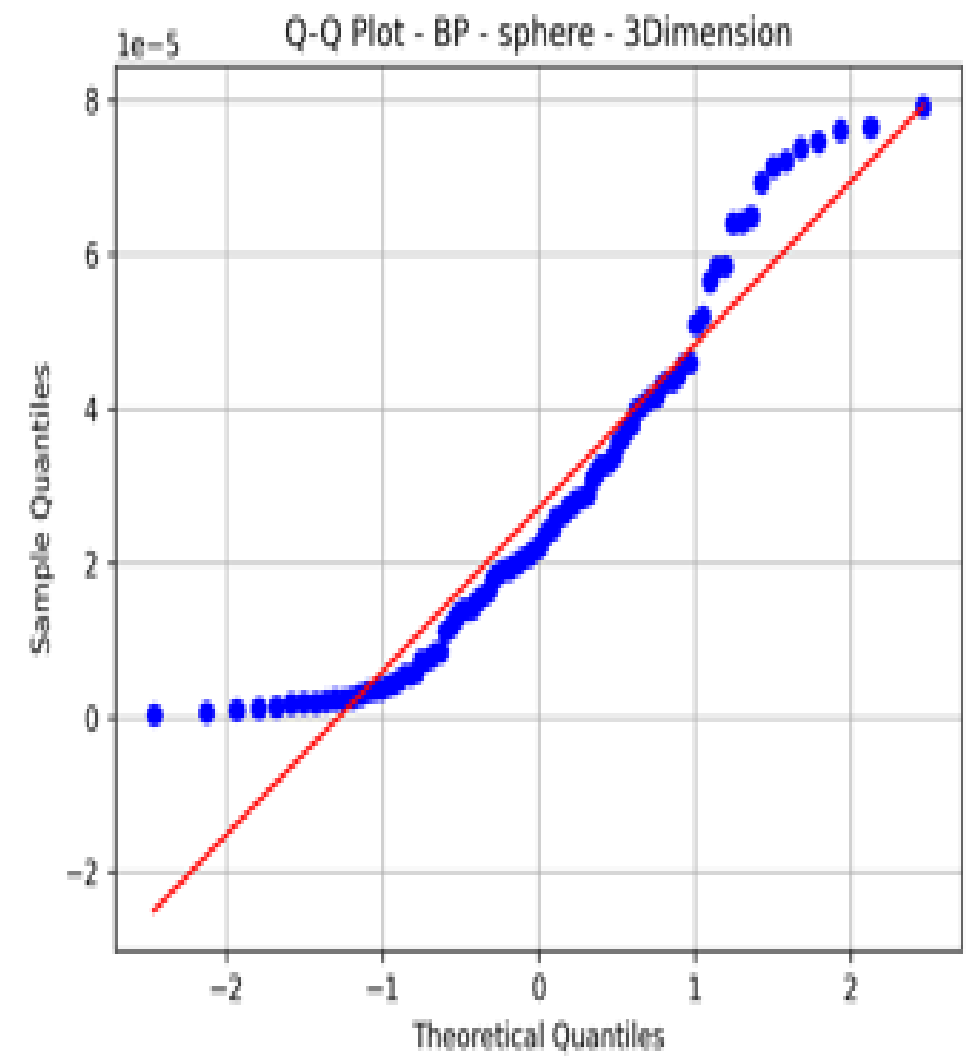
Q-Q Plot - BP - sphere - 3Dimension
1e−5
Sample Quantiles
Theoretical Quantiles
8
6
4
2
0
−2
−2
−1
0
1
2

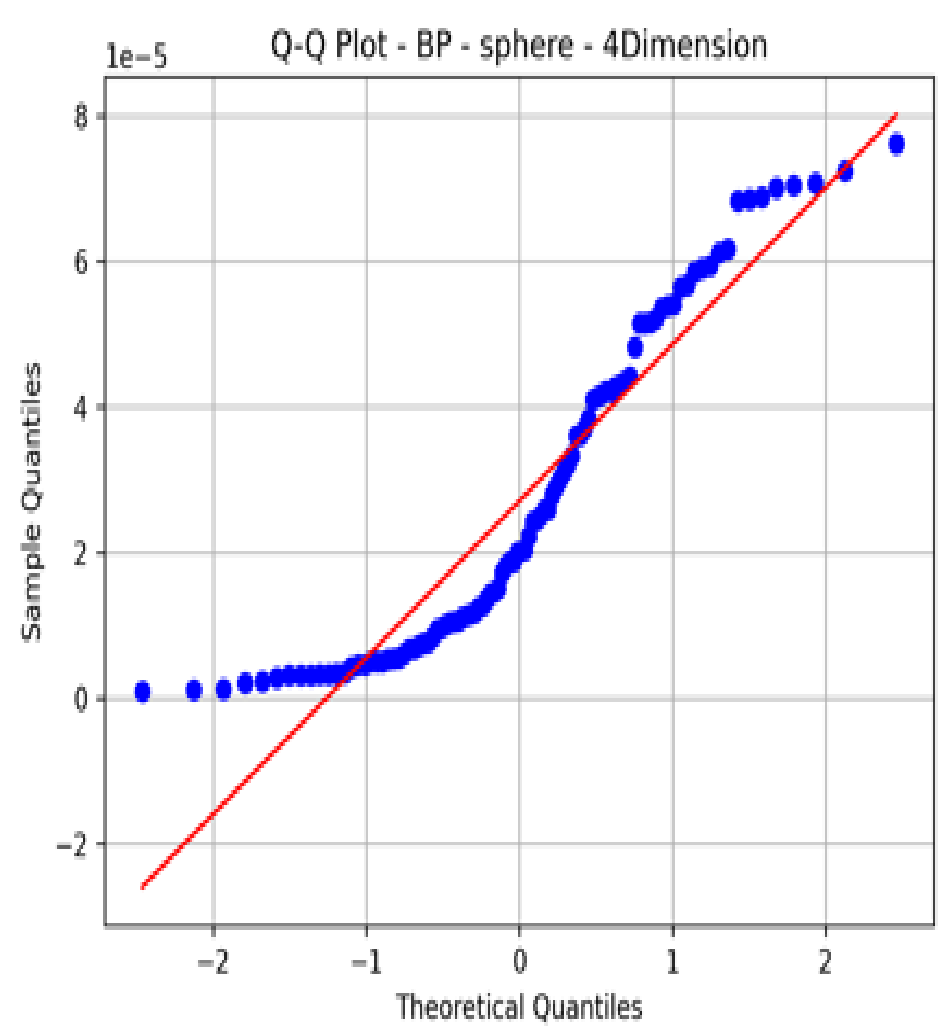
Q-Q Plot - BP - sphere - 4Dimension
1e−5
Sample Quantiles
Theoretical Quantiles
8
6
4
2
0
−2
−2
−1
0
1
2

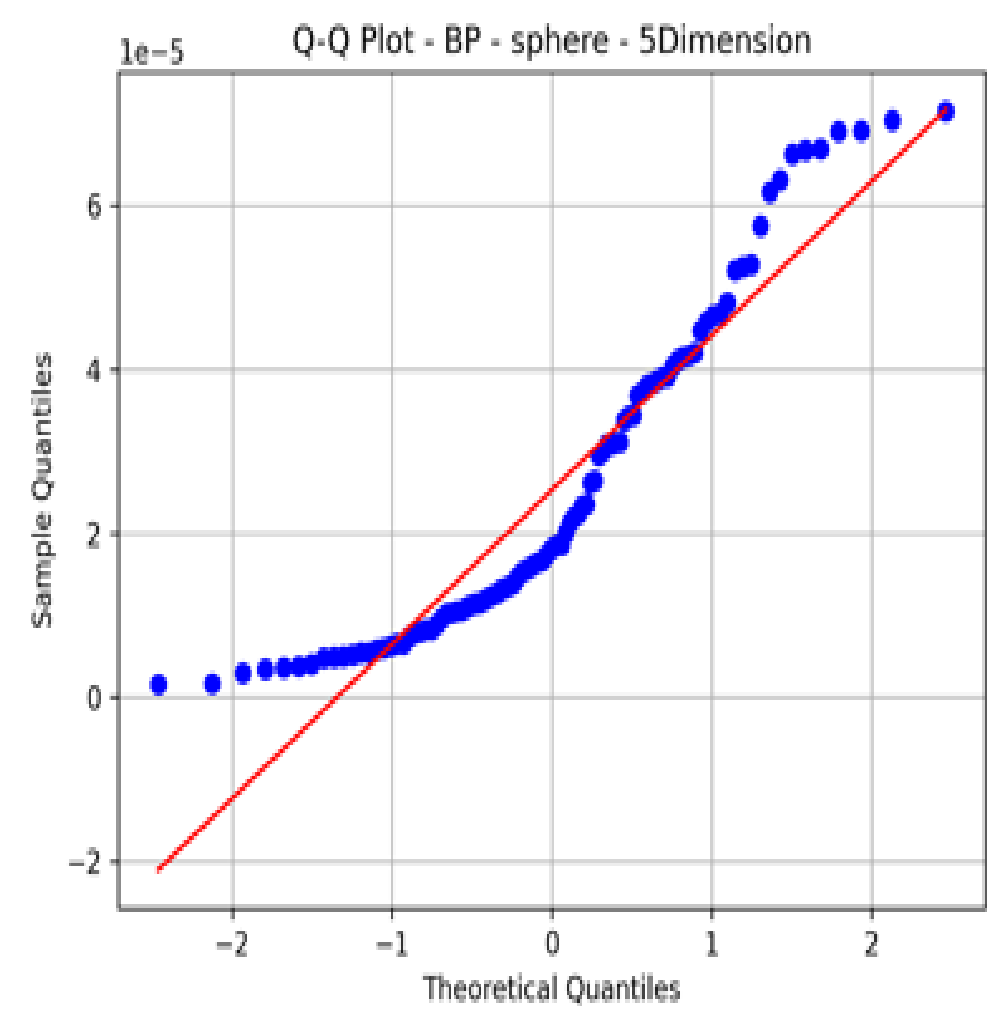
Q-Q Plot - BP - sphere - 5Dimension
1e−5
Sample Quantiles
Theoretical Quantiles
6
4
2
0
−2
−2
−1
0
1
2

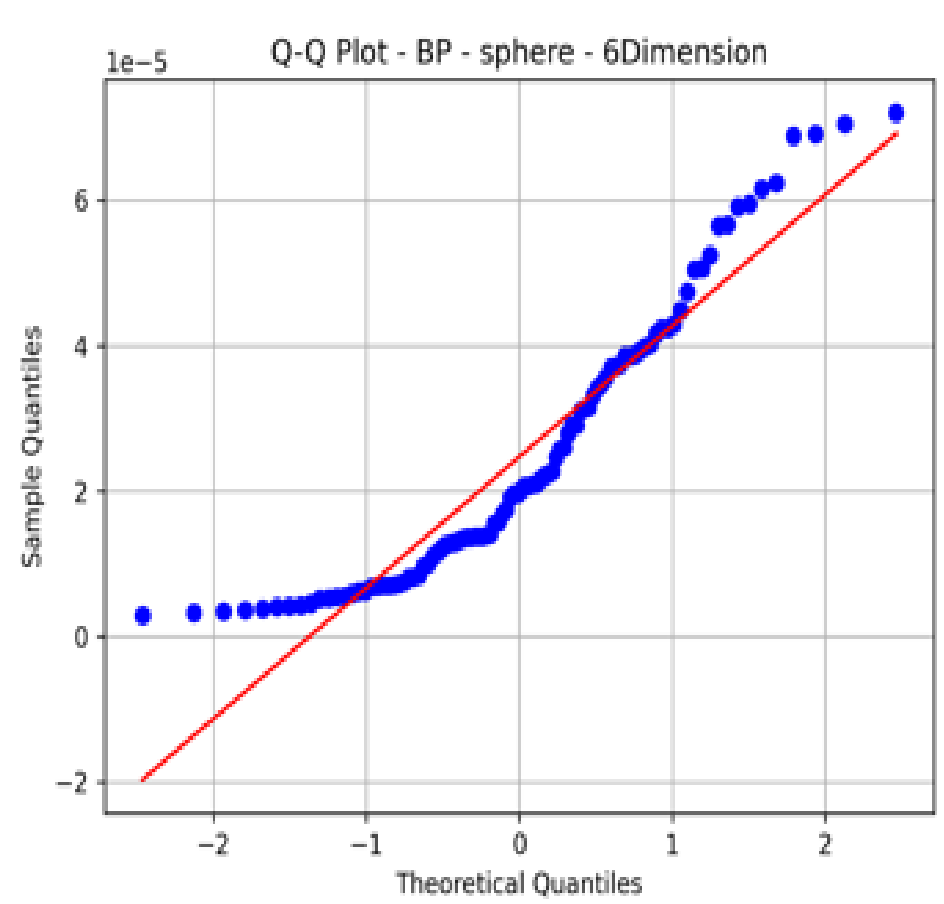
Q-Q Plot - BP - sphere - 6Dimension
1e−5
Sample Quantiles
Theoretical Quantiles
6
4
2
0
−2
−2
−1
0
1
2

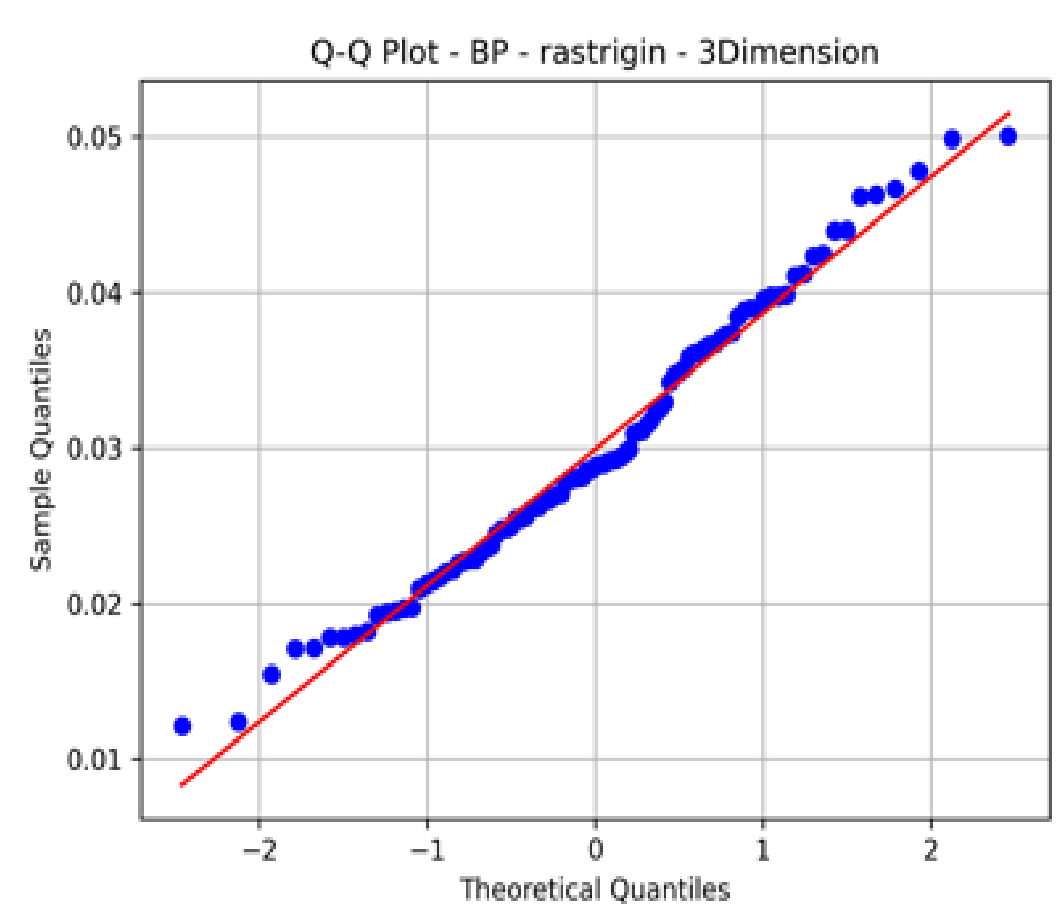
Q-Q Plot - BP - rastrigin - 3Dimension
Sample Quantiles
Theoretical Quantiles
0.05
0.04
0.03
0.02
0.01
−2
−1
0
1
2

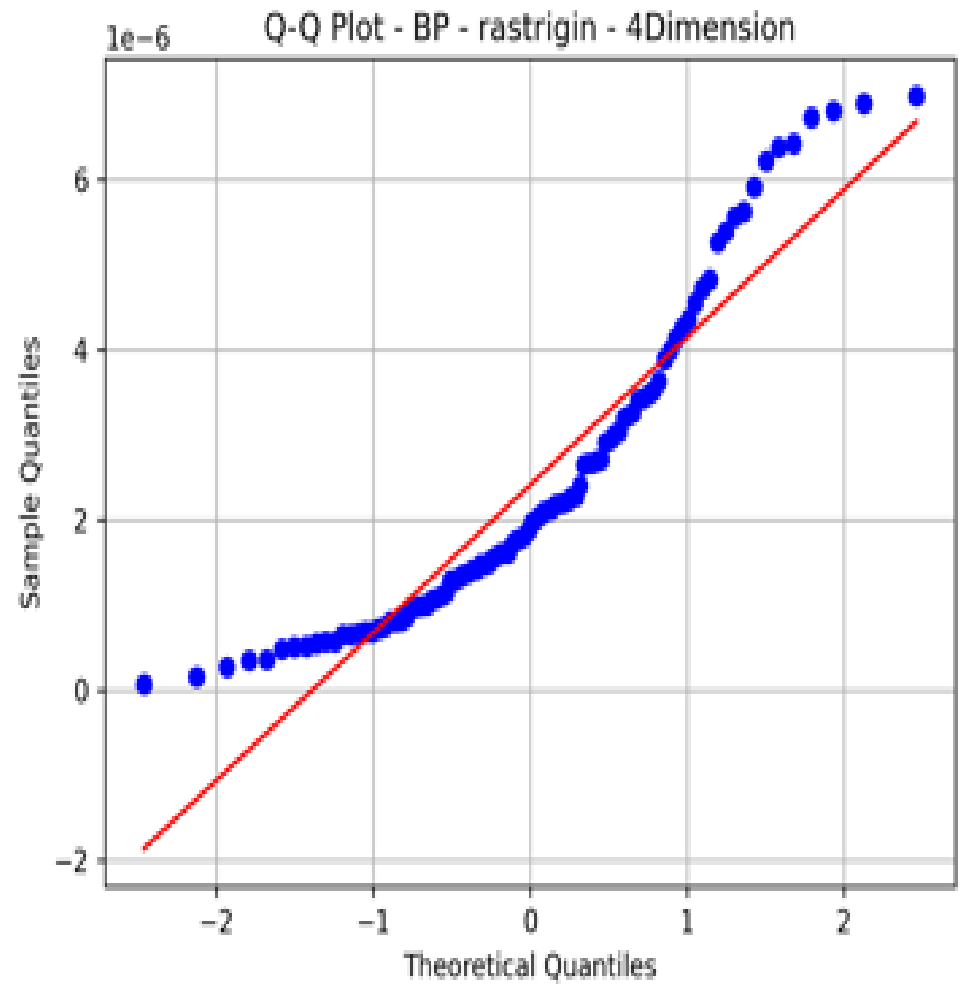

Q-Q Plot - BP - rastrigin - 4Dimension
1e−6
Sample Quantiles
Theoretical Quantiles

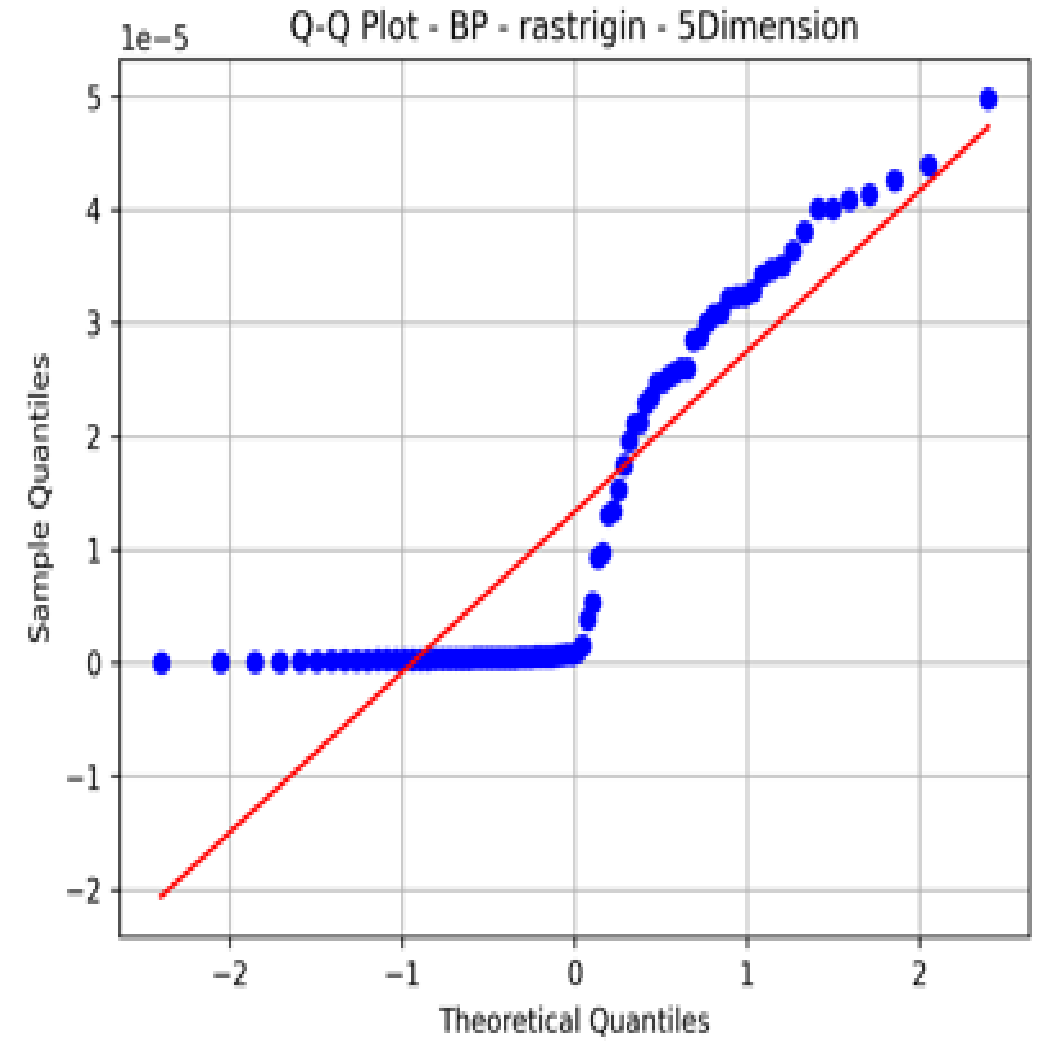

Q-Q Plot - BP - rastrigin - 5Dimension
1e−5
Sample Quantiles
Theoretical Quantiles

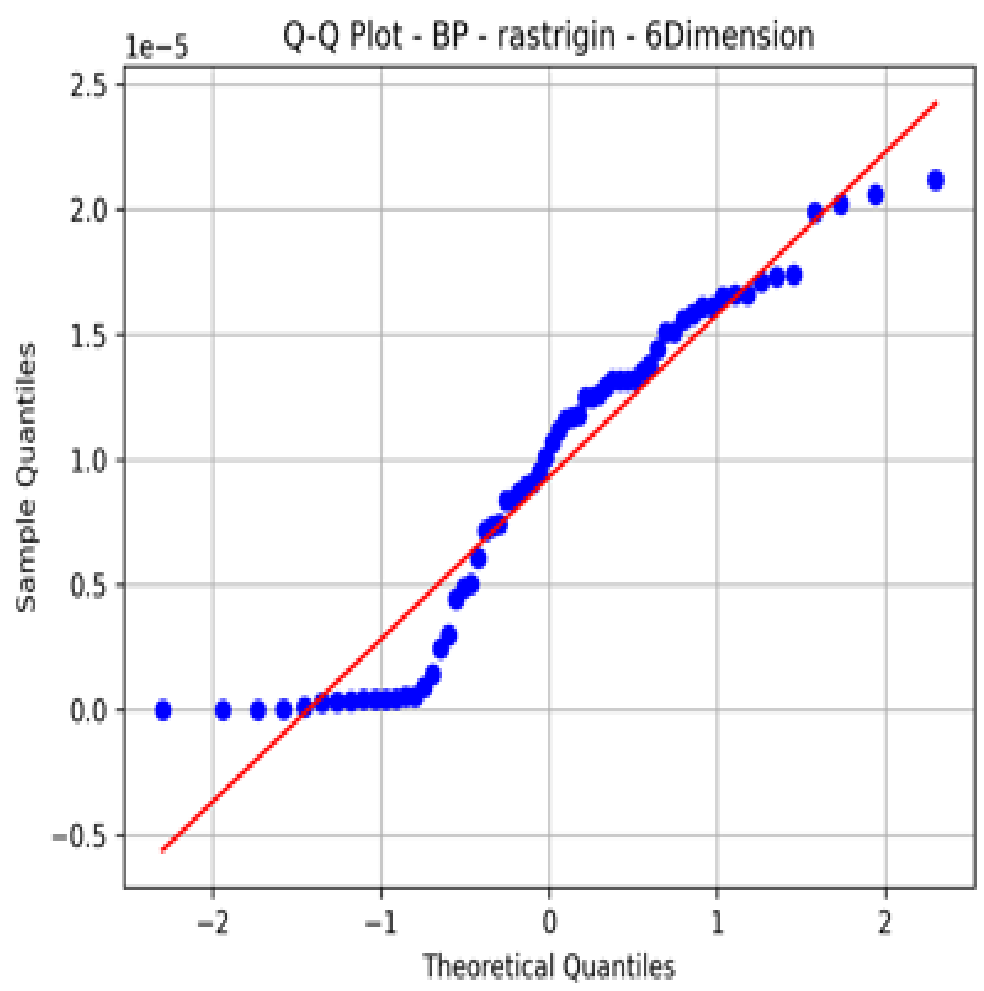

Q-Q Plot - BP - rastrigin - 6Dimension
1e−5
Sample Quantiles
Theoretical Quantiles

BPO Non-Continuous Functions QQ Plots

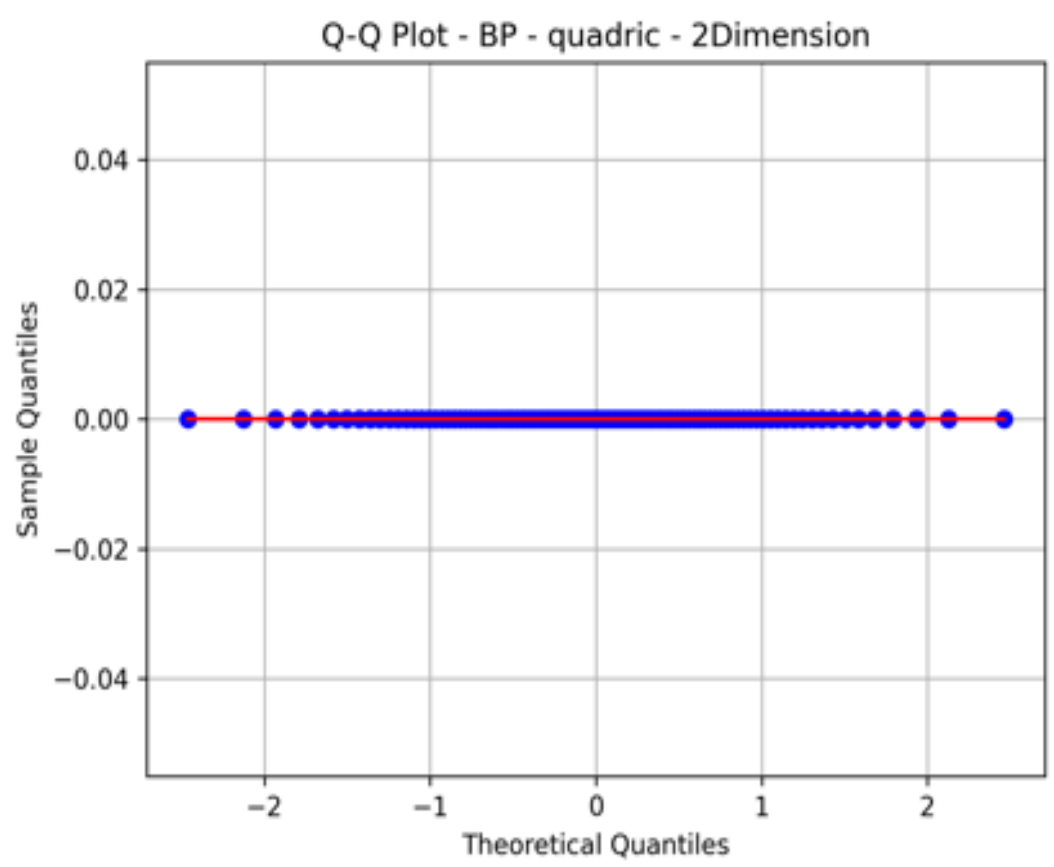

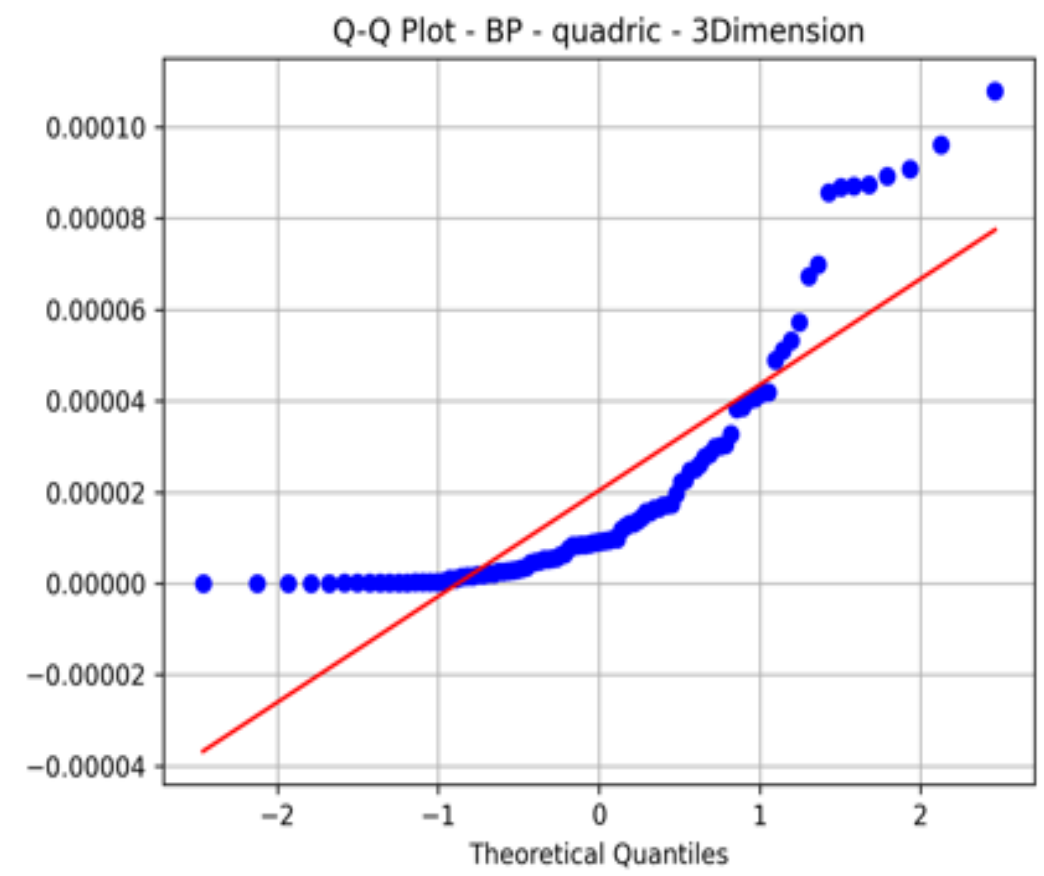

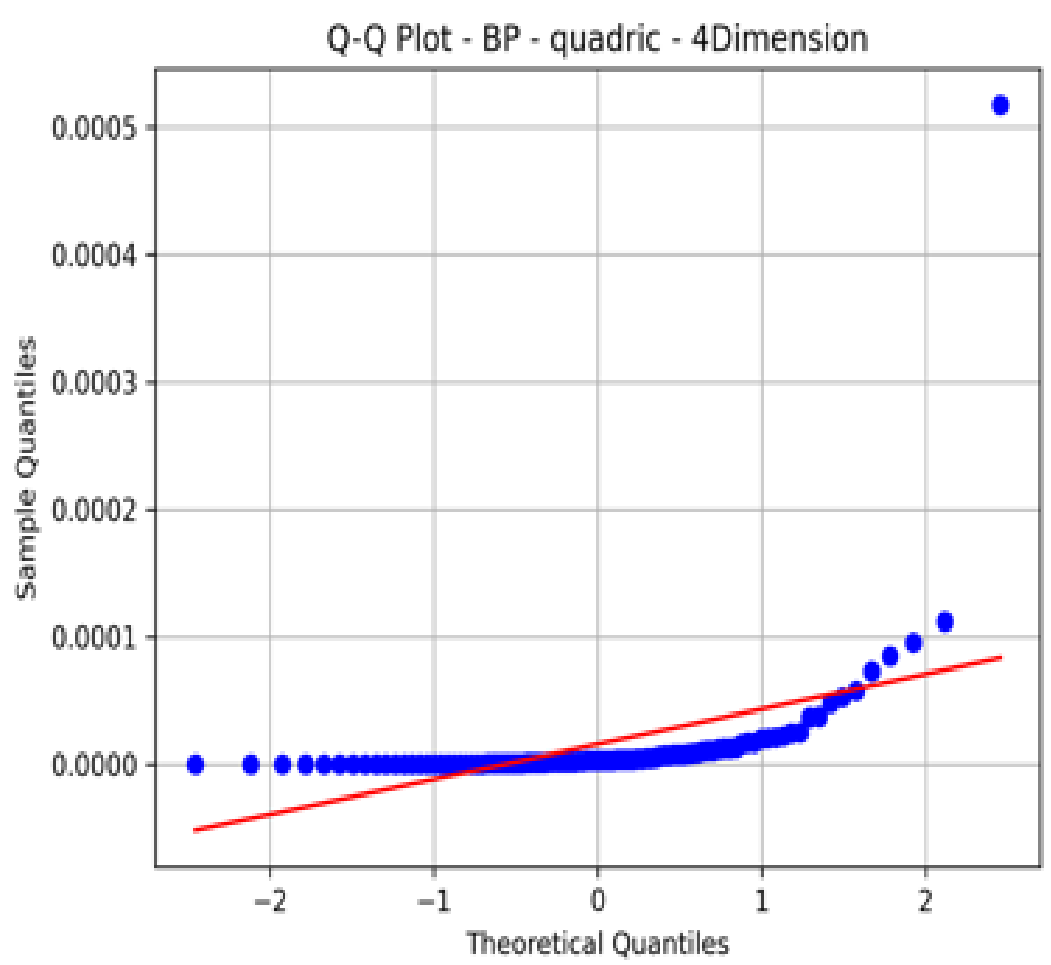

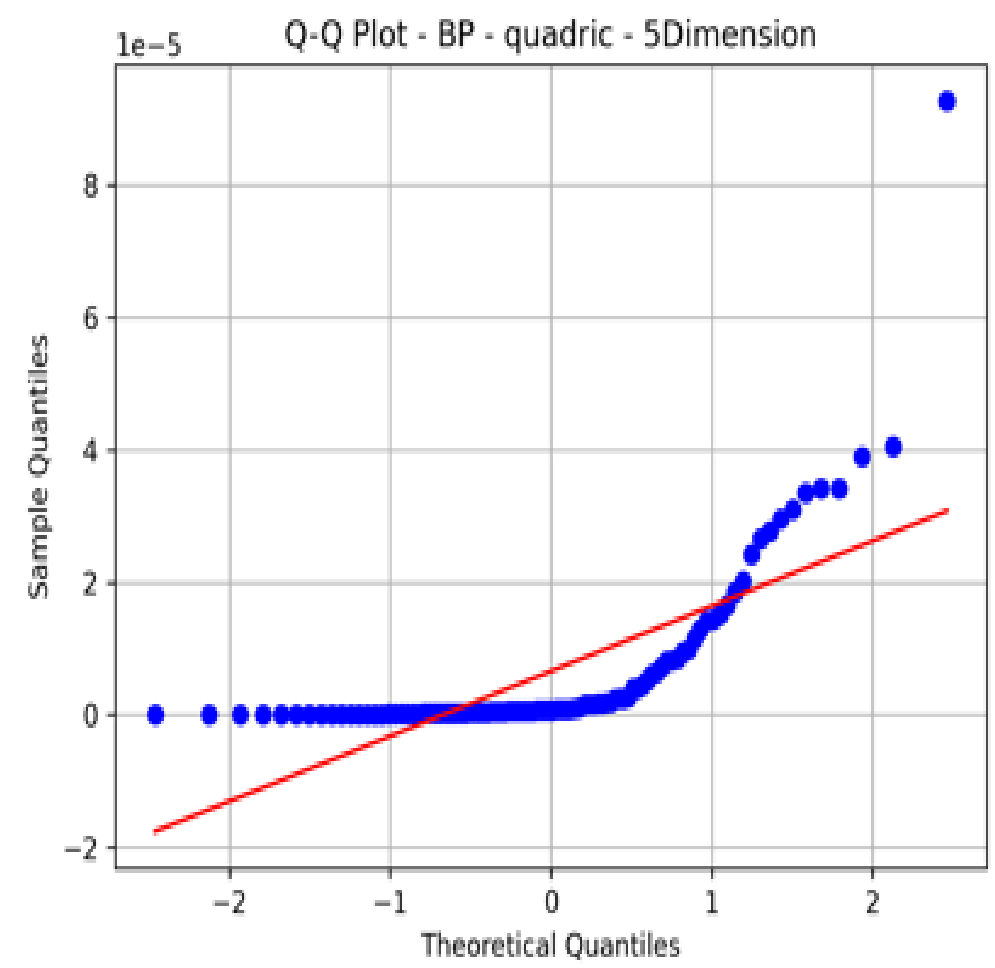

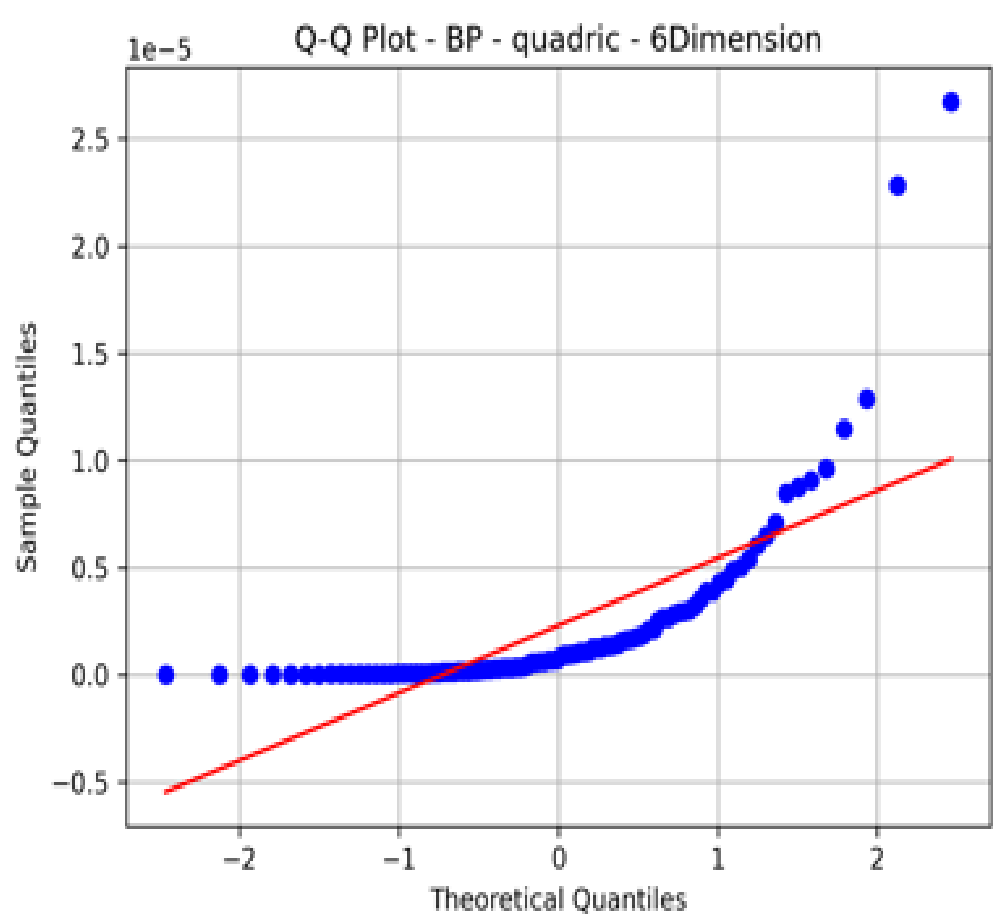

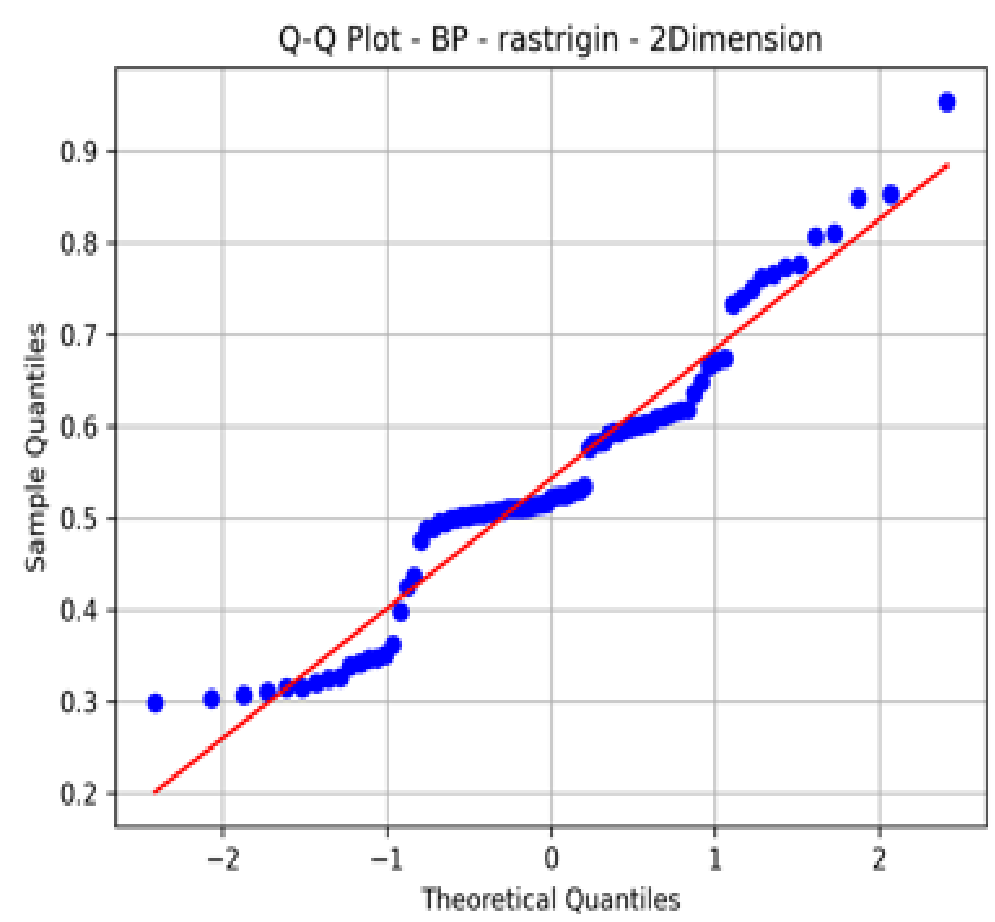

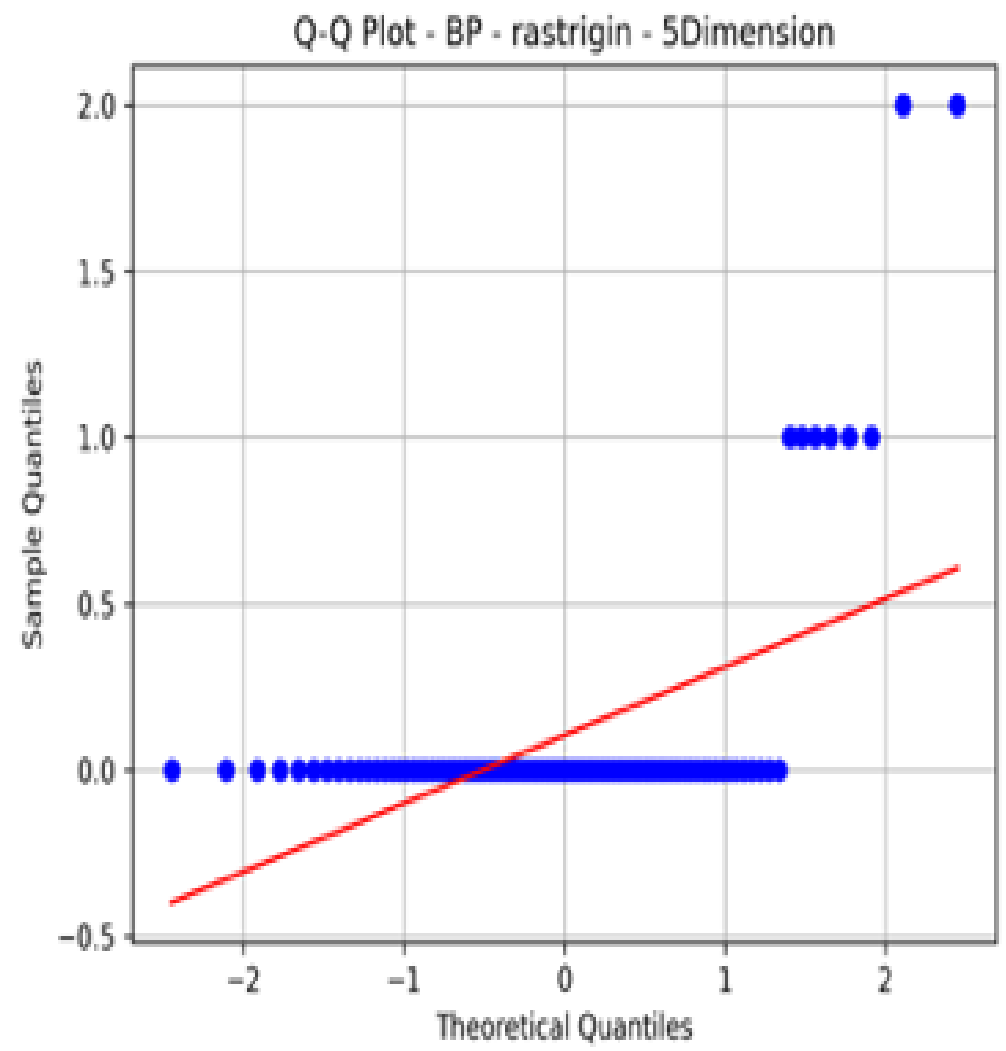
Q-Q Plot - BP - rastrigin - 5Dimension
Sample Quantiles
Theoretical Quantiles
2.0
1.5
1.0
0.5
0.0
−0.5
−2
−1
0
1
2

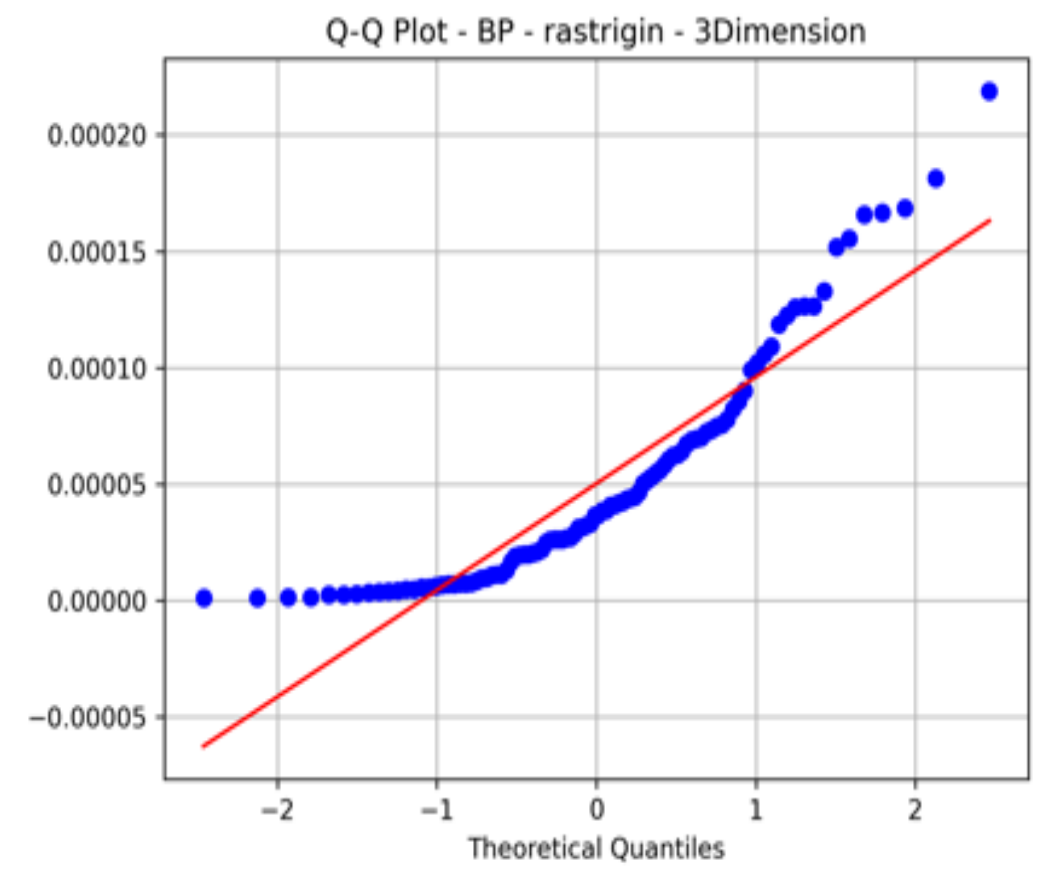
Q-Q Plot - BP - rastrigin - 3Dimension
Theoretical Quantiles
0.00020
0.00015
0.00010
0.00005
0.00000
−0.00005
−2
−1
0
1
2

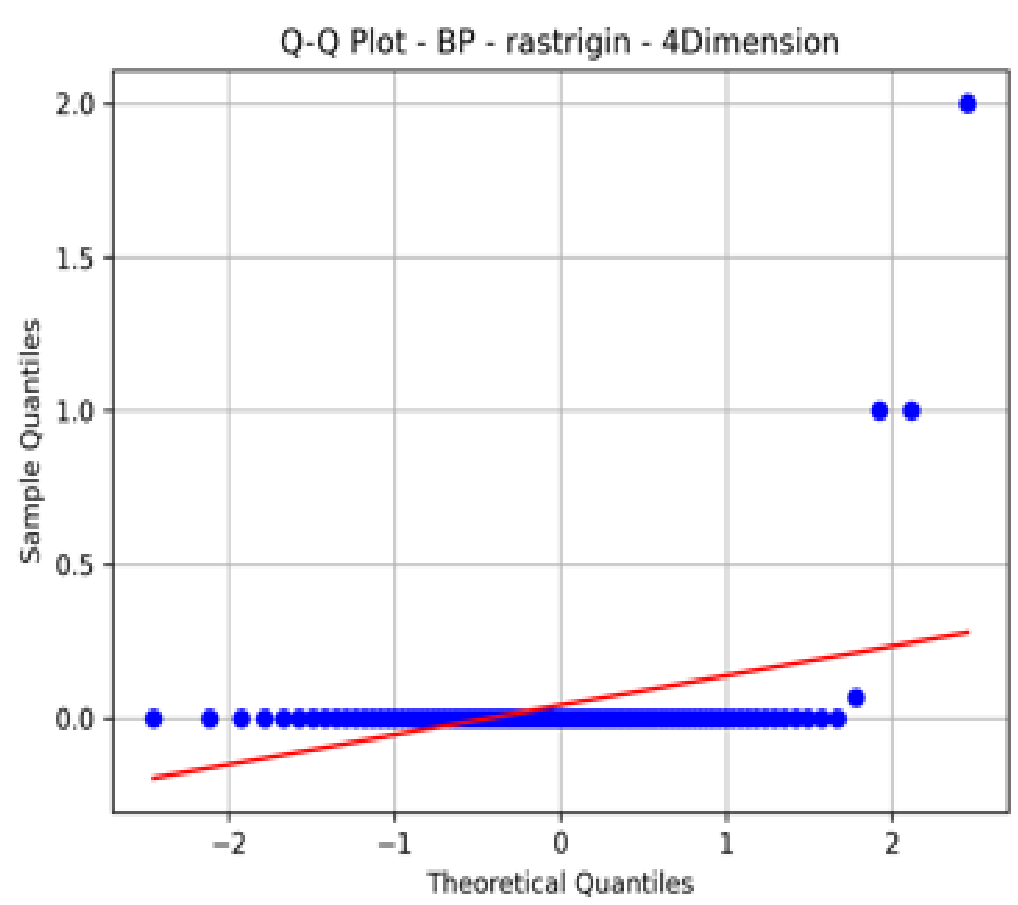
Q-Q Plot - BP - rastrigin - 4Dimension
Sample Quantiles
Theoretical Quantiles
2.0
1.5
1.0
0.5
0.0
−2
−1
0
1
2

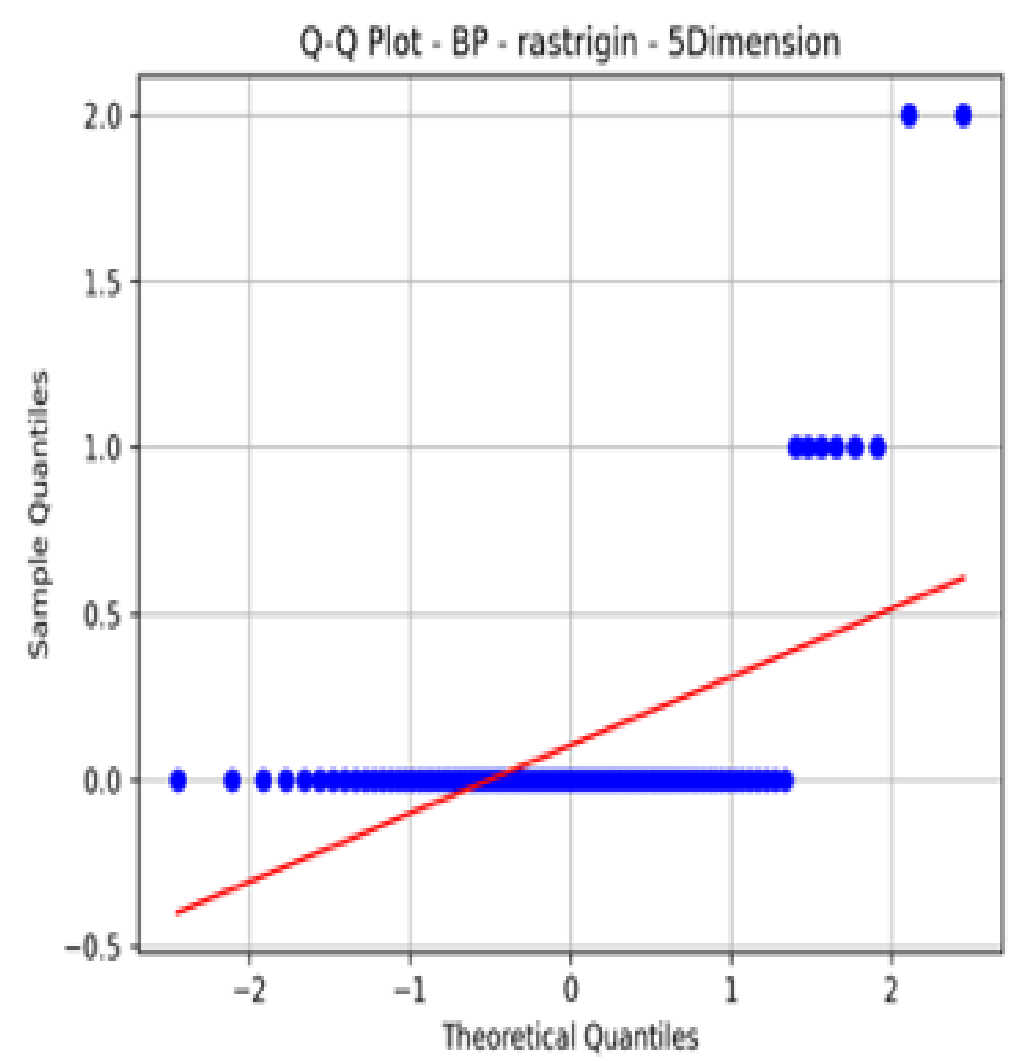
Q-Q Plot - BP - rastrigin - 5Dimension
Sample Quantiles
Theoretical Quantiles
2.0
1.5
1.0
0.5
0.0
−0.5
−2
−1
0
1
2

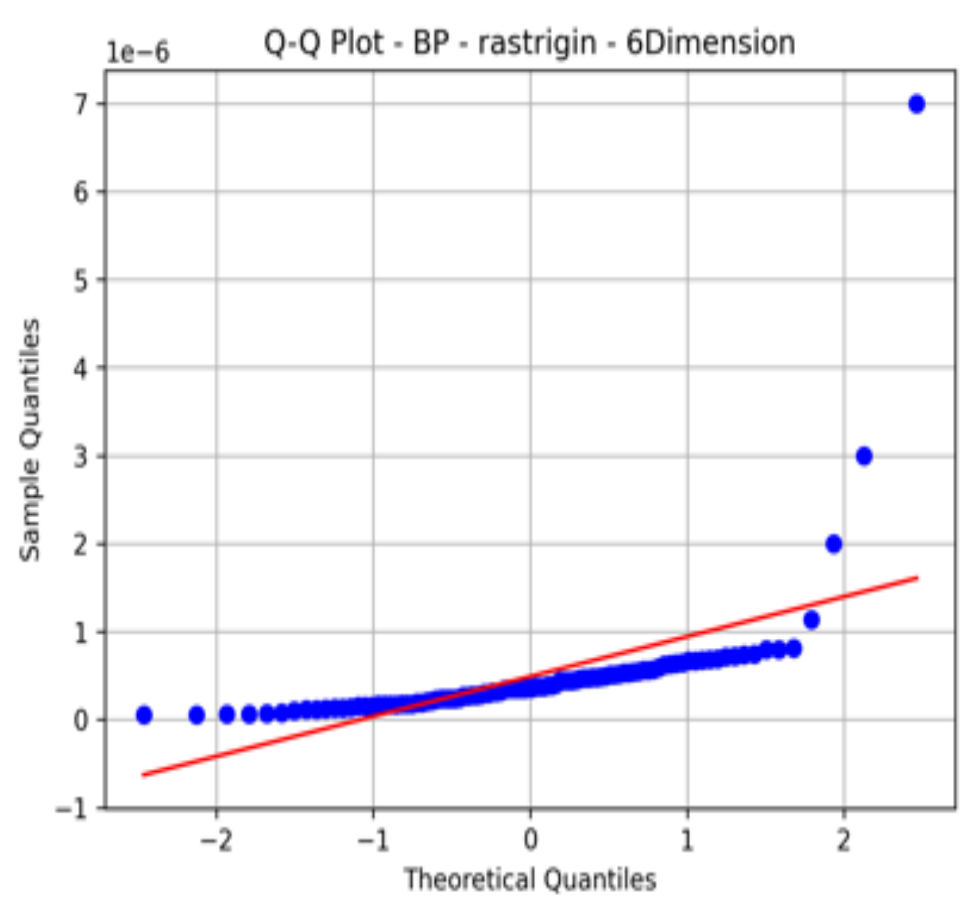
1e−6
Q-Q Plot - BP - rastrigin - 6Dimension
Sample Quantiles
Theoretical Quantiles
7
6
5
4
3
2
1
0
−1
−2
−1
0
1
2

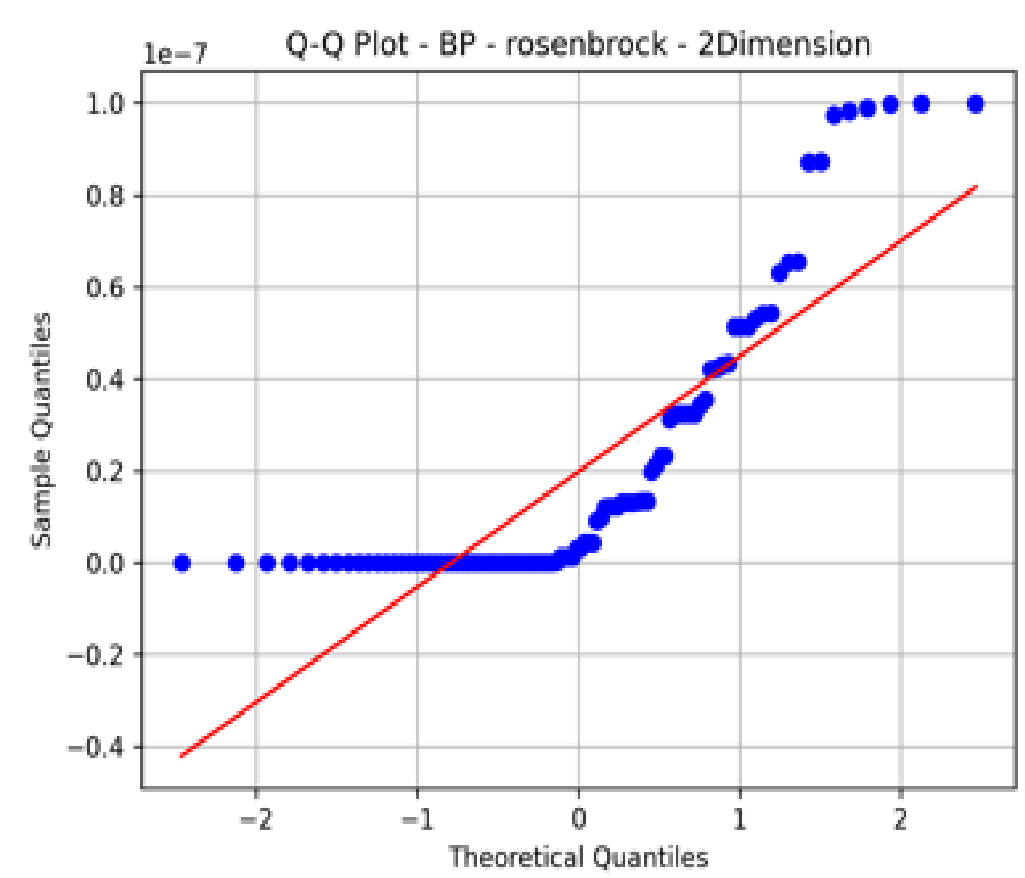
1e−7
Q-Q Plot - BP - rosenbrock - 2Dimension
Sample Quantiles
Theoretical Quantiles
1.0
0.8
0.6
0.4
0.2
0.0
−0.2
−0.4
−2
−1
0
1
2

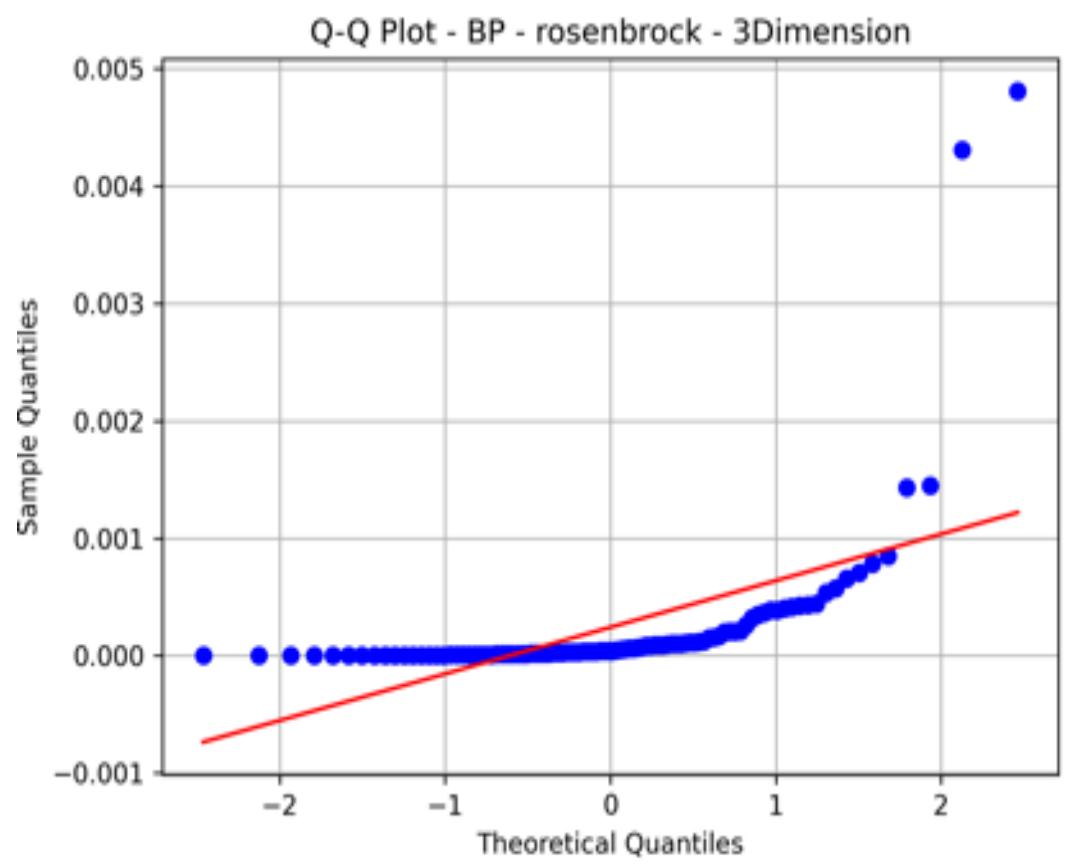
Q-Q Plot - BP - rosenbrock - 3Dimension
Sample Quantiles
Theoretical Quantiles
0.005
0.004
0.003
0.002
0.001
0.000
−0.001
−2
−1
0
1
2

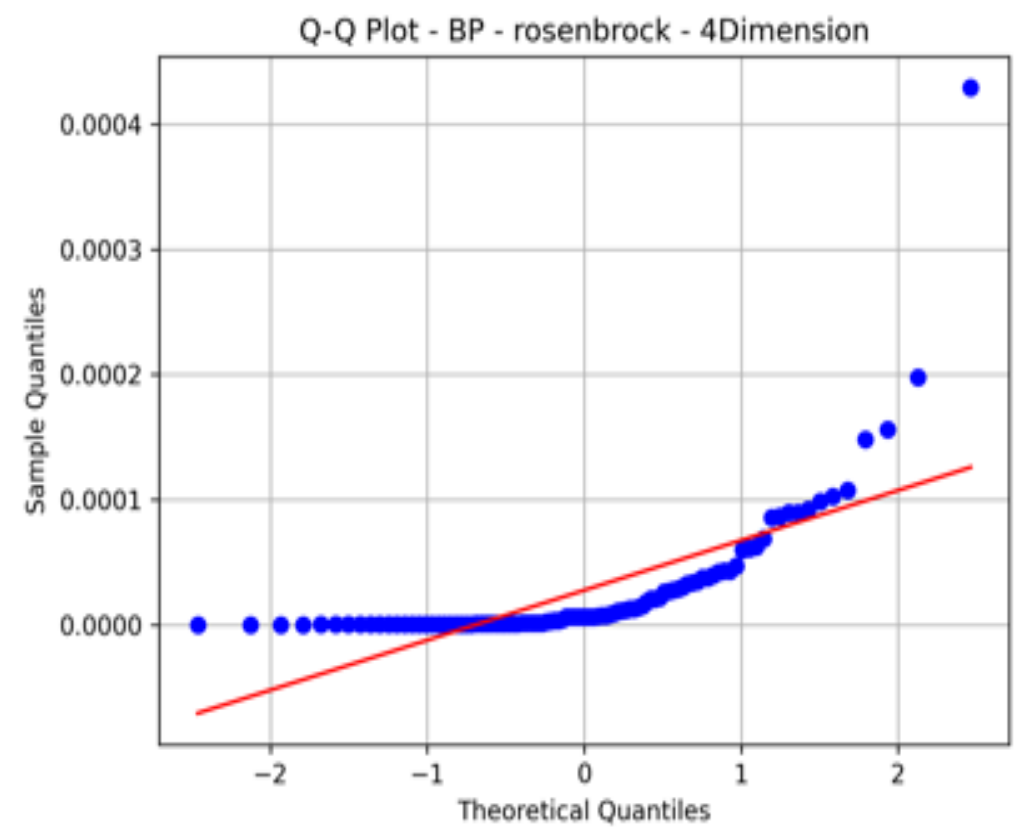
Q-Q Plot - BP - rosenbrock - 4Dimension
Sample Quantiles
Theoretical Quantiles
0.0004
0.0003
0.0002
0.0001
0.0000
−2
−1
0
1
2

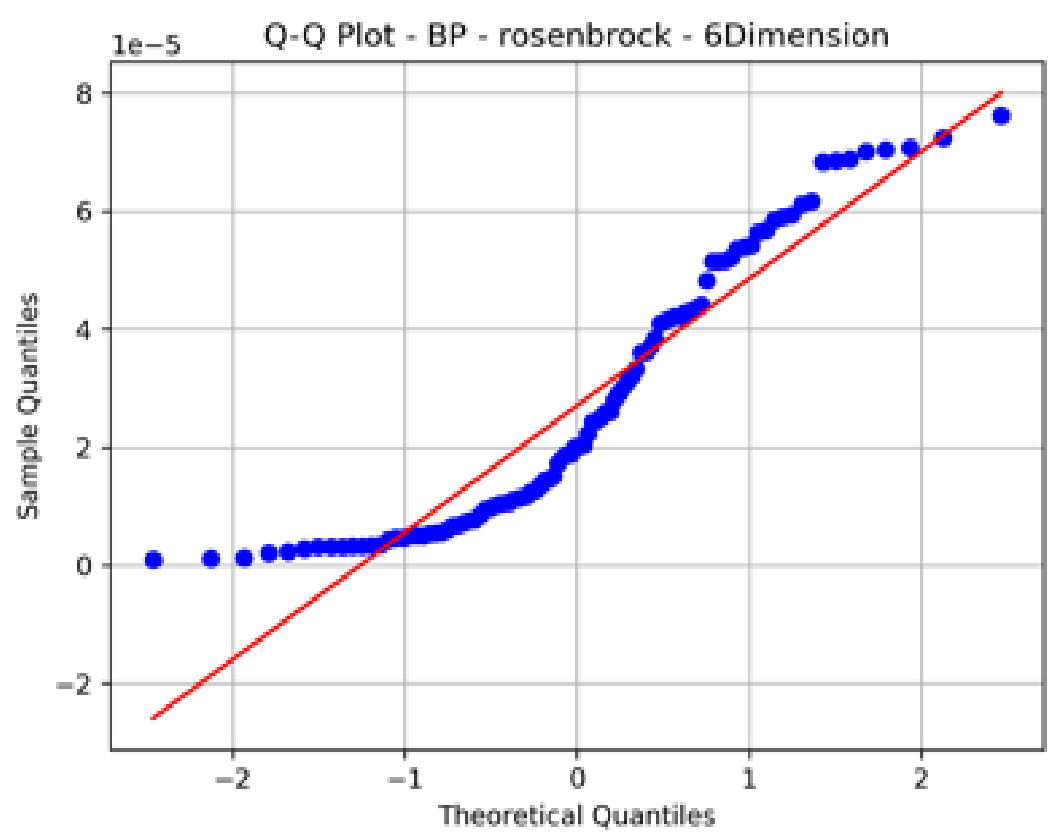
Q-Q Plot - BP - rosenbrock - 6Dimension
1e−5
Sample Quantiles
Theoretical Quantiles
8
6
4
2
0
−2
−2
−1
0
1
2

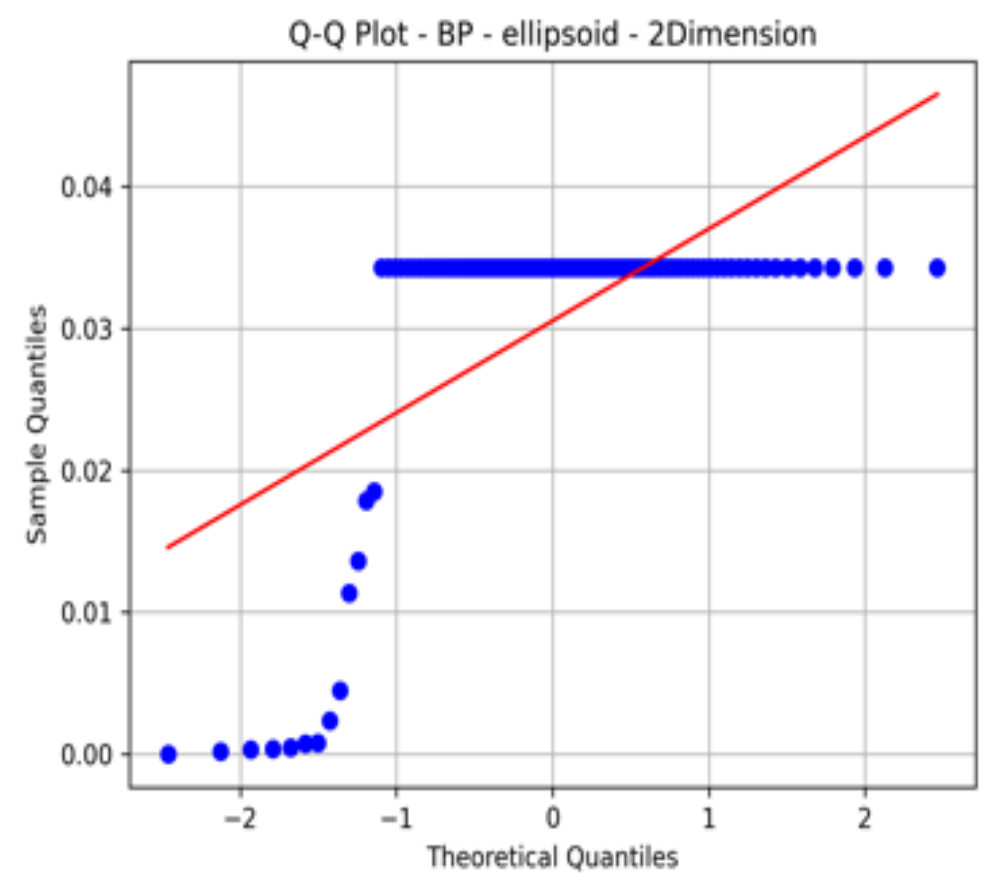
Q-Q Plot - BP - ellipsoid - 2Dimension
Sample Quantiles
Theoretical Quantiles
0.04
0.03
0.02
0.01
0.00
−2
−1
0
1
2

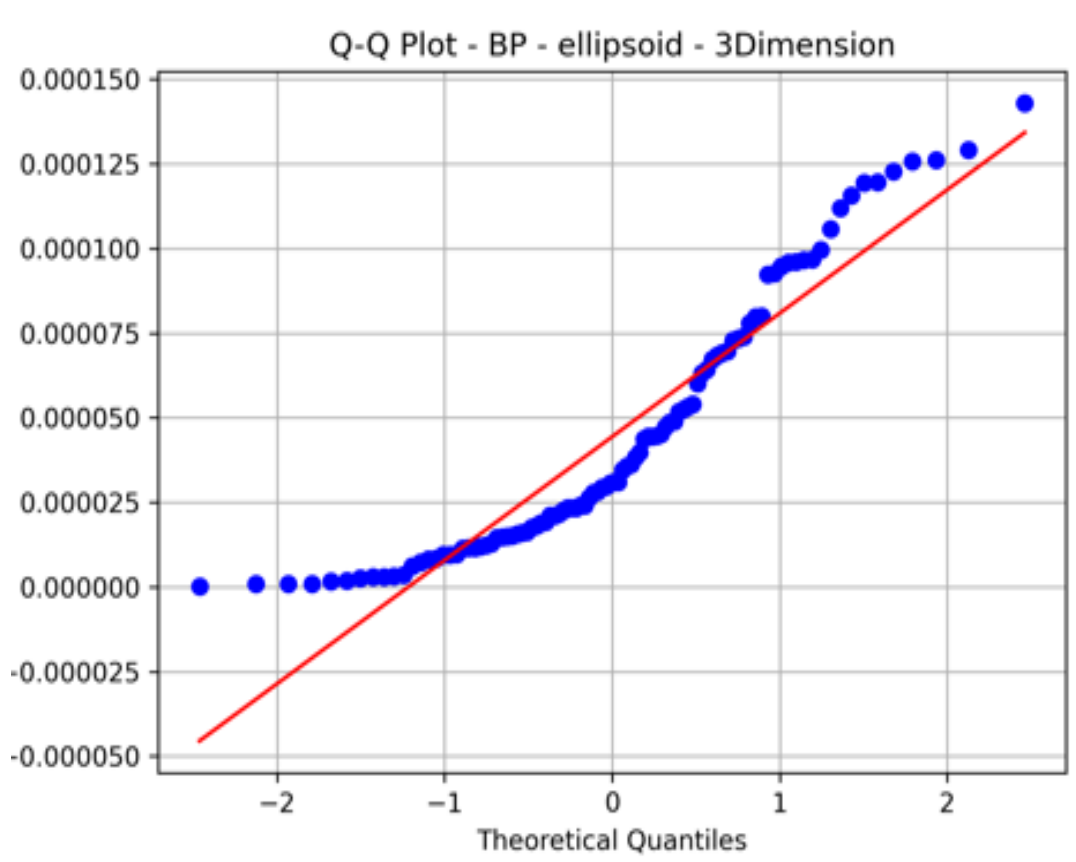
Q-Q Plot - BP - ellipsoid - 3Dimension
Theoretical Quantiles
0.000150
0.000125
0.000100
0.000075
0.000050
0.000025
0.000000
-0.000025
-0.000050
−2
−1
0
1
2

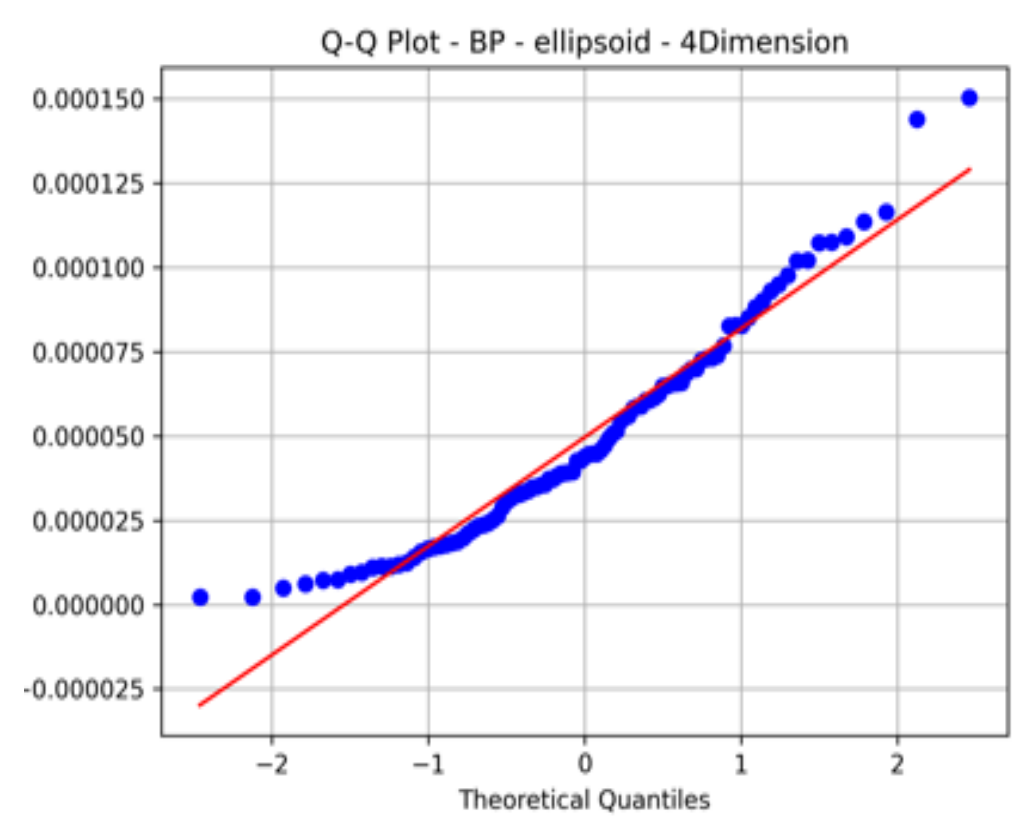
Q-Q Plot - BP - ellipsoid - 4Dimension
Theoretical Quantiles
0.000150
0.000125
0.000100
0.000075
0.000050
0.000025
0.000000
-0.000025
−2
−1
0
1
2

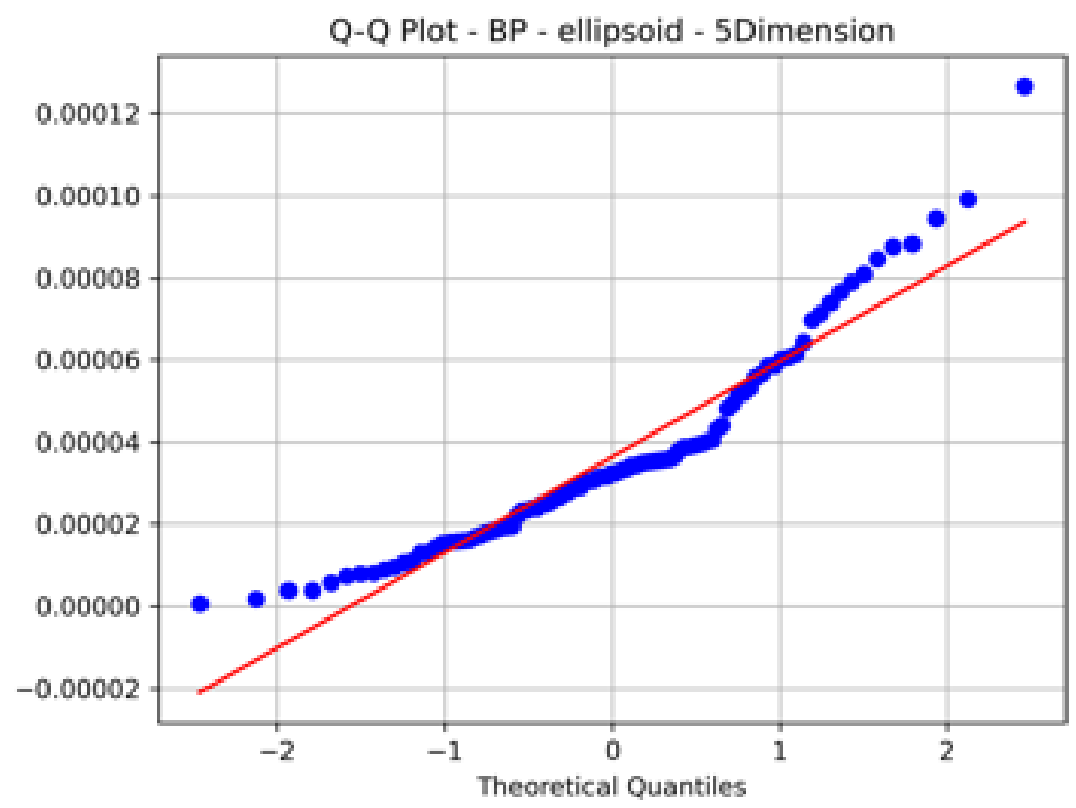
Q-Q Plot - BP - ellipsoid - 5Dimension
Theoretical Quantiles

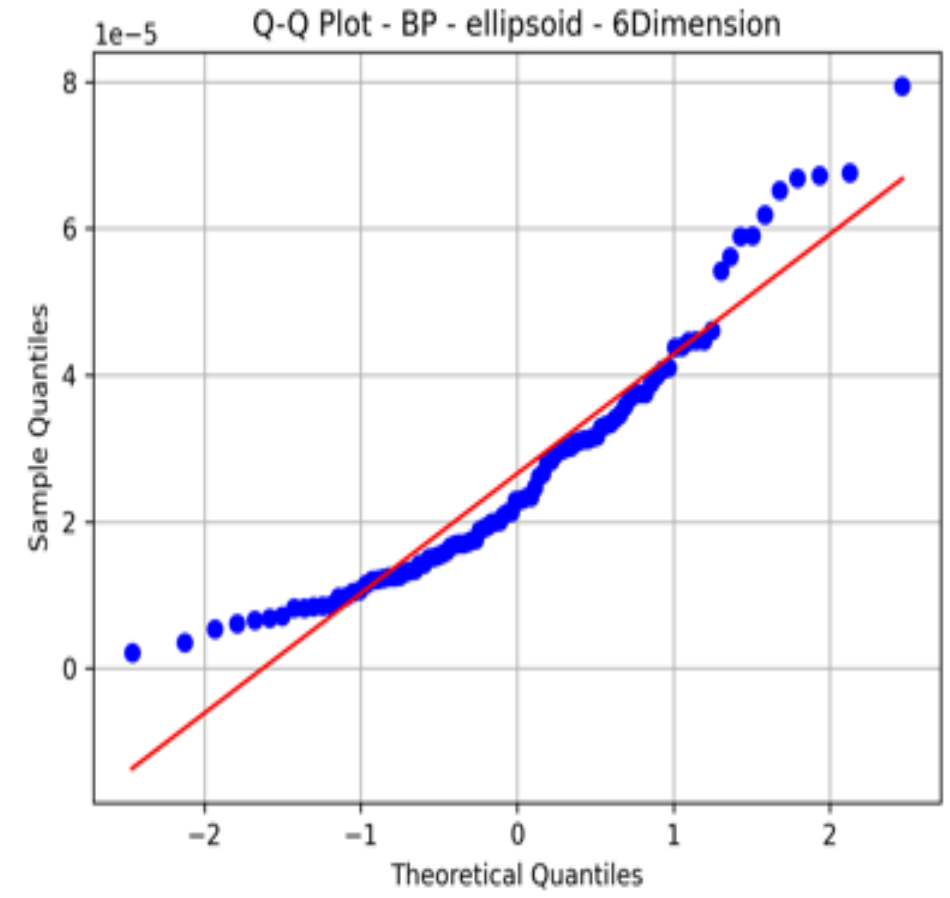
Q-Q Plot - BP - ellipsoid - 6Dimension
1e−5
Sample Quantiles
Theoretical Quantiles

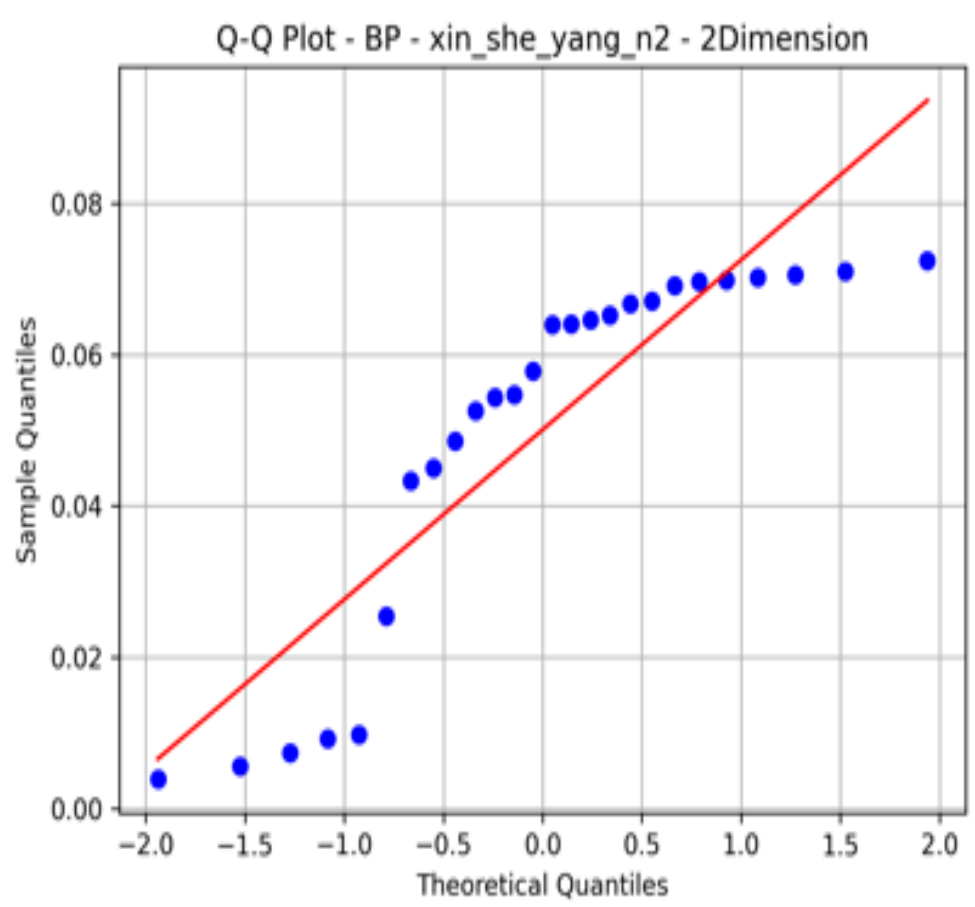
Q-Q Plot - BP - xin_she_yang_n2 - 2Dimension
Sample Quantiles
Theoretical Quantiles

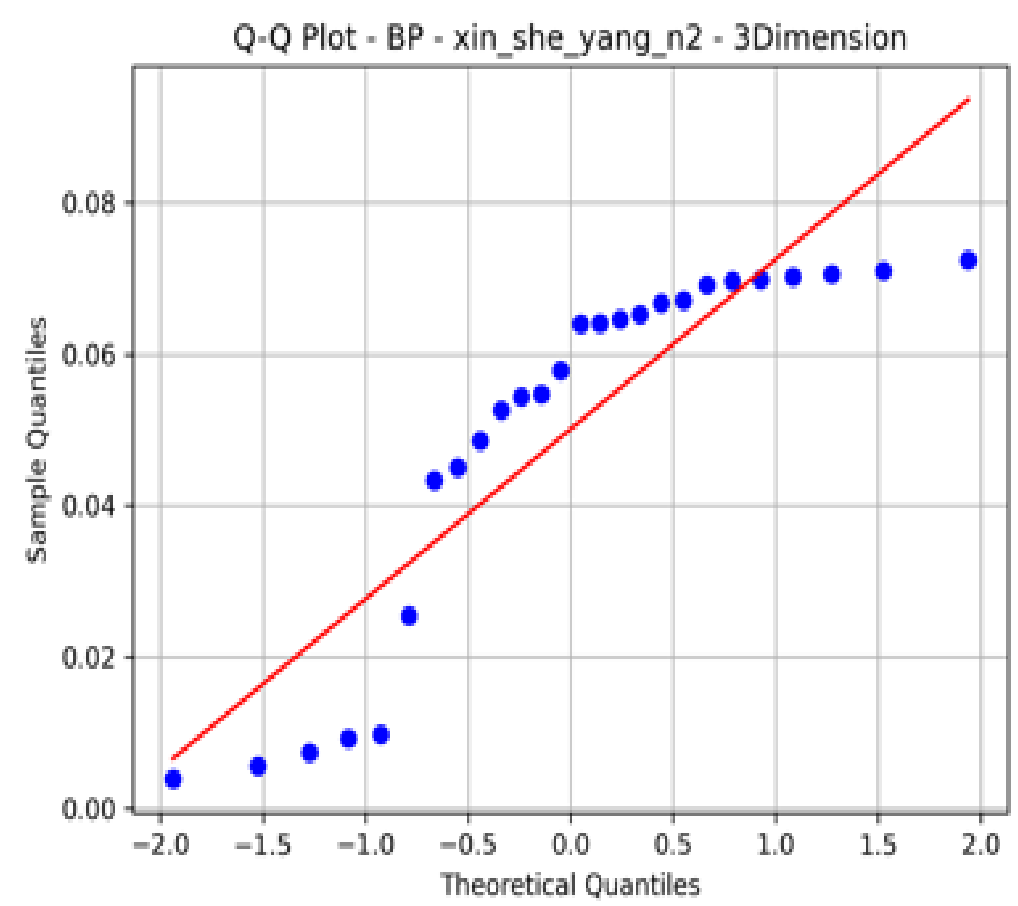
Q-Q Plot - BP - xin_she_yang_n2 - 3Dimension
Sample Quantiles
Theoretical Quantiles

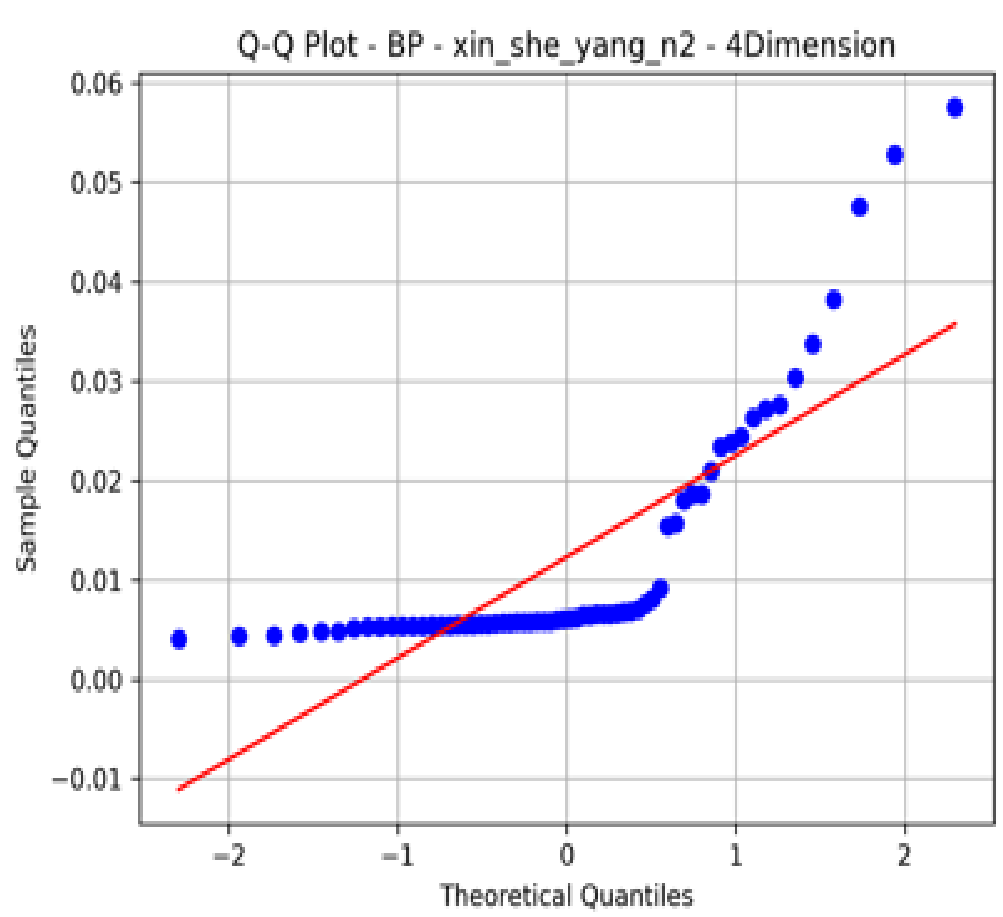
Q-Q Plot - BP - xin_she_yang_n2 - 4Dimension
Sample Quantiles
Theoretical Quantiles

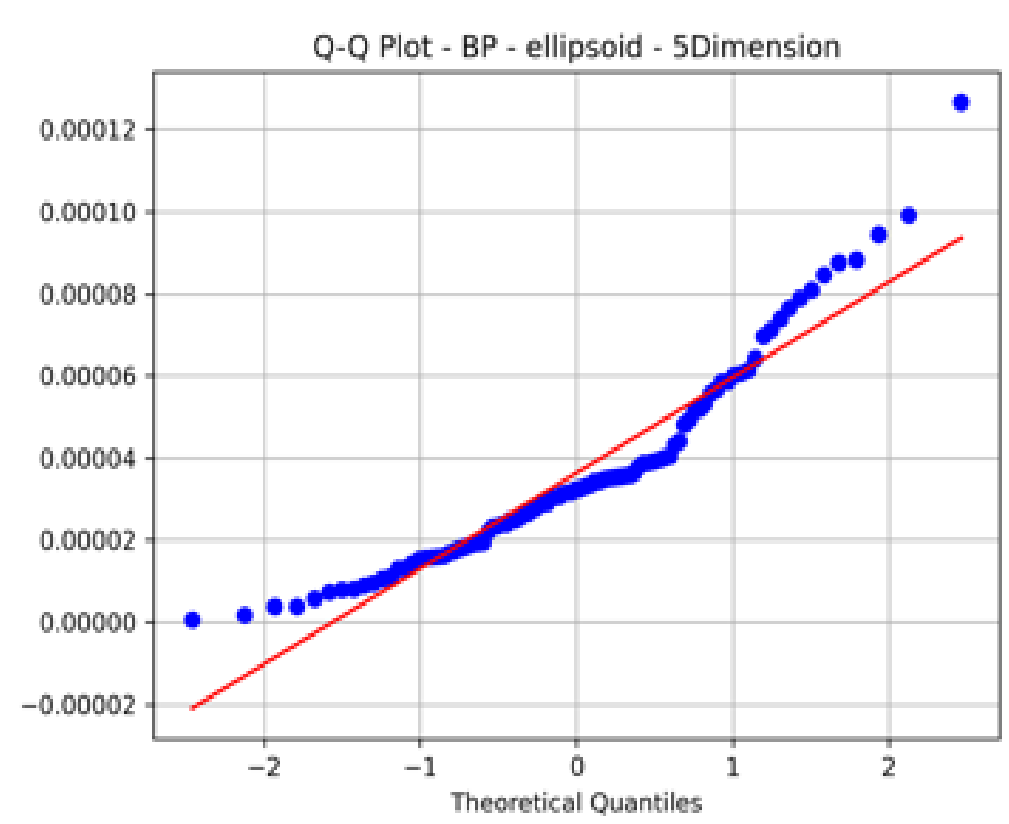
Q-Q Plot - BP - ellipsoid - 5Dimension
Theoretical Quantiles

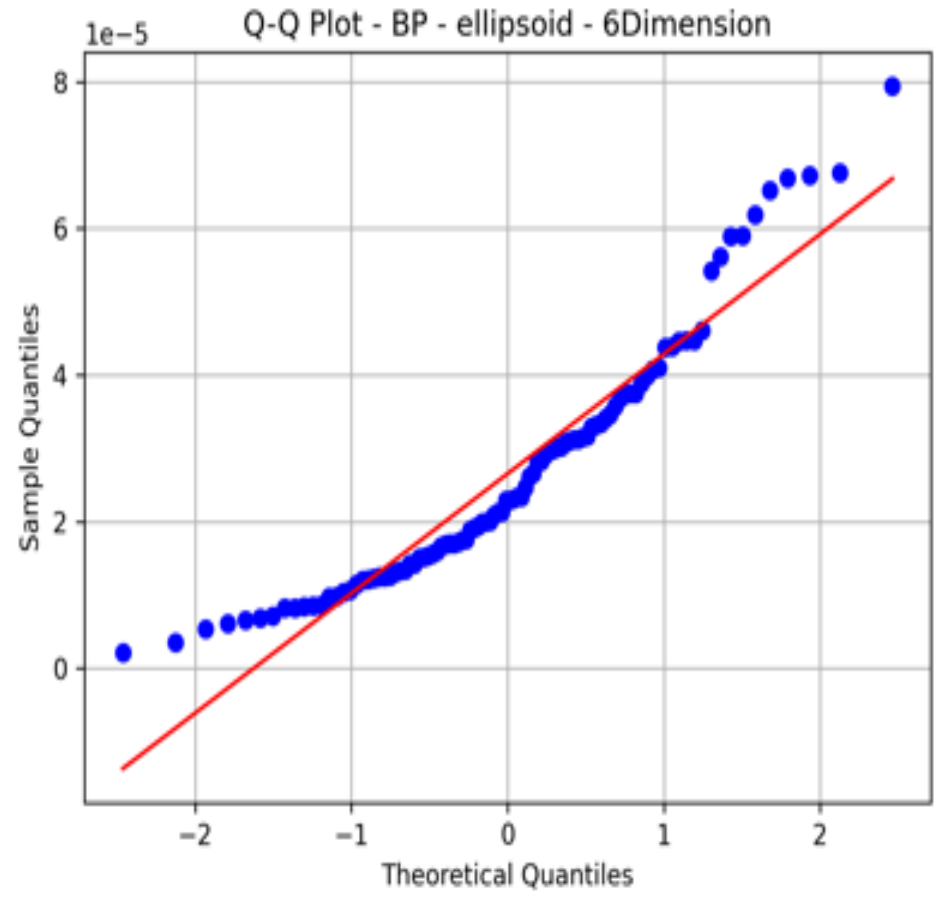
Q-Q Plot - BP - ellipsoid - 6Dimension
1e−5
Sample Quantiles
Theoretical Quantiles

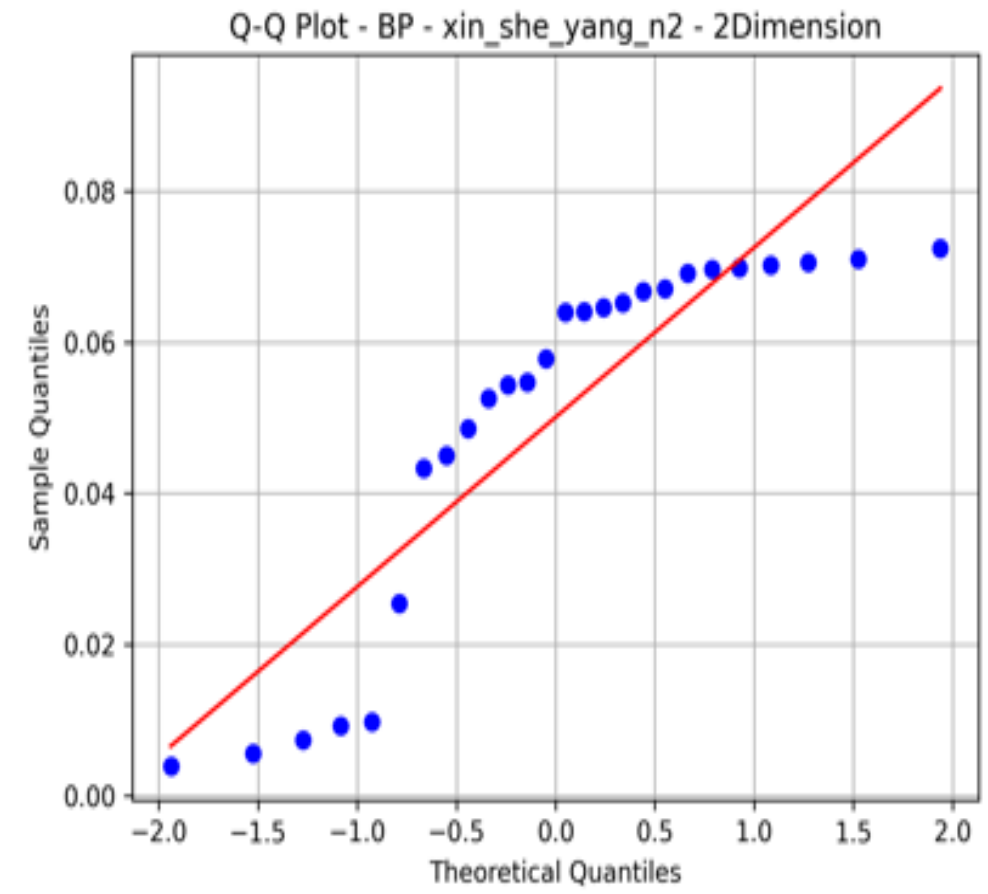
Q-Q Plot - BP - xin_she_yang_n2 - 2Dimension
Sample Quantiles
Theoretical Quantiles

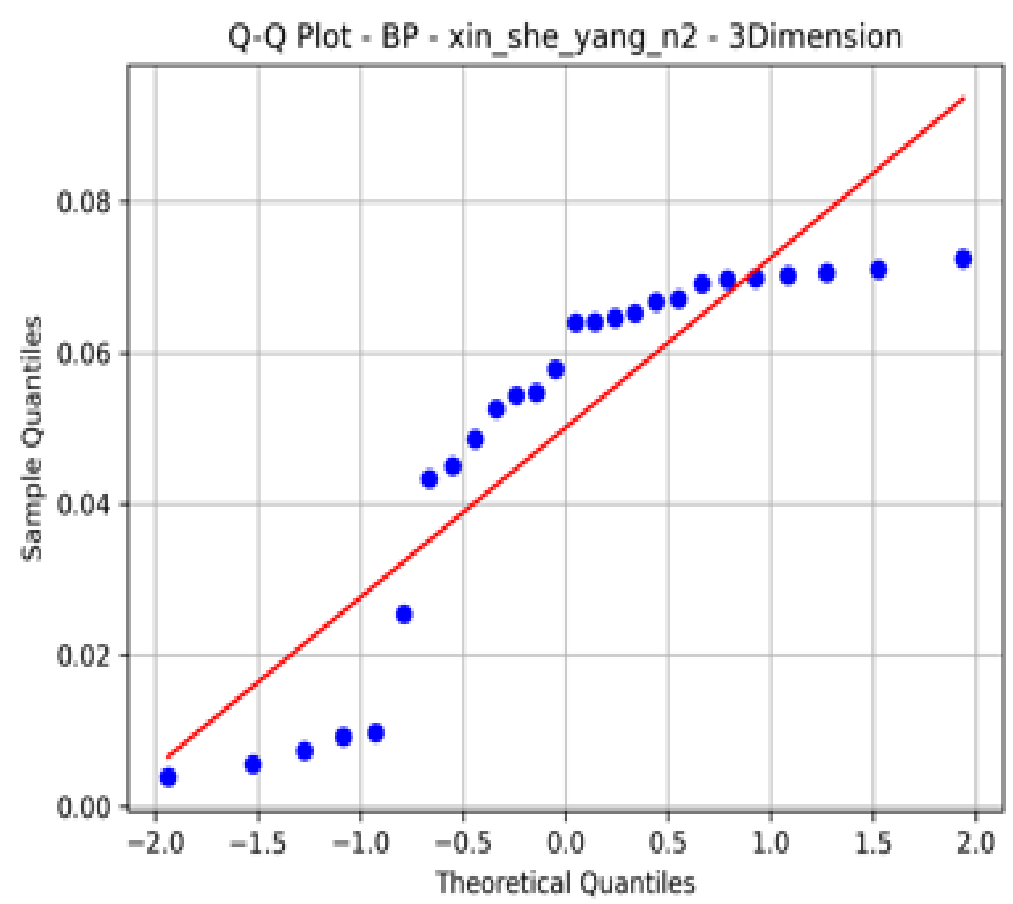
Q-Q Plot - BP - xin_she_yang_n2 - 3Dimension
Sample Quantiles
Theoretical Quantiles

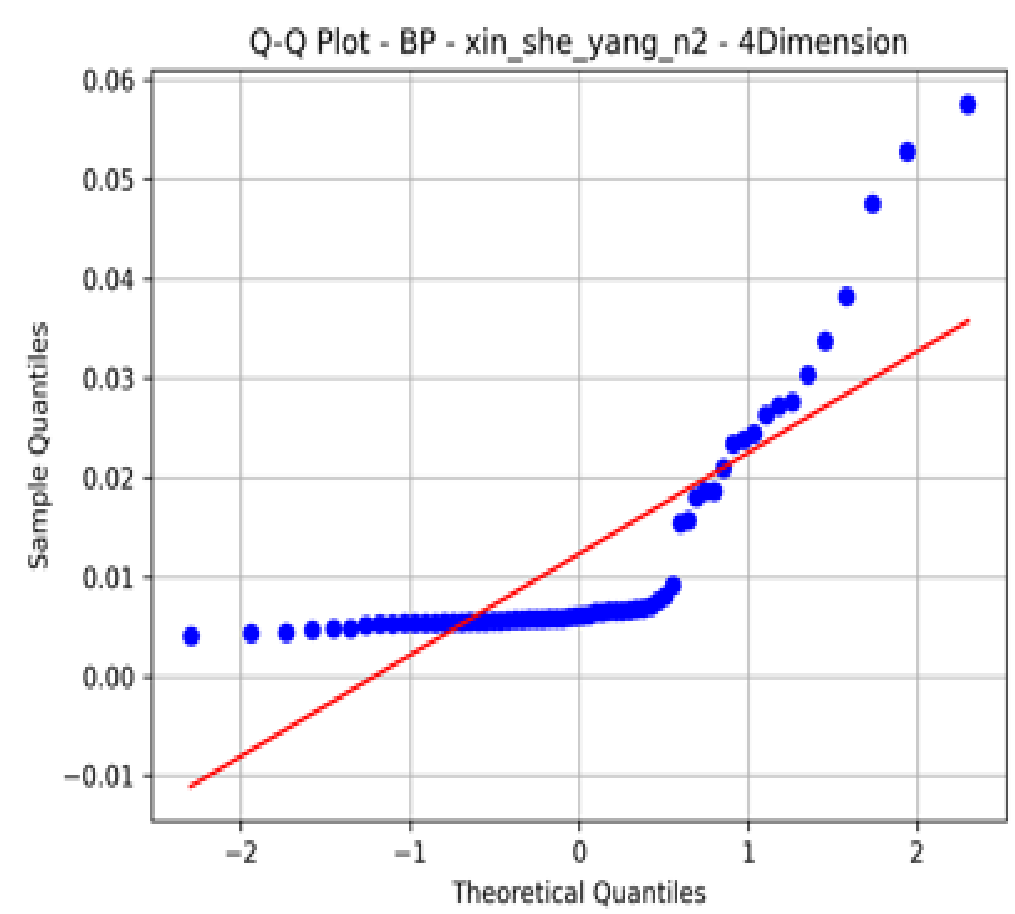
Q-Q Plot - BP - xin_she_yang_n2 - 4Dimension
Sample Quantiles
Theoretical Quantiles

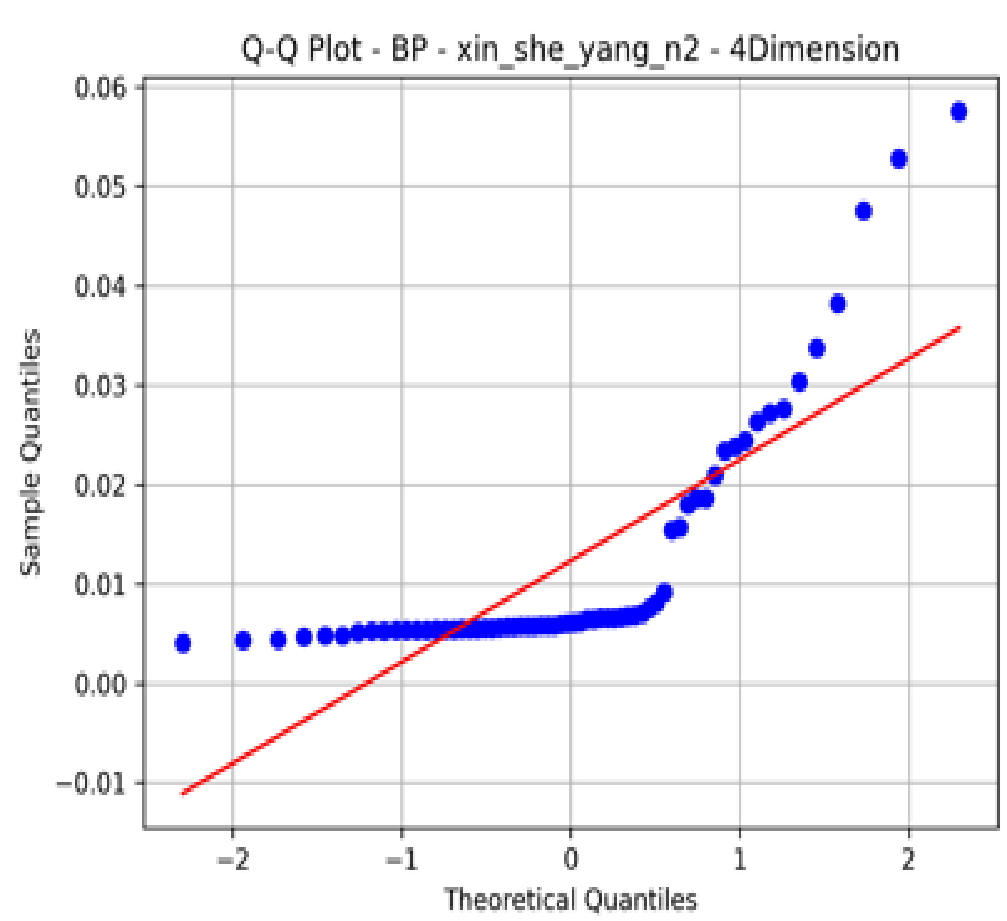
Q-Q Plot - BP - xin_she_yang_n2 - 4Dimension
Sample Quantiles
Theoretical Quantiles

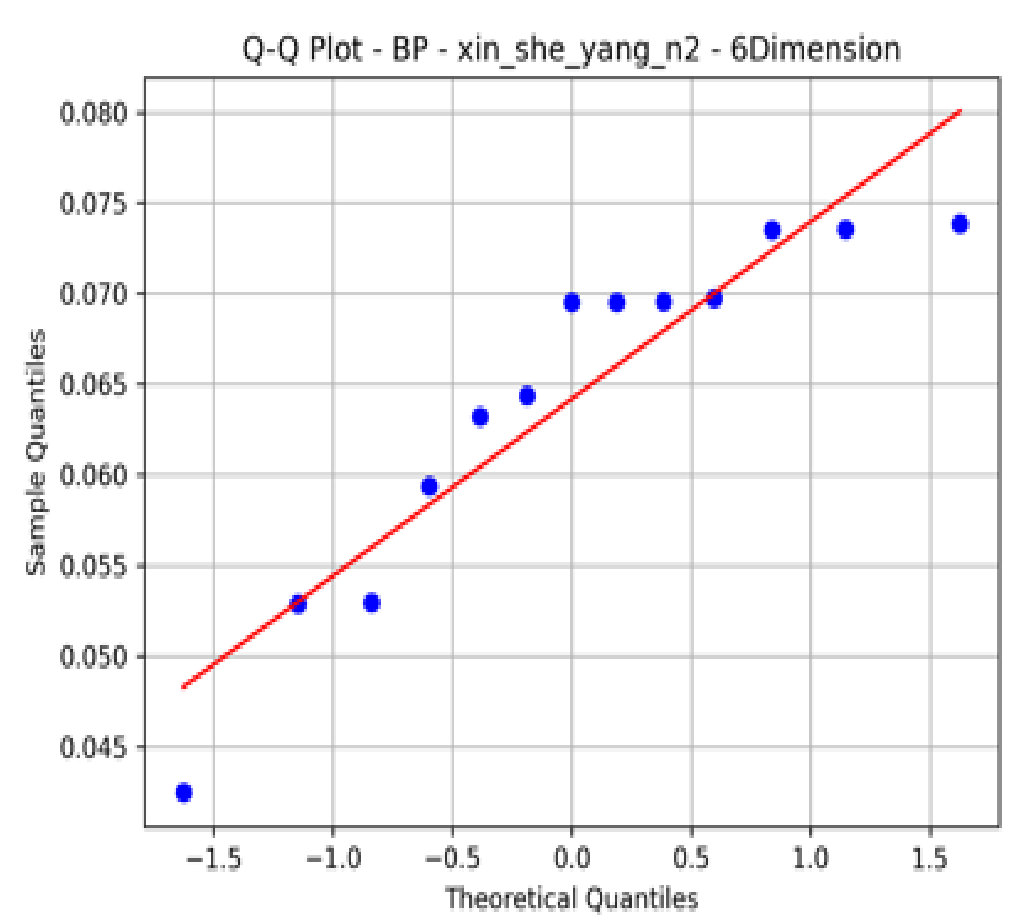
Q-Q Plot - BP - xin_she_yang_n2 - 6Dimension
Sample Quantiles
Theoretical Quantiles

SA Q-Q Plots

Continuous Functions QQ plots

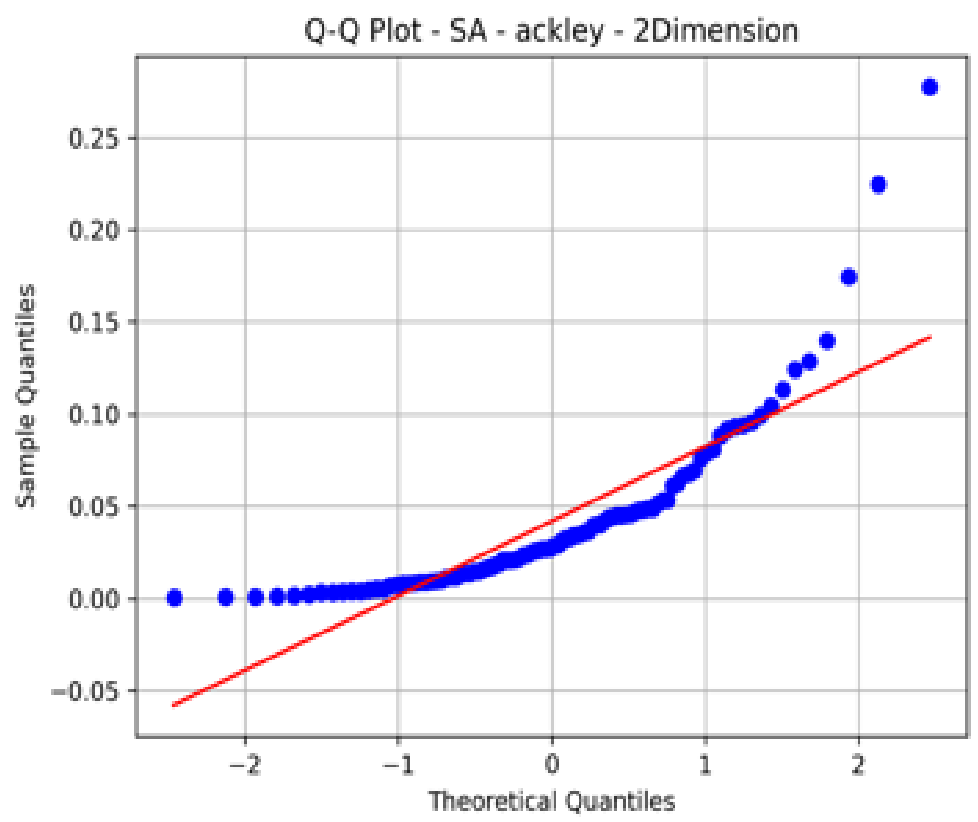

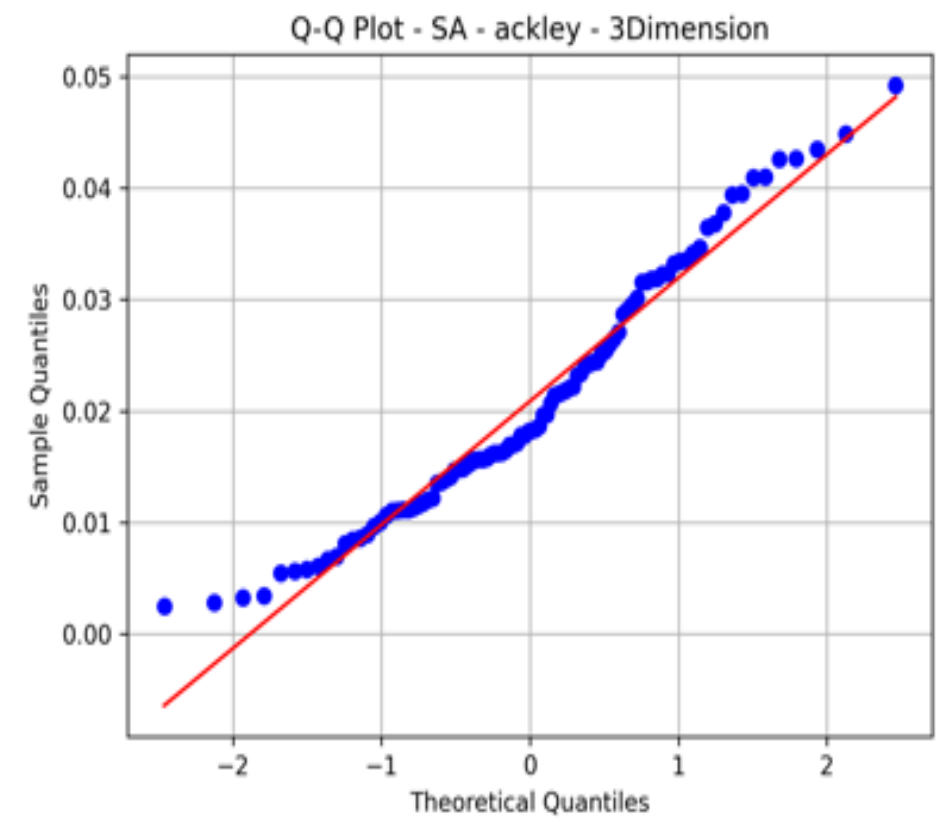

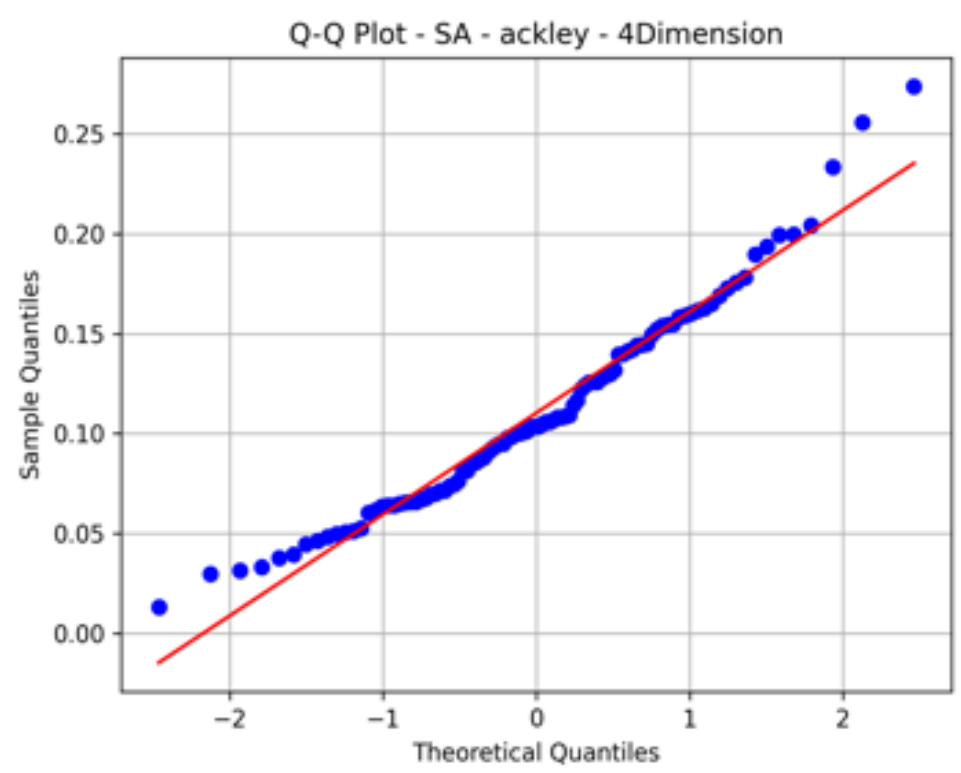

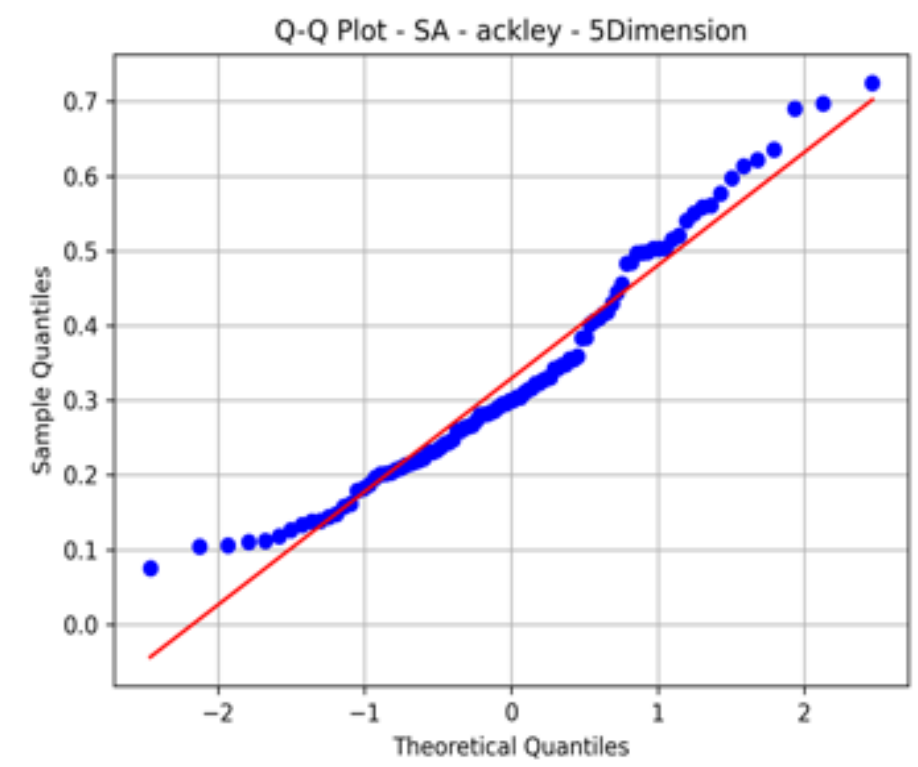

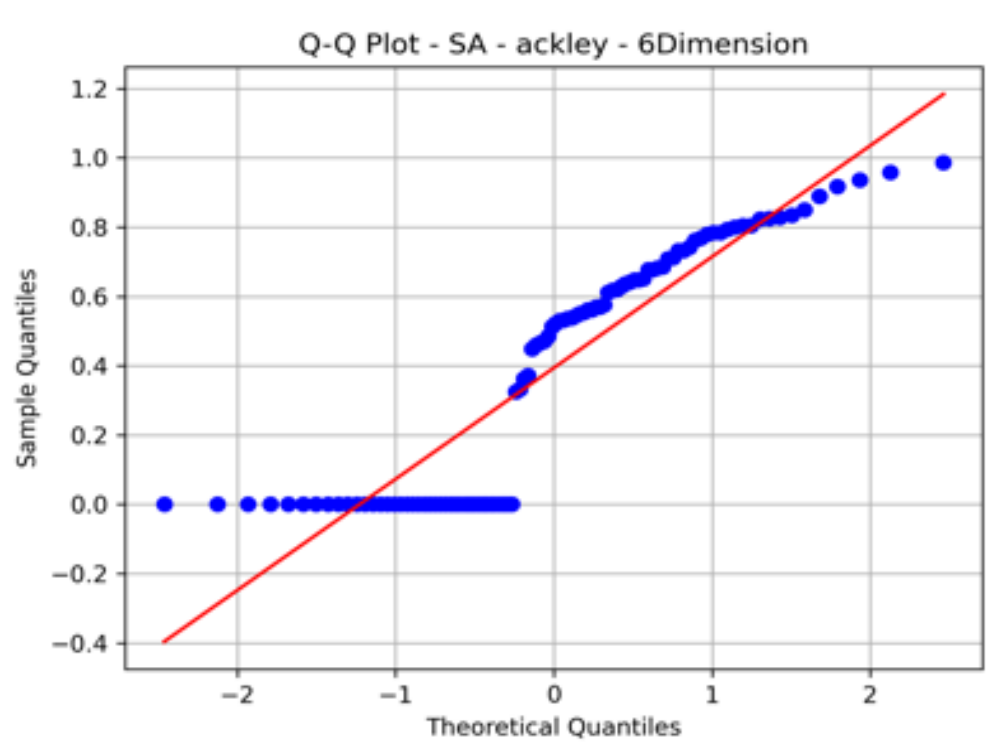

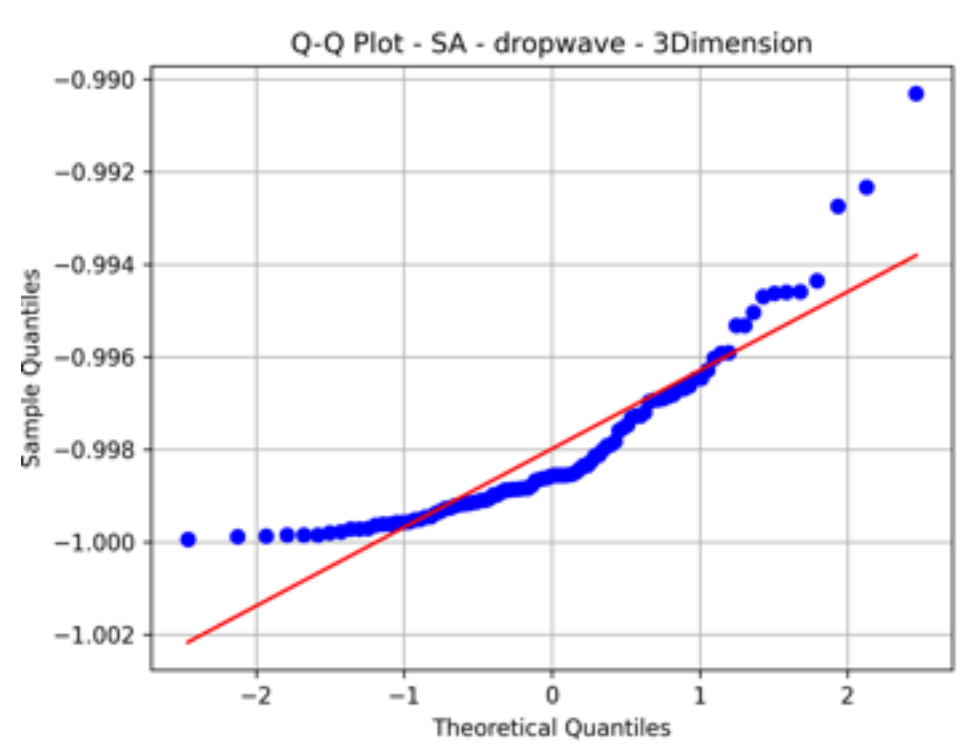

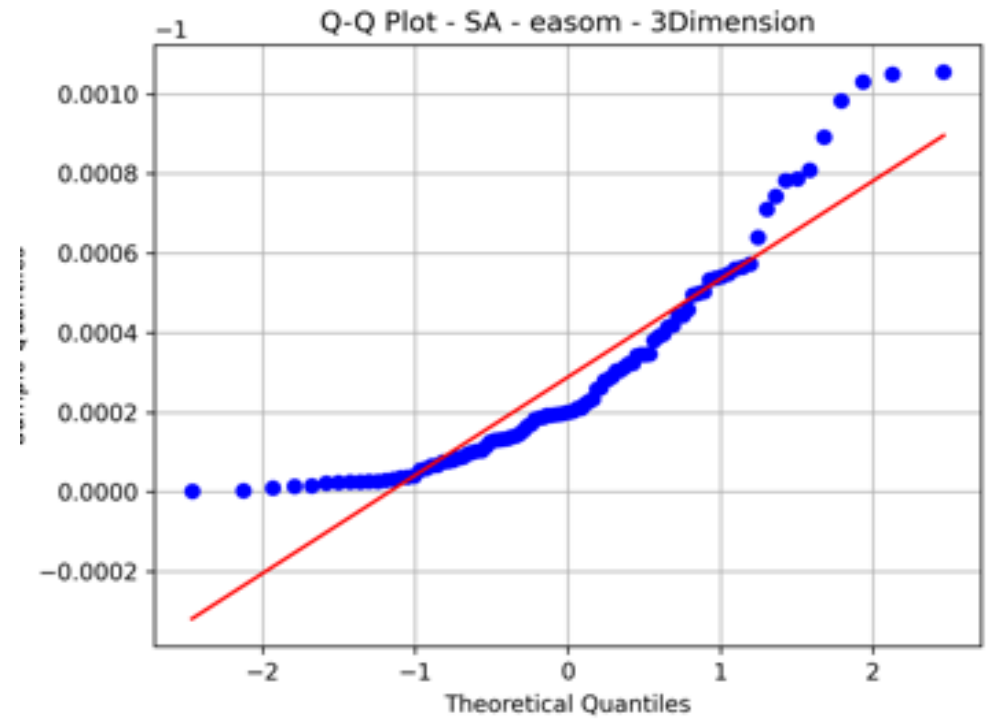
Q-Q Plot - SA - easom - 3Dimension
−1
Theoretical Quantiles

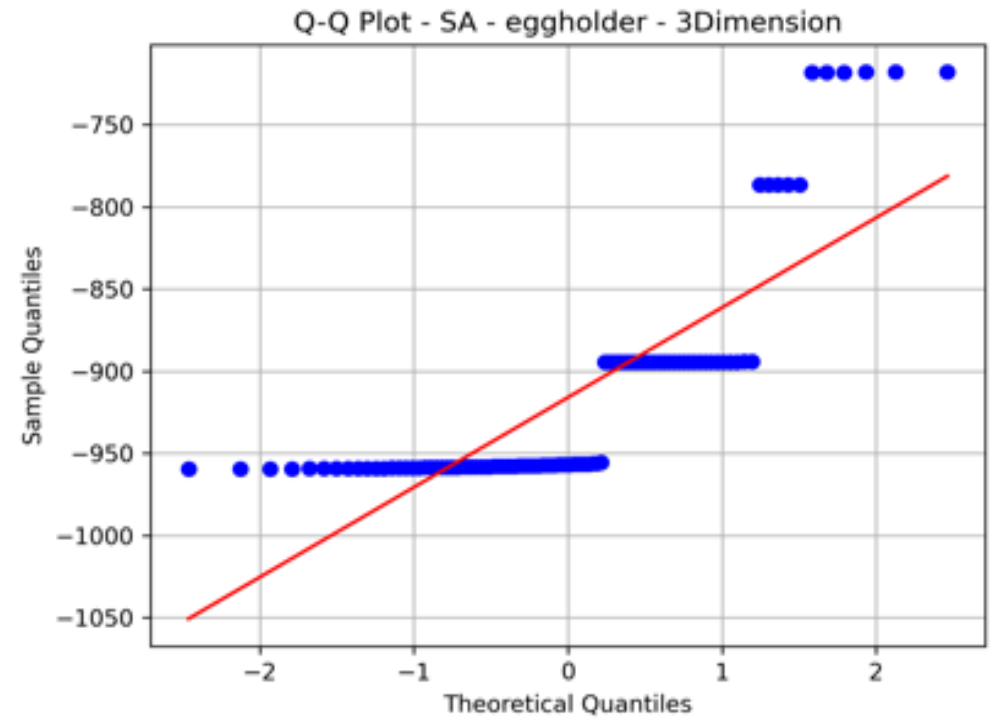
Q-Q Plot - SA - eggholder - 3Dimension
Sample Quantiles
Theoretical Quantiles

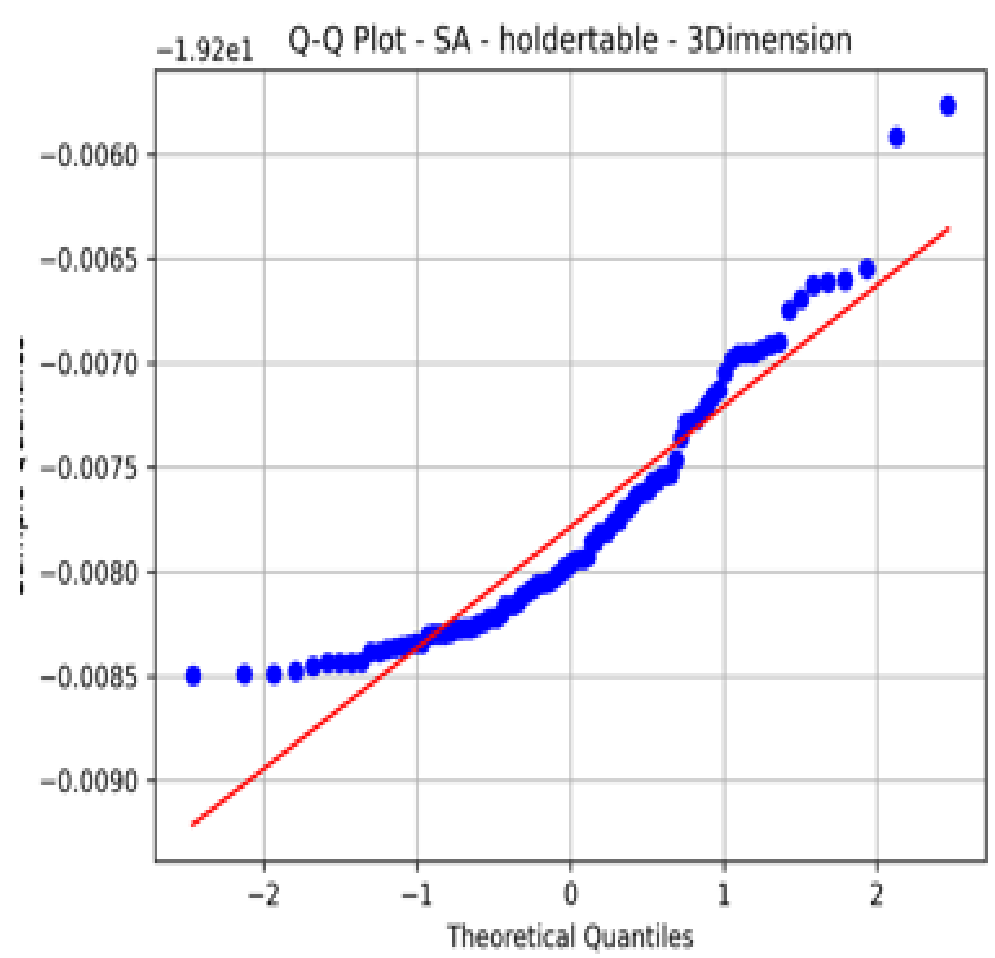
−1.92e1
Q-Q Plot - SA - holdertable - 3Dimension
Theoretical Quantiles

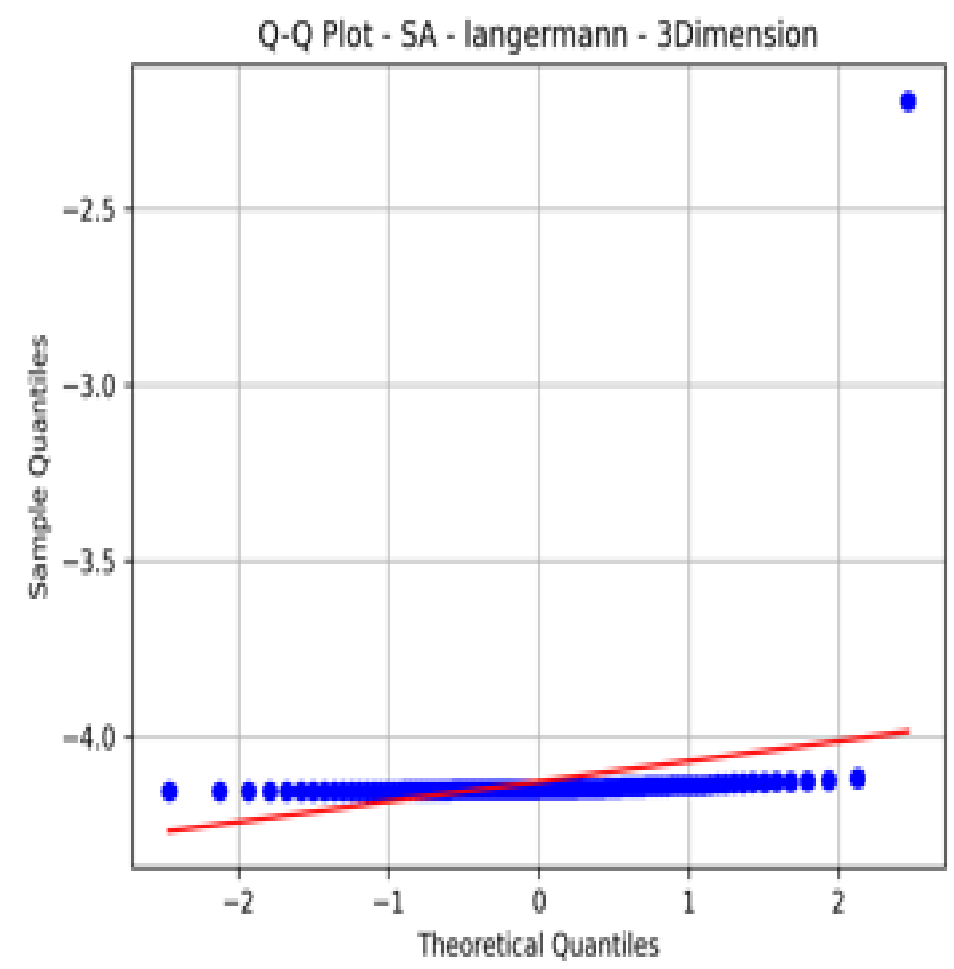
Q-Q Plot - SA - langermann - 3Dimension
Sample Quantiles
Theoretical Quantiles

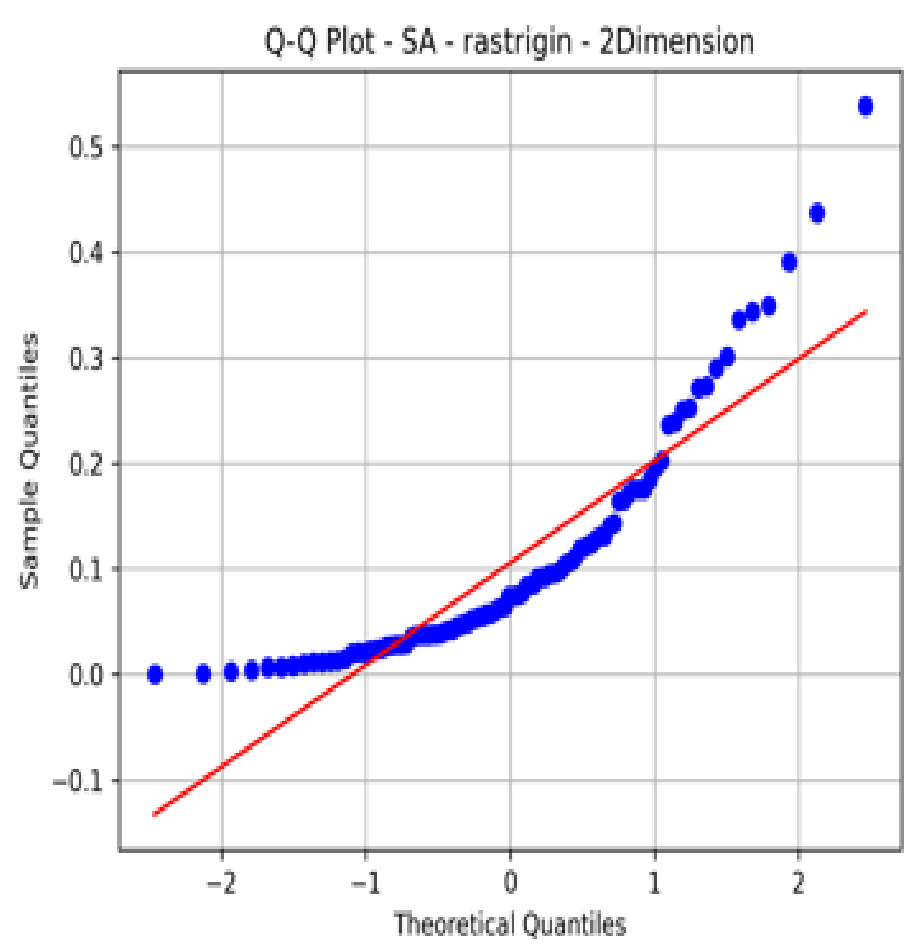
Q-Q Plot - SA - rastrigin - 2Dimension
Sample Quantiles
Theoretical Quantiles

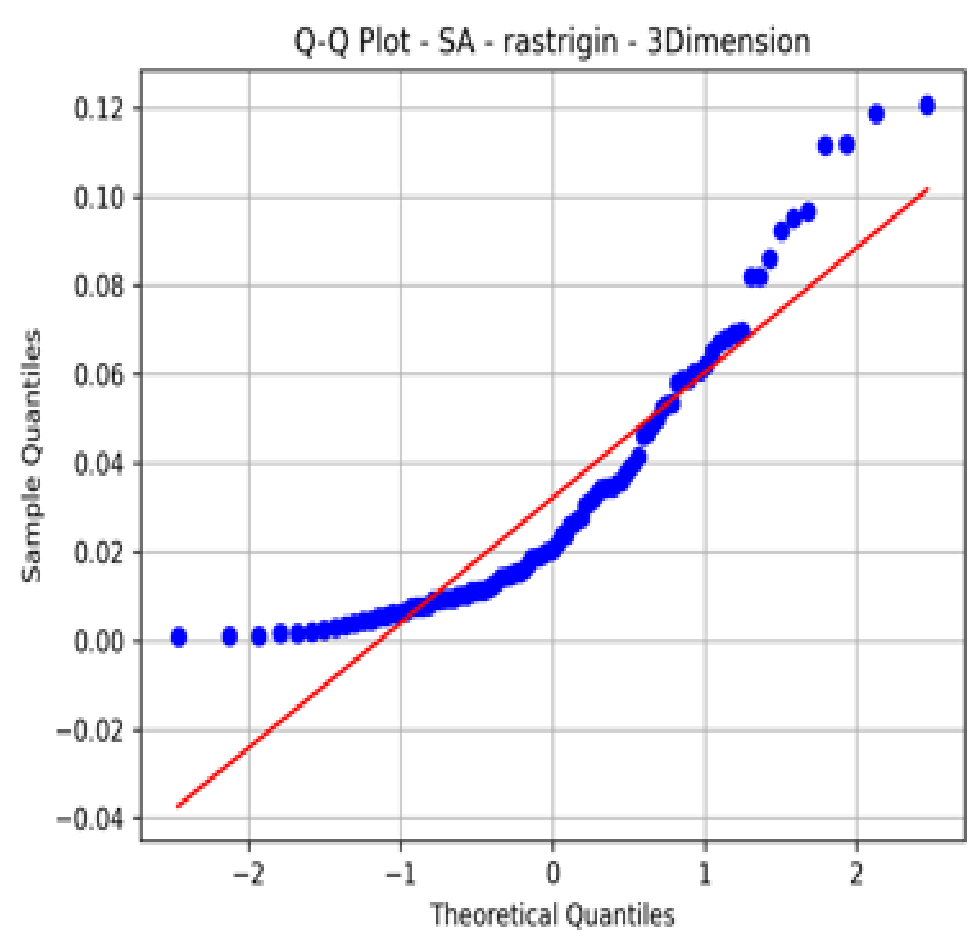
Q-Q Plot - SA - rastrigin - 3Dimension
Sample Quantiles
Theoretical Quantiles

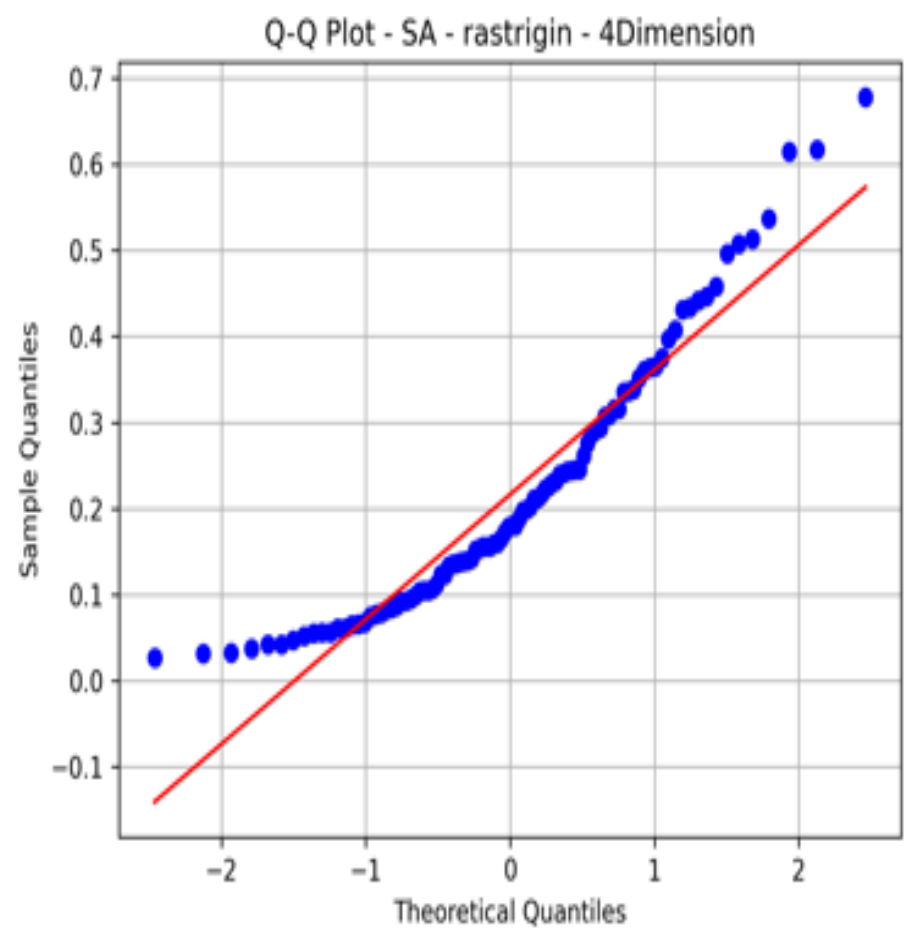
Q-Q Plot - SA - rastrigin - 4Dimension
Sample Quantiles
Theoretical Quantiles

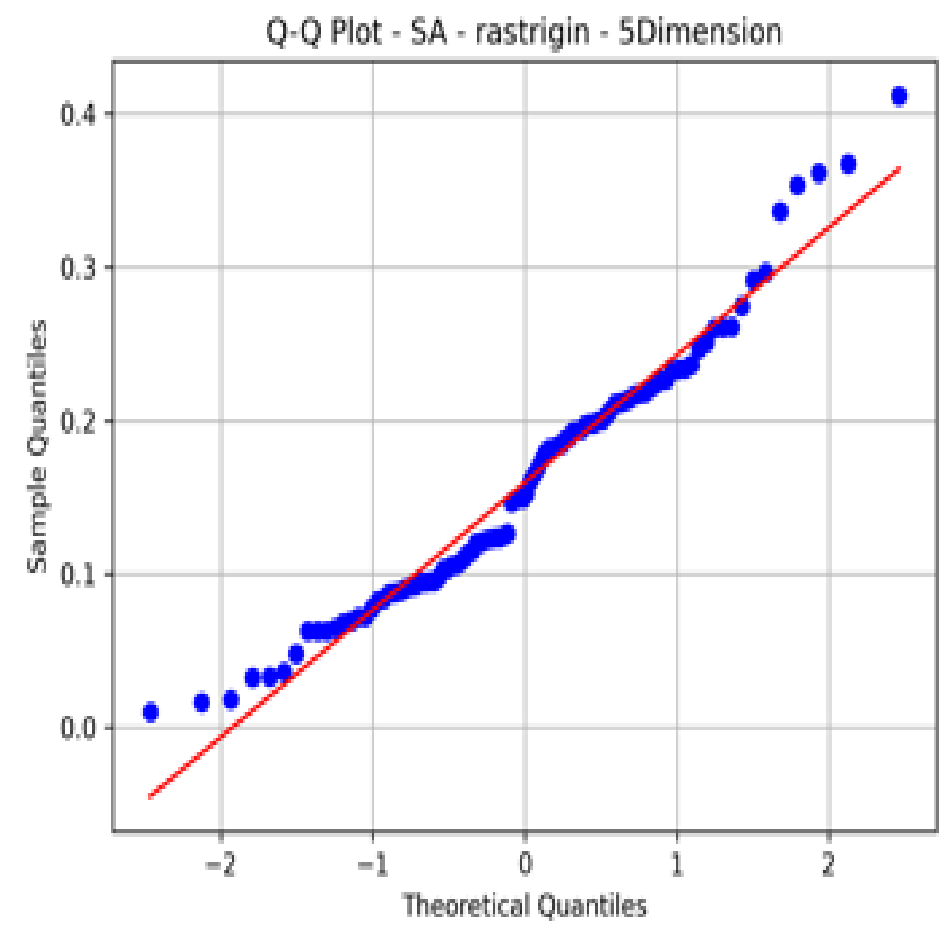
Q-Q Plot - SA - rastrigin - 5Dimension
Sample Quantiles
Theoretical Quantiles

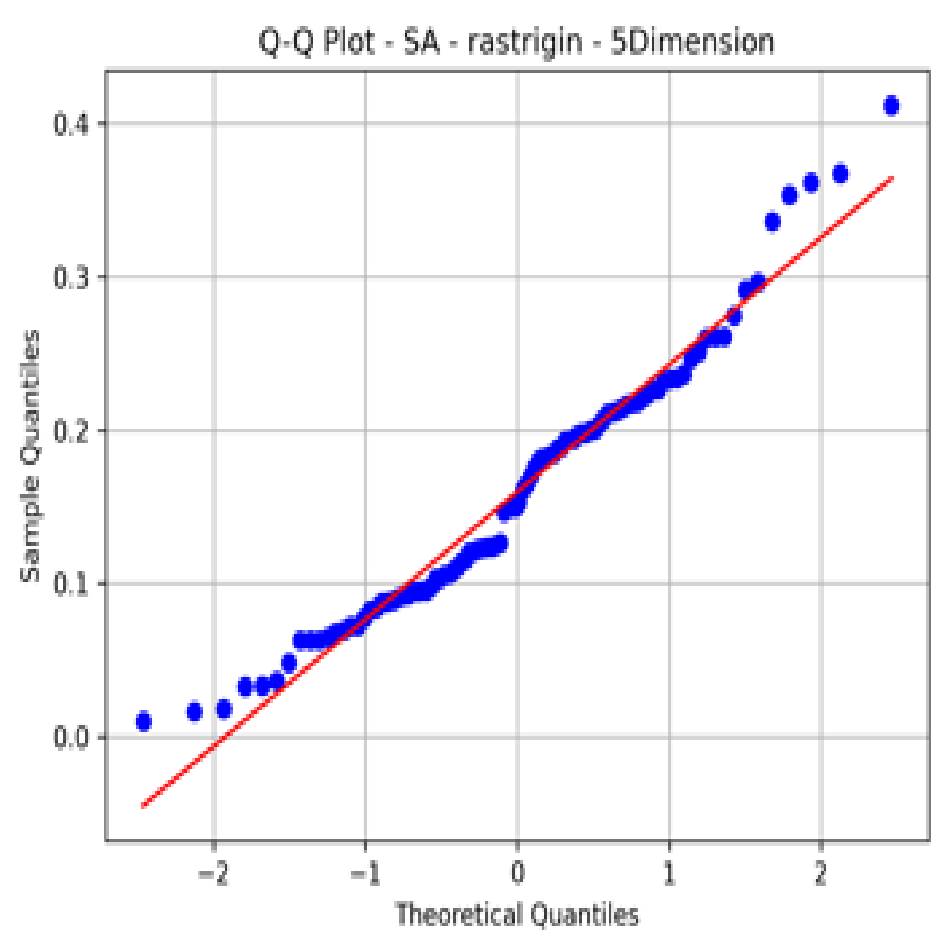
Q-Q Plot - SA - rastrigin - 5Dimension
Sample Quantiles
Theoretical Quantiles

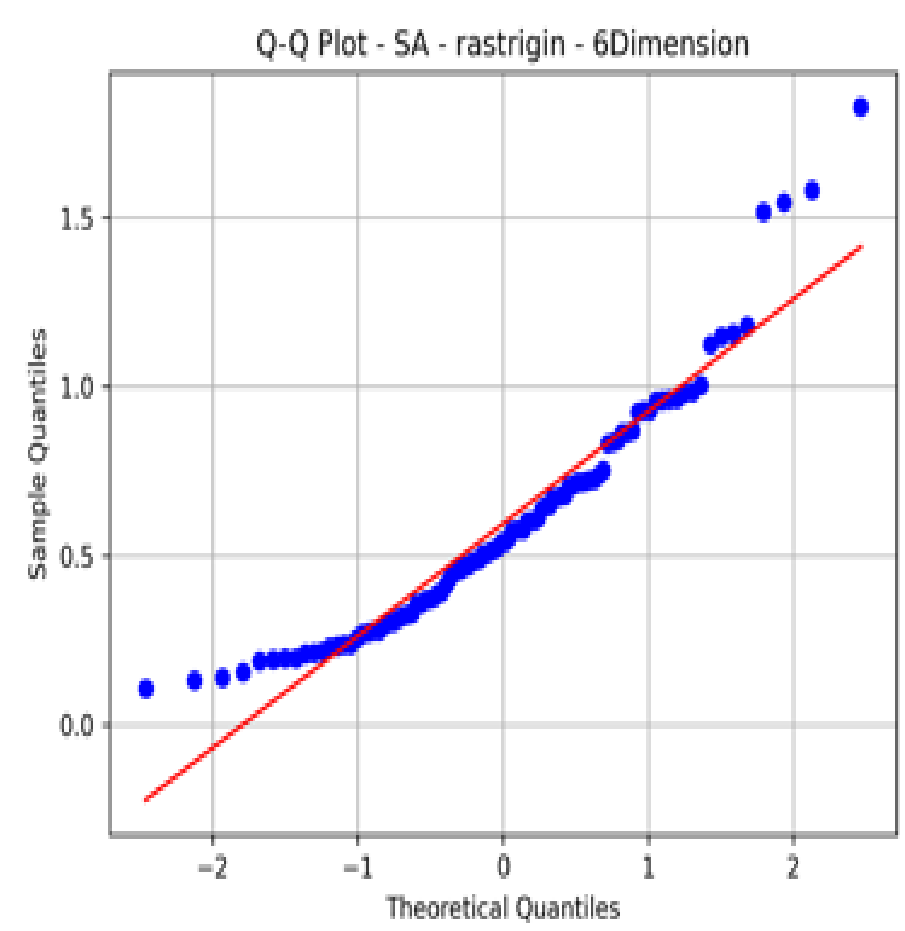
Q-Q Plot - SA - rastrigin - 6Dimension
Sample Quantiles
Theoretical Quantiles

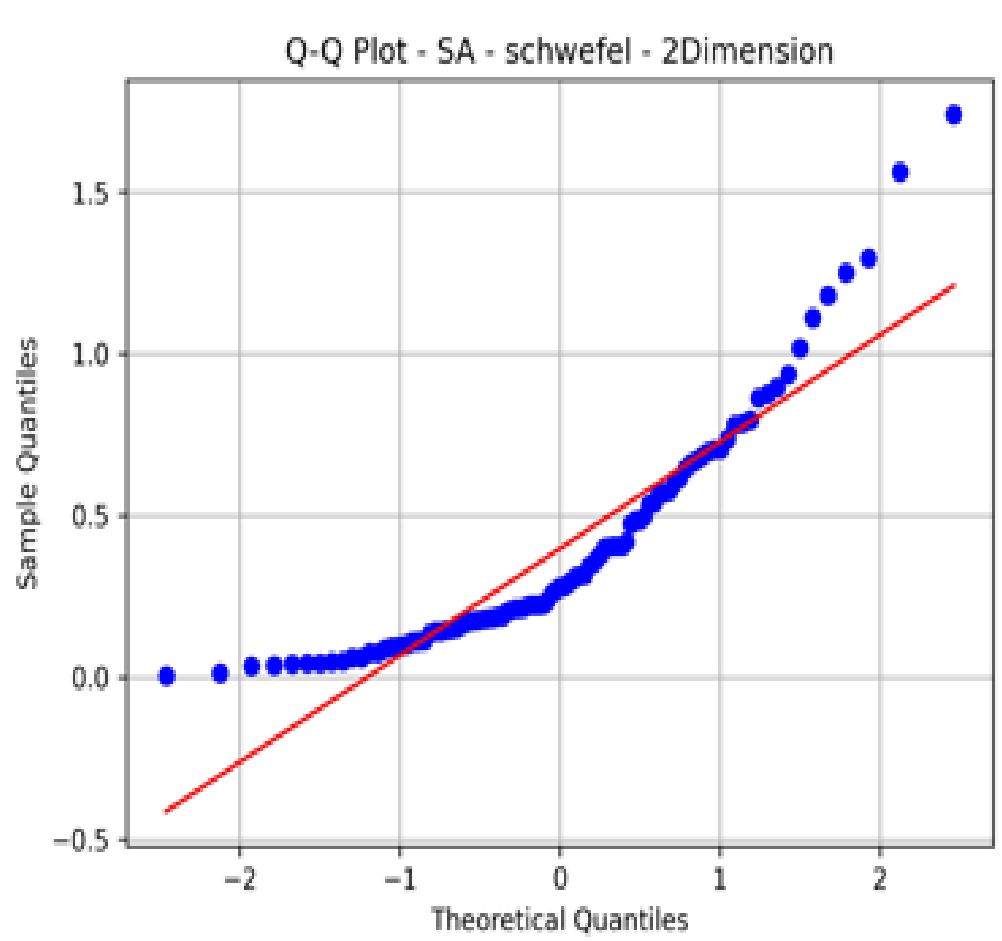
Q-Q Plot - SA - schwefel - 2Dimension
Sample Quantiles
Theoretical Quantiles

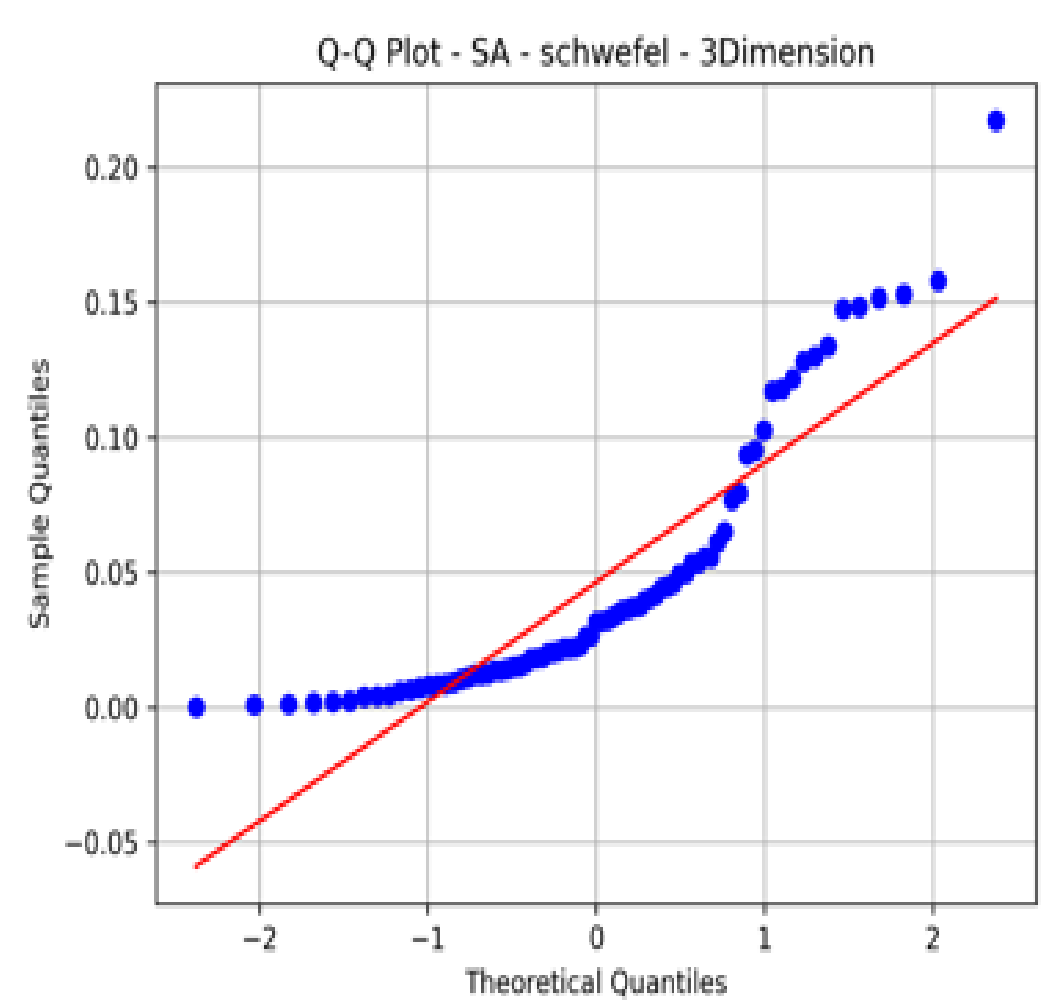
Q-Q Plot - SA - schwefel - 3Dimension
Sample Quantiles
Theoretical Quantiles

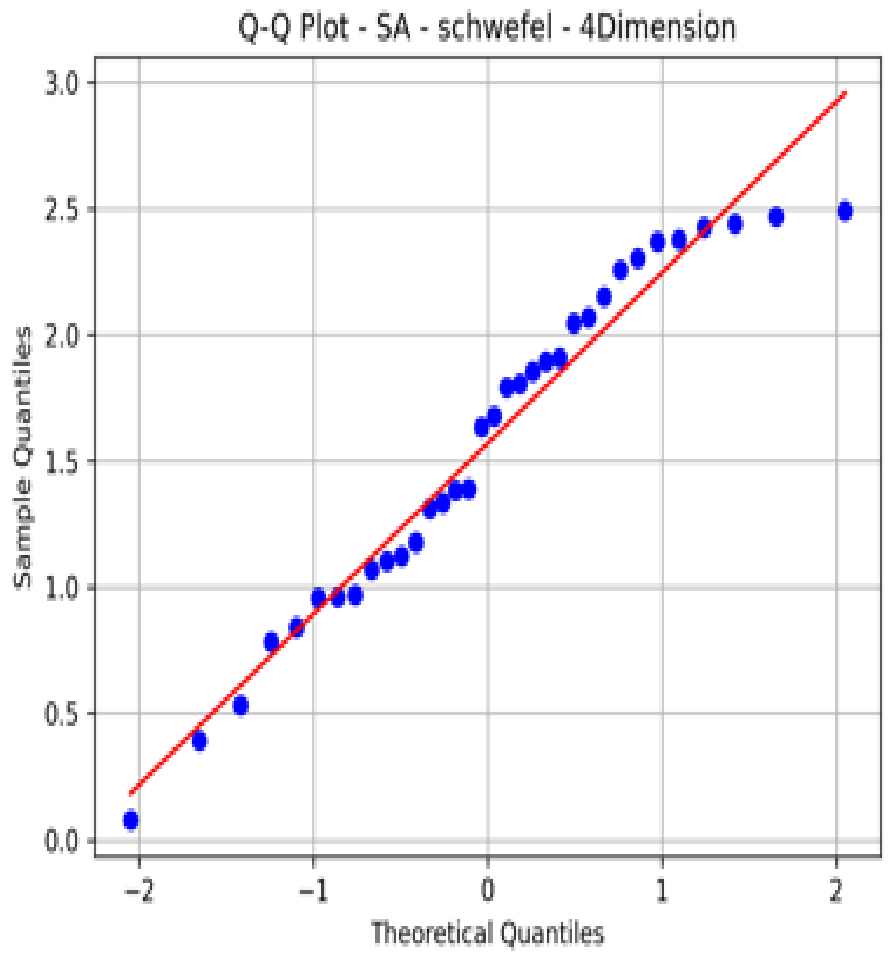
Q-Q Plot - SA - schwefel - 4Dimension
Sample Quantiles
Theoretical Quantiles

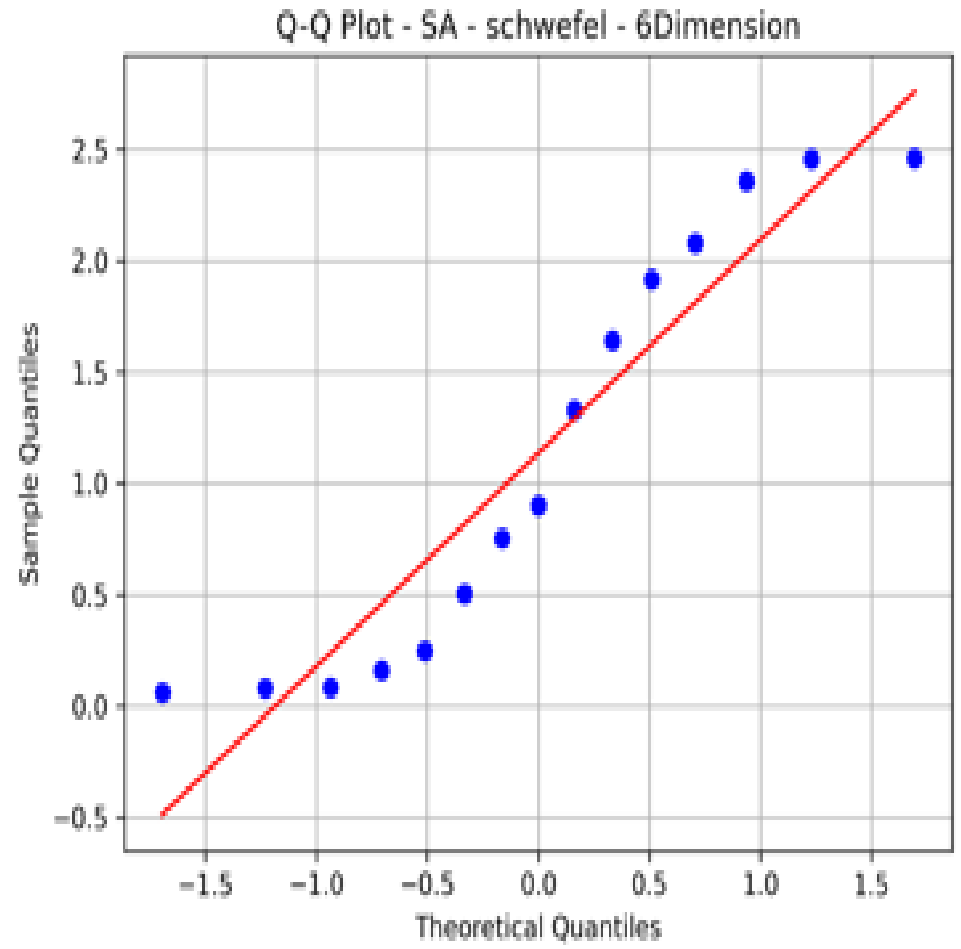
Q-Q Plot - SA - schwefel - 6Dimension
Sample Quantiles
Theoretical Quantiles

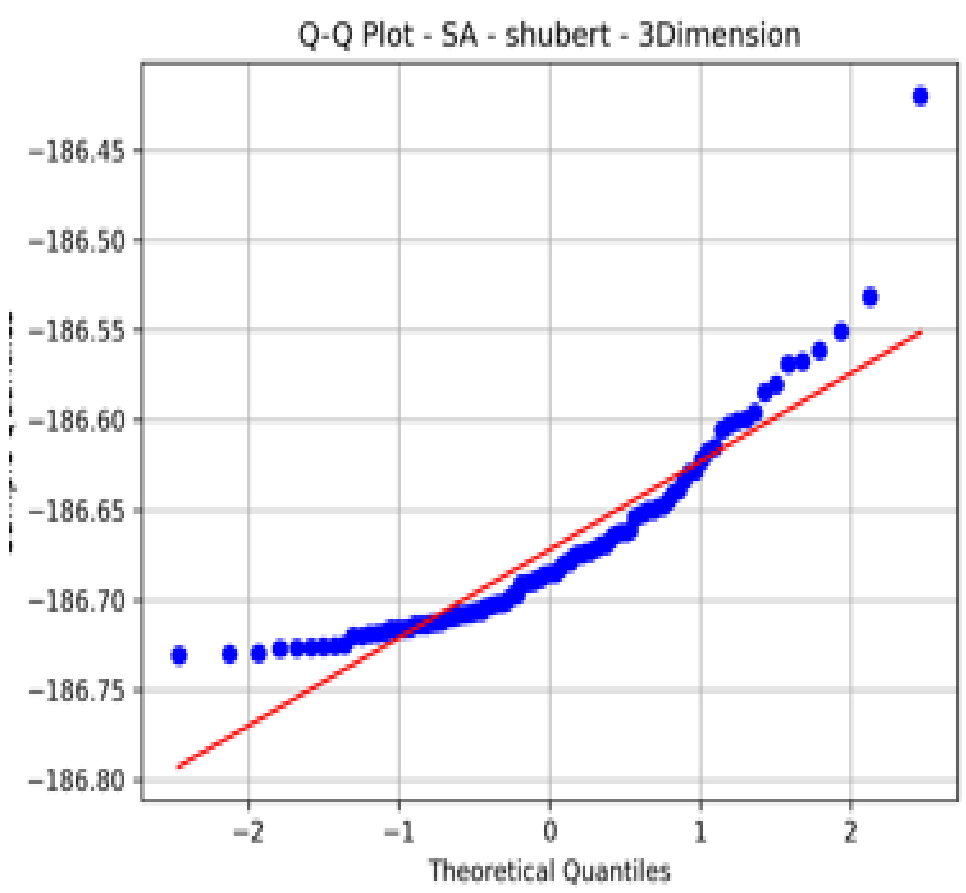
Q-Q Plot - SA - shubert - 3Dimension
Theoretical Quantiles

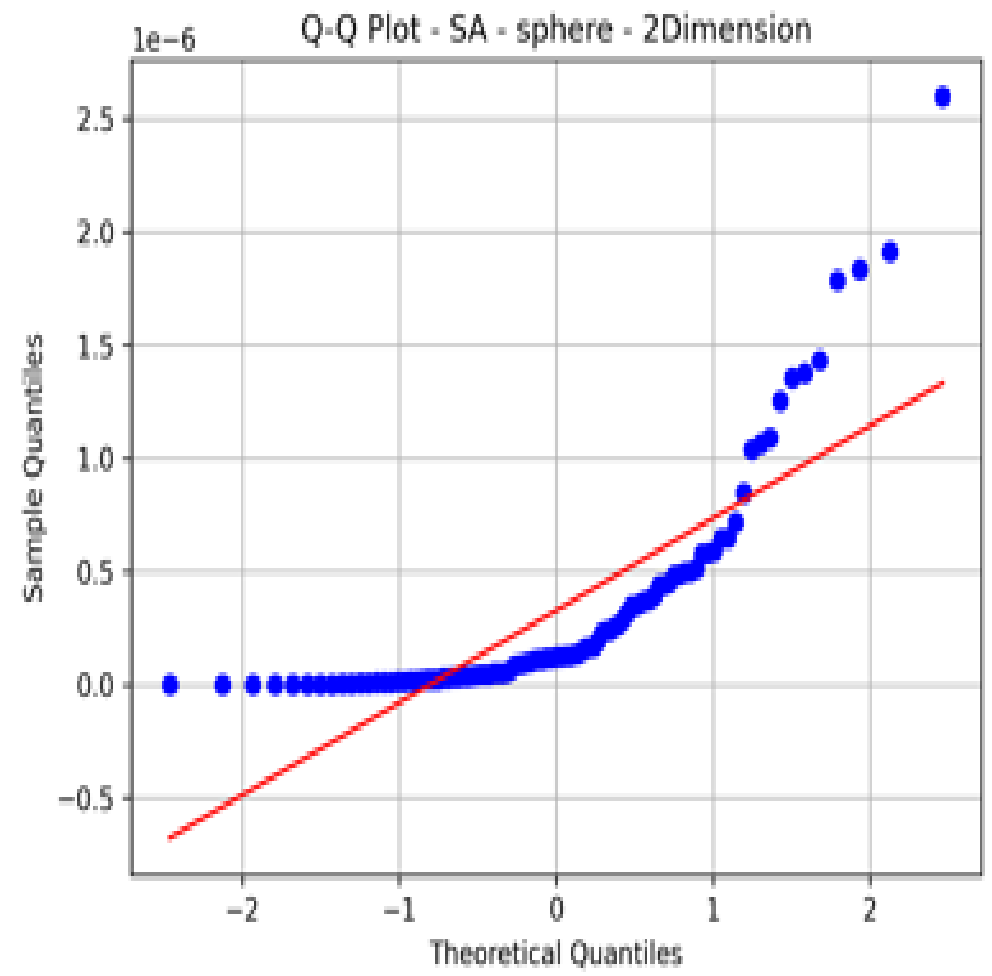
1e−6
Q-Q Plot - SA - sphere - 2Dimension
Sample Quantiles
Theoretical Quantiles

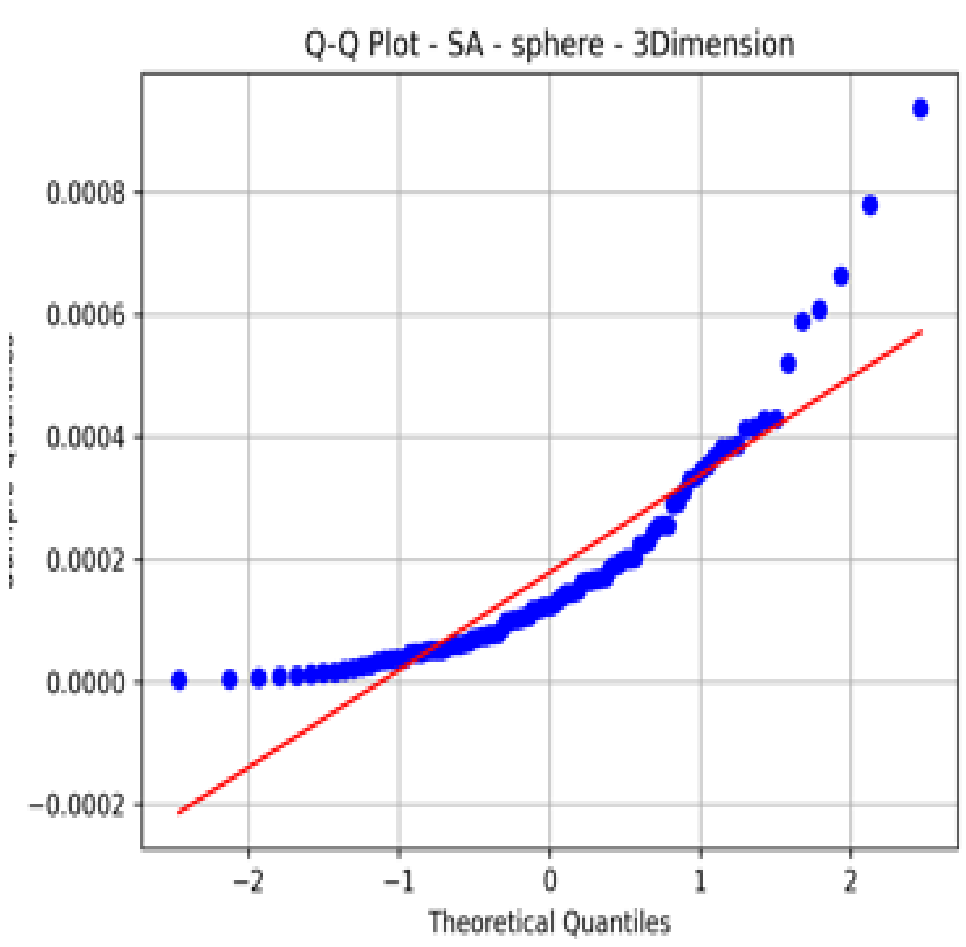
Q-Q Plot - SA - sphere - 3Dimension
Theoretical Quantiles

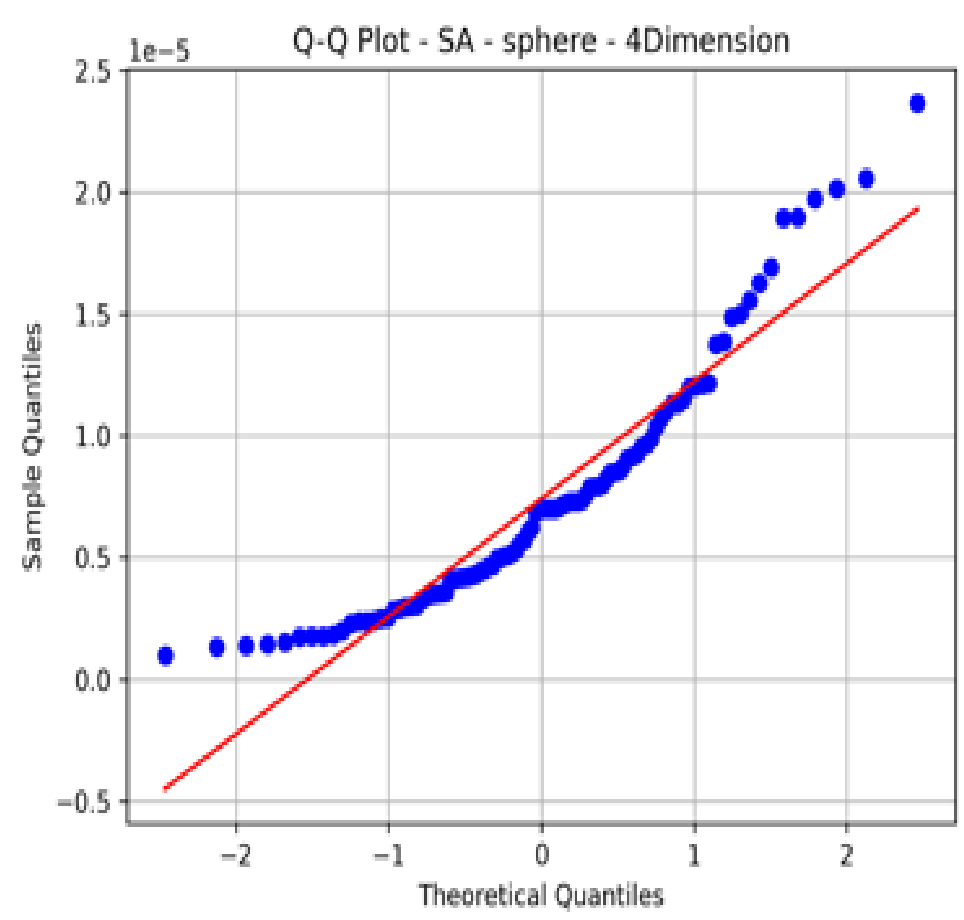
1e−5
Q-Q Plot - SA - sphere - 4Dimension
Sample Quantiles
Theoretical Quantiles

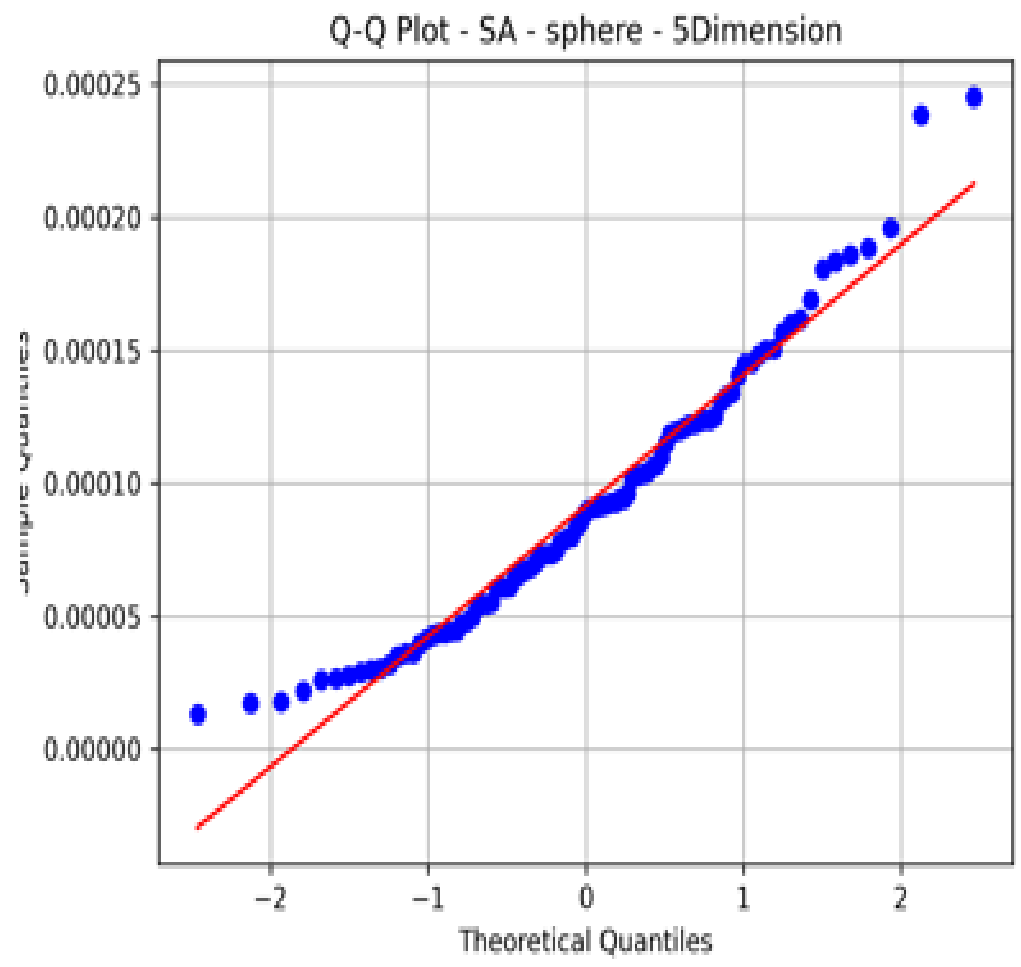

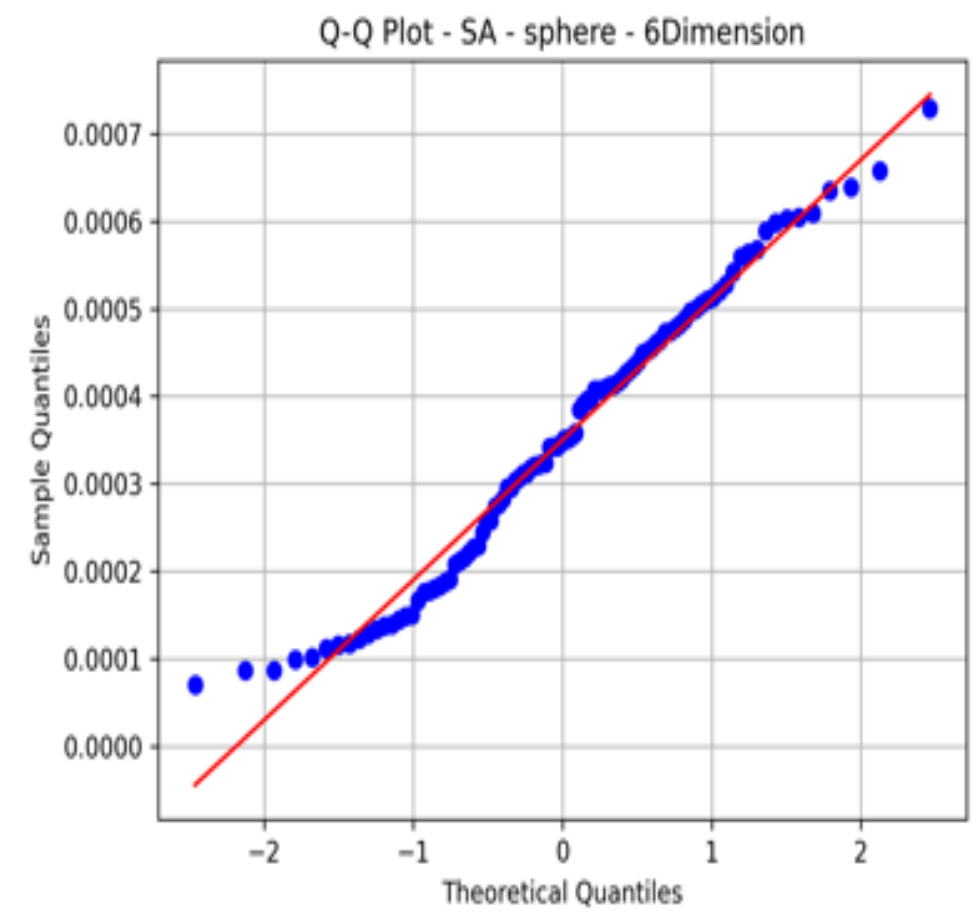

Non-Continuous functions QQ plots

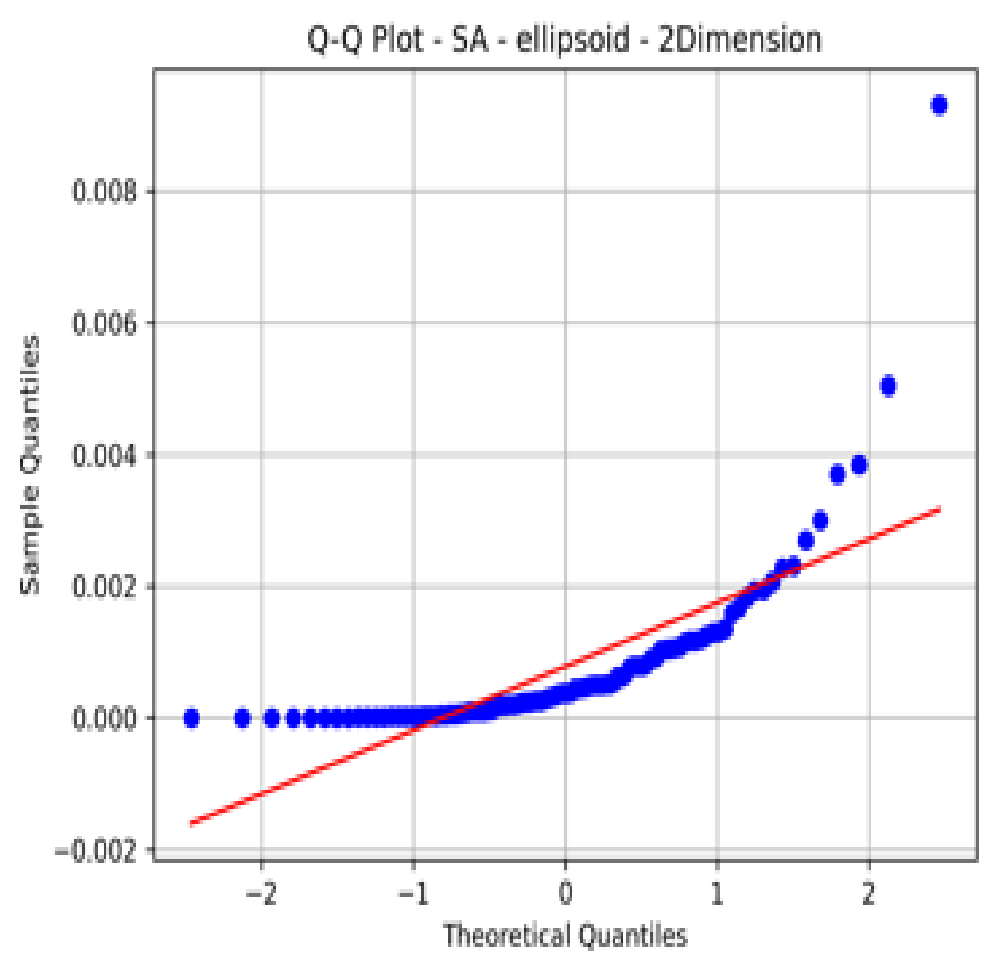

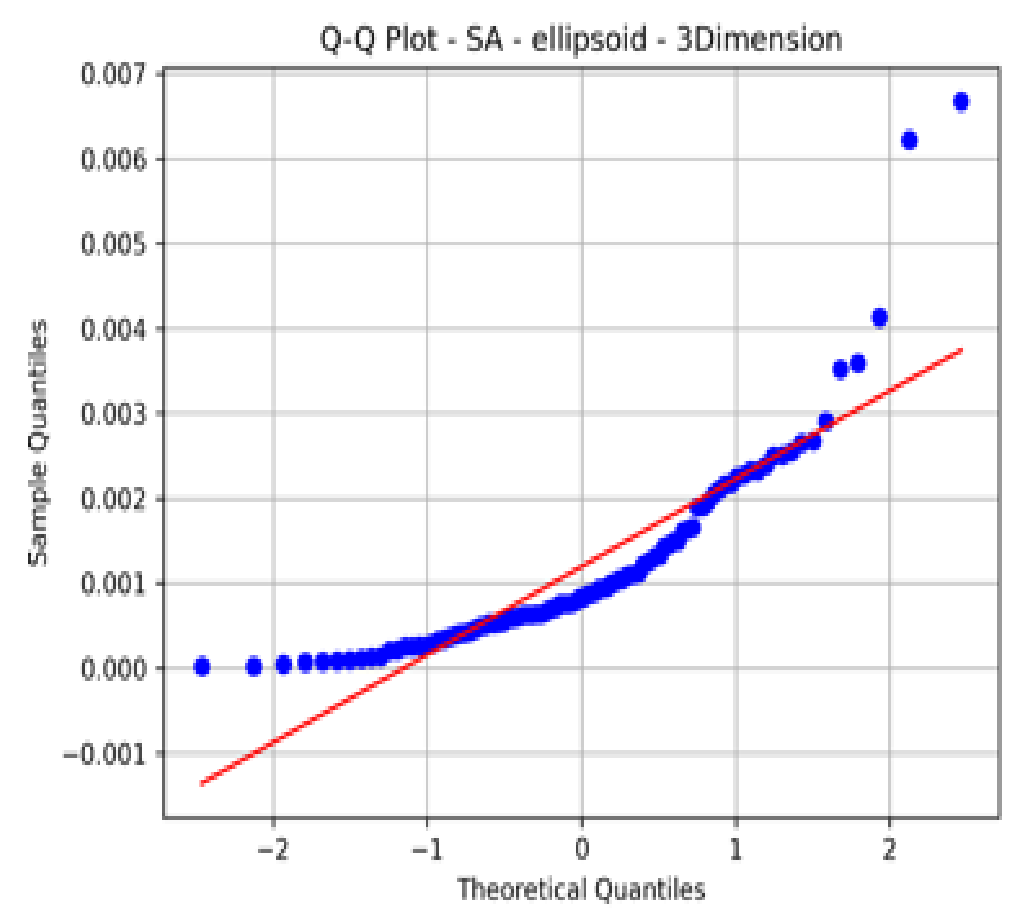

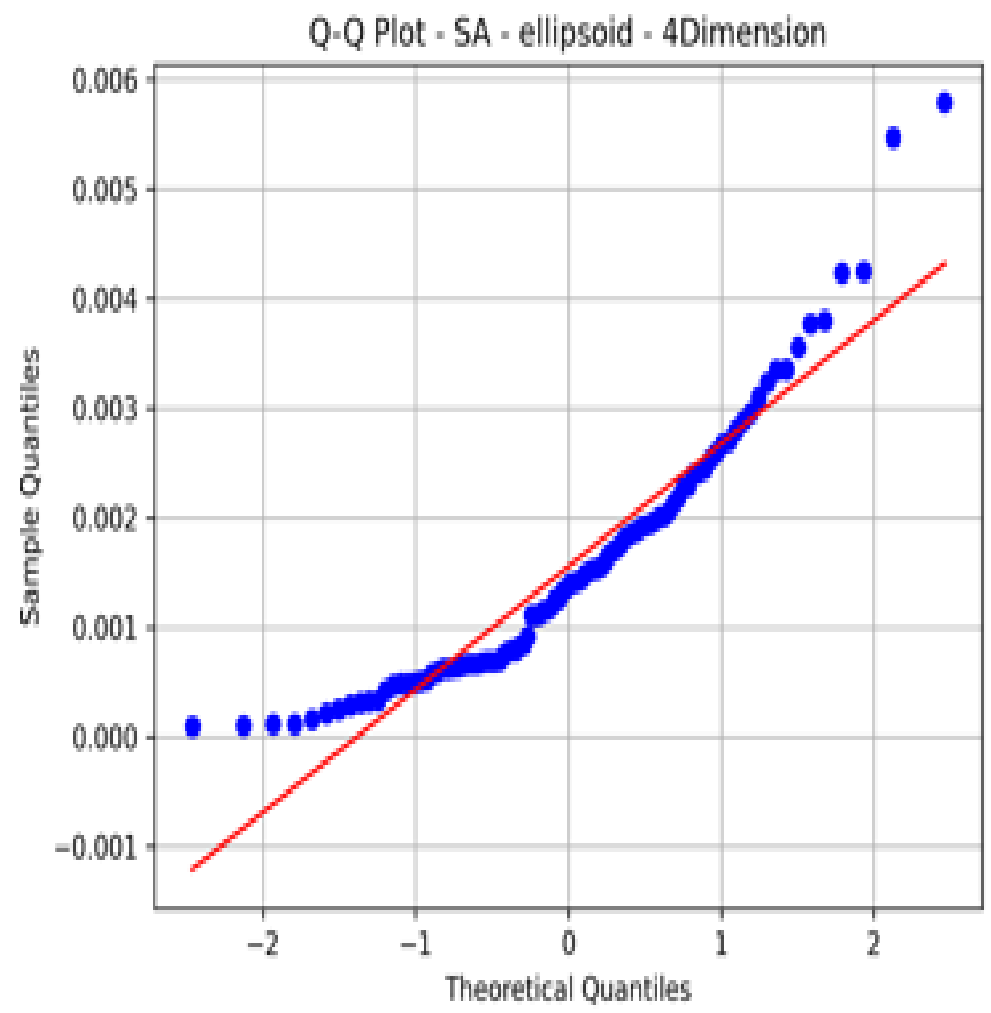

Q-Q Plot - SA - ellipsoid - 4Dimension
Sample Quantiles
Theoretical Quantiles

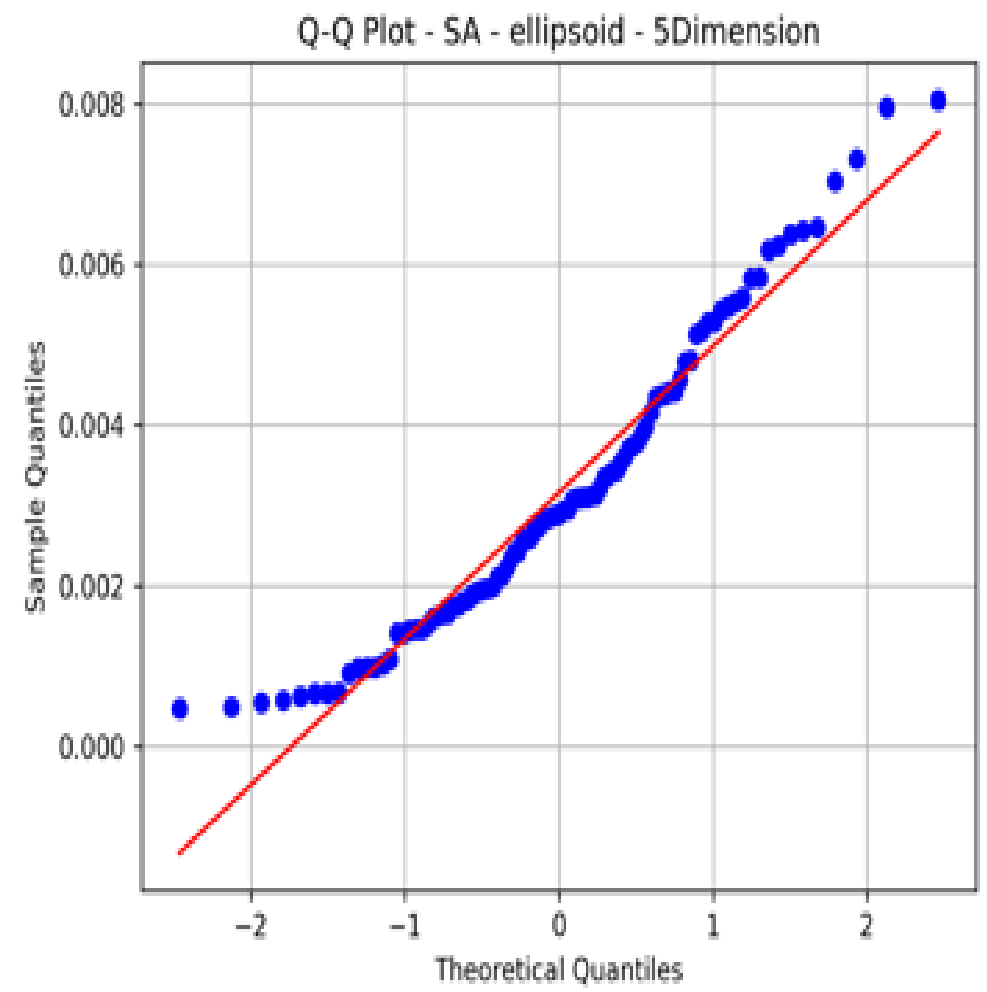

Q-Q Plot - SA - ellipsoid - 5Dimension
Sample Quantiles
Theoretical Quantiles

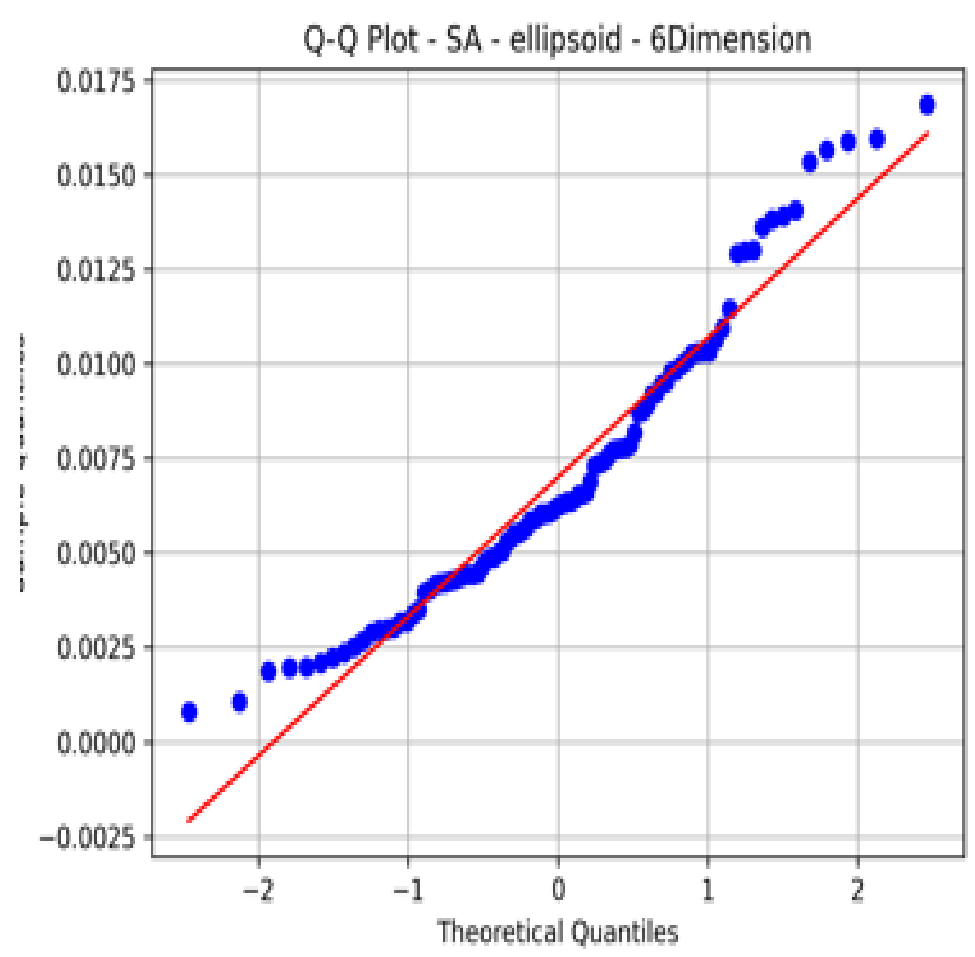

Q-Q Plot - SA - ellipsoid - 6Dimension
Theoretical Quantiles

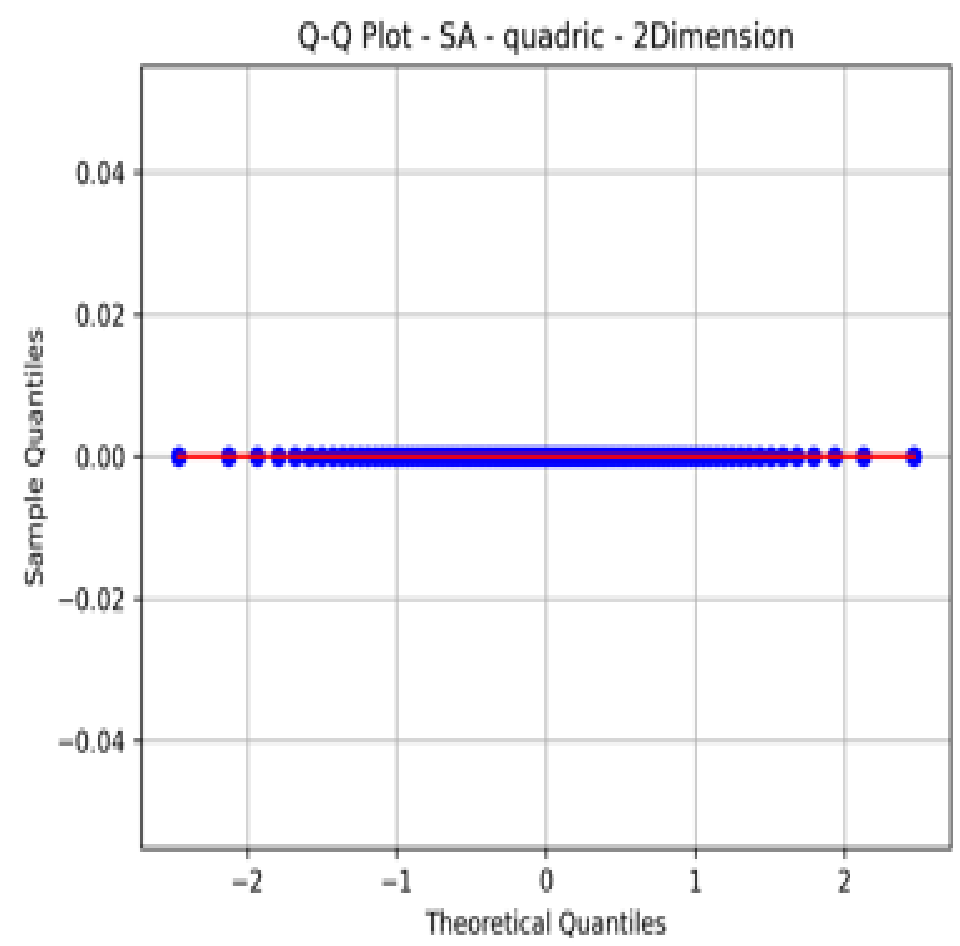

Q-Q Plot - SA - quadric - 2Dimension
Sample Quantiles
Theoretical Quantiles

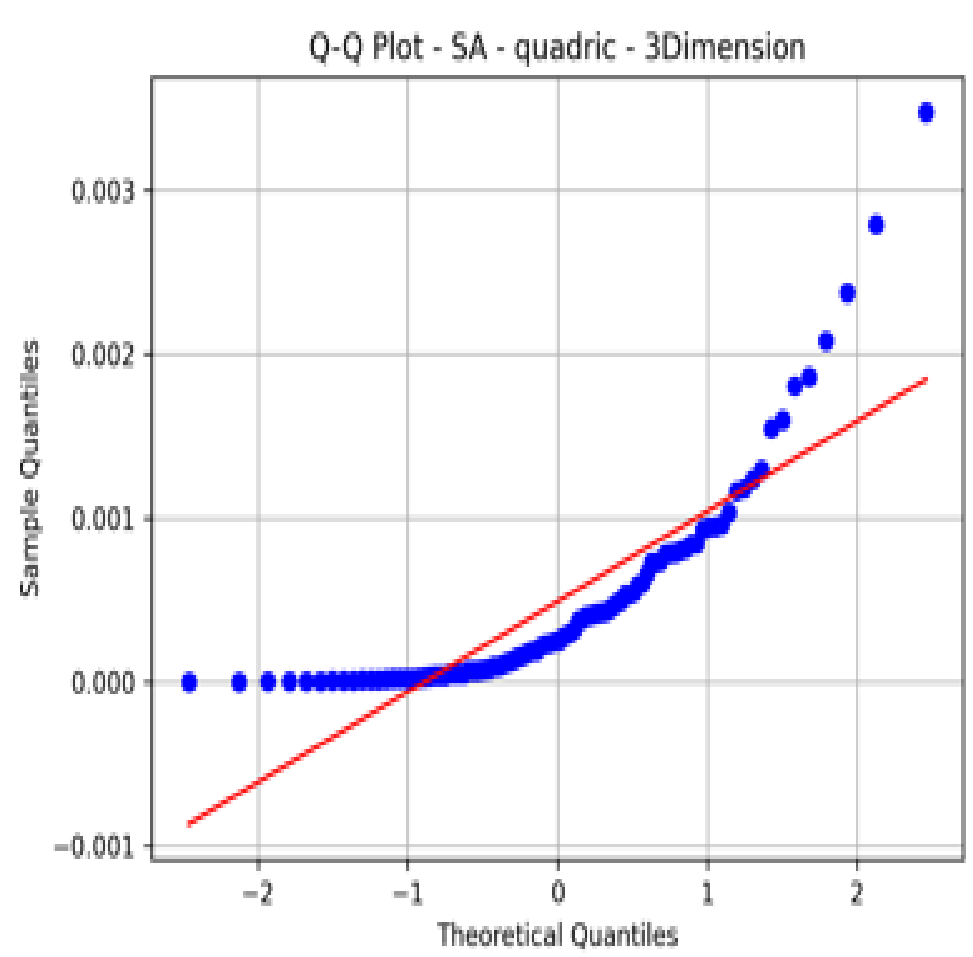

Q-Q Plot - SA - quadric - 3Dimension
Sample Quantiles
Theoretical Quantiles

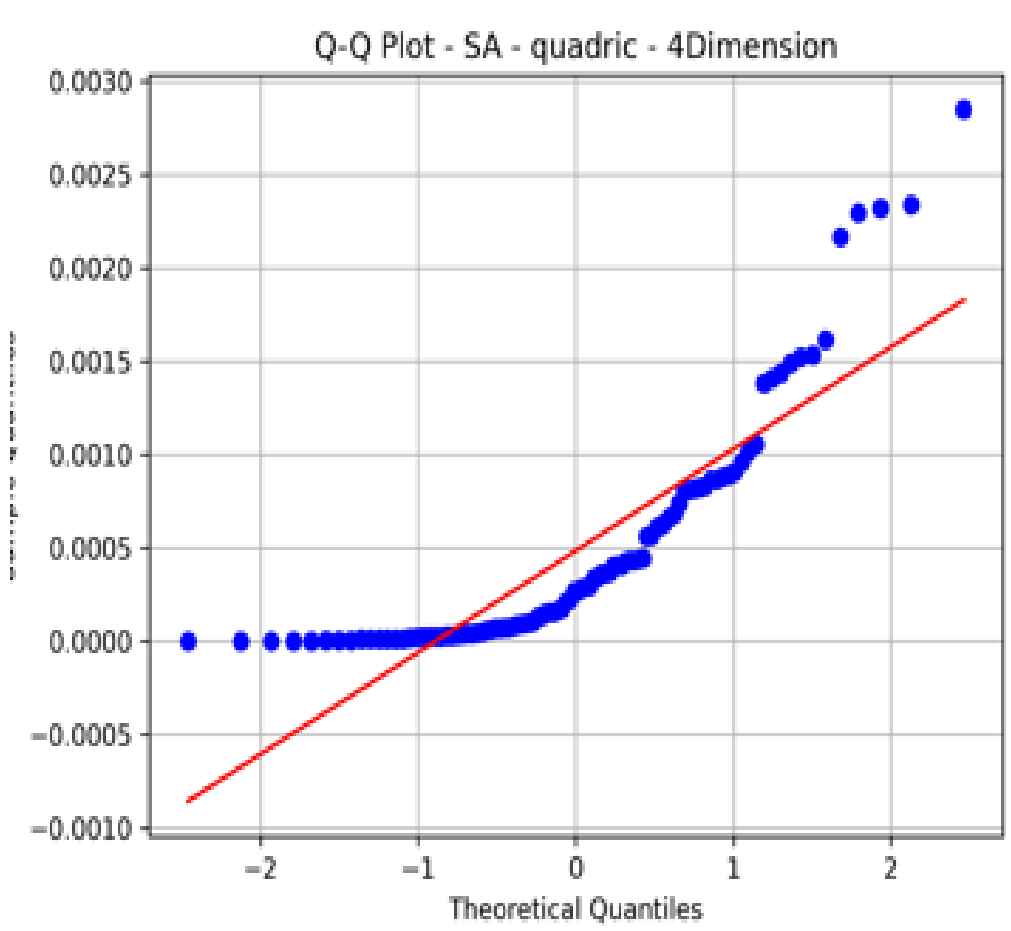

Q-Q Plot - SA - quadric - 4Dimension
Theoretical Quantiles

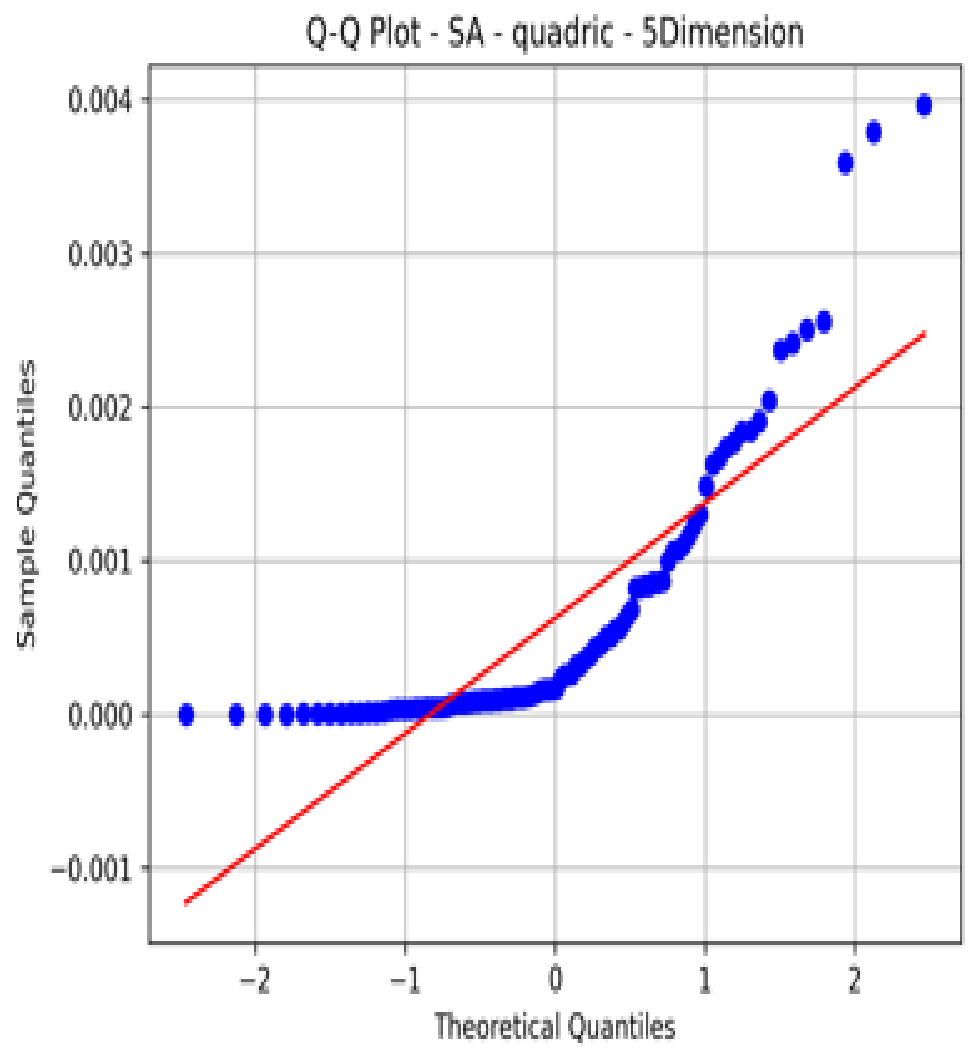
Q-Q Plot - SA - quadric - 5Dimension
Sample Quantiles
Theoretical Quantiles

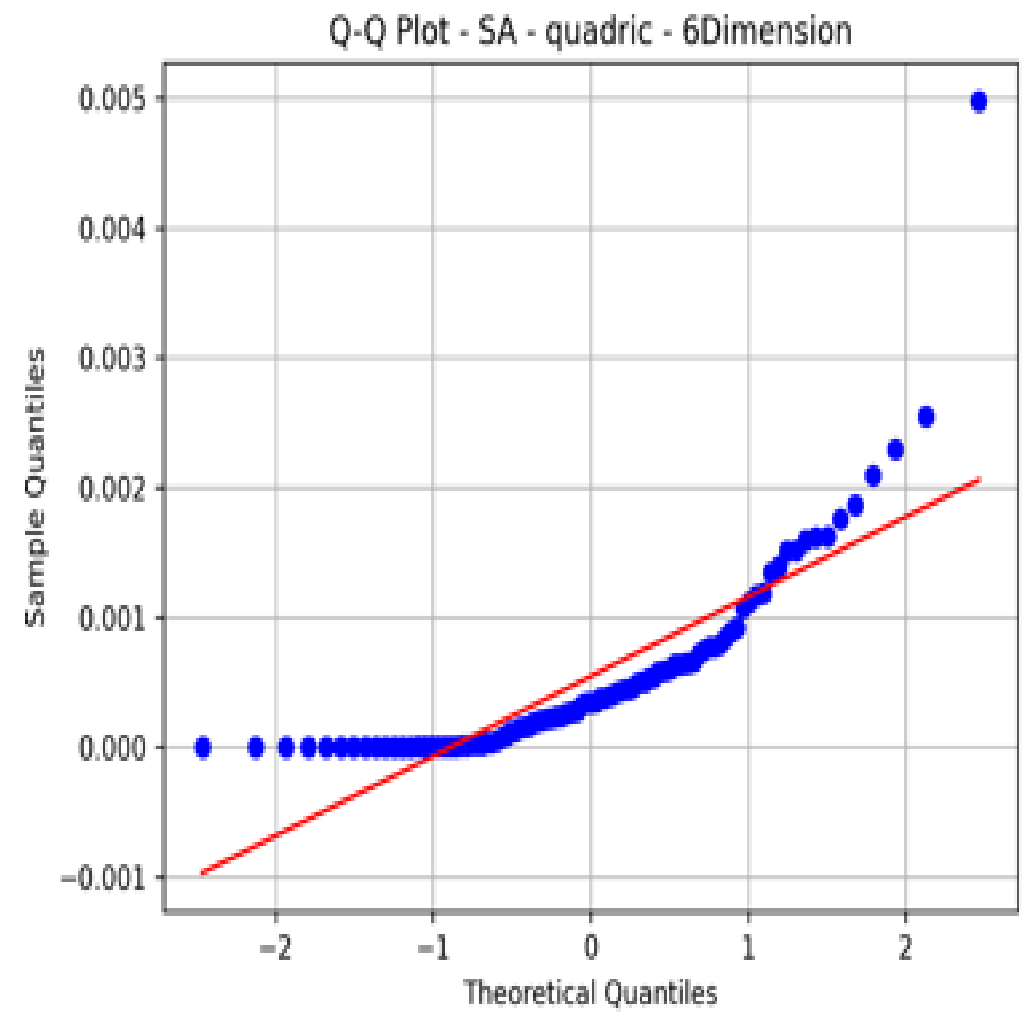
Q-Q Plot - SA - quadric - 6Dimension
Sample Quantiles
Theoretical Quantiles

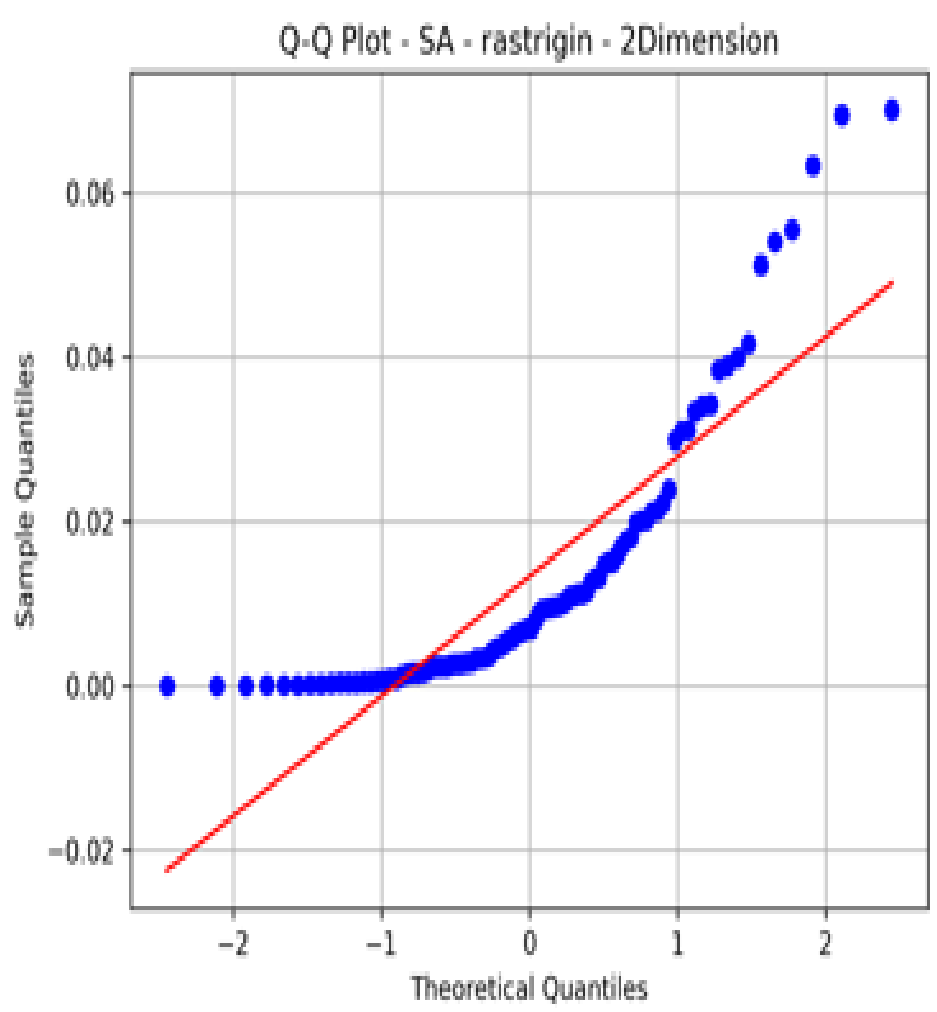
Q-Q Plot - SA - rastrigin - 2Dimension
Sample Quantiles
Theoretical Quantiles

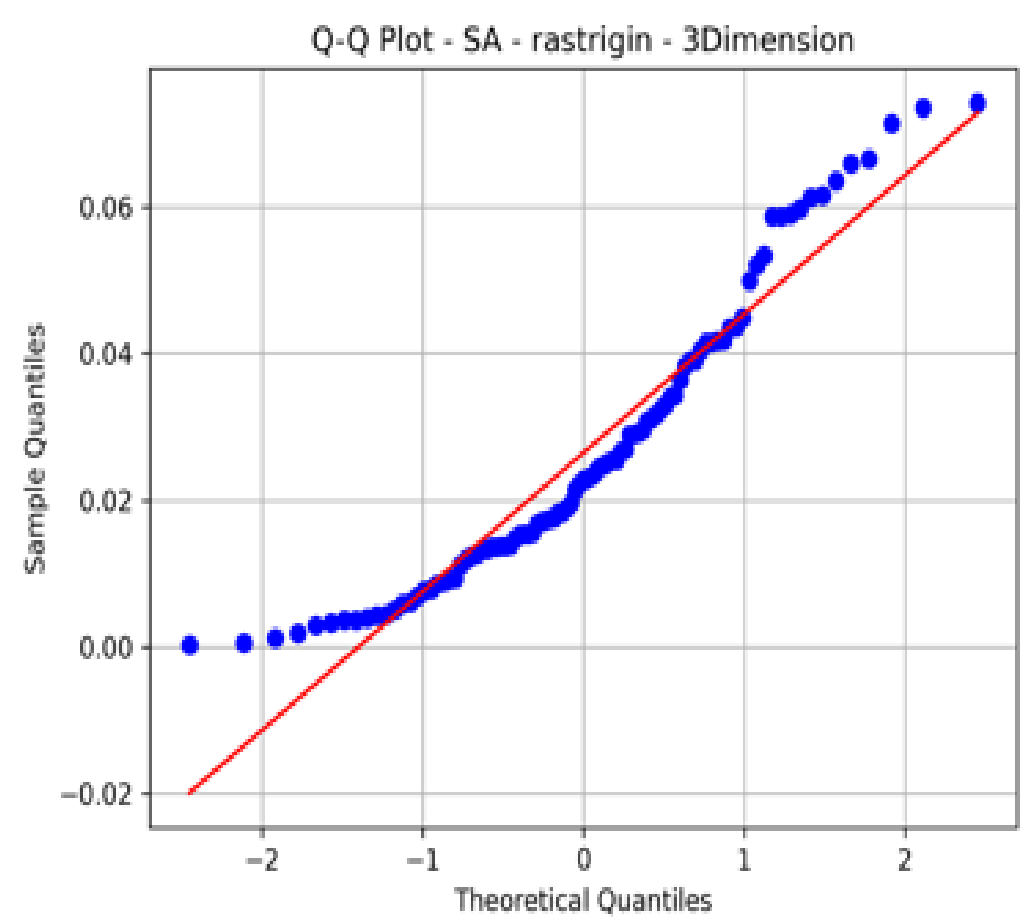
Q-Q Plot - SA - rastrigin - 3Dimension
Sample Quantiles
Theoretical Quantiles

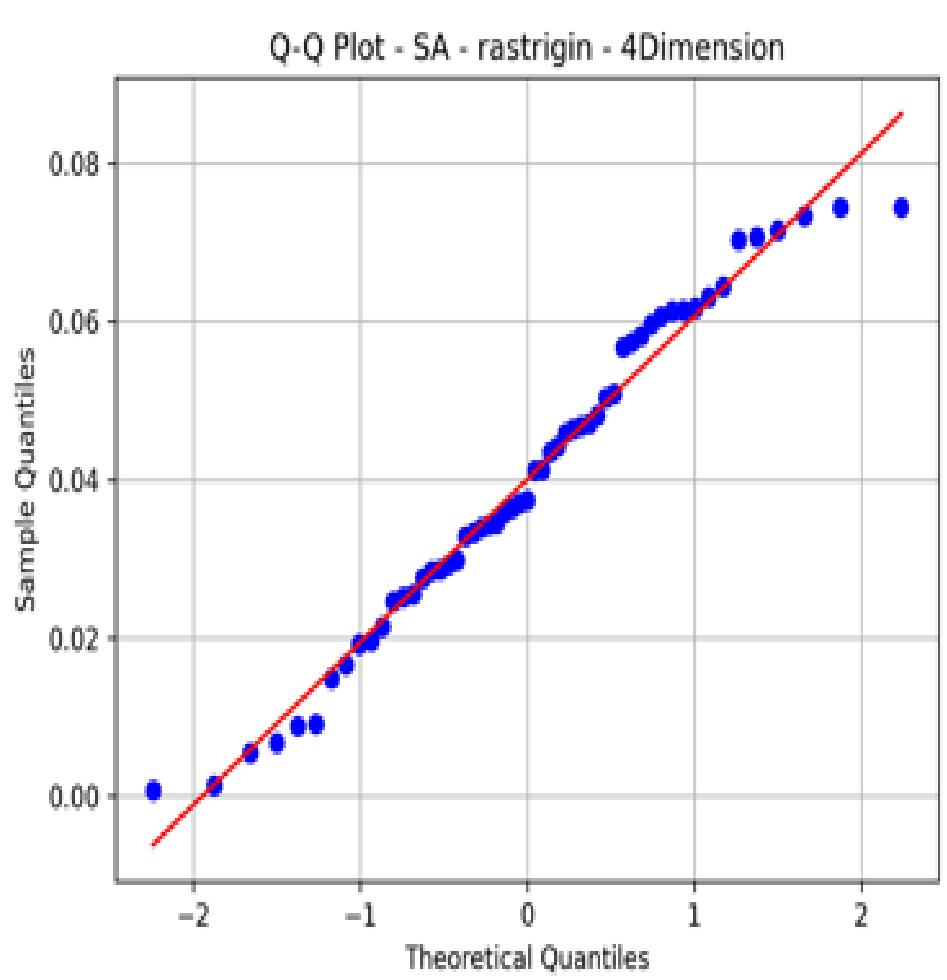
Q-Q Plot - SA - rastrigin - 4Dimension
Sample Quantiles
Theoretical Quantiles

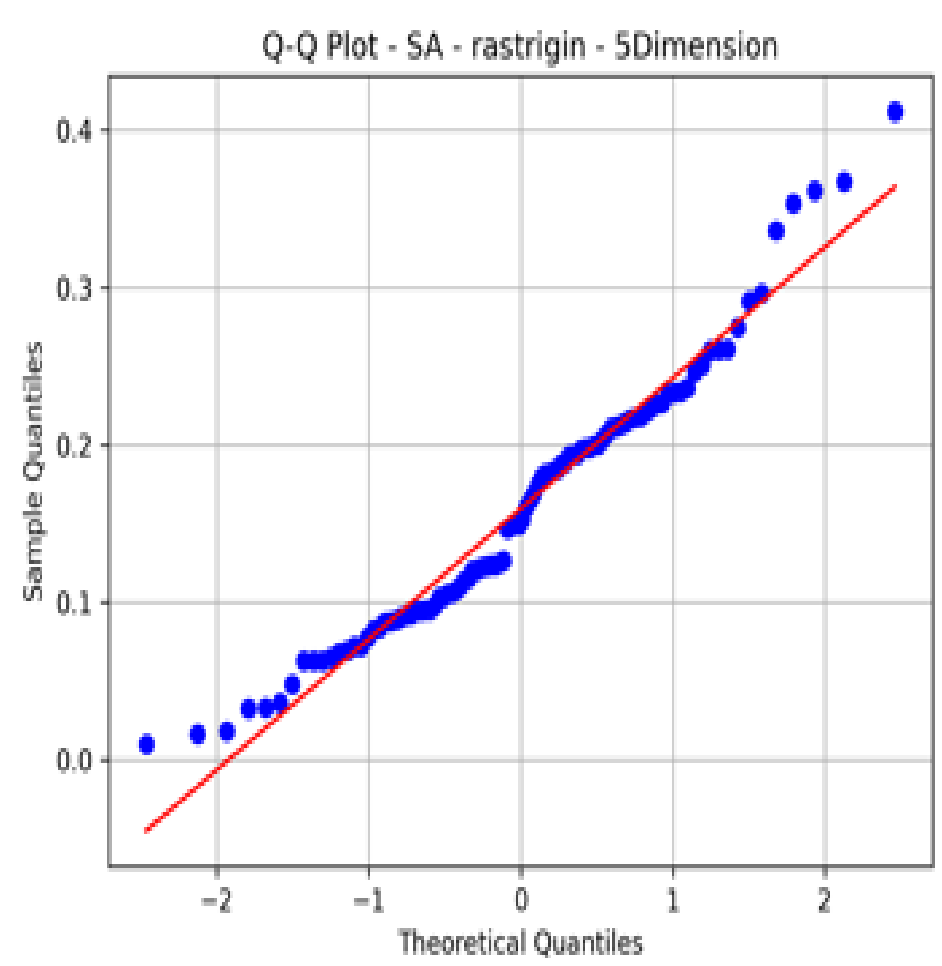
Q-Q Plot - SA - rastrigin - 5Dimension
Sample Quantiles
Theoretical Quantiles

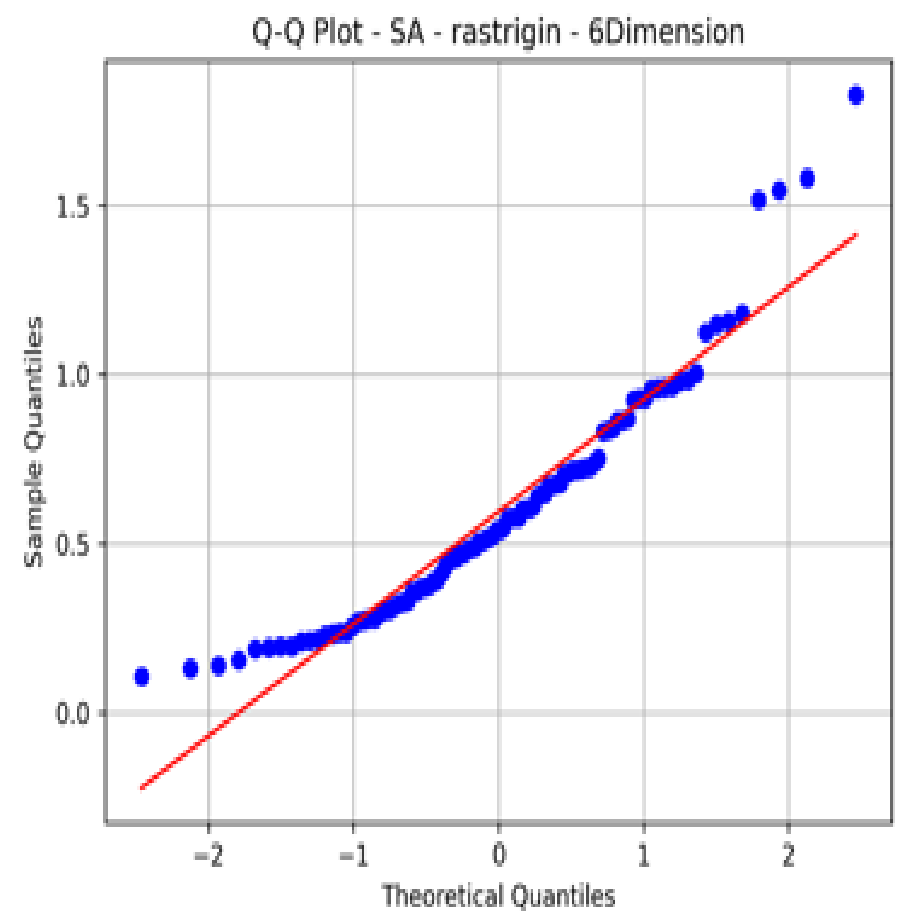
Q-Q Plot - SA - rastrigin - 6Dimension
Sample Quantiles
Theoretical Quantiles

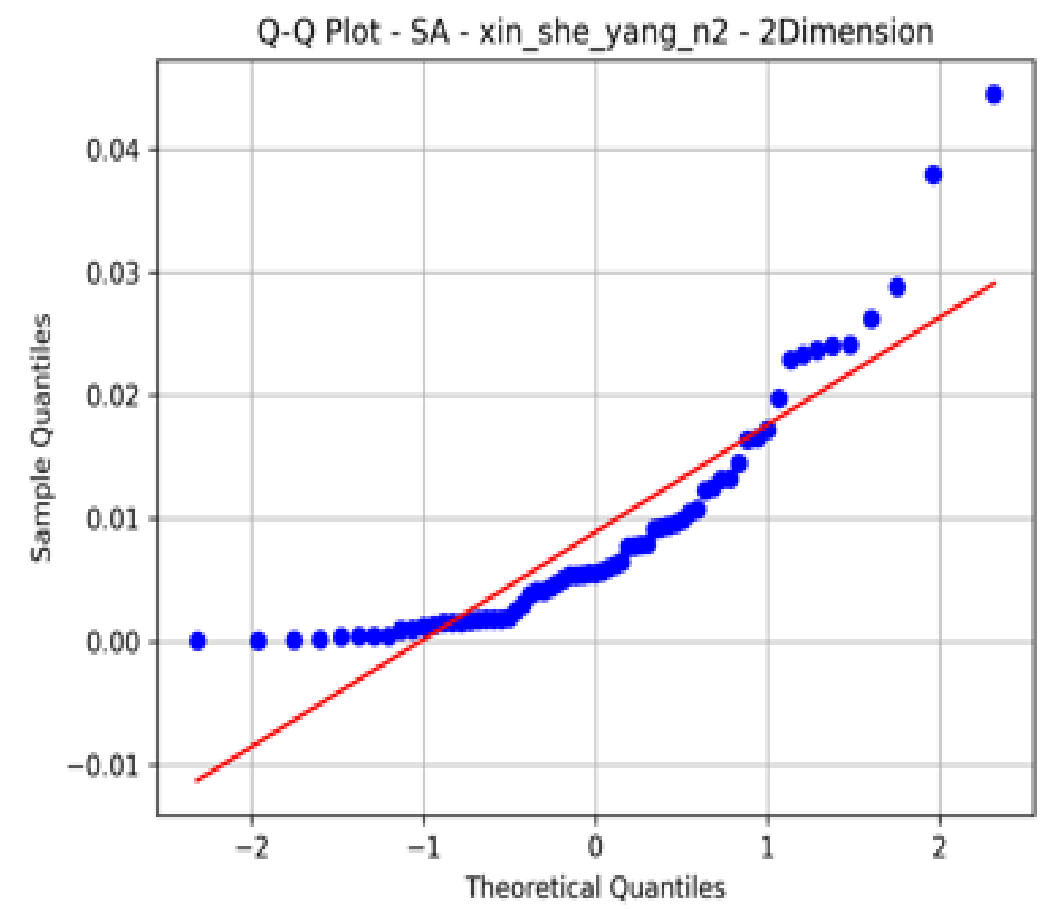
Q-Q Plot - SA - xin_she_yang_n2 - 2Dimension
Sample Quantiles
Theoretical Quantiles

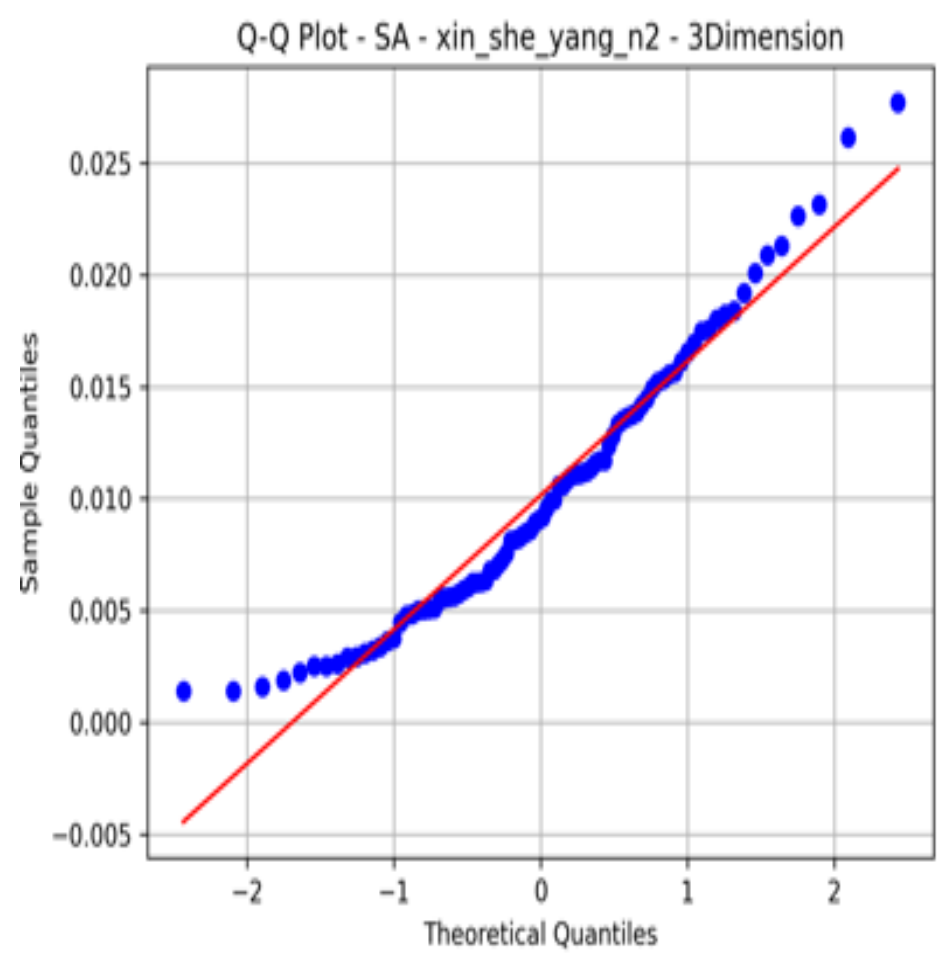
Q-Q Plot - SA - xin_she_yang_n2 - 3Dimension
Sample Quantiles
Theoretical Quantiles

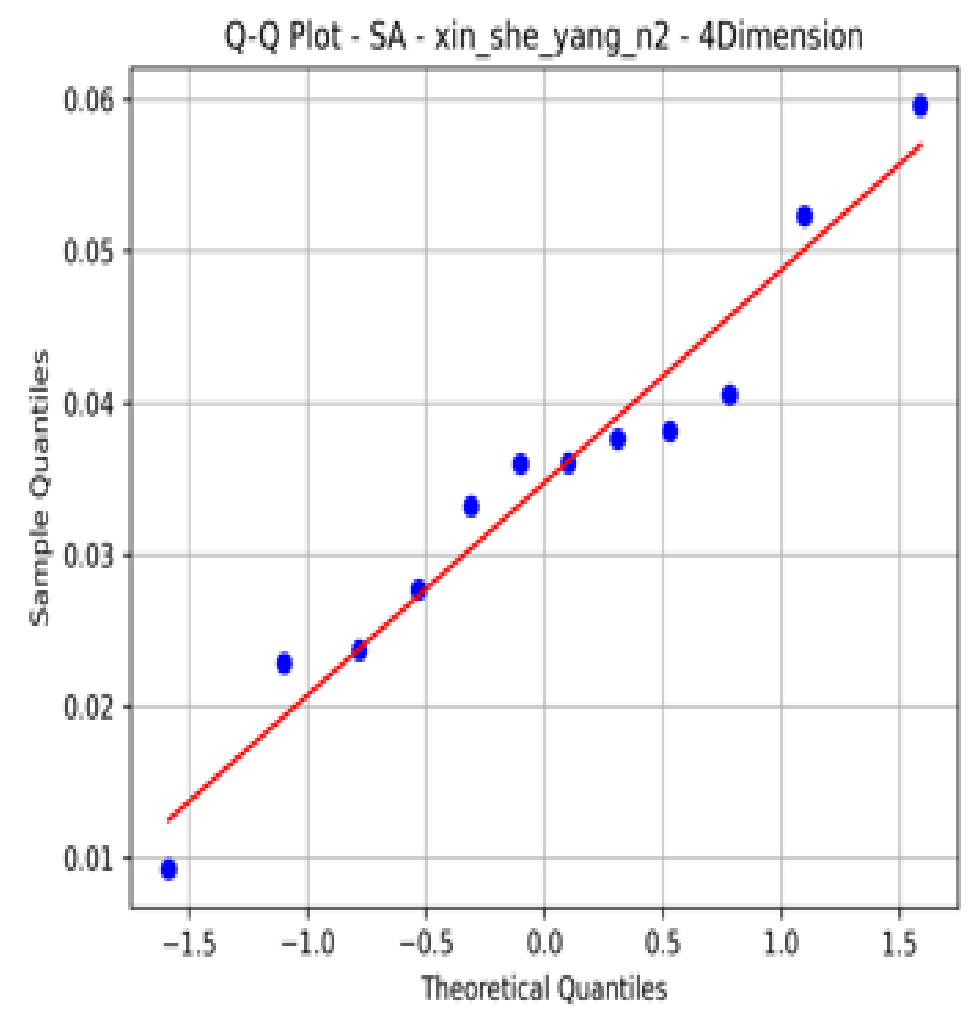
Q-Q Plot - SA - xin_she_yang_n2 - 4Dimension
Sample Quantiles
Theoretical Quantiles

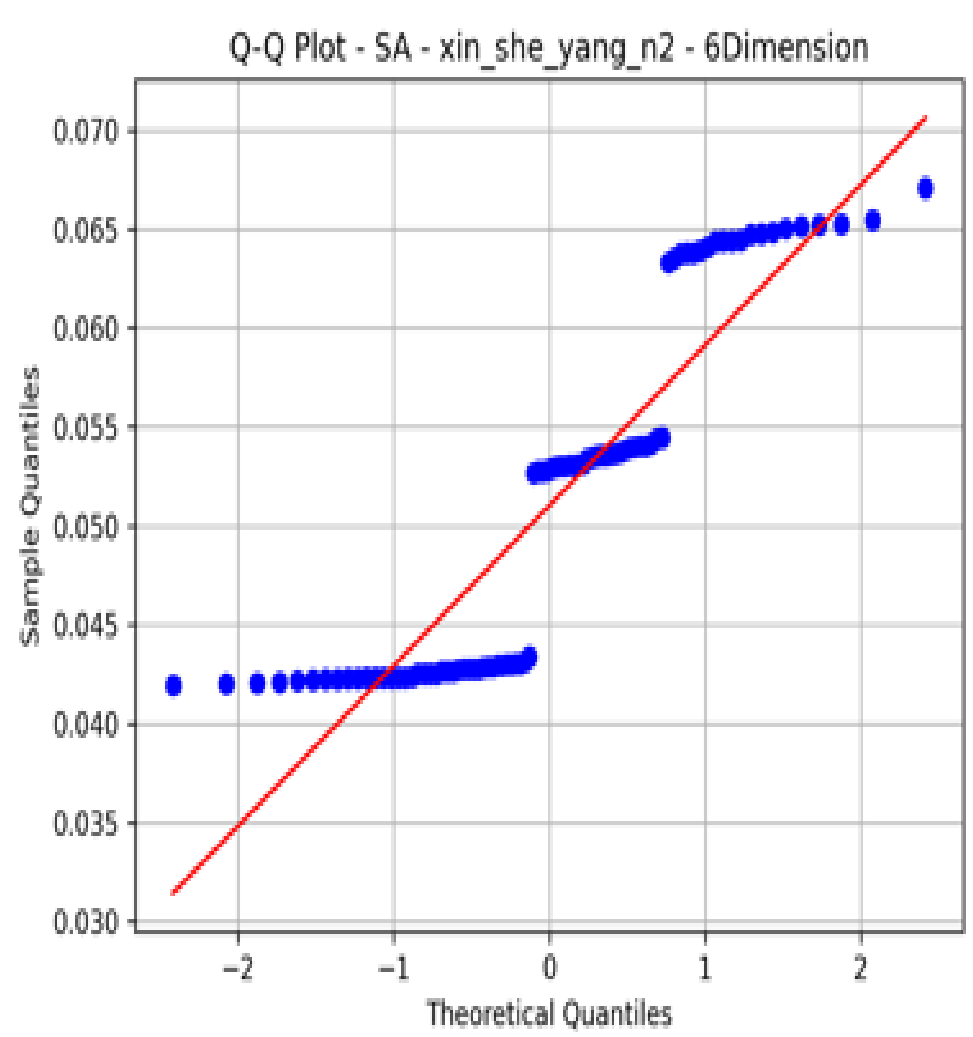
Q-Q Plot - SA - xin_she_yang_n2 - 6Dimension
Sample Quantiles
Theoretical Quantiles

<u>TA Q-Q Plots</u>

Continuous functions QQ Plots

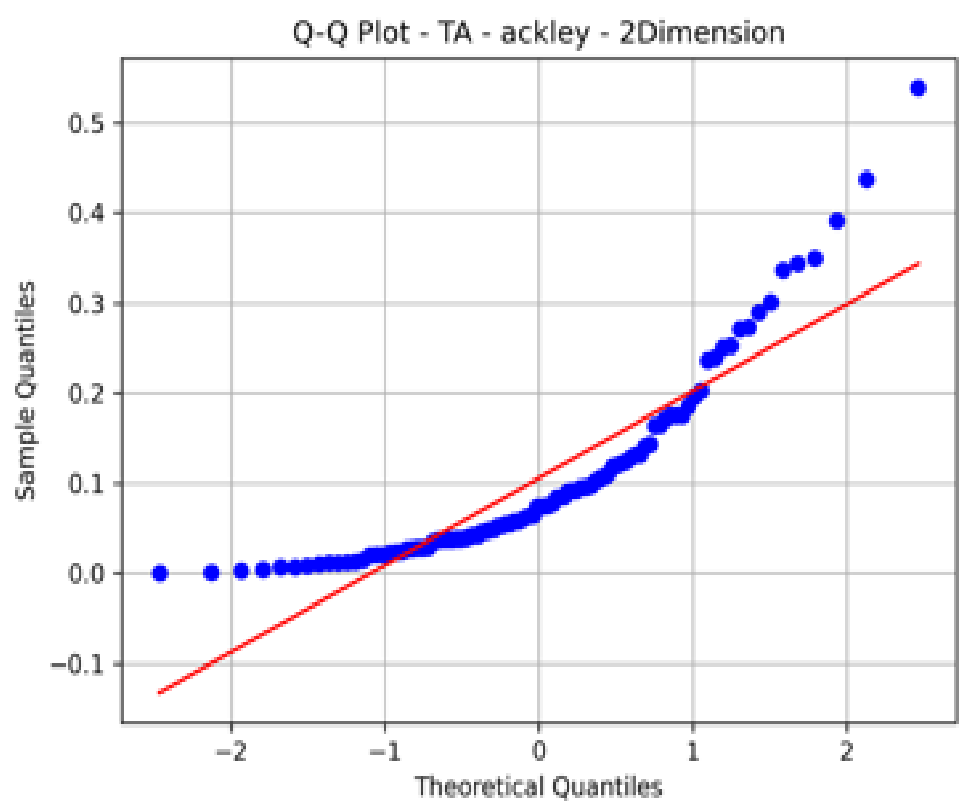

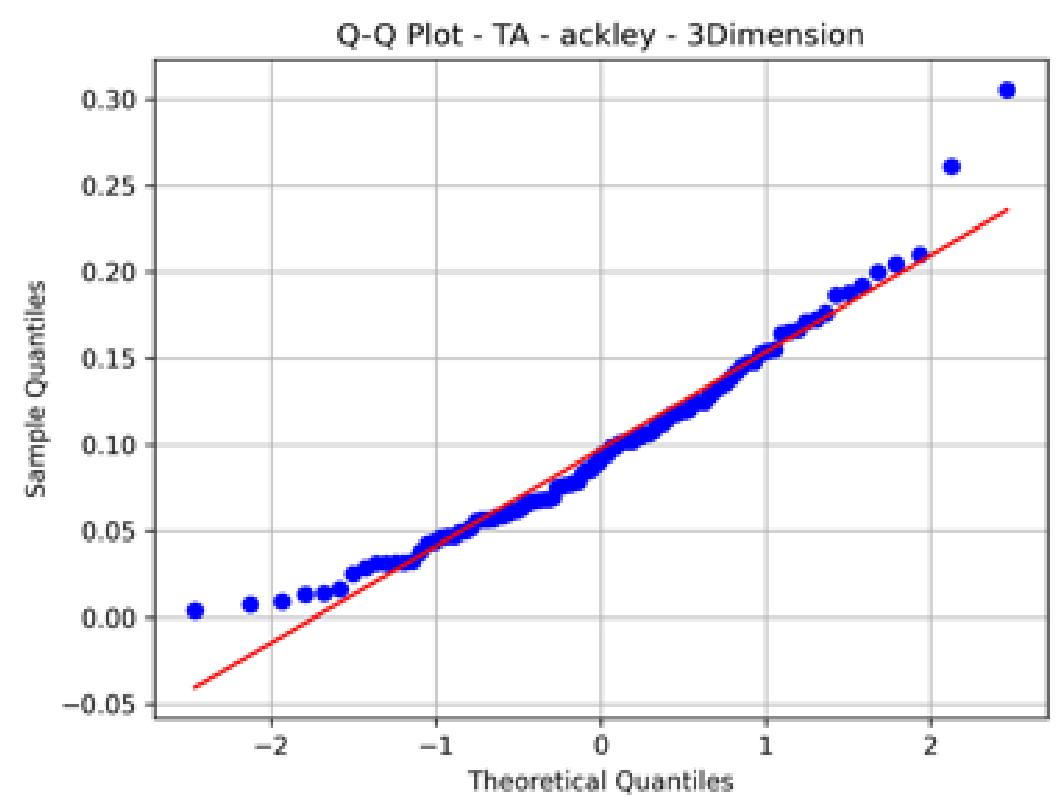

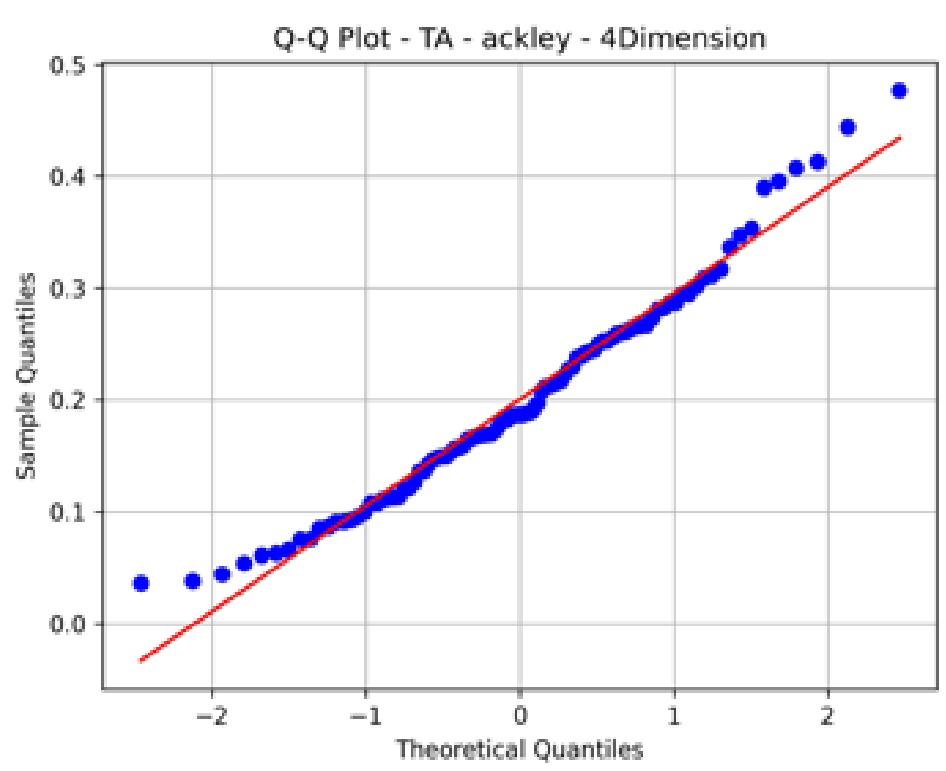

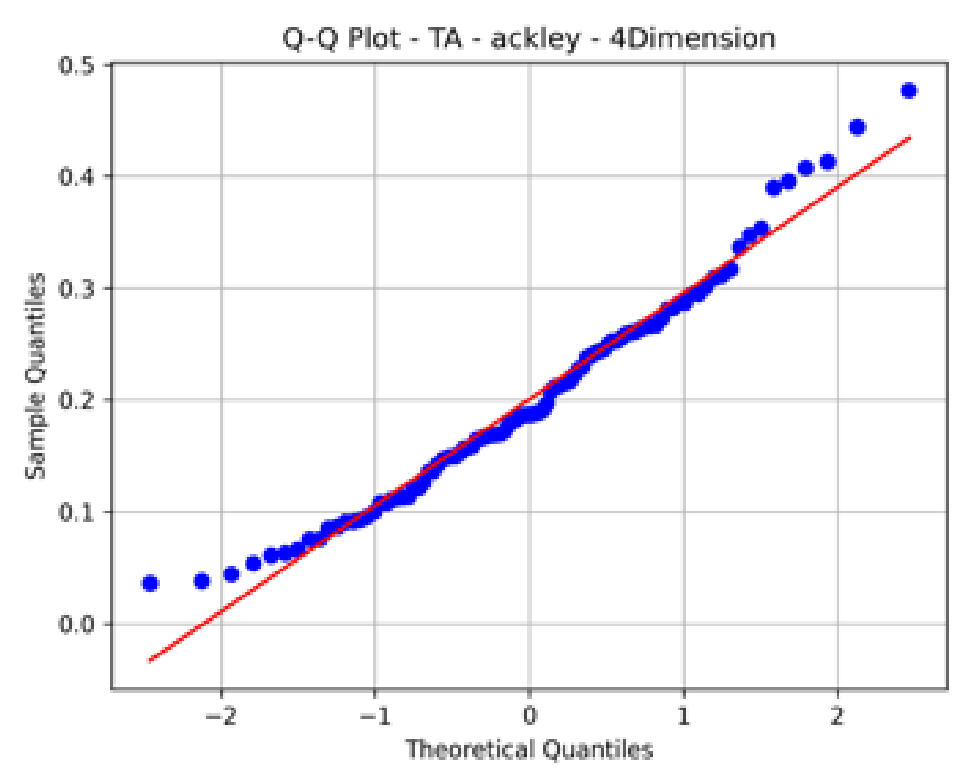

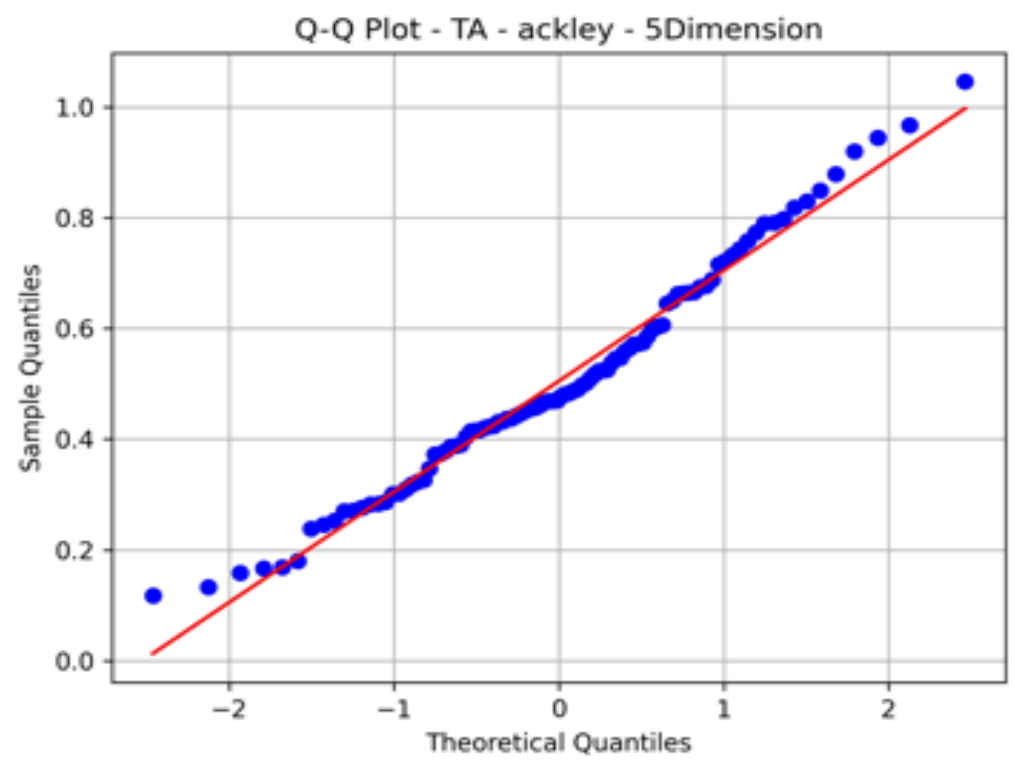
Q-Q Plot - TA - ackley - 5Dimension
Sample Quantiles
Theoretical Quantiles

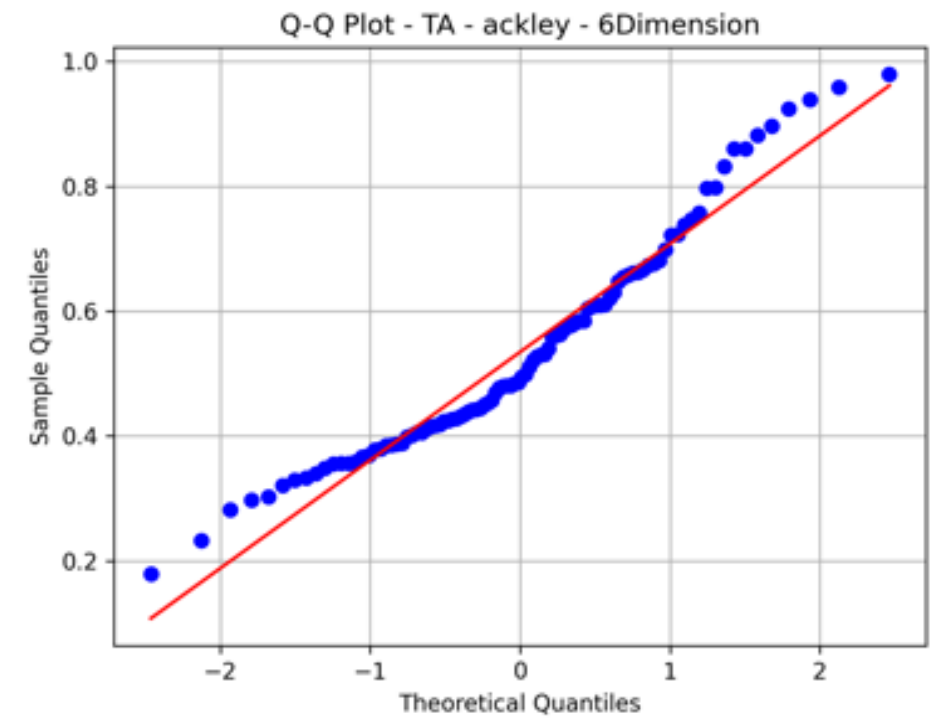
Q-Q Plot - TA - ackley - 6Dimension
Sample Quantiles
Theoretical Quantiles

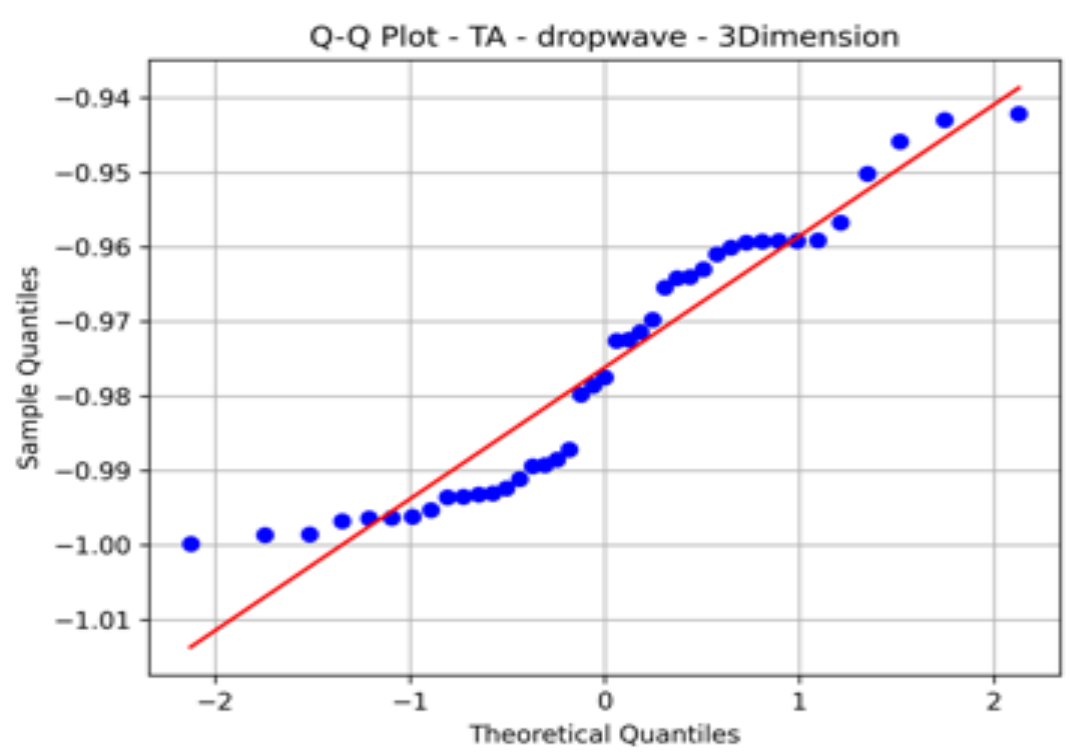
Q-Q Plot - TA - dropwave - 3Dimension
Sample Quantiles
Theoretical Quantiles

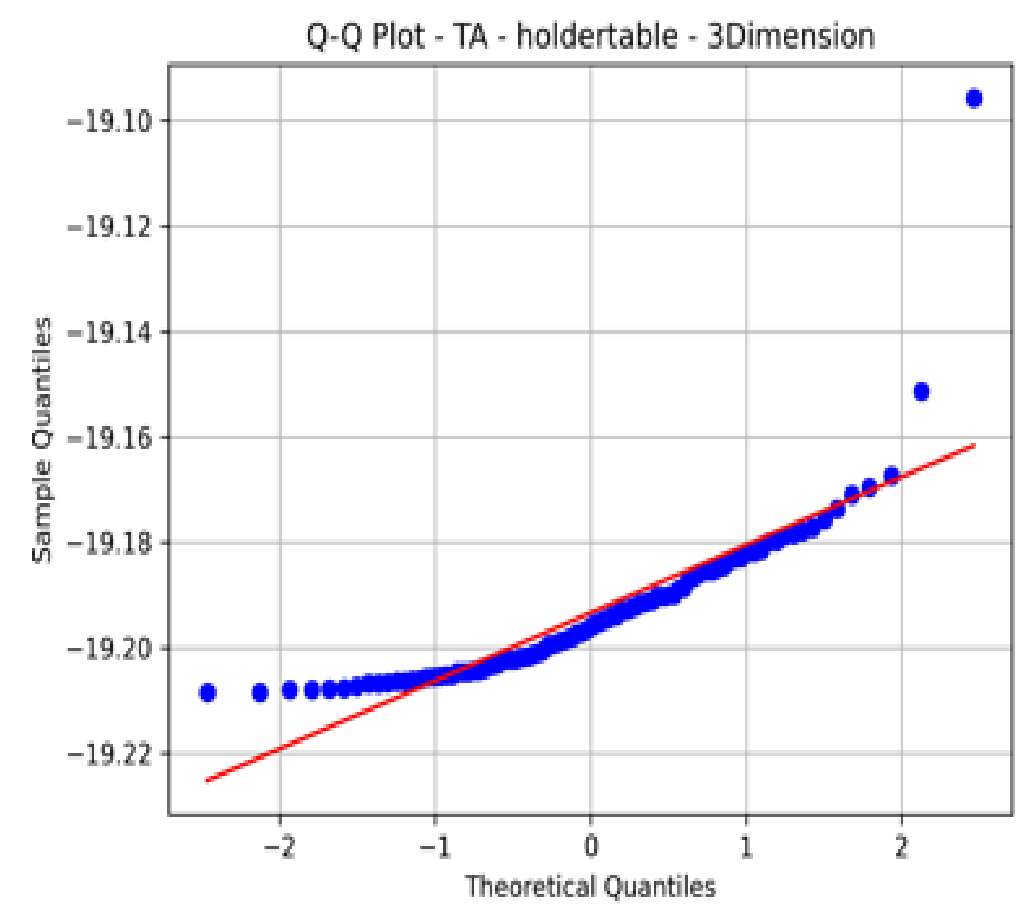
Q-Q Plot - TA - holdertable - 3Dimension
Sample Quantiles
Theoretical Quantiles

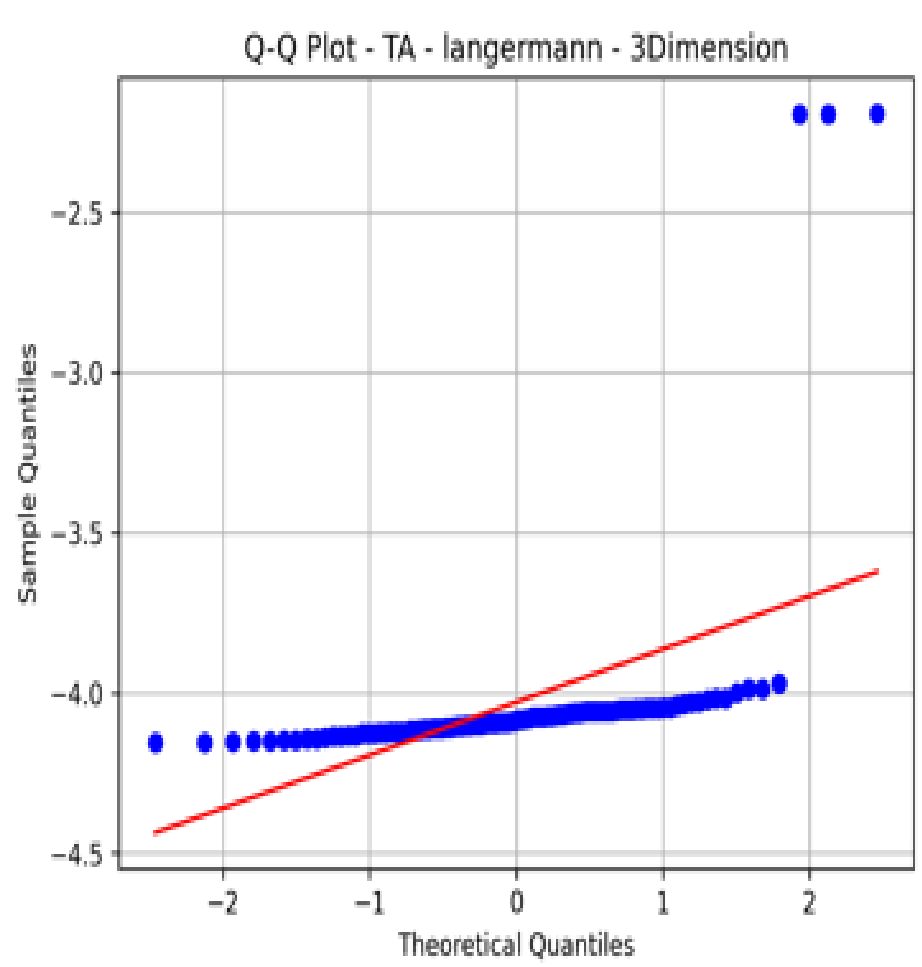
Q-Q Plot - TA - langermann - 3Dimension
Sample Quantiles
Theoretical Quantiles

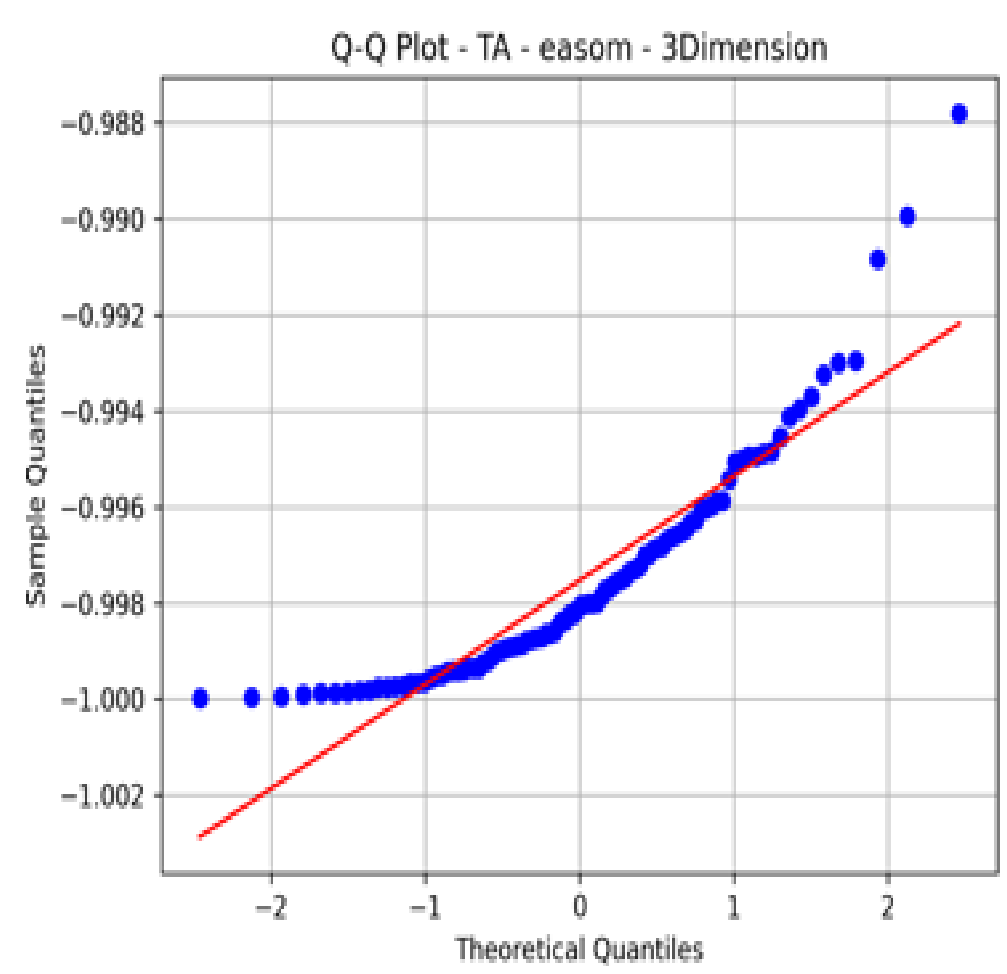
Q-Q Plot - TA - easom - 3Dimension
Sample Quantiles
Theoretical Quantiles

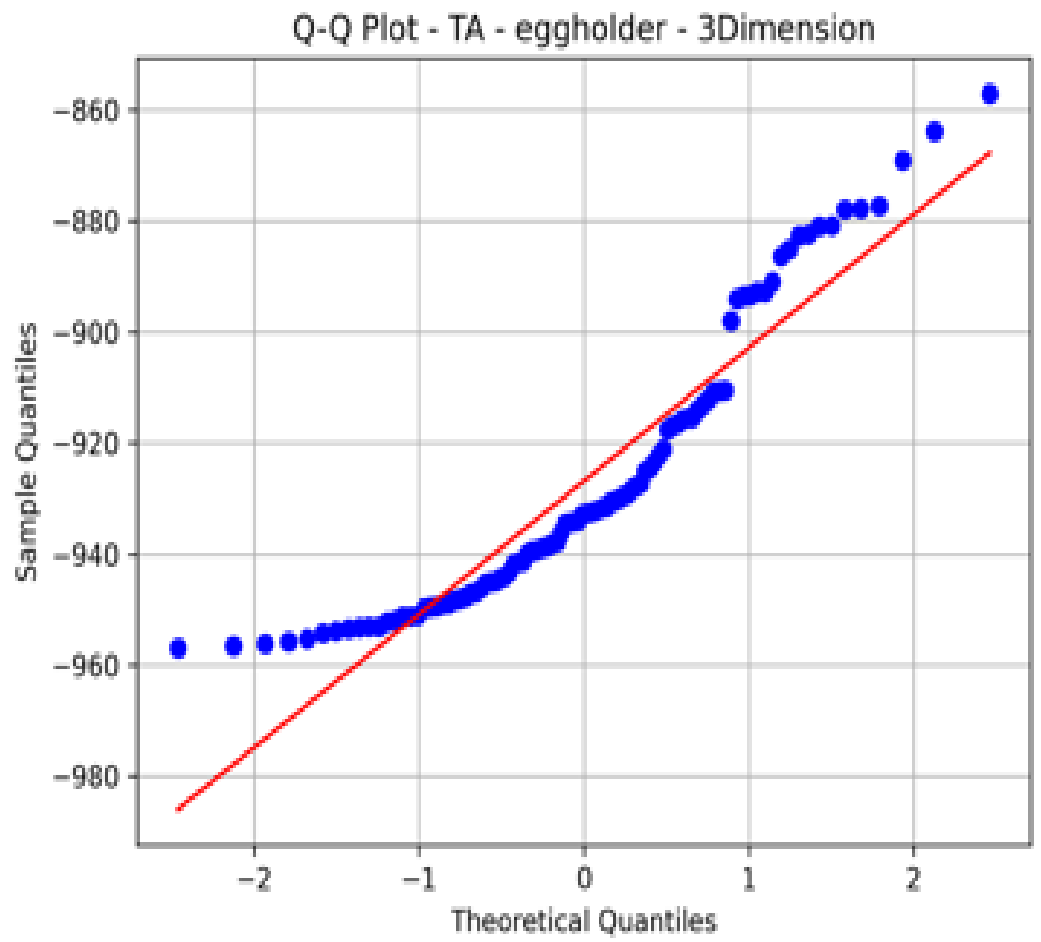
Q-Q Plot - TA - eggholder - 3Dimension
Sample Quantiles
Theoretical Quantiles

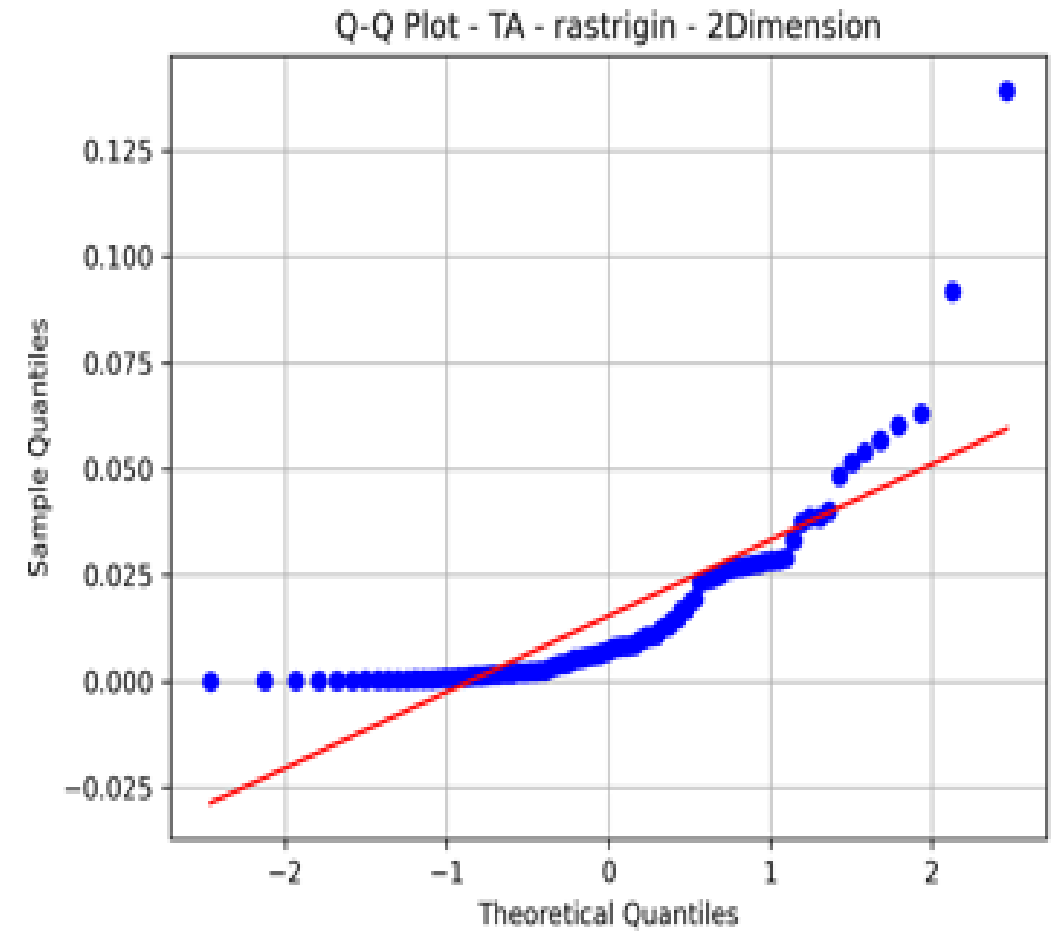
Q-Q Plot - TA - rastrigin - 2Dimension
Sample Quantiles
Theoretical Quantiles

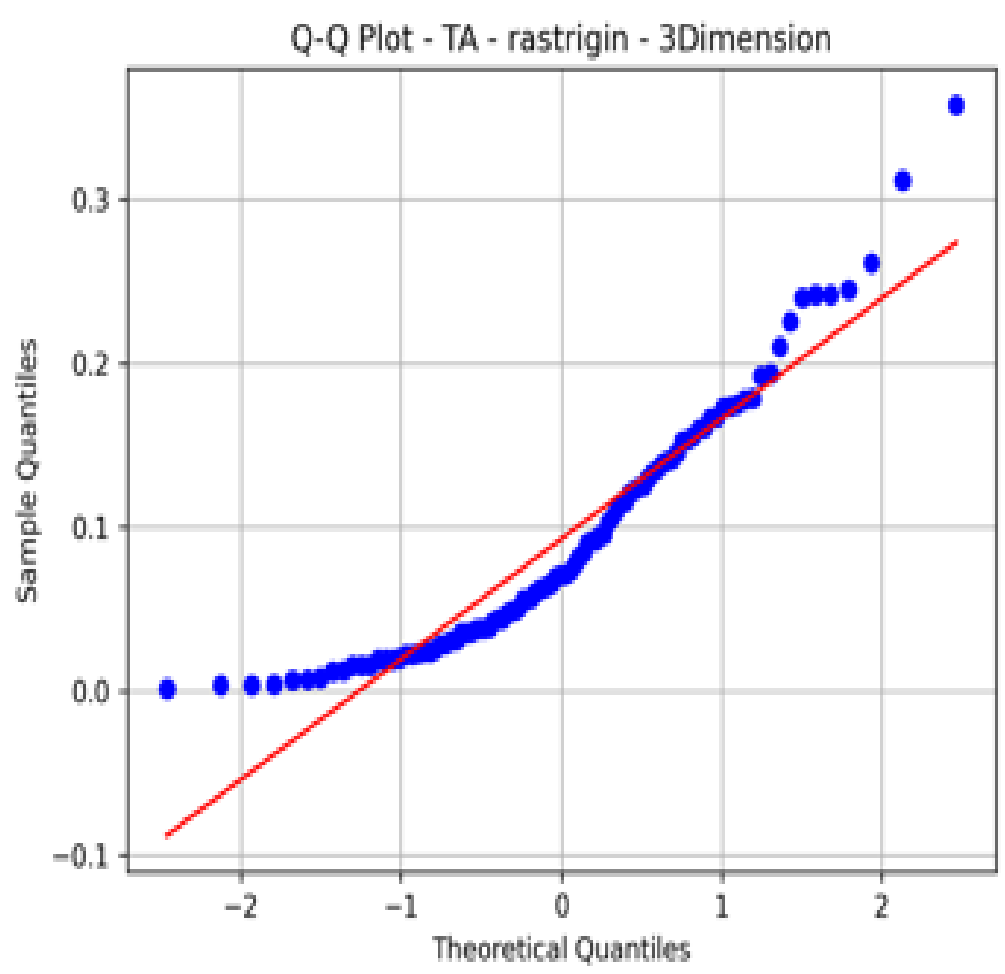
Q-Q Plot - TA - rastrigin - 3Dimension
Sample Quantiles
Theoretical Quantiles

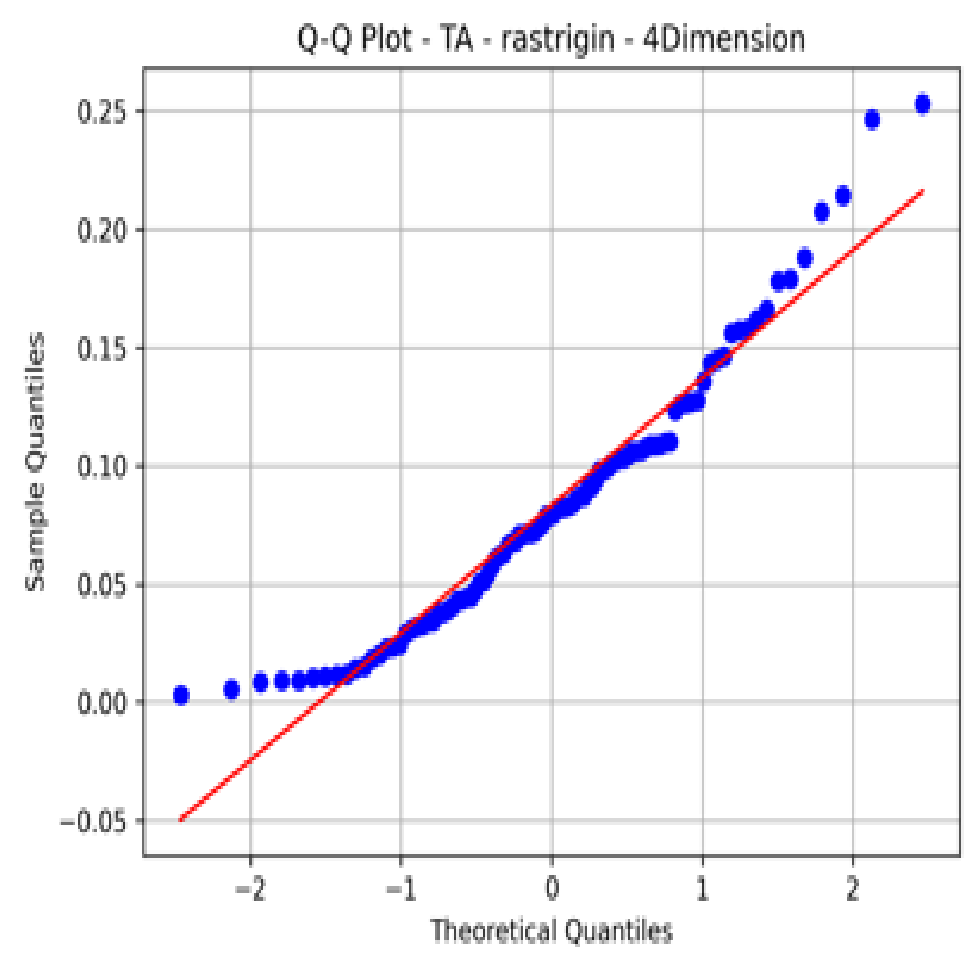
Q-Q Plot - TA - rastrigin - 4Dimension
Sample Quantiles
Theoretical Quantiles

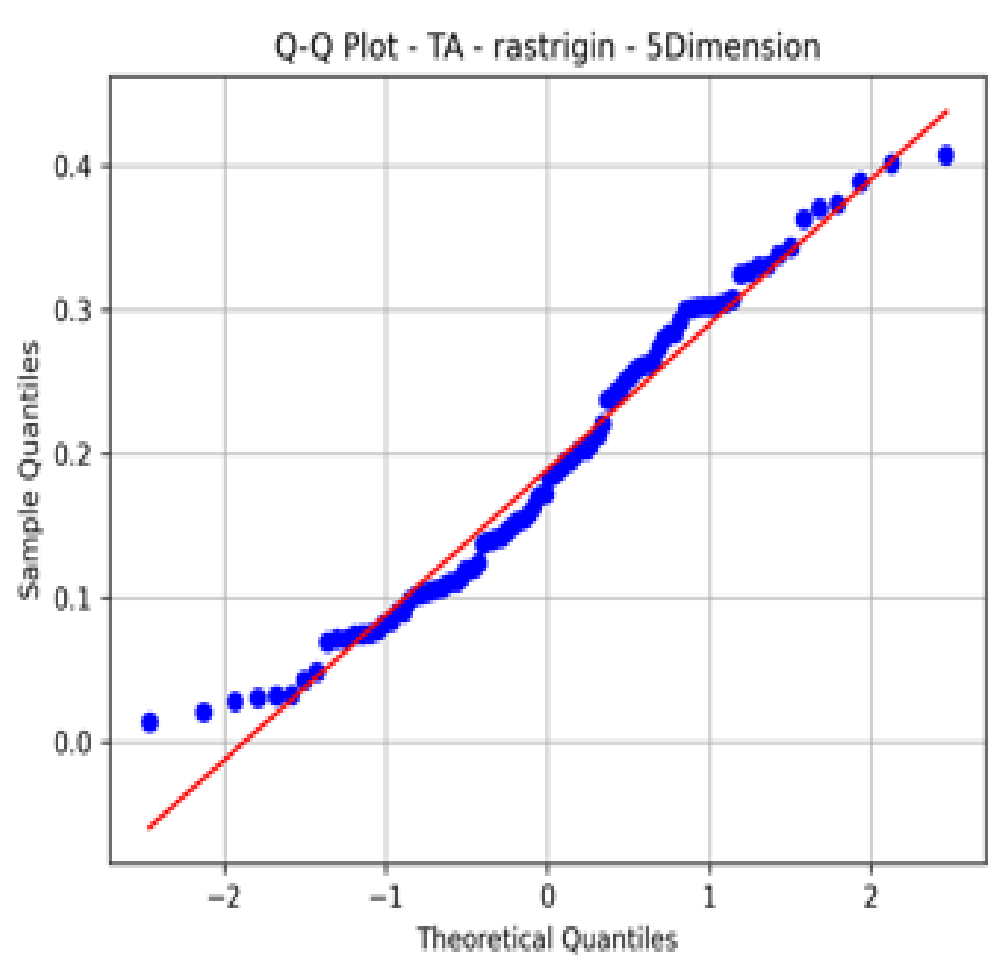
Q-Q Plot - TA - rastrigin - 5Dimension
Sample Quantiles
Theoretical Quantiles

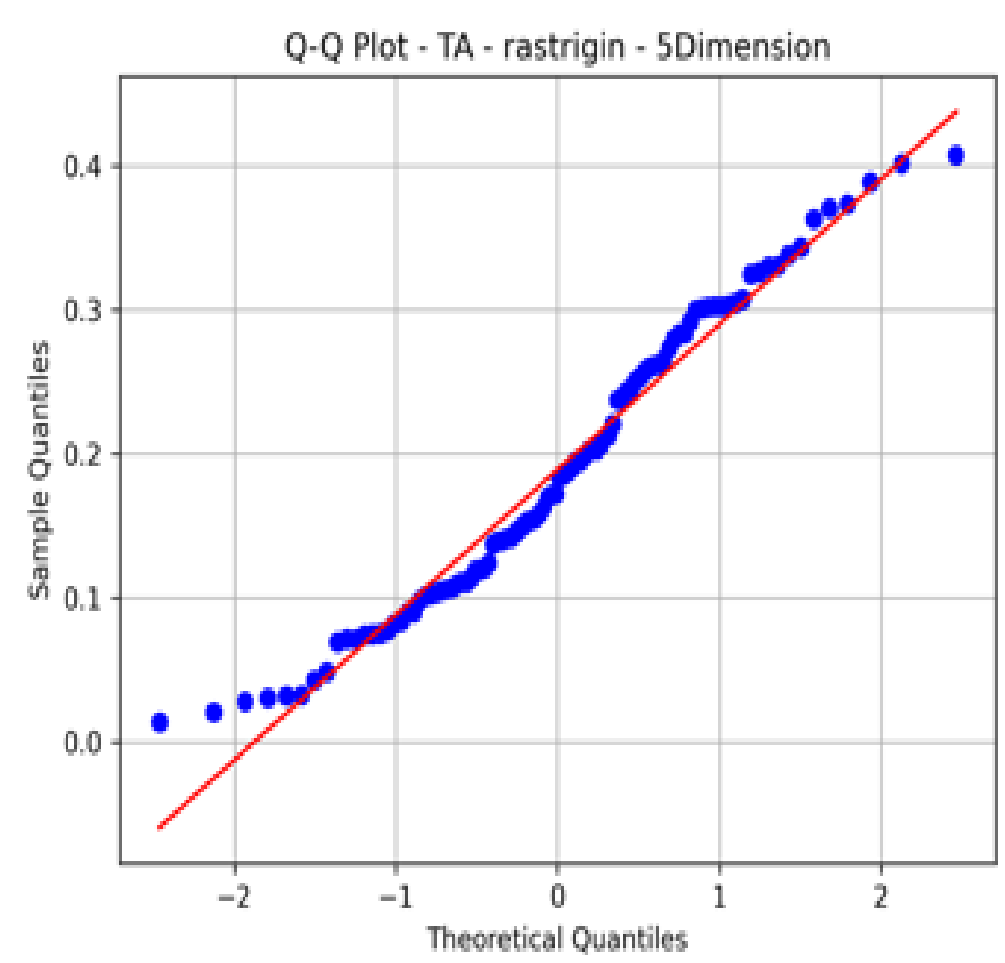
Q-Q Plot - TA - rastrigin - 5Dimension
Sample Quantiles
Theoretical Quantiles

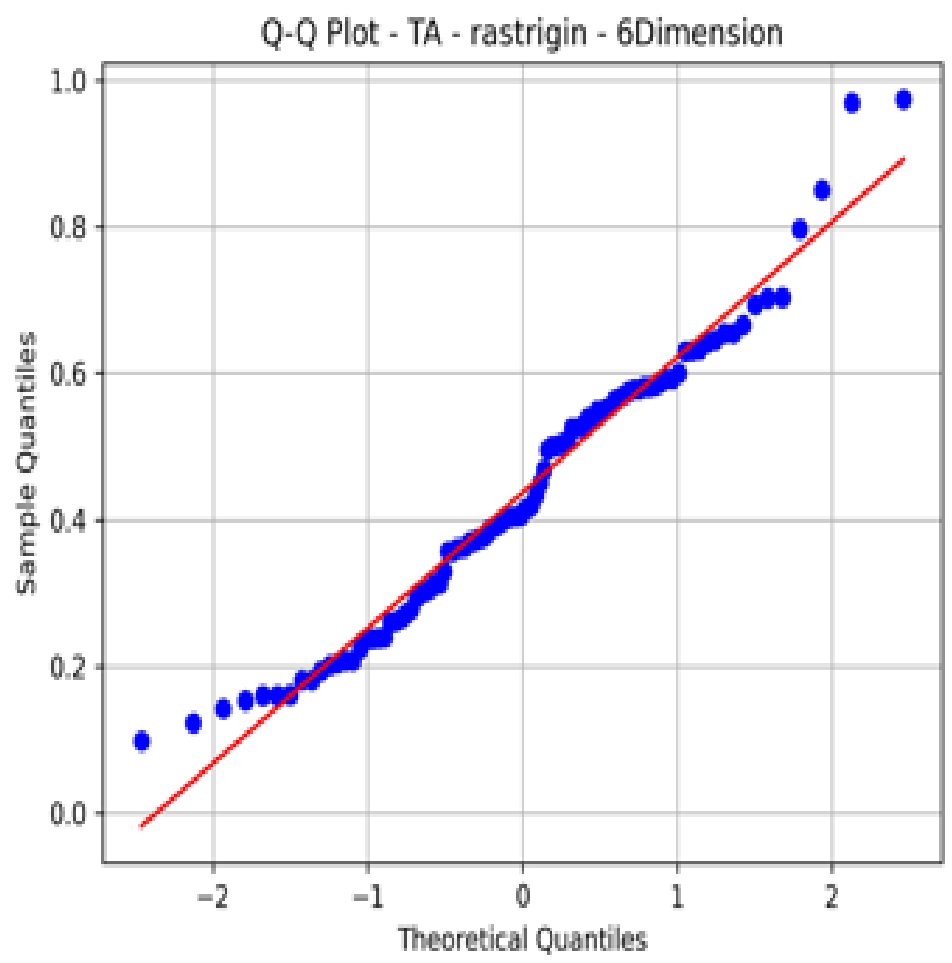
Q-Q Plot - TA - rastrigin - 6Dimension
Sample Quantiles
Theoretical Quantiles

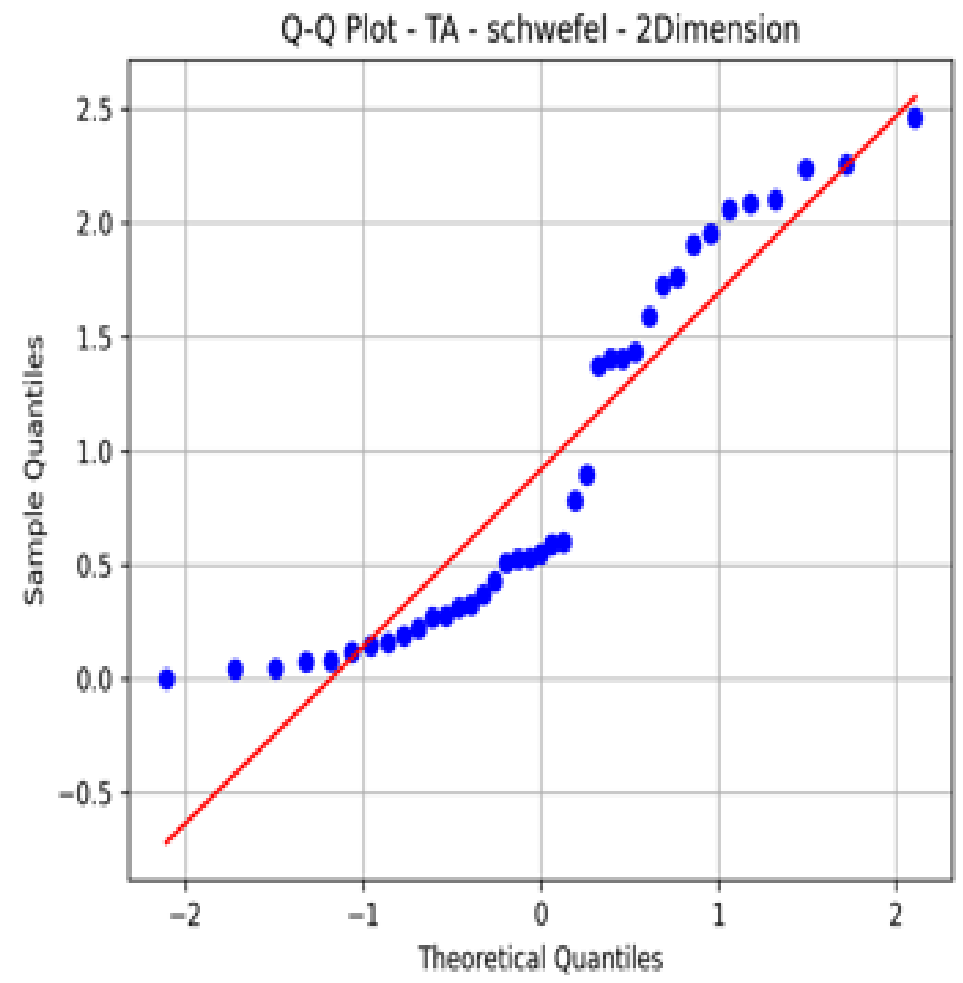
Q-Q Plot - TA - schwefel - 2Dimension
Sample Quantiles
Theoretical Quantiles

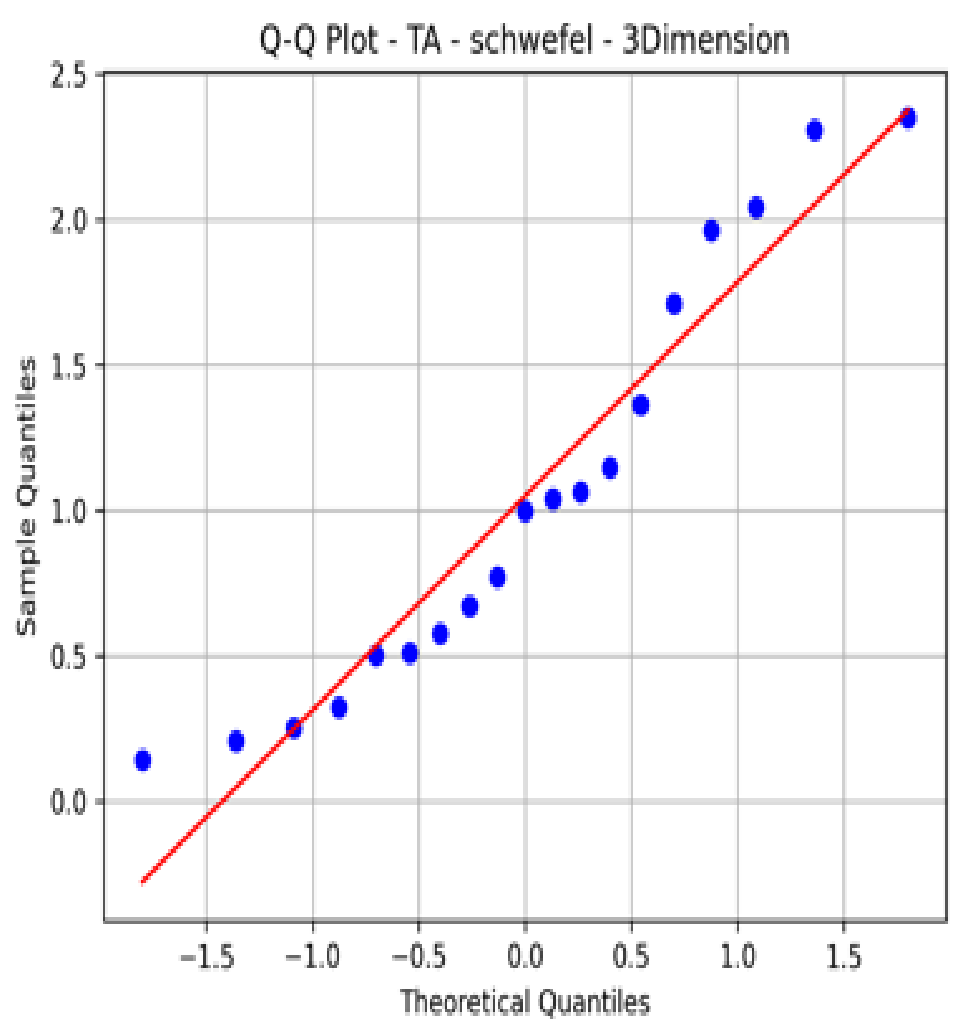
Q-Q Plot - TA - schwefel - 3Dimension
Sample Quantiles
Theoretical Quantiles

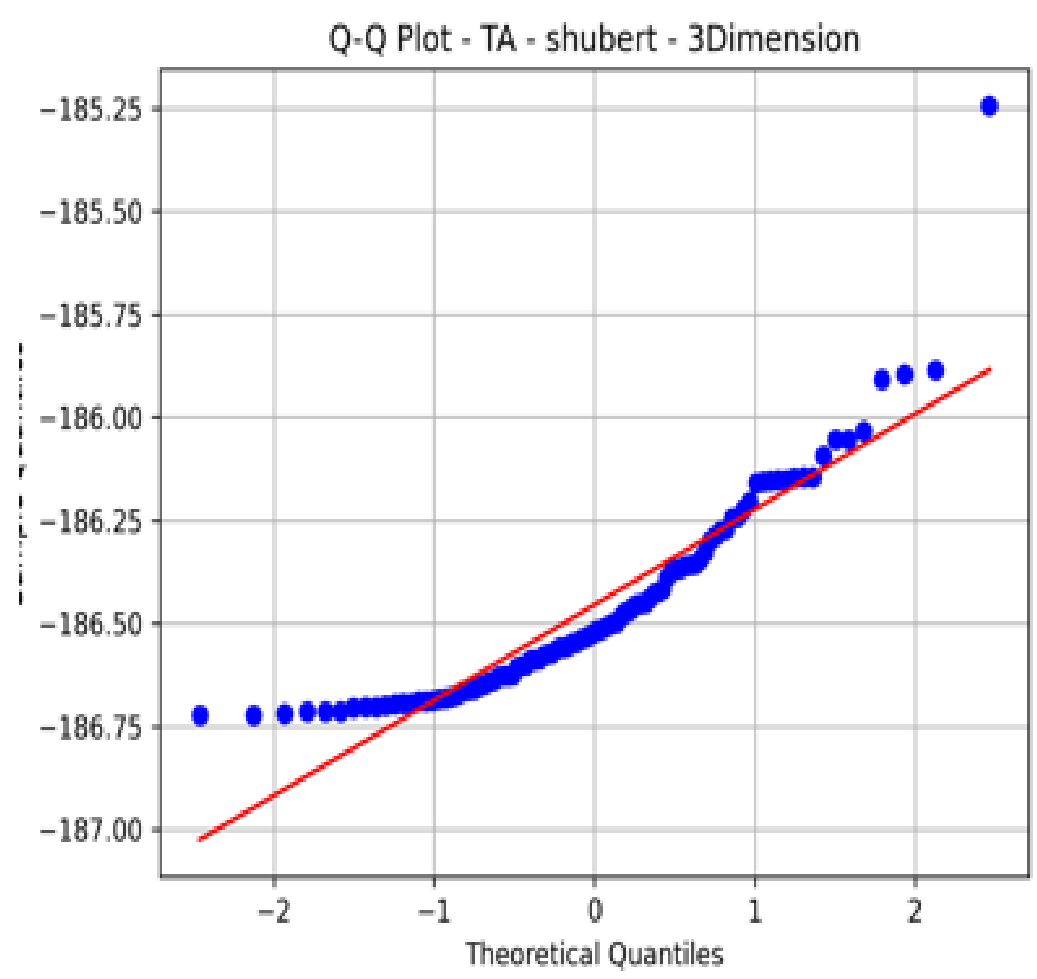
Q-Q Plot - TA - shubert - 3Dimension
Theoretical Quantiles

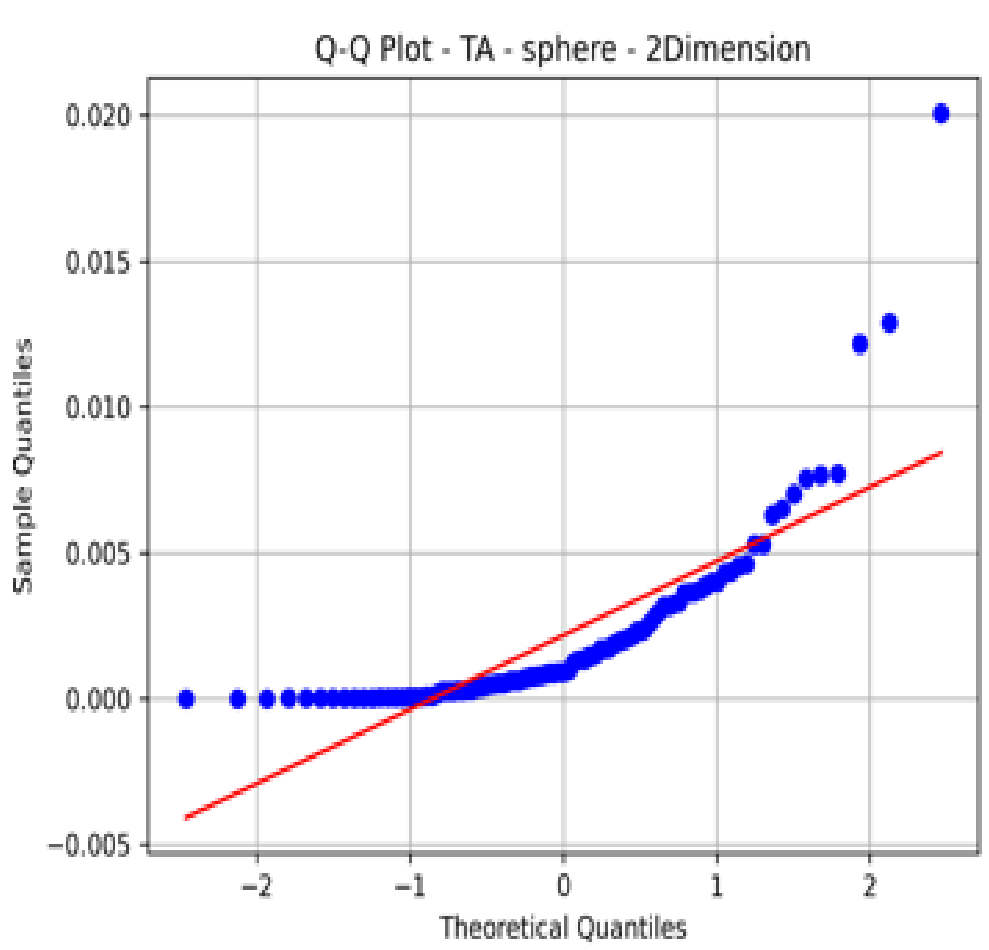
Q-Q Plot - TA - sphere - 2Dimension
Sample Quantiles
Theoretical Quantiles

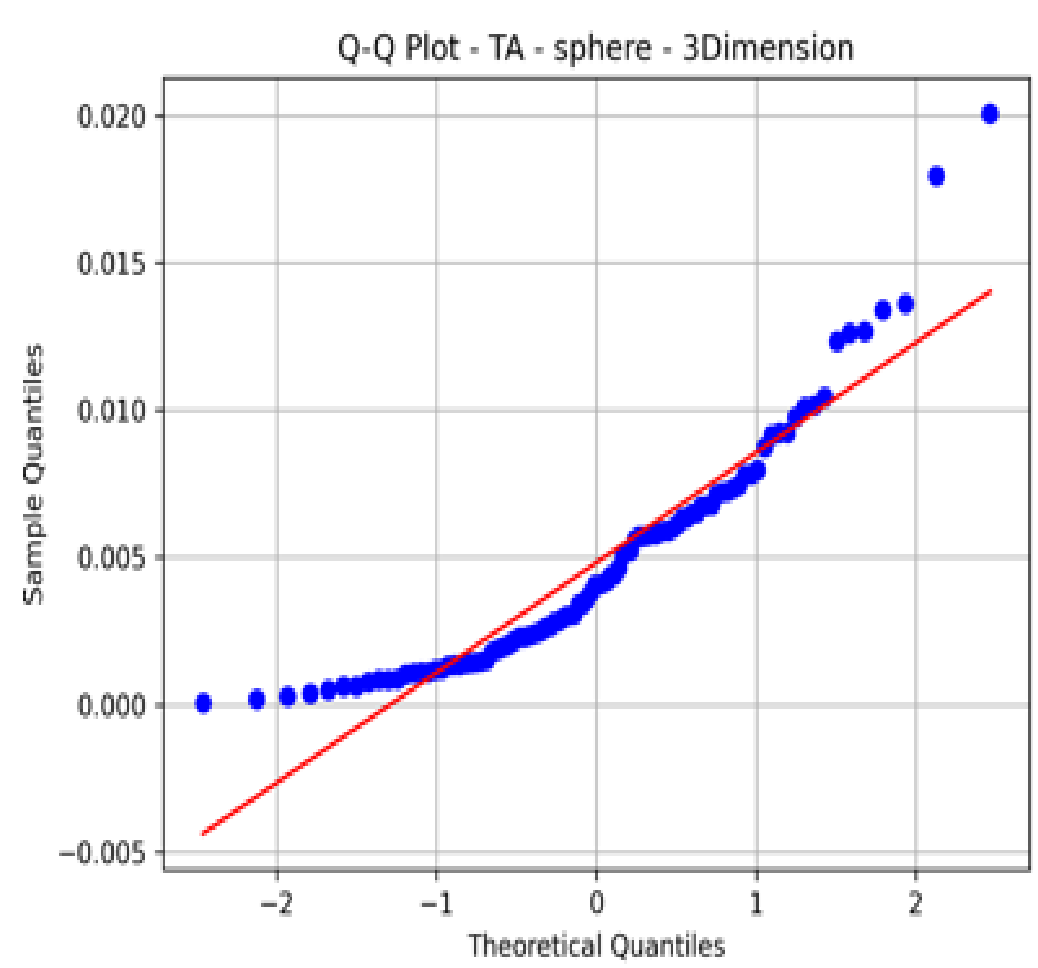
Q-Q Plot - TA - sphere - 3Dimension
Sample Quantiles
Theoretical Quantiles

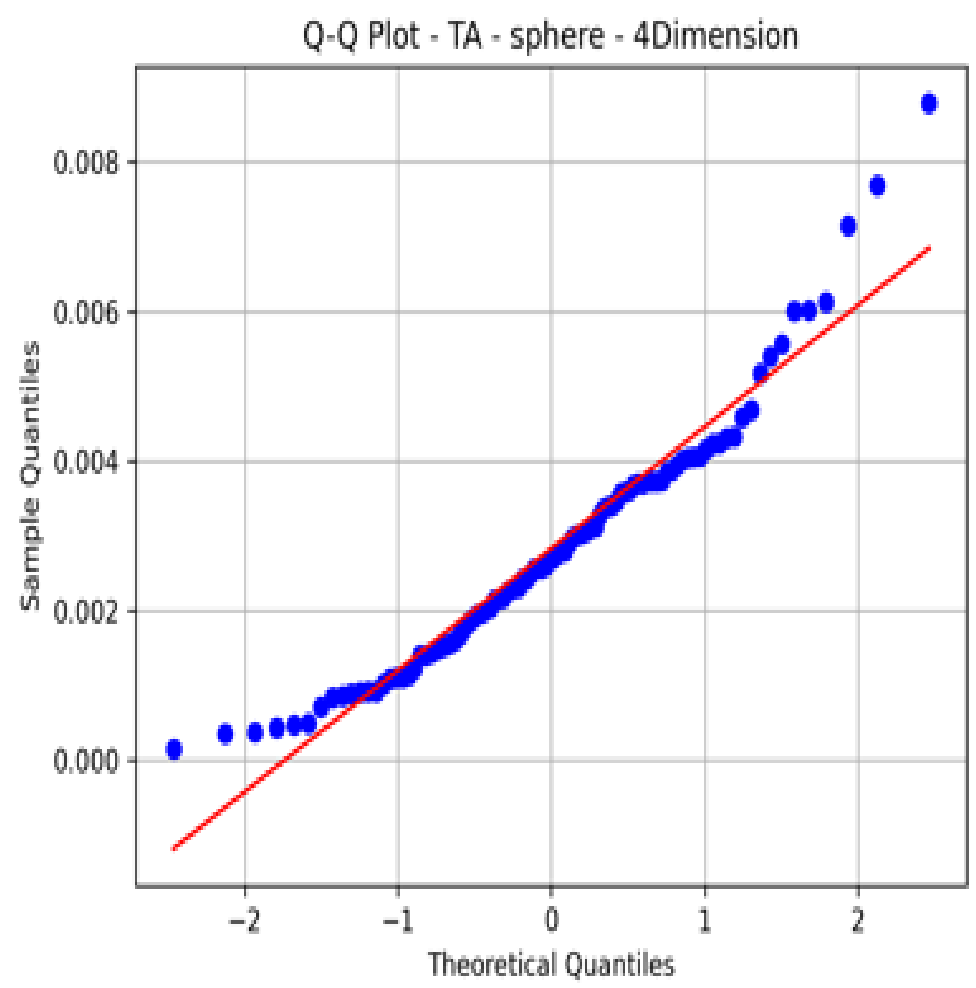

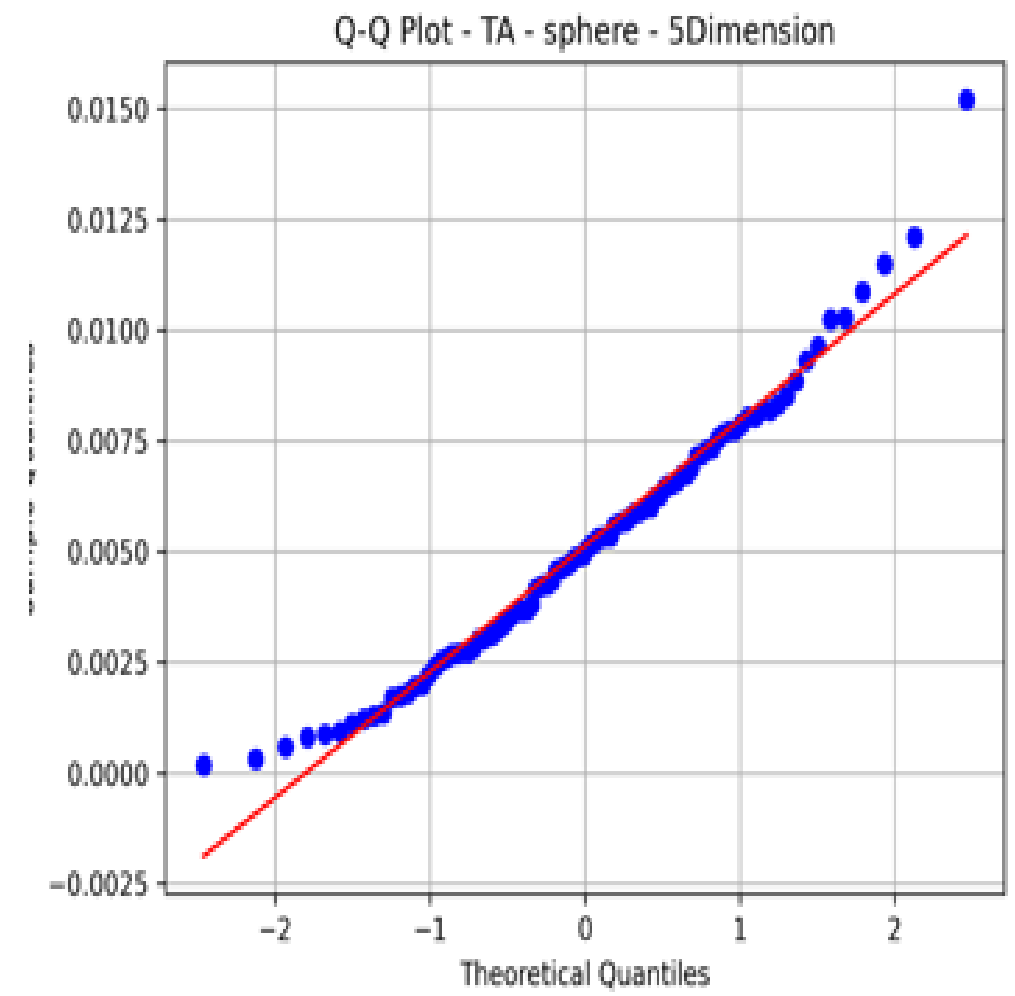

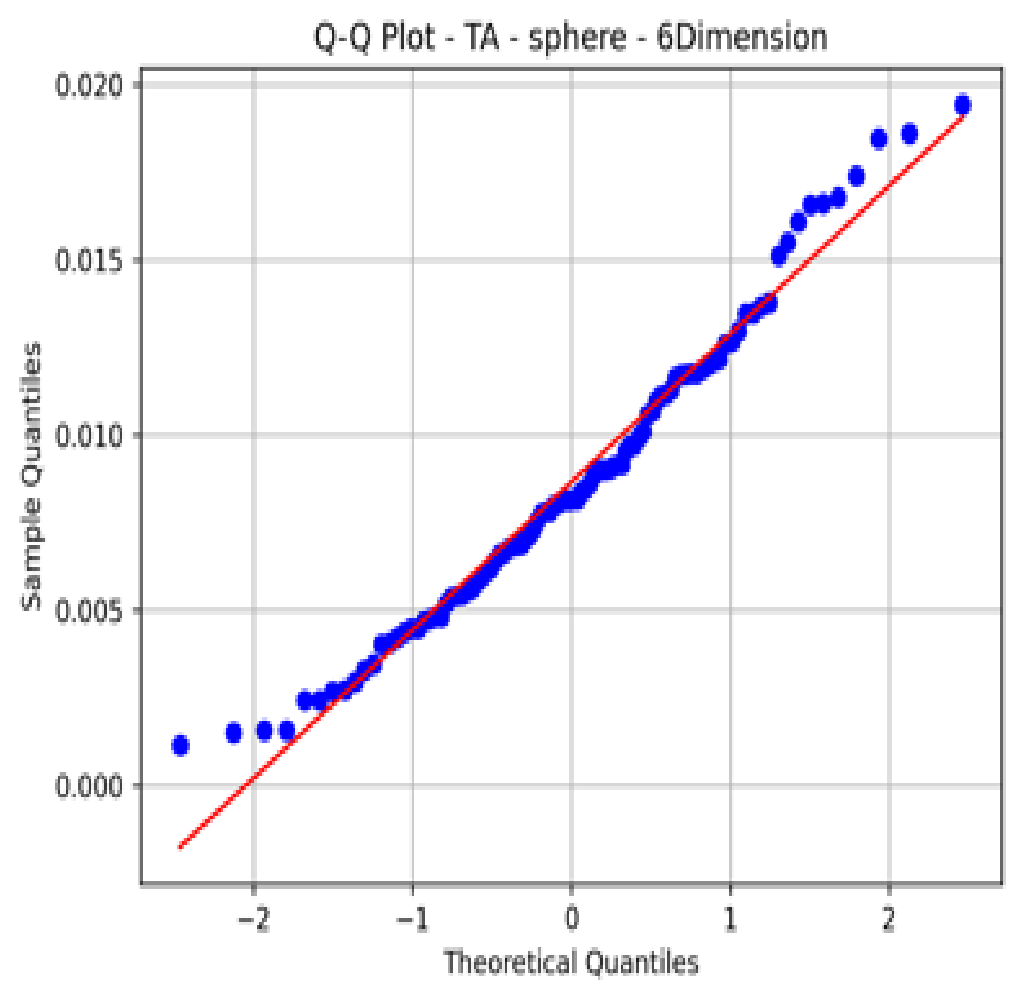

Non-Continuous Functions QQ Plots

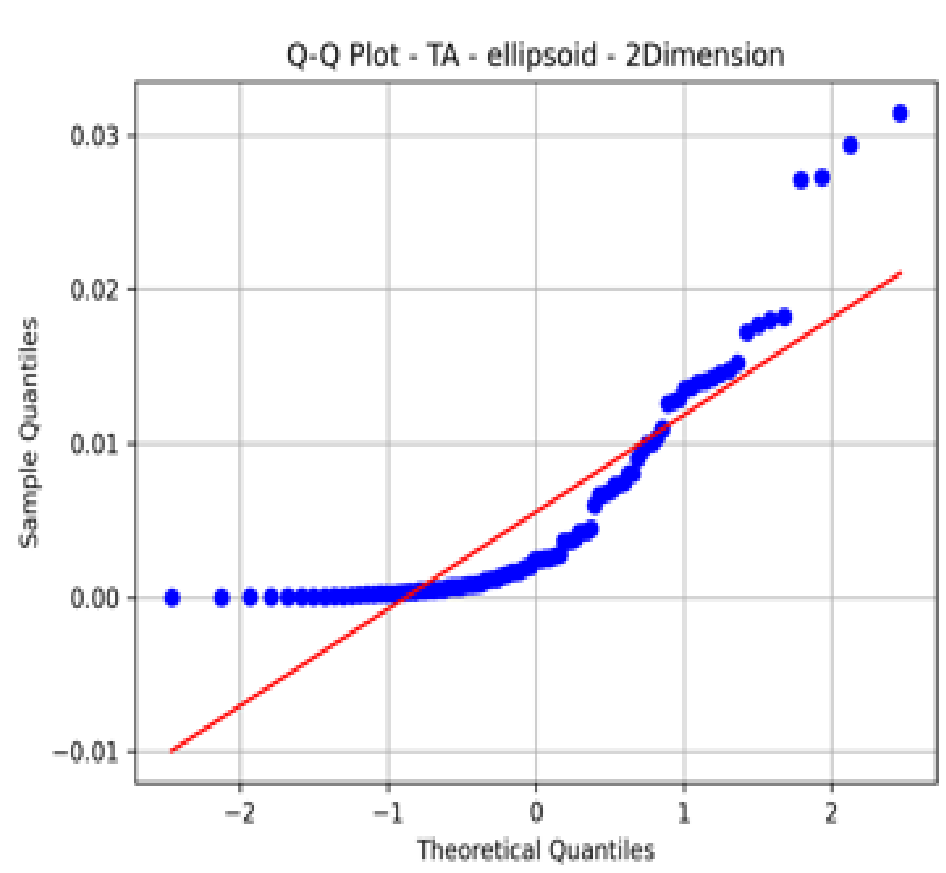

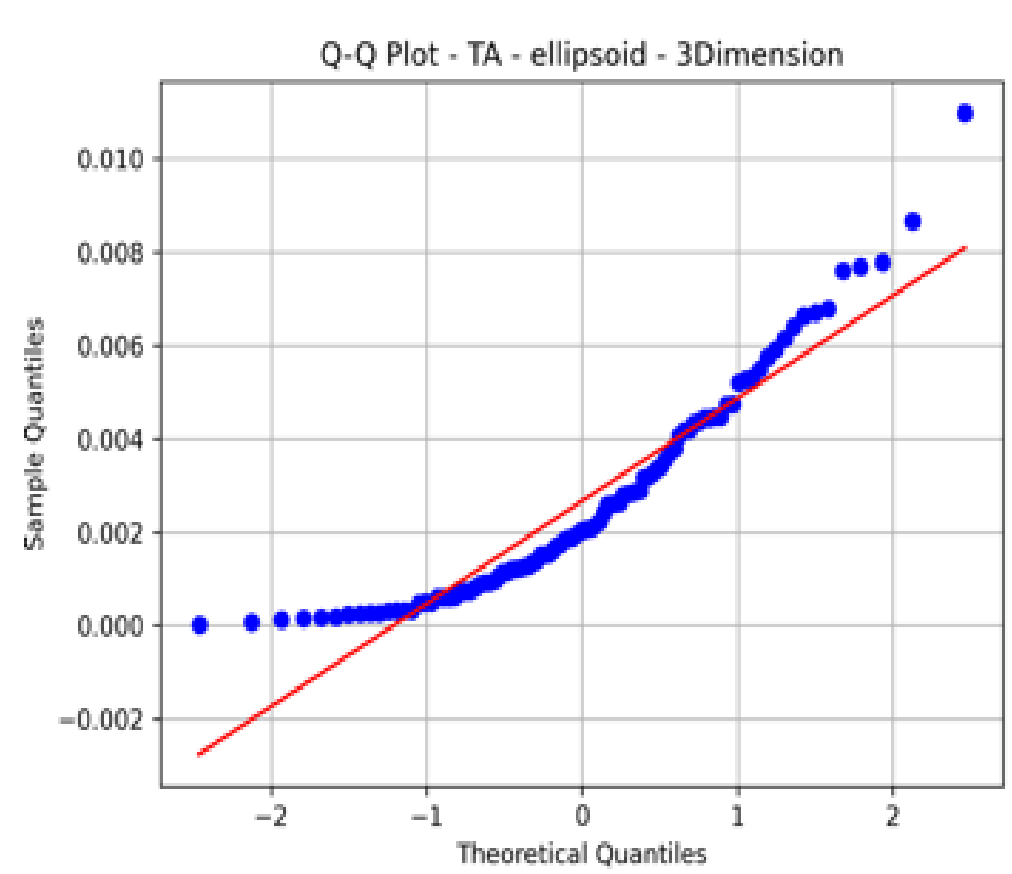

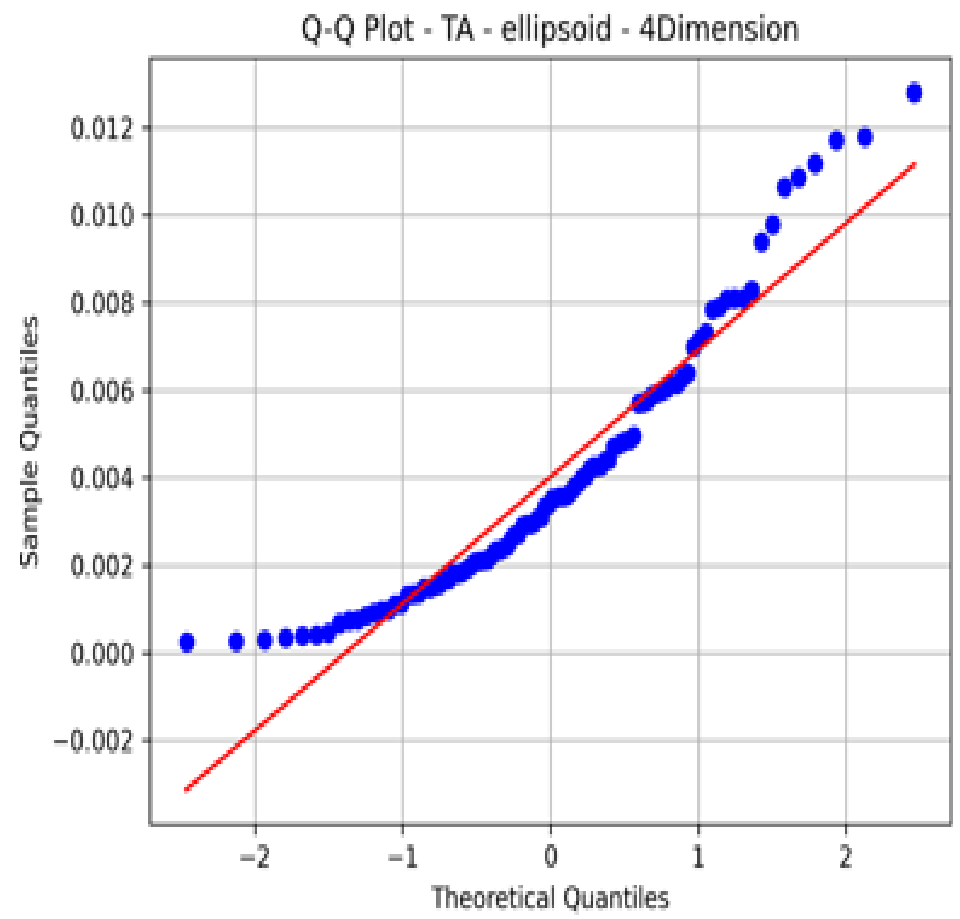
Q-Q Plot - TA - ellipsoid - 4Dimension
0.012
0.010
0.008
0.006
0.004
0.002
0.000
−0.002
Sample Quantiles
−2
−1
0
1
2
Theoretical Quantiles

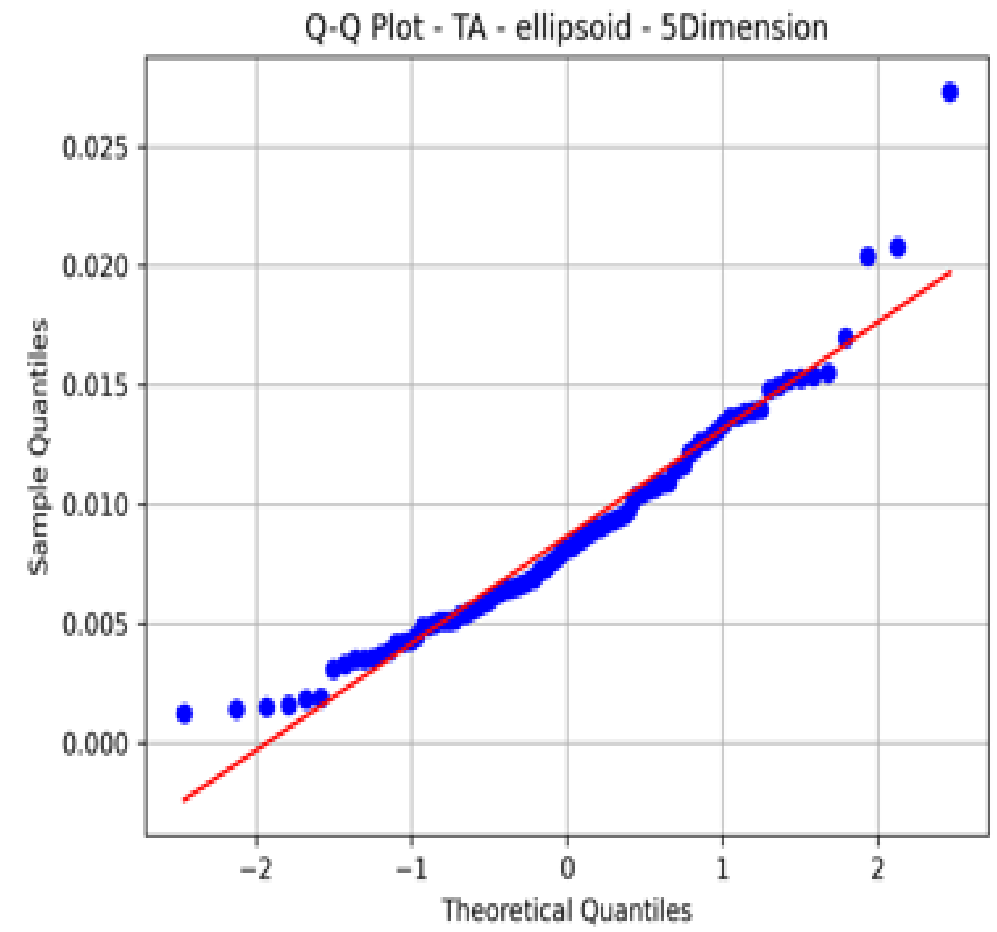
Q-Q Plot - TA - ellipsoid - 5Dimension
0.025
0.020
0.015
0.010
0.005
0.000
Sample Quantiles
−2
−1
0
1
2
Theoretical Quantiles

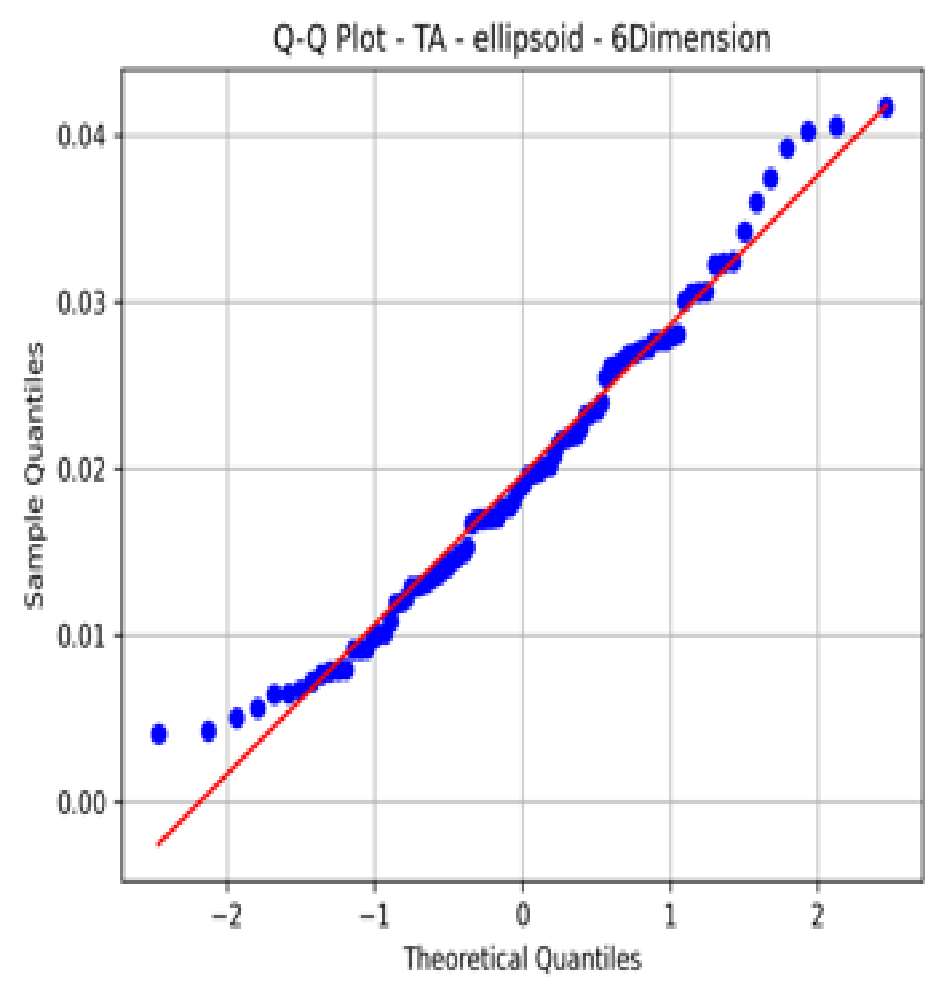
Q-Q Plot - TA - ellipsoid - 6Dimension
0.04
0.03
0.02
0.01
0.00
Sample Quantiles
−2
−1
0
1
2
Theoretical Quantiles

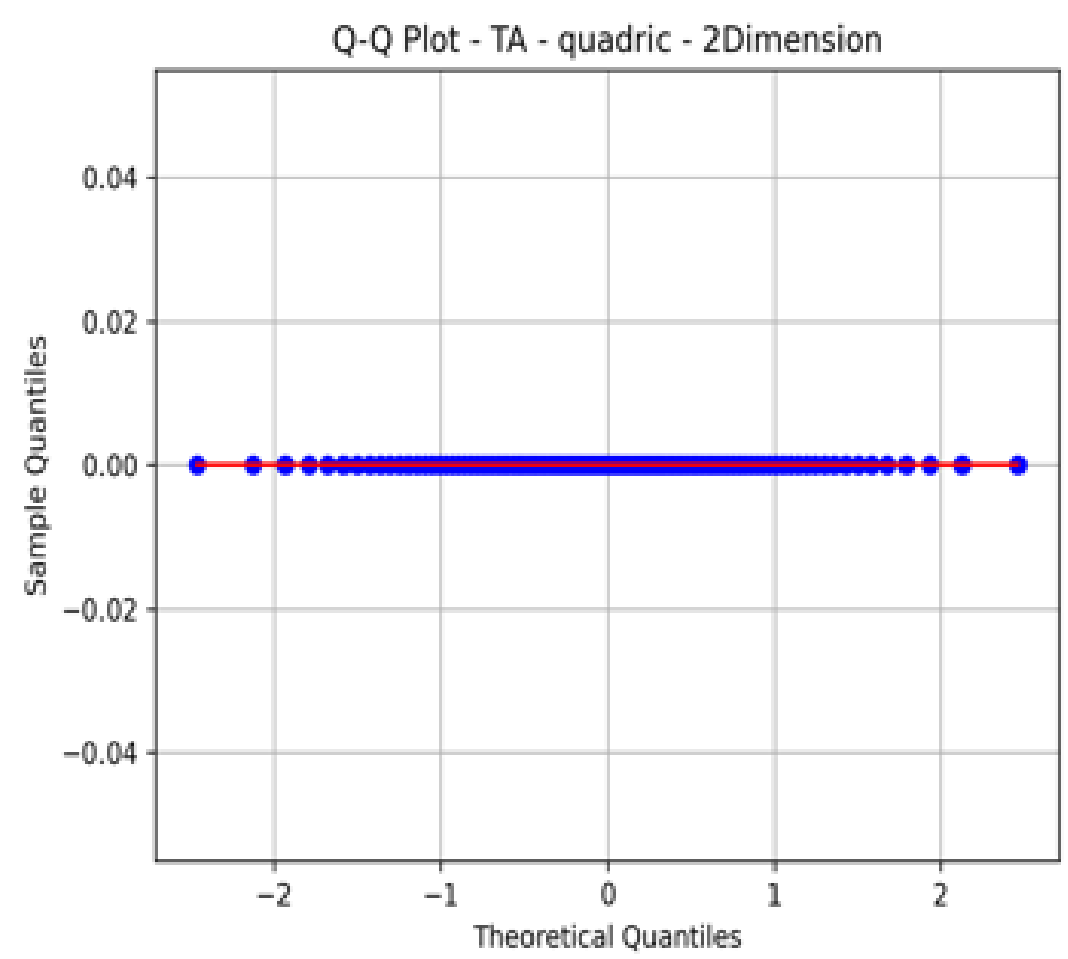
Q-Q Plot - TA - quadric - 2Dimension
0.04
0.02
0.00
−0.02
−0.04
Sample Quantiles
−2
−1
0
1
2
Theoretical Quantiles

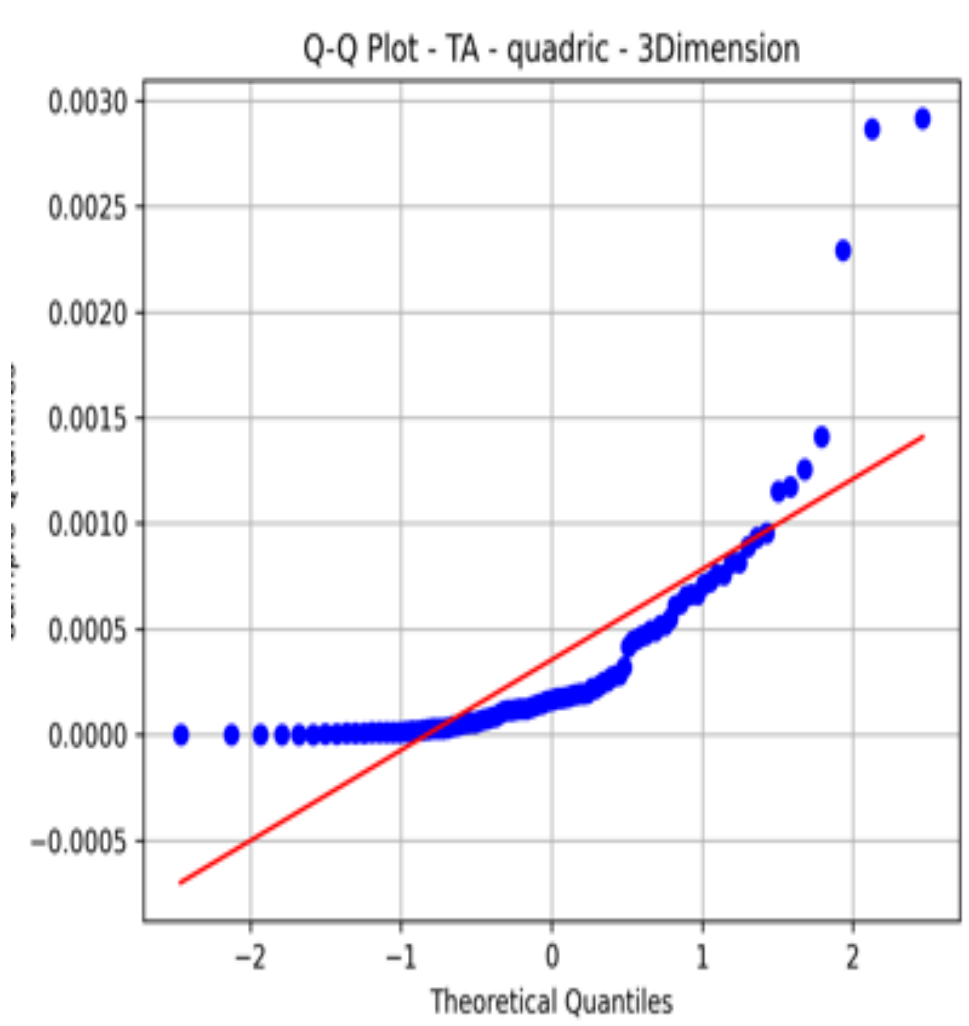
Q-Q Plot - TA - quadric - 3Dimension
0.0030
0.0025
0.0020
0.0015
0.0010
0.0005
0.0000
−0.0005
−2
−1
0
1
2
Theoretical Quantiles

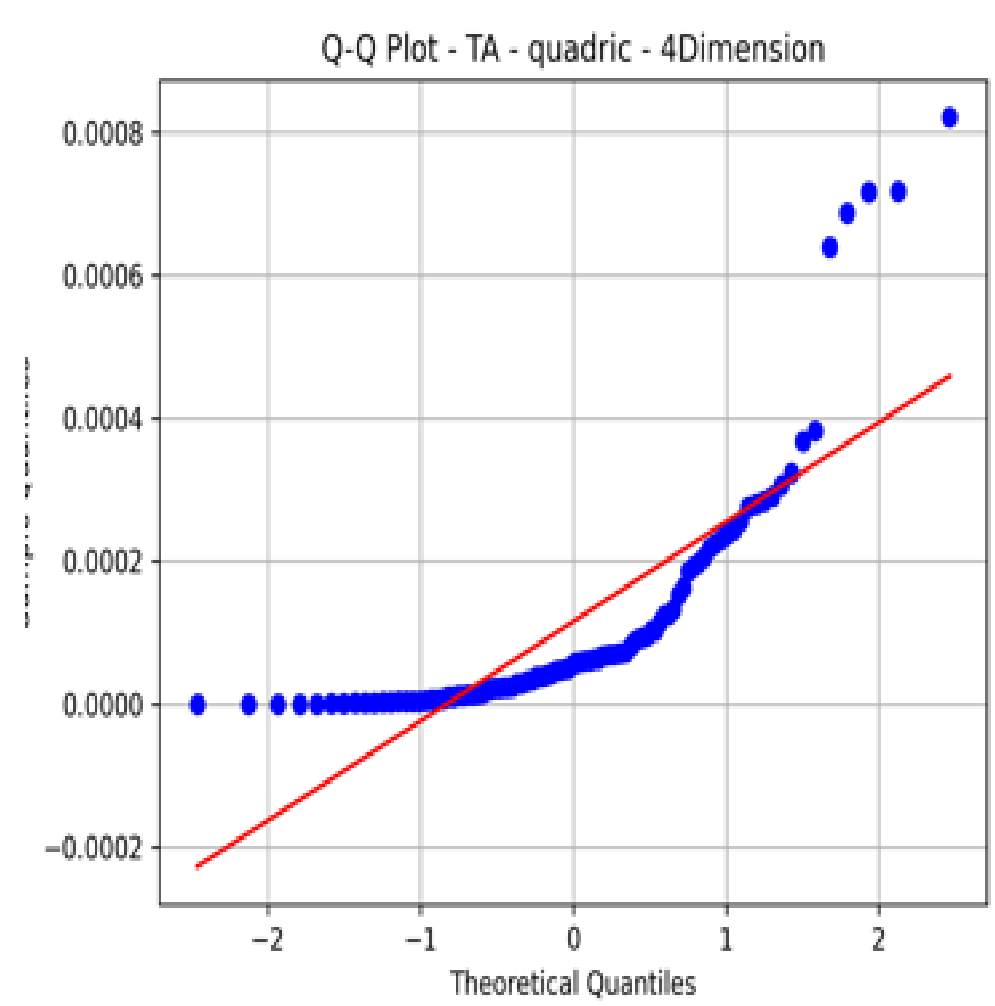
Q-Q Plot - TA - quadric - 4Dimension
0.0008
0.0006
0.0004
0.0002
0.0000
−0.0002
−2
−1
0
1
2
Theoretical Quantiles

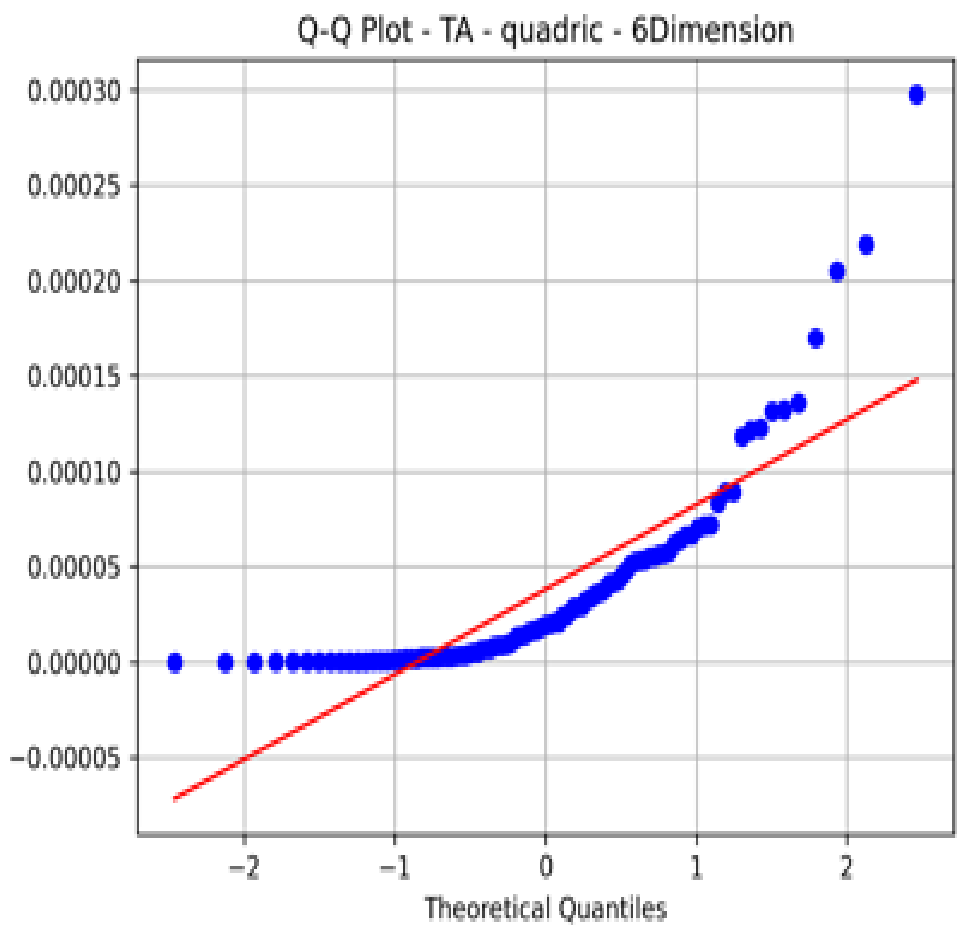
Q-Q Plot - TA - quadric - 6Dimension
0.00030
0.00025
0.00020
0.00015
0.00010
0.00005
0.00000
−0.00005
−2
−1
0
1
2
Theoretical Quantiles

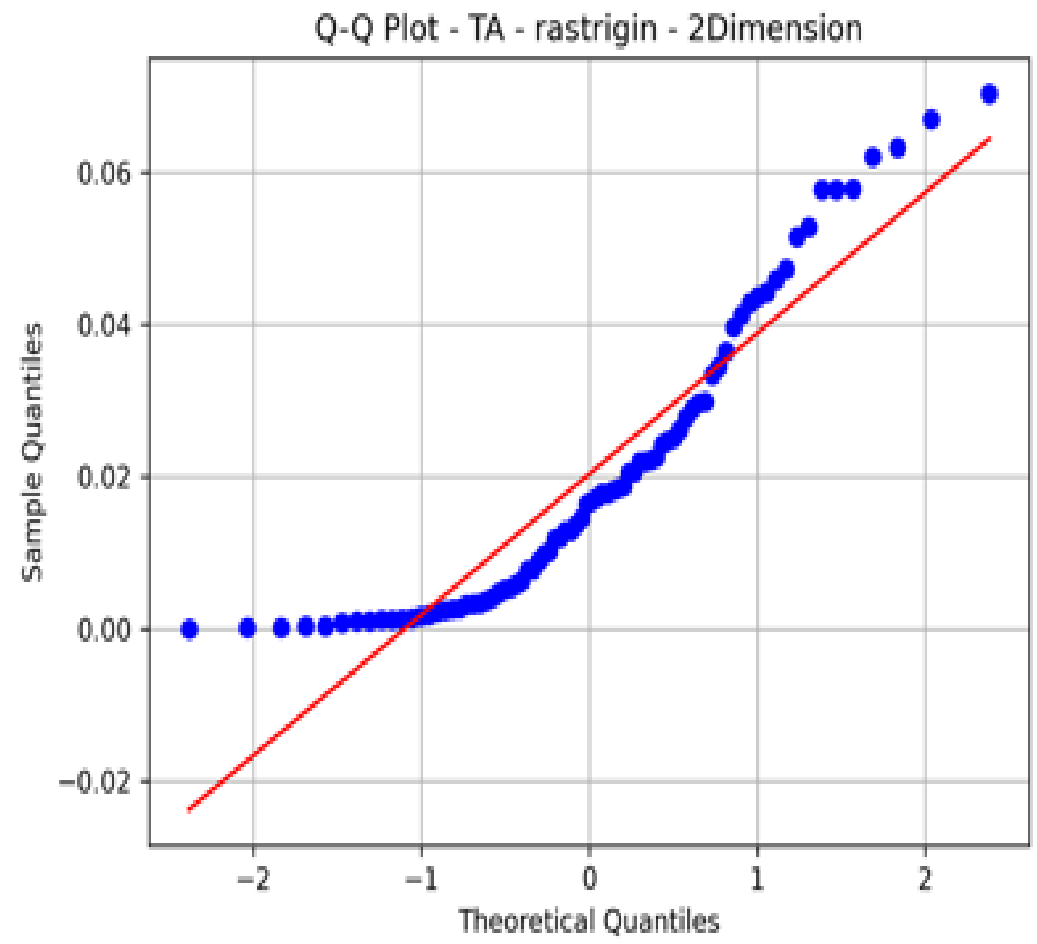
Q-Q Plot - TA - rastrigin - 2Dimension
0.06
0.04
0.02
0.00
−0.02
Sample Quantiles
−2
−1
0
1
2
Theoretical Quantiles

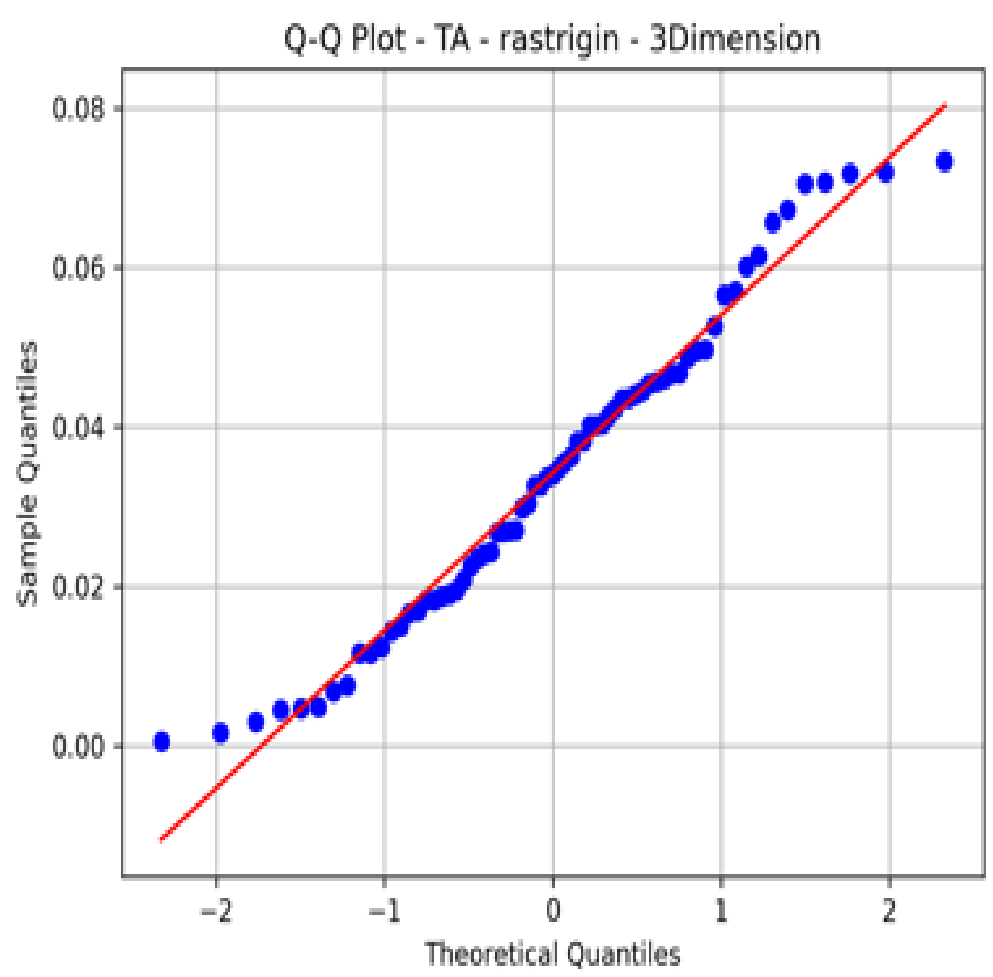
Q-Q Plot - TA - rastrigin - 3Dimension
0.08
0.06
0.04
0.02
0.00
Sample Quantiles
−2
−1
0
1
2
Theoretical Quantiles

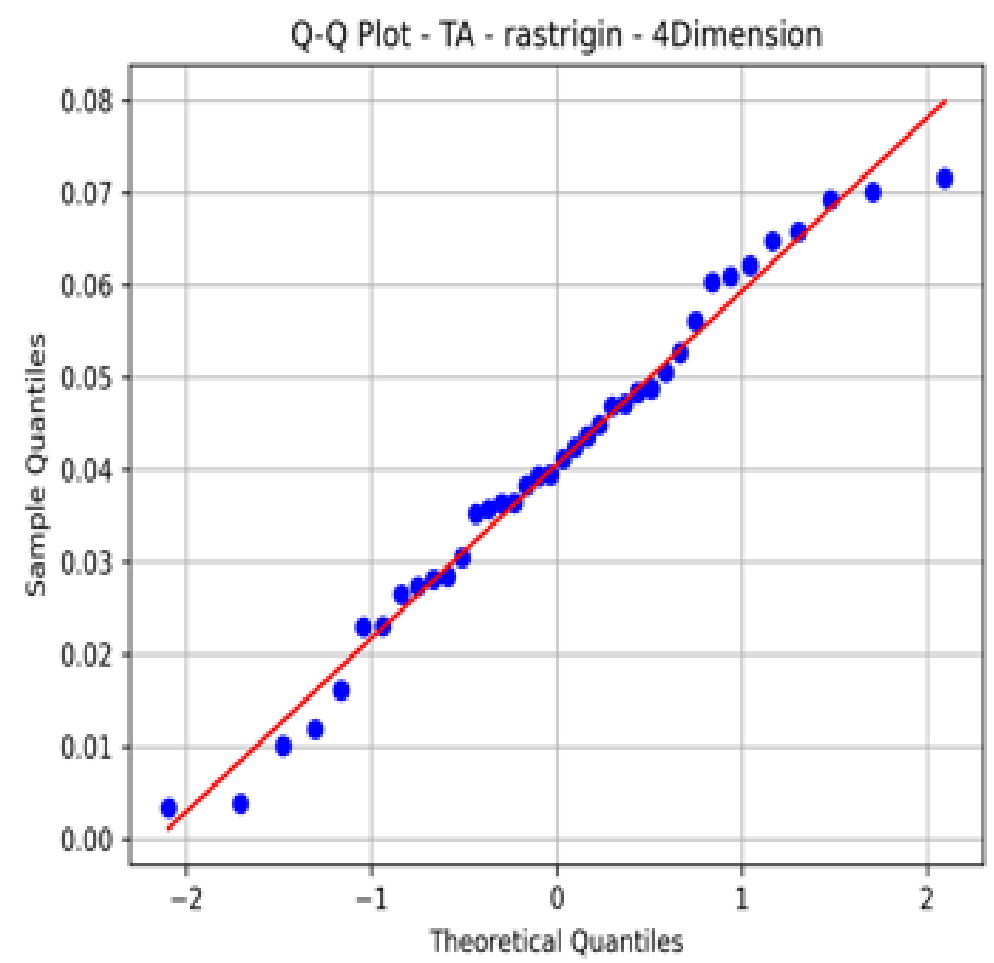
Q-Q Plot - TA - rastrigin - 4Dimension
0.08
0.07
0.06
0.05
0.04
0.03
0.02
0.01
0.00
Sample Quantiles
−2
−1
0
1
2
Theoretical Quantiles

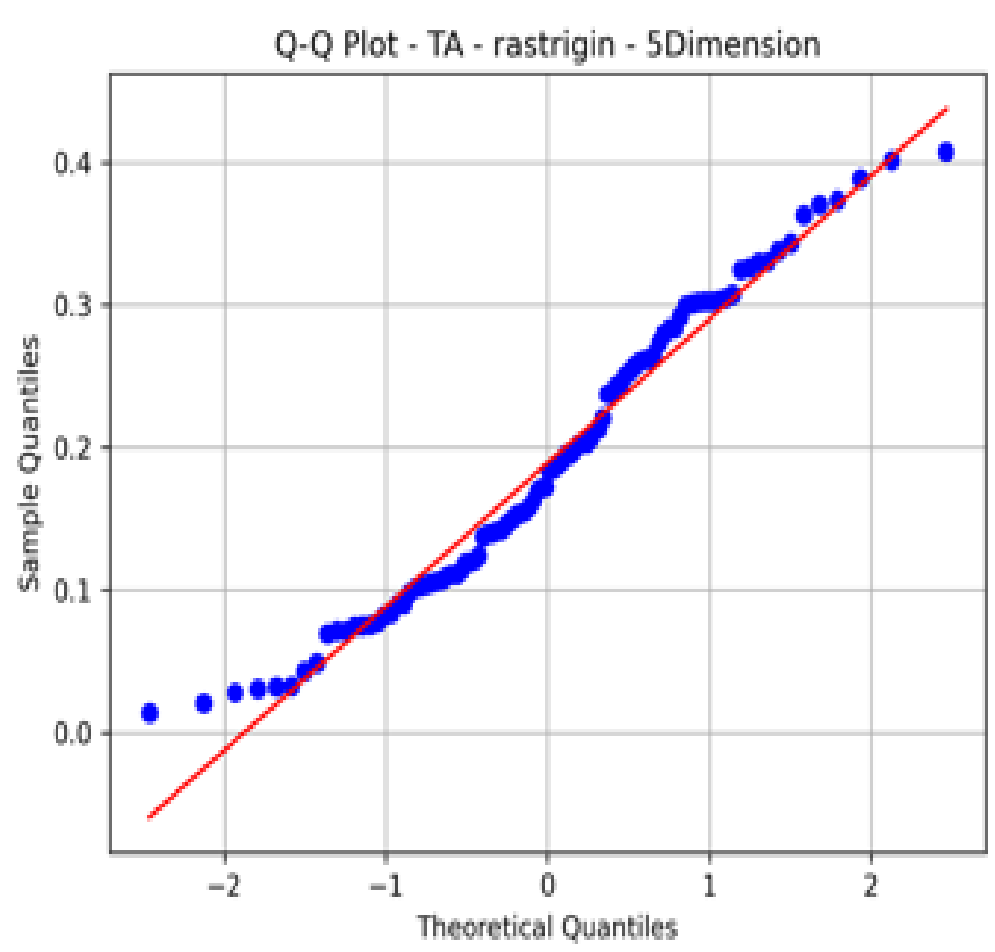
Q-Q Plot - TA - rastrigin - 5Dimension
0.4
0.3
0.2
0.1
0.0
Sample Quantiles
−2
−1
0
1
2
Theoretical Quantiles

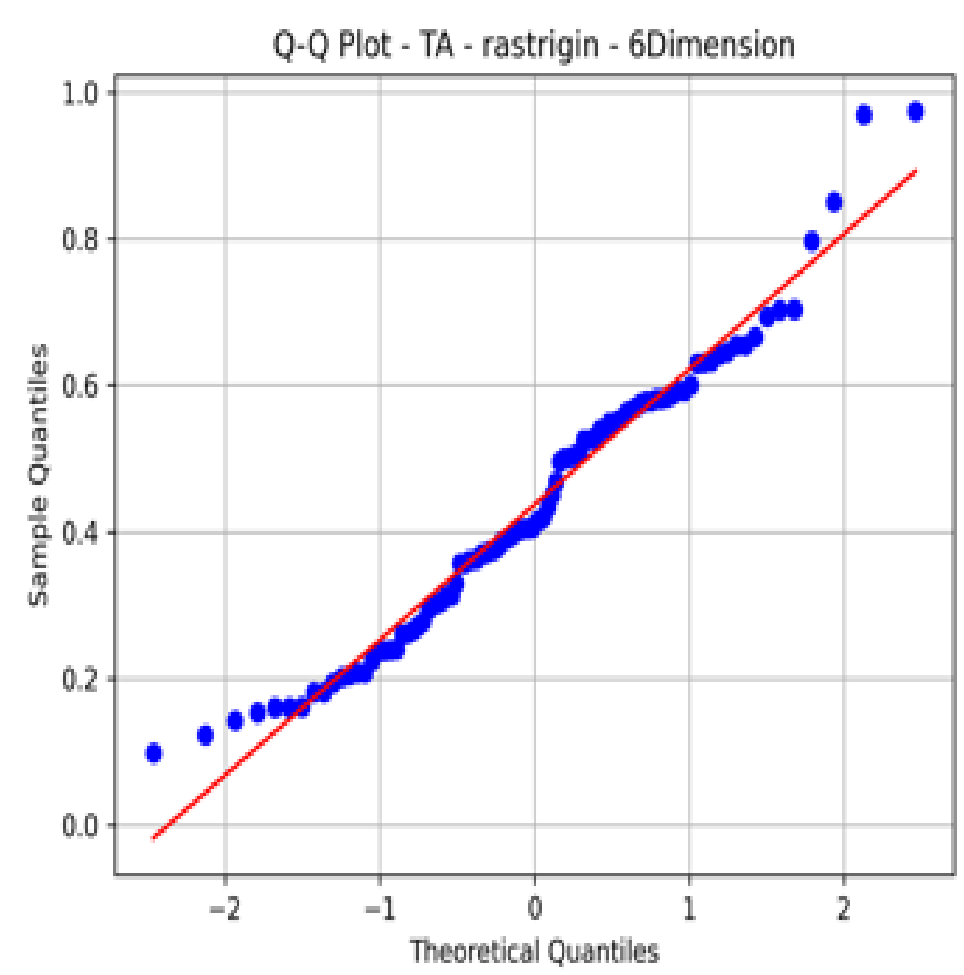
Q-Q Plot - TA - rastrigin - 6Dimension
1.0
0.8
0.6
0.4
0.2
0.0
Sample Quantiles
−2
−1
0
1
2
Theoretical Quantiles

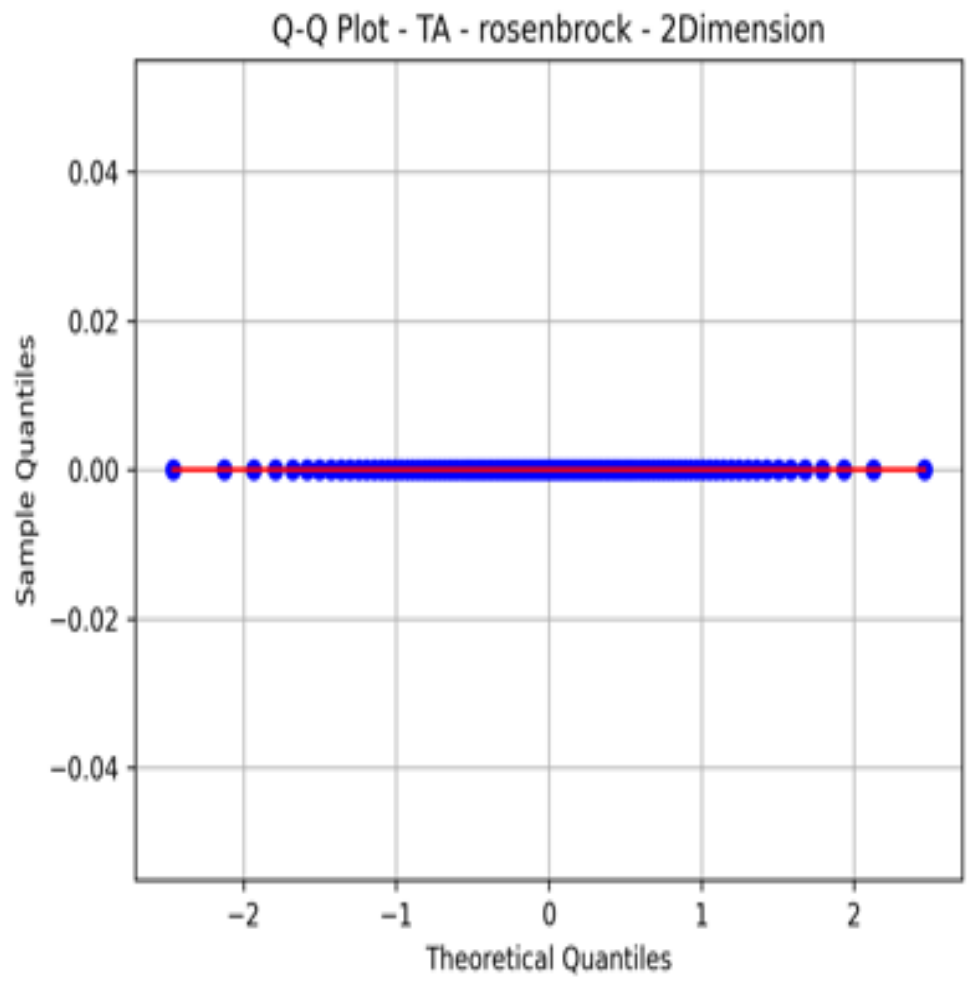
Q-Q Plot - TA - rosenbrock - 2Dimension
Sample Quantiles
Theoretical Quantiles
0.04
0.02
0.00
−0.02
−0.04
−2
−1
0
1
2

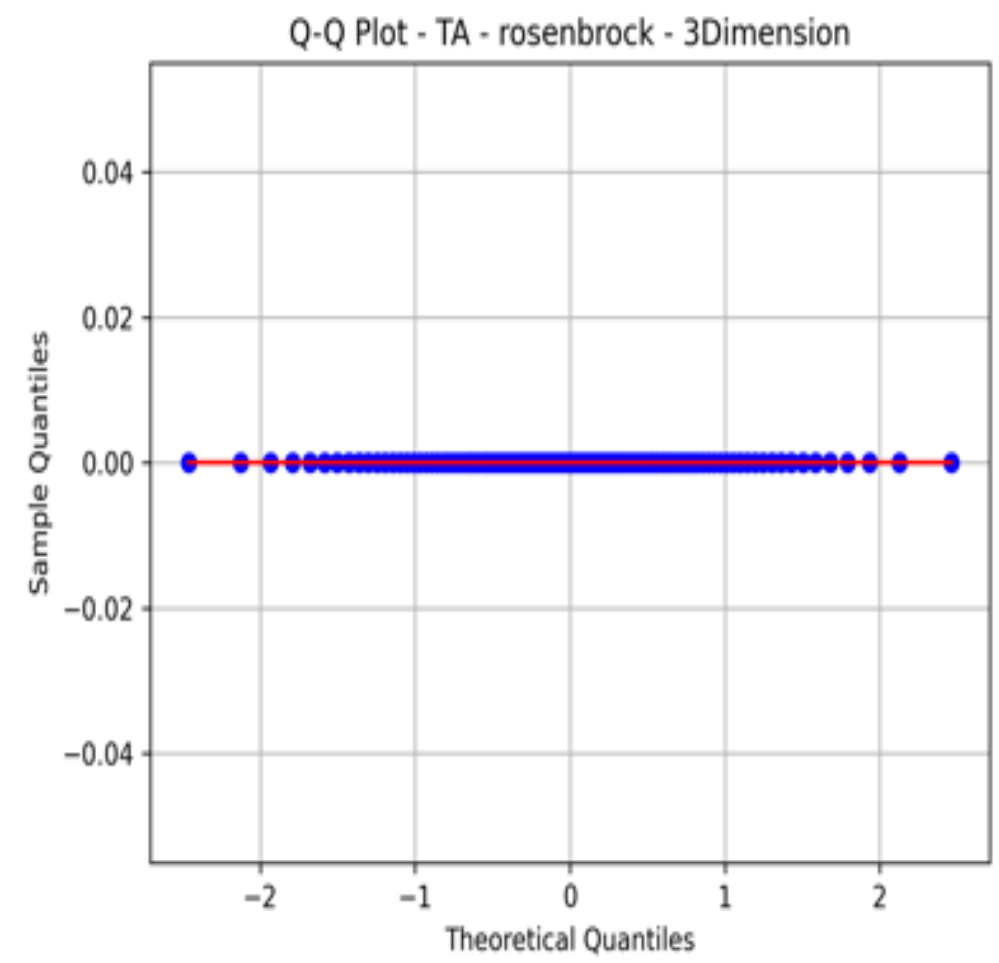
Q-Q Plot - TA - rosenbrock - 3Dimension
Sample Quantiles
Theoretical Quantiles
0.04
0.02
0.00
−0.02
−0.04
−2
−1
0
1
2

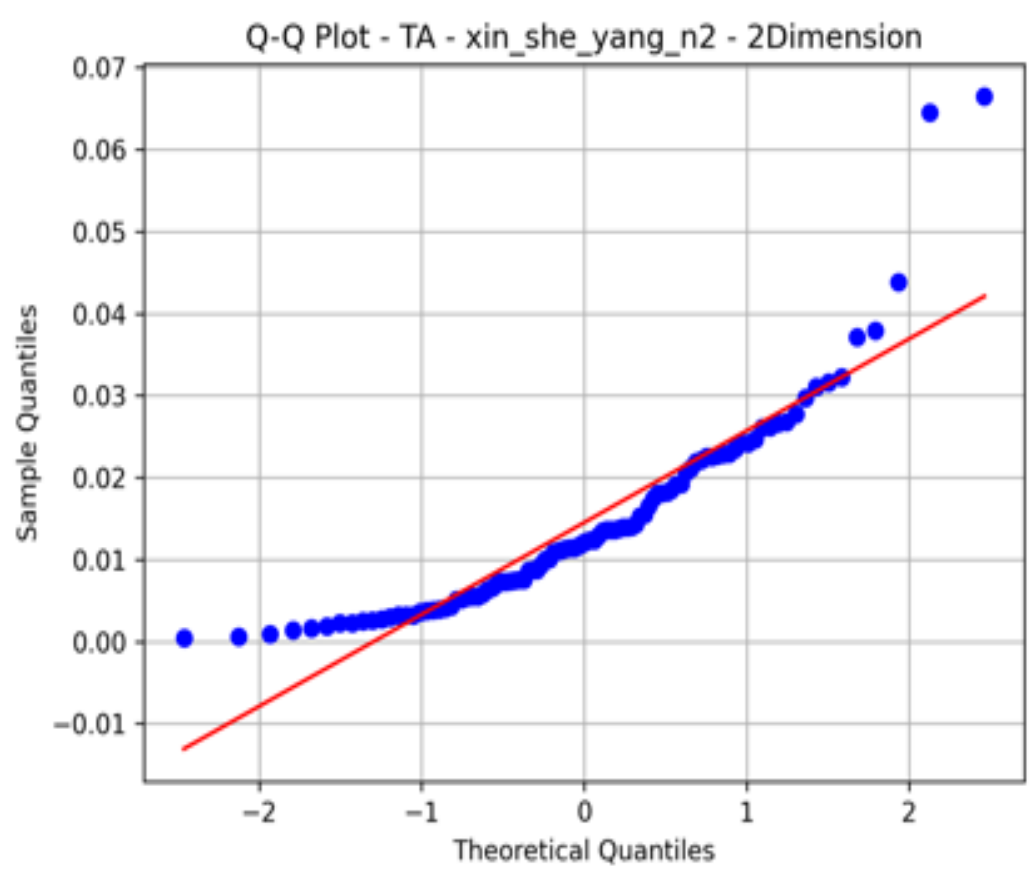
Q-Q Plot - TA - xin_she_yang_n2 - 2Dimension
Sample Quantiles
Theoretical Quantiles
0.07
0.06
0.05
0.04
0.03
0.02
0.01
0.00
−0.01
−2
−1
0
1
2

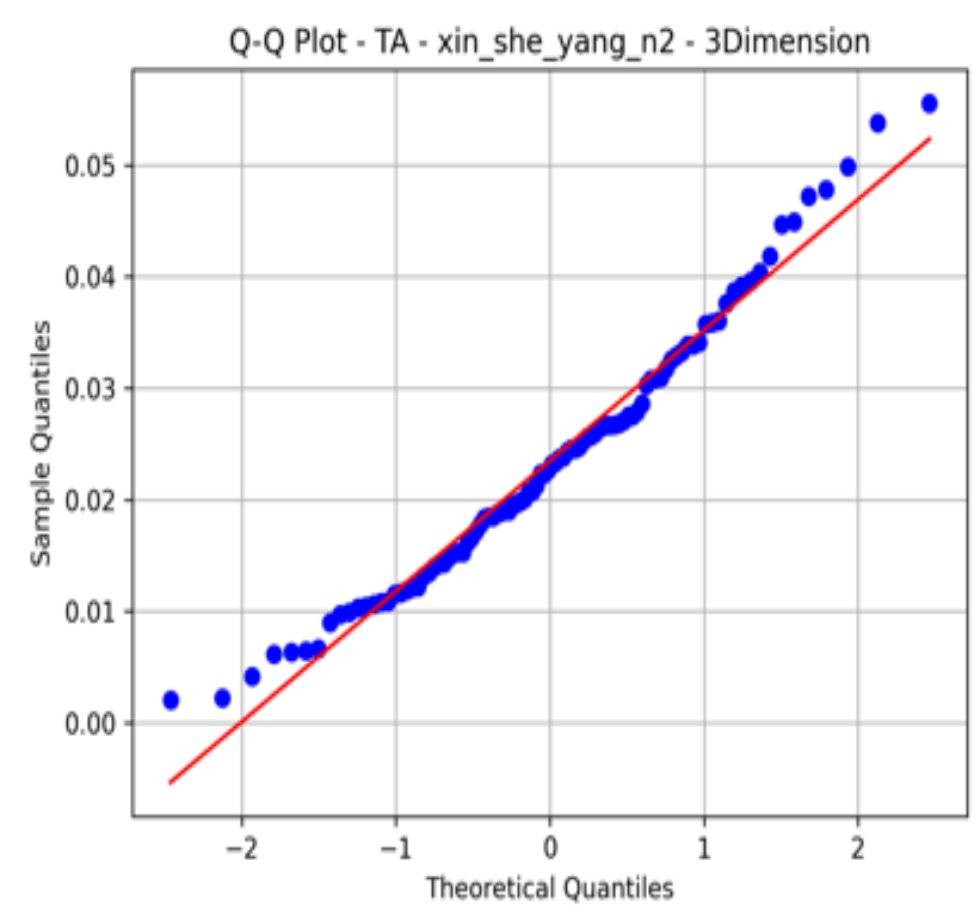
Q-Q Plot - TA - xin_she_yang_n2 - 3Dimension
Sample Quantiles
Theoretical Quantiles
0.05
0.04
0.03
0.02
0.01
0.00
−2
−1
0
1
2

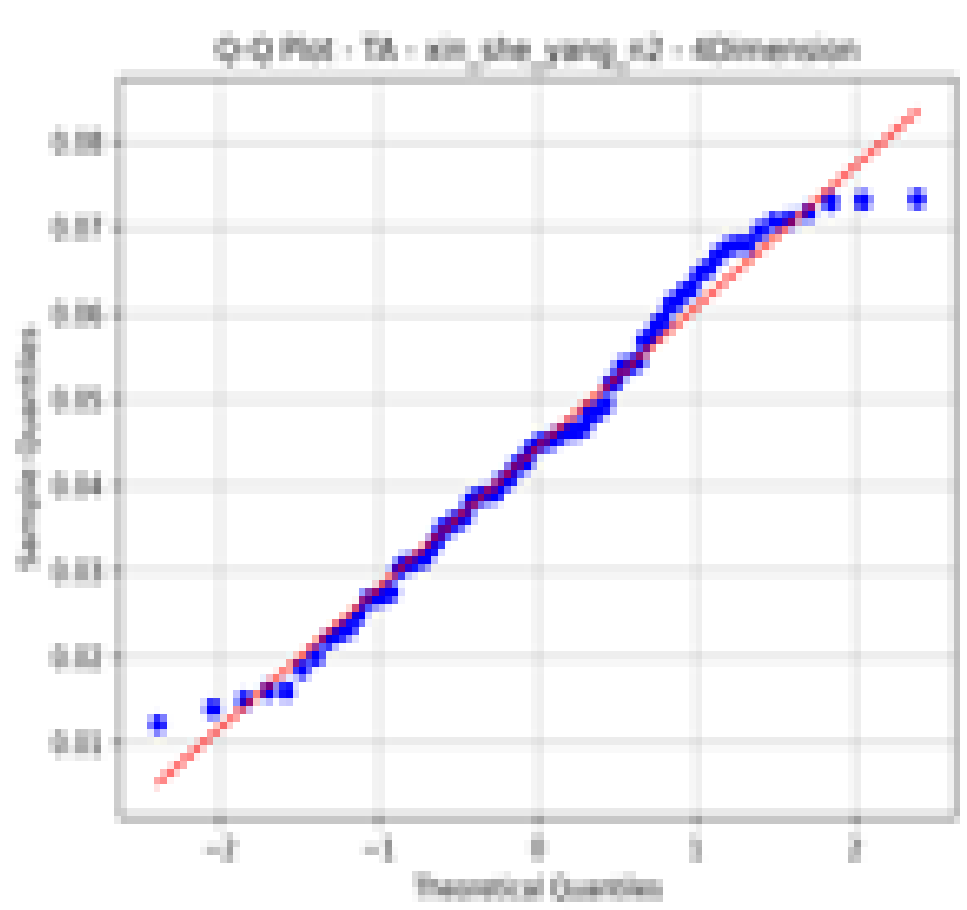

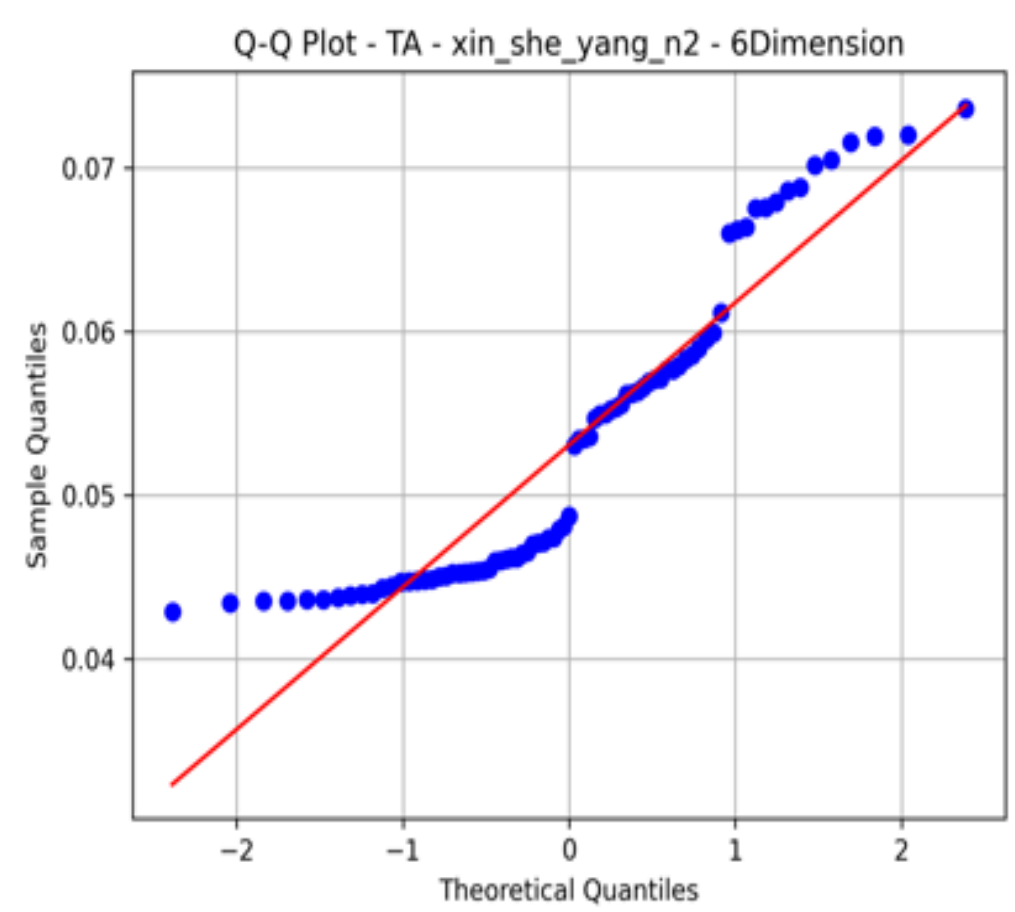
Q-Q Plot - TA - xin_she_yang_n2 - 6Dimension
Sample Quantiles
Theoretical Quantiles
0.07
0.06
0.05
0.04
−2
−1
0
1
2

PSO Q-Q Plots

Below are the  Continuous functions QQ plots

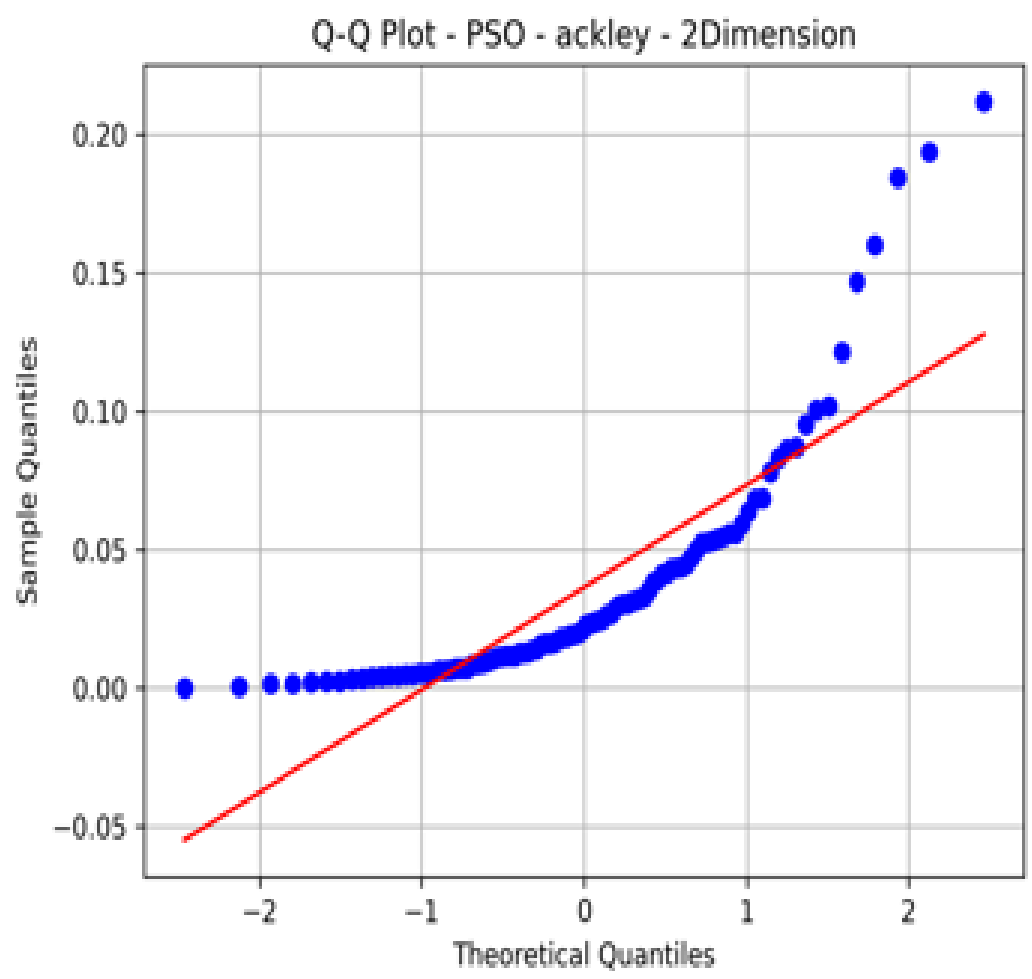

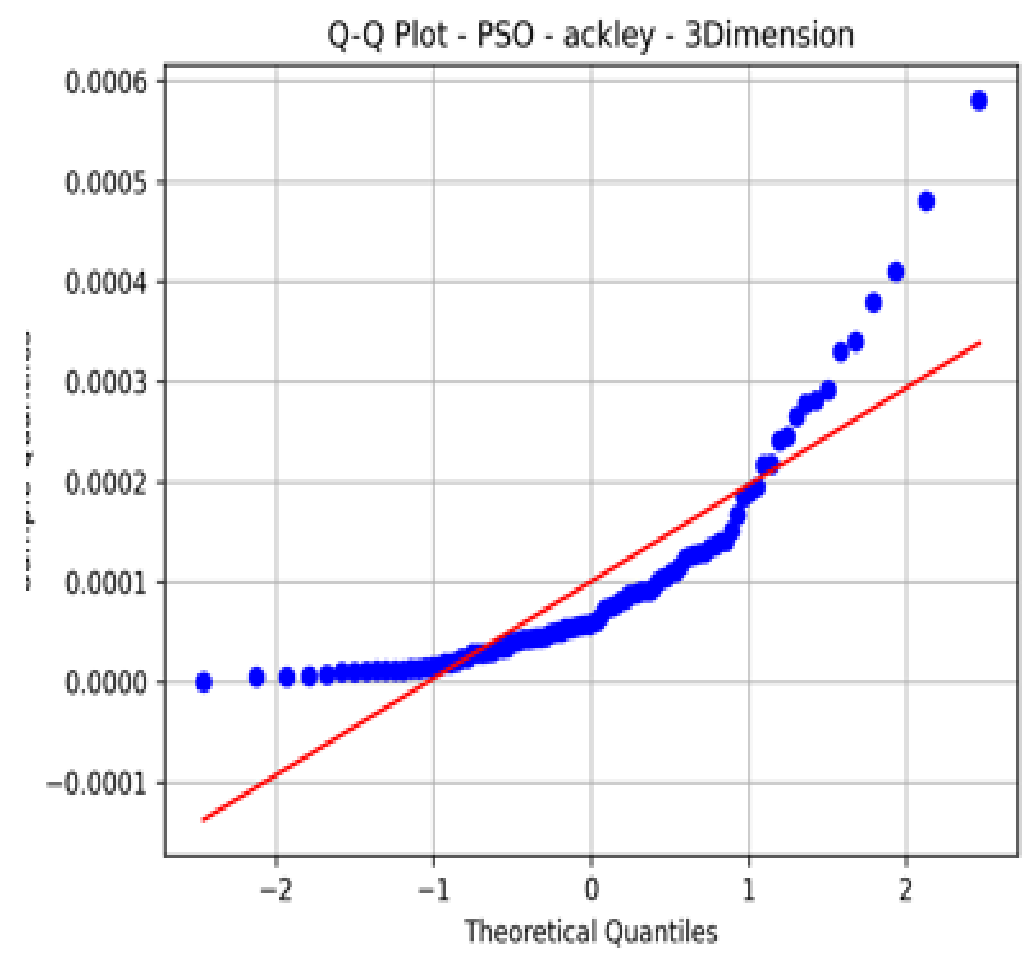

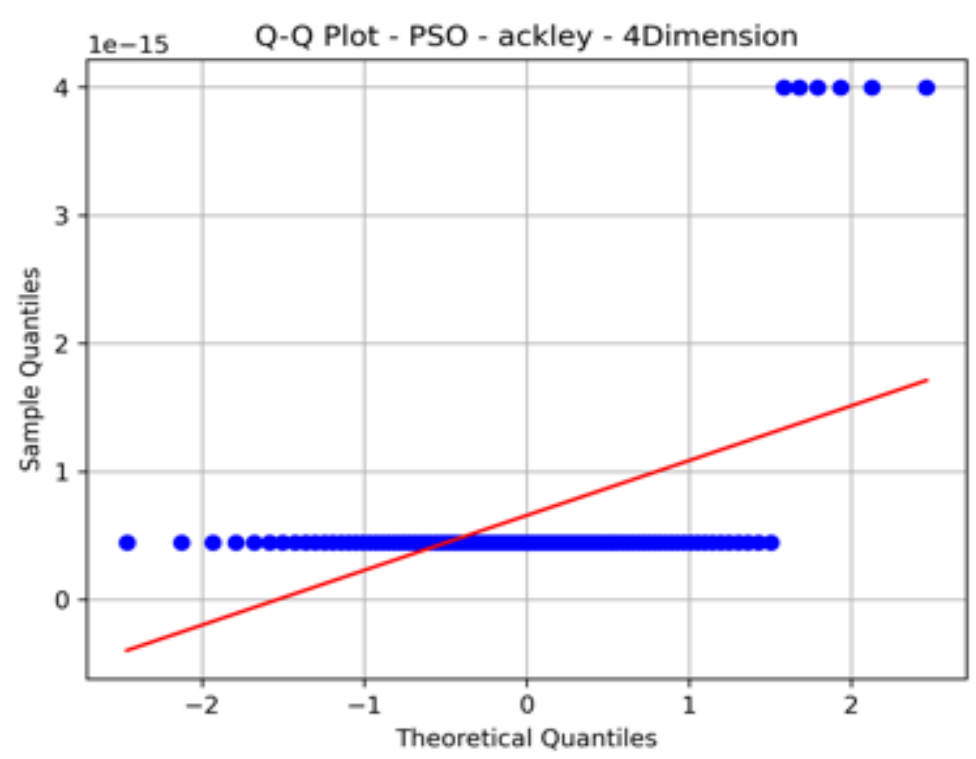

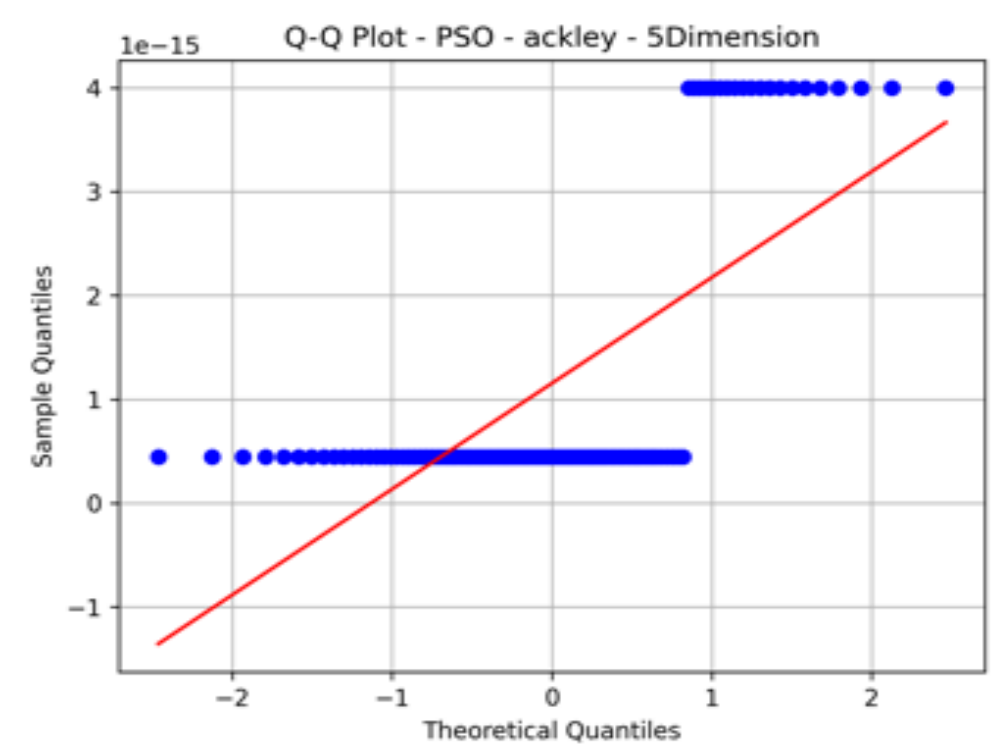

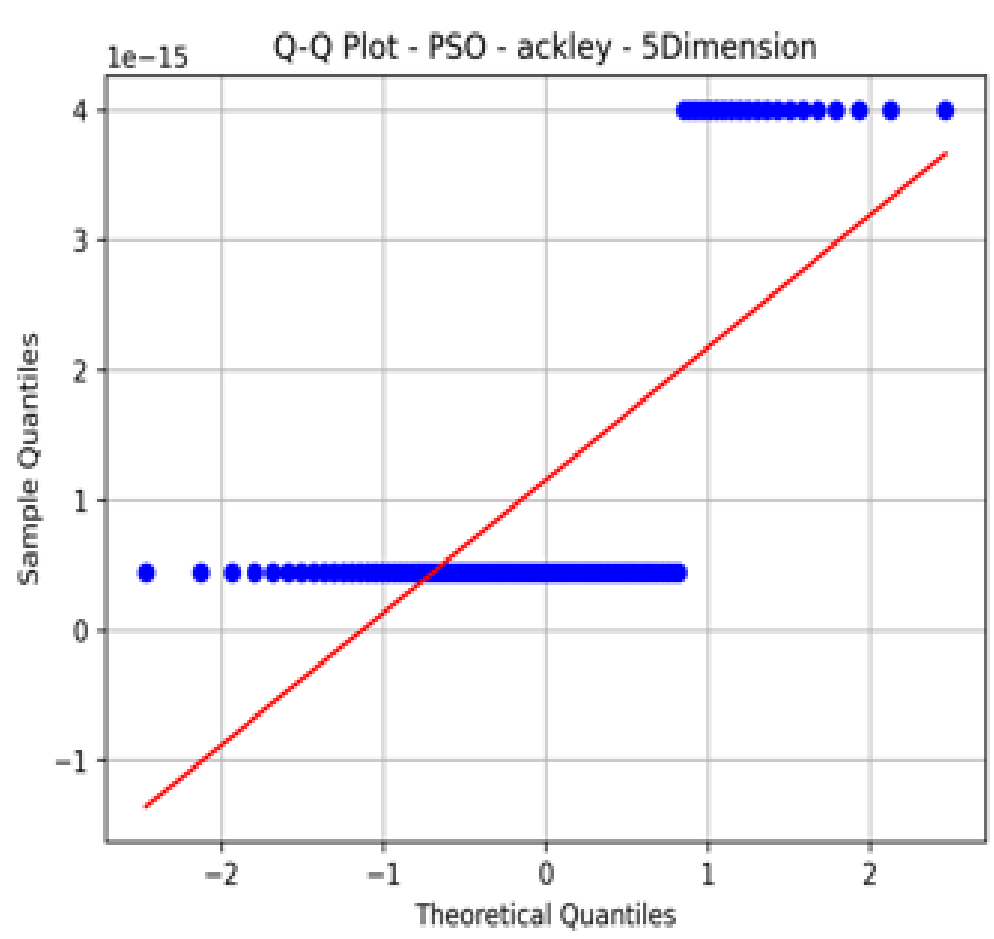

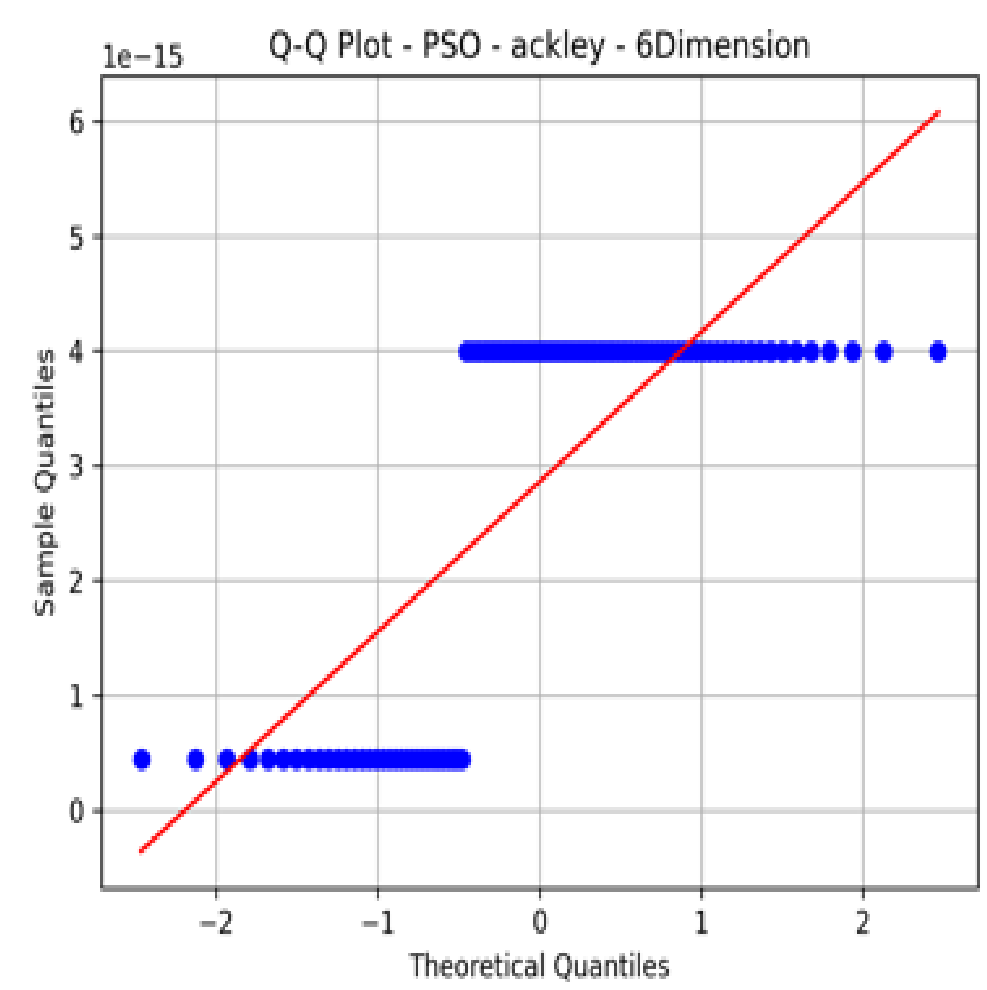

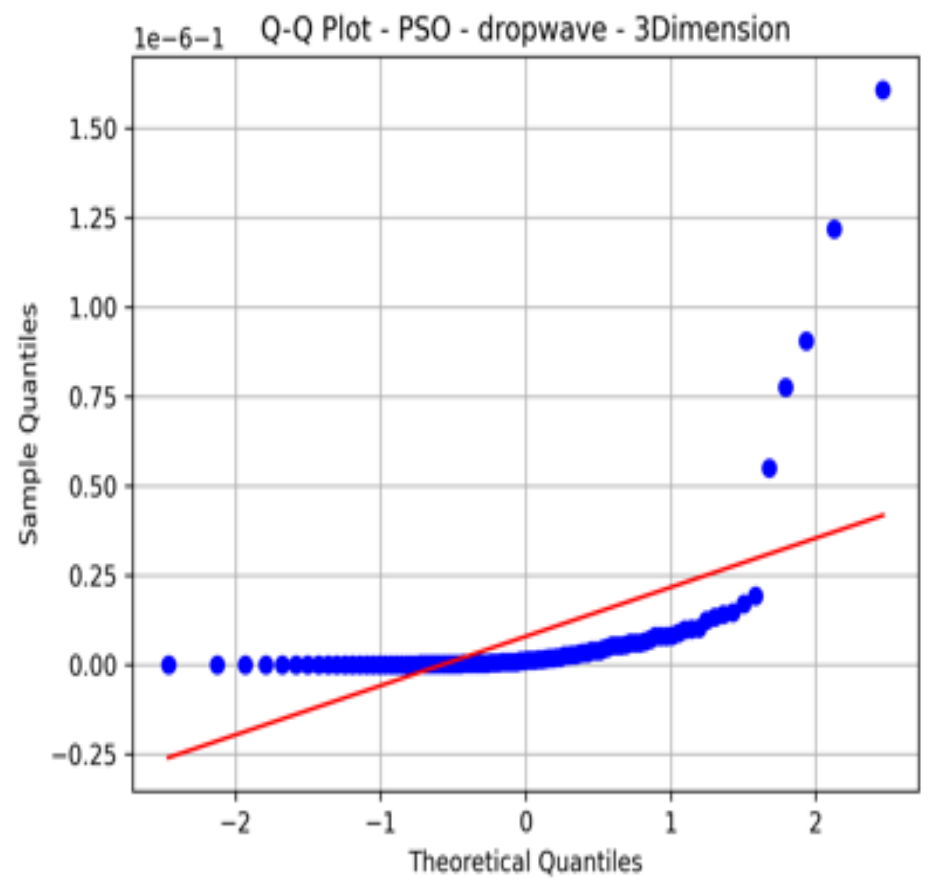
Q-Q Plot - PSO - dropwave - 3Dimension
1e−6−1
1.50
1.25
1.00
0.75
0.50
0.25
0.00
−0.25
Sample Quantiles
−2
−1
0
1
2
Theoretical Quantiles

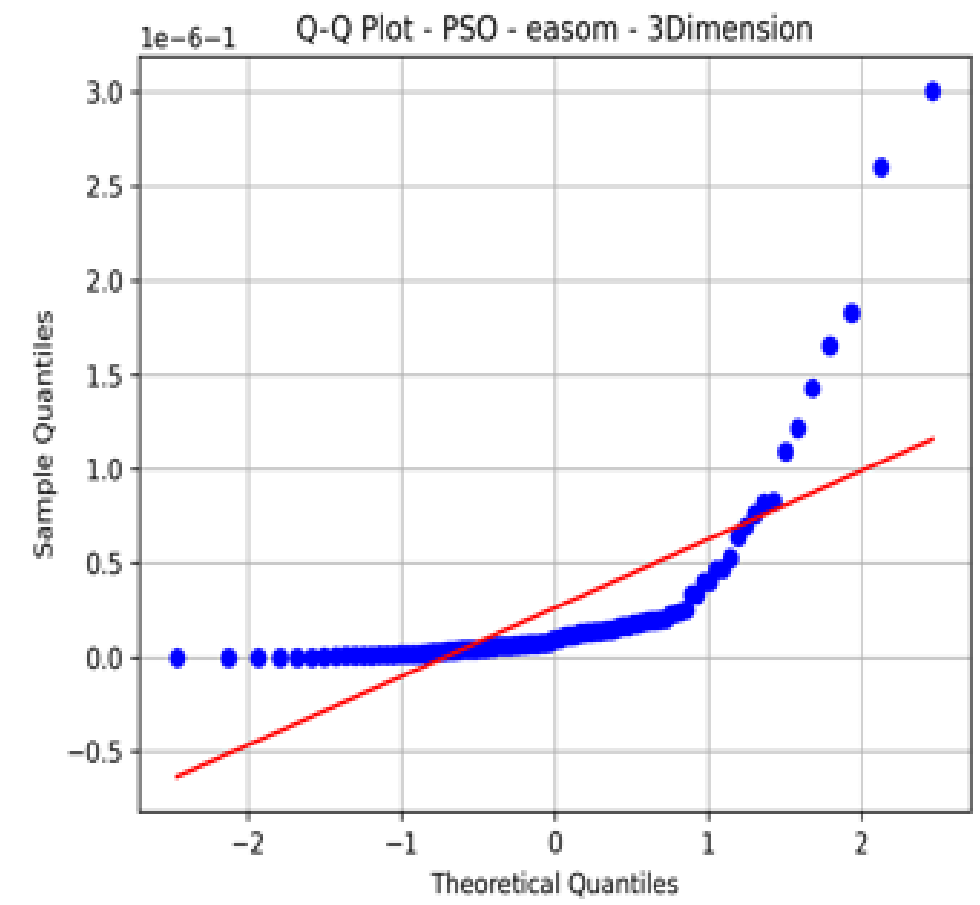
Q-Q Plot - PSO - easom - 3Dimension
1e−6−1
3.0
2.5
2.0
1.5
1.0
0.5
0.0
−0.5
Sample Quantiles
−2
−1
0
1
2
Theoretical Quantiles

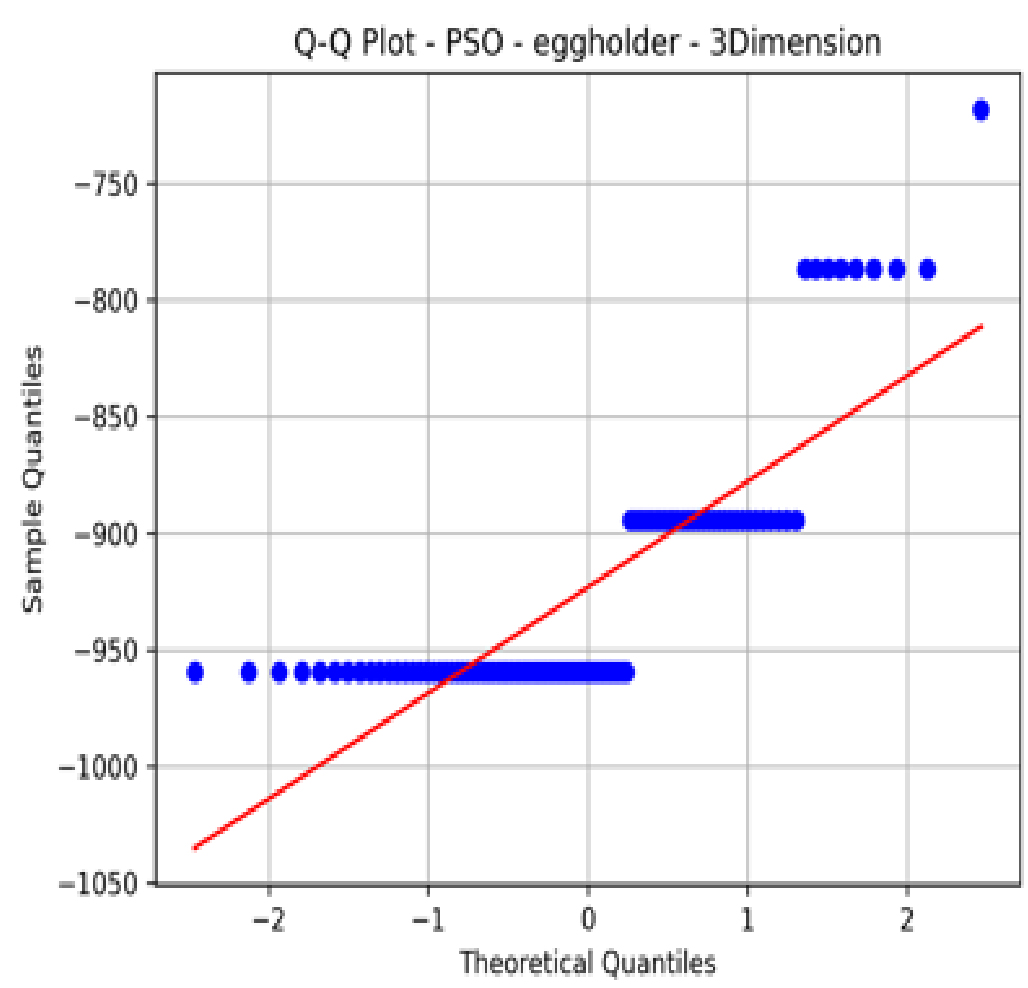
Q-Q Plot - PSO - eggholder - 3Dimension
−750
−800
−850
−900
−950
−1000
−1050
Sample Quantiles
−2
−1
0
1
2
Theoretical Quantiles

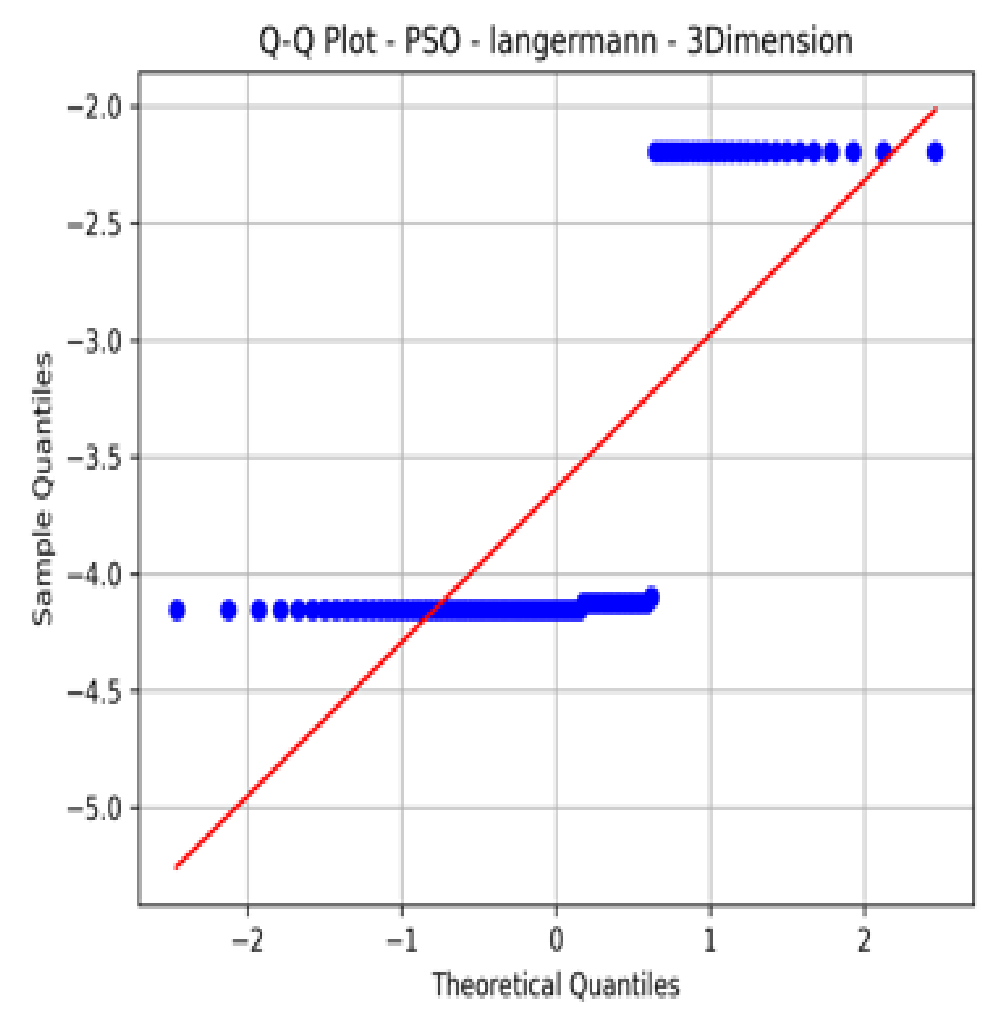
Q-Q Plot - PSO - langermann - 3Dimension
−2.0
−2.5
−3.0
−3.5
−4.0
−4.5
−5.0
Sample Quantiles
−2
−1
0
1
2
Theoretical Quantiles

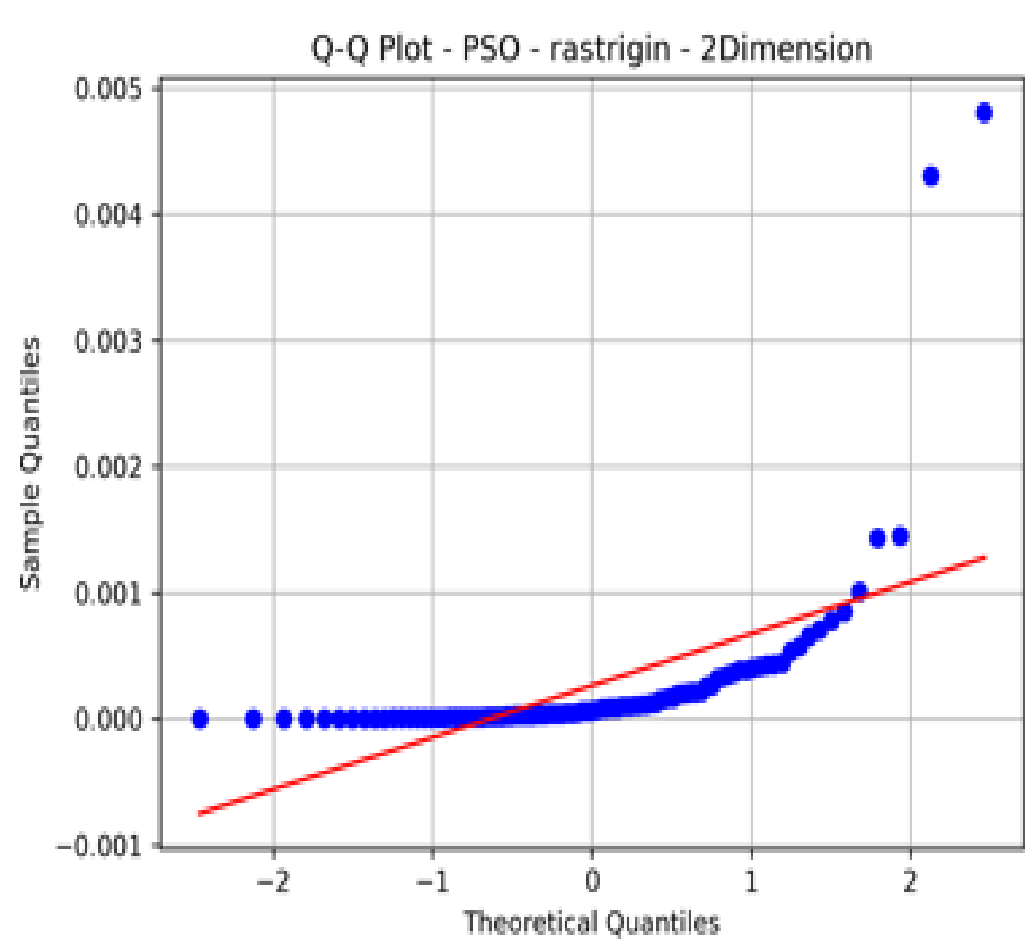
Q-Q Plot - PSO - rastrigin - 2Dimension
0.005
0.004
0.003
0.002
0.001
0.000
−0.001
Sample Quantiles
−2
−1
0
1
2
Theoretical Quantiles

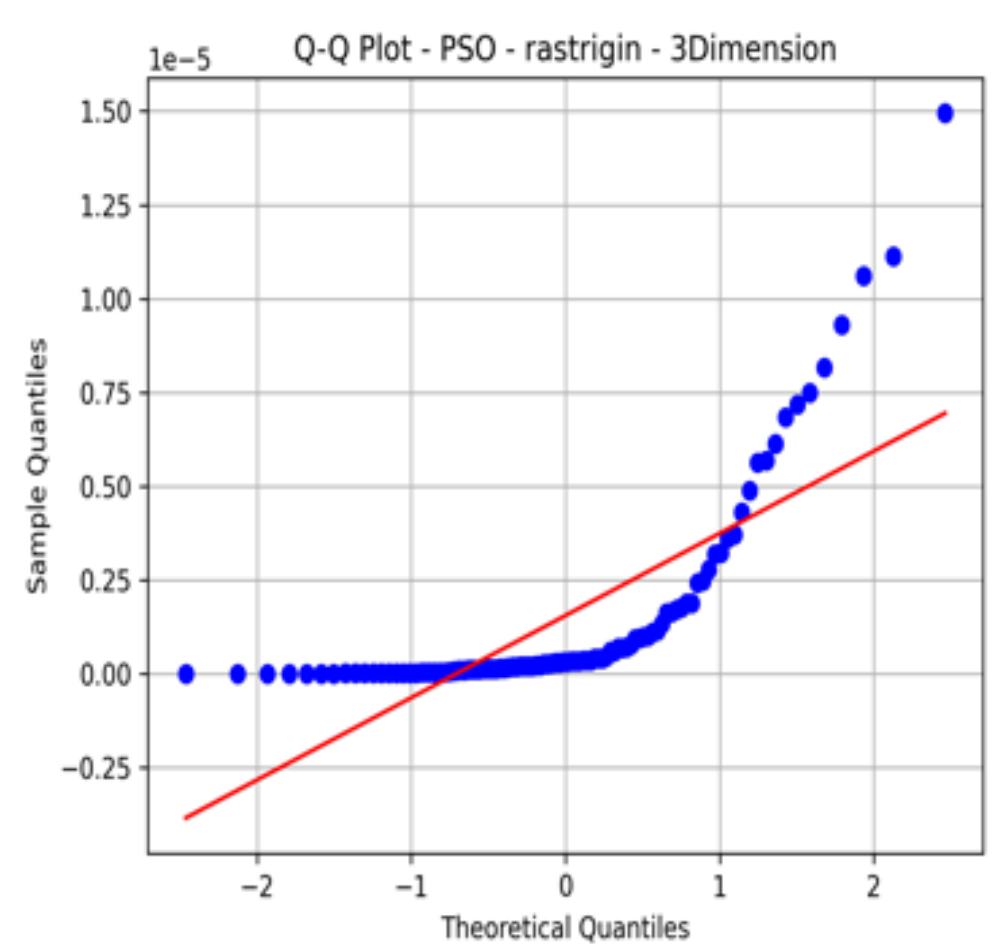
Q-Q Plot - PSO - rastrigin - 3Dimension
1e−5
1.50
1.25
1.00
0.75
0.50
0.25
0.00
−0.25
Sample Quantiles
−2
−1
0
1
2
Theoretical Quantiles

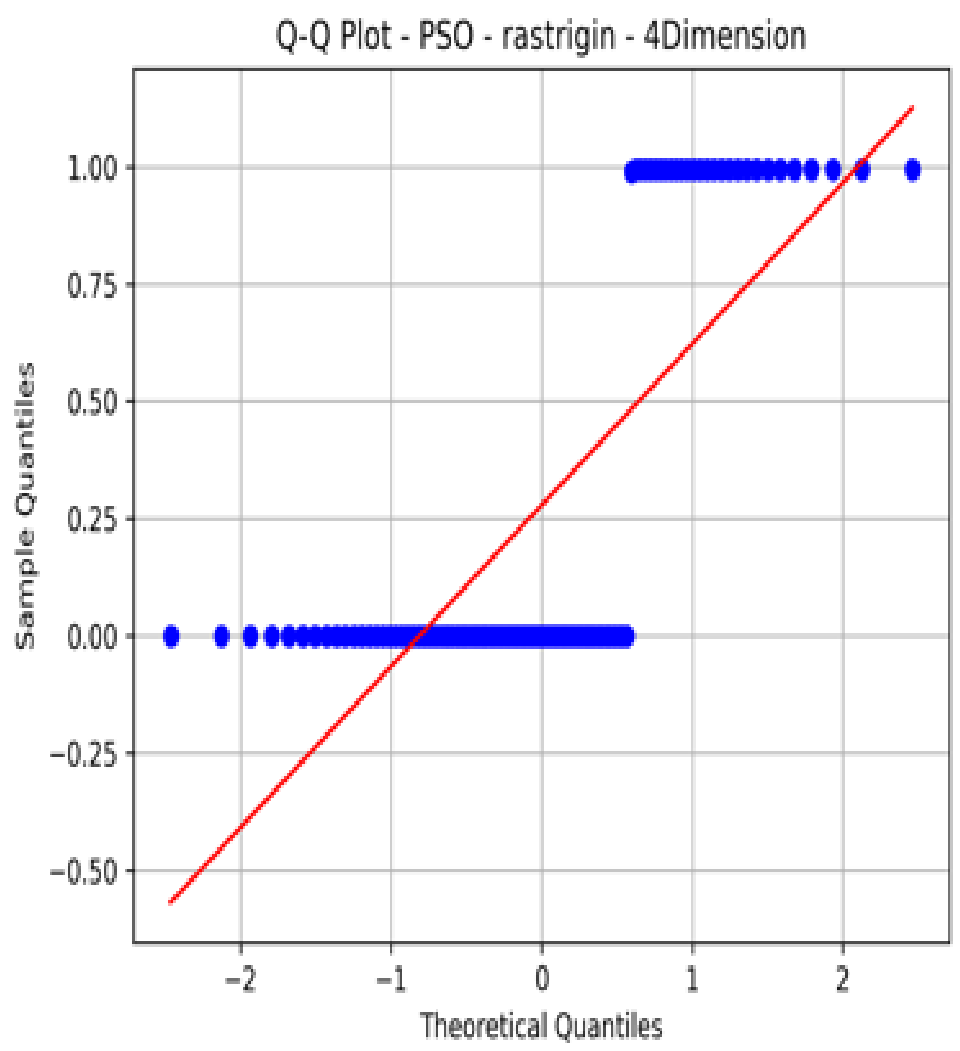
Q-Q Plot - PSO - rastrigin - 4Dimension
1.00
0.75
0.50
0.25
0.00
−0.25
−0.50
Sample Quantiles
−2
−1
0
1
2
Theoretical Quantiles

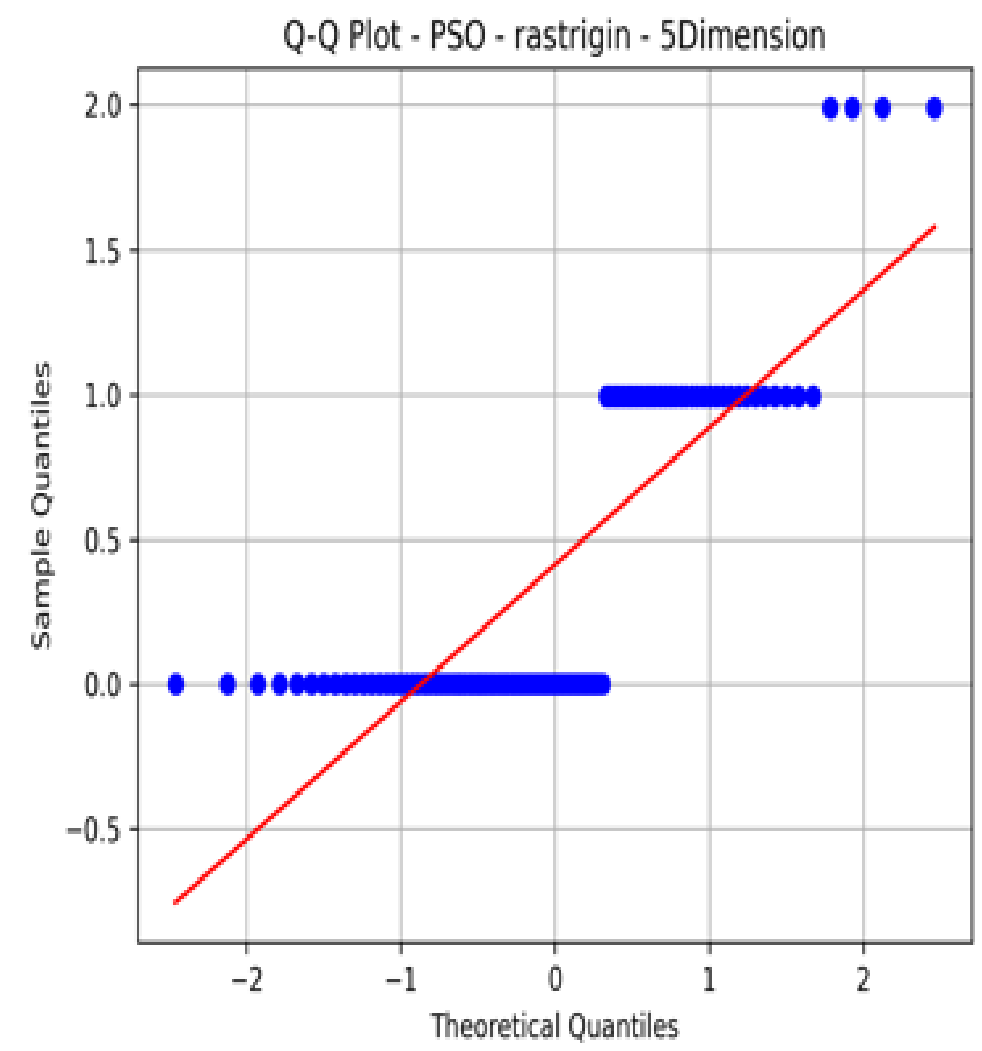
Q-Q Plot - PSO - rastrigin - 5Dimension
2.0
1.5
1.0
0.5
0.0
−0.5
Sample Quantiles
−2
−1
0
1
2
Theoretical Quantiles

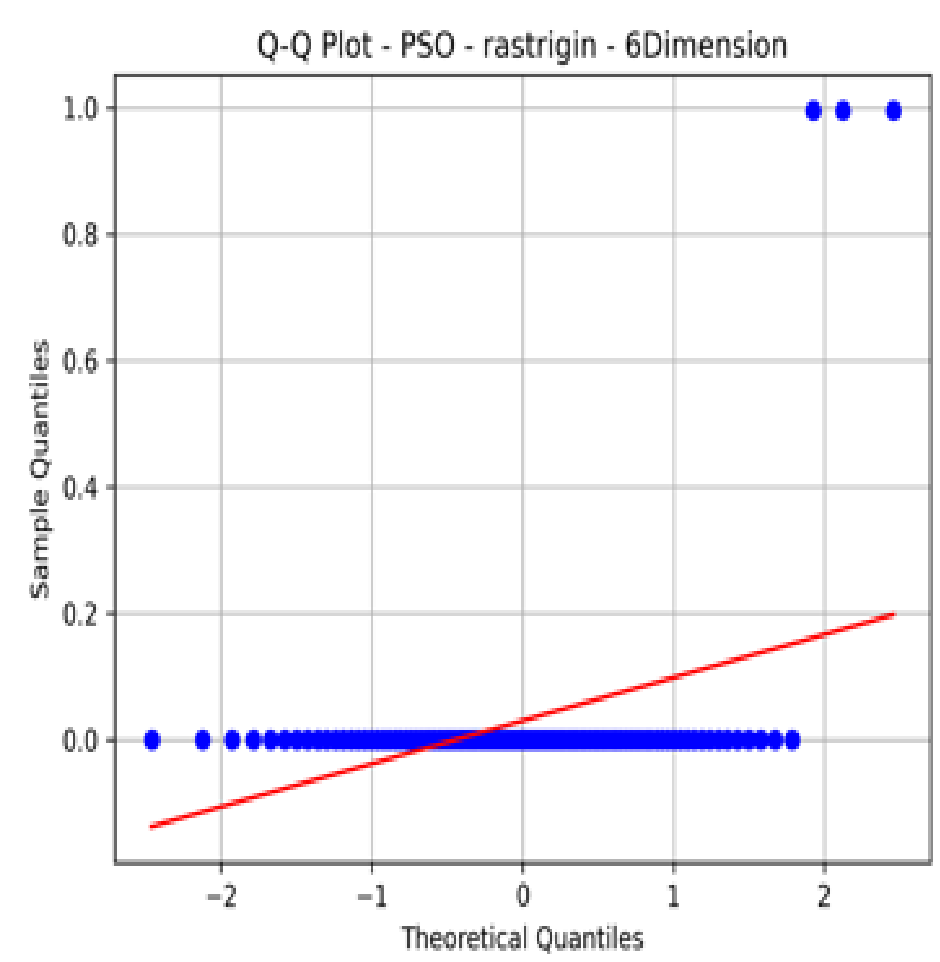
Q-Q Plot - PSO - rastrigin - 6Dimension
1.0
0.8
0.6
0.4
0.2
0.0
Sample Quantiles
−2
−1
0
1
2
Theoretical Quantiles

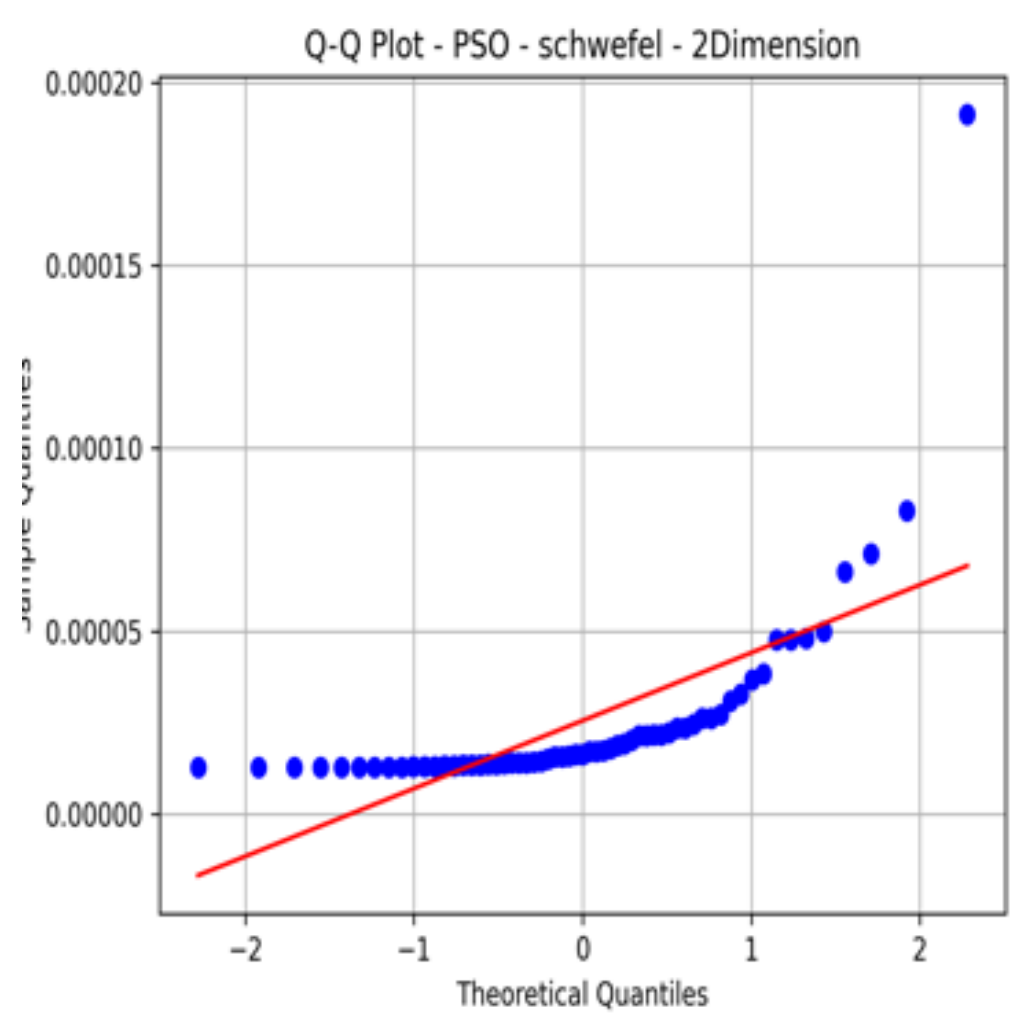
Q-Q Plot - PSO - schwefel - 2Dimension
0.00020
0.00015
0.00010
0.00005
0.00000
−2
−1
0
1
2
Theoretical Quantiles

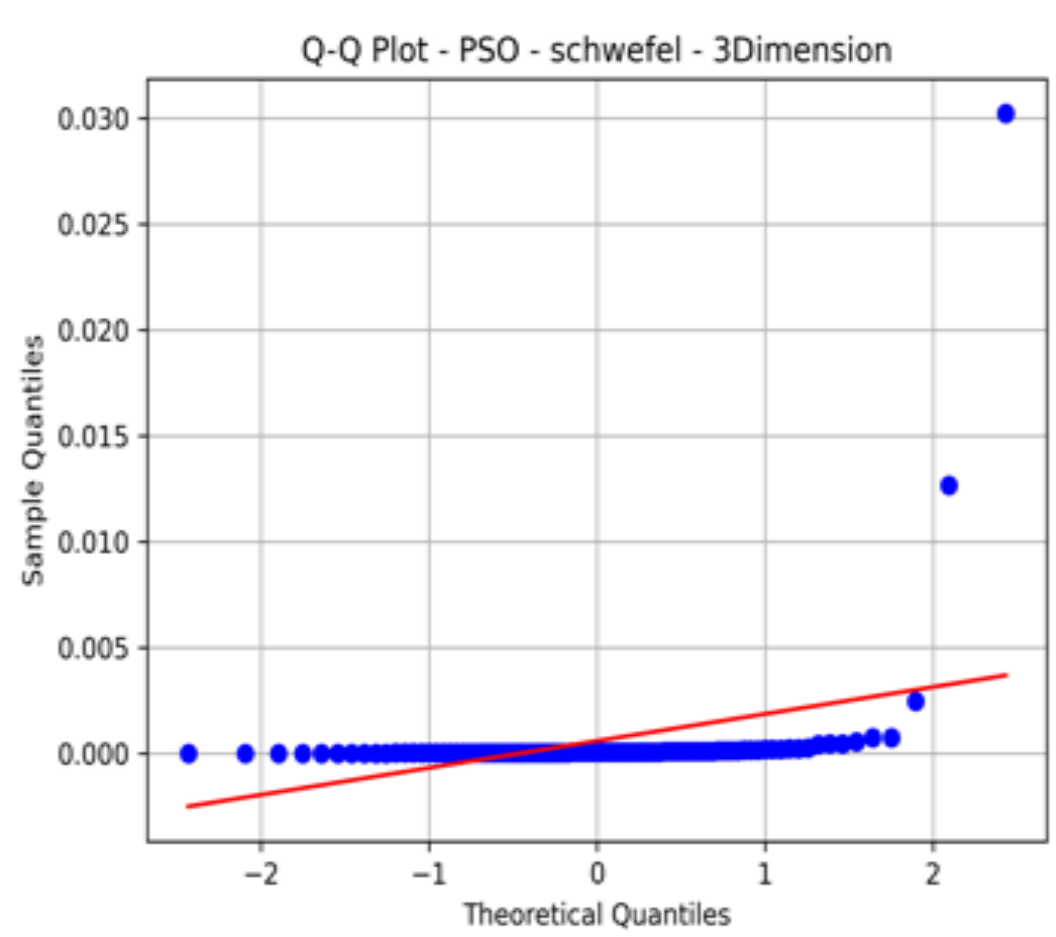
Q-Q Plot - PSO - schwefel - 3Dimension
0.030
0.025
0.020
0.015
0.010
0.005
0.000
Sample Quantiles
−2
−1
0
1
2
Theoretical Quantiles

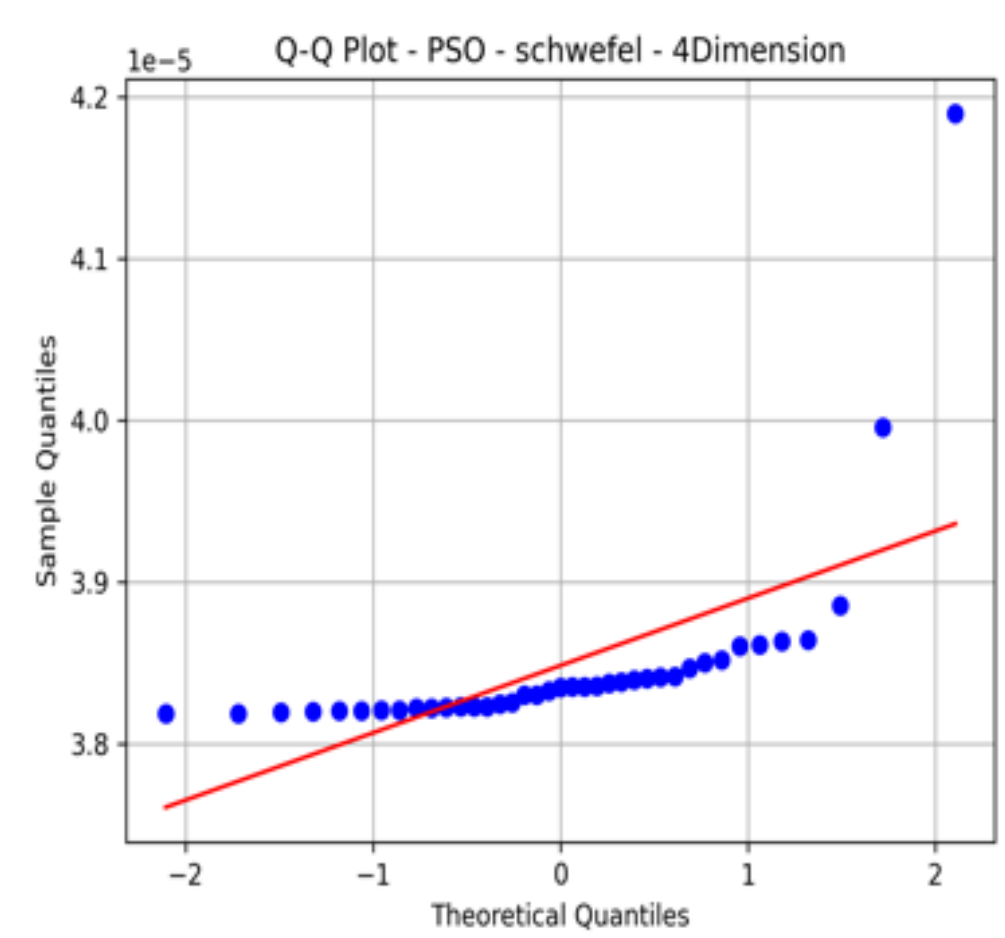
1e−5
Q-Q Plot - PSO - schwefel - 4Dimension
4.2
4.1
4.0
3.9
3.8
Sample Quantiles
−2
−1
0
1
2
Theoretical Quantiles

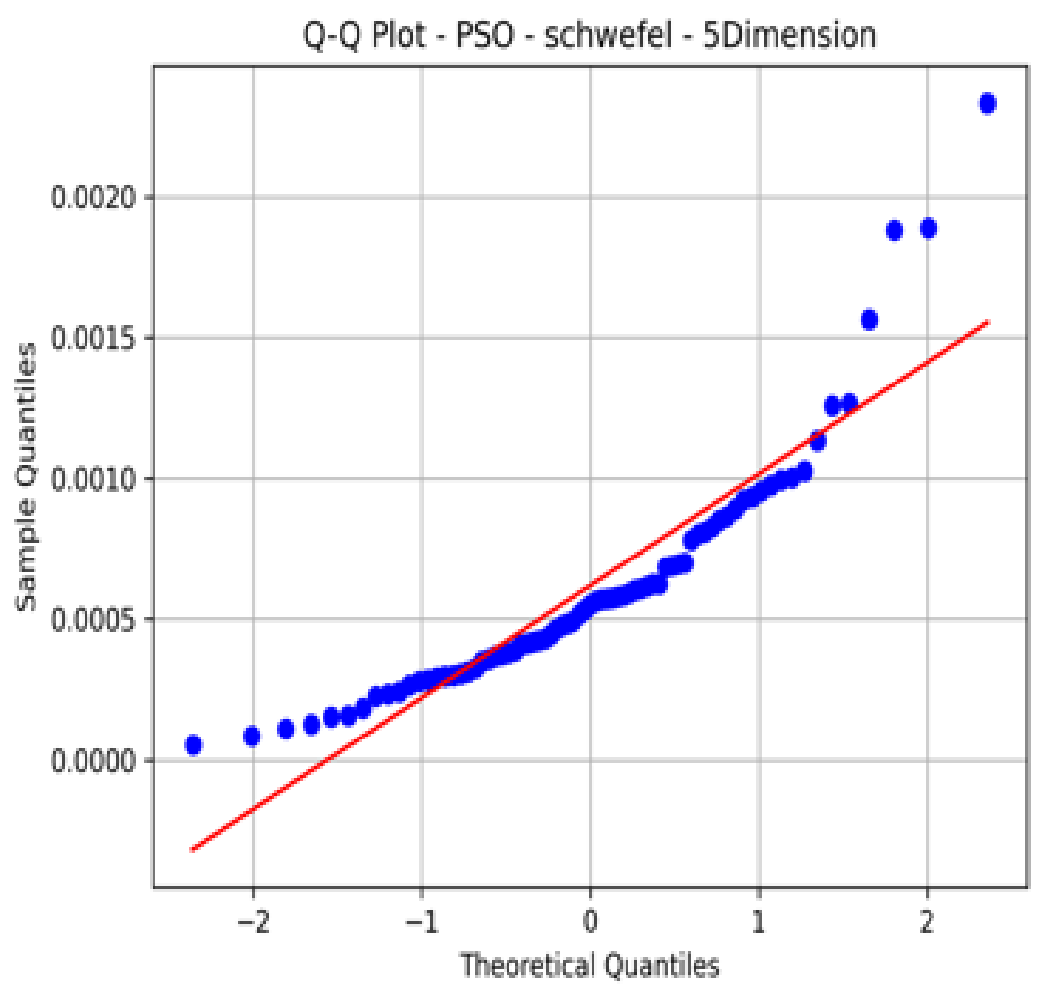
Q-Q Plot - PSO - schwefel - 5Dimension
Sample Quantiles
0.0020
0.0015
0.0010
0.0005
0.0000
−2
−1
0
1
2
Theoretical Quantiles

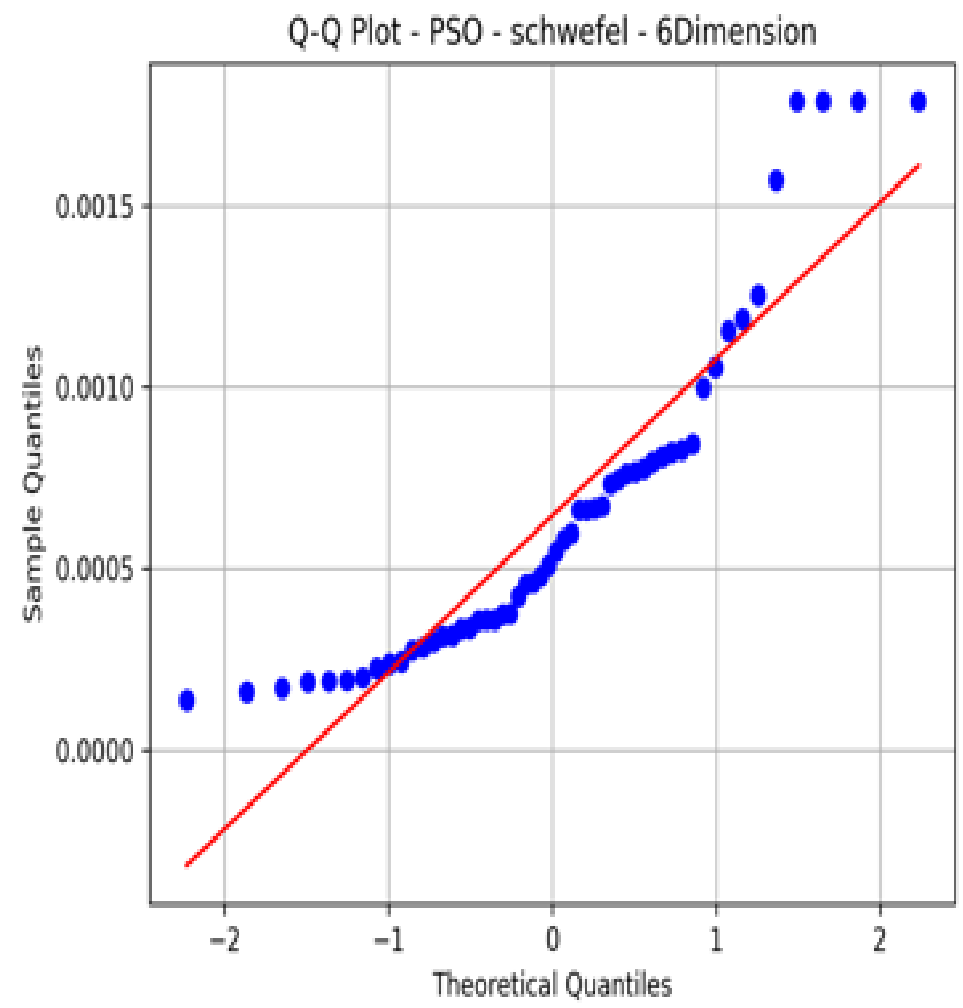
Q-Q Plot - PSO - schwefel - 6Dimension
Sample Quantiles
0.0015
0.0010
0.0005
0.0000
−2
−1
0
1
2
Theoretical Quantiles

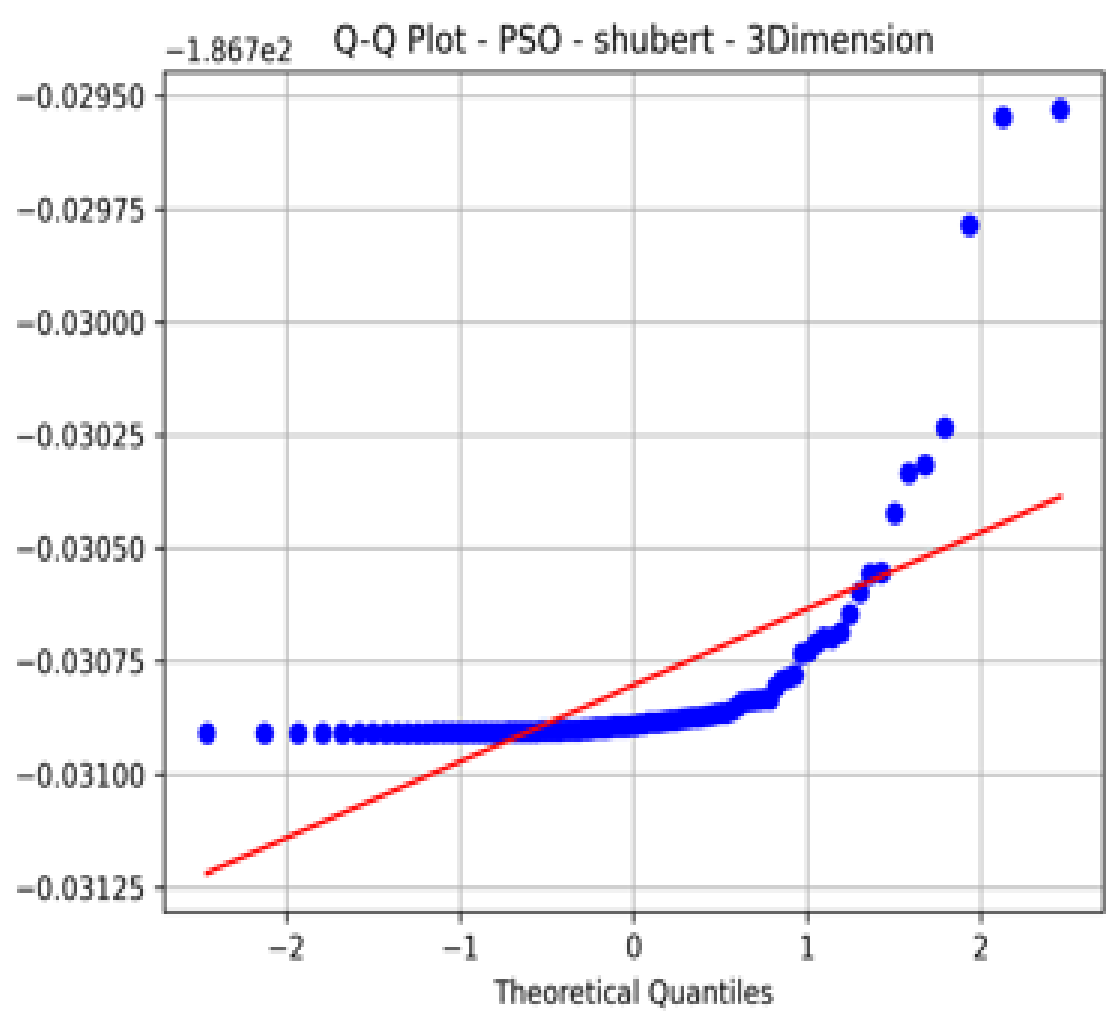
−1.867e2
Q-Q Plot - PSO - shubert - 3Dimension
−0.02950
−0.02975
−0.03000
−0.03025
−0.03050
−0.03075
−0.03100
−0.03125
−2
−1
0
1
2
Theoretical Quantiles

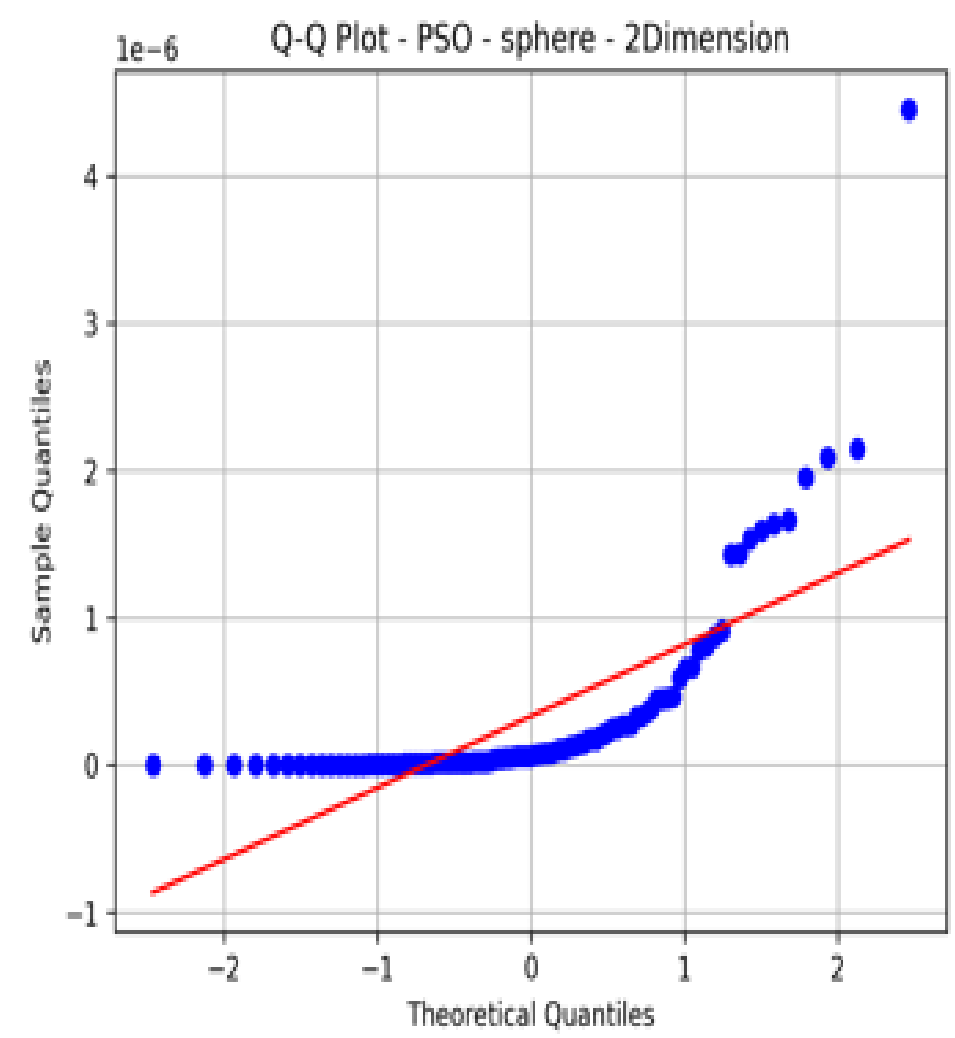
1e−6
Q-Q Plot - PSO - sphere - 2Dimension
Sample Quantiles
4
3
2
1
0
−1
−2
−1
0
1
2
Theoretical Quantiles

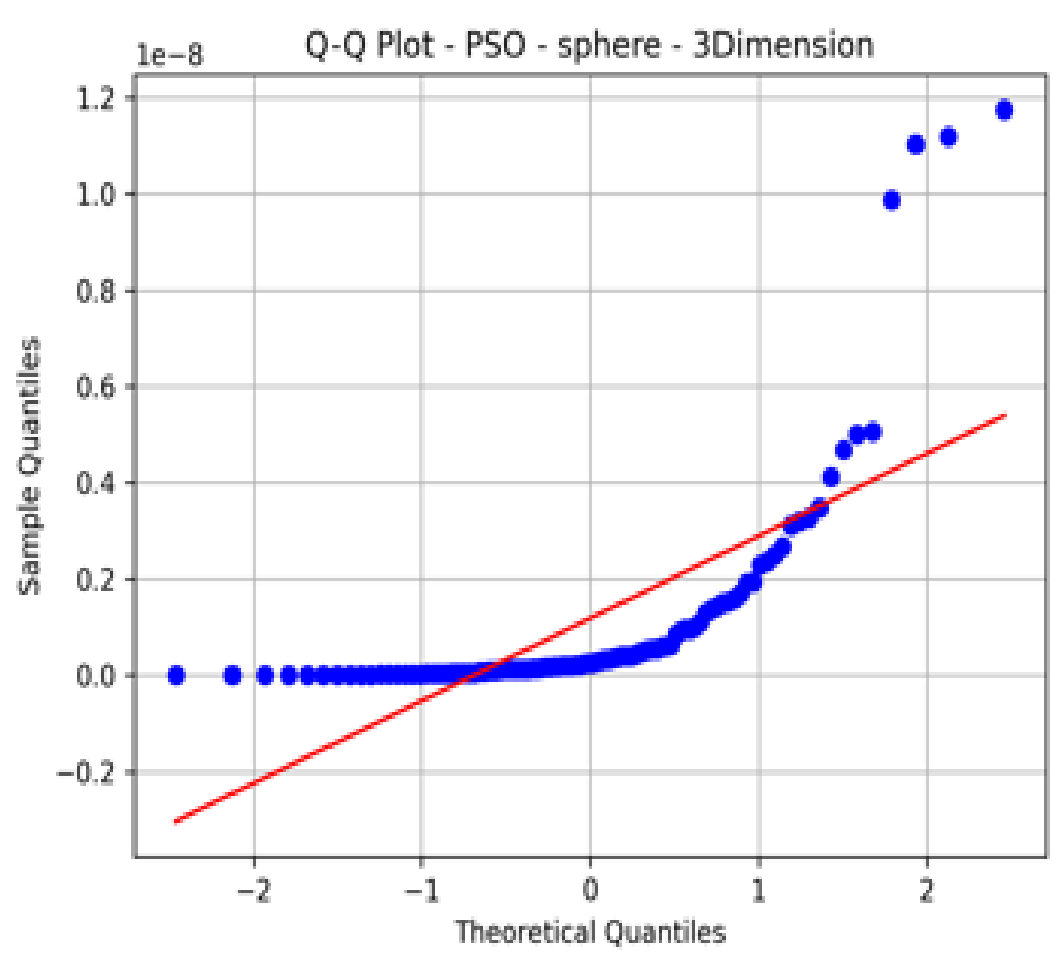
1e−8
Q-Q Plot - PSO - sphere - 3Dimension
Sample Quantiles
1.2
1.0
0.8
0.6
0.4
0.2
0.0
−0.2
−2
−1
0
1
2
Theoretical Quantiles

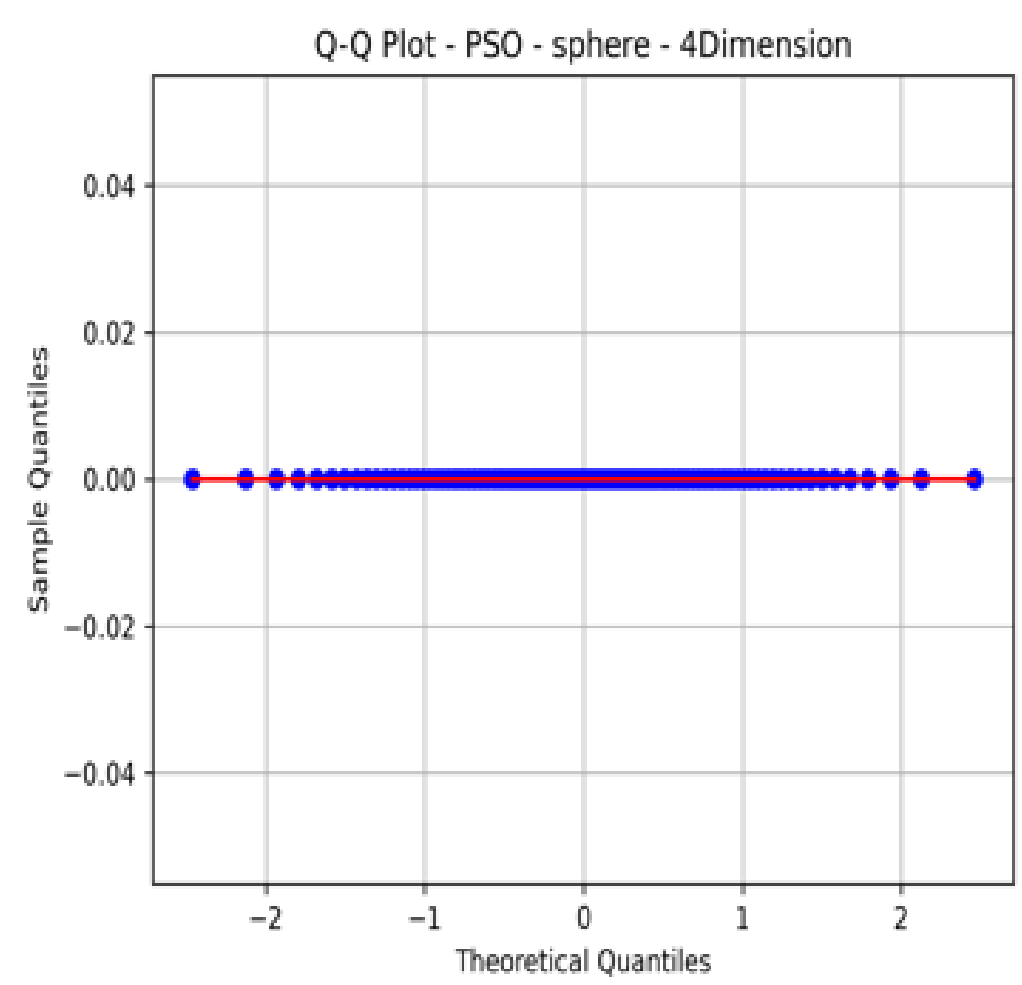
Q-Q Plot - PSO - sphere - 4Dimension
Sample Quantiles
0.04
0.02
0.00
−0.02
−0.04
−2
−1
0
1
2
Theoretical Quantiles

Non-Continuous functions QQ plots

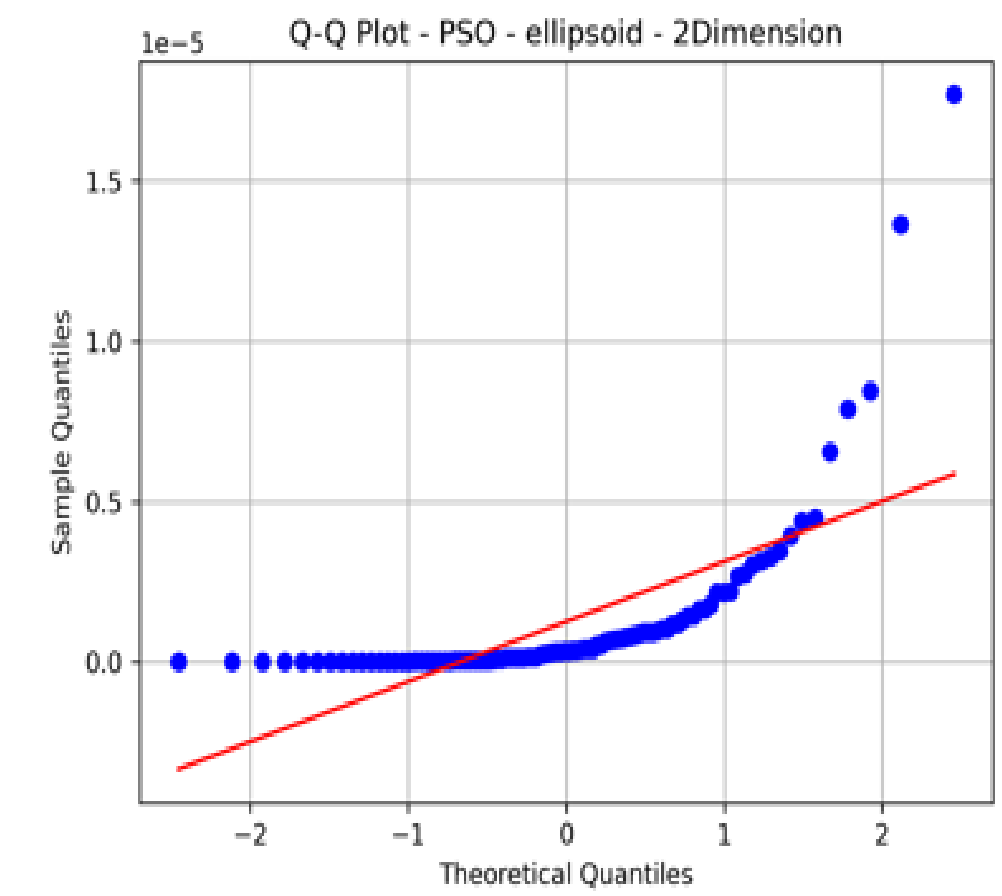

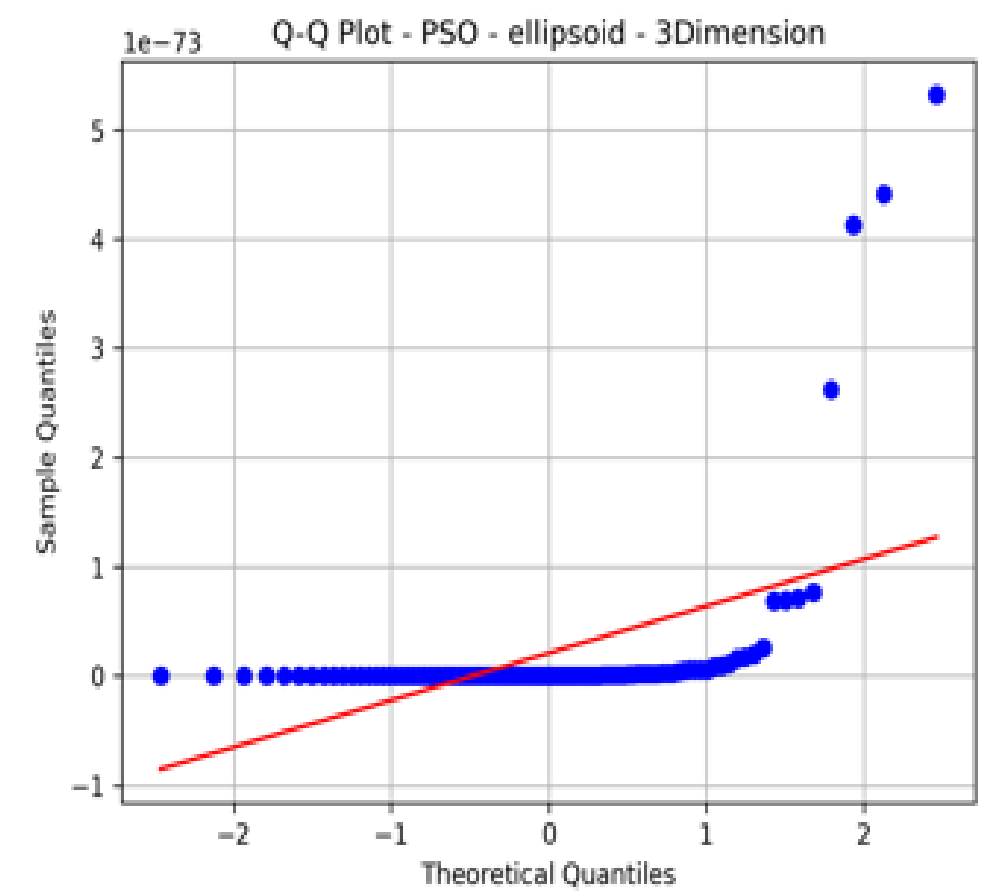

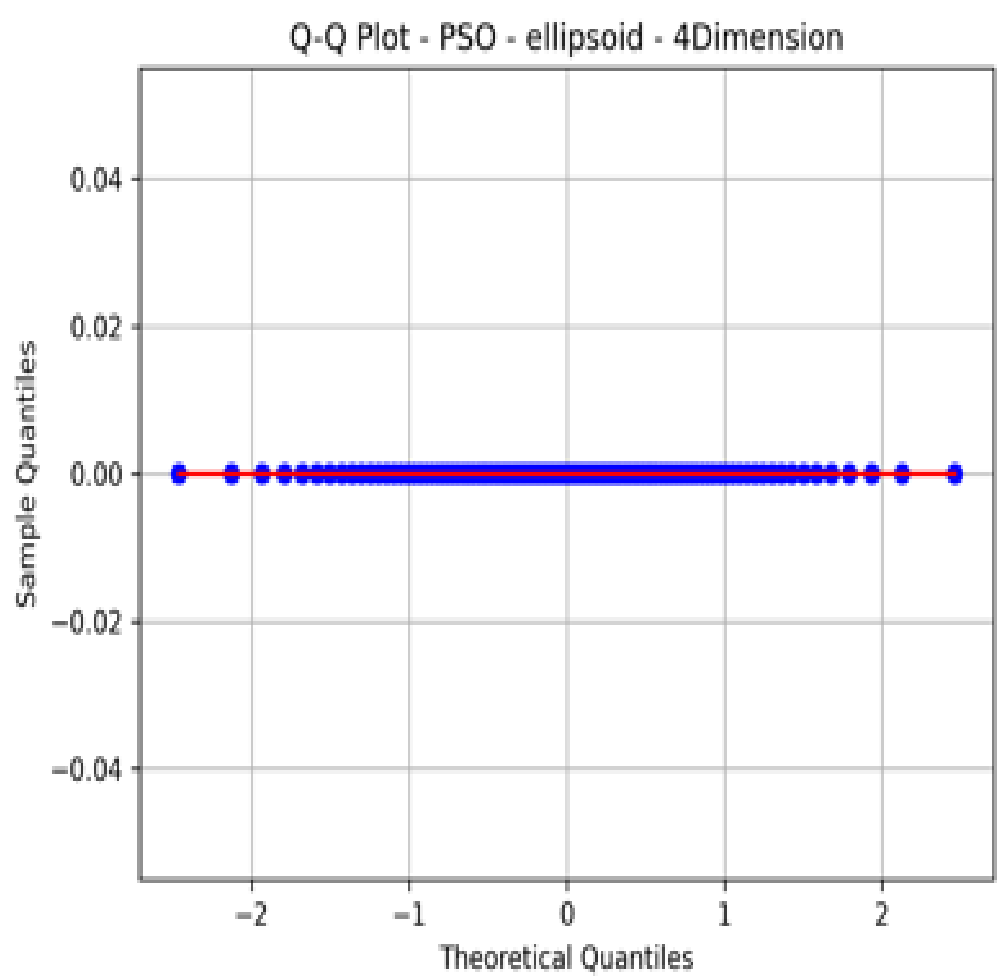

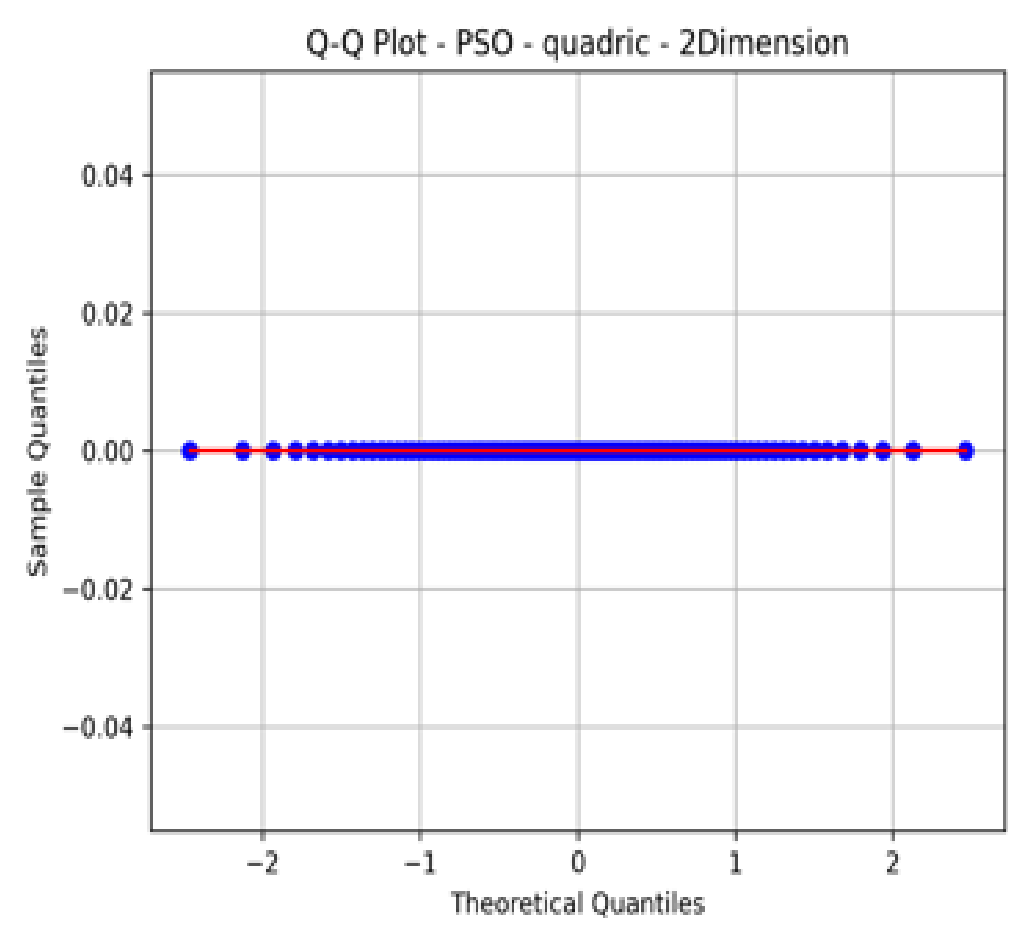

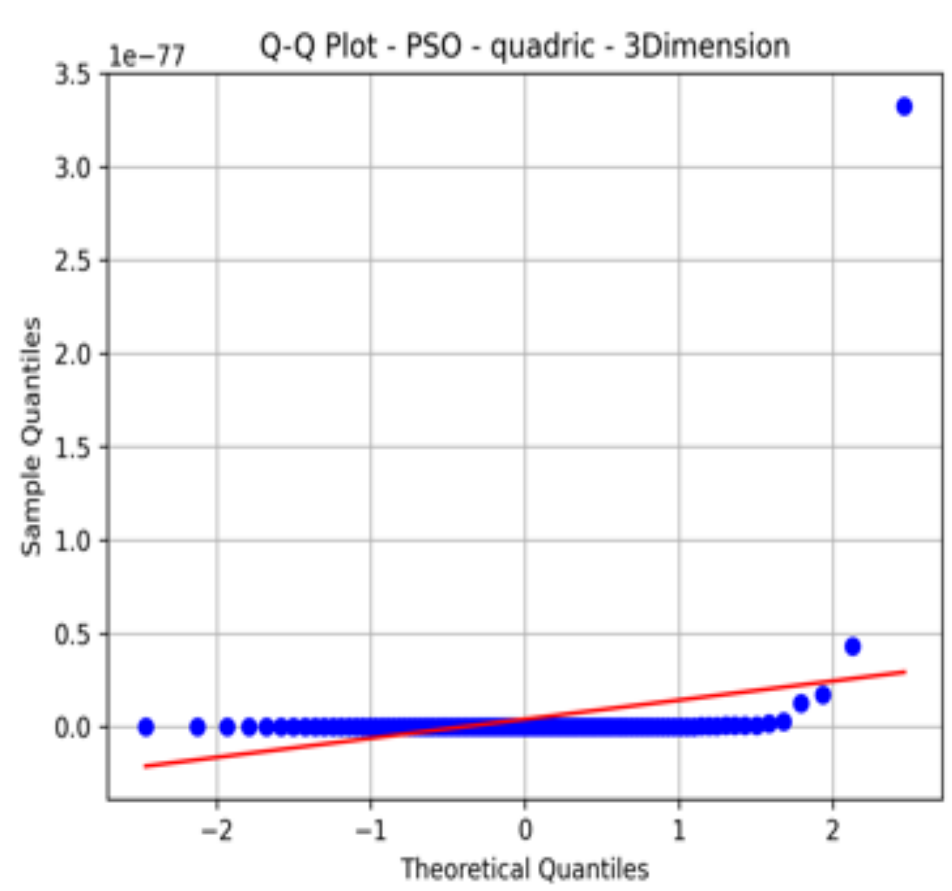

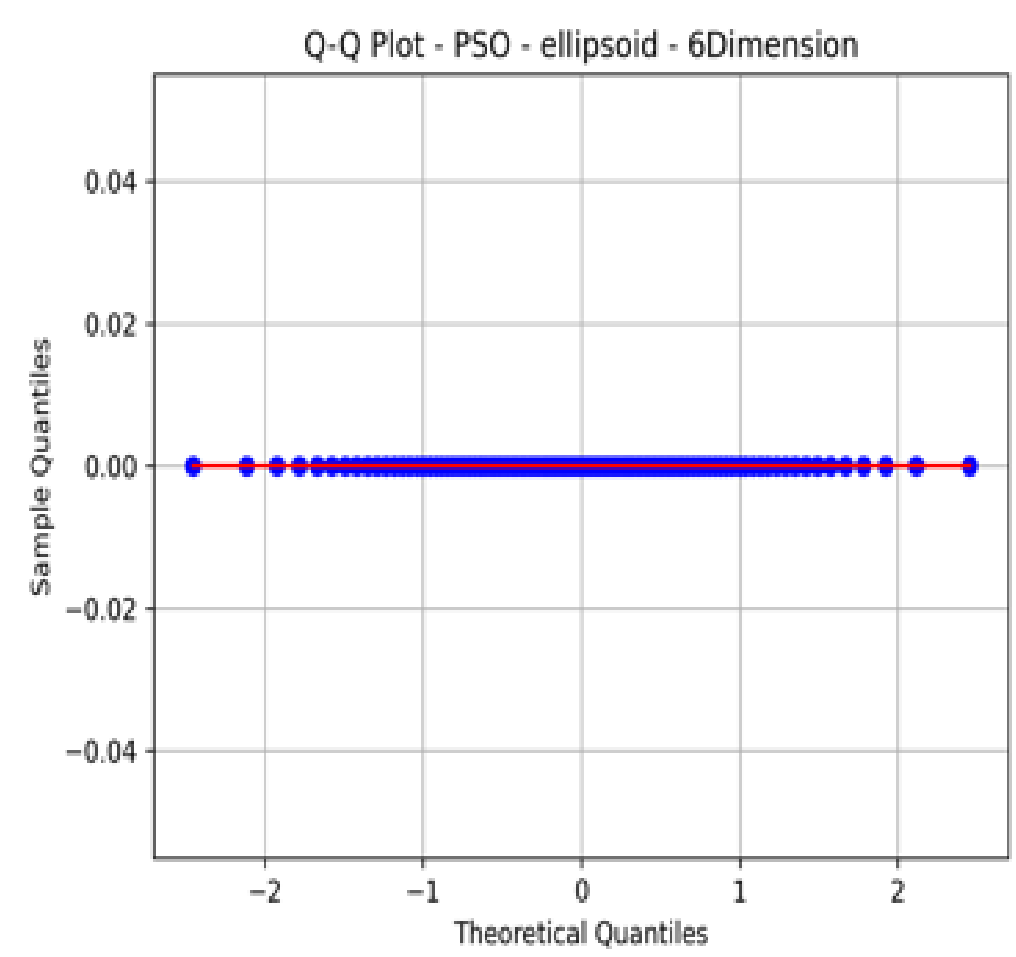

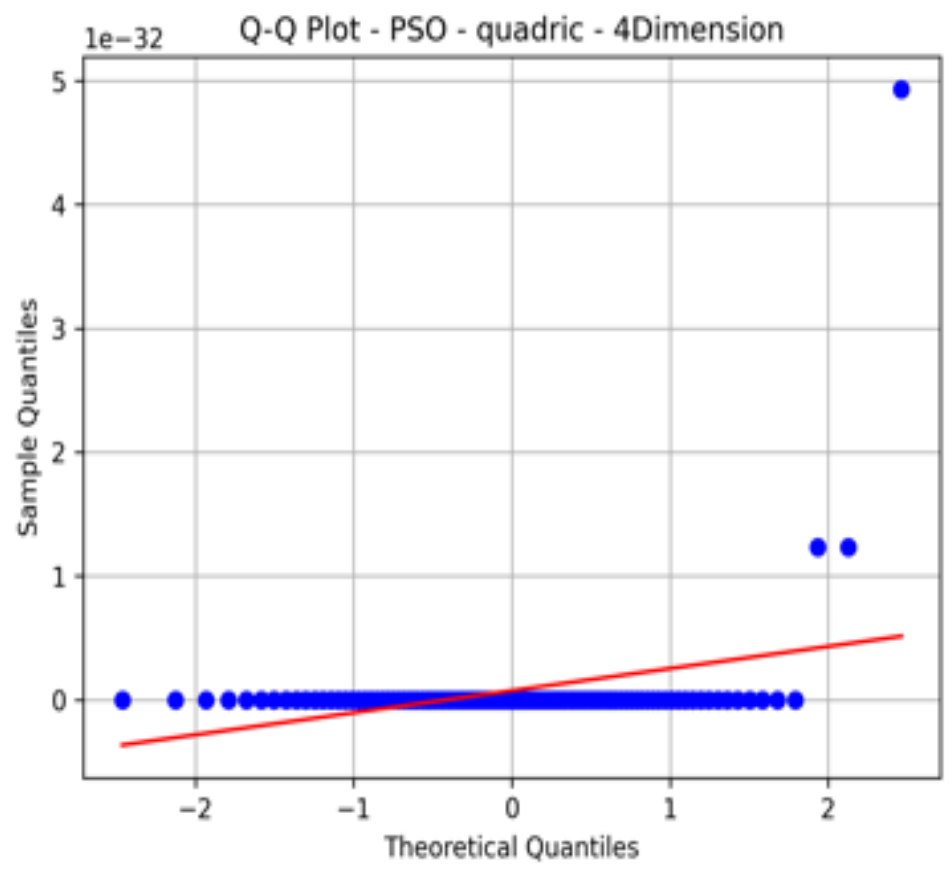
Q-Q Plot - PSO - quadric - 4Dimension
1e−32
Sample Quantiles
Theoretical Quantiles

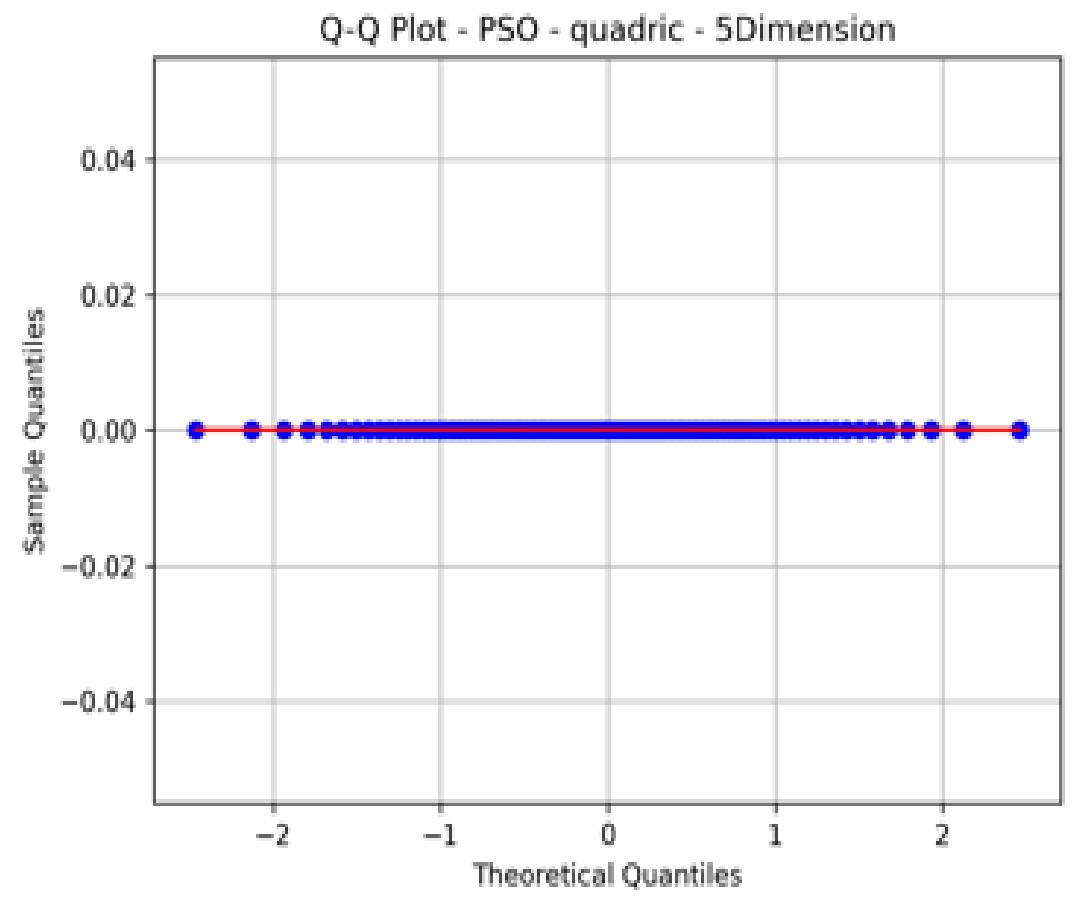
Q-Q Plot - PSO - quadric - 5Dimension
Sample Quantiles
Theoretical Quantiles

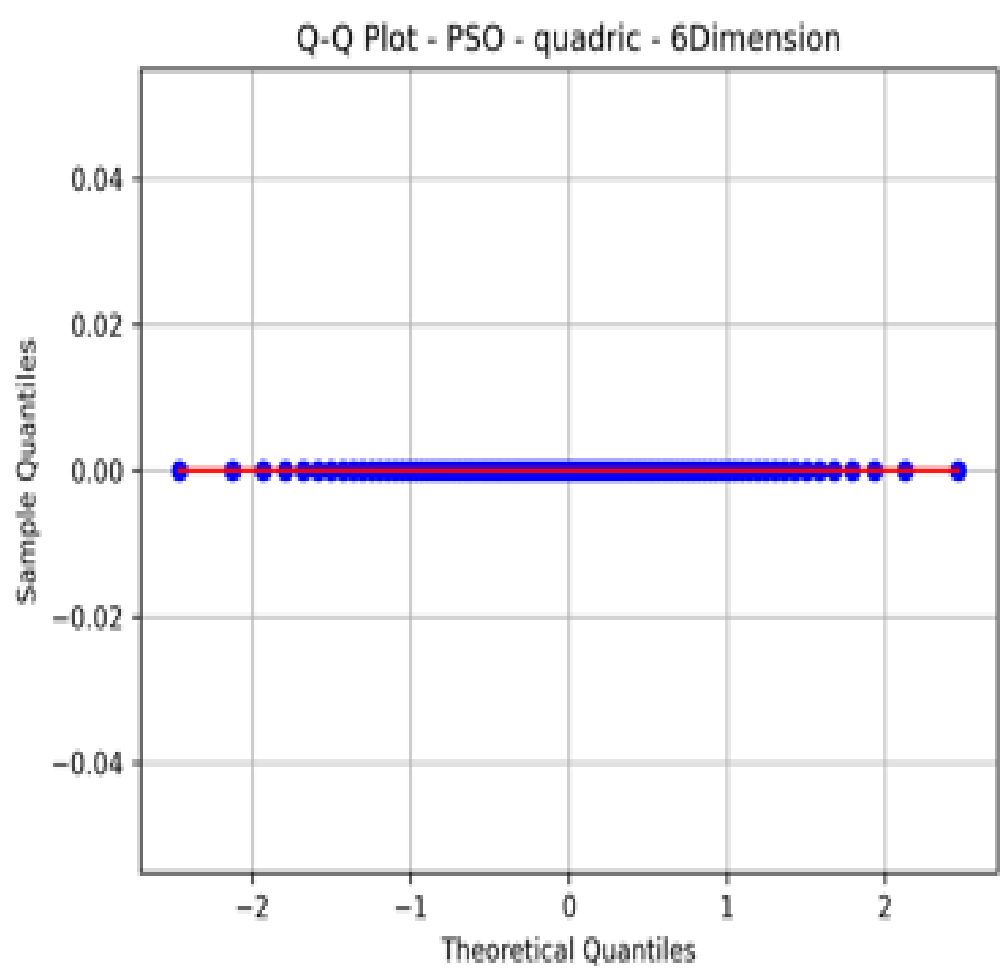
Q-Q Plot - PSO - quadric - 6Dimension
Sample Quantiles
Theoretical Quantiles

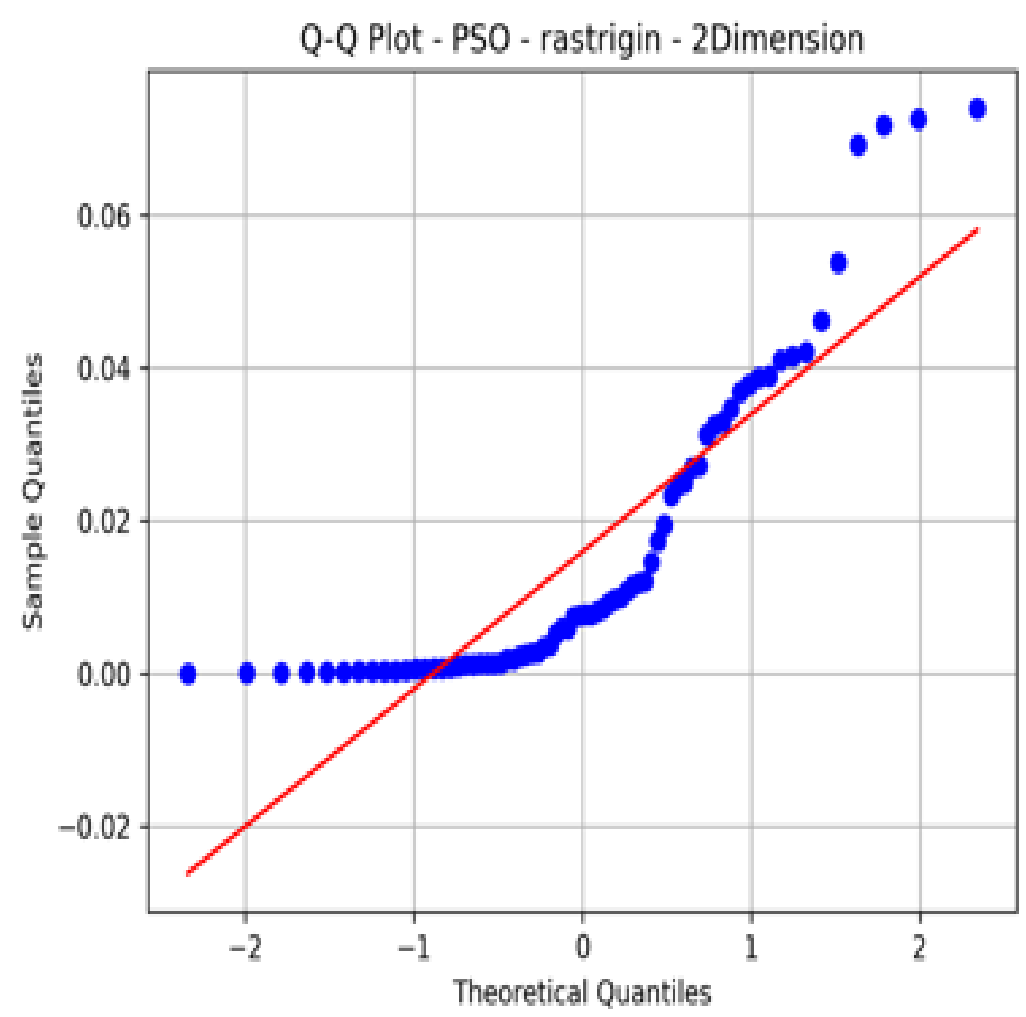
Q-Q Plot - PSO - rastrigin - 2Dimension
Sample Quantiles
Theoretical Quantiles

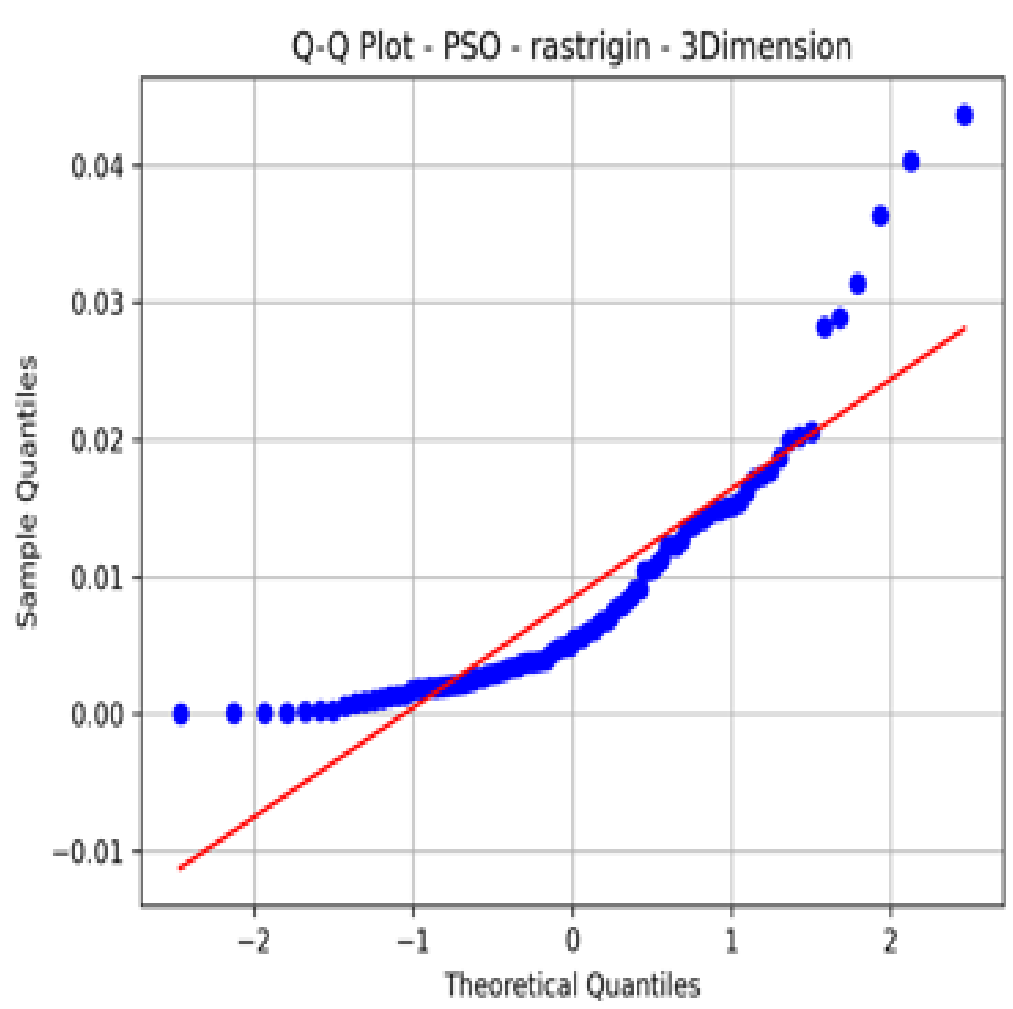
Q-Q Plot - PSO - rastrigin - 3Dimension
Sample Quantiles
Theoretical Quantiles

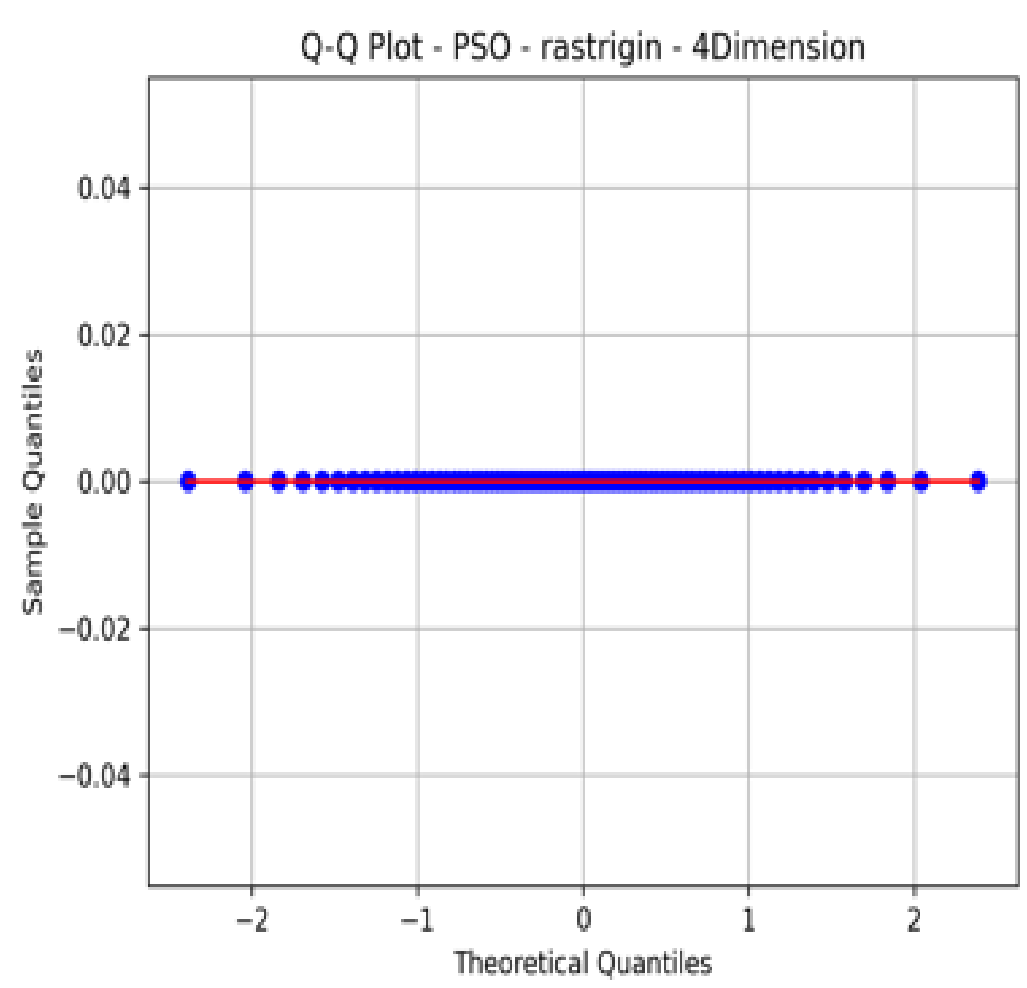
Q-Q Plot - PSO - rastrigin - 4Dimension
Sample Quantiles
Theoretical Quantiles

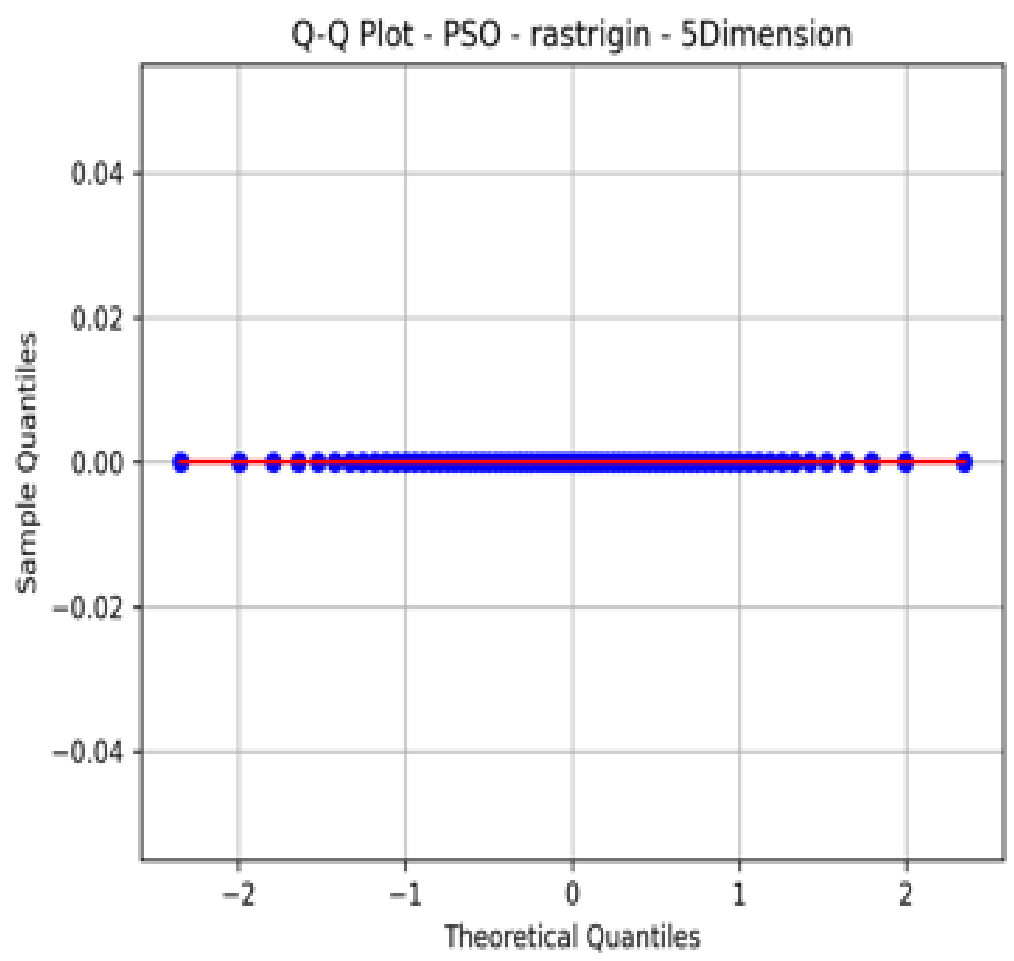
Q-Q Plot - PSO - rastrigin - 5Dimension
Sample Quantiles
Theoretical Quantiles

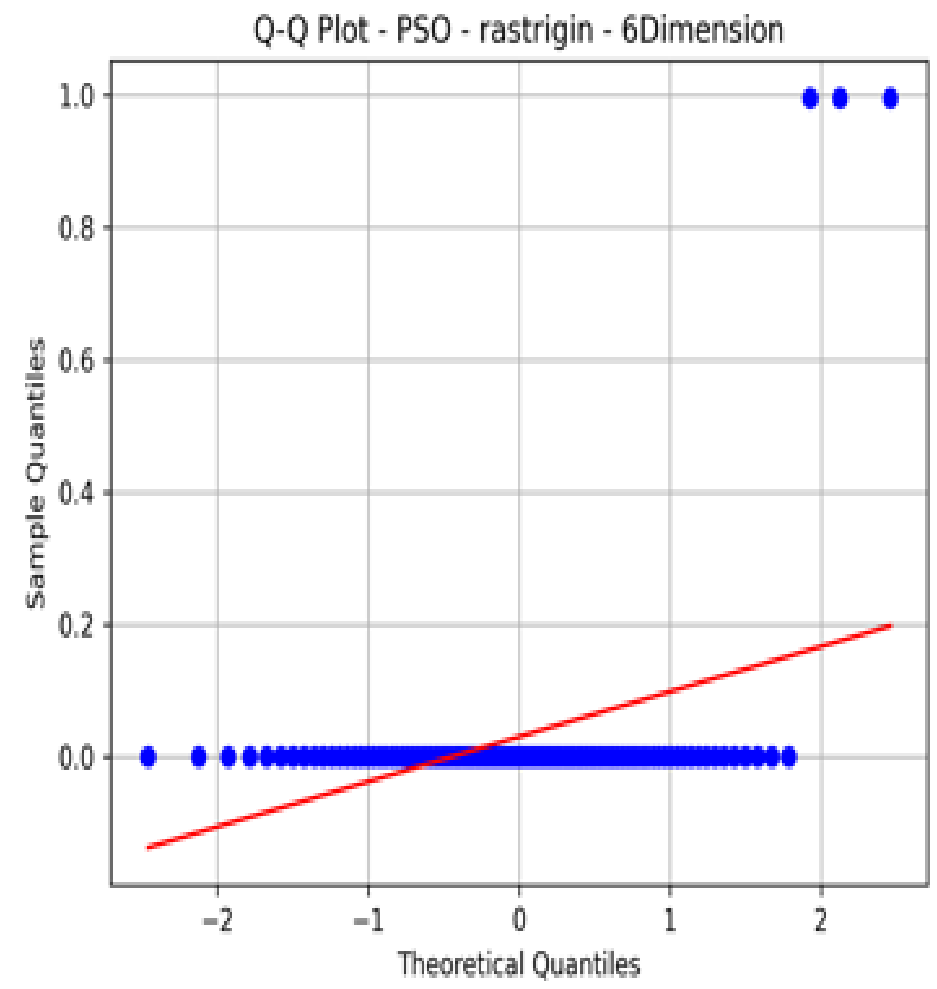
Q-Q Plot - PSO - rastrigin - 6Dimension
Sample Quantiles
Theoretical Quantiles

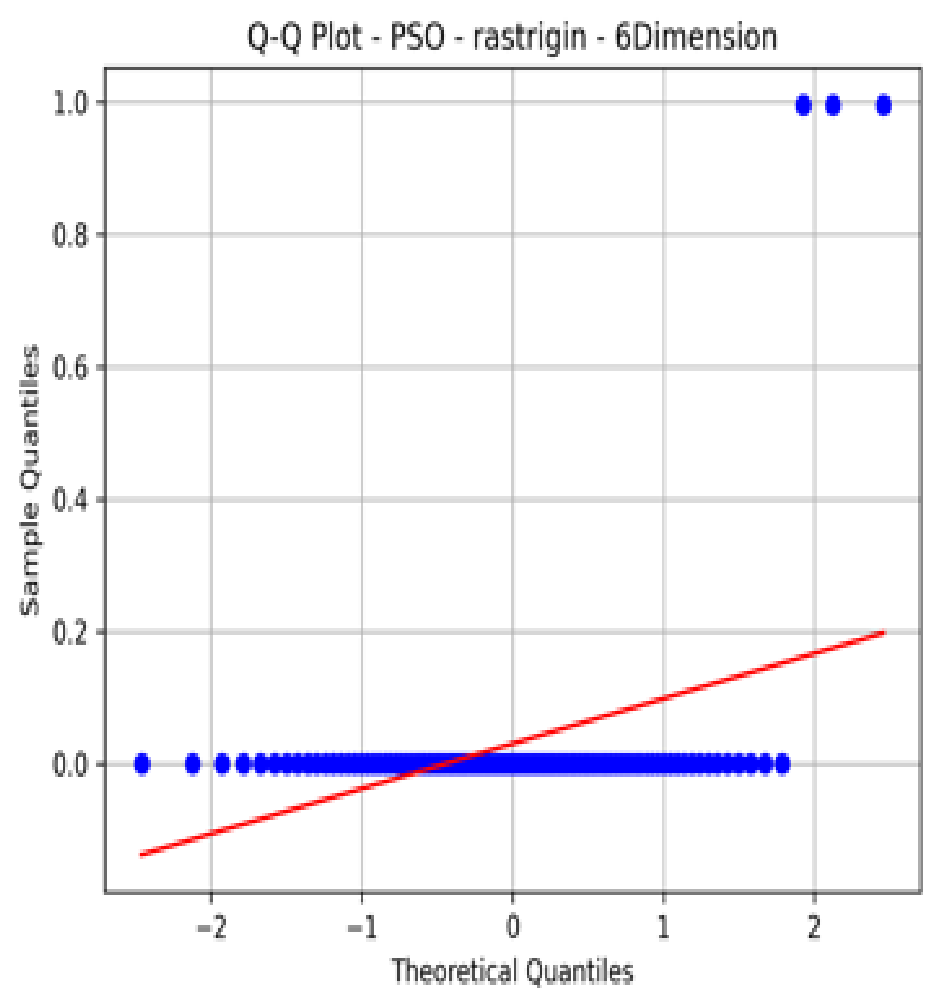
Q-Q Plot - PSO - rastrigin - 6Dimension
Sample Quantiles
Theoretical Quantiles

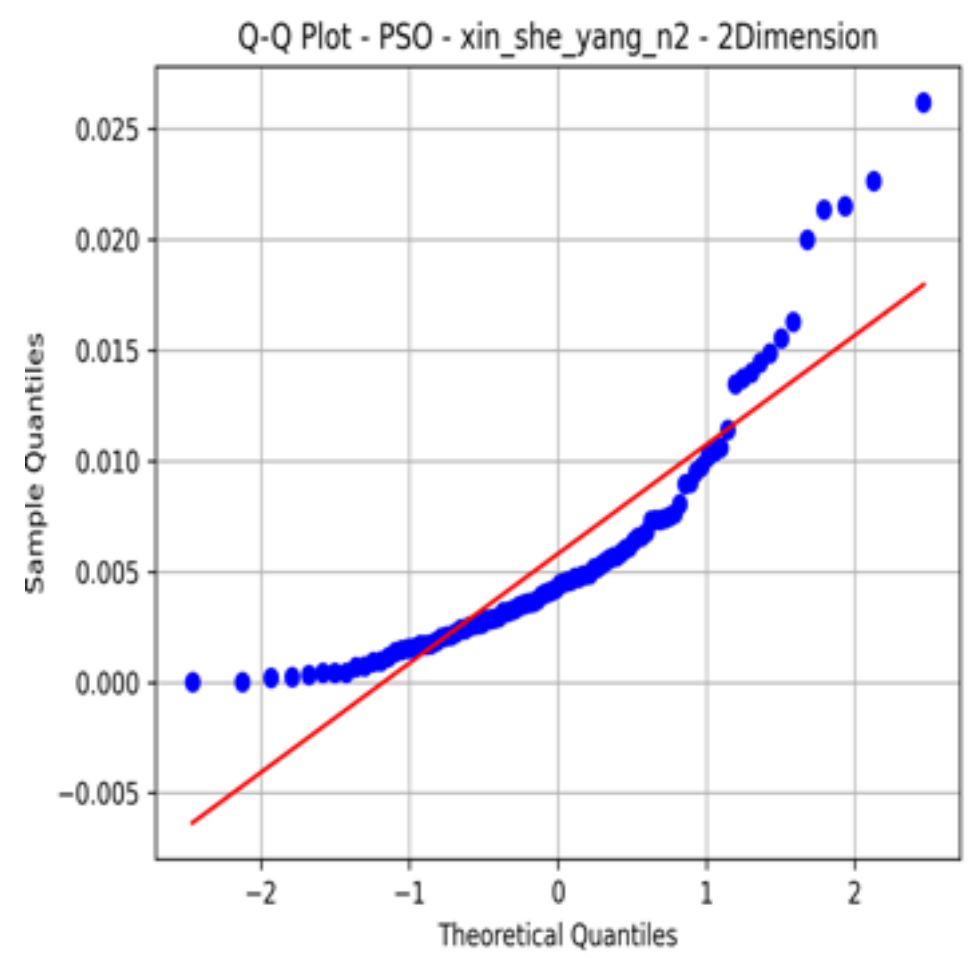
Q-Q Plot - PSO - xin_she_yang_n2 - 2Dimension
Sample Quantiles
Theoretical Quantiles

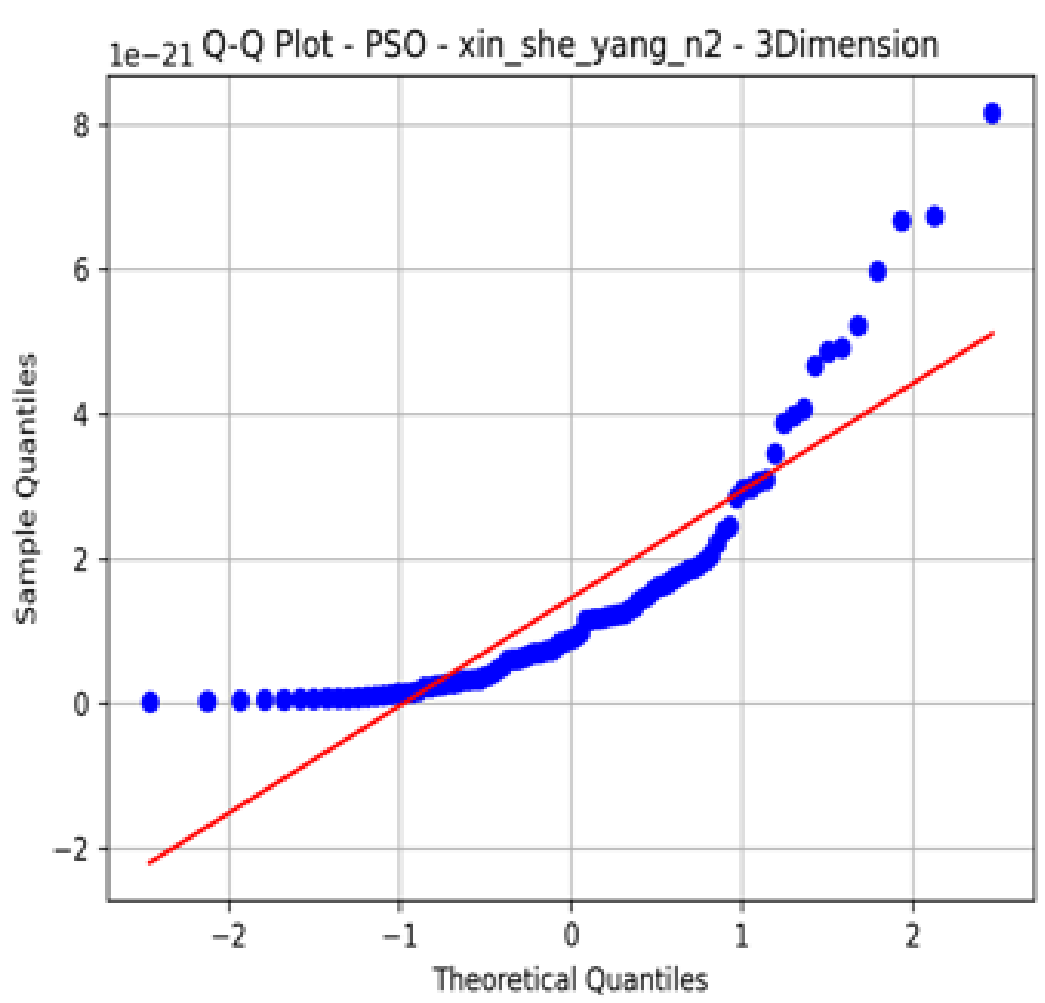
1e−21
Q-Q Plot - PSO - xin_she_yang_n2 - 3Dimension
Sample Quantiles
Theoretical Quantiles

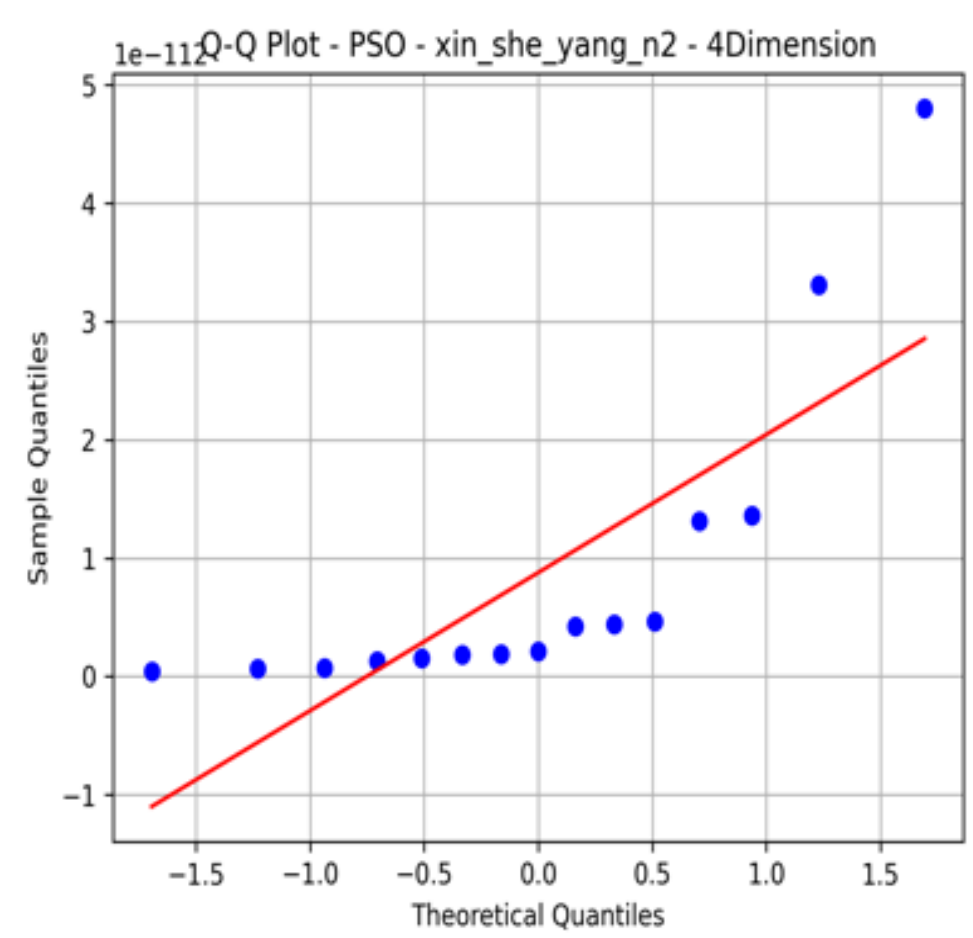
1e−112
Q-Q Plot - PSO - xin_she_yang_n2 - 4Dimension
Sample Quantiles
Theoretical Quantiles

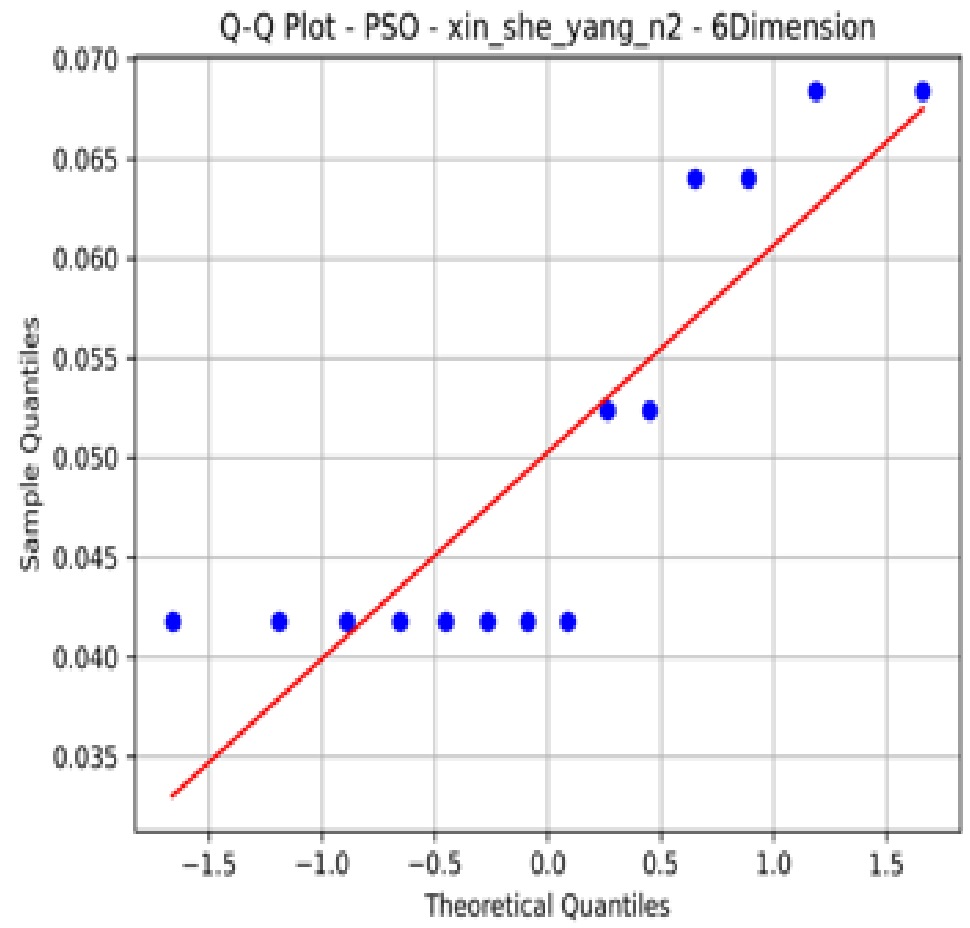

## WO Q-Q plots

### Continuous functions QQ Plots

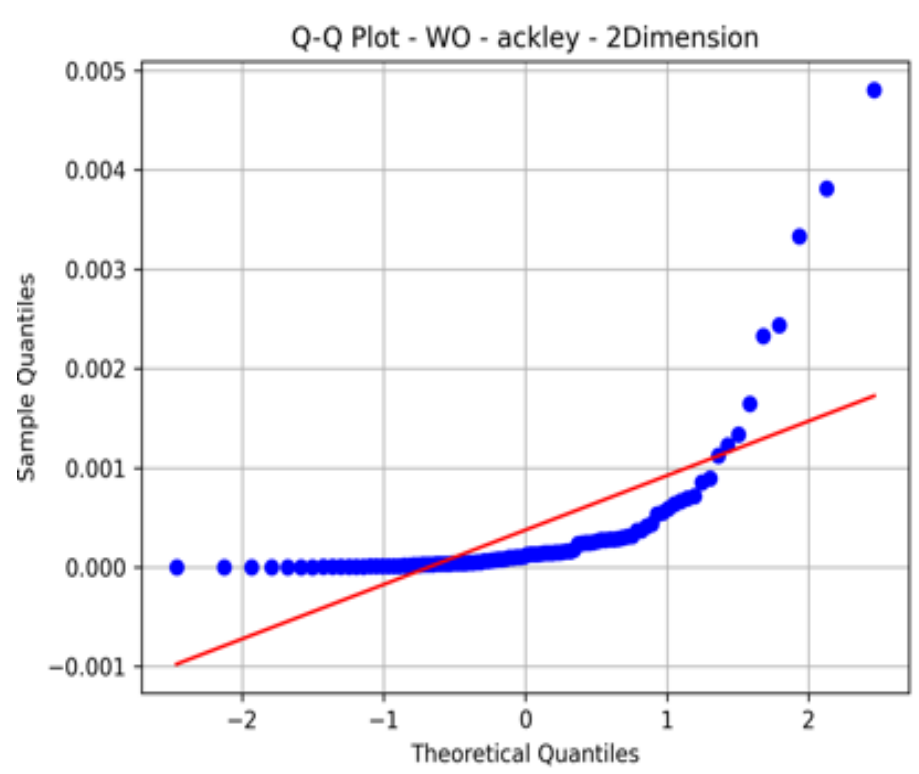

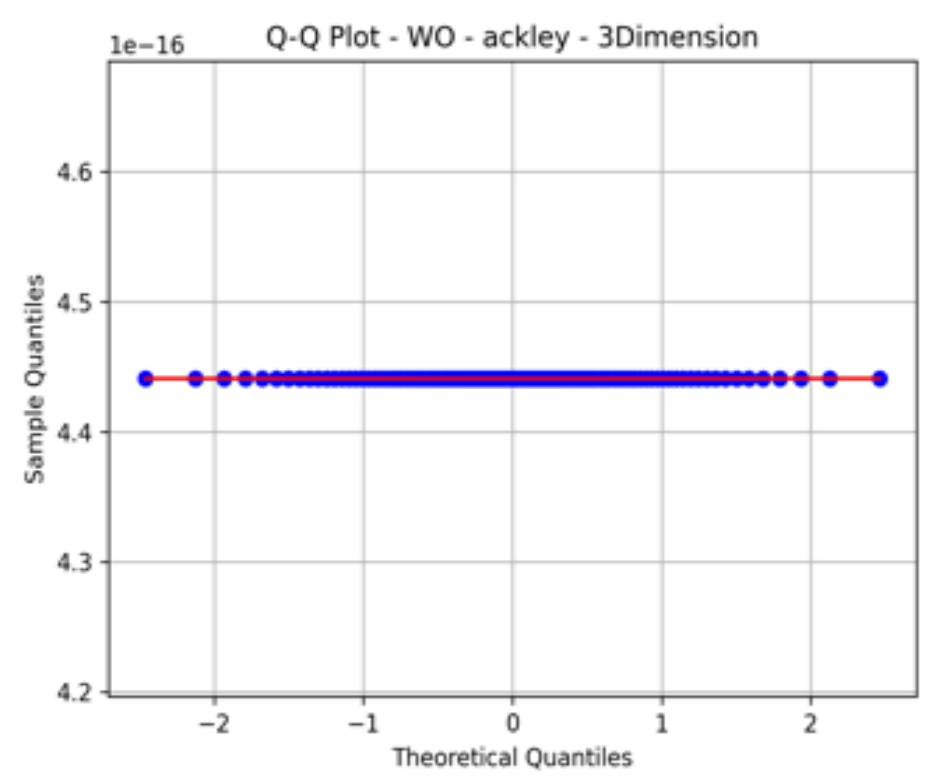

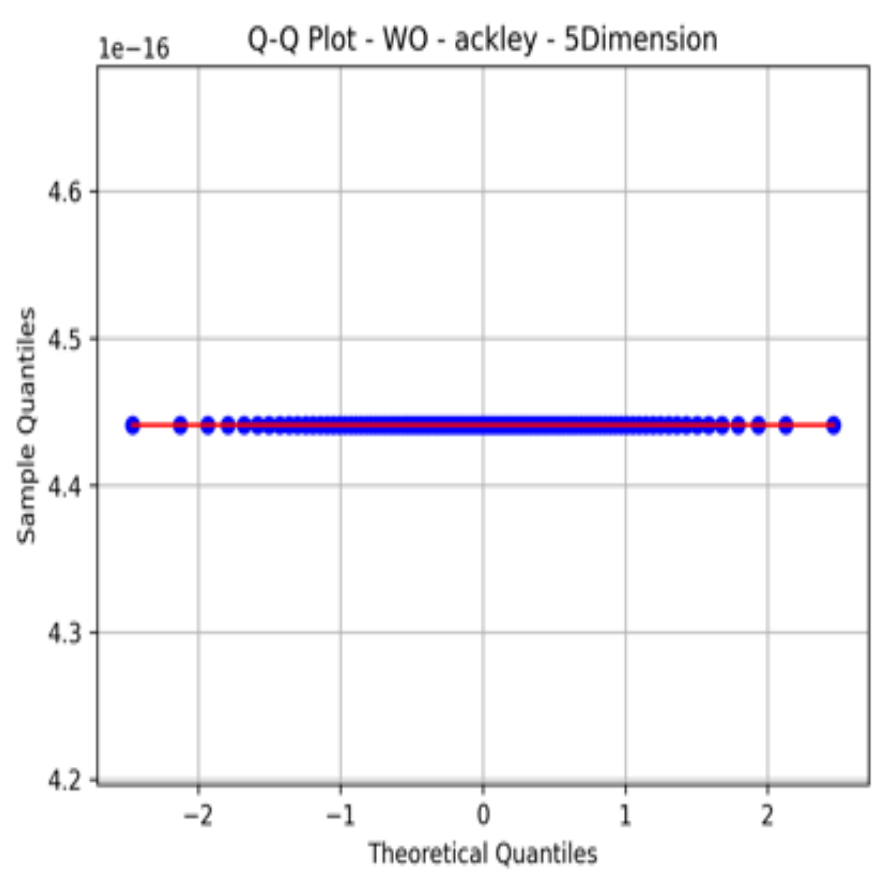

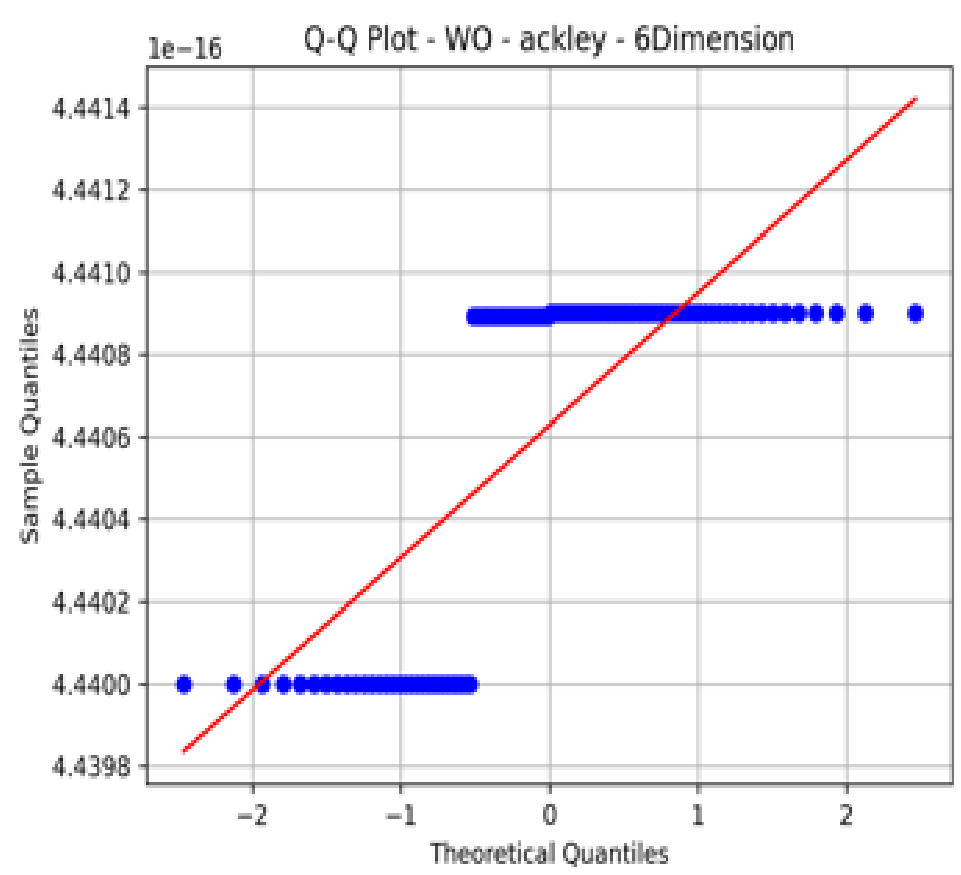

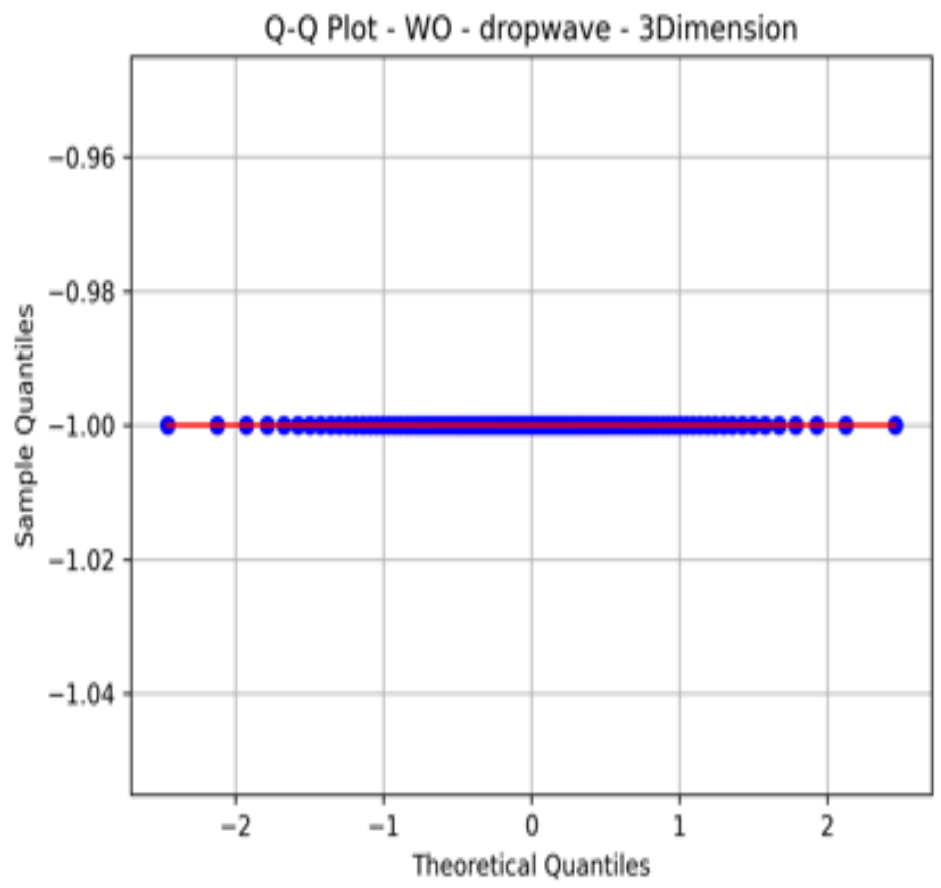
Q-Q Plot - WO - dropwave - 3Dimension
Sample Quantiles
−0.96
−0.98
−1.00
−1.02
−1.04
−2
−1
0
1
2
Theoretical Quantiles

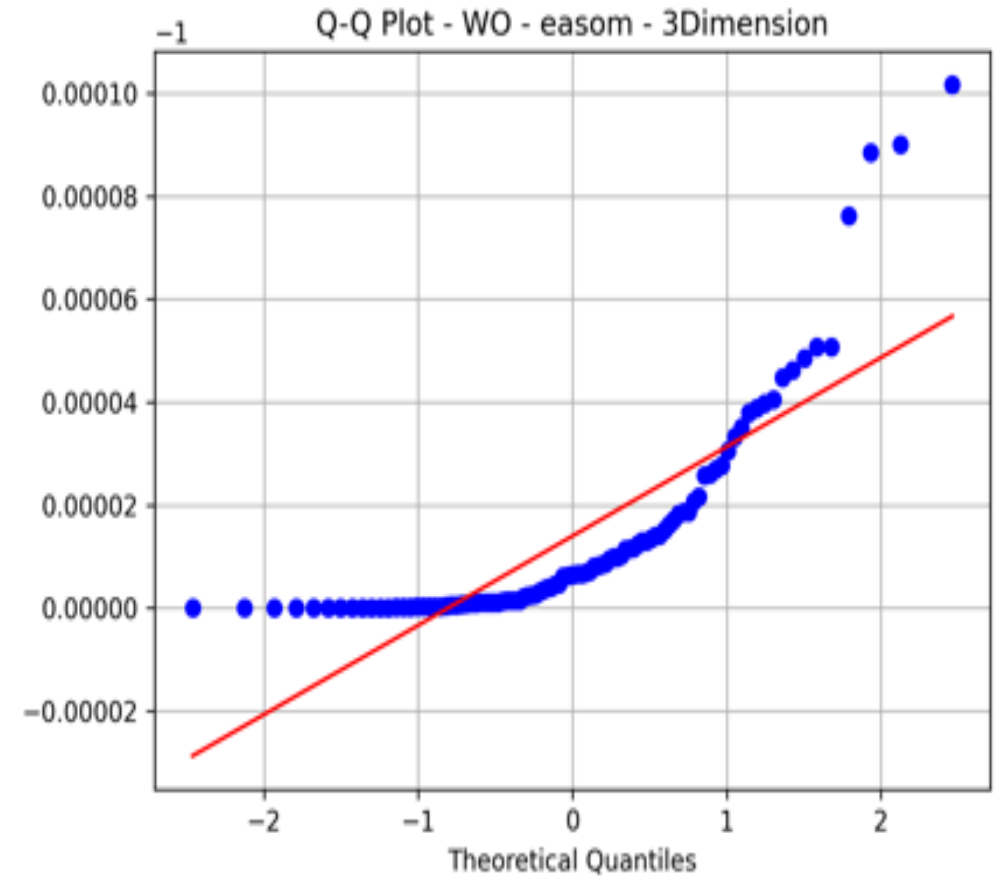
Q-Q Plot - WO - easom - 3Dimension
−1
0.00010
0.00008
0.00006
0.00004
0.00002
0.00000
−0.00002
−2
−1
0
1
2
Theoretical Quantiles

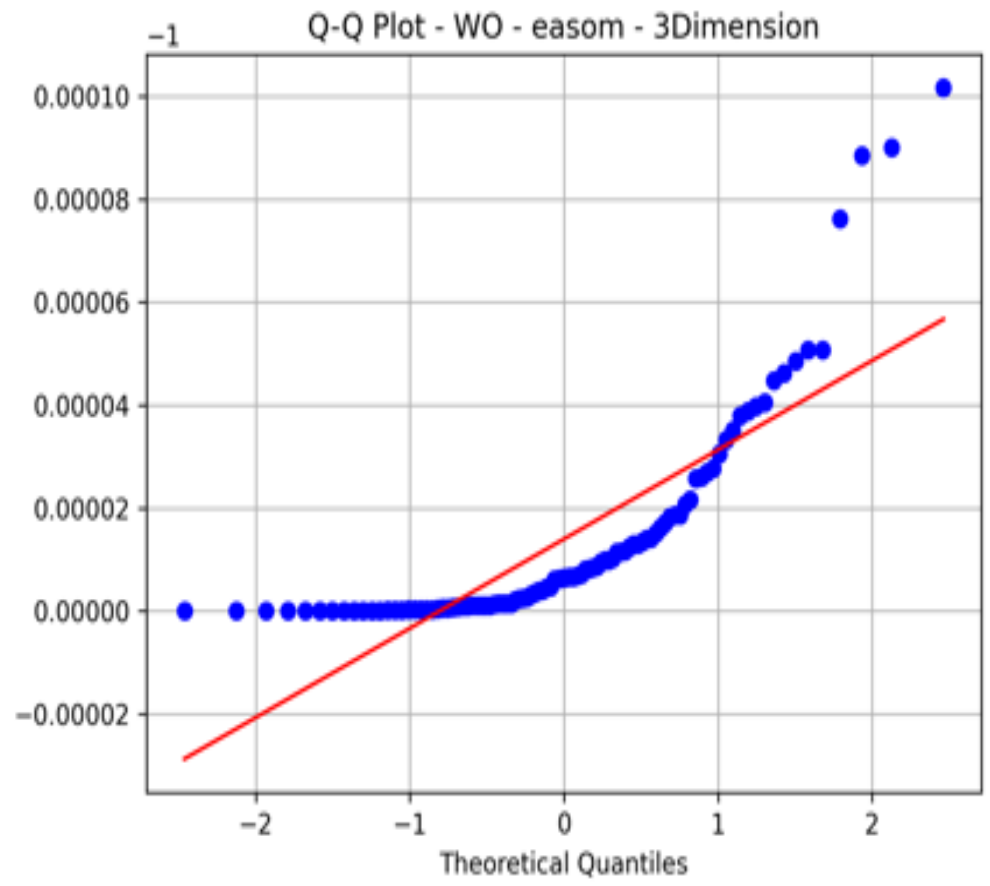
Q-Q Plot - WO - easom - 3Dimension
−1
0.00010
0.00008
0.00006
0.00004
0.00002
0.00000
−0.00002
−2
−1
0
1
2
Theoretical Quantiles

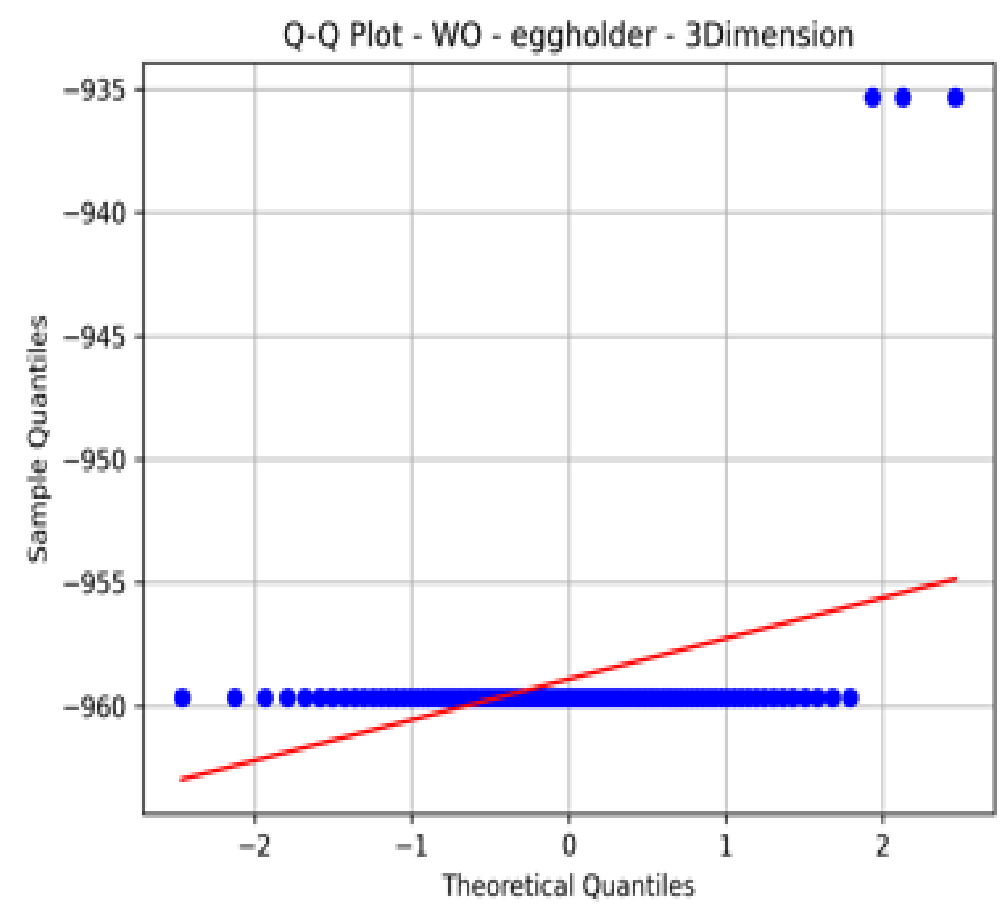
Q-Q Plot - WO - eggholder - 3Dimension
Sample Quantiles
−935
−940
−945
−950
−955
−960
−2
−1
0
1
2
Theoretical Quantiles

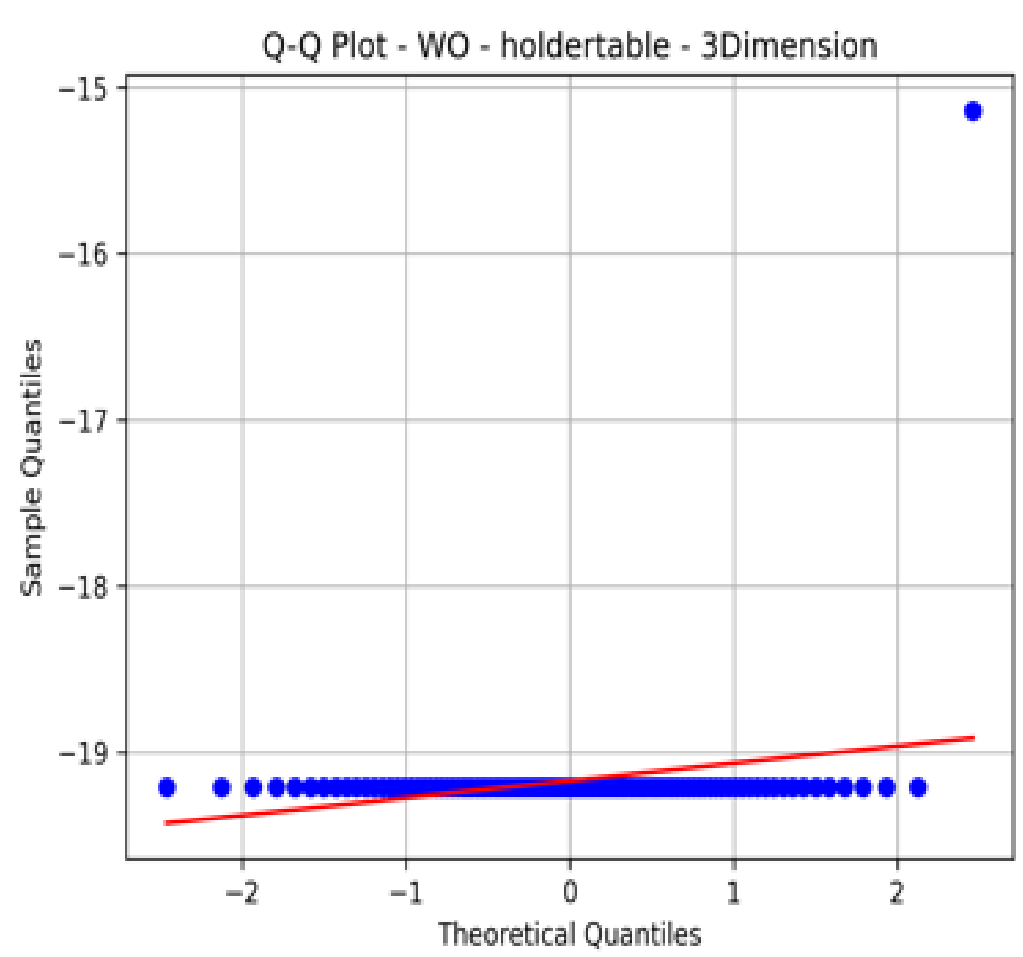
Q-Q Plot - WO - holdertable - 3Dimension
Sample Quantiles
−15
−16
−17
−18
−19
−2
−1
0
1
2
Theoretical Quantiles

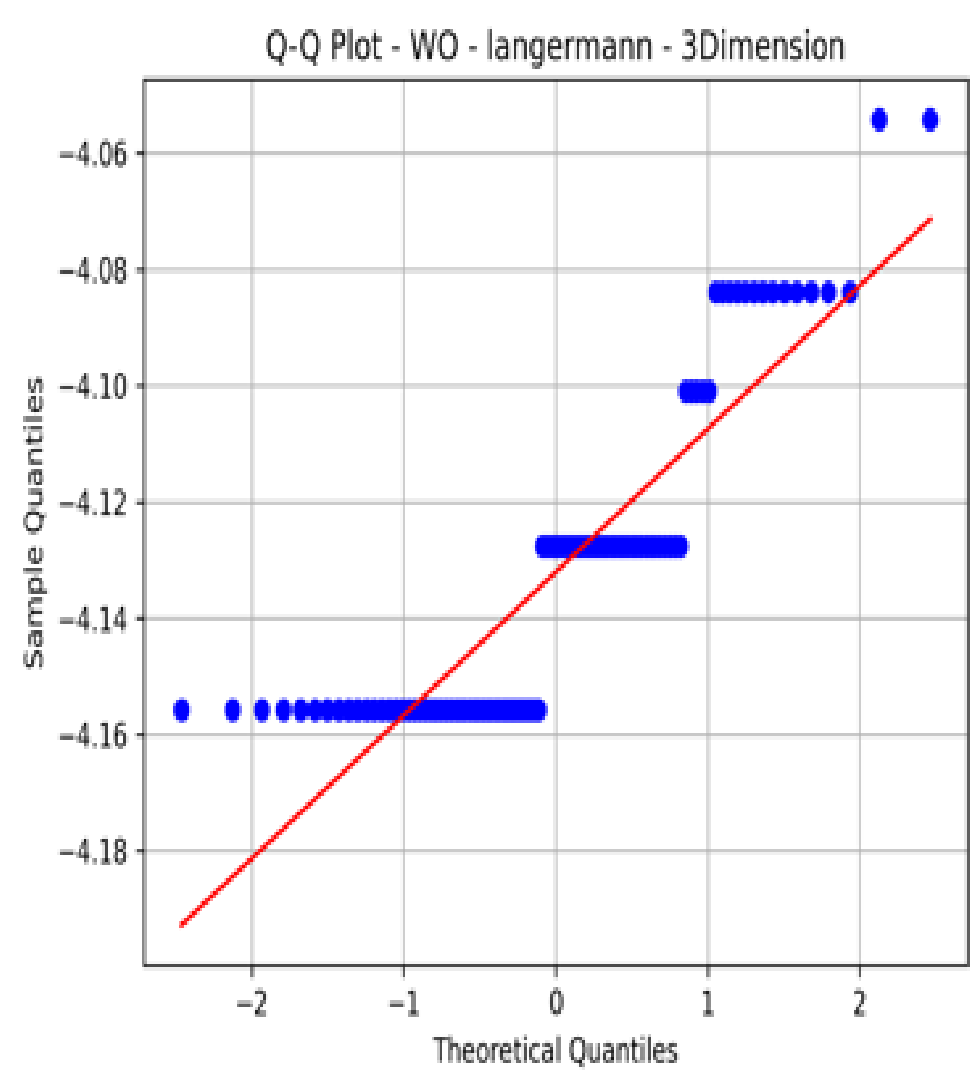
Q-Q Plot - WO - langermann - 3Dimension
Sample Quantiles
−4.06
−4.08
−4.10
−4.12
−4.14
−4.16
−4.18
−2
−1
0
1
2
Theoretical Quantiles

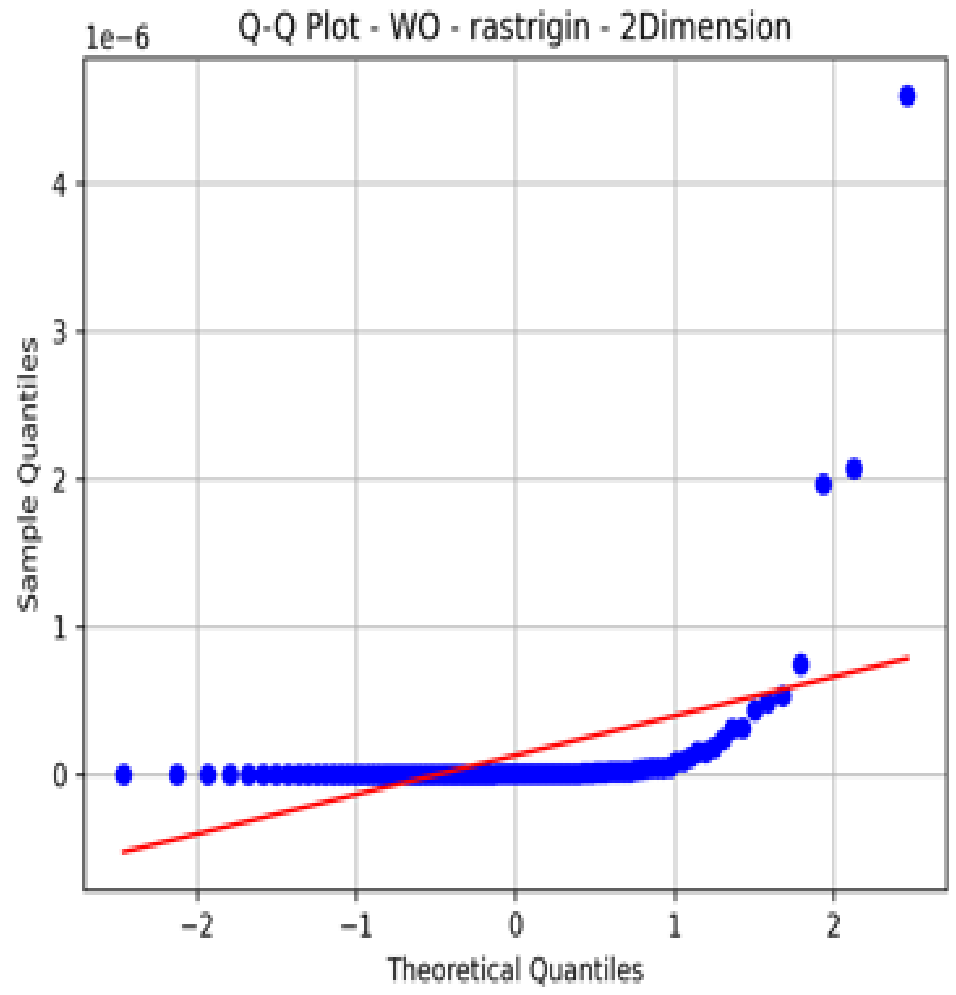
Q-Q Plot - WO - rastrigin - 2Dimension
1e−6
Sample Quantiles
Theoretical Quantiles

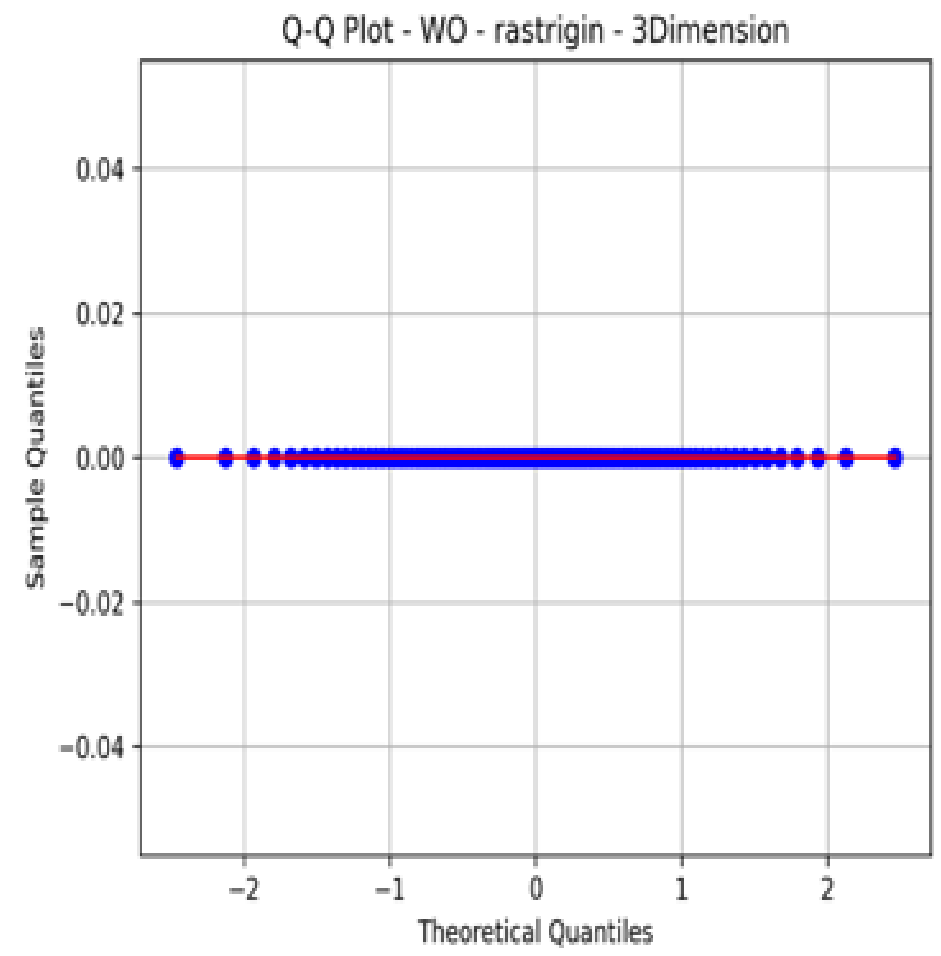
Q-Q Plot - WO - rastrigin - 3Dimension
Sample Quantiles
Theoretical Quantiles

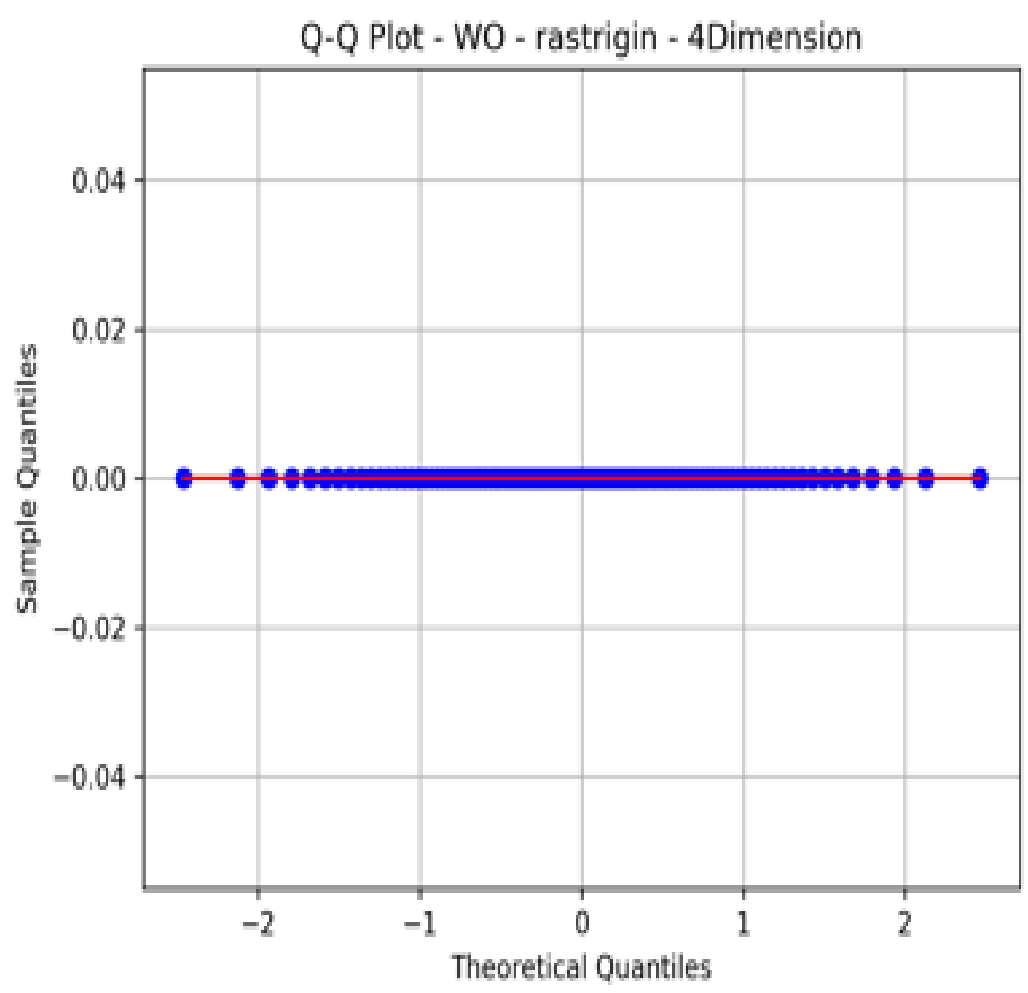
Q-Q Plot - WO - rastrigin - 4Dimension
Sample Quantiles
Theoretical Quantiles

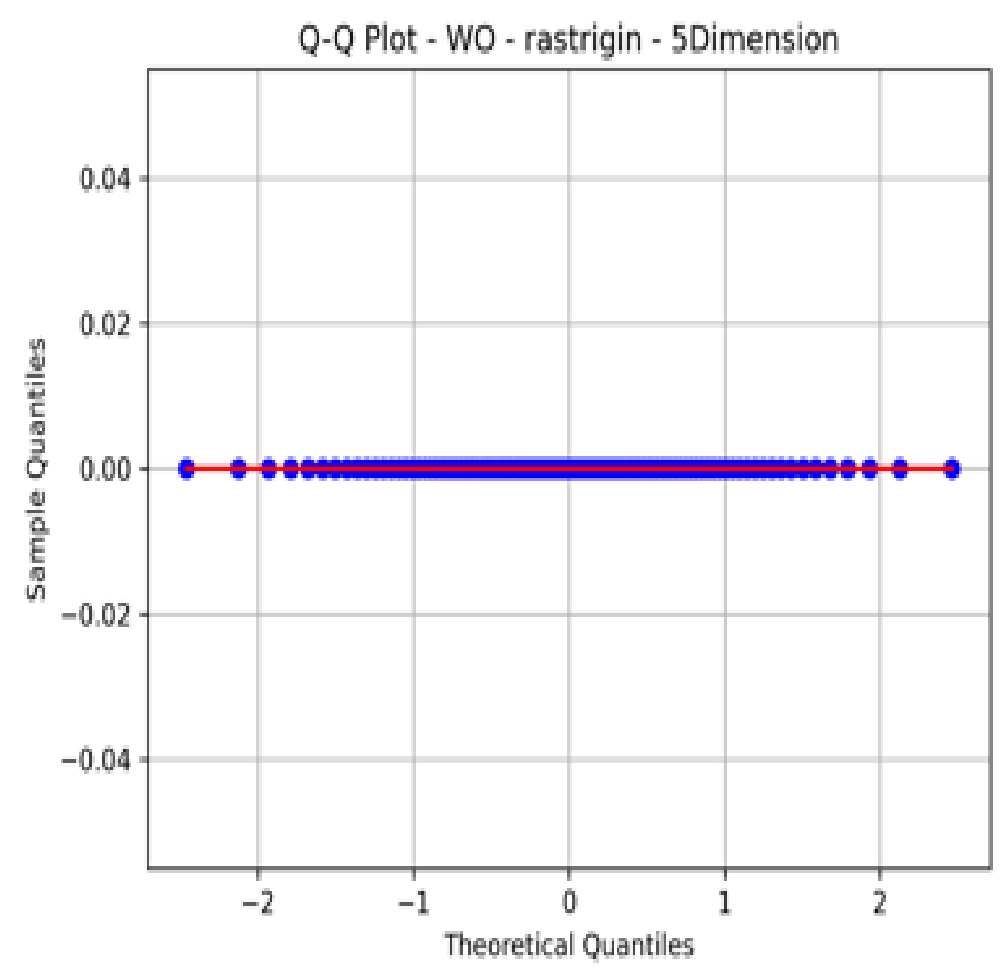
Q-Q Plot - WO - rastrigin - 5Dimension
Sample Quantiles
Theoretical Quantiles

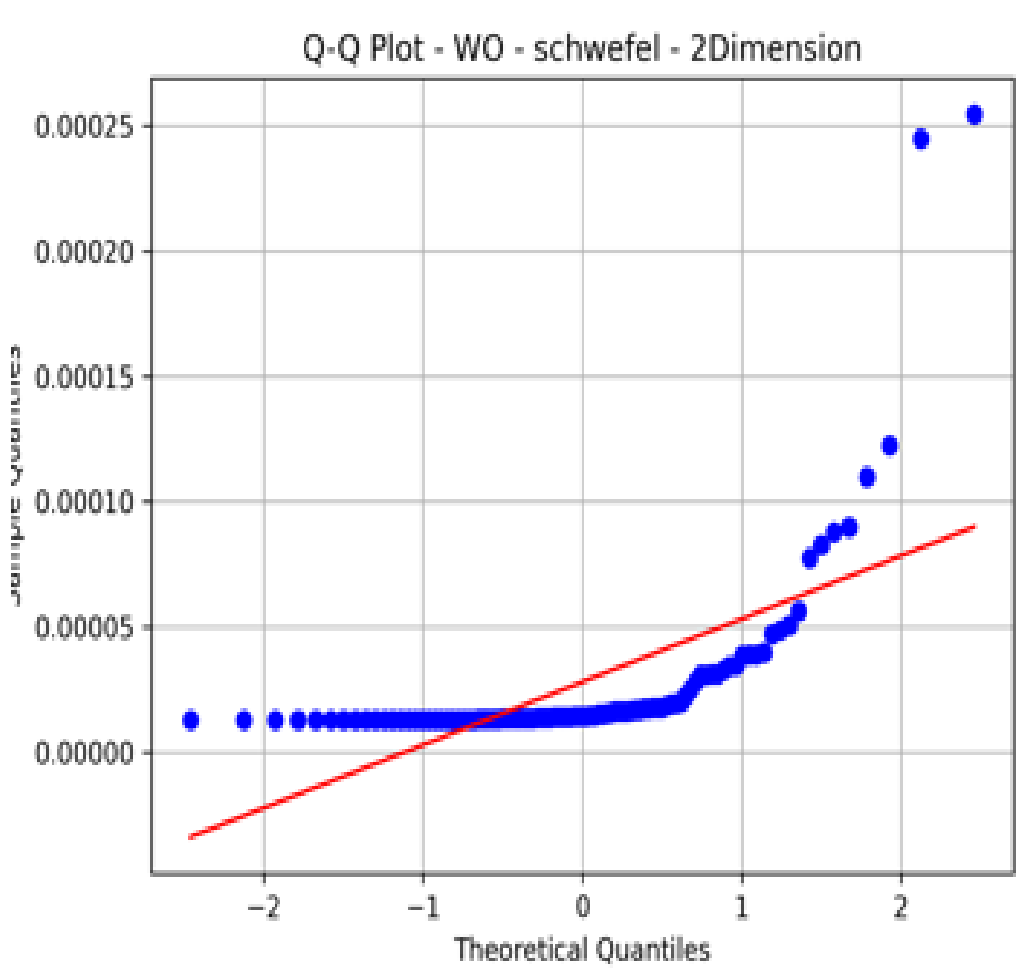
Q-Q Plot - WO - schwefel - 2Dimension
Theoretical Quantiles

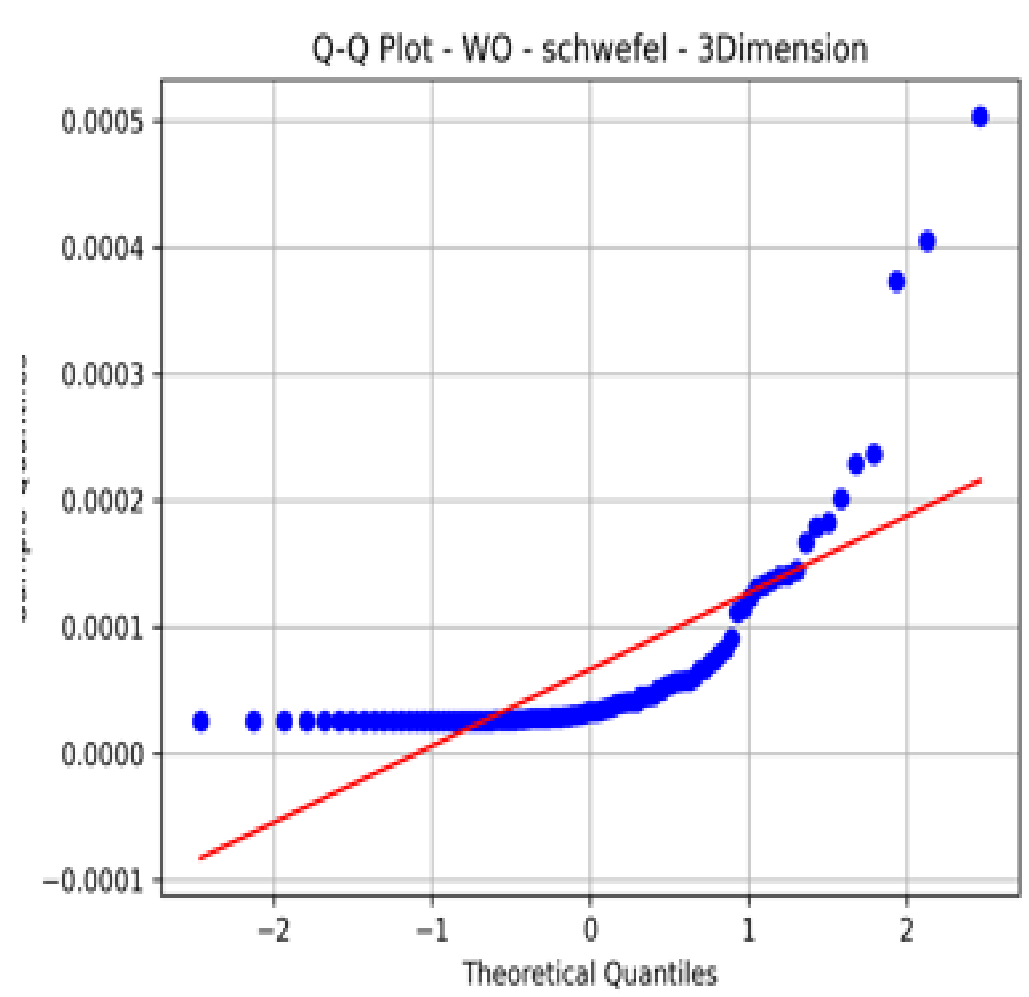
Q-Q Plot - WO - schwefel - 3Dimension
Theoretical Quantiles

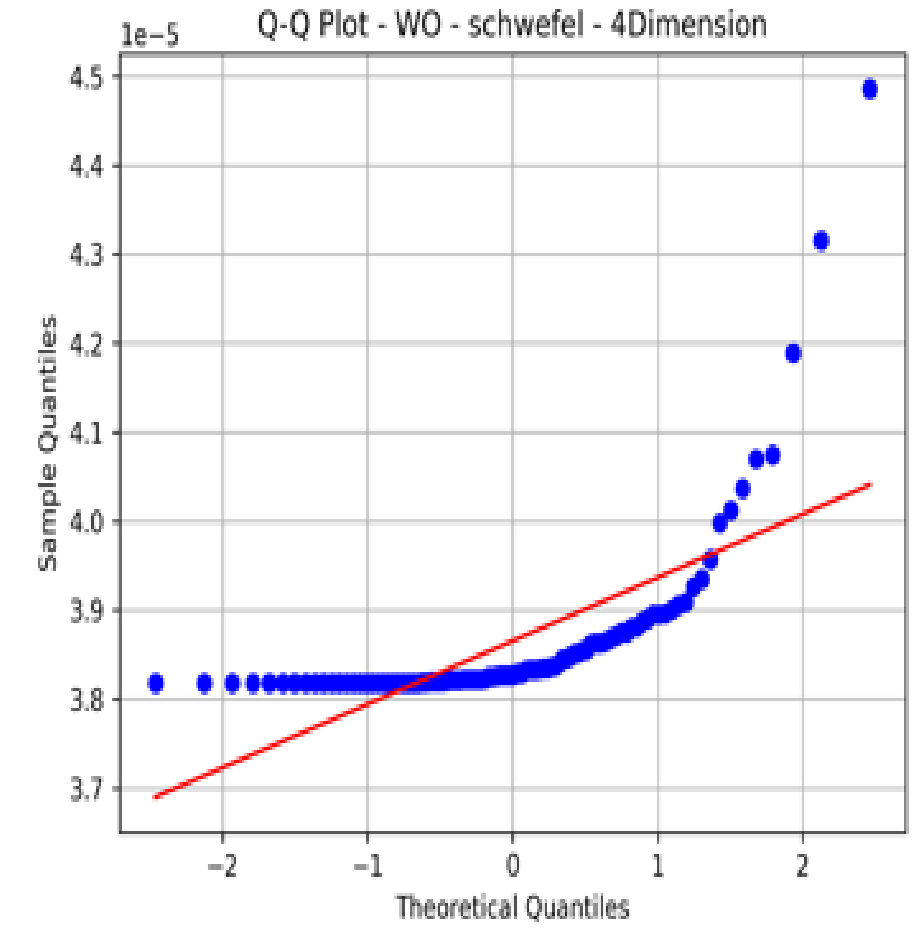
1e−5
Q-Q Plot - WO - schwefel - 4Dimension
Sample Quantiles
Theoretical Quantiles

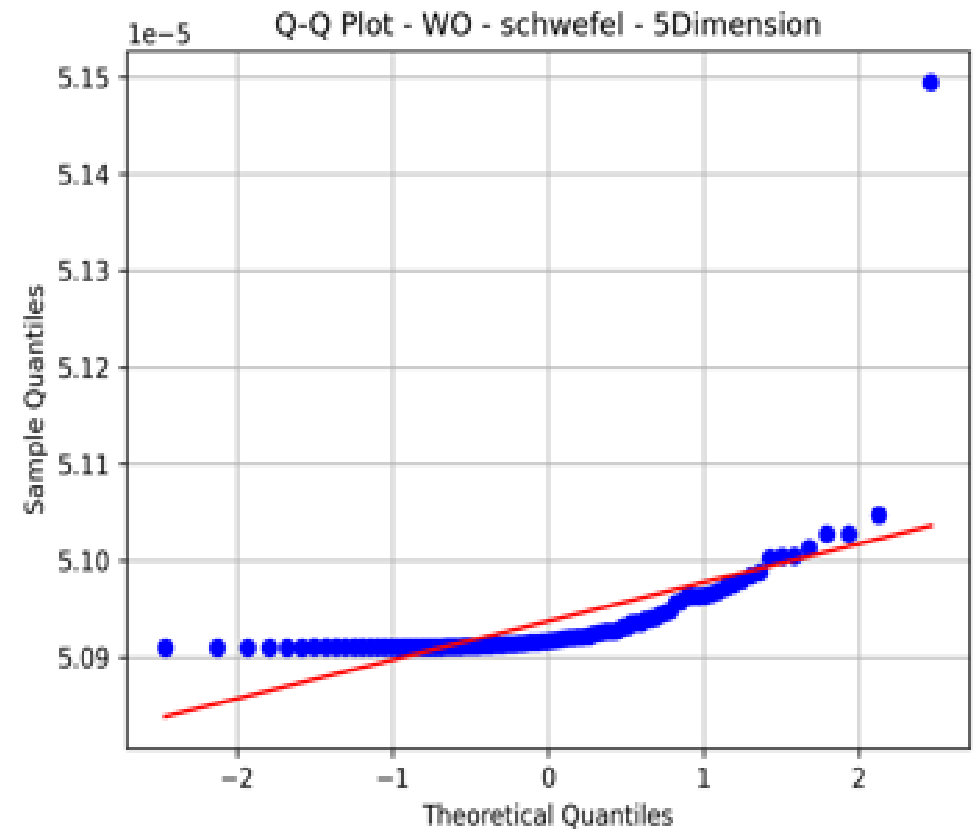
1e−5
Q-Q Plot - WO - schwefel - 5Dimension
Sample Quantiles
Theoretical Quantiles

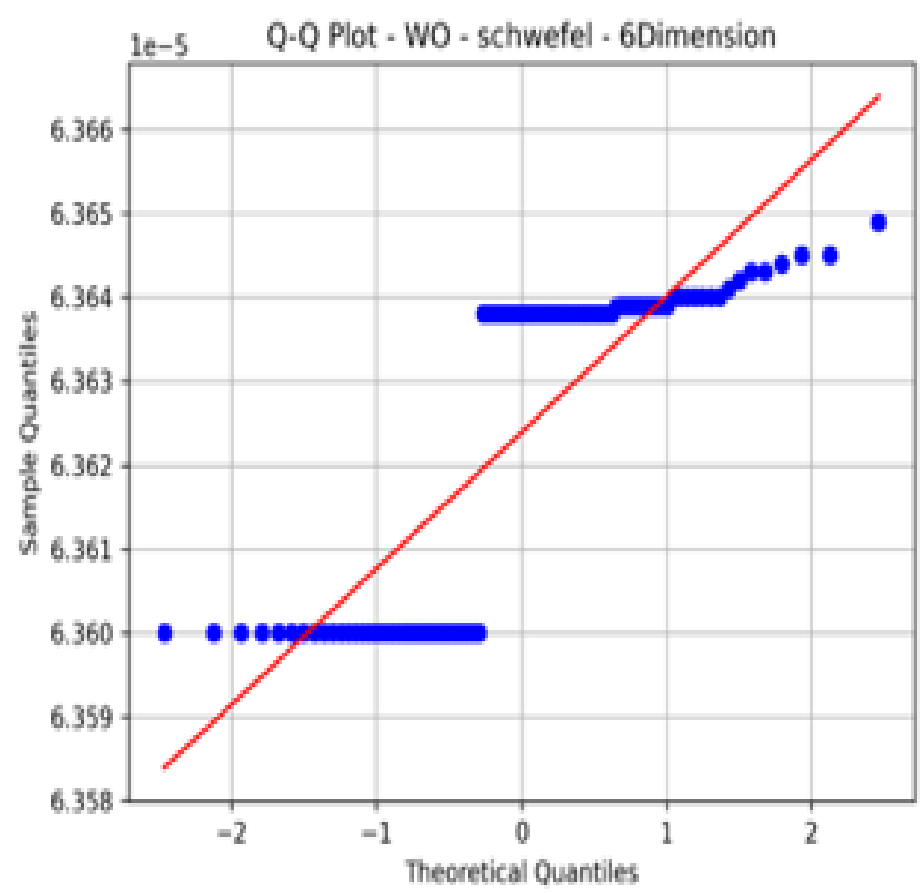
1e−5
Q-Q Plot - WO - schwefel - 6Dimension
Sample Quantiles
Theoretical Quantiles

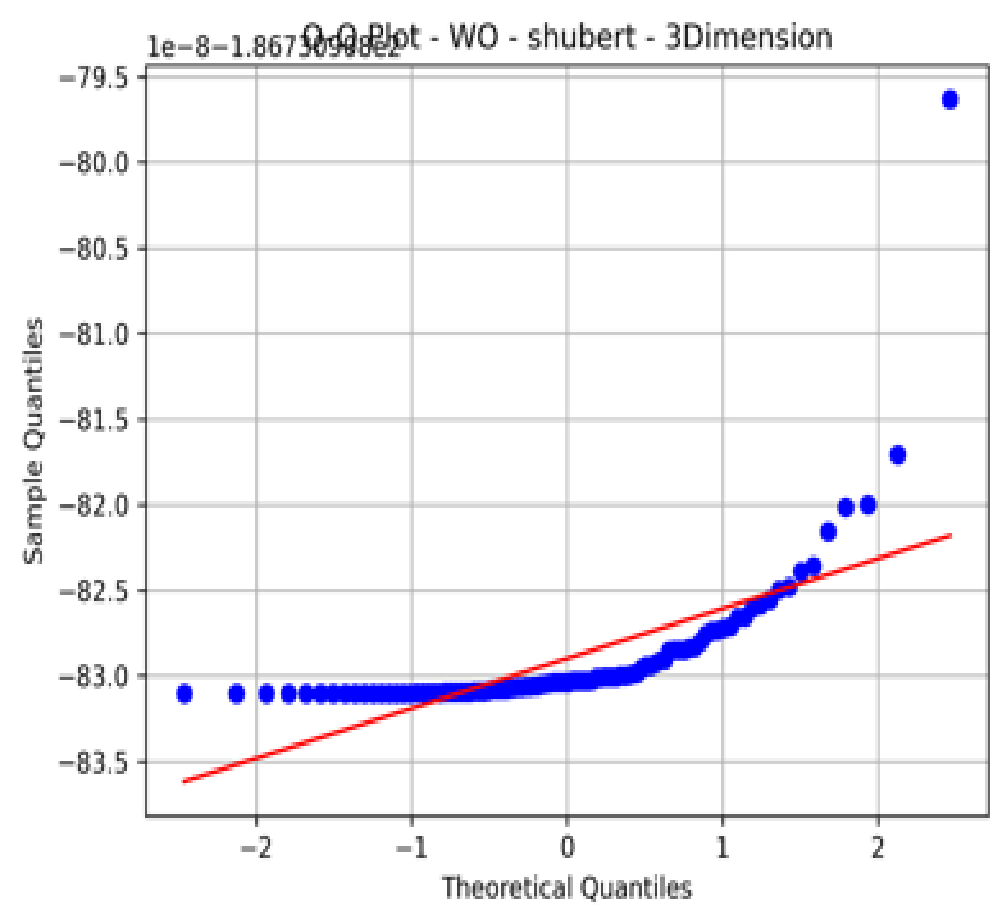
Q-Q Plot - WO - shubert - 3Dimension
Sample Quantiles
Theoretical Quantiles

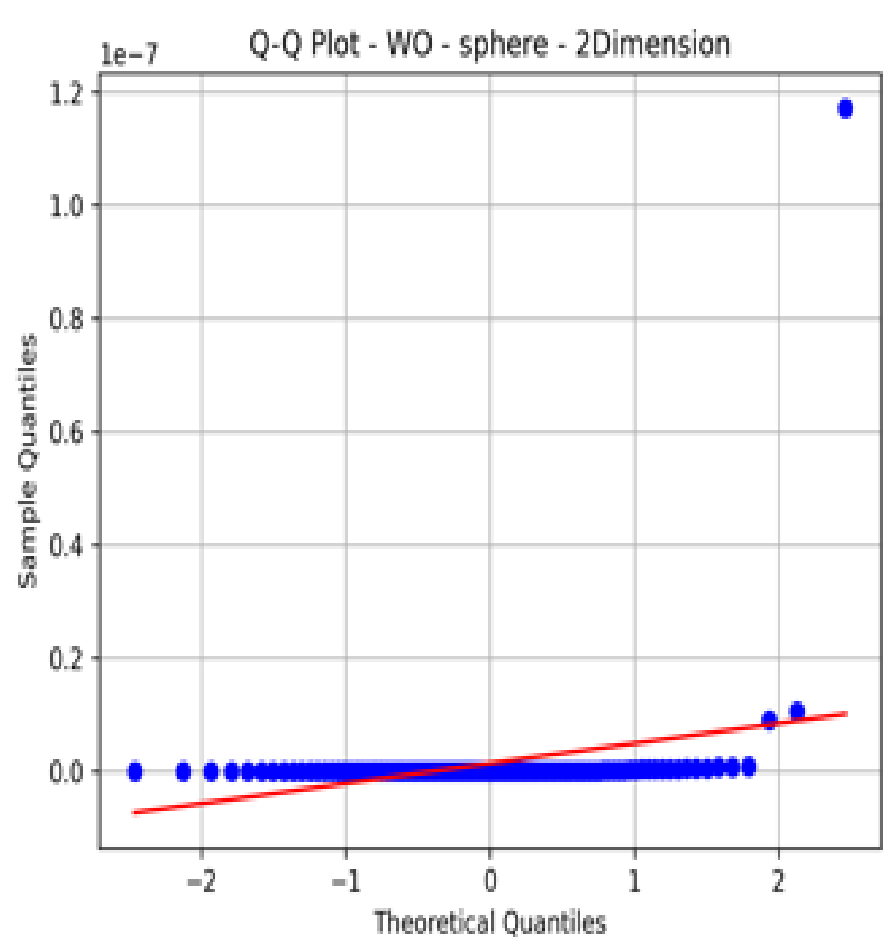
1e−7
Q-Q Plot - WO - sphere - 2Dimension
Sample Quantiles
Theoretical Quantiles

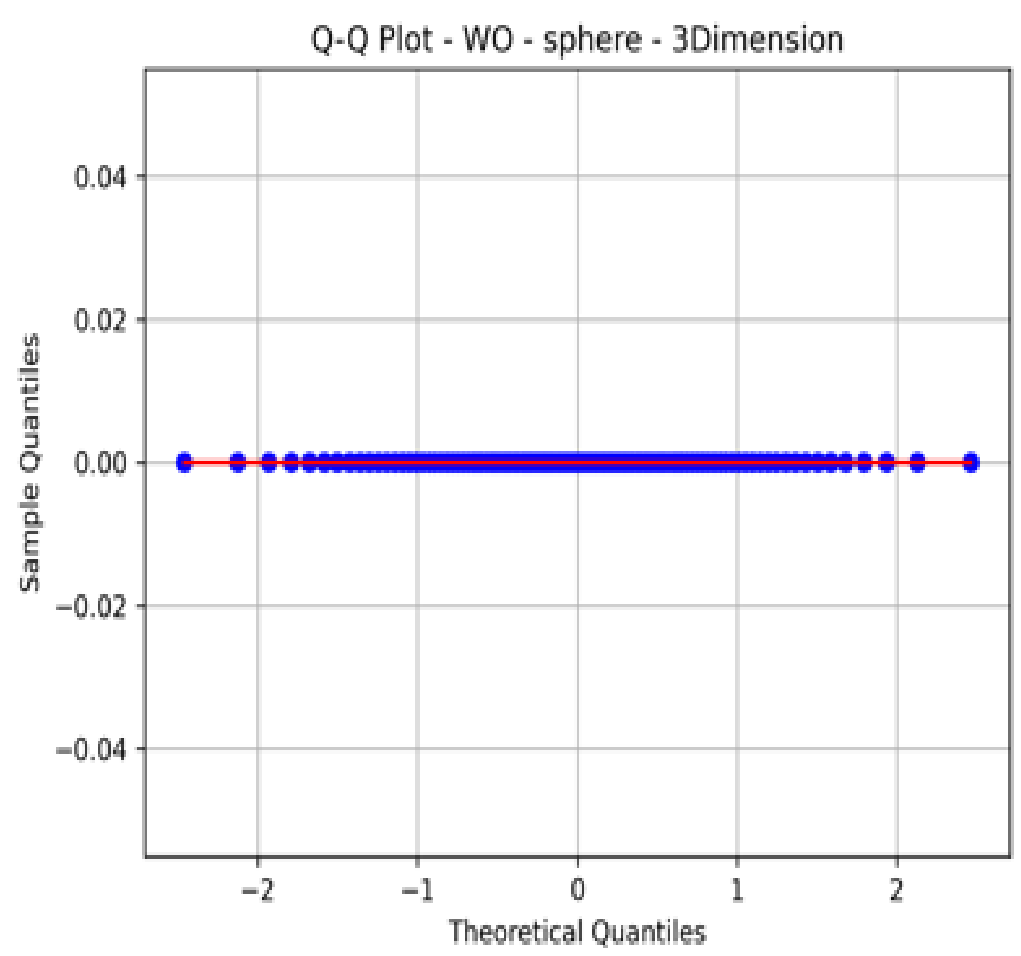
Q-Q Plot - WO - sphere - 3Dimension
Sample Quantiles
Theoretical Quantiles

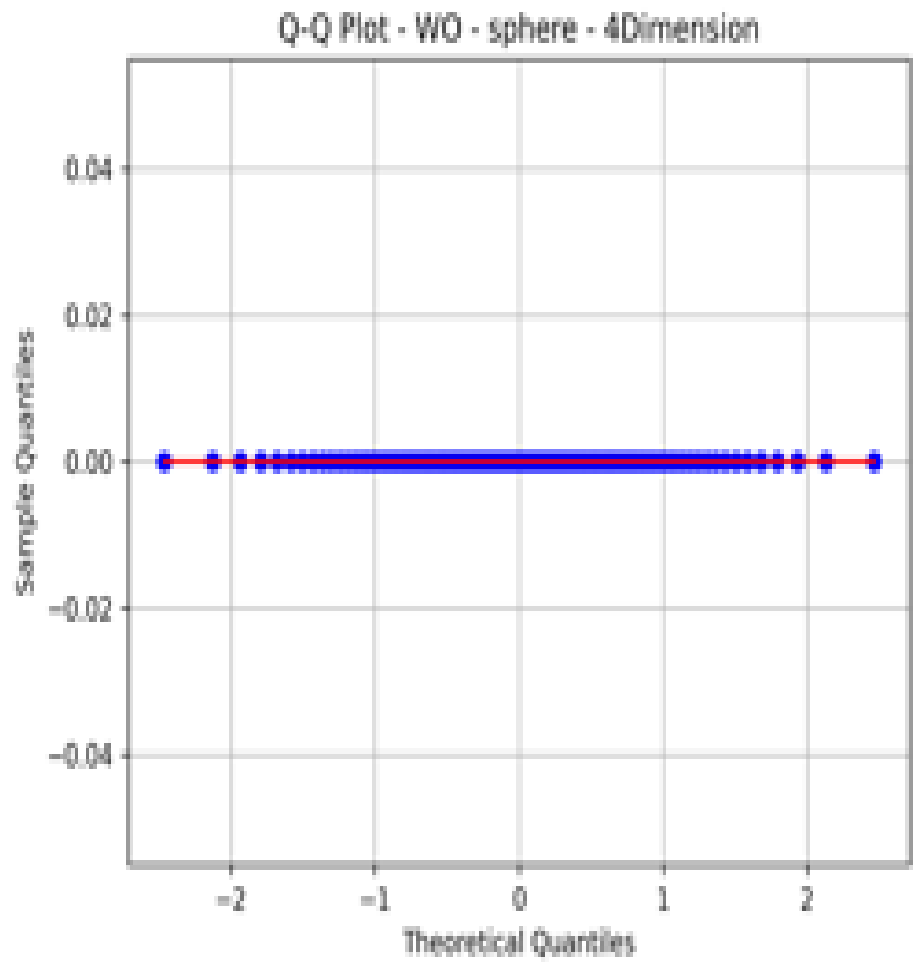

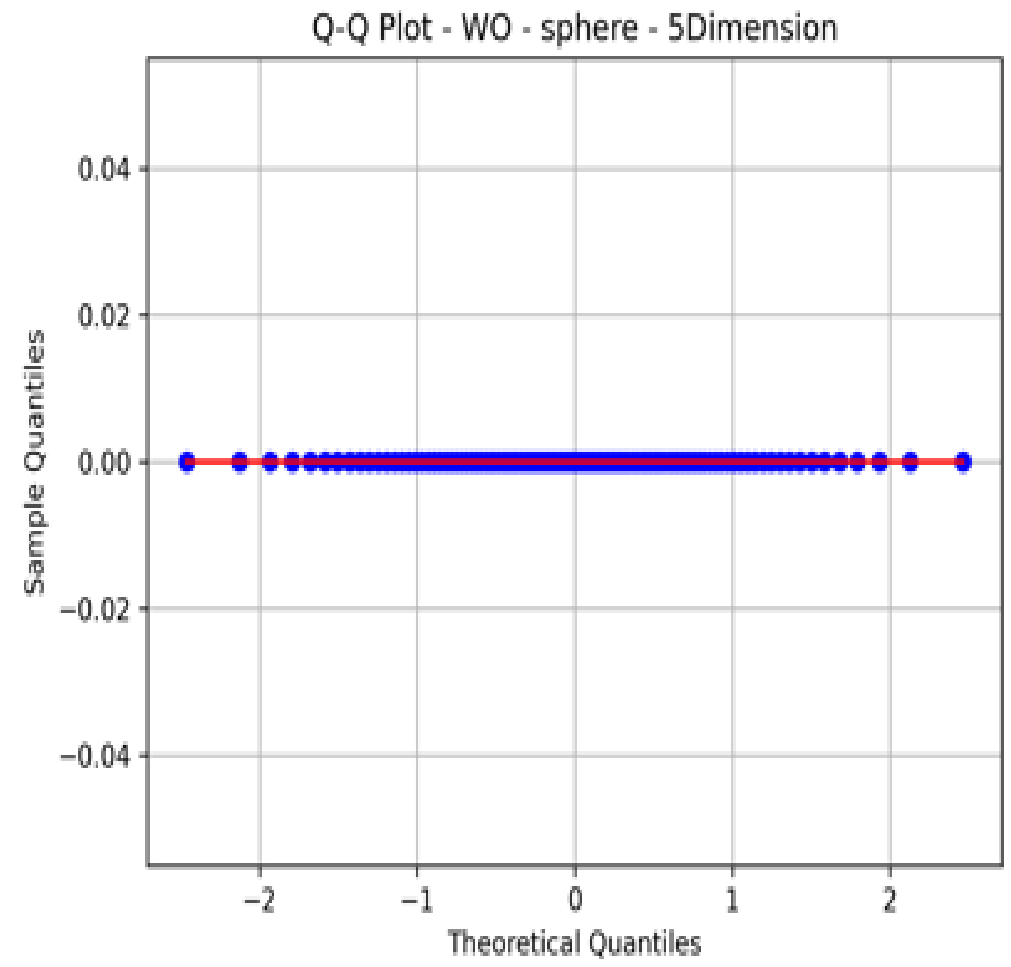

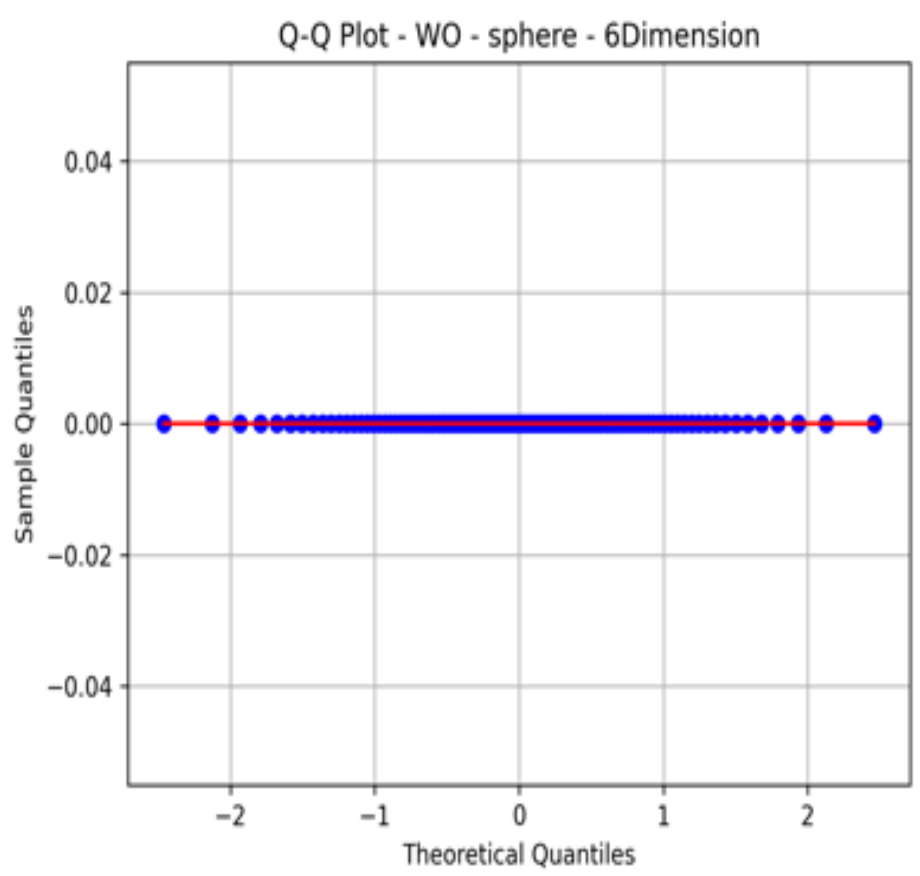

Non-Continuous Functions QQ Plots

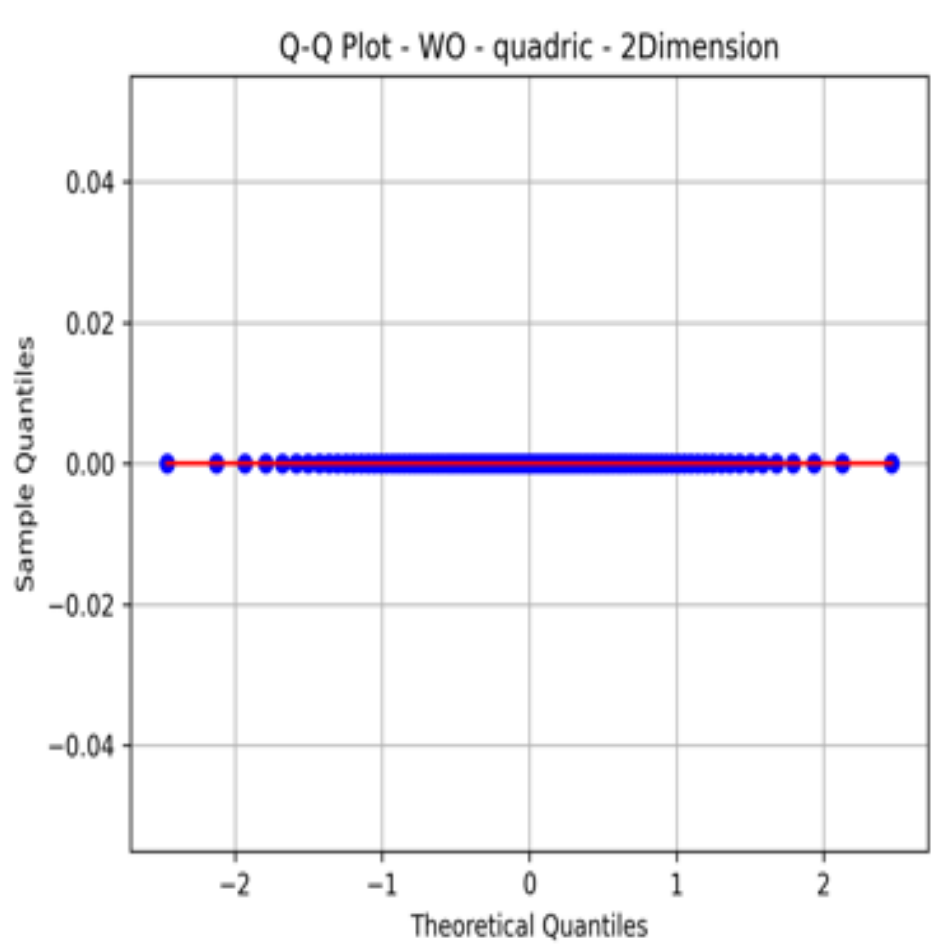

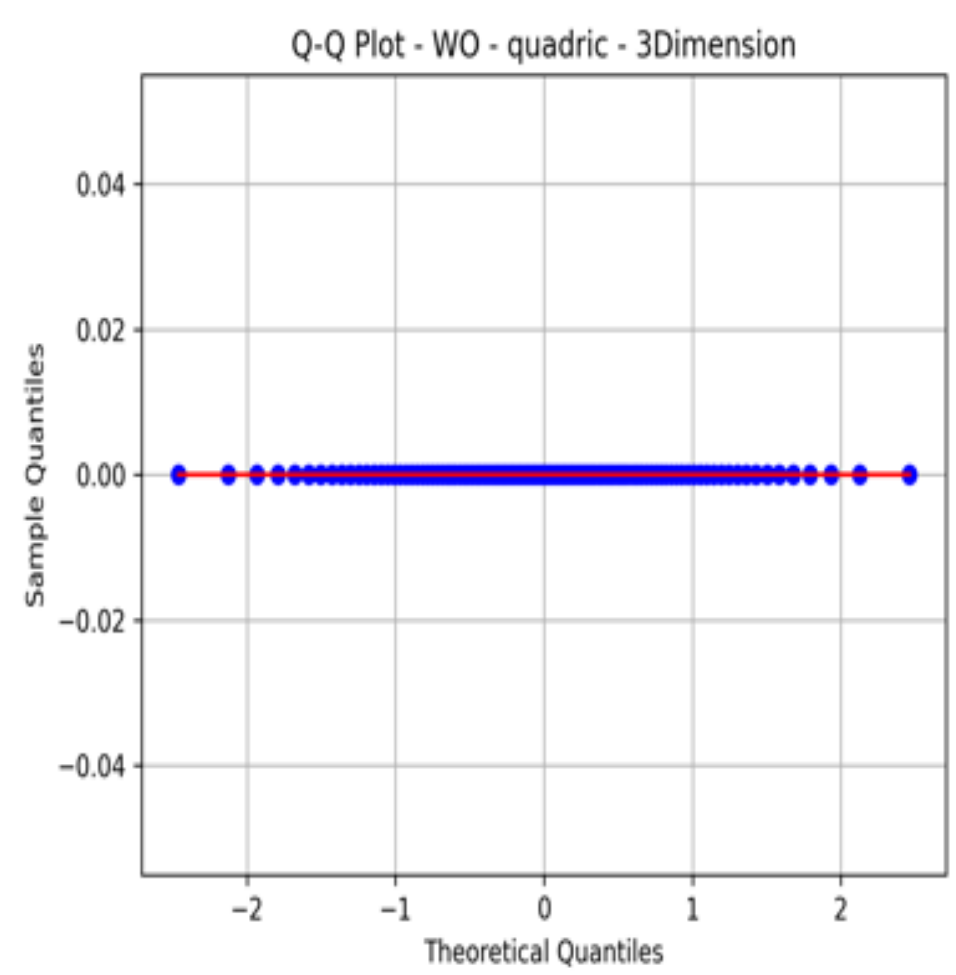

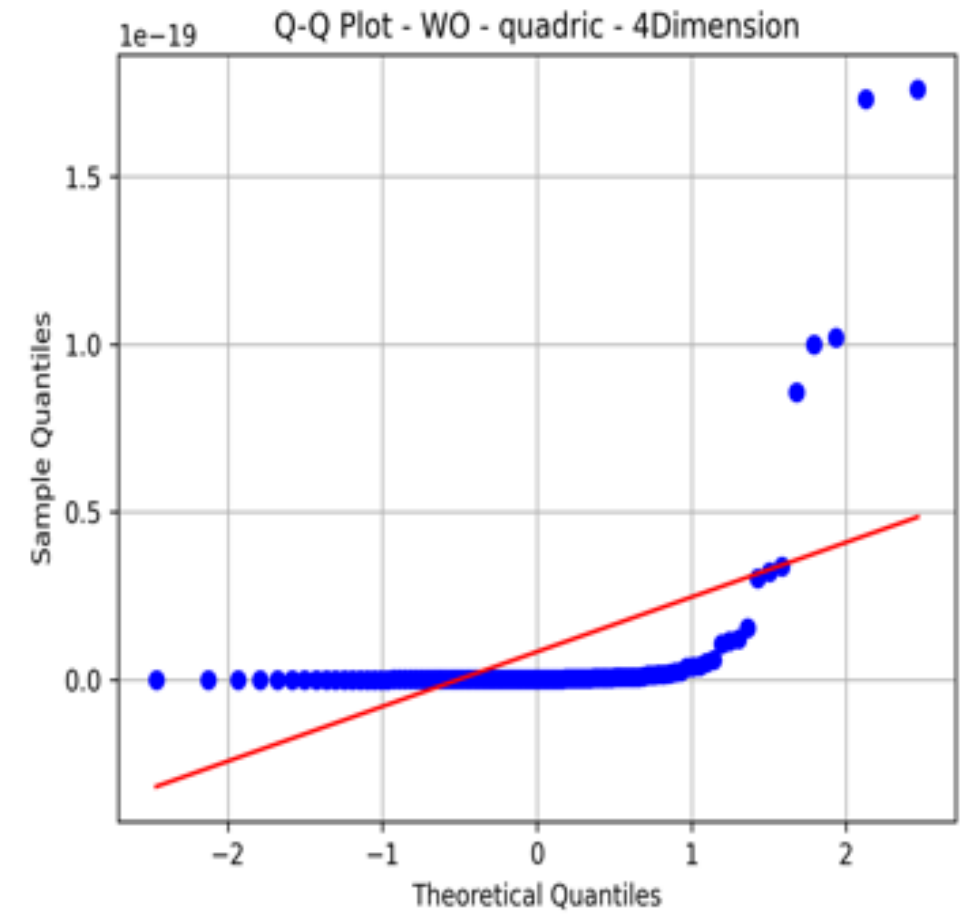
Q-Q Plot - WO - quadric - 4Dimension
1e−19
1.5
1.0
0.5
0.0
Sample Quantiles
−2
−1
0
1
2
Theoretical Quantiles

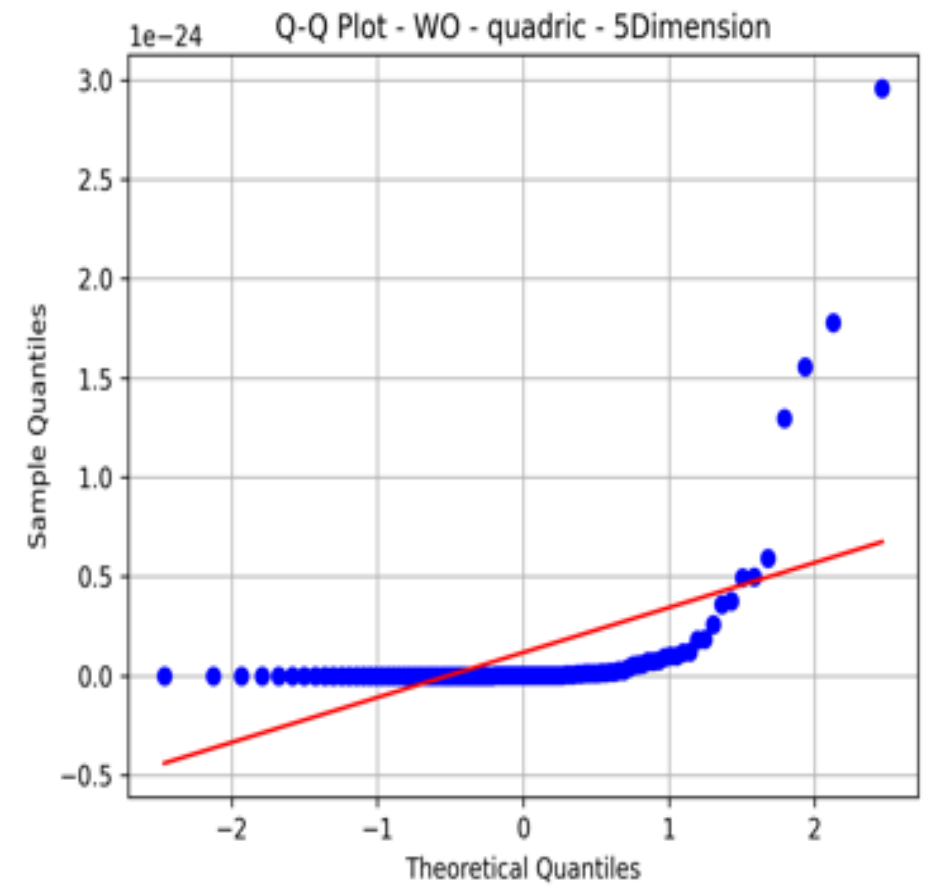
Q-Q Plot - WO - quadric - 5Dimension
1e−24
3.0
2.5
2.0
1.5
1.0
0.5
0.0
−0.5
Sample Quantiles
−2
−1
0
1
2
Theoretical Quantiles

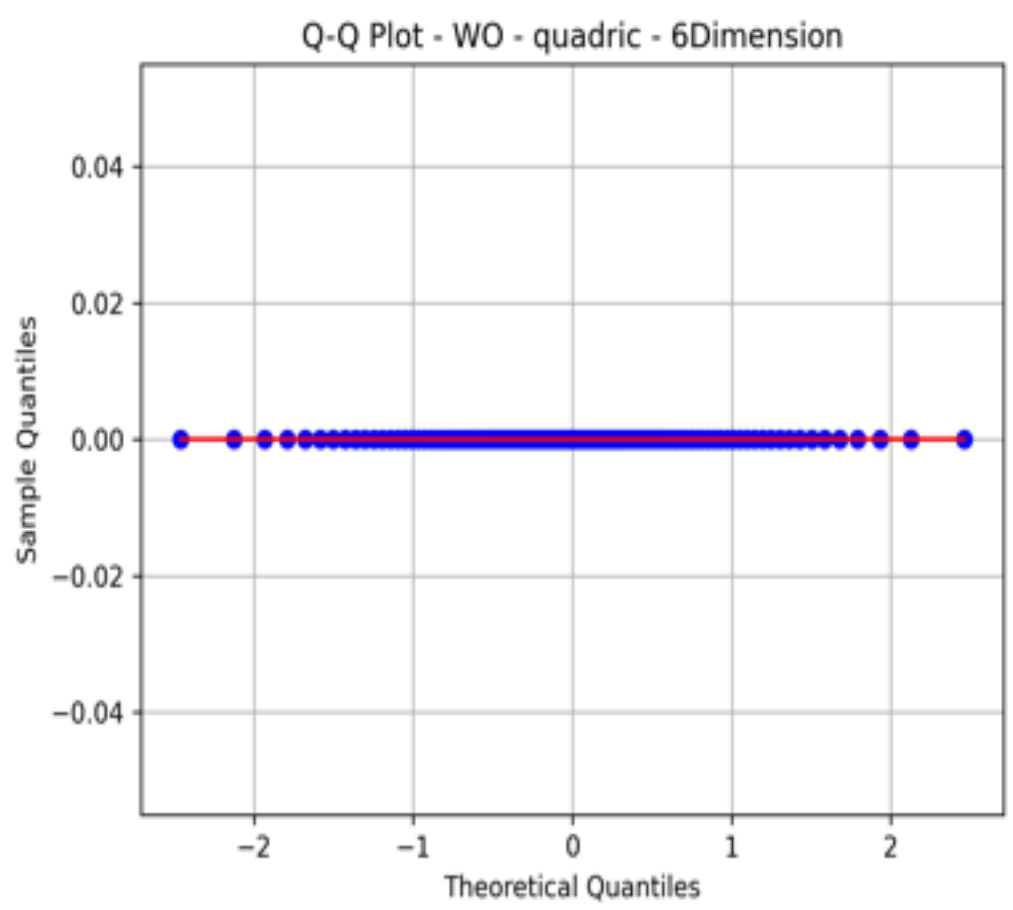
Q-Q Plot - WO - quadric - 6Dimension
0.04
0.02
0.00
−0.02
−0.04
Sample Quantiles
−2
−1
0
1
2
Theoretical Quantiles

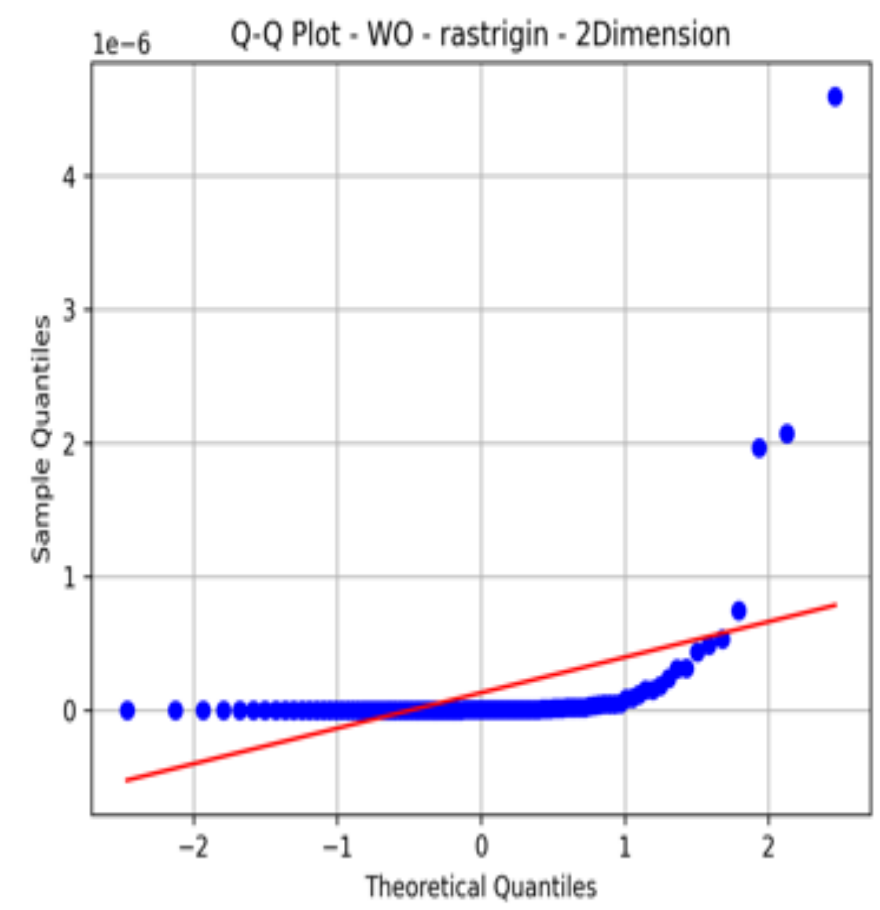
Q-Q Plot - WO - rastrigin - 2Dimension
1e−6
4
3
2
1
0
Sample Quantiles
−2
−1
0
1
2
Theoretical Quantiles

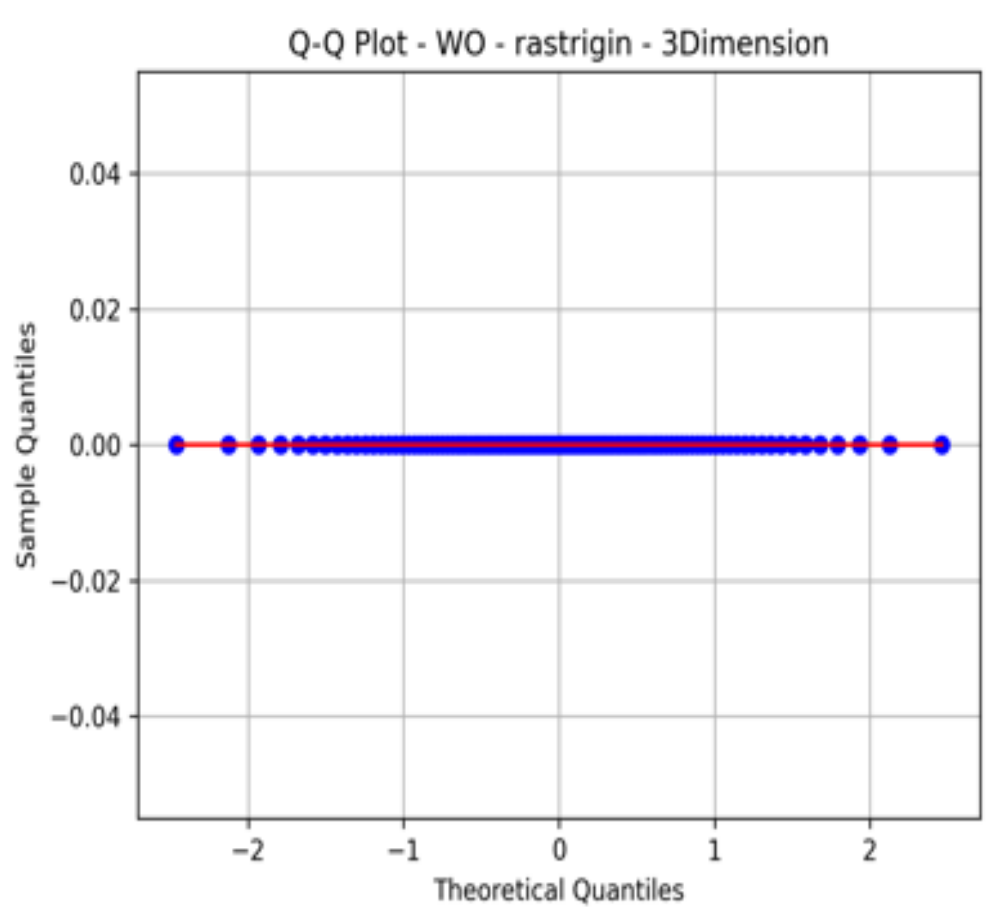
Q-Q Plot - WO - rastrigin - 3Dimension
0.04
0.02
0.00
−0.02
−0.04
Sample Quantiles
−2
−1
0
1
2
Theoretical Quantiles

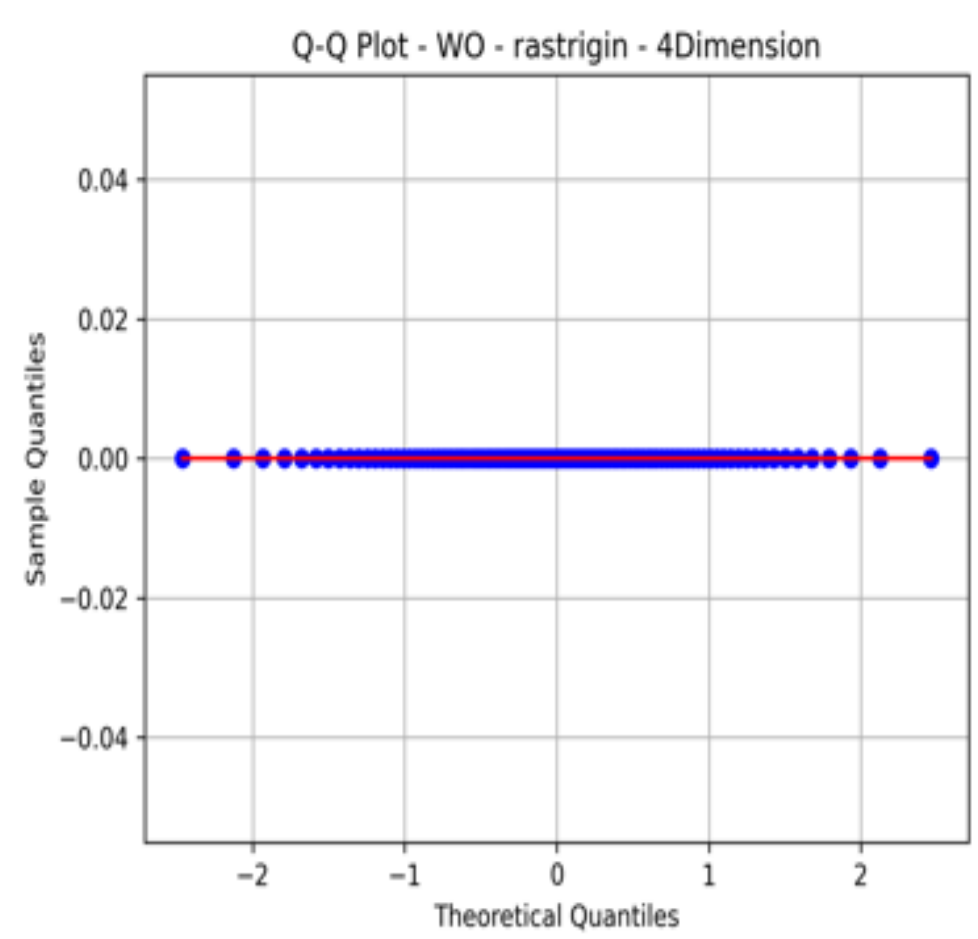
Q-Q Plot - WO - rastrigin - 4Dimension
0.04
0.02
0.00
−0.02
−0.04
Sample Quantiles
−2
−1
0
1
2
Theoretical Quantiles

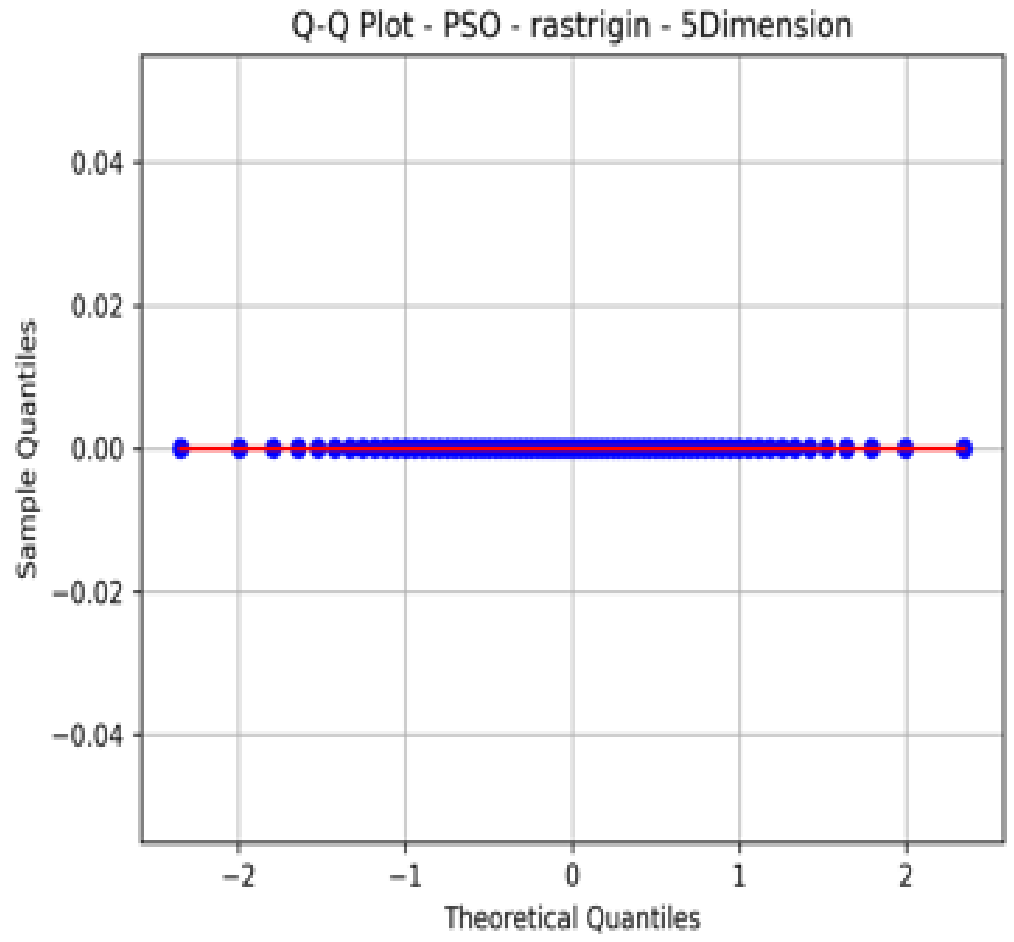
Q-Q Plot - PSO - rastrigin - 5Dimension
Sample Quantiles
Theoretical Quantiles

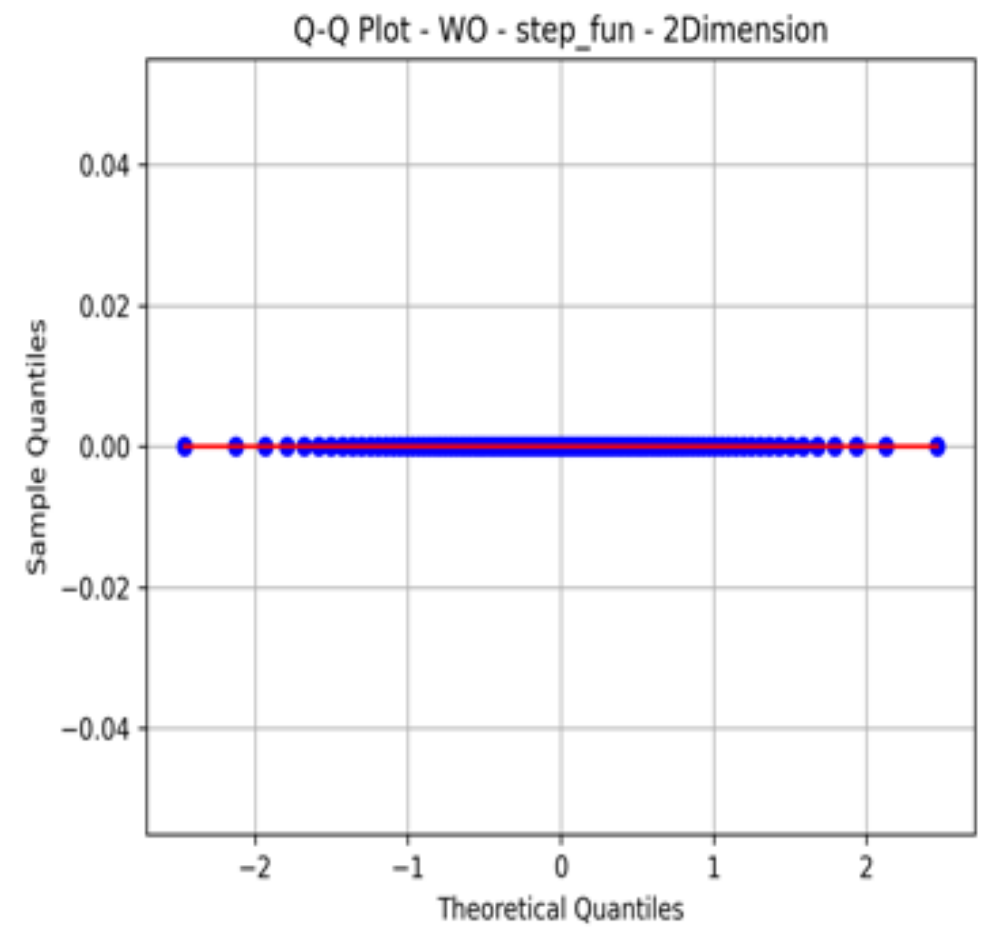
Q-Q Plot - WO - step_fun - 2Dimension
Sample Quantiles
Theoretical Quantiles

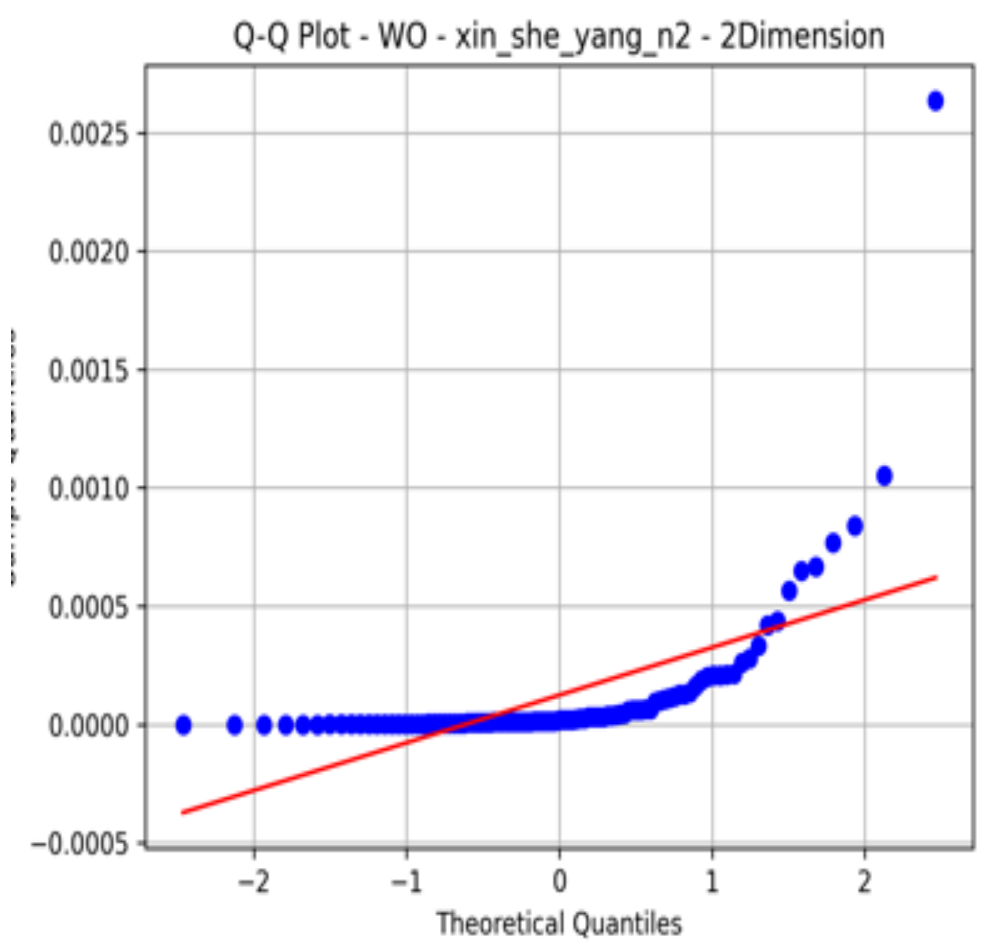
Q-Q Plot - WO - xin_she_yang_n2 - 2Dimension
Theoretical Quantiles

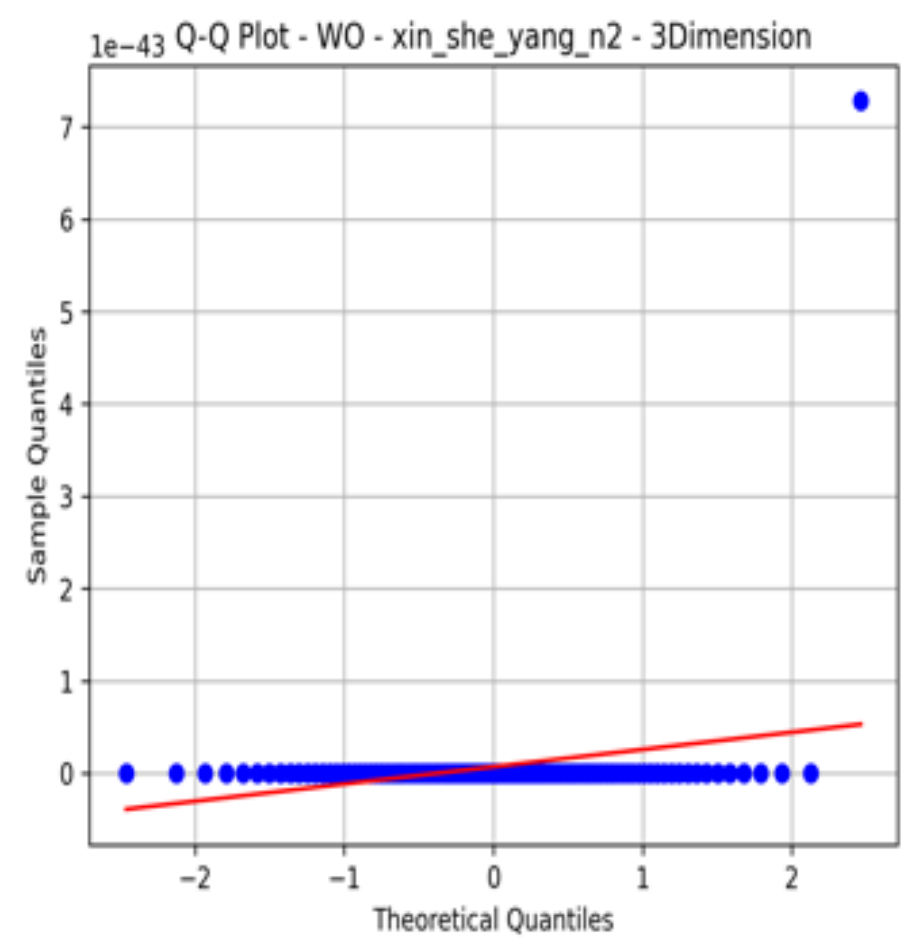
1e−43
Q-Q Plot - WO - xin_she_yang_n2 - 3Dimension
Sample Quantiles
Theoretical Quantiles

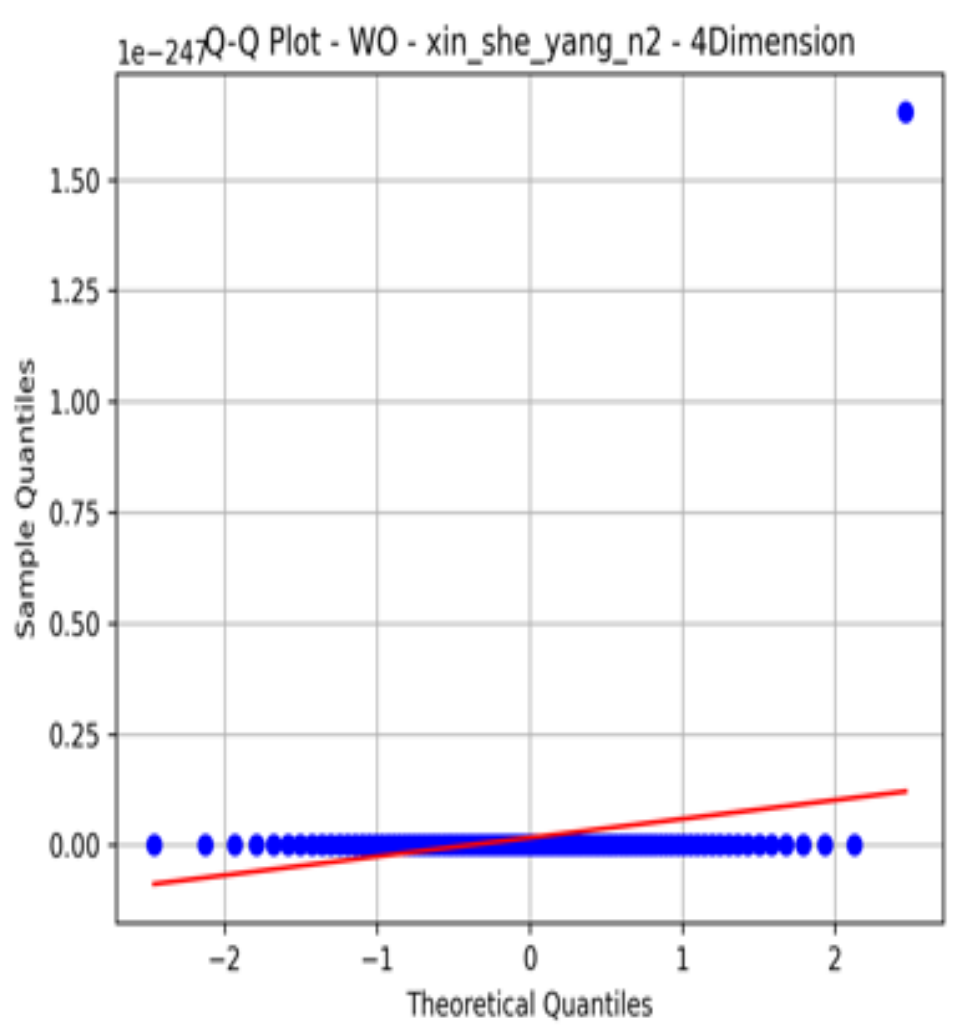
1e−247
Q-Q Plot - WO - xin_she_yang_n2 - 4Dimension
Sample Quantiles
Theoretical Quantiles

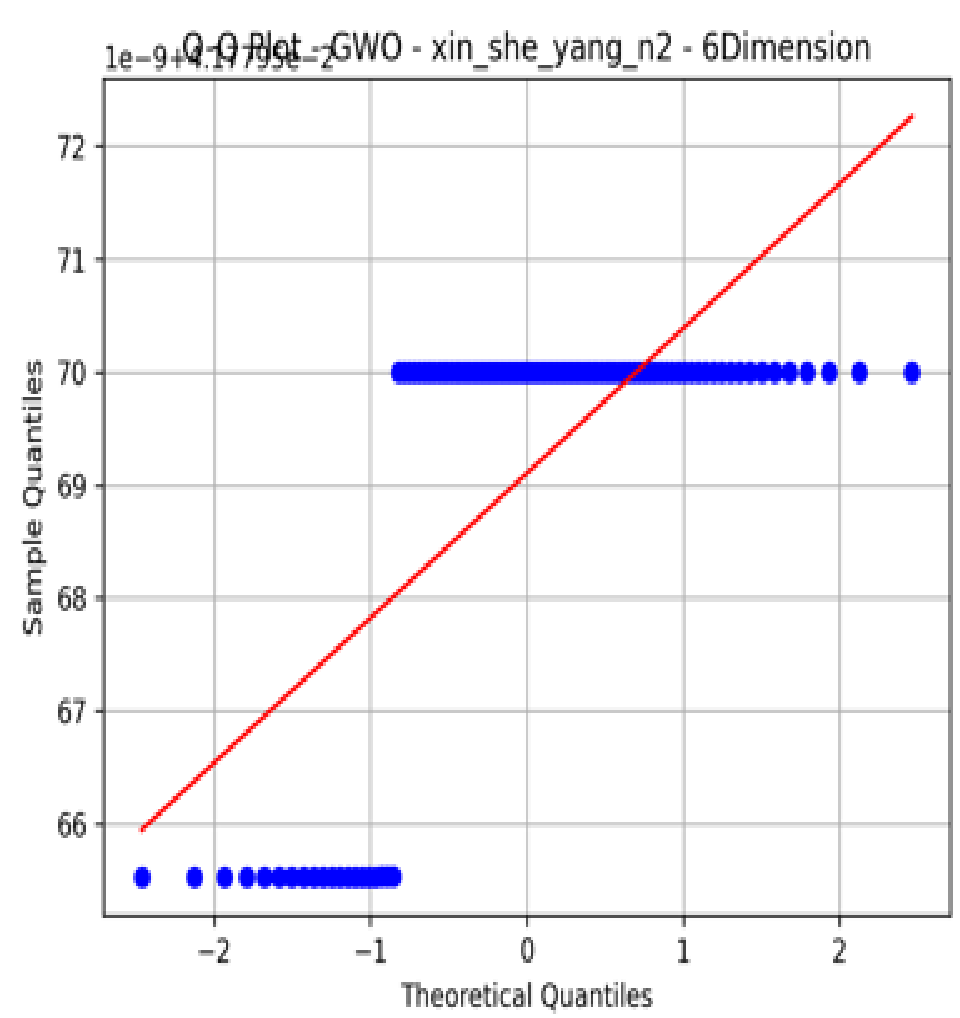
Q-Q Plot - GWO - xin_she_yang_n2 - 6Dimension
Sample Quantiles
Theoretical Quantiles

GWO Q-Q Plots

Continuous functions QQ Plots

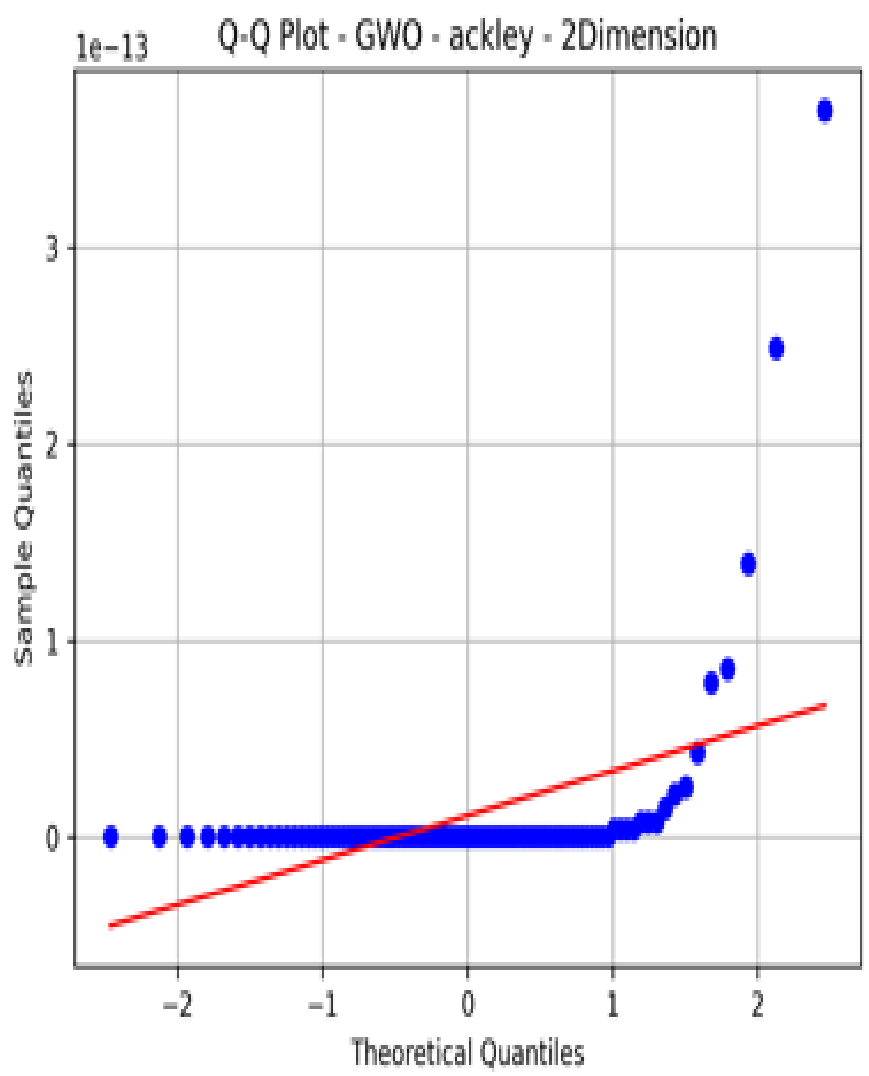

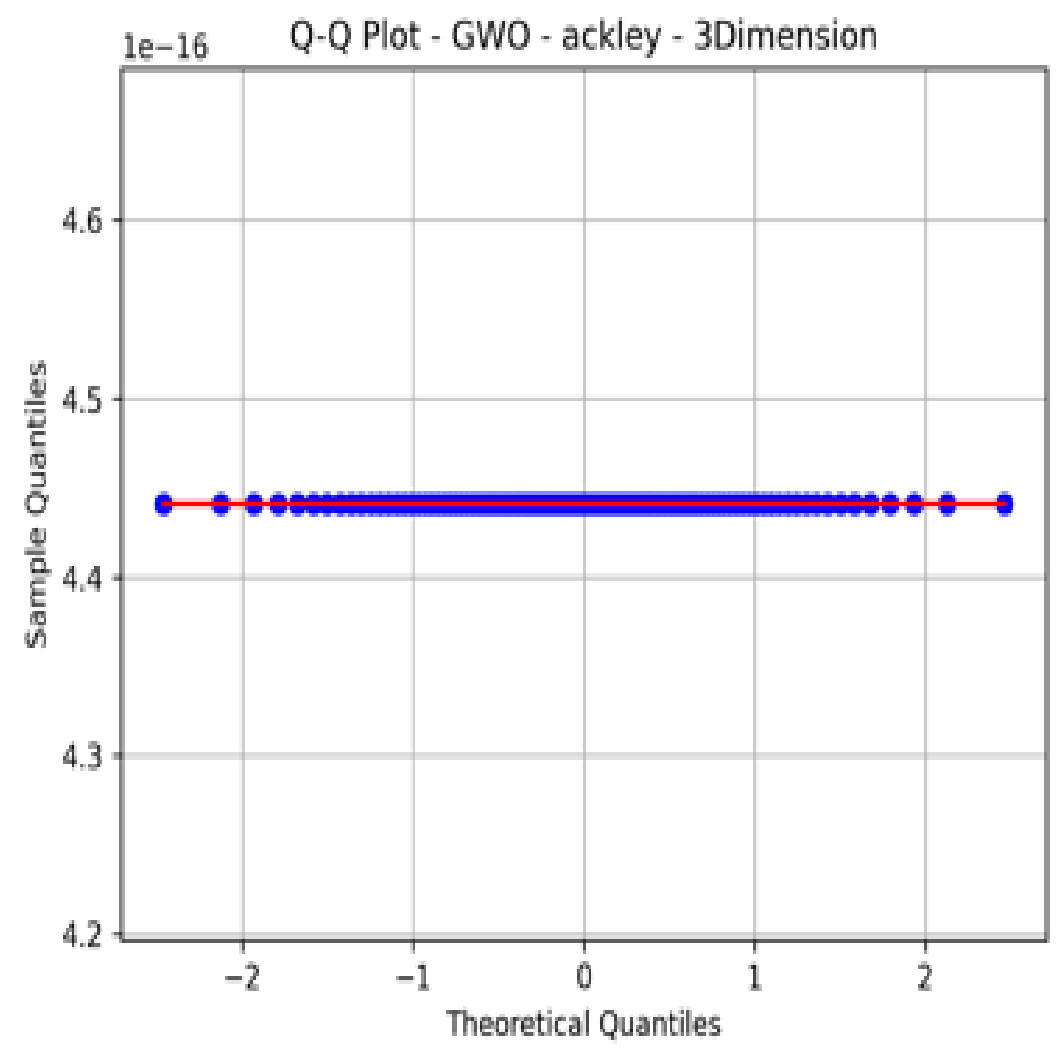

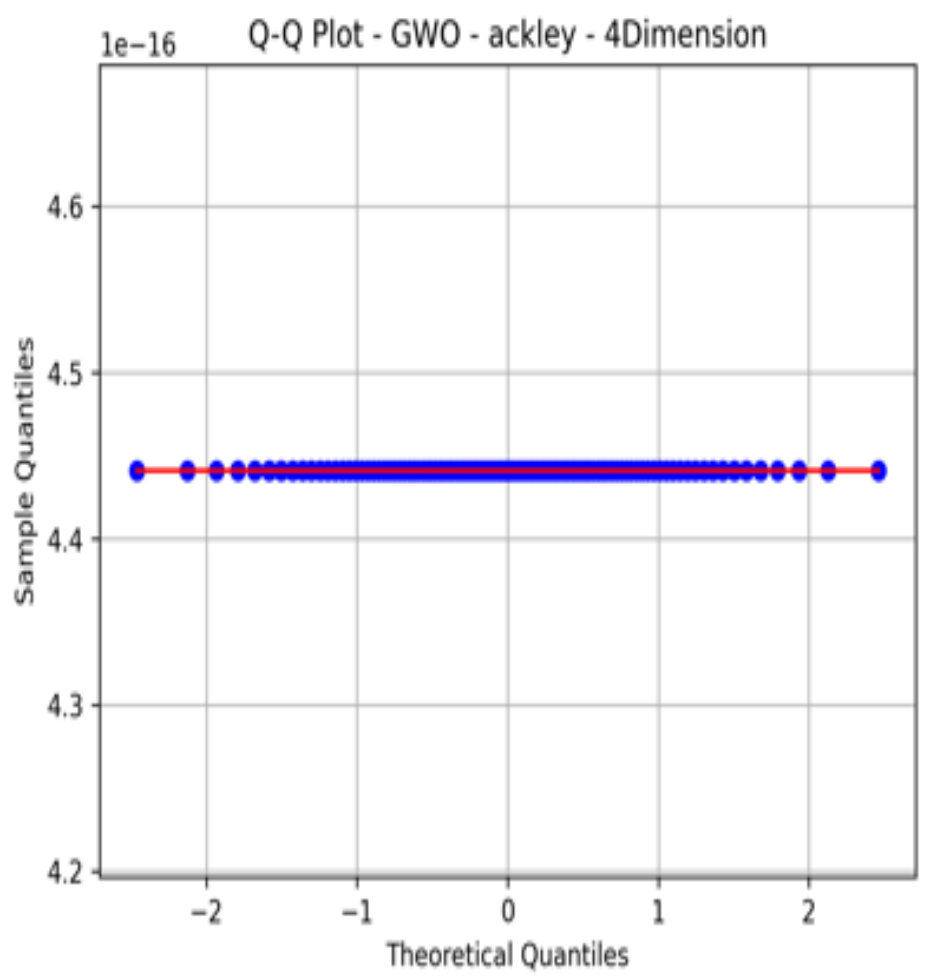

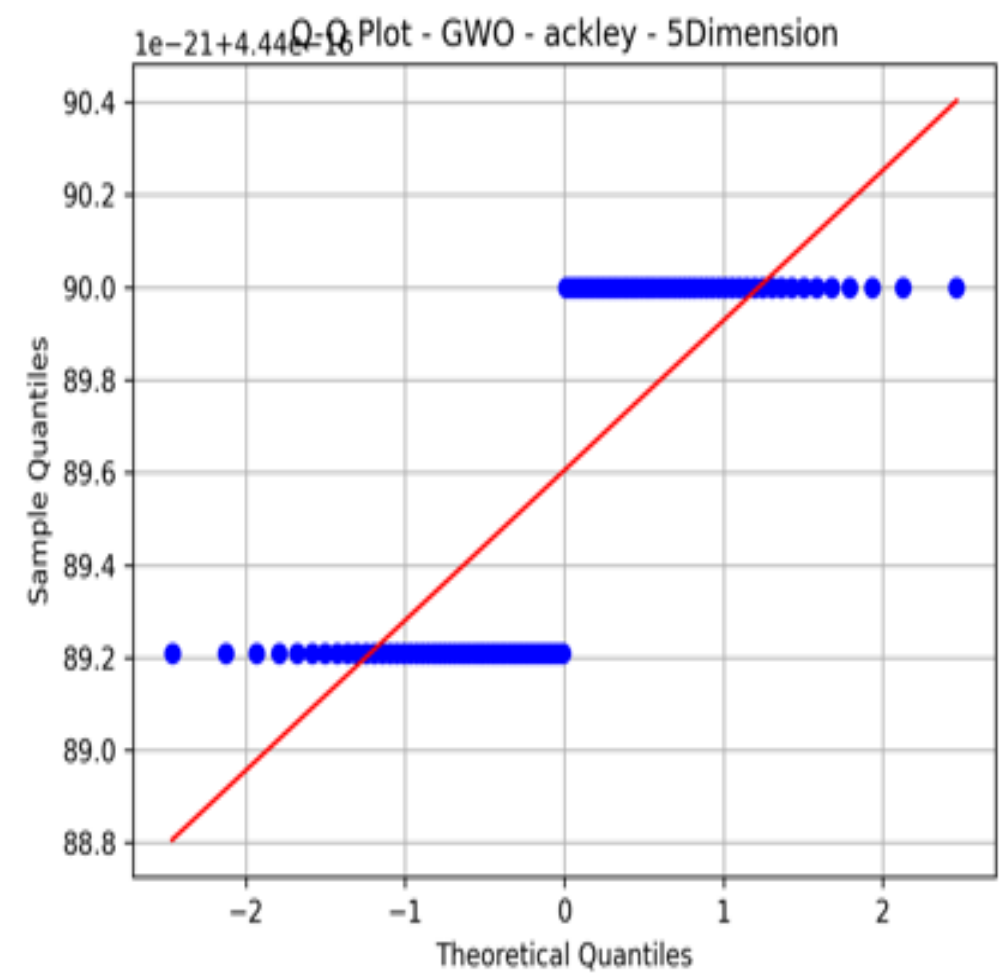

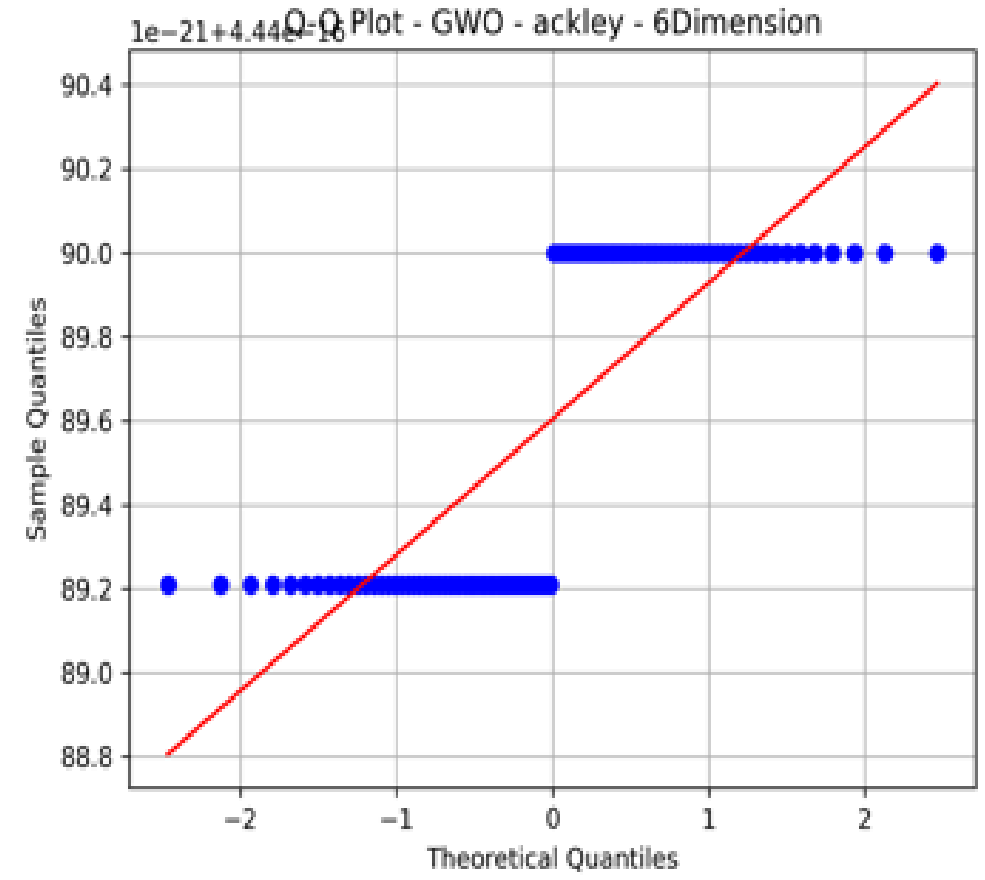
Q-Q Plot - GWO - ackley - 6Dimension
1e−21+4.44e−16
90.4
90.2
90.0
89.8
89.6
89.4
89.2
89.0
88.8
−2
−1
0
1
2
Sample Quantiles
Theoretical Quantiles

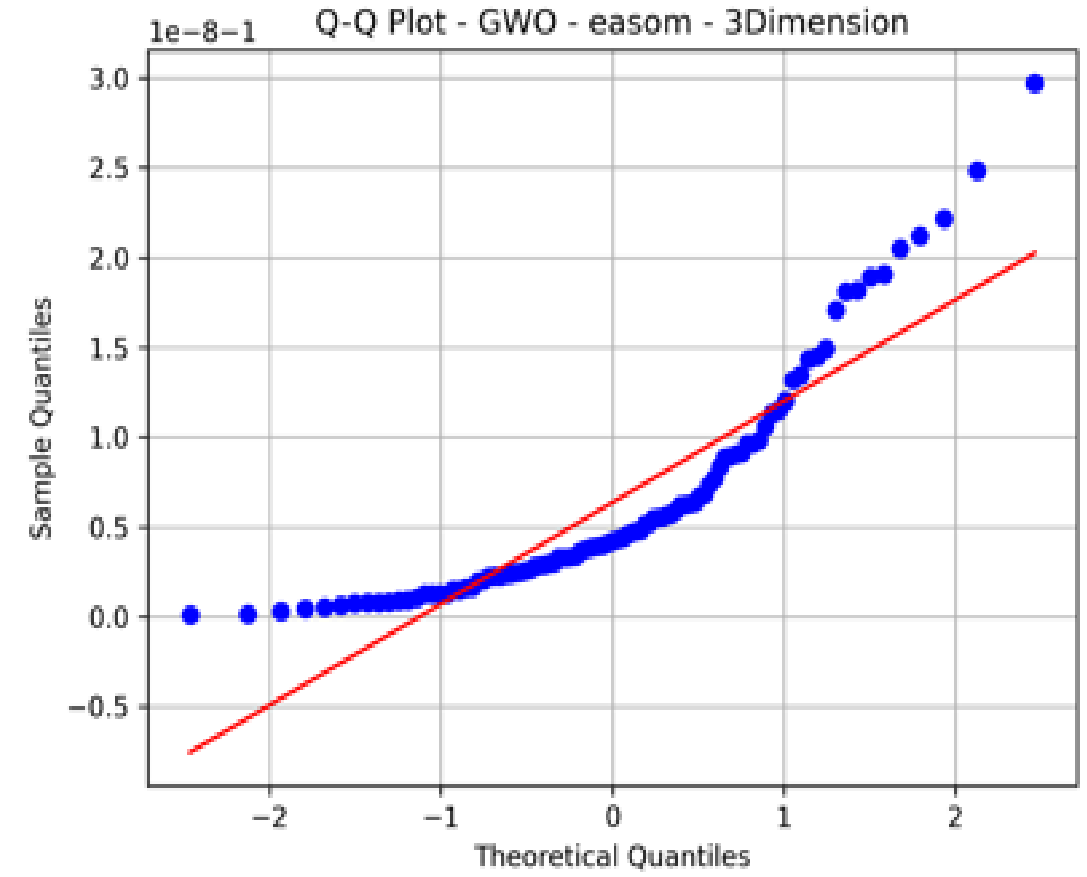
Q-Q Plot - GWO - easom - 3Dimension
1e−8−1
3.0
2.5
2.0
1.5
1.0
0.5
0.0
−0.5
−2
−1
0
1
2
Sample Quantiles
Theoretical Quantiles

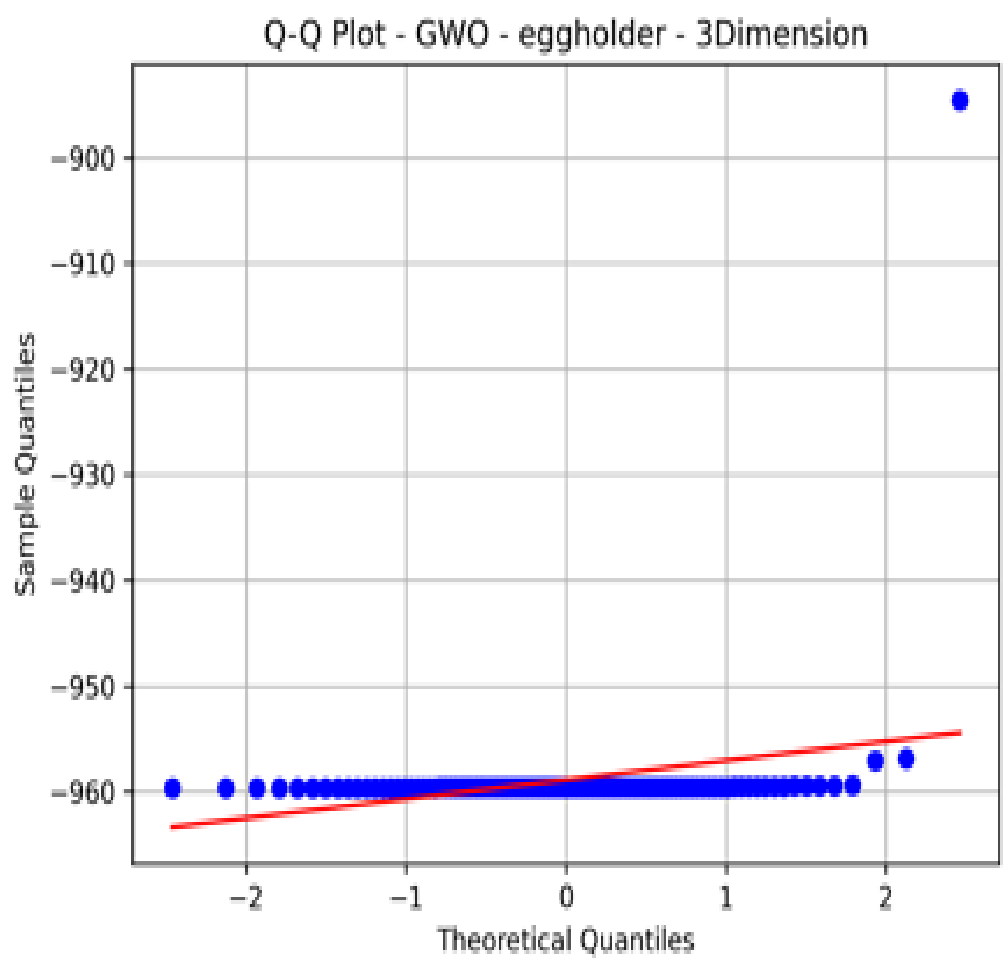
Q-Q Plot - GWO - eggholder - 3Dimension
−900
−910
−920
−930
−940
−950
−960
−2
−1
0
1
2
Sample Quantiles
Theoretical Quantiles

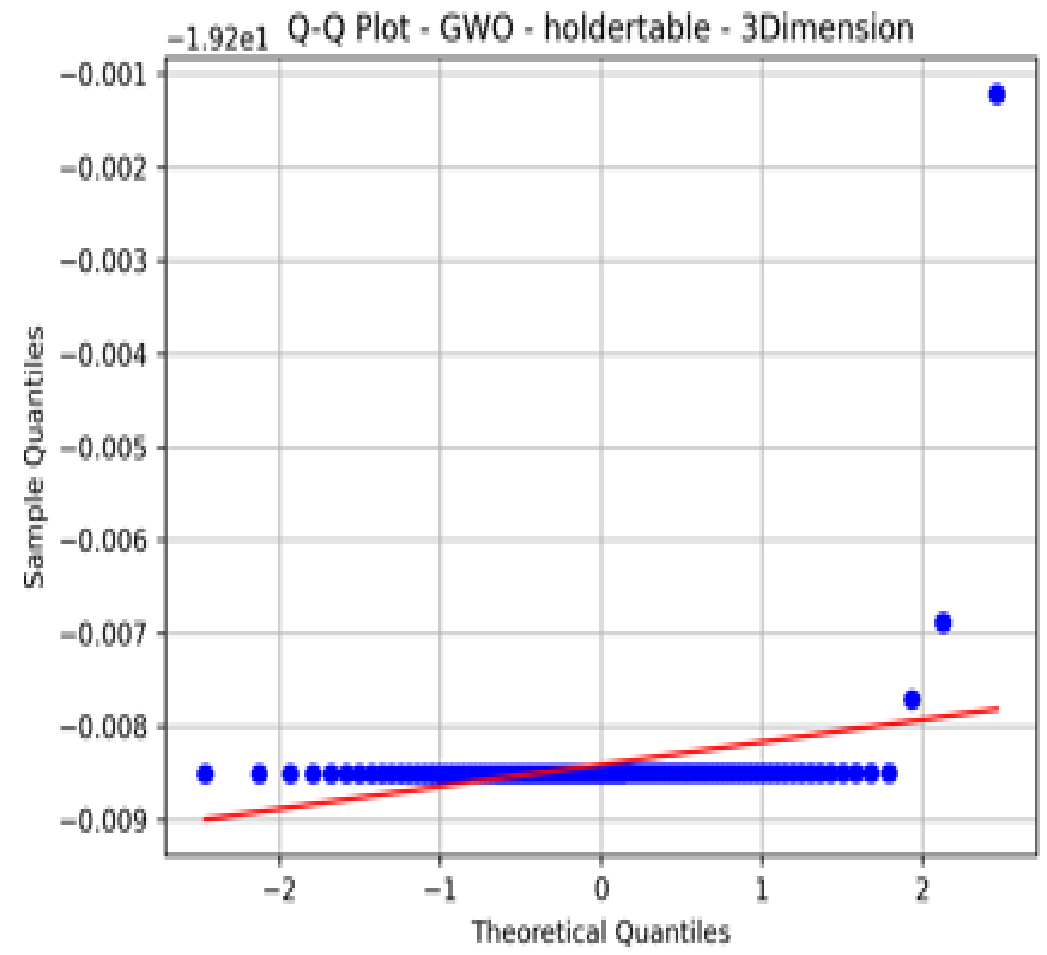
Q-Q Plot - GWO - holdertable - 3Dimension
−1.92e1
−0.001
−0.002
−0.003
−0.004
−0.005
−0.006
−0.007
−0.008
−0.009
−2
−1
0
1
2
Sample Quantiles
Theoretical Quantiles

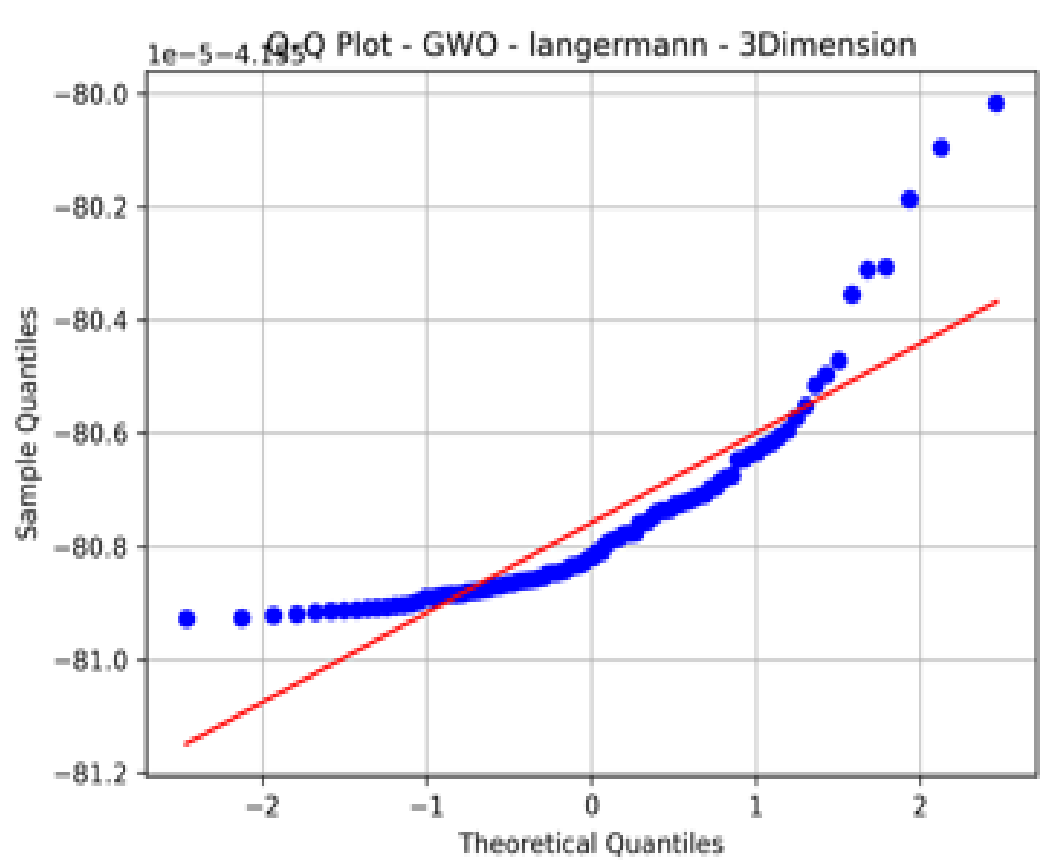
Q-Q Plot - GWO - langermann - 3Dimension
1e−5−4.155
−80.0
−80.2
−80.4
−80.6
−80.8
−81.0
−81.2
−2
−1
0
1
2
Sample Quantiles
Theoretical Quantiles

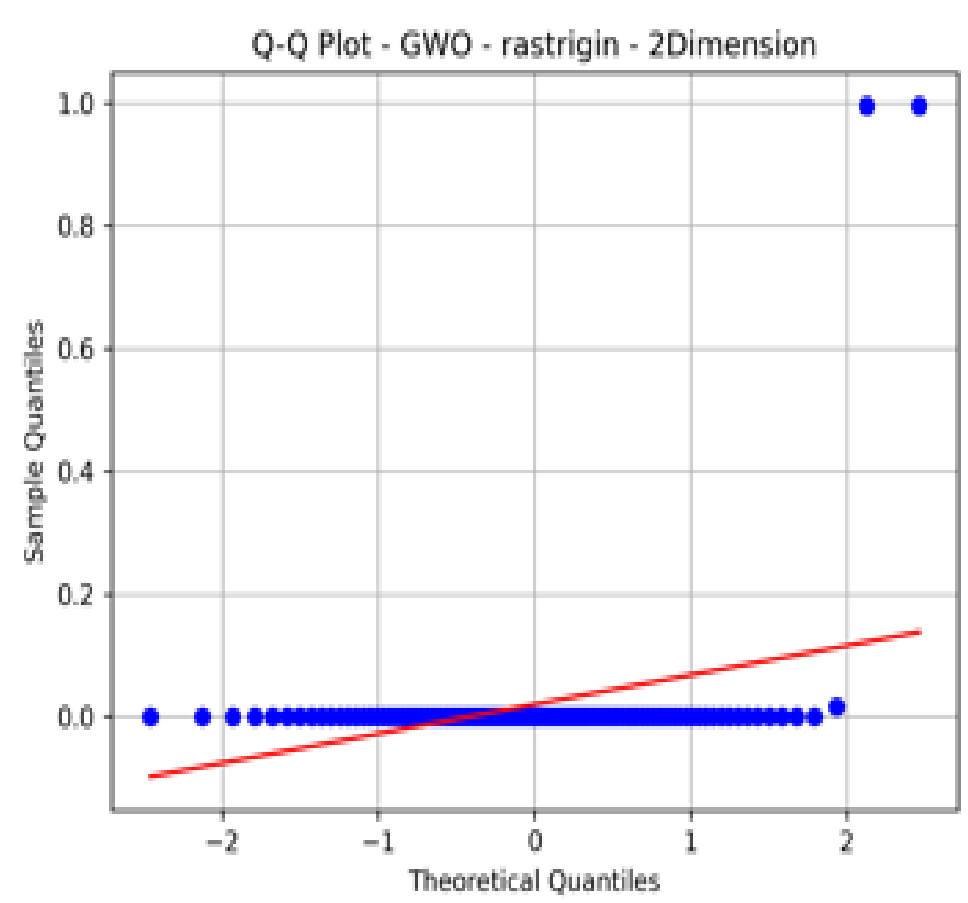
Q-Q Plot - GWO - rastrigin - 2Dimension
1.0
0.8
0.6
0.4
0.2
0.0
−2
−1
0
1
2
Sample Quantiles
Theoretical Quantiles

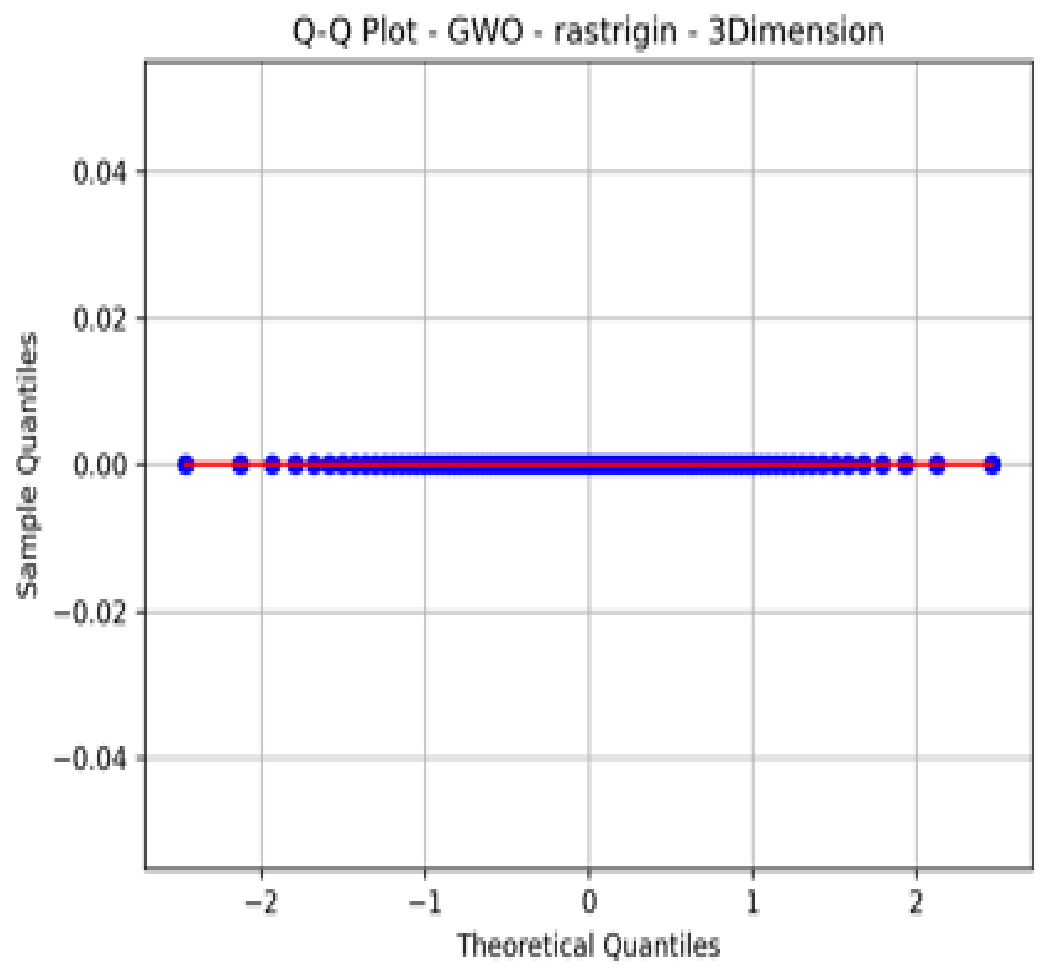
Q-Q Plot - GWO - rastrigin - 3Dimension
Sample Quantiles
Theoretical Quantiles

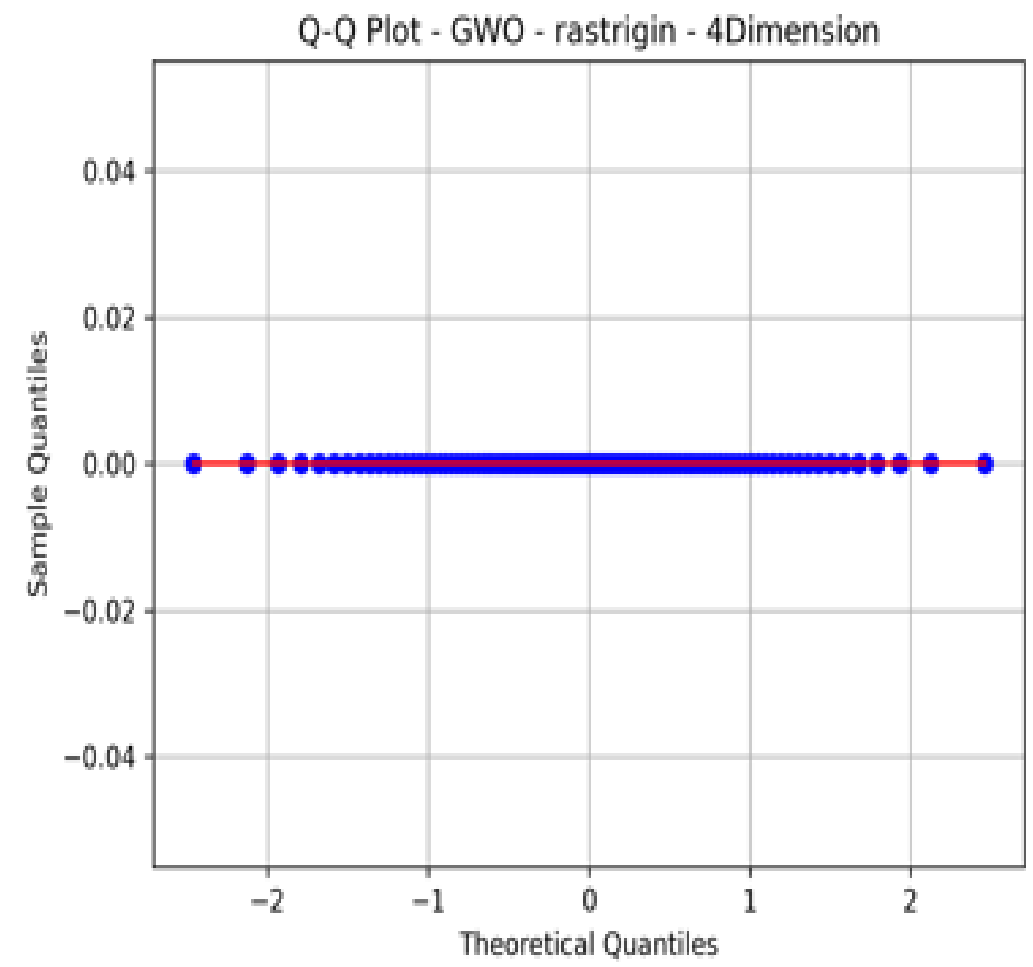
Q-Q Plot - GWO - rastrigin - 4Dimension
Sample Quantiles
Theoretical Quantiles

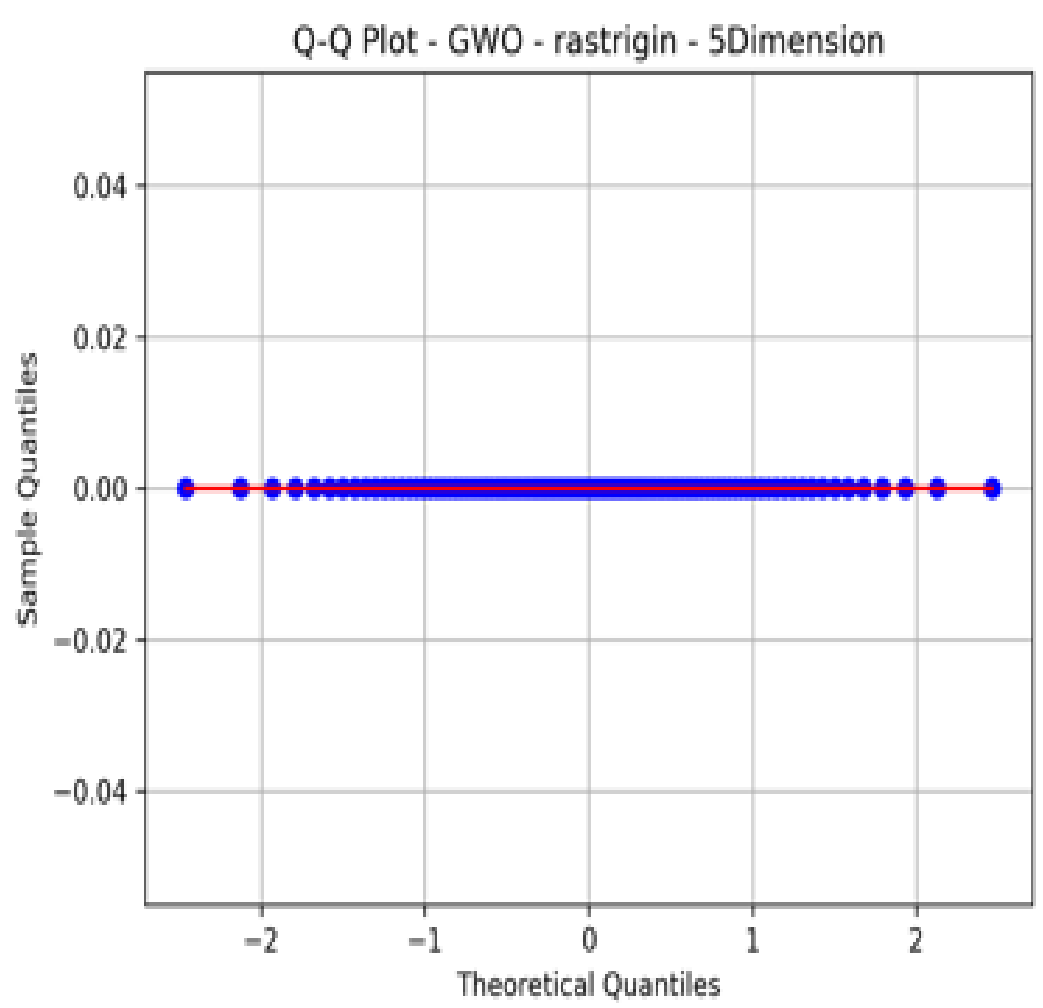
Q-Q Plot - GWO - rastrigin - 5Dimension
Sample Quantiles
Theoretical Quantiles

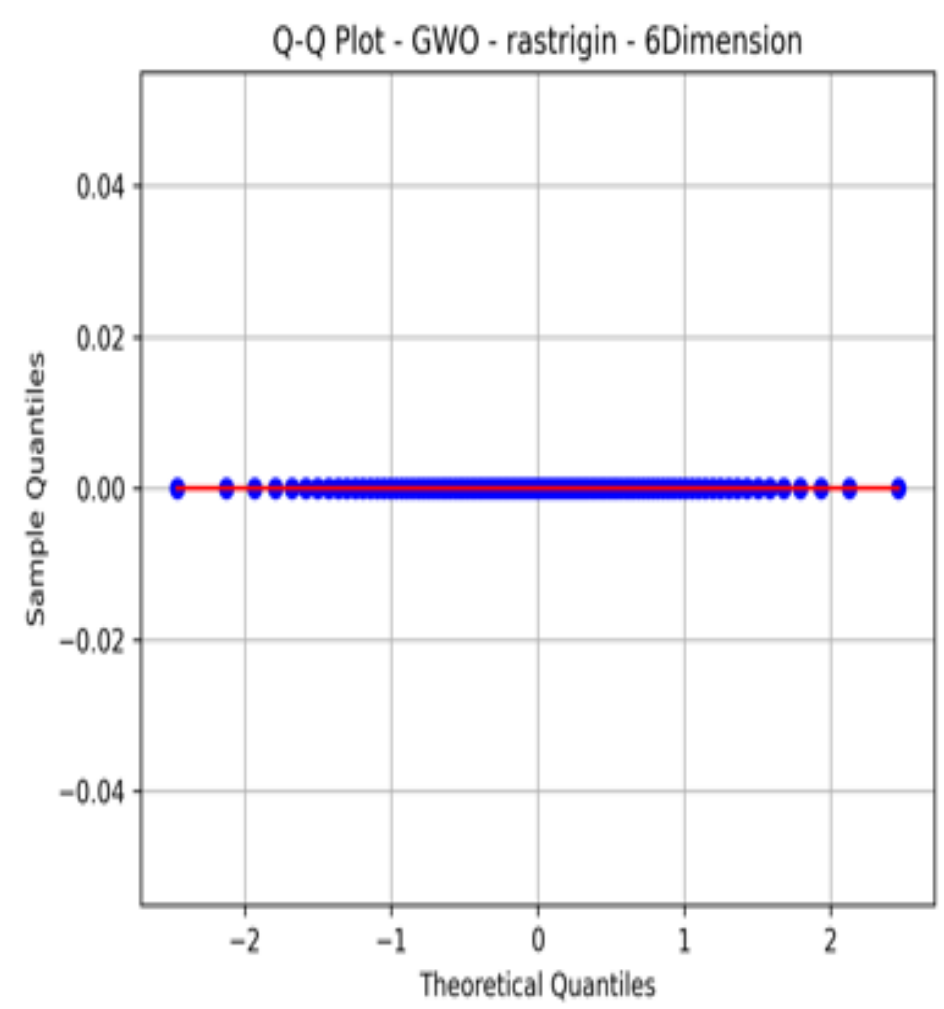
Q-Q Plot - GWO - rastrigin - 6Dimension
Sample Quantiles
Theoretical Quantiles

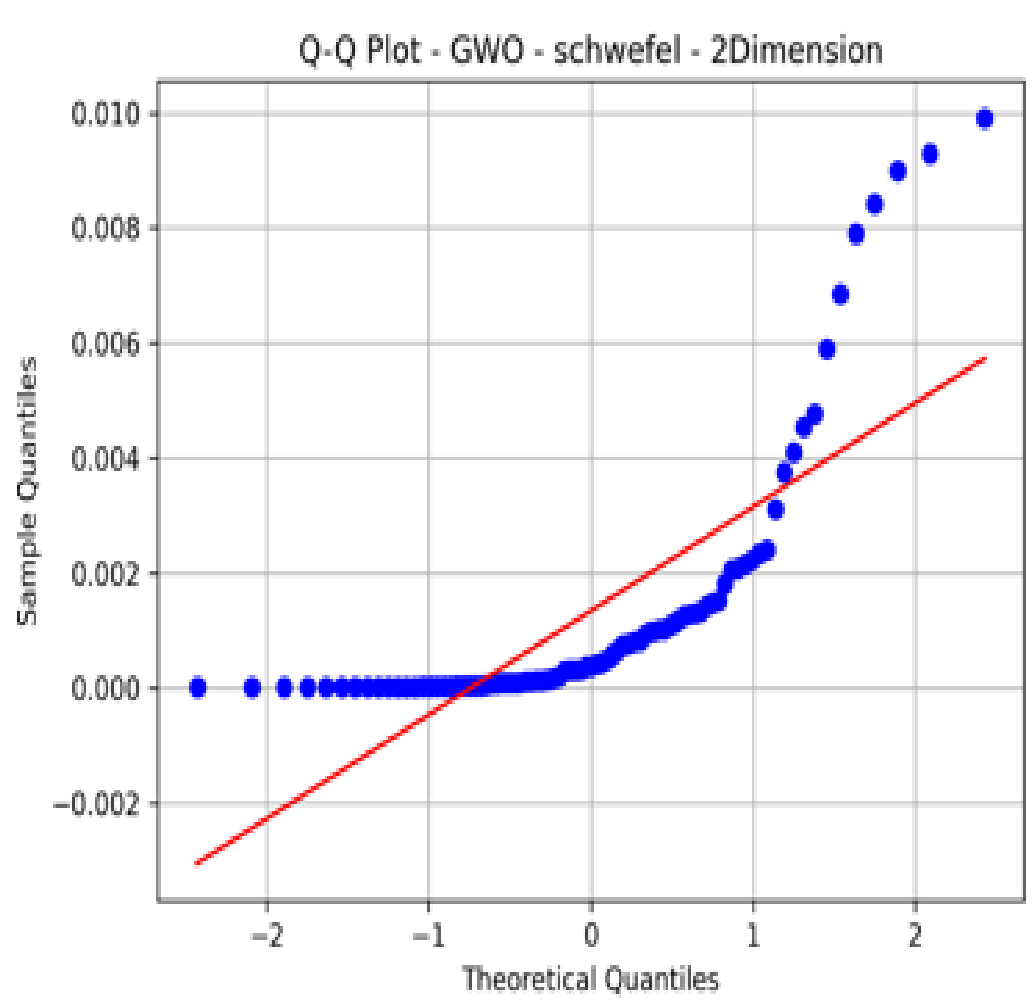
Q-Q Plot - GWO - schwefel - 2Dimension
Sample Quantiles
Theoretical Quantiles

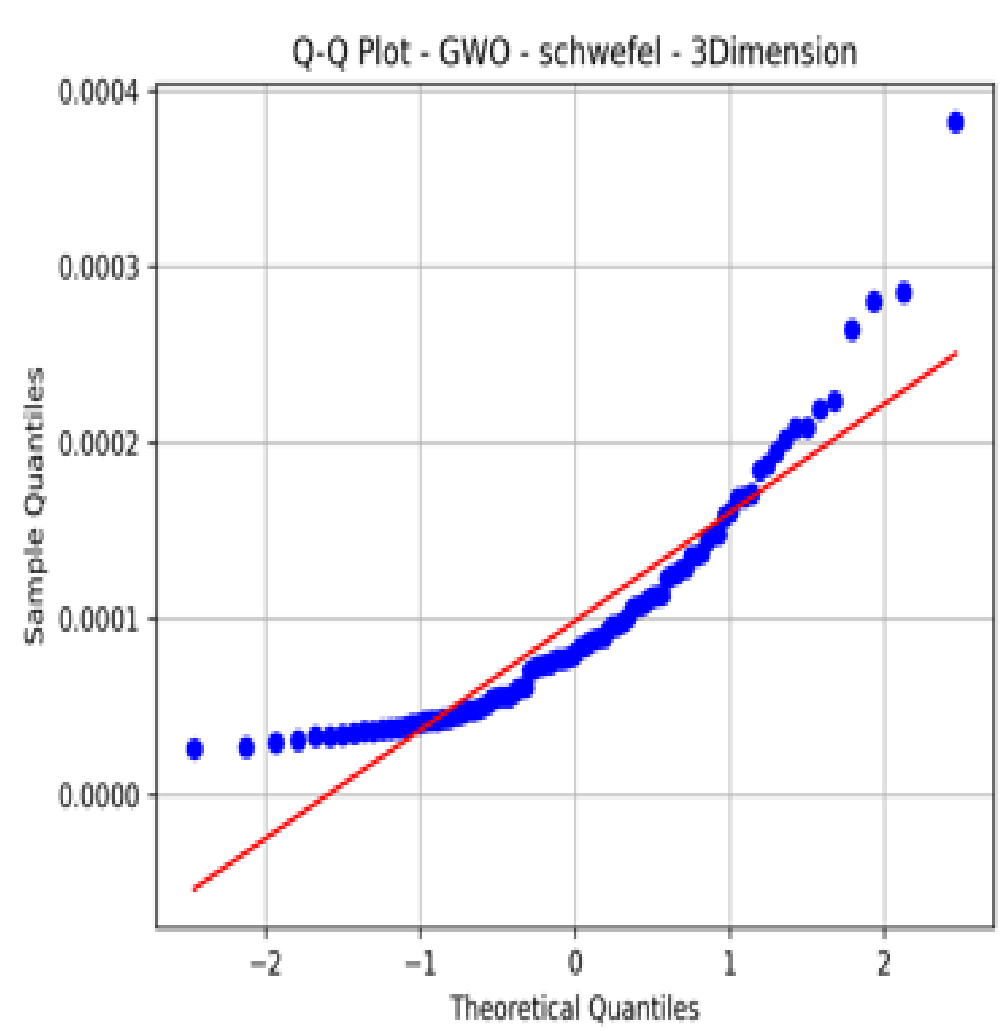
Q-Q Plot - GWO - schwefel - 3Dimension
Sample Quantiles
Theoretical Quantiles

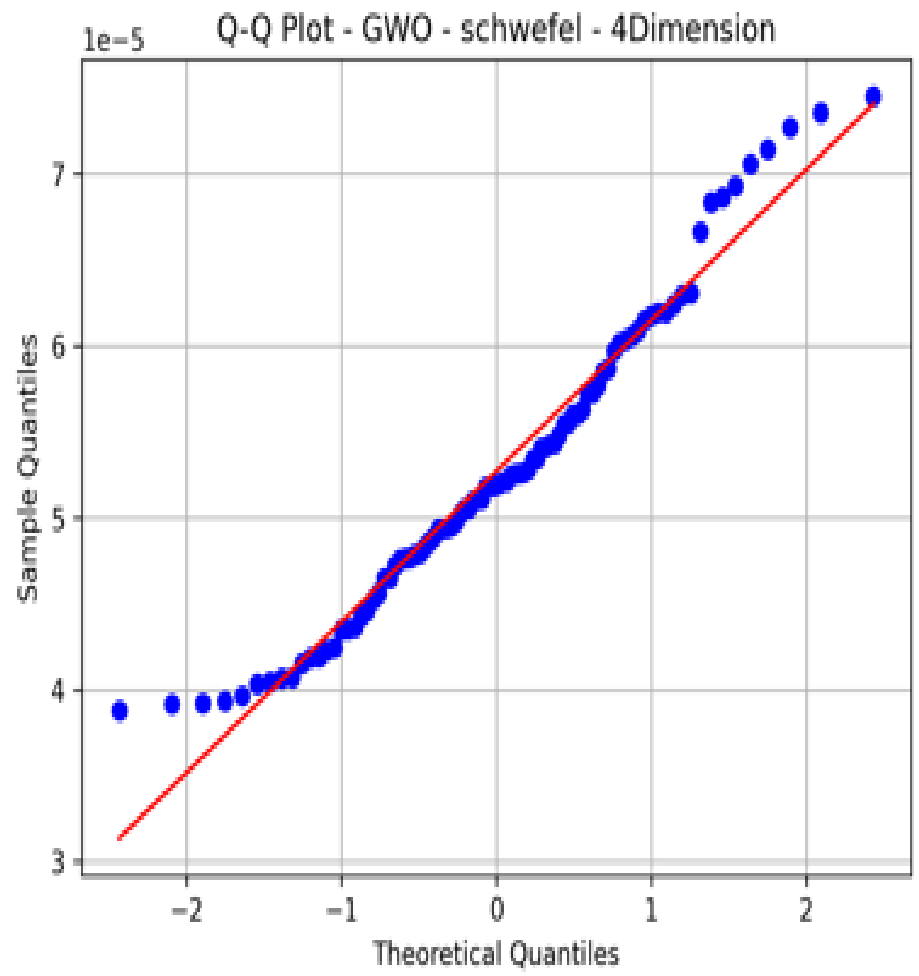
Q-Q Plot - GWO - schwefel - 4Dimension
1e−5
Sample Quantiles
Theoretical Quantiles

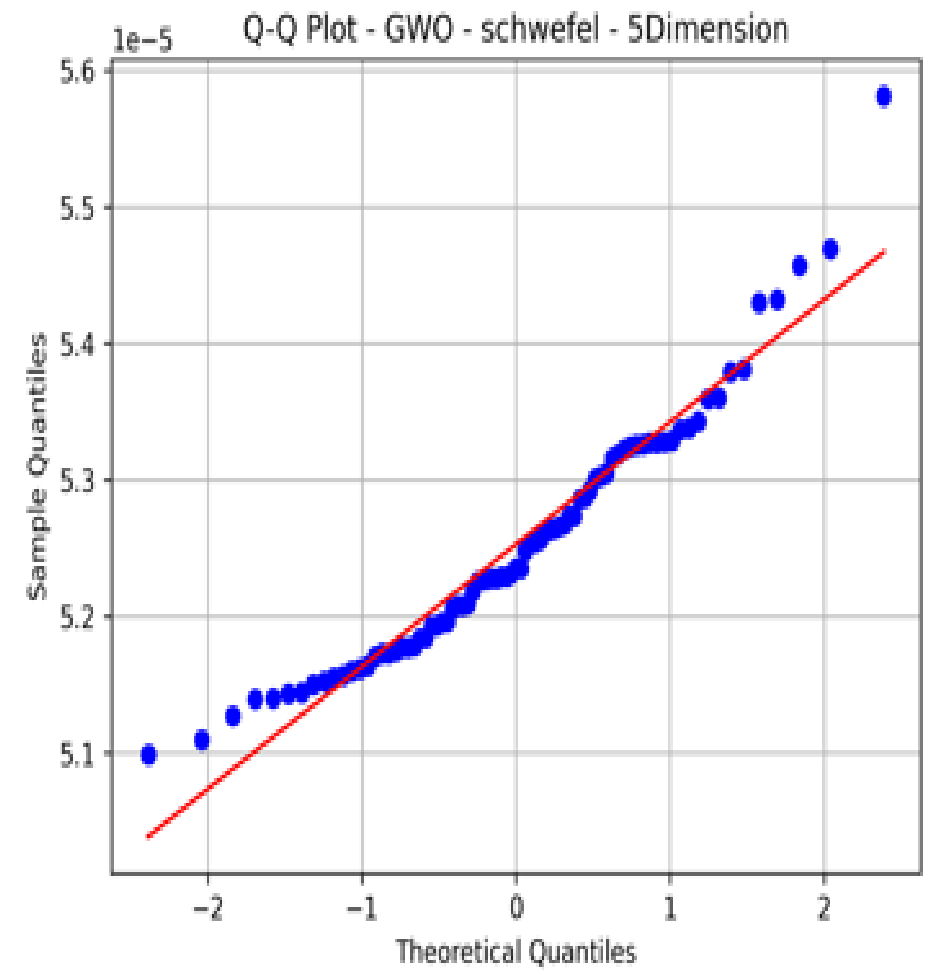
Q-Q Plot - GWO - schwefel - 5Dimension
1e−5
Sample Quantiles
Theoretical Quantiles

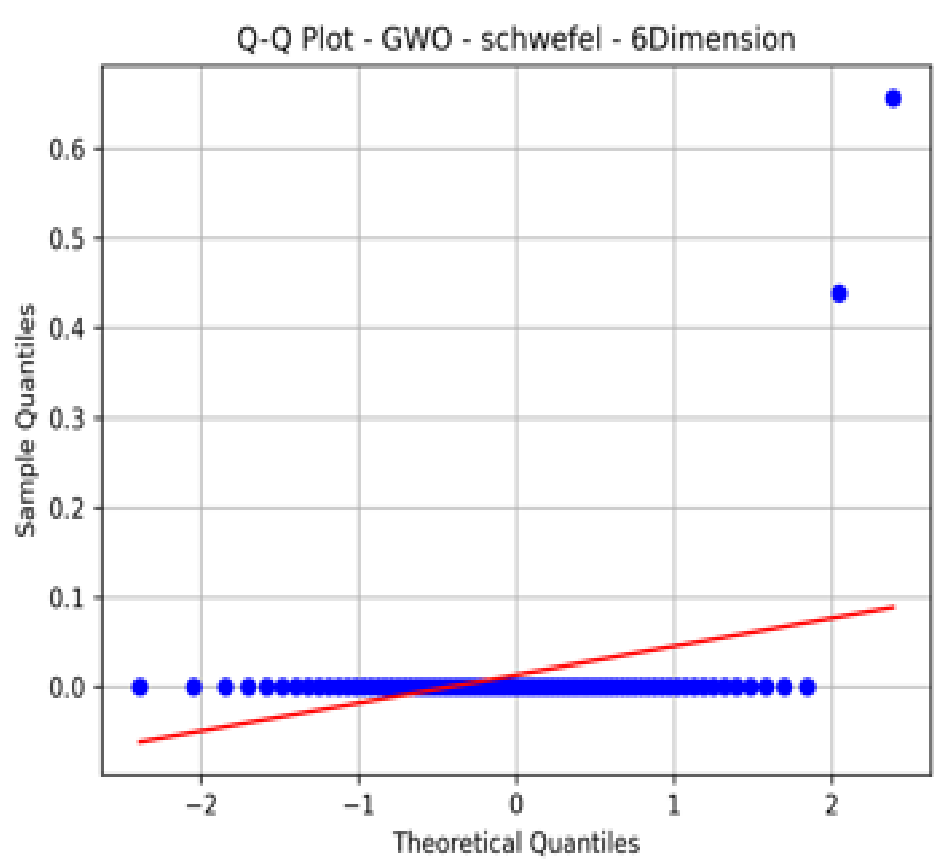
Q-Q Plot - GWO - schwefel - 6Dimension
Sample Quantiles
Theoretical Quantiles

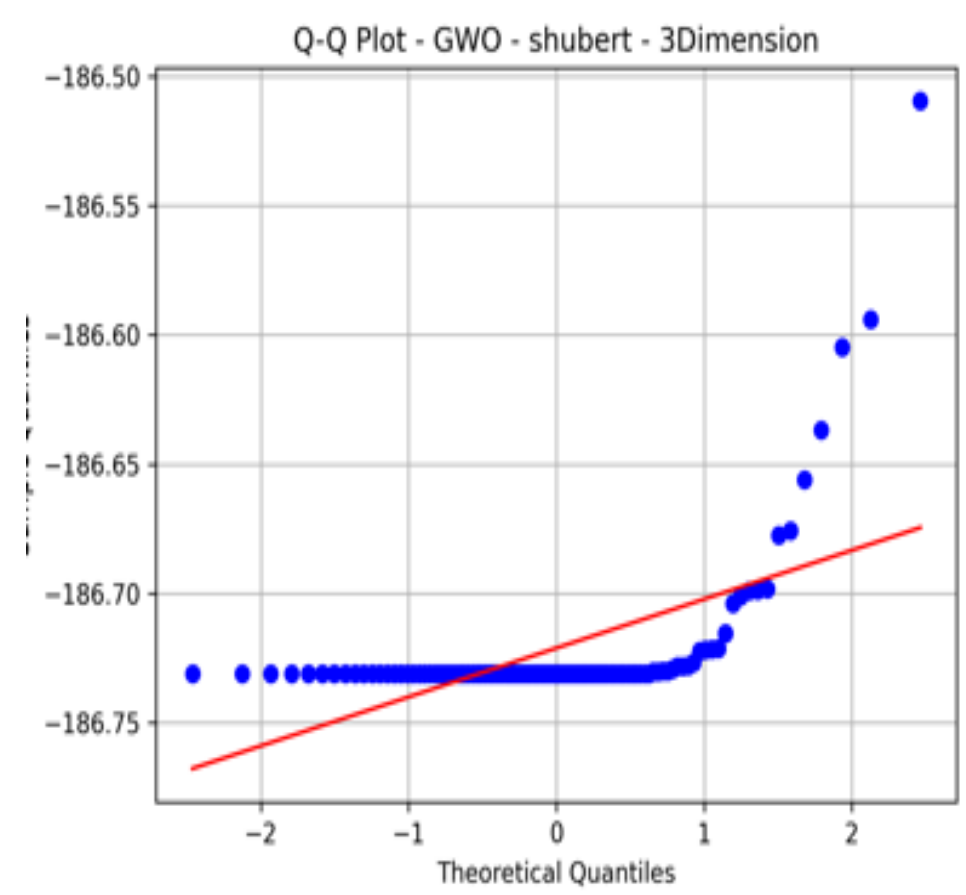
Q-Q Plot - GWO - shubert - 3Dimension
Theoretical Quantiles

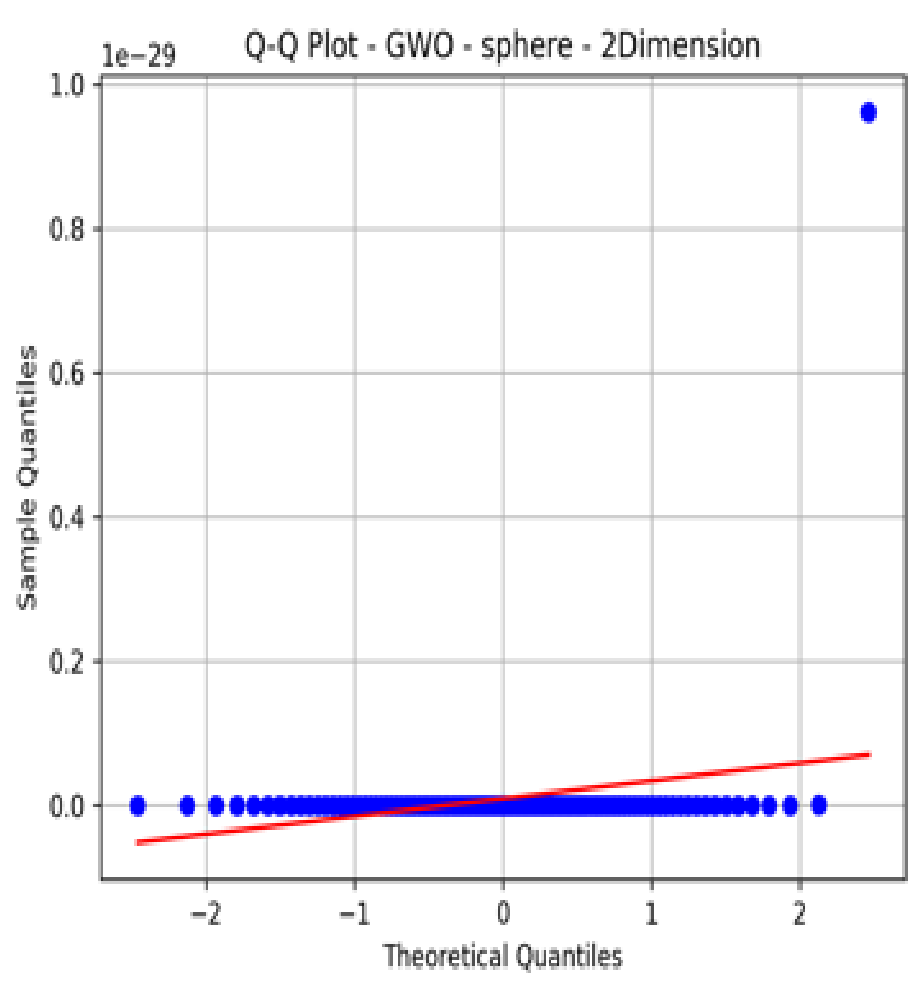
Q-Q Plot - GWO - sphere - 2Dimension
1e−29
Sample Quantiles
Theoretical Quantiles

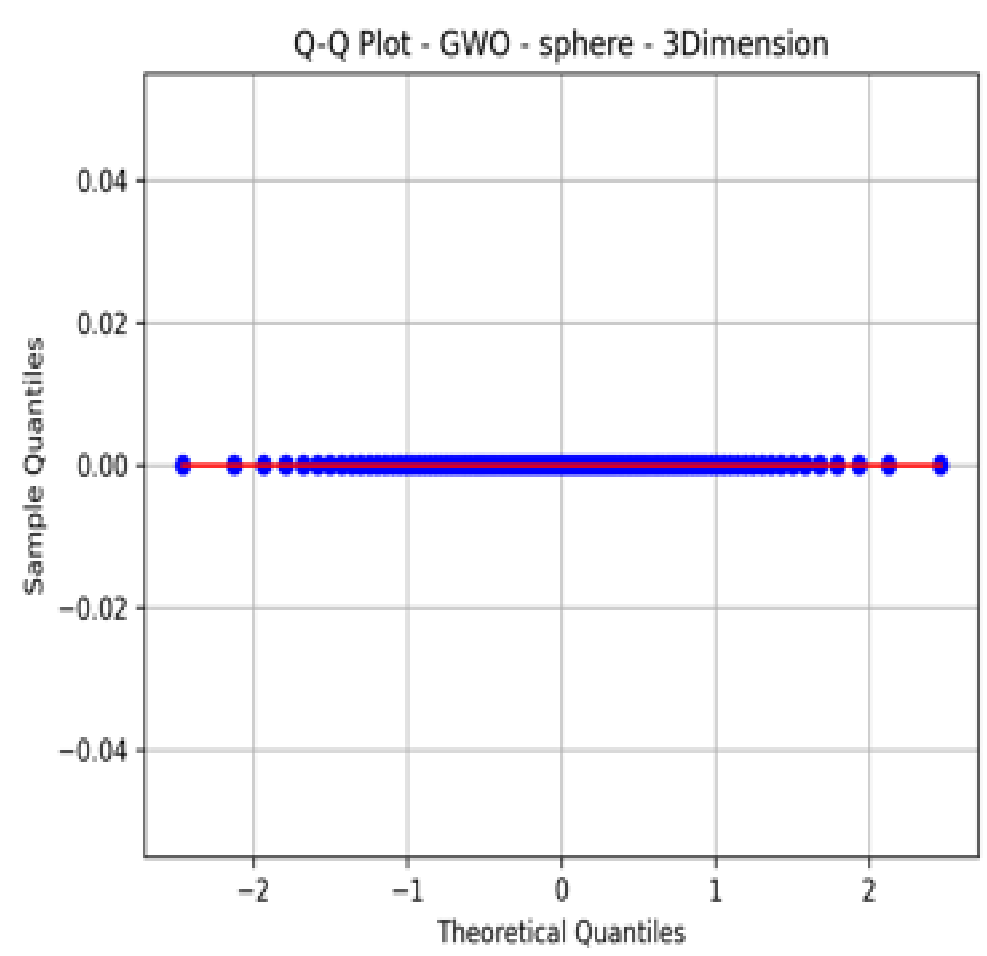
Q-Q Plot - GWO - sphere - 3Dimension
Sample Quantiles
Theoretical Quantiles

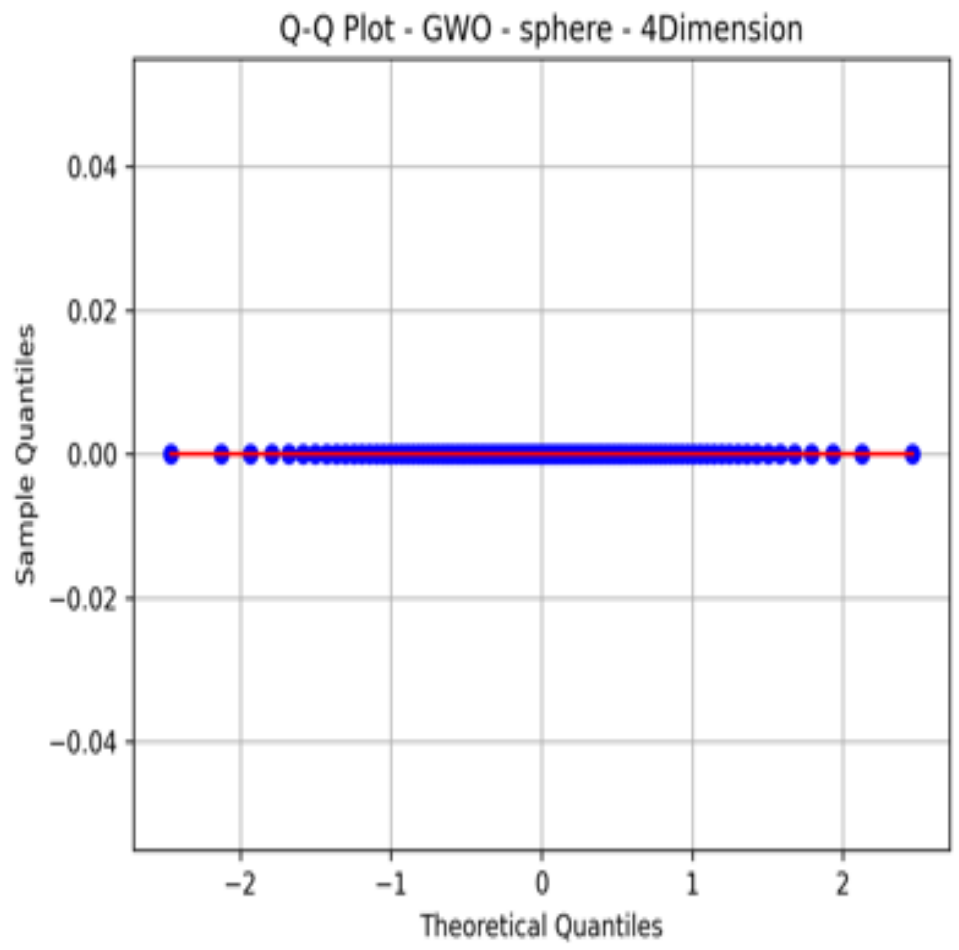

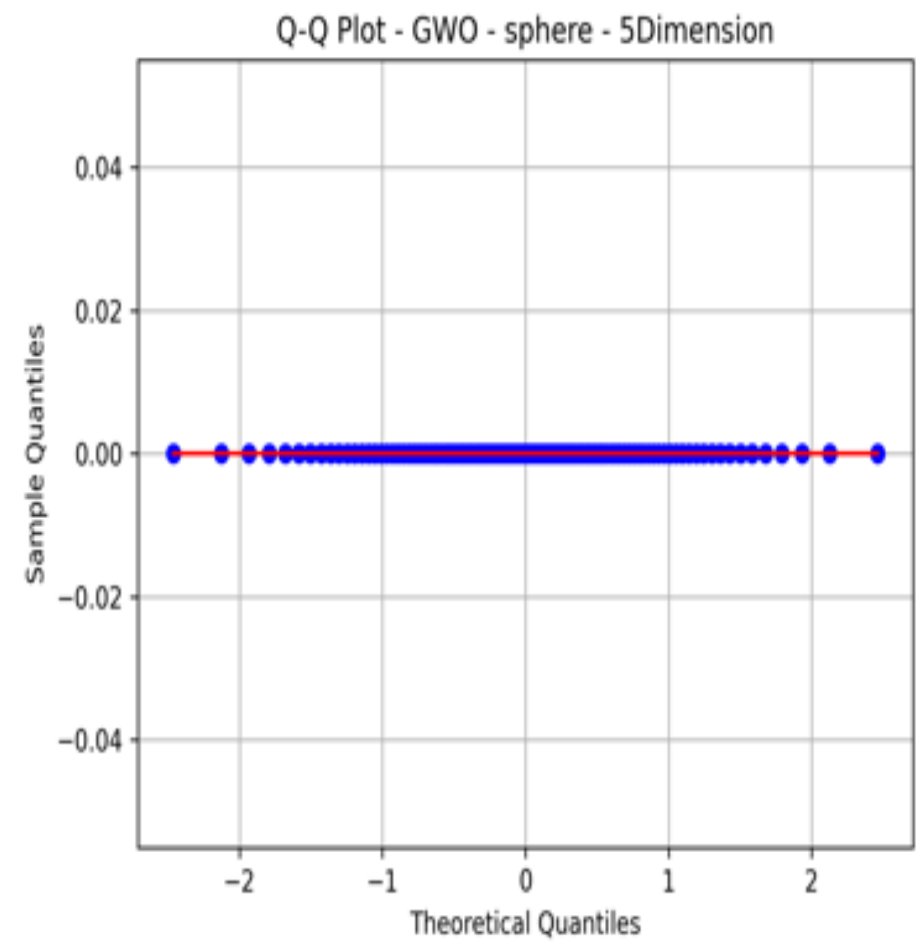

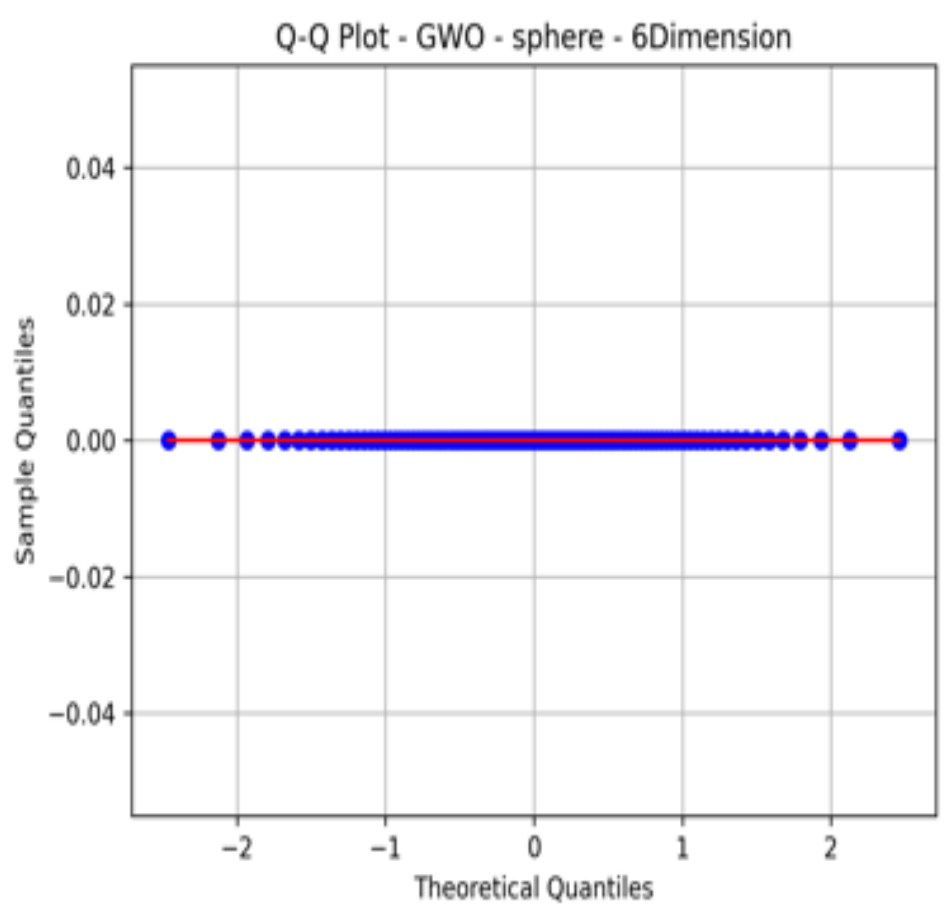

Non-Continuous functions QQ Plots

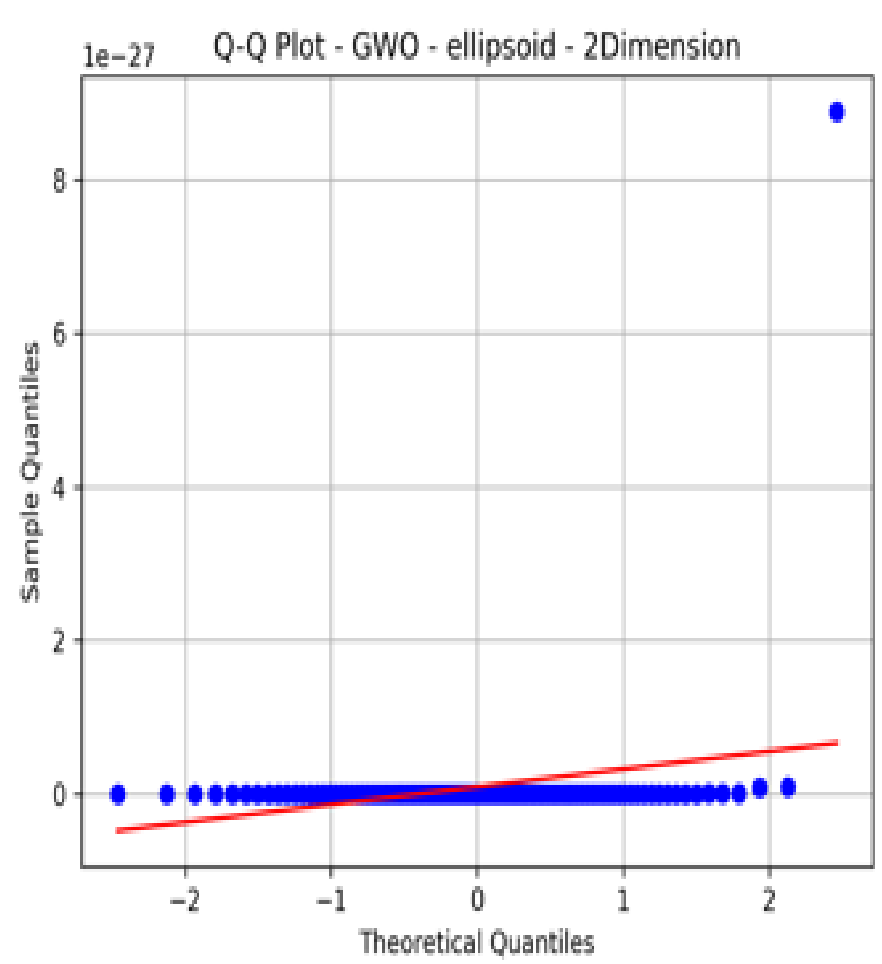

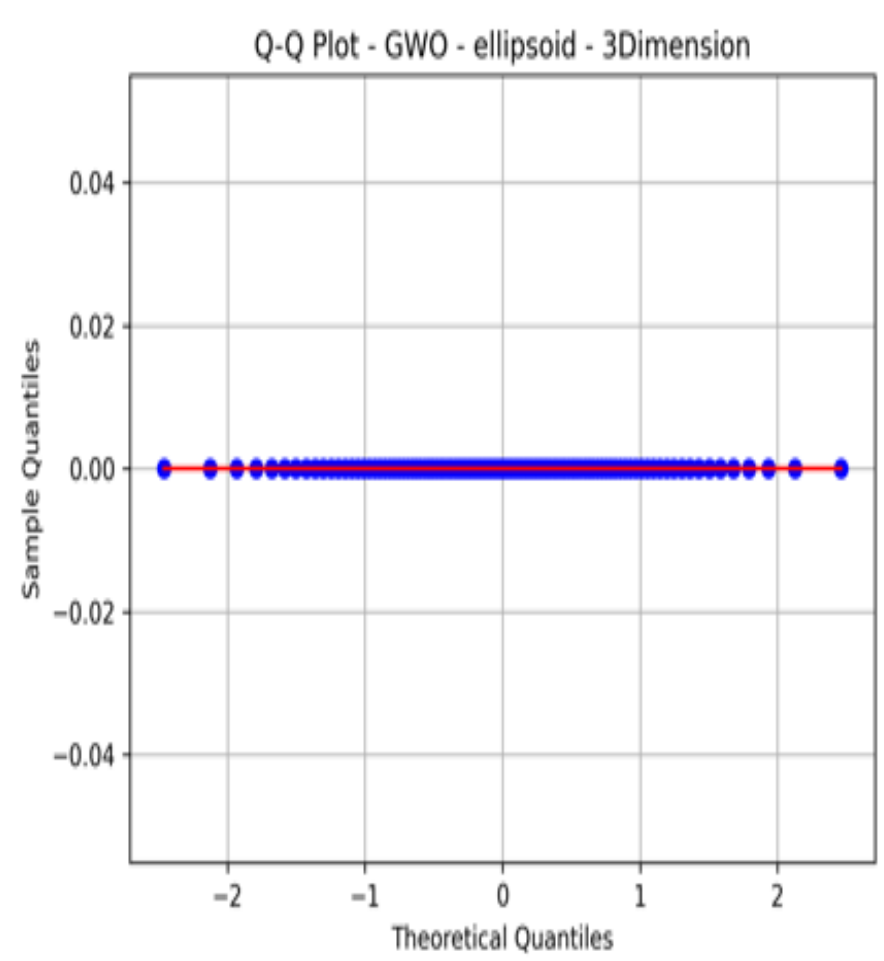

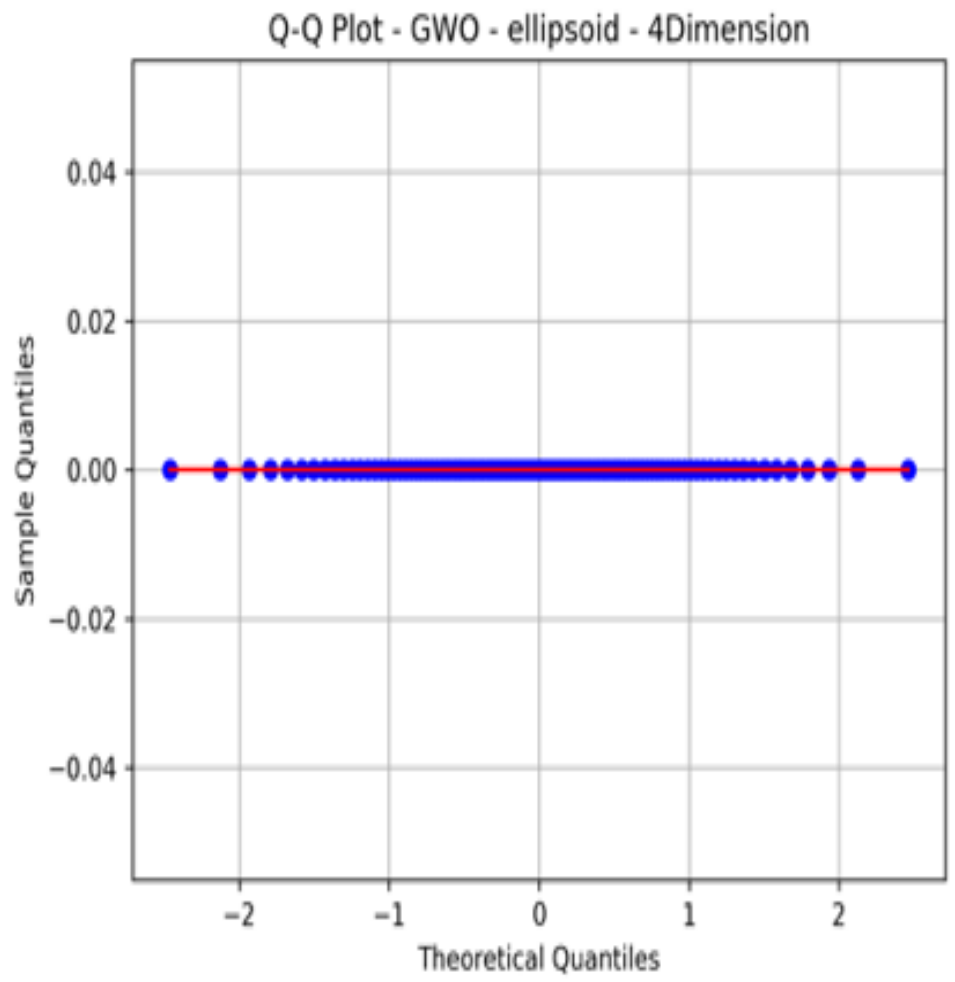
Q-Q Plot - GWO - ellipsoid - 4Dimension
Sample Quantiles
Theoretical Quantiles

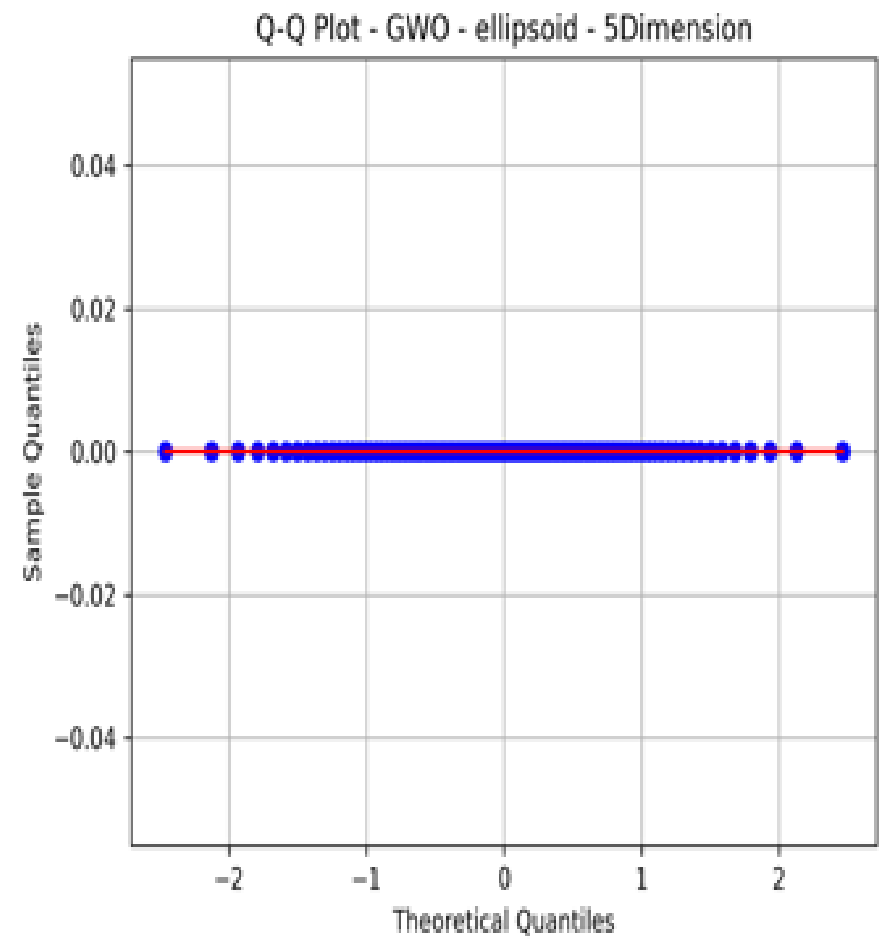
Q-Q Plot - GWO - ellipsoid - 5Dimension
Sample Quantiles
Theoretical Quantiles

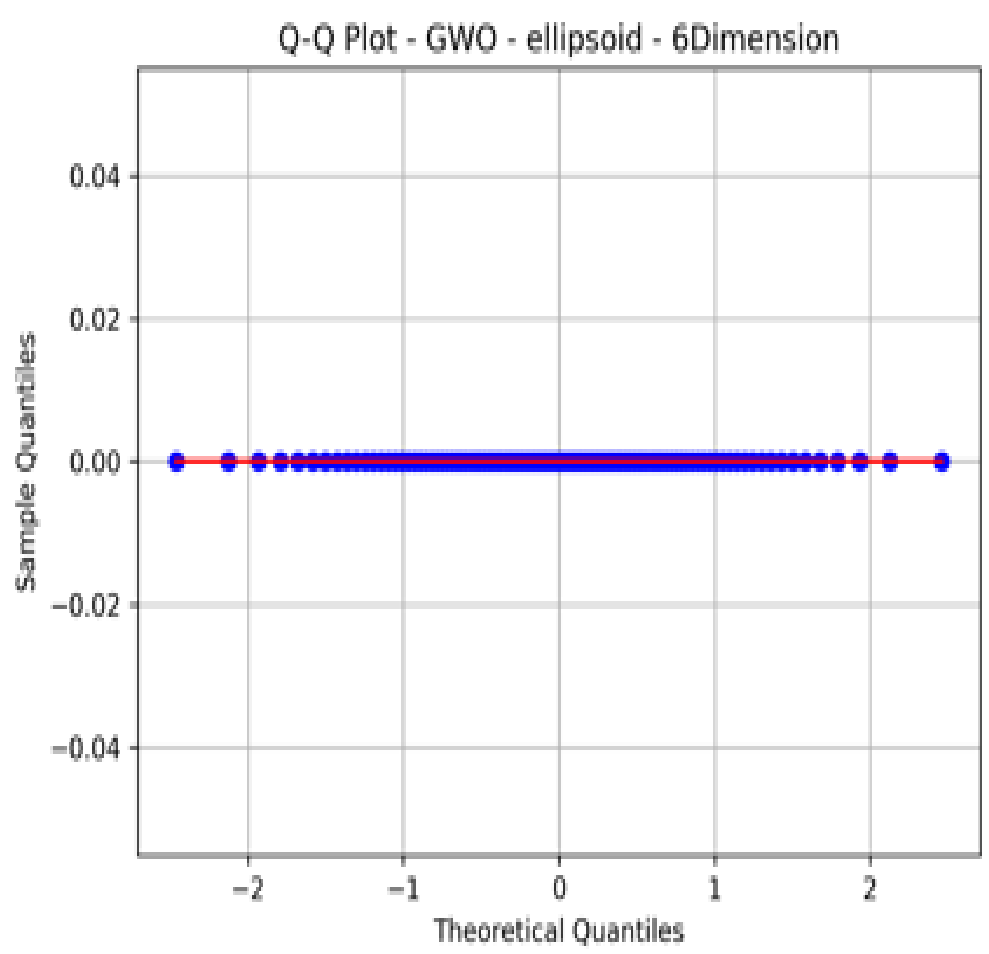
Q-Q Plot - GWO - ellipsoid - 6Dimension
Sample Quantiles
Theoretical Quantiles

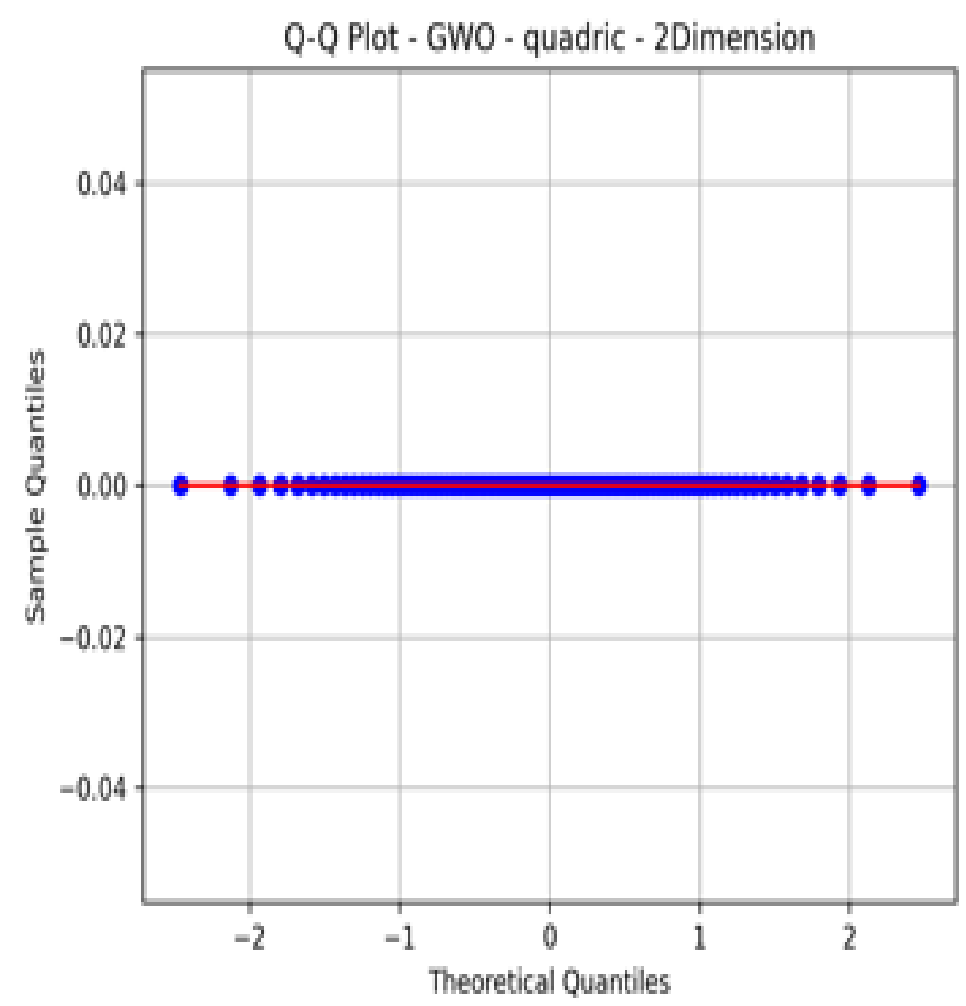
Q-Q Plot - GWO - quadric - 2Dimension
Sample Quantiles
Theoretical Quantiles

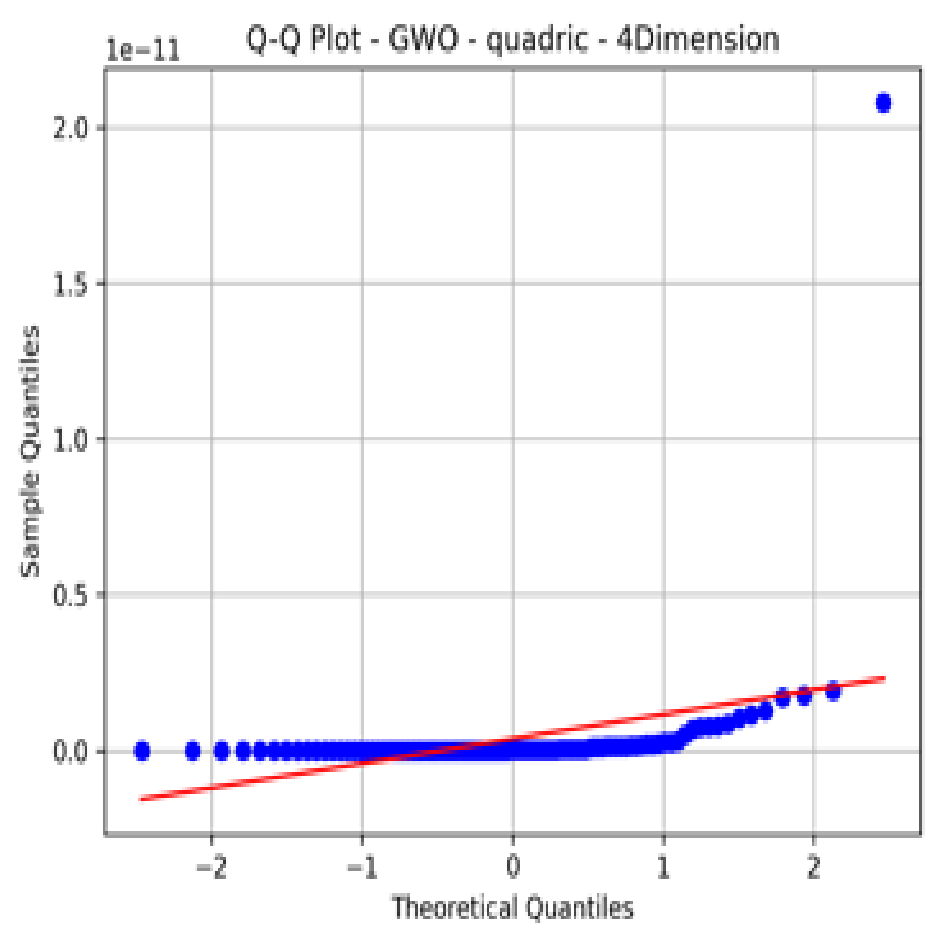
1e−11
Q-Q Plot - GWO - quadric - 4Dimension
Sample Quantiles
Theoretical Quantiles

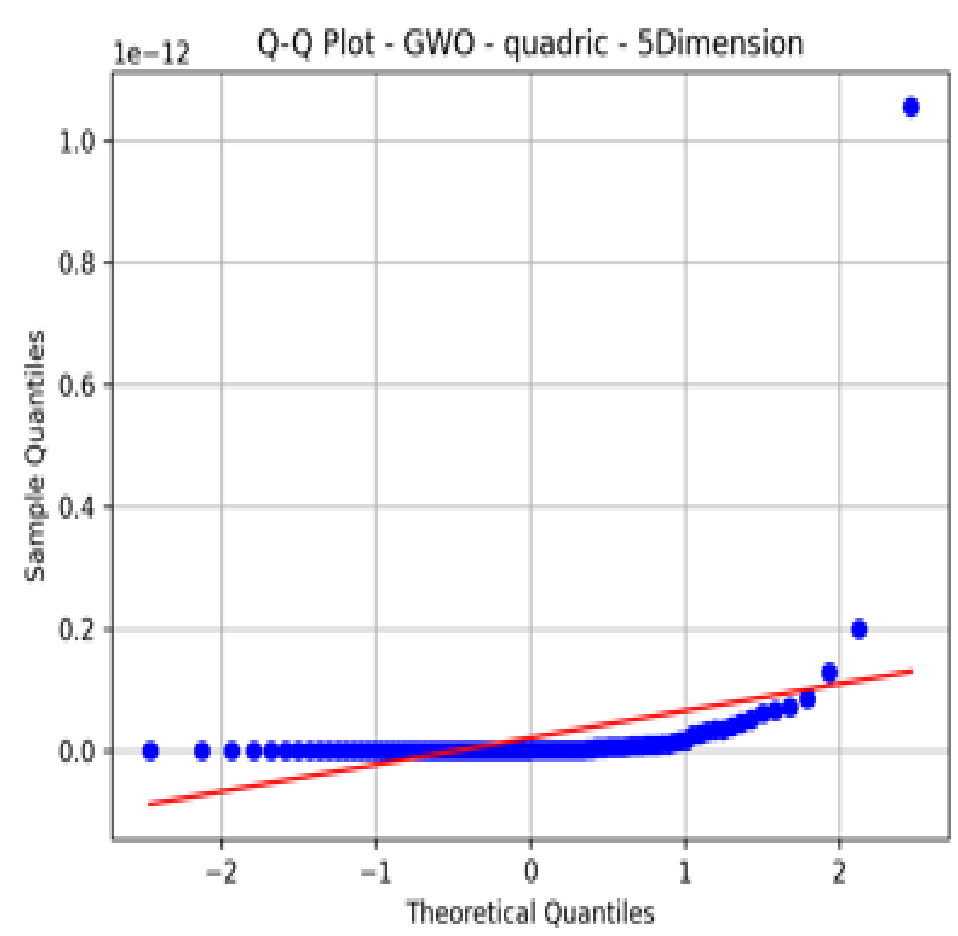
1e−12
Q-Q Plot - GWO - quadric - 5Dimension
Sample Quantiles
Theoretical Quantiles

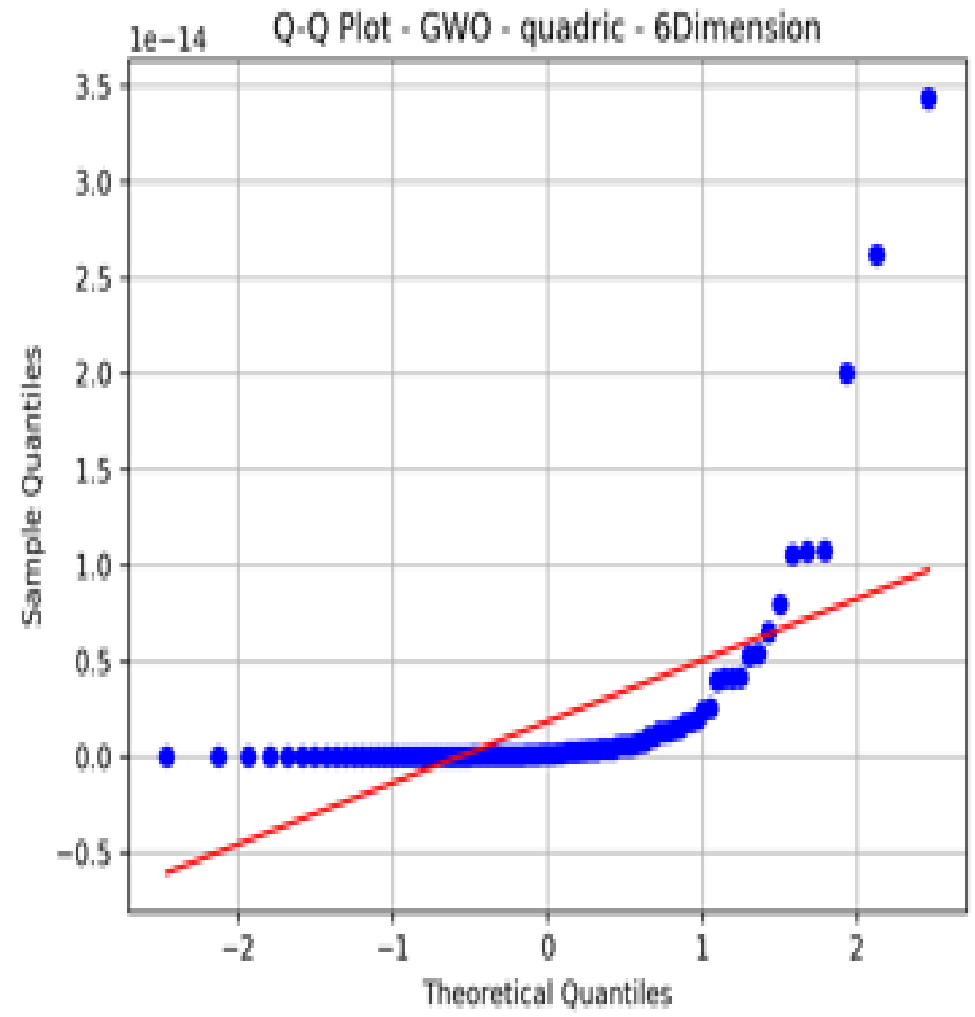
Q-Q Plot - GWO - quadric - 6Dimension
1e−14
Sample Quantiles
Theoretical Quantiles

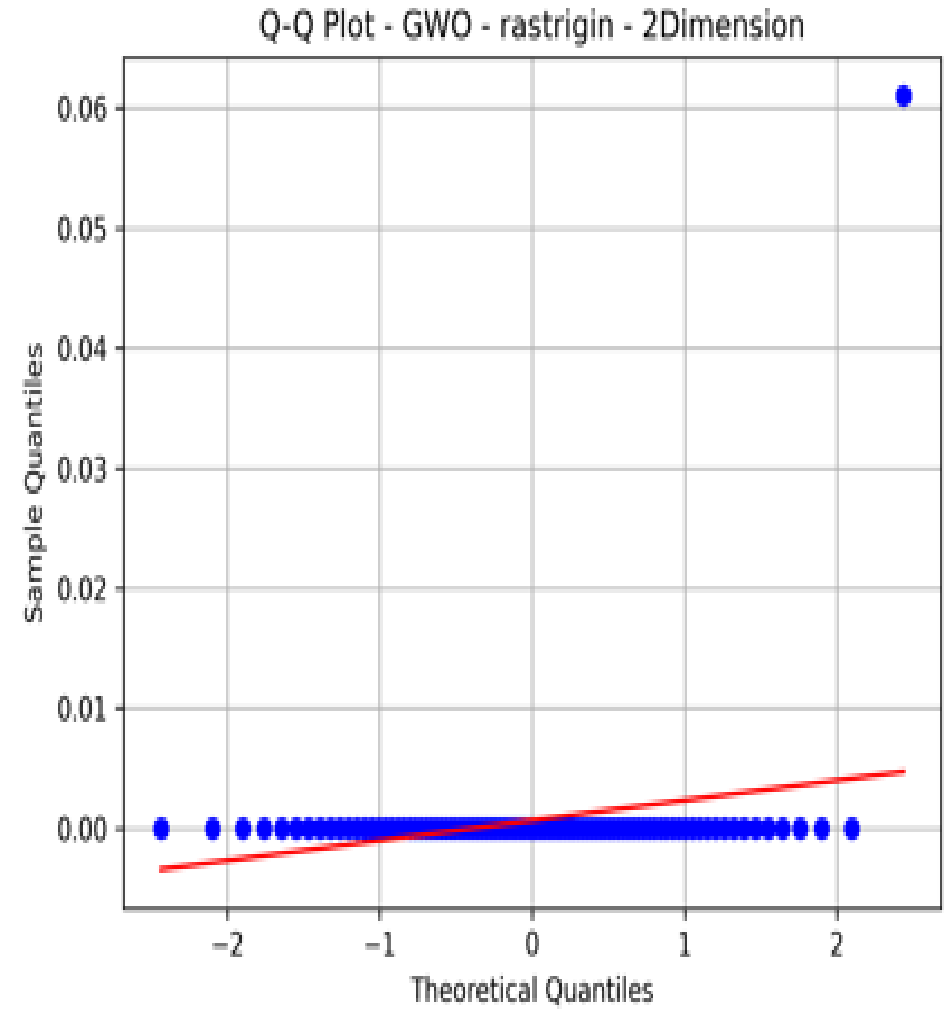
Q-Q Plot - GWO - rastrigin - 2Dimension
Sample Quantiles
Theoretical Quantiles

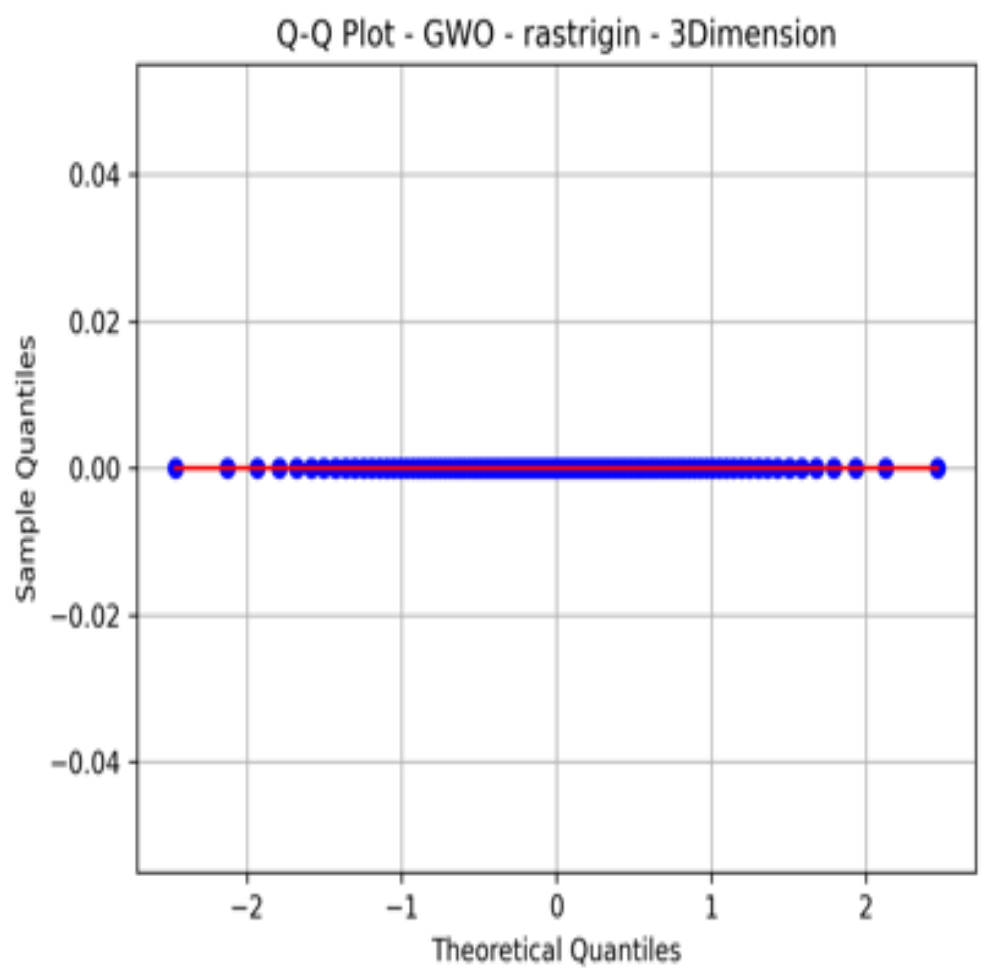
Q-Q Plot - GWO - rastrigin - 3Dimension
Sample Quantiles
Theoretical Quantiles

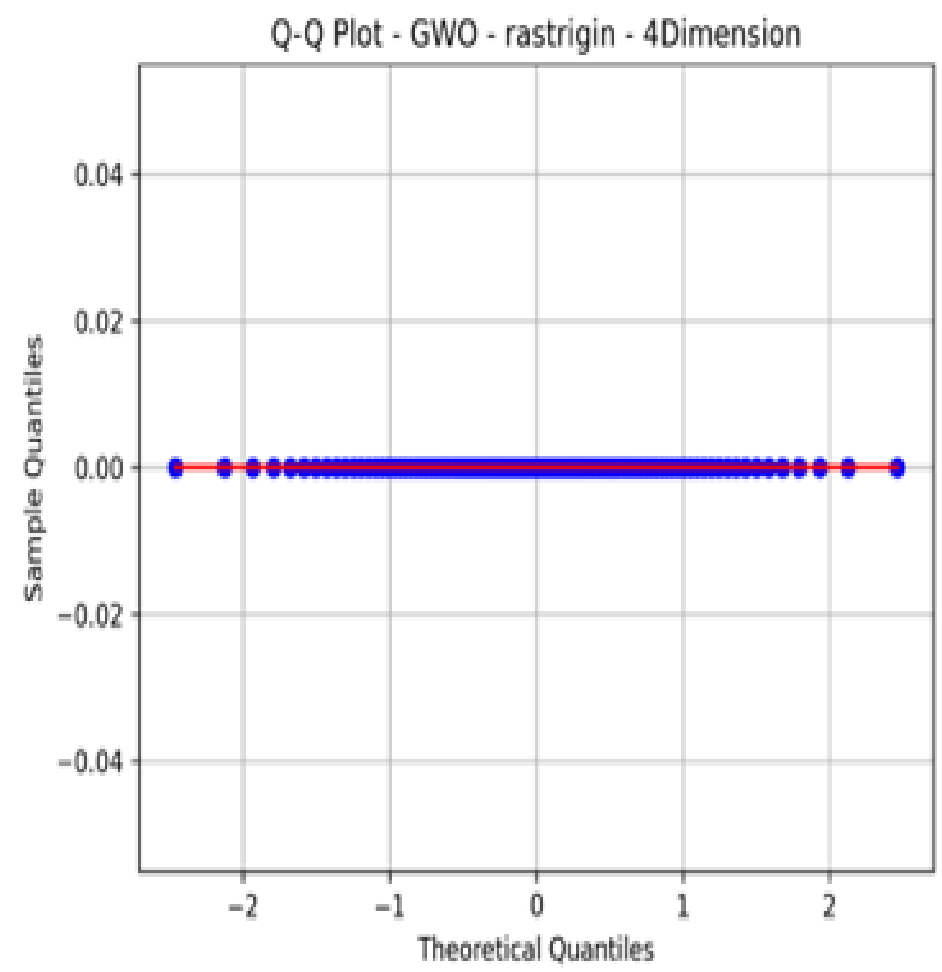
Q-Q Plot - GWO - rastrigin - 4Dimension
Sample Quantiles
Theoretical Quantiles

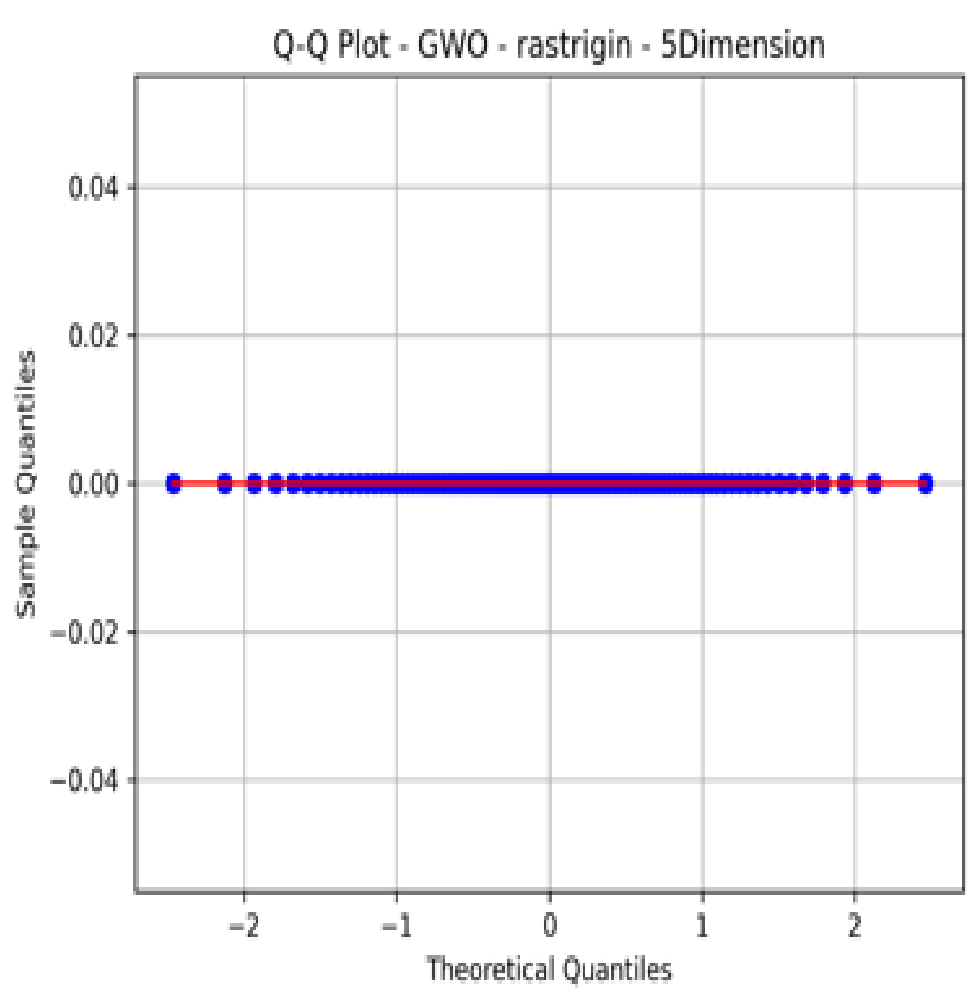
Q-Q Plot - GWO - rastrigin - 5Dimension
Sample Quantiles
Theoretical Quantiles

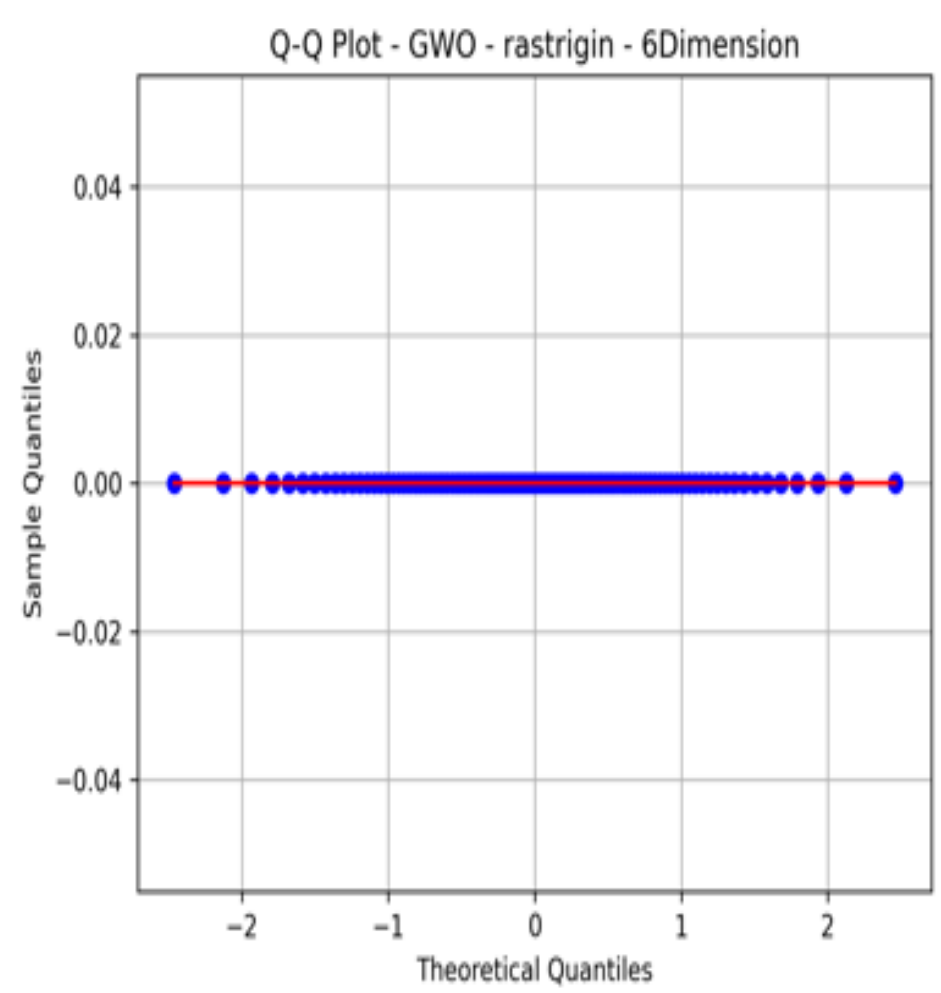
Q-Q Plot - GWO - rastrigin - 6Dimension
Sample Quantiles
Theoretical Quantiles

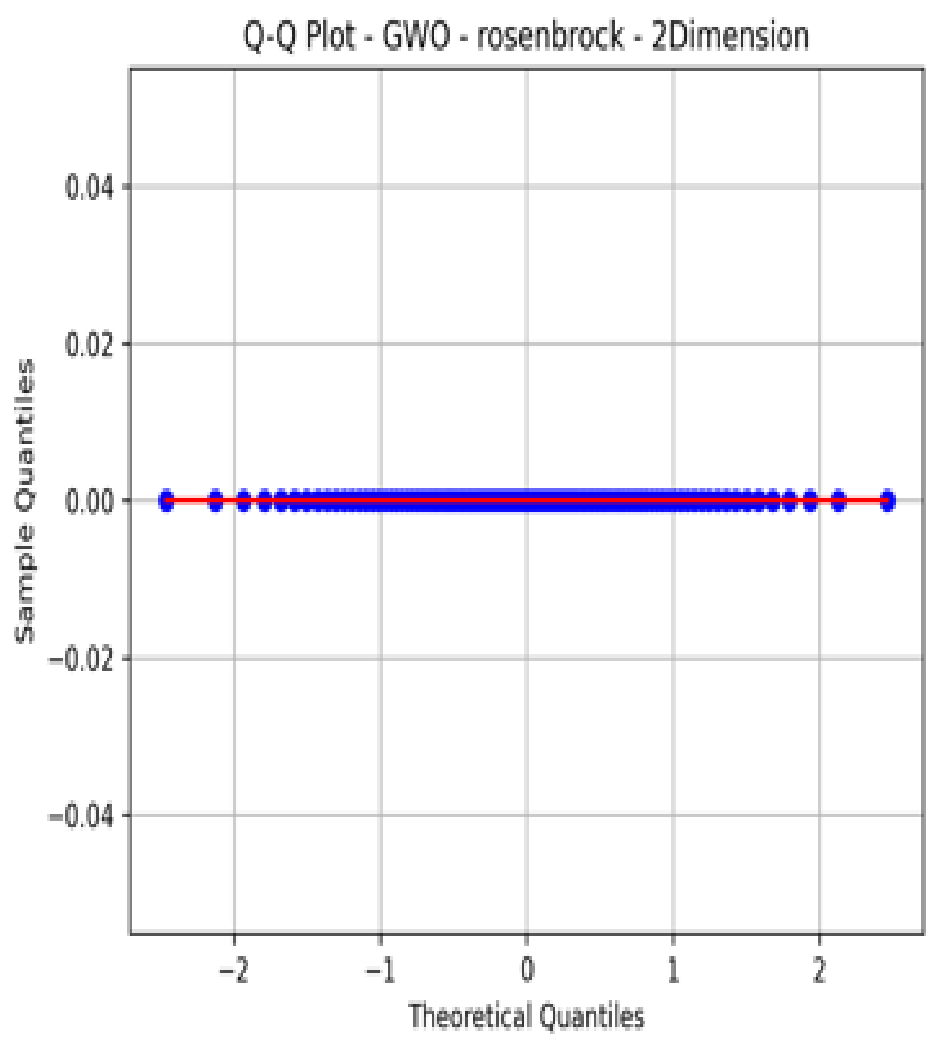
Q-Q Plot - GWO - rosenbrock - 2Dimension
Sample Quantiles
0.04
0.02
0.00
−0.02
−0.04
−2
−1
0
1
2
Theoretical Quantiles

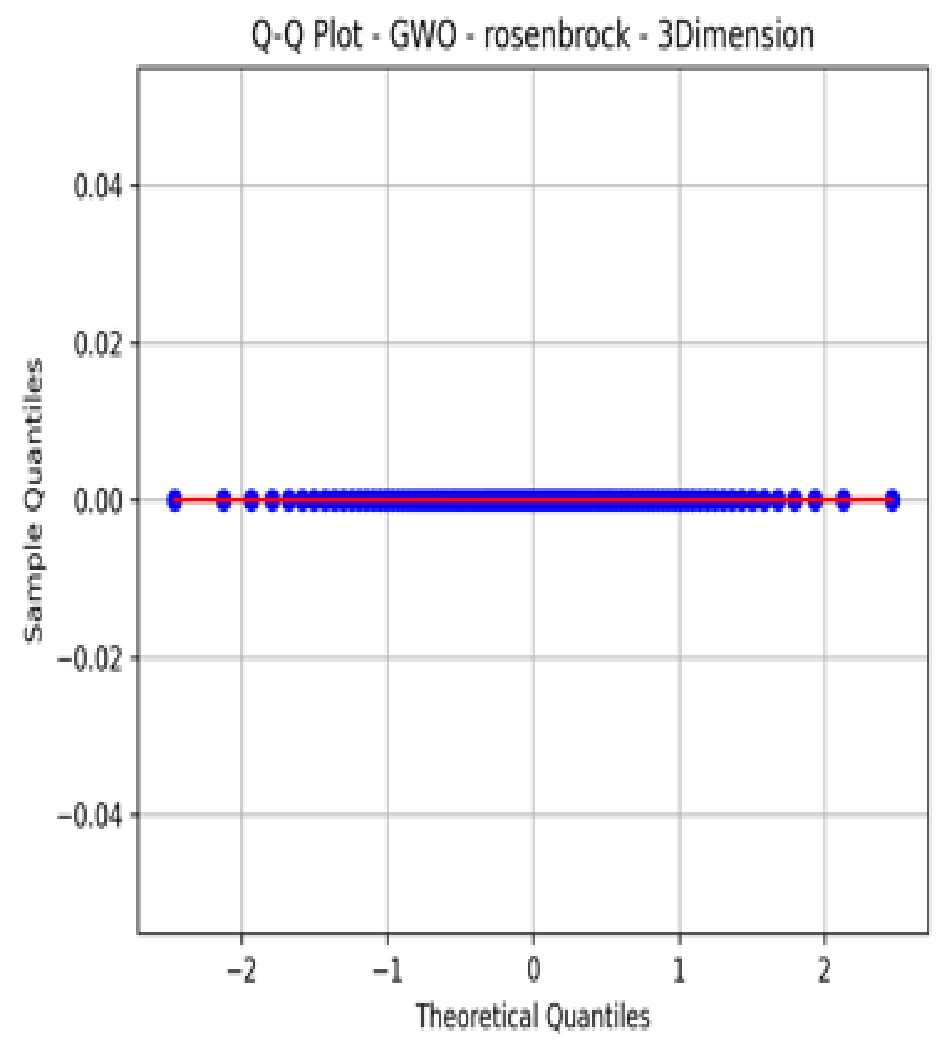
Q-Q Plot - GWO - rosenbrock - 3Dimension
Sample Quantiles
0.04
0.02
0.00
−0.02
−0.04
−2
−1
0
1
2
Theoretical Quantiles

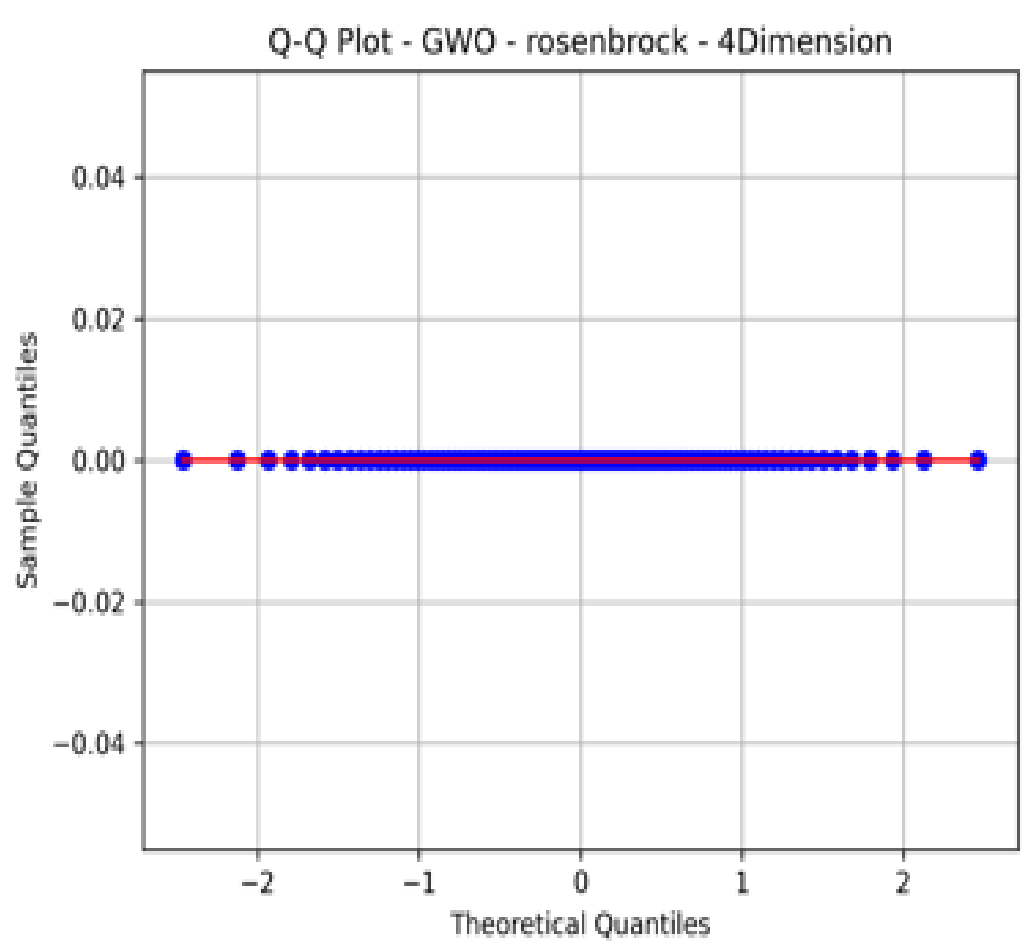
Q-Q Plot - GWO - rosenbrock - 4Dimension
Sample Quantiles
0.04
0.02
0.00
−0.02
−0.04
−2
−1
0
1
2
Theoretical Quantiles

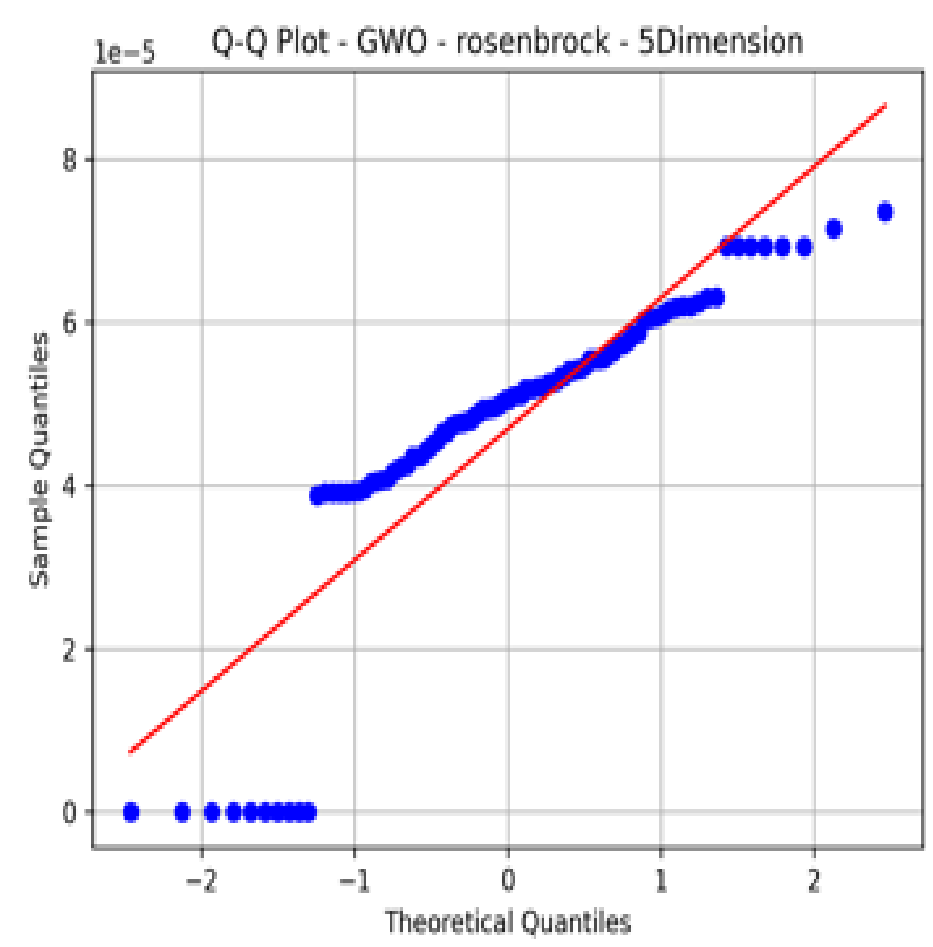
1e−5
Q-Q Plot - GWO - rosenbrock - 5Dimension
Sample Quantiles
8
6
4
2
0
−2
−1
0
1
2
Theoretical Quantiles

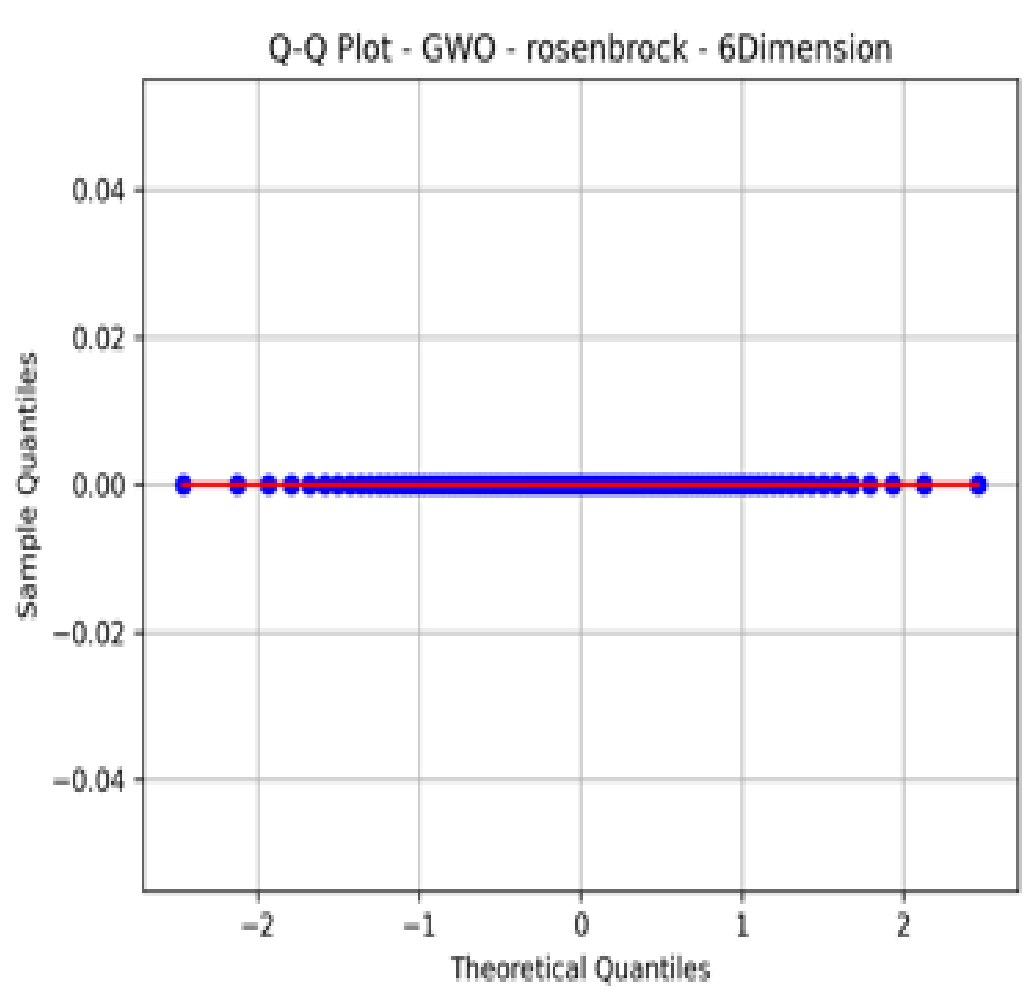
Q-Q Plot - GWO - rosenbrock - 6Dimension
Sample Quantiles
0.04
0.02
0.00
−0.02
−0.04
−2
−1
0
1
2
Theoretical Quantiles

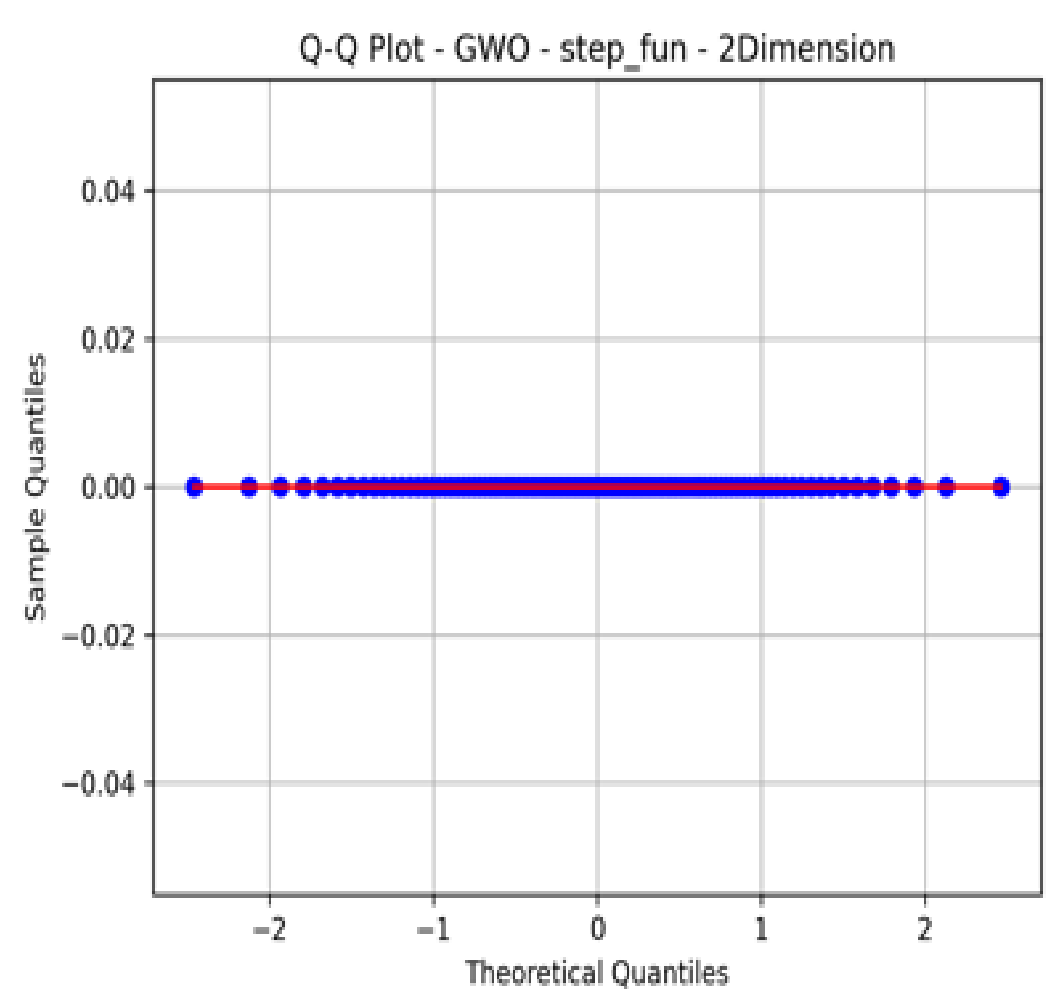
Q-Q Plot - GWO - step_fun - 2Dimension
Sample Quantiles
0.04
0.02
0.00
−0.02
−0.04
−2
−1
0
1
2
Theoretical Quantiles

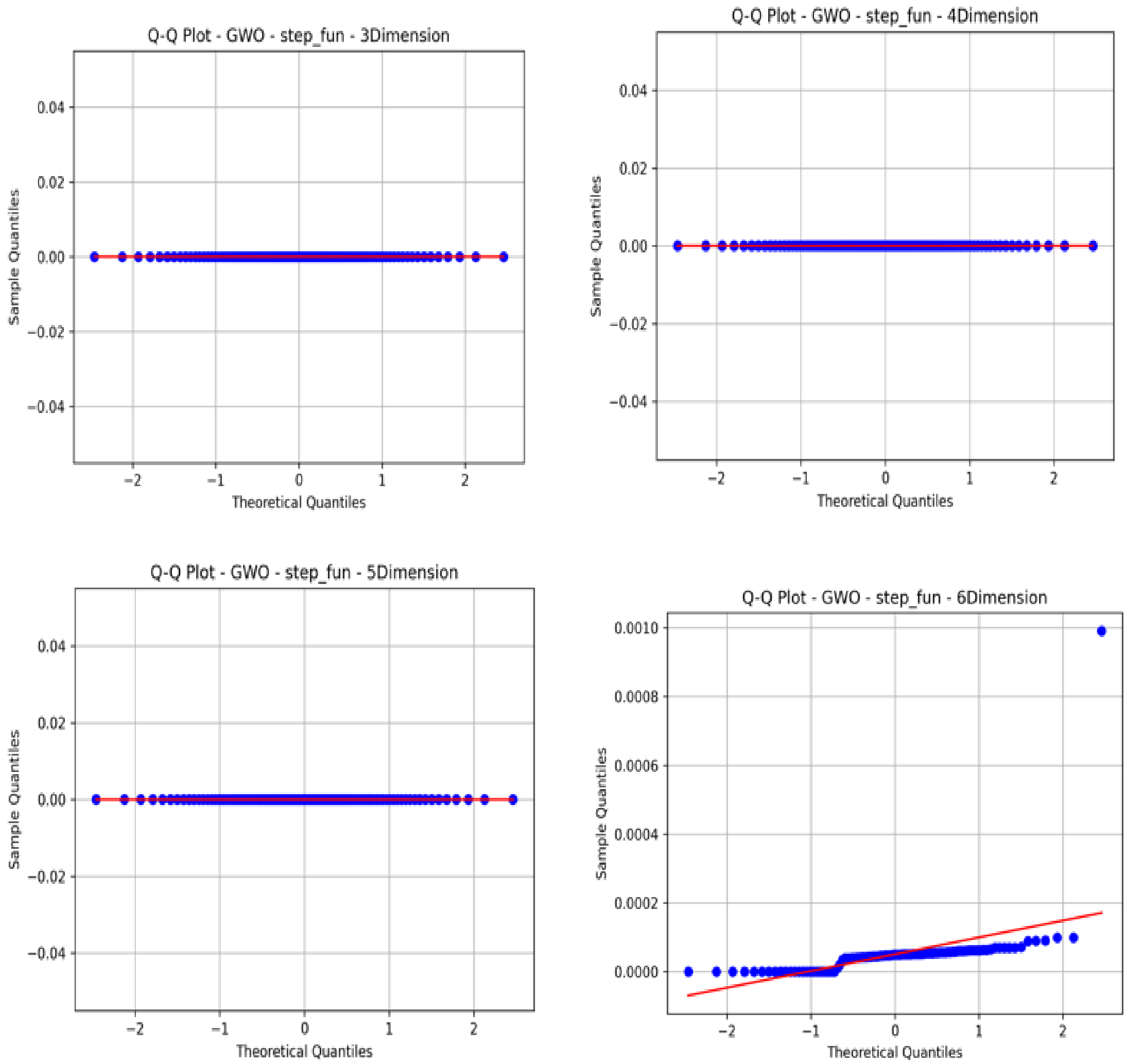
Q-Q Plot - GWO - step_fun - 3Dimension
Q-Q Plot - GWO - step_fun - 4Dimension
Q-Q Plot - GWO - step_fun - 5Dimension
Q-Q Plot - GWO - step_fun - 6Dimension
Sample Quantiles
Theoretical Quantiles

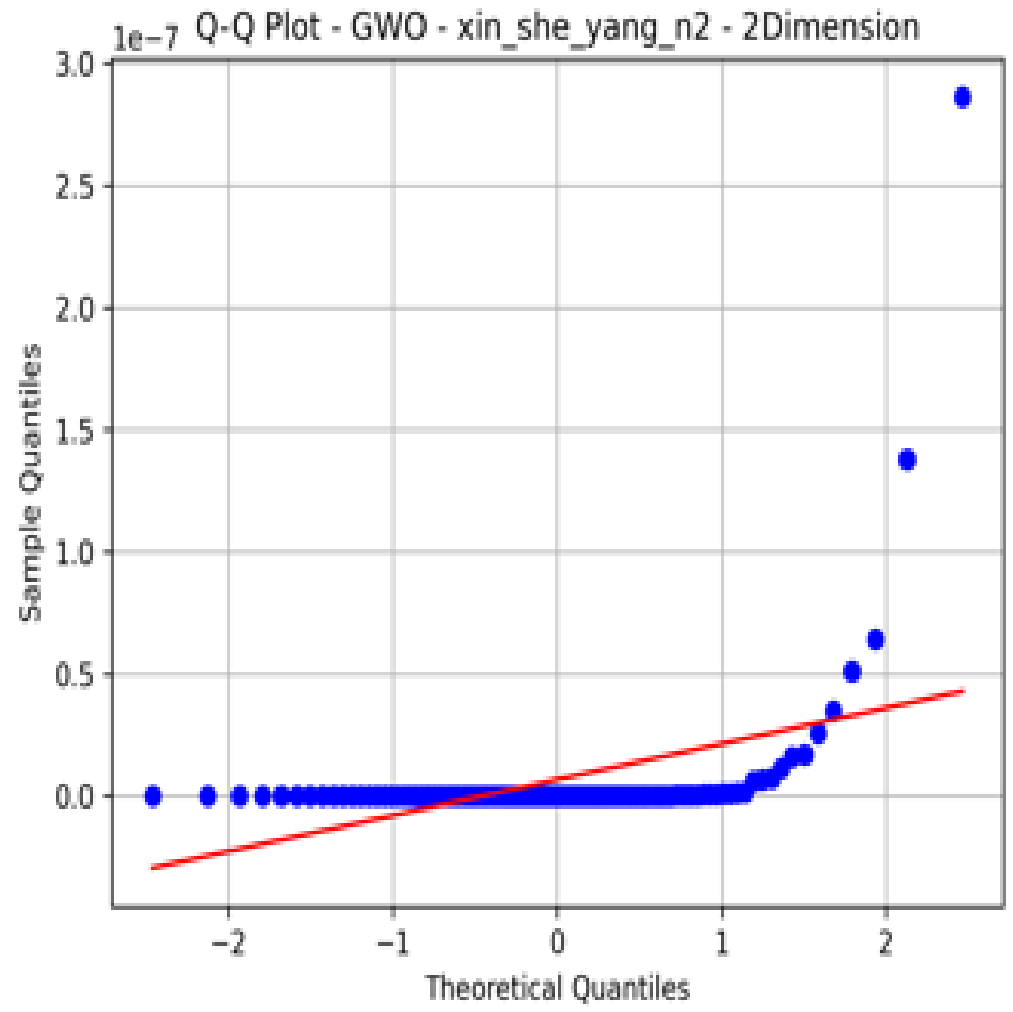
Q-Q Plot - GWO - xin_she_yang_n2 - 2Dimension
1e−7
3.0
2.5
2.0
1.5
1.0
0.5
0.0
Sample Quantiles
−2
−1
0
1
2
Theoretical Quantiles

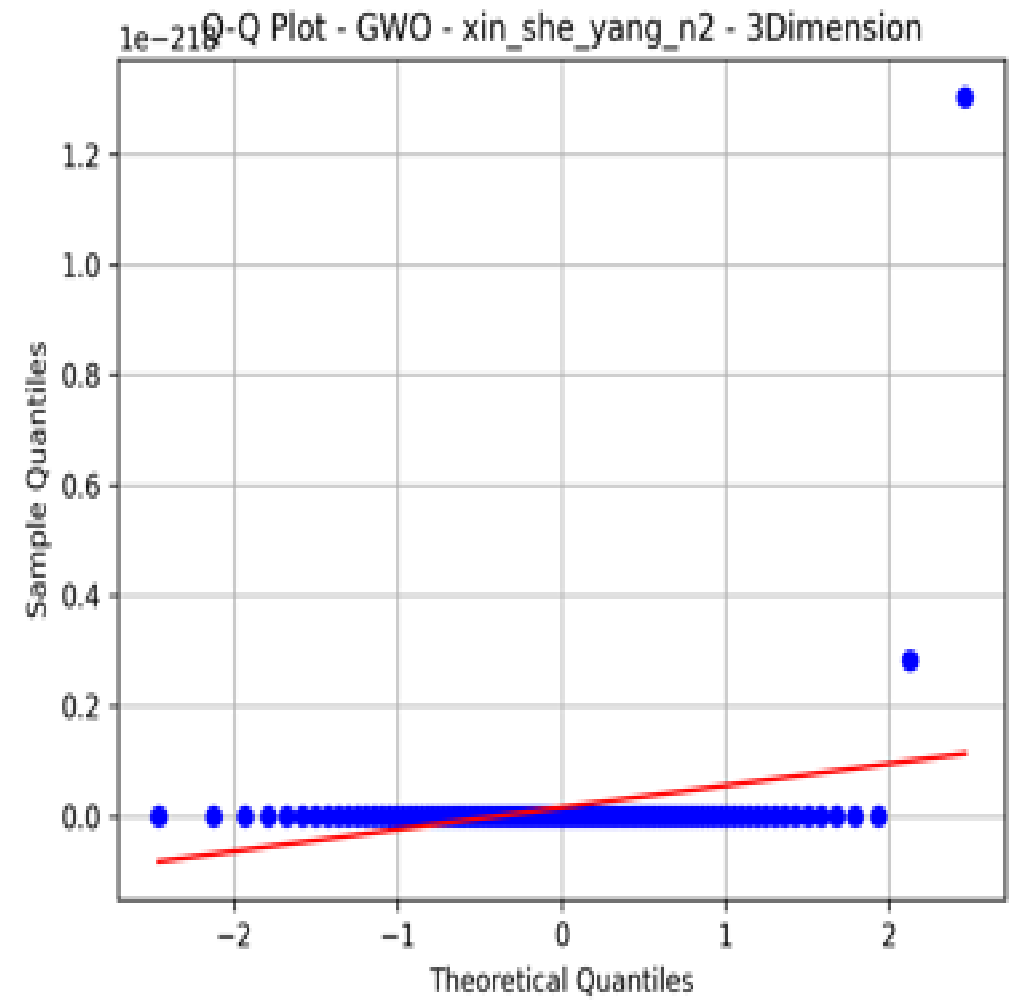
Q-Q Plot - GWO - xin_she_yang_n2 - 3Dimension
1.2
1.0
0.8
0.6
0.4
0.2
0.0
Sample Quantiles
−2
−1
0
1
2
Theoretical Quantiles

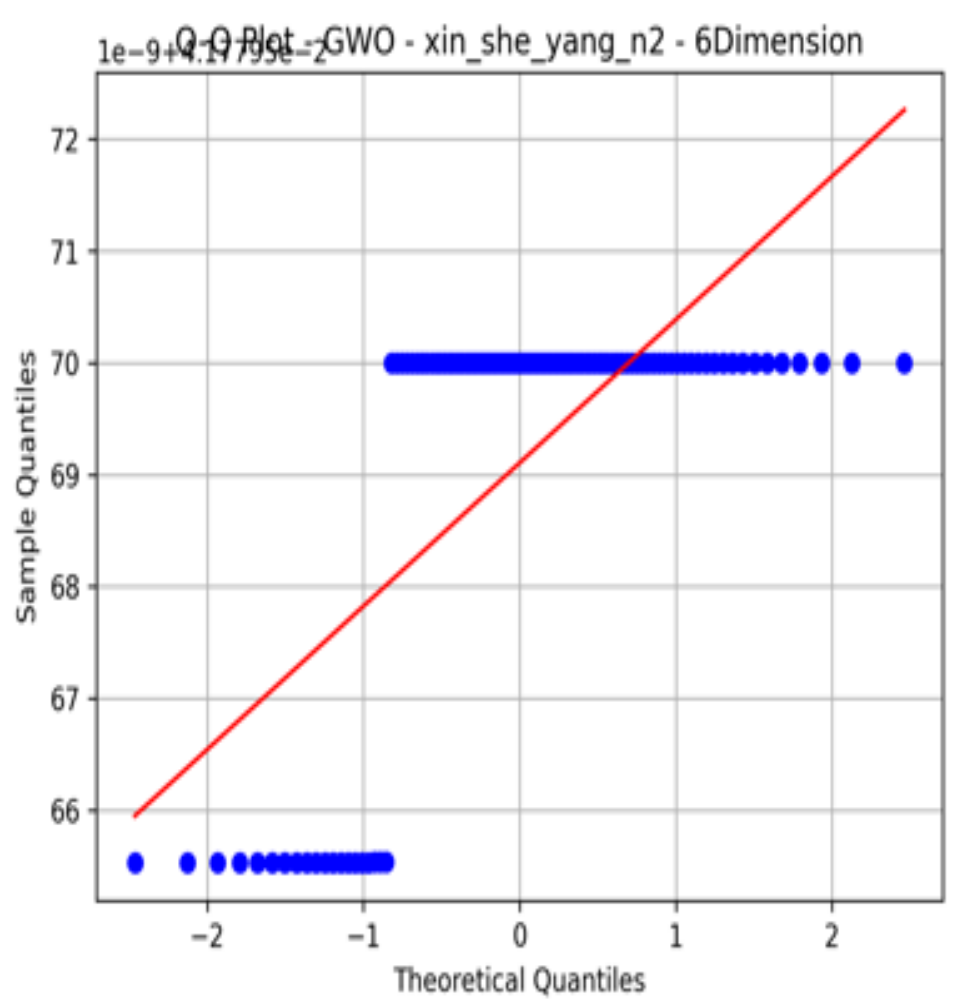
GWO - xin_she_yang_n2 - 6Dimension
72
71
70
69
68
67
66
Sample Quantiles
−2
−1
0
1
2
Theoretical Quantiles

# APPENDIX J: CASE STUDY 2 QUANTILE-QUANTILE(Q-Q) PLOTS FOR RESULT DATASET

In this appendix, we are having the Q-Q plots for optimization problems{continuous and non-continuous} in multiple dimensions(2,3,4,5,6)D as per Case Study 2. The following Q-Q plots are generated for almost all optimization problems{Table 8 and Table 9} in multiple dimensions(2,3,4,5,6)D. which validates that the generated result datasets are not normally distributed. In the plots, blue line represents the data set generated via experiments and red line denotes the benchmark normal-distribution line. Hence, a deviation in blue line from red line represents the non-normally distributed data Each plot is labelled as follows: **metaheuristic name - optimization function name-dimension.** For continuous functions plots we have used the precision results from Table C2 and for non-continuous functions plots we have used the precision results from Table D2

BSO Q-Q Plots

Continuous functions QQ plots

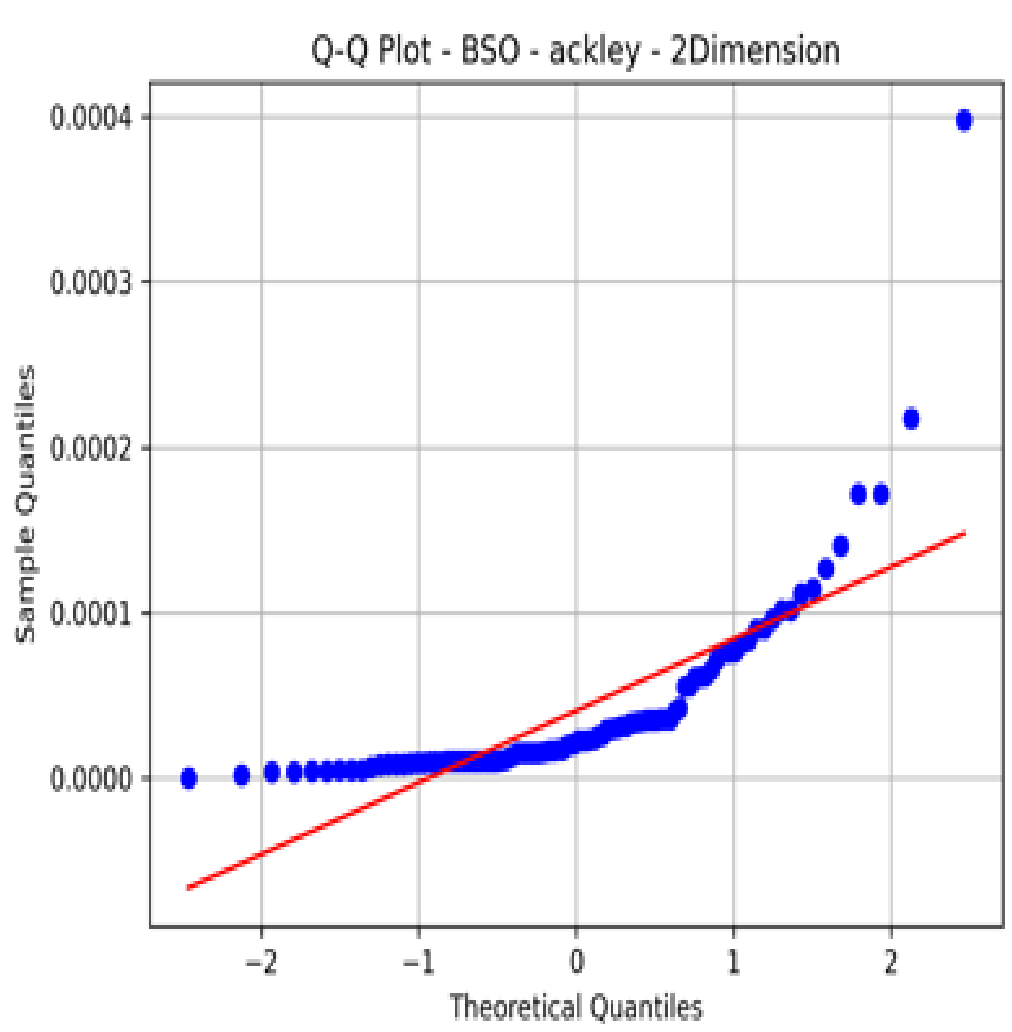

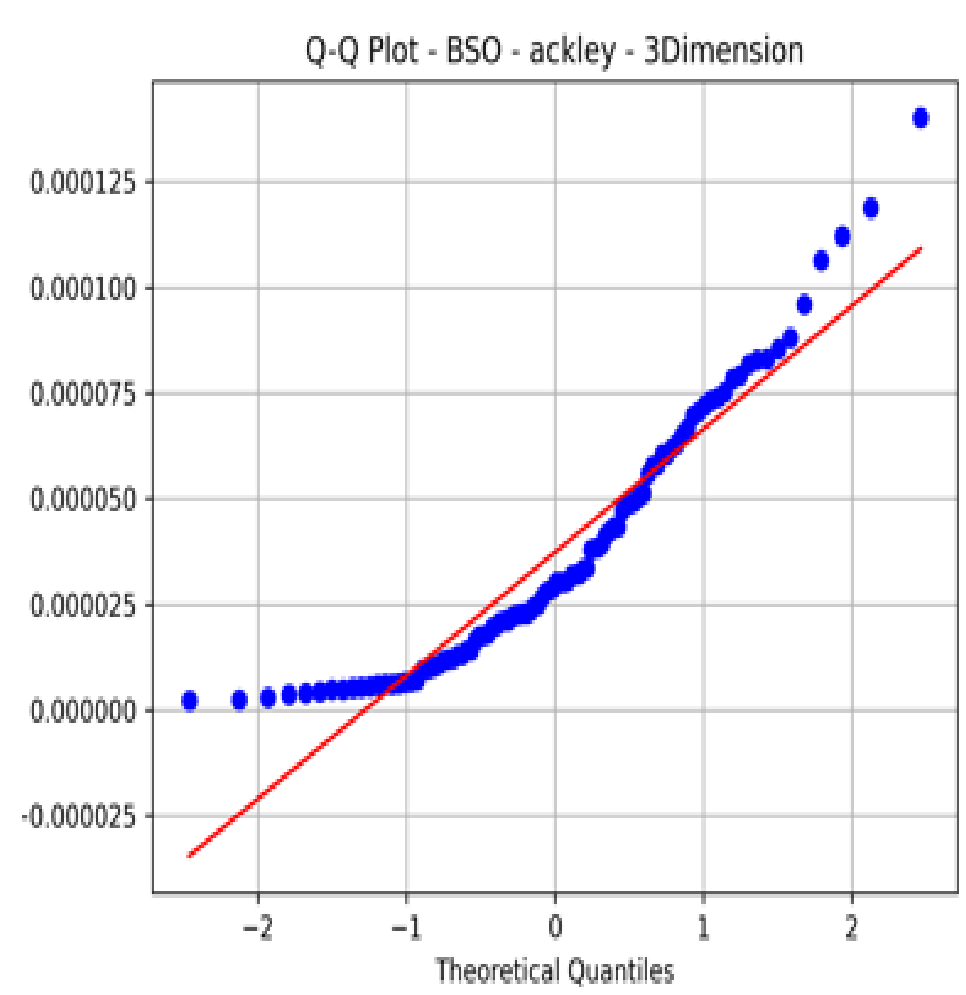

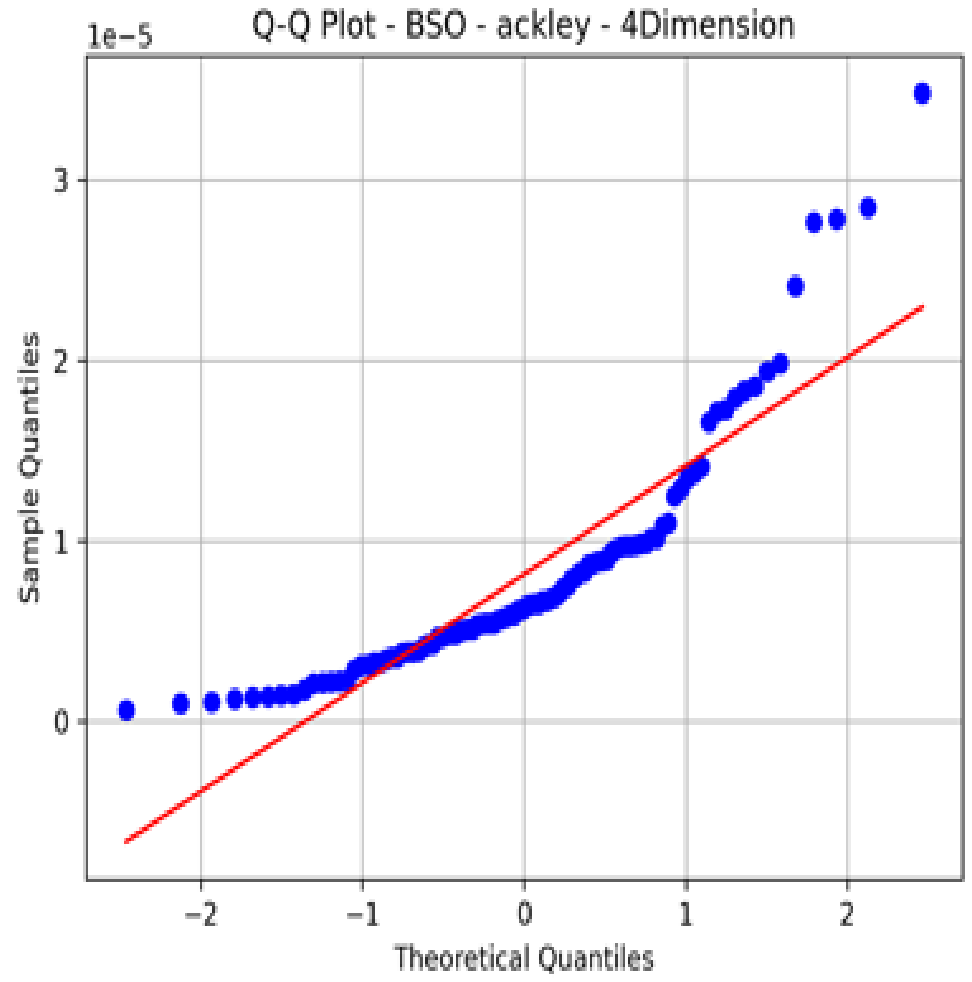
1e−5
Q-Q Plot - BSO - ackley - 4Dimension
Sample Quantiles
Theoretical Quantiles

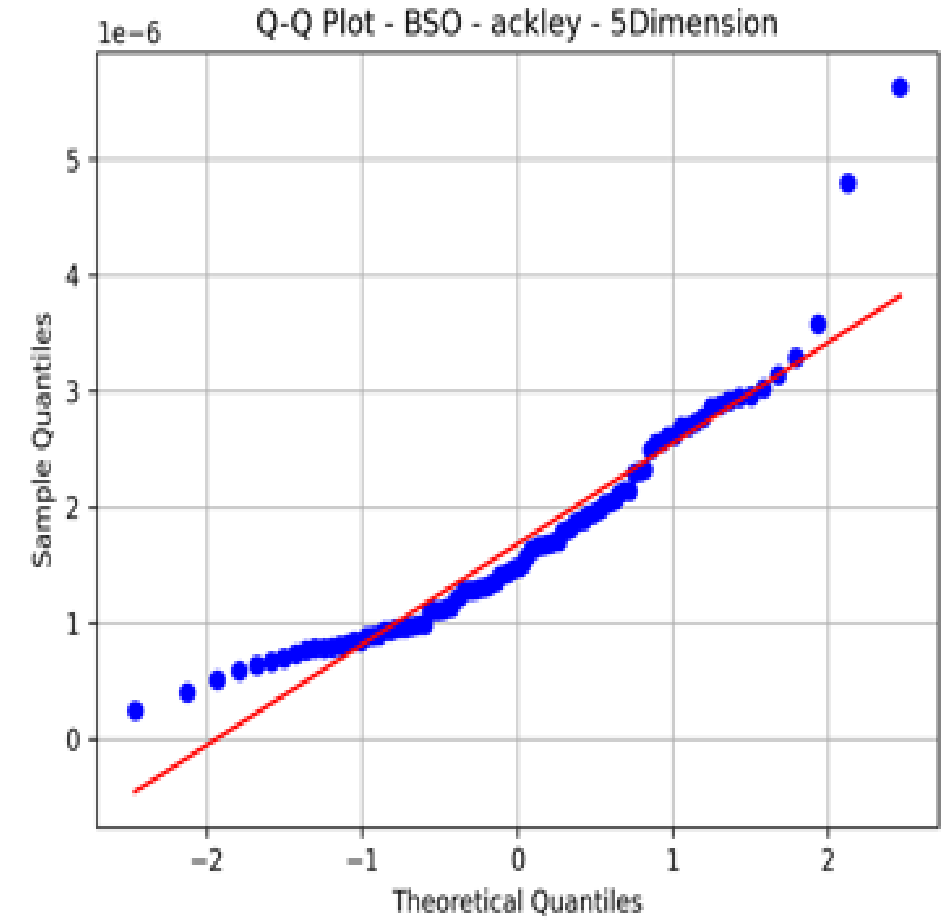
1e−6
Q-Q Plot - BSO - ackley - 5Dimension
Sample Quantiles
Theoretical Quantiles

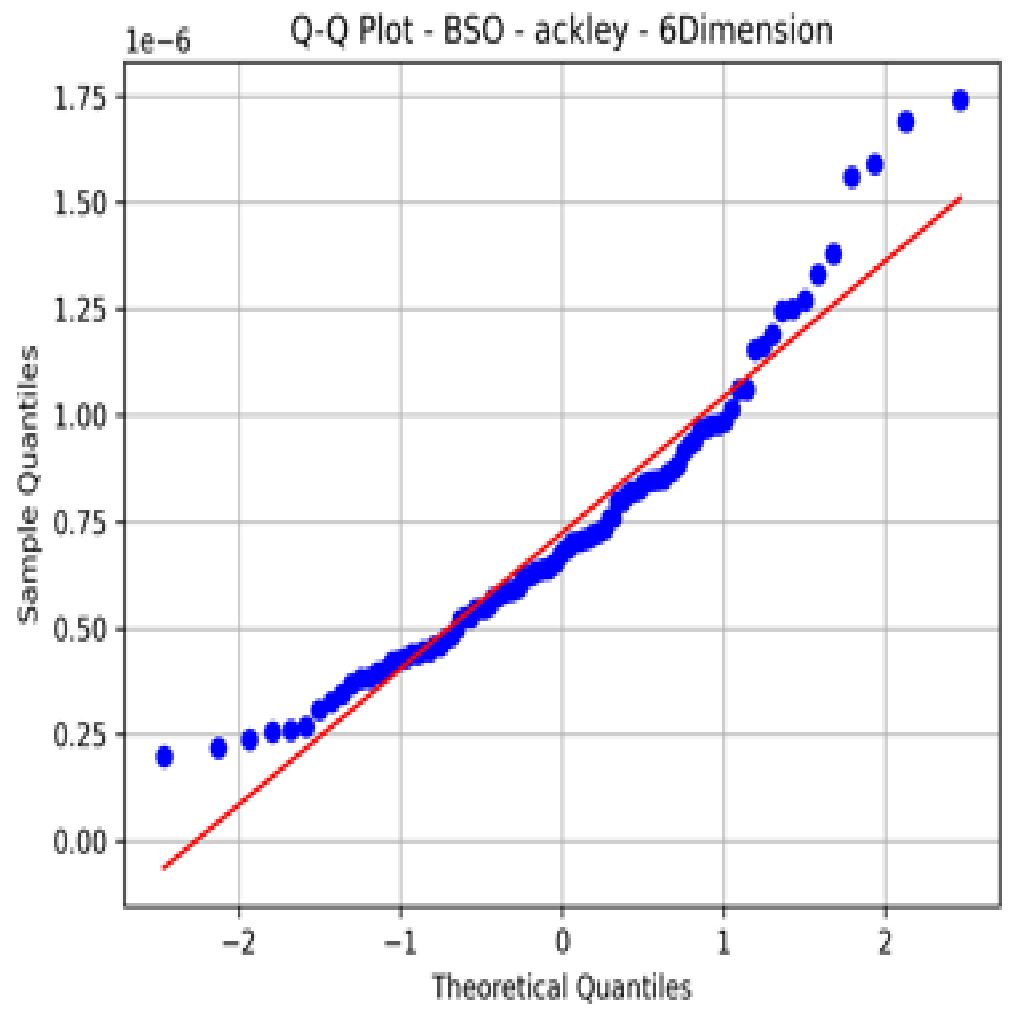
1e−6
Q-Q Plot - BSO - ackley - 6Dimension
Sample Quantiles
Theoretical Quantiles

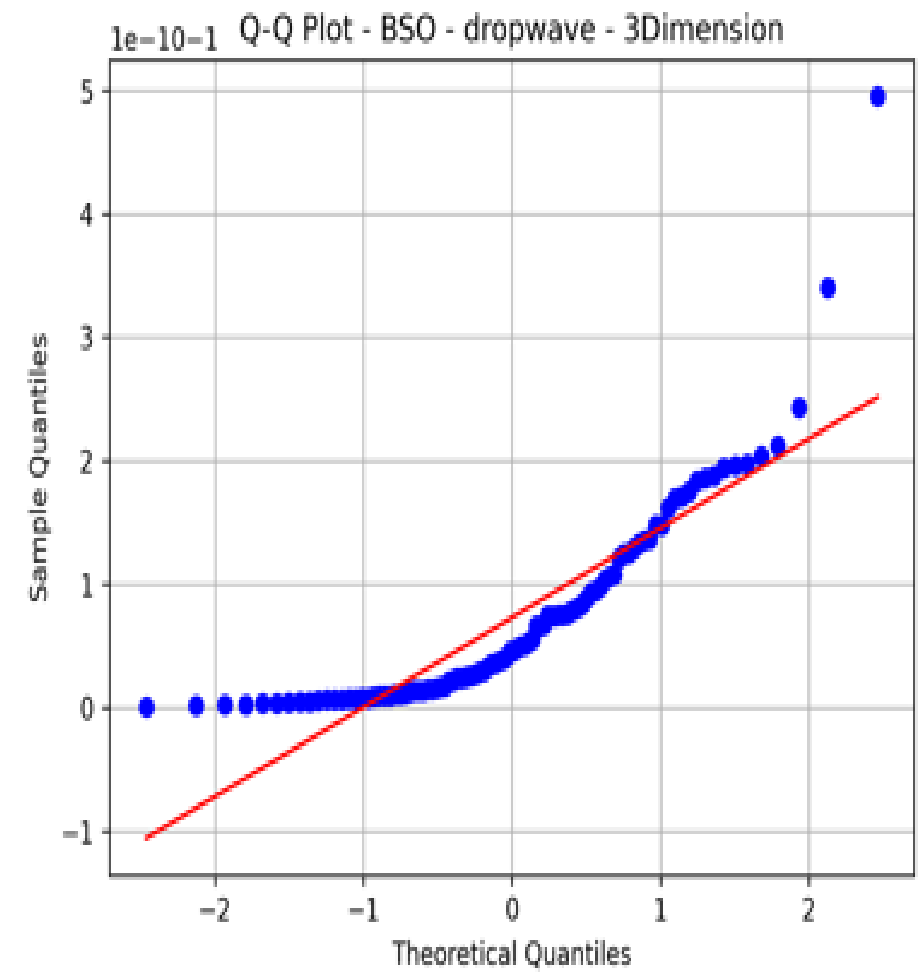
1e−10−1
Q-Q Plot - BSO - dropwave - 3Dimension
Sample Quantiles
Theoretical Quantiles

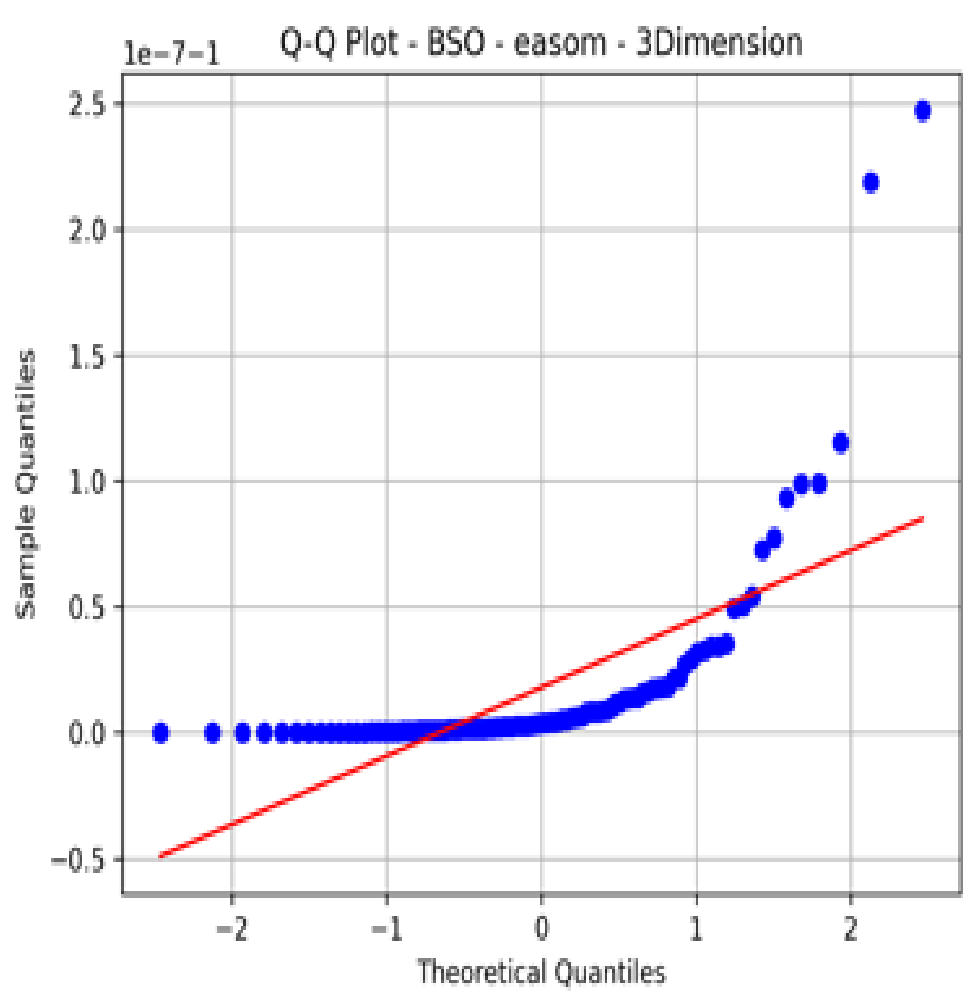
1e−7−1
Q-Q Plot - BSO - easom - 3Dimension
Sample Quantiles
Theoretical Quantiles

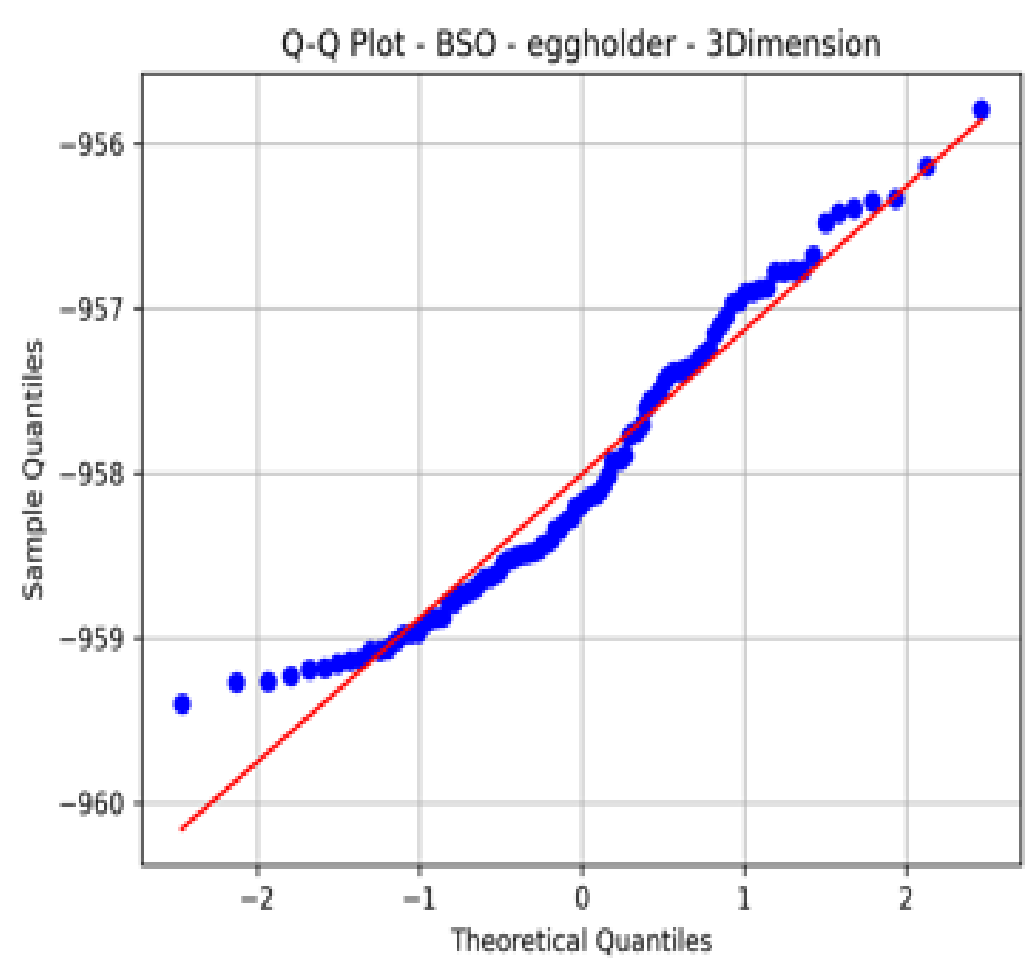
Q-Q Plot - BSO - eggholder - 3Dimension
Sample Quantiles
Theoretical Quantiles

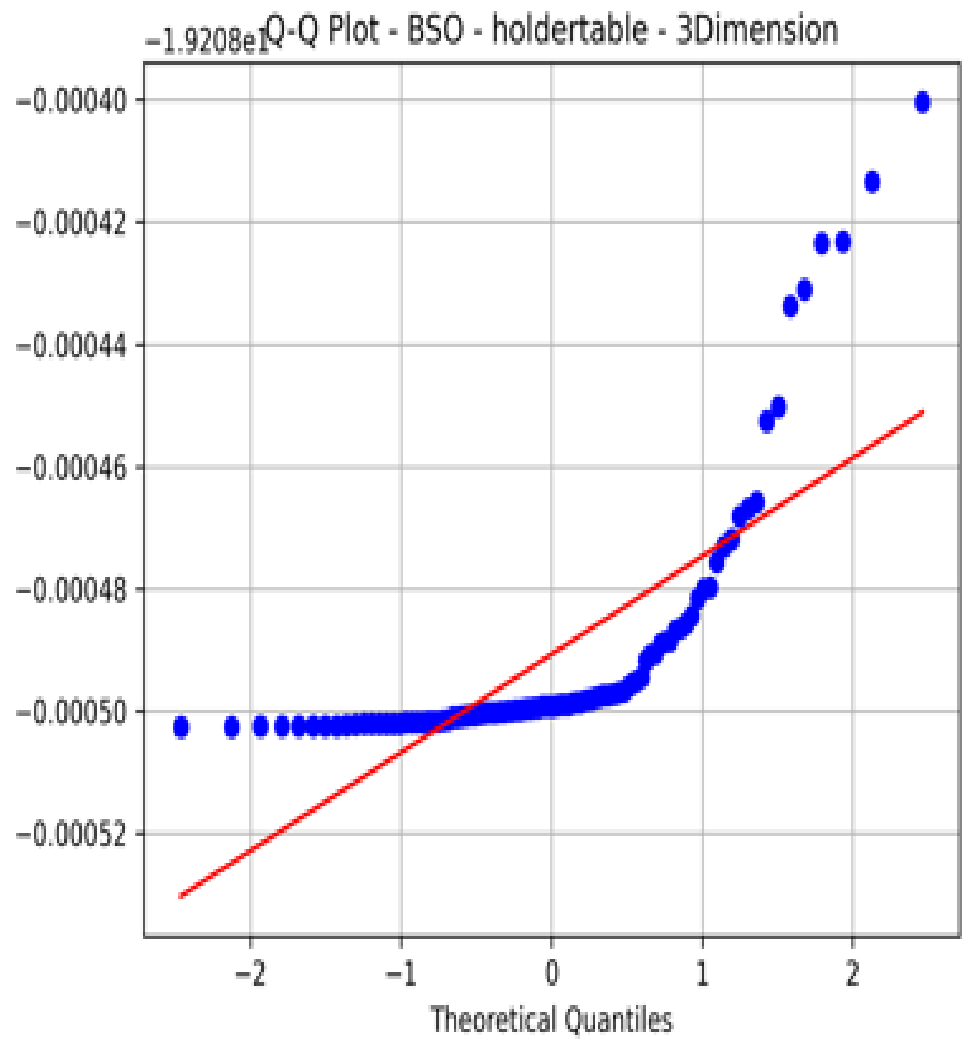

Q-Q Plot - BSO - holdertable - 3Dimension
−1.9208e1
−0.00040
−0.00042
−0.00044
−0.00046
−0.00048
−0.00050
−0.00052
−2
−1
0
1
2
Theoretical Quantiles

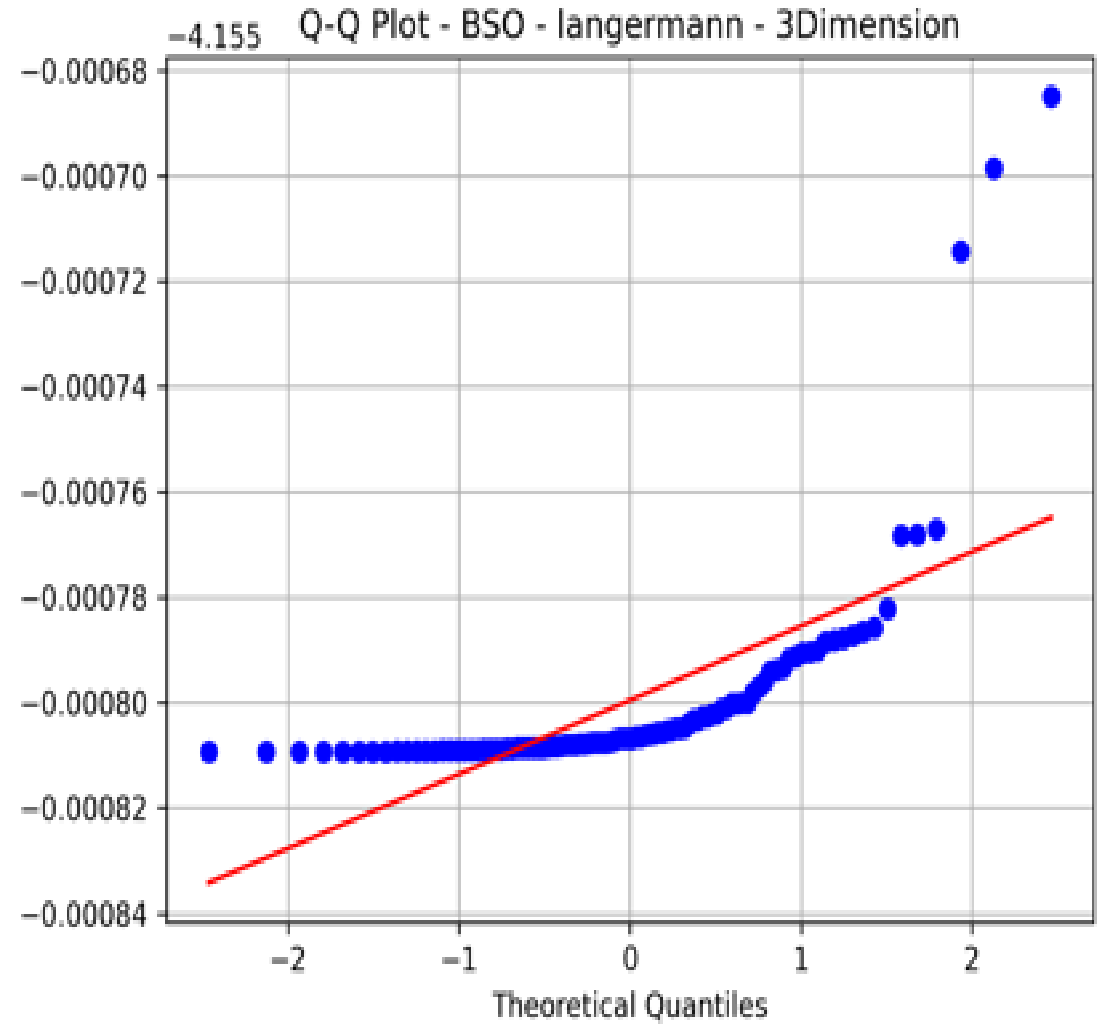

Q-Q Plot - BSO - langermann - 3Dimension
−4.155
−0.00068
−0.00070
−0.00072
−0.00074
−0.00076
−0.00078
−0.00080
−0.00082
−0.00084
−2
−1
0
1
2
Theoretical Quantiles

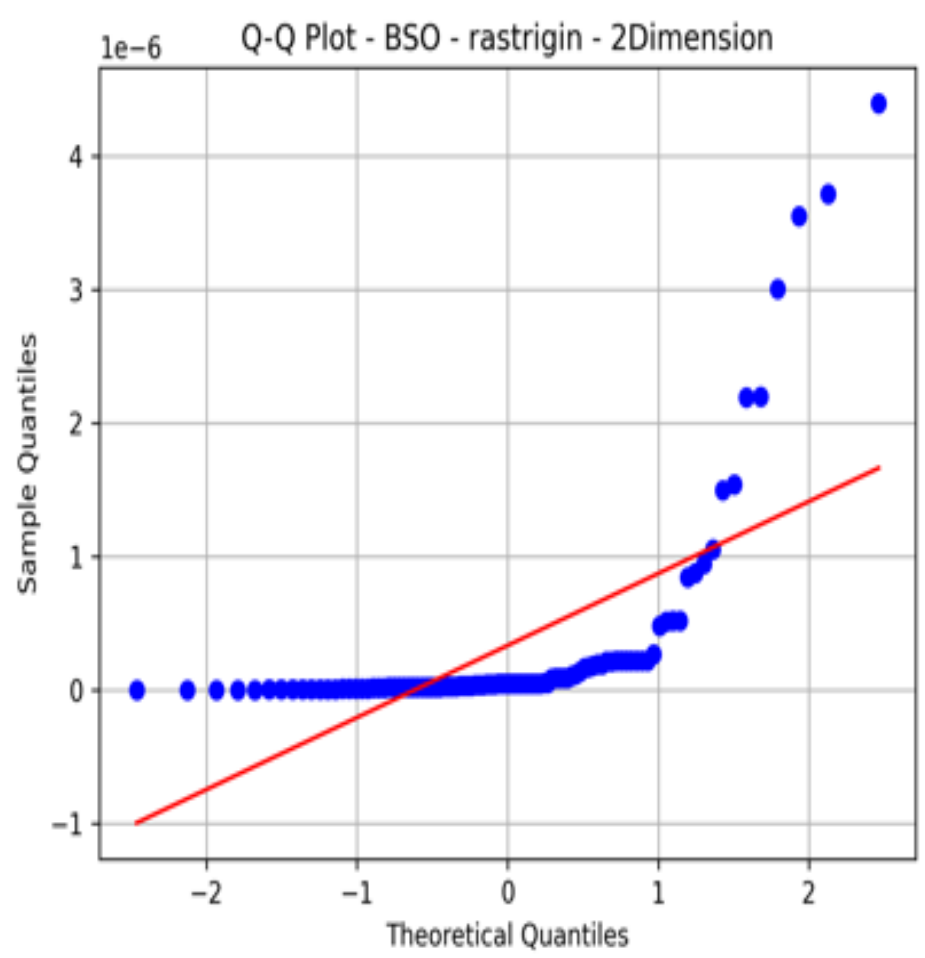

Q-Q Plot - BSO - rastrigin - 2Dimension
1e−6
4
3
2
1
0
−1
Sample Quantiles
−2
−1
0
1
2
Theoretical Quantiles

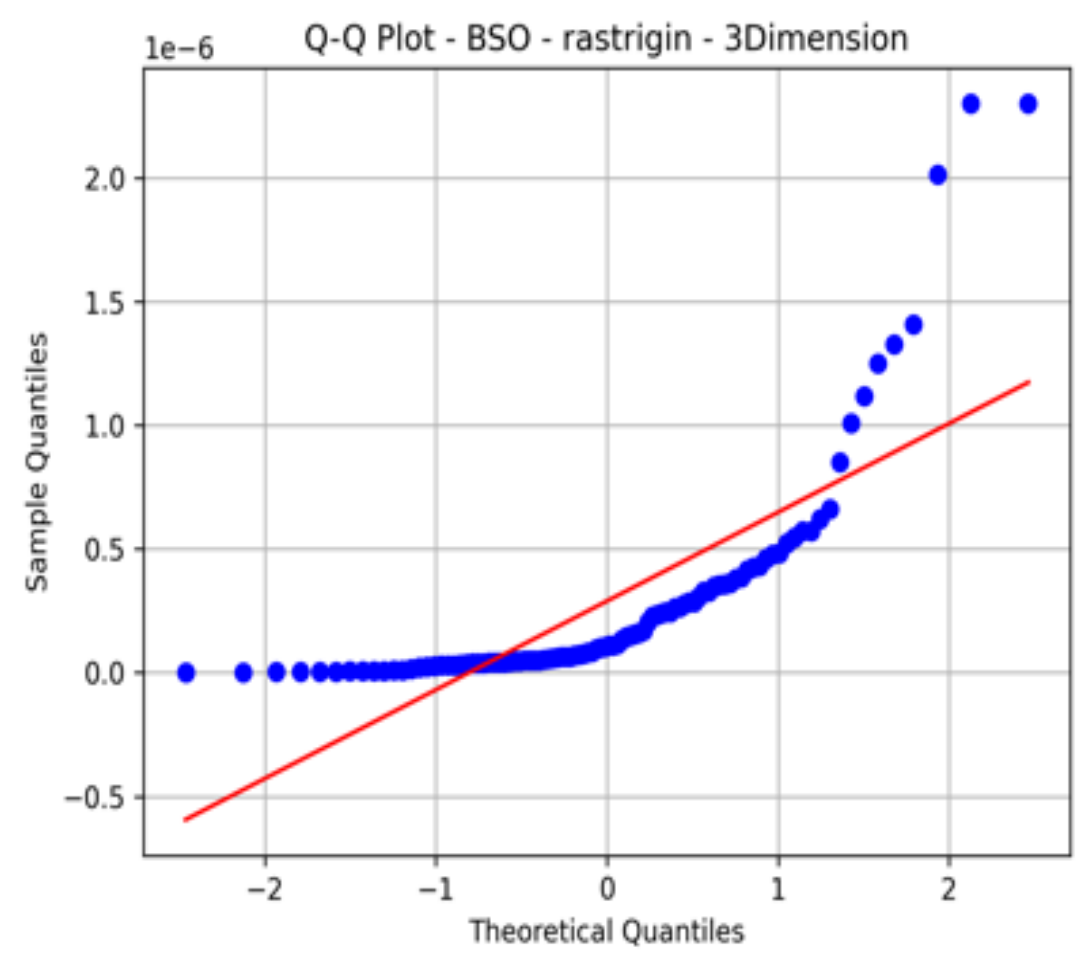

Q-Q Plot - BSO - rastrigin - 3Dimension
1e−6
2.0
1.5
1.0
0.5
0.0
−0.5
Sample Quantiles
−2
−1
0
1
2
Theoretical Quantiles

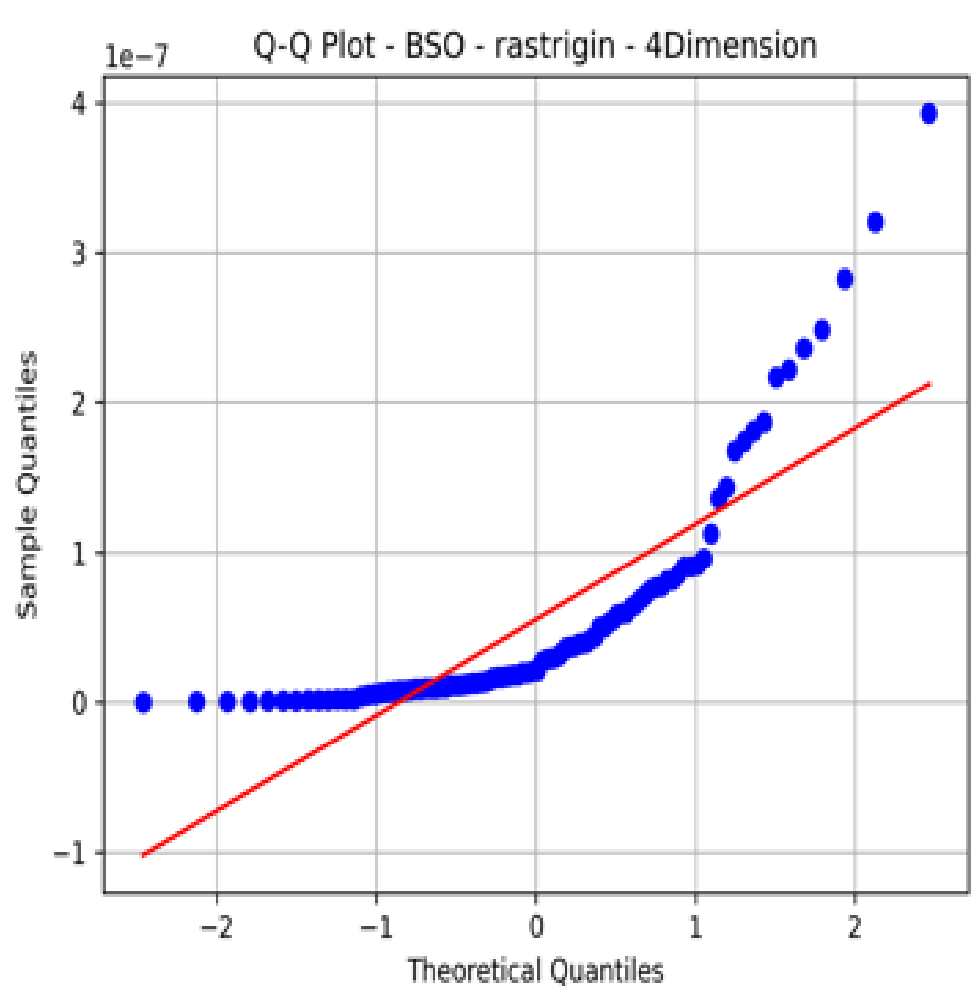

Q-Q Plot - BSO - rastrigin - 4Dimension
1e−7
4
3
2
1
0
−1
Sample Quantiles
−2
−1
0
1
2
Theoretical Quantiles

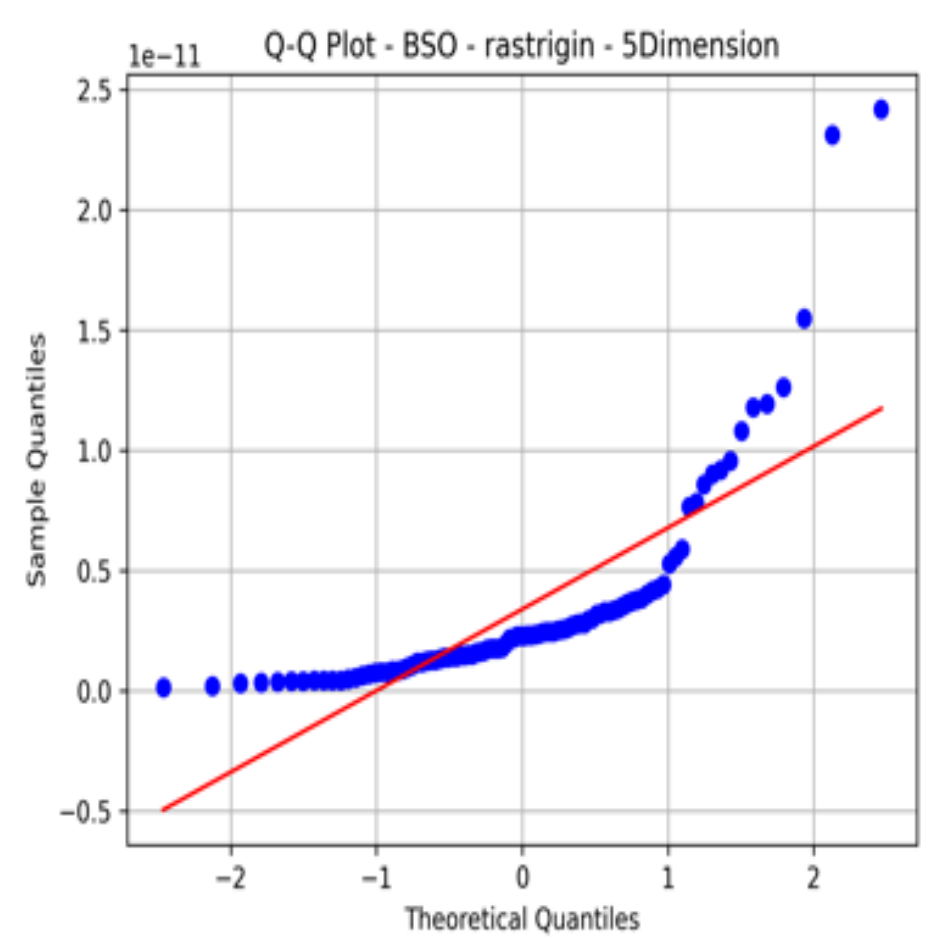

Q-Q Plot - BSO - rastrigin - 5Dimension
1e−11
2.5
2.0
1.5
1.0
0.5
0.0
−0.5
Sample Quantiles
−2
−1
0
1
2
Theoretical Quantiles

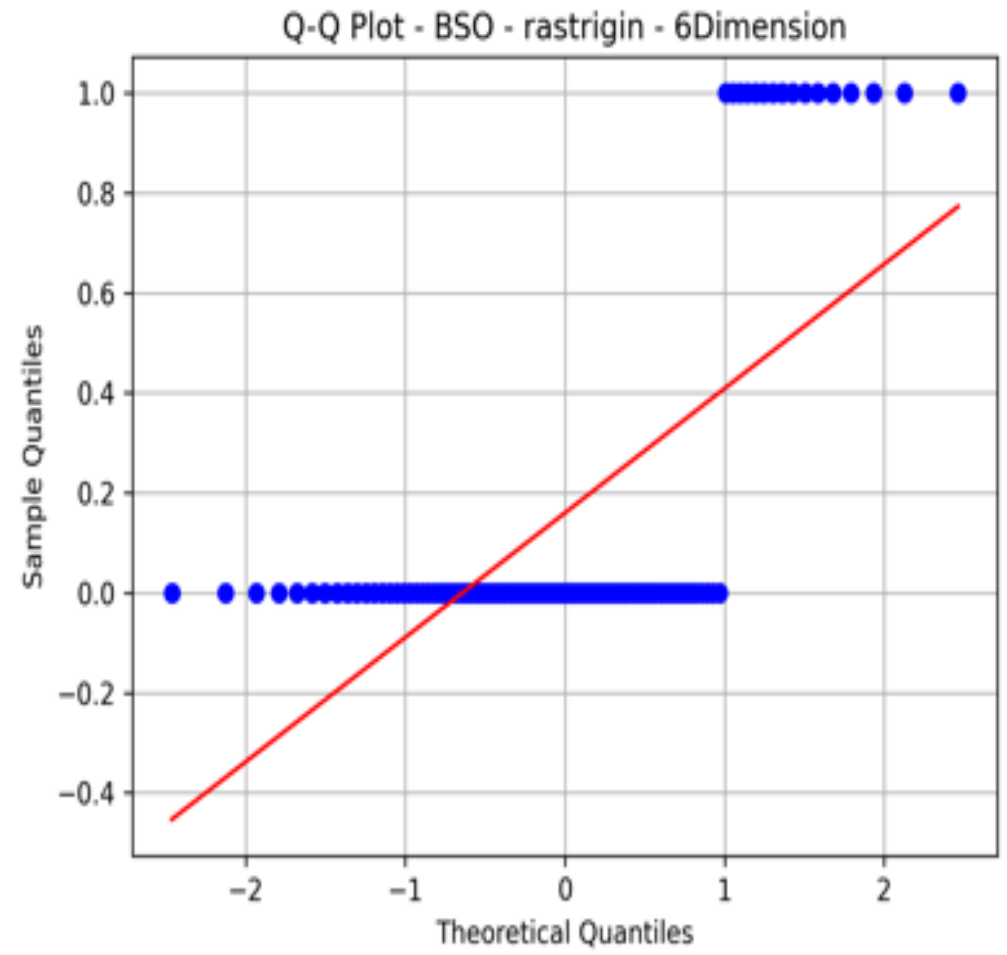
Q-Q Plot - BSO - rastrigin - 6Dimension
Sample Quantiles
1.0
0.8
0.6
0.4
0.2
0.0
−0.2
−0.4
−2
−1
0
1
2
Theoretical Quantiles

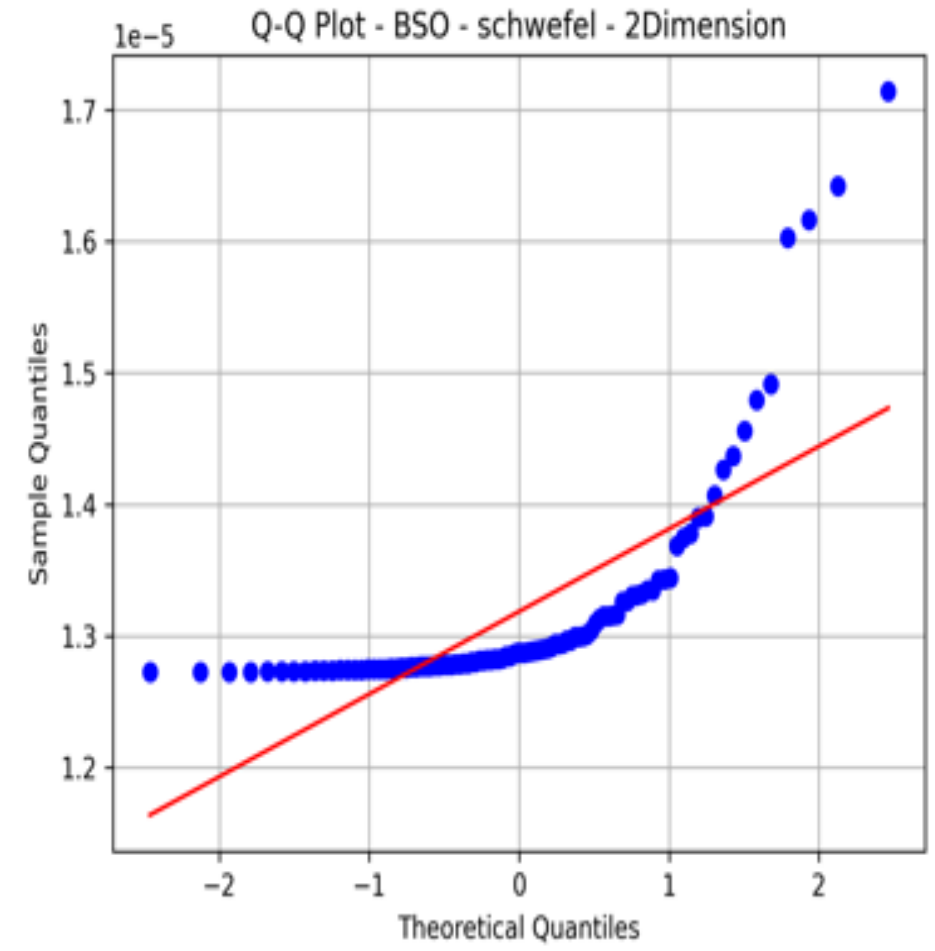
Q-Q Plot - BSO - schwefel - 2Dimension
1e−5
Sample Quantiles
1.7
1.6
1.5
1.4
1.3
1.2
−2
−1
0
1
2
Theoretical Quantiles

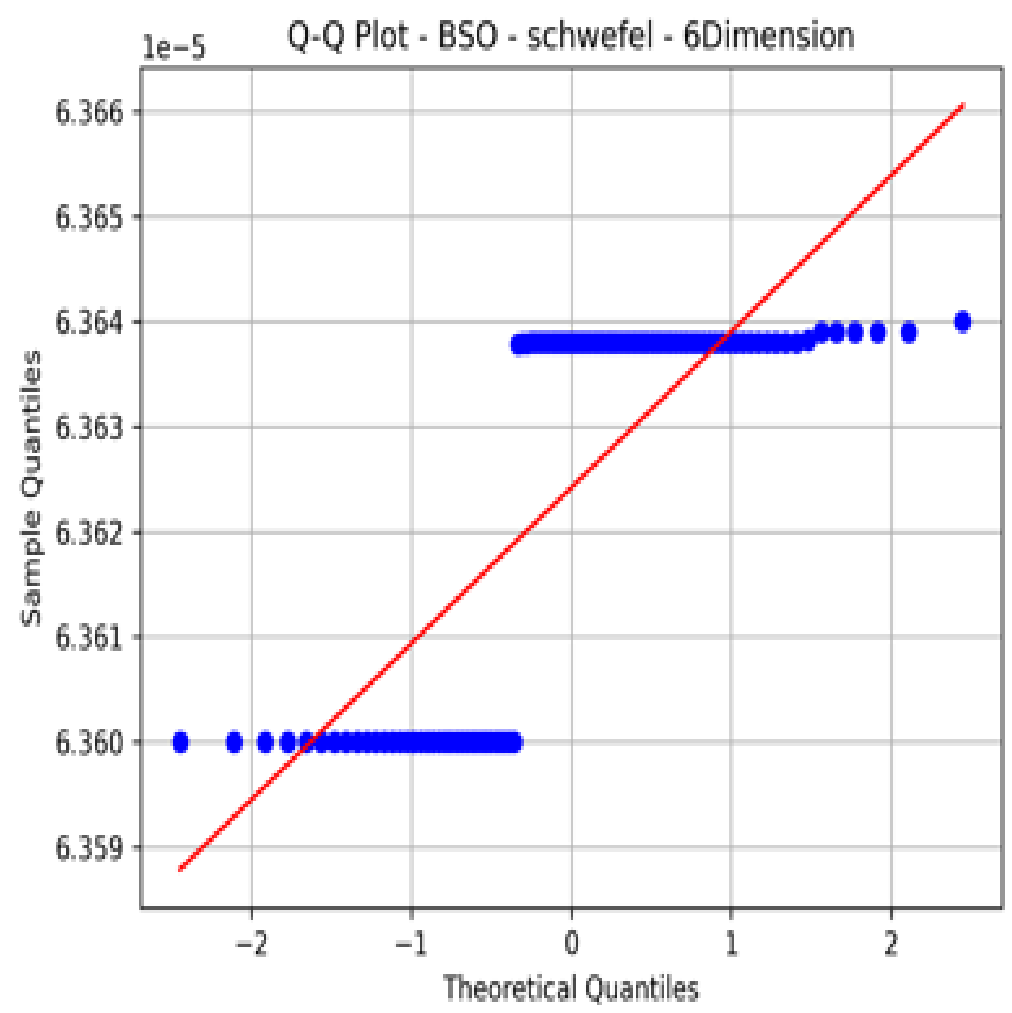
Q-Q Plot - BSO - schwefel - 6Dimension
1e−5
Sample Quantiles
6.366
6.365
6.364
6.363
6.362
6.361
6.360
6.359
−2
−1
0
1
2
Theoretical Quantiles

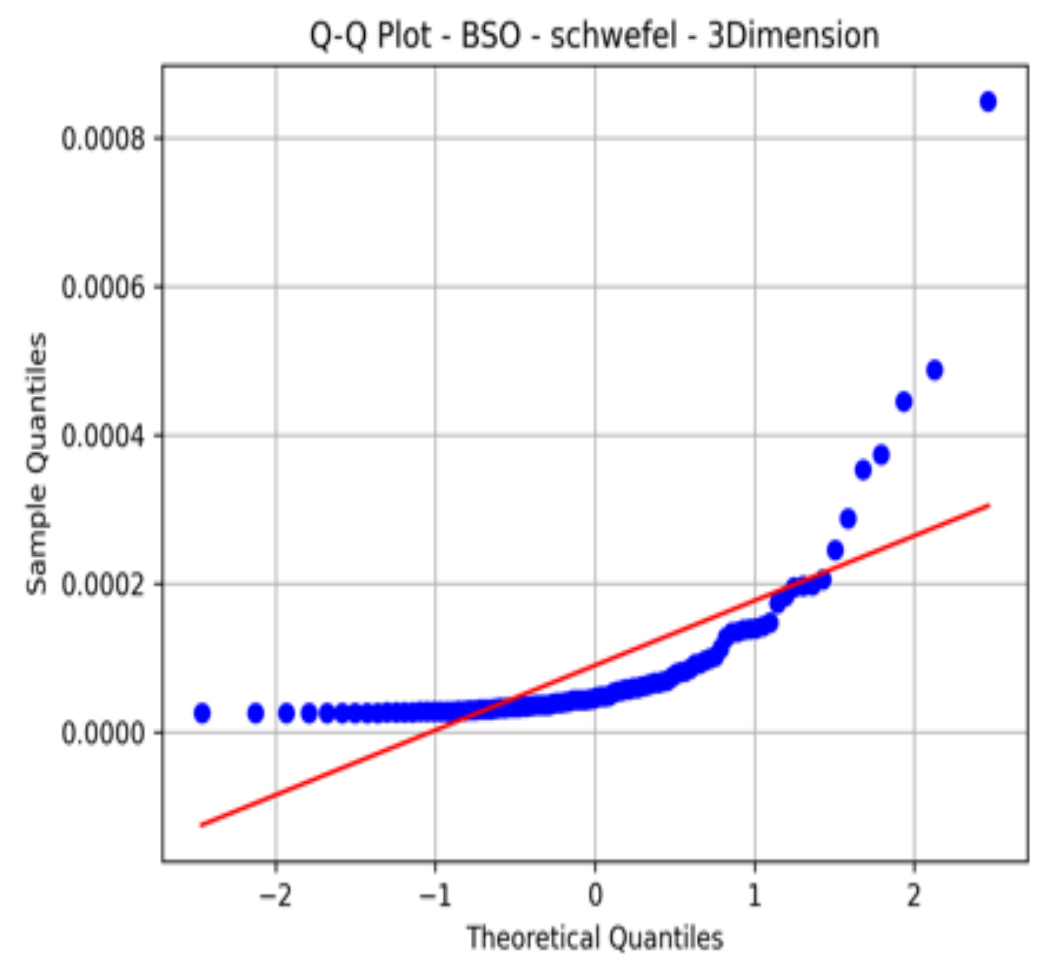
Q-Q Plot - BSO - schwefel - 3Dimension
Sample Quantiles
0.0008
0.0006
0.0004
0.0002
0.0000
−2
−1
0
1
2
Theoretical Quantiles

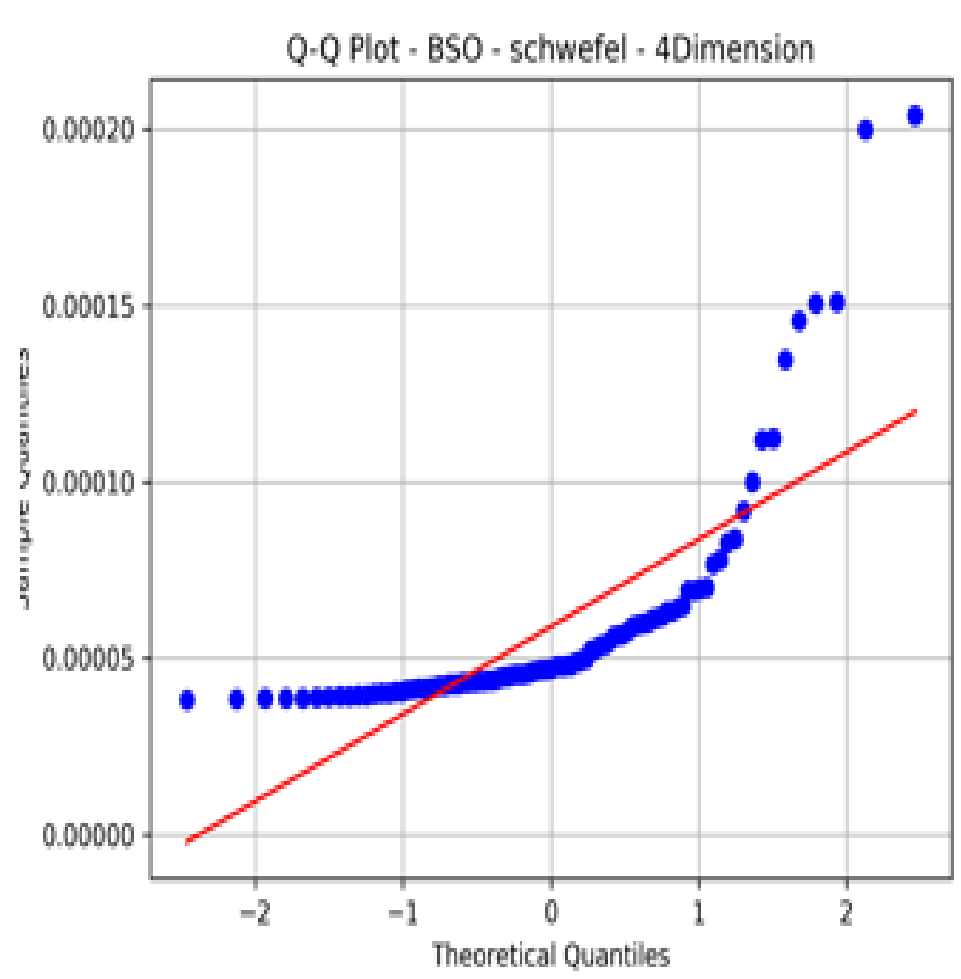
Q-Q Plot - BSO - schwefel - 4Dimension
0.00020
0.00015
0.00010
0.00005
0.00000
−2
−1
0
1
2
Theoretical Quantiles

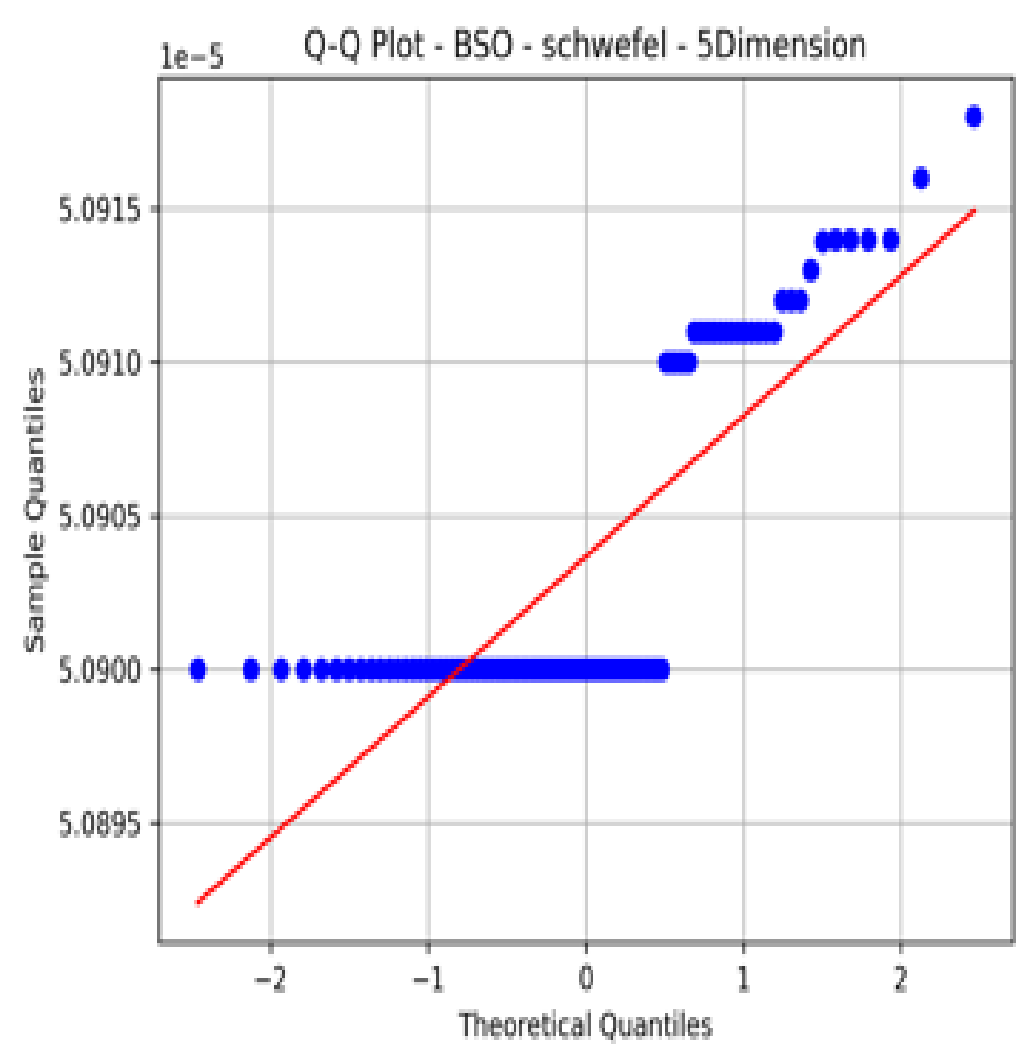
Q-Q Plot - BSO - schwefel - 5Dimension
1e−5
Sample Quantiles
5.0915
5.0910
5.0905
5.0900
5.0895
−2
−1
0
1
2
Theoretical Quantiles

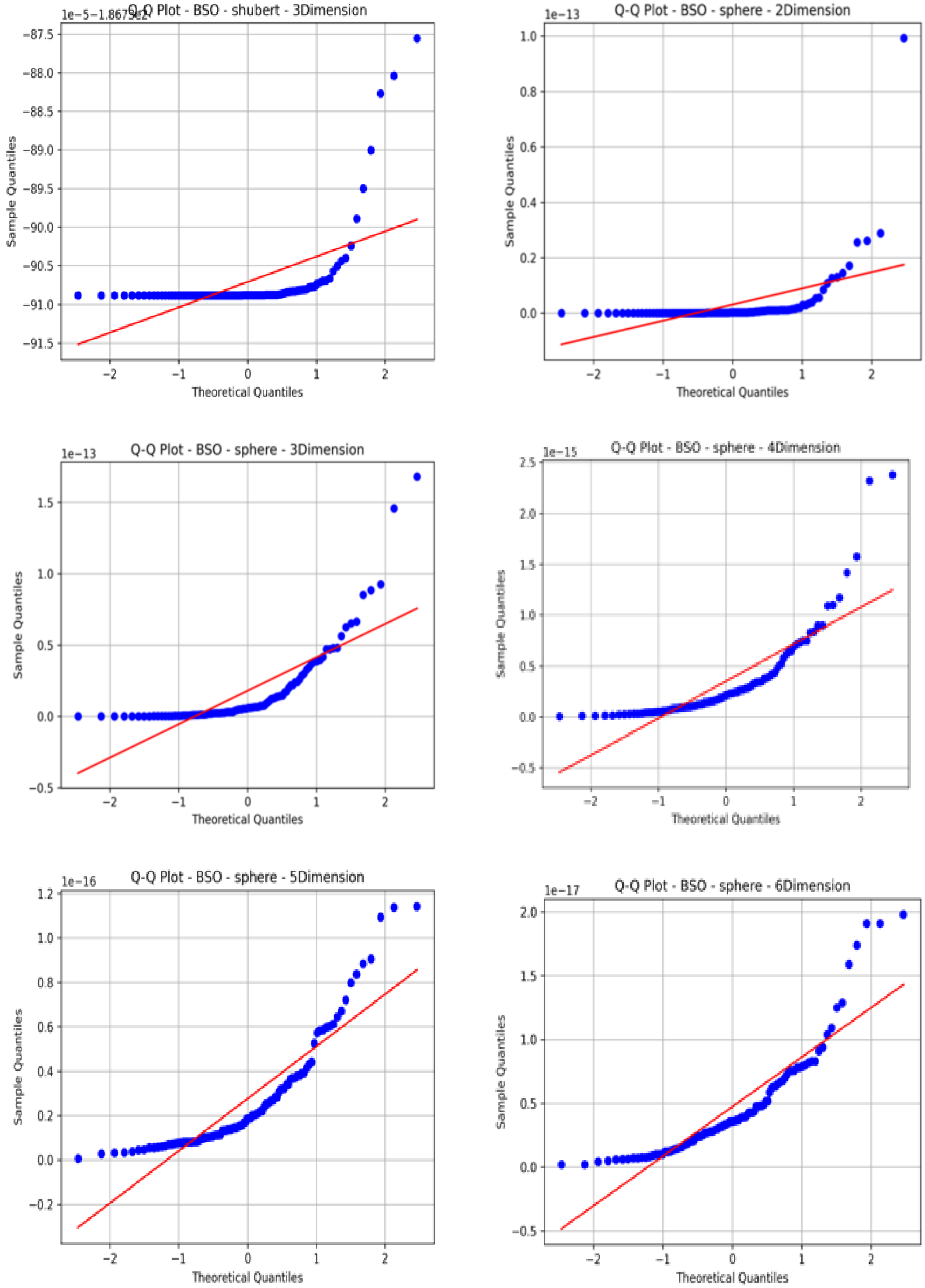

Non-Continuous functions QQ plots

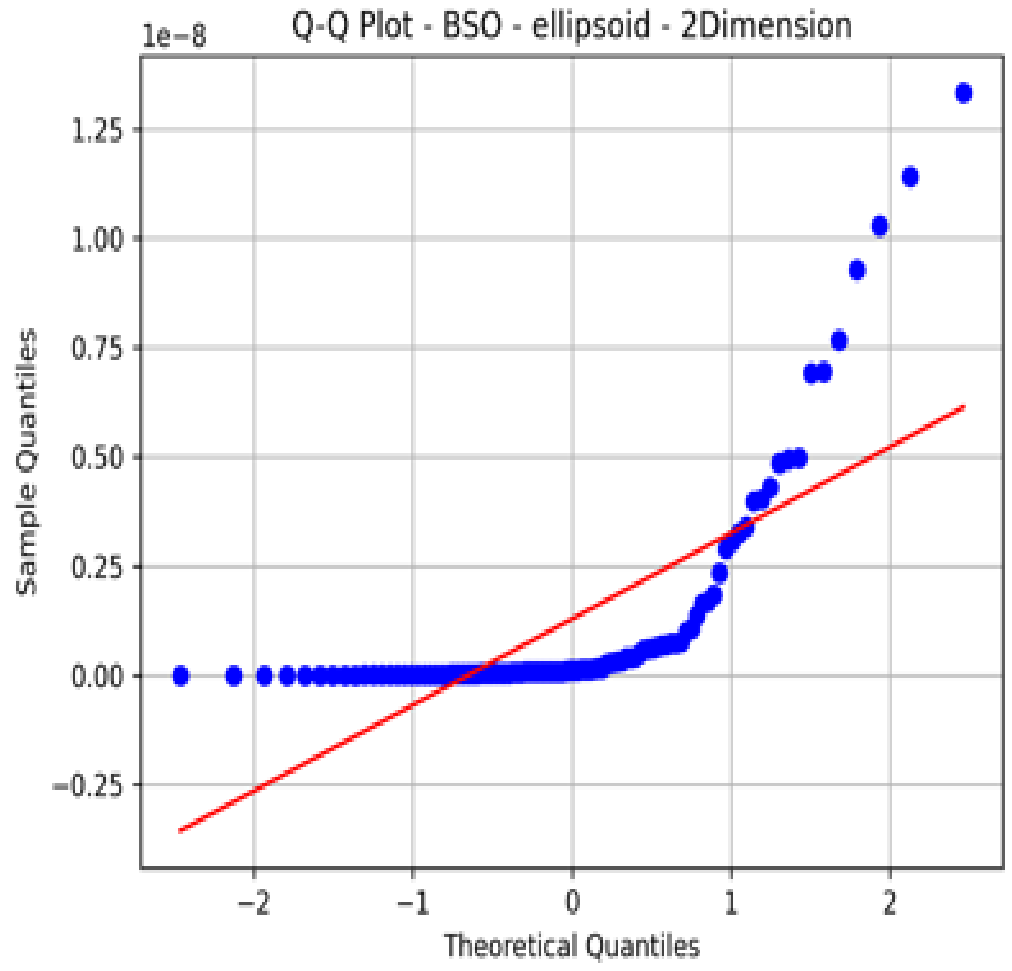
Q-Q Plot - BSO - ellipsoid - 2Dimension
1e−8
Sample Quantiles
Theoretical Quantiles

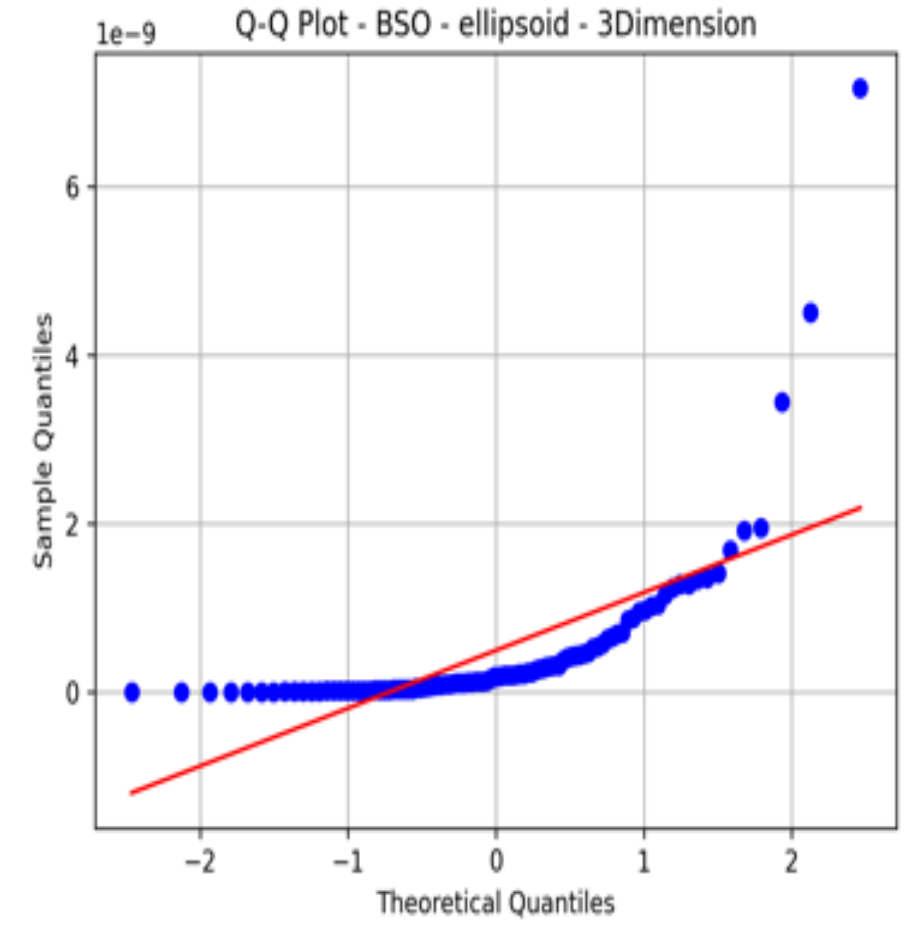
Q-Q Plot - BSO - ellipsoid - 3Dimension
1e−9
Sample Quantiles
Theoretical Quantiles

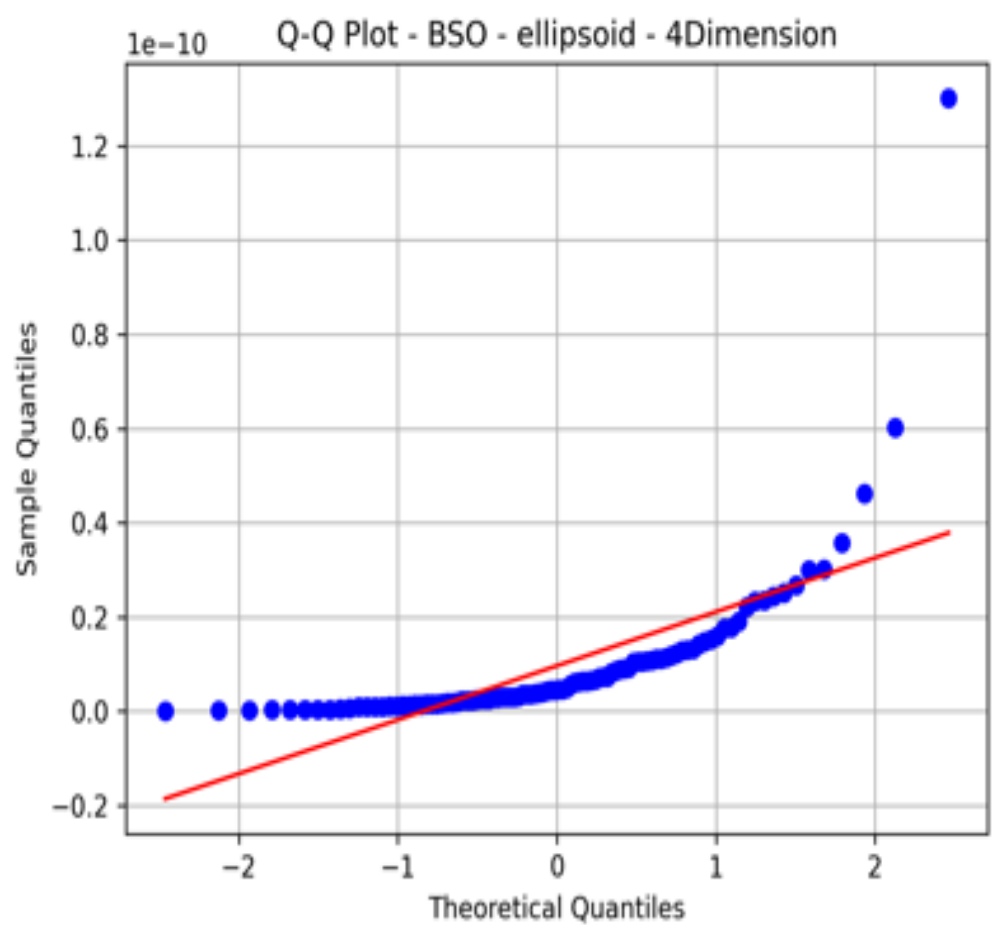
Q-Q Plot - BSO - ellipsoid - 4Dimension
1e−10
Sample Quantiles
Theoretical Quantiles

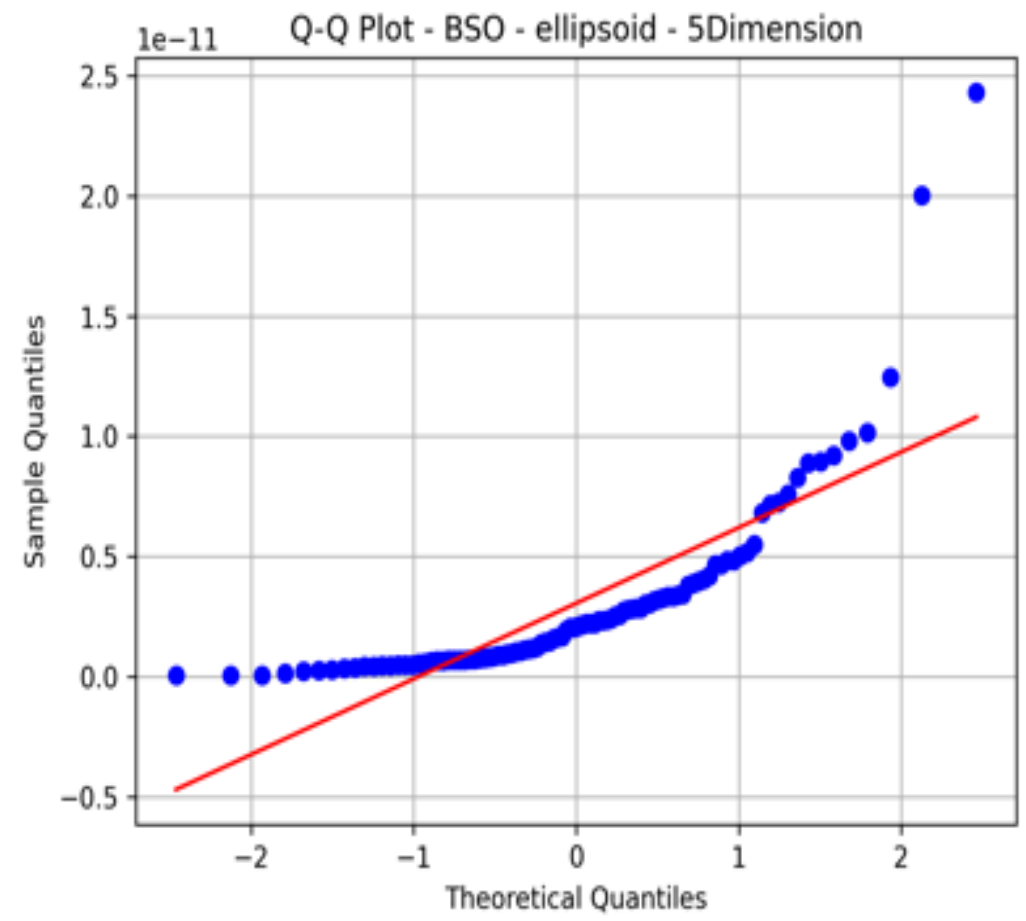
Q-Q Plot - BSO - ellipsoid - 5Dimension
1e−11
Sample Quantiles
Theoretical Quantiles

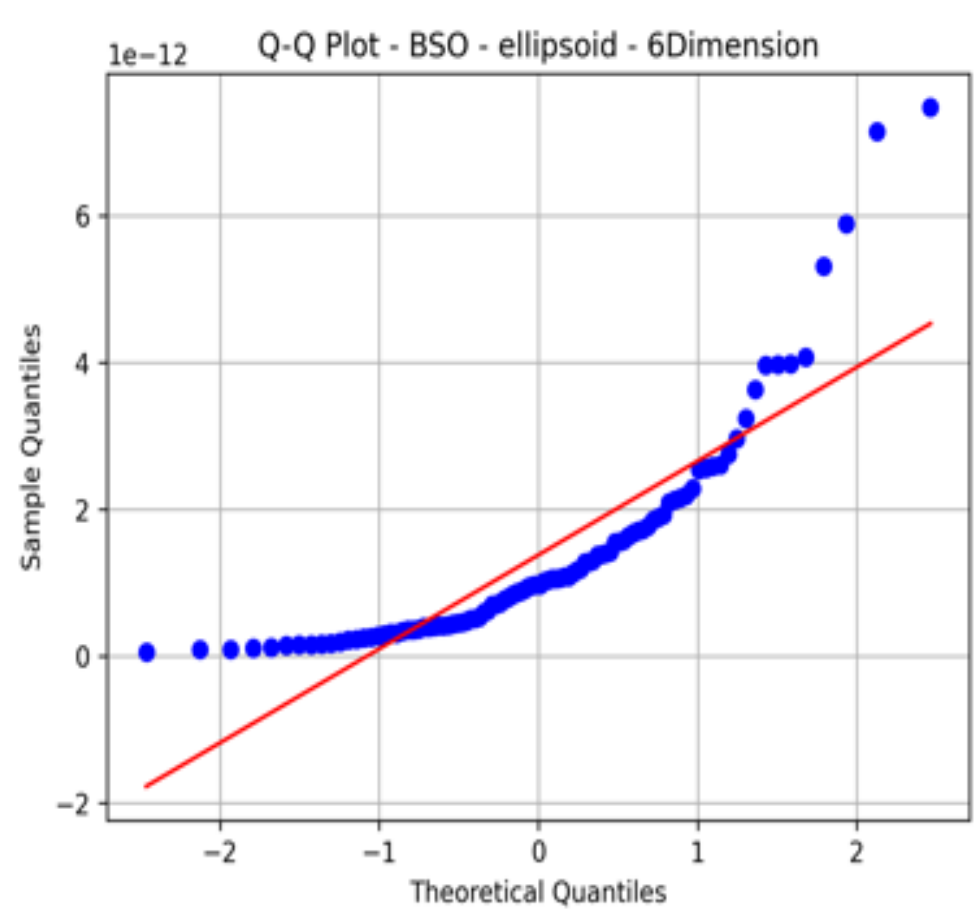
Q-Q Plot - BSO - ellipsoid - 6Dimension
1e−12
Sample Quantiles
Theoretical Quantiles

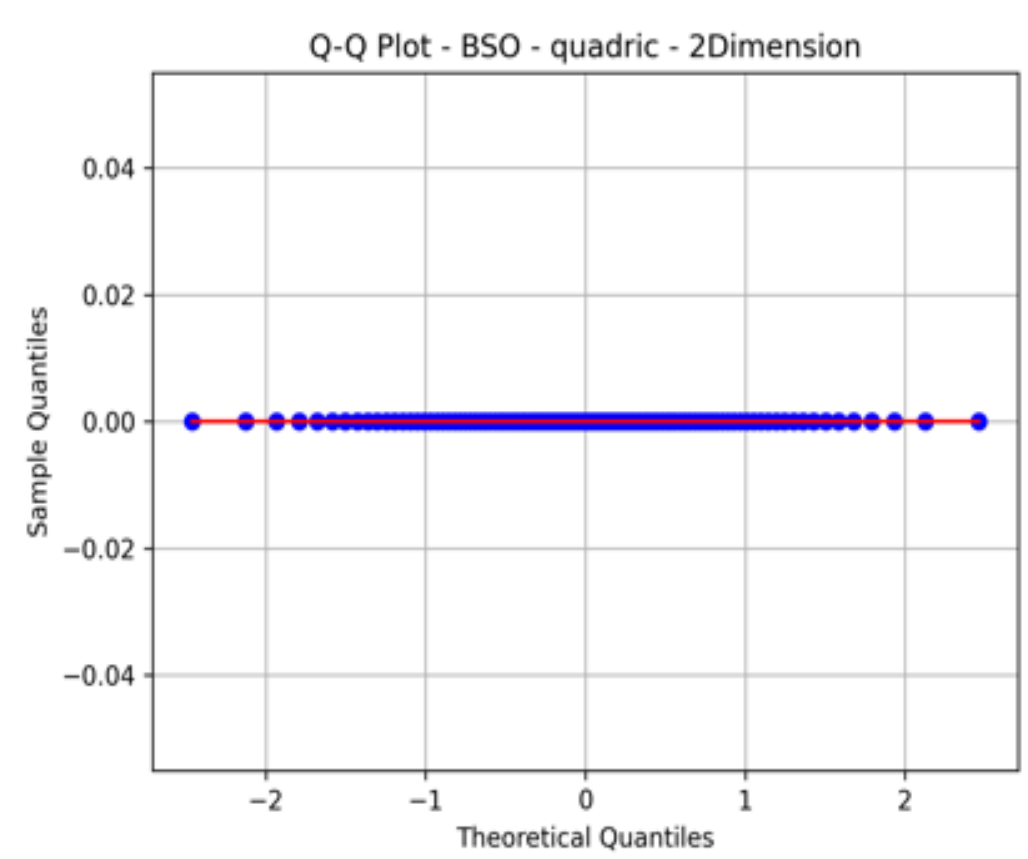
Q-Q Plot - BSO - quadric - 2Dimension
Sample Quantiles
Theoretical Quantiles

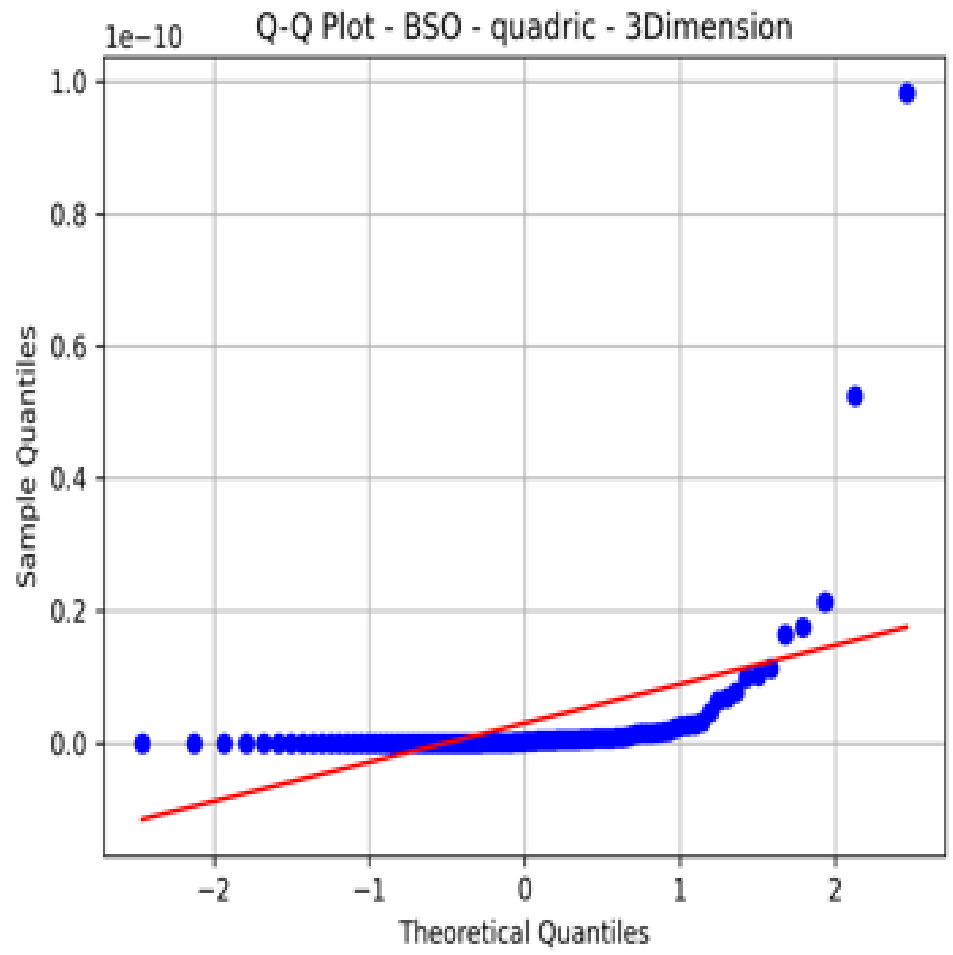
Q-Q Plot - BSO - quadric - 3Dimension
1e−10
Sample Quantiles
Theoretical Quantiles

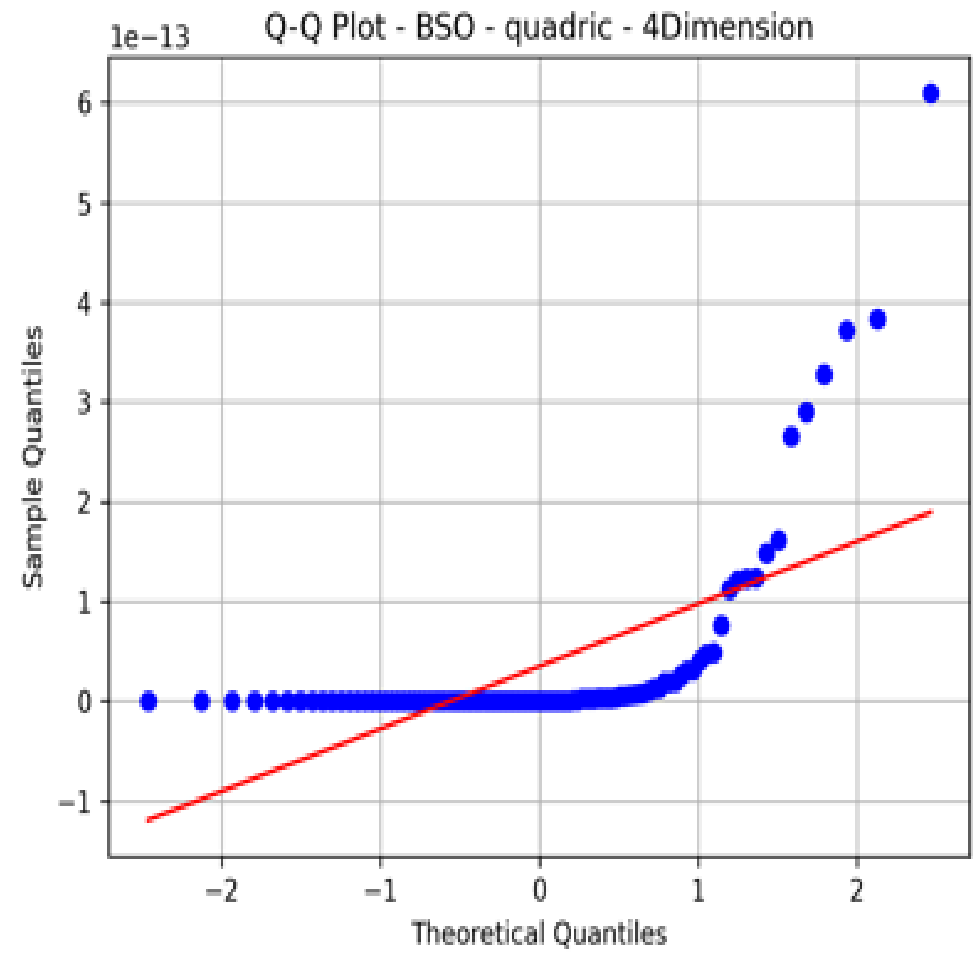
Q-Q Plot - BSO - quadric - 4Dimension
1e−13
Sample Quantiles
Theoretical Quantiles

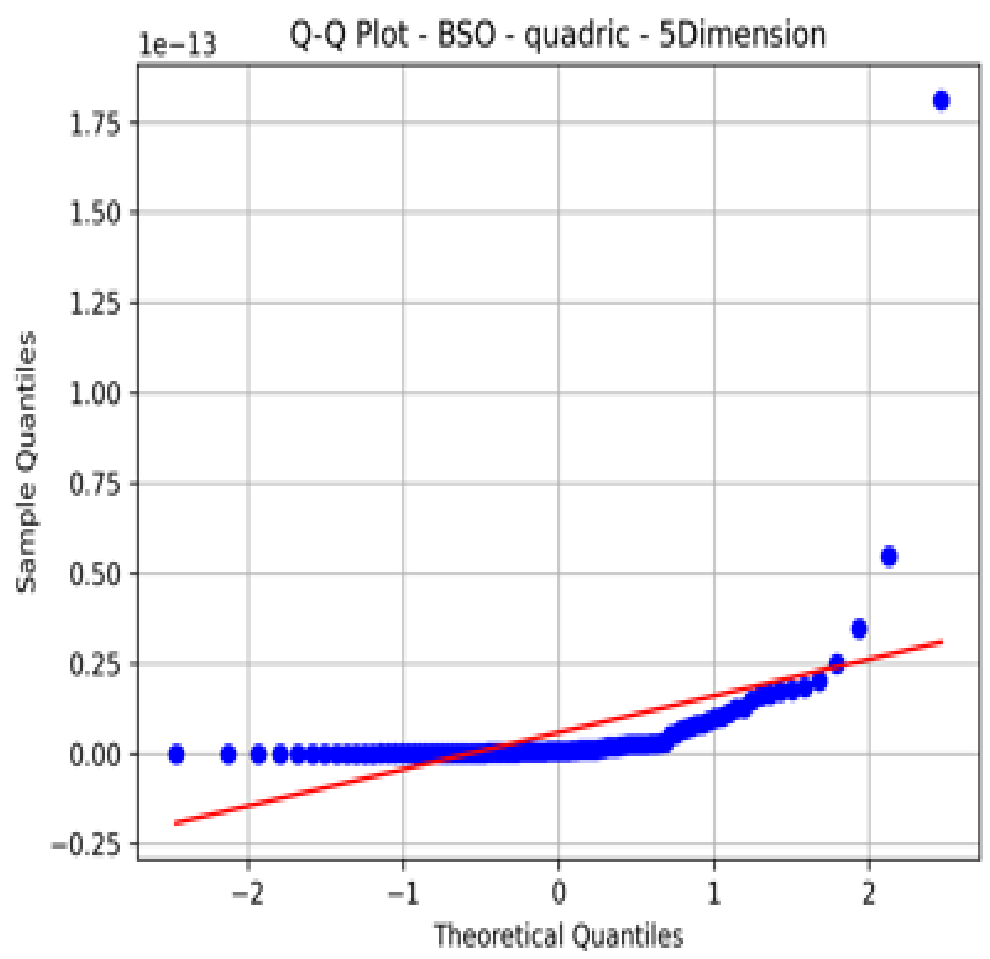
Q-Q Plot - BSO - quadric - 5Dimension
1e−13
Sample Quantiles
Theoretical Quantiles

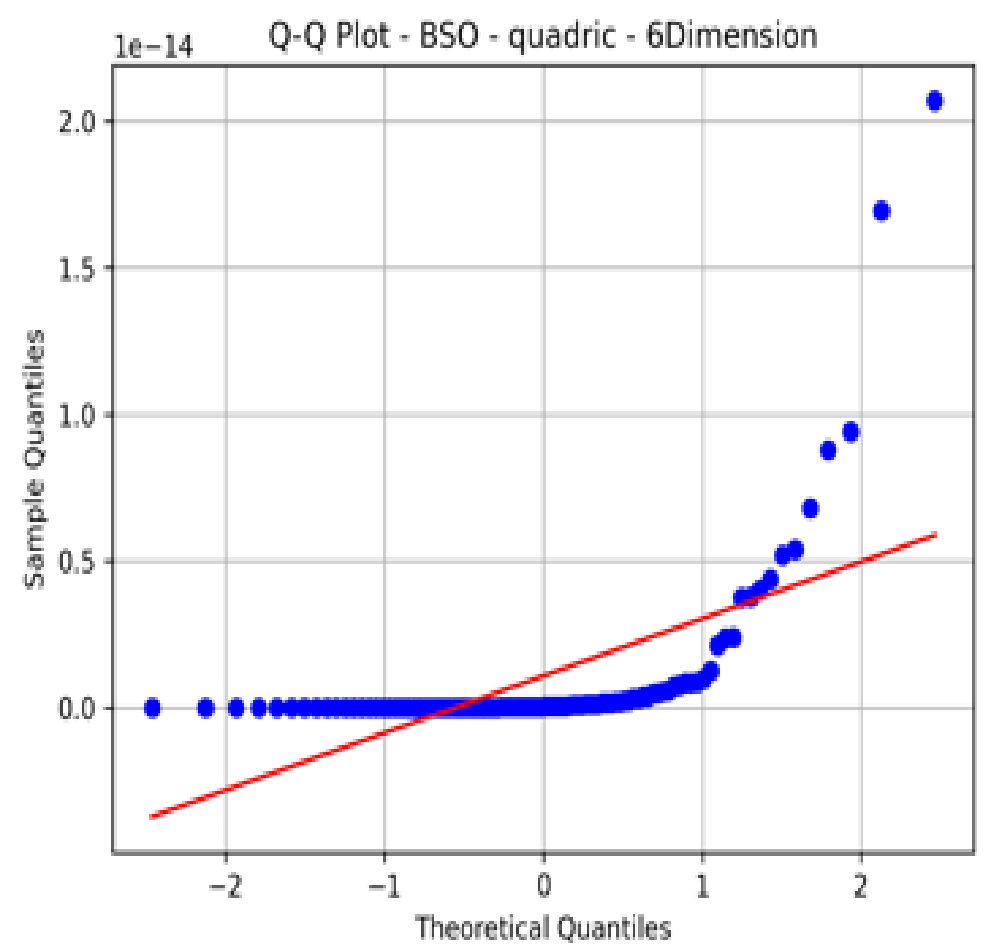
Q-Q Plot - BSO - quadric - 6Dimension
1e−14
Sample Quantiles
Theoretical Quantiles

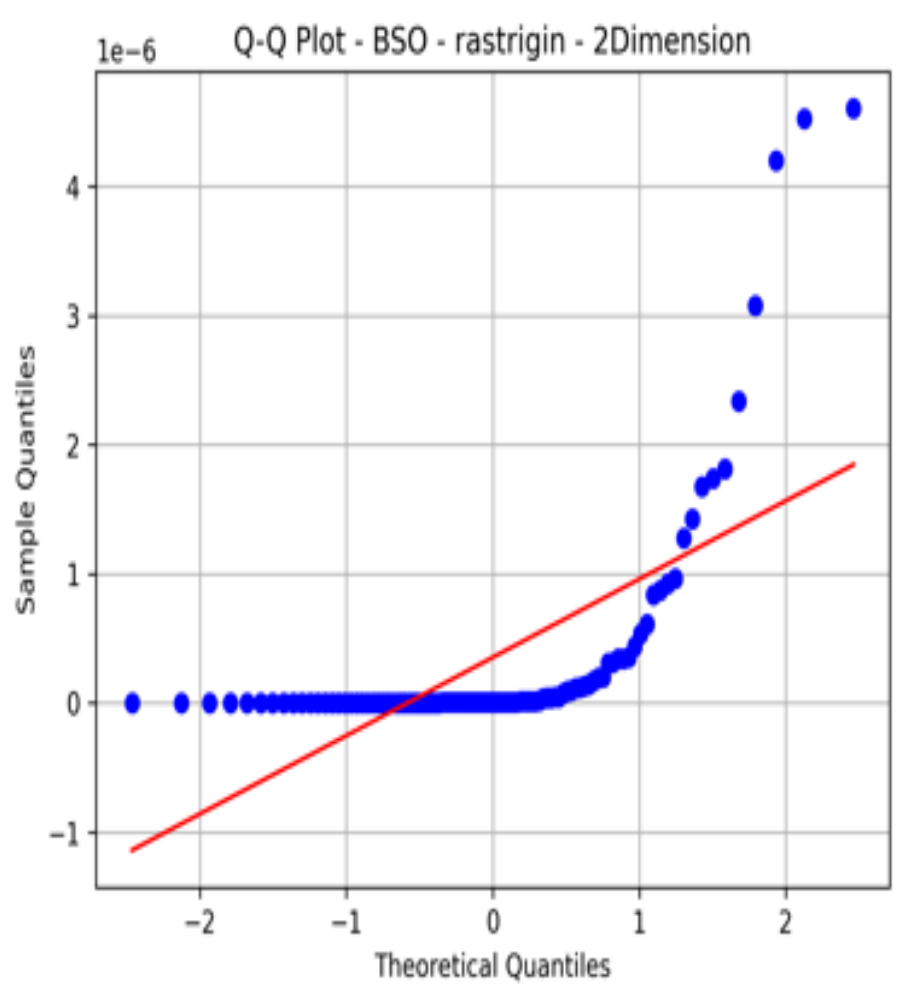
Q-Q Plot - BSO - rastrigin - 2Dimension
1e−6
Sample Quantiles
Theoretical Quantiles

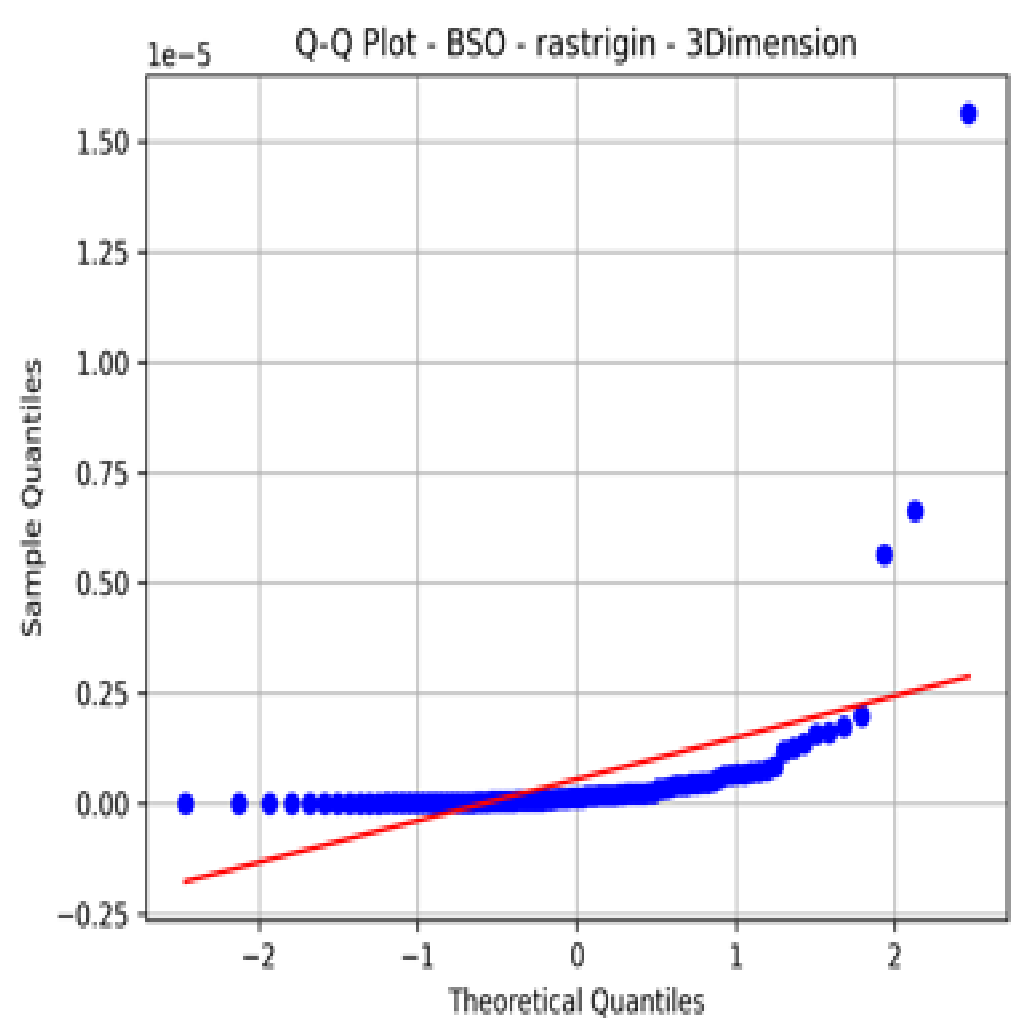
Q-Q Plot - BSO - rastrigin - 3Dimension
1e−5
Sample Quantiles
Theoretical Quantiles

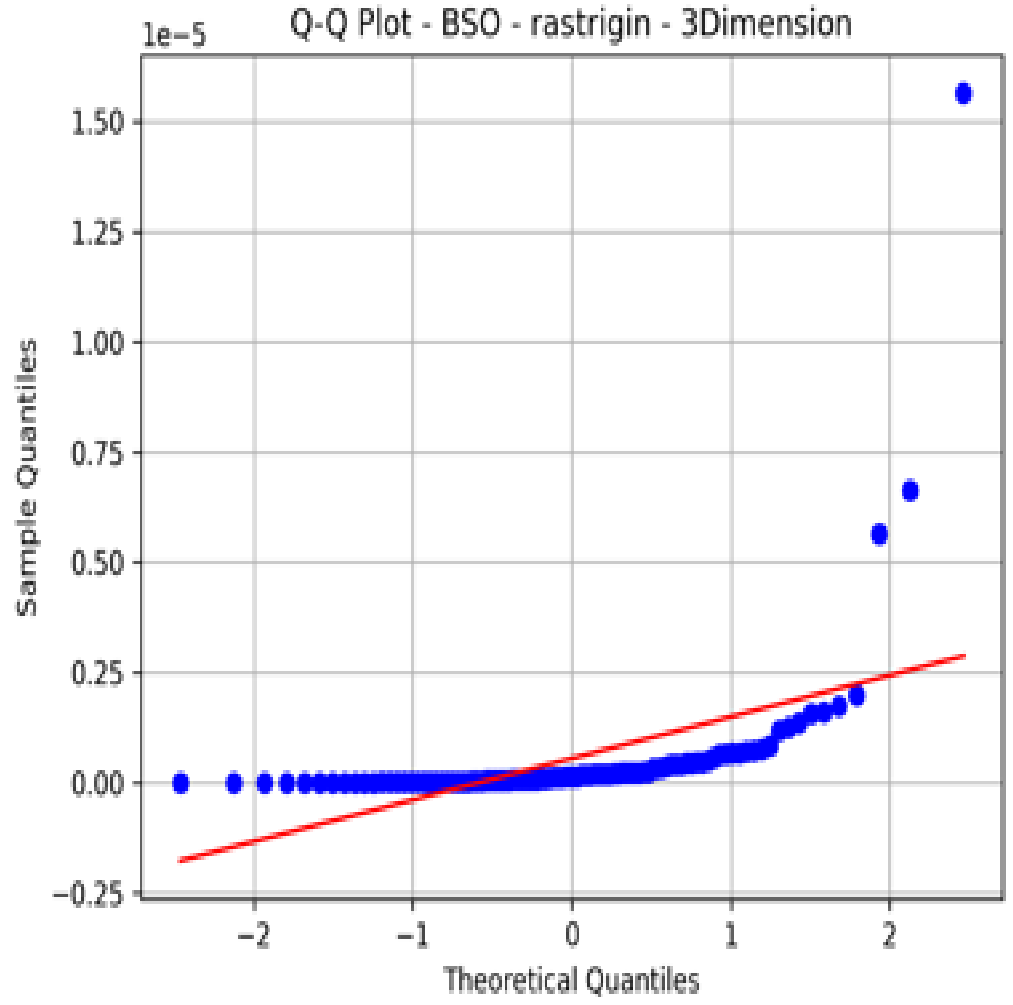
Q-Q Plot - BSO - rastrigin - 3Dimension
1e−5
Sample Quantiles
Theoretical Quantiles

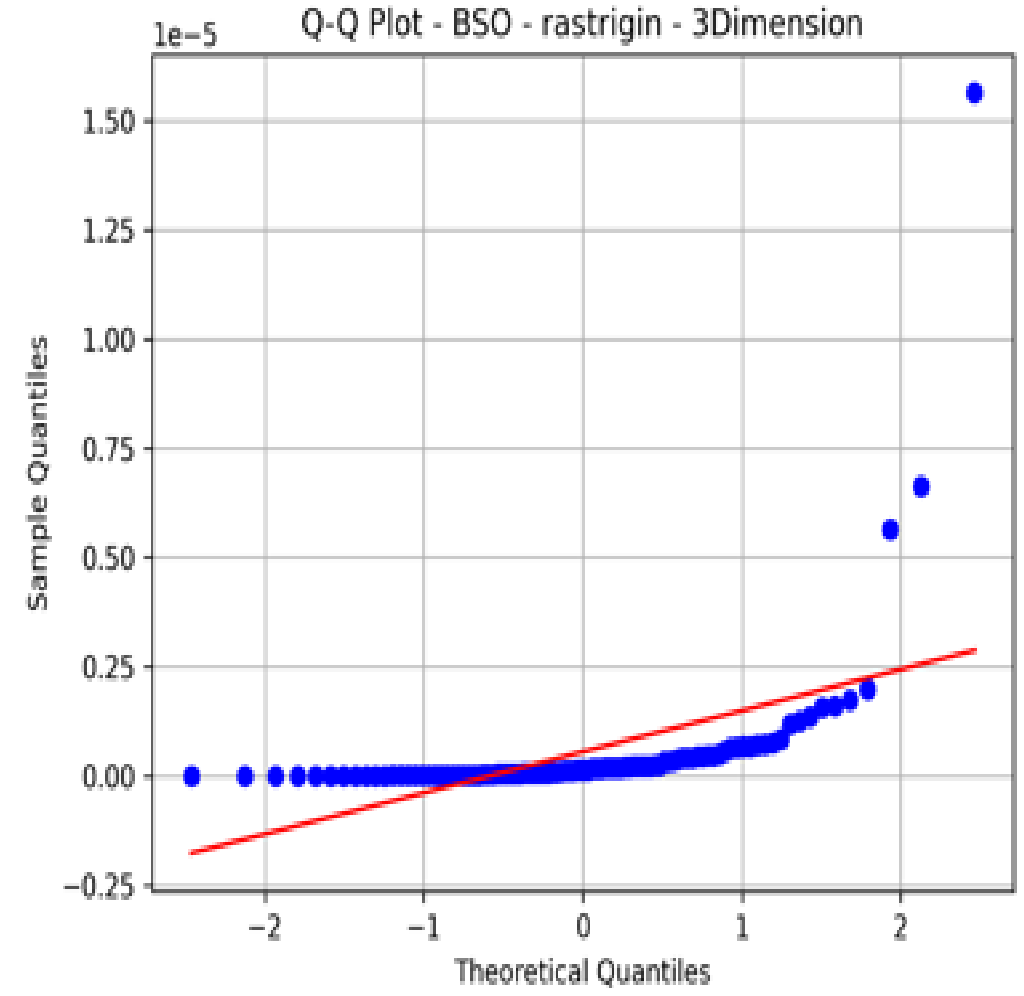
Q-Q Plot - BSO - rastrigin - 3Dimension
1e−5
Sample Quantiles
Theoretical Quantiles

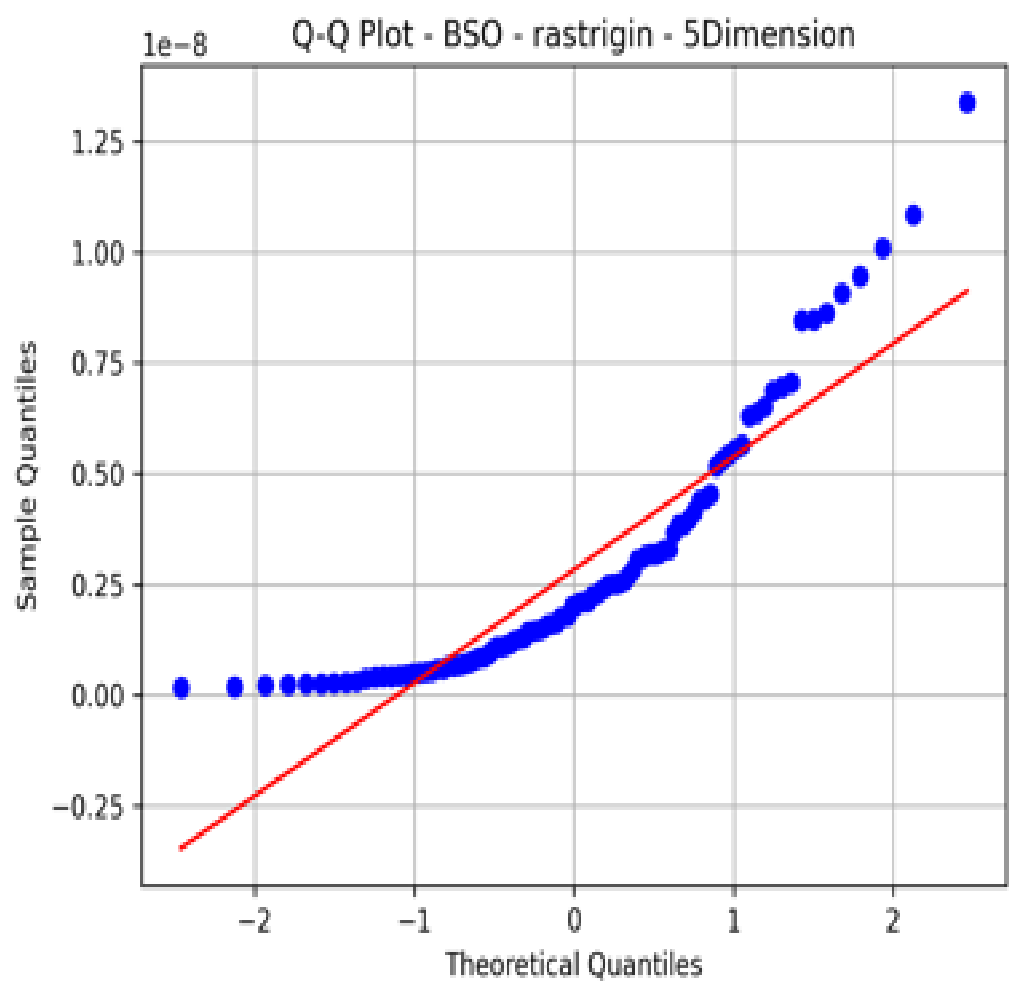
Q-Q Plot - BSO - rastrigin - 5Dimension
1e−8
Sample Quantiles
Theoretical Quantiles

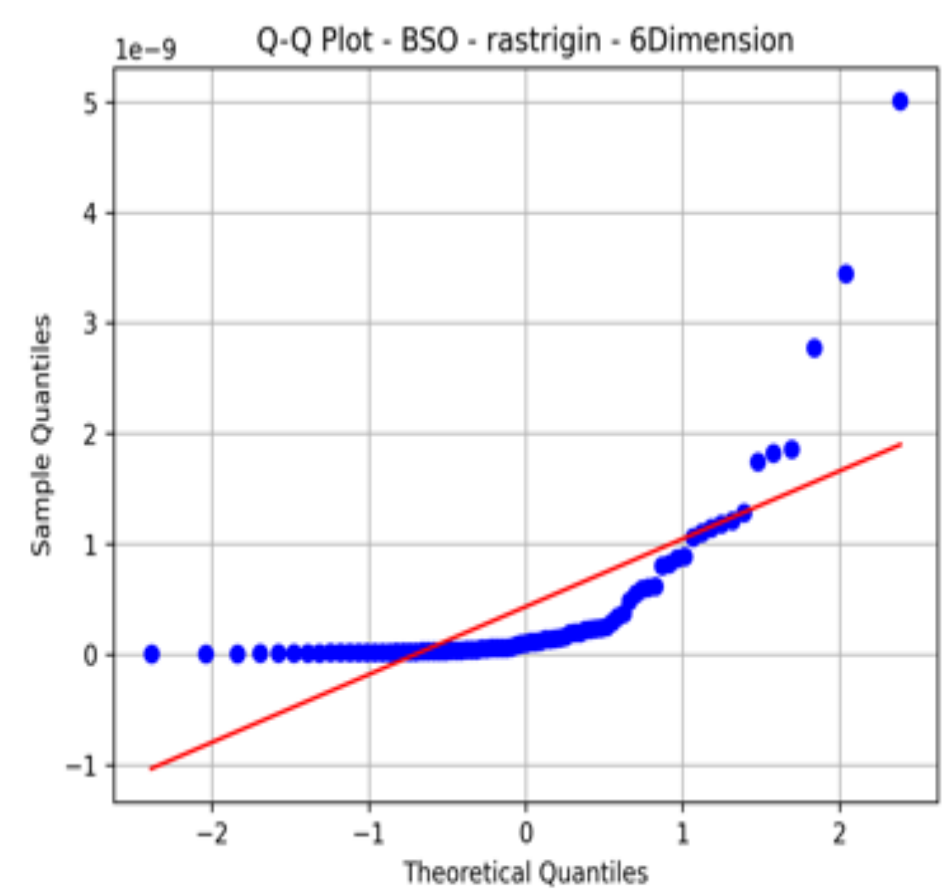
Q-Q Plot - BSO - rastrigin - 6Dimension
1e−9
Sample Quantiles
Theoretical Quantiles

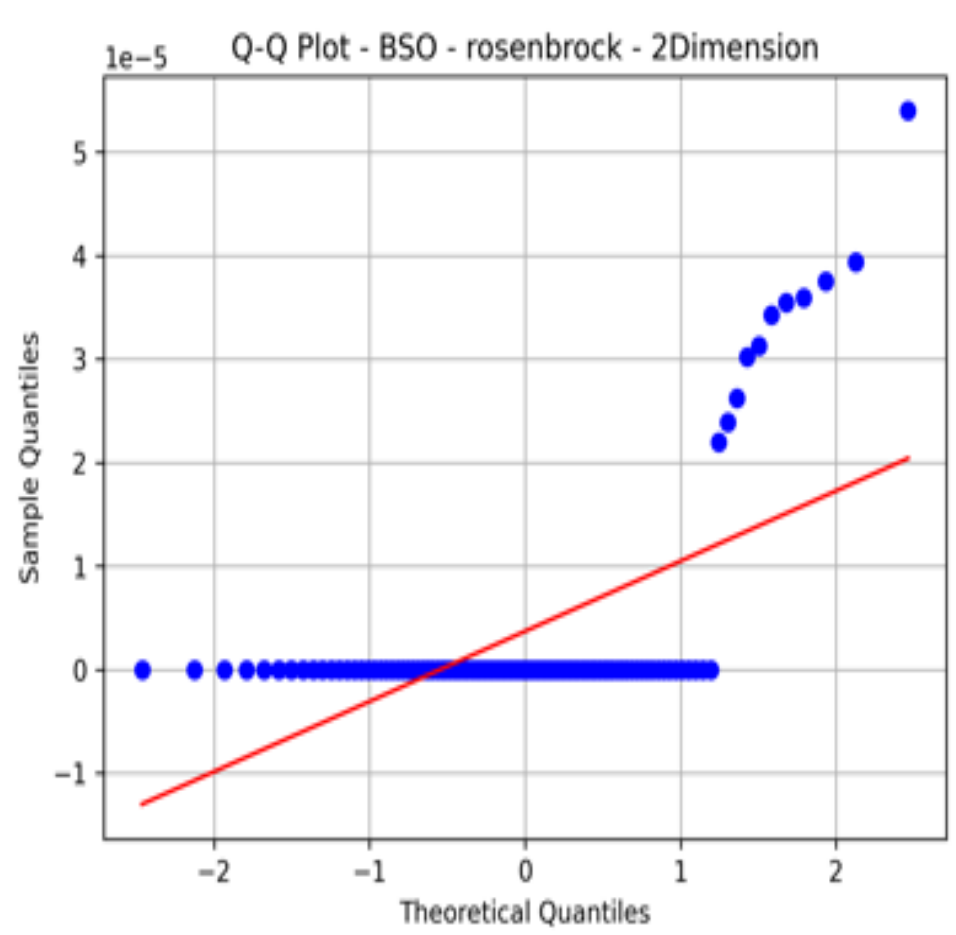
Q-Q Plot - BSO - rosenbrock - 2Dimension
1e−5
Sample Quantiles
Theoretical Quantiles

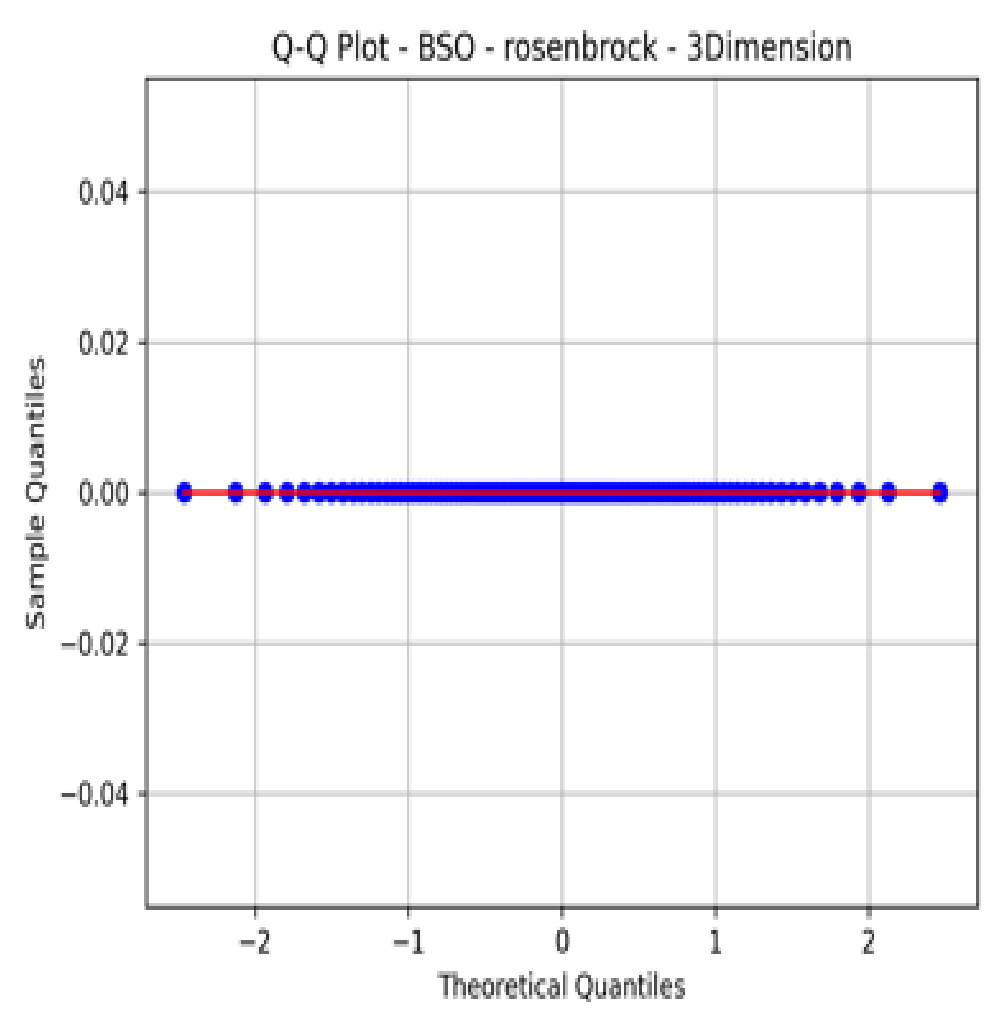
Q-Q Plot - BSO - rosenbrock - 3Dimension
Sample Quantiles
Theoretical Quantiles

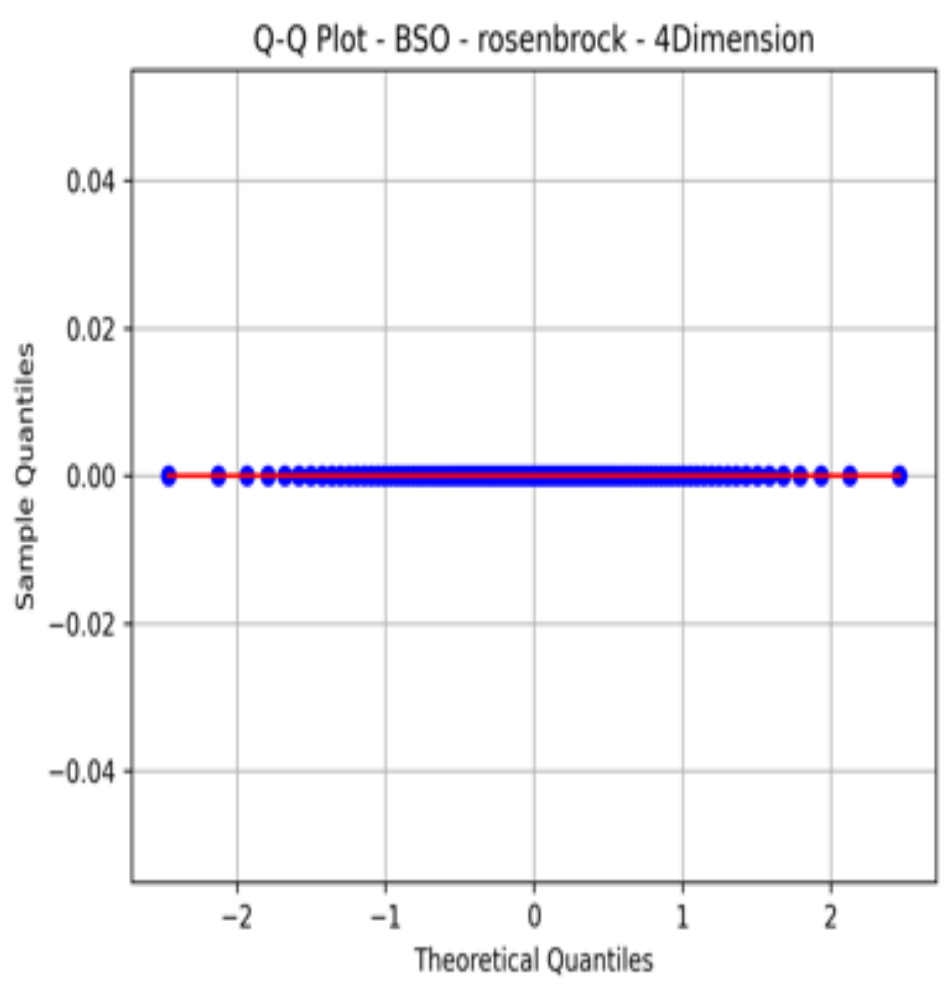
Q-Q Plot - BSO - rosenbrock - 4Dimension
Sample Quantiles
0.04
0.02
0.00
−0.02
−0.04
−2
−1
0
1
2
Theoretical Quantiles

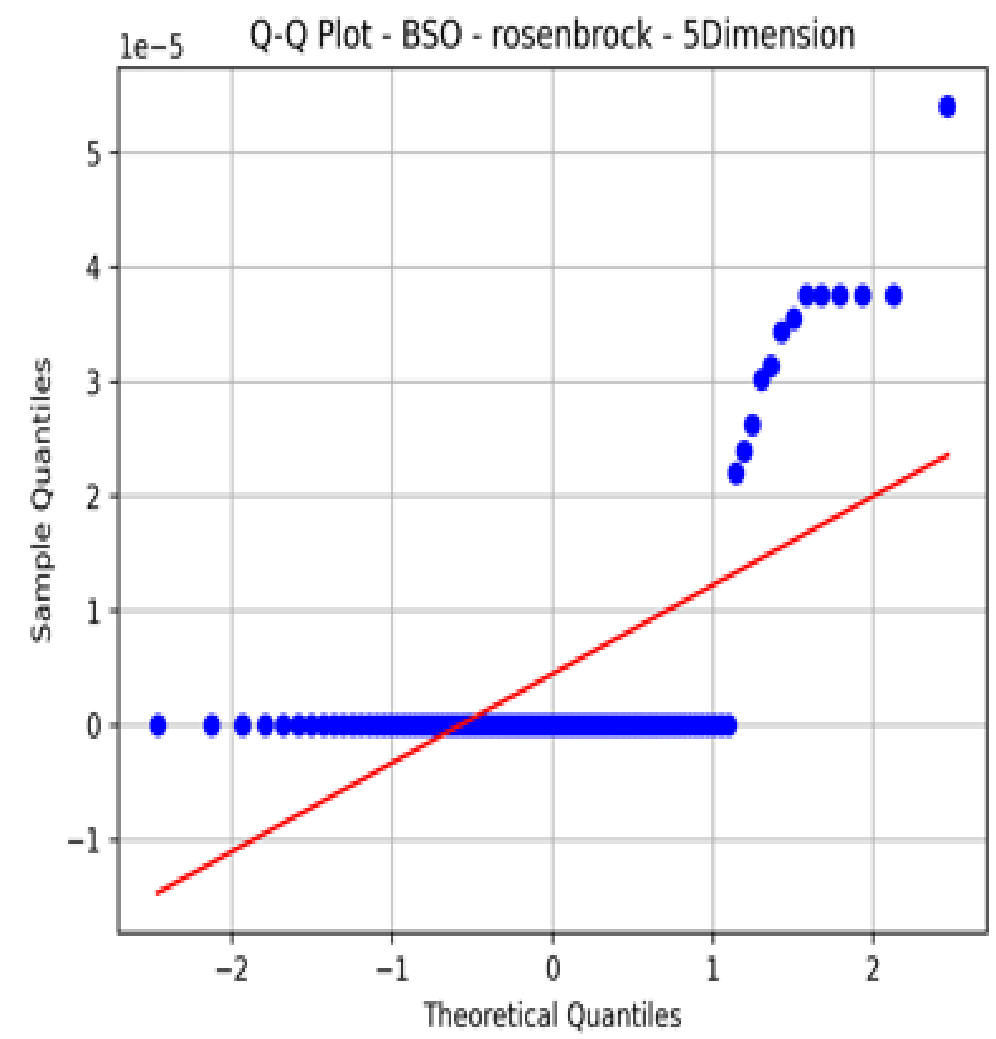
1e−5
Q-Q Plot - BSO - rosenbrock - 5Dimension
Sample Quantiles
5
4
3
2
1
0
−1
−2
−1
0
1
2
Theoretical Quantiles

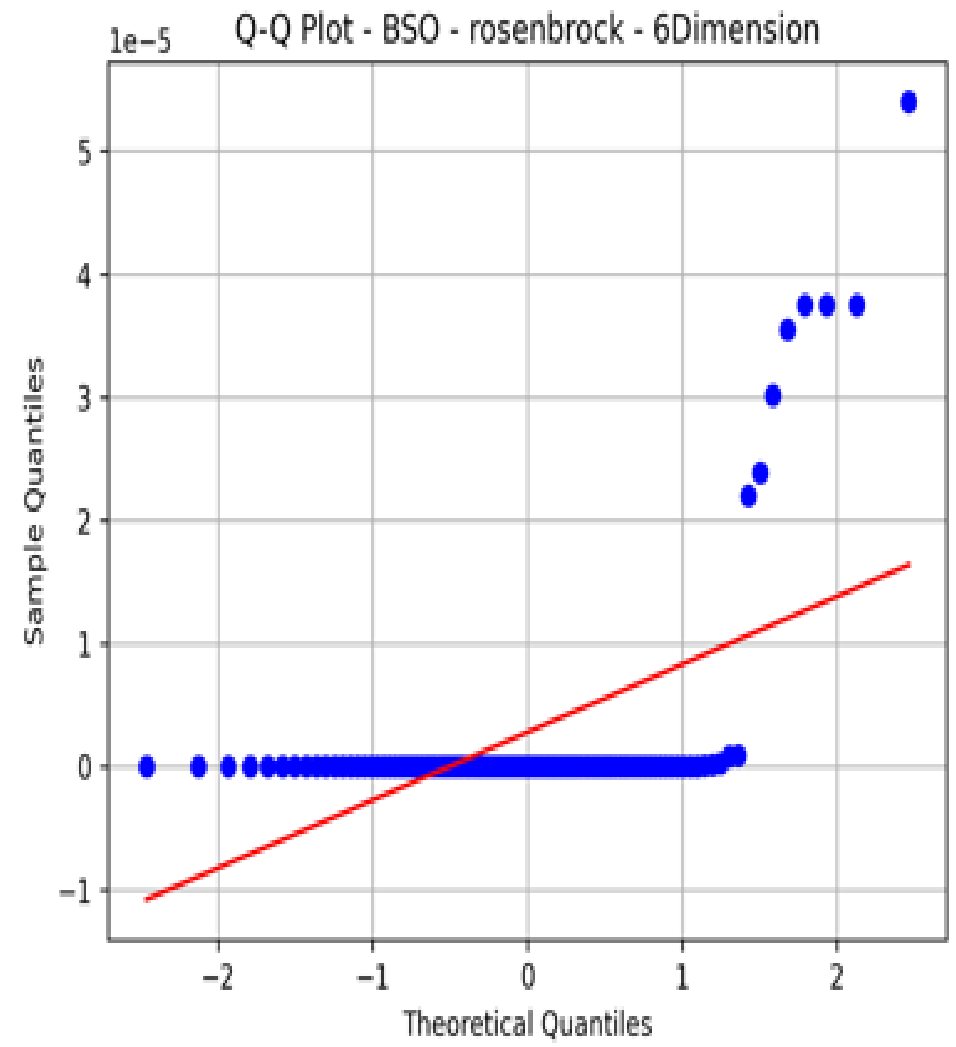
1e−5
Q-Q Plot - BSO - rosenbrock - 6Dimension
Sample Quantiles
5
4
3
2
1
0
−1
−2
−1
0
1
2
Theoretical Quantiles

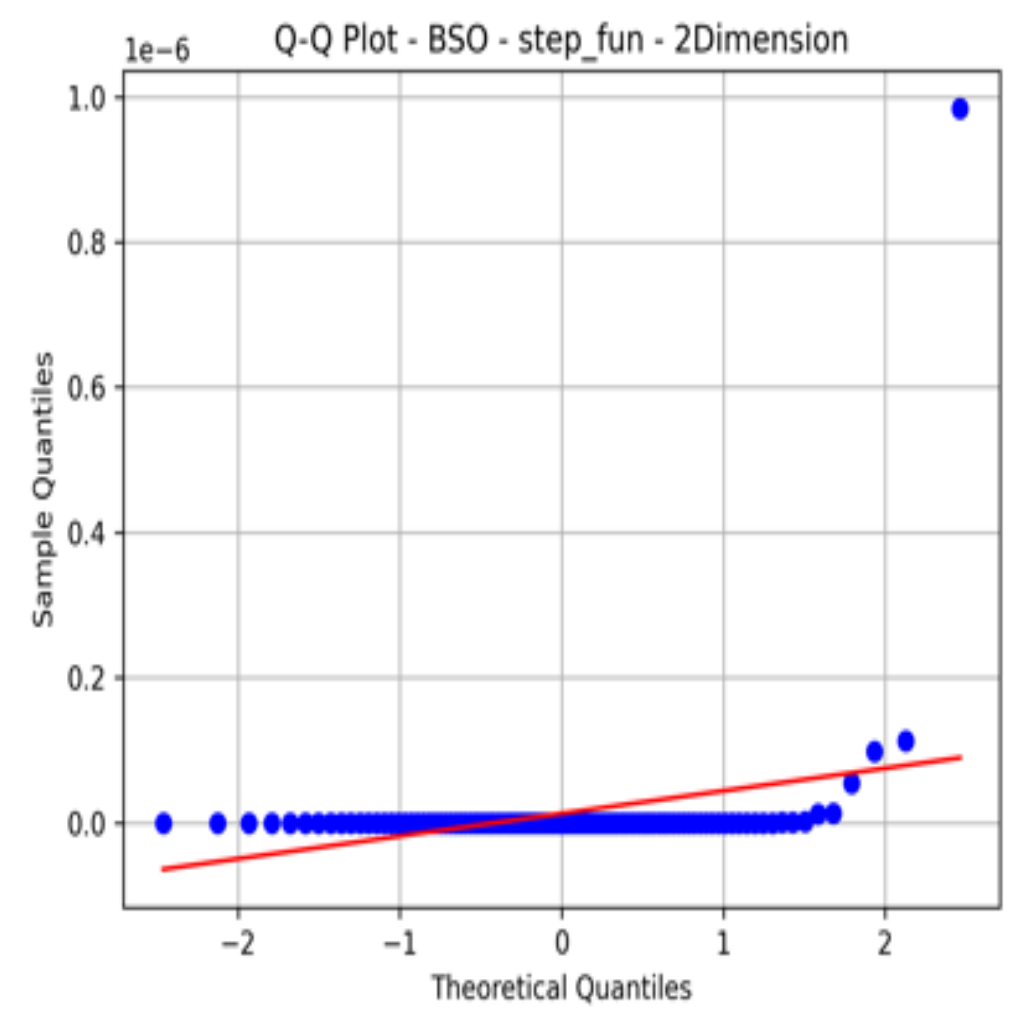
1e−6
Q-Q Plot - BSO - step_fun - 2Dimension
Sample Quantiles
1.0
0.8
0.6
0.4
0.2
0.0
−2
−1
0
1
2
Theoretical Quantiles

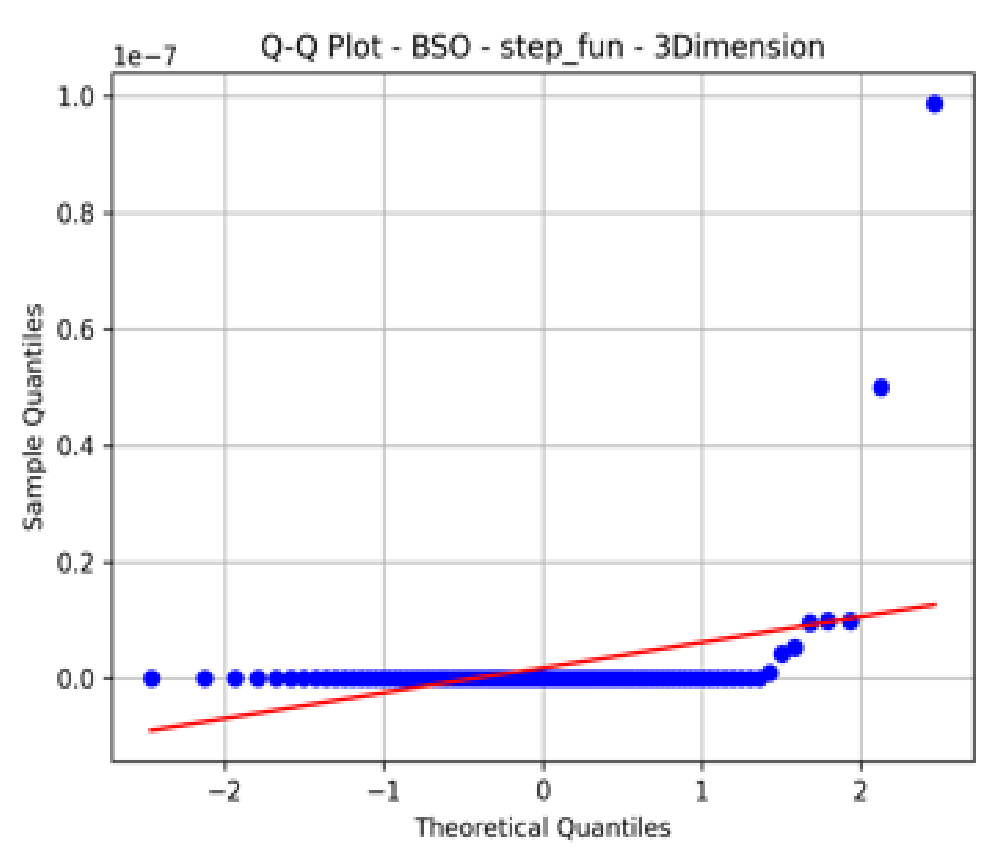
1e−7
Q-Q Plot - BSO - step_fun - 3Dimension
Sample Quantiles
1.0
0.8
0.6
0.4
0.2
0.0
−2
−1
0
1
2
Theoretical Quantiles

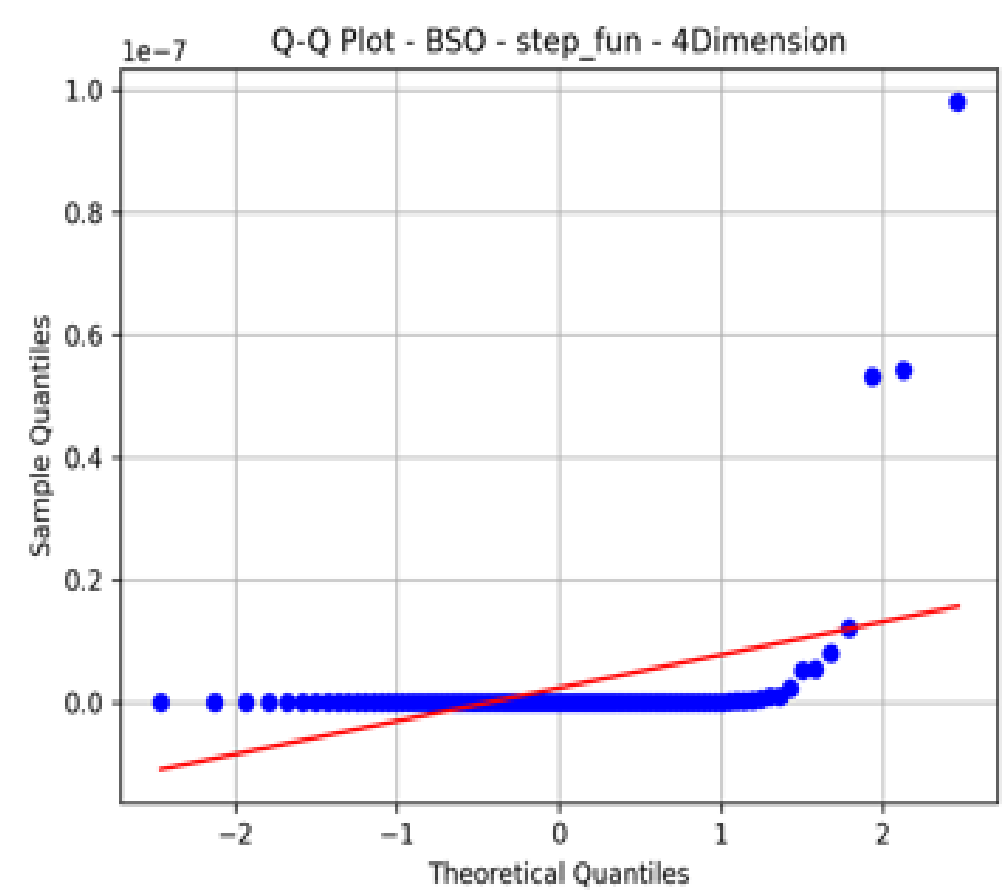
1e−7
Q-Q Plot - BSO - step_fun - 4Dimension
Sample Quantiles
1.0
0.8
0.6
0.4
0.2
0.0
−2
−1
0
1
2
Theoretical Quantiles

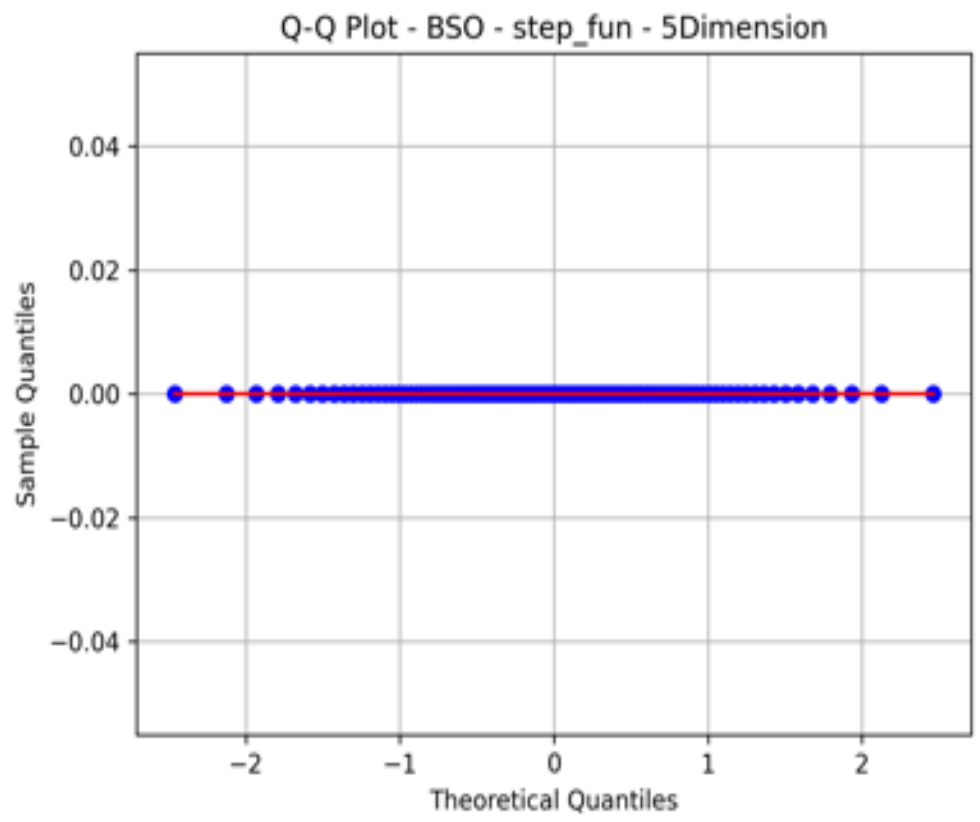
Q-Q Plot - BSO - step_fun - 5Dimension
Sample Quantiles
Theoretical Quantiles

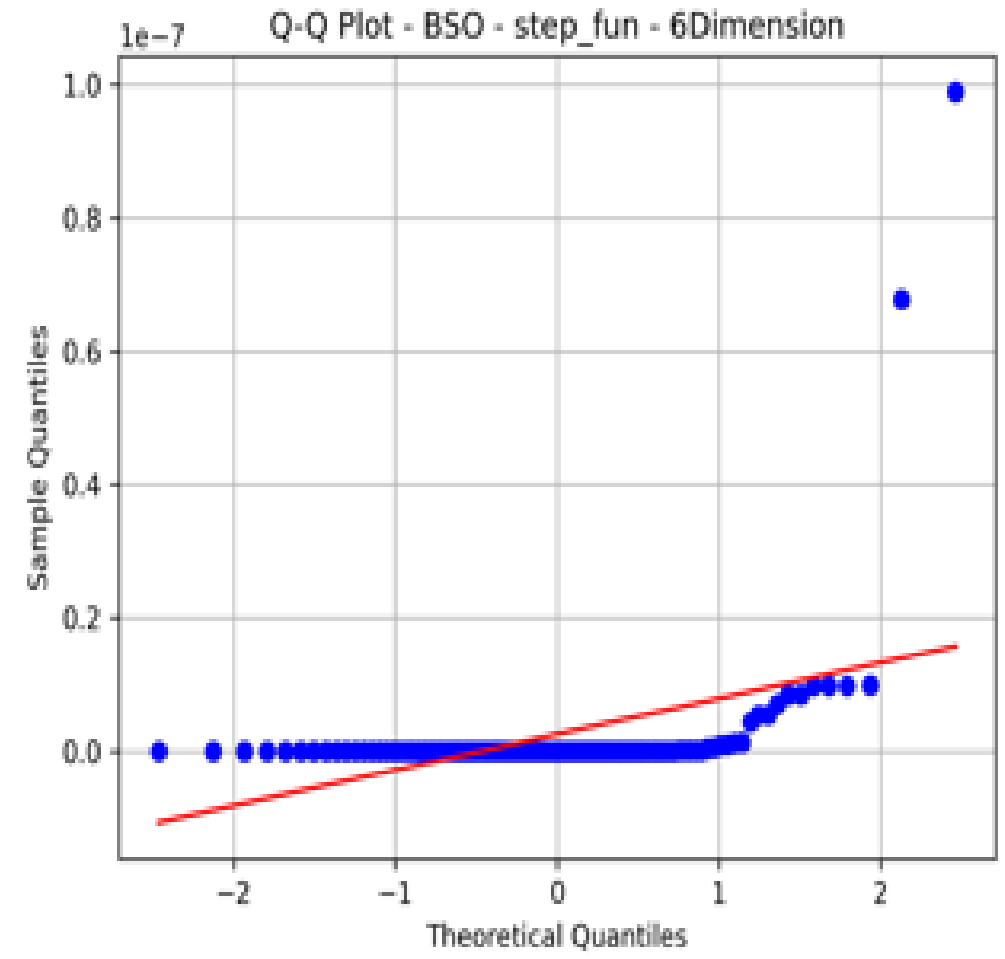
1e−7
Q-Q Plot - BSO - step_fun - 6Dimension
Sample Quantiles
Theoretical Quantiles

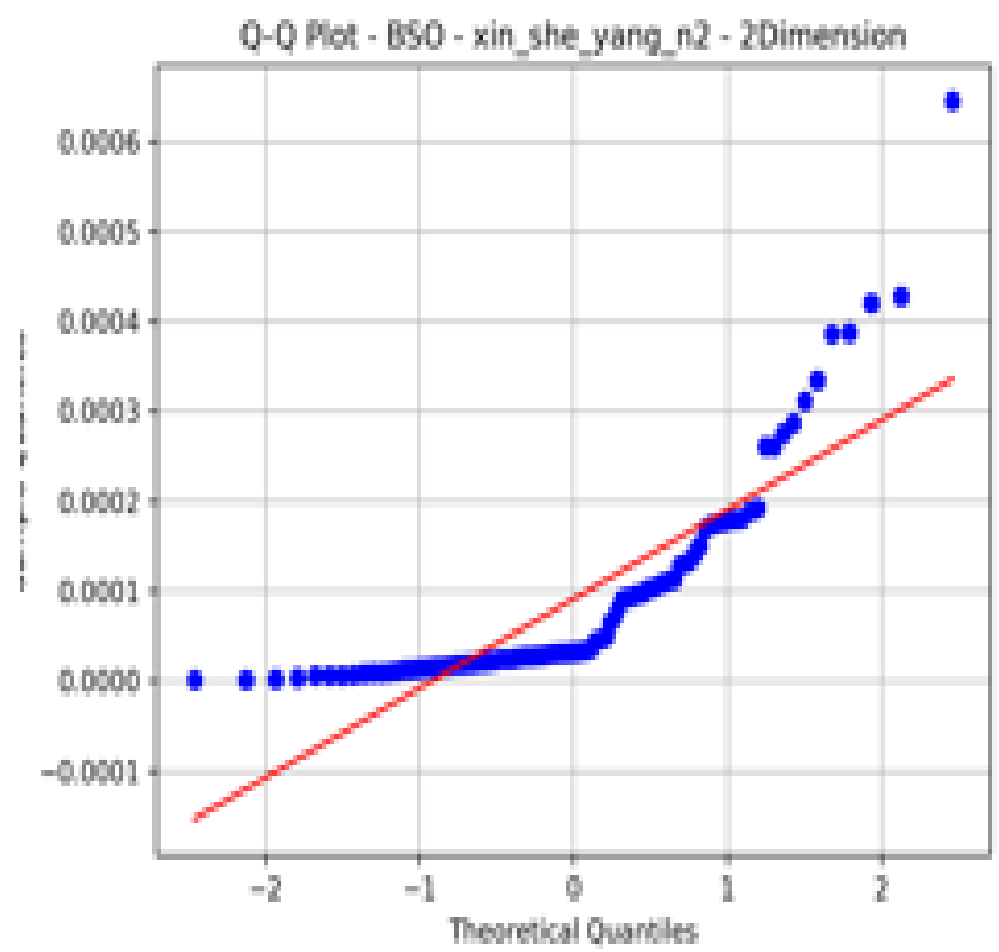
Q-Q Plot - BSO - xin_she_yang_n2 - 2Dimension
Theoretical Quantiles

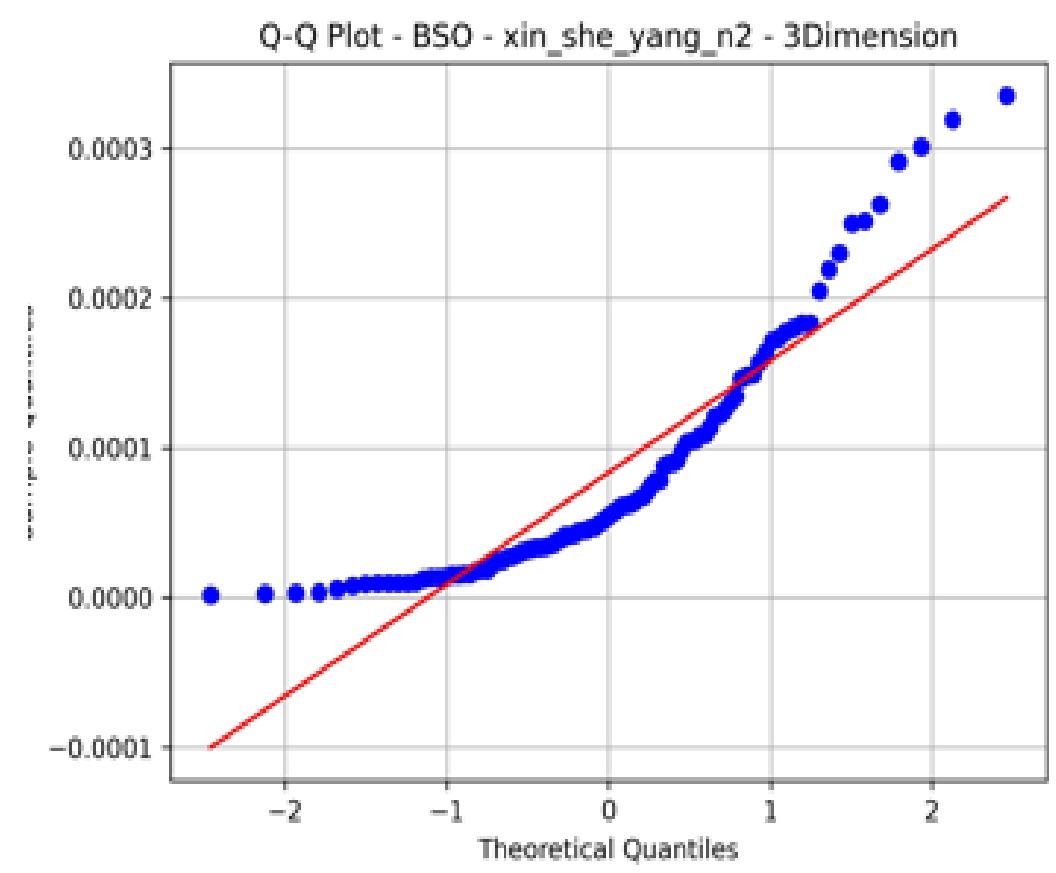
Q-Q Plot - BSO - xin_she_yang_n2 - 3Dimension
Theoretical Quantiles

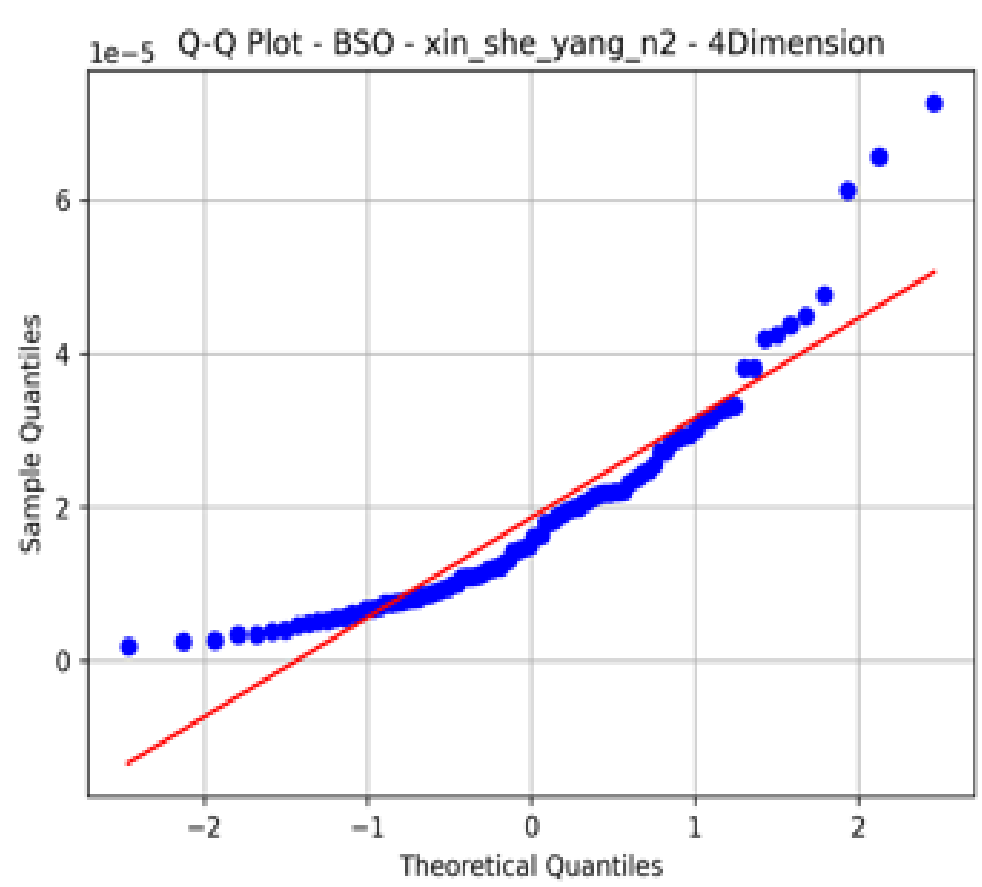
1e−5
Q-Q Plot - BSO - xin_she_yang_n2 - 4Dimension
Sample Quantiles
Theoretical Quantiles

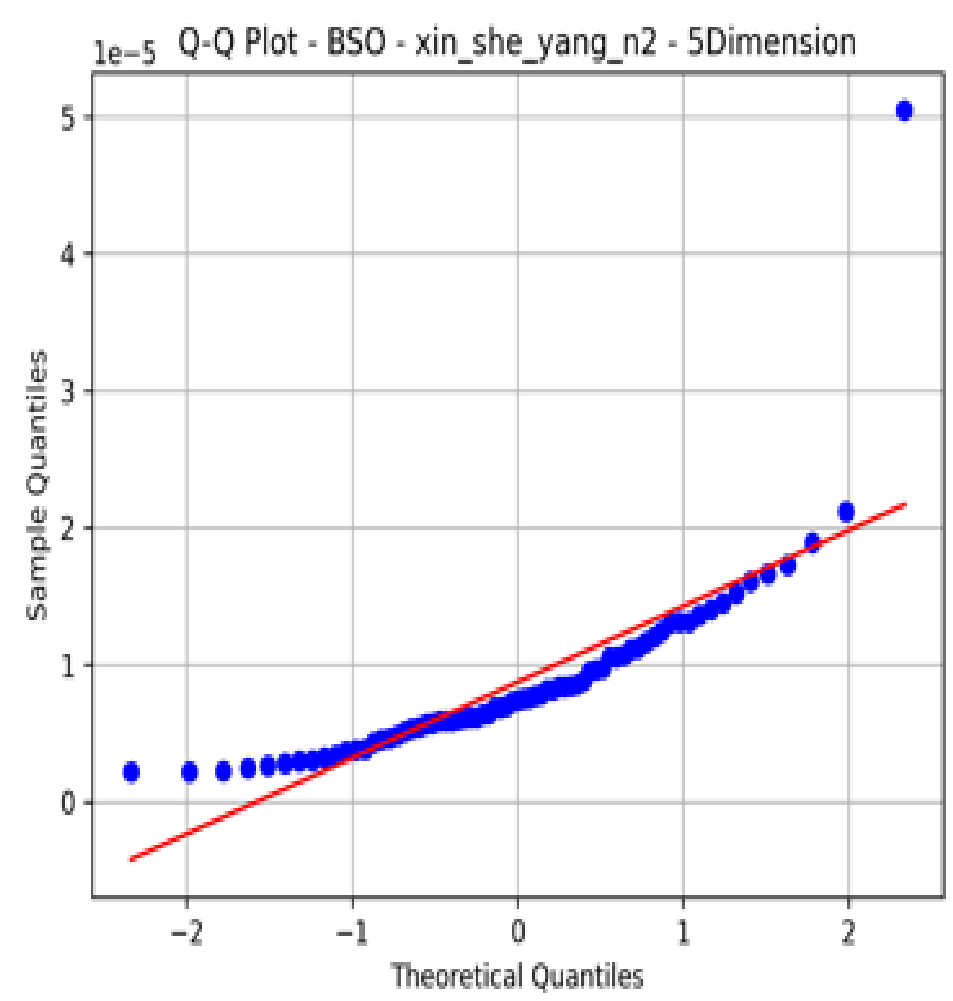
1e−5
Q-Q Plot - BSO - xin_she_yang_n2 - 5Dimension
Sample Quantiles
Theoretical Quantiles

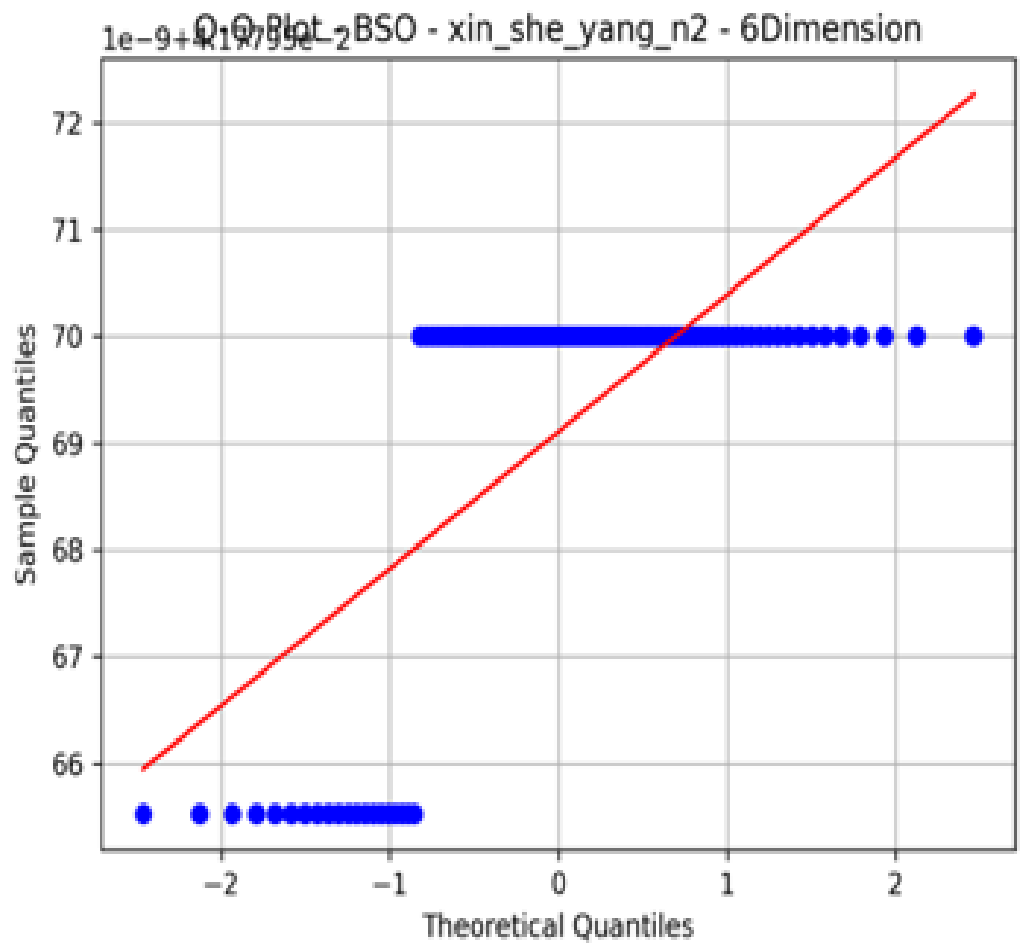

<u>BPO QQ plots</u>

Non-Continuous functions QQ Plots

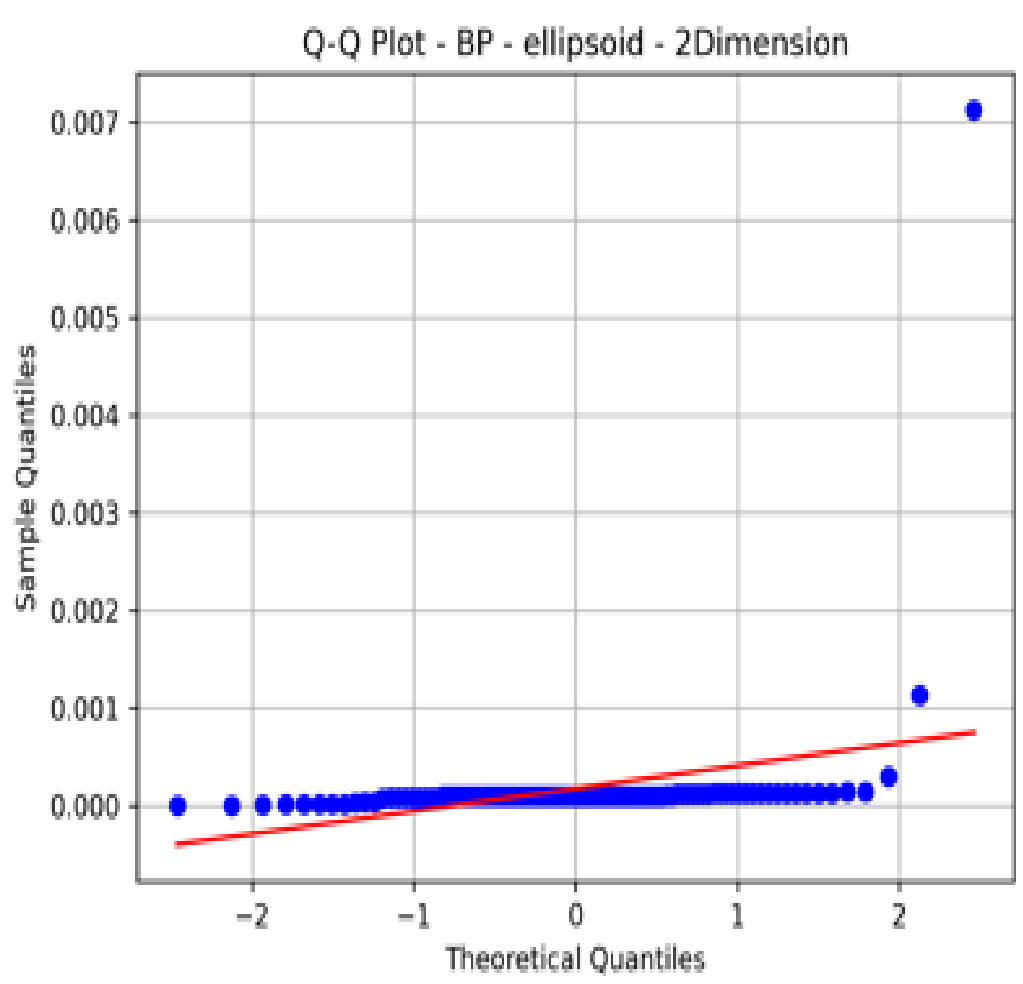

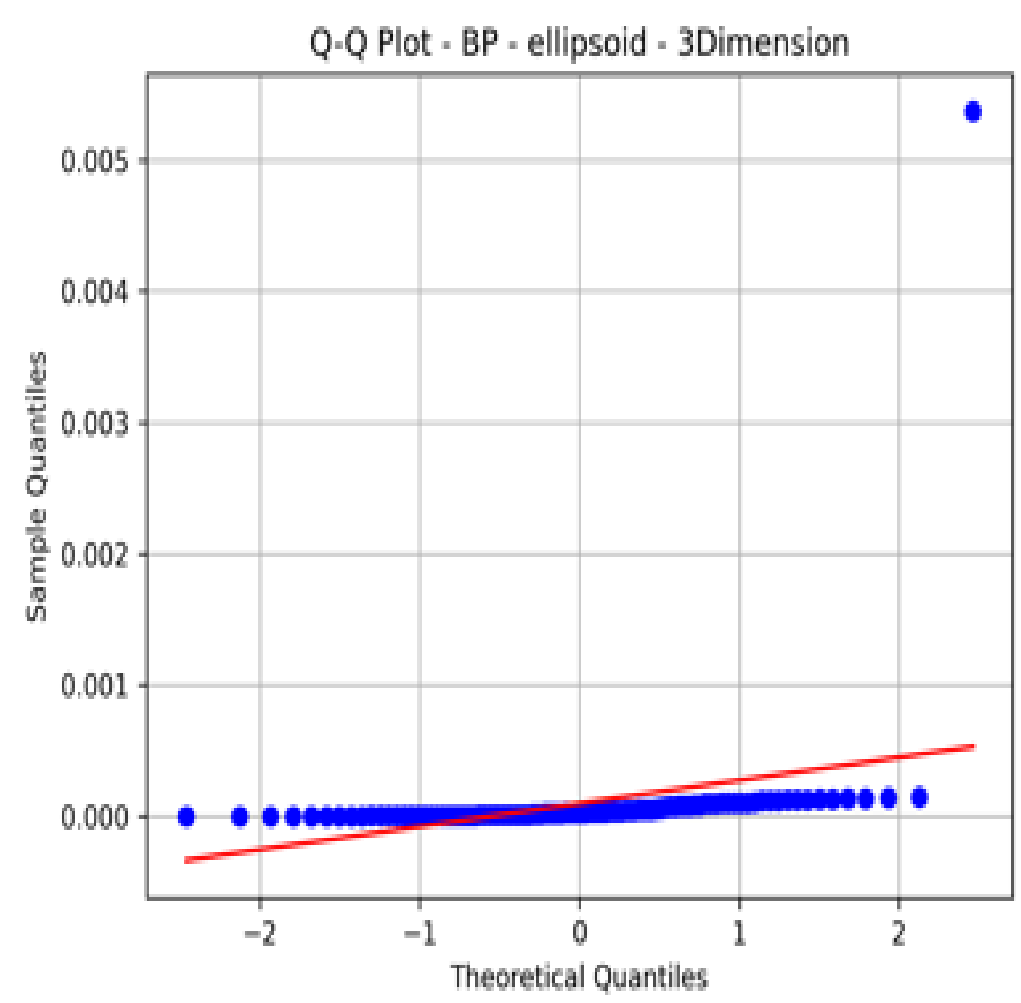

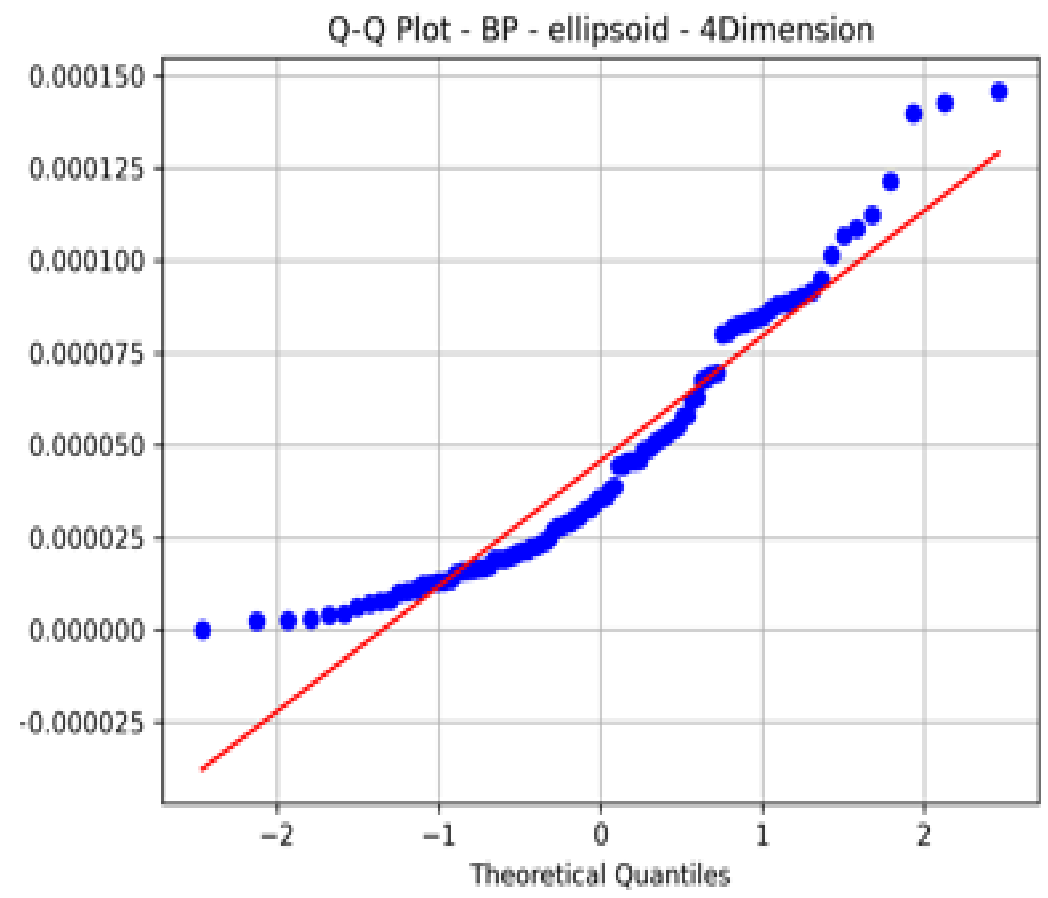
Q-Q Plot - BP - ellipsoid - 4Dimension
Theoretical Quantiles

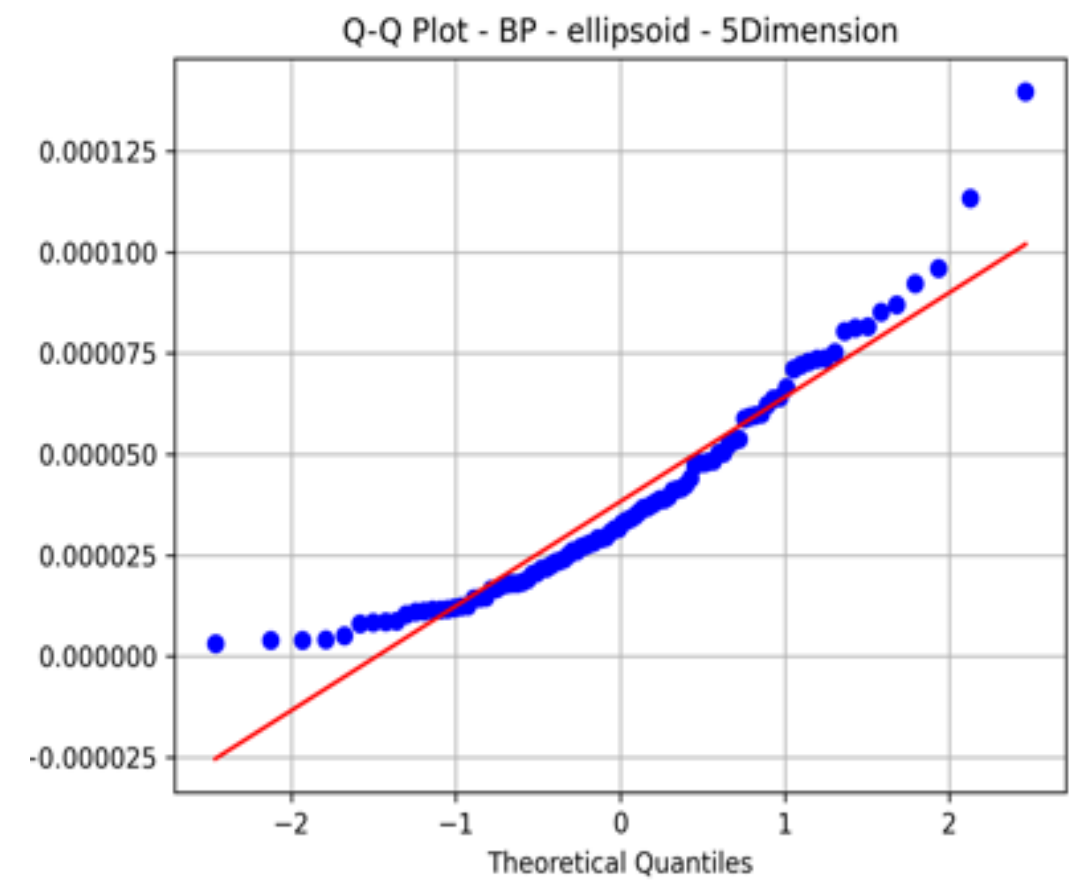
Q-Q Plot - BP - ellipsoid - 5Dimension
Theoretical Quantiles

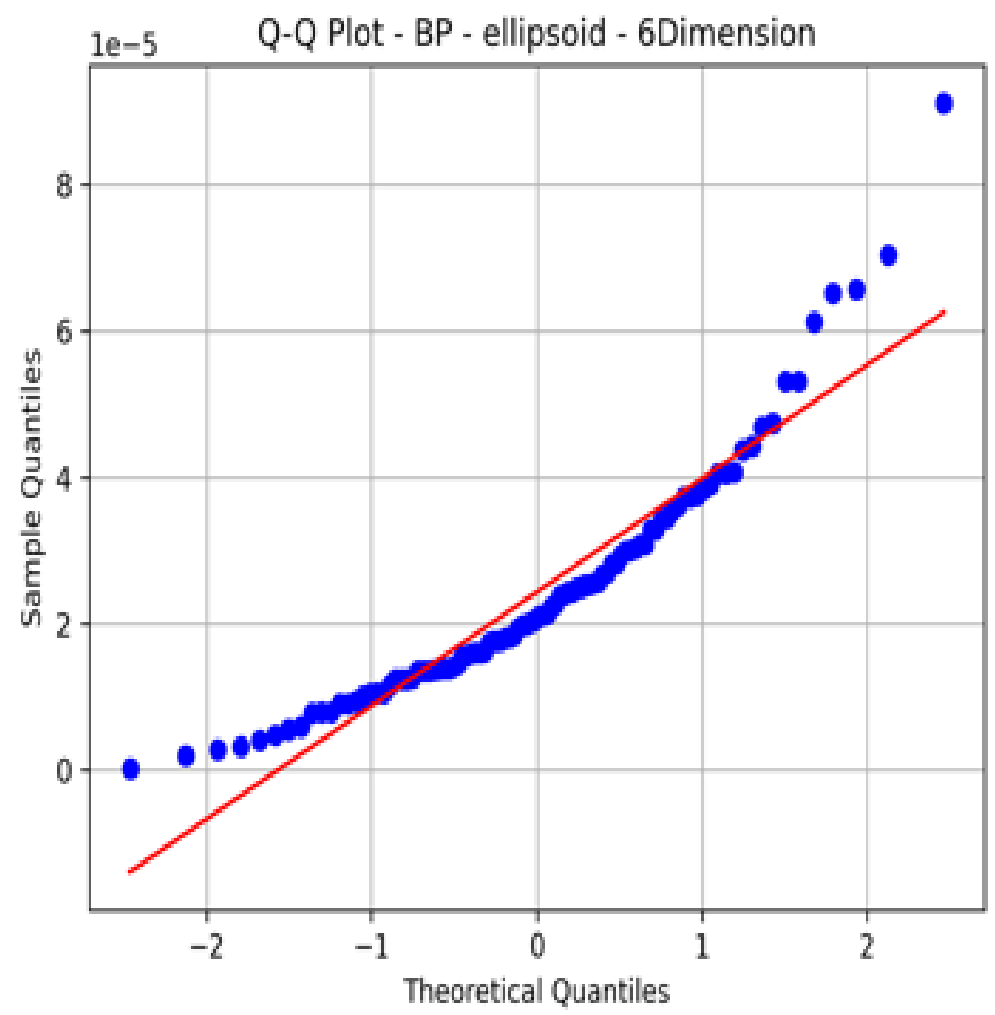
Q-Q Plot - BP - ellipsoid - 6Dimension
1e−5
Sample Quantiles
Theoretical Quantiles

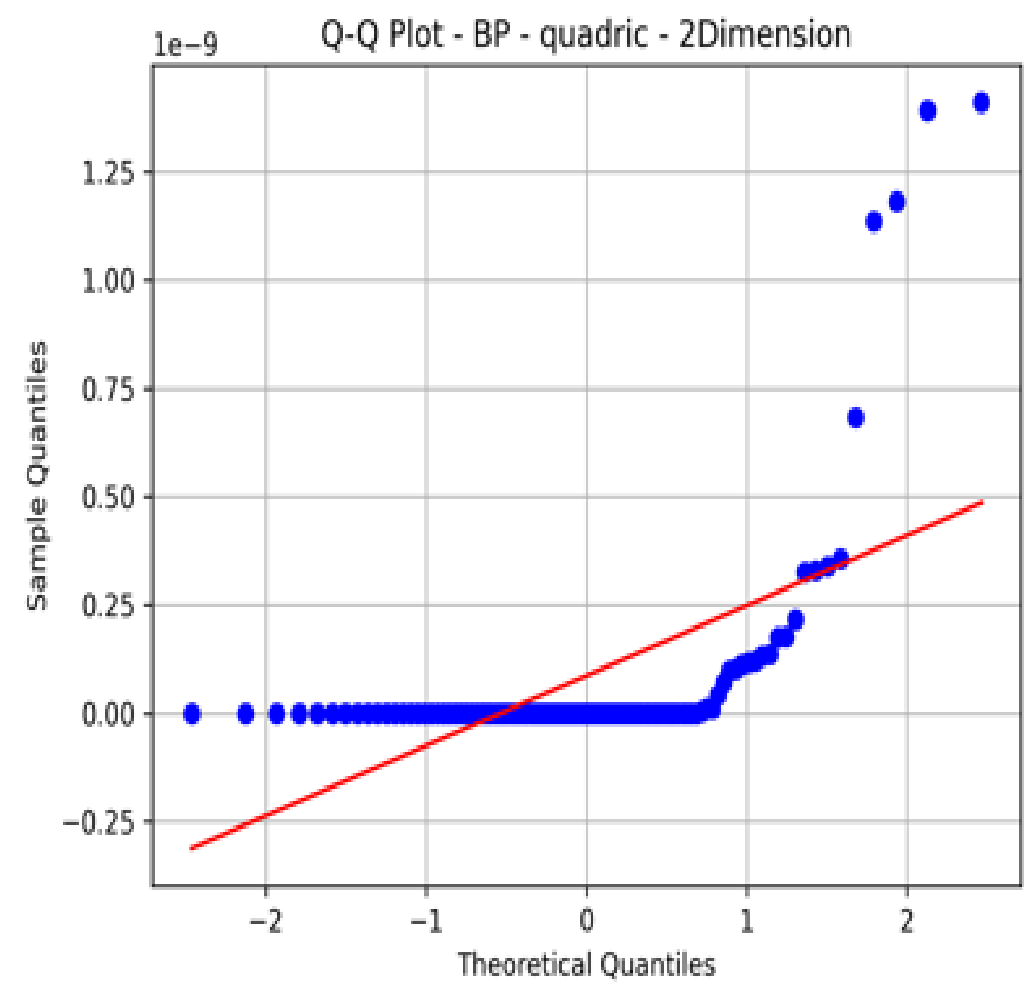
Q-Q Plot - BP - quadric - 2Dimension
1e−9
Sample Quantiles
Theoretical Quantiles

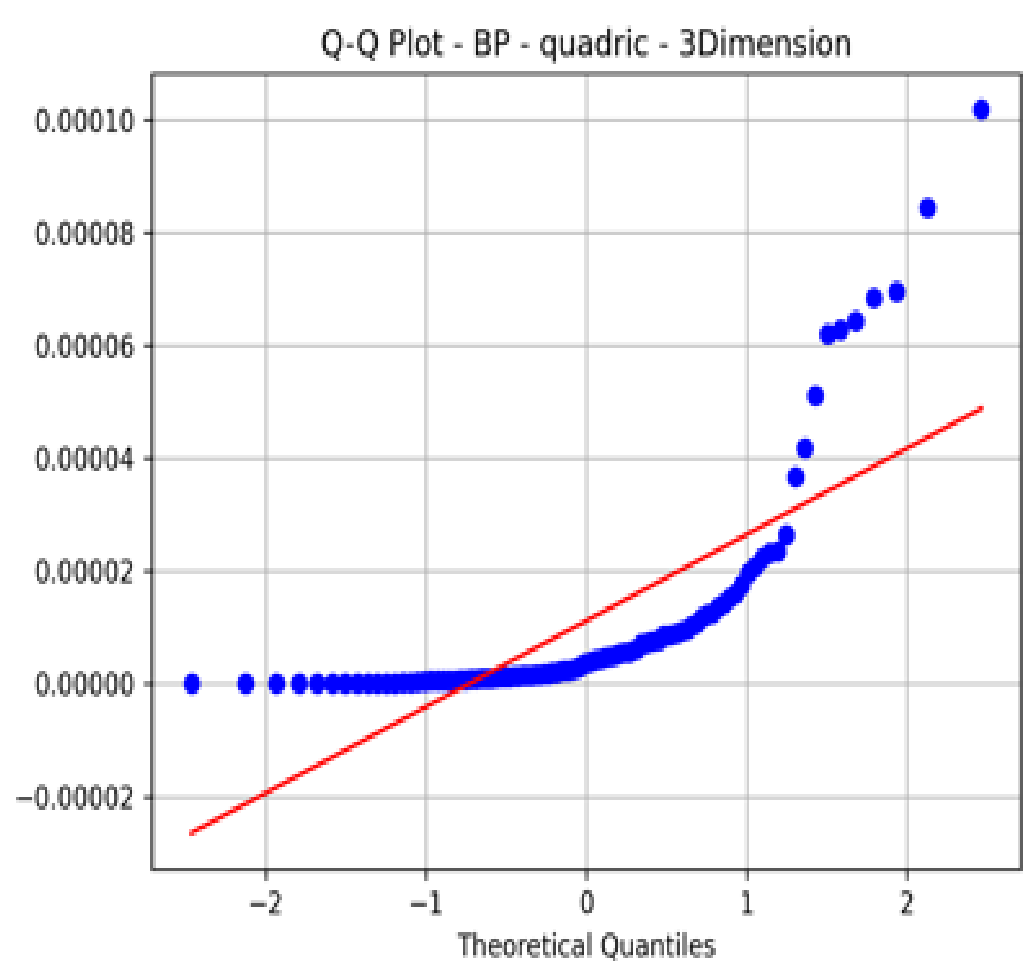
Q-Q Plot - BP - quadric - 3Dimension
Theoretical Quantiles

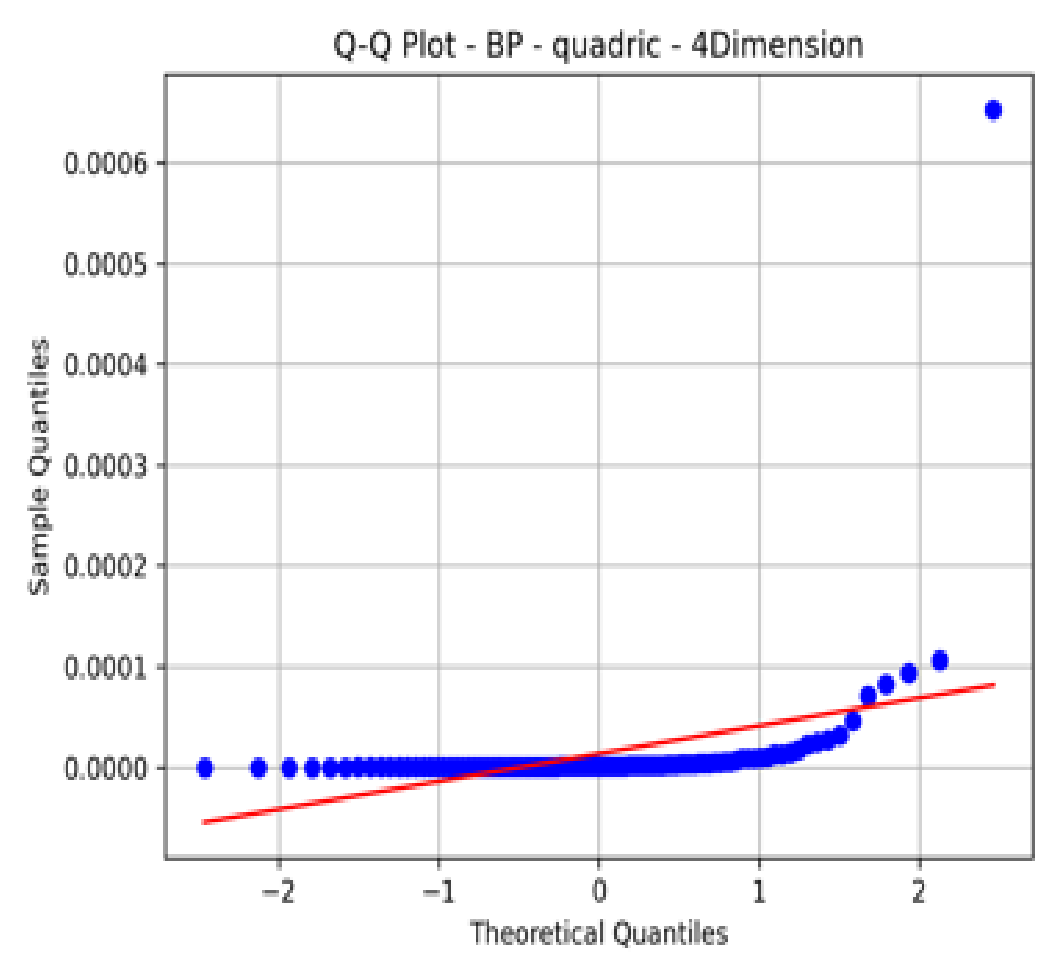
Q-Q Plot - BP - quadric - 4Dimension
Sample Quantiles
Theoretical Quantiles

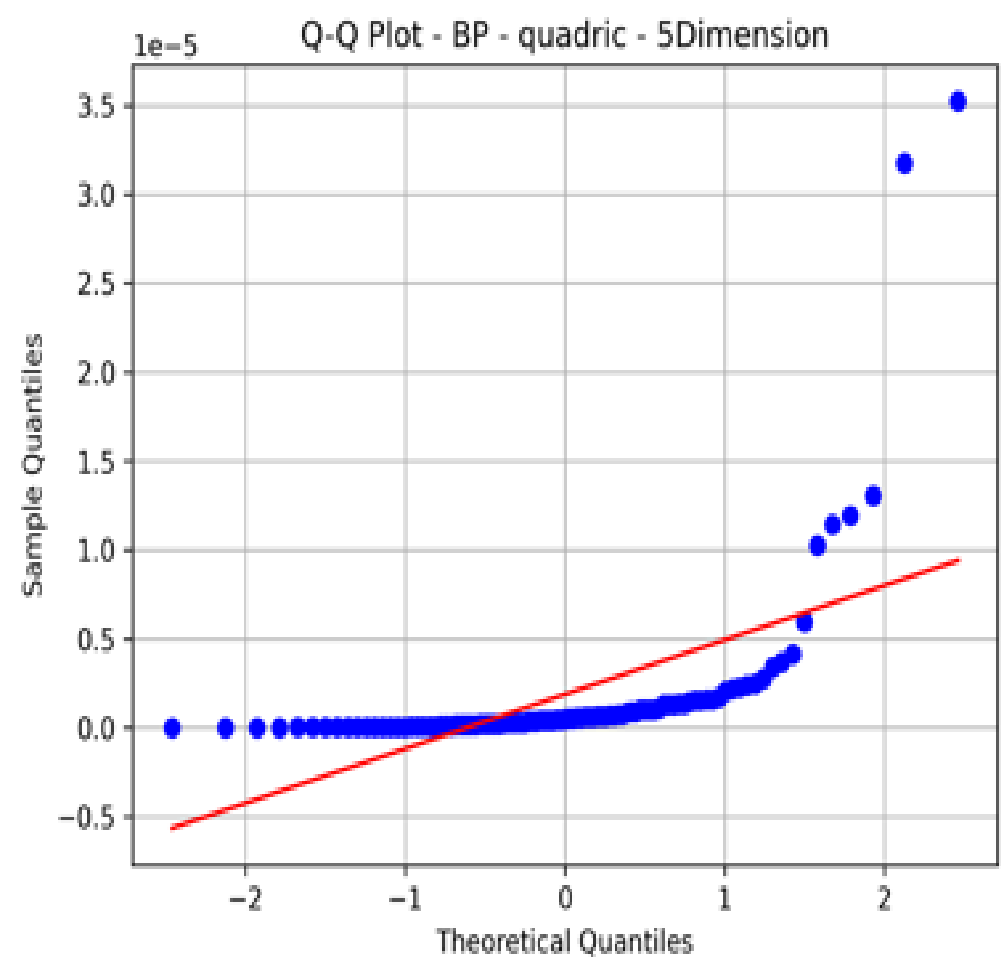
Q-Q Plot - BP - quadric - 5Dimension
1e−5
Sample Quantiles
Theoretical Quantiles

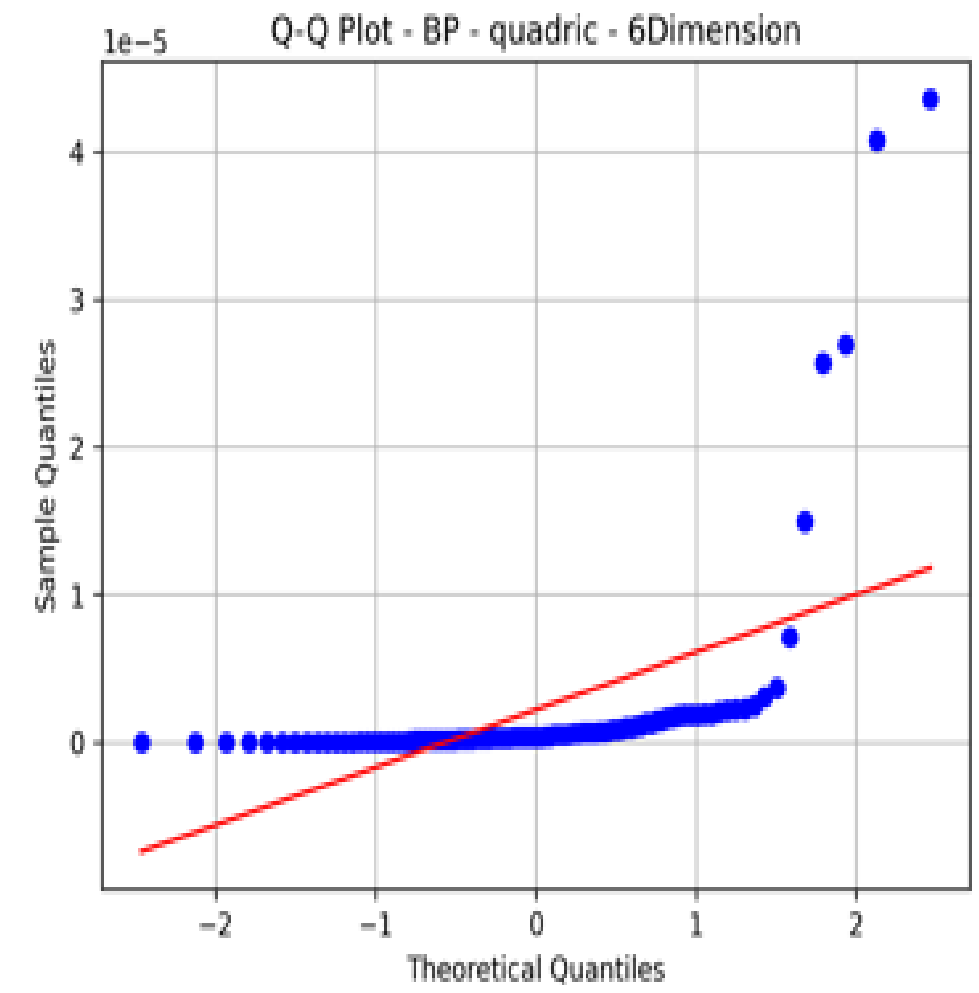
Q-Q Plot - BP - quadric - 6Dimension
1e−5
Sample Quantiles
Theoretical Quantiles

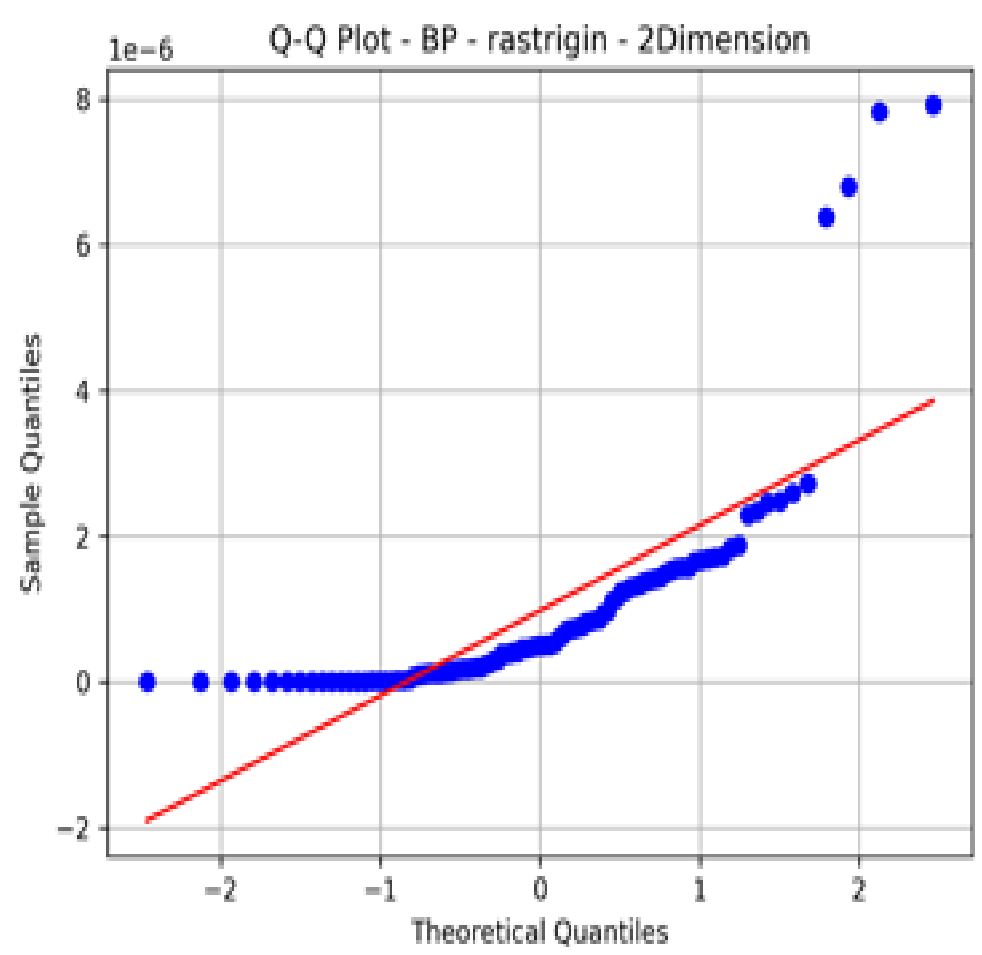
Q-Q Plot - BP - rastrigin - 2Dimension
1e−6
Sample Quantiles
Theoretical Quantiles

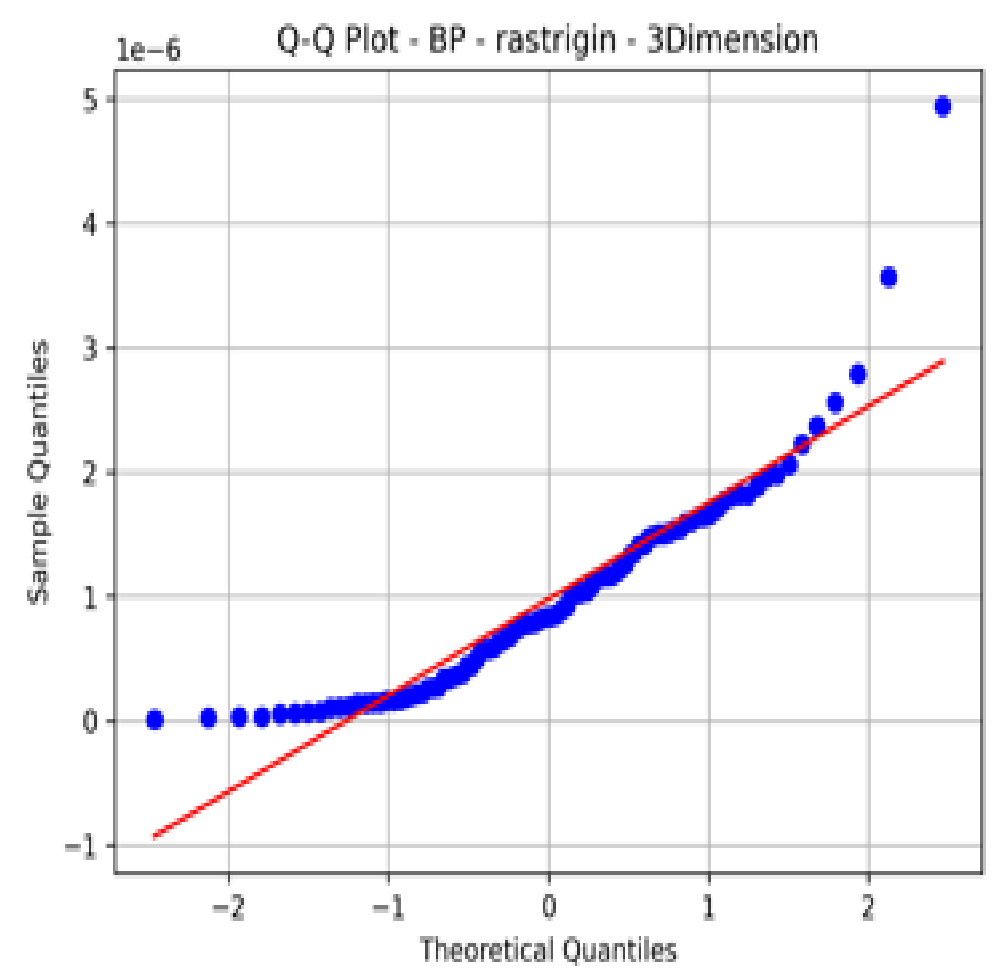
Q-Q Plot - BP - rastrigin - 3Dimension
1e−6
Sample Quantiles
Theoretical Quantiles

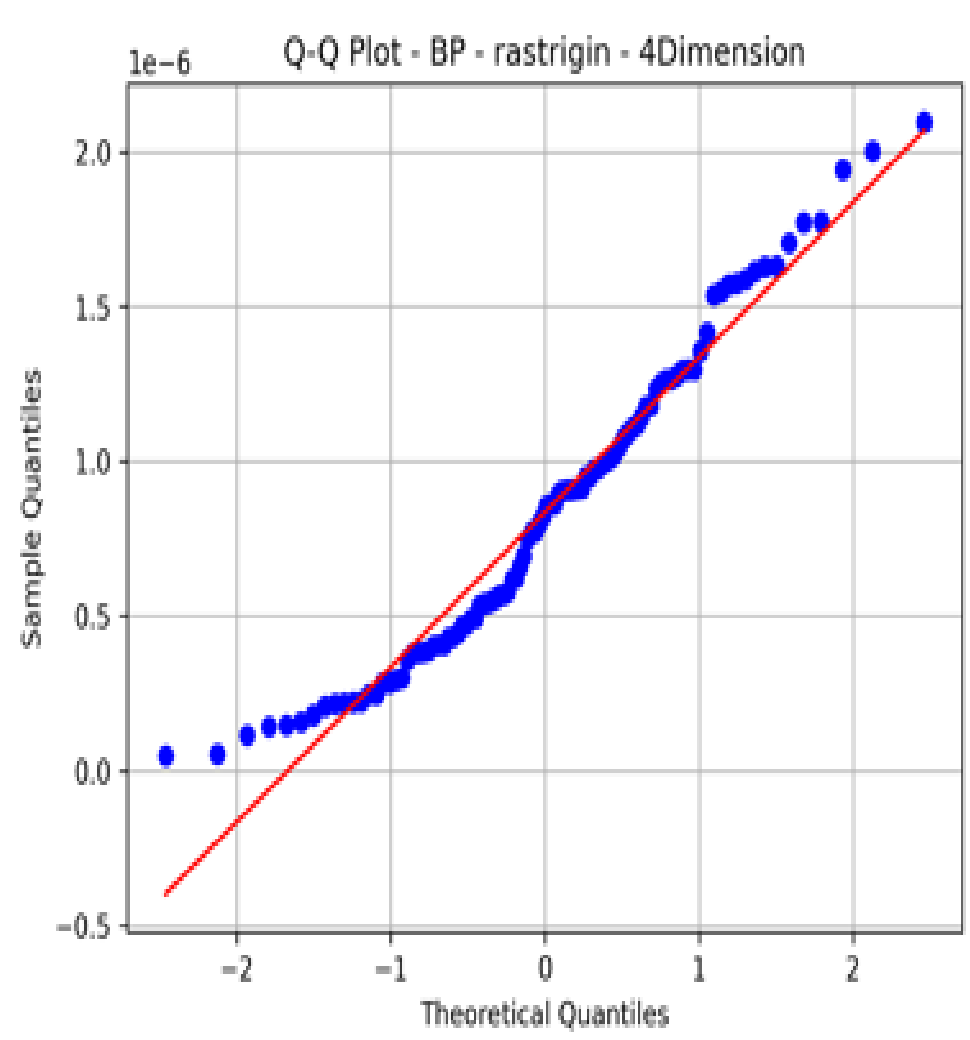
Q-Q Plot - BP - rastrigin - 4Dimension
1e−6
Sample Quantiles
Theoretical Quantiles

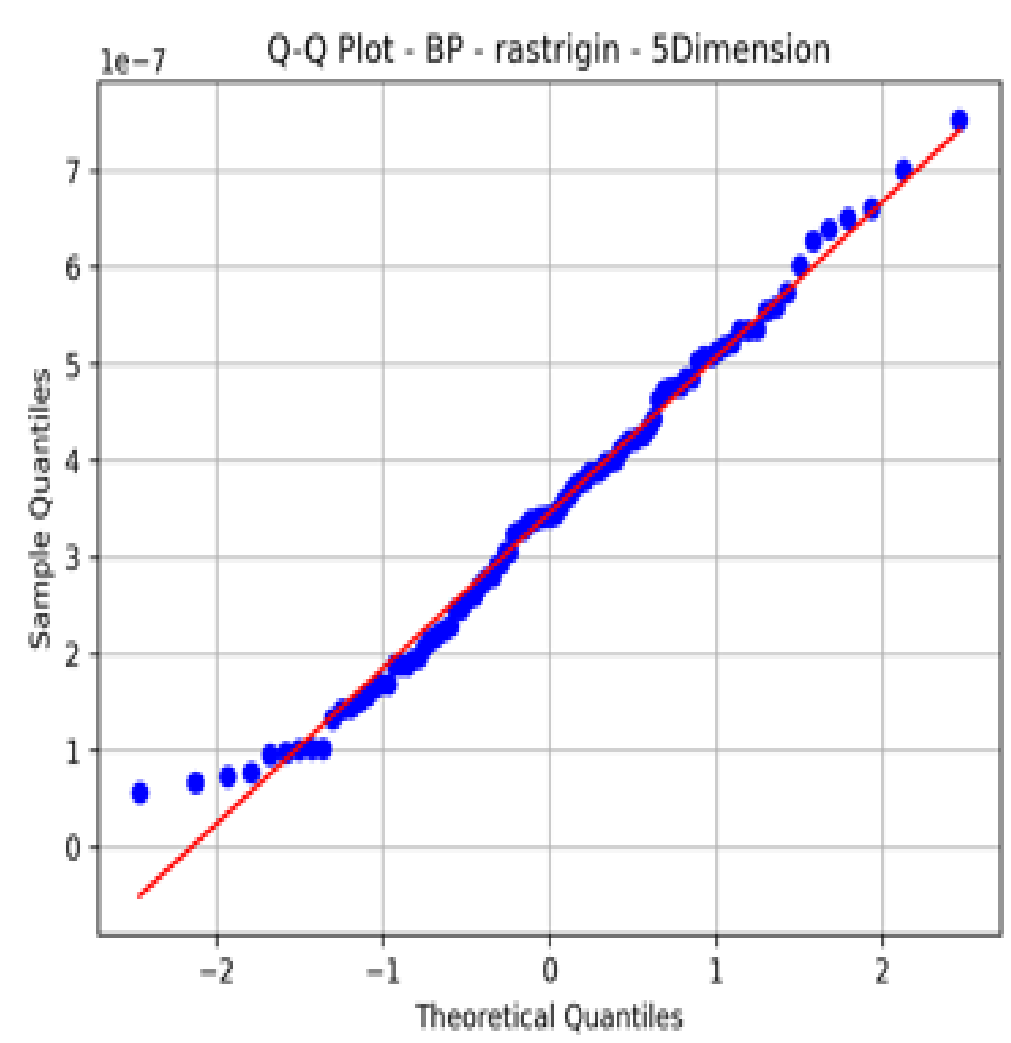
Q-Q Plot - BP - rastrigin - 5Dimension
1e−7
Sample Quantiles
Theoretical Quantiles

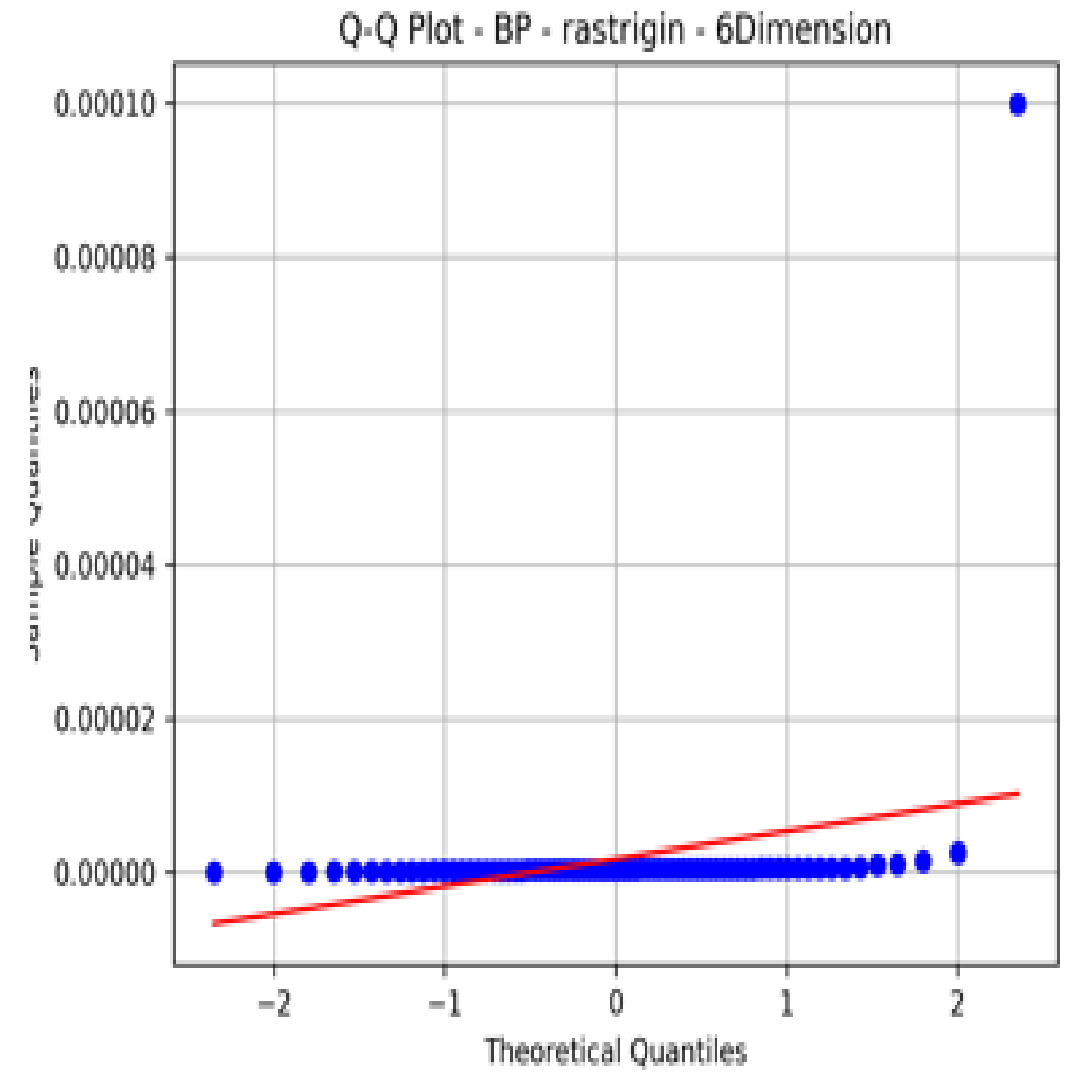
Q-Q Plot - BP - rastrigin - 6Dimension
0.00010
0.00008
0.00006
0.00004
0.00002
0.00000
−2
−1
0
1
2
Theoretical Quantiles

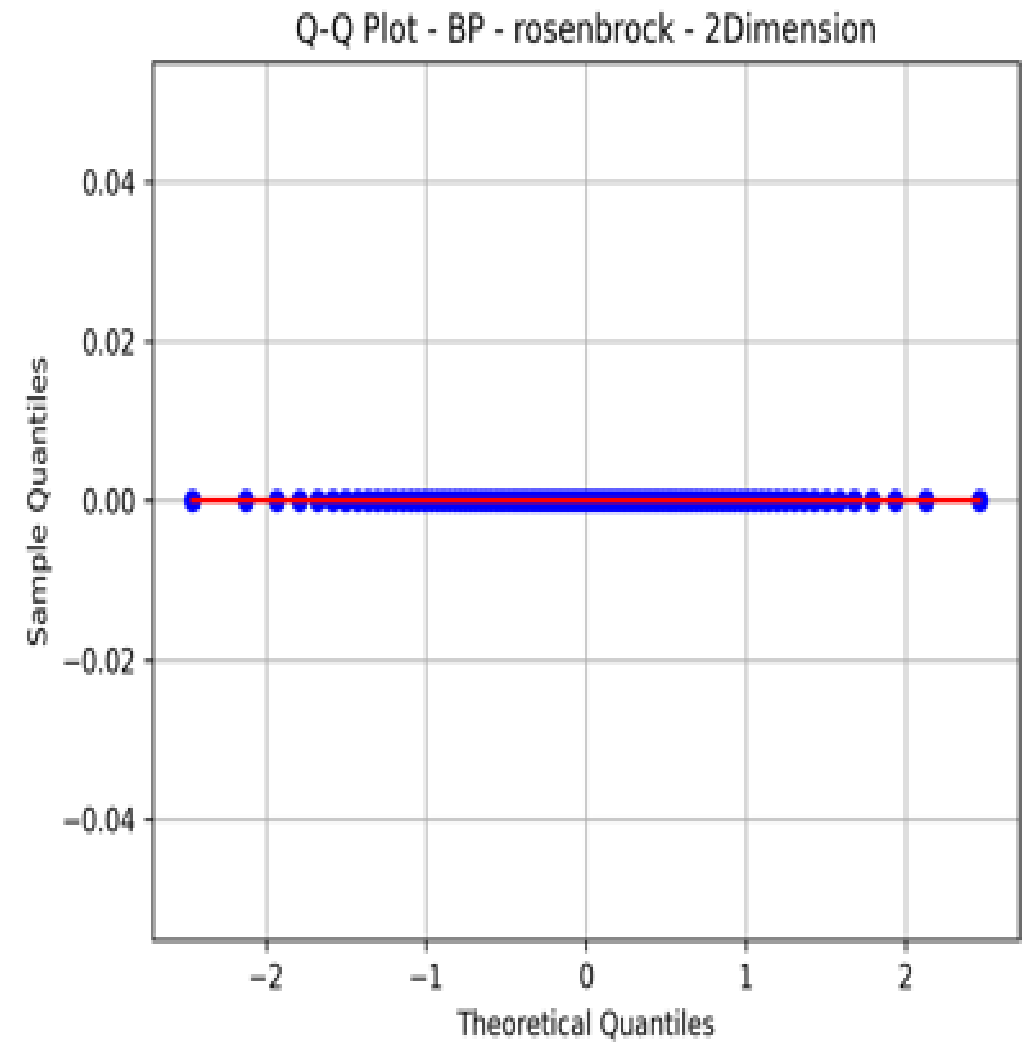
Q-Q Plot - BP - rosenbrock - 2Dimension
Sample Quantiles
0.04
0.02
0.00
−0.02
−0.04
−2
−1
0
1
2
Theoretical Quantiles

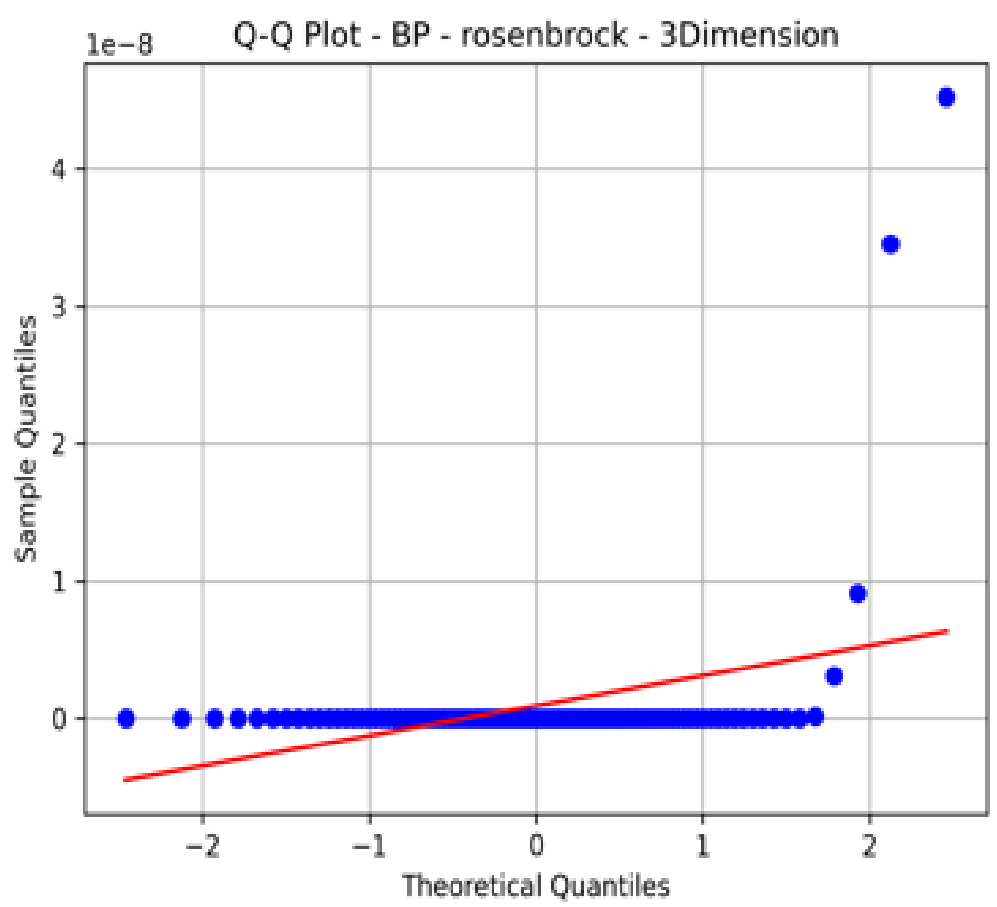
1e−8
Q-Q Plot - BP - rosenbrock - 3Dimension
Sample Quantiles
4
3
2
1
0
−2
−1
0
1
2
Theoretical Quantiles

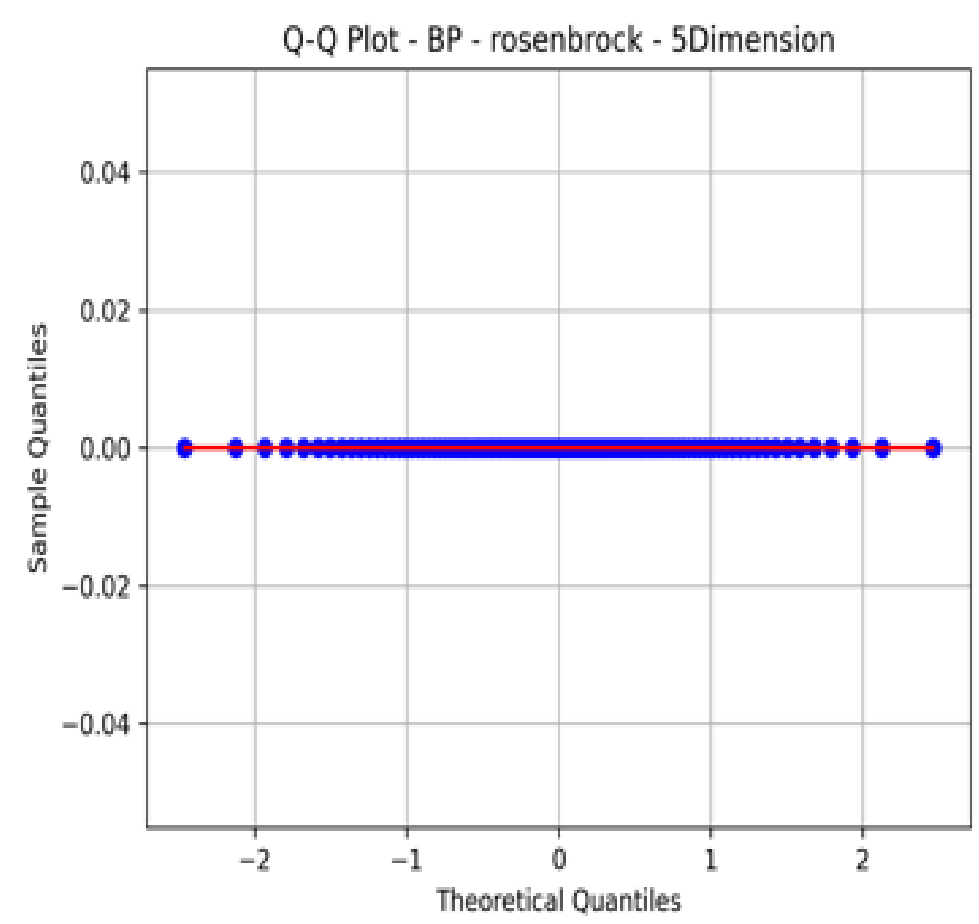
Q-Q Plot - BP - rosenbrock - 5Dimension
Sample Quantiles
0.04
0.02
0.00
−0.02
−0.04
−2
−1
0
1
2
Theoretical Quantiles

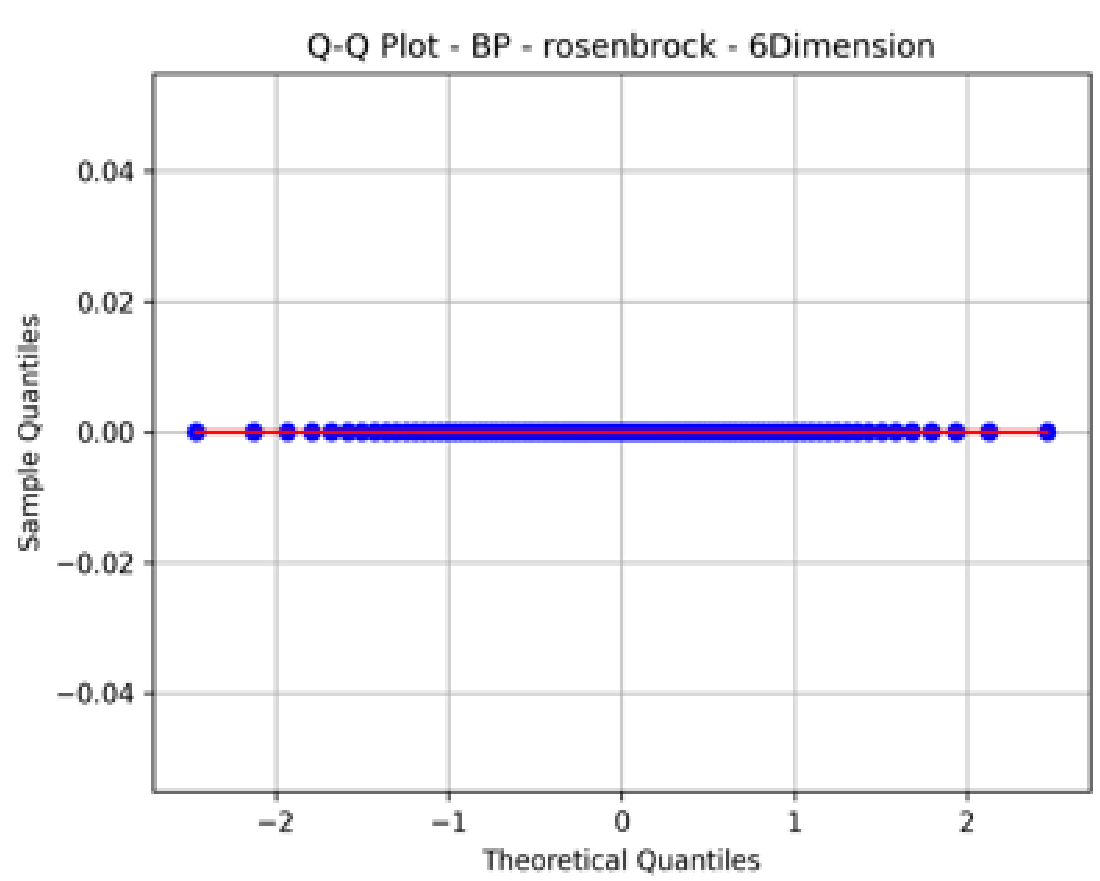
Q-Q Plot - BP - rosenbrock - 6Dimension
Sample Quantiles
0.04
0.02
0.00
−0.02
−0.04
−2
−1
0
1
2
Theoretical Quantiles

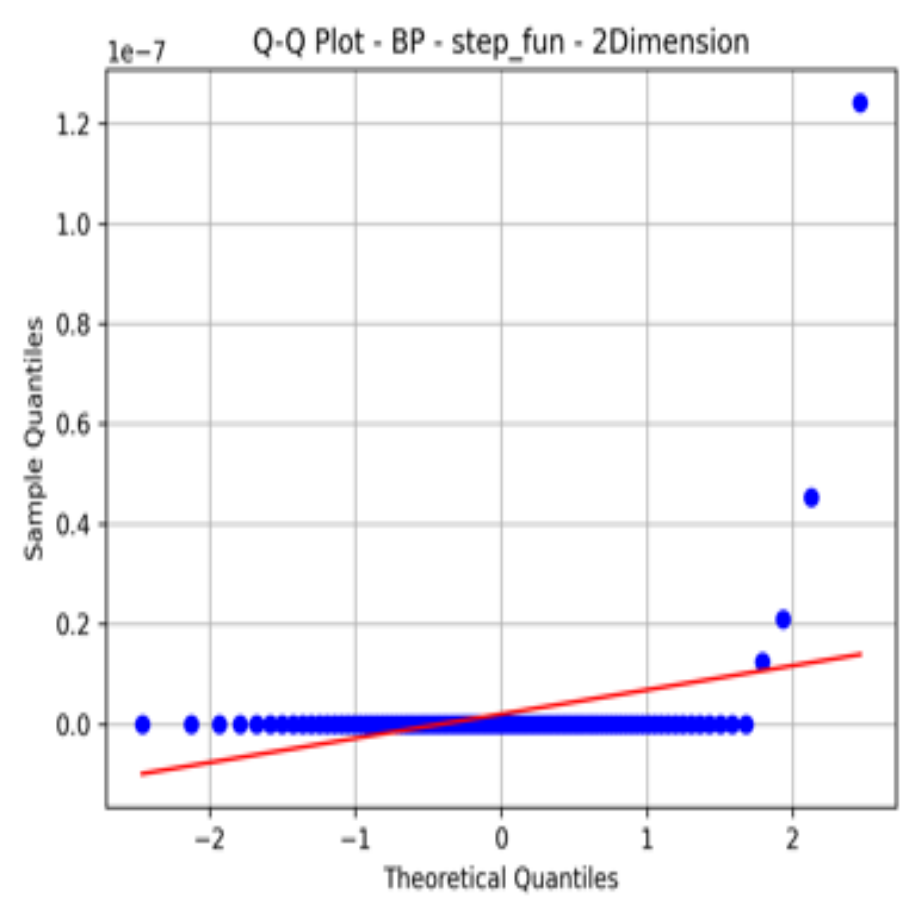
1e−7
Q-Q Plot - BP - step_fun - 2Dimension
Sample Quantiles
1.2
1.0
0.8
0.6
0.4
0.2
0.0
−2
−1
0
1
2
Theoretical Quantiles

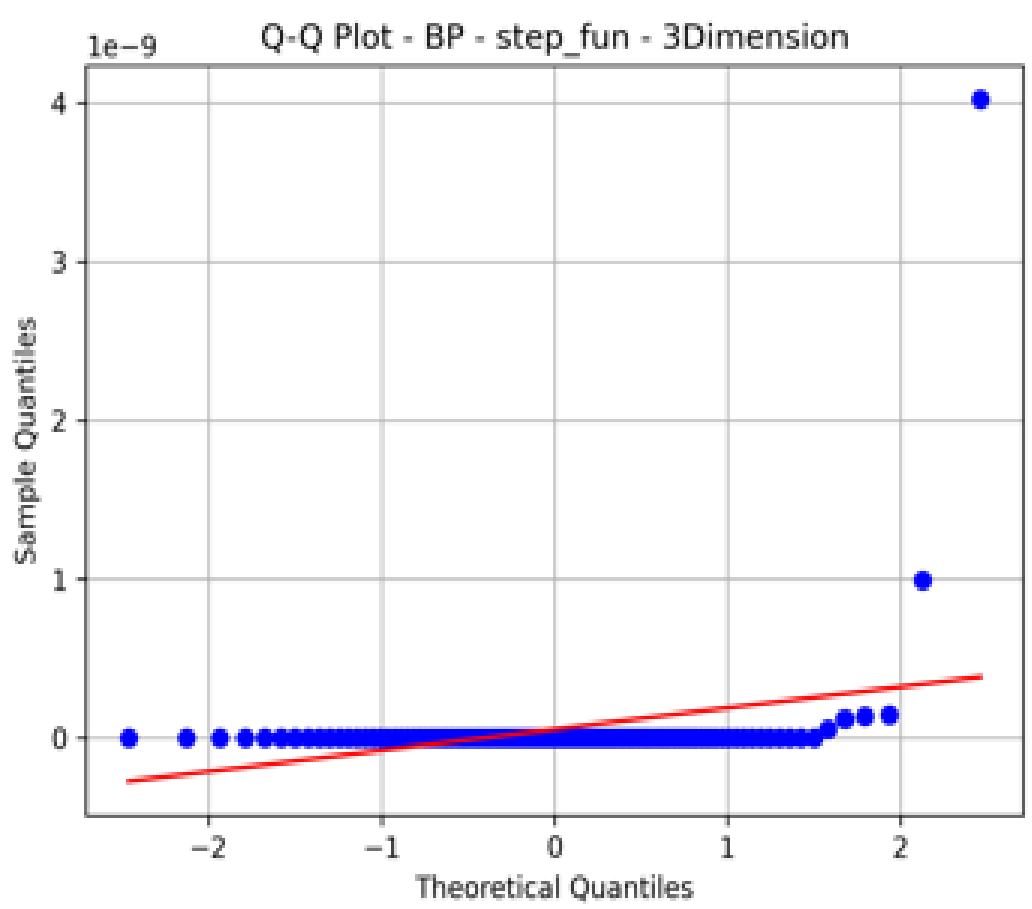
Q-Q Plot - BP - step_fun - 3Dimension
1e−9
Sample Quantiles
Theoretical Quantiles

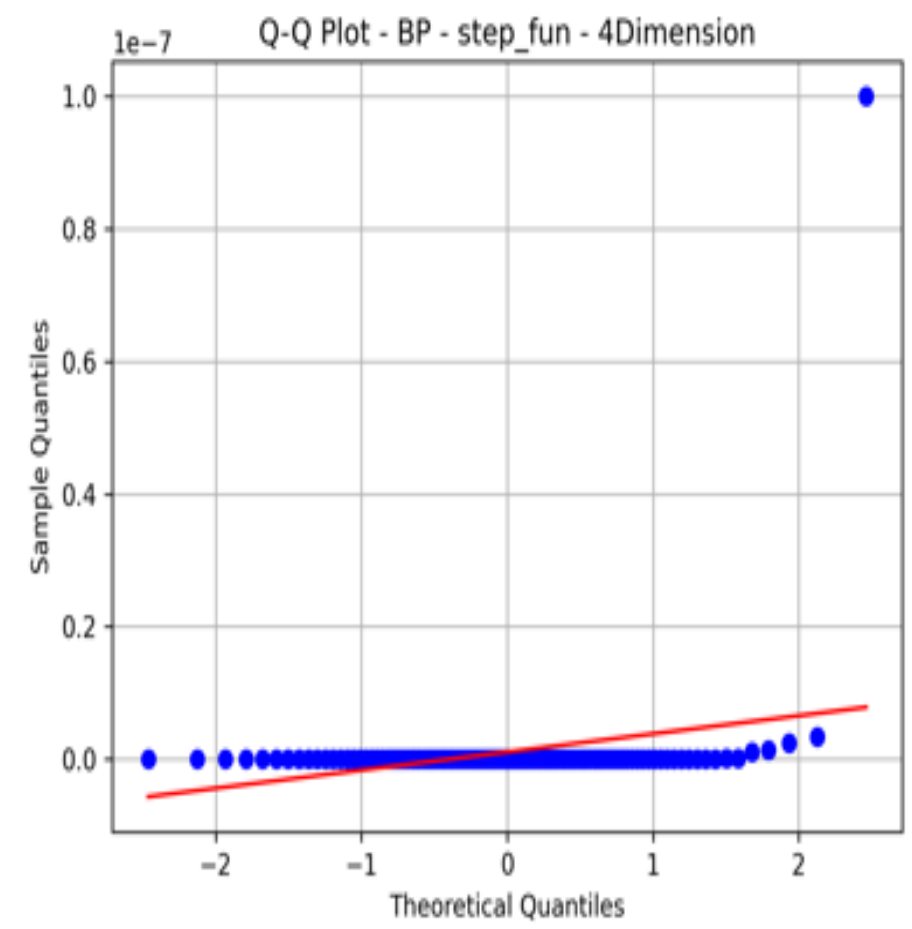
Q-Q Plot - BP - step_fun - 4Dimension
1e−7
Sample Quantiles
Theoretical Quantiles

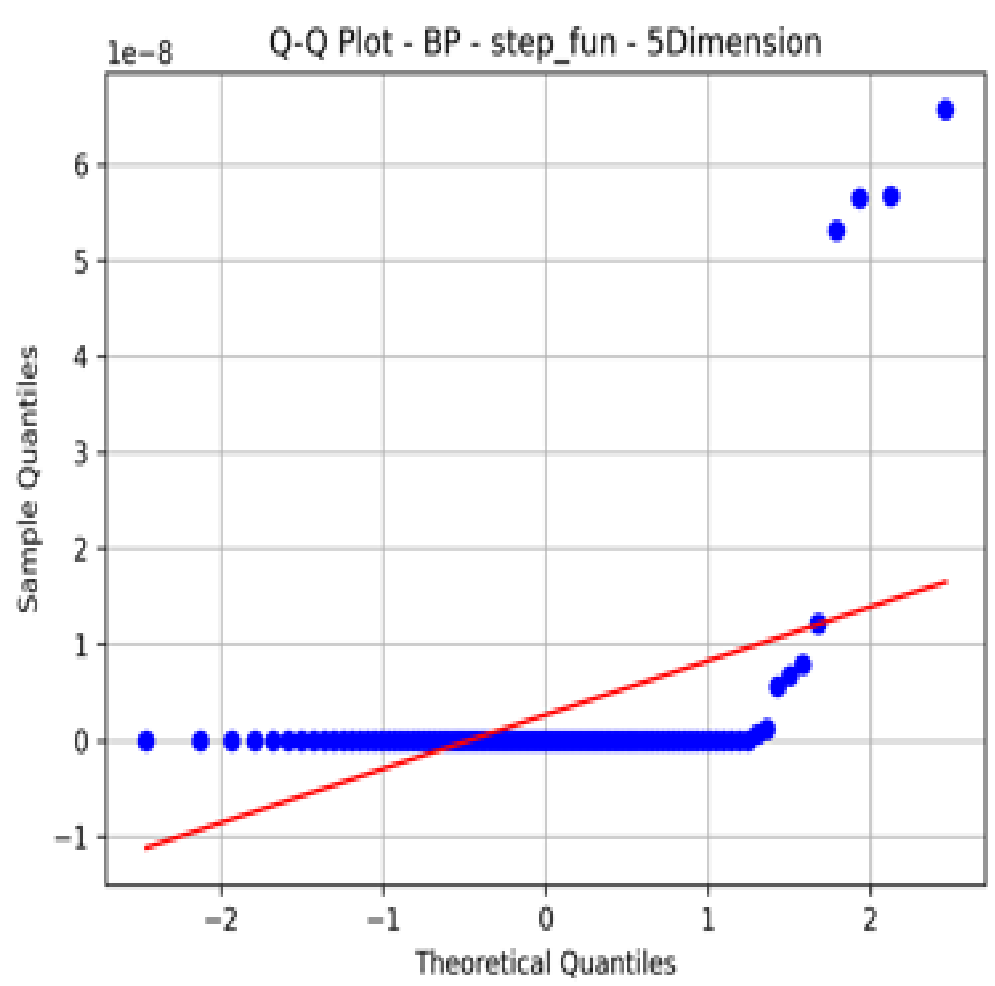
Q-Q Plot - BP - step_fun - 5Dimension
1e−8
Sample Quantiles
Theoretical Quantiles

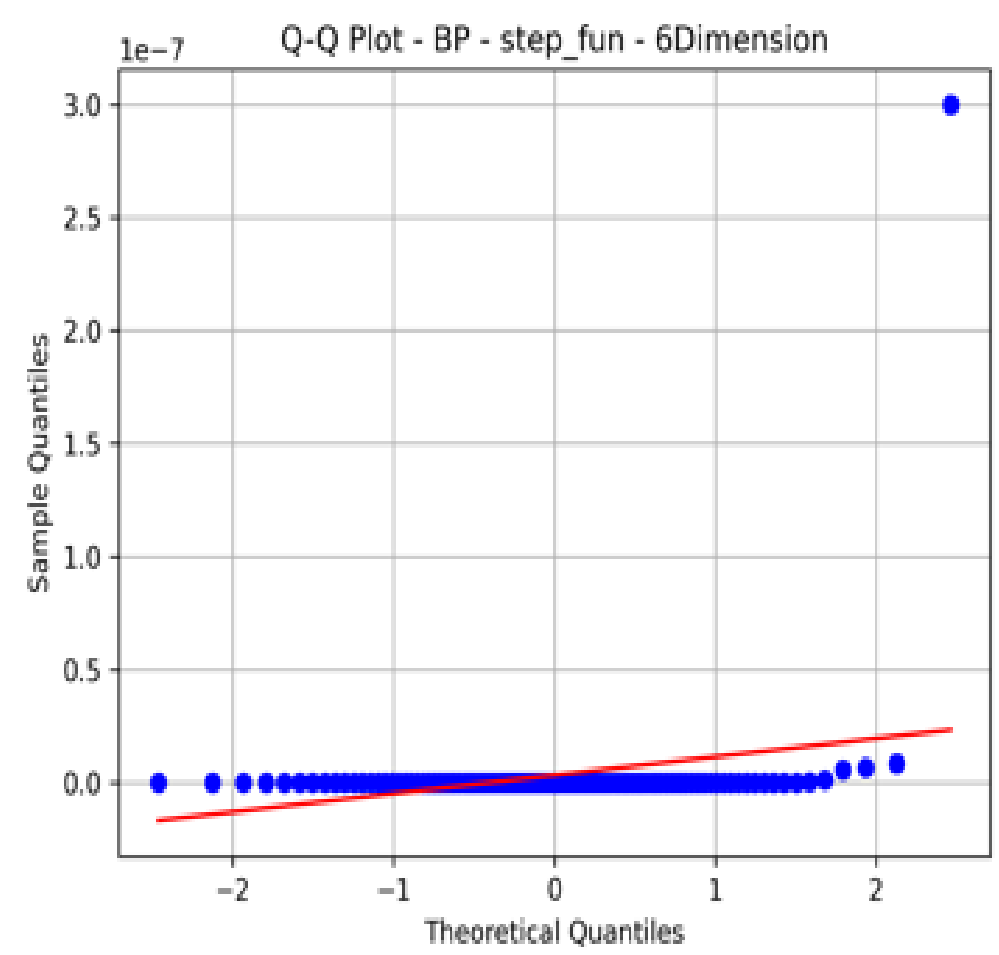
Q-Q Plot - BP - step_fun - 6Dimension
1e−7
Sample Quantiles
Theoretical Quantiles

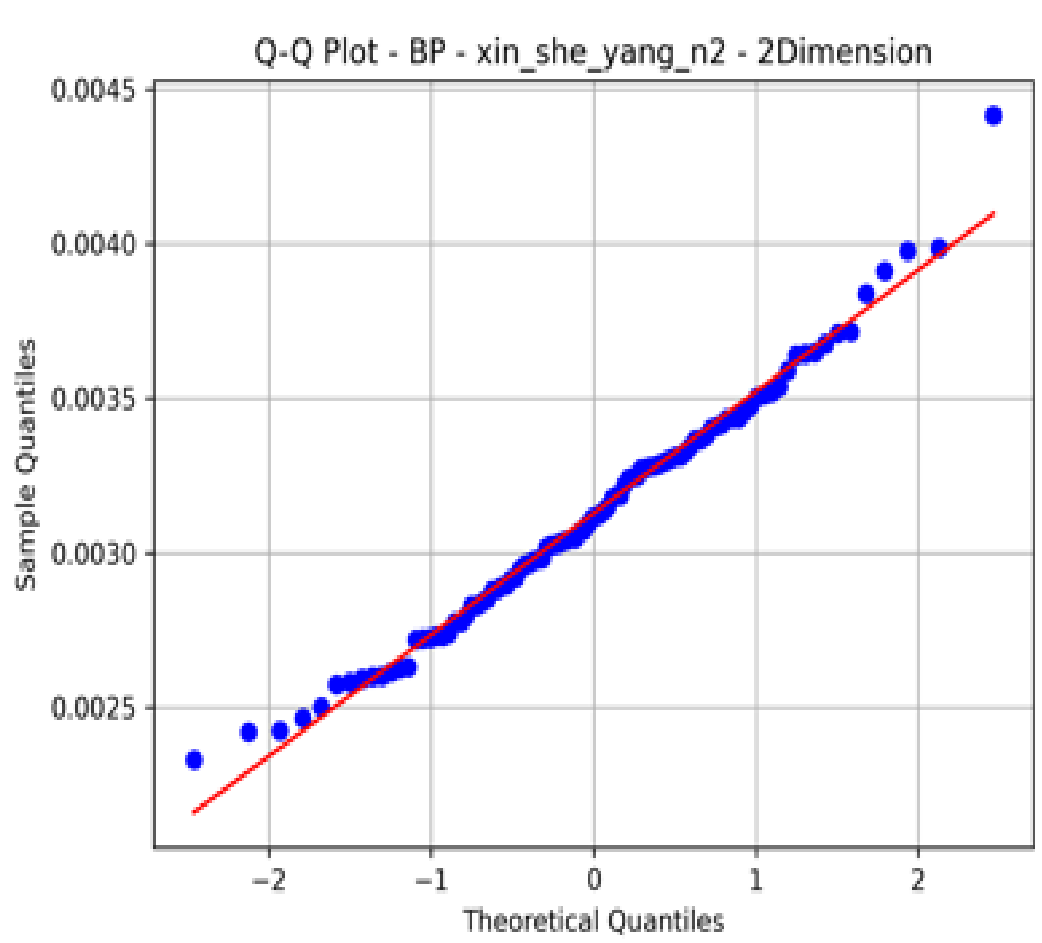
Q-Q Plot - BP - xin_she_yang_n2 - 2Dimension
Sample Quantiles
Theoretical Quantiles

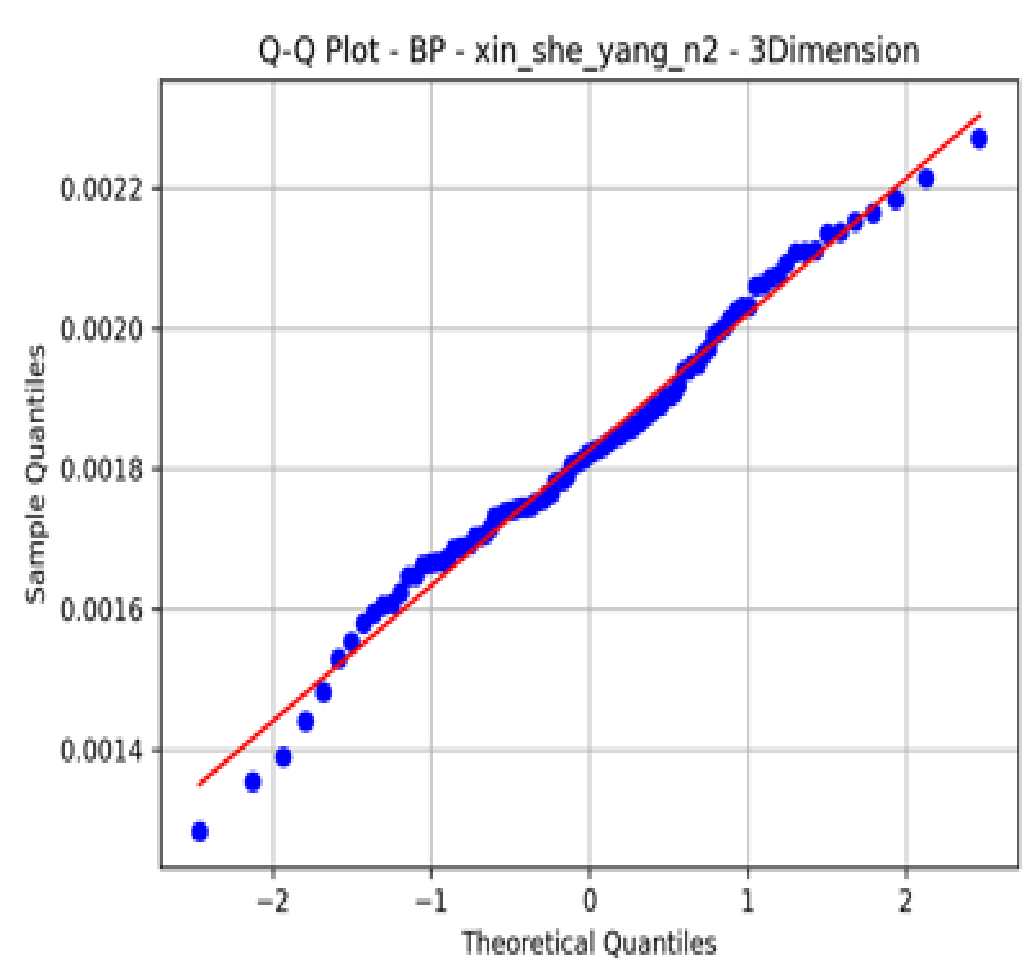
Q-Q Plot - BP - xin_she_yang_n2 - 3Dimension
Sample Quantiles
Theoretical Quantiles

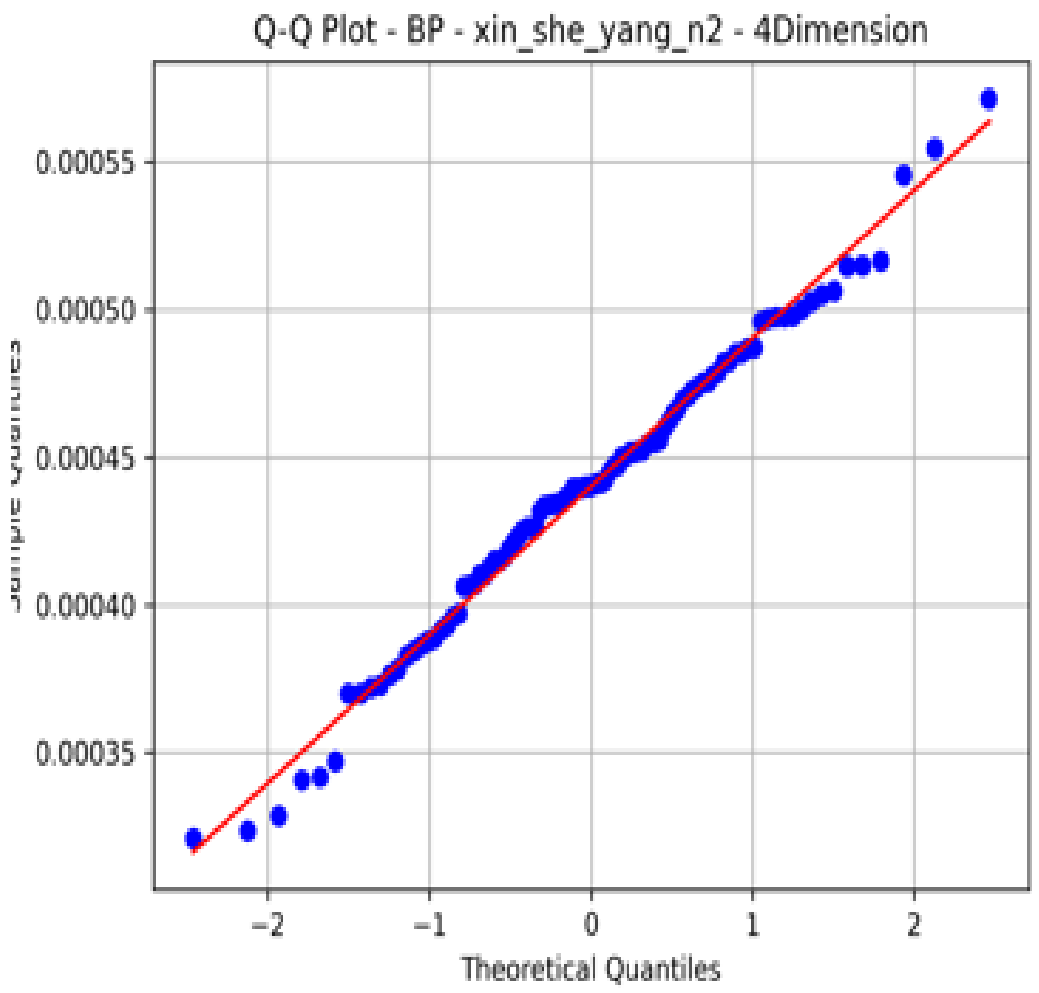

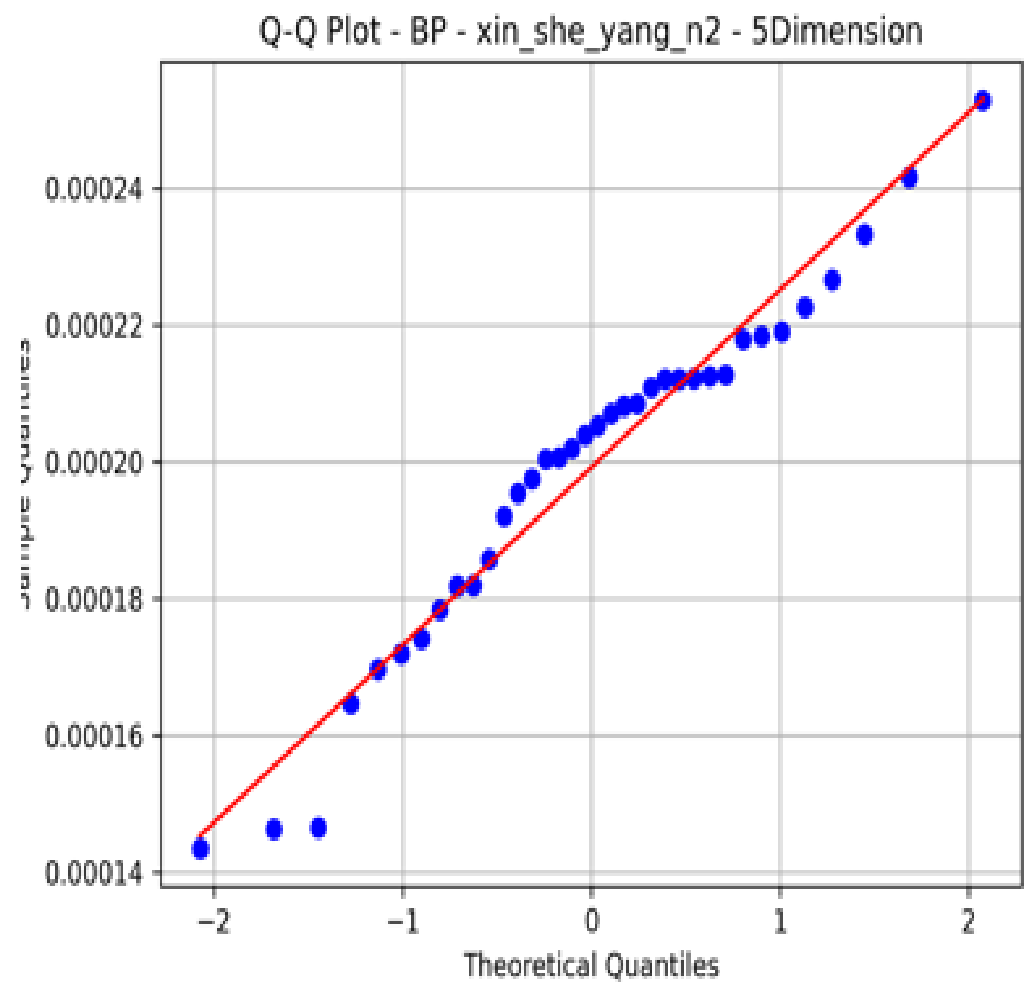

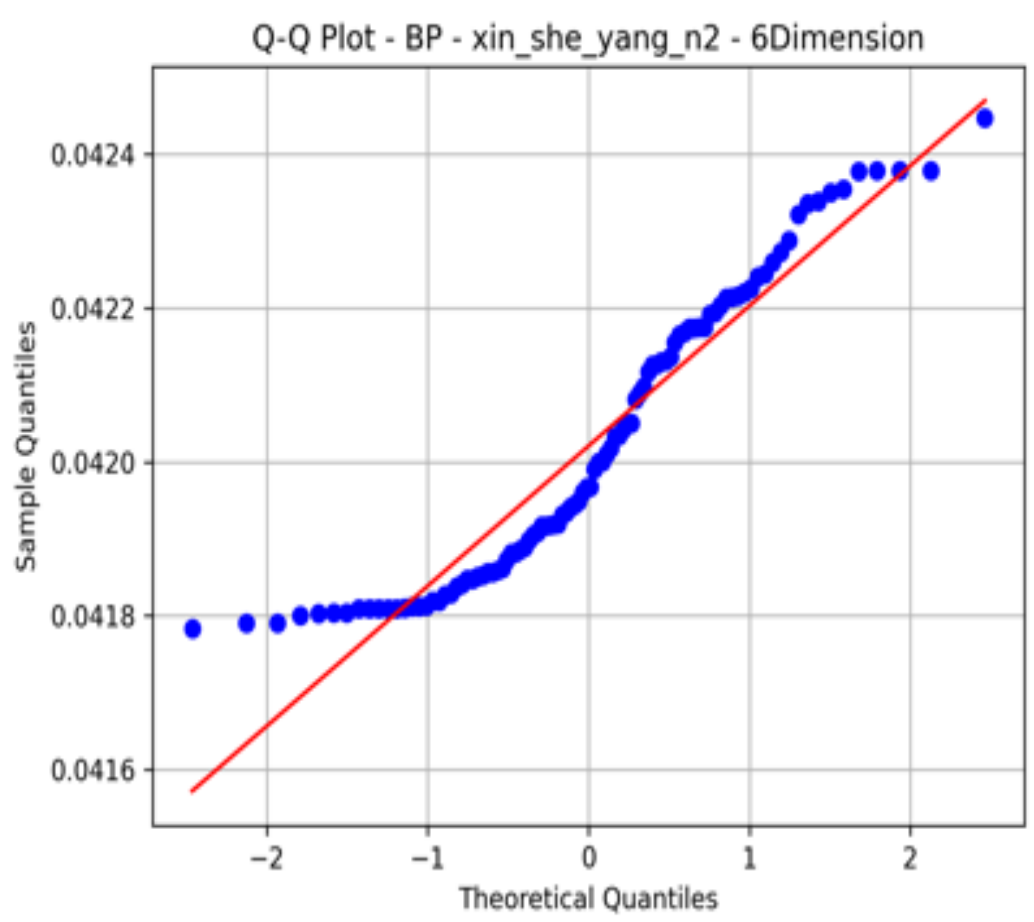

<u>SA Q-Q Plots</u>

Non-Continuous functions QQ plots

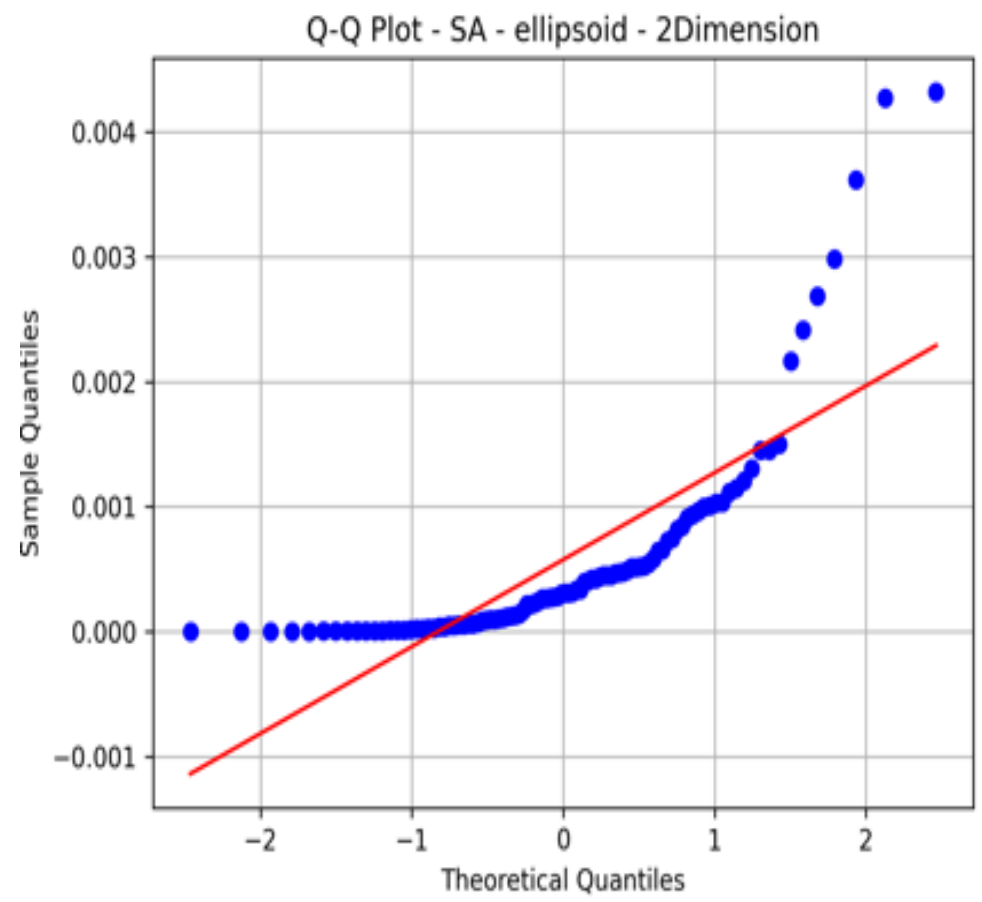
Q-Q Plot - SA - ellipsoid - 2Dimension
Sample Quantiles
Theoretical Quantiles
0.004
0.003
0.002
0.001
0.000
−0.001
−2
−1
0
1
2

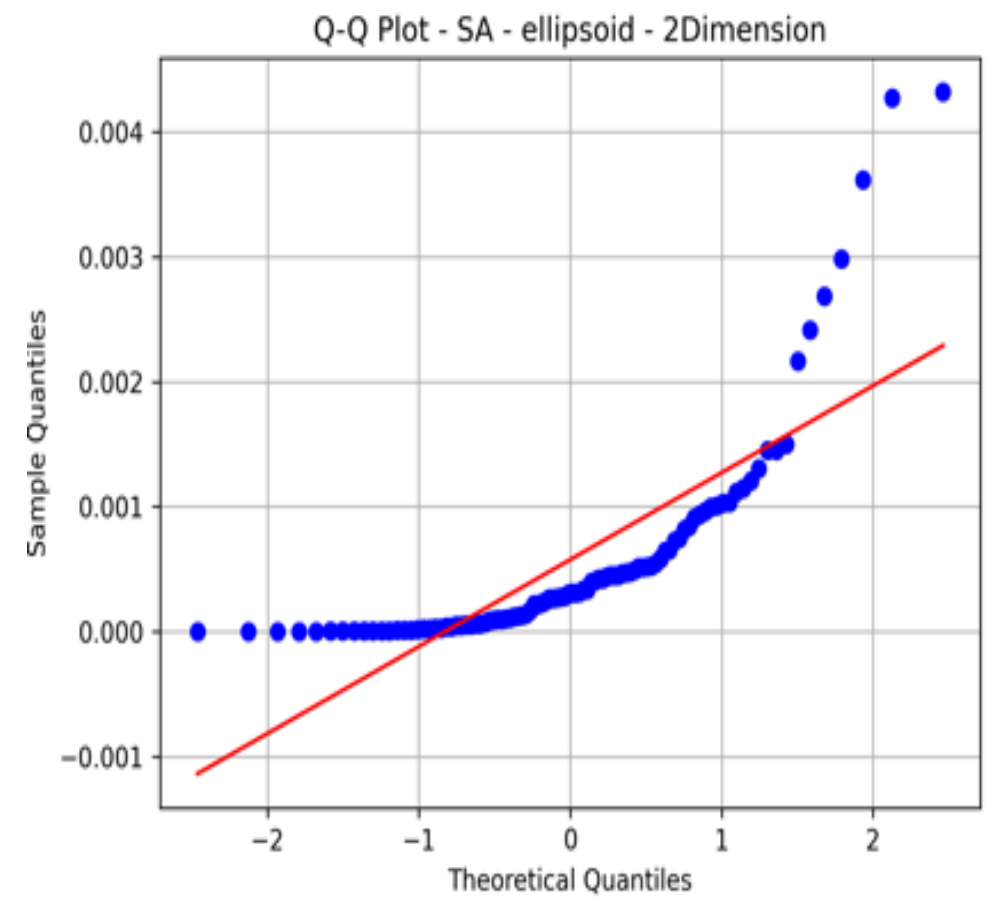
Q-Q Plot - SA - ellipsoid - 2Dimension
Sample Quantiles
Theoretical Quantiles
0.004
0.003
0.002
0.001
0.000
−0.001
−2
−1
0
1
2

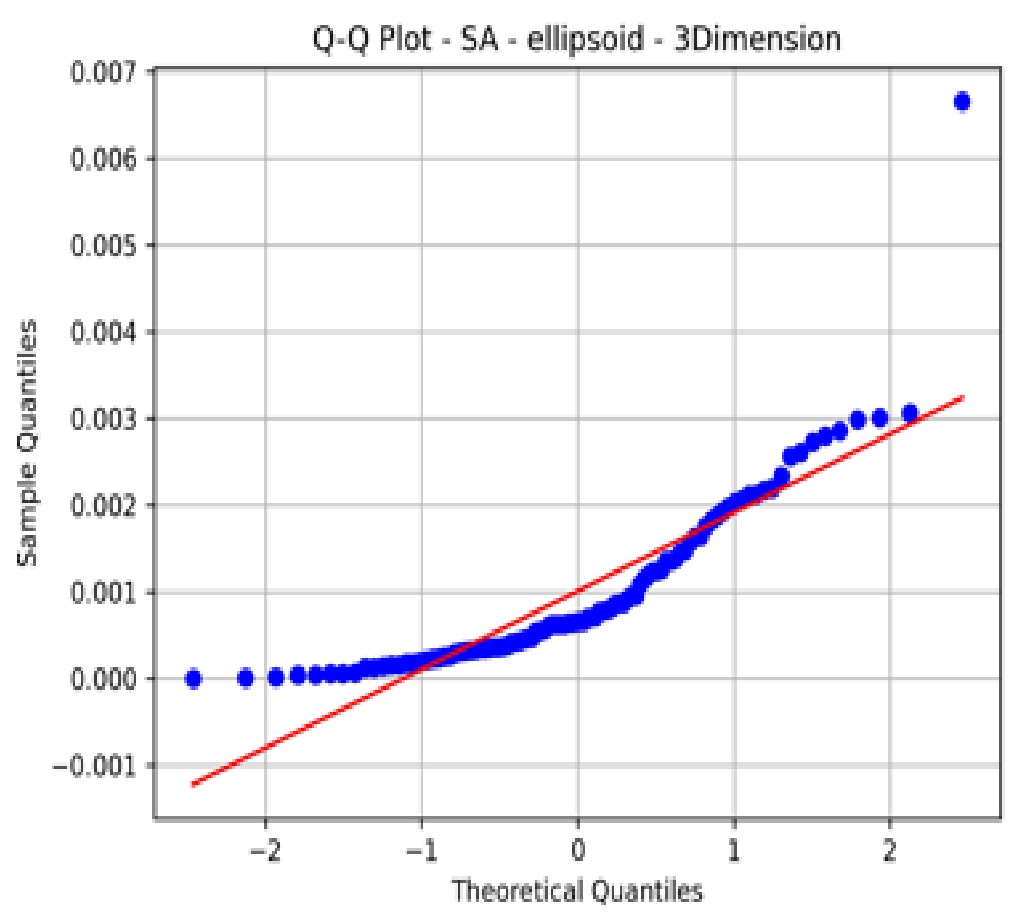
Q-Q Plot - SA - ellipsoid - 3Dimension
Sample Quantiles
Theoretical Quantiles
0.007
0.006
0.005
0.004
0.003
0.002
0.001
0.000
−0.001
−2
−1
0
1
2

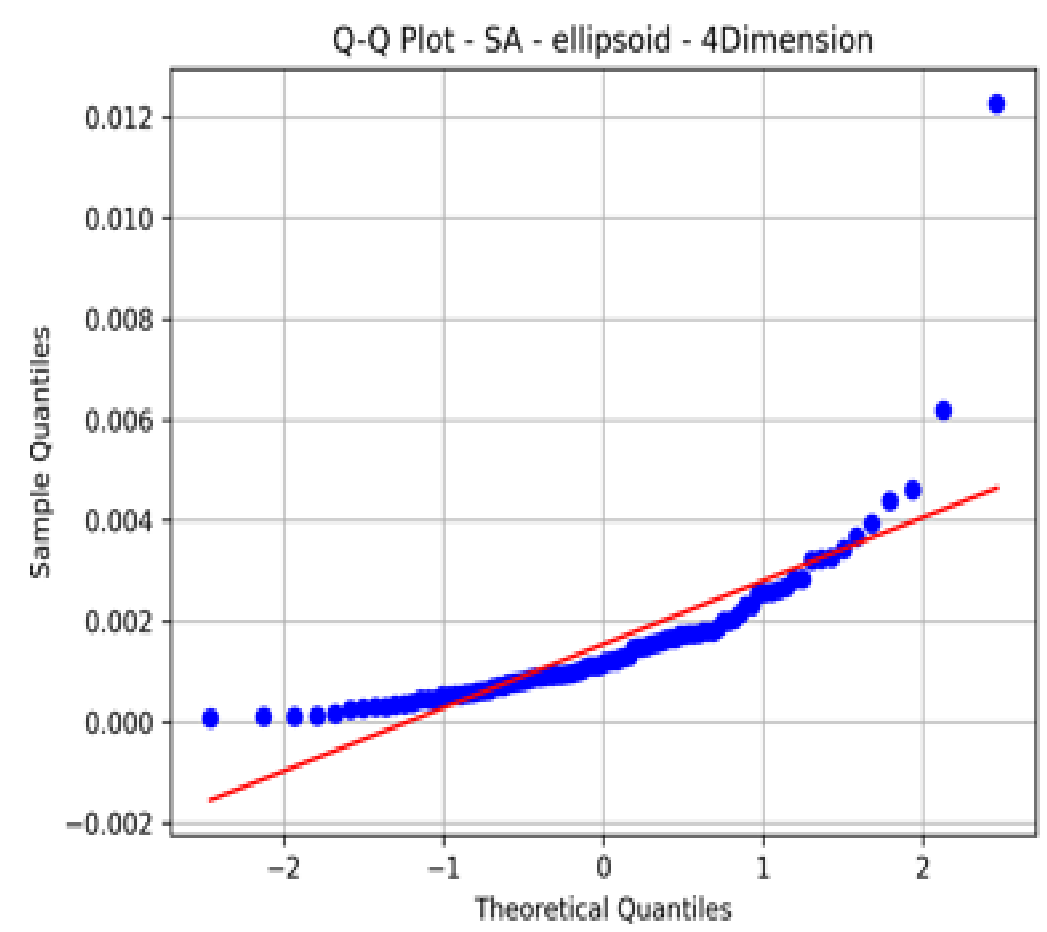
Q-Q Plot - SA - ellipsoid - 4Dimension
Sample Quantiles
Theoretical Quantiles
0.012
0.010
0.008
0.006
0.004
0.002
0.000
−0.002
−2
−1
0
1
2

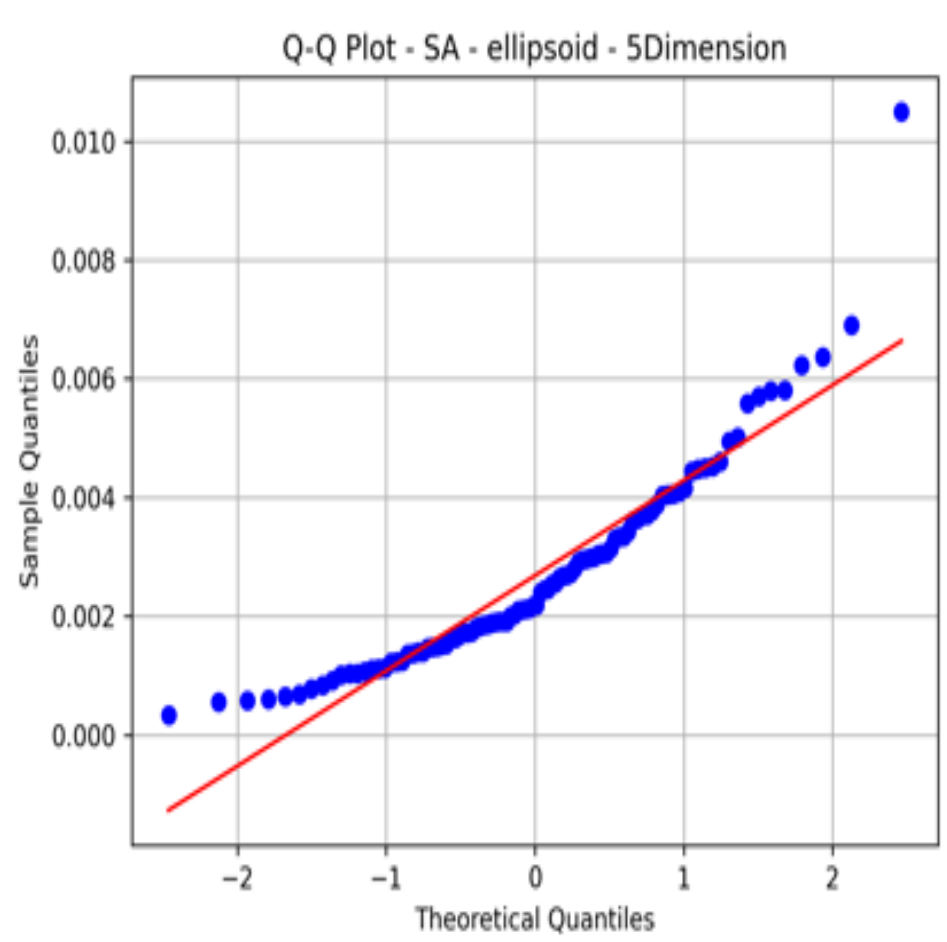
Q-Q Plot - SA - ellipsoid - 5Dimension
Sample Quantiles
Theoretical Quantiles
0.010
0.008
0.006
0.004
0.002
0.000
−2
−1
0
1
2

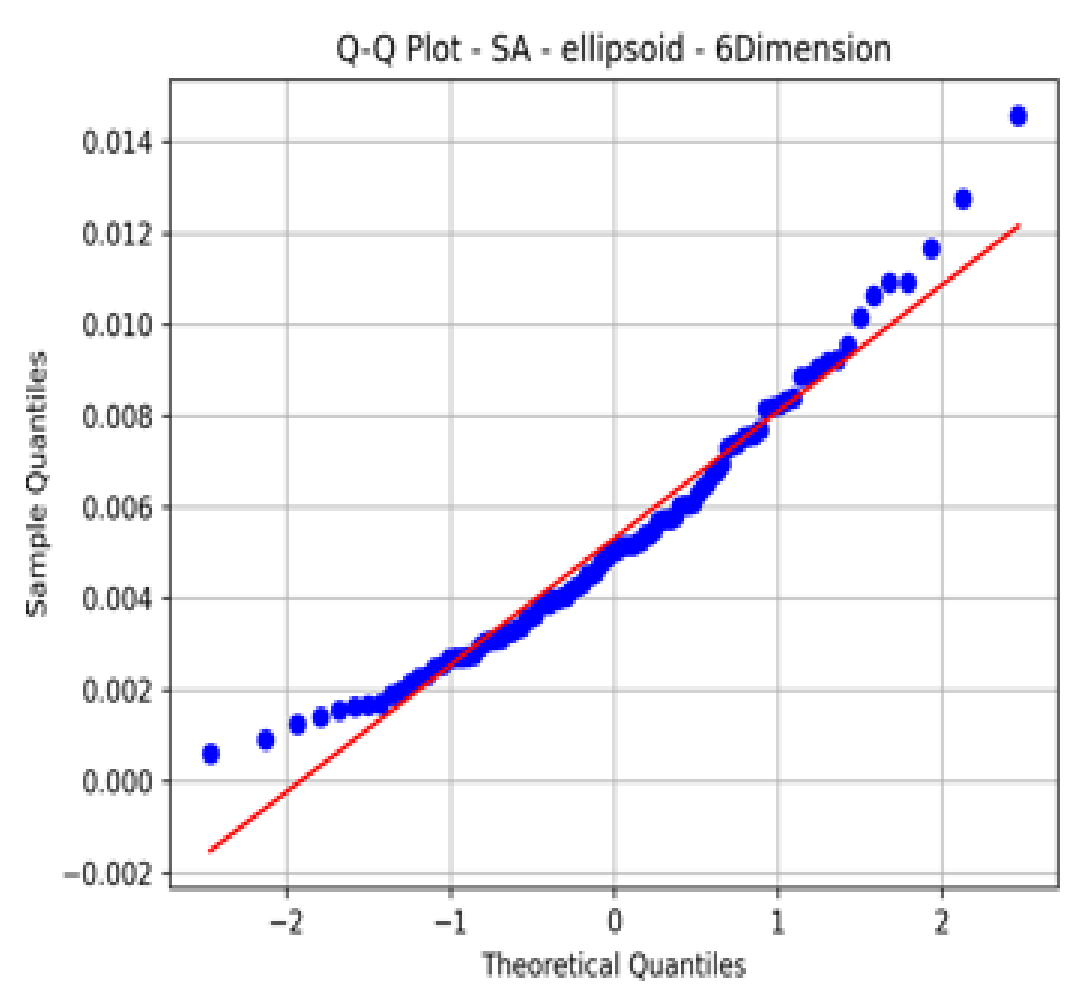
Q-Q Plot - SA - ellipsoid - 6Dimension
Sample Quantiles
Theoretical Quantiles
0.014
0.012
0.010
0.008
0.006
0.004
0.002
0.000
−0.002
−2
−1
0
1
2

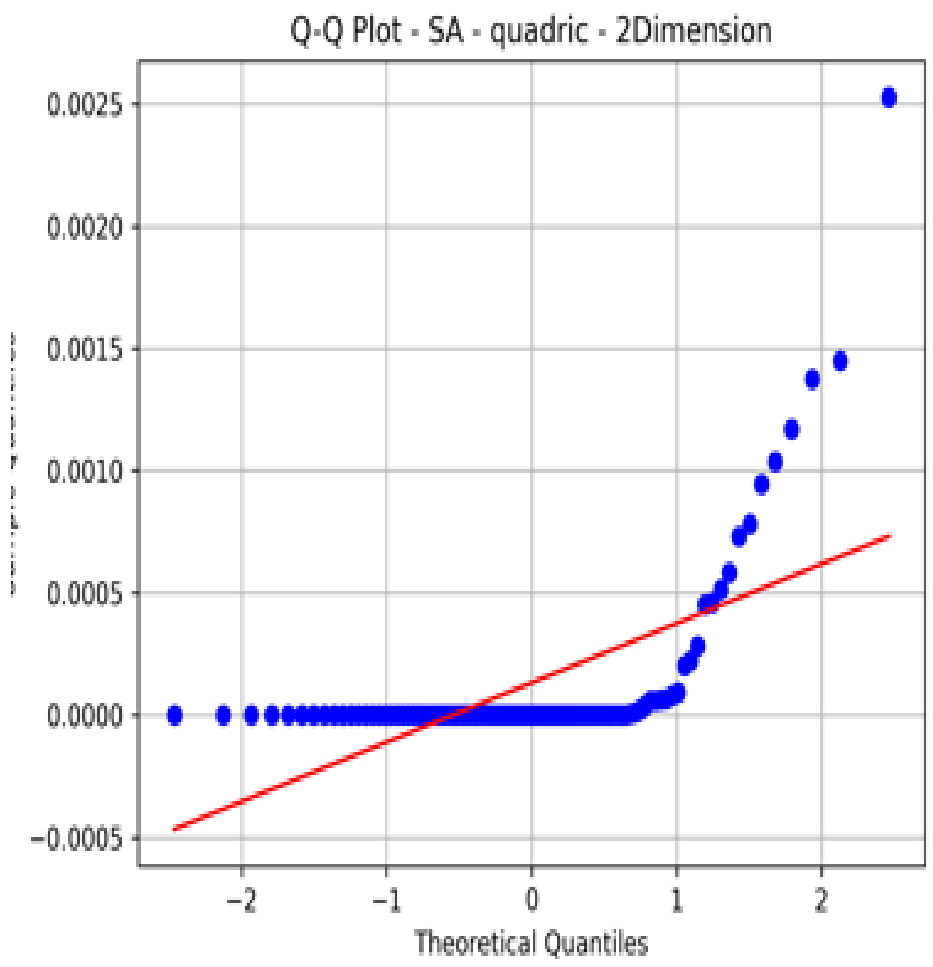
Q-Q Plot - SA - quadric - 2Dimension
0.0025
0.0020
0.0015
0.0010
0.0005
0.0000
−0.0005
−2
−1
0
1
2
Theoretical Quantiles

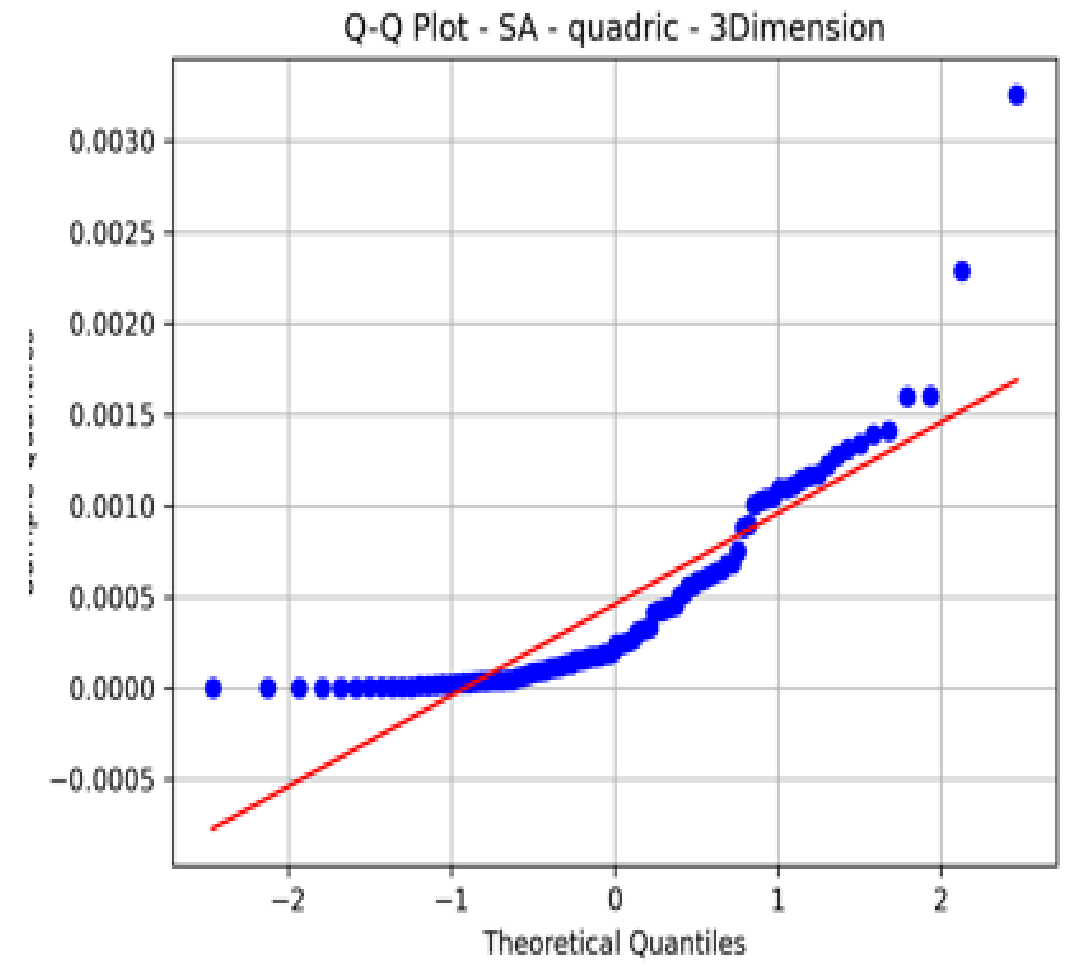
Q-Q Plot - SA - quadric - 3Dimension
0.0030
0.0025
0.0020
0.0015
0.0010
0.0005
0.0000
−0.0005
−2
−1
0
1
2
Theoretical Quantiles

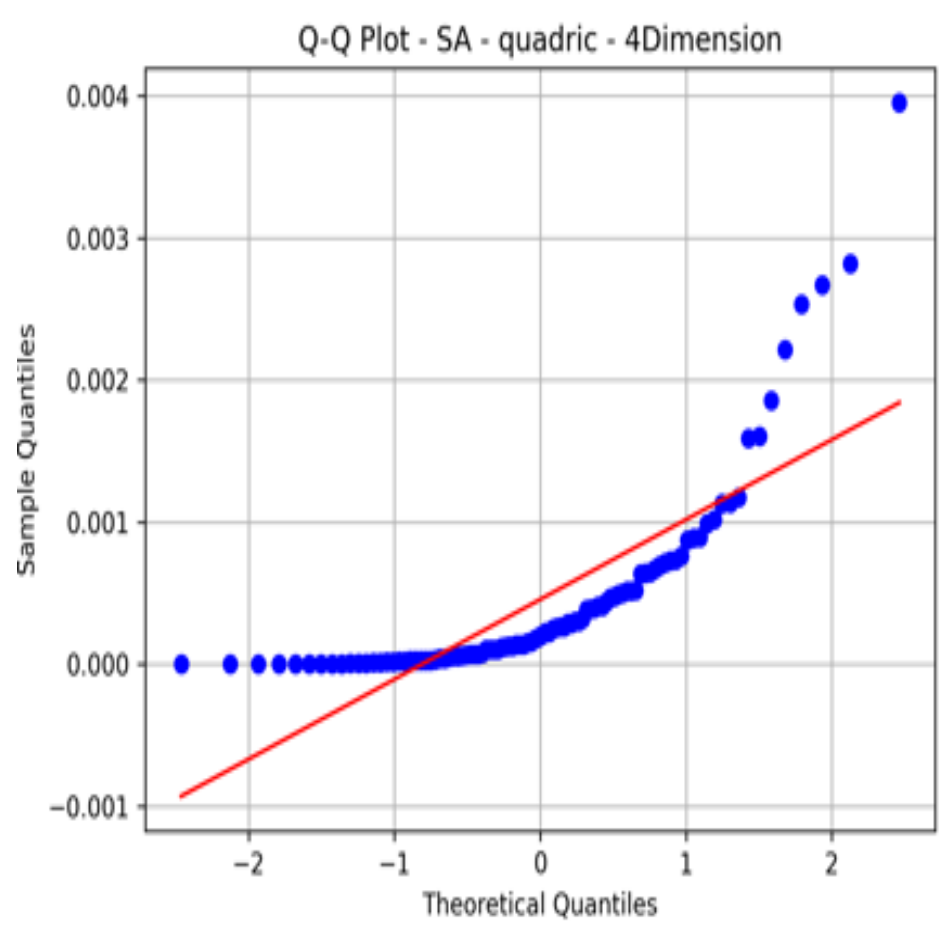
Q-Q Plot - SA - quadric - 4Dimension
Sample Quantiles
0.004
0.003
0.002
0.001
0.000
−0.001
−2
−1
0
1
2
Theoretical Quantiles

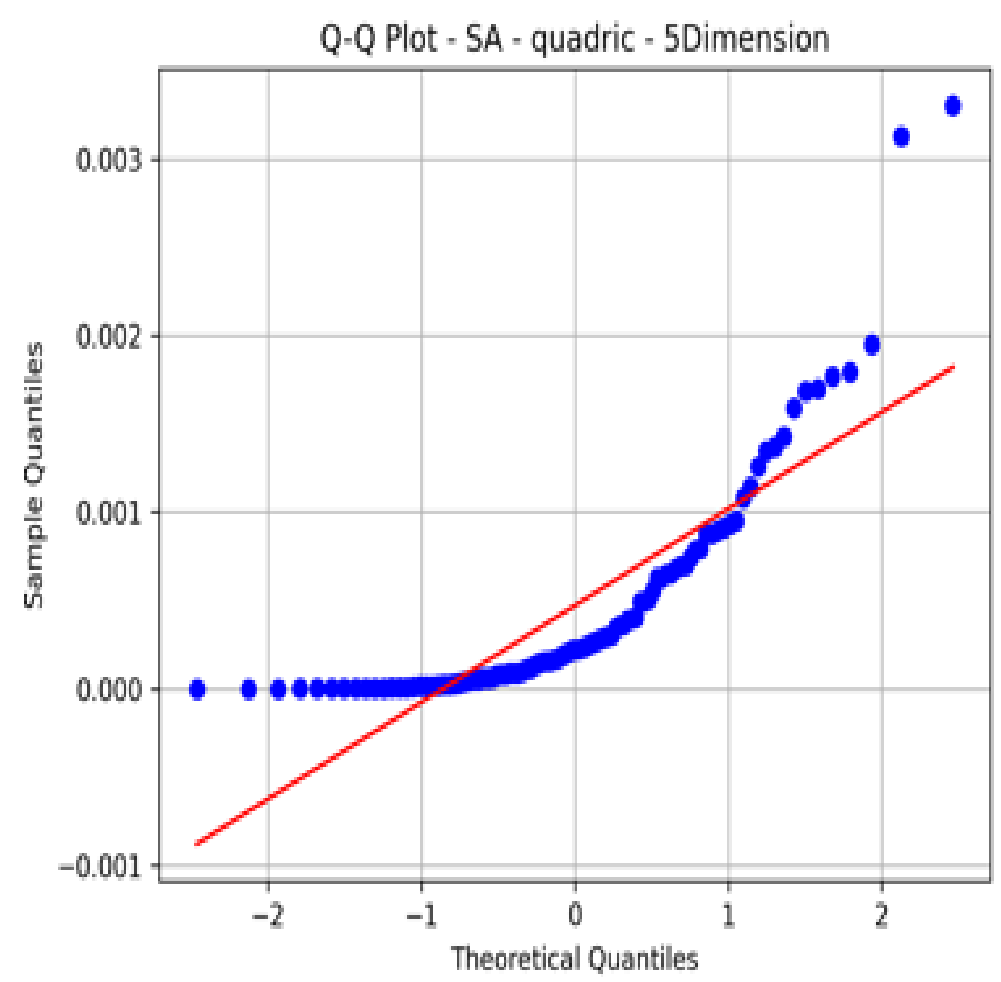
Q-Q Plot - SA - quadric - 5Dimension
Sample Quantiles
0.003
0.002
0.001
0.000
−0.001
−2
−1
0
1
2
Theoretical Quantiles

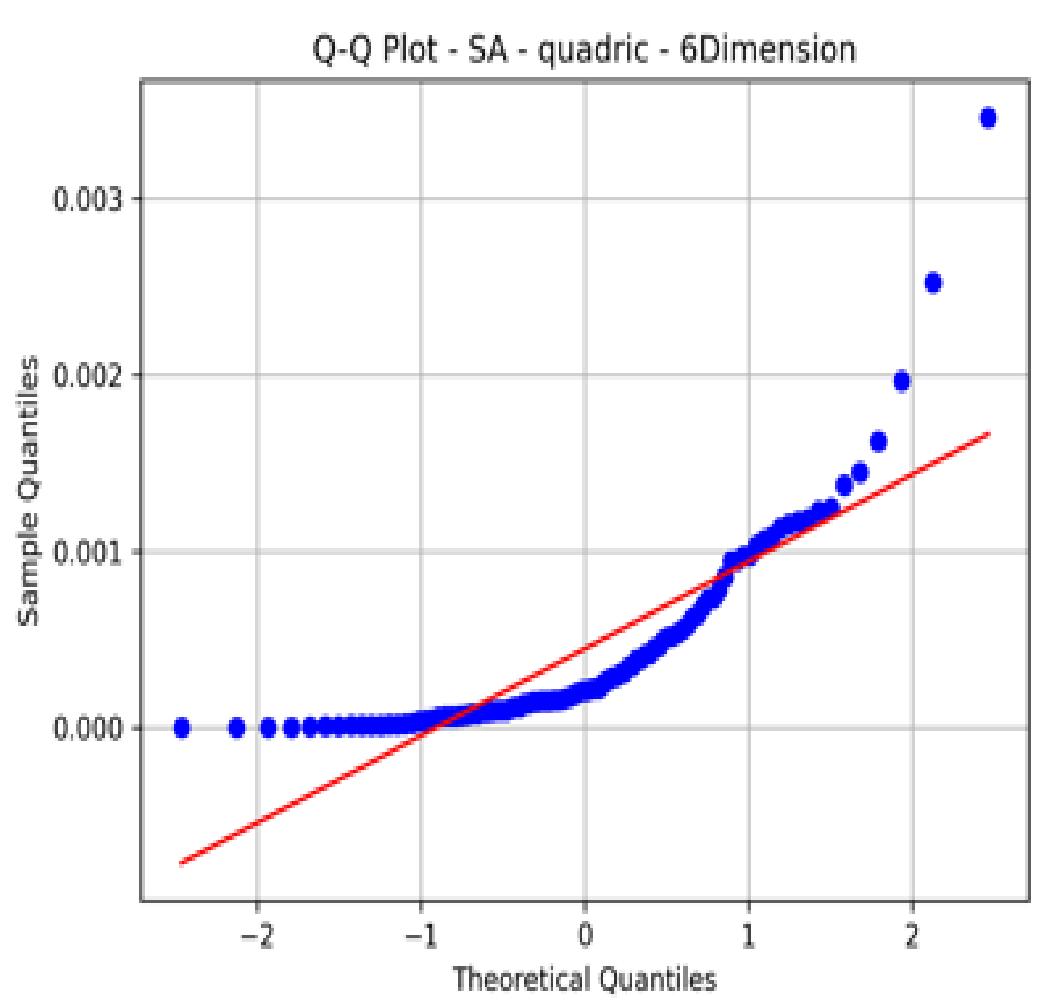
Q-Q Plot - SA - quadric - 6Dimension
Sample Quantiles
0.003
0.002
0.001
0.000
−2
−1
0
1
2
Theoretical Quantiles

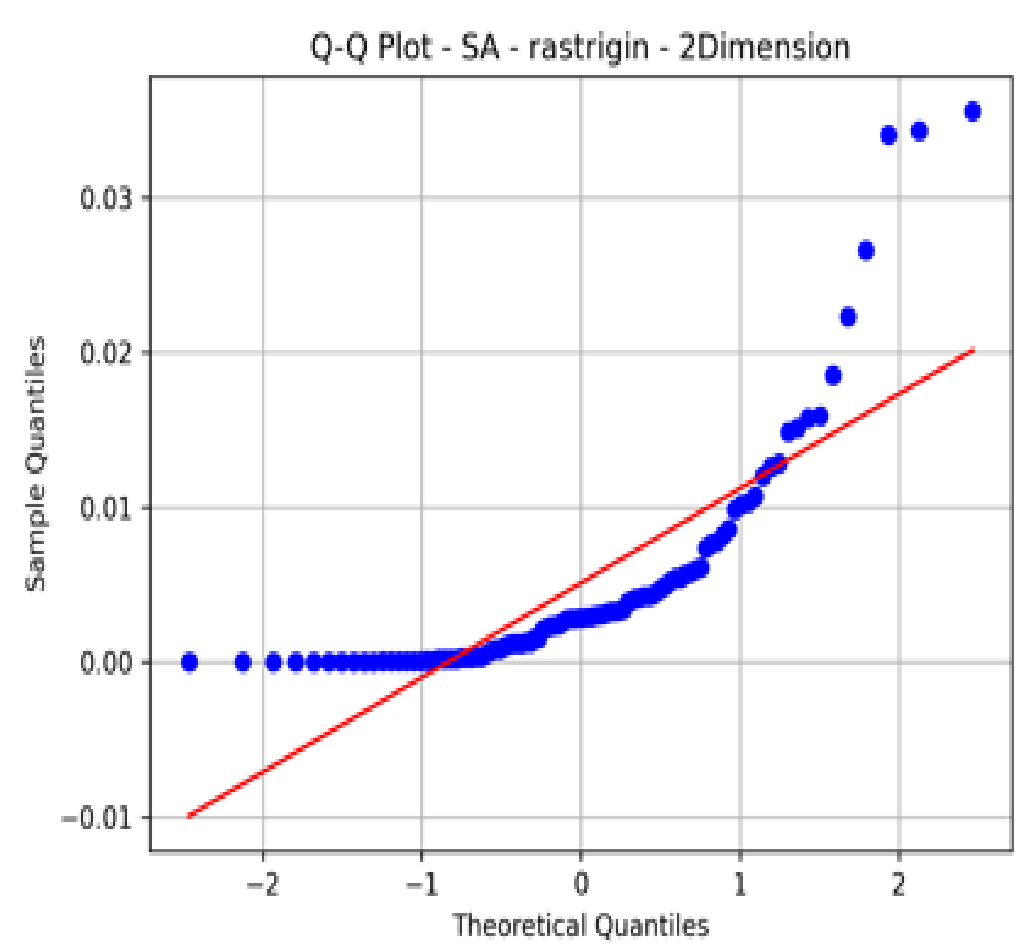
Q-Q Plot - SA - rastrigin - 2Dimension
Sample Quantiles
0.03
0.02
0.01
0.00
−0.01
−2
−1
0
1
2
Theoretical Quantiles

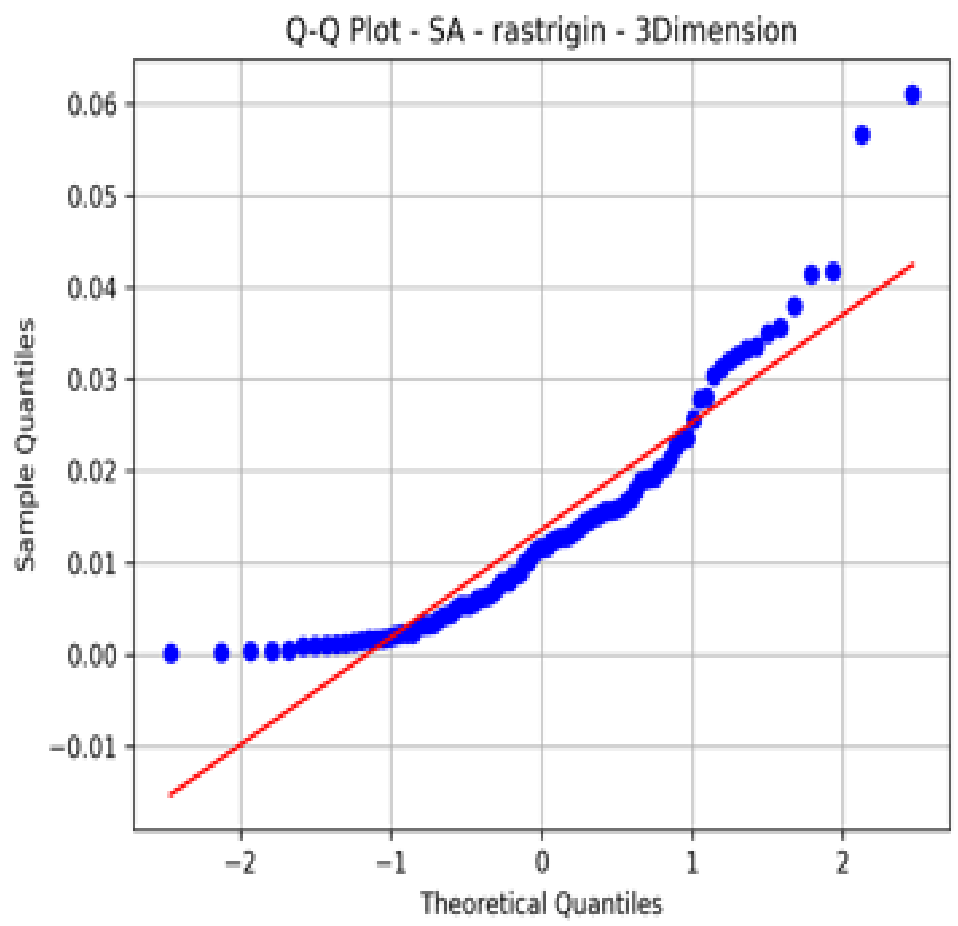
Q-Q Plot - SA - rastrigin - 3Dimension
Sample Quantiles
Theoretical Quantiles
0.06
0.05
0.04
0.03
0.02
0.01
0.00
−0.01
−2
−1
0
1
2

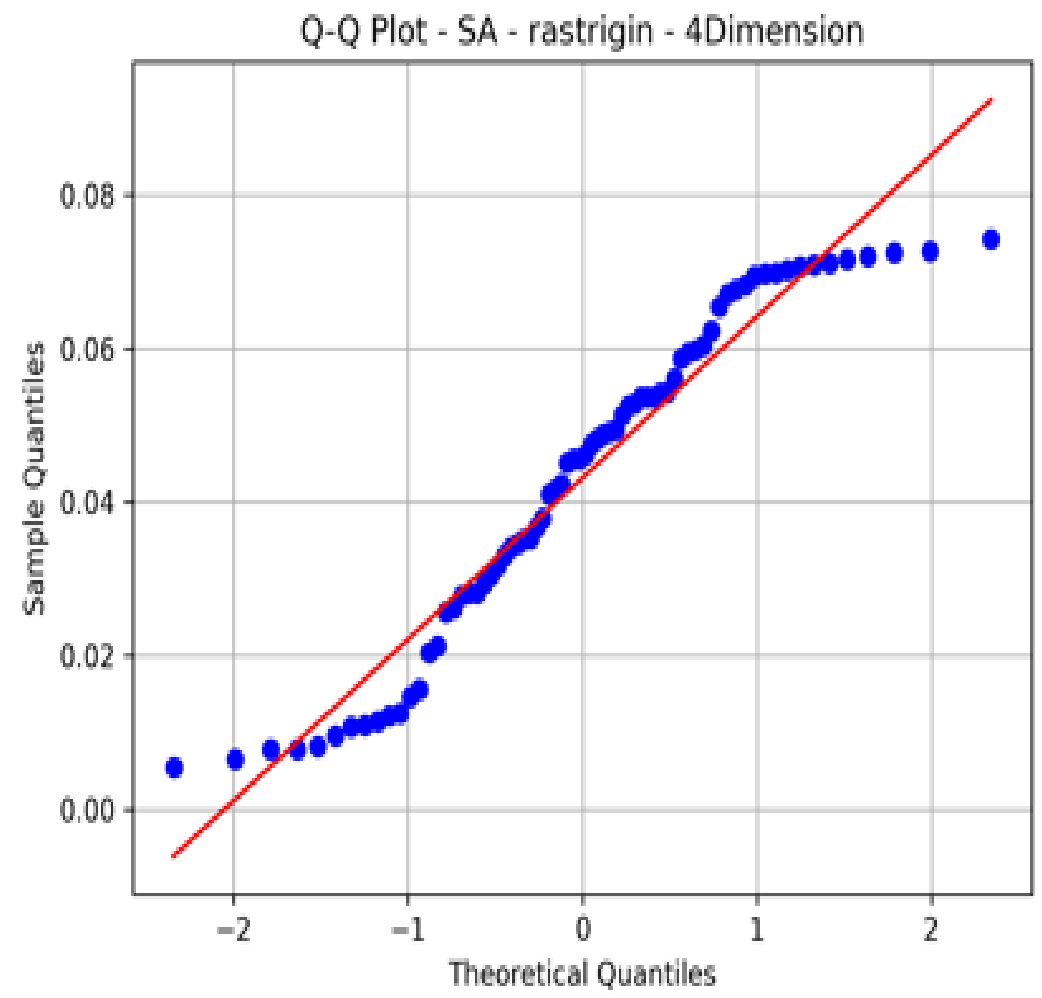
Q-Q Plot - SA - rastrigin - 4Dimension
Sample Quantiles
Theoretical Quantiles
0.08
0.06
0.04
0.02
0.00
−2
−1
0
1
2

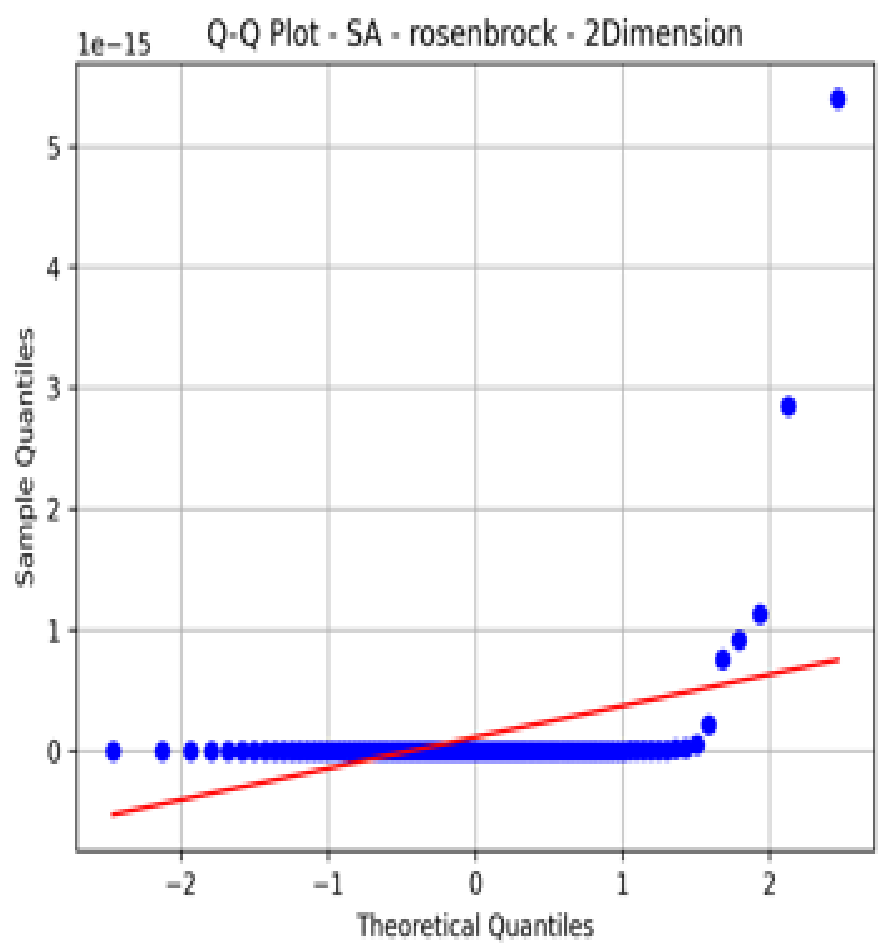
Q-Q Plot - SA - rosenbrock - 2Dimension
1e−15
Sample Quantiles
Theoretical Quantiles
5
4
3
2
1
0
−2
−1
0
1
2

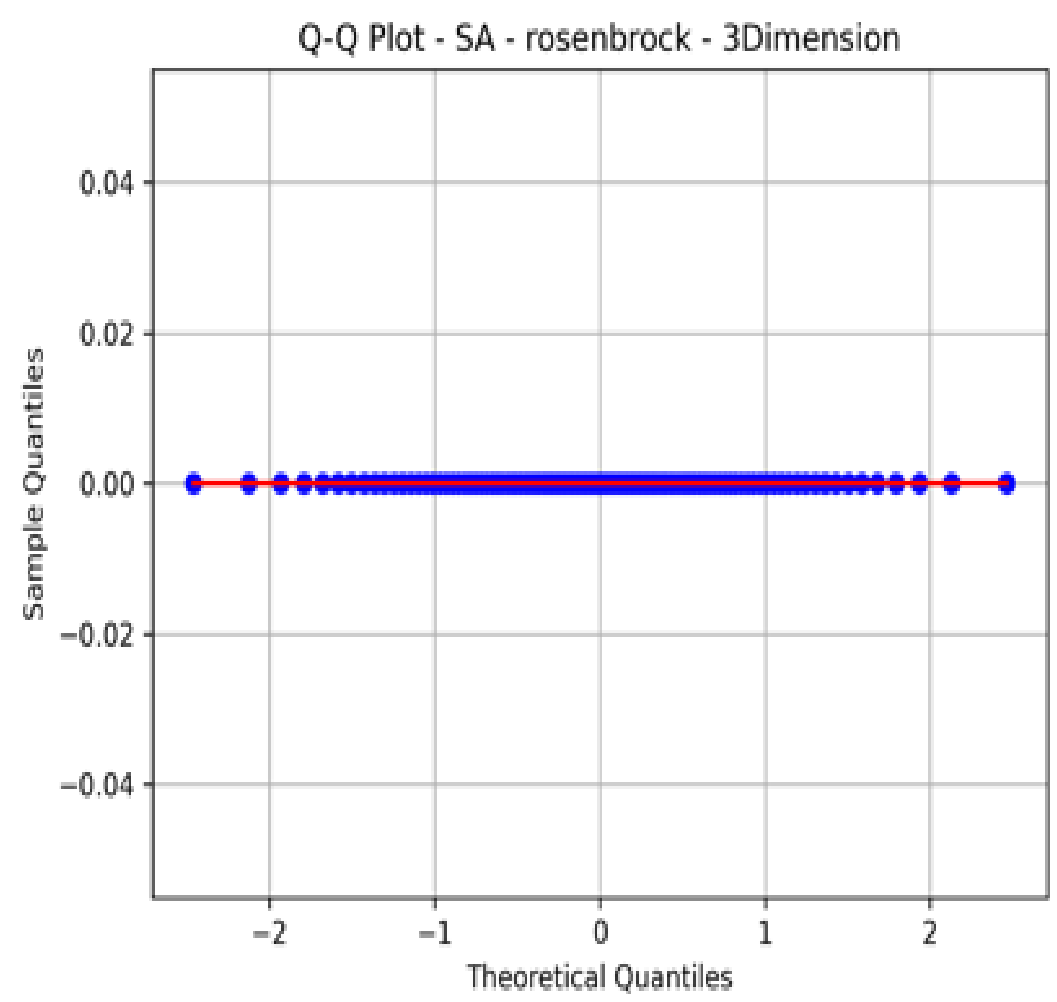
Q-Q Plot - SA - rosenbrock - 3Dimension
Sample Quantiles
Theoretical Quantiles
0.04
0.02
0.00
−0.02
−0.04
−2
−1
0
1
2

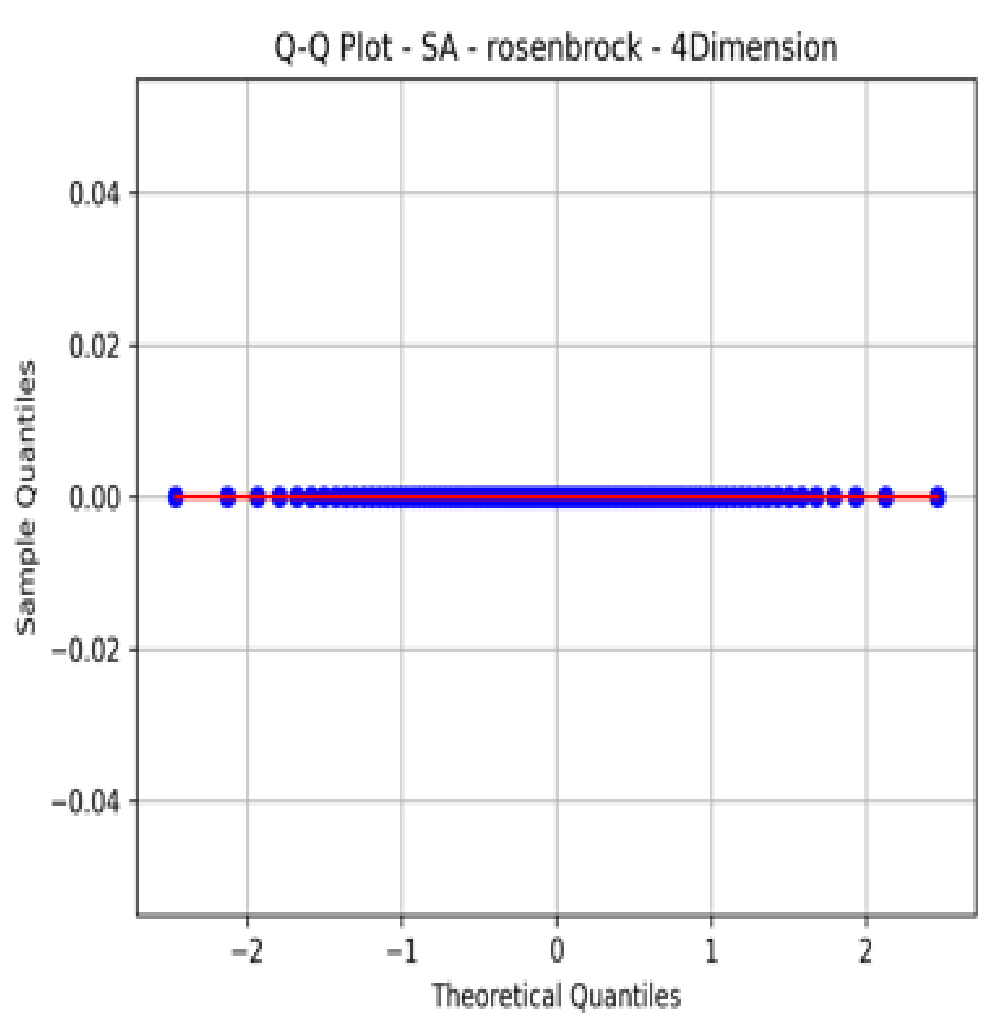
Q-Q Plot - SA - rosenbrock - 4Dimension
Sample Quantiles
Theoretical Quantiles
0.04
0.02
0.00
−0.02
−0.04
−2
−1
0
1
2

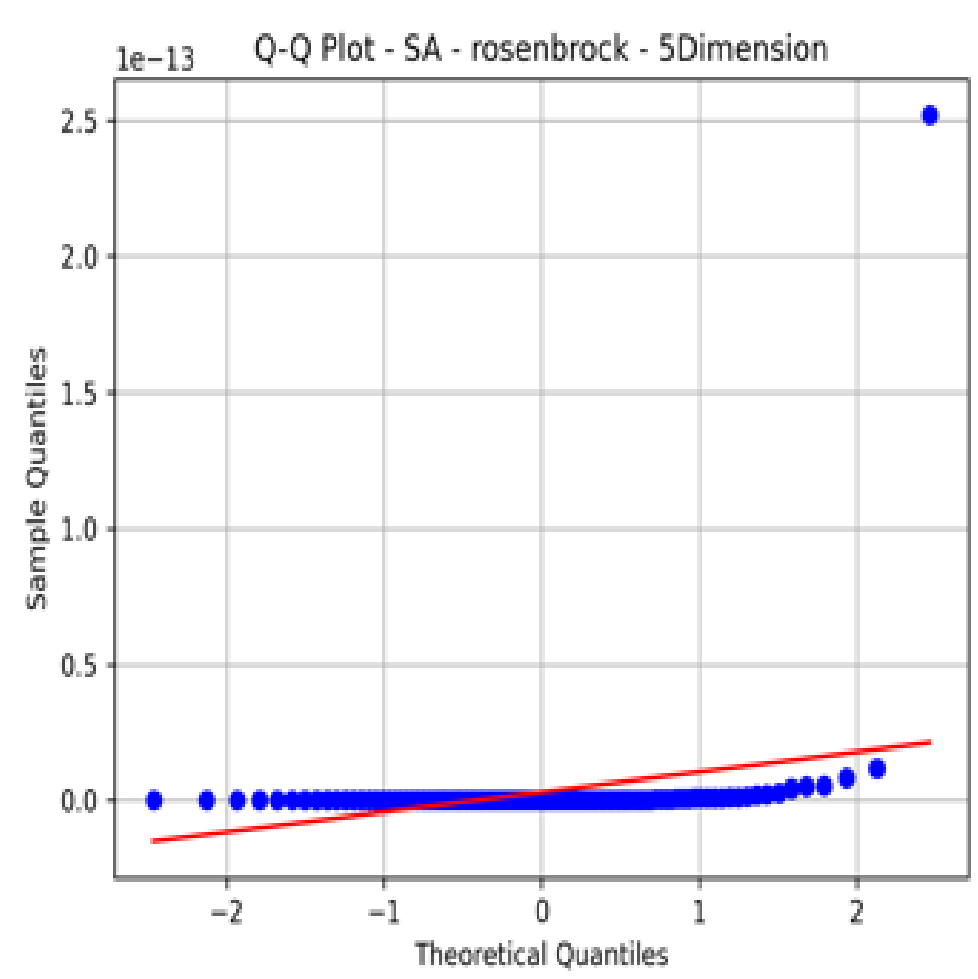
Q-Q Plot - SA - rosenbrock - 5Dimension
1e−13
Sample Quantiles
Theoretical Quantiles
2.5
2.0
1.5
1.0
0.5
0.0
−2
−1
0
1
2

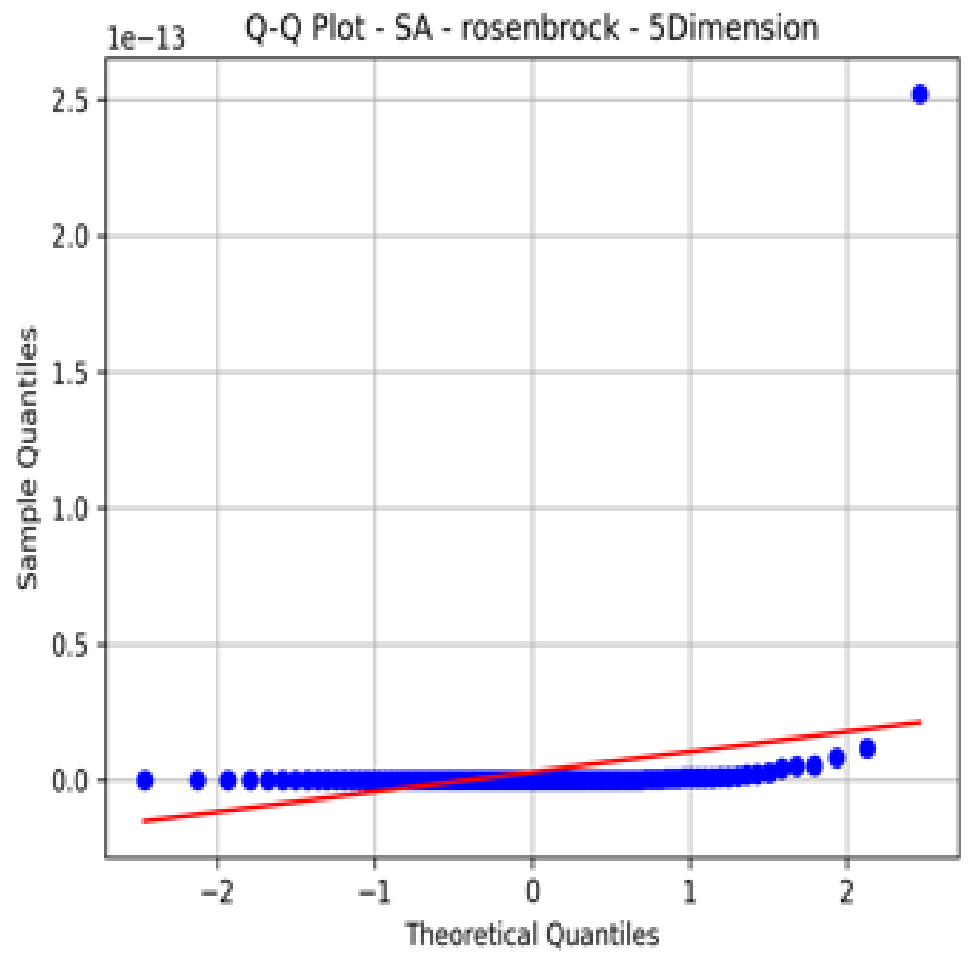
Q-Q Plot - SA - rosenbrock - 5Dimension
1e−13
2.5
2.0
1.5
1.0
0.5
0.0
Sample Quantiles
−2
−1
0
1
2
Theoretical Quantiles

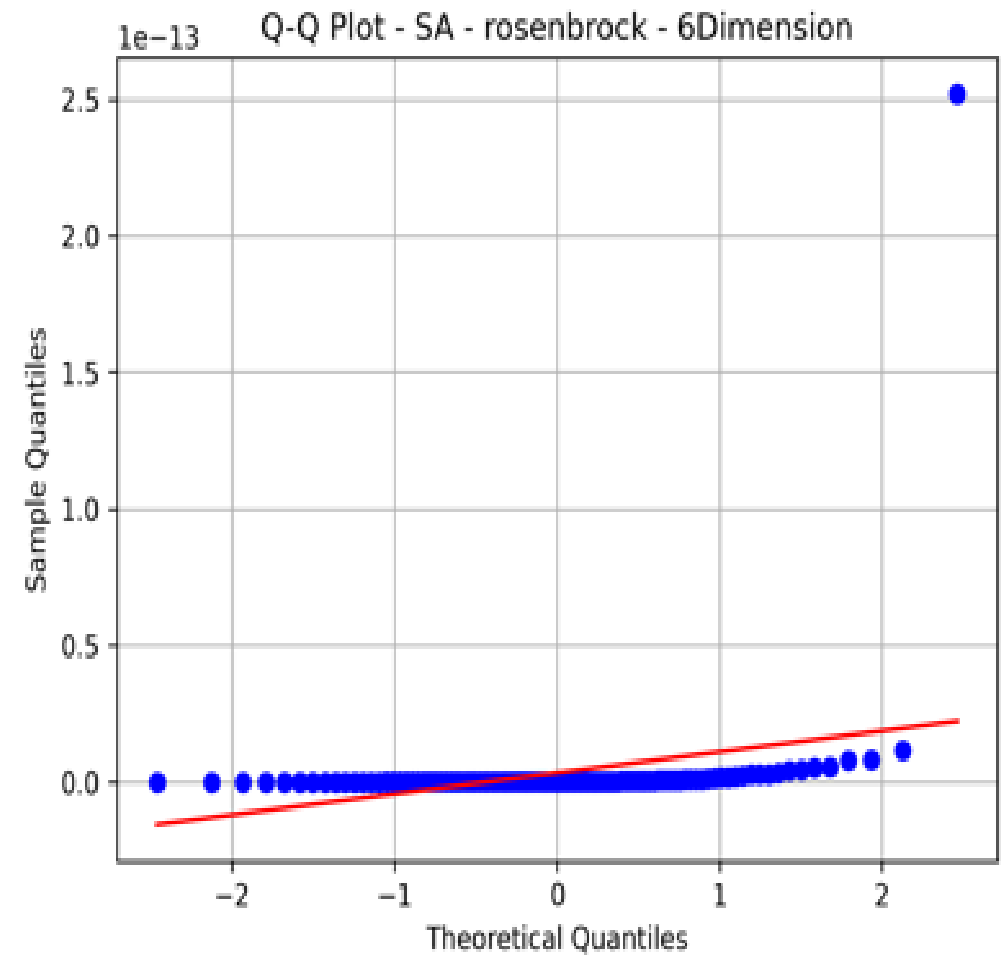
Q-Q Plot - SA - rosenbrock - 6Dimension
1e−13
2.5
2.0
1.5
1.0
0.5
0.0
Sample Quantiles
−2
−1
0
1
2
Theoretical Quantiles

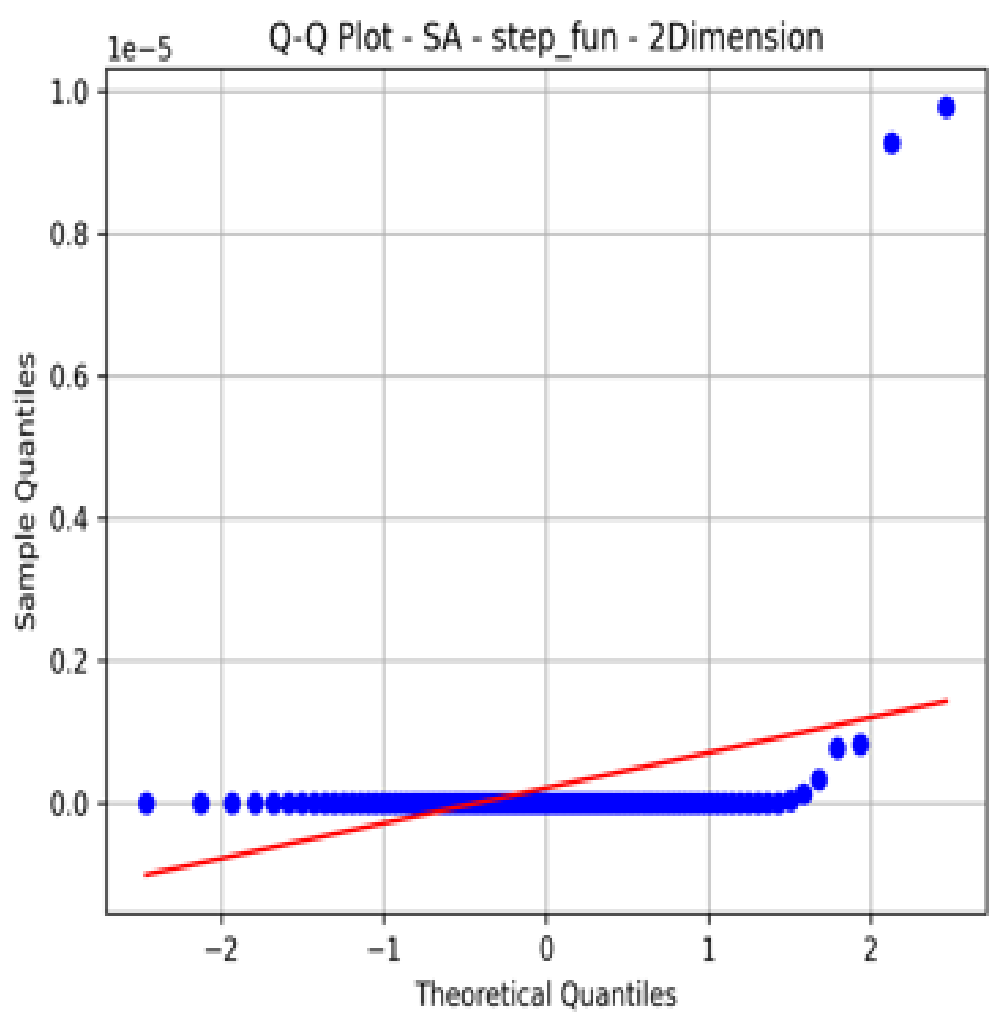
Q-Q Plot - SA - step_fun - 2Dimension
1e−5
1.0
0.8
0.6
0.4
0.2
0.0
Sample Quantiles
−2
−1
0
1
2
Theoretical Quantiles

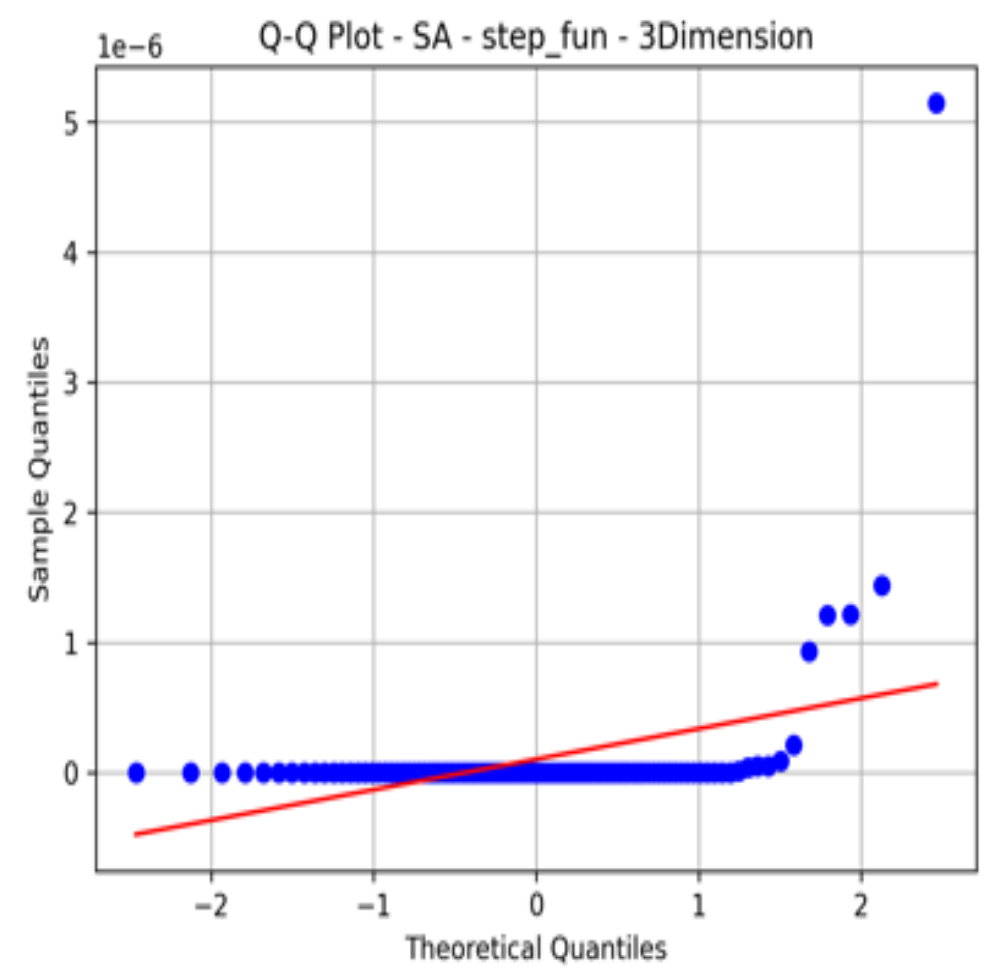
Q-Q Plot - SA - step_fun - 3Dimension
1e−6
5
4
3
2
1
0
Sample Quantiles
−2
−1
0
1
2
Theoretical Quantiles

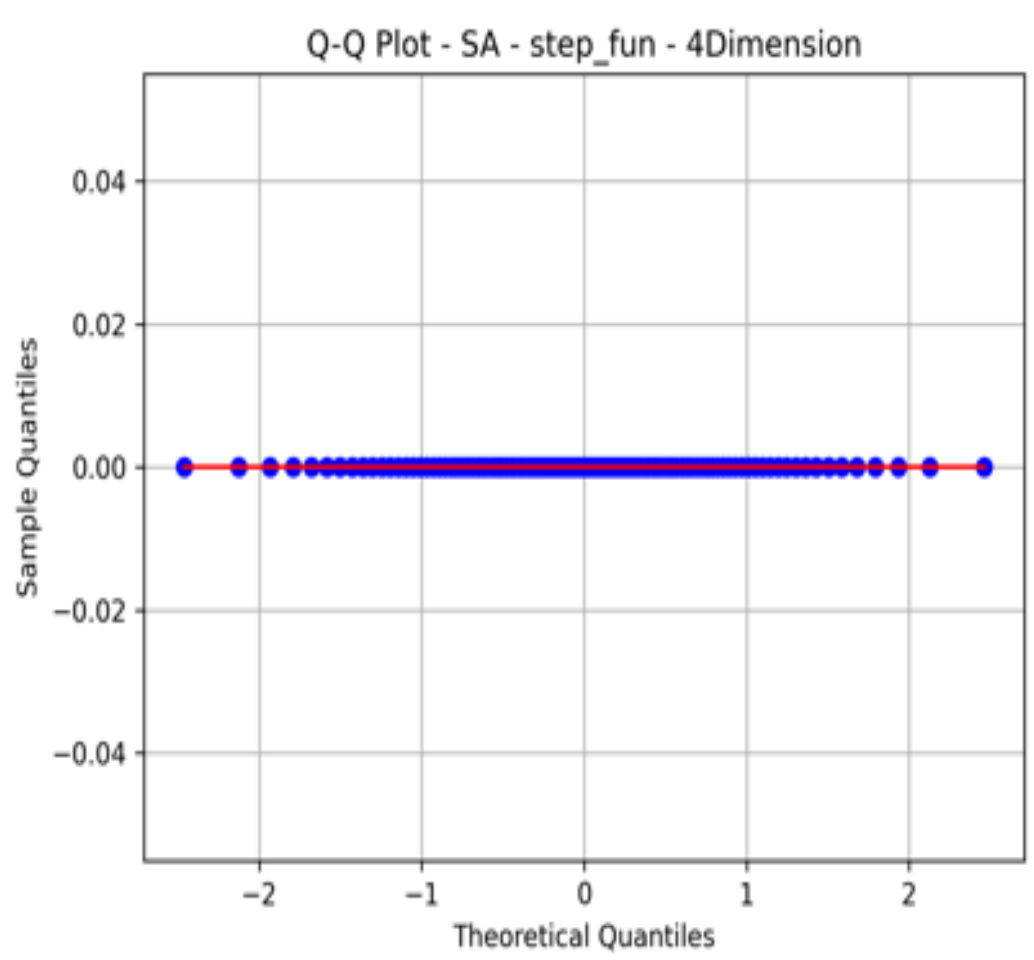
Q-Q Plot - SA - step_fun - 4Dimension
0.04
0.02
0.00
−0.02
−0.04
Sample Quantiles
−2
−1
0
1
2
Theoretical Quantiles

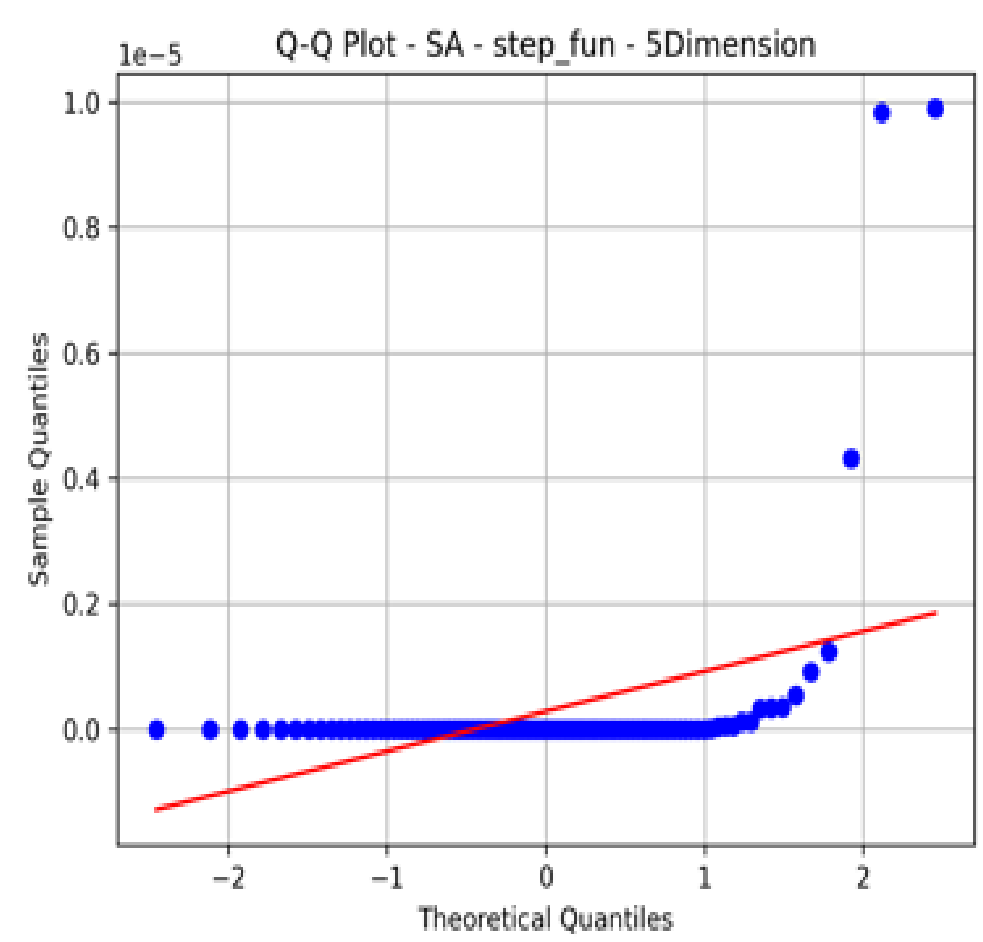
Q-Q Plot - SA - step_fun - 5Dimension
1e−5
1.0
0.8
0.6
0.4
0.2
0.0
Sample Quantiles
−2
−1
0
1
2
Theoretical Quantiles

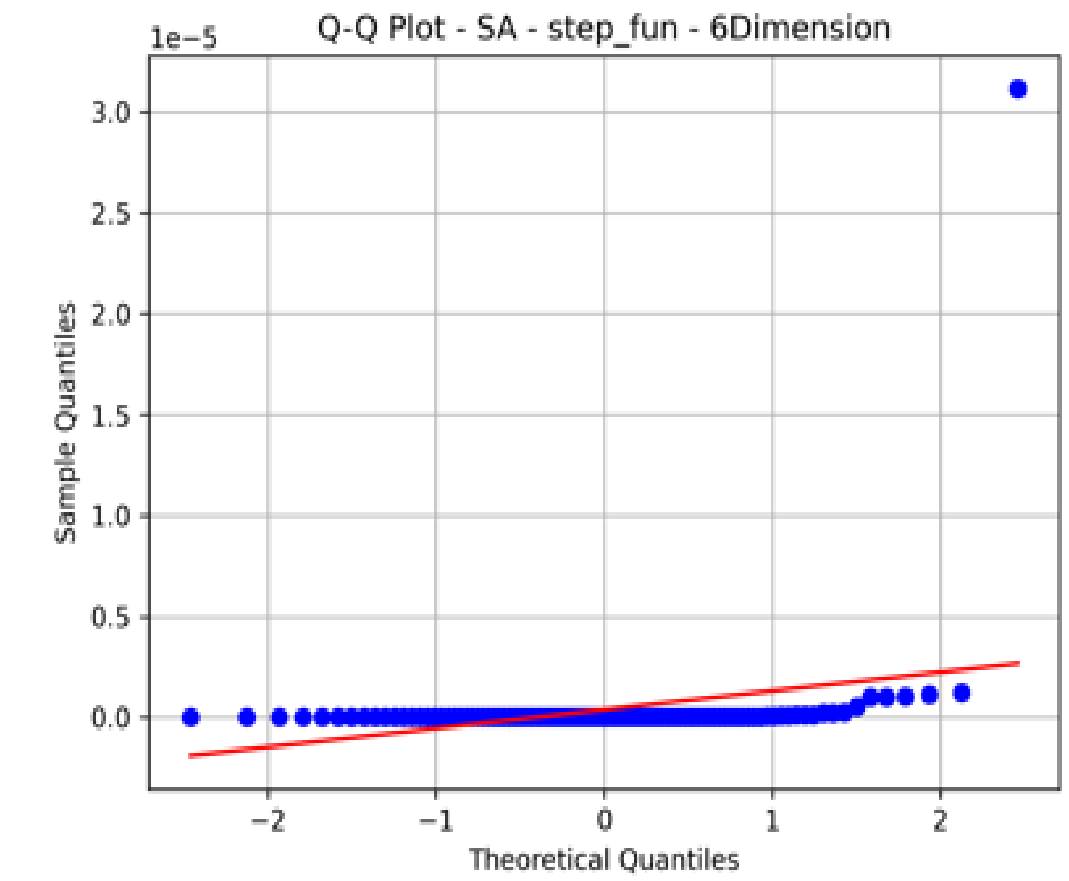

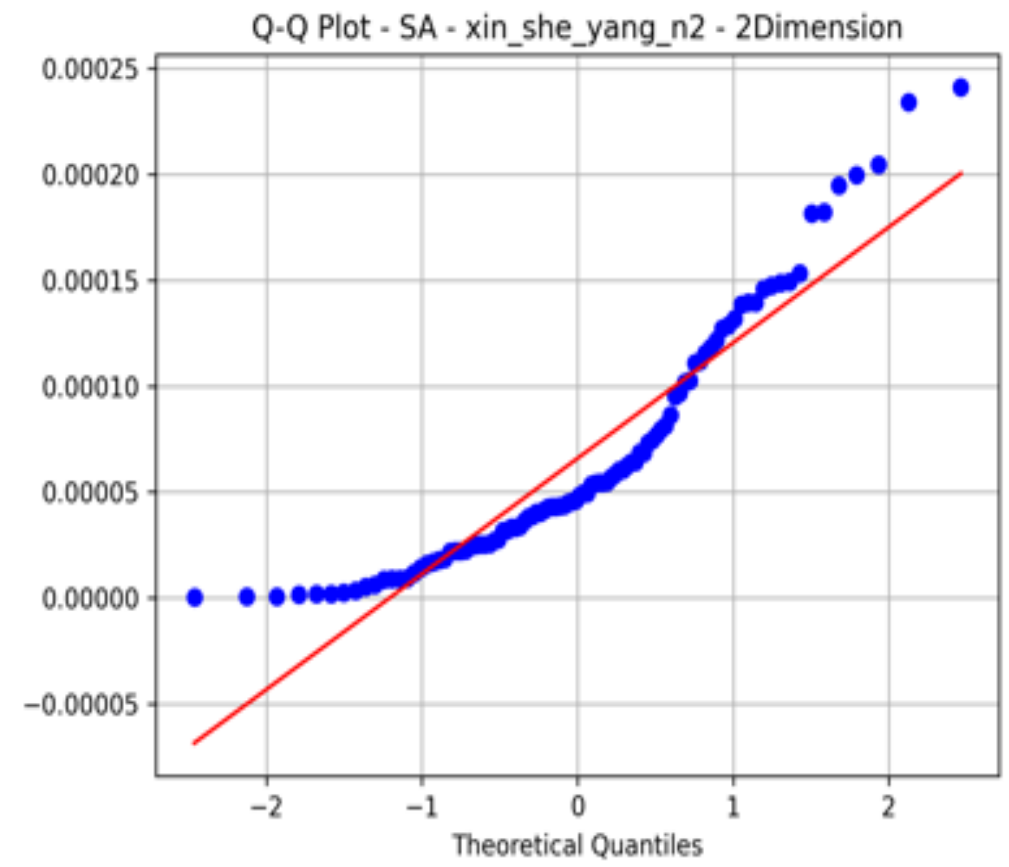

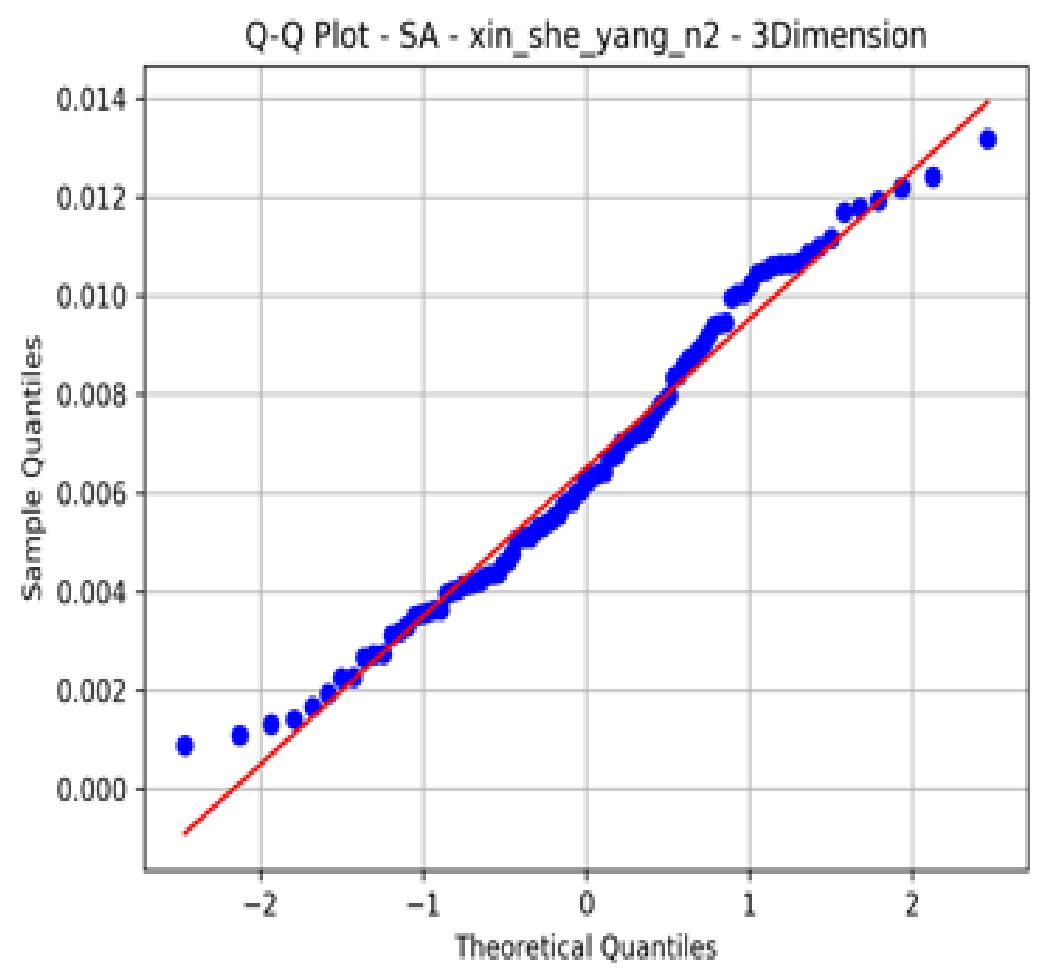

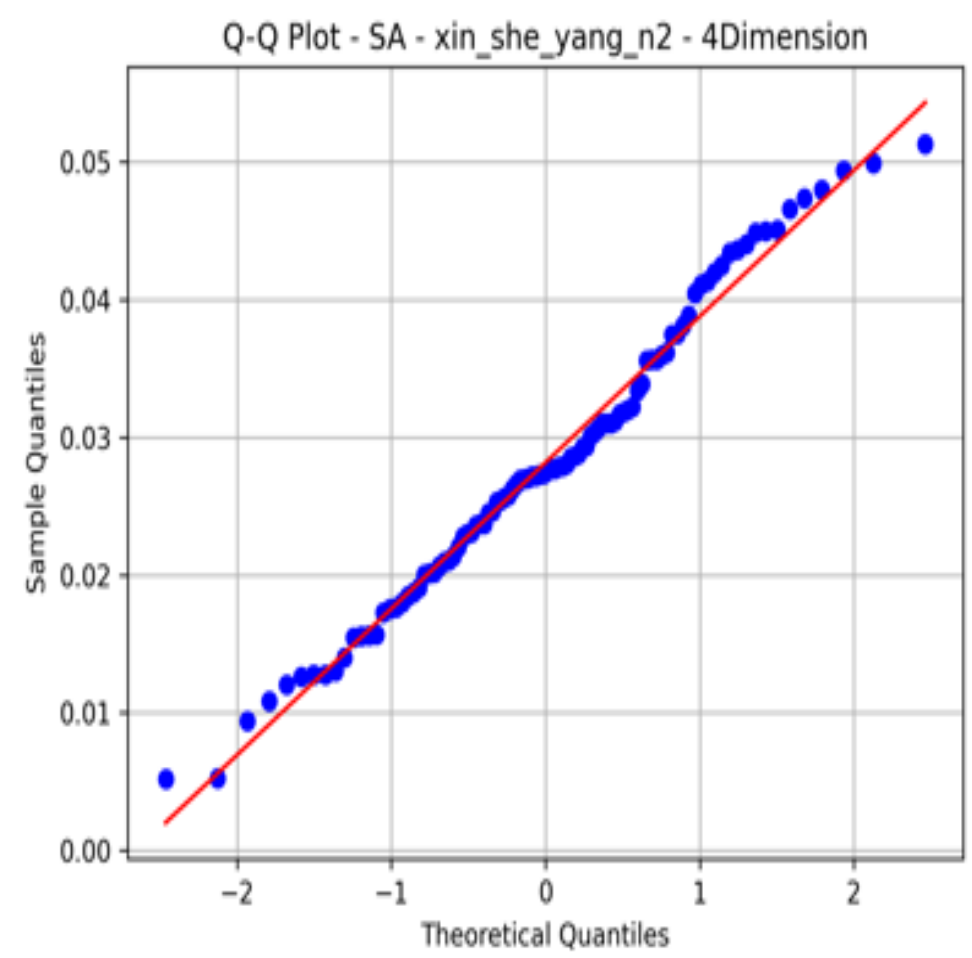

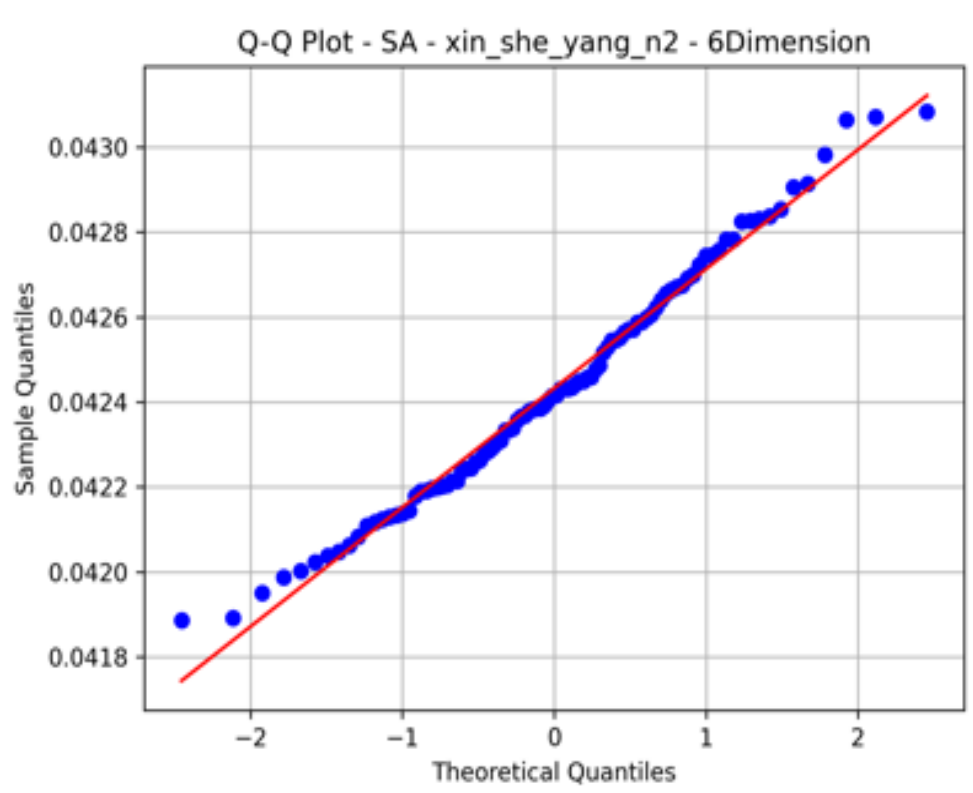

<u>TA Q-Q Plots</u>

Non-Continuous functions QQ Plots

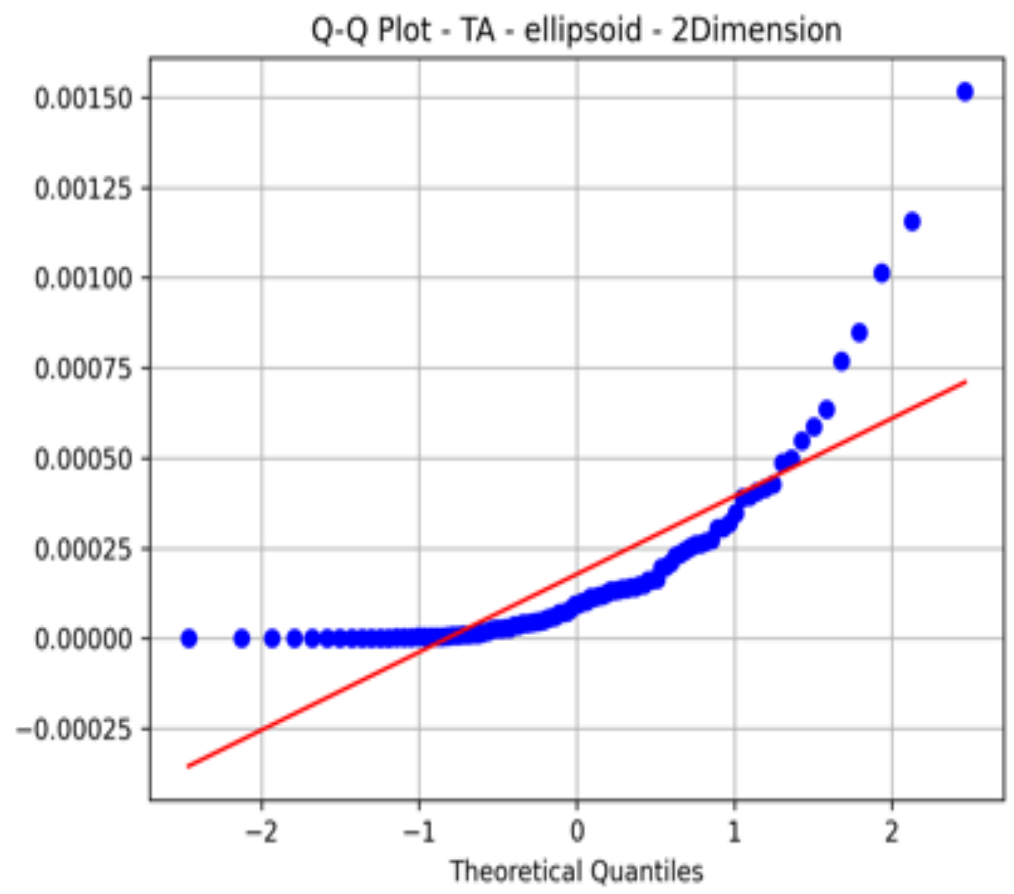
Q-Q Plot - TA - ellipsoid - 2Dimension
0.00150
0.00125
0.00100
0.00075
0.00050
0.00025
0.00000
−0.00025
−2
−1
0
1
2
Theoretical Quantiles

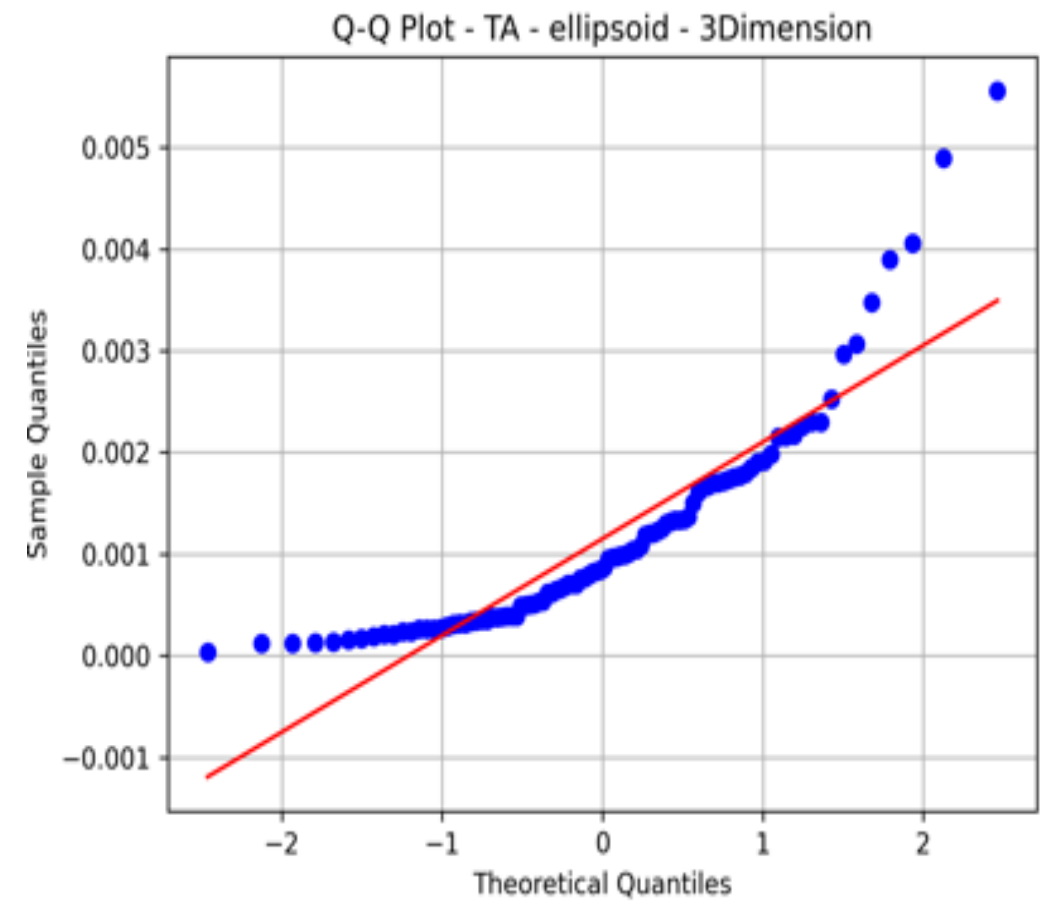
Q-Q Plot - TA - ellipsoid - 3Dimension
Sample Quantiles
0.005
0.004
0.003
0.002
0.001
0.000
−0.001
−2
−1
0
1
2
Theoretical Quantiles

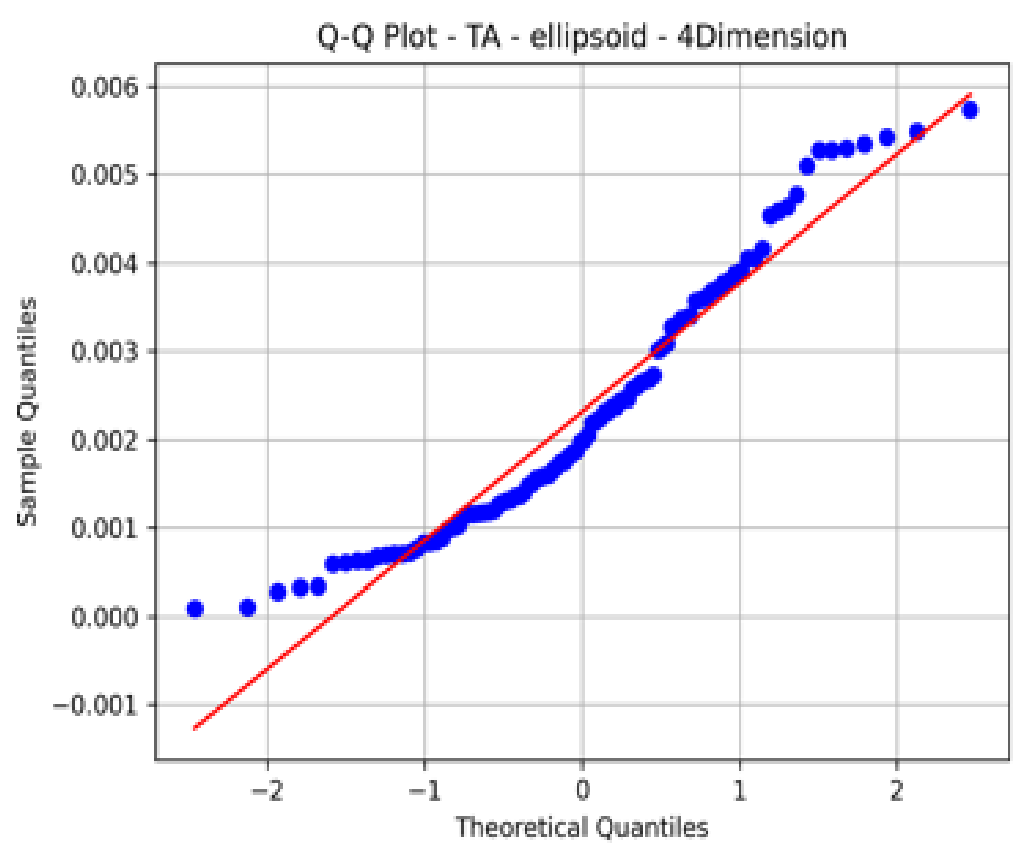
Q-Q Plot - TA - ellipsoid - 4Dimension
Sample Quantiles
0.006
0.005
0.004
0.003
0.002
0.001
0.000
−0.001
−2
−1
0
1
2
Theoretical Quantiles

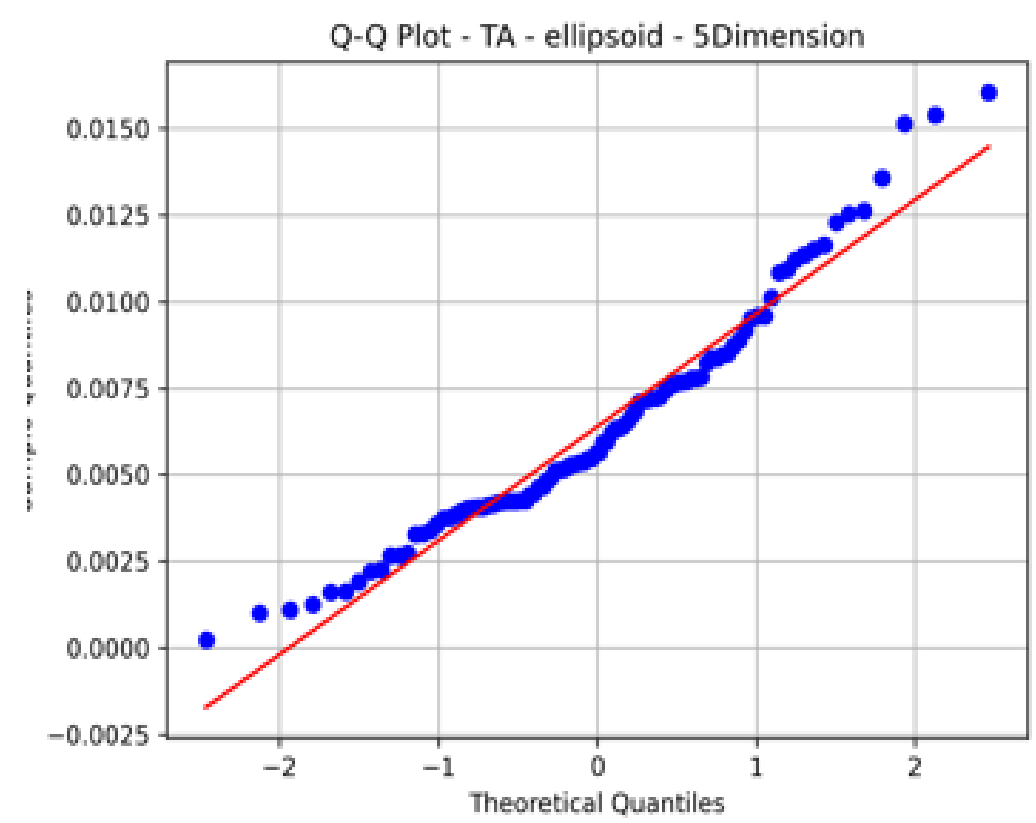
Q-Q Plot - TA - ellipsoid - 5Dimension
0.0150
0.0125
0.0100
0.0075
0.0050
0.0025
0.0000
−0.0025
−2
−1
0
1
2
Theoretical Quantiles

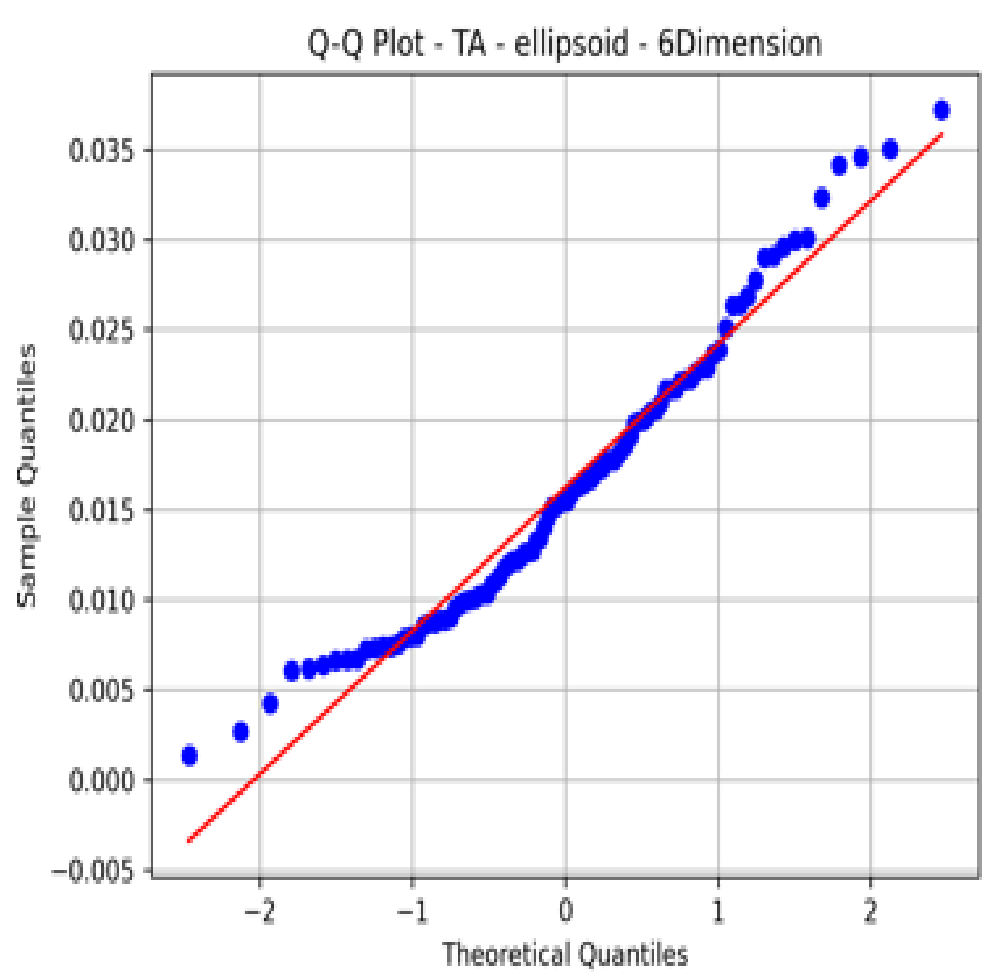
Q-Q Plot - TA - ellipsoid - 6Dimension
Sample Quantiles
0.035
0.030
0.025
0.020
0.015
0.010
0.005
0.000
−0.005
−2
−1
0
1
2
Theoretical Quantiles

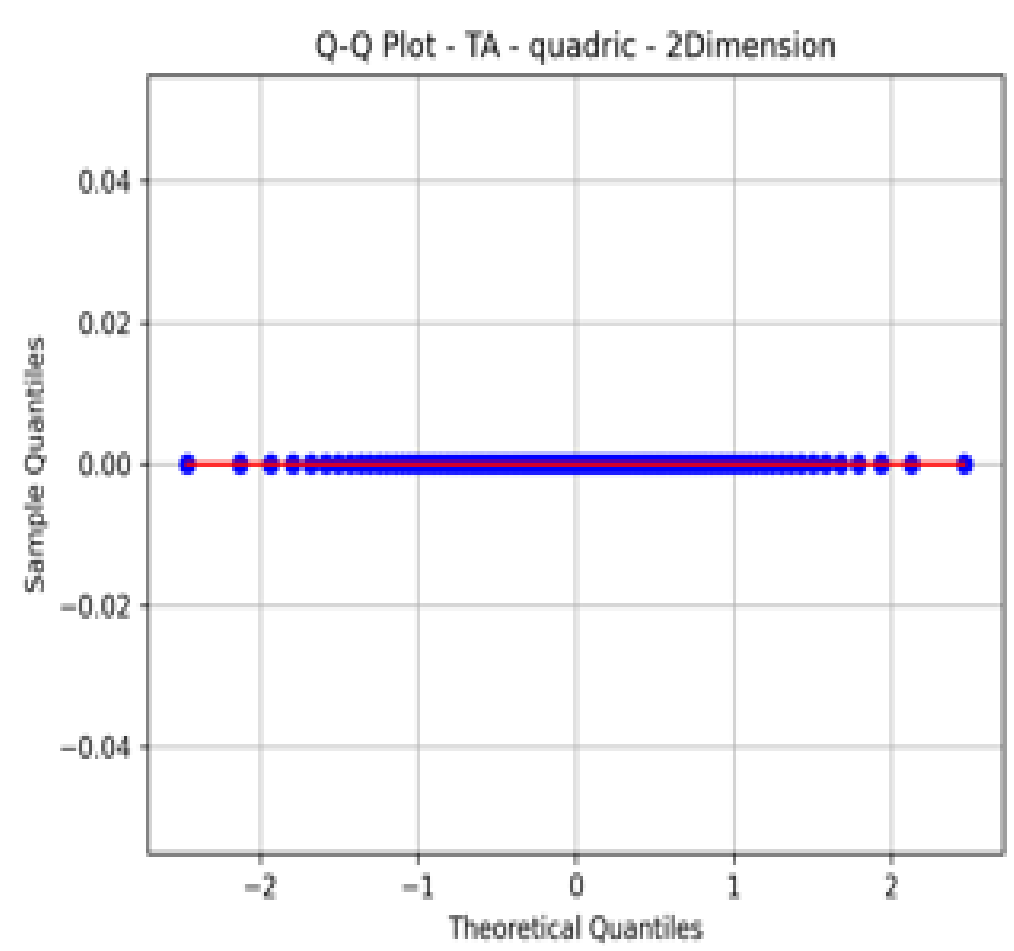
Q-Q Plot - TA - quadric - 2Dimension
Sample Quantiles
0.04
0.02
0.00
−0.02
−0.04
−2
−1
0
1
2
Theoretical Quantiles

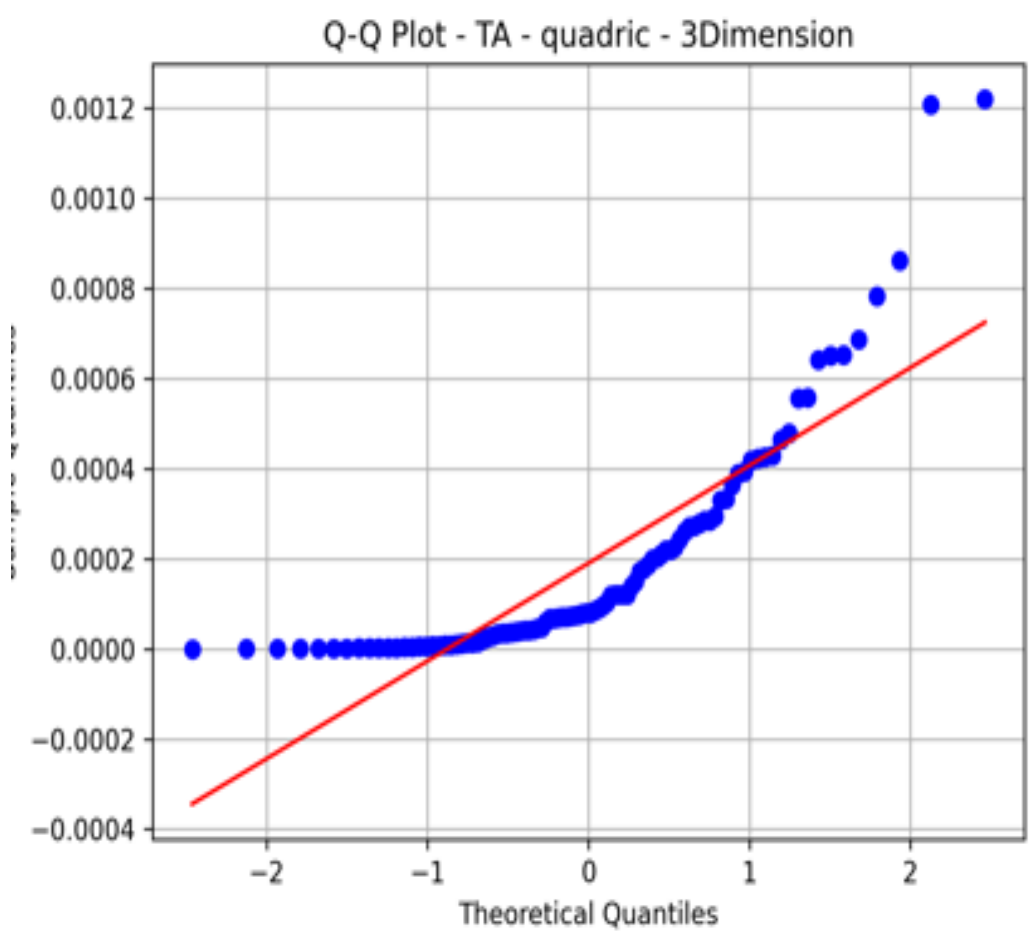
Q-Q Plot - TA - quadric - 3Dimension
Theoretical Quantiles

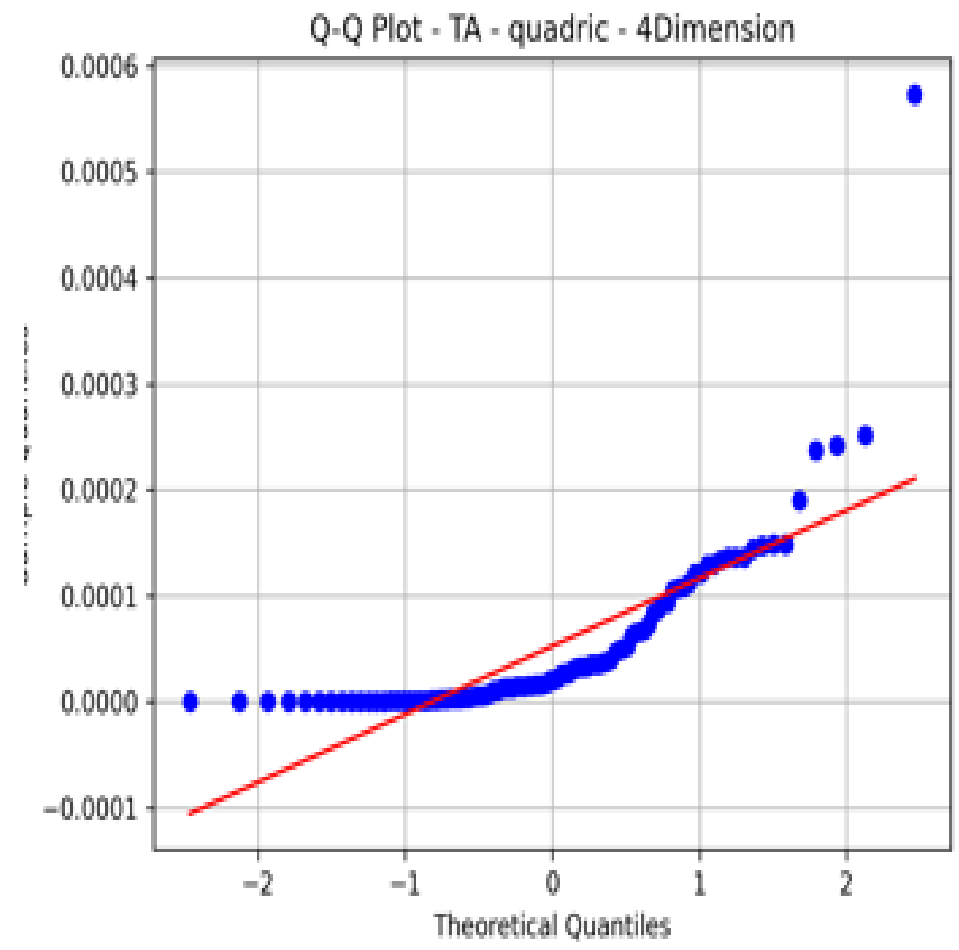
Q-Q Plot - TA - quadric - 4Dimension
Theoretical Quantiles

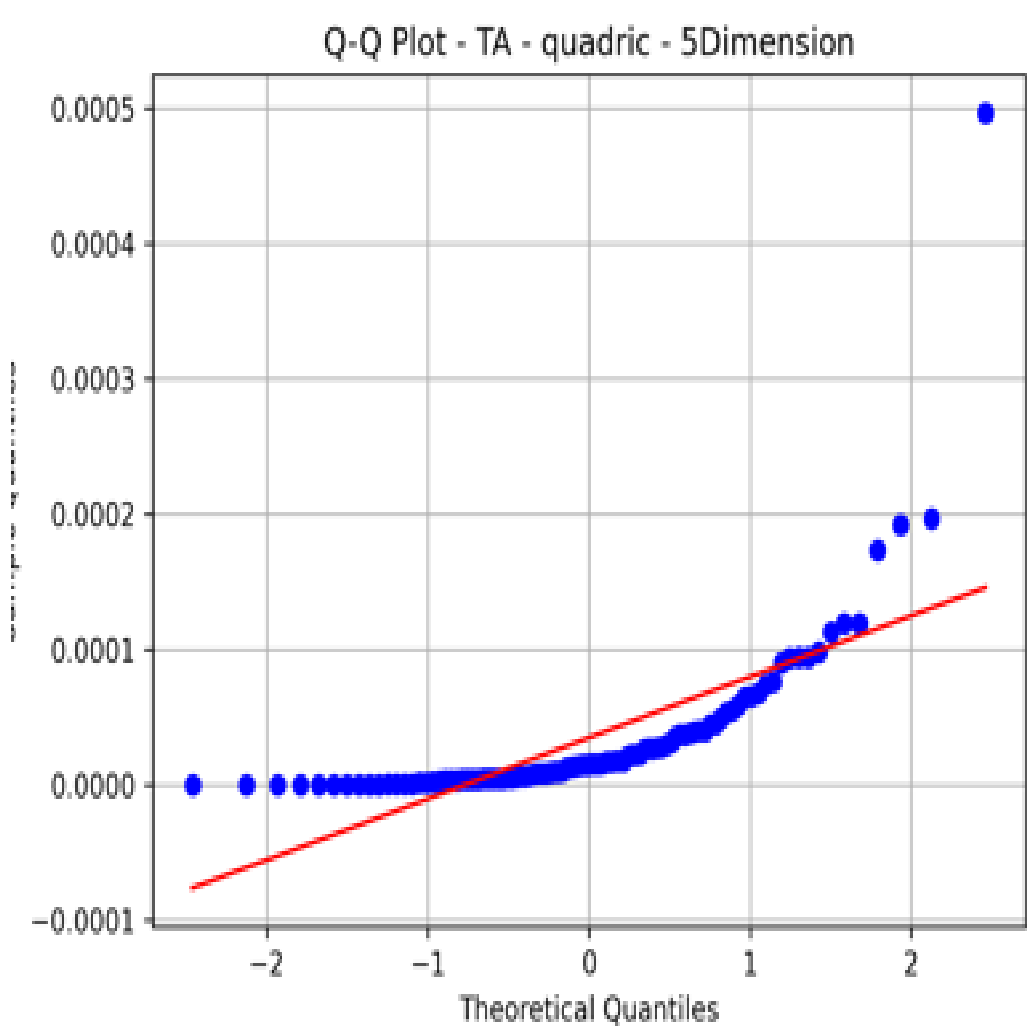
Q-Q Plot - TA - quadric - 5Dimension
Theoretical Quantiles

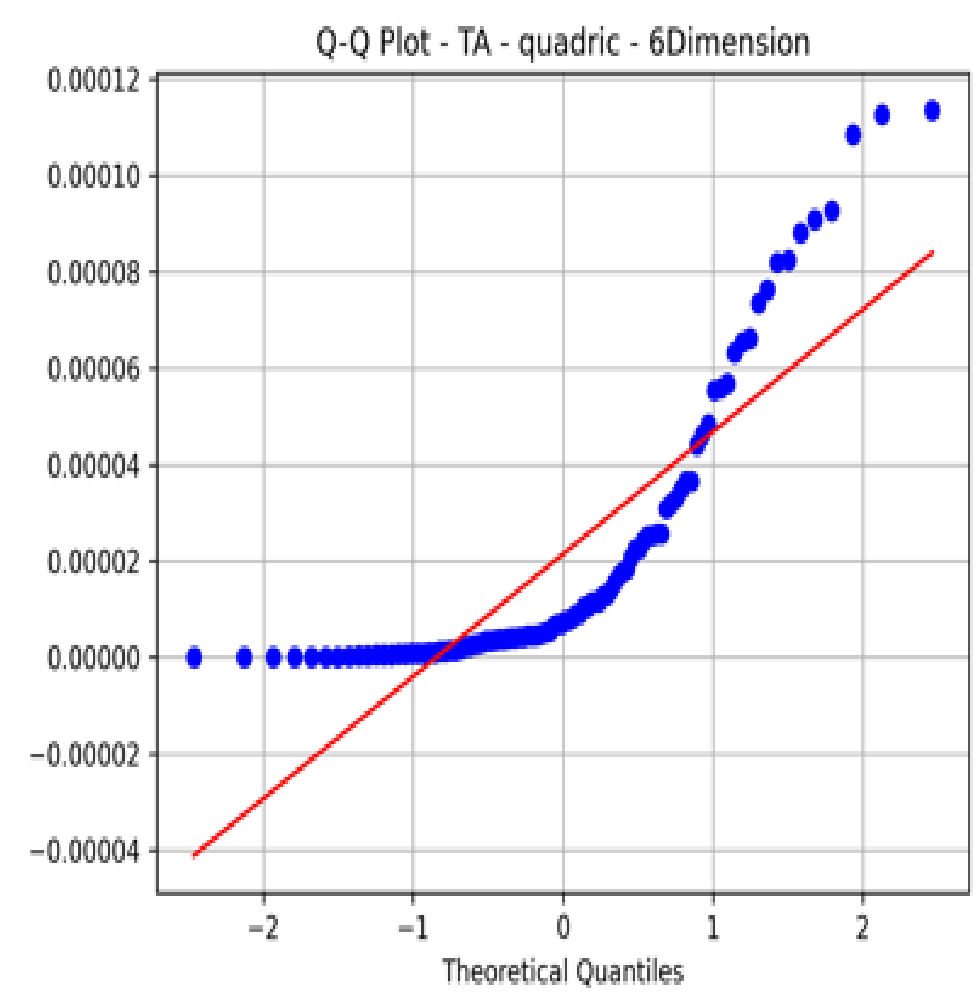
Q-Q Plot - TA - quadric - 6Dimension
Theoretical Quantiles

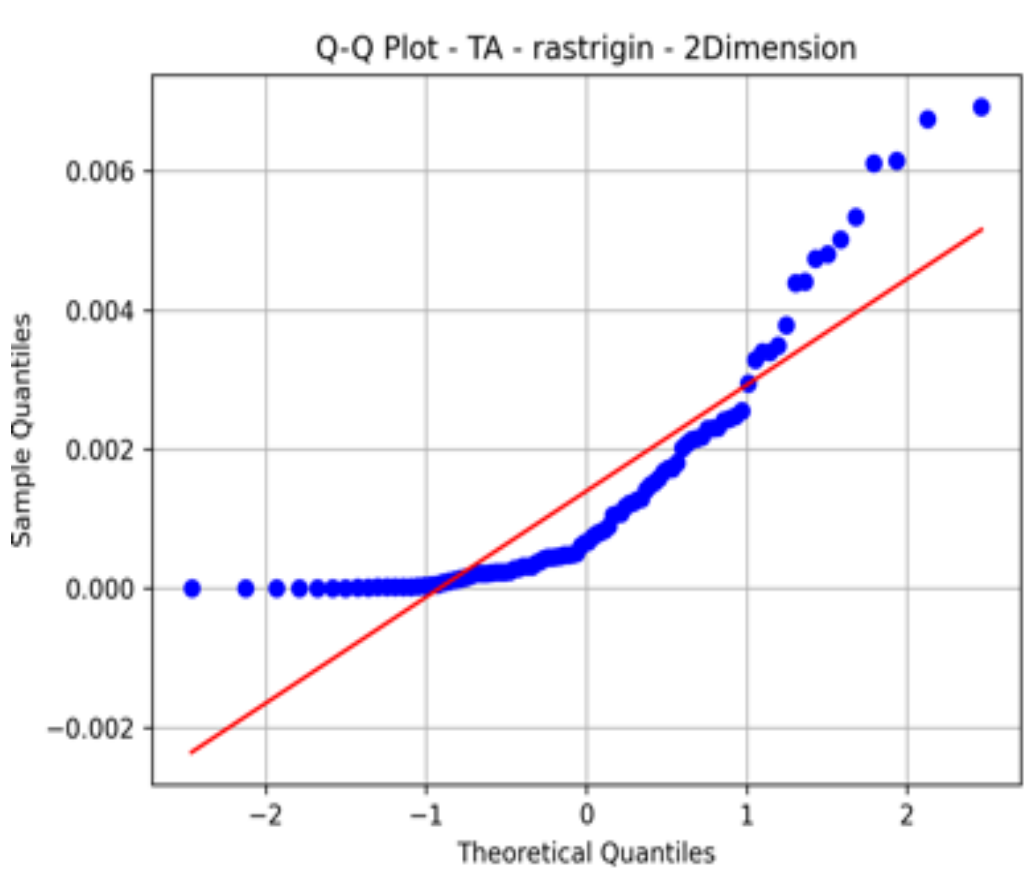
Q-Q Plot - TA - rastrigin - 2Dimension
Sample Quantiles
Theoretical Quantiles

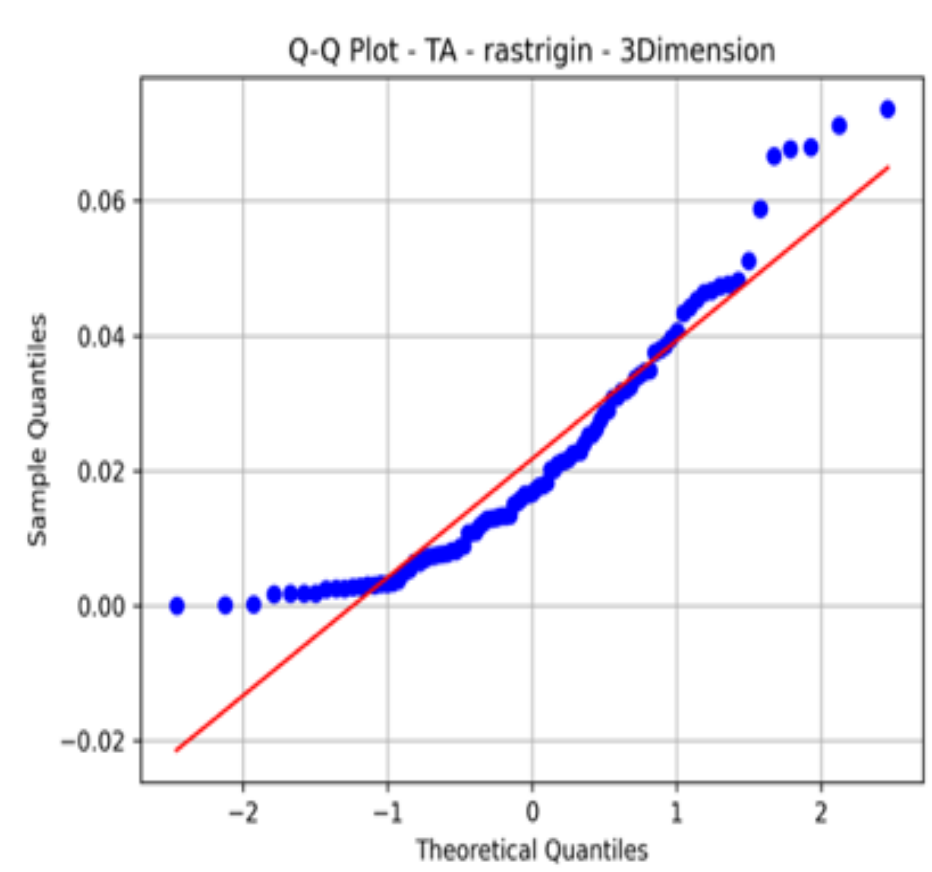
Q-Q Plot - TA - rastrigin - 3Dimension
Sample Quantiles
Theoretical Quantiles

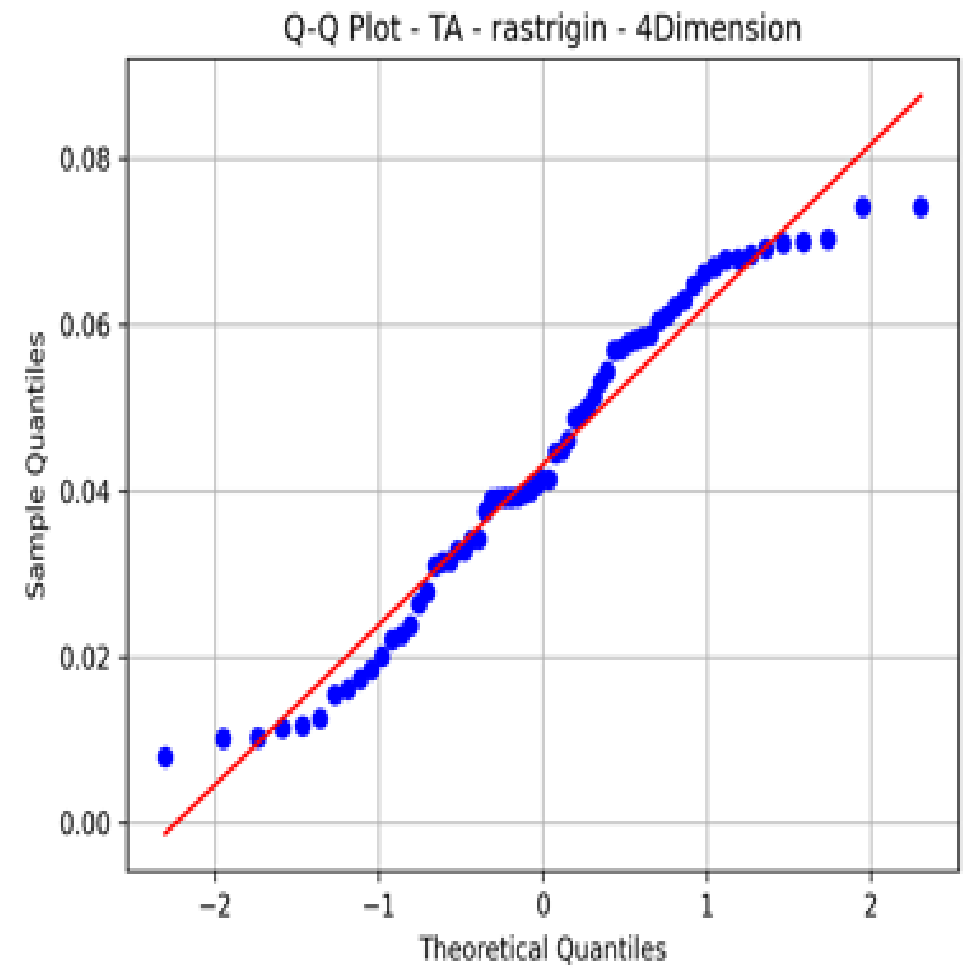
Q-Q Plot - TA - rastrigin - 4Dimension
Sample Quantiles
Theoretical Quantiles
0.08
0.06
0.04
0.02
0.00
−2
−1
0
1
2

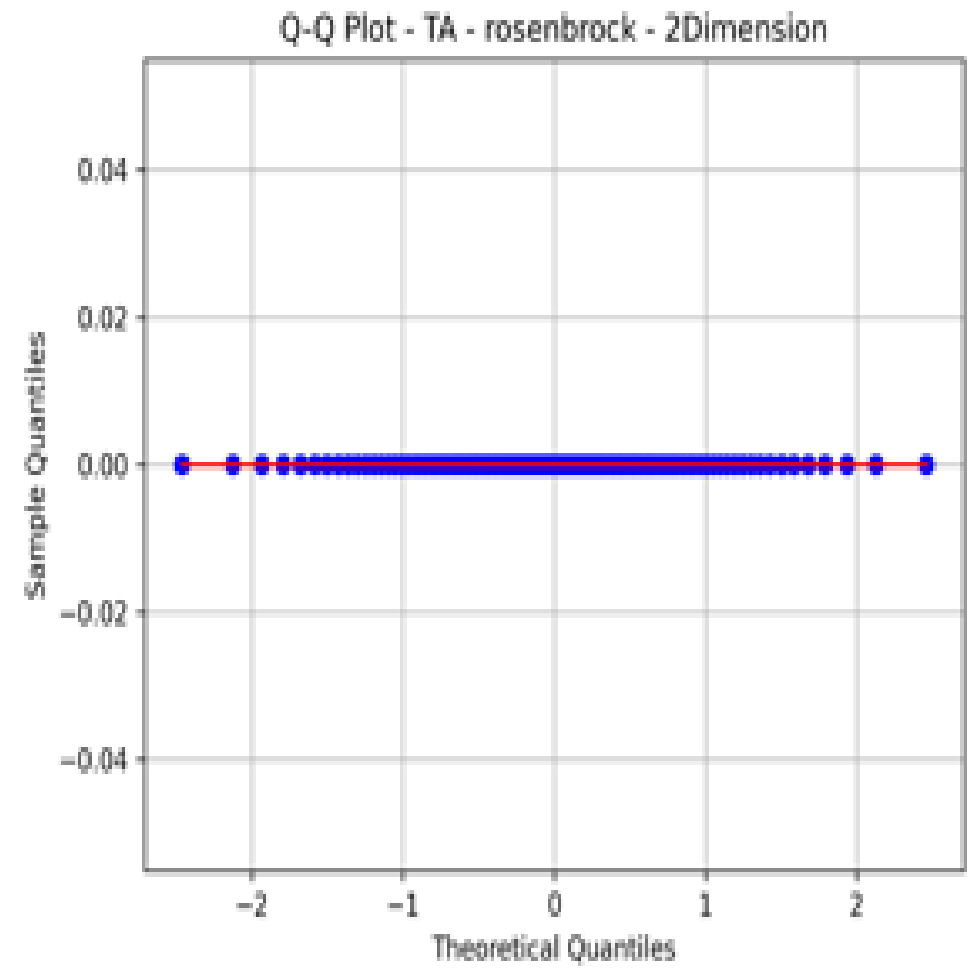
Q-Q Plot - TA - rosenbrock - 2Dimension
Sample Quantiles
Theoretical Quantiles
0.04
0.02
0.00
−0.02
−0.04
−2
−1
0
1
2

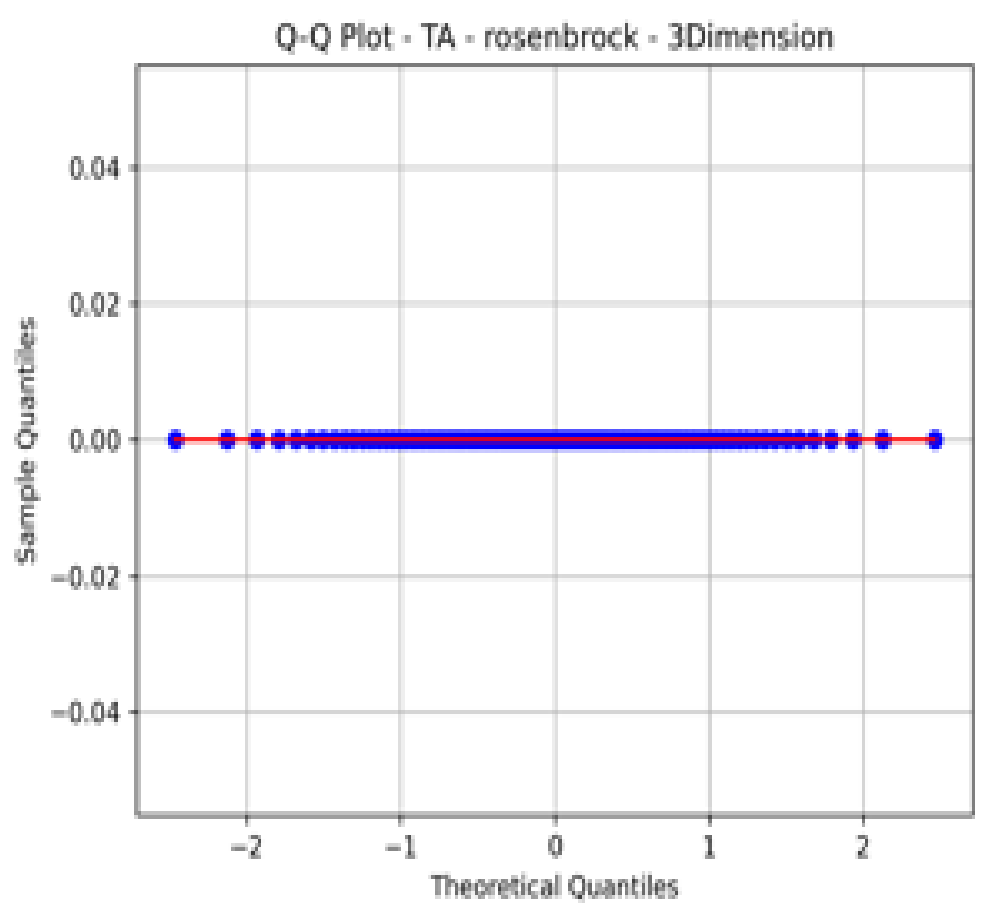
Q-Q Plot - TA - rosenbrock - 3Dimension
Sample Quantiles
Theoretical Quantiles
0.04
0.02
0.00
−0.02
−0.04
−2
−1
0
1
2

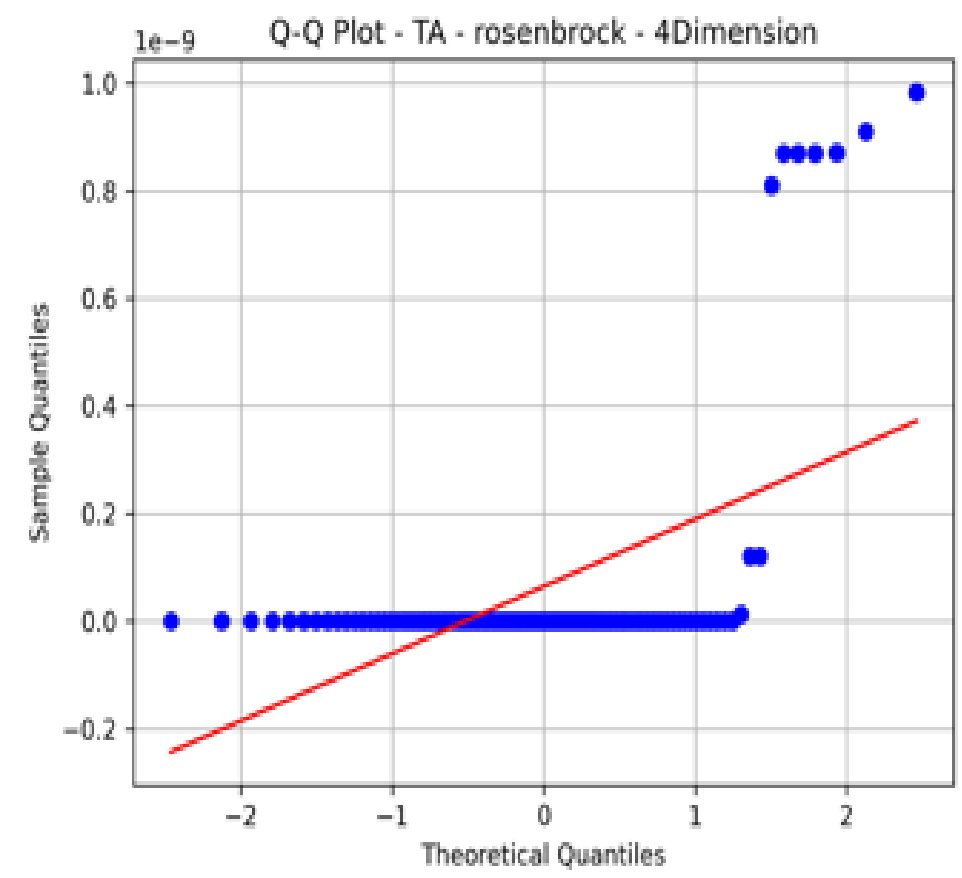
Q-Q Plot - TA - rosenbrock - 4Dimension
1e−9
Sample Quantiles
Theoretical Quantiles
1.0
0.8
0.6
0.4
0.2
0.0
−0.2
−2
−1
0
1
2

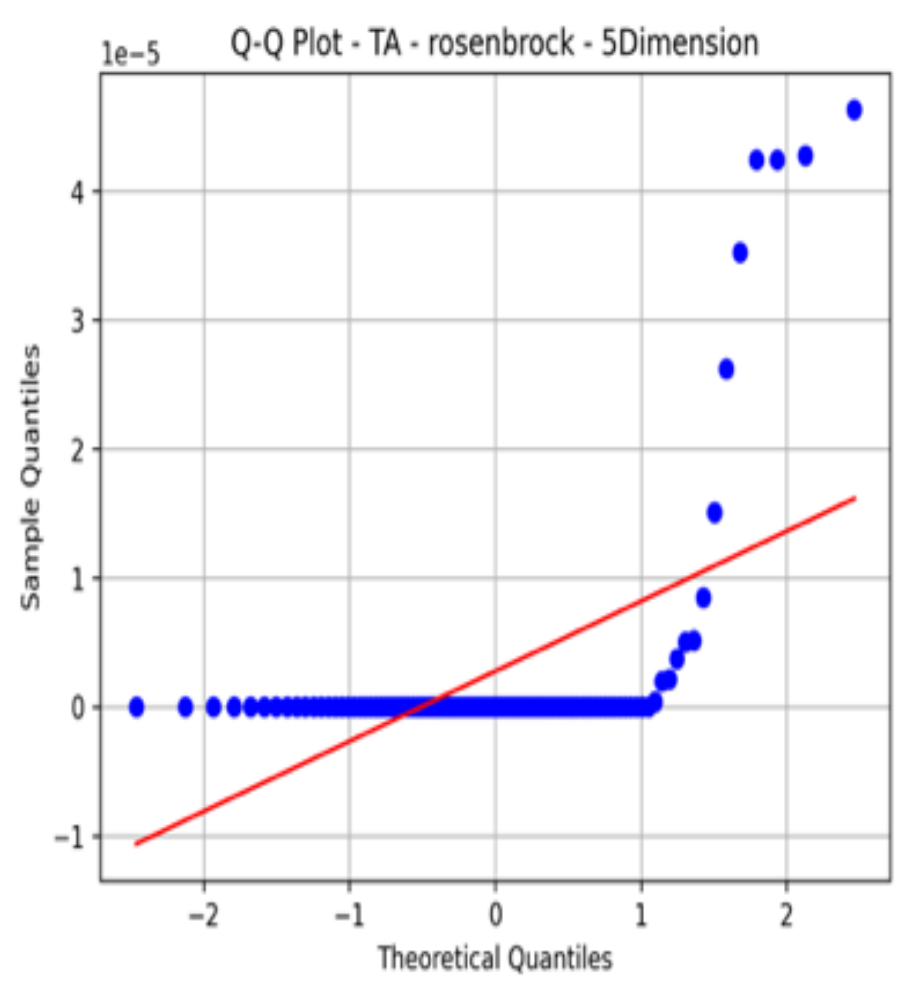
Q-Q Plot - TA - rosenbrock - 5Dimension
1e−5
Sample Quantiles
Theoretical Quantiles
4
3
2
1
0
−1
−2
−1
0
1
2

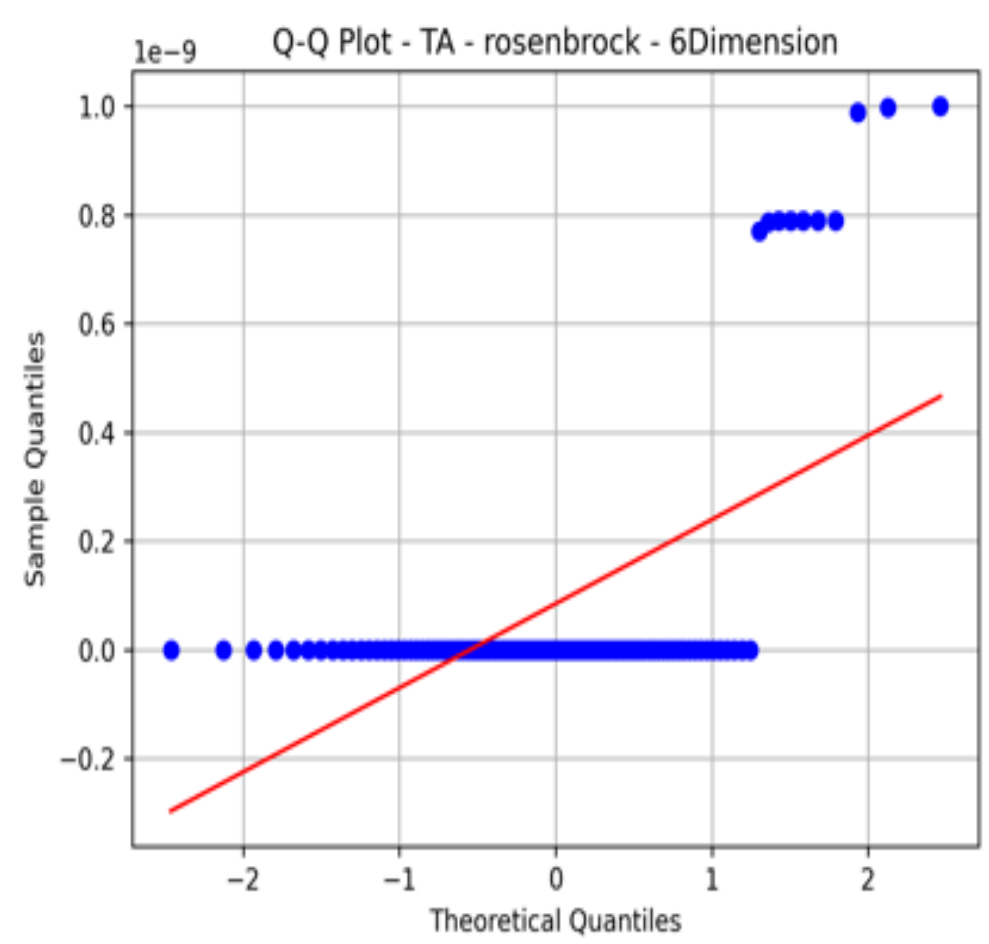
Q-Q Plot - TA - rosenbrock - 6Dimension
1e−9
Sample Quantiles
Theoretical Quantiles
1.0
0.8
0.6
0.4
0.2
0.0
−0.2
−2
−1
0
1
2

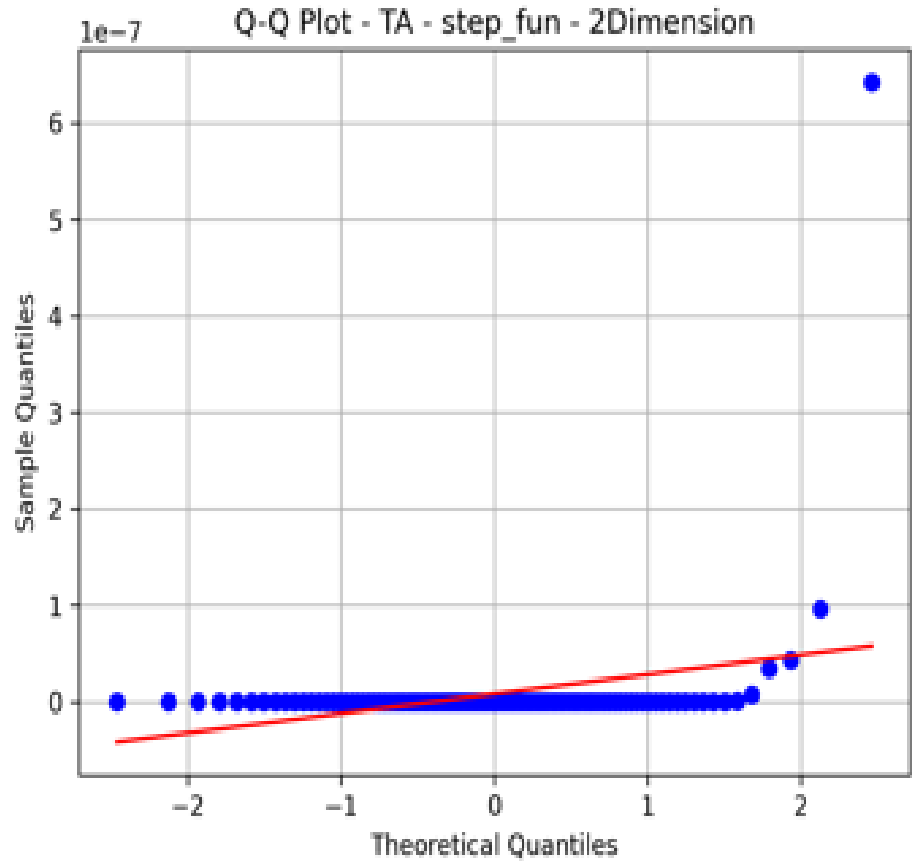
Q-Q Plot - TA - step_fun - 2Dimension
1e−7
Sample Quantiles
Theoretical Quantiles

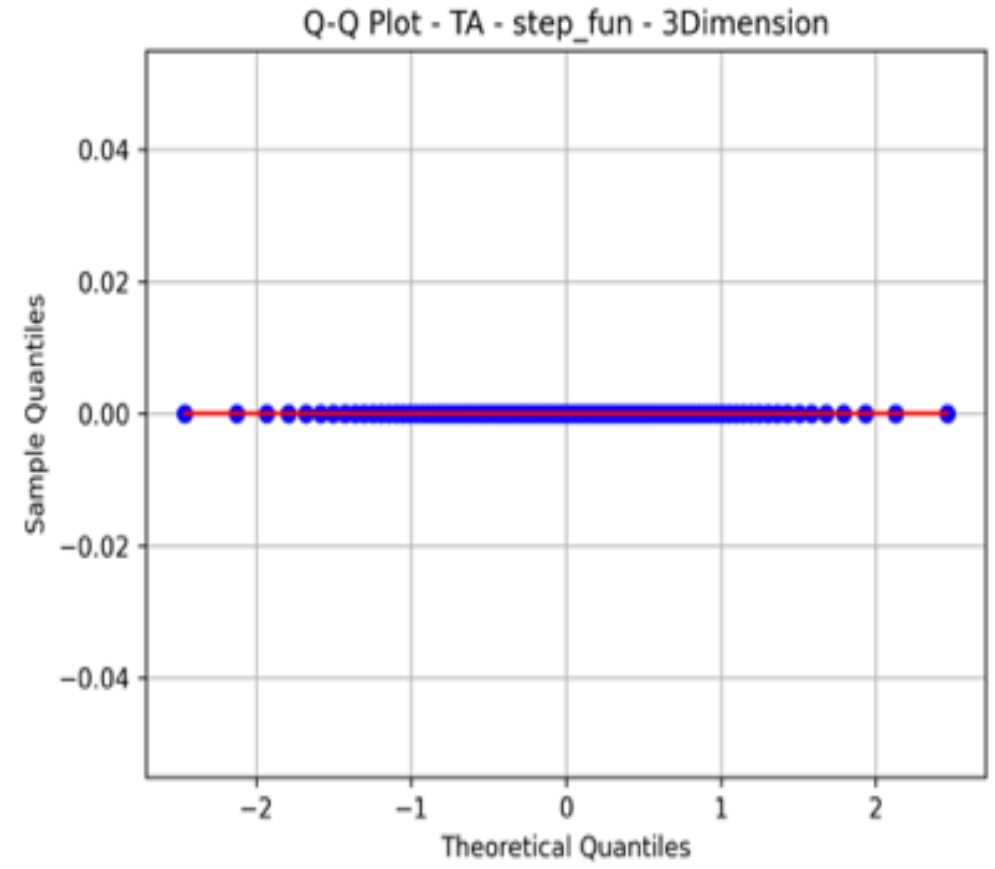
Q-Q Plot - TA - step_fun - 3Dimension
Sample Quantiles
Theoretical Quantiles

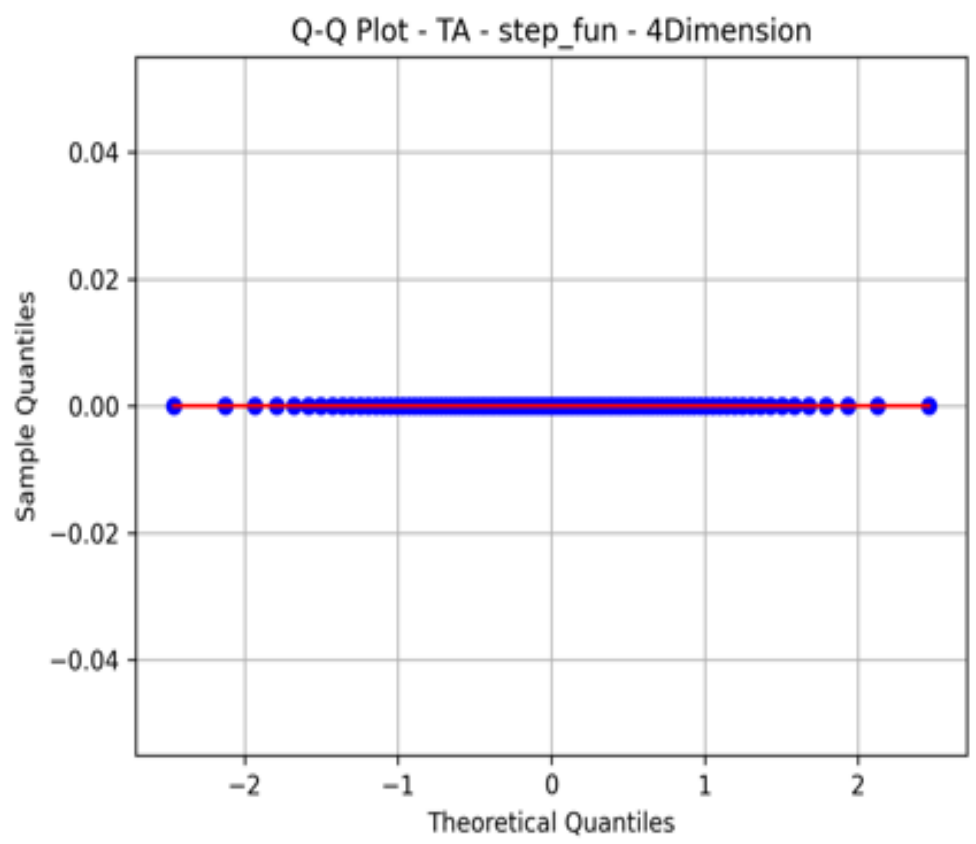
Q-Q Plot - TA - step_fun - 4Dimension
Sample Quantiles
Theoretical Quantiles

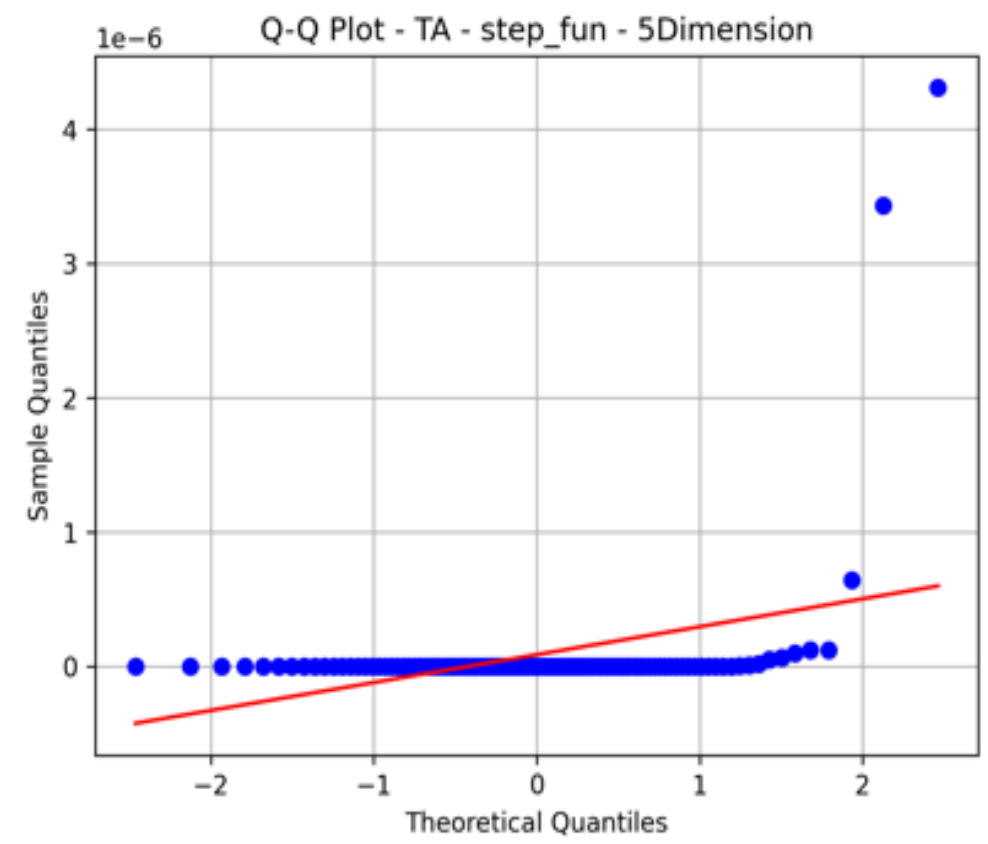
Q-Q Plot - TA - step_fun - 5Dimension
1e−6
Sample Quantiles
Theoretical Quantiles

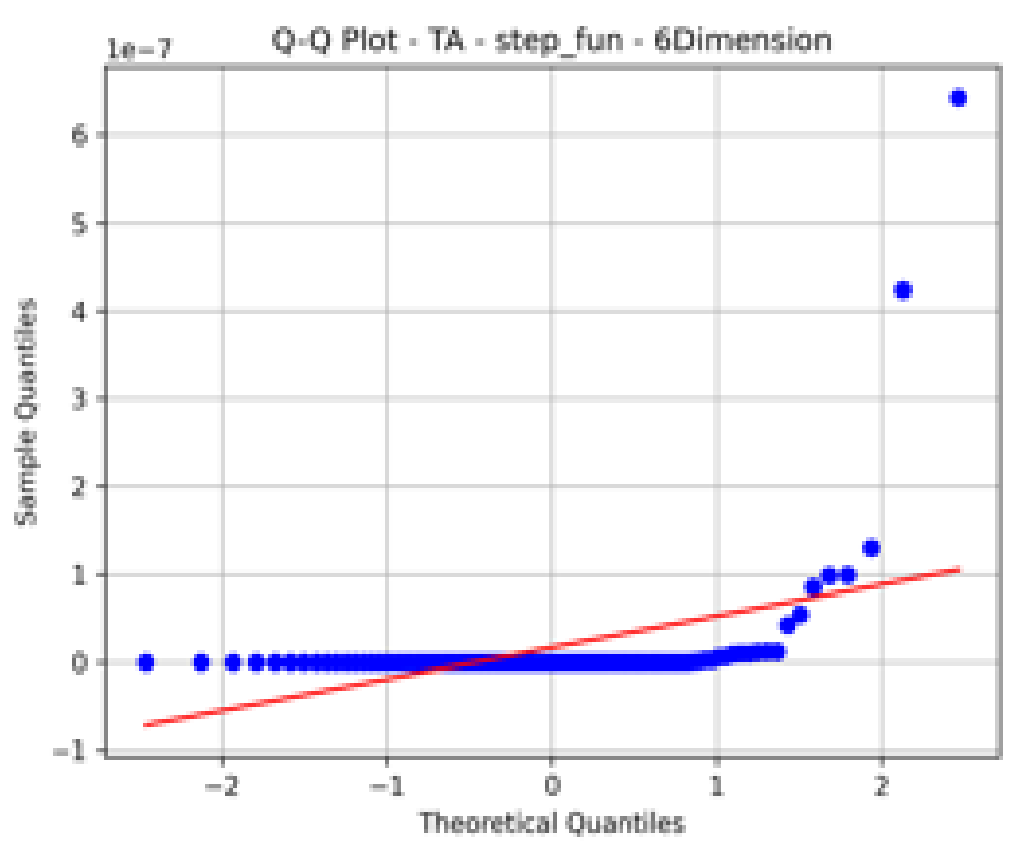
Q-Q Plot - TA - step_fun - 6Dimension
1e−7
Sample Quantiles
Theoretical Quantiles

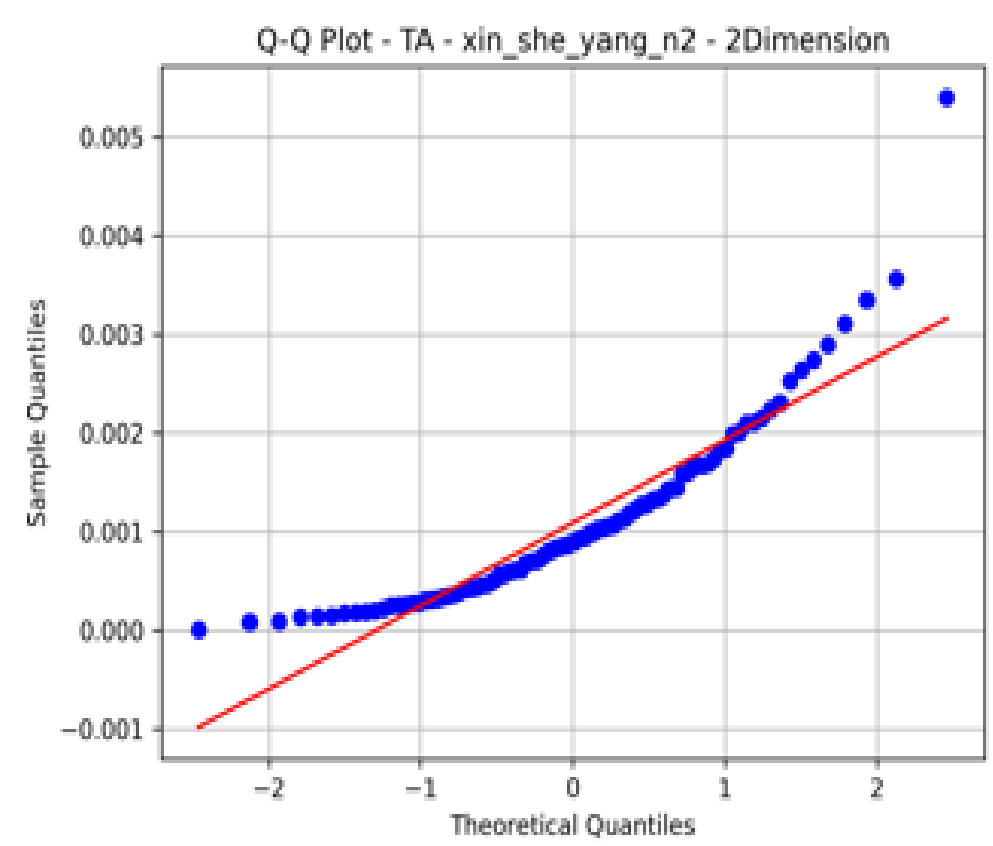
Q-Q Plot - TA - xin_she_yang_n2 - 2Dimension
Sample Quantiles
Theoretical Quantiles

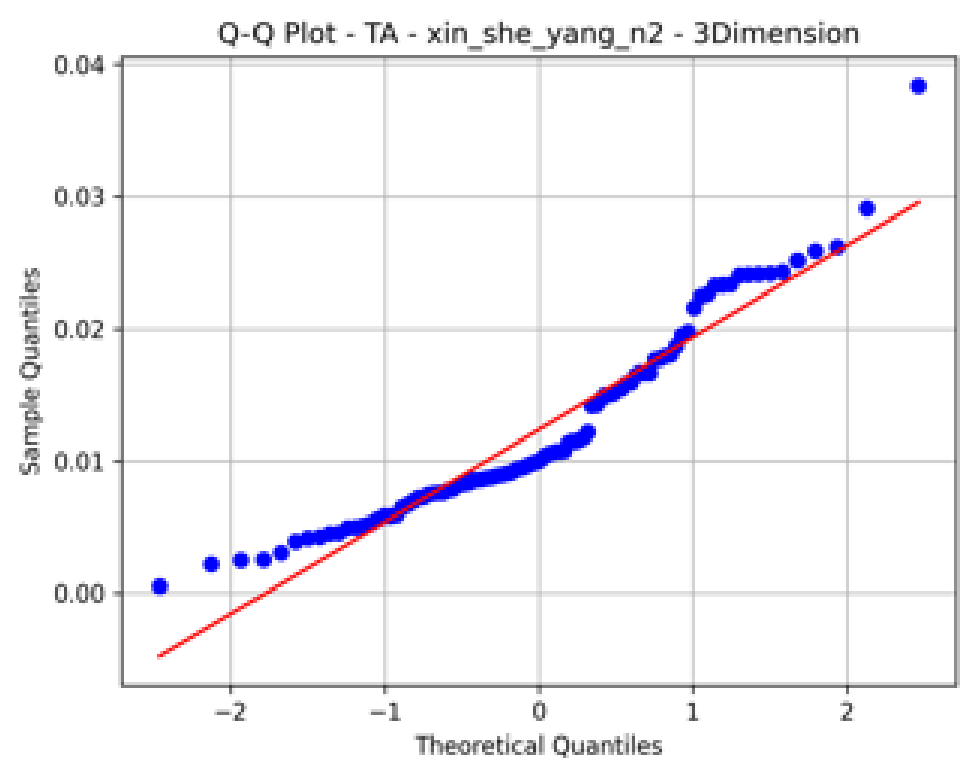

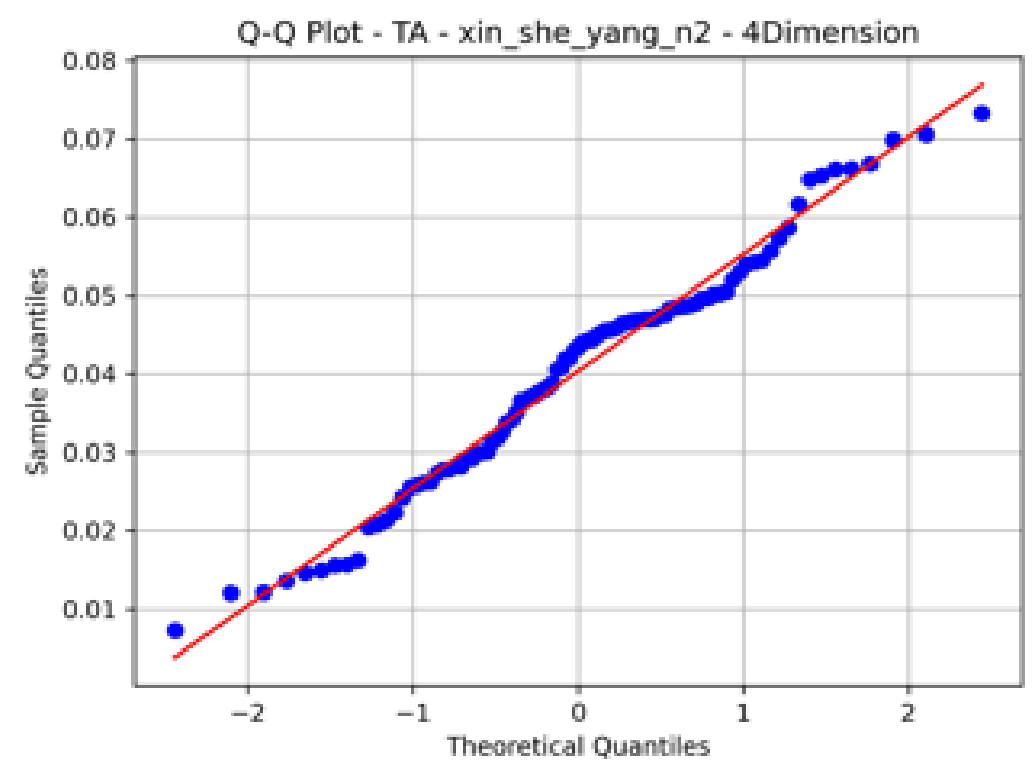

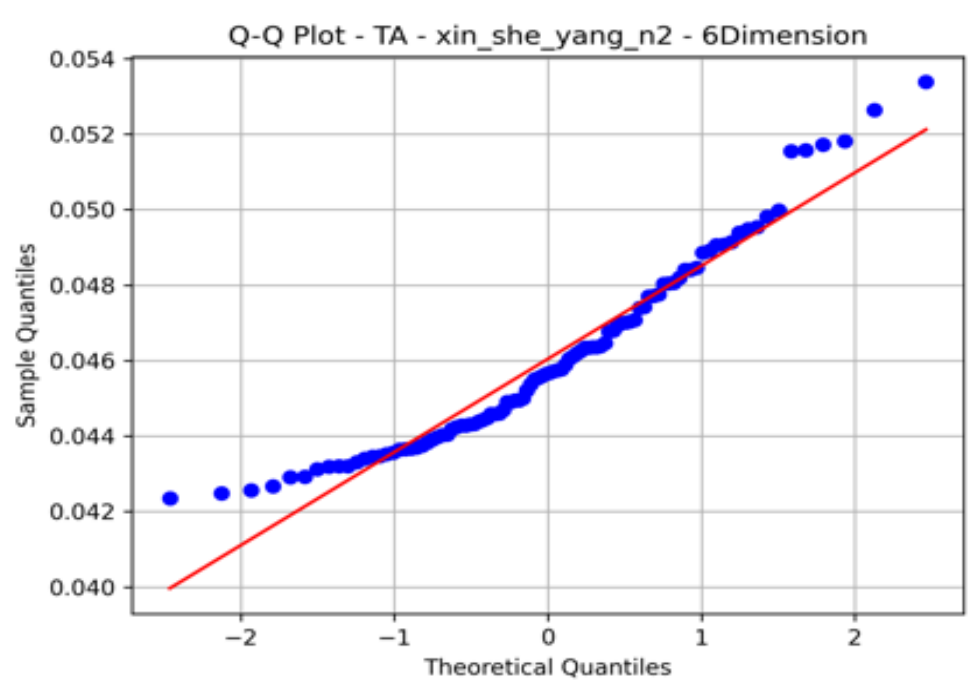

<u>PSO Q-Q Plots</u>

Non-Continuous functions QQ Plots

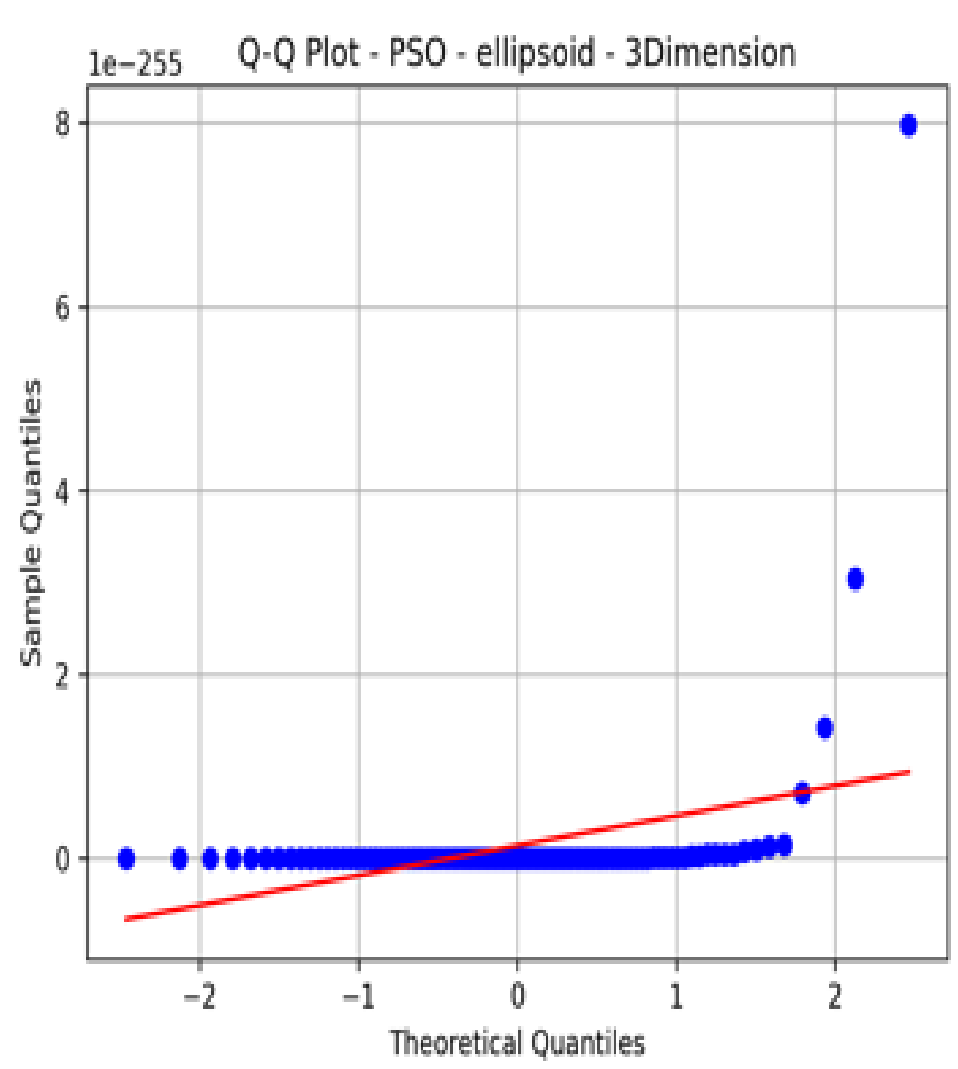

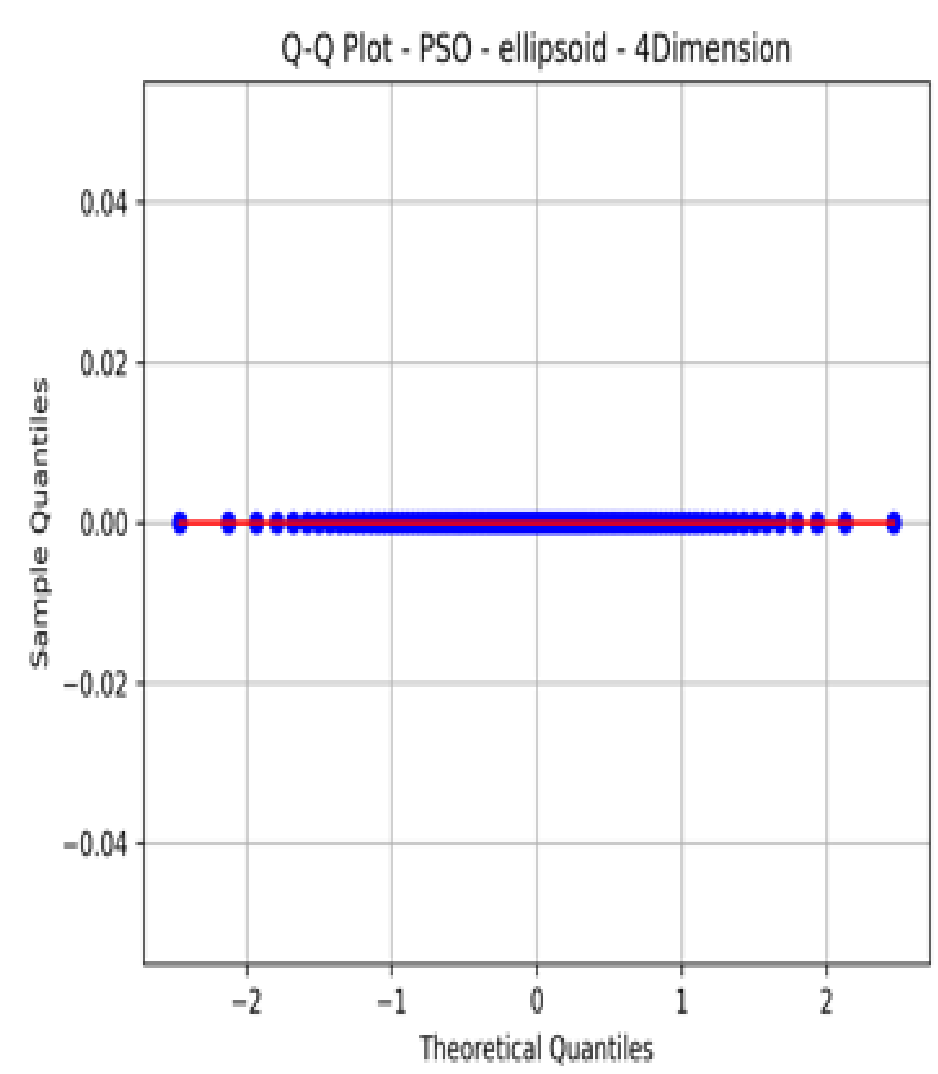

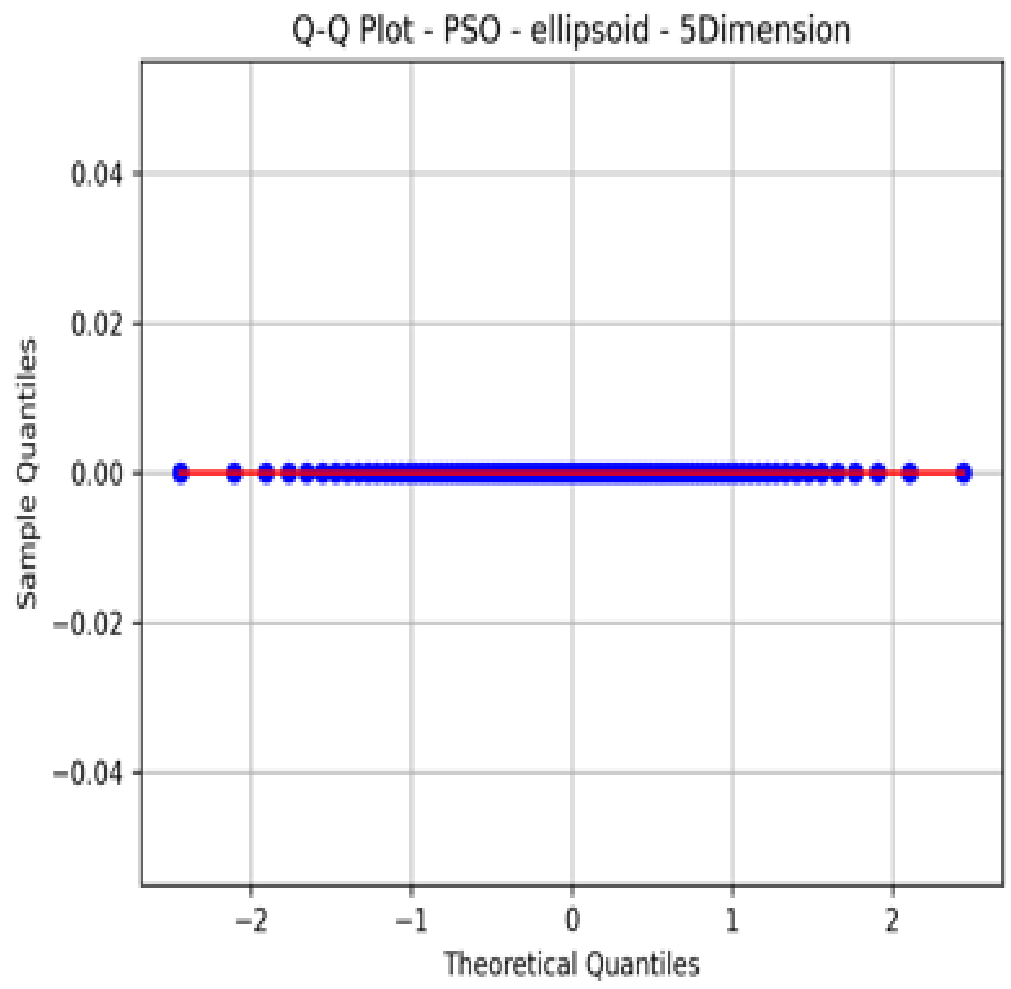
Q-Q Plot - PSO - ellipsoid - 5Dimension
Sample Quantiles
0.04
0.02
0.00
−0.02
−0.04
−2
−1
0
1
2
Theoretical Quantiles

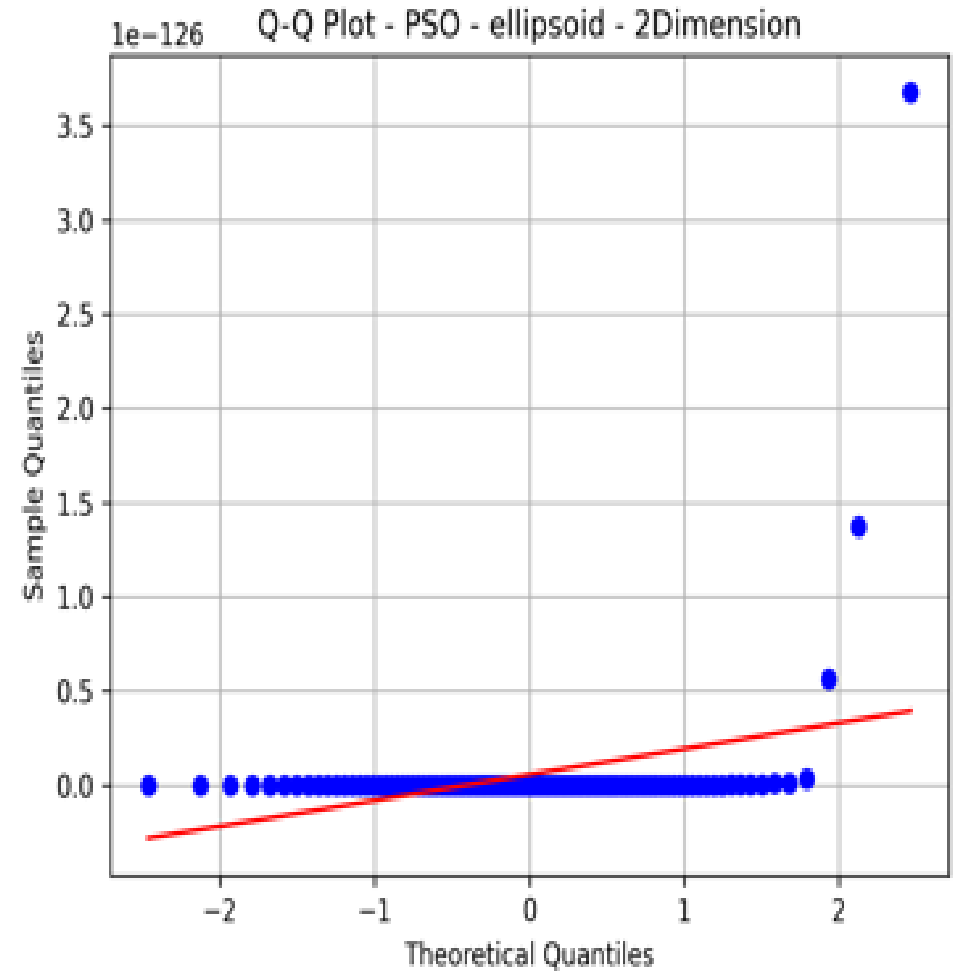
1e−126
Q-Q Plot - PSO - ellipsoid - 2Dimension
Sample Quantiles
3.5
3.0
2.5
2.0
1.5
1.0
0.5
0.0
−2
−1
0
1
2
Theoretical Quantiles

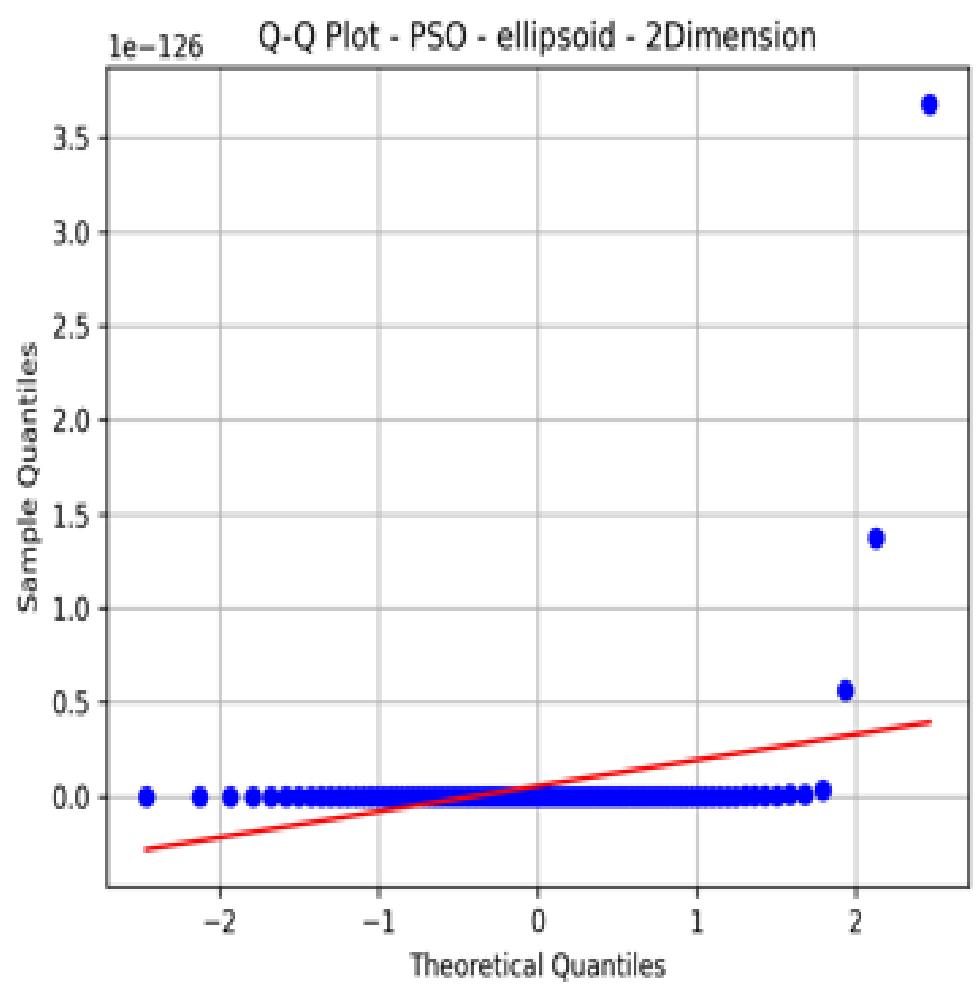
1e−126
Q-Q Plot - PSO - ellipsoid - 2Dimension
Sample Quantiles
3.5
3.0
2.5
2.0
1.5
1.0
0.5
0.0
−2
−1
0
1
2
Theoretical Quantiles

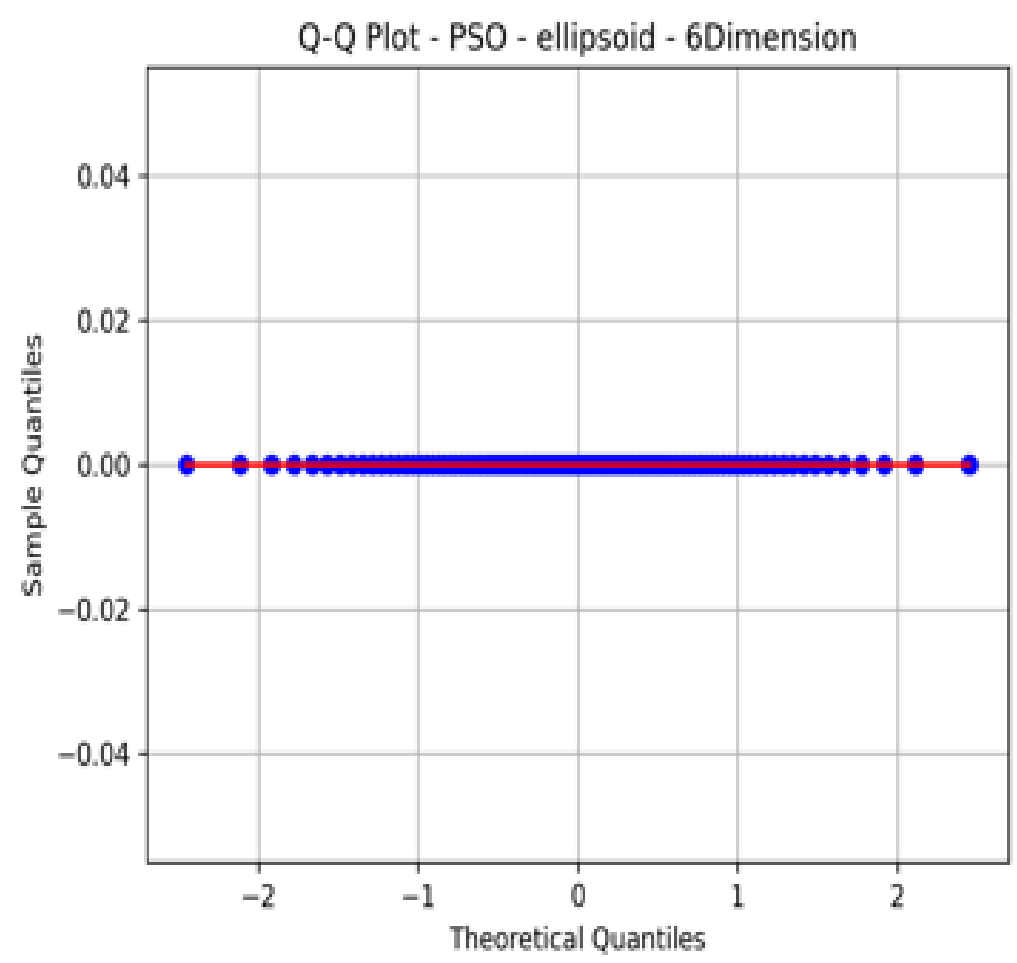
Q-Q Plot - PSO - ellipsoid - 6Dimension
Sample Quantiles
0.04
0.02
0.00
−0.02
−0.04
−2
−1
0
1
2
Theoretical Quantiles

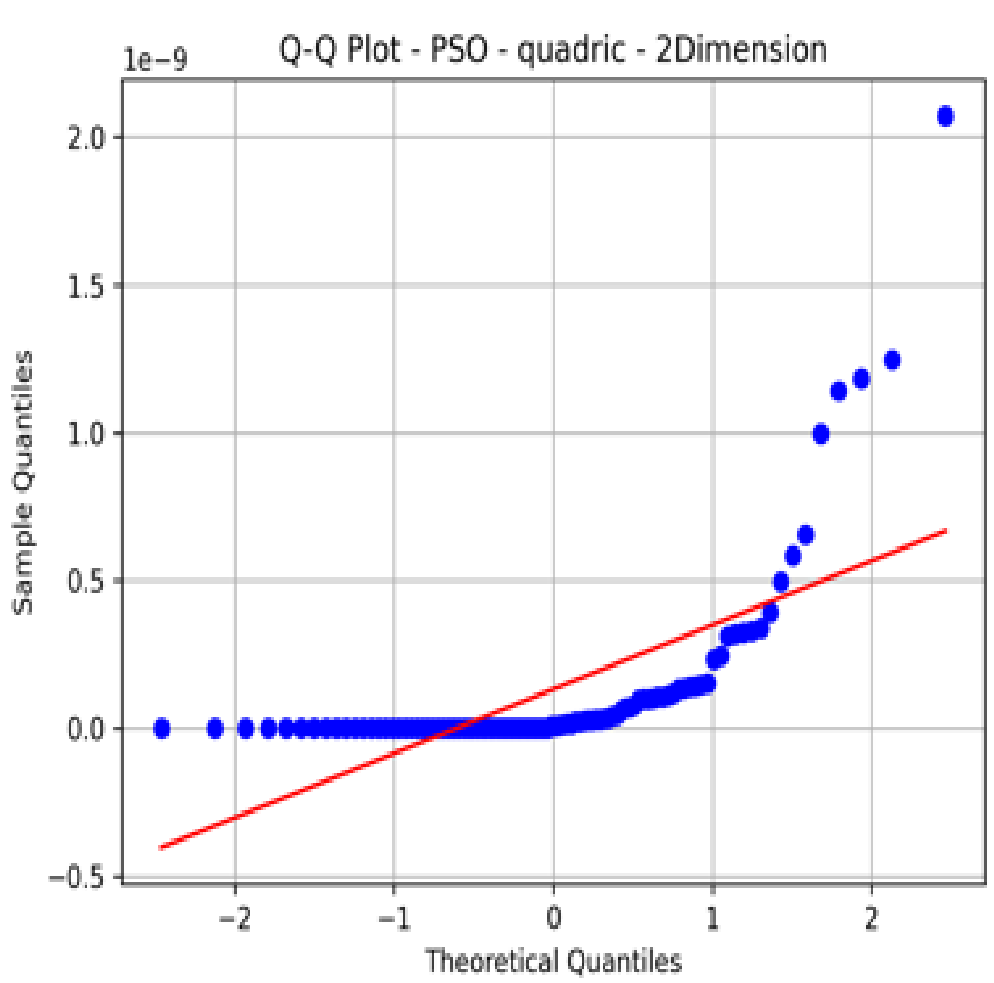
1e−9
Q-Q Plot - PSO - quadric - 2Dimension
Sample Quantiles
2.0
1.5
1.0
0.5
0.0
−0.5
−2
−1
0
1
2
Theoretical Quantiles

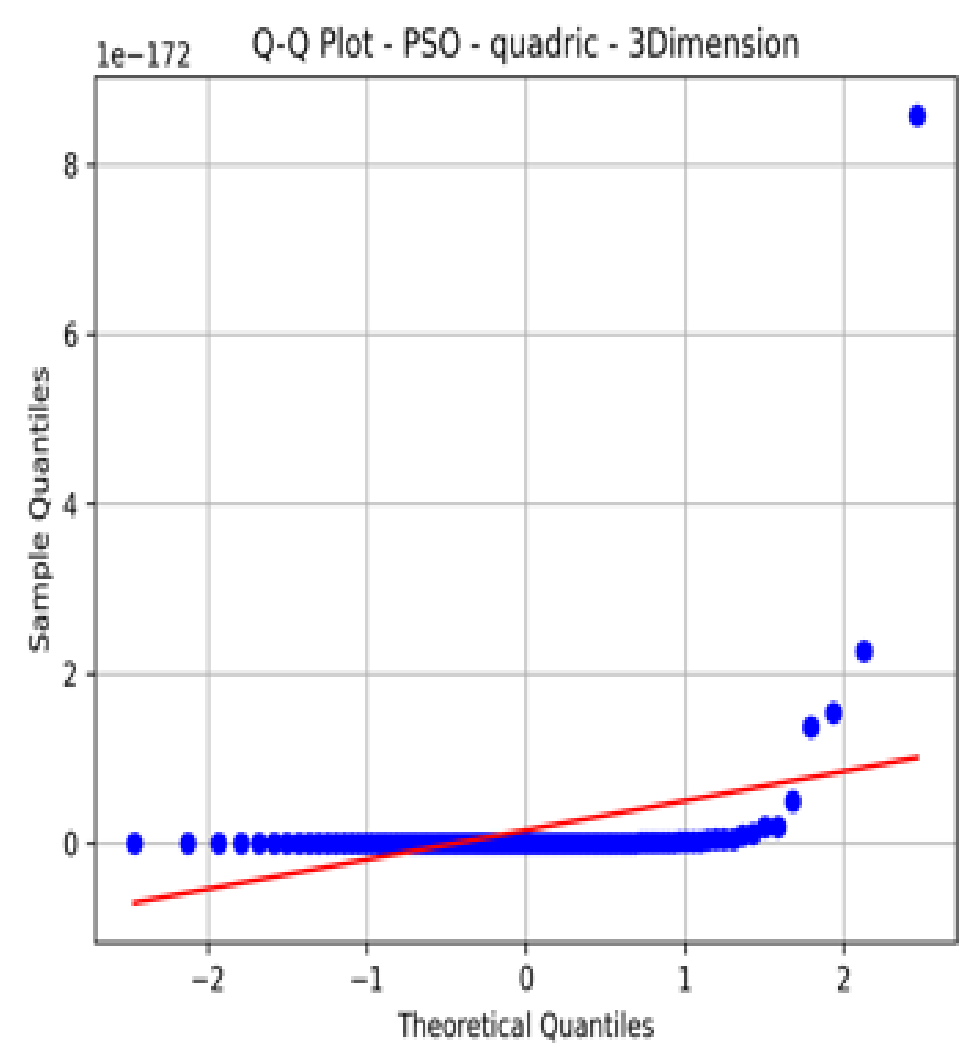
1e−172
Q-Q Plot - PSO - quadric - 3Dimension
Sample Quantiles
8
6
4
2
0
−2
−1
0
1
2
Theoretical Quantiles

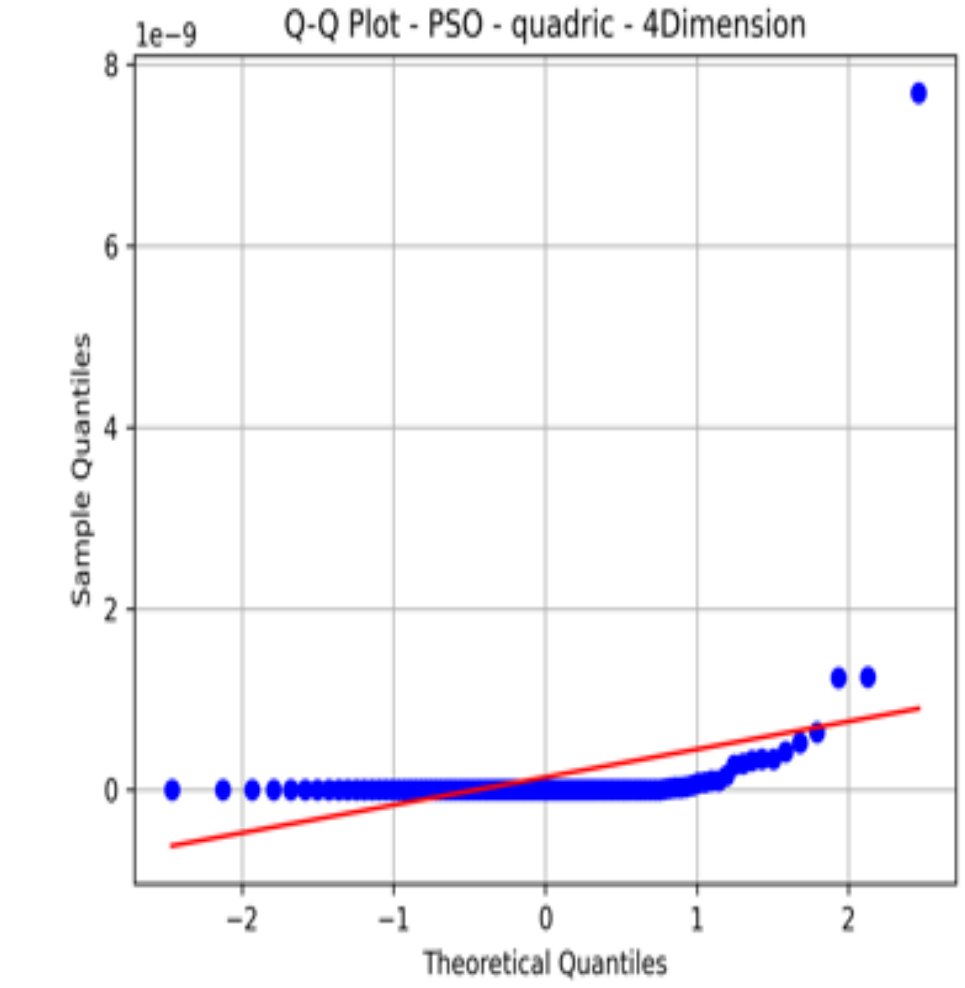
Q-Q Plot - PSO - quadric - 4Dimension
1e−9
Sample Quantiles
Theoretical Quantiles

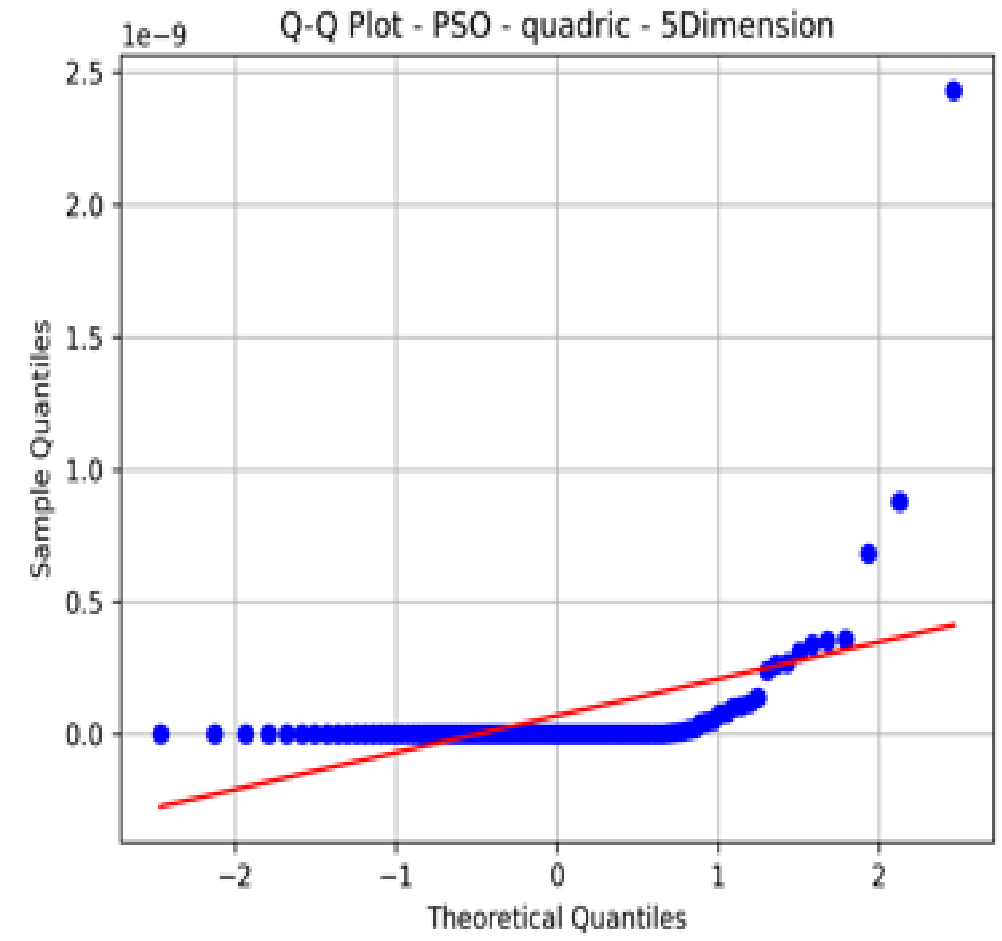
Q-Q Plot - PSO - quadric - 5Dimension
1e−9
Sample Quantiles
Theoretical Quantiles

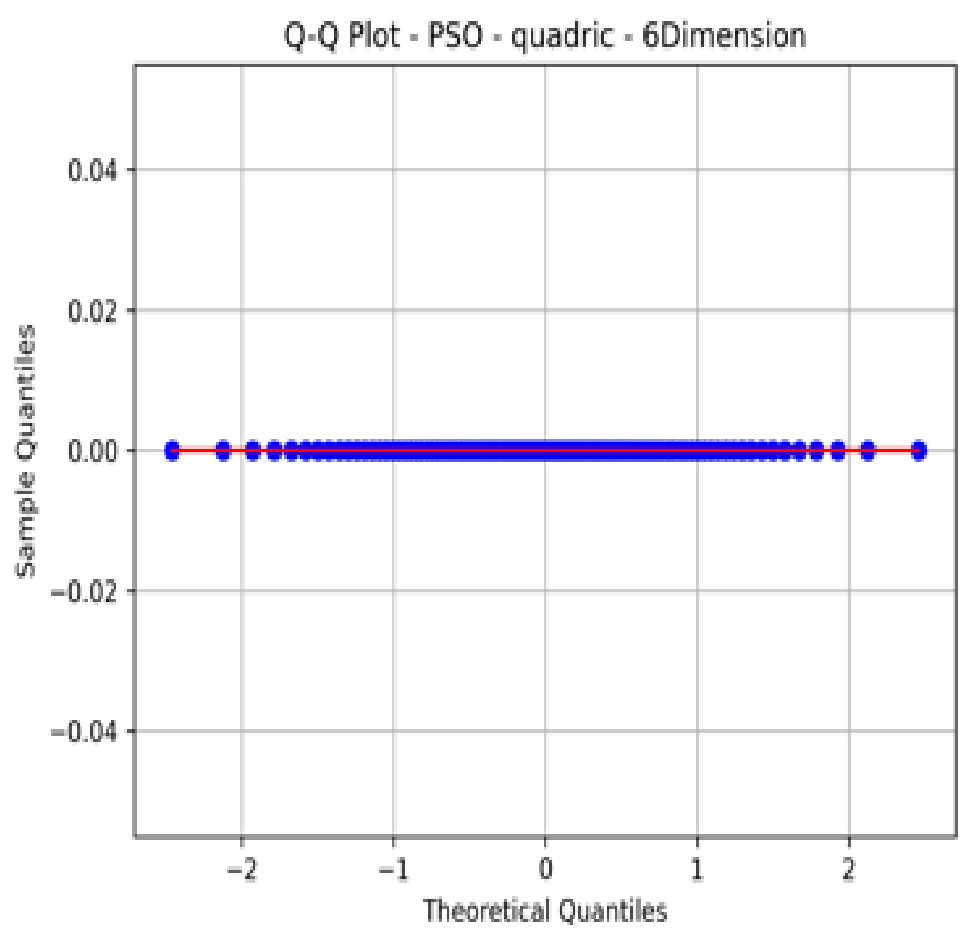
Q-Q Plot - PSO - quadric - 6Dimension
Sample Quantiles
Theoretical Quantiles

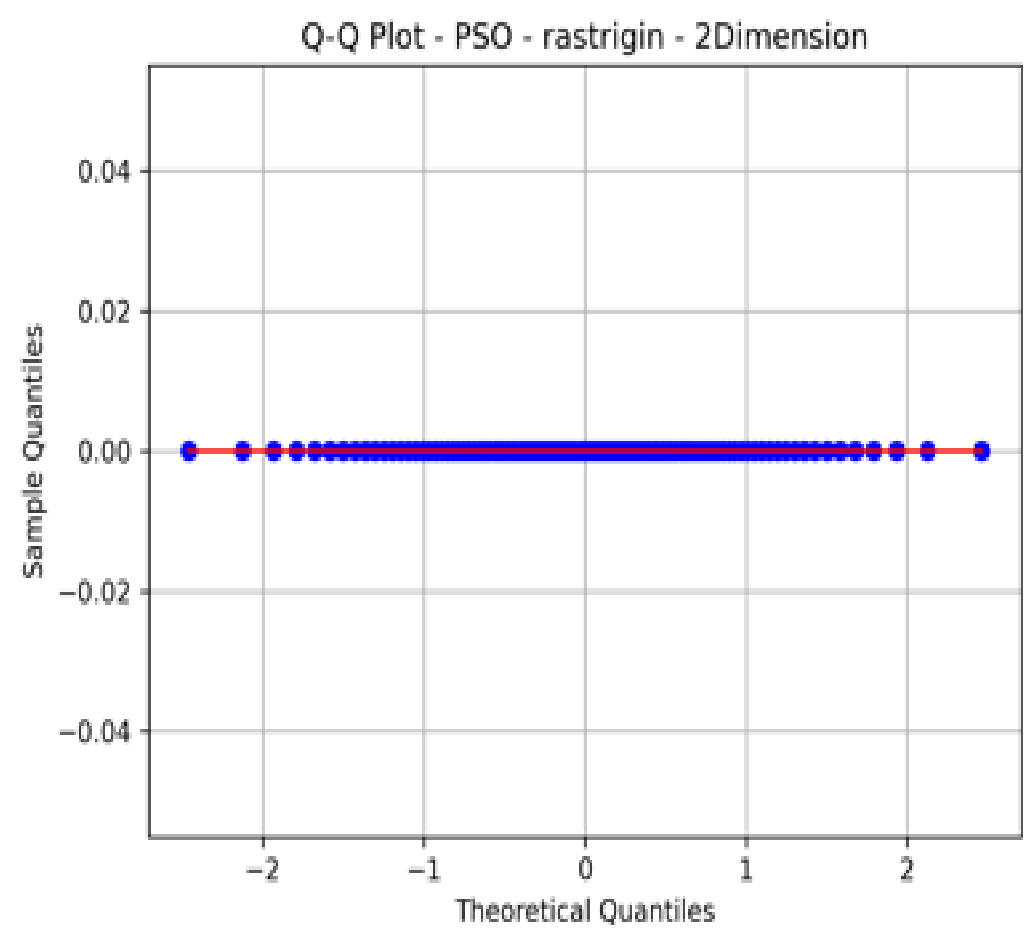
Q-Q Plot - PSO - rastrigin - 2Dimension
Sample Quantiles
Theoretical Quantiles

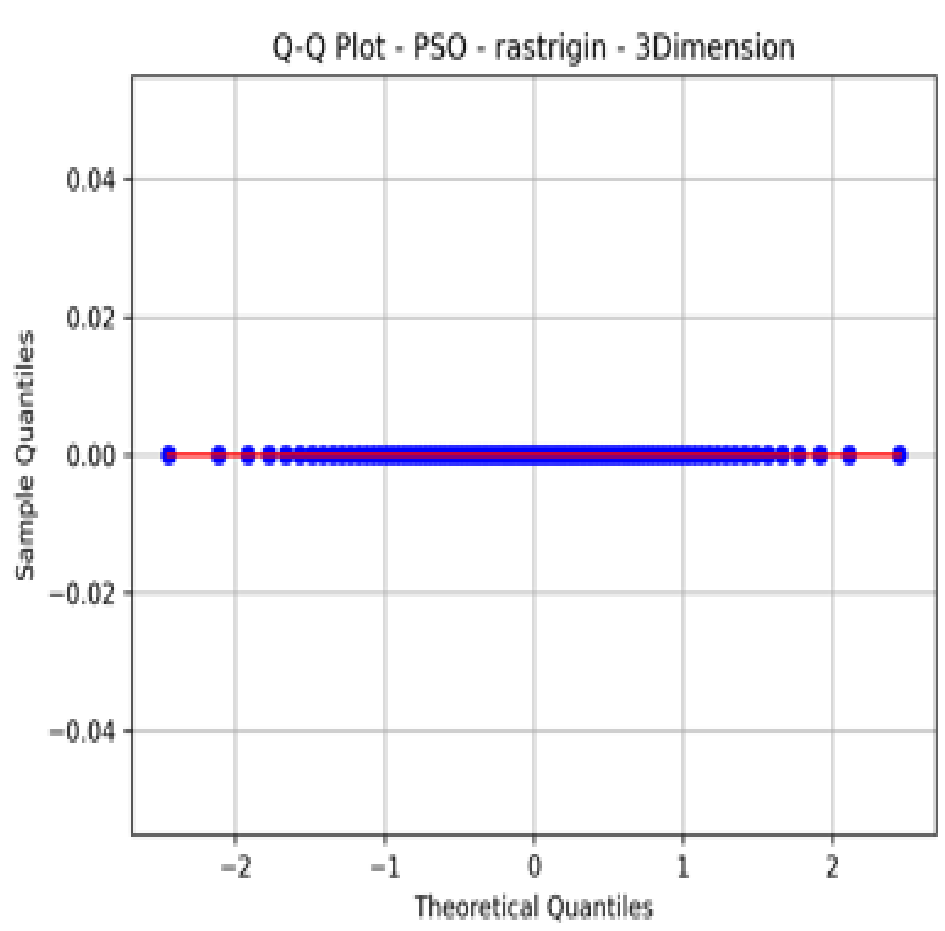
Q-Q Plot - PSO - rastrigin - 3Dimension
Sample Quantiles
Theoretical Quantiles

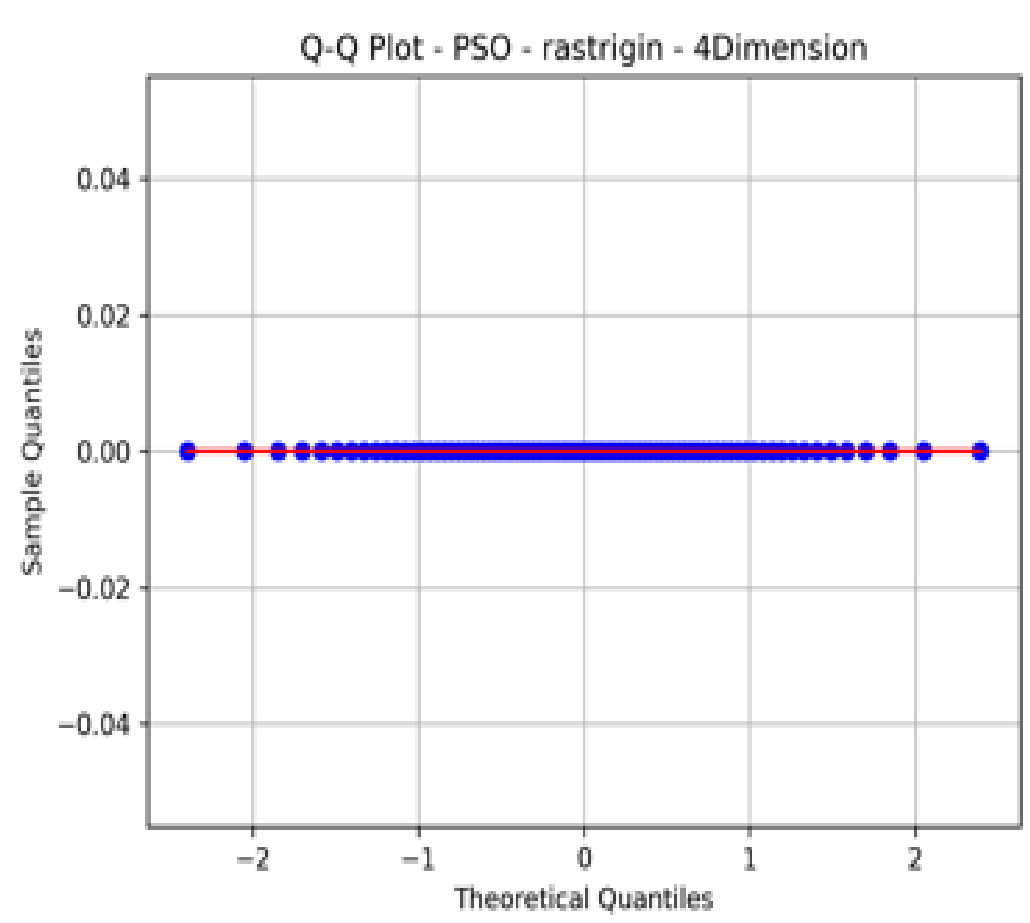
Q-Q Plot - PSO - rastrigin - 4Dimension
Sample Quantiles
Theoretical Quantiles

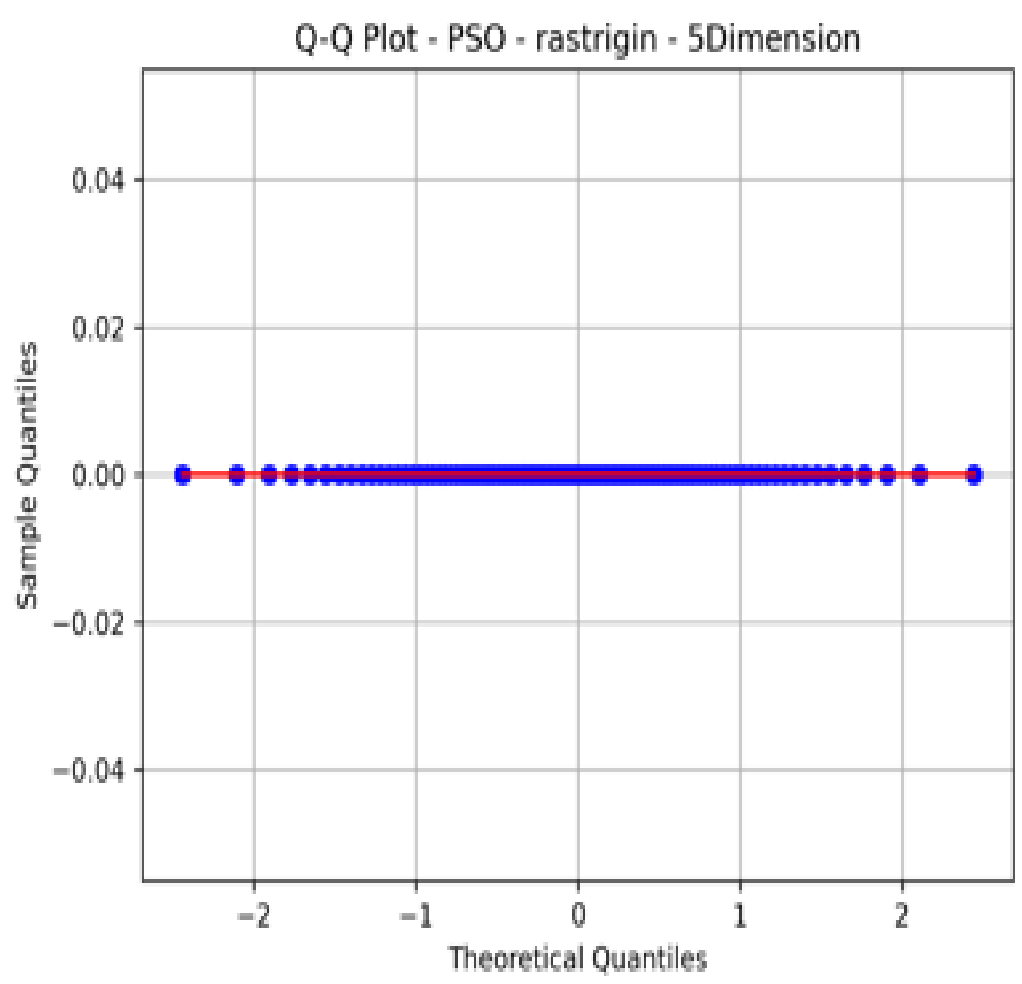
Q-Q Plot - PSO - rastrigin - 5Dimension
0.04
0.02
0.00
−0.02
−0.04
Sample Quantiles
−2
−1
0
1
2
Theoretical Quantiles

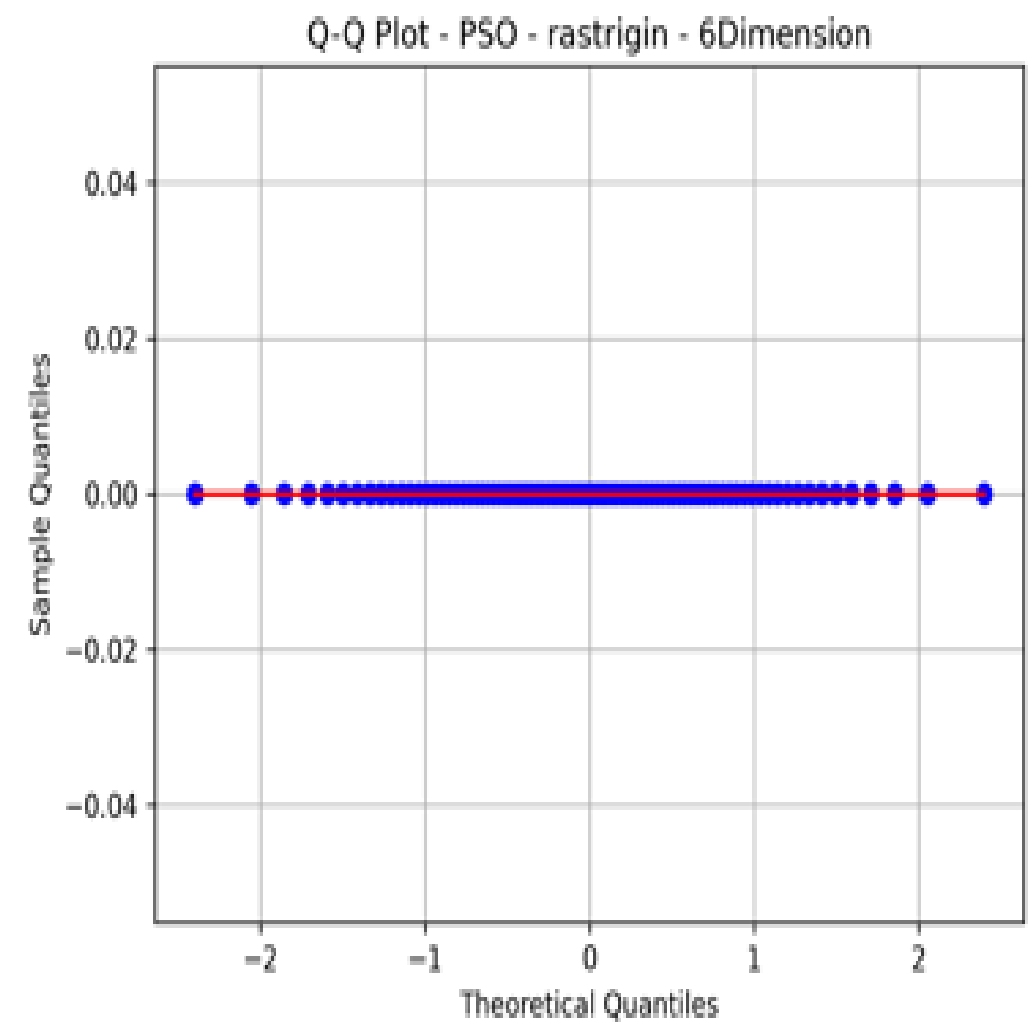
Q-Q Plot - PSO - rastrigin - 6Dimension
0.04
0.02
0.00
−0.02
−0.04
Sample Quantiles
−2
−1
0
1
2
Theoretical Quantiles

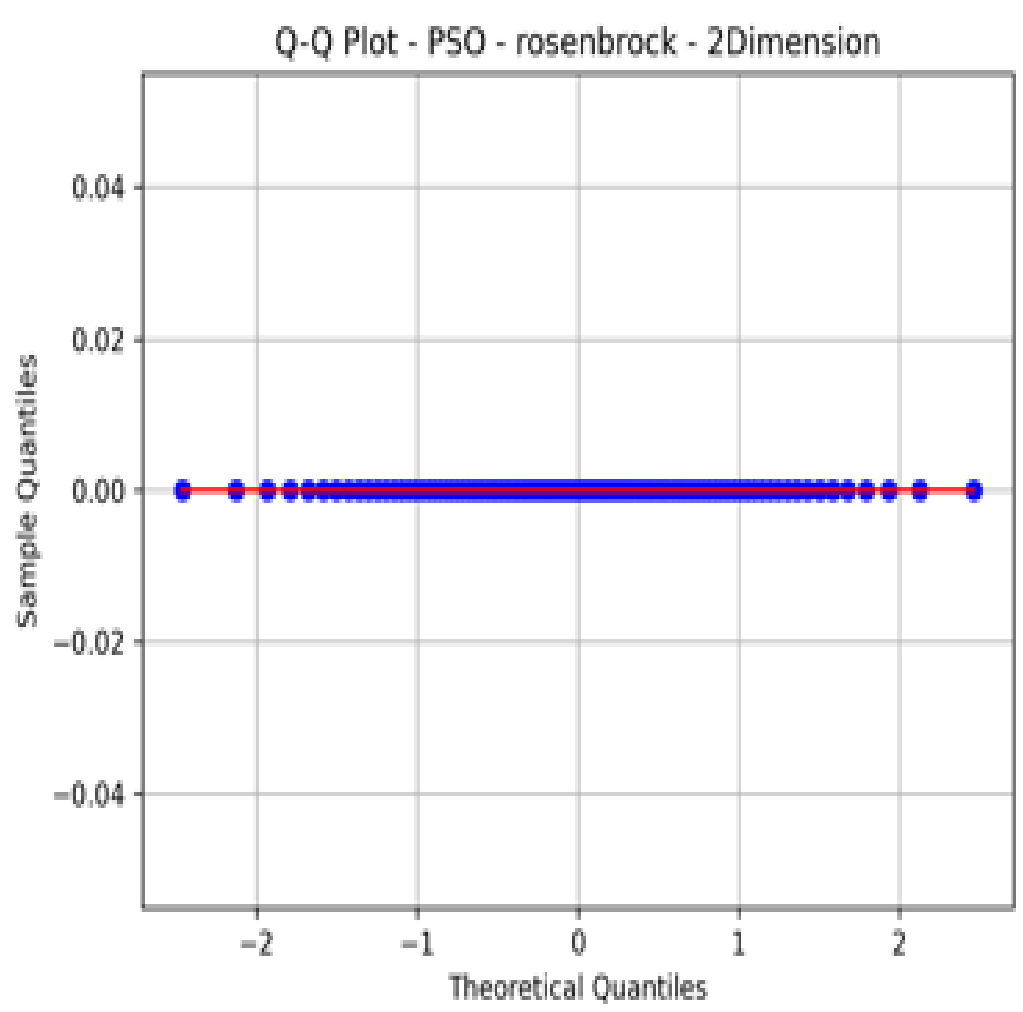
Q-Q Plot - PSO - rosenbrock - 2Dimension
0.04
0.02
0.00
−0.02
−0.04
Sample Quantiles
−2
−1
0
1
2
Theoretical Quantiles

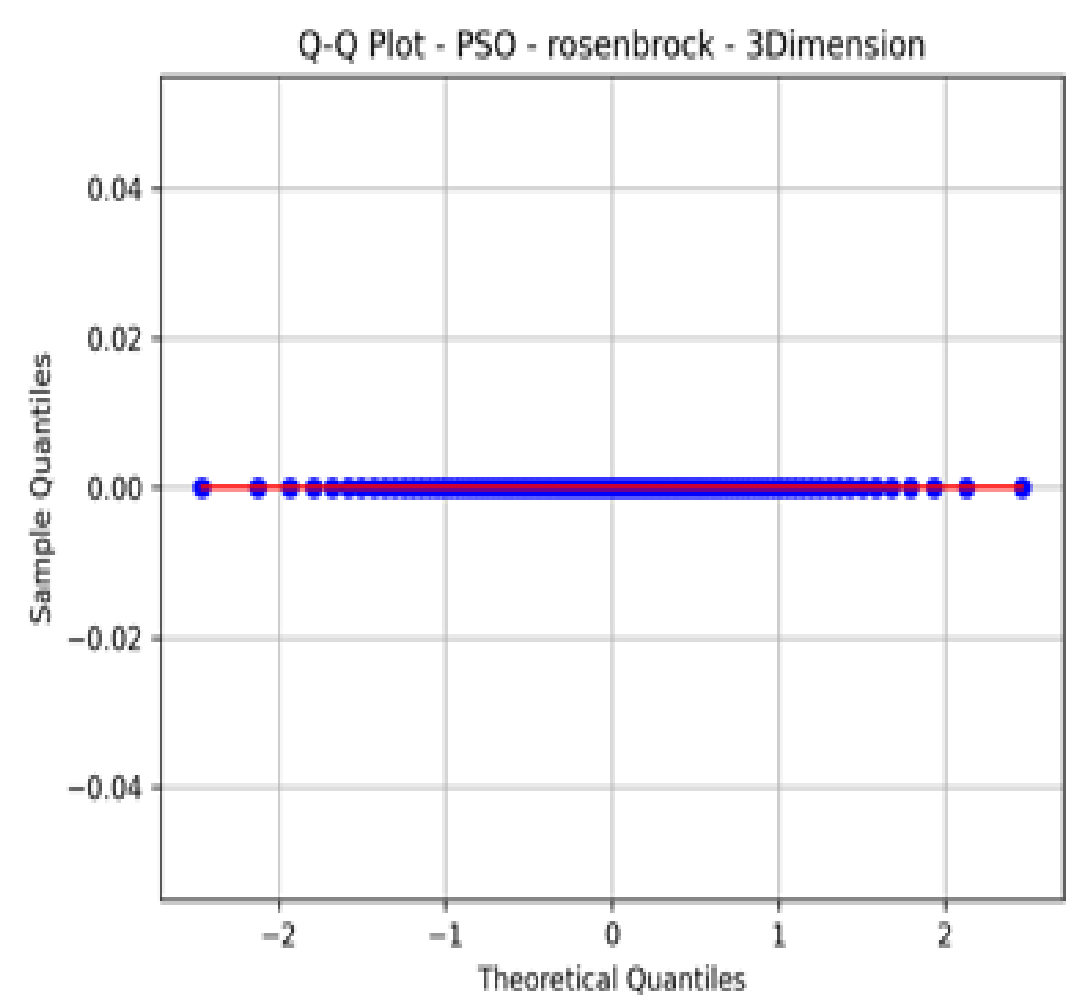
Q-Q Plot - PSO - rosenbrock - 3Dimension
0.04
0.02
0.00
−0.02
−0.04
Sample Quantiles
−2
−1
0
1
2
Theoretical Quantiles

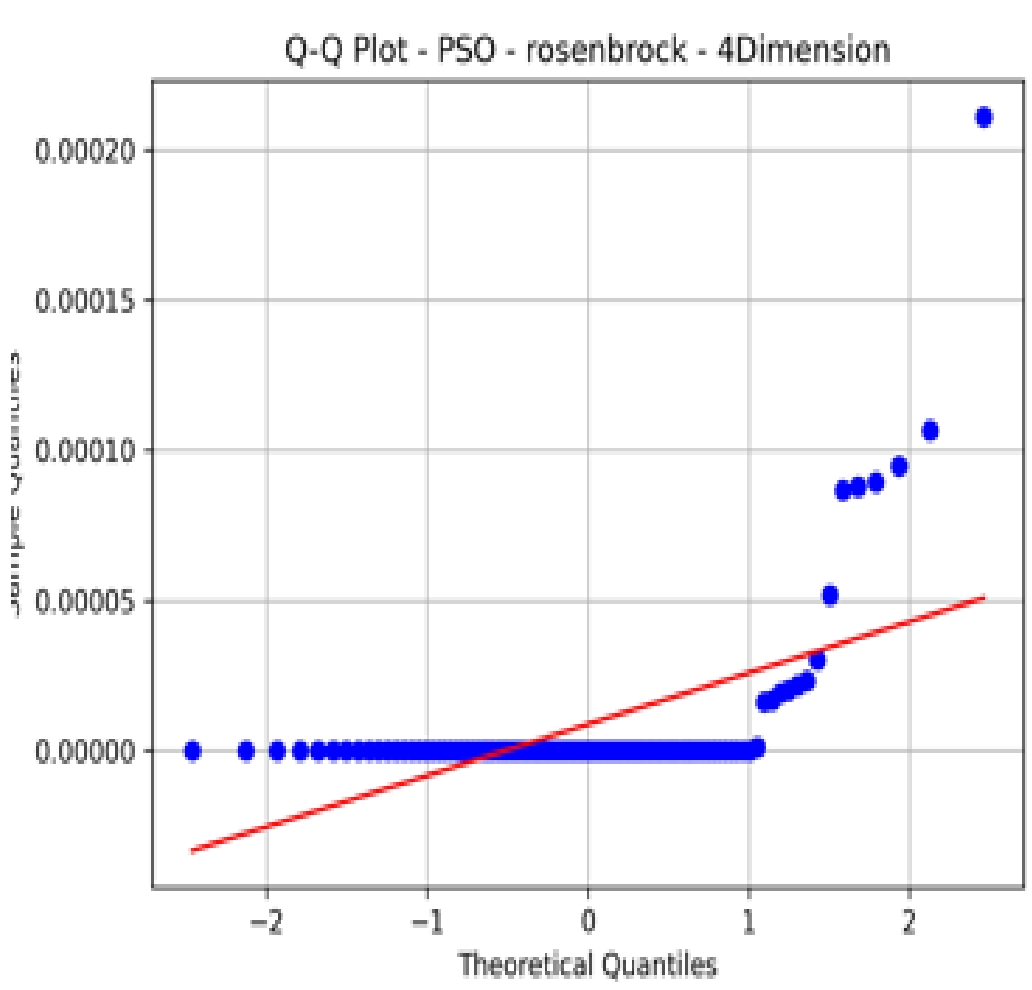
Q-Q Plot - PSO - rosenbrock - 4Dimension
0.00020
0.00015
0.00010
0.00005
0.00000
−2
−1
0
1
2
Theoretical Quantiles

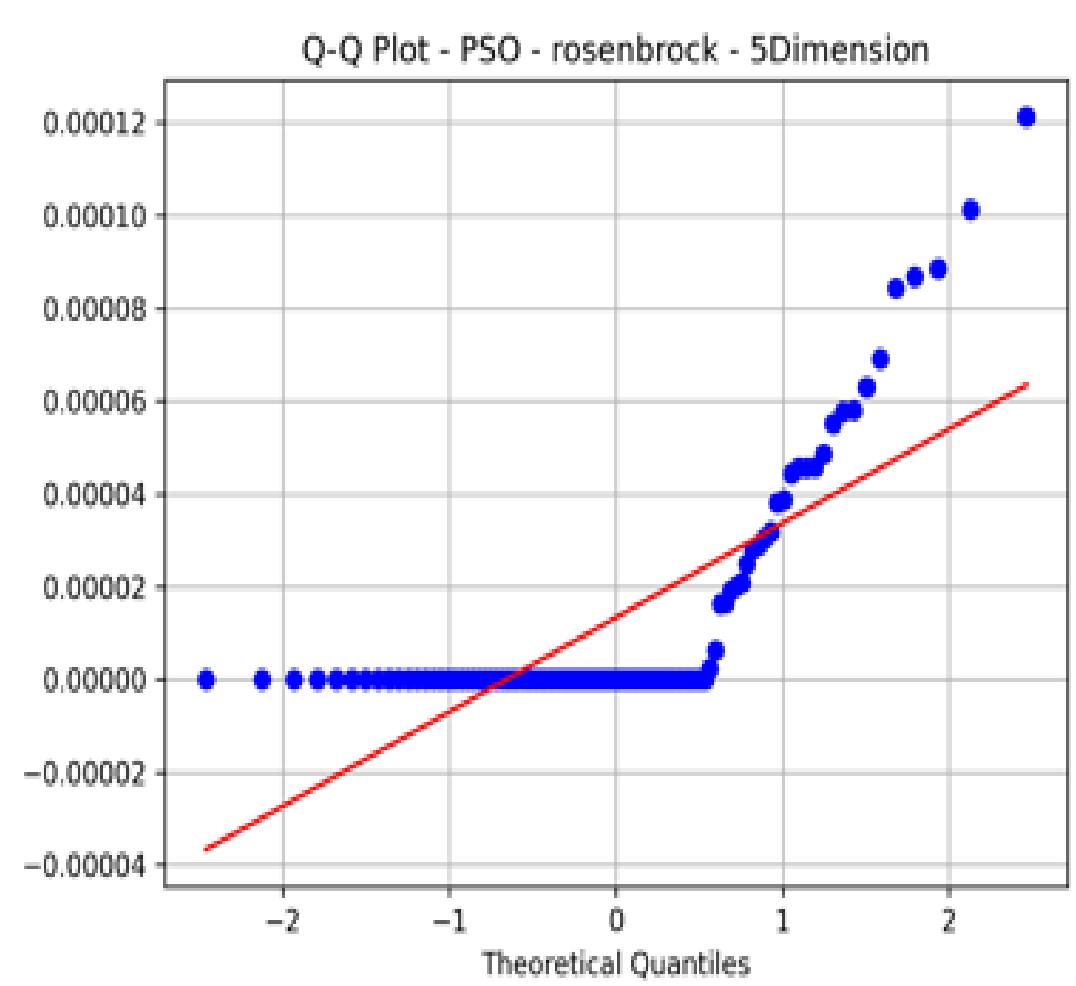
Q-Q Plot - PSO - rosenbrock - 5Dimension
0.00012
0.00010
0.00008
0.00006
0.00004
0.00002
0.00000
−0.00002
−0.00004
−2
−1
0
1
2
Theoretical Quantiles

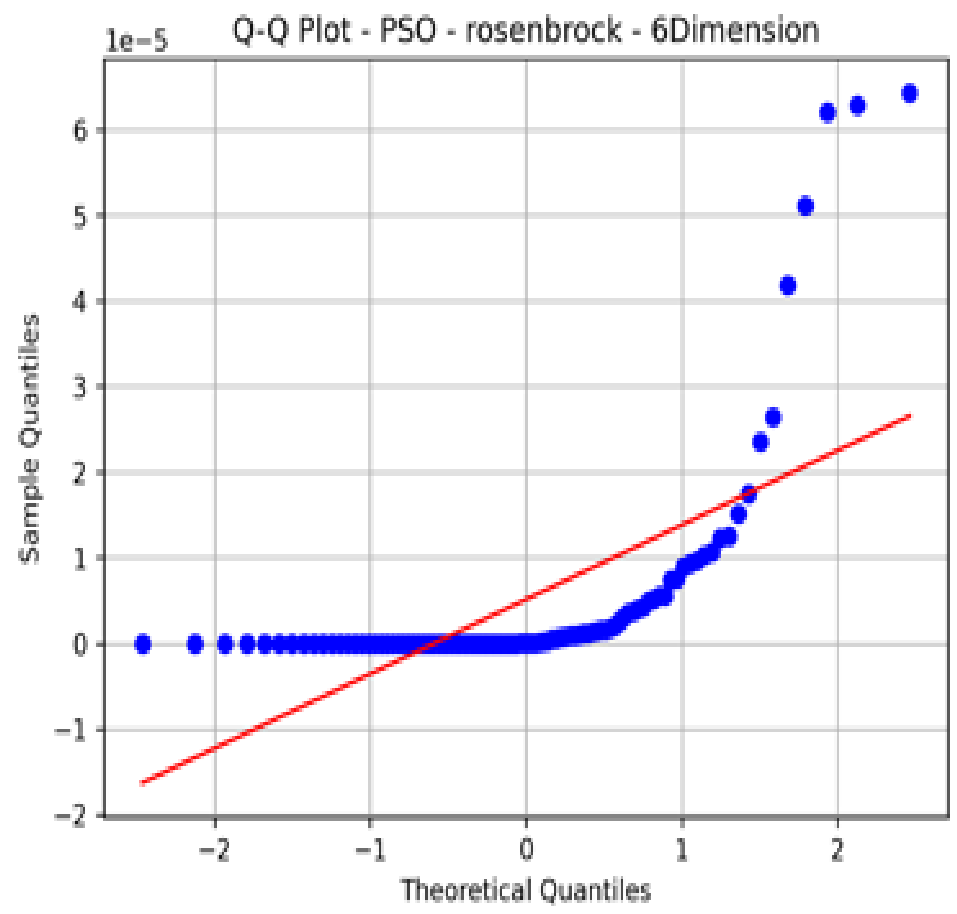
Q-Q Plot - PSO - rosenbrock - 6Dimension
1e−5
Sample Quantiles
Theoretical Quantiles

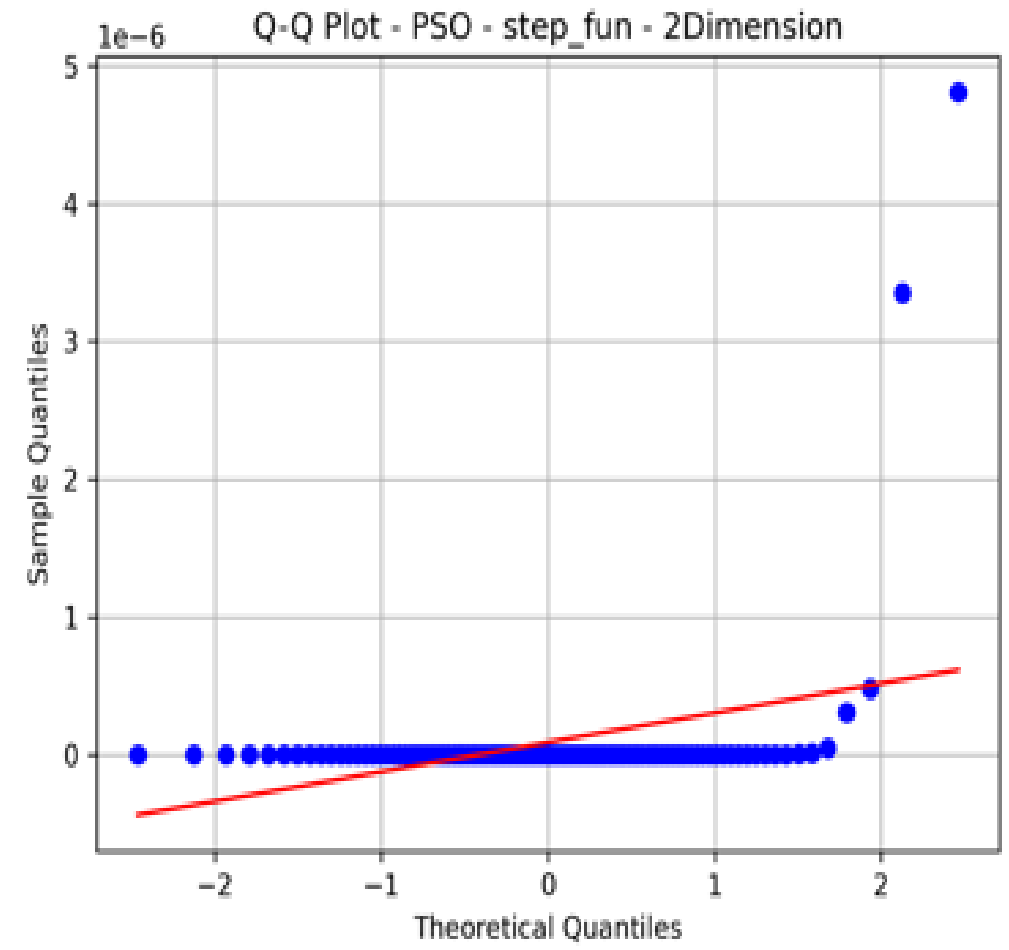
Q-Q Plot - PSO - step_fun - 2Dimension
1e−6
Sample Quantiles
Theoretical Quantiles

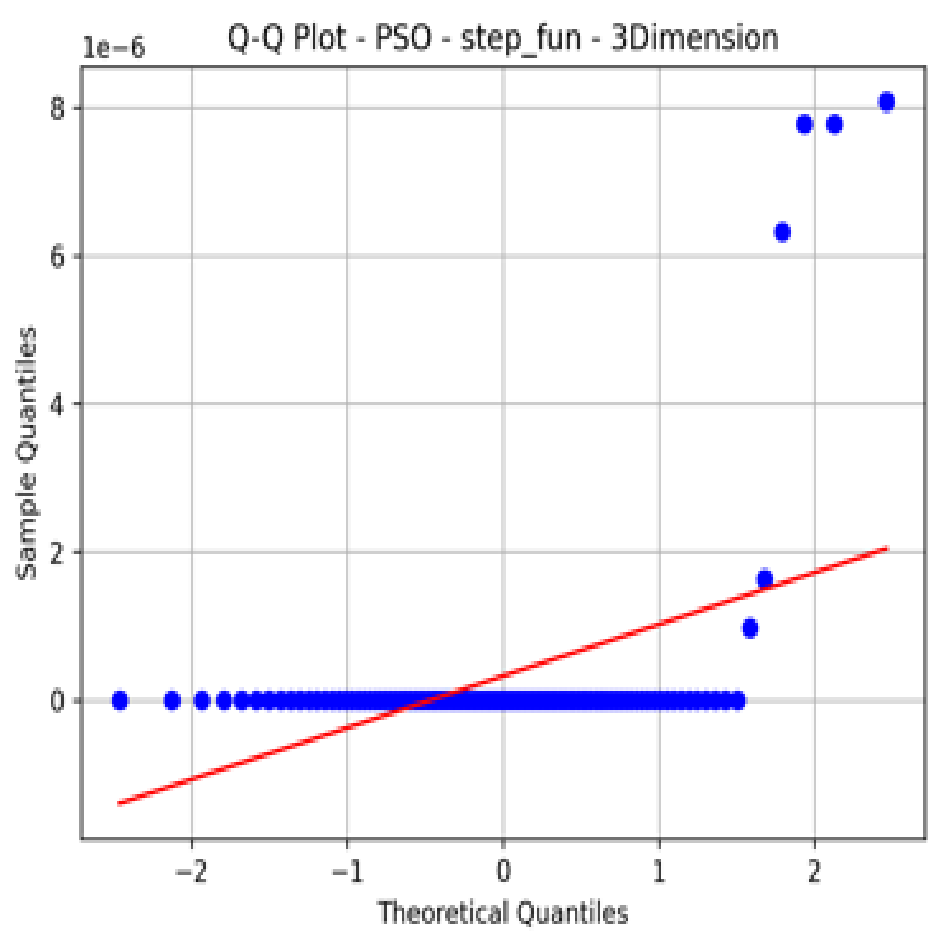
Q-Q Plot - PSO - step_fun - 3Dimension
1e−6
Sample Quantiles
Theoretical Quantiles

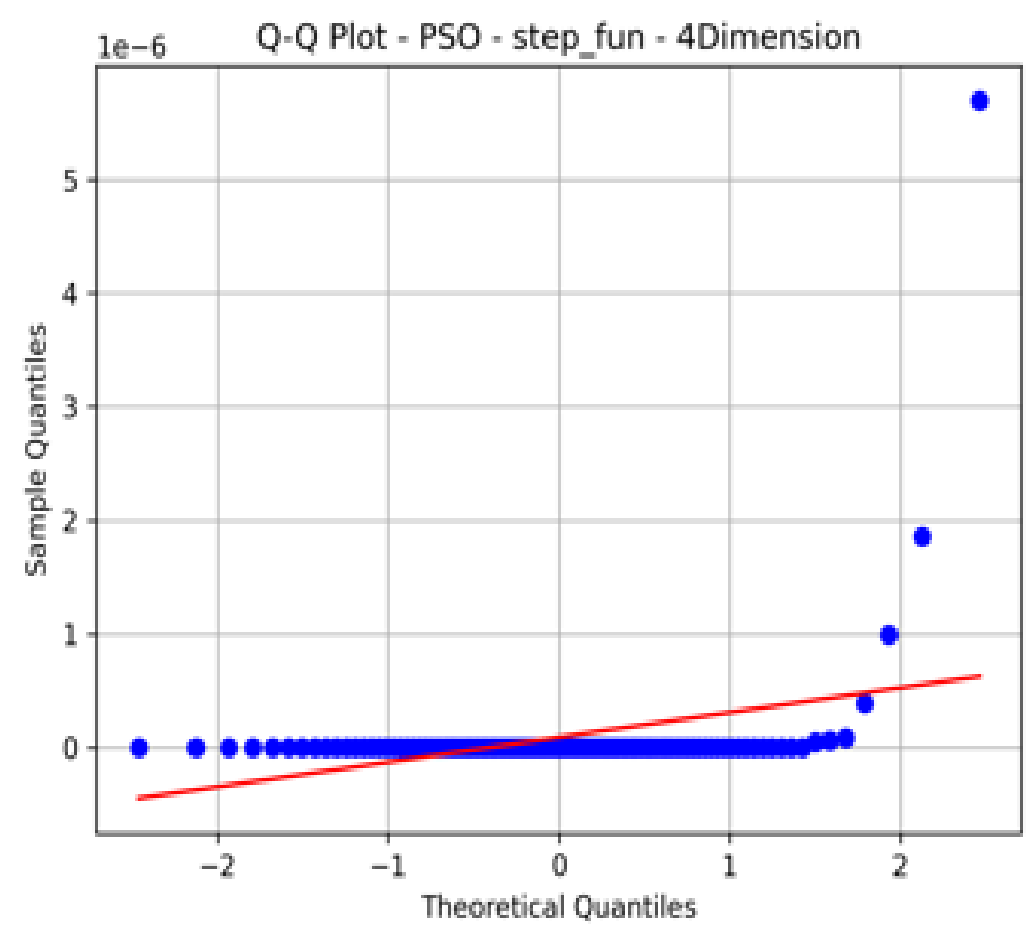
Q-Q Plot - PSO - step_fun - 4Dimension
1e−6
Sample Quantiles
Theoretical Quantiles

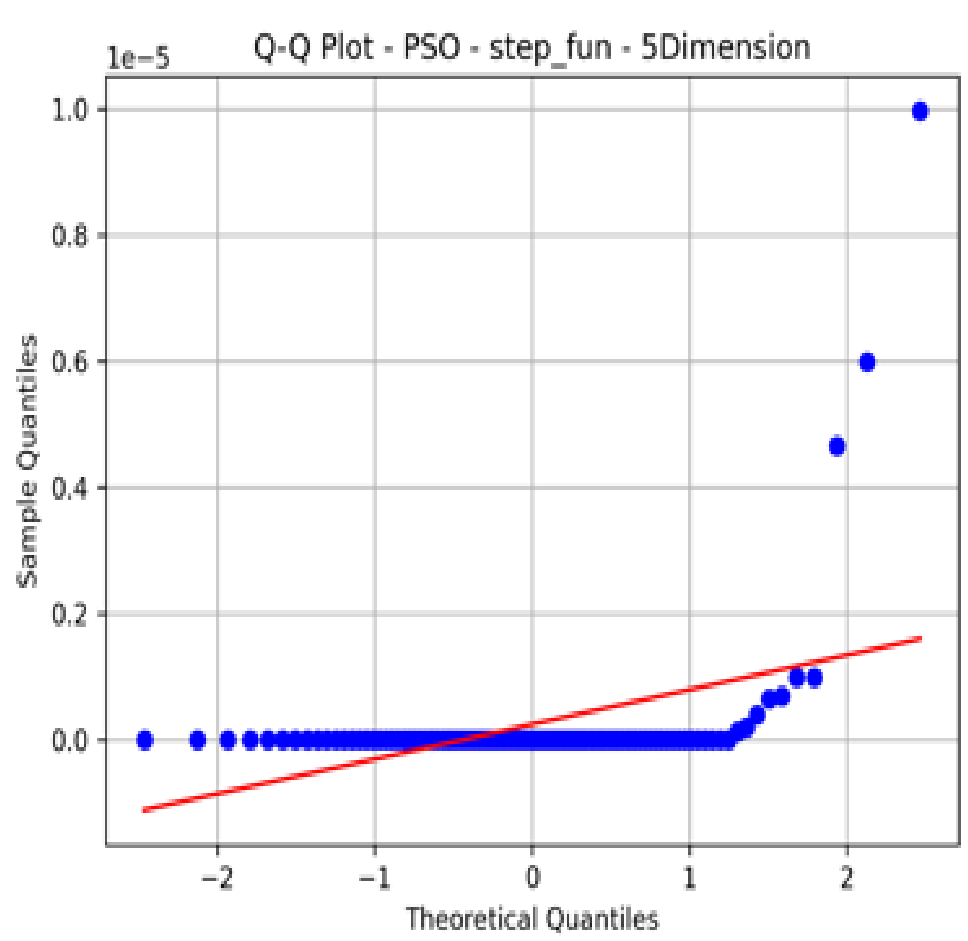
Q-Q Plot - PSO - step_fun - 5Dimension
1e−5
Sample Quantiles
Theoretical Quantiles

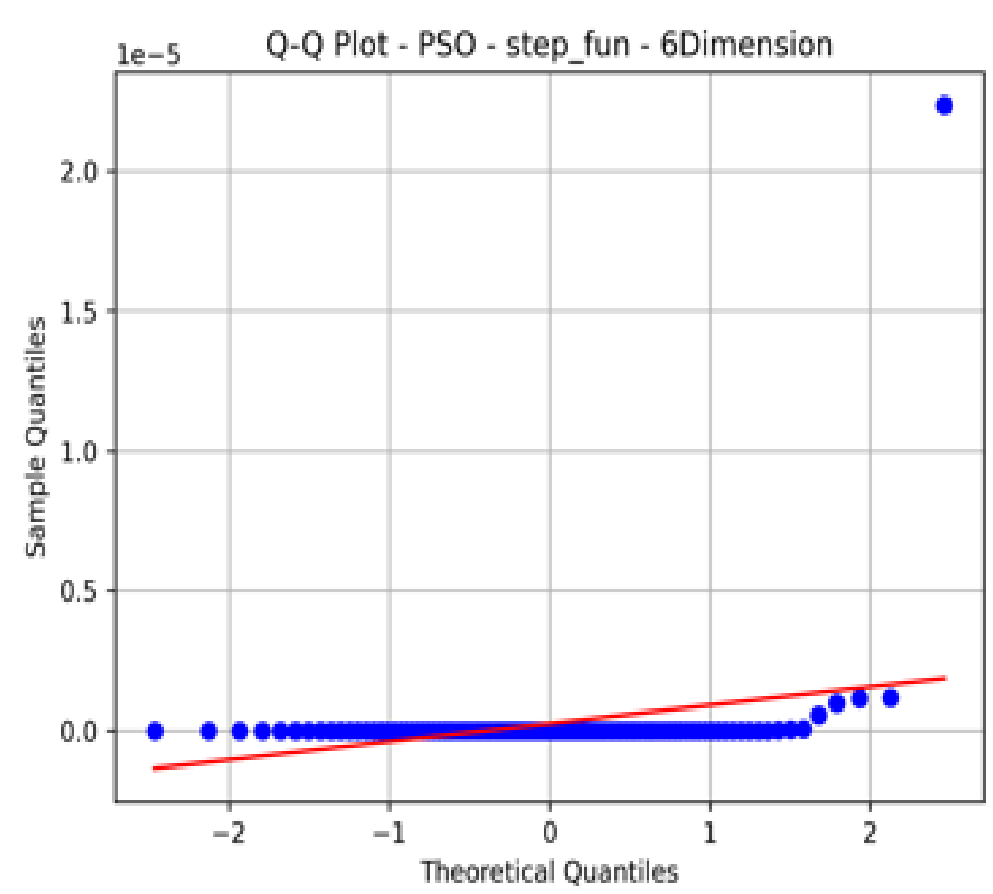
Q-Q Plot - PSO - step_fun - 6Dimension
1e−5
Sample Quantiles
Theoretical Quantiles

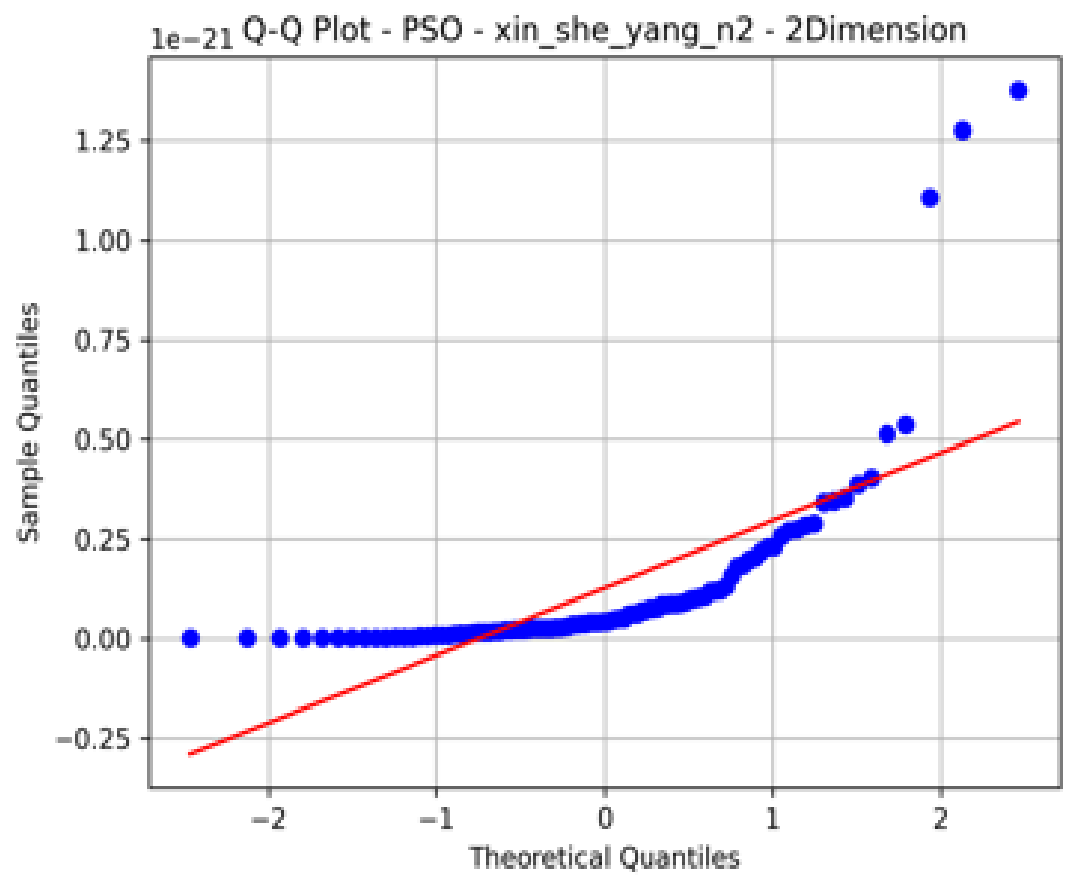

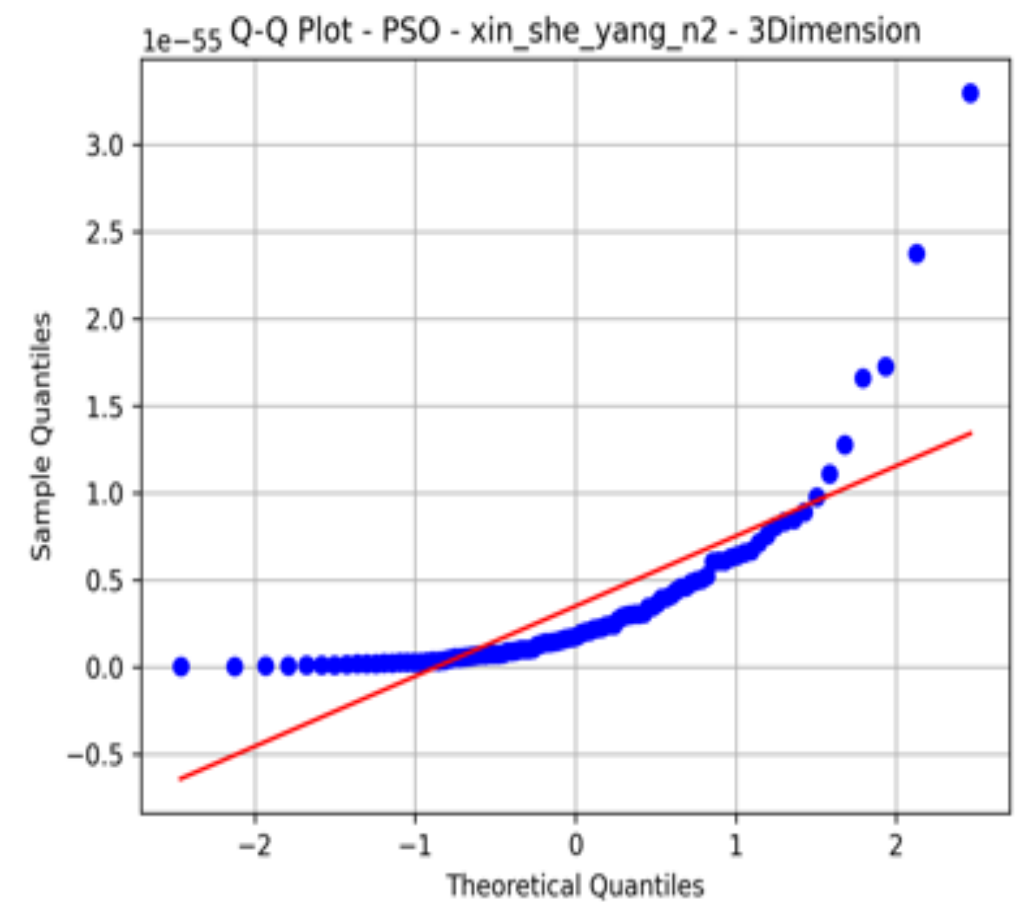

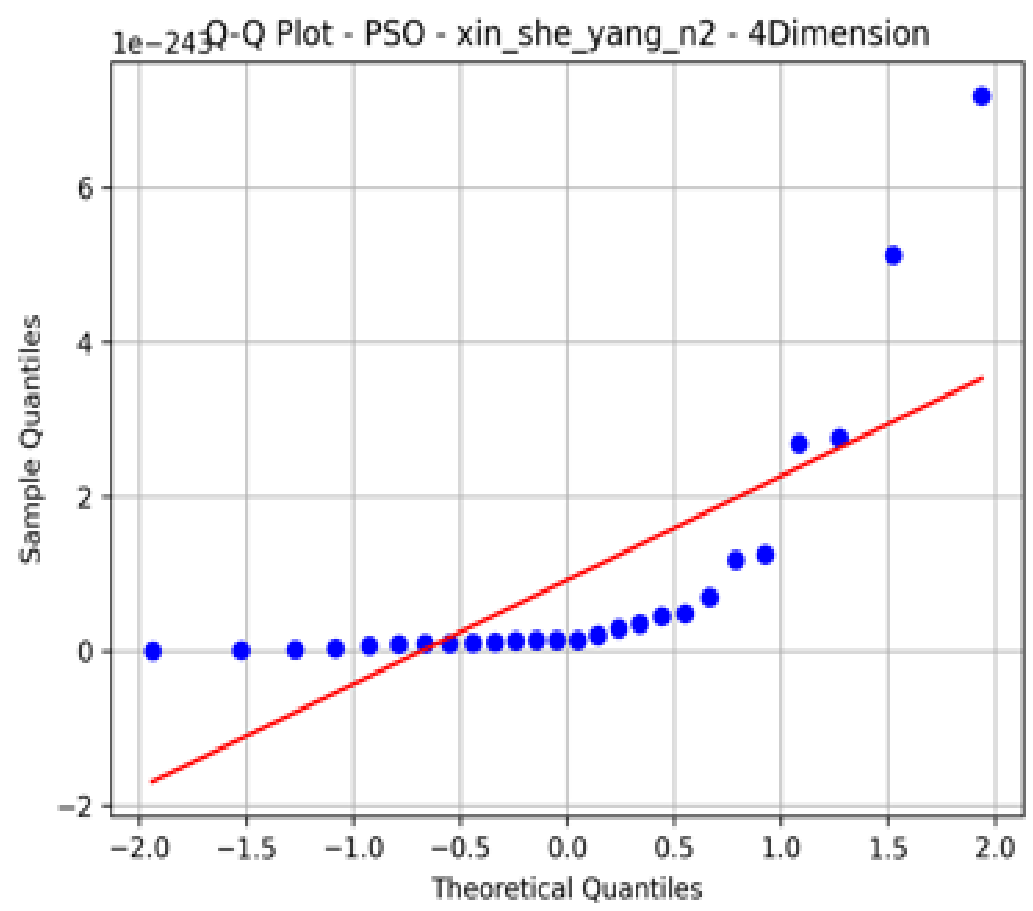

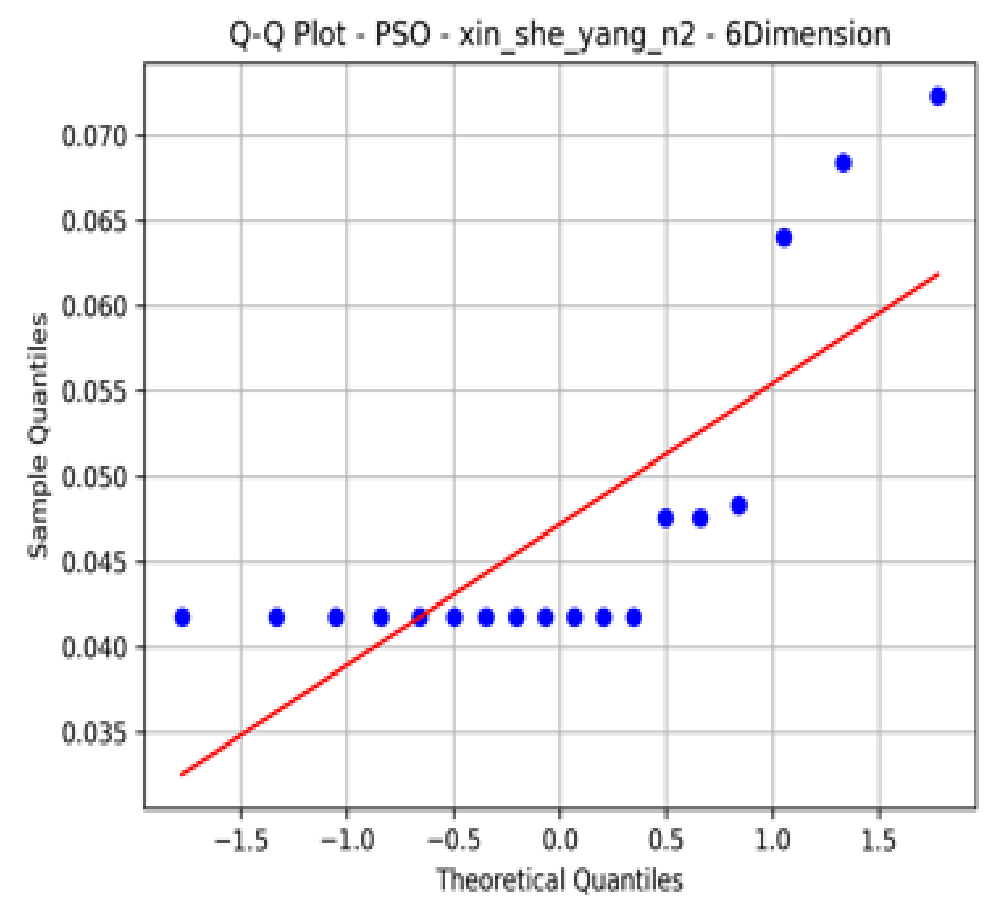

<u>WO Q-Q Plots</u>

Non-Continuous functions QQ Plots

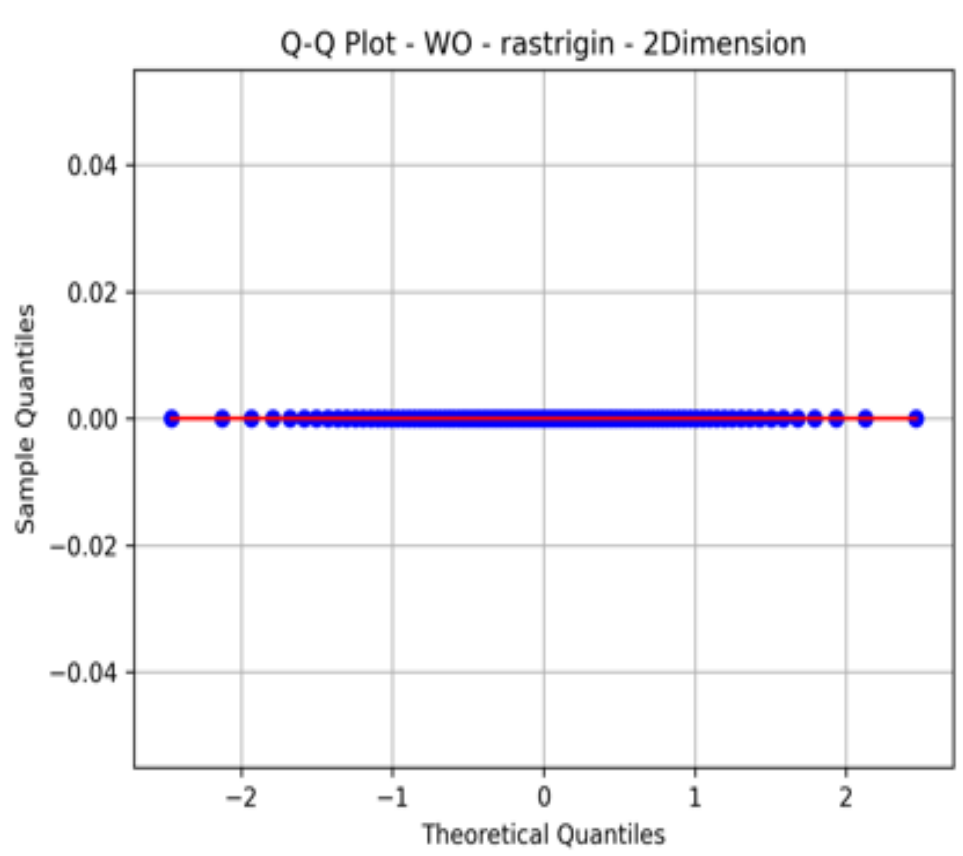

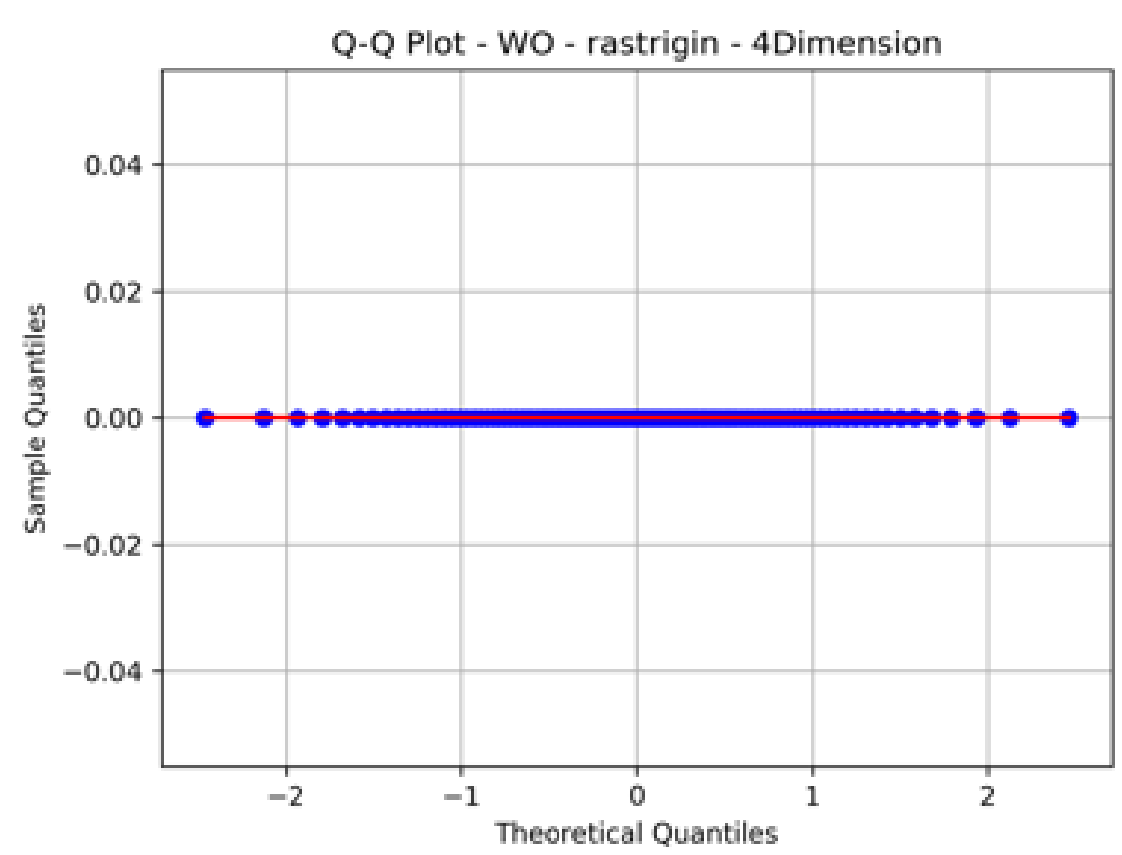

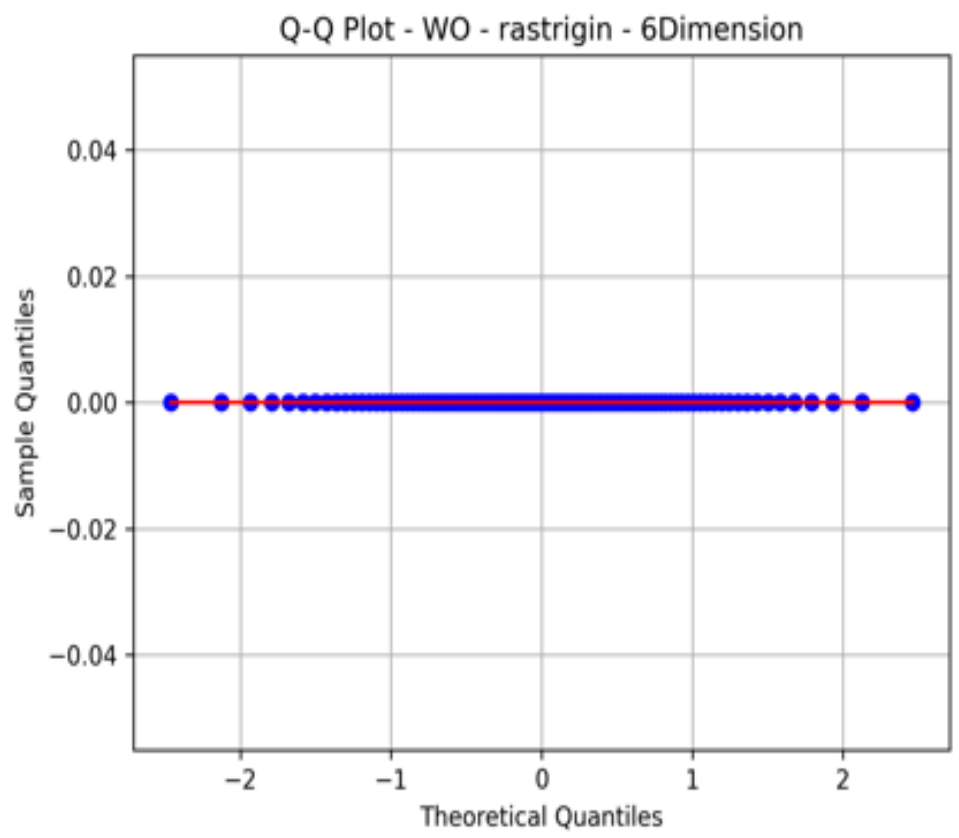
Q-Q Plot - WO - rastrigin - 6Dimension
Sample Quantiles
Theoretical Quantiles

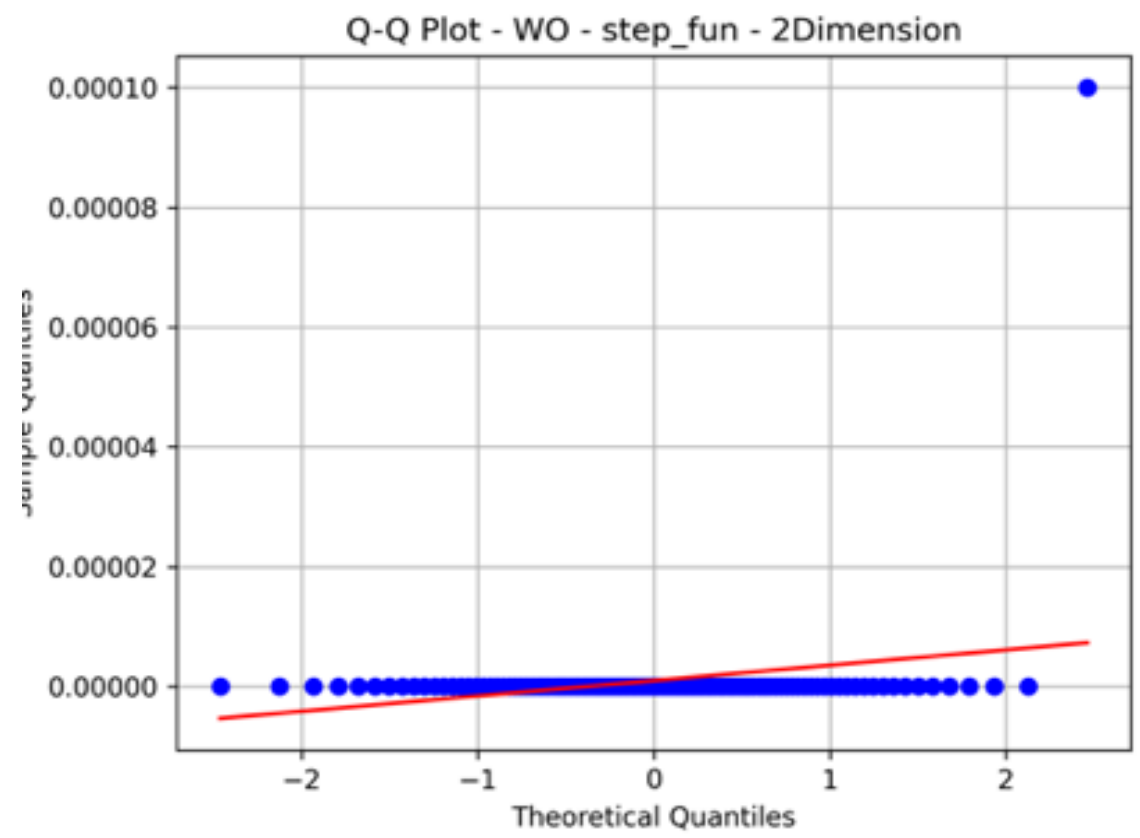
Q-Q Plot - WO - step_fun - 2Dimension
Theoretical Quantiles

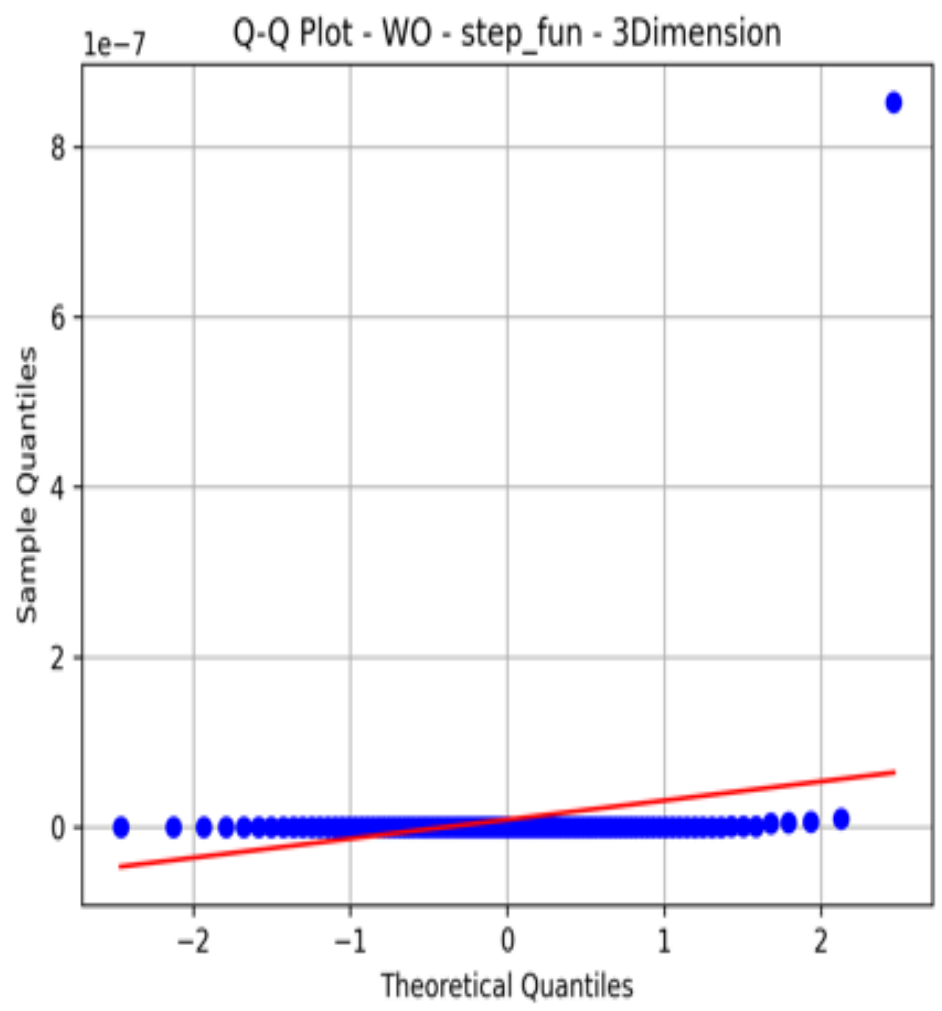
Q-Q Plot - WO - step_fun - 3Dimension
1e−7
Sample Quantiles
Theoretical Quantiles

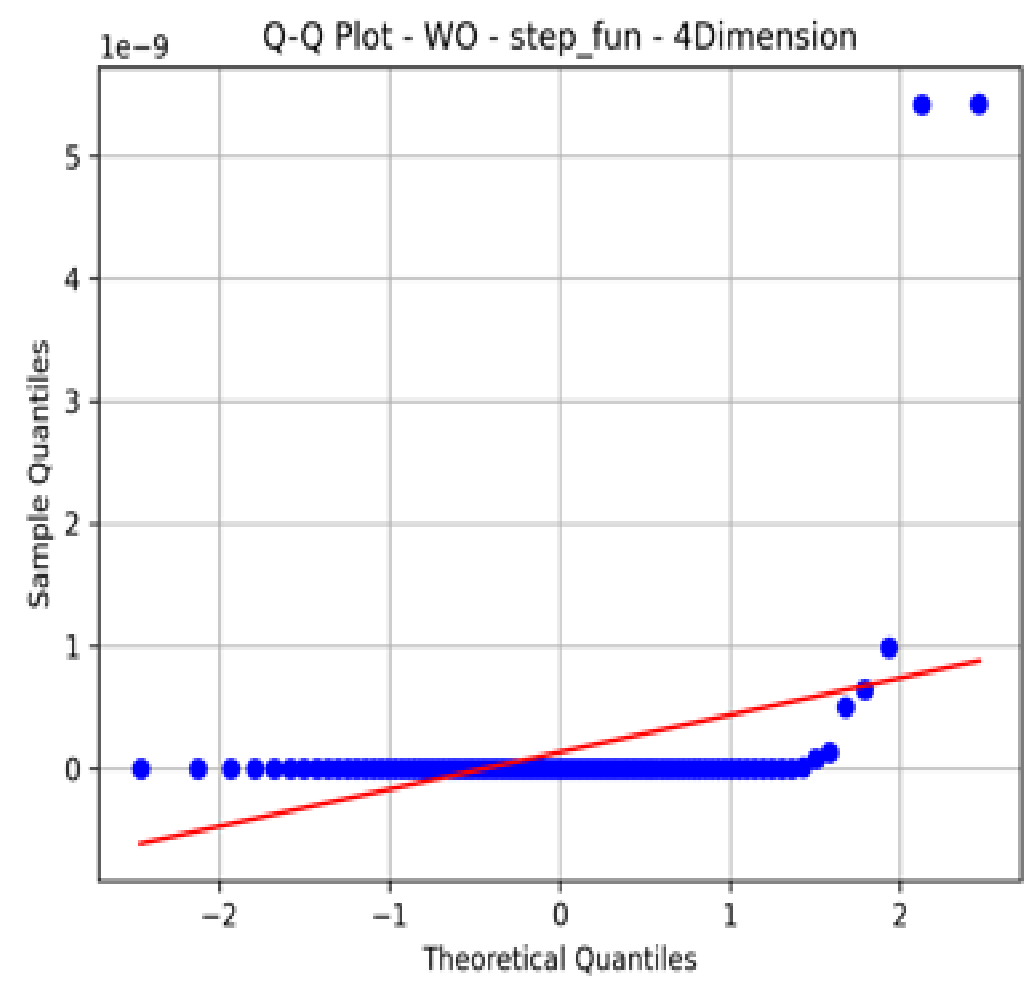
Q-Q Plot - WO - step_fun - 4Dimension
1e−9
Sample Quantiles
Theoretical Quantiles

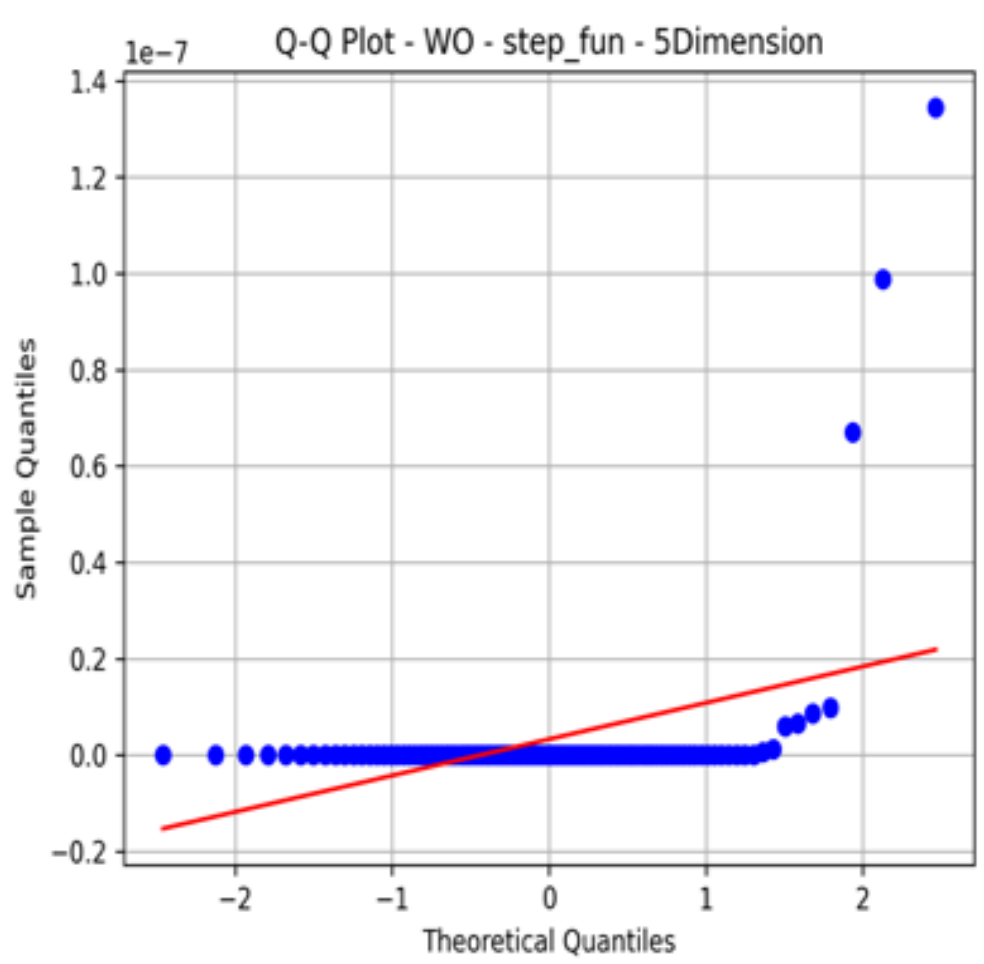
Q-Q Plot - WO - step_fun - 5Dimension
1e−7
Sample Quantiles
Theoretical Quantiles

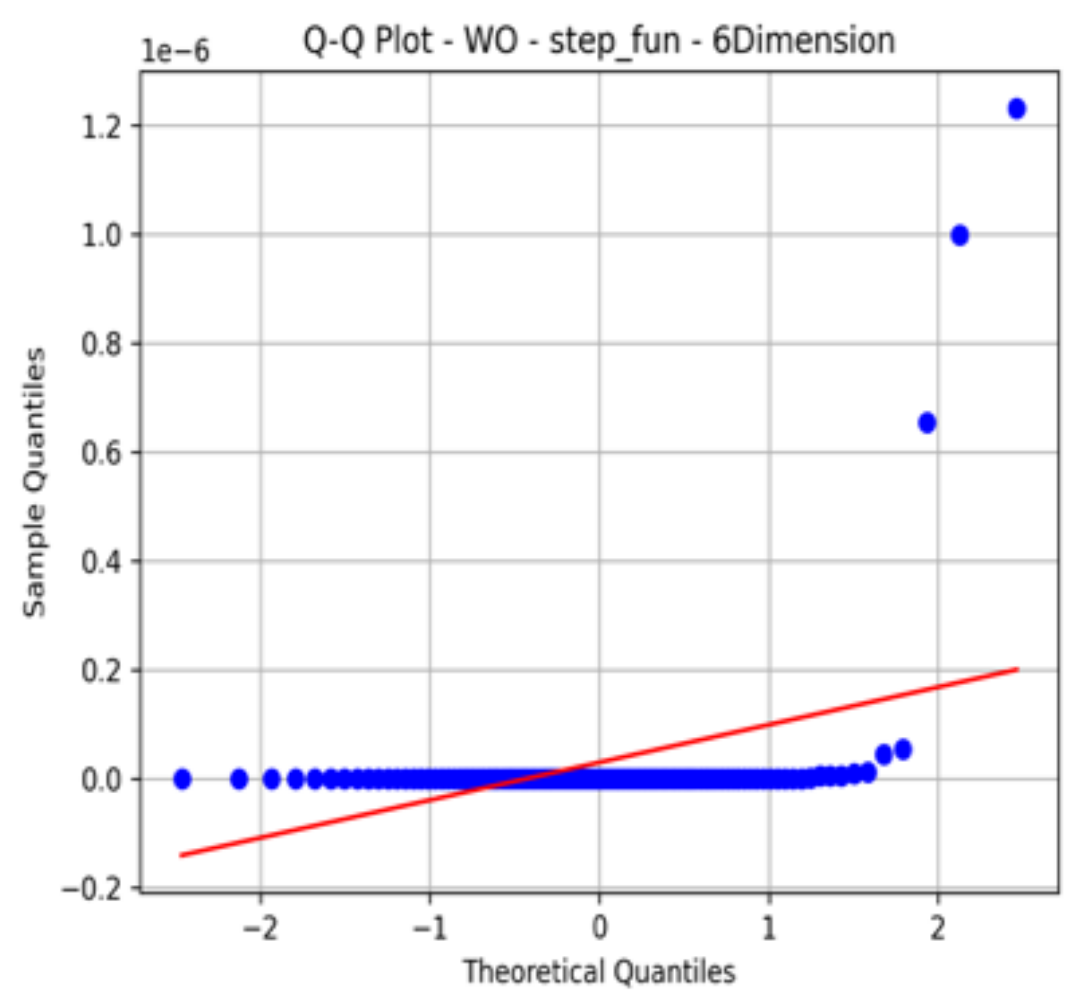
Q-Q Plot - WO - step_fun - 6Dimension
1e−6
Sample Quantiles
Theoretical Quantiles

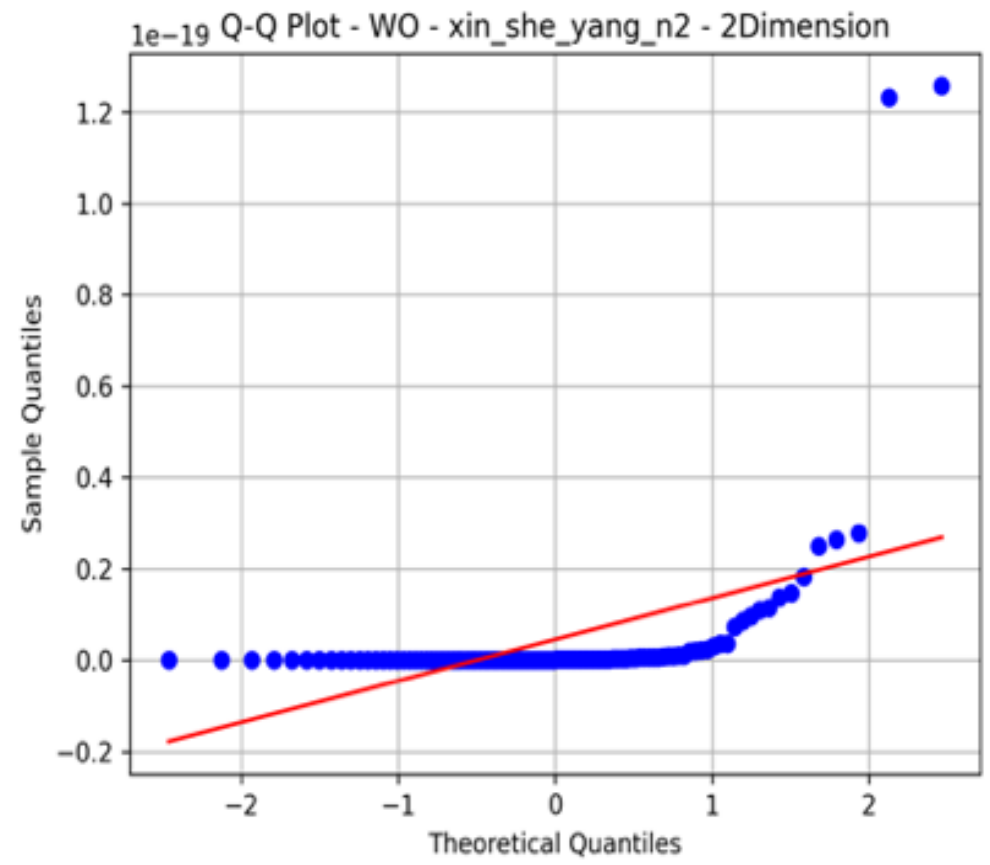
1e−19
Q-Q Plot - WO - xin_she_yang_n2 - 2Dimension
Sample Quantiles
Theoretical Quantiles

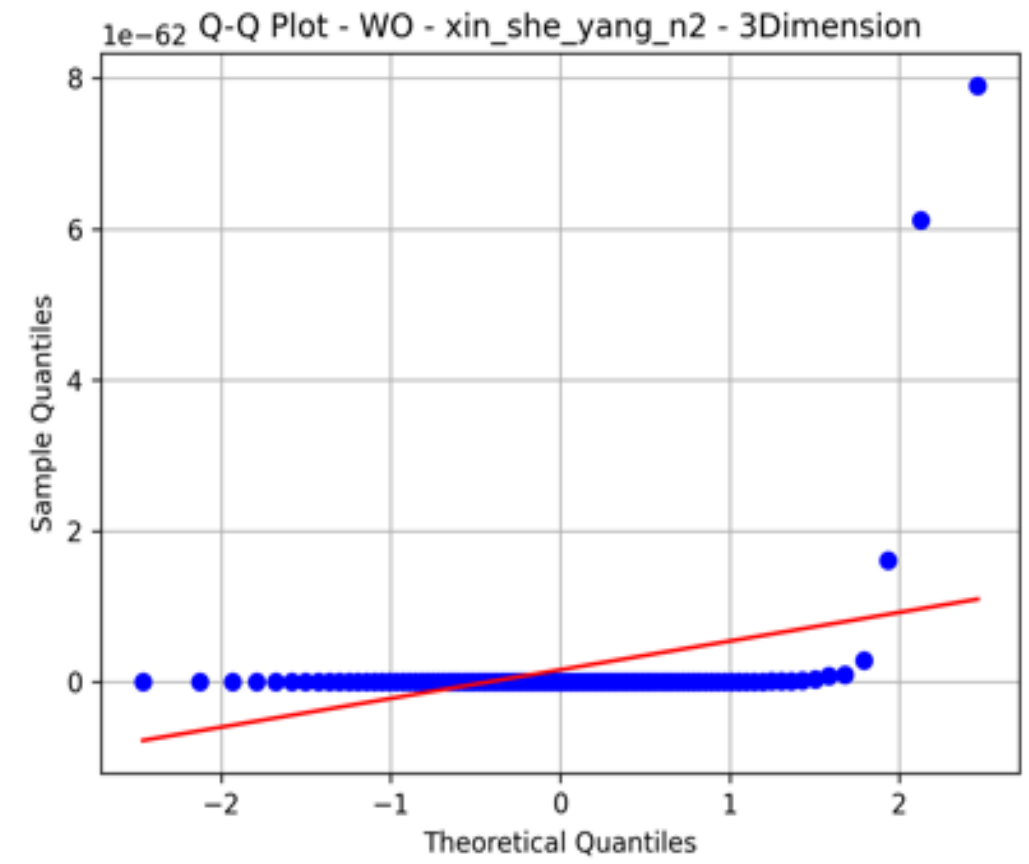
1e−62
Q-Q Plot - WO - xin_she_yang_n2 - 3Dimension
Sample Quantiles
Theoretical Quantiles

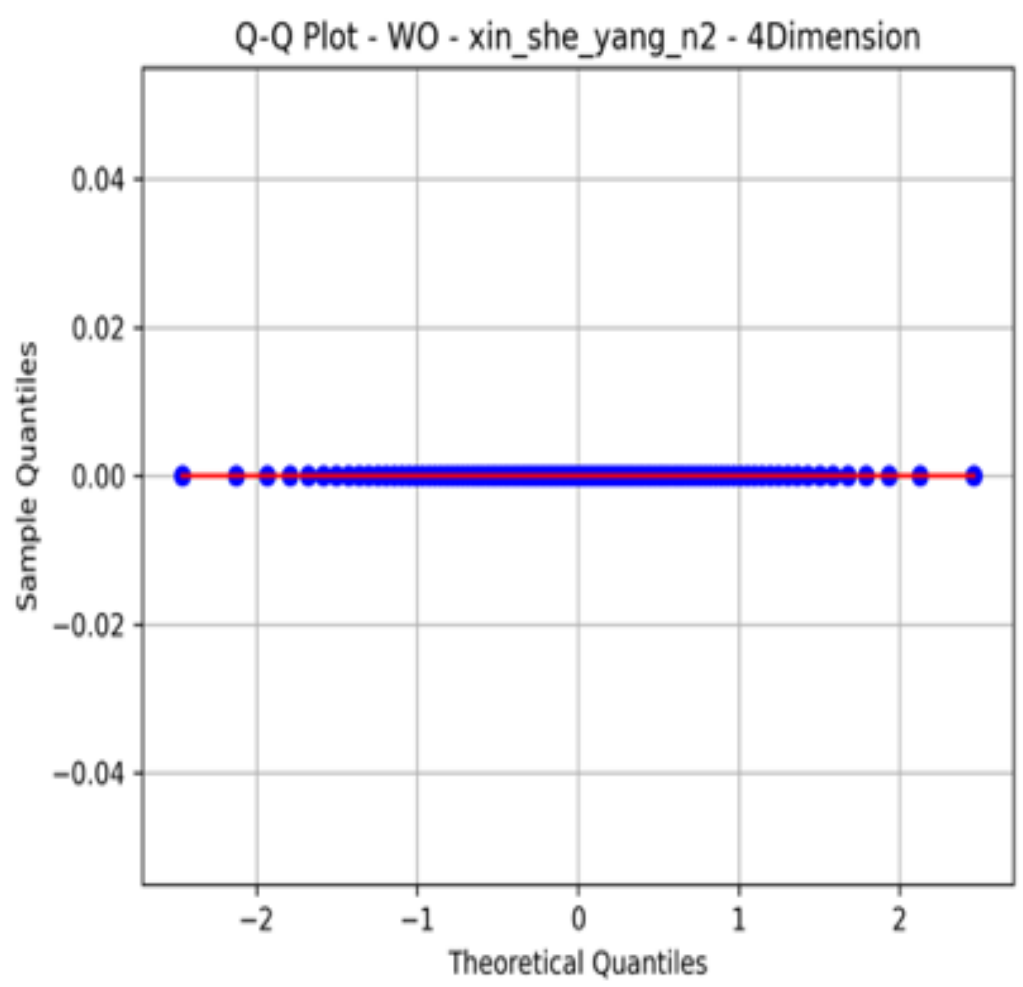
Q-Q Plot - WO - xin_she_yang_n2 - 4Dimension
Sample Quantiles
Theoretical Quantiles

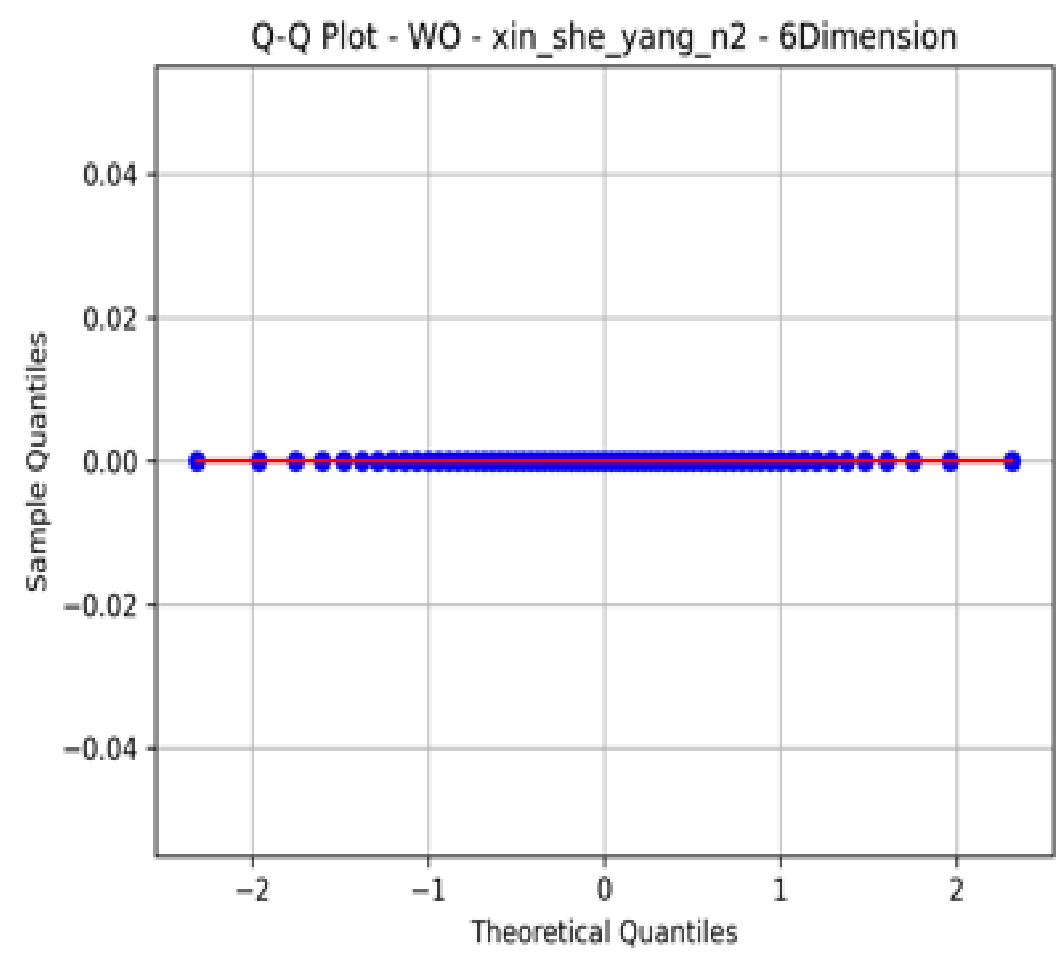
Q-Q Plot - WO - xin_she_yang_n2 - 6Dimension
Sample Quantiles
Theoretical Quantiles

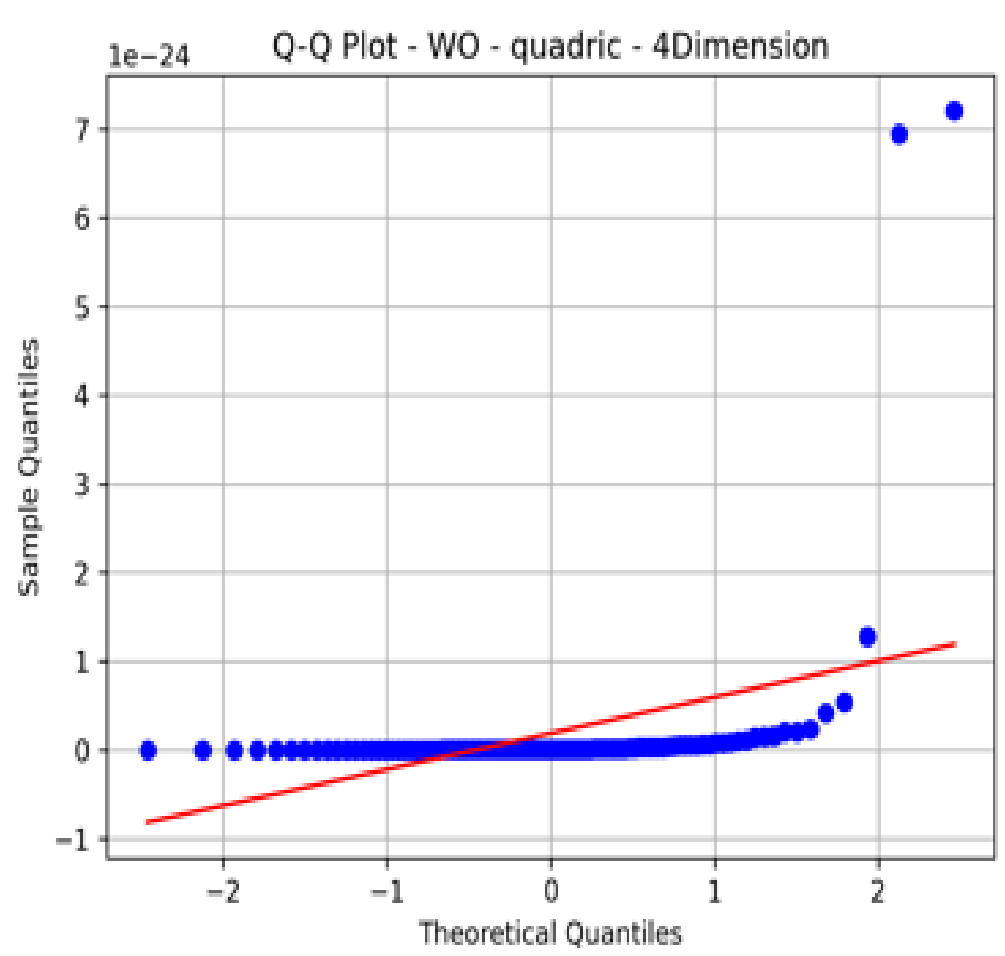
1e−24
Q-Q Plot - WO - quadric - 4Dimension
Sample Quantiles
Theoretical Quantiles

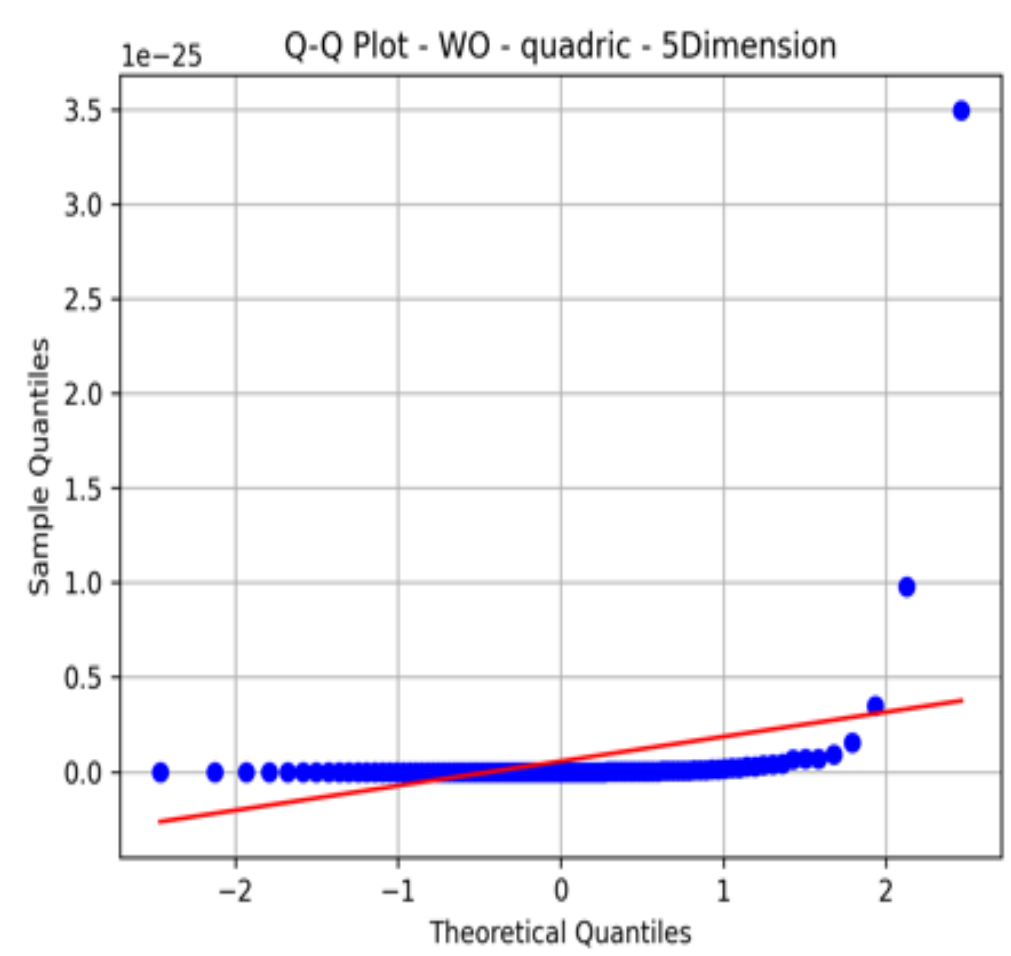
1e−25
Q-Q Plot - WO - quadric - 5Dimension
Sample Quantiles
Theoretical Quantiles

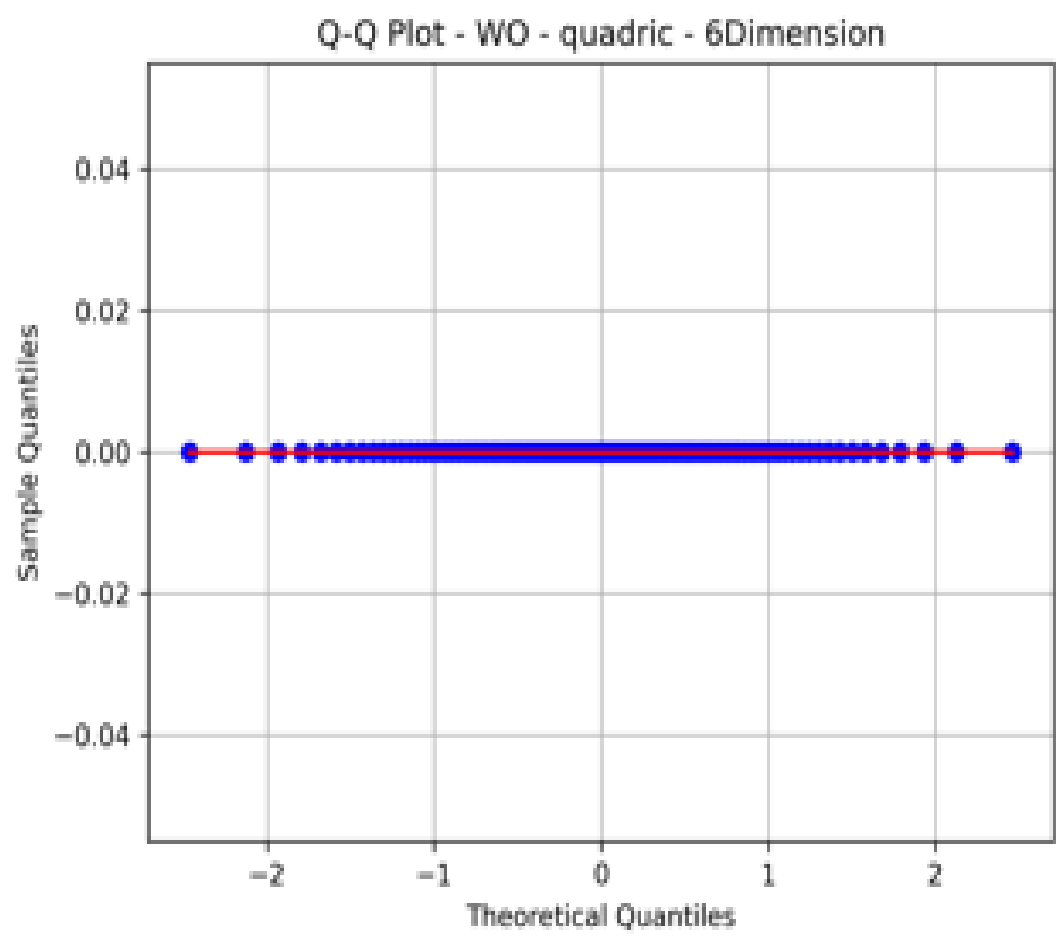

GWO Q-Q plots

Non-Continuous functions QQ Plots

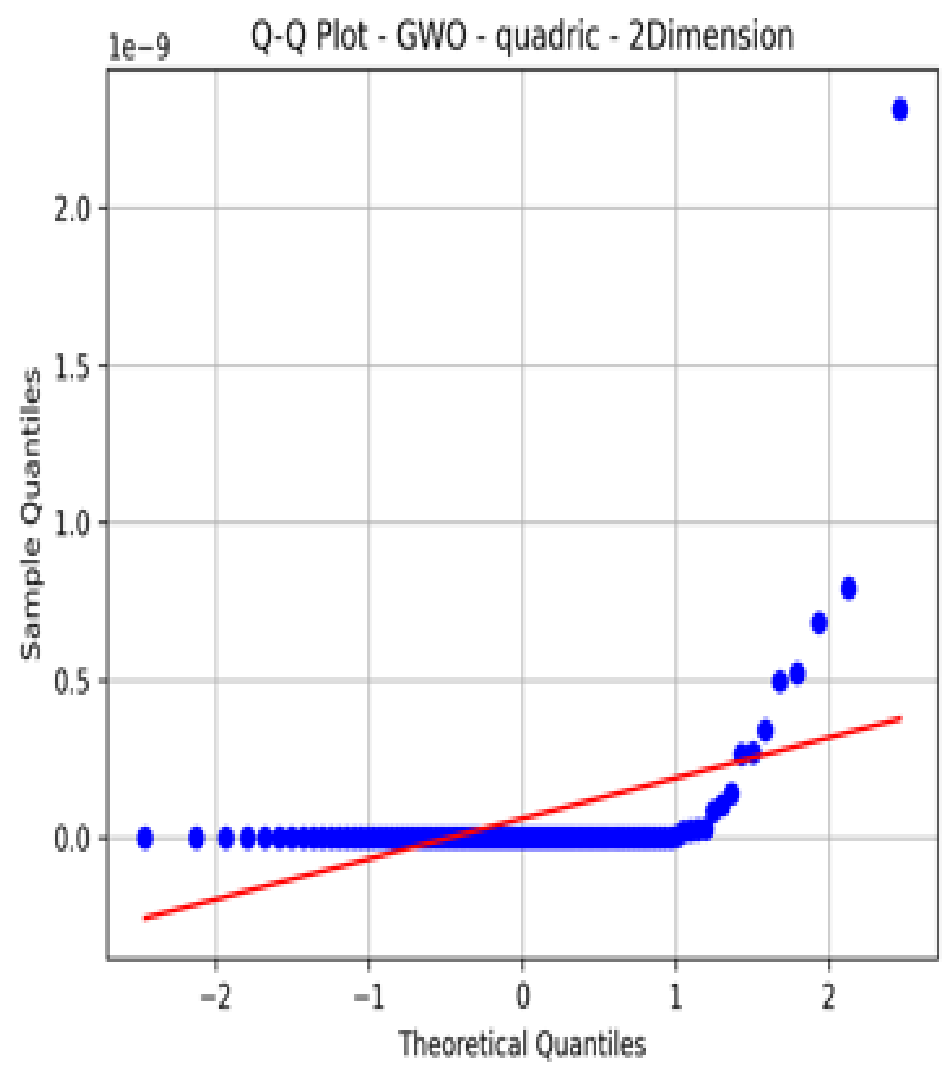

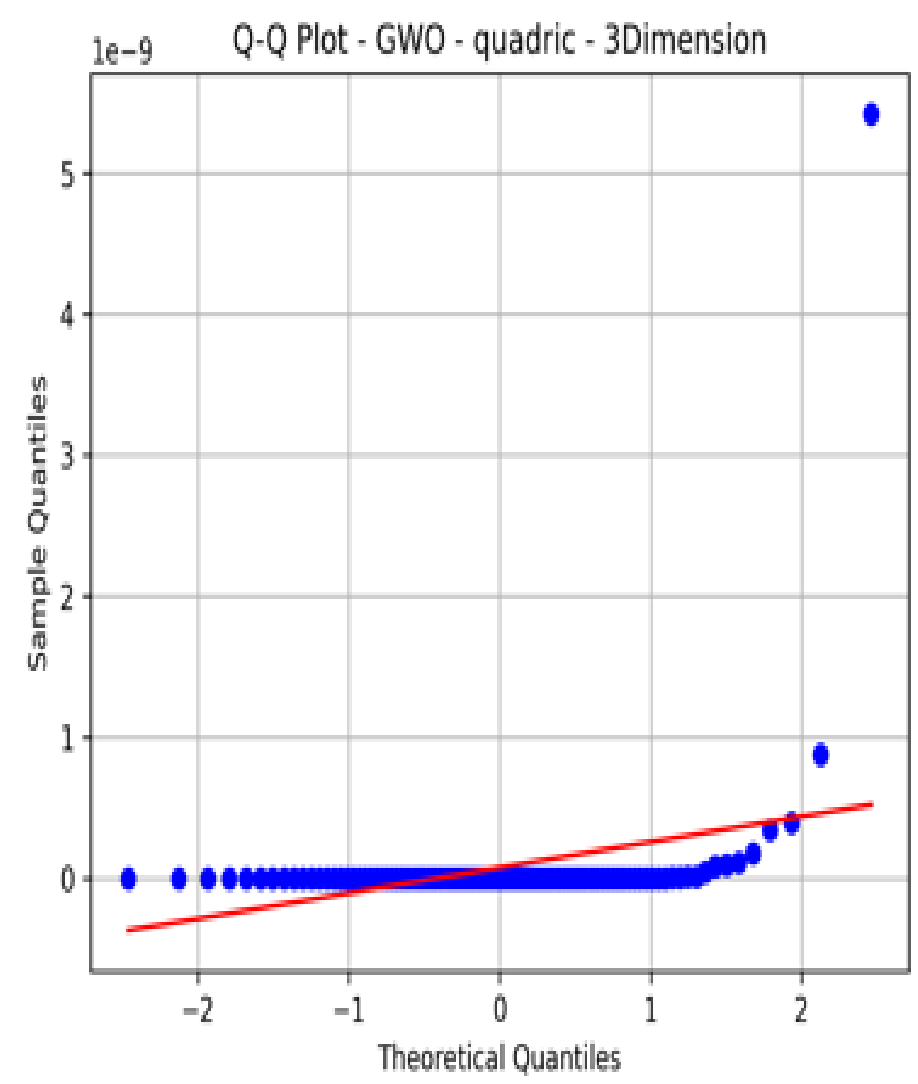

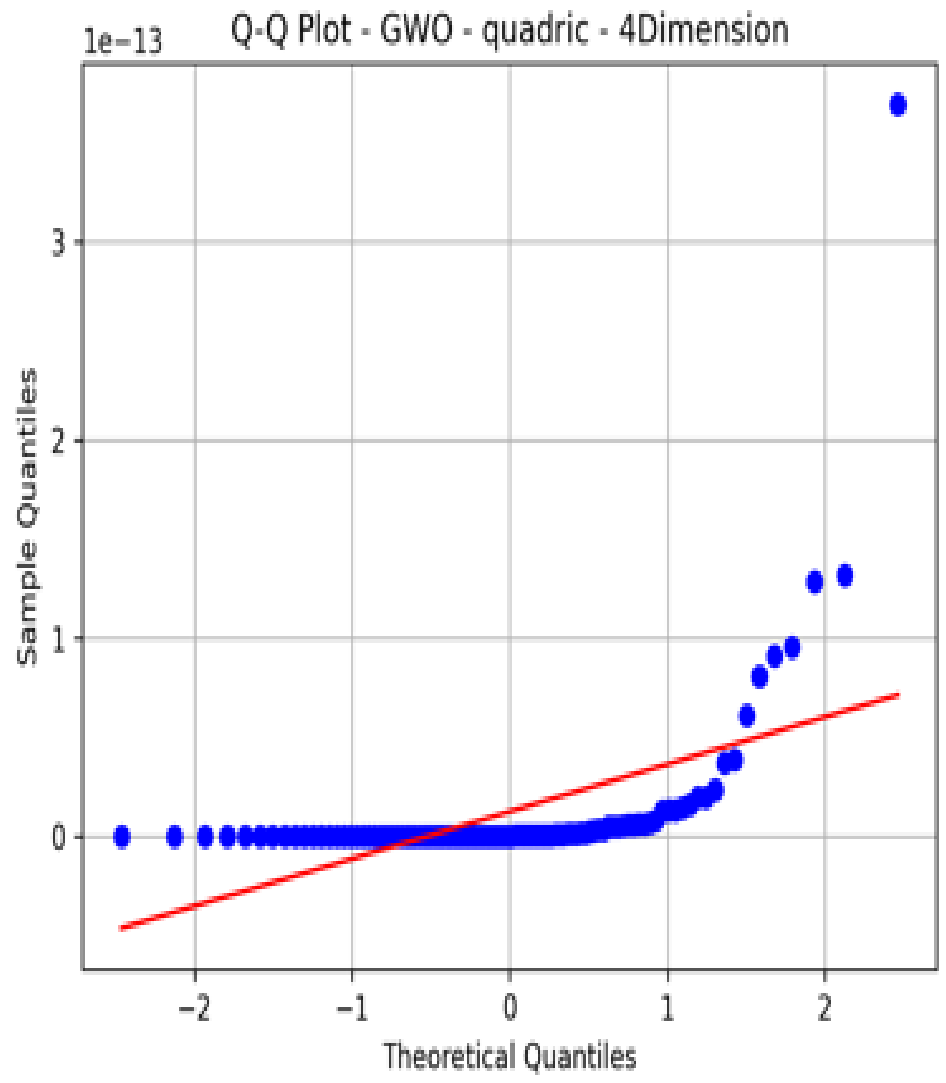
1e−13
Q-Q Plot - GWO - quadric - 4Dimension
Sample Quantiles
Theoretical Quantiles

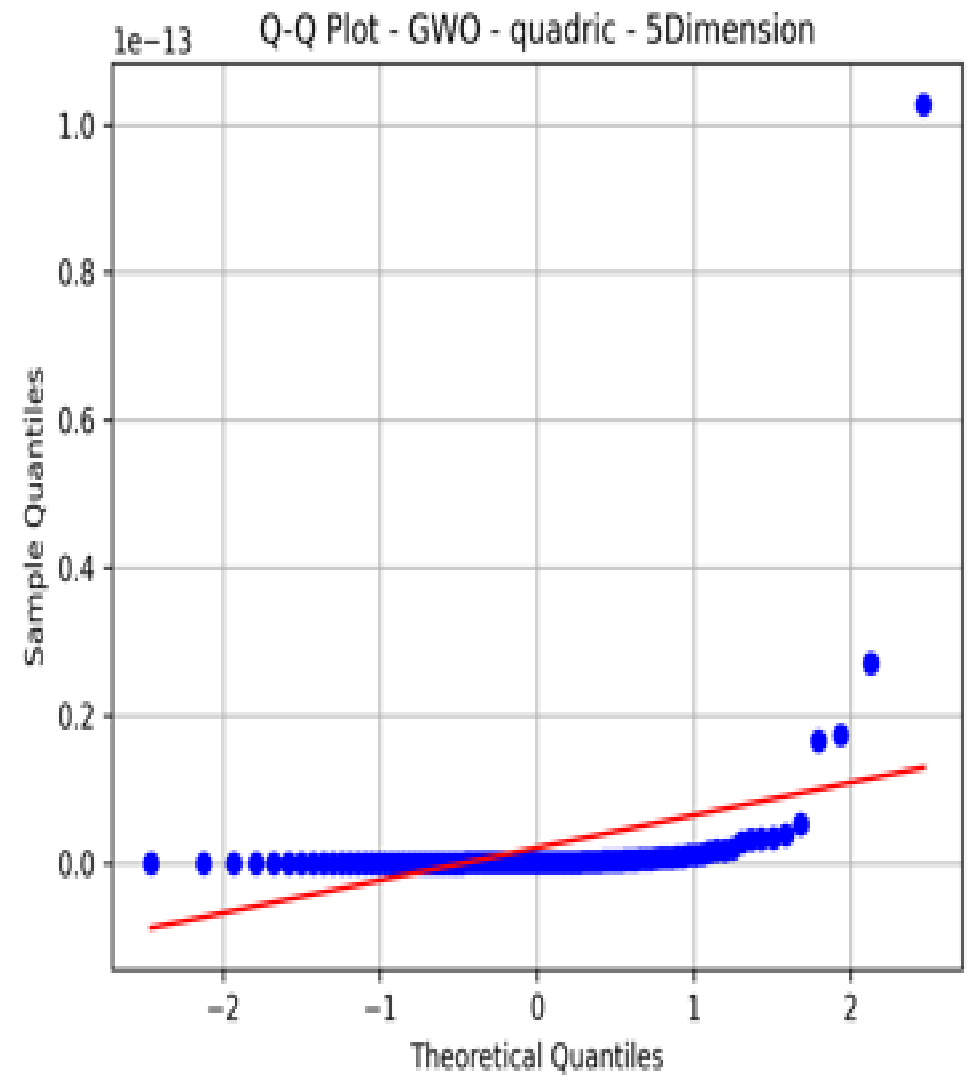
1e−13
Q-Q Plot - GWO - quadric - 5Dimension
Sample Quantiles
Theoretical Quantiles

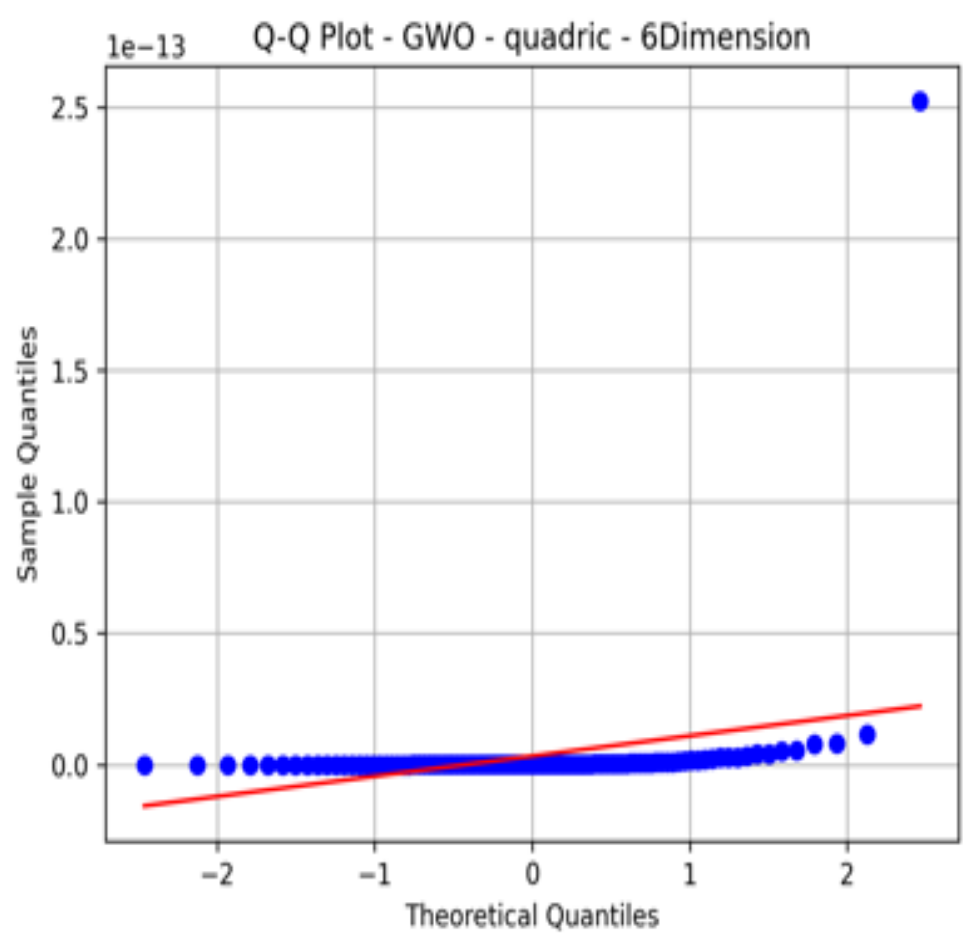
1e−13
Q-Q Plot - GWO - quadric - 6Dimension
Sample Quantiles
Theoretical Quantiles

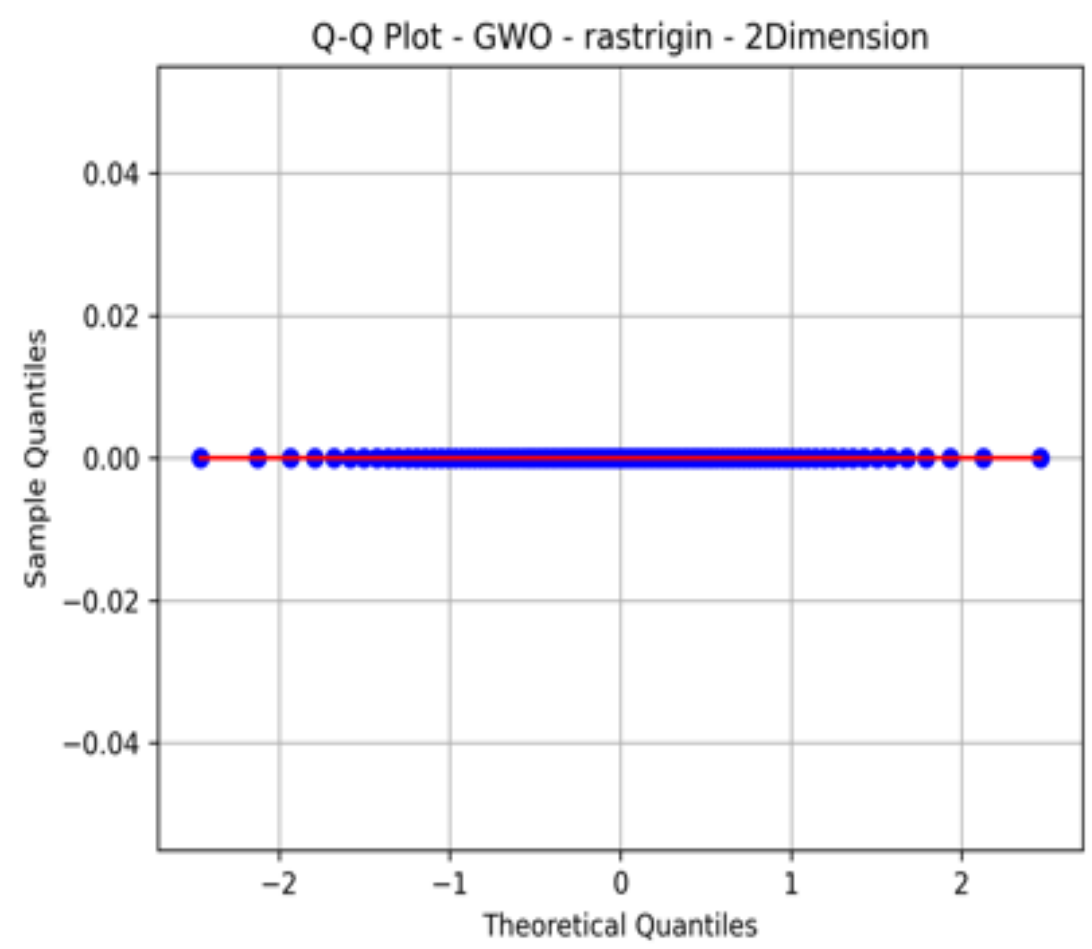
Q-Q Plot - GWO - rastrigin - 2Dimension
Sample Quantiles
Theoretical Quantiles

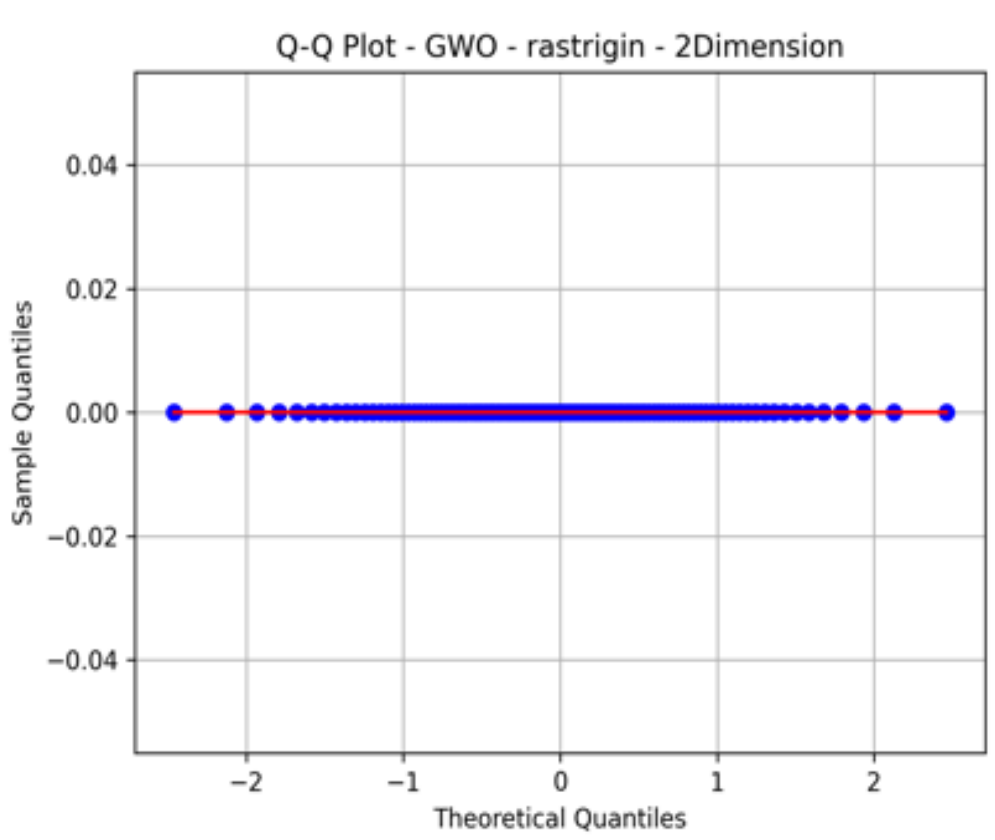
Q-Q Plot - GWO - rastrigin - 2Dimension
Sample Quantiles
Theoretical Quantiles

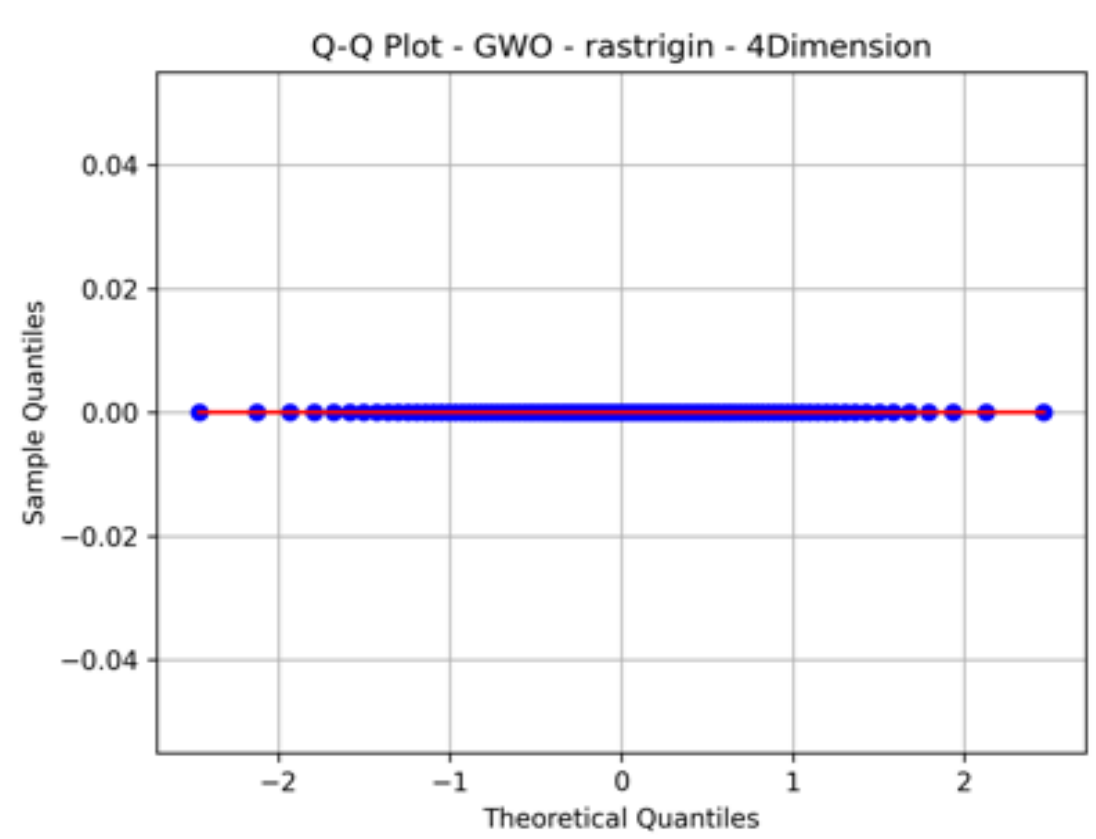
Q-Q Plot - GWO - rastrigin - 4Dimension
Sample Quantiles
Theoretical Quantiles

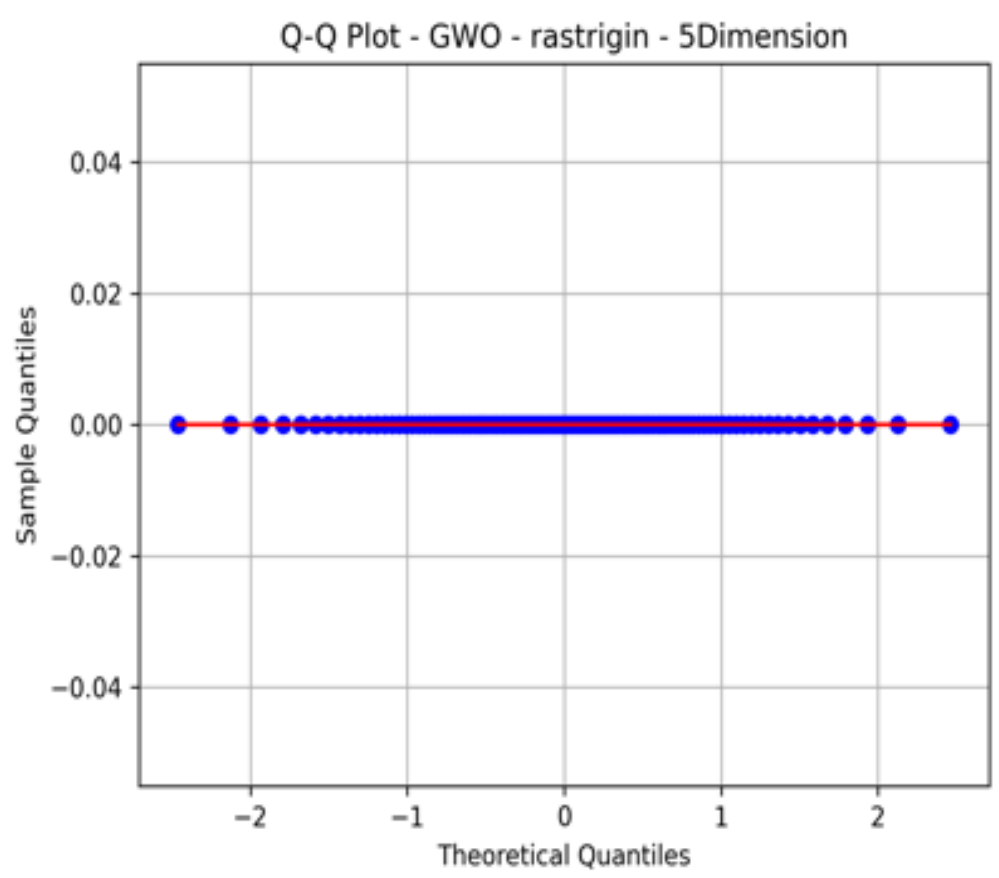
Q-Q Plot - GWO - rastrigin - 5Dimension
Sample Quantiles
Theoretical Quantiles

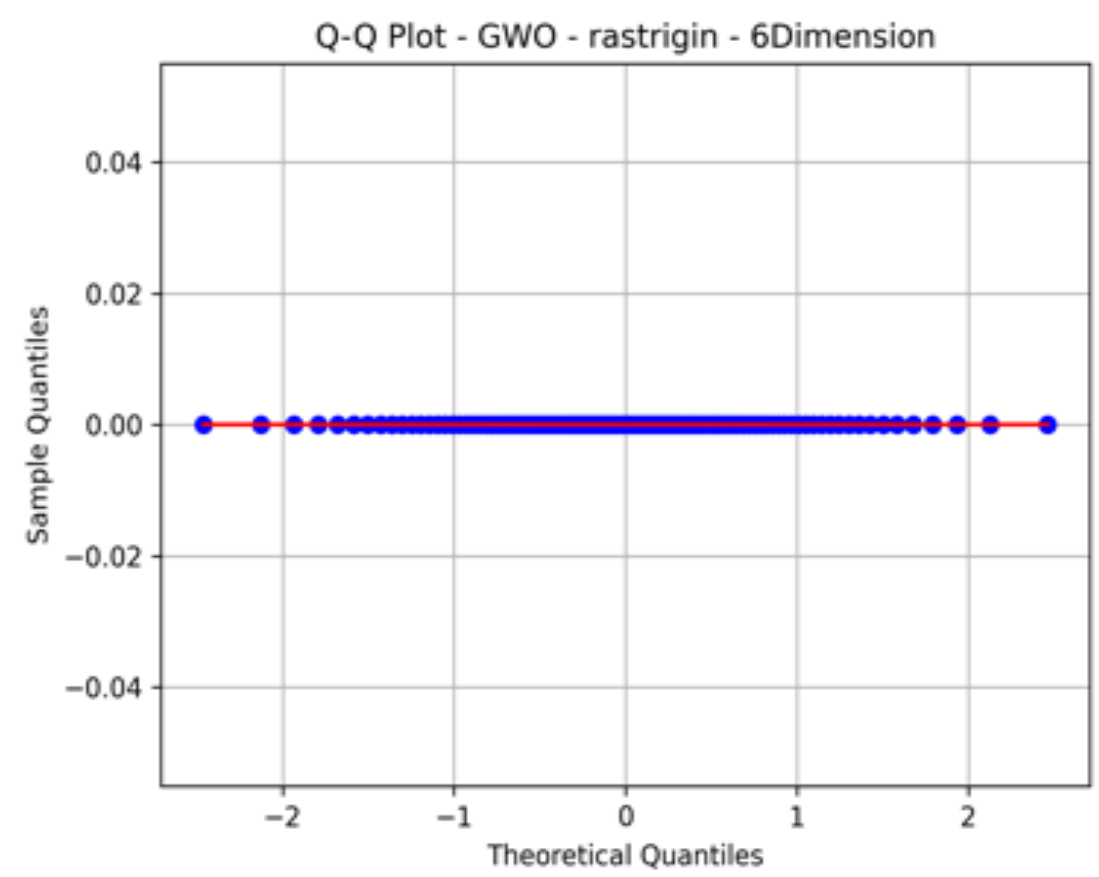
Q-Q Plot - GWO - rastrigin - 6Dimension
Sample Quantiles
Theoretical Quantiles

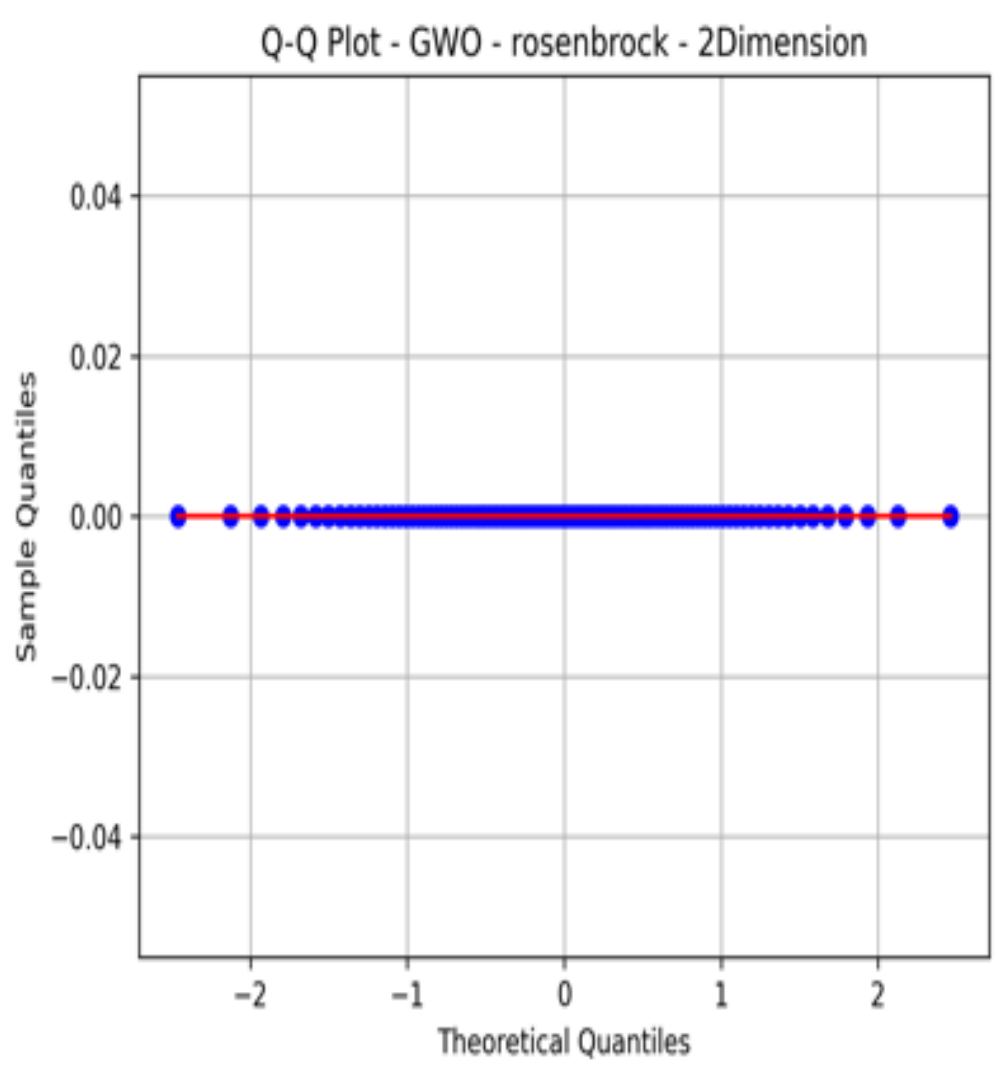
Q-Q Plot - GWO - rosenbrock - 2Dimension
Sample Quantiles
Theoretical Quantiles

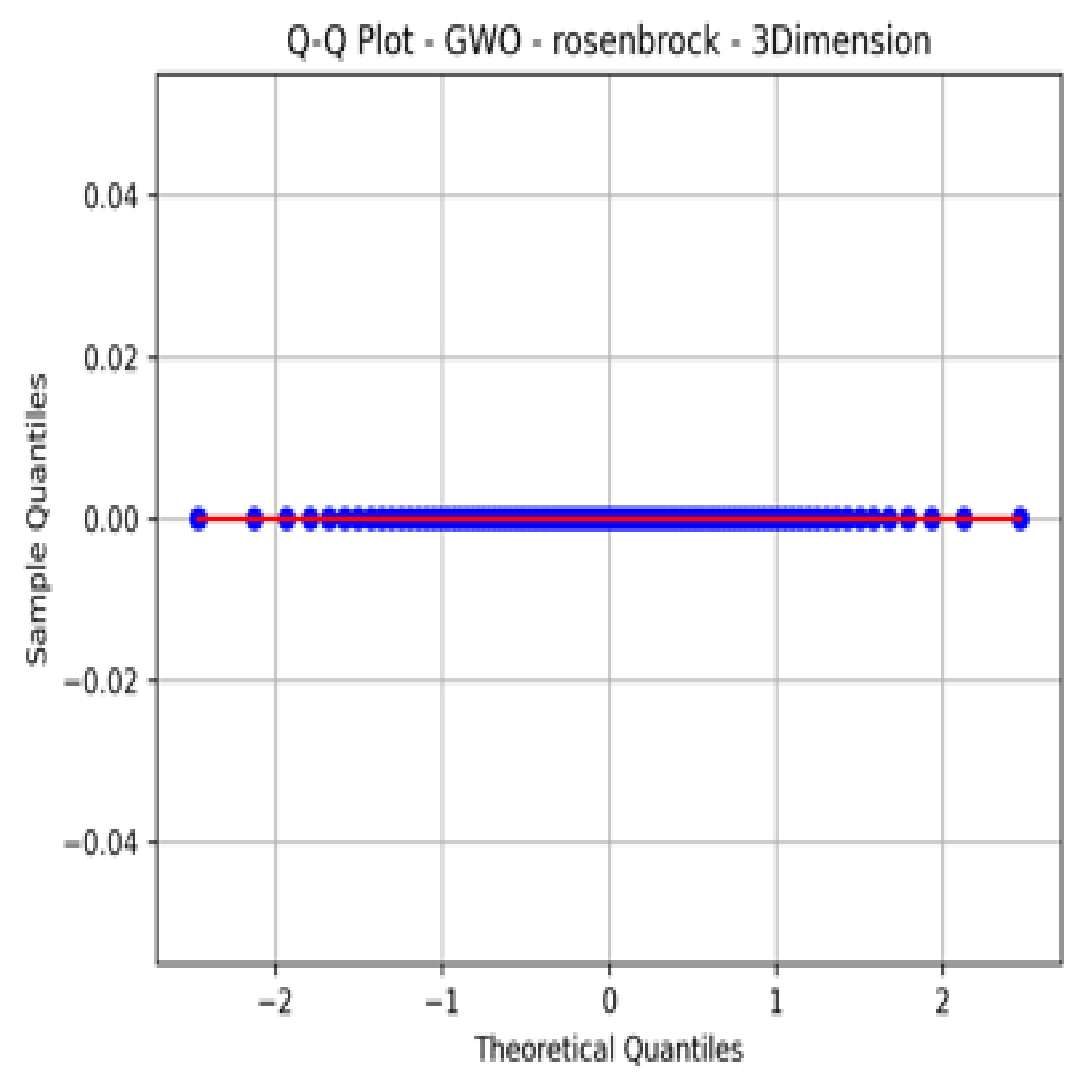
Q-Q Plot - GWO - rosenbrock - 3Dimension
Sample Quantiles
Theoretical Quantiles

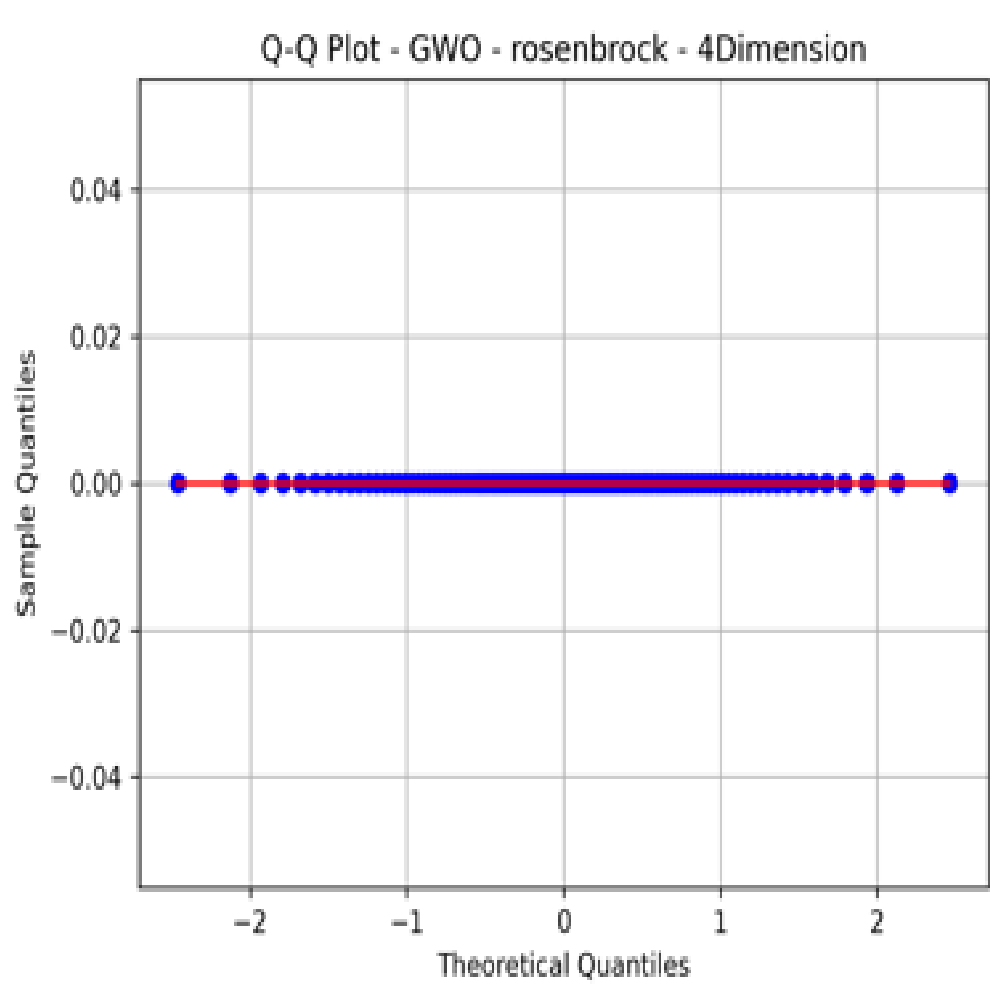
Q-Q Plot - GWO - rosenbrock - 4Dimension
Sample Quantiles
Theoretical Quantiles

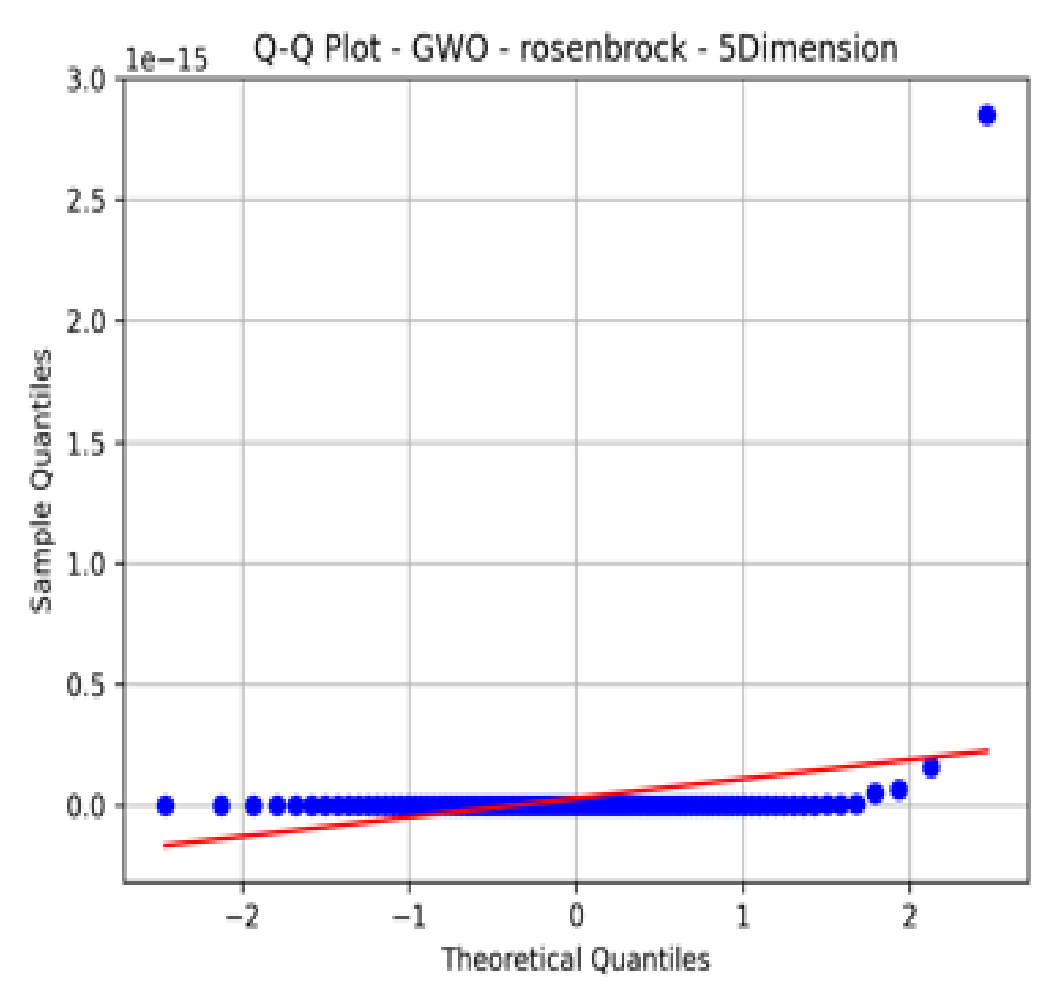
Q-Q Plot - GWO - rosenbrock - 5Dimension
1e−15
Sample Quantiles
Theoretical Quantiles

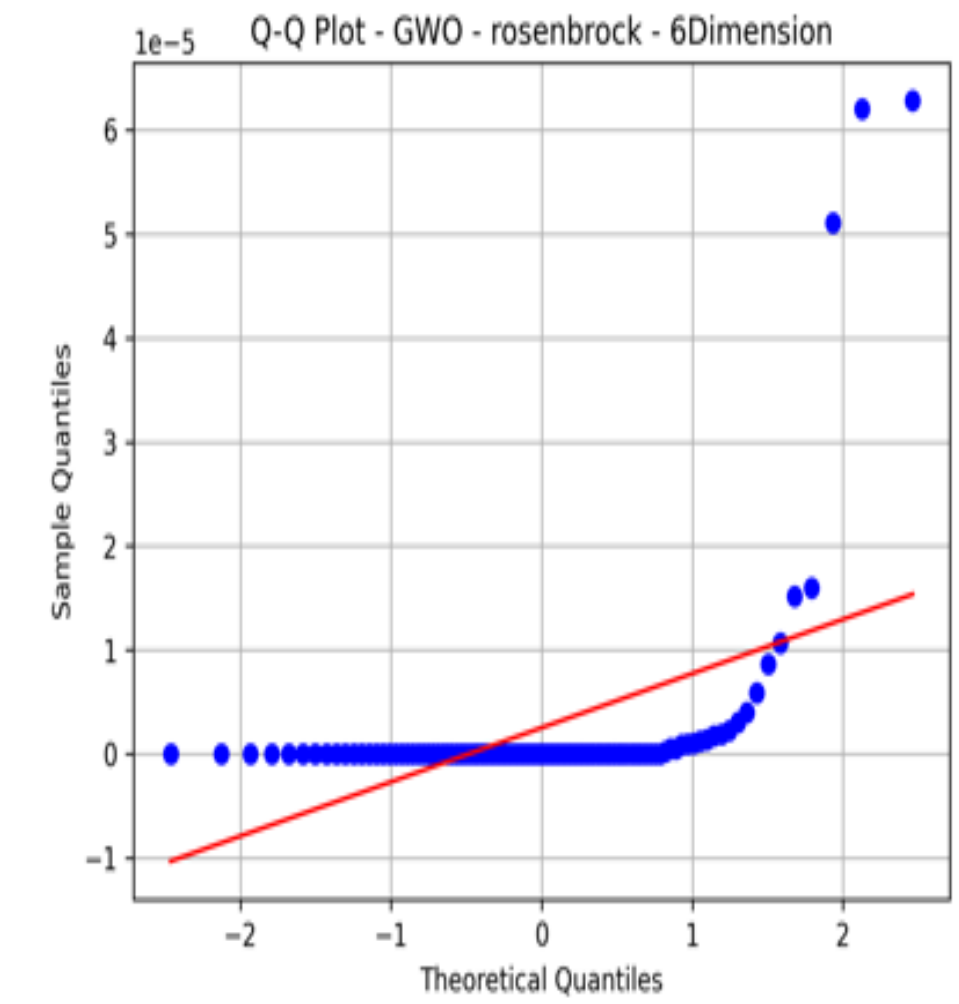
Q-Q Plot - GWO - rosenbrock - 6Dimension
1e−5
Sample Quantiles
Theoretical Quantiles

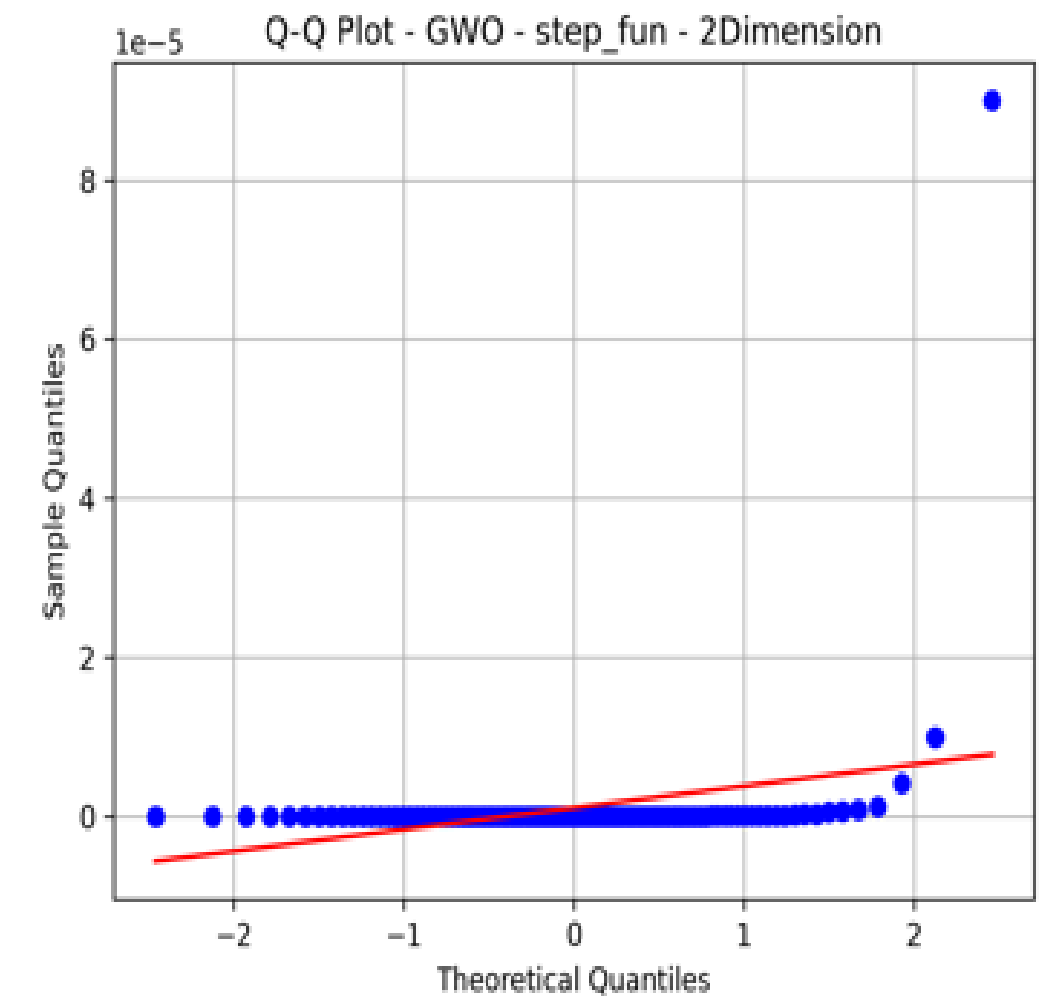
Q-Q Plot - GWO - step_fun - 2Dimension
1e−5
Sample Quantiles
Theoretical Quantiles

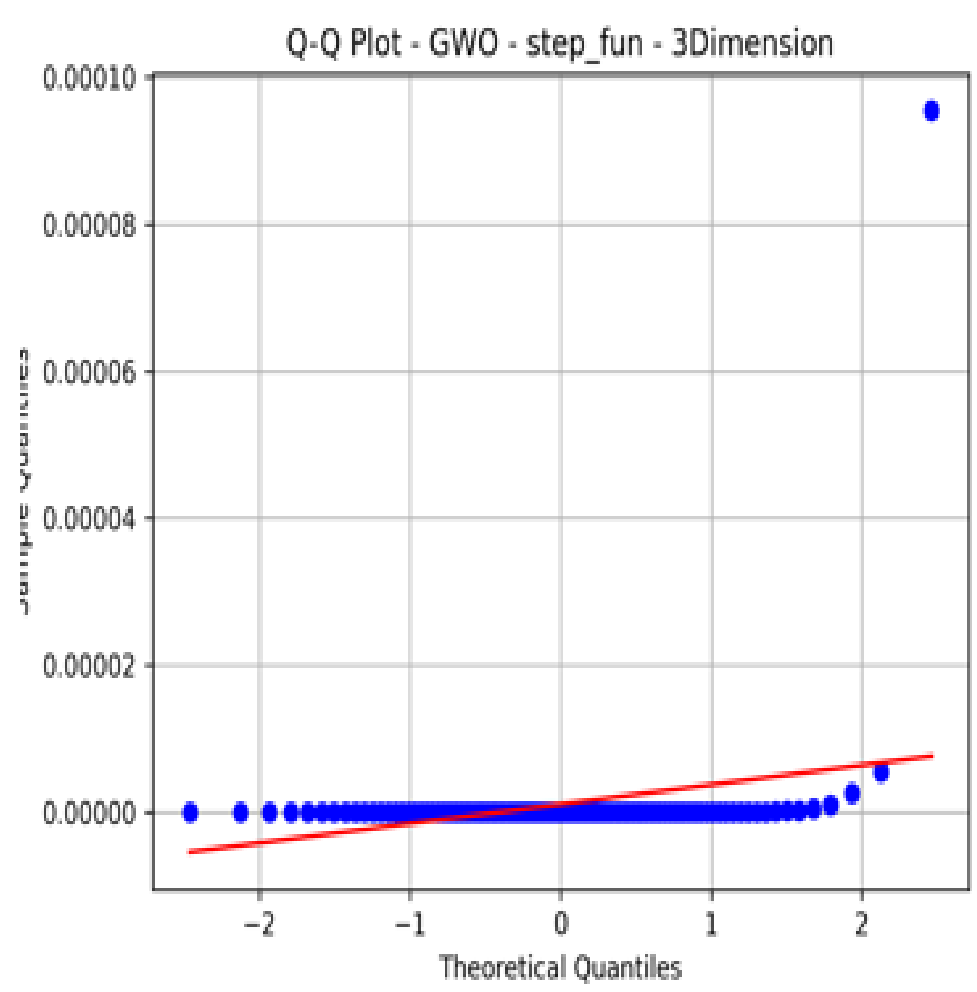
Q-Q Plot - GWO - step_fun - 3Dimension
Theoretical Quantiles

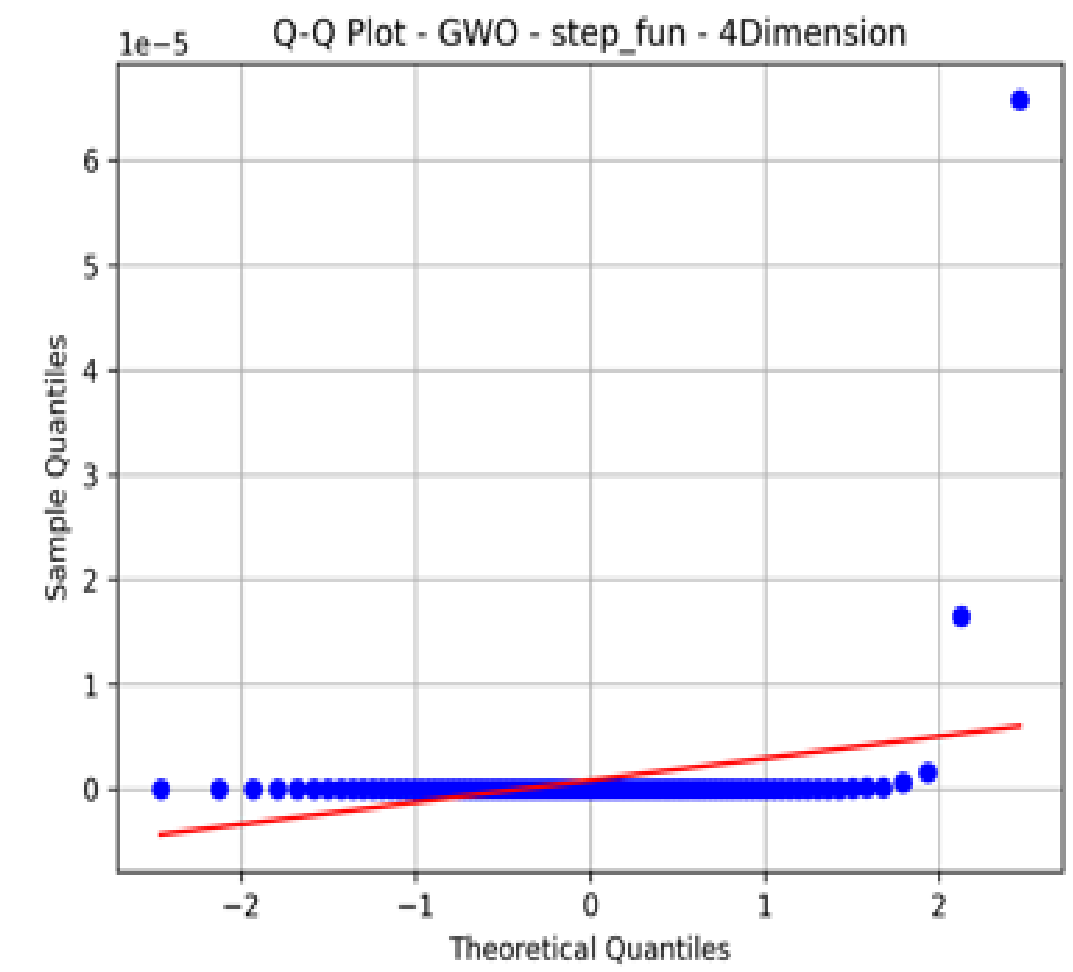
Q-Q Plot - GWO - step_fun - 4Dimension
1e−5
Sample Quantiles
Theoretical Quantiles

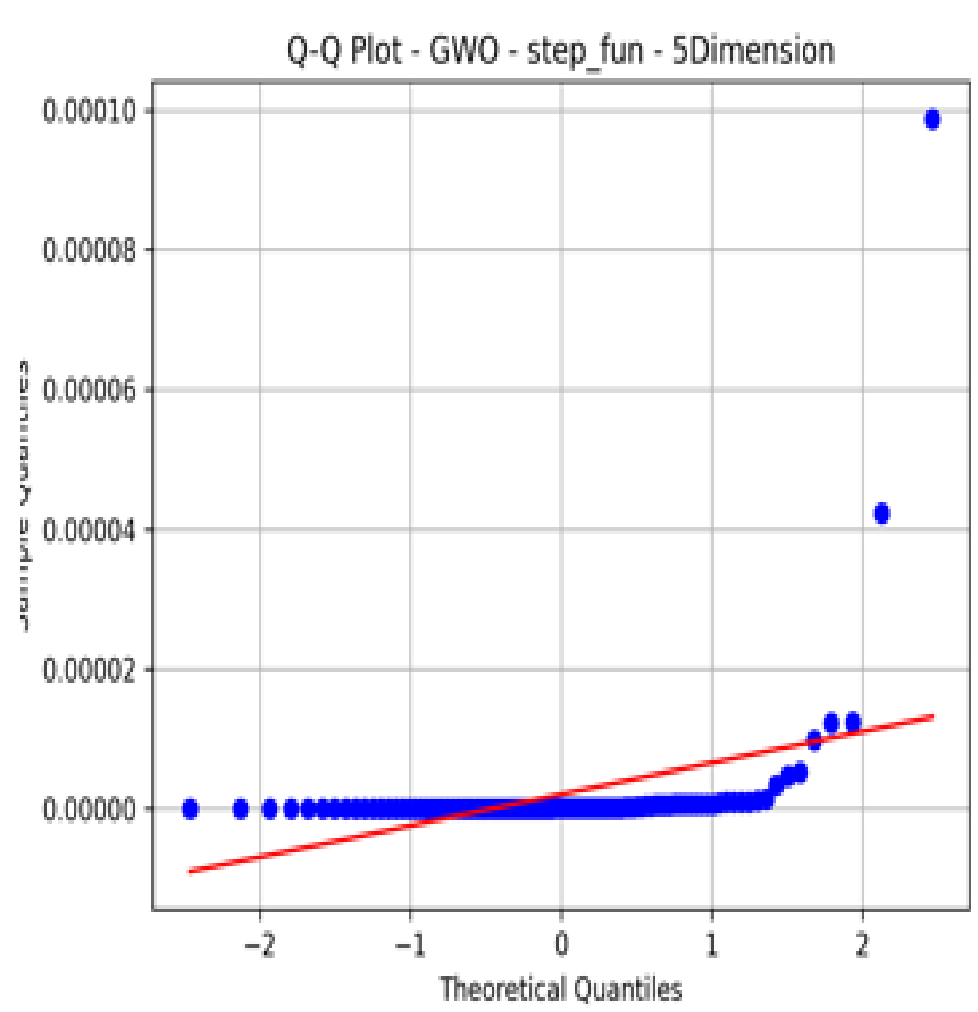
Q-Q Plot - GWO - step_fun - 5Dimension
Theoretical Quantiles

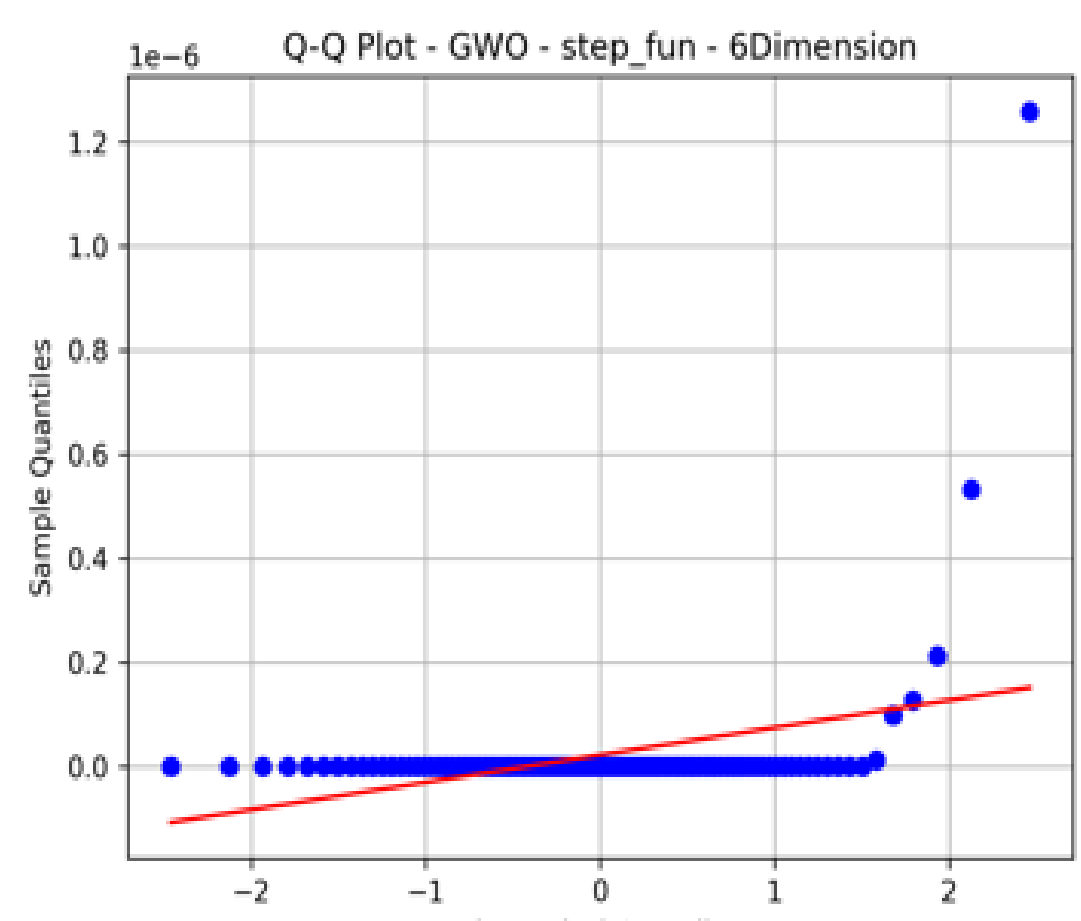
Q-Q Plot - GWO - step_fun - 6Dimension
1e−6
Sample Quantiles
Theoretical Quantiles

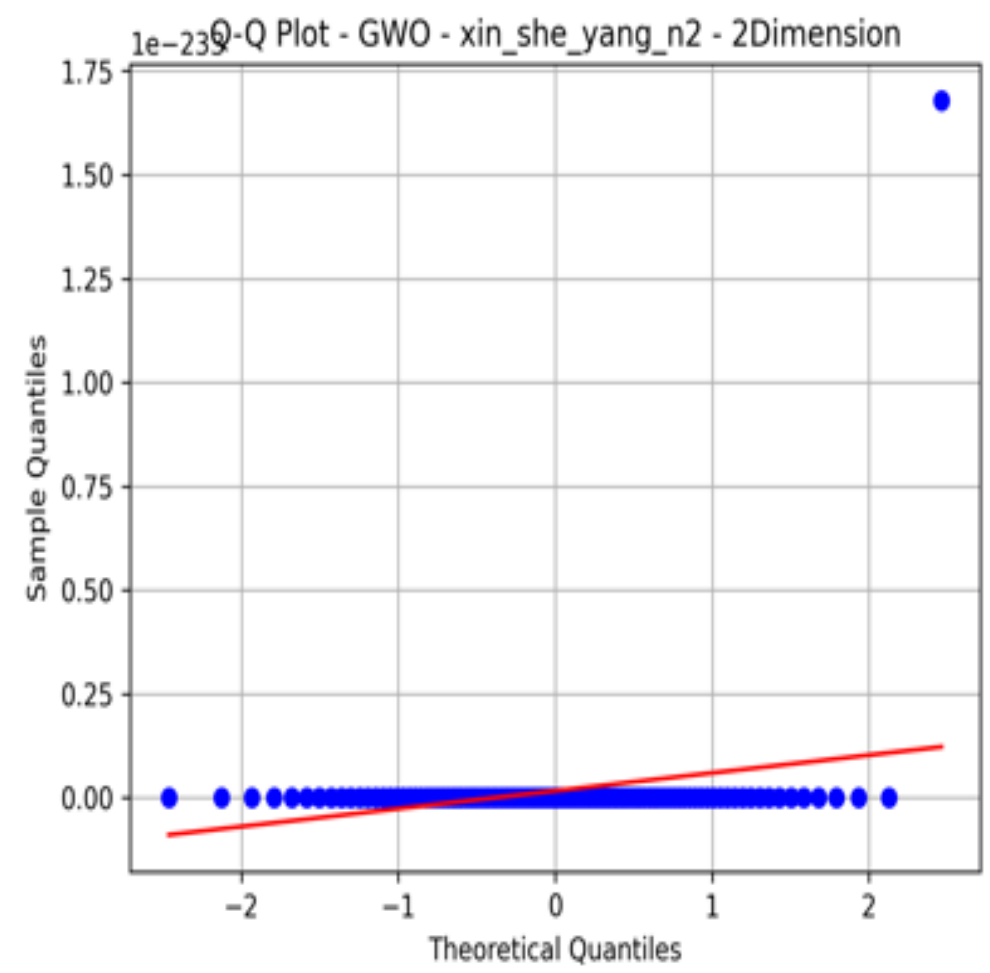
Q-Q Plot - GWO - xin_she_yang_n2 - 2Dimension
Sample Quantiles
Theoretical Quantiles
1.75
1.50
1.25
1.00
0.75
0.50
0.25
0.00
−2
−1
0
1
2

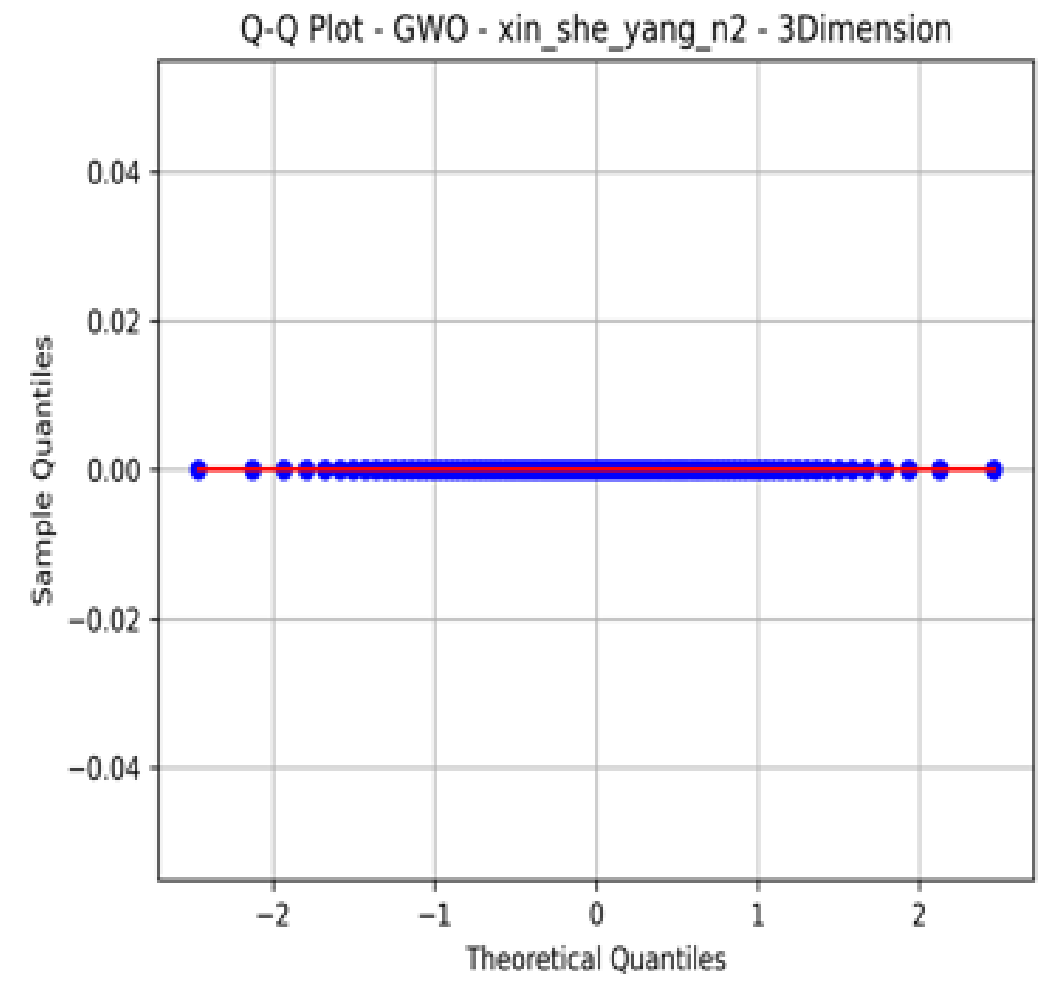
Q-Q Plot - GWO - xin_she_yang_n2 - 3Dimension
Sample Quantiles
Theoretical Quantiles
0.04
0.02
0.00
−0.02
−0.04
−2
−1
0
1
2

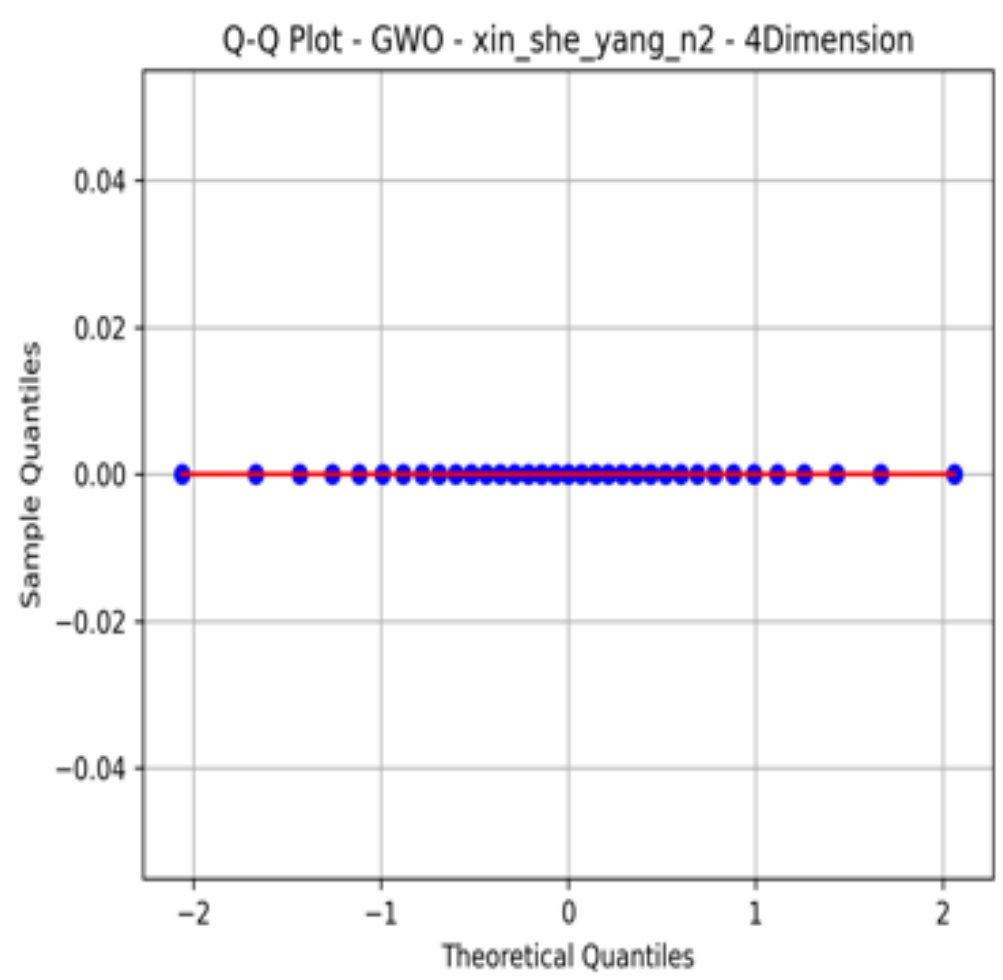
Q-Q Plot - GWO - xin_she_yang_n2 - 4Dimension
Sample Quantiles
Theoretical Quantiles
0.04
0.02
0.00
−0.02
−0.04
−2
−1
0
1
2

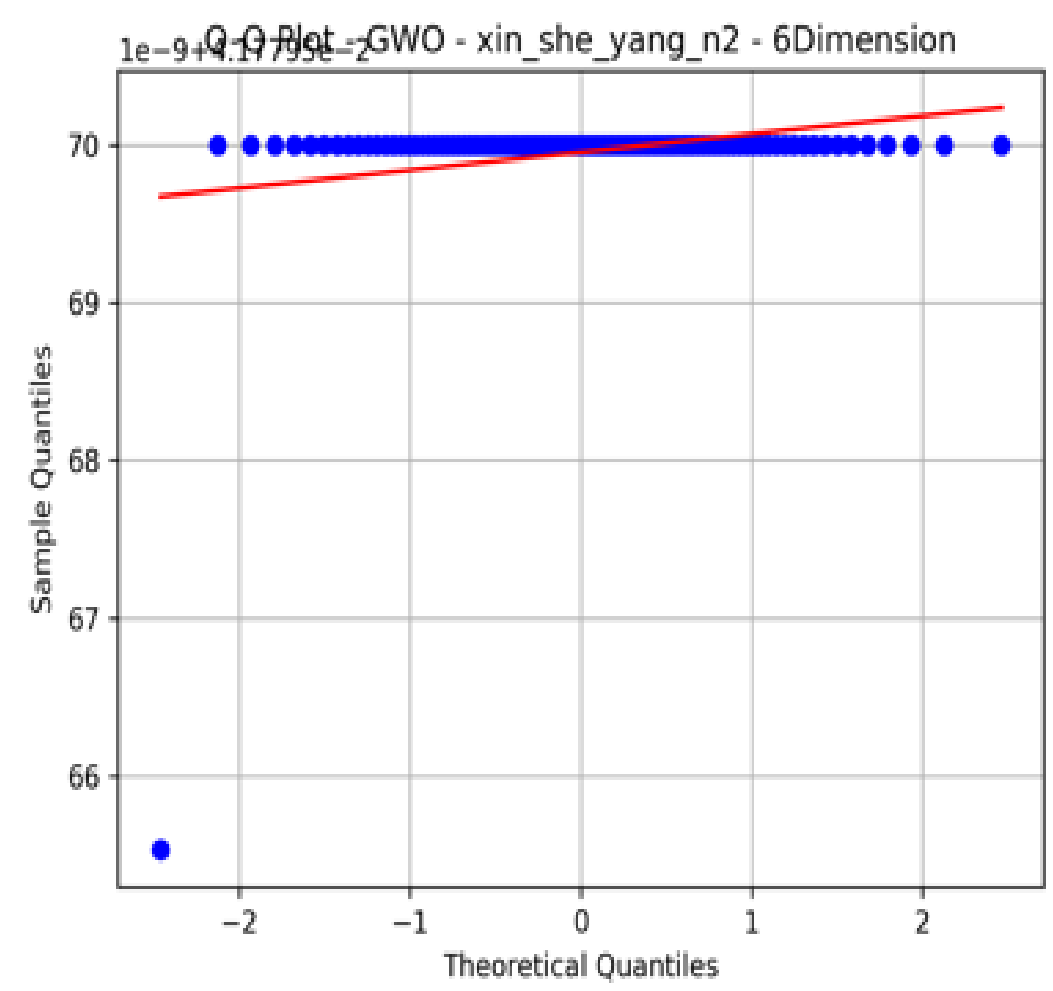
GWO - xin_she_yang_n2 - 6Dimension
Sample Quantiles
Theoretical Quantiles
70
69
68
67
66
−2
−1
0
1
2

# APPENDIX K: CASE STUDY 3 QUANTILE-QUANTILE(Q-Q) PLOTS FOR RESULT DATASET – UNBOUNDED KNAPSACK

In this appendix, we have the Q-Q plots for Unbounded knapsack optimization problem in multiple dimensions(2,3,4,5,6)D as per Case Study 3, which validates that the generated result datasets are not normally distributed . In each plot, the blue line represents the data set generated via experiments and red line denotes the benchmark normal-distribution line. Hence, a deviation in blue line from red line represents the non-normally distributed data. Each plot is labelled as follows: **metaheuristic name - optimization function name-dimension.**

<u>BSO Q-Q Plots</u>

To generate the below mentioned plots we have used Table L1 data.

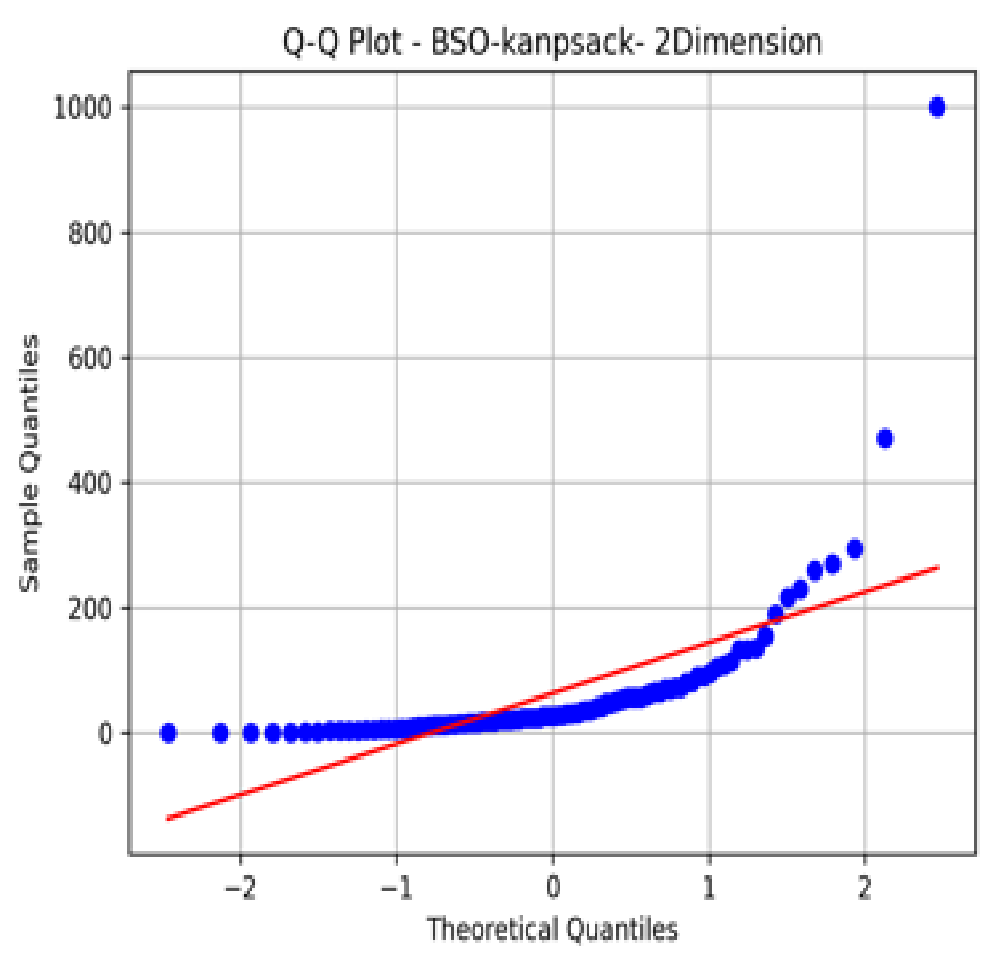

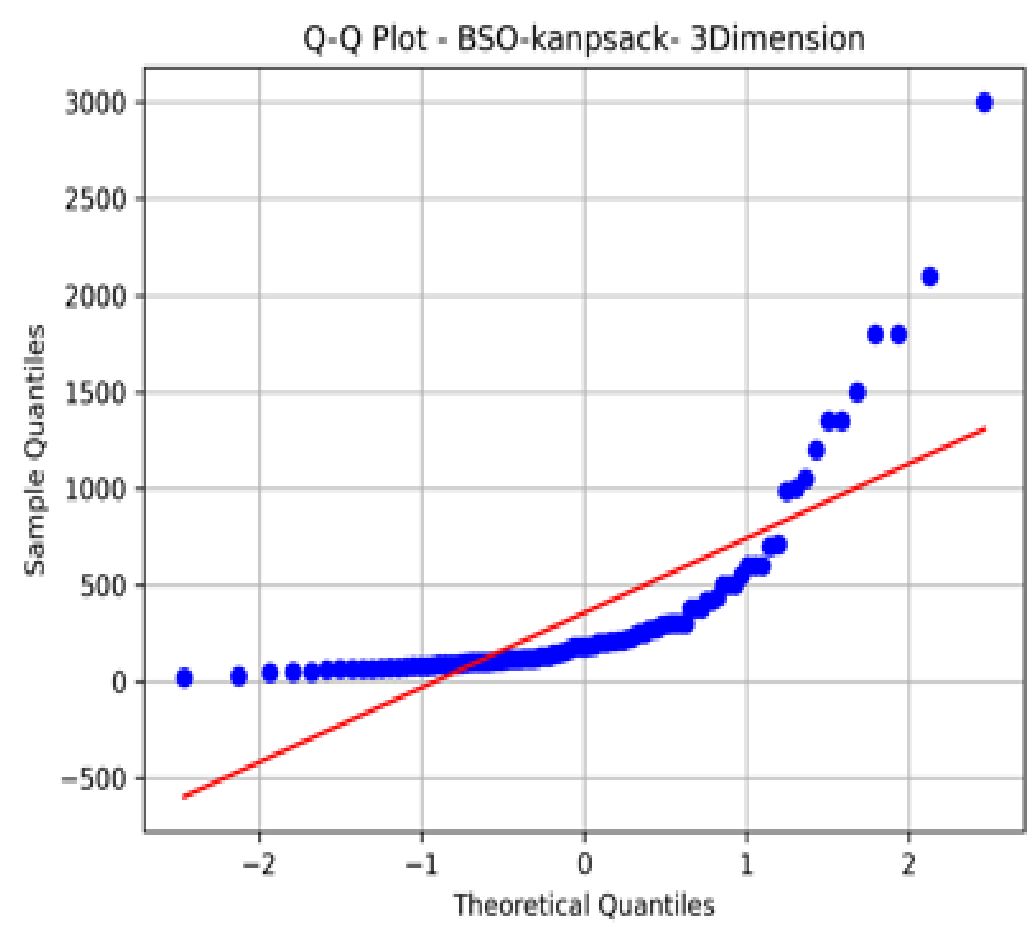

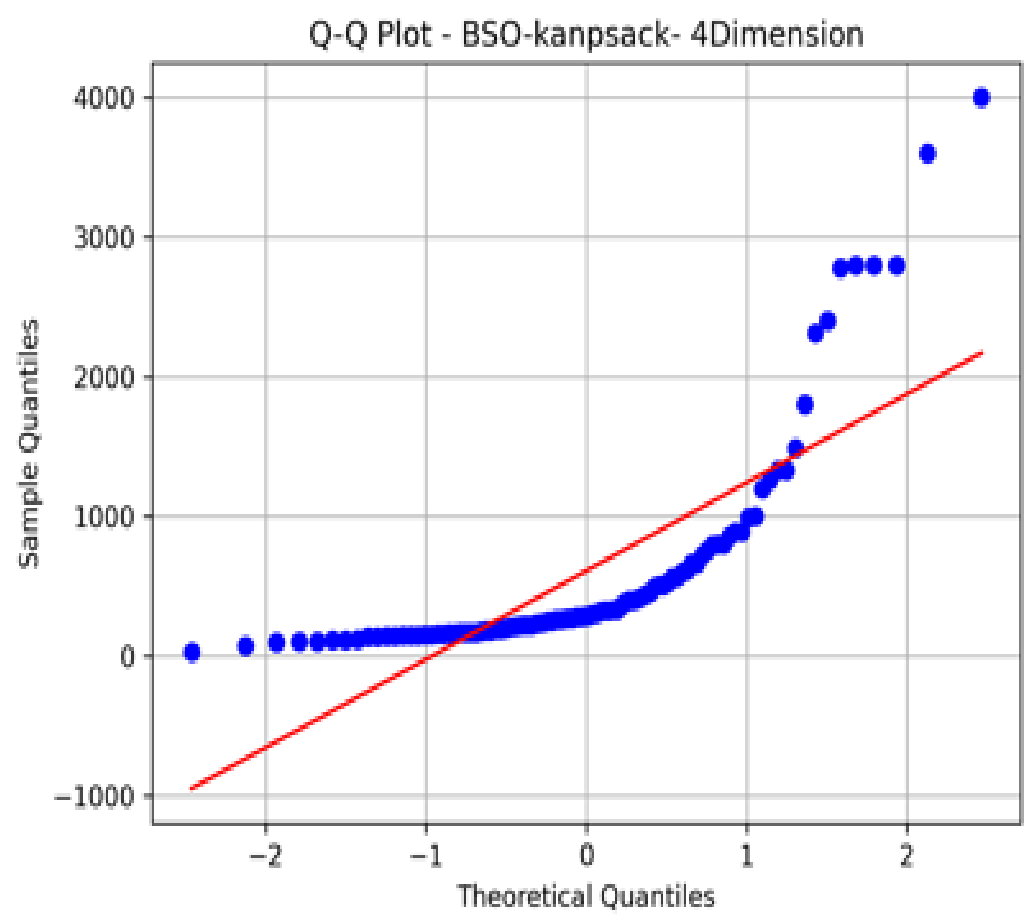

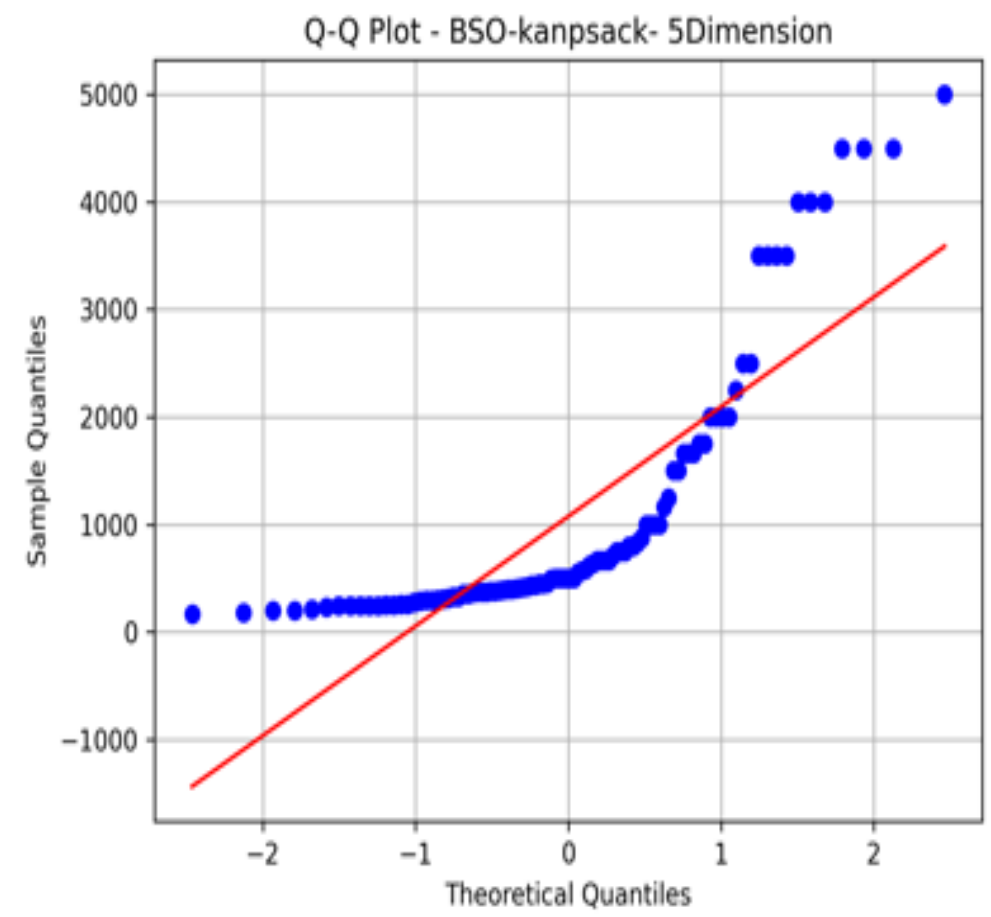

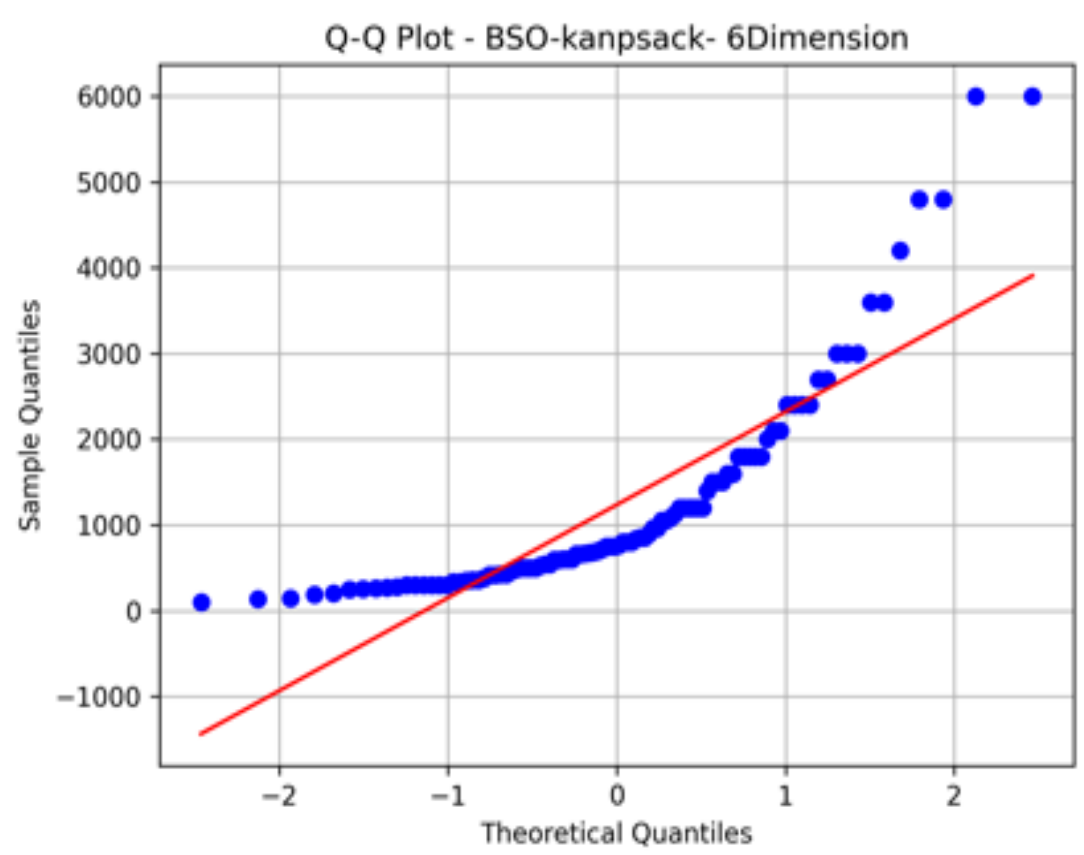

<u>BPO Q-Q Plots</u>

To generate the below mentioned plots we have used Table L2 data

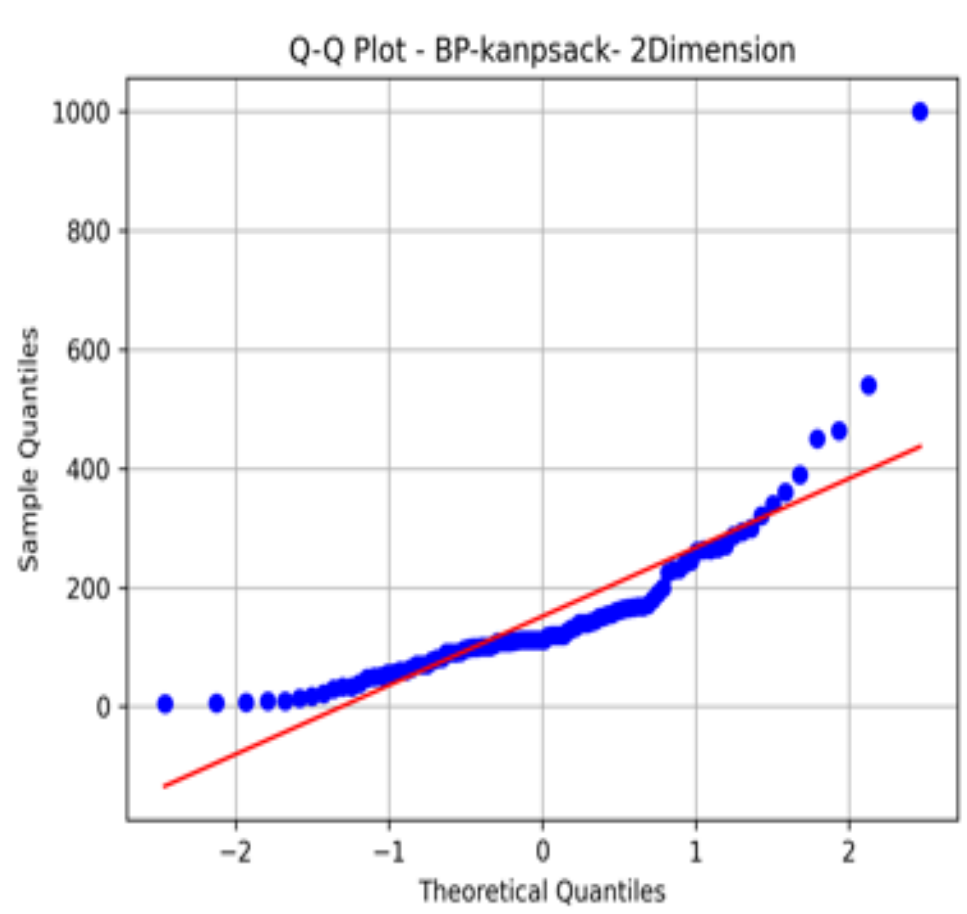

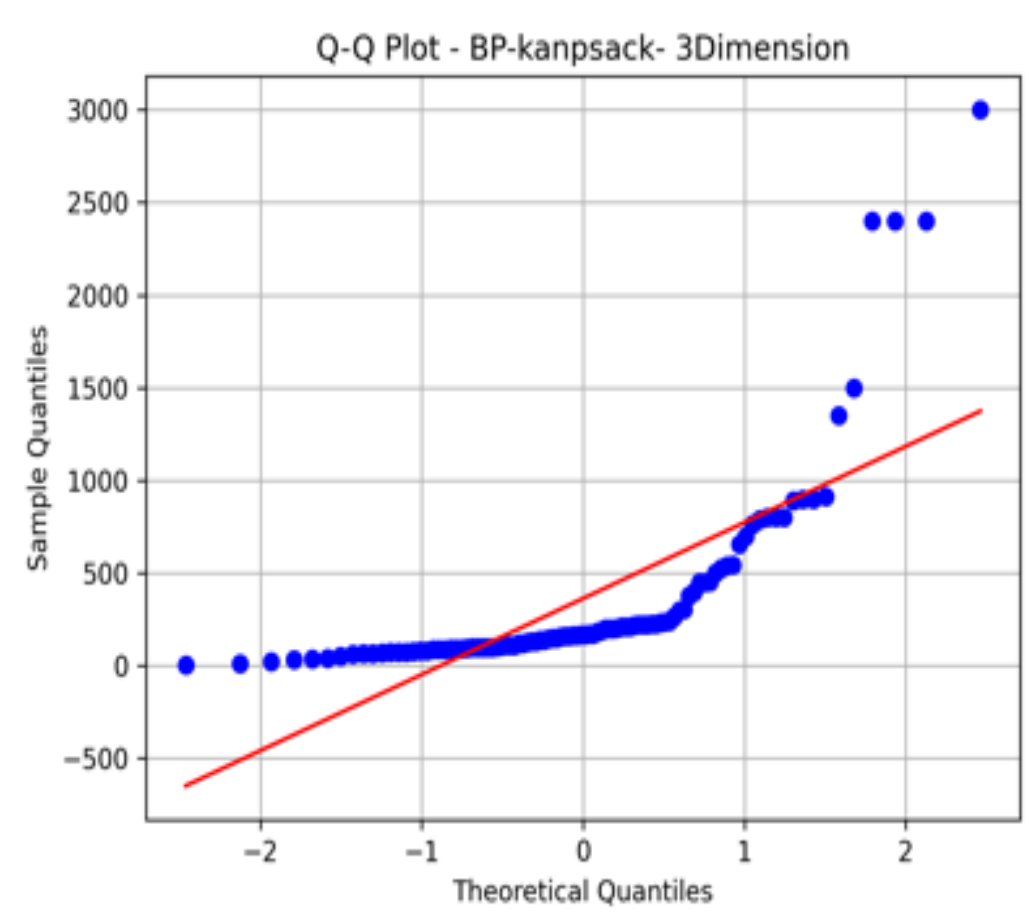

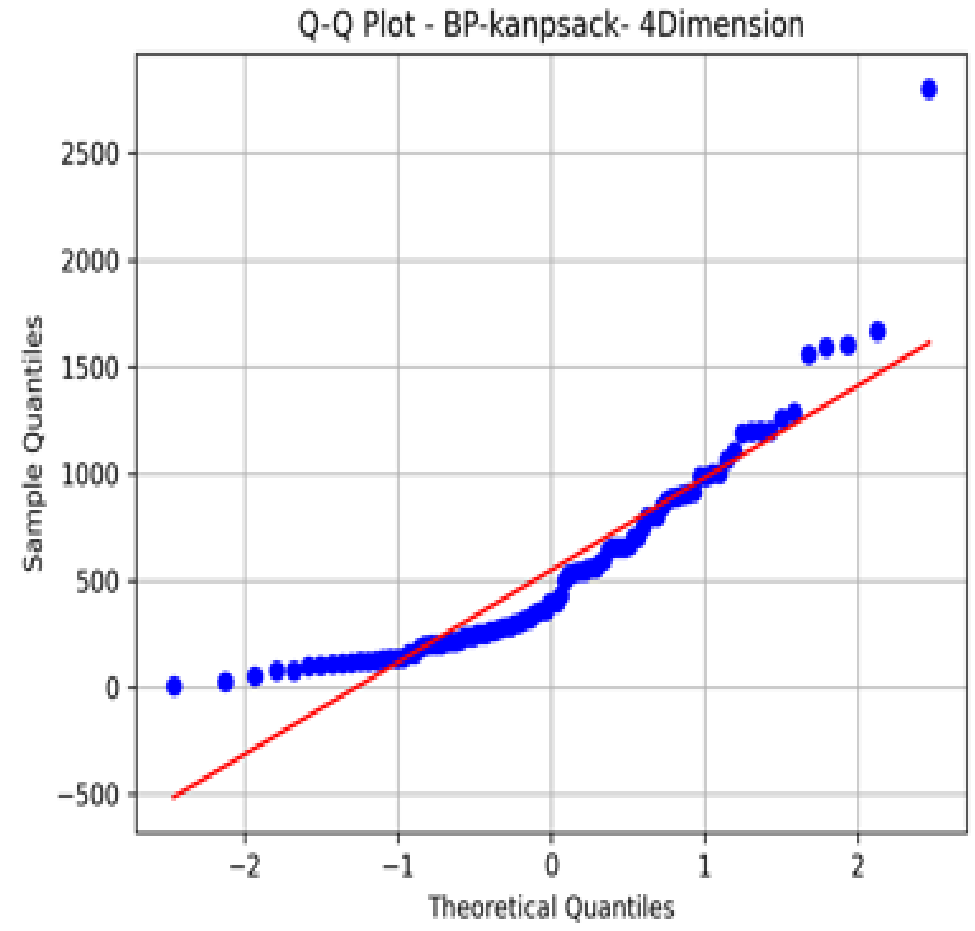

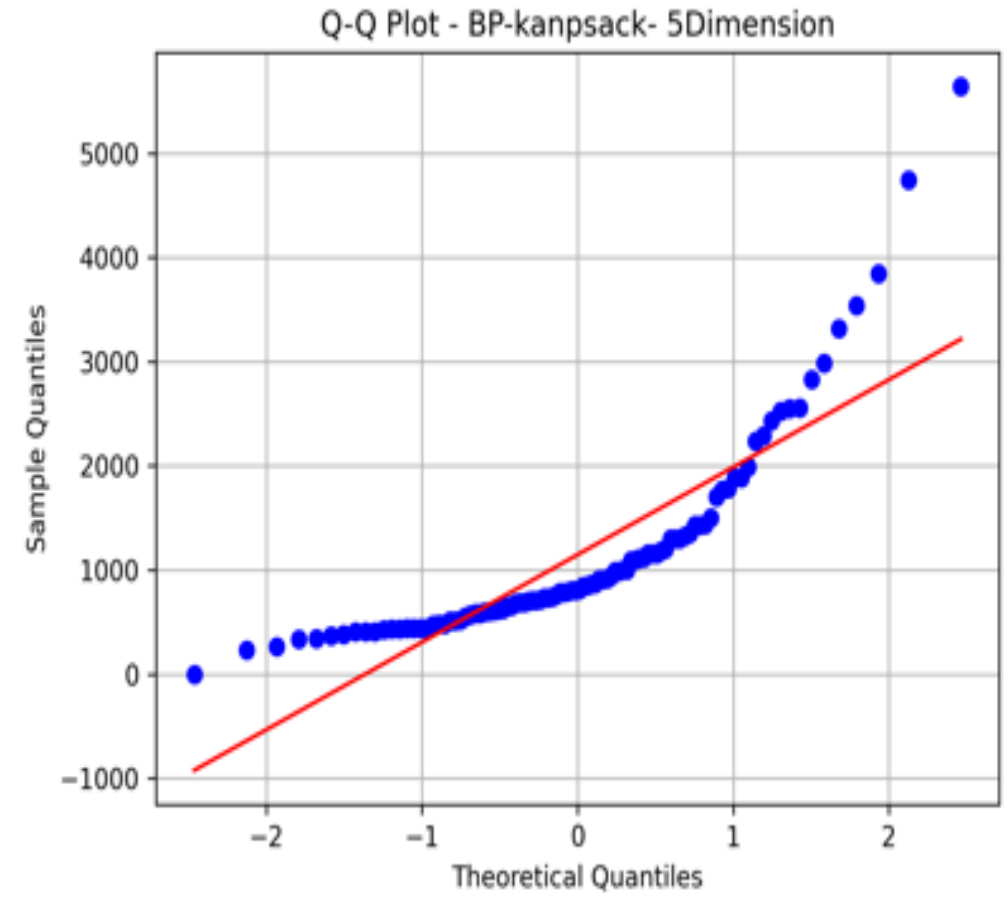

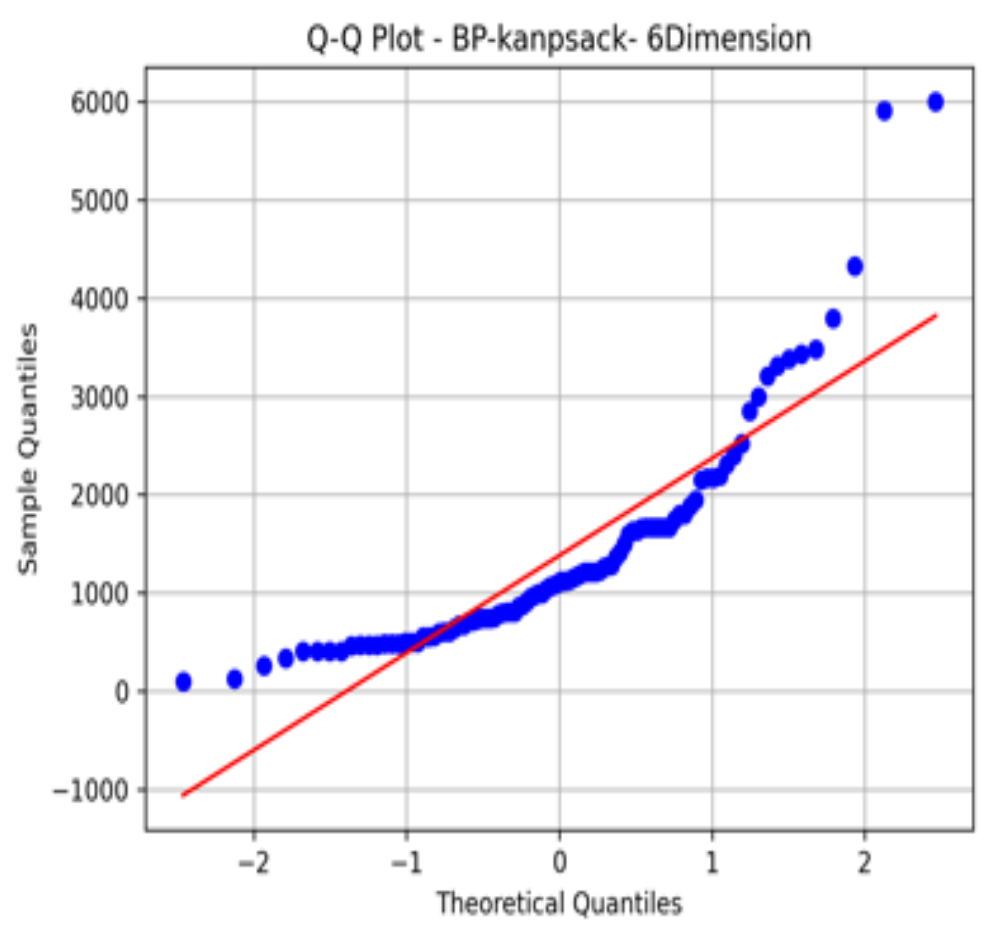

<u>TA Q-Q Plots</u>

To generate the below mentioned plots we have used Table L3 data

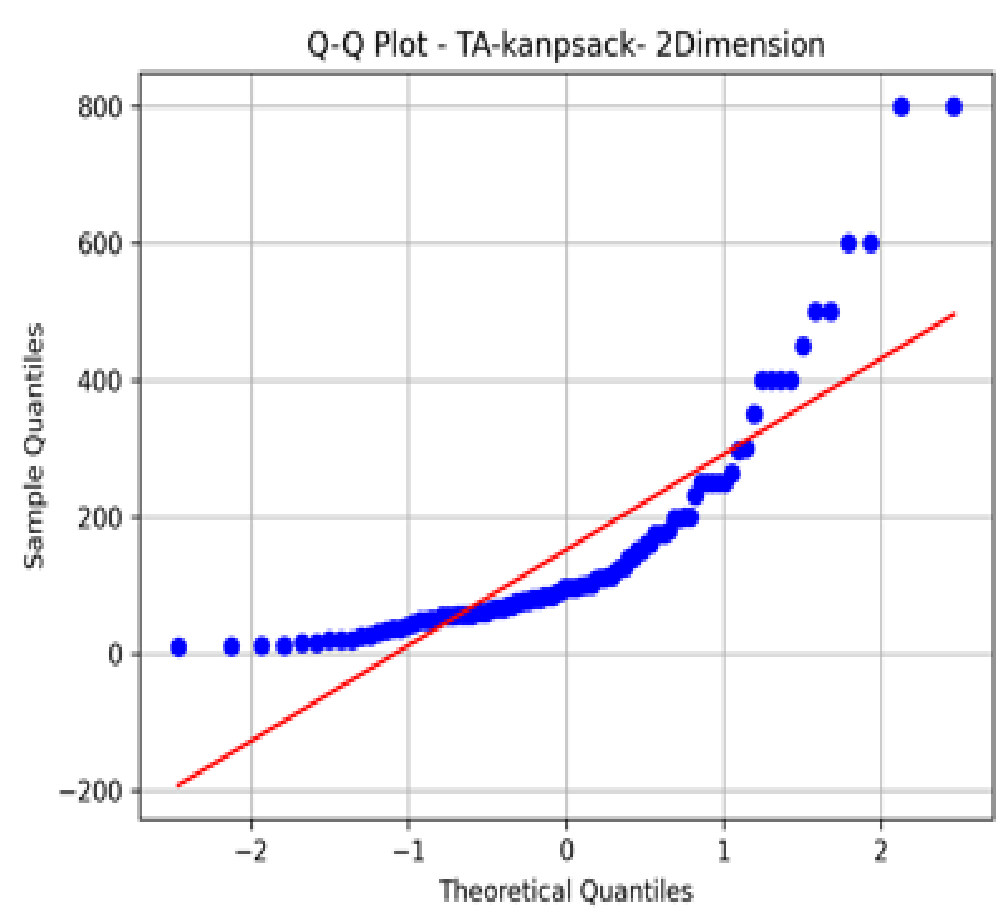

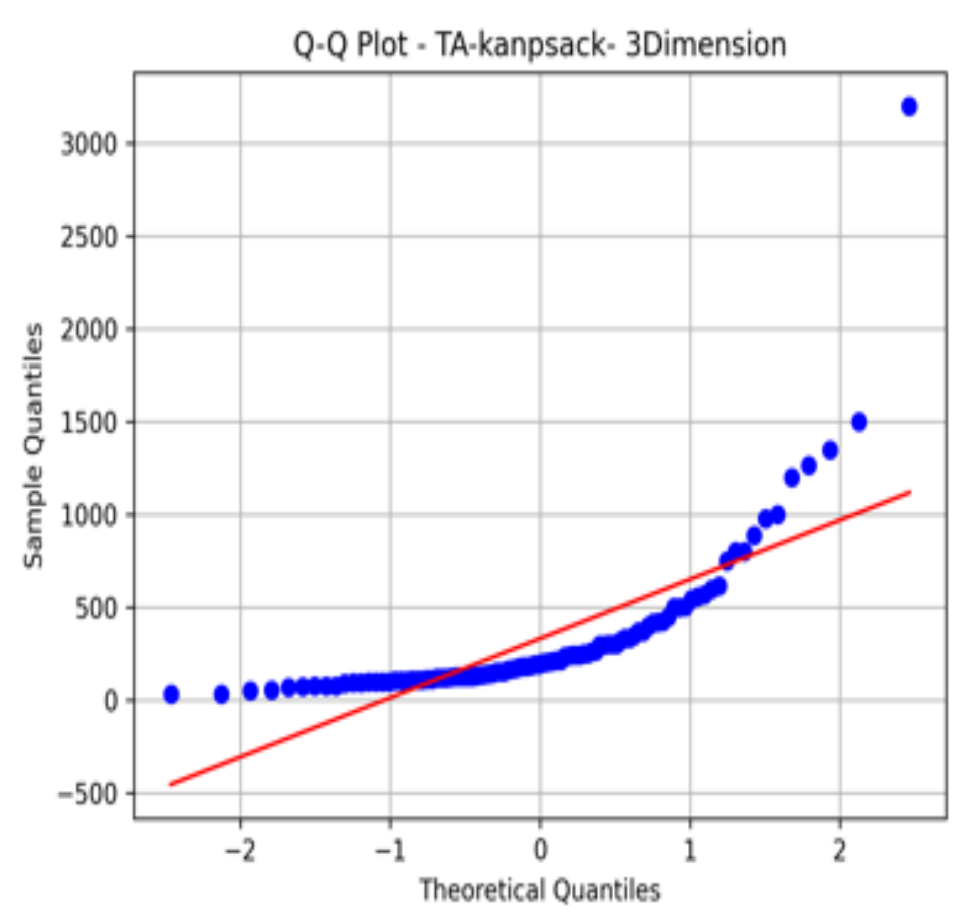

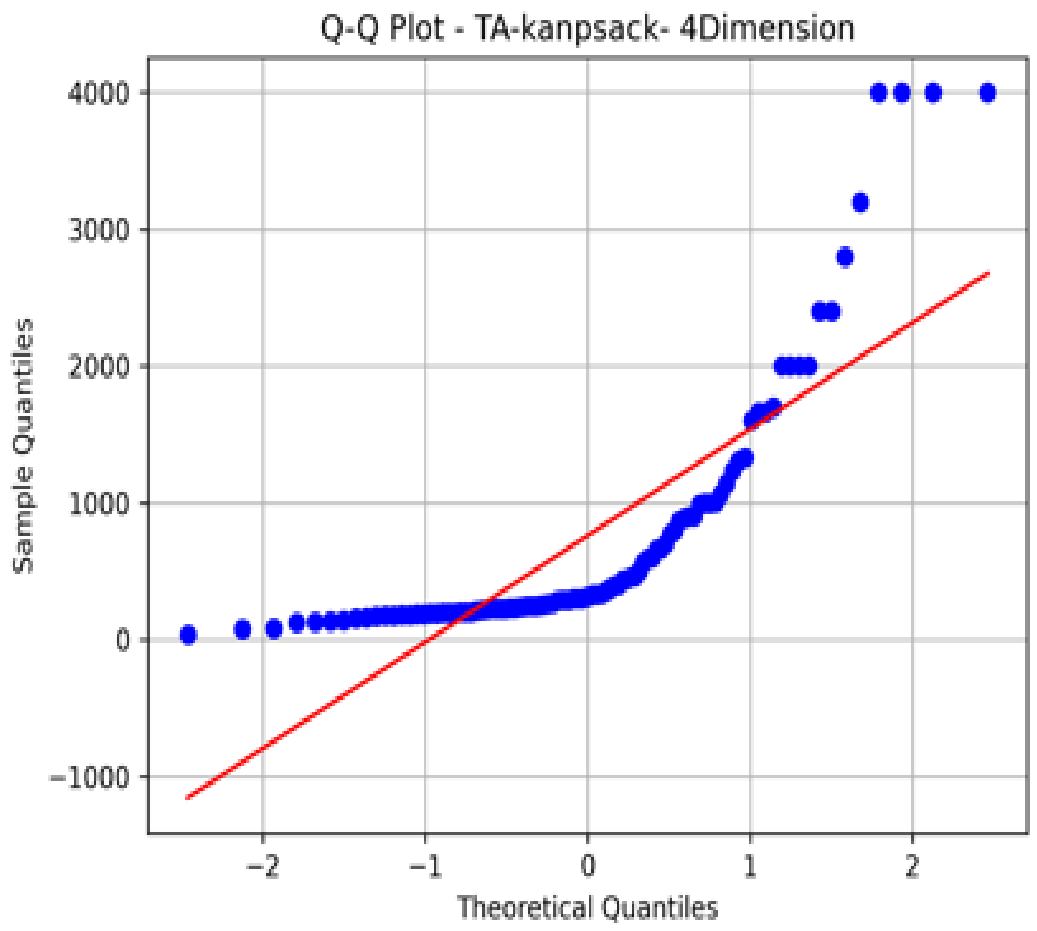

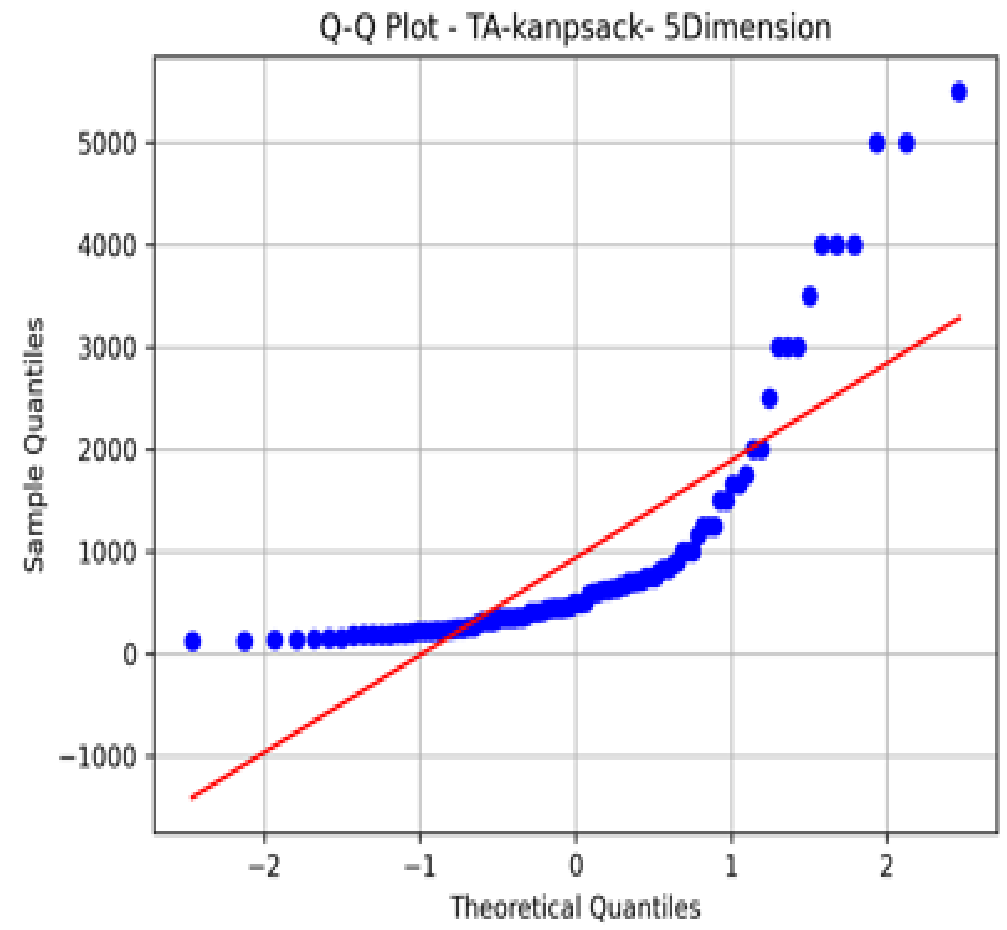

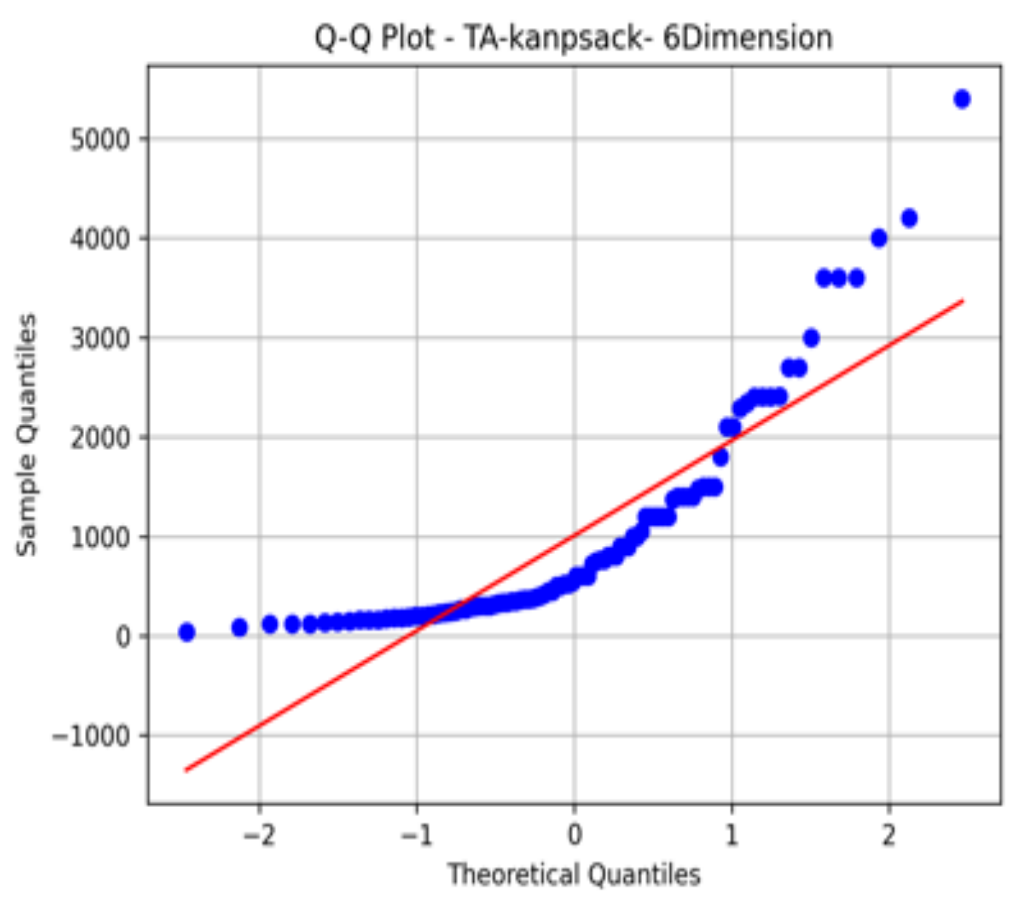

<u>SA Q-Q Plots</u>

To generate the below mentioned plots we have used Table L4 data

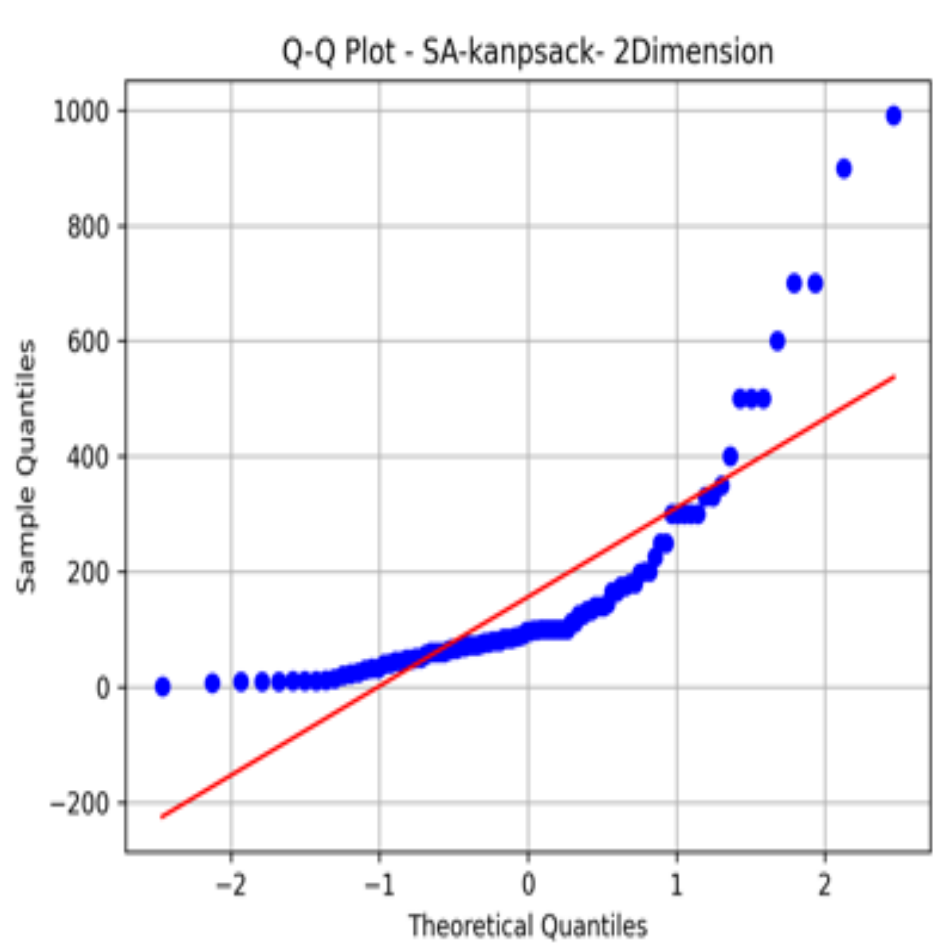

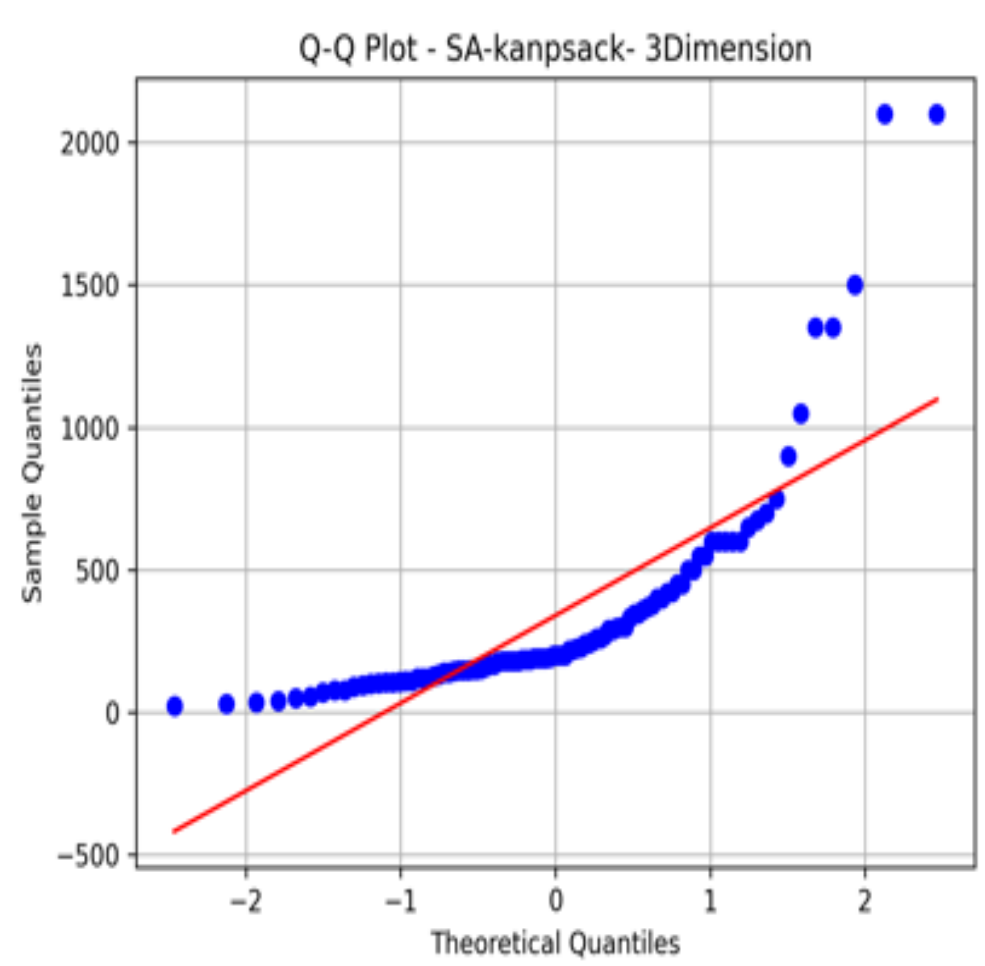

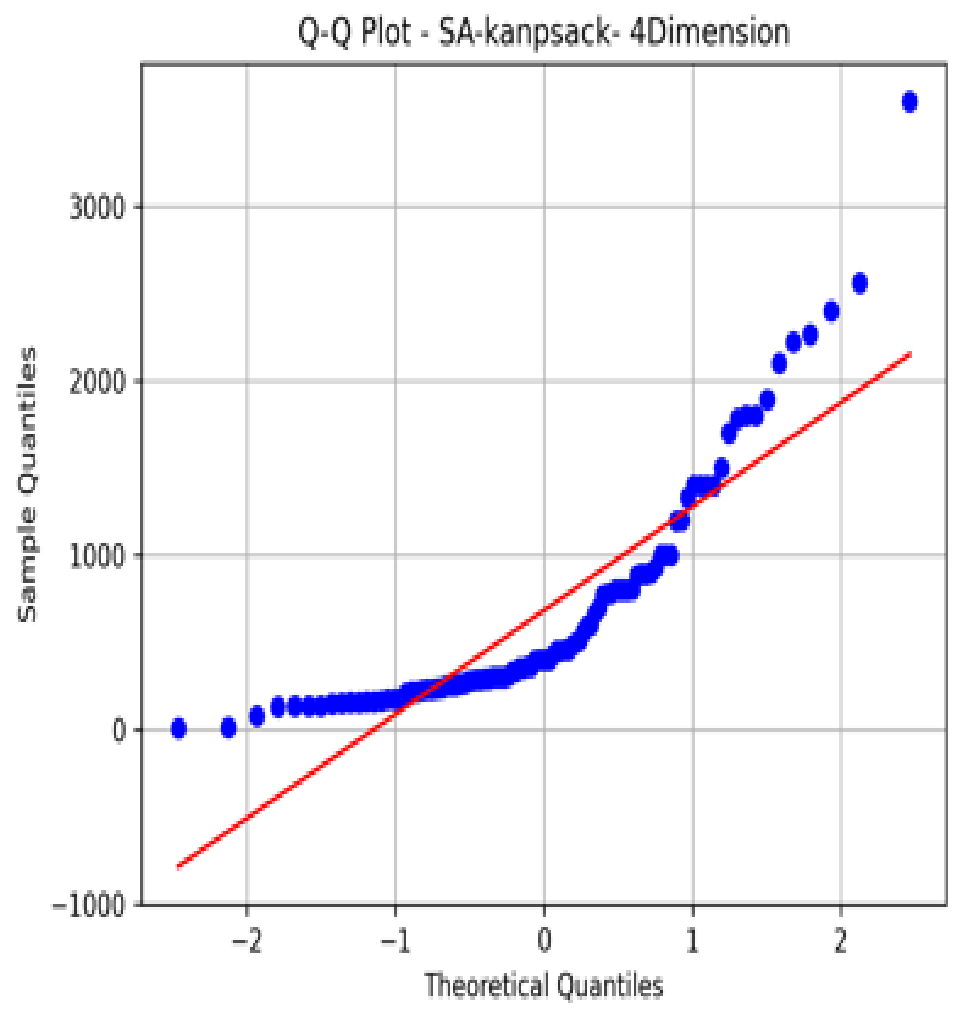

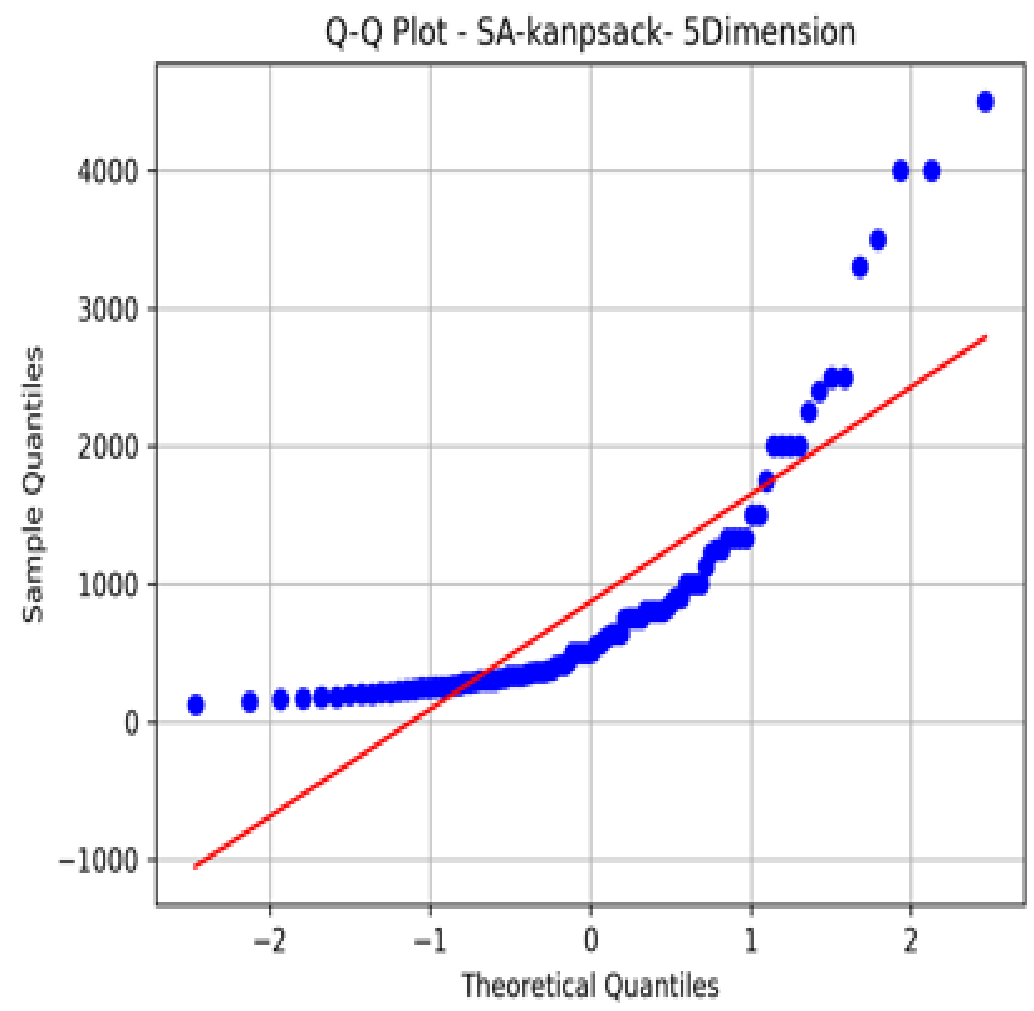

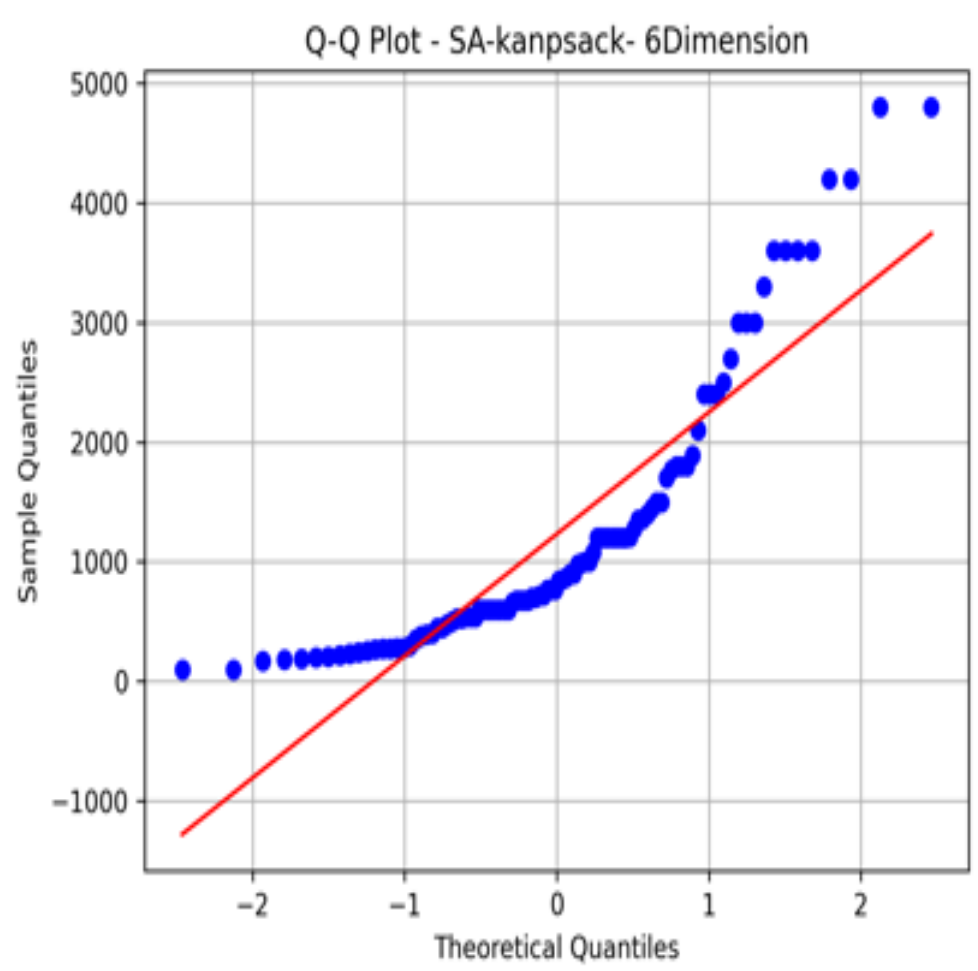

<u>PSO Q-Q Plots</u>

To generate the below mentioned plots we have used Table L5 data

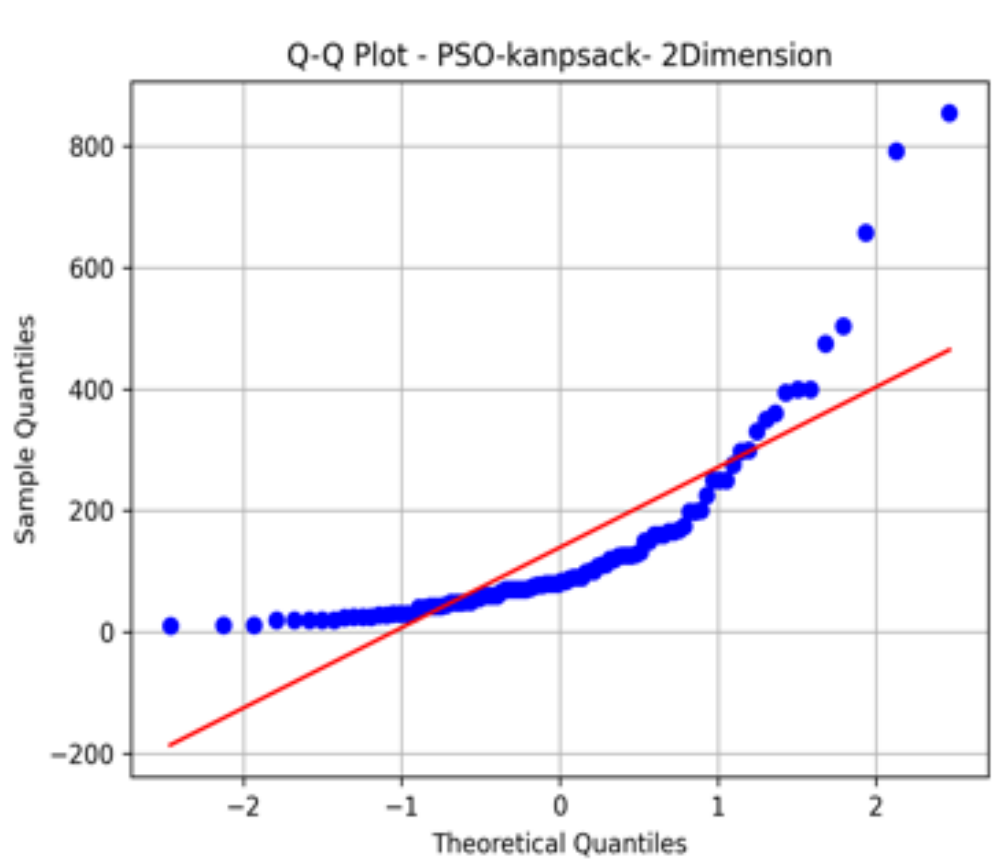

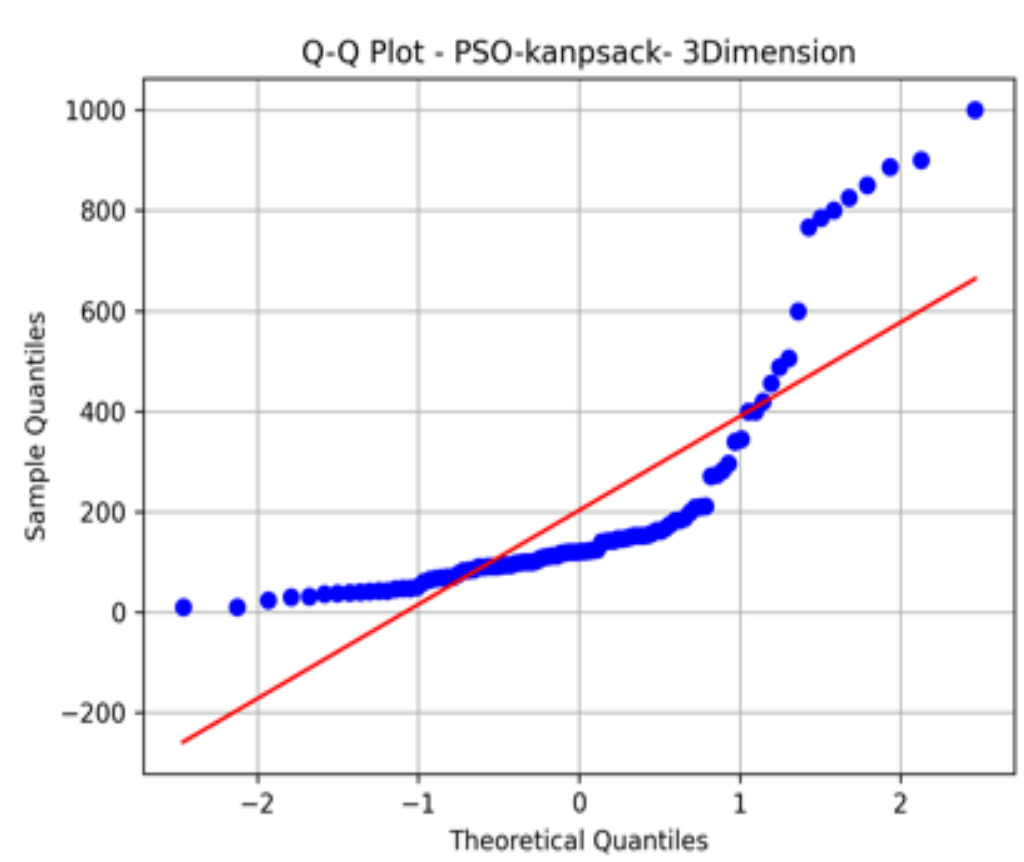

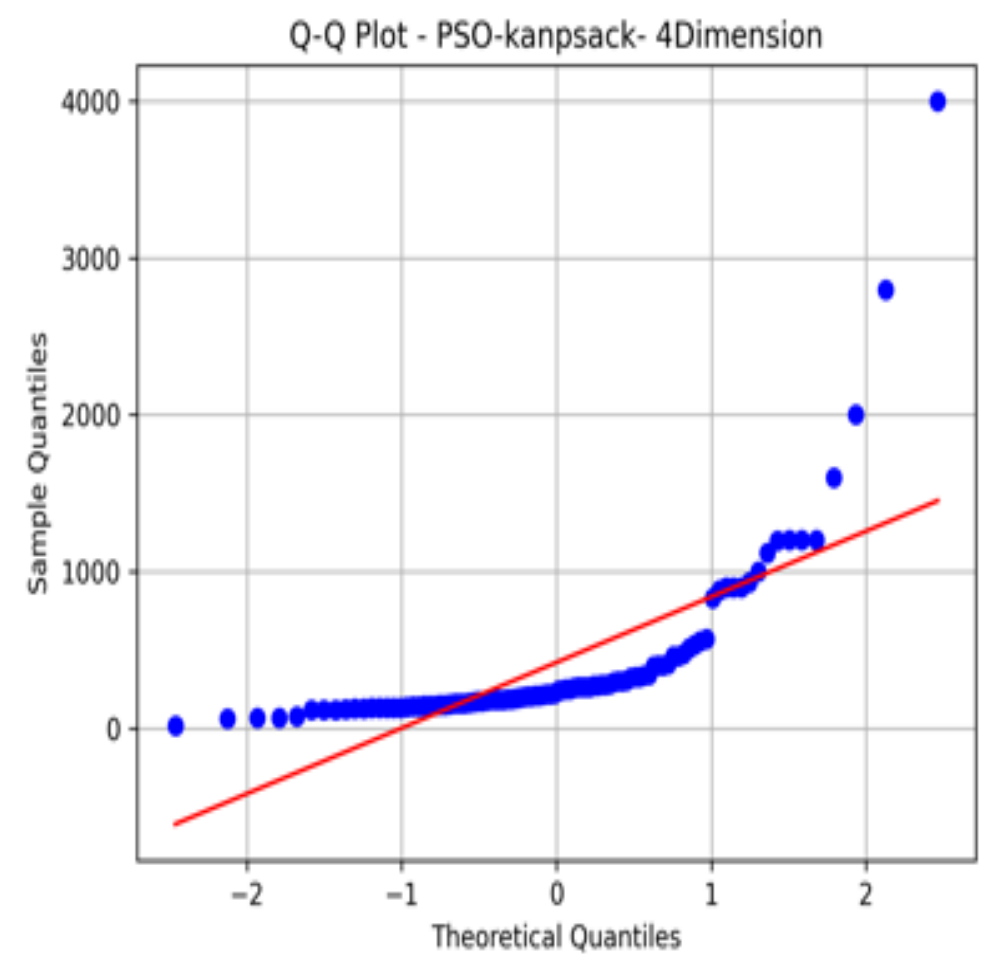

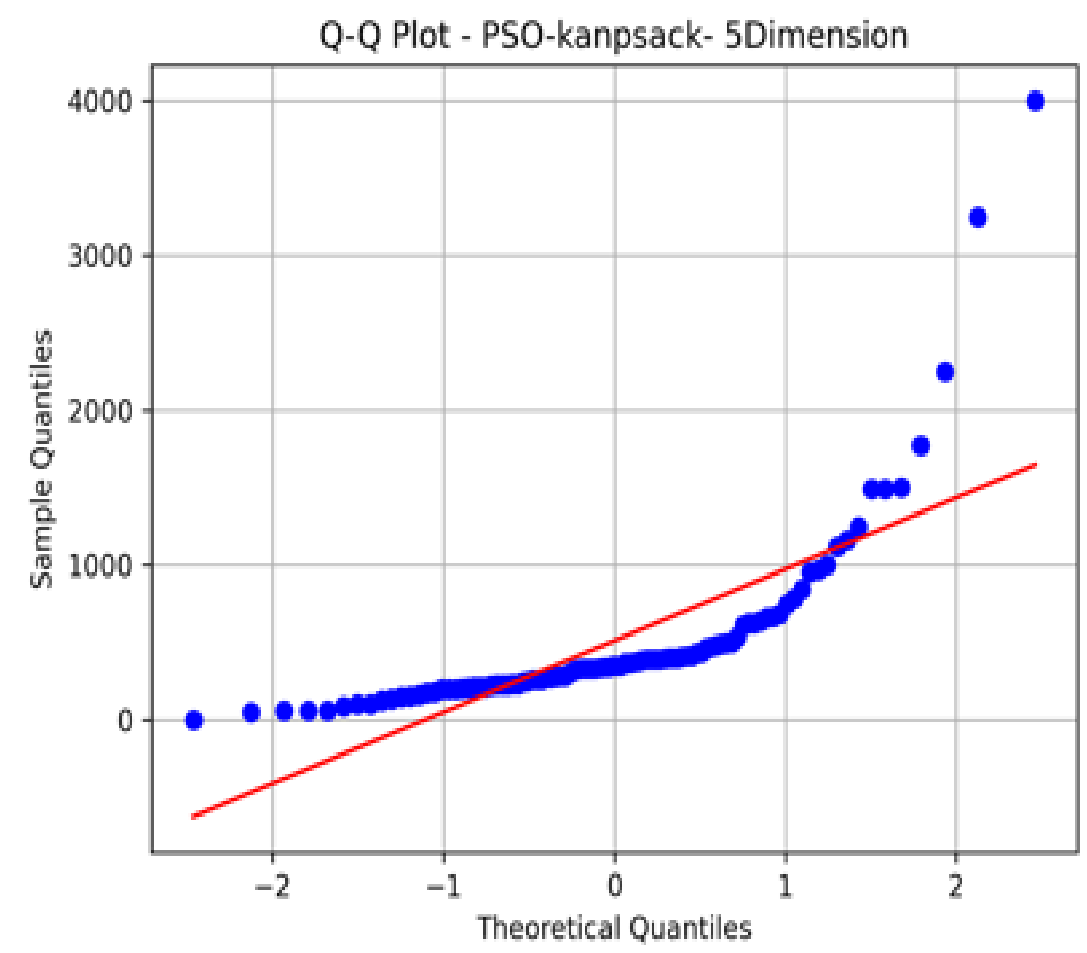

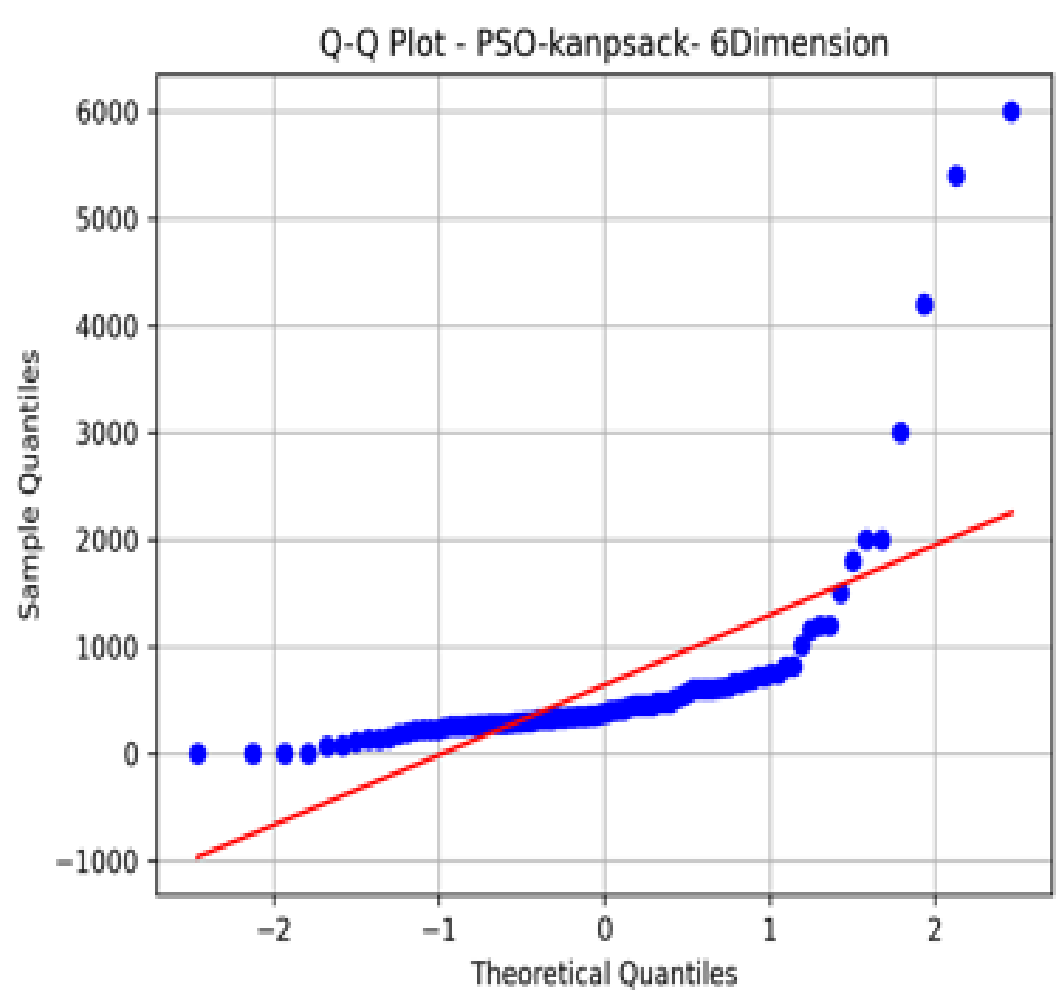

GWO Q-Q Plots

To generate the below mentioned plots we have used Table L7 data

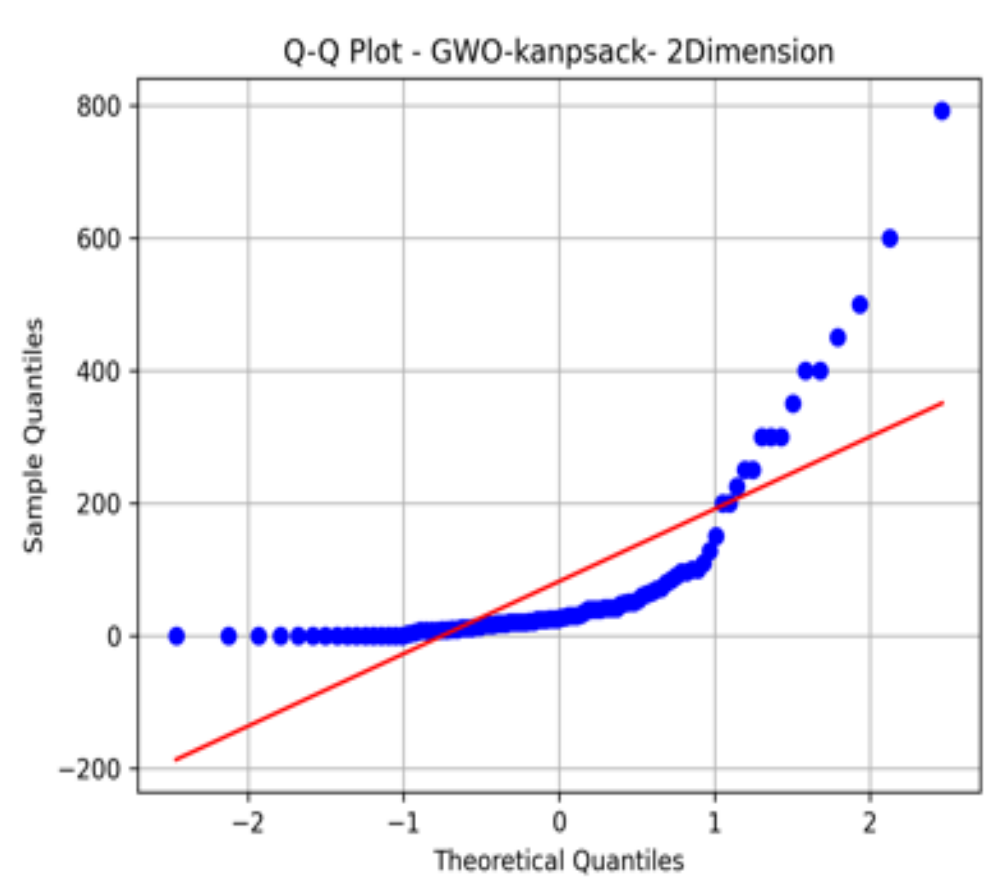

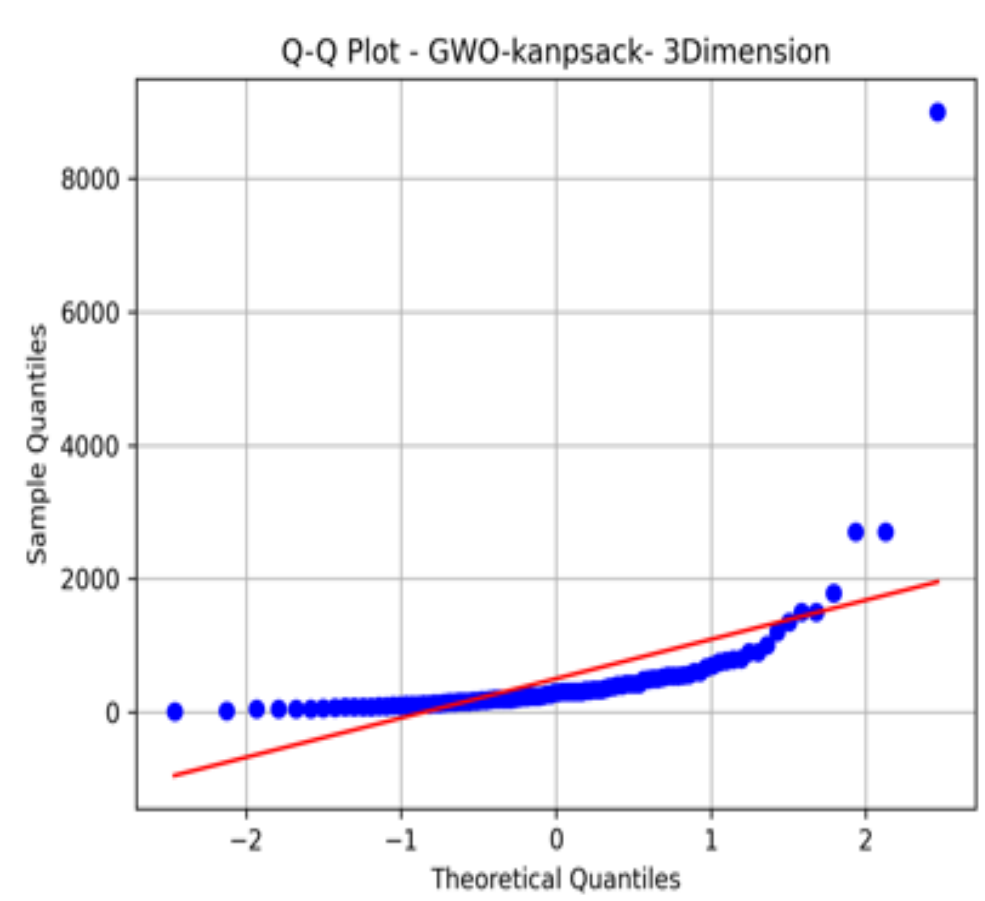

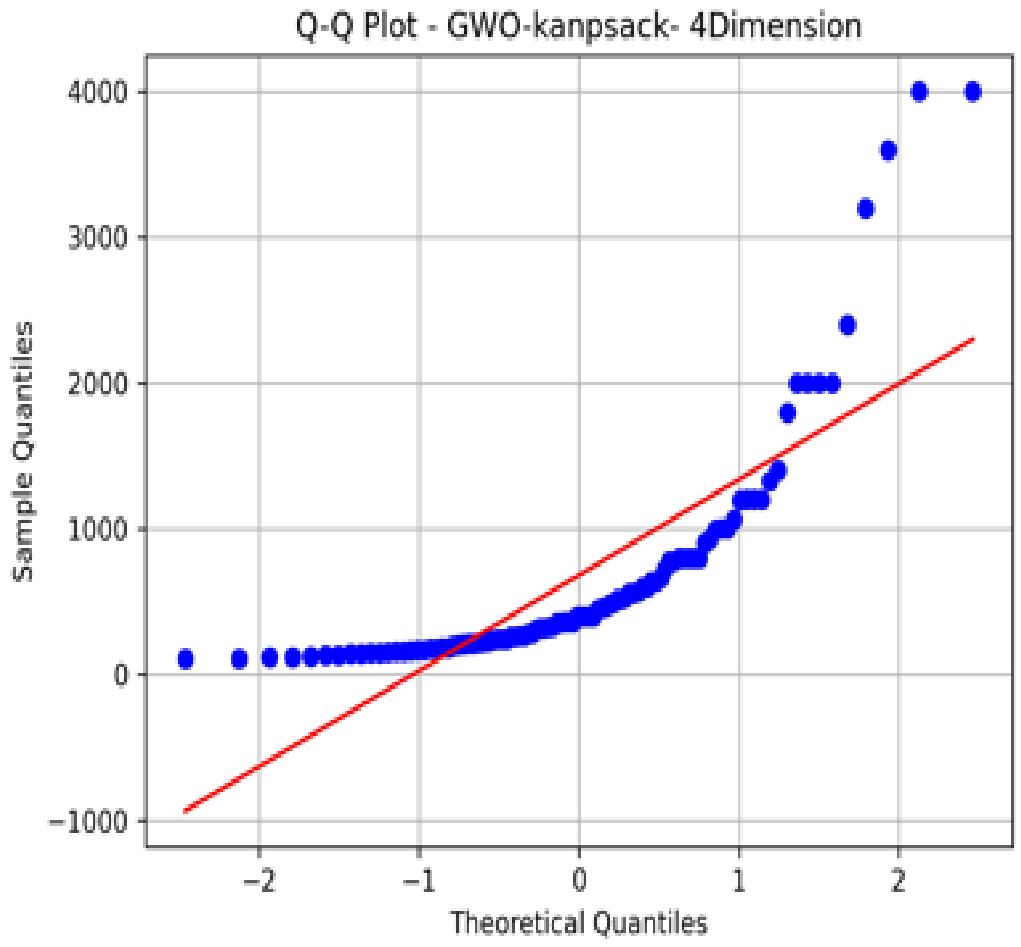

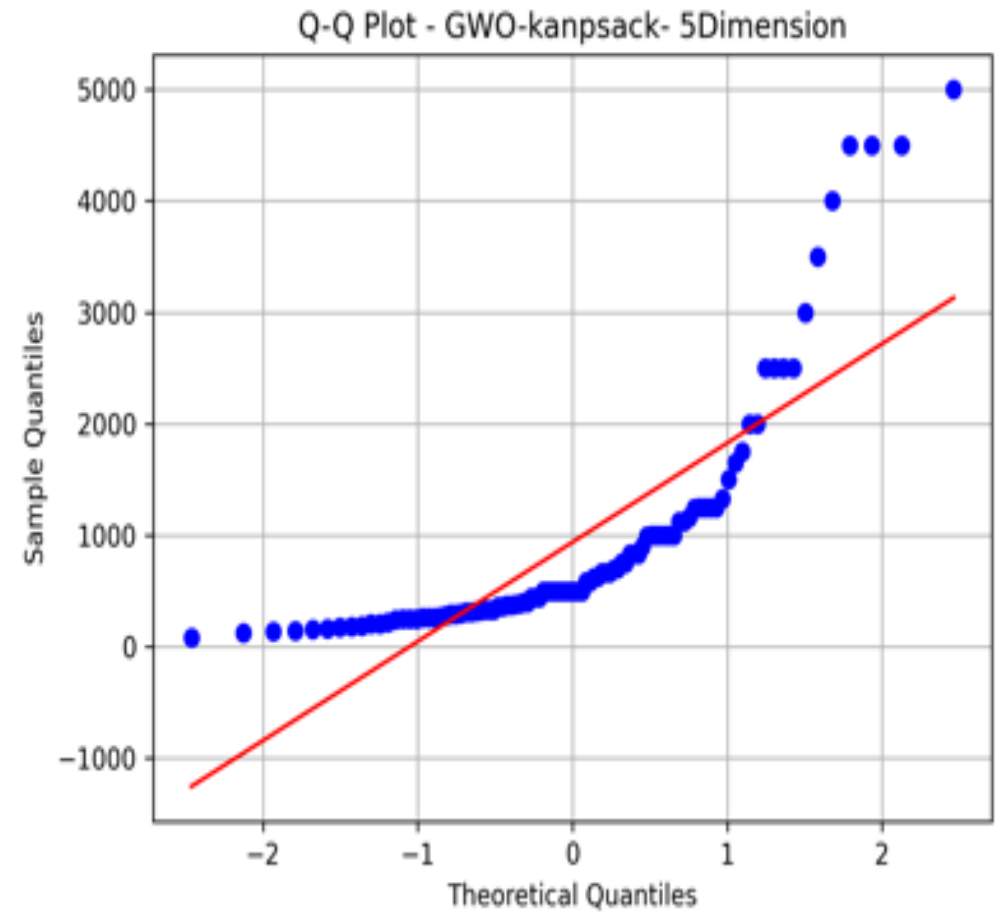

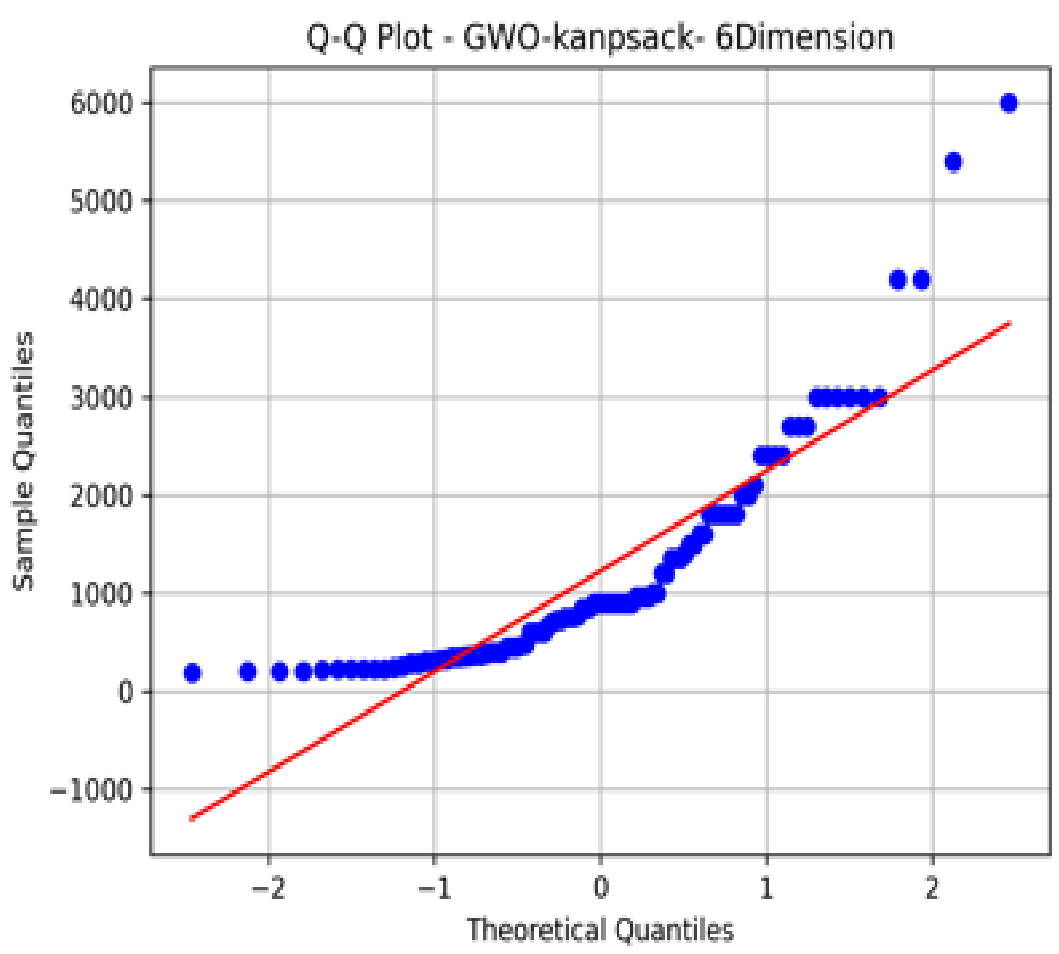

<u>WO Q-Q Plots</u>

To generate the below mentioned plots we have used Table L6 data

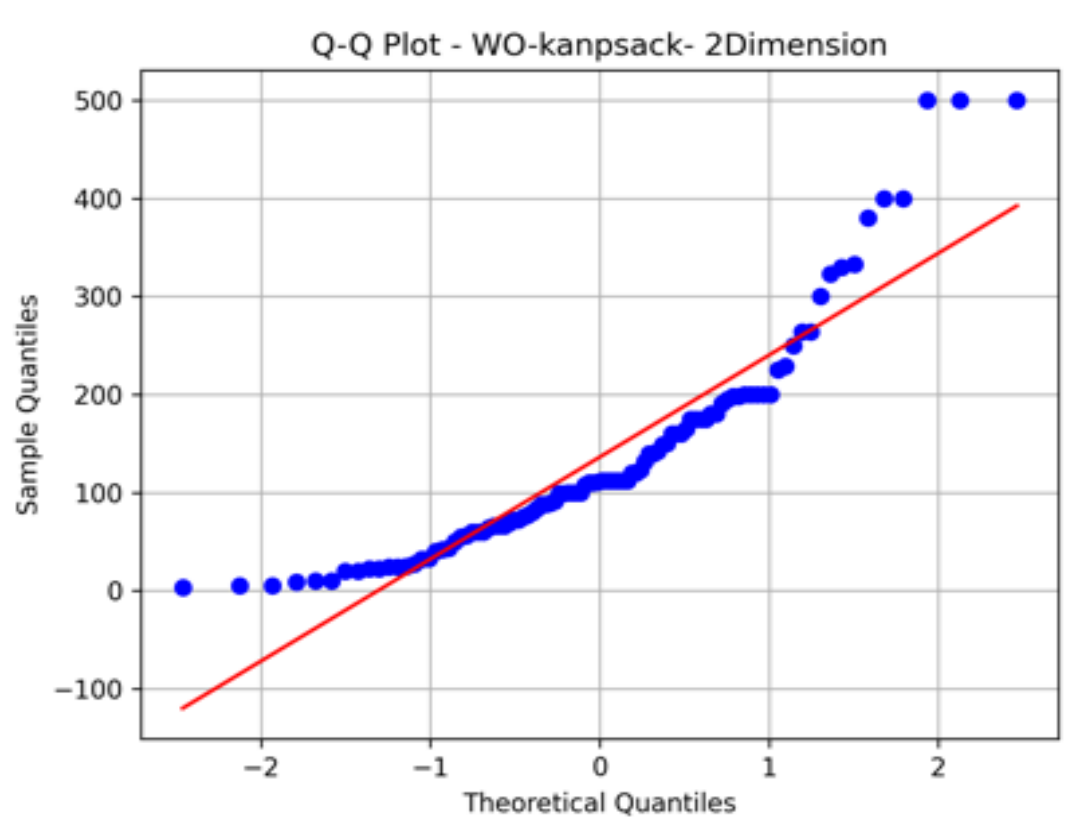

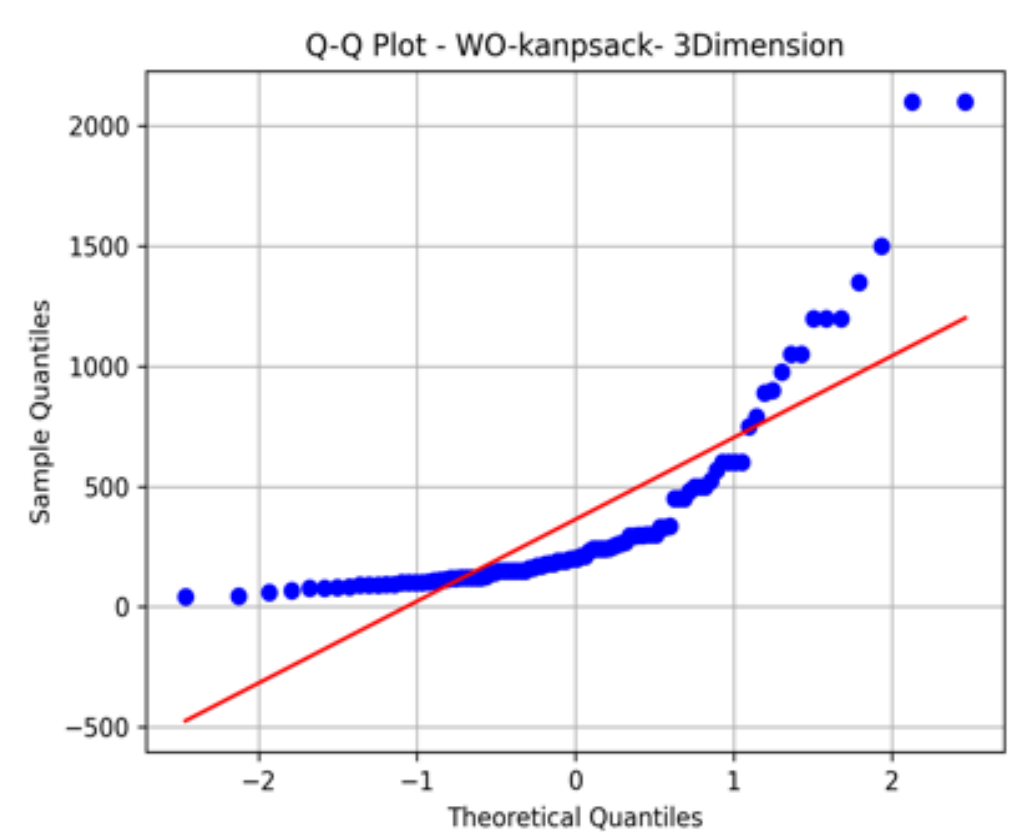

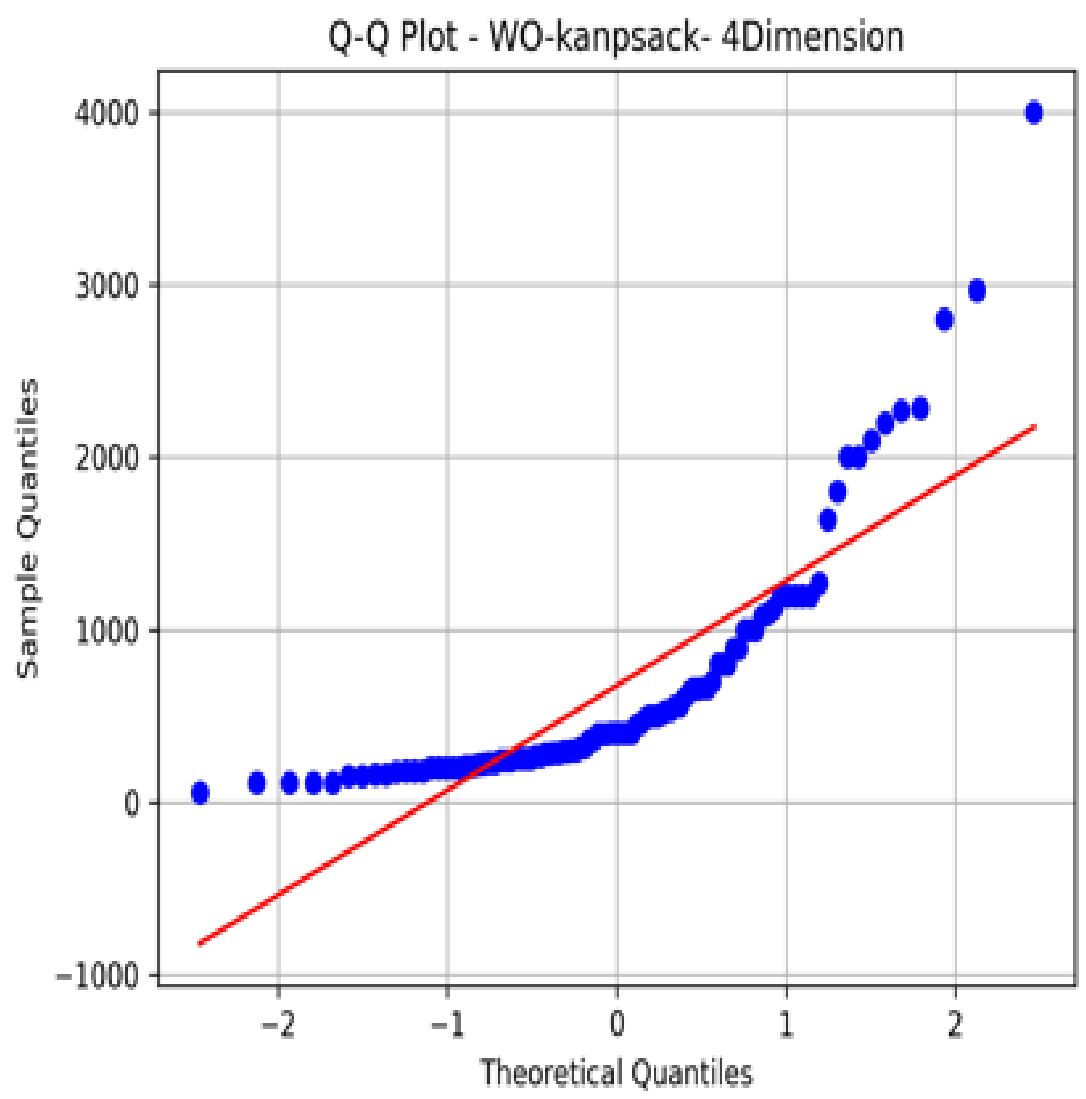
Q-Q Plot - WO-kanpsack- 4Dimension
Sample Quantiles
Theoretical Quantiles

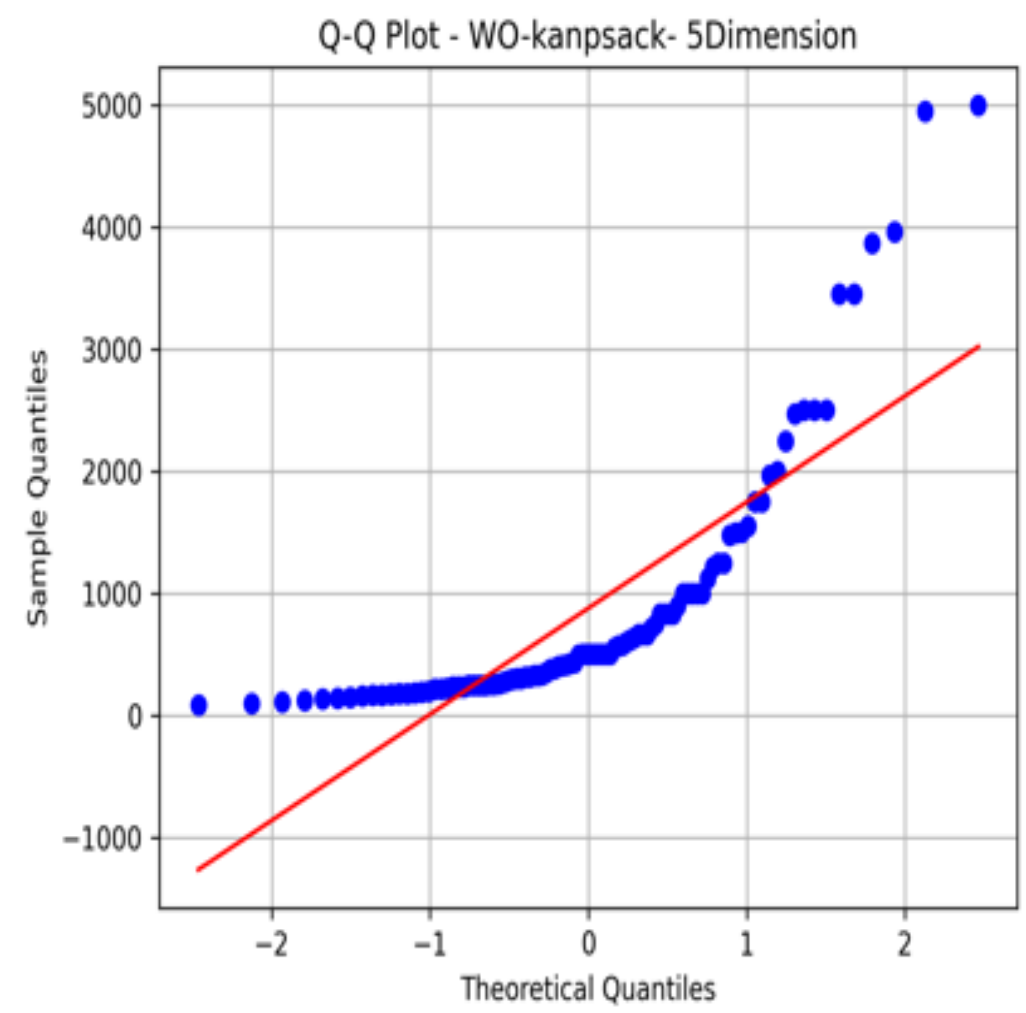
Q-Q Plot - WO-kanpsack- 5Dimension
Sample Quantiles
Theoretical Quantiles

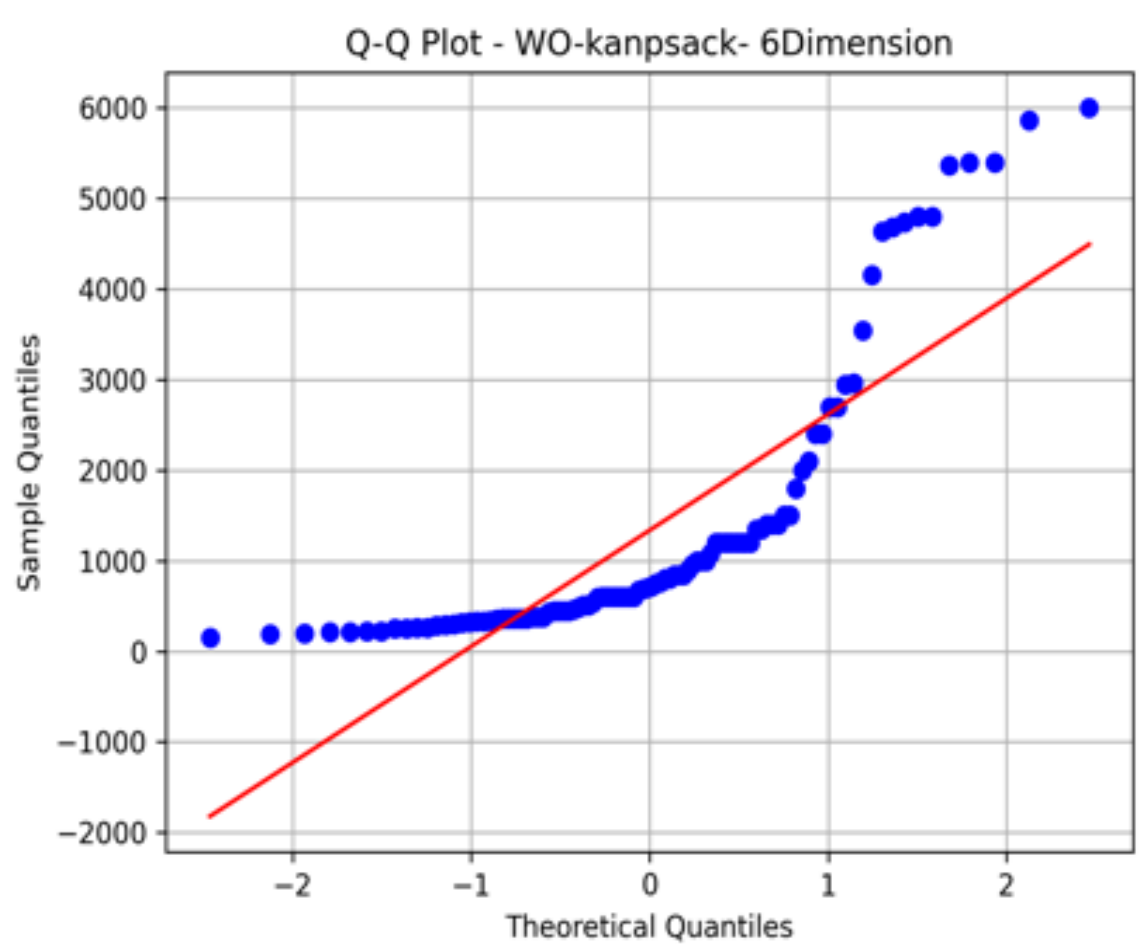
Q-Q Plot - WO-kanpsack- 6Dimension
Sample Quantiles
Theoretical Quantiles

# APPENDIX L: UNBOUNDED KNAPSACK OPTIMIZED RESULT VALUE FOR CASE STUDY 3

In this appendix, we are mentioning the optimized value obtained in unbounded knapsack(Case Study 3) for studied metaheuristics. Below mentioned tables consist of the result value for 100 trials for each dimension as discussed in Case Study 3 (in Chapter 4, section 4.2). The time allotment for each dimension as follows: 2D(0.005sec), 3D(1sec), 4D(15 sec), 5D(2min), 6D(10min)

Optimized Values for BSO

Table L1, shows optimized value achieved by BSO in 2D, 3D, 4D, 5D, 6D for 100 trials in each dimension.

Table L1. Case Study 3 BSO Optimal Values

| 2D | 3D | 4D | 5D | 6D |
|---|---|---|---|---|
| 12 | 180 | 221 | 315 | 500 |
| 72 | 84 | 150 | 710 | 4800 |
| 27 | 200 | 458 | 1660 | 700 |
| 0 | 435 | 1489 | 500 | 300 |
| 36 | 276 | 800 | 747 | 1600 |
| 7 | 300 | 360 | 330 | 660 |
| 48 | 1350 | 264 | 4500 | 340 |
| 0 | 93 | 161 | 299 | 486 |
| 6 | 114 | 324 | 300 | 1050 |
| 96 | 500 | 991 | 330 | 750 |
| 20 | 65 | 223 | 180 | 597 |
| 16 | 75 | 800 | 166 | 600 |
| 14 | 300 | 2800 | 405 | 1800 |
| 21 | 130 | 1800 | 1000 | 960 |
| 24 | 297 | 320 | 458 | 720 |
| 52 | 2100 | 193 | 298 | 1200 |
| 7 | 73 | 2781 | 800 | 420 |
| 4 | 91 | 720 | 500 | 2400 |
| 24 | 116 | 862 | 1162 | 424 |
| 1 | 120 | 300 | 1250 | 594 |
| 70 | 184 | 116 | 750 | 800 |
| 32 | 108 | 440 | 200 | 426 |
| 27 | 207 | 235 | 500 | 280 |
| 24 | 600 | 233 | 747 | 300 |
| 16 | 65 | 112 | 3500 | 300 |
| 18 | 1800 | 324 | 256 | 2700 |
| 60 | 207 | 285 | 376 | 3600 |
| 217 | 66 | 330 | 1000 | 145 |
| 18 | 378 | 2400 | 664 | 255 |
| 133 | 52 | 136 | 369 | 766 |
| 105 | 500 | 179 | 215 | 900 |
| 28 | 170 | 270 | 2000 | 1600 |
| 0 | 549 | 1262 | 245 | 297 |
| 30 | 22 | 330 | 385 | 850 |
| 20 | 250 | 891 | 234 | 6000 |
| 39 | 72 | 147 | 260 | 3000 |
| 3 | 222 | 1000 | 642 | 1800 |
| 189 | 200 | 182 | 355 | 1400 |
| 12 | 144 | 100 | 198 | 750 |
| 35 | 252 | 532 | 250 | 2100 |
| 3 | 80 | 212 | 3500 | 348 |
| 5 | 96 | 421 | 875 | 1200 |
| 65 | 3000 | 3600 | 2000 | 300 |
| 16 | 102 | 198 | 375 | 528 |
| 50 | 500 | 757 | 248 | 1200 |
| 36 | 420 | 215 | 1750 | 357 |
| 42 | 48 | 262 | 274 | 192 |
| 56 | 984 | 75 | 245 | 540 |
| 27 | 100 | 260 | 395 | 1500 |
| 9 | 1000 | 891 | 625 | 800 |
| 6 | 700 | 138 | 496 | 2100 |
| 28 | 180 | 303 | 400 | 1200 |
| 1000 | 210 | 559 | 410 | 1800 |
| 4 | 180 | 4000 | 2000 | 270 |
| 24 | 110 | 154 | 250 | 1050 |
| 230 | 600 | 30 | 434 | 1200 |
| 270 | 128 | 159 | 450 | 2400 |
| 11 | 122 | 396 | 5000 | 2000 |
| 114 | 185 | 1330 | 1500 | 3000 |
| 155 | 1500 | 1200 | 664 | 245 |
| 65 | 180 | 180 | 427 | 4200 |
| 15 | 294 | 616 | 434 | 662 |
| 32 | 100 | 665 | 2500 | 2700 |
| 48 | 272 | 567 | 500 | 206 |
| 24 | 300 | 661 | 1660 | 800 |
| 135 | 200 | 190 | 355 | 360 |
| 470 | 68 | 102 | 310 | 3000 |
| 260 | 130 | 600 | 245 | 840 |
| 18 | 51 | 798 | 500 | 680 |
| 21 | 91 | 222 | 4500 | 1500 |
| 72 | 145 | 409 | 830 | 540 |
| 18 | 1050 | 508 | 4000 | 1500 |
| 133 | 104 | 168 | 415 | 850 |
| 20 | 150 | 168 | 1660 | 960 |
| 12 | 80 | 1330 | 581 | 1200 |
| 54 | 117 | 498 | 1750 | 500 |
| 42 | 600 | 272 | 400 | 600 |
| 80 | 300 | 288 | 552 | 138 |
| 63 | 108 | 142 | 289 | 420 |
| 10 | 100 | 2800 | 375 | 675 |
| 294 | 270 | 166 | 2250 | 1800 |
| 90 | 90 | 2800 | 568 | 500 |
| 70 | 378 | 146 | 2000 | 340 |
| 90 | 30 | 398 | 1500 | 600 |
| 0 | 90 | 161 | 450 | 1800 |
| 6 | 1350 | 207 | 4000 | 102 |
| 15 | 240 | 144 | 664 | 1126 |
| 5 | 210 | 167 | 1000 | 500 |
| 108 | 118 | 384 | 2500 | 2400 |
| 20 | 1800 | 504 | 3500 | 750 |
| 0 | 1200 | 106 | 500 | 260 |
| 27 | 420 | 240 | 3500 | 6000 |
| 10 | 378 | 154 | 4000 | 1080 |
| 56 | 62 | 250 | 380 | 462 |
| 14 | 64 | 118 | 664 | 2400 |
| 1 | 159 | 279 | 1000 | 3600 |
| 56 | 225 | 158 | 800 | 668 |
| 56 | 80 | 280 | 378 | 300 |
| 81 | 710 | 2311 | 246 | 378 |
| 30 | 114 | 252 | 4500 | 4800 |

<u>Optimized Values for BPO</u>

Table L2, shows optimized value achieved by BPO 2D, 3D, 4D, 5D, 6D for 100 trials in each dimension

Table L2. Case Study 3 BPO Optimal Values

| 2D | 3D | 4D | 5D | 6D |
|---|---|---|---|---|
| 6 | 98 | 530 | 2548 | 3799 |
| 9 | 1350 | 80 | 549 | 484 |
| 120 | 200 | 115 | 600 | 784 |
| 14 | 140 | 1590 | 806 | 804 |
| 112 | 800 | 358 | 2522 | 727 |
| 112 | 764 | 1000 | 382 | 977 |
| 30 | 138 | 640 | 425 | 1631 |
| 91 | 225 | 1668 | 759 | 564 |
| 70 | 3000 | 135 | 597 | 471 |
| 104 | 99 | 80 | 2433 | 469 |
| 7 | 131 | 236 | 1436 | 3206 |
| 200 | 378 | 351 | 1430 | 2148 |
| 162 | 789 | 1286 | 738 | 2308 |
| 126 | 64 | 297 | 1000 | 3385 |
| 120 | 234 | 184 | 731 | 745 |
| 90 | 800 | 1200 | 1200 | 807 |
| 154 | 240 | 213 | 853 | 407 |
| 230 | 170 | 800 | 1155 | 1664 |
| 167 | 160 | 360 | 789 | 2522 |
| 464 | 2400 | 746 | 680 | 1367 |
| 111 | 88 | 540 | 1159 | 1417 |
| 112 | 110 | 265 | 985 | 1100 |
| 340 | 76 | 200 | 618 | 471 |
| 240 | 542 | 198 | 2553 | 1948 |
| 108 | 700 | 800 | 860 | 671 |
| 80 | 89 | 126 | 609 | 3481 |
| 5 | 210 | 700 | 267 | 1194 |
| 112 | 120 | 1000 | 2237 | 484 |
| 294 | 450 | 654 | 730 | 1274 |
| 108 | 200 | 846 | 1112 | 1150 |
| 120 | 126 | 132 | 624 | 1598 |
| 96 | 188 | 28 | 3847 | 998 |
| 264 | 69 | 157 | 584 | 807 |
| 99 | 540 | 54 | 438 | 410 |
| 17 | 800 | 400 | 699 | 407 |
| 99 | 23 | 1106 | 1181 | 714 |
| 300 | 98 | 284 | 440 | 1034 |
| 55 | 96 | 280 | 440 | 1173 |
| 33 | 654 | 900 | 370 | 485 |
| 9 | 525 | 263 | 569 | 2192 |
| 140 | 453 | 431 | 410 | 4327 |
| 225 | 168 | 274 | 1766 | 1664 |
| 360 | 304 | 252 | 436 | 998 |
| 156 | 150 | 111 | 233 | 741 |
| 112 | 111 | 9 | 649 | 1210 |
| 57 | 890 | 1558 | 689 | 594 |
| 112 | 225 | 534 | 2286 | 1214 |
| 168 | 150 | 135 | 706 | 1210 |
| 81 | 105 | 214 | 784 | 879 |
| 151 | 222 | 140 | 942 | 3435 |
| 108 | 67 | 212 | 1429 | 2174 |
| 100 | 81 | 160 | 709 | 410 |
| 90 | 1500 | 654 | 5640 | 1668 |
| 51 | 99 | 665 | 3540 | 5907 |
| 264 | 235 | 238 | -1 | 1657 |
| 39 | 32 | 916 | 1350 | 1210 |
| 150 | 90 | 2800 | 336 | 556 |
| 112 | 51 | 121 | 482 | 640 |
| 168 | 210 | 888 | 784 | 1120 |
| 142 | 80 | 555 | 656 | 2168 |
| 1000 | 108 | 324 | 1119 | 862 |
| 22 | 264 | 1600 | 443 | 1258 |
| 77 | 202 | 252 | 846 | 710 |
| 100 | 2400 | 300 | 1297 | 951 |
| 140 | 900 | 222 | 2831 | 1747 |
| 166 | 180 | 104 | 1500 | 2848 |
| 90 | 296 | 561 | 807 | 1668 |
| 190 | 500 | 109 | 1298 | 1883 |
| 287 | 144 | 987 | 1328 | 1284 |
| 133 | 40 | 700 | 911 | 1227 |
| 50 | 912 | 311 | 481 | 1120 |
| 60 | 169 | 1190 | 3318 | 500 |
| 165 | 73 | 320 | 434 | 604 |
| 144 | 38 | 584 | 1702 | 556 |
| 110 | 165 | 500 | 2984 | 909 |
| 321 | 450 | 891 | 690 | 1501 |
| 108 | 210 | 987 | 911 | 3312 |
| 245 | 74 | 1197 | 912 | 3000 |
| 540 | 86 | 560 | 511 | 1068 |
| 231 | 100 | 200 | 4741 | 1668 |
| 160 | 66 | 100 | 869 | 745 |
| 48 | 222 | 876 | 1155 | 802 |
| 33 | 72 | 543 | 410 | 124 |
| 101 | 222 | 656 | 345 | 1631 |
| 261 | 2400 | 396 | 714 | 1668 |
| 60 | 170 | 908 | 475 | 675 |
| 450 | 102 | 285 | 1090 | 1800 |
| 70 | 120 | 654 | 410 | 500 |
| 120 | 163 | 1260 | 1888 | 340 |
| 119 | 90 | 121 | 1000 | 600 |
| 64 | 159 | 543 | 1883 | 1800 |
| 266 | 10 | 252 | 511 | 102 |
| 390 | 219 | 1064 | 1298 | 1126 |
| 140 | 400 | 343 | 583 | 500 |
| 98 | 90 | 1200 | 988 | 2400 |
| 171 | 4 | 800 | 812 | 750 |
| 70 | 158 | 594 | 1987 | 260 |
| 270 | 900 | 236 | 1778 | 6000 |
| 131 | 132 | 400 | 1098 | 1080 |
| 180 | 200 | 200 | 519 | 462 |

Optimized Values for SA

Table L3, shows optimized value achieved by SA 2D, 3D, 4D, 5D, 6D for 100 trials in each dimension

Table L3. Case Study 3 SA Optimal Values

| 2D | 3D | 4D | 5D | 6D |
|---|---|---|---|---|
| 100 | 500 | 1501 | 146 | 1360 |
| 60 | 400 | 2100 | 750 | 600 |
| 350 | 105 | 1700 | 2500 | 1200 |
| 72 | 120 | 300 | 304 | 1800 |
| 700 | 126 | 1890 | 1000 | 1450 |
| 140 | 69 | 431 | 361 | 3600 |
| 300 | 148 | 208 | 318 | 675 |
| 50 | 180 | 160 | 2000 | 1205 |
| 100 | 147 | 513 | 248 | 1283 |
| 60 | 378 | 771 | 3500 | 1000 |
| 198 | 153 | 360 | 203 | 980 |
| 60 | 333 | 1400 | 900 | 525 |
| 66 | 700 | 400 | 750 | 2400 |
| 144 | 119 | 177 | 304 | 282 |
| 70 | 180 | 288 | 410 | 270 |
| 400 | 600 | 300 | 1328 | 396 |
| 700 | 550 | 216 | 1000 | 500 |
| 1 | 650 | 781 | 1125 | 720 |
| 100 | 270 | 9 | 355 | 700 |
| 200 | 360 | 450 | 1328 | 600 |
| 175 | 22 | 456 | 500 | 1200 |
| 14 | 34 | 289 | 500 | 675 |
| 88 | 105 | 225 | 233 | 1200 |
| 180 | 184 | 350 | 747 | 1200 |
| 300 | 600 | 285 | 800 | 1200 |
| 48 | 150 | 234 | 171 | 450 |
| 90 | 180 | 330 | 568 | 765 |
| 70 | 220 | 700 | 332 | 3000 |
| 100 | 139 | 147 | 1750 | 2400 |
| 30 | 189 | 308 | 214 | 4200 |
| 250 | 167 | 1400 | 2500 | 660 |
| 56 | 110 | 1000 | 232 | 900 |
| 96 | 348 | 154 | 287 | 243 |
| 250 | 1350 | 1330 | 830 | 870 |
| 300 | 259 | 400 | 211 | 450 |
| 99 | 750 | 279 | 250 | 540 |
| 11 | 300 | 160 | 1250 | 230 |
| 100 | 259 | 480 | 554 | 600 |
| 225 | 345 | 399 | 498 | 600 |
| 75 | 297 | 132 | 620 | 350 |
| 9 | 140 | 1000 | 125 | 1200 |
| 84 | 108 | 450 | 225 | 1500 |
| 77 | 200 | 289 | 332 | 3600 |
| 12 | 423 | 560 | 440 | 4800 |
| 125 | 243 | 500 | 248 | 204 |
| 11 | 546 | 230 | 1328 | 765 |
| 80 | 144 | 132 | 4000 | 280 |
| 84 | 1500 | 2561 | 1500 | 900 |
| 165 | 228 | 931 | 372 | 276 |
| 900 | 225 | 800 | 410 | 274 |
| 44 | 190 | 152 | 2000 | 184 |
| 72 | 120 | 250 | 1500 | 600 |
| 132 | 400 | 264 | 198 | 1200 |
| 45 | 420 | 216 | 800 | 675 |
| 50 | 370 | 2267 | 600 | 1000 |
| 112 | 600 | 252 | 2000 | 765 |
| 64 | 146 | 130 | 300 | 1400 |
| 991 | 216 | 901 | 1250 | 675 |
| 7 | 675 | 881 | 250 | 255 |
| 80 | 168 | 798 | 4000 | 3600 |
| 140 | 101 | 1781 | 315 | 4800 |
| 99 | 1050 | 1400 | 180 | 1080 |
| 132 | 200 | 352 | 750 | 300 |
| 25 | 450 | 297 | 163 | 400 |
| 80 | 55 | 800 | 4500 | 540 |
| 600 | 900 | 1400 | 1000 | 840 |
| 23 | 77 | 297 | 900 | 3600 |
| 84 | 240 | 174 | 2000 | 1890 |
| 500 | 40 | 891 | 2400 | 600 |
| 330 | 600 | 804 | 639 | 2700 |
| 165 | 91 | 162 | 360 | 381 |
| 20 | 110 | 777 | 639 | 1350 |
| 48 | 184 | 171 | 360 | 540 |
| 72 | 500 | 240 | 405 | 2100 |
| 500 | 2100 | 890 | 2250 | 720 |
| 96 | 200 | 135 | 1000 | 171 |
| 33 | 180 | 665 | 497 | 600 |
| 500 | 178 | 3600 | 332 | 3000 |
| 41 | 184 | 281 | 875 | 851 |
| 10 | 131 | 260 | 500 | 1200 |
| 300 | 180 | 335 | 800 | 480 |
| 10 | 200 | 600 | 290 | 528 |
| 100 | 189 | 1200 | 639 | 190 |
| 200 | 150 | 456 | 268 | 220 |
| 77 | 96 | 251 | 380 | 991 |
| 125 | 50 | 157 | 262 | 701 |
| 300 | 77 | 360 | 332 | 1771 |
| 60 | 1350 | 1800 | 249 | 600 |
| 40 | 30 | 1001 | 282 | 2501 |
| 11 | 600 | 800 | 180 | 3301 |
| 112 | 290 | 399 | 348 | 1800 |
| 100 | 290 | 230 | 3300 | 1500 |
| 33 | 450 | 1197 | 200 | 1700 |
| 87 | 102 | 2400 | 800 | 3000 |
| 140 | 191 | 76 | 750 | 100 |
| 66 | 189 | 1800 | 304 | 4200 |
| 330 | 300 | 10 | 510 | 2400 |
| 100 | 2100 | 396 | 1328 | 100 |
| 175 | 250 | 594 | 800 | 1800 |
| 180 | 160 | 2220 | 1225 | 210 |

Optimized Values for TA

Table L4, shows optimized value achieved by TA 2D, 3D, 4D, 5D, 6D for 100 trials in each dimension

Table L4. Case Study 3 TA Optimal Values

| 2D | 3D | 4D | 5D | 6D |
|---|---|---|---|---|
| 160 | 1264 | 289 | 4000 | 138 |
| 36 | 500 | 1330 | 405 | 2700 |
| 20 | 300 | 462 | 139 | 198 |
| 112 | 162 | 238 | 500 | 178 |
| 55 | 126 | 890 | 415 | 2345 |
| 110 | 153 | 269 | 1660 | 159 |
| 84 | 100 | 4000 | 350 | 299 |
| 300 | 425 | 128 | 498 | 600 |
| 75 | 333 | 594 | 248 | 1200 |
| 200 | 128 | 1140 | 1250 | 2100 |
| 175 | 208 | 456 | 360 | 231 |
| 175 | 545 | 247 | 2000 | 2100 |
| 400 | 370 | 3200 | 668 | 750 |
| 70 | 237 | 192 | 1000 | 1000 |
| 250 | 1350 | 211 | 315 | 2290 |
| 600 | 168 | 332 | 581 | 800 |
| 500 | 138 | 300 | 450 | 600 |
| 231 | 33 | 4000 | 625 | 1500 |
| 96 | 245 | 172 | 900 | 271 |
| 140 | 250 | 230 | 186 | 333 |
| 600 | 191 | 245 | 124 | 900 |
| 20 | 90 | 125 | 714 | 1500 |
| 55 | 68 | 300 | 5000 | 215 |
| 66 | 104 | 700 | 434 | 1200 |
| 450 | 100 | 2000 | 250 | 370 |
| 30 | 800 | 250 | 145 | 243 |
| 108 | 185 | 330 | 405 | 375 |
| 98 | 75 | 2000 | 700 | 185 |
| 175 | 108 | 664 | 200 | 1200 |
| 250 | 297 | 870 | 125 | 297 |
| 180 | 33 | 160 | 450 | 2410 |
| 125 | 399 | 132 | 260 | 207 |
| 120 | 750 | 228 | 625 | 900 |
| 264 | 105 | 2800 | 700 | 500 |
| 800 | 93 | 288 | 353 | 89 |
| 48 | 500 | 198 | 231 | 370 |
| 64 | 450 | 500 | 665 | 500 |
| 84 | 1000 | 800 | 1250 | 1200 |
| 500 | 180 | 180 | 600 | 423 |
| 44 | 245 | 660 | 750 | 171 |
| 198 | 260 | 203 | 497 | 2400 |
| 200 | 122 | 400 | 830 | 326 |
| 25 | 120 | 1000 | 3000 | 1400 |
| 33 | 570 | 352 | 355 | 3600 |
| 110 | 128 | 184 | 830 | 800 |
| 24 | 55 | 1700 | 210 | 198 |
| 350 | 178 | 220 | 456 | 450 |
| 56 | 350 | 1000 | 3000 | 3600 |
| 90 | 153 | 227 | 750 | 1400 |
| 150 | 1200 | 197 | 224 | 1400 |
| 66 | 420 | 396 | 182 | 1400 |
| 96 | 216 | 248 | 350 | 600 |
| 60 | 144 | 216 | 4000 | 360 |
| 36 | 191 | 224 | 218 | 300 |
| 100 | 110 | 187 | 156 | 225 |
| 800 | 120 | 255 | 1000 | 765 |
| 96 | 170 | 37 | 620 | 300 |
| 77 | 77 | 563 | 460 | 1800 |
| 112 | 135 | 196 | 800 | 334 |
| 250 | 112 | 2400 | 3000 | 3600 |
| 55 | 120 | 83 | 639 | 250 |
| 84 | 130 | 289 | 1500 | 2700 |
| 10 | 600 | 328 | 1750 | 160 |
| 12 | 300 | 1306 | 999 | 350 |
| 140 | 417 | 436 | 1162 | 1200 |
| 80 | 3200 | 440 | 150 | 161 |
| 50 | 320 | 180 | 264 | 1050 |
| 56 | 126 | 280 | 315 | 119 |
| 400 | 300 | 320 | 4000 | 1200 |
| 80 | 243 | 225 | 2500 | 400 |
| 77 | 100 | 1006 | 186 | 3000 |
| 160 | 980 | 264 | 369 | 135 |
| 250 | 142 | 2001 | 136 | 147 |
| 150 | 200 | 4000 | 639 | 600 |
| 16 | 297 | 2400 | 600 | 184 |
| 12 | 105 | 200 | 240 | 390 |
| 80 | 890 | 229 | 496 | 120 |
| 88 | 50 | 600 | 225 | 121 |
| 96 | 180 | 244 | 201 | 1500 |
| 75 | 370 | 1064 | 186 | 267 |
| 198 | 210 | 1666 | 235 | 540 |
| 400 | 560 | 138 | 5000 | 285 |
| 100 | 240 | 987 | 3500 | 765 |
| 16 | 202 | 1665 | 189 | 720 |
| 70 | 96 | 772 | 875 | 350 |
| 40 | 617 | 189 | 1500 | 900 |
| 99 | 505 | 900 | 1660 | 1000 |
| 56 | 252 | 1600 | 710 | 800 |
| 66 | 108 | 200 | 1250 | 2400 |
| 11 | 154 | 4000 | 350 | 525 |
| 250 | 216 | 900 | 319 | 4000 |
| 48 | 100 | 78 | 334 | 315 |
| 60 | 264 | 300 | 440 | 1480 |
| 126 | 112 | 307 | 405 | 450 |
| 50 | 333 | 176 | 2000 | 2400 |
| 400 | 77 | 1232 | 750 | 1375 |
| 56 | 128 | 154 | 450 | 5400 |
| 297 | 72 | 176 | 218 | 4200 |
| 20 | 1500 | 377 | 297 | 40 |
| 60 | 800 | 2000 | 5500 | 525 |

Optimized Values for PSO

Table L5, shows optimized value achieved by PSO 2D, 3D, 4D, 5D, 6D for 100 trials in each dimension

Table L5. Case Study 3 PSO Optimal Values

| 2D | 3D | 4D | 5D | 6D |
|---|---|---|---|---|
| 60 | 154 | 240 | 625 | 340 |
| 60 | 85 | 570 | 342 | 750 |
| 42 | 119 | 831 | 438 | 322 |
| 70 | 46 | 400 | 60 | 132 |
| 70 | 100 | 180 | 483 | 335 |
| 25 | 30 | 122 | 1162 | 285 |
| 90 | 43 | 150 | 490 | 608 |
| 504 | 123 | 200 | 472 | 578 |
| 70 | 345 | 133 | 200 | 330 |
| 360 | 142 | 171 | 225 | 454 |
| 24 | 10 | 880 | 396 | 690 |
| 150 | 84 | 144 | 132 | 450 |
| 40 | 826 | 140 | 330 | 2000 |
| 168 | 282 | 67 | 330 | 264 |
| 11 | 175 | 66 | 228 | 225 |
| 30 | 152 | 300 | 270 | 254 |
| 132 | 100 | 204 | 124 | 475 |
| 198 | 107 | 120 | 232 | 1200 |
| 120 | 188 | 158 | 625 | 1010 |
| 48 | 168 | 171 | 350 | 306 |
| 297 | 600 | 462 | 536 | 198 |
| 475 | 90 | 337 | 60 | 720 |
| 20 | 850 | 273 | 338 | 130 |
| 25 | 210 | 1200 | 750 | -1 |
| 160 | 121 | 263 | 246 | 360 |
| 56 | 10 | 279 | 392 | 260 |
| 792 | 111 | 266 | 393 | 270 |
| 100 | 182 | 1200 | -1 | 349 |
| 72 | 158 | 304 | 666 | 450 |
| 855 | 40 | 323 | 385 | 600 |
| 112 | 83 | 264 | 212 | 427 |
| 70 | 65 | 396 | 410 | 272 |
| 48 | 112 | 266 | 663 | 280 |
| 250 | 120 | 246 | 433 | -1 |
| 25 | 39 | 2800 | 372 | 260 |
| 198 | 274 | 175 | 153 | 264 |
| 90 | 90 | 1000 | 369 | 352 |
| 50 | 70 | 250 | 272 | 616 |
| 60 | 38 | 169 | 1250 | 816 |
| 160 | 48 | 139 | 348 | 410 |
| 30 | 100 | 210 | 182 | 4200 |
| 20 | 91 | 120 | 220 | 600 |
| 44 | 102 | 1120 | 956 | 280 |
| 90 | 420 | 116 | 970 | 810 |
| 400 | 69 | 1600 | 325 | 530 |
| 80 | 120 | 272 | 785 | 180 |
| 80 | 68 | 76 | 256 | 217 |
| 20 | 209 | 931 | 1122 | 600 |
| 11 | 37 | 222 | 200 | 409 |
| 150 | 340 | 187 | 500 | 320 |
| 28 | 887 | 412 | 279 | 110 |
| 400 | 120 | 513 | 224 | 480 |
| 42 | 60 | 145 | 392 | 1200 |
| 28 | 146 | 220 | 163 | 5400 |
| 110 | 199 | 62 | 415 | 356 |
| 75 | 296 | 130 | 1494 | 3000 |
| 250 | 400 | 900 | 277 | 450 |
| 20 | 140 | 186 | 1500 | 350 |
| 80 | 110 | 1200 | 150 | 6000 |
| 88 | 488 | 473 | 213 | -1 |
| 276 | 124 | 2000 | 60 | 290 |
| 10 | 70 | 900 | 205 | -1 |
| 350 | 767 | 4000 | 466 | 400 |
| 128 | 456 | 330 | 369 | 148 |
| 160 | 118 | 135 | 844 | 750 |
| 300 | 48 | 130 | 400 | 510 |
| 165 | 24 | 532 | 610 | 680 |
| 98 | 50 | 282 | 333 | 226 |
| 100 | 43 | 340 | 200 | 550 |
| 42 | 146 | 462 | 1000 | 300 |
| 66 | 184 | 900 | 100 | 341 |
| 30 | 271 | 400 | 4000 | 320 |
| 70 | 143 | 181 | 232 | 69 |
| 48 | 42 | 125 | 100 | 606 |
| 77 | 93 | 217 | 88 | 453 |
| 200 | 32 | 128 | 210 | 322 |
| 20 | 94 | 157 | 684 | 300 |
| 126 | 141 | 330 | 415 | 720 |
| 84 | 92 | 258 | 495 | 375 |
| 126 | 162 | 152 | 260 | 80 |
| 330 | 94 | 208 | 249 | 220 |
| 70 | 92 | 180 | 1772 | 290 |
| 77 | 125 | 133 | 174 | 660 |
| 80 | 211 | 156 | 390 | 625 |
| 250 | 99 | 195 | 280 | 2000 |
| 84 | 98 | 180 | 208 | 600 |
| 165 | 1000 | 1197 | 2250 | 445 |
| 30 | 162 | 555 | 300 | 350 |
| 395 | 800 | 204 | 3250 | 333 |
| 225 | 900 | 18 | 260 | 1154 |
| 108 | 121 | 294 | 376 | 240 |
| 56 | 153 | 146 | 400 | 480 |
| 50 | 152 | 280 | 1491 | 280 |
| 40 | 112 | 165 | 349 | 660 |
| 658 | 506 | 301 | 50 | 400 |
| 126 | 786 | 250 | 330 | 476 |
| 120 | 148 | 190 | 342 | 280 |
| 60 | 150 | 132 | 639 | 385 |
| 175 | 400 | 213 | 320 | 1500 |
| 125 | 76 | 160 | 400 | 1800 |

Optimized values for WO

Table L6, shows optimized value achieved by WO 2D, 3D, 4D, 5D, 6D for 100 trials in each dimension

Table L6. Case Study 3 WO Optimal Values

| 2D | 3D | 4D | 5D | 6D |
|---|---|---|---|---|
| 56 | 117 | 400 | 249 | 2400 |
| 10 | 567 | 393 | 600 | 154 |
| 32 | 1199 | 700 | 1550 | 800 |
| 195 | 117 | 500 | 250 | 900 |
| 33 | 450 | 891 | 1217 | 600 |
| 150 | 99 | 114 | 550 | 840 |
| 229 | 300 | 2968 | 250 | 1400 |
| 100 | 189 | 227 | 248 | 2957 |
| 180 | 450 | 1080 | 410 | 210 |
| 142 | 1200 | 660 | 142 | 360 |
| 60 | 300 | 2800 | 190 | 330 |
| 200 | 41 | 250 | 996 | 960 |
| 380 | 240 | 300 | 217 | 800 |
| 123 | 150 | 601 | 830 | 600 |
| 110 | 243 | 2099 | 400 | 765 |
| 225 | 100 | 665 | 198 | 510 |
| 333 | 90 | 111 | 3456 | 4800 |
| 75 | 126 | 1095 | 568 | 252 |
| 264 | 294 | 320 | 750 | 2000 |
| 160 | 2100 | 1197 | 284 | 350 |
| 89 | 120 | 560 | 4949 | 4800 |
| 132 | 500 | 300 | 146 | 1200 |
| 323 | 480 | 160 | 500 | 380 |
| 300 | 100 | 280 | 1000 | 488 |
| 24 | 140 | 500 | 330 | 294 |
| 60 | 198 | 396 | 664 | 600 |
| 264 | 150 | 57 | 162 | 450 |
| 72 | 68 | 240 | 315 | 360 |
| 175 | 108 | 400 | 234 | 1400 |
| 40 | 120 | 2000 | 710 | 300 |
| 500 | 120 | 1800 | 620 | 2400 |
| 330 | 900 | 250 | 2500 | 450 |
| 112 | 200 | 181 | 380 | 333 |
| 3 | 336 | 640 | 165 | 1080 |
| 10 | 150 | 1271 | 500 | 600 |
| 25 | 294 | 396 | 260 | 600 |
| 20 | 330 | 800 | 297 | 600 |
| 500 | 1350 | 660 | 3867 | 1350 |
| 72 | 450 | 224 | 2000 | 1200 |
| 5 | 138 | 297 | 3962 | 380 |
| 112 | 78 | 152 | 180 | 462 |
| 99 | 150 | 216 | 186 | 1200 |
| 66 | 160 | 528 | 180 | 1500 |
| 22 | 230 | 2200 | 2474 | 255 |
| 200 | 112 | 1000 | 1475 | 360 |
| 69 | 111 | 400 | 166 | 595 |
| 88 | 525 | 240 | 500 | 322 |
| 100 | 300 | 4000 | 1494 | 1800 |
| 200 | 976 | 400 | 218 | 5400 |
| 111 | 77 | 359 | 500 | 428 |
| 120 | 150 | 160 | 300 | 2100 |
| 175 | 1500 | 207 | 2500 | 450 |
| 400 | 89 | 800 | 330 | 210 |
| 180 | 890 | 297 | 304 | 1350 |
| 24 | 177 | 200 | 434 | 1200 |
| 20 | 180 | 2285 | 2250 | 1200 |
| 198 | 83 | 456 | 100 | 225 |
| 27 | 92 | 1638 | 830 | 840 |
| 66 | 104 | 253 | 664 | 450 |
| 165 | 180 | 180 | 639 | 684 |
| 112 | 1050 | 210 | 2500 | 260 |
| 500 | 500 | 264 | 132 | 360 |
| 160 | 200 | 480 | 500 | 5369 |
| 50 | 2100 | 560 | 1500 | 675 |
| 88 | 150 | 252 | 350 | 1400 |
| 200 | 120 | 282 | 498 | 360 |
| 160 | 1200 | 360 | 426 | 1200 |
| 120 | 100 | 200 | 500 | 750 |
| 43 | 296 | 1200 | 382 | 4152 |
| 112 | 1050 | 1197 | 1750 | 4639 |
| 91 | 330 | 285 | 830 | 1200 |
| 400 | 189 | 175 | 1000 | 200 |
| 100 | 170 | 198 | 900 | 260 |
| 60 | 600 | 200 | 247 | 1000 |
| 66 | 210 | 2271 | 1250 | 3539 |
| 99 | 600 | 285 | 500 | 700 |
| 64 | 296 | 532 | 1125 | 4740 |
| 9 | 240 | 264 | 1000 | 1000 |
| 110 | 600 | 1197 | 415 | 6000 |
| 77 | 189 | 500 | 315 | 2700 |
| 175 | 210 | 112 | 272 | 5861 |
| 84 | 241 | 1000 | 332 | 840 |
| 198 | 60 | 2000 | 664 | 2700 |
| 100 | 750 | 1000 | 500 | 1000 |
| 108 | 259 | 400 | 830 | 5400 |
| 200 | 89 | 1200 | 1968 | 1500 |
| 22 | 250 | 114 | 248 | 720 |
| 191 | 600 | 800 | 120 | 2947 |
| 140 | 789 | 182 | 232 | 340 |
| 175 | 162 | 320 | 576 | 280 |
| 112 | 500 | 229 | 1250 | 540 |
| 55 | 168 | 441 | 221 | 1200 |
| 149 | 92 | 400 | 1750 | 315 |
| 80 | 45 | 665 | 111 | 218 |
| 42 | 150 | 150 | 1000 | 600 |
| 112 | 79 | 1125 | 3451 | 375 |
| 250 | 120 | 513 | 88 | 4684 |
| 5 | 240 | 890 | 171 | 600 |
| 112 | 264 | 240 | 5000 | 192 |
| 140 | 270 | 198 | 233 | 510 |

Optimized Values for GWO

Table L7, shows optimized value achieved by GWO 2D, 3D, 4D, 5D, 6D for 100 trials in each dimension

Table L7. Case Study 3 GWO Optimal Values

| 2D | 3D | 4D | 5D | 6D |
|---|---|---|---|---|
| 70 | 80 | 1197 | 290 | 1400 |
| 0 | 669 | 513 | 261 | 3000 |
| 100 | 540 | 148 | 1125 | 1800 |
| 8 | 333 | 170 | 830 | 666 |
| 30 | 168 | 798 | 247 | 315 |
| 42 | 108 | 210 | 205 | 1800 |
| 0 | 1200 | 353 | 498 | 1350 |
| 0 | 43 | 226 | 830 | 1800 |
| 20 | 555 | 111 | 900 | 3000 |
| 40 | 324 | 571 | 700 | 1500 |
| 450 | 378 | 900 | 498 | 450 |
| 20 | 9000 | 1200 | 2500 | 2400 |
| 21 | 160 | 132 | 1250 | 400 |
| 8 | 212 | 800 | 1250 | 350 |
| 14 | 44 | 285 | 142 | 2400 |
| 30 | 789 | 491 | 1000 | 215 |
| 90 | 91 | 176 | 500 | 2700 |
| 12 | 1350 | 360 | 2000 | 900 |
| 18 | 135 | 309 | 400 | 1600 |
| 42 | 264 | 330 | 2500 | 960 |
| 42 | 112 | 2000 | 4500 | 357 |
| 28 | 154 | 1200 | 266 | 1800 |
| 24 | 248 | 239 | 165 | 899 |
| 8 | 1000 | 720 | 385 | 720 |
| 20 | 106 | 1400 | 5000 | 900 |
| 9 | 328 | 222 | 1328 | 600 |
| 40 | 300 | 780 | 332 | 900 |
| 3 | 1500 | 800 | 440 | 399 |
| 100 | 150 | 153 | 312 | 480 |
| 30 | 400 | 442 | 380 | 220 |
| 96 | 180 | 2400 | 330 | 200 |
| 25 | 420 | 1064 | 625 | 2400 |
| 350 | 540 | 165 | 250 | 220 |
| 500 | 700 | 180 | 316 | 1200 |
| 14 | 423 | 240 | 830 | 1800 |
| 0 | 600 | 994 | 495 | 360 |
| 66 | 400 | 120 | 187 | 202 |
| 10 | 564 | 660 | 358 | 240 |
| 9 | 336 | 400 | 3000 | 900 |
| 50 | 231 | 462 | 4500 | 381 |
| 12 | 150 | 480 | 2500 | 300 |
| 84 | 232 | 320 | 1250 | 6000 |
| 110 | 106 | 180 | 372 | 1200 |
| 200 | 600 | 1330 | 216 | 3000 |
| 20 | 426 | 1000 | 664 | 960 |
| 8 | 296 | 126 | 498 | 2700 |
| 25 | 10 | 117 | 176 | 1500 |
| 0 | 220 | 640 | 1162 | 2700 |
| 0 | 370 | 168 | 297 | 322 |
| 300 | 198 | 144 | 500 | 300 |
| 36 | 50 | 262 | 750 | 900 |
| 16 | 127 | 931 | 291 | 1000 |
| 0 | 300 | 160 | 80 | 750 |
| 18 | 230 | 159 | 664 | 220 |
| 400 | 100 | 354 | 2500 | 960 |
| 60 | 2700 | 108 | 1000 | 770 |
| 300 | 520 | 399 | 1125 | 186 |
| 0 | 480 | 399 | 2000 | 225 |
| 0 | 264 | 781 | 698 | 900 |
| 792 | 160 | 400 | 1000 | 900 |
| 48 | 1500 | 1200 | 375 | 1350 |
| 12 | 50 | 264 | 992 | 4200 |
| 150 | 177 | 234 | 1000 | 3000 |
| 0 | 495 | 149 | 326 | 900 |
| 0 | 764 | 400 | 500 | 1000 |
| 16 | 750 | 4000 | 400 | 1350 |
| 300 | 162 | 133 | 250 | 1800 |
| 225 | 123 | 280 | 1750 | 600 |
| 5 | 112 | 360 | 1250 | 280 |
| 0 | 200 | 800 | 1240 | 720 |
| 0 | 196 | 200 | 360 | 250 |
| 72 | 210 | 240 | 4500 | 350 |
| 42 | 1784 | 210 | 243 | 840 |
| 54 | 500 | 560 | 440 | 2000 |
| 0 | 100 | 264 | 4000 | 5400 |
| 24 | 80 | 800 | 314 | 2400 |
| 128 | 300 | 600 | 495 | 370 |
| 0 | 500 | 176 | 127 | 600 |
| 250 | 900 | 2000 | 264 | 700 |
| 27 | 350 | 324 | 500 | 750 |
| 40 | 300 | 456 | 620 | 225 |
| 50 | 200 | 198 | 440 | 200 |
| 200 | 900 | 532 | 328 | 960 |
| 96 | 55 | 2000 | 750 | 462 |
| 16 | 300 | 3200 | 500 | 900 |
| 10 | 800 | 3600 | 1000 | 3000 |
| 63 | 182 | 260 | 261 | 750 |
| 0 | 298 | 561 | 1500 | 450 |
| 32 | 222 | 1000 | 497 | 368 |
| 40 | 78 | 216 | 3500 | 4200 |
| 24 | 90 | 144 | 183 | 3000 |
| 18 | 300 | 600 | 1000 | 840 |
| 20 | 420 | 320 | 1653 | 850 |
| 21 | 540 | 640 | 581 | 600 |
| 250 | 78 | 207 | 280 | 2100 |
| 80 | 2700 | 513 | 581 | 396 |
| 600 | 330 | 1800 | 206 | 284 |
| 400 | 200 | 360 | 664 | 450 |
| 49 | 64 | 2000 | 140 | 2000 |
| 25 | 20 | 4000 | 153 | 1600 |